%% file: Deep_Neural_Network.tex
\documentclass[12pt]{book}

\usepackage{tikz}
\usepackage{tcolorbox}
\tcbuselibrary{skins, breakable, listings}  %

\usepackage{tocloft}
\usepackage[calc,en-US]{datetime2}

\usepackage{listings}

\usepackage{color}
 
\usepackage[disable]{todonotes}

\input{Commands.tex}

\input{Argument_command.tex}

\makeatletter
\renewcommand{\frontmatter}{\cleardoublepage\@mainmatterfalse\pagenumbering{arabic}}
\renewcommand{\mainmatter}{\cleardoublepage\@mainmattertrue}
\makeatother

\begin{document}

\frontmatter

\newgeometry{left=4cm,top=5cm}

\pagestyle{empty}

\NewCoffin \result
\NewCoffin \titlec
\NewCoffin \authors

\SetHorizontalCoffin \result {}
\SetVerticalCoffin \titlec {0.7\textwidth} {\raggedright\fontsize{36}{42}\sffamily\bfseries\color{black}Mathematical Introduction to Deep Learning:  Methods, Implementations, and Theory}
\SetVerticalCoffin \authors {0.7\textwidth} {\raggedright\fontsize{16}{20}\sffamily\color{black}Arnulf Jentzen\\Benno Kuckuck\\Philippe von Wurstemberger}

\JoinCoffins \result                \titlec
\JoinCoffins \result[\titlec-b,\titlec-l] \authors[t,l] (0pt, -75pt)
\TypesetCoffin \result

\thispagestyle{empty}

\restoregeometry

\begingroup

\fontsize{10}{12}\selectfont
\newskip\smskipamount \smskipamount=3pt plus 1pt minus 1pt
\newskip\bskipamount \bskipamount=14pt plus 4pt minus 4pt
\newcommand{\smskip}{\par\vskip\smskipamount}
\newcommand{\bskip}{\par\vskip\bskipamount}

\noindent
Arnulf Jentzen

\smskip

\noindent
School of Data Science and Shenzhen Research Institute of Big Data\\
The Chinese University of Hong Kong, Shenzhen (CUHK-Shenzhen)\\
Shenzhen, China\\
email: \url{ajentzen@cuhk.edu.cn}

\smskip

\noindent
Applied Mathematics: Institute for Analysis and Numerics\\
University of Münster\\
Münster, Germany\\
email: \url{ajentzen@uni-muenster.de}

\bskip

\noindent
Benno Kuckuck

\smskip

\noindent
School of Data Science and Shenzhen Research Institute of Big Data\\
The Chinese University of Hong Kong Shenzhen (CUHK-Shenzhen)\\
Shenzhen, China\\
email: \url{bkuckuck@cuhk.edu.cn}

\smskip

\noindent
Applied Mathematics: Institute for Analysis and Numerics\\
University of Münster\\
Münster, Germany\\
email: \url{bkuckuck@uni-muenster.de}

\bskip

\noindent
Philippe von Wurstemberger

\smskip

\noindent
School of Data Science\\
The Chinese University of Hong Kong, Shenzhen (CUHK-Shenzhen)\\
Shenzhen, China\\
email: \url{philippevw@cuhk.edu.cn}

\smskip

\noindent
Risklab, Department of Mathematics\\
ETH Zurich\\
Zurich, Switzerland\\
email: \url{philippe.vonwurstemberger@math.ethz.ch}

\bskip

\noindent
Keywords: deep learning, artificial neural network, stochastic gradient descent, optimization\\
Mathematics Subject Classification (2020): 68T07

\bskip

\noindent
Version of \today

\bskip

\noindent
All \textsc{Python} source codes in this book can be downloaded from 
\begin{equation*}
\begin{split} 
\text{\url{https://github.com/introdeeplearning/book}}
\end{split}
\end{equation*}
or from the arXiv page
of this book (by clicking on ``Other formats'' and then ``Download source'').

\endgroup

\raggedbottom

\cleardoublepage

\listoftodos

\newpage

\pagestyle{fancy}

\input{parts/Preface.tex}

\cleardoublepage
\flushbottom
\setcounter{tocdepth}{2}
\tableofcontents

\mainmatter

\input{parts/Introduction.tex}

\part{Artificial neural networks (ANNs)}
\label{part:ANNs}
\input{parts/Basics_on_ANNs.tex}

\input{parts/ANN_calculus.tex}

\part{Approximation}
\label{part:approx}

\input{parts/One-dimensional_ANN_approximation_results.tex}

\input{parts/Multi-dimensional_ANN_approximation_results.tex}

\part{Optimization}
\label{part:opt}

\input{parts/Optimization_through_flows_of_ODEs.tex}

\input{parts/Optimization_methods_definitions.tex}

\input{parts/Deterministic_GD_type_optimization_methods.tex}

\input{parts/Stochastic_GD_type_optimization_methods.tex}

\input{parts/Backpropagation.tex}

\input{parts/KL-inequalities.tex}

\input{parts/ANNs_with_batch_normalization.tex}

\input{parts/Optimization_through_random_initializations.tex}

\part{Generalization} 
\label{part:generalization}

\input{parts/Generalization_error.tex}

\input{parts/Analysis_of_the_generalization_error.tex}

\part{Composed error analysis}
\label{part:overall}

\input{parts/Overall_error_analysis.tex}

\input{parts/Analysis_of_the_overall_error.tex}

\part{Deep learning for partial differential equations (PDEs)}
\label{part:MLforPDEs}
\todoc{Weinan E PDE Part comments}

\input{parts/PINNs.tex}

\input{parts/Deep_Komogorov_method.tex}

\input{parts/Further_DL_methods.tex}

\cleardoublepage

\backmatter

\makeatletter
\def\toclevel@chapter{-1}
\makeatother

\chapter{Index of abbreviations}
\directoryofabbreviations
\raggedbottom

\cleardoublepage

\noindent
\chapter{List of figures}
\directoryoffigures
\raggedbottom

\cleardoublepage

\noindent
\chapter{List of source codes}
\label{chapter:listofcodes}
\directoryoflistings
\raggedbottom

\cleardoublepage

\chapter{List of definitions}
\directoryofdefinitions
\raggedbottom

\chapter{List of exercises}
\directoryofexercises
\raggedbottom

\cleardoublepage

\phantomsection

\addcontentsline{toc}{chapter}{Bibliography}

\appto\bibfont{\setlength{\emergencystretch}{1em}}
\printbibliography

\end{document}

%% file: Commands.tex
\usepackage[LGR,T1]{fontenc}
\DeclareMathAlphabet{\mathgtt}{LGR}{cmtt}{m}{n}
\DeclareSymbolFont{sfletters}{OML}{cmbrm}{m}{it}

\DeclareMathSymbol{\salpha}{\mathord}{sfletters}{"0B}
\DeclareMathSymbol{\sbeta}{\mathord}{sfletters}{"0C}
\DeclareMathSymbol{\sgamma}{\mathord}{sfletters}{"0D}
\DeclareMathSymbol{\sdelta}{\mathord}{sfletters}{"0E}
\DeclareMathSymbol{\sepsilon}{\mathord}{sfletters}{"0F}
\DeclareMathSymbol{\szeta}{\mathord}{sfletters}{"10}
\DeclareMathSymbol{\seta}{\mathord}{sfletters}{"11}
\DeclareMathSymbol{\stheta}{\mathord}{sfletters}{"12}
\DeclareMathSymbol{\siota}{\mathord}{sfletters}{"13}
\DeclareMathSymbol{\skappa}{\mathord}{sfletters}{"14}
\DeclareMathSymbol{\slambda}{\mathord}{sfletters}{"15}
\DeclareMathSymbol{\smu}{\mathord}{sfletters}{"16}
\DeclareMathSymbol{\snu}{\mathord}{sfletters}{"17}
\DeclareMathSymbol{\sxi}{\mathord}{sfletters}{"18}
\DeclareMathSymbol{\spi}{\mathord}{sfletters}{"19}
\DeclareMathSymbol{\srho}{\mathord}{sfletters}{"1A}
\DeclareMathSymbol{\ssigma}{\mathord}{sfletters}{"1B}
\DeclareMathSymbol{\stau}{\mathord}{sfletters}{"1C}
\DeclareMathSymbol{\supsilon}{\mathord}{sfletters}{"1D}
\DeclareMathSymbol{\sphi}{\mathord}{sfletters}{"1E}
\DeclareMathSymbol{\schi}{\mathord}{sfletters}{"1F}
\DeclareMathSymbol{\spsi}{\mathord}{sfletters}{"20}
\DeclareMathSymbol{\somega}{\mathord}{sfletters}{"21}
\DeclareMathSymbol{\svarepsilon}{\mathord}{sfletters}{"22}
\DeclareMathSymbol{\svartheta}{\mathord}{sfletters}{"23}
\DeclareMathSymbol{\svarpi}{\mathord}{sfletters}{"24}
\DeclareMathSymbol{\svarrho}{\mathord}{sfletters}{"25}
\DeclareMathSymbol{\svarsigma}{\mathord}{sfletters}{"26}
\DeclareMathSymbol{\svarphi}{\mathord}{sfletters}{"27}

\usepackage{amsmath,amsthm,geometry,nicefrac,mathtools,hyperref,comment,%
xcolor,verbatim,graphicx,comment,bbm,bm,wrapfig,datetime,xcoffins,ifthen,xspace}
\usepackage[inline,shortlabels]{enumitem}
\usepackage[capitalise,sort]{cleveref} %

\crefname{enumi}{item}{items}
\crefname{equation}{}{}
\crefname{figure}{Figure}{Figures}
\crefname{listing}{Source code}{Source codes}
\crefname{lstlisting}{Source code}{Source codes}
\crefname{cor}{Corollary}{Corollaries}

\usepackage{multicol}
\usepackage{algorithm}
\usepackage{algpseudocode}

\algtext*{EndFor} %

\newcommand{\todoc}[1]{\ifbool{STDL}
{}
{\ifbool{MIDL}
{}
{\todo[inline]{#1}}
}}

\newcommand{\todosecond}[1]{}

  \def\SourceSansPro@scale{1.02}
\usepackage{amssymb}

\usepackage[scr=boondox]{mathalpha}

\definecolor{darkblue}	{rgb} {0,0,0.75}
\definecolor{thered}    {rgb} {0.65,0.04,0.07}
\definecolor{thegreen}  {rgb} {0.06,0.44,0.08}
\definecolor{theblue}   {rgb} {0.02,0.04,0.48}
\definecolor{sectioning}{gray}{0.44}
\definecolor{thegrey}   {gray}{0.5}
\definecolor{theframe}  {gray}{0.75}
\definecolor{theshade}  {gray}{0.94}

\hypersetup{colorlinks=true,linkcolor=darkblue}

\usepackage{tikz}
\usetikzlibrary{matrix,chains,positioning,decorations.pathreplacing,arrows,math}
\usetikzlibrary{shapes,arrows}
\tikzset{
  font={\fontsize{9pt}{12}\selectfont}}
\usepackage{adjustbox}

\tikzset{
  font={\fontsize{9pt}{12}\selectfont}}

\tikzstyle{every edge}=[draw=black!70]
\tikzstyle{neuron}=[circle,fill=black!25,draw=black!75,minimum size=20pt,inner sep=0pt,thick]
\tikzstyle{input neuron}=[neuron,draw={black}, fill={rgb:red,1;white,5}]
\tikzstyle{hidden neuron}=[neuron, draw={black}, fill={rgb:blue,1;white,5}]
\tikzstyle{output neuron}=[neuron,draw={black}, fill={rgb:green,1;white,5}]
\tikzstyle{annot} = [text width=9em, text centered]
\tikzstyle{annot2} = [text width=4em, text centered]

\usepackage{environ}
\makeatletter
\newsavebox{\measure@tikzpicture}
\NewEnviron{scaletikzpicturetowidth}[1]{%
  \def\tikz@width{#1}%
  \begin{lrbox}{\measure@tikzpicture}%
  \BODY
  \end{lrbox}%
  \pgfmathparse{#1/\wd\measure@tikzpicture}%
  \BODY
}
\makeatother

\newcommand{\deflink}[2]{#2}

\usepackage[utf8]{inputenc}

\usepackage{microtype}

\usepackage{fancyhdr}
\makeatletter
\renewcommand{\chaptermark}[1]{%
  \if@mainmatter
    \markboth{Chapter \thechapter{}: #1}{}%
  \else
    \markboth{#1}{}%
  \fi
}
\makeatother

\renewcommand{\varTheta}{{\boldsymbol\Theta}}

\usepackage{etoolbox}
\makeatletter
\newbool{MIDL}
\@ifundefined{MIDL}{
  \boolfalse{MIDL}
}{
  \setbool{MIDL}{\shortver}
}

\newbool{STDL}
\@ifundefined{STDL}{
  \boolfalse{STDL}
}{
  \setbool{STDL}{\shortver}
}
\makeatother

\newenvironment{introductions}{}{}

\includecomment{notinSTDL}
\includecomment{notinMIDL}
\includecomment{notineither}

\def\sol{DDA4250_MIDL_lecturenotes}
\edef\sol{\meaning\sol}
\edef\tmpjob{\jobname}
\edef\tmpjob{\meaning\tmpjob}
\ifx\sol\tmpjob
\excludecomment{notinMIDL}
\excludecomment{notineither}
\excludecomment{introductions}
\booltrue{MIDL}
\fi

\def\soll{DDA6201_STDDA_lecturenotes}
\edef\soll{\meaning\soll}
\edef\tmpjob{\jobname}
\edef\tmpjob{\meaning\tmpjob}
\ifx\soll\tmpjob
\excludecomment{notineither}
\excludecomment{notinSTDL}
\excludecomment{introductions}
\booltrue{STDL}
\fi

\providecommand\C{}
\renewcommand{\C}{\mathbb{C}}
\newcommand{\Z}{\mathbb{Z}}

\renewcommand{\P}{\mathbb{P}}
\newcommand{\F}{\mathbb{F}}
\newcommand{\Exp}[1]{\E \expbr*{ #1 }}
\newcommand{\sExp}[1]{\E \expbr{ #1 }}
\newcommand{\bExp}[1]{\E\mkern-1mu \expbr[\big]{ #1 }}

\newcommand{\bbbExp}[1]{\E\mkern-2mu \expbr[\bigg]{ #1 }}

\newcommand{\EXp}[1]{\E [ #1 ]}
\newcommand{\EXP}[1]{\E \br[\big]{  #1 }}
\newcommand{\EXPP}[1]{\E \br[\Big]{ #1 }}
\newcommand{\EXPPP}[1]{\E \br[\bigg]{ #1  }}

\newcommand{\qand}{\qquad\text{and}}

\newcommand{\normmm}[1]{\opnorm*{#1}}
\newcommand{\ind}[1]{\mathbbm{1}_{#1}} 

\newcommand{\Hess}{\operatorname{Hess}}

\newcommand{\Real}[1]{\mathfrak{Re}({#1})}
\newcommand{\Ima}[1]{\mathfrak{Im}({#1})}
\renewcommand{\log}{\operatorname{ln}}

\providecommand{\B}{\mathcal{B}}

\newcommand{\indicator}[1]{\mathbbm{1}_{#1}}

\newcommand{\var}{\operatorname{Var}}

\newcommand{\andShort}{\text{ and }}
\newcommand{\forallDist}{\forall\,}

\newcommand{\E}{\mathbb{E}}
\renewcommand{\P}{\mathbb{P}}

\providecommand\cond{}
\newcommand{\conditionSymbol}[1][]{#1\vert\mathopen{}}
\DeclarePairedDelimiterX\expbr[1]{[}{]}{%
	\renewcommand\cond{\conditionSymbol[\delimsize]}
	#1
}

\DeclarePairedDelimiter{\br}{[}{]}

\newcommand{\Ebr}[1]{\E\expbr{#1}}
\newcommand{\bEbr}[1]{\E\expbr[\big]{#1}}

\newcommand{\R}{\mathbb{R}}
\newcommand{\N}{\mathbb{N}}
\newcommand{\pmat}[1]{\mleft(\begin{matrix}#1\end{matrix}\mright)}
\newcommand{\spmat}[1]{\pr*{\begin{smallmatrix}#1\end{smallmatrix}}}

\newcommand{\is}{\curvearrowleft}

\newcommand{\eps}{\varepsilon}

\newcommand{\Borel}{\mathcal{B}}

\newcommand{\idRelu}{\cfadd{def:ReLu:identity}\deflink{def:ReLu:identity}{\mathfrak I}}

\newcommand{\MappingStructuralToVectorized}{\cfadd{def:TranslateStructuredIntoVectorizedDescription}\deflink{def:TranslateStructuredIntoVectorizedDescription}{\mathcal{T}}}

\newcommand{\Exists}{\exists\,}
\newcommand{\Forall}{\forall\,}

\newcommand{\mc}[1]{\mathcal{#1}}
\newcommand{\mf}[1]{\mathfrak{#1}}

\DeclarePairedDelimiter{\abs}{\lvert}{\rvert}

\newcommand{\babs}[1]{\abs[\big]{#1}}

\DeclarePairedDelimiterXPP\lnorm[2]{}\lVert\rVert{_{#2}}{#1}

\newcommand{\pnorm}[2]{\cfadd{def:p-norm}\lnorm{#2}{#1}}
\newcommand{\apnorm}[2]{\cfadd{def:p-norm}\lnorm*{#2}{#1}}
\newcommand{\bpnorm}[2]{\cfadd{def:p-norm}\lnorm[\big]{#2}{#1}}

\newcommand{\bbbpnorm}[2]{\cfadd{def:p-norm}\lnorm[\bigg]{#2}{#1}}
\newcommand{\bbbbpnorm}[2]{\cfadd{def:p-norm}\lnorm[\Bigg]{#2}{#1}}

\newcommand{\hilbertSchmidtNorm}[1]{\cfadd{def:hilbertSchmidtNorm}\lnorm{#1}{HS}}

\DeclarePairedDelimiter{\scp}{\langle\cfadd{def:scalarproduct}}{\rangle}
\newcommand{\bscp}[1]{\scp[\big]{#1}}

\newcommand{\altscp}[1]{\langle\!\langle #1\rangle\!\rangle}
\newcommand{\baltscp}[1]{\bigl\langle\!\bigl\langle #1\bigr\rangle\!\bigr\rangle}

\newcommand{\aaltscp}[1]{\mleft\langle\!\mleft\langle #1\mright\rangle\!\mright\rangle}

\usepackage{mleftright}
\renewenvironment{pmatrix}{\mleft(\begin{matrix}}{\end{matrix}\mright)}

\DeclareFontFamily{U}{matha}{\hyphenchar\font45}
\DeclareFontShape{U}{matha}{m}{n}{
<-6> matha5 <6-7> matha6 <7-8> matha7
<8-9> matha8 <9-10> matha9
<10-12> matha10 <12-> matha12
}{}
\DeclareSymbolFont{matha}{U}{matha}{m}{n}

\DeclareFontFamily{U}{mathx}{\hyphenchar\font45}
\DeclareFontShape{U}{mathx}{m}{n}{
<-6> mathx5 <6-7> mathx6 <7-8> mathx7
<8-9> mathx8 <9-10> mathx9
<10-12> mathx10 <12-> mathx12
}{}
\DeclareSymbolFont{mathx}{U}{mathx}{m}{n}

\DeclareMathDelimiter{\vvvert} {0}{matha}{"7E}{mathx}{"17}%

\DeclarePairedDelimiterX{\opnorm}[1]
{\vvvert}
{\vvvert}
{\ifblank{#1}{\:\cdot\:}{#1}}

\newcommand{\shat}[1]{\smash[t]{\hat{#1}}}

\makeatletter
\newcommand{\oast}{\mathbin{\mathpalette\make@circled\ast}}
\newcommand{\make@circled}[2]{%
  \ooalign{$\m@th#1\smallbigcirc{#1}$\cr\hidewidth$\m@th#1#2$\hidewidth\cr}%
}
\newcommand{\smallbigcirc}[1]{%
  \vcenter{\hbox{\scalebox{0.77778}{$\m@th#1\bigcirc$}}}%
}
\makeatother

\newcommand{\infnorm}[1]{\pnorm\infty{#1}}
\newcommand{\asinfnorm}[1]{\apnorm\infty{#1}}

\newcommand{\Aff}{\cfadd{def:affine}\deflink{def:affine}{\mathcal{A}}}
\newcommand{\RealV}[4]{\cfadd{def:FFNN}\deflink{def:FFNN}{\mathcal{N}}^{#1,#3}_{#4}}
\newcommand{\RealVshort}{\cfadd{def:FFNN}\deflink{def:FFNN}{\mathcal{N}}}

\newcommand{\ClippedRealV}[4]{\cfadd{def:rectclippedFFANN}\deflink{def:rectclippedFFANN}{\mathscr{N}}^{#1,#2}_{\mkern-2mu #3,#4}}
\newcommand{\UnclippedRealV}[2]{\cfadd{def:rectclippedFFANN}\deflink{def:rectclippedFFANN}{\mathscr{N}}^{#1,#2}_{\!\!-\infty,\infty}}

\newcommand{\clippedNN}[4]{\ClippedRealV{#1}{#2}{#3}{#4}}

\newcommand{\CovNum}[1]{\cfadd{def:covering_number}\deflink{def:covering_number}{\mathcal C}^{#1}}
\newcommand{\CovRad}[1]{\cfadd{def:covering_radius}\deflink{def:covering_radius}{\mathcal C}_{#1}}
\newcommand{\PackNum}[1]{\cfadd{def:packing_number}\deflink{def:packing_number}{\mathcal P}^{#1}}
\newcommand{\PackRad}[1]{\cfadd{def:packing_radius}\deflink{def:packing_radius}{\mathcal P}_{#1}}

\newcommand{\logistic}{\cfadd{def:logistic1}\deflink{def:logistic1}{\mathfrak{l}}}

\newcommand{\logfunction}{standard logistic activation function\cfadd{def:logistic1}}

\newcommand{\tanhh}{\cfadd{def:hyperbolic_tangent1}\deflink{def:hyperbolic_tangent1}{\tanh}}

\newcommand{\softplus}{\cfadd{def:softplus1}\deflink{def:softplus1}{\mathfrak{s}}}

\newcommand{\softplusfunc}{softplus activation function\cfadd{def:softplus1}}

\newcommand{\heavisidefunction}{Heaviside activation function\cfadd{def:heaviside}}

\newcommand{\eLU}{\ELU\ activation function\cfadd{def:eLU}}

\newcommand{\rePU}{\RePU\ activation function\cfadd{def:rePU}}

\newcommand{\idMatrix}{\cfadd{def:identityMatrix}\operatorname{\deflink{def:identityMatrix}{I}}}

\newcommand{\ANNs}{\cfadd{def:ANN}\deflink{def:ANN}{\mathbf{N}}}

\newcommand{\activation}{a}
\newcommand{\Activation}{A}
\newcommand{\activationDim}[1]{\multdim_{\activation,#1}}
\newcommand{\functionANN}[1]{\cfadd{def:ANNrealization}\deflink{def:ANNrealization}{\mathcal{R}}^{\mathbf{N}}_{#1}}
\newcommand{\paramANN}{\cfadd{def:ANN}\deflink{def:ANN}{\mathcal{P}}}

\newcommand{\lengthANN}{\cfadd{def:ANN}\deflink{def:ANN}{\mathcal{L}}}
\newcommand{\inDimANN}{\cfadd{def:ANN}\deflink{def:ANN}{\mathcal{I}}}
\newcommand{\compANN}[2]{\cfadd{def:ANNcomposition}{#1 \deflink{def:ANNcomposition}{\bullet} #2}}
\newcommand{\powANN}[2]{\cfadd{def:iteratedANNcomposition}#1^{\deflink{def:iteratedANNcomposition}{\bullet} #2}}
\newcommand{\compANNbullet}{\cfadd{def:ANNcomposition} \deflink{def:ANNcomposition}{\bullet}} %

\newcommand{\paraANN}[1]{\parallelizationSpecial_{#1}}

\newcommand{\outDimANN}{\cfadd{def:ANN}\deflink{def:ANN}{\mathcal{O}}}
\newcommand{\longerANN}[1]{\cfadd{def:ANNenlargement}\mathcal{E}_{#1}}

\newcommand{\idANNshort}[1]{\mathbb{I}_{#1}}
\newcommand{\dims}{\cfadd{def:ANN}\deflink{def:ANN}{\mathcal{D}}}
\newcommand{\hiddenLength}{\cfadd{def:ANN}\deflink{def:ANN}{\mathcal{H}}}
\newcommand{\hiddenDimId}{\mathfrak{i}}
\newcommand{\parallelization}{\cfadd{def:generalParallelization}\operatorname{P}}
\newcommand{\parallelizationSpecial}{\cfadd{def:simpleParallelization}\deflink{def:simpleParallelization}{\mathbf{P}}}

\newcommand{\scalarMultANN}[2]{\cfadd{def:ANNscalar}#1\deflink{def:ANNscalar}{\circledast}#2}

\newcommand{\qandq}{\qquad\text{and}\qquad}
\newcommand{\andq}{\text{and}\qquad}

\newcommand{\rect}{\cfadd{def:relu1}\deflink{def:relu1}{\mathfrak r}}
\newcommand{\Rect}{\cfadd{def:relu}\deflink{def:relu}{\mathfrak R}}
\newcommand{\multdim}{\cfadd{def:multidim_version}\deflink{def:multidim_version}{\mathfrak M}}

\newcommand{\clip}[2]{\cfadd{def:clip1}\deflink{def:clip1}{\mf c}_{#1,#2}}
\newcommand{\Clip}[3]{\cfadd{def:clip}\deflink{def:clip}{\mf C}_{#1,#2,#3}}

\DeclareMathOperator{\id}{id}

\newcommand{\card}[1]{\abs{#1}}

\newcommand{\transpose}[1]{\cfadd{def:Transpose}#1^{\deflink{def:Transpose}*}}

\newcommand{\ANNsum}{\mathbin{\cfadd{def:ANNsum:same}\deflink{def:ANNsum:same}\oplus}}
\newcommand{\bigANNsum}{\cfadd{def:ANNsum:same}\deflink{def:ANNsum:same}{\bigoplus}}
\newcommand{\smallbigANNsum}{\cfadd{def:ANNsum:same}\mathop{\textstyle\deflink{def:ANNsum:same}{\bigoplus}}\limits}
\newcommand{\ssmallbigANNsum}{\cfadd{def:ANNsum:same}\mathop{\textstyle\deflink{def:ANNsum:same}{\bigoplus}}\nolimits}

\newcommand{\sumANN}{\cfadd{def:ANN:sum}\deflink{def:ANN:sum}{\mathbb{S}}}
\newcommand{\extensionANN}{\cfadd{def:ANN:extension}\deflink{def:ANN:extension}{\mathbb{T}}}

\newcommand{\dimANNlevel}{\cfadd{def:ANN}\deflink{def:ANN}{\mathbb{D}}}

\newcommand{\cB}{\mathcal{B}}

\newcommand{\cD}{\mathcal{D}}
\newcommand{\cE}{\mathcal{E}}
\newcommand{\cF}{\mathcal{F}}
\newcommand{\cG}{\mathcal{G}}

\newcommand{\cK}{\mathcal{K}}
\newcommand{\cL}{\mathcal{L}}
\newcommand{\cM}{\mathcal{M}}
\newcommand{\cN}{\mathcal{N}}

\newcommand{\cS}{\mathcal{S}}

\newcommand{\cX}{\mathcal{X}}
\newcommand{\cY}{\mathcal{Y}}

\newcommand{\fB}{\mathfrak{B}}
\newcommand{\fC}{\mathfrak{C}}
\newcommand{\fD}{\mathfrak{D}}

\newcommand{\fG}{\mathfrak{G}}

\newcommand{\fK}{\mathfrak{K}}

\newcommand{\fN}{\mathfrak{N}}

\newcommand{\fW}{\mathfrak{W}}

\newcommand{\fZ}{\mathfrak{Z}}

\newcommand{\fc}{\mathfrak{c}}
\newcommand{\fd}{\mathfrak{d}}

\newcommand{\ff}{\mathfrak{f}}

\newcommand{\fh}{\mathfrak{h}}

\newcommand{\fx}{\mathfrak{x}}
\newcommand{\fy}{\mathfrak{y}}

\newcommand{\bfd}{\mathbf{d}}

\newcommand{\bfk}{\mathbf{k}}
\newcommand{\bfl}{\mathbf{l}}
\newcommand{\bfm}{\mathbf{m}}

\newcommand{\bfB}{\mathbf{B}}

\newcommand{\bfD}{\mathbf{D}}

\newcommand{\bfJ}{\mathbf{J}}

\newcommand{\bfL}{\mathbf{L}}

\newcommand{\bfR}{\mathbf{R}}

\newcommand{\bfT}{\mathbf{T}}

\newcommand{\bfW}{\mathbf{W}}

\newcommand{\scrC}{\mathscr{C}}

\newcommand{\scrR}{\mathscr{R}}

\newcommand{\bbB}{\ensuremath{\mathbb B}}

\newcommand{\diff}{\mathrm{d}}

\newcommand{\defaultNetDim}{\mathfrak d}
\newcommand{\defaultParamDim}{\mathfrak{d}}
\newcommand{\defaultInputDim}{d} 
\newcommand{\defaultLossFunction}{\mathscr{L}}
\newcommand{\defaultGradientFunction}{\mathscr{G}}

\newcommand{\defaultx}{\mathscr{x}}
\newcommand{\defaulty}{\mathscr{y}}

\newcommand{\defaultStochLoss}{\mathscr{l}}
\newcommand{\defaultStochGradient}{\mathscr{g}}

\newcommand{\Lone}{\mathscr{l}^1}

\newcommand{\emprisk}{\mathscr{R}}

\newcommand{\unif}[1]{#1-uniformly distributed}

\newcommand{\llfloor}{\lfloor}
\newcommand{\rrfloor}{\rfloor}

\DeclareMathOperator*{\ssuml}{\textstyle\sum}

\DeclareMathOperator{\ssumnl}{\textstyle\sum}

\DeclareMathOperator*{\smallsum}{\textstyle\sum}
\newcommand{\SmallSum}[2]{ {\textstyle\sum\limits_{#1}^{#2}}}
\DeclareMathOperator*{\smalltimes}{\textstyle\bigtimes}
\let\smallprod\undefined
\DeclareMathOperator*{\smallprod}{\textstyle\prod}
\DeclareMathOperator*{\smallbigcup}{\textstyle\bigcup}

\DeclareMathOperator{\Trace}{Trace}

\newcommand{\thisbook}{these lecture notes }

\usepackage{scalerel}
\newcommand{\medint}[1]{{\stretchrel*{\scalerel*[2ex]{\int}{\int}}{\sum}}_{ \mkern-10mu \smash{#1} }}

\usepackage[backend=biber,style=numeric-comp,giveninits=true,sorting=nyt,maxbibnames=99,useprefix=true,isbn=false]{biblatex}
\DeclareNameAlias{default}{family-given}
\renewbibmacro{in:}{\ifentrytype{article}{}{\printtext{\bibstring{in}\space}}} %
 
\renewbibmacro*{author}{%
  \ifboolexpr{
    test \ifuseauthor
    and
    not test {\ifnameundef{author}}
  }
    {\textsc{\printnames{author}}%
     \iffieldundef{authortype}
       {}
       {\setunit{\printdelim{authortypedelim}}%
        \usebibmacro{authorstrg}}}
    {}}

\DeclareFieldFormat[article,inbook,incollection,inproceedings,patent,thesis,unpublished]{title}{#1}

\DeclareSourcemap{
  \maps[datatype=bibtex]{
    \map[overwrite]{
      \step[fieldsource=eprint,final=true]
      \step[fieldset=url,fieldvalue={arxiv.org/abs/}]
      \step[fieldset=url,append,origfieldval]
    }
    \map[overwrite]{
      \step[fieldsource=doi,final=true]
      \step[fieldset=url,fieldvalue={doi.org/}]
      \step[fieldset=url,append,origfieldval]
    }
  }
}
 
\newcommand{\urlhttps}[1]{\href{https://#1}{\nolinkurl{#1}}}

\ExplSyntaxOn 

\NewDocumentCommand{\urlnohttps}{m}{
  \tl_set:Nn \l_tmpa_tl {#1}
  \regex_replace_once:nnN { \A https?:// } { } \l_tmpa_tl
  \urlhttps{\l_tmpa_tl}
}
\ExplSyntaxOff

\makeatletter
\newcommand{\transform}[1]{\transform@#1::\@nil}
\def\transform@@prefix{https}
\def\transform@#1://#2://#3\@nil{%
  \if\relax\detokenize{#2}\relax
    \href{https://#1}{\nolinkurl{#1}}%
  \else
    \lowercase{\def\transform@@start{#1}}%
      \ifx\transform@@start\transform@@prefix
        \href{https://#2}{\nolinkurl{#2}}%
      \else
        \href{https://#2}{\nolinkurl{#2}}%
      \fi 
  \fi
}
\makeatother 

\DeclareFieldFormat{url}{\textsc{url:} \urlnohttps{#1}}

\renewbibmacro*{eprint}{}
\DeclareFieldFormat{doi}{}
 
\renewbibmacro*{volume+number+eid}{%
  \printfield{volume}%
  \setunit*{\addcomma\space}%
  \printfield{number}%
}

\DeclareFieldFormat[book]{series}{}

\renewbibmacro*{note+pages}{%
  \printfield{note}%
  \setunit{\bibpagespunct}%
  \printfield{eid}
  \setunit{\addcomma\space}%
  \iffieldundef{pages}%
    {\printfield{pagetotal}}%
    {\printfield{pages}}%
  \newunit}

\NewBibliographyString{artno}
\DefineBibliographyStrings{english}{artno = {Art\adddotspace No\adddot}}

\DeclareFieldFormat[article,periodical]{eid}{\bibstring{artno}\addabbrvspace #1}

\makeatletter
\newcommand{\pushright}[1]{\ifmeasuring@#1\else\omit\hfill$\displaystyle#1$\fi\ignorespaces}
\makeatother

\newcommand\yesnumber{\refstepcounter{equation}\tag{\theequation}}

\usepackage{caption}
\usepackage{newfloat}
\DeclareFloatingEnvironment[fileext=frm,placement={!ht},name=Source Code]{listing}
\DeclareCaptionType{code}[Listing][List of Code Listings]

\usepackage{listings}

\usepackage{color}
 
\definecolor{codeblue}{rgb}{0,0,0.8}
\definecolor{codegray}{rgb}{0.5,0.5,0.5}
\definecolor{codebrown}{rgb}{0.56,0.28,0.16}
\definecolor{backcolour}{rgb}{1,1,1}
\definecolor{codegreen}{rgb}{0.4,0.8,0.3}

\lstdefinestyle{mystyle}{
	language = Python,
	morekeywords = {as, with}, %
	abovecaptionskip=\bigskipamount,
	backgroundcolor=\color{backcolour},   
	keywordstyle=\color{codeblue},
	commentstyle=\color{codegray},
	numberstyle=\tiny,
	stringstyle=\color{codegreen},
	basicstyle=\footnotesize\ttfamily,
	breakatwhitespace=false,         
	breaklines=true,                 
	captionpos=b,
  abovecaptionskip=5pt,              
	keepspaces=true,                 
	numbers=left,  
	frame = single,                  
	numbersep=8pt,                  
	xleftmargin=.25in,
	xrightmargin=.25in,
	showspaces=false,                
	showstringspaces=false,
	showtabs=false,                  
	tabsize=2,
  extendedchars=true,
  literate=%
  {φ}{$\mathgtt{f}$}1%
  {ψ}{$\mathgtt{y}$}1%
  {ρ}{$\mathgtt{r}$}1%
  {𝔼}{$\mathbb{E}$}1%
  {𝓧}{$\mathcal{X}$}1%
  {𝓣}{$\mathcal{T}$}1%
  {∂}{$\partial$}1%
  {√}{$\sqrt{}$}1%
  {τ}{$\mathgtt{t}$}1%
  {ξ}{$\mathgtt{x}$}1%
  {Δ}{$\mathgtt{D}$}1%
  {ᵢ}{${}_i$}1%
  {²}{${}^2$}1%
  {ₓ}{${}_x$}1%
  {𝓝}{$\mathcal{N}$}1%
  {¹}{${}^{1}$}1%
  {ℓ}{$\ell$}1%
}
 
\DeclareCaptionFormat{listing}{#1 (\listingfilename)#2#3}
\ExplSyntaxOn

\seq_new:N \g_listing_seq

\NewDocumentCommand{\addlisting}{m}{
  \seq_gput_right:Nn \g_listing_seq { #1 }
}

\NewDocumentCommand{\directoryoflistings}{}{
  \seq_map_inline:Nn \g_listing_seq {
    \noindent ##1\par
  }
}

\newcommand{\filelisting}[3]{
  \gdef\listingfilename{\url{#2}}
  \seq_gput_right:Nn \g_listing_seq {\cref{#1}:~\url{#2}~\dotfill~\pageref{#1}}
  \lstinputlisting[label=#1,caption={#3}]{#2}
}

\seq_new:N \g_fig_seq

\NewDocumentCommand{\addfig}{m}{
  \seq_gput_right:Nn \g_fig_seq { #1 }
}

\NewDocumentCommand{\directoryoffigures}{}{
  \seq_map_inline:Nn \g_fig_seq {
    \noindent\cref{##1}\tl_if_in:nnTF{##1}{.pdf}{:~\url{##1}}{}~\dotfill~\pageref{##1}\par
  }
}

\seq_new:N \g_exercise_seq

\NewDocumentEnvironment{exercise} { o m } {
  \seq_gput_right:Nn \g_exercise_seq {
    \cref{#2}
    \IfValueTF{#1}{~(#1)}{}
    ~\dotfill~\pageref{#2}
  }
  \question
  \label{#2}
} {
  \endquestion
}

\NewDocumentCommand{\directoryofexercises}{}{
  \seq_map_inline:Nn \g_exercise_seq {
    \noindent ##1\par
  }
}

\cs_new:Nn \parifnotblank:n {
  \tl_if_blank:nTF{#1}{}{(#1)}
}

\cs_generate_variant:Nn \parifnotblank:n { x }

\ExplSyntaxOff

\makeatletter
\newcommand{\parifnotblank}[1]{%
  \protected@edef\@tempa{#1}%
  \expandafter\notblank\expandafter{\@tempa}{ (\url{#1})}{}%
}
\let\includegraphics@old=\includegraphics
\renewcommand{\includegraphics}[2][]{\gdef\figurefilename{#2}\addfig{#2}\includegraphics@old[#1]{#2}}
\let\figure@old=\figure
\renewcommand{\figure}{\gdef\figurefilename{}\figure@old}
\makeatother

\DeclareCaptionFormat{figure}{#1\parifnotblank{\figurefilename}\label{\figurefilename}#2#3}
\newcommand{\AffineANN}{\cfadd{def:ANN:affine}\deflink{def:ANN:affine}{\mathbf{A}}}
\newcommand{\weight}{W}
\newcommand{\bias}{B}
\newcommand{\ii}{\cfadd{def:padding}\deflink{def:padding}{\mathfrak{i}}}
\newcommand{\basicANN}{\mathbf{H}}
\newcommand{\interpolatingDNN}{\mathbf{F}}
\newcommand{\modcont}[1]{\cfadd{mod_cont_def}\deflink{mod_cont_def}{w}_{\mkern-1mu #1}}
\newcommand{\Rr}{\functionANN{\rect}}
\newcommand{\1}{\mathbb{1}}
\newcommand{\interpol}[2]{\cfadd{def:lin_interp}\deflink{def:lin_interp}{\mathscr{L}}_{\!#1}^{#2}}

\newcommand{\convfunc}{\cfadd{def:cont_convolution}\deflink{def:cont_convolution}{\oast}}

\newcommand{\CNNs}{\cfadd{def:CNN}\deflink{def:CNN}{\mathbf{C}}}
\newcommand{\convolution}{\cfadd{def:convolution}\deflink{def:convolution}{\ast}}
\newcommand{\onetensor}{\cfadd{def:onetensor}\deflink{def:onetensor}{\mathbf{I}}}
\newcommand{\CNNRealisation}[1]{\cfadd{def:CNNrealisation}\deflink{def:CNNrealisation}{\mathcal{R}}^{\mathbf{C}}_{#1}}

\newcommand{\unrolling}[3]{\cfadd{def:unrolling}\deflink{def:unrolling}{\mathfrak{R}}_{#1, #2,#3}}

\newcommand{\ResNets}{\cfadd{def:ResNet}\deflink{def:ResNet}{\mathbf{R}}}

\newcommand{\ResNetRealisation}[1]{\cfadd{def:ResNetrealization}\deflink{def:ResNetrealization}{\mathcal{R}}^{\mathbf{R}}_{#1}}

\newcommand{\batchmean}{\cfadd{def:batch_mean}\operatorname{\deflink{def:batch_mean}{Batchmean}}}
\newcommand{\batchvar}{\cfadd{def:batch_var}\operatorname{\deflink{def:batch_var}{Batchvar}}}
\newcommand{\Batchnormop}{\cfadd{def:batch_norm_op}\operatorname{\deflink{def:batch_norm_op}{batchnorm}}}
\newcommand{\Batchnorm}{\cfadd{def:batch_norm}\operatorname{\deflink{def:batch_norm}{Batchnorm}}}
\newcommand{\BNANNs}{\cfadd{def:BNANN}\deflink{def:BNANN}{\mathbf{B}}}
\newcommand{\BNANNsData}{\cfadd{def:BNANNData}\deflink{def:BNANNData}{\mathbf{b}}}
\newcommand{\Normlayers}{N}
\newcommand{\BNANNDataRealisation}[1]{\cfadd{def:BNANNgivenrealisation}\deflink{def:BNANNgivenrealisation}{\mathcal{R}}^{\mathbf{b}}_{#1}}
\newcommand{\BNANNRealisation}[1]{\cfadd{def:BNANNrealisation}\deflink{def:BNANNrealisation}{\mathcal{R}}^{\mathbf{B}}_{#1}}

\newcommand{\variance}{V}

\newcommand{\Targetfunction}{\defaultLossFunction}
\newcommand{\thetavar}[1]{\theta_{#1}}
\newcommand{\xvar}[1]{x_{#1}}
\newcommand{\diag}{\cfadd{def:diag_matrix}\deflink{def:ResNet}{\operatorname{diag}}}

\DeclarePairedDelimiter{\rbr}{(}{)}

\newcommand{\FrechetSubdiff}{\cfadd{def:limit:subdiff}\deflink{def:limit:subdiff}{\mathcal{D}}}
\newcommand{\limitingFrechetSubdiff}{\cfadd{def:limit:subdiff}\deflink{def:limit:subdiff}{\mathscr{D}}}
\newcommand{\NSslope}{\cfadd{def:nonsmoothSlope}\deflink{def:nonsmoothSlope}{\mathbb{S}}}

\newcommand{\const}{\fC}
\newcommand{\constt}{\fD}
\newcommand{\consttt}{\fc}
\newcommand{\with}{\curvearrowleft}
\newcommand{\bL}{\mathbb{L}}
\newcommand{\bB}{\mathbb{B}}

\newcommand{\KKc}[1]{\fK_{#1}\cfadd{def:Kahane_Khintchine}}

\DeclarePairedDelimiterXPP\Pnorm[2]{}\lVert\rVert{_{#1}}{\cfadd{def:p-norm} #2}

\makeatletter
\ExplSyntaxOn

\seq_new:N \g_def_seq
\newcounter{def@label@counter}

\bool_new:N \g_newchapter_bool
\bool_gset_true:N \g_newchapter_bool
\tl_new:N \g_chapter_tl

\def\dotfil{\cleaders\hbox{$\mkern1.5mu.\mkern1.5mu$}\hfil}

\NewDocumentCommand{\adddef}{m m}{
  \bool_if:NT \g_newchapter_bool {
    \seq_gput_right:Nn \g_def_seq {\smallskip}
    \seq_gput_right:Nx \g_def_seq {\noexpand\textbf{\noexpand\cref{\g_chapter_tl}}}
    \seq_gput_right:Nn \g_def_seq {\par}
    \bool_gset_false:N \g_newchapter_bool
  }
  \seq_gput_right:Nn \g_def_seq {\hspace{1.5em}\cref{#1}:~#2
  \dotfil\penalty0 \null\nobreak\dotfill ~ 
  \pageref{#1}\par}
}
\def\dotfil{\cleaders\hbox{$\mkern1.5mu.\mkern1.5mu$}\hfil}

\NewDocumentCommand{\cchapter}{m m}{
  \chapter{#1}
  \label{#2}
  \tl_gset:Nn \g_chapter_tl {#2}
  \bool_gset_true:N \g_newchapter_bool
}

\NewDocumentCommand{\directoryofdefinitions}{}{
  \leftskip=3em
  \parindent=-3em
  \hspace{-3em}\seq_map_inline:Nn \g_def_seq {##1}
}

\ExplSyntaxOff
\makeatother

\usepackage{mdframed}
\newmdenv[
  skipabove=\topsep,
  skipbelow=\topsep,
  backgroundcolor=gray!10,
  linecolor=gray!80,
  linewidth=1pt,
  innertopmargin=0pt,
  innerbottommargin=0.5\baselineskip,
  innerleftmargin=0.5\baselineskip,
  innerrightmargin=0.5\baselineskip
]{boxedthm}

\AtBeginEnvironment{theorem}{\begin{boxedthm}}
\AtEndEnvironment{theorem}{\end{boxedthm}}

\AtBeginEnvironment{lemma}{\begin{boxedthm}}
\AtEndEnvironment{lemma}{\end{boxedthm}}

\AtBeginEnvironment{prop}{\begin{boxedthm}}
\AtEndEnvironment{prop}{\end{boxedthm}}

\AtBeginEnvironment{cor}{\begin{boxedthm}}
\AtEndEnvironment{cor}{\end{boxedthm}}

\newmdenv[
  skipabove=\topsep,
  skipbelow=\topsep,
  backgroundcolor=blue!10,
  linecolor=gray!80,
  linewidth=1pt,
  innertopmargin=0pt,
  innerbottommargin=0.5\baselineskip,
  innerleftmargin=0.5\baselineskip,
  innerrightmargin=0.5\baselineskip
]{boxeddef}

\AtBeginEnvironment{definition}{\begin{boxeddef}}
\AtEndEnvironment{definition}{\end{boxeddef}}

\newtheoremstyle{plainn}
  {.5\topsep}   %
  {\topsep}   %
  {\itshape}  %
  {0pt}       %
  {\bfseries} %
  {.}         %
  {5pt plus 1pt minus 1pt} %
  {}          %

\theoremstyle{plainn}
\newtheorem{theorem}{Theorem}[section]
\newtheorem{lemma}[theorem]{Lemma}
\newtheorem{prop}[theorem]{Proposition}
\newtheorem{cor}[theorem]{Corollary}

\newtheorem{definition} [theorem]{Definition}

\theoremstyle{remark}
\newtheorem{remark}[theorem]{Remark}

\newtheorem{question}{Exercise}[section]

\DeclarePairedDelimiter{\pr}{(}{)}
\DeclarePairedDelimiter{\cu}{\{}{\}}

\newcommand{\bpr}[1]{\pr[\big]{#1}}
\newcommand{\bbpr}[1]{\pr[\Big]{#1}}
\newcommand{\bbbpr}[1]{\pr[\bigg]{#1}}
\newcommand{\bbbbpr}[1]{\pr[\Bigg]{#1}}
\newcommand{\bbr}[1]{\br[\big]{#1}}
\newcommand{\bbbr}[1]{\br[\Big]{#1}}
\newcommand{\bbbbr}[1]{\br[\bigg]{#1}}
\newcommand{\bbbbbr}[1]{\br[\Bigg]{#1}}
\newcommand{\bcu}[1]{\cu[\big]{#1}}
\newcommand{\bbcu}[1]{\cu[\Big]{#1}}

\usepackage{xparse}

\ExplSyntaxOn

\cs_generate_variant:Nn \str_range_ignore_spaces:nnn { onn }
\cs_generate_variant:Nn \str_range_if_eq:nnTF { onTF }
\cs_generate_variant:Nn \str_range_if_eq:nnTF { vnTF }
\cs_generate_variant:Nn \str_range_if_eq:nnTF { VnTF }
\cs_generate_variant:Nn \str_range_if_eq:nnTF { enTF }
\cs_generate_variant:Nn \str_range_if_eq:nnTF { fnTF }

\NewDocumentCommand{\Henceex}{ o m }{
  \IfValueT{#1}{
    \str_if_eq:noTF {hence} {#1} {
      \bool_gset_true:N \g_hencetherefore
    } {
      \str_if_eq:noTF {Hence} {#1} {
        \bool_gset_true:N \g_hencetherefore
      } {
        \bool_gset_false:N \g_hencetherefore
      }
    }
  }
  \bool_if:nTF { \g_hencetherefore } {
    \bool_gset_false:N \g_hencetherefore
    Hence,~
  } {
    \bool_gset_true:N \g_hencetherefore
    Therefore,~
  }
  we~obtain
  \str_if_eq:eeTF {\str_range_ignore_spaces:onn { #2 } {1} {6}} {forall} 
  {#2 that}{F}
  \IfValueF{#1}{~}
}

\int_new:N \g_furthermore

\NewDocumentCommand{\Moreover}{ o o }{
  \IfValueT{#1}{
    \str_case:nn {#1} {
      {Next} {\int_gset:Nn {\g_furthermore} {0}}      
      {Furthermore} {\int_gset:Nn {\g_furthermore} {1}}
      {Moreover} {\int_gset:Nn {\g_furthermore} {2}}
      {In~addition} {\int_gset:Nn {\g_furthermore} {3}}
      {note} {\bool_gset_true:N \g_noteobserve}
      {observe} {\bool_gset_false:N \g_noteobserve}
    }
    \IfValueT{#2}{
      \str_case:nn {#2} {
        {Next} {\int_gset:Nn {\g_furthermore} {0}}        
        {Furthermore} {\int_gset:Nn {\g_furthermore} {1}}
        {Moreover} {\int_gset:Nn {\g_furthermore} {2}}
        {In~addition} {\int_gset:Nn {\g_furthermore} {3}}
        {note} {\bool_gset_true:N \g_noteobserve}
        {observe} {\bool_gset_false:N \g_noteobserve}
      }
    }
  }
  \int_case:nn { \int_mod:nn {\g_furthermore} {4} } {
    { 0 } { Next~\nobs that}    
    { 1 } { Furthermore,~\nobs that}
    { 2 } { Moreover,~\nobs that}
    { 3 } { In~addition,~\nobs that}
  }
  \int_incr:N \g_furthermore
  \peek_charcode:NTF , {  } { ~ }
}

\ExplSyntaxOff

\NewDocumentEnvironment {adef} {m o} {%
\IfNoValueTF{#2} {
  \begin{definition}\label{#1}\global\def\loc{#1}%
} {
  \begin{definition}[#2]\label{#1}\global\def\loc{#1}%
  \adddef{#1}{#2}
}
}{%
\end{definition}%
}

\ExplSyntaxOn

\NewDocumentCommand{\Itref}{ m m }{
  \clist_set:No \l_localreflist {#2}
  \clist_clear:N \l_reflist
  \clist_map_inline:Nn \l_localreflist { \clist_put_right:Nn \l_reflist {#1.##1} }
  \Cref{\l_reflist}~in~\cref{#1}
}

\ExplSyntaxOff

\newenvironment{eqsplit}{\equation\aligned%
}{\endaligned\endequation%
}

\ExplSyntaxOn

\tl_new:N \l_tmpc_tl

\NewDocumentCommand{\easymath}{ o m }{
  \tl_set:Nn \l_tmpa_tl {#2}
  \tl_replace_all:Nnn \l_tmpa_tl {(} {\mleft(}
  \tl_replace_all:Nnn \l_tmpa_tl {)} {\mright)}
  \tl_replace_all:Nnn \l_tmpa_tl {[} {\mleft[}
  \tl_replace_all:Nnn \l_tmpa_tl {]} {\mright]}
  \tl_replace_all:Nnn \l_tmpa_tl {\{} {\mleft\{}
  \tl_replace_all:Nnn \l_tmpa_tl {\}} {\mright\}}
  \tl_replace_all:Nnn \l_tmpa_tl {||<} {\mleft\lVert}
  \tl_replace_all:Nnn \l_tmpa_tl {>||} {\mright\rVert}
  \tl_replace_all:Nnn \l_tmpa_tl {|<} {\mleft\lvert}
  \tl_replace_all:Nnn \l_tmpa_tl {>|} {\mright\rvert}
  \tl_set_eq:NN \l_tmpb_tl \l_tmpa_tl
  \tl_reverse:N \l_tmpb_tl
  \tl_set:Nx \l_tmpc_tl {\tl_head:N \l_tmpb_tl}
  \str_case:nnTF {#1} {
    { t } {
      \tl_replace_all:Nnn \l_tmpa_tl {&} {}
      \tl_replace_all:Nnn \l_tmpa_tl {\\} {}
      \str_if_in:nVTF {.,} {\l_tmpc_tl} {
        \tl_set:Nx \l_tmpa_tl {\tl_tail:N \l_tmpb_tl}
        \tl_reverse:N \l_tmpa_tl
        \begin{math}
          \tl_use:N \l_tmpa_tl
        \end{math}
        \tl_use:N \l_tmpc_tl
      } {
        \begin{math}
          \tl_use:N \l_tmpa_tl
        \end{math}
      }
    }
    { a } {
      \equation
        \split
          \tl_use:N \l_tmpa_tl
        \endsplit
      \endequation
    } 
  } { } {
    \tl_replace_all:Nnn \l_tmpa_tl {&} {}
    \equation
    \tl_use:N \l_tmpa_tl
    \endequation
  }
}

\NewDocumentCommand{\am}{ o m m }{
  \tl_set:Nn \l_tmpa_tl {#2}
  \tl_replace_all:Nnn \l_tmpa_tl {(} {\mleft(}
  \tl_replace_all:Nnn \l_tmpa_tl {)} {\mright)}
  \tl_replace_all:Nnn \l_tmpa_tl {[} {\mleft[}
  \tl_replace_all:Nnn \l_tmpa_tl {]} {\mright]}
  \tl_replace_all:Nnn \l_tmpa_tl {\{} {\mleft\{}
  \tl_replace_all:Nnn \l_tmpa_tl {\}} {\mright\}}
  \tl_replace_all:Nnn \l_tmpa_tl {||<} {\mleft\lVert}
  \tl_replace_all:Nnn \l_tmpa_tl {>||} {\mright\rVert}
  \tl_replace_all:Nnn \l_tmpa_tl {|<} {\mleft\lvert}
  \tl_replace_all:Nnn \l_tmpa_tl {>|} {\mright\rvert}
  \str_case:nnTF {#1} {
    { t } {
      \tl_replace_all:Nnn \l_tmpa_tl {&} {}
      \tl_replace_all:Nnn \l_tmpa_tl {\\} {}
      \begin{math}
        \tl_use:N \l_tmpa_tl
      \end{math}
      #3
    }
    { a } {
      \equation
        \split
          \tl_use:N \l_tmpa_tl
          \str_if_in:nnT {.,} {#3} {
            #3
          }
        \endsplit
      \endequation
    } 
  } { } {
    \tl_replace_all:Nnn \l_tmpa_tl {&} {}
    \equation
      \tl_use:N \l_tmpa_tl
      \str_if_in:nnT {.,} {#3} {
        #3
      }
    \endequation
  }
}

\cs_new_protected:Nn \check_punct:n {
  \peek_charcode_remove_ignore_spaces:NT . { . #1 }
}

\NewDocumentCommand{\testpunct}{ m }{
  #1 \check_punct:n {something else}
}

\ExplSyntaxOff

\ExplSyntaxOn

\bool_new:N \g_forexample

\NewDocumentCommand{\eg}{ o }{
\IfValueT{#1}{
\str_if_eq:noTF {fe} {#1} {
\bool_gset_true:N \g_forexample
} {\bool_gset_false:N \g_forexample}
}
\bool_if:nTF { \g_forexample } {
\bool_gset_false:N \g_forexample
for~example
}{
\bool_gset_true:N \g_forexample
for~instance
}
}

\ExplSyntaxOff

\makeatletter
\ExplSyntaxOn
\seq_new:N \g_abbrs
\prop_new:N \g_abbr_counts
\tl_new:N \l_abbr_count_tl

\NewDocumentCommand{\abbr}{m m O{#1} m m O{#4}}{
\expandafter\newcommand\csname#3\endcsname[1][]{
\seq_if_in:NnTF \g_abbrs {#1}{
  \prop_get:NnN \g_abbr_counts {#1} \l_abbr_count_tl
  \prop_gput:Nnx \g_abbr_counts {#1} {\int_eval:n {\l_abbr_count_tl + 1}}
  \hyperref[directoryofabbreviations]{#1}
} {
  \seq_gput_left:Nn \g_abbrs {#1}
  \prop_gput:Nnn \g_abbr_counts {#1} {1}
  \expandafter\gdef\csname#1@def\endcsname{#2}
  \phantomsection\label{#1}
  \str_if_eq:nnTF{##1}{}{\emph{#2}}{##1}~(\hyperref[directoryofabbreviations]{#1})
}}
\expandafter\newcommand\csname#6\endcsname[1][]{
\seq_if_in:NnTF \g_abbrs {#1} {
  \prop_get:NnN \g_abbr_counts {#1} \l_abbr_count_tl
  \prop_gput:Nnx \g_abbr_counts {#1} {\int_eval:n {\l_abbr_count_tl + 1}}
  \hyperref[directoryofabbreviations]{#4}
} {
  \expandafter\gdef\csname#1@def\endcsname{#2}
  \seq_gput_left:Nn \g_abbrs {#1}
  \prop_gput:Nnn \g_abbr_counts {#1} {1}
  \phantomsection\label{#1}
  \str_if_eq:nnTF{##1}{}{\emph{#5}}{##1}~(\hyperref[directoryofabbreviations]{#4})
}}}

\newcommand{\directoryofabbreviations}{
  \seq_sort:Nn \g_abbrs {
    \str_compare:nNnTF { ##1 } > { ##2 }
    { \sort_return_swapped: }
    { \sort_return_same: }
  }
  \label{directoryofabbreviations}
  \seq_map_inline:Nn \g_abbrs {##1~(\csname##1@def\endcsname) \dotfill \pageref{##1}
  \\}
}
\ExplSyntaxOff
\makeatother

\abbr{ANN}{artificial neural network}[ann]{ANNs}{artificial neural networks}[anns]
\abbr{ODE}{ordinary differential equation}{ODEs}{ordinary differential equations}
\abbr{PDE}{partial differential equation}{PDEs}{partial differential equations}
\abbr{GD}{gradient descent}{GDs}{gradient descents}
\abbr{SGD}{stochastic gradient descent}{SGDs}{stochastic gradient descents}
\abbr{GF}{gradient flow}{GFs}{gradient flows}
\abbr{ReLU}{rectified linear unit}{ReLUs}{rectified linear units}
\abbr{GELU}{Gaussian error linear unit}{GELUs}{Gaussian error linear units}
\abbr{SiLU}{sigmoid linear unit}{SiLUs}{sigmoid linear units}
\abbr{ELU}{exponential linear unit}{ELUs}{exponential linear units}
\abbr{RePU}{rectified power unit}{RePUs}{rectified power units}[RePUs]
\abbr{KL}{Kurdyka--\L ojasiewicz}{KLs}{Kurdyka--\L ojasiewicz}
\abbr{BN}{batch normalization}{BNs}{batch normalizations}
\abbr{PINN}{physics-informed neural network}{PINNs}{physics-informed neural networks}
\abbr{DGM}{deep Galerkin method}{DGMs}{deep Galerkin methods}
\abbr{DKM}{deep Kolmogorov method}{DKMs}{deep Kolmogorov methods}
\abbr{FNO}{Fourier neural operator}{FNOs}{Fourier neural operators}
\abbr{deepONet}{deep operator network}{deepONets}{deep operator networks}
\abbr{RNN}{recurrent \ann}{RNNs}{recurrent \anns}
\abbr{CNN}{convolutional \ann}[cnn]{CNNs}{convolutional \anns}[cnns]
\abbr{ResNet}{residual \ann}[resnet]{ResNets}{residual \anns}[resnets]
\abbr{LSTM}{long short-term memory}{LSTMs}{?????}
\abbr{NLP}{natural language processing}{NLPs}{?????}
\abbr{PCA}{principal component analysis}{PCAs}{?????}
\abbr{LLM}{large language model}{LLMs}{large language models}
\abbr{GPT}{generative pre-trained transformer}{GPTs}{generative pre-trained transformers}
\abbr{BERT}{Bidirectional Encoder Representations from Transformers}{BERTs}{?????}
\abbr{GNN}{graph neural network}{GNNs}{graph neural networks}
\abbr{COD}{curse of dimensionality}{CODs}{?????}
\abbr{SDE}{stochastic differential equation}{SDEs}{stochastic differential equations}
\abbr{BSDE}{backward stochastic differential equation}{BSDEs}{backward stochastic differential equations}
\abbr{FBSDE}{forward backward stochastic differential equation}{FBSDEs}{forward backward stochastic differential equations}
\abbr{PIDE}{partial integro-differential equation}{PIDEs}{partial integro-differential equations}
\abbr{CV}{computer vision}{CVs}{?????}
\abbr{cPINN}{conservative \PINN}{cPINNs}{conservative \PINNs}
\abbr{BNN}{Bayesian neural network}{BNNs}{Bayesian neural networks}
\abbr{fPINN}{fractional \PINN}{fPINNs}{fractional \PINNs}
\abbr{PPINN}{parareal \PINN}{PPINNs}{parareal \PINNs}
\abbr{XPINN}{extended \PINN}{XPINNs}{extended \PINNs}
\abbr{TGNN}{theory-guided neural network}{TGNNs}{theory-guided neural networks}
\abbr{WAN}{weak adversarial network}{WANs}{weak adversarial networks}
\abbr{VPINN}{variational \PINN}{VPINNs}{variational \PINNs}
\abbr{NSFnet}{Navier-Stokes flow net}{NSFnets}{Navier-Stokes flow nets}
\abbr{DST}{discrete sine transform}{DSTs}{discrete sine transforms}
\abbr{DCT}{discrete cosine transform}{DCTs}{discrete cosine transforms}
\abbr{D3M}{deep domain decomposition method}[DDDM]{D3Ms}{deep domain decomposition methods}[DDDMs]
\abbr{MscaleDNN}{multi-scale deep neural network}{MscaleDNNs}{multi-scale deep neural networks}

\abbr{Adagrad}{adaptive gradient}{Adagrads}{adaptive gradients}
\abbr{RMSprop}{root mean square propagation}{RMSprops}{root mean square propagations}
\abbr{Adam}{adaptive moment estimation}{ADAMs}{adaptive moment estimations}
\abbr{AdamW}{\Adam\ with decoupled weight decay}{AdamWs}{\Adam\ with decoupled weight decays}
\abbr{Nadam}{Nesterov-accelerated adaptive moment estimation}{NADAMs}{Nesterov-accelerated adaptive moment estimations}
\abbr{Muon}{momentum orthogonalized by Newton-Schulz}{Muons}{momentum orthogonalized by Newton-Schulz}

\newtcolorbox[auto counter, number within=section]{myalgorithm}[3][]{%
    enhanced,
    breakable,
    fonttitle=\bfseries,
    title=Algorithm~\refstepcounter{theorem}\thetheorem: #2,
    label={#3},  %
    label type=algorithmcounter,
    #1,
    colframe=black,
    colback=white,
    coltitle=black,
    colbacktitle=white,
    sharp corners,
    boxrule=0.5pt,
    boxsep=2mm,
    top=0mm,
    bottom=2mm,
    left=0mm,
    right=2mm
}
\newcommand{\myalgorithmLine}{
\noindent\hspace*{-2mm}\rule{\dimexpr\linewidth+6mm}{0.4pt}
\vspace{-3mm}
}
\crefname{algorithmcounter}{Algorithm}{Algorithms}
\Crefname{algorithmcounter}{Algorithm}{Algorithms}
\crefname{line}{line}{lines}
\Crefname{line}{Line}{Lines}
\algtext*{EndFor}
\algtext*{EndFunction}
\algtext*{EndIf}

\newcommand{\mycomment}[1]{\hfill \textcolor{gray}{\textit{\# #1}}}

\newcommand{\first}{1\textsuperscript{st}\xspace}
\newcommand{\second}{2\textsuperscript{nd}\xspace}
\newcommand{\third}{3\textsuperscript{rd}\xspace}
\newcommand{\fourth}{4\textsuperscript{th}\xspace}
\newcommand{\ith}{$i$\textsuperscript{th}\xspace}
\newcommand{\nth}[1]{#1\textsuperscript{th}\xspace}

\newcommand{\oneNormANN}{\deflink{def:dnn:l1norm}{\mathbb{L}}\cfadd{def:dnn:l1norm}}
\newcommand{\maxANN}{\cfadd{def:max_d}\deflink{def:max_d}{\mathbb{M}}}

\DeclarePairedDelimiter{\ceil}{\lceil\cfadd{def:ceiling}}{\rceil}
\newcommand{\bceil}[1]{\ceil[\big]{#1}}
\DeclarePairedDelimiter{\floor}{\lfloor\cfadd{def:ceiling}}{\rfloor}

\DeclareMathOperator{\logg}{log}

\newcommand{\biasANN}[2]{ \cfadd{def:ANN}\deflink{def:ANN}{\mathcal{B}}_{#1, #2}}
\newcommand{\weightANN}[2]{ \cfadd{def:ANN}\deflink{def:ANN}{\mathcal{W}}_{#1, #2}}

\newcommand{\providecommandordefault}[2]{%
    \providecommand{#1}{}%
    \renewcommand{#1}{#2}%
}

\def\ifUnDefinedCs#1{\expandafter\ifx\csname#1\endcsname\relax}

\DTMnewdatestyle{monthyear}{

}
\newcommand{\monthyearnow}{
  \DTMsetdatestyle{monthyear}
  \today
}

\excludecomment{firstVcomment}
\excludecomment{overviewComment}

\newcommand{\beq}{\begin{equation}}
\newcommand{\eeq}{\end{equation}}

\newcommand{\I}{{\rm{I}}}

\ExplSyntaxOn

\NewDocumentCommand{\mEE}{ o m }{
  \IfValueTF{#1}{
    \str_case:on {#1} {
      {0}{\mathbb E\br{#2}}
      {1}{\mathbb E\br[\big]{#2}}
      {2}{\mathbb E\mkern-1.1mu\br[\Big]{#2}}
      {3}{\mathbb E\mkern-1.3mu\br[\bigg]{#2}}
      {4}{\mathbb E\mkern-1.5mu\br[\Bigg]{#2}}
    }
  } {
    \mathbb E\br{#2}
  }
}

\ExplSyntaxOff

\usepackage{mleftright}

\usepackage{mathtools}

\makeatletter
\newcommand{\mylabel}[2]{#2\def\@currentlabel{#2}\label{#1}}
\makeatother

\renewcommand{\P}{\mathbb{P}}

\usepackage{mleftright}

\renewenvironment{pmatrix}{\mleft(\begin{matrix}}{\end{matrix}\mright)}

\usepackage{subcaption}

\usepackage{marginfix}

    \def\cK{{\mathcal K}}

\DeclarePairedDelimiterX{\lossmetric}[1]{\lvert\!\lvert\!\lvert}{\rvert\!\rvert\!\rvert}{#1}

\newcommand{\loss}{L_\text{Base}}

\abbr{MLMC}{multilevel Monte Carlo}{MLMCs}{multilevel Monte Carlos}
\abbr{QMC}{quasi--Monte Carlo}{QMCs}{quasi--Monte Carlos}
\abbr{LRV}{learning the random variables}{LRVs}{learning the random variables}
\abbr{IKNO}{integral kernel neural operator}{IKNOs}{integral kernel neural operators}
\abbr{PINO}{physics--informed neural operator}{PINOs}{physics--informed neural operators}

\newcommand{\cpoint}{\vartheta}
\newcommand{\altpoint}{\theta}
\newcommand{\altpointTwo}{v}
\newcommand{\altpointThree}{w}
\newcommand{\altpointFour}{u}

\newcommand{\Altpoint}{\Theta}
\newcommand{\AltpointTwo}{V}

%% file: Argument_command.tex
\ExplSyntaxOn

\seq_new:N \g_cflist_loaded
\seq_new:N \g_cflist_pending

\NewDocumentCommand{\cfadd} { m } {
	\seq_if_in:NnF \g_cflist_loaded { #1 } {
		\seq_if_in:NnF \g_cflist_pending { #1 } {
			\seq_gput_right:Nn \g_cflist_pending { #1 }
		}
	}
}

\NewDocumentCommand{\cfconsiderloaded} { m } {
	\seq_gput_right:Nn \g_cflist_loaded {#1}
}

\NewDocumentCommand{\cfremove} { m } {
	\seq_gremove_all:Nn \g_cflist_pending { #1 }
}

\NewDocumentCommand{\cfload} { o } {
	\seq_if_empty:NTF \g_cflist_pending {
		\IfValueTF{#1}{\ignorespaces}{\unskip}
	} {
		(cf.\ \cref{\seq_use:Nn \g_cflist_pending {,}})\IfValueTF{#1}{#1~}{\unskip}
		\seq_gconcat:NNN \g_cflist_loaded \g_cflist_loaded \g_cflist_pending
		\seq_gclear:N \g_cflist_pending
		\IfValueT{#1}{\ignorespaces}
	}
}

\NewDocumentCommand{\cfclear} {} {
	\seq_gclear:N \g_cflist_loaded
	\seq_gclear:N \g_cflist_pending
}

\NewDocumentCommand{\cfout} { o } {
	\seq_if_empty:NTF \g_cflist_pending {\unskip\IfValueT{#1}{\ignorespaces}} {
		(cf.\ \cref{\seq_use:Nn \g_cflist_pending {,}})\IfValueTF{#1}{#1~}{\unskip}
		\seq_gclear:N \g_cflist_pending
		\IfValueT{#1}{\ignorespaces}
	}
}

\NewDocumentCommand{\ifnocf} { m } {
	\seq_if_empty:NT \g_cflist_pending { #1 }
}

\ExplSyntaxOff

\ExplSyntaxOn

\bool_new:N \g_noteobserve

\NewDocumentCommand{\setnote}{}{
  \bool_gset_true:N \g_noteobserve
}

\NewDocumentCommand{\setobserve}{}{
  \bool_gset_false:N \g_noteobserve
}

\NewDocumentCommand{\nobs}{ o }{
  \IfValueT{#1}{
    \str_if_eq:noTF {note} {#1} {
      \bool_gset_true:N \g_noteobserve
    } {
      \str_if_eq:noTF {Note} {#1} {
        \bool_gset_true:N \g_noteobserve
      } {
        \bool_gset_false:N \g_noteobserve
      }
    }
  }
  \bool_if:nTF { \g_noteobserve } {
    \bool_gset_false:N \g_noteobserve
    note
  } {
    \bool_gset_true:N \g_noteobserve
    observe
  }
  \IfValueF{#1}{~}
}

\NewDocumentCommand{\Nobs}{ o }{
  \IfValueT{#1}{
    \str_if_eq:noTF {note} {#1} {
      \bool_gset_true:N \g_noteobserve
    } {
      \str_if_eq:noTF {Note} {#1} {
        \bool_gset_true:N \g_noteobserve
      } {
        \bool_gset_false:N \g_noteobserve
      }
    }
  }
  \bool_if:nTF { \g_noteobserve } {
    \bool_gset_false:N \g_noteobserve
    Note
  } {
    \bool_gset_true:N \g_noteobserve
    Observe
  }
  \IfValueF{#1}{~}
}

\ExplSyntaxOff

\ExplSyntaxOn

\bool_new:N \g_hencetherefore

\NewDocumentCommand{\hence}{ o }{
  \IfValueT{#1}{
    \str_if_eq:noTF {hence} {#1} {
      \bool_gset_true:N \g_hencetherefore
    } {
      \str_if_eq:noTF {Hence} {#1} {
        \bool_gset_true:N \g_hencetherefore
      } {
        \bool_gset_false:N \g_hencetherefore
      }
    }
  }
  \bool_if:nTF { \g_hencetherefore } {
    \bool_gset_false:N \g_hencetherefore
    hence
  } {
    \bool_gset_true:N \g_hencetherefore
    therefore
  }
  \IfValueF{#1}{~}
}

\NewDocumentCommand{\Hence}{ o }{
  \IfValueT{#1}{
    \str_if_eq:noTF {hence} {#1} {
      \bool_gset_true:N \g_hencetherefore
    } {
      \str_if_eq:noTF {Hence} {#1} {
        \bool_gset_true:N \g_hencetherefore
      } {
        \bool_gset_false:N \g_hencetherefore
      }
    }
  }
  \bool_if:nTF { \g_hencetherefore } {
    \bool_gset_false:N \g_hencetherefore
    Hence,~we~obtain
  } {
    \bool_gset_true:N \g_hencetherefore
    Therefore,~we~obtain
  }
  \IfValueF{#1}{~}
}

\ExplSyntaxOff

\ExplSyntaxOn

\seq_const_from_clist:Nn \g_prove_mru {
	establish,
	demonstrate,
	prove,
	show,
	imply,
	ensure
}

\prop_new:N \l__verbs
\prop_put:Nnn \l__verbs {show} {shows}
\prop_put:Nnn \l__verbs {imply} {implies}
\prop_put:Nnn \l__verbs {demonstrate} {demonstrates}
\prop_put:Nnn \l__verbs {prove} {proves}
\prop_put:Nnn \l__verbs {establish} {establishes}
\prop_put:Nnn \l__verbs {ensure} {ensures}
\prop_put:Nnn \l__verbs {assure} {assures}

\tl_new:N \g_wordtmp
\seq_new:N \l_mytmps

\cs_generate_variant:Nn \str_if_in:nnTF { nVTF }
\cs_generate_variant:Nn \str_if_in:nnTF { xVTF }

\NewDocumentCommand{\prove}{ o }{
	\IfValueTF{#1}{
		\seq_clear:N \l_mytmps
		\seq_map_inline:Nn \g_prove_mru {
			\str_if_eq:nnTF {##1} {ensure} {
				\str_set:Nn \l_temps {n}
			} {
				\str_set:Nx \l_temps {\str_head_ignore_spaces:n {##1}}
			}
			\str_if_in:xVTF {#1} \l_temps {
				\seq_put_right:Nn \l_mytmps {##1}
			} { }
		}
		\seq_get_right:NN \l_mytmps \g_wordtmp
	} {
		\seq_get_right:NN \g_prove_mru \g_wordtmp
	}
	\tl_use:N \g_wordtmp
	\IfValueTF{#1}{}{~}
	\seq_gput_left:NV \g_prove_mru \g_wordtmp
	\seq_gremove_duplicates:N \g_prove_mru
}

\NewDocumentCommand{\proves}{ o }{
	\IfValueTF{#1}{
		\seq_clear:N \l_mytmps
		\seq_map_inline:Nn \g_prove_mru {
			\str_if_eq:nnTF {##1} {ensure} {
				\str_set:Nn \l_temps {n}
			} {
				\str_set:Nx \l_temps {\str_head_ignore_spaces:n {##1}}
			}
			\str_if_in:xVTF {#1} \l_temps {
				\seq_put_right:Nn \l_mytmps {##1}
			} { }
		}
		\seq_get_right:NN \l_mytmps \g_wordtmp
	} {
		\seq_get_right:NN \g_prove_mru \g_wordtmp
	}
	\str_set:NV \l_tmpa_str \g_wordtmp
	\prop_get:NVN \l__verbs \l_tmpa_str \l_tmpa_tl
	\tl_use:N \l_tmpa_tl
	\IfValueTF{#1}{}{~}
	\seq_gput_left:NV \g_prove_mru \g_wordtmp
	\seq_gremove_duplicates:N \g_prove_mru
}

\ExplSyntaxOff

\ExplSyntaxOn

\newcommand{\llabel}[1]{\savelabel{#1}\label{\loc.#1}\ignorespaces}

\clist_new:N \l_localreflist
\clist_new:N \l_reflist

\NewDocumentCommand{\lref} { m } {
  \clist_set:No \l_localreflist {#1}
  \clist_clear:N \l_reflist
  \clist_map_inline:Nn \l_localreflist { \clist_put_right:Nn \l_reflist {\loc.##1} }
  \cref{\l_reflist}
}

\NewDocumentCommand{\Lref} { m } {
  \clist_set:No \l_localreflist {#1}
  \clist_clear:N \l_reflist
  \clist_map_inline:Nn \l_localreflist { \clist_put_right:Nn \l_reflist {\loc.##1} }
  \Cref{\l_reflist}
}

\NewDocumentCommand{\itref}{ m m }{
  \clist_set:No \l_localreflist {#2}
  \clist_clear:N \l_reflist
  \clist_map_inline:Nn \l_localreflist { \clist_put_right:Nn \l_reflist {#1.##1} }
  \cref{\l_reflist}~in~\cref{#1}
}

\ExplSyntaxOff

\ExplSyntaxOn

\seq_new:N \l_enum_seq
\int_new:N \l_num_items

\bool_new:N \g_commaused_bool

\providecommand{\comma}{}

\cs_new:Nn \enum_it:nn {
  \int_case:nnF {\l_num_items - #1} {
    {0} {
      \renewcommand{\comma}{}
      #2\space
    }
    {1} {
      \bool_gset_false:N \g_commaused_bool
      \renewcommand{\comma}{,~\bool_gset_true:N \g_commaused_bool}
      #2
      \bool_if:NTF \g_commaused_bool {} {,~}
      and~
    }
  } {
    \bool_gset_false:N \g_commaused_bool
    \renewcommand{\comma}{,~\bool_gset_true:N \g_commaused_bool}
    #2
    \bool_if:NTF \g_commaused_bool {} {,~}
  }
}

\cs_new:Nn \enum_it_U:nn {
  \int_case:nnF {\l_num_items - #1} {
    {0} {
      \renewcommand{\comma}{}
      #2
      \space
    }
    {1} {
      \bool_gset_false:N \g_commaused_bool
      \renewcommand{\comma}{,~\bool_gset_true:N \g_commaused_bool}
      #2
      \bool_if:NTF \g_commaused_bool {} {,~}
      and~
    }
  } {
    \bool_gset_false:N \g_commaused_bool
    \renewcommand{\comma}{,~\bool_gset_true:N \g_commaused_bool}
    \int_compare:nTF {#1=1} {
      \text_titlecase_first:n {#2}
    } {
      #2
    }
    \bool_if:NTF \g_commaused_bool {} {,~}
  }
}

\cs_generate_variant:Nn \tl_if_eq:nnTF {onTF}

\cs_new:Nn \enum:nnnn {
  \seq_set_split:Nnn \l_enum_seq ; {#1}
  \seq_remove_all:Nn \l_enum_seq { }
  \seq_remove_all:Nn \l_enum_seq {#2}
  \seq_log:N \l_enum_seq
  \int_set:Nn \l_num_items {\seq_count:N \l_enum_seq}
  \int_log:N \l_num_items
  \int_case:nnF {\l_num_items} {
    { 0 } { 0 }
    { 1 } {
      \IfBooleanTF{#4} {
        \tl_set:Nn \l_text_case_exclude_arg_tl {\cref}
        \bool_set_false:N \l_text_titlecase_check_letter_bool
        \text_titlecase_first:n {\seq_use:Nn \l_enum_seq {}}
      } {
        \seq_use:Nn \l_enum_seq {}
      }
      \space
      \tl_if_eq:onTF{#3}{-}{}{
        \bool_if:NTF \l_plural_bool {
          \prove[#3]~
        } {
          \proves[#3]~
        }
      }
    }
    { 2 } {
      \IfBooleanTF{#4} {
        \tl_set:Nn \l_text_case_exclude_arg_tl {\cref}
        \bool_set_false:N \l_text_titlecase_check_letter_bool
        \text_titlecase_first:n {\seq_item:Nn \l_enum_seq {1}}
        {} ~and~
        \seq_item:Nn \l_enum_seq {2}
      } {
        \seq_use:Nn \l_enum_seq {~and~}
      }
      \space
      \tl_if_eq:onTF{#3}{-}{}{
        \prove[#3]~
      }
    }
  } {
    \IfBooleanTF{#4} {
      \tl_set:Nn \l_text_case_exclude_arg_tl {\cref}
      \seq_indexed_map_function:NN \l_enum_seq \enum_it_U:nn
    } {
      \seq_indexed_map_function:NN \l_enum_seq \enum_it:nn
    }
    \tl_if_eq:onTF{#3}{-}{}{
      \prove[#3]~
    }
  }
}

\cs_generate_variant:Nn \enum:nnnn {nxnn}
\cs_generate_variant:Nn \enum:nnnn {nxxn}

\NewDocumentCommand{\enum}{O{} m O{-} s}{
  \IfBooleanTF{#4}{
    \enum:nxnn {#2} {#1} {sindep} \BooleanFalse
  } {
    \enum:nxxn {#2} {#1} {#3} \BooleanFalse
  }
}

\NewDocumentCommand{\dott}{}{\ifnocf{.}\space}

\bool_new:N \g_arg_start_bool
\bool_gset_true:N \g_arg_start_bool

\NewDocumentCommand{\startnewargseq}{}{\bool_gset_true:N \g_arg_start_bool \tl_set:Nn \g_label_tl {}}

\cs_generate_variant:Nn \seq_if_in:NnTF {NxTF}
\cs_generate_variant:Nn \seq_remove_all:Nn {Nx}

\int_new:N \l_random_int

\cs_generate_variant:Nn \tl_if_head_eq_catcode:nNTF {oNTF}
\cs_generate_variant:Nn \tl_if_head_eq_catcode:nNTF {VNTF}

\cs_generate_variant:Nn \tl_if_head_eq_catcode:nNTF {eNTF}
\cs_generate_variant:Nn \tl_log:n {o}
\cs_generate_variant:Nn \tl_log:n {f}
\cs_generate_variant:Nn \tl_log:n {x}
\cs_generate_variant:Nn \tl_log:n {e}

\cs_generate_variant:Nn \tl_if_in:nnTF {onTF}
\cs_generate_variant:Nn \tl_if_in:NnTF {NeTF}

\cs_generate_variant:Nn \tl_if_head_eq_meaning:nNTF {VNTF}

\seq_const_from_clist:Nn \g_arg_mru_this {
	Ahpr,
	Tapr,
	Ctapr,
	H
}

\seq_const_from_clist:Nn \g_arg_mru_nothis {
	Ia,
	Nwc,
	N,
	Itns,
	Fm,
	Itnswc,
	Mo
}

\seq_new:N \l_arg_seq
\tl_new:N \l_cons_tl
\tl_new:N \l_dummy_tl

\bool_new:N \g_debug_bool
\bool_gset_false:N \g_debug_bool

\bool_new:N \l_insidearg_bool

\bool_new:N \g_firstargletter_bool

\sys_gset_rand_seed:n {0903}

\bool_new:N \l_plural_bool
\tl_new:N \l_arg_verbs_tl

\NewDocumentCommand{\argument}{mom}{
	\bool_set_false:N \l_plural_bool
	\tl_set:Nn \l_arg_verbs_tl {sindep}
	\keys_define:nn { benno/argument } {
		plural .value_forbidden:n = true,
		plural .code:n = {\bool_set_true:N \l_plural_bool},
		verbs .value_required:n = false,
		verbs .tl_set:N = \l_arg_verbs_tl,
	}
	\IfValueT{#2}{
		\keys_set:nn { benno/argument } {#2}
	}
	\bool_log:N \l_plural_bool
	\bool_gset_true:N \l_insidearg_bool
	\seq_set_split:Nnn \l_arg_seq ; {#1}
	\seq_remove_all:Nn \l_arg_seq { }
	\seq_log:N \l_arg_seq
	\tl_set:Nn \l_cons_tl {#3}
	\tl_trim_spaces:N \l_cons_tl
	\seq_if_in:NxTF \l_arg_seq {\lref{\g_label_tl}} {
		\seq_remove_all:Nx \l_arg_seq {\lref{\g_label_tl}}
		\seq_get_left:NNTF \l_arg_seq \l_dummy_tl {
			\tl_trim_spaces:N \l_dummy_tl
			\bool_gset_false:N \g_firstargletter_bool
			\tl_if_head_eq_catcode:nNTF \l_dummy_tl a {
				\bool_gset_true:N \g_firstargletter_bool
			} {
				\tl_if_head_eq_meaning:VNTF \l_dummy_tl {\cref} {
					\tl_set:Nx \l_tmpa_tl {\tl_tail:N \l_dummy_tl}
					\tl_set:Nx \l_tmpb_tl {\tl_head:N \l_tmpa_tl}
					\bool_gset_true:N \g_firstargletter_bool
					\tl_if_in:NeTF \l_tmpb_tl {lem\c_colon_str} {} {
						\tl_if_in:NeTF \l_tmpb_tl {thm\c_colon_str} {} {
							\tl_if_in:NeTF \l_tmpb_tl {prop\c_colon_str} {} {
								\tl_if_in:NeTF \l_tmpb_tl {cor\c_colon_str} {} {
									\bool_gset_false:N \g_firstargletter_bool
								}
							}
						}
					}
				} {
				}
			}
			\bool_if:NTF \g_firstargletter_bool {
				\seq_set_eq:NN \l_tmpa_seq \g_arg_mru_this
				\seq_remove_all:Nn \l_tmpa_seq {H}
				\seq_get_right:NN \l_tmpa_seq \l_tmpa_tl
				\int_case:nnF {\seq_count:N \l_arg_seq} {
					{1} {
						\str_case:VnF {\l_tmpa_tl} {
							{Ahpr} {
								\bool_if:NT \g_debug_bool {C1.1}
								\seq_gput_left:Nn \g_arg_mru_this {Ahpr}
								\seq_gremove_duplicates:N \g_arg_mru_this
								\enum:nxnn {#1} {\lref{\g_label_tl}} {-} {\BooleanTrue}
								\hence~
								\bool_if:NTF \l_plural_bool {
									\prove[\l_arg_verbs_tl]~\ignorespaces #3
								} {
									\proves[\l_arg_verbs_tl]~\ignorespaces #3
								}
							}
							{Tapr} {
								\bool_if:NT \g_debug_bool {C1.2}
								\seq_gput_left:Nn \g_arg_mru_this {Tapr}
								\seq_gremove_duplicates:N \g_arg_mru_this
								\enum[\lref{\g_label_tl}]{
									This;
									#1
								}[\l_arg_verbs_tl]\ignorespaces #3
							}
							{Ctapr} {
								\bool_if:NT \g_debug_bool {C1.3}
								\seq_gput_left:Nn \g_arg_mru_this {Ctapr}
								\seq_gremove_duplicates:N \g_arg_mru_this
								Combining~
								\enum[\lref{\g_label_tl}]{
									this;
									#1
								} \proves[\l_arg_verbs_tl]~\ignorespaces #3
							}
						} {}
					}
				} {
					\str_case:VnF {\l_tmpa_tl} {
						{Ahpr} {
							\bool_if:NT \g_debug_bool {C2.1}
							\seq_gput_left:Nn \g_arg_mru_this {Ahpr}
							\seq_gremove_duplicates:N \g_arg_mru_this
							\enum:nxnn {#1} {\lref{\g_label_tl}} {-} {\BooleanTrue}
							\hence~
							\prove[\l_arg_verbs_tl]~\ignorespaces #3
						}
						{Tapr} {
							\bool_if:NT \g_debug_bool {C2.2}
							\seq_gput_left:Nn \g_arg_mru_this {Tapr}
							\seq_gremove_duplicates:N \g_arg_mru_this
							\enum[\lref{\g_label_tl}]{
								This;
								#1
							}[\l_arg_verbs_tl]\ignorespaces #3
						}
						{Ctapr} {
							\int_case:nn {\int_rand:nn {0} {1}} {
								{0} {
									\bool_if:NT \g_debug_bool {C2.3}
									\seq_gput_left:Nn \g_arg_mru_this {Ctapr}
									\seq_gremove_duplicates:N \g_arg_mru_this
									Combining~
									\enum[\lref{\g_label_tl}]{
										this;
										#1
									} \proves[\l_arg_verbs_tl]~\ignorespaces #3
								}
								{1} {
									\bool_if:NT \g_debug_bool {C2.4}
									\seq_gput_left:Nn \g_arg_mru_this {Ctapr}
									\seq_gremove_duplicates:N \g_arg_mru_this
									Combining~
									\enum:nxnn {#1} {\lref{\g_label_tl}} {-} {\BooleanFalse}
									\hence~
									\proves[\l_arg_verbs_tl]~\ignorespaces #3
								}
							}
						}
					} {}
				}
			} {
				\seq_set_eq:NN \l_tmpa_seq \g_arg_mru_this
				\seq_remove_all:Nn \l_tmpa_seq {H}
				\seq_remove_all:Nn \l_tmpa_seq {Ahpr}
				\seq_get_right:NN \l_tmpa_seq \l_tmpa_tl
				\int_case:nnF {\seq_count:N \l_arg_seq} {
					{1} {
						\str_case:VnF {\l_tmpa_tl} {
							{Tapr} {
								\bool_if:NT \g_debug_bool {C3.1}
								\seq_gput_left:Nn \g_arg_mru_this {Tapr}
								\seq_gremove_duplicates:N \g_arg_mru_this
								\enum[\lref{\g_label_tl}]{
									This;
									#1
								}[\l_arg_verbs_tl]\ignorespaces #3
							}
							{Ctapr} {
								\bool_if:NT \g_debug_bool {C3.2}
								\seq_gput_left:Nn \g_arg_mru_this {Ctapr}
								\seq_gremove_duplicates:N \g_arg_mru_this
								Combining~
								\enum[\lref{\g_label_tl}]{
									this;
									#1
								} \proves[\l_arg_verbs_tl]~\ignorespaces #3
							}
						} {}
					}
				} {
					\str_case:VnF {\l_tmpa_tl} {
						{Tapr} {
							\bool_if:NT \g_debug_bool {C4.1}
							\seq_gput_left:Nn \g_arg_mru_this {Tapr}
							\seq_gremove_duplicates:N \g_arg_mru_this
							\enum[\lref{\g_label_tl}]{
								This;
								#1
							}[\l_arg_verbs_tl]\ignorespaces #3		
						}
						{Ctapr} {
							\int_case:nn {\int_rand:nn {0} {1}} {
								{0} {
									\bool_if:NT \g_debug_bool {C4.2}
									\seq_gput_left:Nn \g_arg_mru_this {Ctapr}
									\seq_gremove_duplicates:N \g_arg_mru_this
									Combining~
									\enum[\lref{\g_label_tl}]{
										this;
										#1
									} \proves[\l_arg_verbs_tl]~\ignorespaces #3		
								}
								{1} {
									\bool_if:NT \g_debug_bool {C4.3}
									\seq_gput_left:Nn \g_arg_mru_this {Ctapr}
									\seq_gremove_duplicates:N \g_arg_mru_this
									Combining~
									\enum:nxnn {#1} {\lref{\g_label_tl}} {-} {\BooleanFalse}
									\hence~
									\proves[\l_arg_verbs_tl]~\ignorespaces #3    
								}
							}
						}
					} {}
				}
			}
		} {
			\tl_if_head_eq_catcode:oNTF \l_cons_tl a {
				\seq_set_eq:NN \l_tmpa_seq \g_arg_mru_this
				\seq_remove_all:Nn \l_tmpa_seq {Ctapr}
				\seq_remove_all:Nn \l_tmpa_seq {Ahpr}
				\seq_get_right:NN \l_tmpa_seq \l_tmpa_tl
				\str_case:VnF {\l_tmpa_tl} {
					{H} {
						\bool_if:NT \g_debug_bool {C5.1}
						\seq_gput_left:Nn \g_arg_mru_this {H}
						\seq_gremove_duplicates:N \g_arg_mru_this
						Hence,~we~obtain~\ignorespaces #3
					}
					{Tapr} {
						\bool_if:NT \g_debug_bool {C5.2}
						\seq_gput_left:Nn \g_arg_mru_this {Tapr}
						\seq_gremove_duplicates:N \g_arg_mru_this
						This~\proves[\l_arg_verbs_tl]~\ignorespaces #3
					}
				} {}
			} {
				\bool_if:NT \g_debug_bool {C6.1}
				\seq_gput_left:Nn \g_arg_mru_this {Tapr}
				\seq_gremove_duplicates:N \g_arg_mru_this
				This~\proves[\l_arg_verbs_tl]~\ignorespaces #3
			}
		} 
	} {
		\int_compare:nNnTF {\seq_count:N \l_arg_seq} = {0} {
			\bool_if:NTF \g_arg_start_bool {
				\bool_if:NT \g_debug_bool {C7.1}
				\Nobs\unskip
				#3
			} {
				\bool_if:NT \g_debug_bool {C7.2}
				\Moreover~
				#3
			}
		} {
			\bool_if:NTF \g_arg_start_bool {
				\bool_if:NT \g_debug_bool {C8.1}
				\tl_log:N \l_arg_verbs_tl
				\Nobs~that~
				\enum{
					#1
				}[\l_arg_verbs_tl]\ignorespaces #3
			} {
				\int_compare:nNnTF {\seq_count:N \l_arg_seq} = {1} {
					\seq_set_eq:NN \l_tmpa_seq \g_arg_mru_nothis
					\seq_remove_all:Nn \l_tmpa_seq {Nwc}
					\seq_remove_all:Nn \l_tmpa_seq {Itnswc}
					\seq_get_right:NN \l_tmpa_seq \l_tmpa_tl
				} {
					\seq_get_right:NN \g_arg_mru_nothis \l_tmpa_tl
				}
				\str_case:VnF {\l_tmpa_tl} {
					{Mo} {
						\bool_if:NT \g_debug_bool {C9.1}
						\seq_gput_left:Nn \g_arg_mru_nothis {Mo}
						\seq_gremove_duplicates:N \g_arg_mru_nothis
						Moreover,~\nobs~that~
						\enum{
							#1
						}[\l_arg_verbs_tl]\ignorespaces #3		
					}
					{Fm} {
						\bool_if:NT \g_debug_bool {C9.2}
						\seq_gput_left:Nn \g_arg_mru_nothis {Fm}
						\seq_gremove_duplicates:N \g_arg_mru_nothis
						Furthermore,~\nobs~that~
						\enum{
							#1
						}[\l_arg_verbs_tl]\ignorespaces #3		
					}
					{Ia} {
						\bool_if:NT \g_debug_bool {C9.3}
						\seq_gput_left:Nn \g_arg_mru_nothis {Ia}
						\seq_gremove_duplicates:N \g_arg_mru_nothis
						In~addition,~\nobs~that~
						\enum{
							#1
						}[\l_arg_verbs_tl]\ignorespaces #3		
					}
					{N} {
						\bool_if:NT \g_debug_bool {C9.4}
						\seq_gput_left:Nn \g_arg_mru_nothis {N}
						\seq_gremove_duplicates:N \g_arg_mru_nothis
						Next,~\nobs~that~
						\enum{
							#1
						}[\l_arg_verbs_tl]\ignorespaces #3		
					}
					{Itns} {
						\bool_if:NT \g_debug_bool {C9.5}
						\seq_gput_left:Nn \g_arg_mru_nothis {Itnswc}
						\seq_gput_left:Nn \g_arg_mru_nothis {Itns}
						\seq_gremove_duplicates:N \g_arg_mru_nothis
						In~the~next~step~we~\nobs~that~
						\enum{
							#1
						}[\l_arg_verbs_tl]\ignorespaces #3		
					}
					{Nwc} {
						\bool_if:NT \g_debug_bool {C9.6}
						\seq_gput_left:Nn \g_arg_mru_nothis {Nwc}
						\seq_gremove_duplicates:N \g_arg_mru_nothis
						Next~we~combine~
						\enum{
							#1
						}to~obtain~\ignorespaces #3
					}
					{Itnswc} {
						\bool_if:NT \g_debug_bool {C9.7}
						\seq_gput_left:Nn \g_arg_mru_nothis {Itns}
						\seq_gput_left:Nn \g_arg_mru_nothis {Itnswc}
						\seq_gremove_duplicates:N \g_arg_mru_nothis
						In~the~next~step~we~combine~
						\enum{
							#1
						}to~obtain~\ignorespaces #3
					}
				} {}
			}
		}
	}
	\bool_gset_false:N \g_arg_start_bool
	\bool_gset_false:N \l_insidearg_bool
	\cfload[.]%
}

\tl_new:N \g_label_tl
\tl_gset:Nn \g_label_tl { }

\NewDocumentCommand{\savelabel}{m}{
	\bool_if:NTF \l_insidearg_bool {
		\tl_gset:Nn \g_label_tl {#1}
	} {
		\tl_gset:Nn \g_label_tl { }
	}
}

\ExplSyntaxOff

\ExplSyntaxOn

\NewDocumentEnvironment {athm} {m m o} {
\str_if_eq:noTF {example} {#1} {
  \bool_gset_true:N \g_example_bool
} {
  \bool_gset_false:N \g_example_bool
}
\cfclear
\IfNoValueTF{#3}{
\begin{#1}\label{#2}\global\def\loc{#2}
}{
\begin{#1}[#3]\label{#2}\global\def\loc{#2}
}
}{
\end{#1}
}

\NewDocumentEnvironment{aproof} {} {
\bool_if:NTF \g_example_bool {
  \bool_gset_true:N \g_arg_start_bool
  \begin{proof}[Proof~for~\cref{\loc}]
} {
  \bool_gset_true:N \g_arg_start_bool
  \begin{proof}[Proof~of~\cref{\loc}]
}
\bool_gset_false:N \g_finishproof_bool
}{
\bool_if:NTF \g_finishproof_bool {}
{\finishproofthus}
\end{proof}
}

\NewDocumentCommand{\finishproofthus} {} {
  \bool_gset_true:N \g_finishproof_bool 
  \bool_if:NTF \g_example_bool {
    The~proof~for~\cref{\loc}~is~thus~complete.
  } {
    The~proof~of~\cref{\loc}~is~thus~complete.
  }
}
\NewDocumentCommand{\finishproofthis} {} {
  \bool_gset_true:N \g_finishproof_bool 
  \bool_if:NTF \g_example_bool {
    This~completes~the~proof~for~\cref{\loc}.
  } {
    This~completes~the~proof~of~\cref{\loc}.
  }
}

\ExplSyntaxOff

%% file: parts/Preface.tex
\section*{Preface}
\addcontentsline{toc}{chapter}{Preface}

This book aims to provide an introduction to the topic of deep learning algorithms.
Very roughly speaking, when we speak of a \emph{deep learning algorithm} we think of a computational scheme which aims to approximate certain relations, functions, or quantities by means of so-called deep \anns\ and the iterated use of some kind of data.
\anns, in turn, can be thought of as classes of functions that consist of multiple compositions of certain nonlinear functions, which are referred to as \emph{activation functions}, and certain affine functions. 
Loosely speaking, the depth of such \anns\ corresponds to the number of involved iterated compositions in the \ann\ and one starts to speak of \emph{deep} \anns\ when the number of involved compositions of nonlinear and affine functions is larger than two.

We hope that this book will be useful for students and scientists who do not yet have any background in deep learning at all and would like to gain a solid foundation as well as for practitioners who would like to obtain a firmer mathematical understanding of the objects and methods considered in deep learning.

After a brief \hyperref[sec:intro]{introduction},
this book is divided into six parts (see \cref{part:ANNs,part:approx,part:opt,part:generalization,part:overall,part:MLforPDEs}).
In \cref{part:ANNs} we introduce in \cref{chapter:dnns} different types of \anns\
including 
\emph{fully-connected feedforward \anns},
\cnns,
\RNNs, and
\resnets\
in all mathematical details
and in \cref{chapter:ANN_calc} we present a certain calculus for fully-connected feedforward \anns.

In \cref{part:approx} we present several mathematical results that analyze how well \anns\ can approximate given functions.
To make this part more accessible, we first restrict ourselves in \cref{sect:onedApprox} to one-dimensional functions from the reals to the reals
and, 
thereafter, 
we study \ann\ approximation results for multivariate functions in \cref{sect:multidApprox}.

A key aspect of deep learning algorithms is usually to model or reformulate the problem under consideration as a suitable optimization problem involving deep \anns.
It is precisely the subject of \cref{part:opt} to study such and related optimization problems and the corresponding optimization algorithms to approximately solve such problems in detail.
In particular, in the context of deep learning methods such optimization problems -- typically given in the form of a minimization problem -- are usually solved by means of appropriate \emph{gradient based} optimization methods.
Roughly speaking, we think of a gradient based optimization method as a computational scheme which aims to solve the considered optimization problem  by performing successive steps based on the direction of the (negative) gradient of the function which one wants to optimize.
Deterministic variants of such gradient based optimization methods such as the \GD\ optimization method
are reviewed and studied in \cref{chapter:deterministic} %
and 
stochastic variants of such gradient based optimization methods such as the \SGD\ optimization method
are reviewed and studied in \cref{chapter:stochastic}. %
\GD-type and \SGD-type optimization methods
can, roughly speaking, be viewed as time-discrete approximations of 
solutions of suitable \GF\ \ODEs. 
To develop intuitions for \GD-type and \SGD-type optimization methods
and for some of the tools which we employ to analyze such methods, 
we study in \cref{chapter:flow} such \GF\ \ODEs. 
In particular, we show in \cref{chapter:flow} how such \GF\ \ODEs\ can be used to approximately 
solve appropriate optimization problems.
Implementations of the gradient based methods discussed in \cref{chapter:deterministic,chapter:stochastic} require efficient computations of gradients.
The most popular and in some sense most natural method to explicitly compute such gradients in the case of the training of \anns\ is the \emph{backpropagation} method, which we derive and present in detail in \cref{chapter:backprop}. %
The mathematical analyses for gradient based optimization methods that we present in \cref{chapter:flow,chapter:deterministic,chapter:stochastic} are in almost all cases too restrictive to cover optimization problems associated to the training of \anns. 
However, such optimization problems %
can be covered by the \KL\ approach which we discuss in detail in \cref{chapter:KL}. %
In \cref{chapter:BN} we rigorously review \BN\ methods, which are popular methods that aim to accelerate \ann\ training procedures in data-driven learning problems.
In \cref{chapter:optimization_error} we review and study the approach to optimize an objective function through different random initializations.

The mathematical analysis of deep learning algorithms does not only consist of
error estimates for approximation capacities of \anns\ (cf.\ \cref{part:approx}) 
and of error estimates for the involved optimization methods (cf.\ \cref{part:opt})
but also requires estimates for the \emph{generalization error} which, roughly speaking, arises 
when the probability distribution associated to the learning problem cannot be accessed explicitly but is approximated by a finite number of realizations/data.
It is precisely the subject of \cref{part:generalization} to study the generalization error. 
Specifically, 
in \cref{sect:probabilistic_generalization} we review suitable probabilistic generalization error estimates
and 
in \cref{sec:generalisation_error} we review suitable strong $L^p$-type generalization error estimates.

In \cref{part:overall} we illustrate how to combine 
parts of the \emph{approximation error} estimates from \cref{part:approx},
parts of the \emph{optimization error} estimates from \cref{part:opt}, and 
parts of the \emph{generalization error} estimates from \cref{part:generalization}
to establish estimates for the overall error
in the exemplary situation of the training of \anns\ based on \SGD-type optimization methods with many independent random initializations.
Specifically, in \cref{sect:overall_error_decomp} we present a suitable overall
error decomposition for supervised learning problems, 
which we employ in \cref{sec:composed_error} together with some of the findings of \cref{part:approx,part:opt,part:generalization} 
to establish the aforementioned illustrative overall error analysis.

Deep learning methods have not only become very popular for data-driven learning problems, but are nowadays also heavily used for approximately solving \PDEs.
In \cref{part:MLforPDEs} we review and implement three popular variants of such deep learning methods for \PDEs.
Specifically, in \cref{subsec:dgm} we treat \PINNs\ and \DGMs\
and in \cref{sect:deepKolmogorov} we treat \DKMs.

This book contains a number of \hyperref[chapter:listofcodes]{\textsc{Python} source codes}, 
which can be downloaded from two sources, namely from
the public GitHub repository at
\begin{equation*}
\begin{split} 
\text{\url{https://github.com/introdeeplearning/book}}
\end{split}
\end{equation*} 
and from
the arXiv page of this
book (by
clicking on the link ``Other formats'' and then on ``Download
source''). For ease of reference, the caption of each source listing in this book
contains the filename of the corresponding source file.

This book grew out of a series of lectures held by the authors at ETH Zurich, University of M\"unster, and the Chinese University of Hong Kong, Shenzhen.
It is in parts based on recent joint articles of
Christian Beck,
Sebastian Becker,
Weinan E,
Lukas Gonon,
Robin Graeber,
Philipp Grohs,
Fabian Hornung,
Martin Hutzenthaler,
Nor Jaafari,
Joshua Lee Padgett,
Adrian Riekert,
Diyora Salimova,
Timo Welti, 
and
Philipp Zimmermann
with the authors of this book.
We thank all of our aforementioned co-authors for very fruitful collaborations.
Special thanks are due to Timo Welti for his permission to integrate slightly modified extracts of the article \cite{JentzenWelti2023} into this book.
We also thank Lukas Gonon, Timo Kr\"oger, Siyu Liang, and Joshua Lee Padget for several insightful discussions and useful suggestions. 
Finally, we thank the students of the courses 
that we held on the basis of preliminary material of this book
for bringing several typos to our notice.

This work has been partially funded by the National Science Foundation of China (NSFC) under grant number 12250610192.
Moreover, this work was supported by the internal project fund from the Shenzhen Research Institute of Big Data under grant T00120220001.
The first author gratefully acknowledges the support of the Cluster of Excellence EXC 2044-390685587, Mathematics M\"unster: Dynamics-Geometry-Structure funded by the Deutsche Forschungsgemeinschaft (DFG, German Research Foundation).

\vspace{1cm}

\noindent
Shenzhen and M\"unster, \hfill Arnulf Jentzen\\
\monthyearnow \hfill  Benno Kuckuck\\ 
\phantom{a} \hfill Philippe von Wurstemberger\\

%% file: parts/Introduction.tex
\chapter*{Introduction}
\addcontentsline{toc}{chapter}{Introduction}
\label{sec:intro}

\begingroup
\providecommandordefault{\inputDim}{\defaultInputDim}
\providecommandordefault{\netDim}{{\defaultNetDim}}
\providecommandordefault{\LossFunction}{\defaultLossFunction}
\providecommandordefault{\targetFunction}{\cE}
\providecommandordefault{\empRiskInifite}{\mathfrak{L}}
\providecommandordefault{\x}{\defaultx}
\providecommandordefault{\y}{\defaulty}

Very roughly speaking, 
the field \emph{deep learning} can 
be divided into three subfields, 
deep \emph{supervised learning}, deep \emph{unsupervised learning}, 
and deep \emph{reinforcement learning}. 
Algorithms in deep supervised learning often seem to be 
most accessible for a mathematical analysis. 
In the following we briefly sketch in a simplified situation some ideas of deep supervised 
learning.

Let $ \inputDim, M \in \N = \{ 1, 2, 3, \dots \} $, 
$ \targetFunction \in C( \R^\inputDim, \R ) $,
$ \x_1, \x_2, \dots, \x_{ M + 1 } \in \R^\inputDim $, 
$ \y_1, \y_2, \dots, \y_M \in \R $ 
satisfy for all $ m \in \{ 1, 2, \dots, M \} $ that
\begin{equation}
\label{eq:f_function}
  \y_m = \targetFunction( \x_m )
  .
\end{equation}
In the framework described in the previous sentence 
we think of $ M \in \N $ as the number of available known input-output data pairs, 
we think of $ \inputDim \in \N $ as the dimension of the input data, 
we think of $ \targetFunction \colon \R^\inputDim \to \R $ as an unknown function 
which relates input and output data through \cref{eq:f_function}, 
we think of $ \x_1, \x_2, \dots, \x_{ M + 1 } \in \R^\inputDim $ as the available known 
input data, and we think of $ \y_1, \y_2, \dots, \y_M \in \R $ as the available 
known output data. 

In the context of a learning problem of the type \cref{eq:f_function}
the objective then is  to approximately compute the output 
$ \targetFunction( \x_{ M + 1 } ) $ of the $ ( M + 1 ) $-th input data $ \x_{ M + 1 } $ 
without using explicit knowledge of 
the function $ \targetFunction \colon \R^\inputDim \to \R $ 
but instead by using the knowledge of 
the $ M $ input-output data pairs 
\begin{equation}
\begin{split} 
  ( \x_1, \y_1 ) = ( \x_1, \targetFunction( \x_1 ) ),\,\,
  ( \x_2, \y_2 ) = ( \x_2, \targetFunction( \x_2 ) ),
  \,\,\dots\,\, , 
  ( \x_M, \y_M ) = ( \x_M, \targetFunction( \x_M ) )
  \in \R^\inputDim \times \R.
\end{split}
\end{equation}
To accomplish this, 
one considers the 
optimization problem of
computing approximate minimizers
of the function 
$ \empRiskInifite \colon C( \R^\inputDim, \R ) \to [0,\infty) $
which satisfies for all 
$
  \phi 
  \in C( \R^\inputDim, \R )
$
that
\begin{equation}
	\label{eq:intro.2}
  \empRiskInifite( \phi )
  =
  \frac1M\bbbbbr{\sum_{ m = 1 }^M 
  \abs*{ \phi( \x_m ) - \y_m }^2}
  .
\end{equation}
Observe that \cref{eq:f_function} ensures that 
$ \empRiskInifite( \targetFunction ) = 0 $ 
and, in particular, we have that 
the unknown function 
$ \targetFunction \colon \R^\inputDim \to \R $ in \cref{eq:f_function} above 
is a minimizer of the function 
\begin{equation}
\begin{split} 
\empRiskInifite \colon C( \R^\inputDim, \R ) \to [0,\infty). 
\end{split}
\end{equation}
The optimization problem of
computing approximate minimizers
of the function $ \empRiskInifite $ is not 
suitable for discrete numerical computations 
on a computer as the function $ \empRiskInifite $
is defined on the infinite-dimensional vector 
space $ C( \R^\inputDim, \R ) $. 

To overcome this we introduce a spatially 
discretized version of this optimization problem. 
More specifically, let 
$ \netDim \in \N $,
let
$ 
  \psi = ( \psi_{ \theta } )_{ \theta \in \R^{ \netDim } } 
  \colon 
  \R^{ \netDim }
  \to
  C( \R^\inputDim, \R ) 
$
be a function, 
and let 
$
  \LossFunction \colon \R^{ \netDim }
  \to [0,\infty)
$
satisfy
\begin{equation}
\begin{split} 
  \LossFunction = \empRiskInifite \circ \psi.
\end{split}
\end{equation}
We think of the set 
\begin{equation}
\label{eq:set}
  \cu*{
    \psi_{ \theta }  \colon \theta \in \R^{ \netDim }
  }
  \subseteq C( \R^\inputDim, \R )
\end{equation}
as a parametrized set of functions 
which we employ to approximate the infinite-dimensional  
vector space $ C( \R^\inputDim, \R ) $
and we think of the function 
\begin{equation}
\label{intro:eq1}
\begin{split} 
\R^{ \netDim } \ni \theta \mapsto \psi_{ \theta } \in C( \R^\inputDim, \R )
\end{split}
\end{equation}
as the parametrization function associated to this set. 
For example, in the case $d=1$ one could think of \cref{intro:eq1} as the parametrization function associated to polynomials in the sense that for all
	$\theta = (\theta_1, \ldots, \theta_{\netDim}) \in \R^\netDim$,
	$x \in \R$
it holds that
\begin{equation}
\begin{split} 
	\psi_{ \theta }(x)
=
	\sum_{k = 0}^{\netDim-1}
	\theta_{k+1} x^{k}
\end{split}
\end{equation}
or one could think of \cref{intro:eq1} as the parametrization associated to trigonometric polynomials.
However, in the context of \emph{deep supervised learning} one neither chooses \cref{intro:eq1} as parametrization of polynomials
nor as parametrization of trigonometric polynomials, but instead one chooses \cref{intro:eq1} as a parametrization associated to \emph{deep} \anns.
In \cref{chapter:dnns} in \cref{part:ANNs} we present different types of such deep \ann\ parametrization functions in all mathematical details.

Taking the set in \cref{eq:set} and its parametrization function 
in \cref{intro:eq1}
into account, we then intend to  
compute approximate minimizers of the function 
$ 
  \empRiskInifite
$
restricted to the set 
$
  \cu*{
    \psi_{ \theta } \colon \theta \in \R^{ \netDim }
  }
$, that is, we consider the optimization problem 
of computing approximate minimizers
of the function
\begin{equation}
\label{intro:eq2}
  \cu*{
    \psi_{ \theta } 
    \colon
    \theta \in \R^{ \netDim }
  }
  \ni 
  \phi 
  \mapsto 
  \empRiskInifite( \phi )
  =
	\frac1M
  \br*{
    \sum_{ m = 1 }^M 
    \abs*{\phi( \x_m ) - \y_m }^2
  }
  \in [0,\infty)
  .
\end{equation}
Employing the parametrization function 
in \cref{intro:eq1},
one can also reformulate the optimization problem 
in \cref{intro:eq2}
as the 
optimization problem of 
computing approximate minimizers of the function 
\begin{equation}
\label{eq:optimization_final}
  \R^{ \netDim }
  \ni 
  \theta 
  \mapsto 
  \LossFunction( \theta )
  =
  \empRiskInifite( \psi_{ \theta } )
  =
	\frac1M
  \br*{
    \sum_{ m = 1 }^M 
    \abs*{ \psi_{ \theta }( \x_m ) - \y_m }^2
  }
  \in [0,\infty)
\end{equation}
and this optimization problem now has the potential to be amenable for discrete numerical computations.
In the context of deep supervised learning,
where one chooses
the parametrization function 
in \cref{intro:eq1}
as deep \ann\ parametrizations, 
one would apply an \SGD-type optimization algorithm 
to the optimization problem in \cref{eq:optimization_final} to 
compute approximate minimizers of \cref{eq:optimization_final}. 
In \cref{chapter:stochastic} in \cref{part:opt} we present the most common variants of such \SGD-type optimization algorithms.
If $\vartheta \in \R^\netDim$ is an approximate minimizer of \eqref{eq:optimization_final} in the sense that
$
	\LossFunction(\vartheta) 
\approx
	\inf_{\theta \in \R^{\netDim}} \LossFunction( \theta )
$,
one then considers $\psi_{ \vartheta }( \x_{M+1} )$ as an approximation
\begin{equation}
\begin{split} 
	\psi_{ \vartheta }( \x_{M+1} ) \approx  \targetFunction( \x_{M+1} )
\end{split}
\end{equation}
of the unknown output $\targetFunction( \x_{M+1} )$ of the $(M+1)$-th input data $\x_{M+1}$.
We note that in deep supervised learning algorithms one typically aims to compute an approximate minimizer $\vartheta \in \R^{\netDim}$ of \cref{eq:optimization_final} in the sense that
$
	\LossFunction(\vartheta) 
\approx
	\inf_{\theta \in \R^{\netDim}} \LossFunction( \theta )
$,
which is, however, typically not a minimizer of \cref{eq:optimization_final} in the sense that 
$
	\LossFunction(\vartheta) 
=
	\inf_{\theta \in \R^{\netDim}} \LossFunction( \theta )
$
(cf.\ \cref{sec:KL_for_ANNs}).

In \cref{eq:intro.2} above we have set up an optimization problem for the learning problem by using the standard mean squared error function to measure the loss. This \emph{mean squared error loss function} is just one possible example in the formulation of deep learning optimization problems.
In particular, in image classification problems other loss functions such as the \emph{cross-entropy loss function} are often used and we refer to \cref{chapter:flow} of \cref{part:opt} for a survey of commonly used loss function in deep learning algorithms (see \cref{sect:MSE}).
We also refer to \cref{chapter:KL} for convergence results in the above framework where the parametrization function in \cref{intro:eq1} corresponds to \emph{fully-connected feedforward} \anns\ (see \cref{sec:KL_for_ANNs}).

\endgroup

%% file: parts/Basics_on_ANNs.tex
\cchapter{Basics on ANNs}{chapter:dnns}

In this chapter
	we review different types of architectures of \anns\ such as 
		fully-connected feedforward \anns\ (see \cref{subsec:vectorized_description,subsec:structured_description}), 
		\cnns\  (see \cref{section:cnns}), 
		\resnets\ (see \cref{section:resnets}), and 
		\RNNs\ (see \cref{section:rnns}), 
we review different types of popular activation functions used in applications such as 
	the \ReLU\ activation (see \cref{subsect:Relu}), 
	the \GELU\ activation (see \cref{subsect:GELU}),
	and
	the standard logistic activation (see \cref{subsect:logistic})
	among others,
and
we review different procedures for how \anns\ can be formulated in rigorous mathematical terms 
	(see \cref{subsec:vectorized_description} for a vectorized description and \cref{subsec:structured_description} for a structured description).

In the literature different types of \ann\ architectures and activation functions have been reviewed in several excellent works; cf., for example, 
\cite{bach2023learning,shalev2014understanding,Goodfellow2016,zhang2023dive,bishop1995neural,Deisenroth2020,hastie2009elements,grohs2023mathematical,sutton2018reinforcement,Schmidhuber2015,Caterini2018,alpaydin2020introduction,Calin2020,Blum2020} and the references therein.
The specific presentation of \cref{subsec:vectorized_description,subsec:structured_description} is based on
\cite{Beck2019published,GononGraeberJentzen2023,BeckJafaari21,GrohsHornung2023,BeckWeinanJentzen2019}.

\begingroup

\newcommand{\netDim}{\defaultNetDim}

\section{Fully-connected feedforward ANNs (vectorized description)}
\label{subsec:vectorized_description} 

We start the mathematical content of this book with a review of fully-connected feedforward \anns, the most basic type of \anns.
Roughly speaking, fully-connected feedforward \anns\ can be thought of as parametric functions resulting from successive compositions of affine functions followed by nonlinear functions,
where the parameters of a fully-connected feedforward \ann\ correspond to all the entries of the linear transformation matrices and translation vectors of the involved affine functions
(cf.\ \cref{def:FFNN} below for a precise definition of fully-connected feedforward \anns\ and \cref{fig:deepANN} below for a graphical illustration of fully-connected feedforward \anns).
The linear transformation matrices and translation vectors are sometimes called \emph{weight matrices} and \emph{bias vectors}, respectively, and can be thought of as the \emph{trainable parameters} of fully-connected feedforward \anns\ (cf.\ \cref{weight_and_bias} below).

In this section we introduce in \cref{def:FFNN} below a \emph{vectorized description} of fully-connected feedforward \anns\  in the sense that all the trainable parameters of a fully-connected feedforward \ann\ are represented by the components of a single Euclidean vector.
In \cref{subsec:structured_description} below we will discuss an alternative way to describe fully-connected feedforward \anns\ in which the trainable parameters of a fully-connected feedforward \ann\ are represented by a tuple of matrix-vector pairs corresponding to the weight matrices and bias vectors of the fully-connected feedforward \anns\ (cf.\ \cref{def:ANN,def:ANNrealization} below).

\begin{figure}[!ht]
	\hspace{-1cm}
	\begin{tikzpicture}[shorten >=1pt,-latex,draw=black!100, node distance=\layersep,auto, scale=0.92]
			\def\layersep{2.9cm}
			\def\neuronsep{1.8cm}
			\def\ninput{3}
			\def\nhidden{5}
			\def\noutput{3}

			\tikzmath{
				\ninputl=\ninput-1;
				\nhiddenl=\nhidden-1;
				\noutputl=\noutput-1;
				\nmax=max(\ninput,\nhidden,\noutput);
			}

			\foreach \inode in {1,...,\ninputl}
				\node[input neuron] (I-\inode) at (0,.5*\ninput*\neuronsep-\inode*\neuronsep+\neuronsep) {$\inode$};
			\node (I-dots) at (0,.5*\ninput*\neuronsep-\ninputl*\neuronsep) {$\vdots$};
			\node[input neuron] (I-\ninput) at (0,.5*\ninput*\neuronsep-\ninput*\neuronsep) {$l_0$};

			\foreach \inode in {1,...,\nhiddenl}
				\node[hidden neuron] (H1-\inode) at (\layersep,.5*\nhidden*\neuronsep-\inode*\neuronsep+\neuronsep) {$\inode$};
			\node (H1-dots) at (\layersep,.5*\nhidden*\neuronsep-\nhiddenl*\neuronsep) {$\vdots$};
			\node[hidden neuron] (H1-\nhidden) at (\layersep,.5*\nhidden*\neuronsep-\nhidden*\neuronsep) {$l_1$};

			\foreach \inode in {1,...,\nhiddenl}
				\node[hidden neuron] (H2-\inode) at (2*\layersep,.5*\nhidden*\neuronsep-\inode*\neuronsep+\neuronsep) {$\inode$};
			\node (H2-dots) at (2*\layersep,.5*\nhidden*\neuronsep-\nhiddenl*\neuronsep) {$\vdots$};
			\node[hidden neuron] (H2-\nhidden) at (2*\layersep,.5*\nhidden*\neuronsep-\nhidden*\neuronsep) {$l_2$};

			\foreach \inode in {1,...,\nhiddenl}
				\node (Hdots-\inode) at (3*\layersep,.5*\nhidden*\neuronsep-\inode*\neuronsep+\neuronsep) {$\cdots$};
			\node (Hdots-dots) at (3*\layersep,.5*\nhidden*\neuronsep-\nhiddenl*\neuronsep) {$\ddots$};
			\node (Hdots-\nhidden) at (3*\layersep,.5*\nhidden*\neuronsep-\nhidden*\neuronsep) {$\cdots$};

			\foreach \inode in {1,...,\nhiddenl}
				\node[hidden neuron] (H3-\inode) at (4*\layersep,.5*\nhidden*\neuronsep-\inode*\neuronsep+\neuronsep) {$\inode$};
			\node (H3-dots) at (4*\layersep,.5*\nhidden*\neuronsep-\nhiddenl*\neuronsep) {$\vdots$};
			\node[hidden neuron] (H3-\nhidden) at (4*\layersep,.5*\nhidden*\neuronsep-\nhidden*\neuronsep) {$l_{L-1}$};

			\foreach \inode in {1,...,\noutputl}
				\node[output neuron] (O-\inode) at (5*\layersep,.5*\noutput*\neuronsep-\inode*\neuronsep+\neuronsep) {$\inode$};
			\node (O-dots) at (5*\layersep,.5*\noutput*\neuronsep-\noutputl*\neuronsep) {$\vdots$};
			\node[output neuron] (O-\noutput) at (5*\layersep,.5*\noutput*\neuronsep-\noutput*\neuronsep) {$l_L$};

			\foreach \inode in {1,...,\ninput}
				\foreach \hnode in {1,...,\nhidden}
					\path (I-\inode) edge (H1-\hnode);

			\foreach \inode in {1,...,\nhidden}
				\foreach \hnode in {1,...,\nhidden}
					\path (H1-\inode) edge (H2-\hnode);

			\foreach \inode in {1,...,\nhidden}
				\foreach \hnode in {1,...,\nhidden}
					\path (H2-\inode) edge [-] (Hdots-\hnode);

			\foreach \inode in {1,...,\nhidden}
				\foreach \hnode in {1,...,\nhidden}
					\path (Hdots-\inode) edge (H3-\hnode);

			\foreach \inode in {1,...,\nhidden}
				\foreach \hnode in {1,...,\noutput}
					\path (H3-\inode) edge (O-\hnode);

			\node[annot] (input) at (0,.5*\nmax*\neuronsep+.5*\neuronsep) {Input layer\\(\first layer)}; 
			\node[annot] (hidden1) at (\layersep,.5*\nmax*\neuronsep+.5*\neuronsep) {\first hidden layer\\(\second layer)}; 
			\node[annot] (hidden2) at (2*\layersep,.5*\nmax*\neuronsep+.5*\neuronsep) {\second hidden layer\\(\third layer)}; 
			\node[annot] (dots) at (3*\layersep,.5*\nmax*\neuronsep+.5*\neuronsep) {$\cdots$}; 
			\node[annot] (hidden3) at (4*\layersep,.5*\nmax*\neuronsep+.5*\neuronsep) {\nth{$(L-1)$} hidden layer\\(\nth{$L$} layer)}; 
			\node[annot] (output) at (5*\layersep,.5*\nmax*\neuronsep+.5*\neuronsep) {Output layer\\(\nth{$(L+1)$} layer)}; 
	\end{tikzpicture}
	\addfig{fig:deepANN}
\caption{\label{fig:deepANN}Graphical illustration of a fully-connected feedforward \ann\ consisting of
$L\in\N$ affine transformations (i.e., consisting of $L+1$ layers: one input layer, $L-1$ hidden layers, and one output layer) 
with $l_0\in\N$ neurons on the input layer (i.e., with $l_0$-dimensional input layer), with
$l_1\in\N$ neurons on the \first hidden layer (i.e., with $l_1$-dimensional \first hidden layer),
with $l_2\in\N$ neurons on the \second hidden layer (i.e., with $l_2$-dimensional \second hidden layer),
$\dots$, with $l_{L-1}$ neurons on the \nth{$(L-1)$} hidden layer (i.e., with $(l_{L-1})$-dimensional \nth{$(L-1)$} hidden layer),
and with $l_L$ neurons in the output layer (i.e., with $l_L$-dimensional output layer).}
\end{figure}

\subsection{Affine functions}

\begingroup
\newcommand{\s}{s}
\begin{adef}{def:affine}[Affine functions]
  Let $\netDim,m,n \in \N$, $\s \in \N_0$, 
  $ \theta = ( \theta_1, \dots, \theta_\netDim ) \in \R^\netDim $ 
  satisfy $\netDim \geq \s + m n + m$.
  Then we denote by $\Aff_{m,n}^{\theta, \s}\colon \R^{n} \to \R^{m}$ the function which satisfies 
  for all $x = (x_1,\ldots, x_{n}) \in \R^{n}$ that
  \begin{equation}
  \label{def:affine:eq1}
		\begin{split}
     \Aff_{m,n}^{\theta,\s}( x ) 
  &= 
    \begin{pmatrix}
        \theta_{ \s + 1 }
      &
        \theta_{ \s + 2 }
      &
        \cdots
      &
        \theta_{ \s + n }
      \\
        \theta_{ \s + n + 1 }
      &
        \theta_{ \s + n + 2 }
      &
        \cdots
      &
        \theta_{ \s + 2 n }
      \\
        \theta_{ \s + 2 n + 1 }
      &
        \theta_{ \s + 2 n + 2 }
      &
        \cdots
      &
        \theta_{ \s + 3 n }
      \\
        \vdots
      &
        \vdots
      &
        \ddots
      &
        \vdots
      \\
        \theta_{ \s + ( m - 1 ) n + 1 }
      &
        \theta_{ \s + ( m - 1 ) n + 2 }
      &
        \cdots
      &
        \theta_{ \s + m n }
      \end{pmatrix}
      \begin{pmatrix}
        x_1
      \\
        x_2
      \\
        x_3
      \\
        \vdots 
      \\
        x_{n}
      \end{pmatrix}
    +
      \begin{pmatrix}
        \theta_{ \s + m n + 1 }
      \\
        \theta_{ \s + m n + 2 }
      \\
        \theta_{ \s + m n + 3 }
      \\
        \vdots 
      \\
        \theta_{ \s + m n + m }
      \end{pmatrix}
    \\
  &=\begin{multlined}[t]\mathtoolsset{firstline-afterskip=1cm}
    \pr[\Big]{ \textstyle
      \br[\big]{ \sum_{k = 1}^n x_k \theta_{\s+k} } + \theta_{\s + m n + 1} ,
      \br[\big]{ \sum_{k = 1}^n  x_k\theta_{\s+n+k} } + \theta_{\s + m n + 2} , 
      \ldots,\\\textstyle
      \br[\big]{ \sum_{k = 1}^n  x_k \theta_{\s+(m-1)n+k} } + \theta_{\s + m n + m}
	}
	\end{multlined}
	\end{split}
  \end{equation}
and we call $\Aff_{m,n}^{\theta,\s}$ the affine function from $\R^{n}$ to $\R^{m}$ associated to $(\theta,\s)$.
\end{adef}
\endgroup

\cfclear
\begin{athm}{example}{affine_example}[Example for \cref{def:affine}]
Let
	$\theta = (0, 1, 2, 0, 3, 3, 0, 1, 7) \in \R^9$.
Then
\begin{equation}
\begin{split} 
	\Aff_{2, 2}^{\theta,1}( (1, 2) )
=
	(8, 6)	 
\end{split}
\end{equation}
\cfout.
\end{athm}

\begin{aproof}
\Nobs that 
\enum{
	\eqref{def:affine:eq1}
}[ensure]
that
\begin{equation}
\begin{split} 
	\Aff_{2, 2}^{\theta,1}( (1, 2) )
=
	\begin{pmatrix}
		1 & 2 \\
		0 & 3
	\end{pmatrix}
	\begin{pmatrix}
		1\\2
	\end{pmatrix}
	+
	\begin{pmatrix}
		3\\0
	\end{pmatrix}
=
	\begin{pmatrix}
		1 + 4 \\
		0 + 6
	\end{pmatrix}
	+
	\begin{pmatrix}
		3\\0
	\end{pmatrix}
=
	\begin{pmatrix}
		8\\6
	\end{pmatrix}.
\end{split}
\end{equation}
\end{aproof}

\cfclear
\begin{exercise}{quest:affine}
Let $\theta = ( 3,1, -2, 1, -3, 0, 5, 4, -1, -1, 0)  \in \R^{11}$.
Specify
$
	\Aff_{2, 3}^{\theta,2}( (-1, 1, -1) )
$
explicitly and prove that your result is correct \cfload!
\end{exercise}

\subsection{Vectorized description of fully-connected feedforward ANNs}

\begingroup
\newcommand{\s}{s}
\begin{adef}{def:FFNN}[Vectorized description of fully-connected feedforward \anns]
  Let 
		$\netDim,L \in \N$, 
		$l_0,l_1,\ldots, l_L \in \N$, 
		$\theta \in \R^\netDim$ 
	satisfy
  \begin{equation}
    \netDim \geq \sum_{k=1}^{L} l_k(l_{k-1} + 1)
  \end{equation}
  and 
  for every 
  	$k \in \{1,2,\ldots,L\}$
  let 
		$\Psi_k \colon \R^{l_k} \to \R^{l_k}$
	be a function.
  Then we denote by 
    $\RealV{\theta}{\s}{l_0}{\Psi_1,\Psi_2,\ldots,\Psi_L} \colon\R^{l_0} \to \R^{l_L}$ 
  the function given by 
  \begin{multline}
  \label{eq:FFNN}
		\RealV{\theta}{\s}{l_0}{ \Psi_1, \Psi_2, \ldots, \Psi_L } 
  =  
      \Psi_L
      \circ \Aff_{l_L,l_{L-1}}^{\theta,\sum_{k = 1}^{L-1} l_k (l_{k-1} + 1)}
      \circ \Psi_{L-1} 
      \circ \Aff_{l_{L-1},l_{L-2}}^{\theta,\sum_{k = 1}^{L-2} l_k (l_{k-1} + 1)}
      \circ 
      \ldots  \\
      \ldots
      \circ \Psi_{2}  
      \circ  \Aff_{l_2,l_1}^{\theta,l_1(l_0 + 1)}
      \circ \Psi_{1}  
      \circ \Aff_{l_1,l_0}^{\theta,0}
  \end{multline}
  and we call 
  $
    \RealV{\theta}{\s}{l_0}{ \Psi_1, \Psi_2, \ldots, \Psi_L } 
  $ 
  the realization function of 
  the fully-connected feedforward \ann\ 
  associated to $ \theta $
  with $ L + 1 $ layers with dimensions $ ( l_0, l_1, \ldots, l_L ) $ 
  and
  activation functions $ ( \Psi_1, \Psi_2, \ldots, \Psi_L ) $
  	(we call 
  	  $
  	    \RealV{\theta}{\s}{l_0}{ \Psi_1, \Psi_2, \ldots, \Psi_L } 
  	  $ 
  	  the realization of 
  	  the fully-connected feedforward \ann\ 
  	  associated to $ \theta $
  	  with $ L + 1 $ layers with dimensions $ ( l_0, l_1, \ldots, l_L ) $ 
  	  and
  	  activations $ ( \Psi_1, \Psi_2, \ldots, \Psi_L ) $)
  (cf.\ \cref{def:affine}).
\end{adef}
\endgroup

\cfclear
\begin{athm}{example}{ANNs_example}[Example for \cref{def:FFNN}]
Let $\theta = ( 1, -1, 2,\allowbreak -2,\allowbreak 3, \allowbreak -3,\allowbreak 0,\allowbreak 0,\allowbreak 1)  \in \R^9$ 
and let 
$\Psi \colon \R^2 \to \R^2$ satisfy for all
	$x = (x_1, x_2) \in \R^2$
that
\begin{equation}
\label{ANNs_example:ass1}
\begin{split} 
	\Psi(x)
=
	(\max\{x_1, 0\}, \max\{x_2, 0 \}).
\end{split}
\end{equation}
Then
\begin{equation}
	\bigl( 
		\RealV{\theta}{\s}{1}{\Psi, \id_\R}
	\bigr)(2)
=
	12
\end{equation}
\cfout.
\end{athm}

\begin{aproof}
\Nobs that
\enum{
	\eqref{def:affine:eq1};
	\eqref{eq:FFNN};
	\eqref{ANNs_example:ass1}
}[assure]
that
\begin{equation}
\begin{split} 
	\big( 
		\RealV{\theta}{\s}{1}{\Psi, \id_\R}
	\big)(2)
&=
	\bpr{
		\id_\R
		\circ  \Aff_{1,2}^{\theta,4}
		\circ \Psi 
		\circ \Aff_{2,1}^{\theta,0}
	}(2)
=
	\bpr{
		\Aff_{1,2}^{\theta,4}
		\circ \Psi 
	}
	\pr*{
		\begin{pmatrix}
			1\\-1
		\end{pmatrix}
		\begin{pmatrix}
			2
		\end{pmatrix}
		+
		\begin{pmatrix}
			2\\-2
		\end{pmatrix}
	}\\
&=
	\bpr{
		\Aff_{1,2}^{\theta,4}
		\circ \Psi 
	}
	\pr*{
		\begin{pmatrix}
			4\\-4
		\end{pmatrix}
	}
=
	\Aff_{1,2}^{\theta,4}
	\pr*{
		\begin{pmatrix}
			4\\0
		\end{pmatrix}
	}
=
	\begin{pmatrix}
		3 & -3
	\end{pmatrix}
	\begin{pmatrix}
		4\\0
	\end{pmatrix}
	+
	\begin{pmatrix}
		0
	\end{pmatrix}
=
	12
\end{split}
\end{equation}
\cfload.
\end{aproof}

\cfclear
\begin{exercise}{quest:max}
Let $\theta = ( 1, -1, 0, 0, 1, -1, 0)  \in \R^7$ 
and let 
$\Psi \colon \R^2 \to \R^2$ satisfy for all
	$x = (x_1, x_2) \in \R^2$
that
\begin{equation}
\label{quest:max:eq1}
\begin{split} 
	\Psi(x)
=
	(\max\{x_1, 0\}, \min\{x_2, 0 \}).
\end{split}
\end{equation}
Prove or disprove the following statement:
It holds that
\begin{equation}
	\big( 
		\RealV{\theta}{\s}{1}{\Psi, \id_\R}
	\big)(-1)
=
	-1
\end{equation}
\cfload.
\end{exercise}

\cfclear
\begin{exercise}{neural_network_1d}
Let \(\theta=(\theta_1,\ldots,\theta_{10})\in\R^{10}\) satisfy
\[
	\theta = (\theta_1,\ldots,\theta_{10}) =
	(
	1,
	0,
	2,
	-1,
	2,
	0,
	-1,
	1,
	2,
	1
	)
\]
and let
$ m \colon \R \to \R $
and $ q \colon \R \to \R $
satisfy
for all $ x \in \R $ that
\begin{equation}
  m( x ) = \max\{ -x, 0 \}
	\qquad\text{and}\qquad
	q( x ) = x^2.
\end{equation}
Specify
\( \bigl( \RealV{\theta}{\s}{1}{q,m,q} \bigr) ( 0 )\),
\( \bigl( \RealV{\theta}{\s}{1}{q,m,q} \bigr) ( 1 )\),
and
\( \bigl( \RealV{\theta}{\s}{1}{q,m,q} \bigr) ( \nicefrac{1}{2} )\)
explicitly and prove that your results are correct \cfload!
\end{exercise}

\cfclear
\begin{exercise}{quest:evaluate_NN2}
Let $\theta = (\theta_1, \ldots, \theta_{15}) \in \R^{15}$ satisfy
\begin{equation}
  (\theta_1, \ldots, \theta_{15})
=
  (1, -2, 0, 3, 2, -1, 0, 3, 1, -1, 1, -1, 2, 0, -1)
\end{equation} 
and let $\Phi \colon \R^2 \to \R^2$ and $\Psi \colon \R^2 \to \R^2$ satisfy 
for all $x , y \in \R$ that
$
  \Phi(x, y) = (y, x)
$
and
$
  \Psi(x, y) 
=
  (x y, x y)
$.
\begin{enumerate}[label = \textbf{\alph*)}]
\item  \label{quest:evaluate_NN2:item1}   %
Prove or disprove the following statement:
It holds that
$ 
	\big(
		\RealV{\theta}{\s}{2}{\Phi, \Psi}
	\big)(1,-1) 
=  
	(4, 4)
$ \cfout.

\item  \label{quest:evaluate_NN2:item2}  %
Prove or disprove the following statement:
It holds that
$ 
	\big(
		\RealV{\theta}{\s}{2}{\Phi, \Psi}
	\big)(-1,1) 
= 
	(-4, -4)
$ \cfout.
\end{enumerate}
\end{exercise}

\subsection{Weight and bias parameters of fully-connected feedforward ANNs}

\begin{remark}[Weights and biases for fully-connected feedforward \anns]
\label{weight_and_bias}
Let $L \in \{2,3,\allowbreak 4,\ldots \} $, 
$ v_0, v_1, \ldots, v_{L-1} \in \N_0 $,
$ l_0, l_1, \ldots, l_L $, $ \netDim \in \N $, 
$ \theta = ( \theta_1, \dots, \theta_\netDim ) \in \R^\netDim $ 
satisfy for all $k \in \{0,1,\ldots,L-1\}$ that
\begin{equation}
	\netDim \geq \sum_{i=1}^{L} l_i(l_{i-1} + 1) 
\qandq 
	v_k = \sum_{i=1}^k l_i(l_{i-1} + 1),
\end{equation}
let 
$ W_k \in \R^{ l_k \times l_{ k - 1 } } $,
$ k \in \{ 1, 2, \dots, L \} $,
and 
$ b_k \in \R^{ l_k } $,
$ k \in \{ 1, 2, \dots, L \} $, 
satisfy for all $ k \in \{ 1, 2, \dots, L \} $ that
\begin{align}
  W_k 
&=
\underbrace{
\begin{pmatrix}
      \theta_{ v_{ k - 1 } + 1 }
    &
      \theta_{ v_{ k - 1 } + 2 }
    &
      \dots
    &
      \theta_{ v_{ k - 1 } + l_{k-1} }
    \\
      \theta_{ v_{ k - 1 } + l_{k-1} + 1 }
    &
      \theta_{ v_{ k - 1 } + l_{k-1} + 2 }
    &
      \dots
    &
      \theta_{ v_{ k - 1 } + 2 l_{k-1} }
    \\
      \theta_{ v_{ k - 1 } + 2 l_{k-1} + 1 }
    &
      \theta_{ v_{ k - 1 } + 2 l_{k-1} + 2 }
    &
      \dots
    &
      \theta_{ v_{ k - 1 } + 3 l_{k-1} }
    \\
      \vdots
    &
      \vdots
    &
      \vdots
    &
      \vdots
    \\
      \theta_{ v_{ k - 1 } + ( l_k - 1 ) l_{k-1} + 1 }
    &
      \theta_{ v_{ k - 1 } + ( l_k - 1 ) l_{k-1} + 2 }
    &
      \dots
    &
      \theta_{ v_{ k - 1 } + l_k l_{k-1} }
\end{pmatrix}
}_{
  \text{weight parameters}
}\\
\andq
	b_k
	&=
\underbrace{
	\pr*{
      \theta_{ v_{ k - 1 } + l_k l_{k-1} + 1 }
    ,
      \theta_{ v_{ k - 1 } + l_k l_{k-1} + 2 }
    ,
      \ldots 
    ,
      \theta_{ v_{ k - 1 } + l_k l_{k-1} + l_k }
  }
}_{
  \text{bias parameters}
},
\end{align}
and let $\Psi_k \colon \R^{l_k} \to \R^{l_k}$, $k \in \{1,2,\ldots,L\}$, be functions.
Then
\begin{enumerate}[label=(\roman{*})]
\item \label{weight_and_bias:item1}
it holds that 
\begin{equation}
	\RealV{\theta}{v_0}{l_0}{\Psi_1,\Psi_2,\ldots,\Psi_L} 
= 
	\Psi_L
	\circ \Aff^{\theta,v_{L-1}}_{l_L,l_{L-1}} 
	\circ \Psi_{L-1}
	\circ \Aff^{\theta,v_{L-2}}_{l_{L-1},l_{L-2}} 
	\circ \Psi_{L-2}
	\circ 
	\ldots 
	\circ  \Aff^{\theta,v_1}_{l_2,l_1} 
	\circ \Psi_1
	\circ \Aff^{\theta,v_0}_{l_1,l_0}
\end{equation}
and
\item \label{weight_and_bias:item2}
it holds for all $k \in \{1,2,\ldots, L \}$, $x \in \R^{l_{k-1}}$ that
$
  \Aff^{\theta,v_{k-1}}_{l_k,l_{k-1}}(x) = W_k x + b_{k}
$
\end{enumerate}
(cf.\ \cref{def:affine,def:FFNN}).
\end{remark}

\def\layersep{4cm}
\begin{figure}
    \centering
    \begin{adjustbox}{width=\textwidth}
    \begin{tikzpicture}[shorten >=1pt,-latex,draw=black!100, node distance=\layersep,auto]

        \foreach \name / \y in {1,...,3}
            \node[input neuron] (I-\name) at (0,-3*\y+2.75) {};
            \path[yshift=1.5cm]
								node[hidden neuron] (H-1) at (\layersep,-1 cm) {}
								node[hidden neuron] (Q-1) at (\layersep,-2.5 cm) {}
								node[hidden neuron] (H-2) at (\layersep,-4 cm) {}
								node[hidden neuron] (Q-2) at (\layersep,-5.5 cm) {}
								node[hidden neuron] (H-3) at (\layersep,-7 cm) {}
								node[hidden neuron] (Q-3) at (\layersep,-8.5 cm) {};
                 \node[hidden neuron] (H2-1) at (2*\layersep,-3*1+2.75) {};
								 \node[hidden neuron] (H2-2) at (2*\layersep,-3*2+2.75) {};
				         \node[hidden neuron] (H2-3) at (2*\layersep,-3*3+2.75) {};
           \node[output neuron] (H3-1) at (3*\layersep,-3.25 cm) {};
					 \node[annot,right of=H3-1, node distance=1.7cm, align=center] () {};
                
				\foreach \y in {1,...,3}
							\foreach \target in {H,Q}
									\foreach \source in {1,...,3}
                \path (I-\source) edge (\target-\y);
				\foreach \y in {1,...,3}
							\foreach \source in {H,Q}
									\foreach \target in {1,...,3}
									\path (\source-\y) edge (H2-\target);
				\foreach \y in {1,...,3}
					\path (H2-\y) edge (H3-1);

        \node[annot,above of=H-1, node distance=1.5cm, align=center] (hl) {\first hidden layer\\(\second layer)};
        \node[annot,above of=H2-1, node distance=2.25cm, align=center] (hl2) {\second hidden layer\\(\third layer)};
        \node[annot,above of=H3-1, node distance=5.25cm, align=center] (hl3) {Output layer\\(\fourth layer)};
        \node[annot,left of=hl, align=center] {Input layer\\ (\first layer)};
				
    \end{tikzpicture}
    \end{adjustbox}
		\addfig{figure_1}
	\caption{\label{figure_1}Graphical illustration of an \ann.
    The \ann\
    has $2$ hidden layers and length $L=3$
    with $3$ neurons in the input layer (corresponding to ${l}_0 = 3$), 
    $6$ neurons in the first hidden layer (corresponding to ${l}_1 = 6$), 
    $3$ neurons in the second hidden layer (corresponding to ${l}_2 = 3$), 
    and one neuron in the output layer (corresponding to ${l}_3 = 1$). 
    In this situation we have an \ann\ with $39$ weight parameters and $10$ bias parameters 
    adding up to $49$ parameters overall.
    The realization of this \ann\ is a
	function from $\R^3$ to $\R$.}
\end{figure}

\section{Activation functions}
\label{sec:activation}

In this section we review a few popular activation functions from the literature
(cf.\ \cref{def:FFNN} above and \cref{def:ANNrealization} below for the use of activation functions in the context of fully-connected feedforward \anns, 
cf.\ \cref{def:CNNrealisation} below for the use of activation functions in the context of \cnns,
cf.\ \cref{def:ResNetrealization} below for the use of activation functions in the context of \resnets, 
and
cf.\ \cref{def:RNNNode,def:VanillaRNN} below for the use of activation functions in the context of \RNNs).

\subsection{Multi-dimensional versions}

To describe multi-dimensional activation functions, 
we frequently employ the concept of the multi-dimensional 
version of a function. This concept is the subject 
of the next notion.

\begin{adef}{def:multidim_version}[Multi-dimensional versions of one-dimensional functions]
Let $T \in \N$, $d_1, d_2, \allowbreak \ldots, d_T \in \N$ and let $\psi \colon \R \to \R$ be a function.
Then we denote by 
\begin{equation}
	\multdim_{\psi, d_1, d_2, \ldots, d_T} \colon \R^{d_1 \times d_2 \times \ldots \times d_T} \to \R^{d_1 \times d_2 \times \ldots \times d_T}
\end{equation}
the function which satisfies for all 
$ 
	x 
= 
	( x_{k_1, k_2, \ldots, k_T} )_{
		(k_1, k_2, \ldots, k_T) \in (\bigtimes_{t = 1}^T  \{1, 2, \ldots, d_t\})
	} \in \R^{d_1 \times d_2 \times \ldots \times d_T}
$,
$ 
	y
= 
	( y_{k_1, k_2, \ldots, k_T} )_{
		(k_1, k_2, \ldots, k_T) \in (\bigtimes_{t = 1}^T  \{1, 2, \ldots, d_t\})
	} \in \R^{d_1 \times d_2 \times \ldots \times d_T}
$
with 
$\forall \, k_1 \in \{1, 2, \ldots, d_1\}$,  $k_2 \in \{1, 2, \ldots, d_2\}$, $\dots$, $k_T \in \{1, 2, \ldots, d_T\} \colon y_{k_1, k_2, \ldots, k_T} = \psi(x_{k_1, k_2, \ldots, k_T})$
that
\begin{equation}
\label{multidim_version:Equation}
\multdim_{\psi, d_1, d_2, \ldots, d_T}( x ) 
=
y
\end{equation}
and we call $\multdim_{\psi, d_1, d_2, \ldots, d_T} $ the $d_1 \times d_2 \times \ldots \times d_T$-dimensional version of $\psi$.
\end{adef}

\cfclear
\begin{athm}{example}{mulidim_version_example}[Example for \cref{def:multidim_version}]
Let $A \in \R^{3 \times 1 \times 2}$ satisfy
\begin{equation}
\begin{split} 
	A 
=
	\pr*{
		\begin{pmatrix}
			1 & -1 
		\end{pmatrix},
		\begin{pmatrix}
			-2 & 2 
		\end{pmatrix},
		\begin{pmatrix}
			3& -3 
		\end{pmatrix}
	}
\end{split}
\end{equation}
and let $\psi \colon \R \to \R$ satisfy for all 
	$x \in \R$
that
$
	\psi(x) = x^2
$.
Then
\begin{equation}
\label{mulidim_version_example:concl1}
\begin{split} 
	\multdim_{\psi, 3, 1, 2}(A)
=
	\pr*{
		\begin{pmatrix}
			1 & 1 
		\end{pmatrix},
		\begin{pmatrix}
			4 & 4 
		\end{pmatrix},
		\begin{pmatrix}
			9& 9 
		\end{pmatrix}
	}
\end{split}
\end{equation}
\end{athm}

\begin{aproof}
Note that \eqref{multidim_version:Equation} establishes \eqref{mulidim_version_example:concl1}.
\end{aproof}

\cfclear
\begin{exercise}{ex:mulidim_version_example}
Let $A \in \R^{2 \times 3}$, $B \in \R^{2 \times 2 \times 2}$ satisfy
\begin{equation}
\begin{split} 
	A 
=
	\begin{pmatrix}
		3 & -2 & 5\\
		1 & 0 & -2
	\end{pmatrix} 
\qandq
	B
=
	\pr*{
		\begin{pmatrix}
			0 & 1 \\
			-1 & 0 
		\end{pmatrix},
		\begin{pmatrix}
			-3 & -4 \\
			5 & 2 
		\end{pmatrix}	
	}
\end{split}
\end{equation}
and let $\psi \colon \R \to \R$ satisfy for all 
	$x \in \R$
that
$
	\psi(x) = \abs{x}
$.
Specify
$
	\multdim_{\psi, 2, 3}(A)
$
and
$
	\multdim_{\psi, 2, 2, 2}(B)
$
explicitly and prove that your results are correct \cfload!
\end{exercise}

\cfclear
\begin{exercise}{quest:evaluate_NN1}
Let $\theta = (\theta_1, \theta_2, \ldots, \theta_{14}) \in \R^{14}$ satisfy
\begin{equation}
  (\theta_1, \theta_2, \ldots, \theta_{14})
=
  (0, 1, 2, 2, 1, 0, 1, 1, 1, -3, -1, 4, 0, 1)
\end{equation} 
and let $f \colon \R \to \R$ and $g \colon \R \to \R$ satisfy 
for all $x \in \R$ that
\begin{equation}
\begin{split} 
  f(x) = \frac{1}{1 + |x|}
\qandq
  g(x) 
=
  x^2.
\end{split}
\end{equation}
Specify
$
  \big(\RealV{\theta}{\s}{1}{\multdim_{f,3}, \multdim_{g,2}} \big)(1)
$
and
$
  \big(\RealV{\theta}{\s}{1}{\multdim_{g,2}, \multdim_{f,3}} \big)(1)
$
explicitly and prove that your results are correct \cfload!
\end{exercise}

\subsection{Single hidden layer fully-connected feedforward ANNs}

\begin{figure}[!ht]
    \centering
    \begin{tikzpicture}[shorten >=1pt,-latex,draw=black!100, node distance=\layersep,auto]
				\def\layersep{4cm}
				\def\neuronsep{2cm}
				\def\ninput{3}
				\def\nhidden{4}

				\tikzmath{
					\ninputl=\ninput-1;
					\nhiddenl=\nhidden-1;
					\nmax=max(\ninput,\nhidden);
				}

				\foreach \inode in {1,...,\ninputl}
					\node[input neuron] (I-\inode) at (0,.5*\ninput*\neuronsep-\inode*\neuronsep+\neuronsep) {\inode};
				\node (I-dots) at (0,.5*\ninput*\neuronsep-\ninputl*\neuronsep) {$\vdots$};
				\node[input neuron] (I-\ninput) at (0,.5*\ninput*\neuronsep-\ninput*\neuronsep) {$\mathcal I$};

				\foreach \inode in {1,...,\nhiddenl}
					\node[hidden neuron] (H-\inode) at (\layersep,.5*\nhidden*\neuronsep-\inode*\neuronsep+\neuronsep) {\inode};
				\node (H-dots) at (\layersep,.5*\nhidden*\neuronsep-\nhiddenl*\neuronsep) {$\vdots$};
				\node[hidden neuron] (H-\nhidden) at (\layersep,.5*\nhidden*\neuronsep-\nhidden*\neuronsep) {$\mathcal H$};

				\node[output neuron] (O) at (2*\layersep,0) {};

				\foreach \inode in {1,...,\ninput}
				  \foreach \hnode in {1,...,\nhidden}
						\path (I-\inode) edge (H-\hnode);

				\foreach \hnode in {1,...,\nhidden}
  				\path (H-\hnode) edge (O);

				\node[annot] (input) at (0,.5*\nmax*\neuronsep+.5*\neuronsep) {Input layer}; 
				\node[annot] (hidden) at (\layersep,.5*\nmax*\neuronsep+.5*\neuronsep) {Hidden layer}; 
				\node[annot] (output) at (2*\layersep,.5*\nmax*\neuronsep+.5*\neuronsep) {Output layer}; 
    \end{tikzpicture}
		\addfig{fig:single_hidden_layer_ann}
		\caption{\label{fig:single_hidden_layer_ann}Graphical illustration of a fully-connected feedforward \ann\ consisting of
		two affine transformations (i.e., consisting of $3$ layers: one input layer, one hidden layer, and one output layer) 
		with $\mathcal I\in\N$ neurons on the input layer (i.e., with $\mathcal I$-dimensional input layer), with
		$\mathcal H\in\N$ neurons on the hidden layer (i.e., with $\mathcal H$-dimensional hidden layer),
		and with one neuron in the output layer (i.e., with one-dimensional output layer).}
	\end{figure}

\begin{athm}{lemma}{ex:single_hidden}[Fully-connected feedforward \ann\ with one hidden layer]
Let 
$
  \mathcal{I}, \mathcal{H} \in \N
$, 
$ 
  \theta = ( \theta_1, \dots, \theta_{ \mathcal{H} \mathcal{I} + 2 \mathcal{H} + 1 } ) 
  \in \R^{ \mathcal{H}\mathcal{I} + 2 \mathcal{H} + 1 }
$,
$x = (x_1, \dots, x_\mathcal{I}) \in \R^{\mathcal{I}}$
and
let $\psi \colon \R \to \R$ be a function.
Then 
\begin{equation}
	\RealV{ \theta}{ 0}{ \mathcal{I} }{ \multdim_{\psi,\mathcal{H}}, \id_{\R}} (x)\\
=
	\br*{
		\sum_{k = 1}^\mathcal{H}
		\theta_{ \mathcal{H}\mathcal{I}  + \mathcal{H} + k} 
		\,
		\psi\pr*{
			\br*{
				{\textstyle\sum\limits_{i = 1}^\mathcal{I}}
				x_i\theta_{(k-1)\mathcal{I} + i} 
			}
			+
			\theta_{ \mathcal{H} \mathcal{I} + k} 
		}
	}
	+
	\theta_{\mathcal{H}\mathcal{I} + 2 \mathcal{H} + 1}
	.
\end{equation}
(cf.\ \cref{def:multidim_version,def:affine,def:FFNN}).
\end{athm}
\begin{aproof}
	\Nobs that
		\cref{eq:FFNN}
		and \cref{multidim_version:Equation}
	show that
	\begin{equation}
		\begin{split}
			&\RealV{ \theta}{ 0}{ \mathcal{I} }{ \multdim_{\psi,\mathcal{H}}, \id_{\R}} (x)\\
		&=
			\bbpr{
				\id_{\R}
				\circ \Aff^{\theta, \mathcal{H} \mathcal{I}+ \mathcal{H}}_{1,\mathcal{H}}
				\circ \multdim_{\psi,\mathcal{H}}
				\circ \Aff^{\theta,0}_{\mathcal{H}, \mathcal{I}} 
			} (x)\\
		&=
			\Aff^{\theta, \mathcal{H}\mathcal{I} + \mathcal{H}}_{1,\mathcal{H}} \bpr{
			\multdim_{\psi,\mathcal{H}} \bpr{
			\Aff^{\theta,0}_{\mathcal{H}, \mathcal{I}} (x)
			}}\\
		&=
			\br*{
				\sum_{k = 1}^\mathcal{H}
				\theta_{ \mathcal{H}\mathcal{I}  + \mathcal{H} + k} 
				\,
				\psi\pr*{
					\br*{
						{\textstyle\sum\limits_{i = 1}^\mathcal{I}}
						x_i\theta_{(k-1)\mathcal{I} + i} 
					}
					+
					\theta_{ \mathcal{H} \mathcal{I} + k} 
				}
			}
			+
			\theta_{\mathcal{H}\mathcal{I} + 2 \mathcal{H} + 1}
			.
		\end{split}
		\end{equation}
\end{aproof}

\subsection{Rectified linear unit (ReLU) activation}
\label{subsect:Relu}

In this subsection we formulate the \ReLU\ function 
which is one of the most frequently used activation functions
in deep learning applications 
(cf., for example, LeCun et al.~\cite{LeCunBengioHinton15}).

\begin{adef}{def:relu1}[\ReLU\ activation function]
  We denote by $ \rect \colon \R \to \R $ the function which satisfies 
  for all $ x \in \R $ that 
  \begin{equation}
  \label{def:relu1:eq1}
    \rect (x) = \max\{ x, 0 \}
  \end{equation}
and we call $\rect$ the \ReLU\ activation function
(we call $\rect$ the rectifier function).
\end{adef}

\begin{figure}[!ht]
	\centering
	\includegraphics[width=0.5\linewidth]{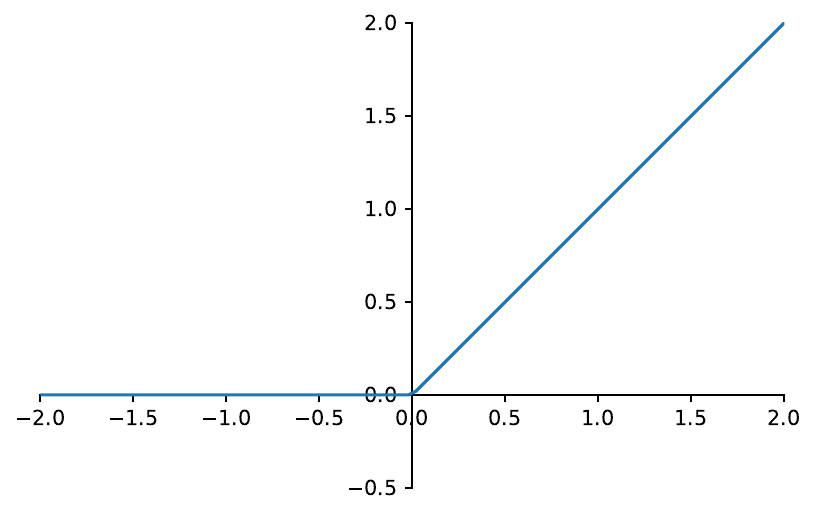}
	\caption{\label{fig:relu_plot}A plot of the \ReLU\ activation function}
\end{figure}

\filelisting{code:plot_util}{code/activation_functions/plot_util.py}{{\sc Python} code for the {\sc plot\textunderscore util} module %
used in the code listings throughout this subsection}

\filelisting{code:relu_plot}{code/activation_functions/relu_plot.py}{{\sc Python} code used to create \cref{fig:relu_plot}}

\begin{adef}{def:relu}[Multi-dimensional \ReLU\ activation functions]
  Let $d \in \N$. Then we denote by $ \Rect_{d} \colon \R^{d} \to \R^{d} $ the function given by
  \begin{equation}
  \label{eq:relu}
    \Rect_d
  =
    \multdim_{\rect, d}
  \end{equation}
  and 
  we call $\Rect_{d}$ the $d$-dimensional \ReLU\ activation function
  (we call $\Rect_{d}$ the $d$-dimensional rectifier function)
  (cf.\ \cref{def:multidim_version,def:relu1}).
\end{adef}

\begin{lemma}[An \ann\ with the \ReLU\ activation function as the activation function]
\label{prop:identity_rep}
Let $ W_1 = w_1 = 1 $, $ W_2 = w_2 = - 1 $, $ b_1 = b_2 = B = 0 $. 
Then it holds for all $ x \in \R $ that
\begin{equation}
  x 
  =
  W_1
  \max\{ w_1 x + b_1, 0 \} 
  +
  W_2
  \max\{ w_2 x + b_2, 0 \} 
  +
  B
  .
\end{equation}
\end{lemma}

\begin{proof}[Proof of \cref{prop:identity_rep}]
Observe that for all $ x \in \R $ it holds that 
\begin{equation}
\begin{split}
&
  W_1
  \max\{ w_1 x + b_1, 0 \} 
  +
  W_2
  \max\{ w_2 x + b_2, 0 \} 
  +
  B
\\ &
  =
  \max\{ w_1 x + b_1, 0 \} 
  -
  \max\{ w_2 x + b_2, 0 \} 
=
  \max\{ x , 0 \} 
  -
  \max\{ - x , 0 \} 
\\ &
=
  \max\{ x , 0 \} 
  +
  \min\{ x , 0 \} 
  =
  x
  .
\end{split}
\end{equation}
The proof of \cref{prop:identity_rep} is thus complete.
\end{proof}

\begin{exercise}[Real identity]{quest:identity1}
Prove or disprove the following statement: 
There exist 
$ \netDim,H \in \N $, $ l_1, \allowbreak l_2, \allowbreak \dots, l_H \in \N $, 
$ \theta %
\in \R^\netDim $
with $\netDim\geq 2l_1+\br[\big]{\sum_{k=2}^Hl_k(l_{k-1}+1)}+l_H+1$
such that for all $ x \in \R $ 
it holds that
\begin{equation}
  \pr[\big]{ \RealV{ \theta}{ 0}{ 1 }{ \Rect_{ l_1 }, \Rect_{ l_2 }, \dots, \Rect_{ l_H } , \id_{ \R } } }( x )
  = x 
\end{equation}
(cf.\ \cref{def:FFNN,def:relu}).
\end{exercise}

The statement of the next lemma, \cref{lem:identity1}, provides 
a partial answer to \cref{quest:identity1}. 

\cfclear
\begin{athm}{lemma}{lem:identity1}[Real identity]
Let
$ \theta = ( 1, -1, 0, 0, 1, -1, 0) \in \R^7 $. 
Then it holds for all $ x \in \R $ that
\begin{equation}
\label{eq:identity1}
  \pr[\big]{ \RealV{ \theta}{ 0}{ 1 }{ \Rect_{ 2 }, \id_{ \R } } }( x )
  = x 
\end{equation}
\cfout.
\end{athm}

\begin{aproof}
\Nobs that
\enum{
  \cref{def:affine:eq1};
  \cref{eq:FFNN};
  \cref{def:relu1:eq1};
}\prove that
for all $x \in \R$ it holds that
\begin{equation}
\begin{split}
  \pr[\big]{ \RealV{ \theta}{ 0}{ 1 }{ \Rect_{ 2 }, \id_{ \R } } }( x )
&=
  \max\{x + 0, 0\} - \max\{-x + 0, 0\} + 0\\
&=
  \max\{x, 0\} - \max\{-x, 0\}
= 
  x
\end{split}
\end{equation}
\cfload.
\end{aproof}

\begin{exercise}[Absolute value]{quest:abs}
  Prove or disprove the following statement: 
	There exist $\netDim,H\in\N$, $ l_1 , \allowbreak l_2 , \allowbreak \dots , l_H \in \N $, 
	$ \theta %
	\in \R^\netDim $
	with $\netDim\geq 2l_1+\br[\big]{\sum_{k=2}^Hl_k(l_{k-1}+1)}+l_H+1$
	such that for all $ x \in \R $ it holds that
	\begin{equation}
	\label{eq:abs}
	\pr[\big]{ \RealV{ \theta}{ 0}{ 1 }{ \Rect_{ l_1 }, \Rect_{ l_2 }, \dots, \Rect_{ l_H } , \id_{ \R } } }( x )
	= \abs{x}
	\end{equation}
	(cf.\ \cref{def:FFNN,def:relu}).
\end{exercise}

\begin{exercise}[Exponential]{quest:exp}
	Prove or disprove the following statement:
	There exist $\netDim,H\in\N$, $ l_1 , \allowbreak l_2 , \allowbreak \dots , l_H \in \N $, 
	$ \theta %
	\in \R^\netDim $
	with $\netDim\geq 2l_1+\br[\big]{\sum_{k=2}^Hl_k(l_{k-1}+1)}+l_H+1$
	such that for all $ x \in \R $ it holds that
	\begin{equation}
	\label{eq:exp}
	\pr[\big]{ \RealV{ \theta}{ 0}{ 1 }{ \Rect_{ l_1 }, \Rect_{ l_2 }, \dots, \Rect_{ l_H } , \id_{ \R } } }( x )
	= e^x 
	\end{equation}
	(cf.\ \cref{def:FFNN,def:relu}).
\end{exercise}

\begin{exercise}[Two-dimensional maximum]{quest:max2}
	Prove or disprove the following statement:
	There exist $ \netDim, H \in\N$, $l_1, \allowbreak l_2, \allowbreak \dots, l_H \in \N $, 
	$ \theta %
	\in \R^\netDim $
	with $\netDim\geq 3l_1+\br[\big]{\sum_{k=2}^Hl_k(l_{k-1}+1)}+l_H+1$
	such that for all $ x, y \in \R $ it holds that
	\begin{equation}
	\label{eq:max2}
	\pr[\big]{ \RealV{ \theta}{ 0}{ 2 }{ \Rect_{ l_1 }, \Rect_{ l_2 }, \dots, \Rect_{ l_H } , \id_{ \R } } }( x, y )
	= \max\{ x, y \} 
	\end{equation}
	(cf.\ \cref{def:FFNN,def:relu}).
\end{exercise}

\begin{exercise}[Real identity with two hidden layers]{quest:identity2}
	Prove or disprove the following statement:
	There exist $ \netDim, l_1, l_2 \in \N $, 
	$ \theta %
	\in	\R^\netDim $
	with $\netDim\geq 2l_1+l_1l_2+2l_2+1$
	such that for all $x\in\R$ it holds that
	\begin{equation}
	\label{eq:identity2}
	\pr[\big]{ \RealV{ \theta}{ 0}{ 1 }{ \Rect_{ l_1 },  \Rect_{ l_2 } , \id_{ \R } } }( x )
	= x 
	\end{equation}
	(cf.\ \cref{def:FFNN,def:relu}).
\end{exercise}

The statement of the next lemma, \cref{lem:identity2}, provides 
a partial answer to \cref{quest:identity2}.

\begin{athm}{lemma}{lem:identity2}[Real identity with two hidden layers]
Let 
$ \theta = ( 1 $, $ -1 $, $ 0 $, $ 0 $, $ 1 $, $ -1 $, $ -1 $, $ 1 $, $ 0 $, $ 0 $, $ 1 $, $ -1 $, $ 0) \in \R^{13} $. 
Then it holds for all $ x \in \R $ that
\begin{equation}
\label{eq:}
	\pr[\big]{ \RealV{ \theta}{0}{1}{ \Rect_2,\Rect_2, \id_{ \R } } }( x )
	= x 
\end{equation}
(cf.\ \cref{def:FFNN,def:relu}).
\end{athm}

\begin{aproof}
\Nobs that
\enum{
  \cref{def:affine:eq1};
  \cref{eq:FFNN};
  \cref{def:relu1:eq1};
}\prove that
for all $x \in \R$ it holds that
\begin{equation}
\begin{split}
	\pr[\big]{ \RealV{ \theta}{0}{1}{ \Rect_2,\Rect_2, \id_{ \R } } }( x )
&=
	\begin{pmatrix}
		1 & -1
	\end{pmatrix}
	\Rect_2\pr*{
		\begin{pmatrix}
			1 & -1 \\ -1 & 1
		\end{pmatrix}
		\Rect_2\pr*{
			\begin{pmatrix}
				1 \\ -1
			\end{pmatrix}
			x 
			+
			\begin{pmatrix}
				0 \\ 0
			\end{pmatrix}
		}
		+
		\begin{pmatrix}
			0 \\ 0
		\end{pmatrix}
	}
	+
	0 \\
&=
	\begin{pmatrix}
		1 & -1
	\end{pmatrix}
	\Rect_2\pr*{
		\begin{pmatrix}
			\max\{x, 0\} - \max\{-x, 0\} \\
			-\max\{x, 0\} + \max\{-x, 0\}
		\end{pmatrix}
	} \\
&=
	\begin{pmatrix}
		1 & -1
	\end{pmatrix}
	\Rect_2
		\begin{pmatrix}
			x \\
			x
		\end{pmatrix}
	\\
&=
	\max\{x, 0\} - \max\{-x, 0\}
=
	x.
\end{split}
\end{equation}
\cfload.
\end{aproof}

\cfclear
\begin{exercise}[Three-dimensional maximum]{quest:max3}
	Prove or disprove the following statement:
	There exist $ \netDim, H\in\N$, $l_1, l_2, \dots, l_H \in \N $, 
	$ \theta %
	\in \R^\netDim $
	with $\netDim\geq 4l_1+\br[\big]{\sum_{k=2}^Hl_k(l_{k-1}+1)}+l_H+1$
	such that for all $ x, y, z \in \R $ it holds that
	\begin{equation}
	\label{eq:max3}
	\pr[\big]{ \RealV{\theta}{0}{3}{ \Rect_{ l_1 }, \Rect_{ l_2 }, \dots, \Rect_{ l_H } , \id_{ \R } } }( x, y, z )
	= \max\{ x, y, z \} 
	\end{equation}
	\cfload.
\end{exercise}

\begin{exercise}[Multi-dimensional maxima]{quest:maxk}
	Prove or disprove the following statement:
	For every $ k \in \N $ there exist $ \netDim, H\in\N$, $l_1, l_2, \dots, l_H \in \N $, 
	$ \theta %
	\in \R^\netDim $
	with $\netDim\geq (k+1)l_1+\br[\big]{\sum_{k=2}^Hl_k(l_{k-1}+1)}+l_H+1$
	such that for all $ x_1, x_2, \dots, x_k \in \R $ it holds that
	\begin{equation}
	\label{eq:maxd}
	\pr[\big]{ \RealV{\theta}{0}{k}{ \Rect_{ l_1 }, \Rect_{ l_2 }, \dots, \Rect_{ l_H } , \id_{ \R } } }( x_1, x_2, \dots, x_k )
	= \max\{ x_1, x_2, \dots, x_k \} 
	\end{equation}
	(cf.\ \cref{def:FFNN,def:relu}).
\end{exercise}

\cfclear
\begin{exercise}{quest:maxx}
Prove or disprove the following statement:
There exist 
$\netDim,H \in \N$, $l_1,l_2,\ldots,\allowbreak l_H \in \N$, $\theta \in \R^\netDim$ with
$
  \netDim \geq 2 \,  l_1 + \big[\sum_{k=2}^H l_k(l_{k-1} + 1) \big] + (l_H + 1)
$
such that for all $x \in \R$ it holds that
\begin{equation}
	\big( 
		\RealV{\theta}{\s}{1}{\Rect_{l_1}, \Rect_{l_2}, \ldots, \Rect_{l_H}, \id_\R}
	\big)(x)
=
 	\max\{x, \tfrac{x}{2}\}
\end{equation}
\cfload.
\end{exercise}

\cfclear
\begin{exercise}[Hat function]{quest:hat1}
	Prove or disprove the following statement:
	There exist $ \netDim, l \in \N $, 
	$ \theta %
	\in \R^\netDim $
	with $\netDim\geq 3l+1$
	such that for all $ x \in \R $ it holds that
	\begin{equation}
	\label{eq:hat1}
	\bigl( \RealV{\theta}{0}{1}{ \Rect_l, \id_{ \R } } \bigr)( x )
	= 
	\begin{cases}
	  1 & \colon x \leq 2
	\\
	  x - 1 & \colon 2 < x \leq 3
	\\
	  5 - x & \colon 3 < x \leq 4
	\\
	  1 & \colon x > 4
	\end{cases}
	\end{equation}
	\cfload.
\end{exercise}

\cfclear
\begin{exercise}{quest:hat2}
	Prove or disprove the following statement:
	There exist $ \netDim, l \in \N $, 
	$ \theta %
	\in \R^\netDim $
	with $\netDim\geq 3l+1$
	such that for all $ x \in \R $ it holds that
	\begin{equation}
	\label{eq:hat2}
	\pr[\big]{ \RealV{\theta}{0}{1}{ \Rect_l, \id_{ \R } } }( x )
	= 
	\begin{cases}
	  - 2 & \colon x \leq 1
	\\
	  2 x - 4 & \colon 1 < x \leq 3
	\\
	  2 & \colon x > 3
	\end{cases}
	\end{equation}
	\cfload.
\end{exercise}

\cfclear
\begin{exercise}{quest:step_function}
Prove or disprove the following statement:
There exists $\netDim,H \in \N$, $l_1,l_2,\ldots, \allowbreak l_H \in \N$, $\theta \in \R^\netDim$ with
$
	\netDim \geq 2 \,  l_1 + \big[\sum_{k=2}^H l_k(l_{k-1} + 1) \big] + (l_H + 1)
$
such that for all $x \in \R$ it holds that
\begin{equation}
	\big( 
		\RealV{\theta}{\s}{1}{\Rect_{l_1}, \Rect_{l_2}, \ldots, \Rect_{l_H}, \id_\R}
	\big)(x)
=
  \begin{cases}
    0 & \colon x \leq 1   \\
    x-1 & \colon 1 \leq x  \leq 2 \\
    1 & \colon x \geq 2
  \end{cases}
\end{equation} 
\cfout.
\end{exercise}

\cfclear
\begin{exercise}{quest:square}
	Prove or disprove the following statement:
	There exist $ \netDim, l \in \N $, 
	$ \theta %
	\in \R^\netDim $
	with $\netDim\geq 3l+1$
	such that for all $ x \in [0,1] $ it holds that
	\begin{equation}
	\pr[\big]{ \RealV{\theta}{0}{1}{ \Rect_l, \id_{ \R } } }( x )
	= 
	x^2
	\end{equation}
	\cfload.
\end{exercise}

\cfclear
\begin{exercise}{quest:square_approx}
Prove or disprove the following statement:
There exists $\netDim,H \in \N$, $l_1,l_2,\ldots,\allowbreak  l_H \in \N$, $\theta \in \R^\netDim$ with
$
	\netDim \geq 2 \,  l_1 + \br[\big]{\sum_{k=2}^H l_k(l_{k-1} + 1) } + (l_H + 1)
$
such that 
\begin{equation}
	\textstyle
  \sup_{x \in \left[-3, -2 \right]} 
    \big|  \big( 
		\RealV{\theta}{\s}{1}{\Rect_{l_1}, \Rect_{l_2}, \ldots, \Rect_{l_H}, \id_\R}
    \big)(x) - (x+2)^2 
    \big|
\leq
 \frac{1}{4}
\end{equation} 
\cfout.
\end{exercise}

\subsection{Clipping activation}

\begin{adef}{def:clip1}[Clipping activation functions]
  Let $u\in[-\infty,\infty)$, $v\in(u,\infty]$.
  Then we denote by $ \clip uv \colon \R \to \R $ the function which satisfies 
  for all $ x \in \R $ that 
  \begin{equation}
  \label{def:clip1:eq1}
    \clip uv (x) = \max\{u,\min\{x,v\}\}.
  \end{equation}
  and we call $\clip uv$ the $ (u,v) $-clipping activation function.
\end{adef}

\begin{figure}[!ht]
	\centering
	\includegraphics[width=0.5\linewidth]{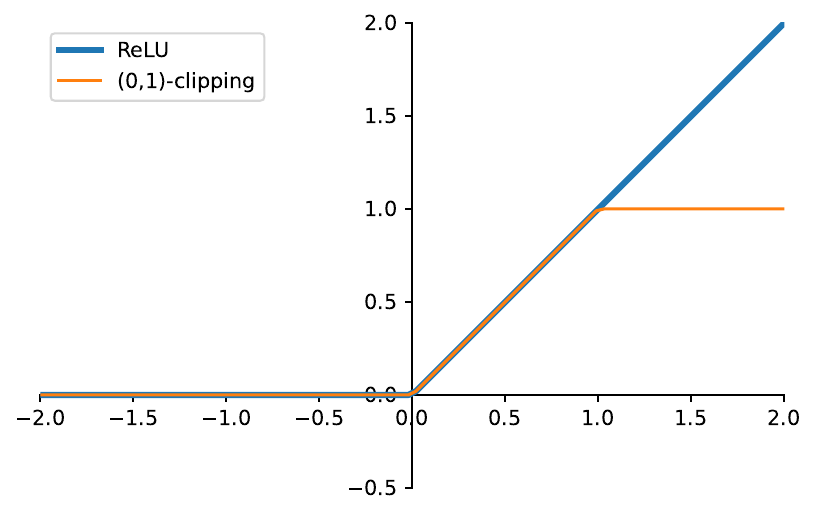}
	\caption{\label{fig:clipping_plot}A plot of the $(0,1)$-clipping activation function and the \ReLU\ activation function}
\end{figure}

\filelisting{code:clipping_plot}{code/activation_functions/clipping_plot.py}{{\sc Python} code used to create \cref{fig:clipping_plot}}

\begin{adef}{def:clip}[Multi-dimensional clipping activation functions]
  Let 
    $d \in \N$,
	$u\in [-\infty,\infty)$,
	$v\in (u,\infty]$.
  Then we denote by $ \Clip uvd \colon \R^{d} \to \R^{d} $ the function given by
  \begin{equation}
    \Clip uvd
  =
    \multdim_{\clip uv, d}
  \end{equation}
  and we call $\Clip uvd$ the $ d $-dimensional $ (u,v) $-clipping activation function
  (cf.\ \cref{def:multidim_version,def:clip1}).
\end{adef}

\subsection{Softplus activation}
\label{subsect:Softplus}

\begin{adef}{def:softplus1}[Softplus activation function]
We say that 
$\activation$
is the \softplusfunc{} if and only if
it holds that 
$
  \activation \colon \R \to \R
$
is the function from $\R$ to $\R$ which satisfies for all 
$ x \in \R $
that
\begin{equation}
  \label{eq:softplus1.def}
  \activation( x )
  =
  \log(1 + \exp(x)).
\end{equation}
\end{adef}

\begin{figure}[!ht]
	\centering
	\includegraphics[width=0.5\linewidth]{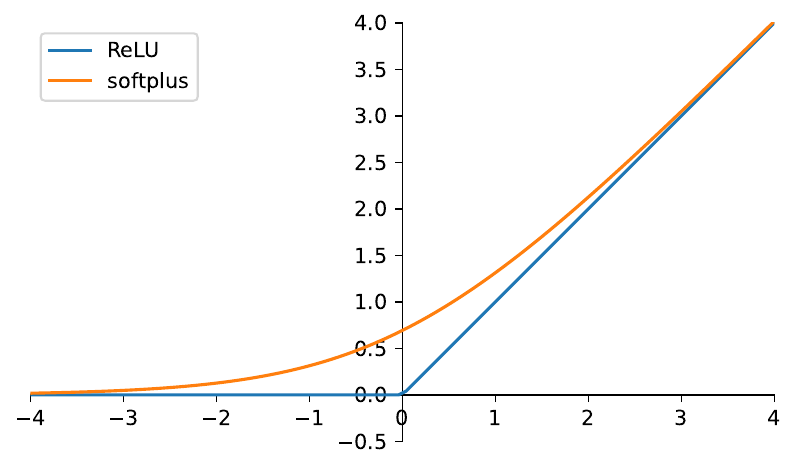}
	\caption{\label{fig:softplus_plot}A plot of the softplus activation function and the \ReLU\ activation function}
\end{figure}

\filelisting{code:softplus_plot}{code/activation_functions/softplus_plot.py}{{\sc Python} code used to create \cref{fig:softplus_plot}}

The next result, \cref{lem:softplus} below, 
presents a few elementary properties of the softplus function.

\cfclear
\begin{lemma}[Properties of the softplus function] 
\label{lem:softplus}
Let $\activation$ be the \softplusfunc{}
\cfload.
Then  
\begin{enumerate}[label=(\roman{*})]
\item 
it holds for all $ x \in [0,\infty) $ that
$
  x \leq \activation( x ) \leq x + 1 
$,
\item
it holds that
$
  \lim_{ x \to - \infty }
  \activation( x )
  = 0
$,
\item
it holds that
$
  \lim_{ x \to \infty }
  \activation( x )
  = \infty
$,
and
\item 
it holds that
$
  \activation( 0 )
  =
  \ln( 2 )
$
\end{enumerate}
(cf.\ \cref{def:softplus1}).
\end{lemma}

\begin{proof}[Proof of \cref{lem:softplus}]
Observe that the fact that $ 2\leq \exp(1)$ ensures that for all $ x \in [0,\infty) $ it holds that
\begin{equation}
\begin{split}
  x 
& =
  \ln\pr*{
    \exp( x )
  }
  \leq
  \ln\pr*{
    1 +
    \exp( x )
  }
  =
  \ln\pr*{
    \exp( 0 )
    +
    \exp( x )
  }
\\ & \leq
  \ln\pr*{
    \exp( x )
    +
    \exp( x )
  }
  =
  \ln\pr*{
    2
    \exp( x )
  }
  \leq
  \ln\pr*{
    \exp( 1 )
    \exp( x )
  }
\\ & =
  \ln\pr*{
    \exp( x + 1 )
  }
  =
  x + 1
  .
\end{split}
\end{equation}
The proof of \cref{lem:softplus} is thus complete.
\end{proof}

Note that \cref{lem:softplus} ensures that
$
  \softplus( 0 )
  =
  \ln( 2 )
  =
  0.693\dots
$
(cf.\ \cref{def:softplus1}). 
In the next step we introduce the multi-dimensional version 
of the softplus function 
(cf.\ \cref{def:multidim_version,def:softplus1} above).

\cfclear
\begin{adef}{def:softplus}[Multi-dimensional softplus activation functions]
Let $d \in \N$ and let $\activation$ be the \softplusfunc{}
\cfload.
Then we say that $\Activation$ is the $d$-dimensional softplus activation function if and only if
$
	\Activation
=
  	\multdim_{\activation, d}
$
\cfclear\cfadd{def:multidim_version}\cfout.
\end{adef}

\cfclear
\begin{lemma}
\label{softplus_explicit}
Let $d \in \N$ and let 
$\Activation \colon \R^d \to \R^d$ be a function.
Then  
$\Activation$ is the $d$-dimensional softplus activation function\cfadd{def:softplus} if and only if
it holds for all 
	$x = (x_1, \ldots, x_d) \in \R^d$
that
\begin{equation}
\label{softplus_explicit:concl1}
\begin{split} 
	A(x)
=
	(
		\log(1 + \exp(x_1)),
		\log(1 + \exp(x_2)),
		\ldots, 
		\log(1 + \exp(x_d))
	)
\end{split}
\end{equation}
\cfout.
\end{lemma}

\begin{proof}[Proof of \cref{softplus_explicit}]
Throughout this proof, let $a$ be the \softplusfunc{}
\cfload.
\Nobs that
\enum{
	\eqref{multidim_version:Equation};
	\eqref{eq:softplus1.def}
}[ensure]
that for all 
	$x = (x_1, \ldots, x_d) \in \R^d$
it holds that 
\begin{equation}
\begin{split} 
	\multdim_{\activation, d}(x)
=
	(
		\log(1 + \exp(x_1)),
		\log(1 + \exp(x_2)),
		\ldots, 
		\log(1 + \exp(x_d))
	)
\end{split}
\end{equation}
\cfload. 
The fact that 
$\Activation$ is the $d$-dimensional softplus activation function\cfadd{def:softplus} \cfload{} if and only if
$
	\Activation
=
  	\multdim_{\activation, d}
$
hence 
implies 
\eqref{softplus_explicit:concl1}.
The proof of \cref{softplus_explicit} is thus complete.
\end{proof}

\cfclear
\begin{exercise}[Real identity]{quest:identitySoftplus}
For every $d \in \N$ let $\Activation_d$ be the $d$-dimensional\cfadd{def:softplus} softplus activation function
\cfload.
Prove or disprove the following statement: 
There exist 
$ \netDim,H \in \N $, $ l_1, \allowbreak l_2, \allowbreak \dots, l_H \in \N $, 
$ \theta %
\in \R^\netDim $
with $\netDim\geq 2l_1+\br[\big]{\sum_{k=2}^Hl_k(l_{k-1}+1)}+l_H+1$
such that for all $ x \in \R $ 
it holds that
\begin{equation}
  \pr[\big]{ \RealV{ \theta}{ 0}{ 1 }{ \Activation_{ l_1 }, \Activation_{ l_2 }, \dots, \Activation_{ l_H } , \id_{ \R } } }( x )
  = x 
\end{equation}
\cfout.
\end{exercise}

The statement of the next lemma, \cref{lem:identity1_softplus}, provides 
a partial answer to \cref{quest:identitySoftplus} (cf.\ also \cref{lem:softplus:identity} below).

\cfclear
\begin{athm}{lemma}{lem:identity1_softplus}[Real identity for the softplus activation function]
Let 
$a$ be the \softplusfunc{} 
and let 
$ \theta = ( 1, -1, 0, 0, 1, -1, 0) \in \R^7 $
\cfload.
Then it holds for all $ x \in \R $ that
\begin{equation}
  \pr[\big]{ \RealV{ \theta}{ 0}{ 1 }{ \multdim_{\activation, 2}, \id_{ \R } } }( x )
  = x 
\end{equation}
\cfout.
\end{athm}

\begin{aproof}
\Nobs that
\enum{
  \cref{def:affine:eq1};
  \cref{eq:FFNN};
  \cref{eq:softplus1.def};
}\prove that heere
for all $x \in \R$ it holds that
\begin{equation}
\begin{split}
  \pr[\big]{ \RealV{ \theta}{ 0}{ 1 }{ \multdim_{\activation, 2}, \id_{ \R } } }( x )
&=
  \log(1 + \exp(x + 0)) - \log(1 + \exp(-x + 0)) + 0\\
&=
  \log(1 + \exp(x)) - \log(1 + \exp(-x))\\
&=
  \log\pr*{
    \tfrac{1 + \exp(x)}{1 + \exp(-x)}
  }\\
&=
  \log\pr*{
    \tfrac{\exp(x)(1 + \exp(-x))}{1 + \exp(-x)}
  }\\
&=
  \log\pr*{
    \exp(x)
  }
= 
  x
\end{split}
\end{equation}
\cfload.
\end{aproof}

\subsection{Gaussian error linear unit (GELU) activation}
\label{subsect:GELU}

Another popular activation function is the \GELU\ activation function first introduced in 
Hendrycks \& Gimpel~\cite{Hendrycks2016}.
This activation function is the subject of the next definition.

\newcommand{\gelu}{\GELU\ activation function\cfadd{def:gELU}}

\begin{adef}{def:gELU}[\GELU\ activation function]
We say that $a$ is the \GELU\ unit activation function (we say that $a$ is the \gelu{})
if and only if 
it holds that $a \colon \R \to \R$ is the function from $\R$ to $\R$ which satisfies for all
	$x \in \R$
that
\begin{equation}
\label{def:gELU:eq1}
\begin{split} 
	a(x)
=
	\frac{x}{\sqrt{2\pi}} 
	\br*{
		\int_{-\infty}^x 
			\exp\pr{-\tfrac{z^2}{2}}
		\,\diff z
	}.
\end{split}
\end{equation}
\end{adef}

\begin{figure}[!ht]
	\centering
	\includegraphics[width=0.5\linewidth]{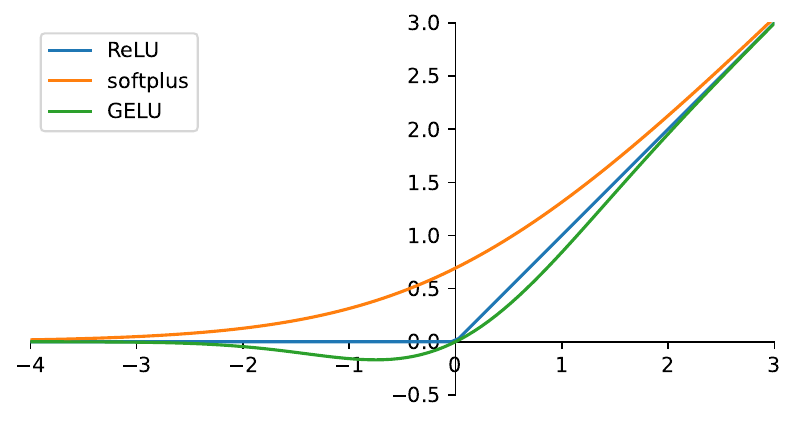}
	\caption{\label{fig:gelu_plot}A plot of the \GELU\ activation function, the \ReLU\ activation function, and the softplus activation function}
\end{figure}

\filelisting{code:gelu_plot}{code/activation_functions/gelu_plot.py}{{\sc Python} code used to create \cref{fig:gelu_plot}}

\cfclear
\begin{lemma}
\label{Gelu_relu}
Let $x \in \R$ and
let $a$ be the \gelu{} 
\cfload.
Then the following two statements are equivalent:
\begin{enumerate}[label=(\roman *)]
\item 
\label{Gelu_relu:item1}
It holds that $a(x) >0$.
\item 
\label{Gelu_relu:item2}
It holds that $\rect(x) > 0$ \cfout.
\end{enumerate}
\end{lemma}

\begin{proof}[Proof of \cref{Gelu_relu}]
Note that
\enum{
	\eqref{def:relu1:eq1};
	\eqref{def:gELU:eq1}
}[establish]
that
(\ref{Gelu_relu:item1} $\leftrightarrow$ \ref{Gelu_relu:item2}).
The proof of \cref{Gelu_relu} is thus complete.
\end{proof}

\cfclear
\begin{adef}{def:GELU}[Multi-dimensional \GELU\ activation functions]
Let $d\in \N$ and let $\activation$ be the \gelu\ \cfload.
Then we say that $\Activation$ is the $d$-dimensional \GELU\ activation function
if and only if
$
	\Activation
=
  	\multdim_{\activation, d}
$
\cfout.
\end{adef}

\subsection{Standard logistic activation}
\label{subsect:logistic}

\begin{adef}{def:logistic1}[Standard logistic activation function]
We say that 
$\activation$
is the \logfunction{} if and only if
it holds that 
$
  \activation \colon \R \to \R
$
is the function from $\R$ to $\R$ which satisfies for all 
$ x \in \R $
that
\begin{equation}
\label{def:logistic1:eq1}
  \activation( x )
  =
  \frac{1}{ 1 + \exp(-x) }
  =
  \frac{\exp(x)}{ \exp(x) + 1 }.
\end{equation}
\end{adef}

\begin{figure}[!ht]
	\centering
	\includegraphics[width=0.5\linewidth]{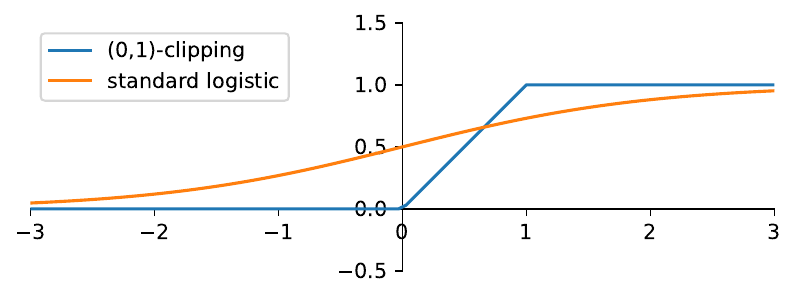}
	\caption{\label{fig:logistic_plot}A plot of the standard logistic activation function and the $(0,1)$-clipping activation function}
\end{figure}

\filelisting{code:logistic_plot}{code/activation_functions/logistic_plot.py}{{\sc Python} code used to create \cref{fig:logistic_plot}}

\cfclear
\begin{adef}{def:logistic}[Multi-dimensional standard logistic activation functions]
Let $d \in \N$
and let $\activation$ be the \logfunction{}
\cfload.
Then we say that $\Activation$ is the $d$-dimensional standard logistic activation function if and only if
$
	\Activation
=
  	\multdim_{\activation, d}
$
\cfout.
\end{adef}

\subsubsection{Derivative of the standard logistic activation function}

\cfclear
\begin{prop}[Logistic \ODE]
\label{logistic_equation}
Let $\activation$ be the \logfunction{} 
\cfload.
Then 
\begin{enumerate}[label=(\roman *)]
\item 
\label{logistic_equation:item1}
it holds that $ \activation \colon \R \to \R $ 
is infinitely often differentiable
and

\item 
\label{logistic_equation:item2}
it holds for all $ x \in \R $ that
\begin{equation}
  \activation( 0 ) = \nicefrac{ 1 }{ 2 } 
  ,
  \qquad
  \activation'(x) 
  =
  \activation( x )
  \pr*{ 1 - \activation( x ) }
  =
  \activation( x )
  -
  \br*{ 
    \activation( x )
  }^2
  ,
  \qquad 
  \text{and}
\end{equation}
\begin{equation}
  \activation''(x) 
  =
  \activation( x )
  \pr*{ 1 - \activation( x ) }
  \pr*{ 1 - 2 \, \activation( x ) }
  =
  2 
  \br*{ \activation( x ) }^3
  -
  3
  \br*{ \activation( x ) }^2
  +
  \activation( x ).
\end{equation}
\end{enumerate}
\end{prop}

\begin{proof}[Proof of 
\cref{logistic_equation}]
Note that \eqref{def:logistic1:eq1} implies \cref{logistic_equation:item1}.
\Moreover
\eqref{def:logistic1:eq1} ensures that for all $ x \in \R $ it holds that
\begin{equation}
\begin{split}
\label{eq:logistic_derivative}
  \activation'(x) 
& =
  \frac{\exp(-x)}{
  	\pr*{ 1 + \exp(-x) }^{ 2 }
  }
  =
  \activation( x )
  \pr*{ 
  	\frac{ \exp(-x) }{
  		1 + \exp(-x) 
  	}
  }
\\
&
  =
  \activation( x )
  \pr*{
  \frac{ 
    1 + \exp(-x) - 1
  }{
    1 + \exp(-x) 
  }
  }
  =
  \activation( x )
  \pr*{
    1 
    -
    \frac{ 
      1
    }{
      1 + \exp(-x) 
    }
  } \\
 & =
  \activation( x )
  \pr*{
    1 
    -
    \activation( x )
  }
  .
\end{split}
\end{equation}
Hence, we obtain that for all $ x \in \R $ it holds that
\begin{equation}
\begin{split}
\activation''(x)
& =
\bigl[
\activation( x )
\pr*{
1 
-
\activation( x )
}
\bigr]'
=
\activation'( x )
\pr*{
1 
-
\activation( x )
}
+
\activation( x )
\pr*{
1 
-
\activation( x )
}'
\\ & =
\activation'( x )
\pr*{
1 
-
\activation( x )
}
-
\activation( x )
\,
\activation'( x )
=
\activation'( x )
\pr*{
1 
-
2 \,
\activation( x )
}
\\ & =
\activation( x )
\pr*{
1 
-
\activation( x )
}
\pr*{
1 
-
2 \,
\activation( x )
}
\\ & =
\pr*{
\activation( x )
-
\br*{ \activation( x ) }^2
}
\pr*{
1 
-
2 \,
\activation( x )
}
=
\activation( x )
-
\br*{ \activation( x ) }^2
-
2
\br*{ \activation( x ) }^2
+
2
\br*{ \activation( x ) }^3
\\ & =
2
\br*{ \activation( x ) }^3
-
3
\br*{ \activation( x ) }^2
+
\activation( x )
.
\end{split}
\end{equation}
This establishes \cref{logistic_equation:item2}.
The proof of \cref{logistic_equation}
is thus complete.
\end{proof}

\subsubsection{Integral of the standard logistic activation function}

\cfclear
\begin{lemma}[Primitive of the standard logistic activation function]
\label{primitive_logistic}
Let $\softplus{}$ be the \softplusfunc{} and 
let $\logistic{}$ be the \logfunction{} 
\cfload.
Then it holds for all $ x \in \R $ that
\begin{equation}
\label{primitive_logistic:concl1}
  \int_{ - \infty }^x
  \logistic( y ) 
  \, \diff y
  =
  \int_{ - \infty }^x
  \pr*{
  \frac{
    1
  }{
    1 + e^{ - y }
  }
  } 
  \diff y
  =
  \log\pr*{
    1 + \exp(x)
  }
  =
  \softplus(x).
\end{equation}
\end{lemma}

\begin{proof}[Proof of \cref{primitive_logistic}]
Observe that \eqref{eq:softplus1.def} implies that for all $ x \in \R $  it holds
that 
\begin{equation}
  \softplus'(x) 
  =
  \br*{
  \frac{ 1 }{ 1 + \exp(x) }
  } 
  \exp(x)
  = \logistic(x) 
  .
\end{equation}
The fundamental theorem of calculus hence 
shows that for all $ w, x \in \R $ with $ w \leq x $
it holds that
\begin{equation}
  \int_w^x \underbrace{ \logistic( y ) }_{ \geq 0 } \diff y
  =
  \softplus( x ) - \softplus( w )
  .
\end{equation}
Combining this with the fact that $ \lim_{ w \to - \infty } \softplus( w ) = 0 $
establishes \cref{primitive_logistic:concl1}.
The proof of \cref{primitive_logistic}
is thus complete.
\end{proof}

\subsection{Swish and sigmoid linear unit (SiLU) activation}
\label{sec:swish}

\newcommand{\silufunc}{\SiLU\ activation function\cfadd{def:silu1}}
\newcommand{\swishfunc}{swish activation function\cfadd{def:swish1}}

In this section we introduce the \swishfunc{} and the \silufunc{} which is a special case of the \swishfunc{}.

\begin{adef}{def:swish1}[Swish activation functions]
Let $\beta \in \R$.
Then we say that $\activation$ is the \swishfunc{} with parameter $\beta$
if and only if 
it holds that $\activation \colon \R \to \R$ is the function from $\R$ to $\R$ which satisfies for all
	$x \in \R$
that
\begin{equation}
\label{def:swish1:eq1}
	\activation( x )
=
	\frac{x}{1 + \exp(-\beta x)}.
\end{equation}
\end{adef}

\begin{adef}{def:silu1}[\SiLU\ activation functions]
We say that $\activation$ is the \silufunc{}
if and only if 
it holds that $\activation \colon \R \to \R$ is the function from $\R$ to $\R$ which satisfies for all
	$x \in \R$
that
\begin{equation}
\label{def:silu1:eq1}
	\activation( x )
=
	\frac{x}{1 + \exp(- x)}.
\end{equation}
\end{adef}

\begin{athm}{lemma}{silu_swish}[Relation between the \silufunc{} and the \swishfunc{}]
\cfclear
Let $\mathfrak{S}$ be the \silufunc{} and 
let $\mathfrak{s}$ be the \swishfunc{} with parameter $1$ 
\cfload.
Then $\mathfrak{S} = \mathfrak{s}$.
\end{athm}

\begin{aproof}
\Nobs[Observe] that 
\enum{
	\eqref{def:silu1:eq1};
	\eqref{def:swish1:eq1}
}[e]
that
$\mathfrak{S} = \mathfrak{s}$.
The proof of \cref{silu_swish} is thus complete.
\end{aproof}

\begin{figure}[!ht]
	\centering
	\includegraphics[width=0.5\linewidth]{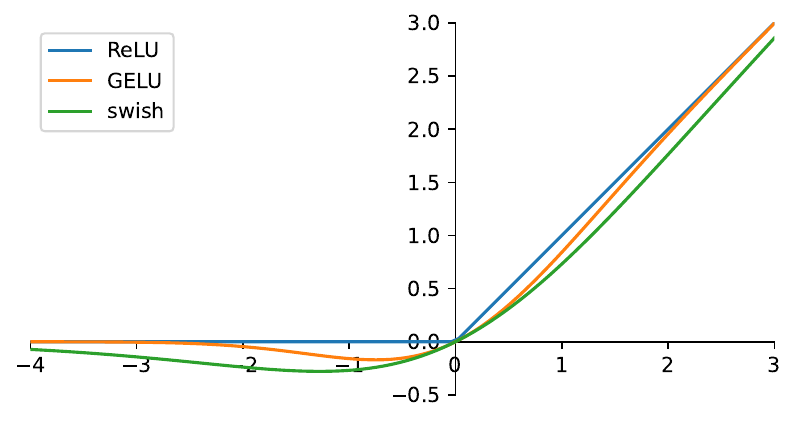}
	\caption{\label{fig:swish_plot}A plot of the swish activation function with parameter $1$, the \GELU\ activation function, and the \ReLU\ activation function}
\end{figure}

\filelisting{code:swish_plot}{code/activation_functions/swish_plot.py}{{\sc Python} code used to create \cref{fig:swish_plot}}

\cfclear
\begin{lemma}[Relation between swish activation functions  and the logistic activation function]
\label{swish_logistic}
Let $\beta \in \R$,
let $\mathfrak{s}$ be the \swishfunc{} with parameter $\beta$, and 
let $\logistic{}$ be the \logfunction{}
\cfload.
Then it holds for all 
	$x \in \R$
that
\begin{equation}
\label{swish_logistic:concl1}
\begin{split} 
	\mathfrak{s}(x)
=
	x\logistic(\beta x).
\end{split}
\end{equation}
\end{lemma}

\begin{proof}[Proof of \cref{swish_logistic}]
\Nobs[Observe] that 
\enum{
	\eqref{def:swish1:eq1};
	\eqref{def:logistic1:eq1}
}[e]
\eqref{swish_logistic:concl1}.
The proof of \cref{swish_logistic} is thus complete.
\end{proof}

\cfclear
\begin{adef}{def:swish_mult}[Multi-dimensional swish activation functions]
Let $d \in \N$, $\beta \in \R$ and
let $\activation$ be the \swishfunc{} with parameter $\beta$
\cfload.
Then we say that $\Activation$ is the 
$ d $-dimensional swish activation function with parameter $\beta$
if and only if
$
	\Activation
=
  	\multdim_{\activation, d}
$
\cfout.
\end{adef}

\cfclear
\begin{adef}{def:silu_mult}[Multi-dimensional \SiLU\ activation functions]
Let $d \in \N$ and
let $\activation$ be the \silufunc{}
\cfload.
Then we say that $\Activation$ is the 
$ d $-dimensional \silufunc{} activation function
if and only if
$
	\Activation
=
  	\multdim_{\activation, d}
$
\cfout.
\end{adef}

\subsection{Hyperbolic tangent activation}

\begin{adef}{def:hyperbolic_tangent1}[Hyperbolic tangent activation function]
We denote by
$
  \tanhh \colon \R \to \R
$
the function which satisfies for all 
$ x \in \R $
that
\begin{equation}
\label{def:hyperbolic_tangent1:eq1}
  \tanhh( x )
  =
  \frac{ \exp(x) - \exp(-x)}{\exp(x) + \exp(-x)}
\end{equation}
and we call $\tanhh$ the hyperbolic tangent activation function
(we call $ \tanh $ the hyperbolic tangent).
\end{adef}

\begin{figure}[!ht]
	\centering
	\includegraphics[width=0.5\linewidth]{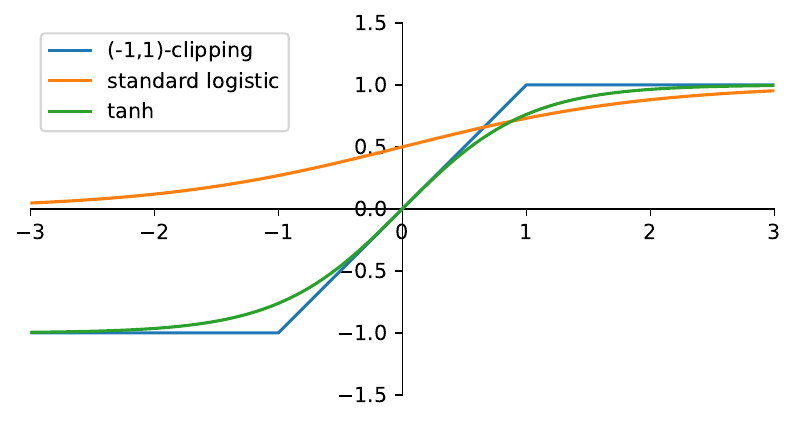}
	\caption{\label{fig:tanh_plot}A plot of the hyperbolic tangent, the $(-1,1)$-clipping activation function, and the standard logistic activation function}
\end{figure}

\filelisting{code:tanh_plot}{code/activation_functions/tanh_plot.py}{{\sc Python} code used to create \cref{fig:tanh_plot}}

\begin{adef}{def:hyperbolic_tangent_mult}[Multi-dimensional hyperbolic tangent activation functions]
Let $d \in \N$.
Then we say that $\Activation$ is the $d$-dimensional hyperbolic tangent activation function if and only if
$
	\Activation
=
  	\multdim_{\tanhh, d}
$
\cfout.
\end{adef}

\cfclear
\begin{lemma}
\label{lem:hyperbolic_tangent_logistic}
Let $\activation$ be the \logfunction{}
\cfload.
Then it holds for all $ x \in \R $ that
\begin{equation}
\tanhh( x )
=
2 \, \activation( 2 x ) - 1
\end{equation}
(cf.\ \cref{def:logistic1,def:hyperbolic_tangent1}).
\end{lemma}
\begin{proof}[Proof of 
\cref{lem:hyperbolic_tangent_logistic}]
Observe that \eqref{def:logistic1:eq1} and \eqref{def:hyperbolic_tangent1:eq1} ensure that
for all $ x \in \R $ it holds that
\begin{equation}
\begin{split}
2 \, \activation( 2 x ) - 1
& =
2 \pr*{
\frac{\exp(2x)}{ \exp(2x) + 1 }
}
-1
=
\frac{2 \exp(2x) - ( \exp(2x) + 1 )}{\exp(2x) + 1}
\\ & =
\frac{\exp(2x) - 1}{\exp(2x) + 1}
=
\frac{\exp(x)( \exp(x) - \exp(-x))}{\exp(x)( \exp(x) + \exp(-x))}
\\ & =
\frac{\exp(x) - \exp(-x)}{\exp(x) + \exp(-x)}
=
\tanhh( x )
.
\end{split}
\end{equation}
The proof of \cref{lem:hyperbolic_tangent_logistic} is thus complete.
\end{proof}

\cfclear
\begin{exercise}{quest:tanh}
Let $\activation$ be the \logfunction{} 
\cfload.
Prove or disprove the following statement:
There exists $L \in \{2,3,\ldots \}$, $\netDim, l_1, l_2, \ldots, l_{L-1} \in \N$, $\theta \in \R^\netDim$ with
$
	\netDim \geq 2 \,  l_1 + \br[\big]{\sum_{k=2}^{L-1} l_k(l_{k-1} + 1) } + (l_{L-1} + 1)
$
such that for all $x \in \R$ it holds that
\begin{equation}
	\big( 
		\RealV{\theta}{\s}{1}{\multdim_{\activation, l_1}, \multdim_{\activation, l_2}, \ldots, \multdim_{\activation, l_{L-1}}, \id_\R}
	\big)(x)
=
 	\tanhh(x)
\end{equation}
\cfout.
\end{exercise}

\subsection{Softsign activation}

\newcommand{\softsign}{softsign activation function\cfadd{def:softsign}}

\begin{adef}{def:softsign}[Softsign activation function]
We say that $\activation$ is the \softsign{}
if and only if 
it holds that $\activation \colon \R \to \R$ is the function from $\R$ to $\R$ which satisfies for all
	$x \in \R$
that
\begin{equation}
\label{def:softsign:eq1}
	\activation( x )
=
	\frac{x}{\abs{x} + 1}.
\end{equation}
\end{adef}

\begin{figure}[!ht]
	\centering
	\includegraphics[width=0.5\linewidth]{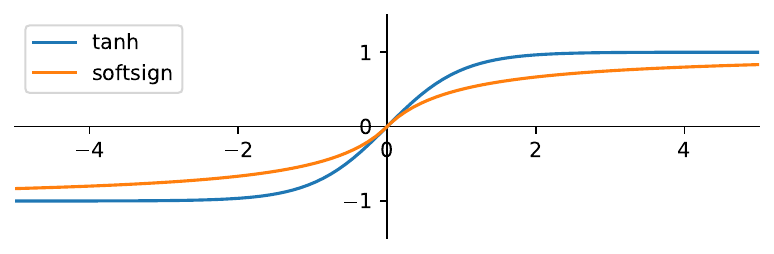}
	\caption{\label{fig:softsign_plot}A plot of the softsign activation function and the hyperbolic tangent}
\end{figure}

\filelisting{code:softsign_plot}{code/activation_functions/softsign_plot.py}{{\sc Python} code used to create \cref{fig:softsign_plot}}

\cfclear
\begin{adef}{def:softsine_mult}[Multi-dimensional softsign activation functions]
Let $d \in \N$ and
let $\activation$ be the \softsign{}
\cfload.
Then we say that $\Activation$ is the 
$ d $-dimensional softsign activation function
if and only if
$
	\Activation
=
  	\multdim_{\activation, d}
$
\cfout.
\end{adef}

\subsection{Leaky rectified linear unit (leaky ReLU) activation}
\label{subsect:leaky_relu}

\newcommand{\lReLU}{leaky \ReLU\ activation function\cfadd{def:lReLU}}
\begin{adef}{def:lReLU}[Leaky \ReLU\ activation functions]
Let $\gamma \in [0,\infty)$.
Then
we say that $a$ is the \lReLU{} with leak factor $\gamma$
if and only if 
it holds that $a \colon \R \to \R$ is the function from $\R$ to $\R$ which satisfies for all
	$x \in \R$
that
\begin{equation}
\label{def:lReLU:eq1}
\begin{split} 
	a(x)
=
	\begin{cases}
		x &\colon x > 0 \\
		\gamma x &\colon x \leq 0.
	\end{cases}
\end{split}
\end{equation}
\end{adef}

\begin{figure}[!ht]
	\centering
	\includegraphics[width=0.5\linewidth]{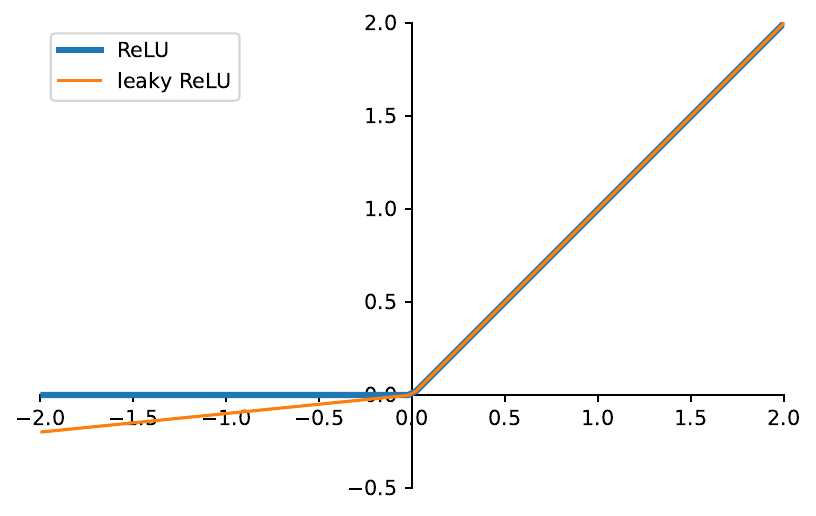}
	\caption{\label{fig:leaky_relu_plot}A plot of the leaky \ReLU\ activation function with leak factor $\nicefrac1{10}$ and the \ReLU\ activation function}
\end{figure}

\filelisting{code:leaky_relu_plot}{code/activation_functions/leaky_relu_plot.py}{{\sc Python} code used to create \cref{fig:leaky_relu_plot}}

\cfclear
\begin{lemma}
\label{LReLU_alternative}
Let $\gamma \in [0,1]$ and let $\activation \colon \R \to \R$ be a function.
Then
$\activation$ is the \lReLU{} with leak factor $\gamma$ if and only if it holds for all
	$x \in \R$
that
\begin{equation}
\label{LReLU_alternative:concl1}
\begin{split} 
	\activation(x)
=
	\max\{x, \gamma x \}
\end{split}
\end{equation}
\cfout.
\end{lemma}

\begin{proof}[Proof of \cref{LReLU_alternative}]
Note that 
\enum{
	the fact that $\gamma \leq 1$;
	\eqref{def:lReLU:eq1} 
}[establish]
\eqref{LReLU_alternative:concl1}.
The proof of \cref{LReLU_alternative} is thus complete.
\end{proof}

\cfclear
\begin{athm}{lemma}{leaky_derivative_as}
Let $u, \beta \in \R$, $v \in (u, \infty)$, $\alpha \in (-\infty, 0]$, 
let $a_1$ be the \softplusfunc{},
let $a_2$ be the \gelu{},
let $a_3$ be the \logfunction{},
let $a_4$ be the \swishfunc{} with parameter $\beta$,
let $a_5$ be the \softsign{},
and  
let $l$ be the \lReLU{} with leaky parameter $\gamma$
\cfload.
Then
\begin{enumerate}[label=(\roman *)]
\item 
\label{leaky_derivative_as:item1}
	it holds for all $f \in \{\rect, \clip uv, \tanhh, a_1, a_2, \ldots, a_5\}$ that
	$\limsup_{x \to -\infty} |f'(x)| = 0$
	and
\item 
\label{leaky_derivative_as:item2}
it holds that 
$\lim_{x \to -\infty} l'(x) = \gamma$
\end{enumerate}
\cfout.
\end{athm}

\begin{aproof}
\Nobs that
\cref{def:relu1:eq1,def:clip1:eq1,eq:softplus1.def,def:gELU:eq1,def:logistic1:eq1,def:swish1:eq1,def:hyperbolic_tangent1:eq1,def:softsign:eq1}
prove
\cref{leaky_derivative_as:item1}.
\Nobs
that
\cref{def:lReLU:eq1}
establishes
\cref{leaky_derivative_as:item2}.
\end{aproof}

\cfclear
\begin{adef}{def:LReLU}[Multi-dimensional leaky \ReLU\ activation functions]
Let $d \in \N$, $\gamma \in [0,\infty)$ and let $\activation$ be the \lReLU{} with leak factor $\gamma$
\cfload.
Then we say that $\Activation$ is the $d$-dimensional leaky \ReLU\ activation function with leak factor $\gamma$
if and only if
$
	\Activation
=
  	\multdim_{\activation, d}
$
\cfclear\cfadd{def:multidim_version}\cfout.
\end{adef}

\subsection{Exponential linear unit (ELU) activation}

Another popular activation function is the so-called \ELU\ activation function which has been introduced in Clevert et al.~\cite{Clevert2015}.
This activation function is the subject of the next notion.

\begin{adef}{def:eLU}[\ELU\ activation functions]
Let $\gamma \in (-\infty,0]$.
Then
we say that $a$ is the \eLU{} with asymptotic $\gamma$
if and only if 
it holds that $a \colon \R \to \R$ is the function from $\R$ to $\R$ which satisfies for all
	$x \in \R$
that
\begin{equation}
\label{def:eLU:eq1}
\begin{split} 
	a(x)
=
	\begin{cases}
	x &\colon x > 0 \\
	\gamma(1 - \exp(x)) &\colon x \leq 0.
	\end{cases}
\end{split}
\end{equation}
\end{adef}

\begin{figure}[!ht]
	\centering
	\includegraphics[width=0.5\linewidth]{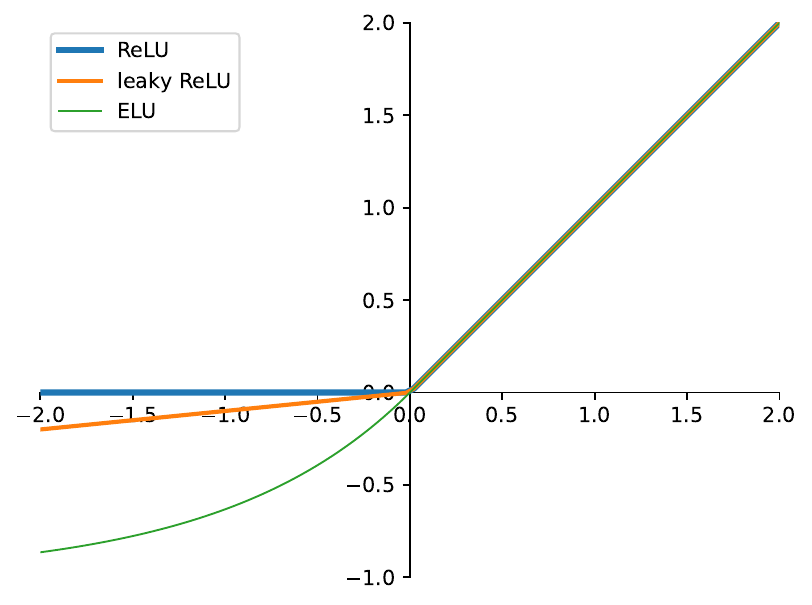}
	\caption{\label{fig:elu_plot}A plot of the \ELU\ activation function with asymptotic $-1$, the leaky \ReLU\ activation function with leak factor $\nicefrac1{10}$, and the \ReLU\ activation function}
\end{figure}

\filelisting{code:elu_plot}{code/activation_functions/elu_plot.py}{{\sc Python} code used to create \cref{fig:elu_plot}}

\cfclear
\begin{lemma}
\label{eLU_assymptotic}
Let $\gamma \in (-\infty,0]$ and let $\activation$ be the \eLU{} with asymptotic $\gamma$
\cfload.
Then 
\begin{equation}
\label{eLU_assymptotic:concl1}
\begin{split} 
	\limsup_{x \to -\infty} \activation(x)
=
	\liminf_{x \to -\infty} \activation(x)
=
	\gamma.
\end{split}
\end{equation}

\end{lemma}

\begin{proof}[Proof of \cref{eLU_assymptotic}]
Observe that 
\enum{
	\eqref{def:eLU:eq1}
}[establish]
\eqref{eLU_assymptotic:concl1}.
The proof of \cref{eLU_assymptotic} is thus complete.
\end{proof}

\cfclear
\begin{adef}{def:ELU}[Multi-dimensional \ELU\ activation functions]
Let $d\in \N$, $\gamma \in (-\infty,0]$ and let $\activation$ be the \eLU{} with asymptotic $\gamma$
\cfload.
Then 
we say that $\Activation$ is the $d$-dimensional \ELU\ activation function with asymptotic $\gamma$
if and only if
$
	\Activation
=
  	\multdim_{\activation, d}
$
\cfclear\cfadd{def:multidim_version}\cfout.
\end{adef}

\subsection{Rectified power unit (RePU) activation}

Another popular activation function is the so-called \RePU\ activation function. %
This concept is the subject of the next notion.

\begin{adef}{def:rePU}[\RePU\ activation functions]
Let $p \in \N$. 
Then
we say that $\activation$ is the \RePU\ activation function with power $p$
if and only if 
it holds that $\activation \colon \R \to \R$ is the function from $\R$ to $\R$ which satisfies for all
	$x \in \R$
that
\begin{equation}
\label{RePU:eq1}
\begin{split} 
	\activation(x)
=
	(\max\{x, 0\})^p.
\end{split}
\end{equation}
\end{adef}

\begin{figure}[!ht]
	\centering
	\includegraphics[width=0.5\linewidth]{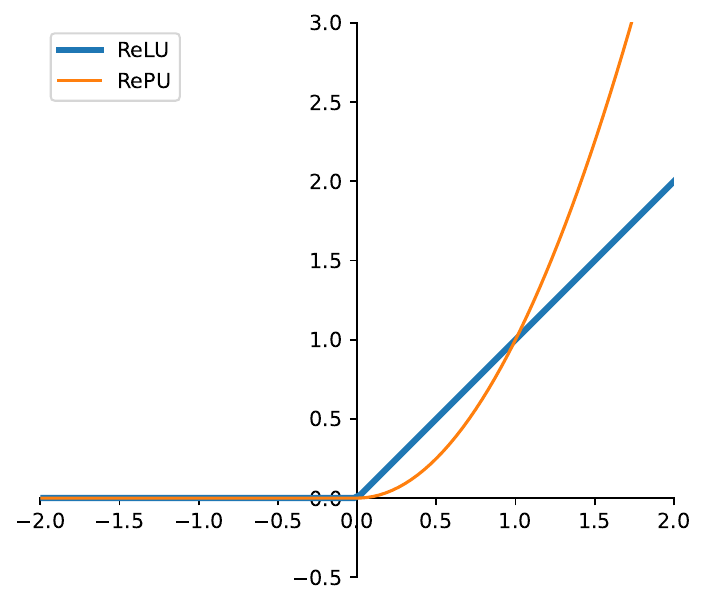}
	\caption{\label{fig:repu_plot}A plot of the \RePU\ activation function with power $2$ and the \ReLU\ activation function}
\end{figure}
  
\filelisting{code:repu_plot}{code/activation_functions/repu_plot.py}{{\sc Python} code used to create \cref{fig:repu_plot}}

\cfclear
\begin{adef}{def:RePU}[Multi-dimensional \RePU\ activation functions]
Let $d, p \in \N$ and let $\activation$ be the \rePU{} with power $p$
\cfload.
Then we say that $\Activation$ is the $d$-dimensional \RePU\ activation function with power $p$ if and only if it holds that
$
	\Activation
=
  	\multdim_{\activation, d}
$
\cfclear\cfadd{def:multidim_version}\cfout.
\end{adef}

\subsection{Sine activation}

The sine function has been proposed as activation function in Sitzmann et al.~\cite{Sitzmann2020}.
This is formulated in the next notion.

\newcommand{\sinefunc}{sine activation function\cfadd{def:sine}}

\begin{adef}{def:sine}[Sine activation function]
We say that $\activation$ is the \sinefunc{}
if and only if 
it holds that $\activation \colon \R \to \R$ is the function from $\R$ to $\R$ which satisfies for all
	$x \in \R$
that
\begin{equation}
	\activation( x )
=
	\sin(x).
\end{equation}
\end{adef}

\begin{figure}[!ht]
	\centering
	\includegraphics[width=0.5\linewidth]{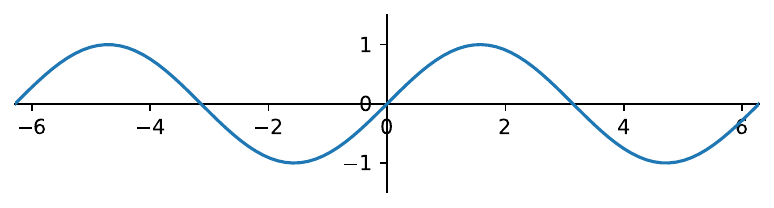}
	\caption{\label{fig:sine_plot}A plot of the sine activation function}
\end{figure}

\filelisting{code:sine_plot}{code/activation_functions/sine_plot.py}{{\sc Python} code used to create \cref{fig:sine_plot}}

\cfclear
\begin{adef}{def:sine_mult}[Multi-dimensional sine activation functions]
Let $d \in \N$ and
let $\activation$ be the \sinefunc{}
\cfload.
Then we say that $\Activation$ is the 
$ d $-dimensional sine activation function
if and only if it holds that
$
	\Activation
=
  	\multdim_{\activation, d}
$
\cfout.
\end{adef}

\subsection{Heaviside activation}

\begin{adef}{def:heaviside}[Heaviside activation function]
We say that $\activation$ is the Heaviside activation function 
(we say that $ \activation $ is the Heaviside step function, we say that $ \activation $ is the unit step function)
if and only if 
it holds that $\activation \colon \R \to \R$ is the function from $\R$ to $\R$ which satisfies for all
	$x \in \R$
that
\begin{equation}
  \activation( x )
  =
  \ind{ [0,\infty) }( x )
  =
  \begin{cases}
    1
  &
    \colon 
    x \geq 0
  \\
    0
  &
    \colon
    x < 0.
  \end{cases}
\end{equation}
\end{adef}

\begin{figure}[!ht]
	\centering
	\includegraphics[width=.5\linewidth]{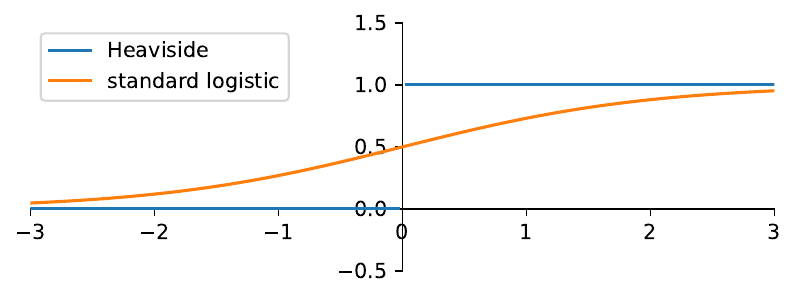}
	\caption{\label{fig:heaviside_plot}A plot of the Heaviside activation function and the standard logistic activation function}
\end{figure}

\filelisting{code:heaviside_plot}{code/activation_functions/heaviside_plot.py}{{\sc Python} code used to create \cref{fig:heaviside_plot}}

\cfclear
\begin{adef}{def:heaviside_mult}[Multi-dimensional Heaviside activation functions]
Let $d \in \N$ and
let $\activation$ be the \heavisidefunction{}
\cfload.
Then we say that $\Activation$ is the 
$ d $-dimensional Heaviside activation function 
(we say that $ \Activation $ is the $ d $-dimensional Heaviside step function, we say that $ \Activation $ is the $ d $-dimensional unit step function)
if and only if it holds that
$
	\Activation
=
  	\multdim_{\activation, d}
$
\cfout.
\end{adef}

\subsection{Softmax activation}

\newcommand{\softmaxfun}{softmax activation function\cfadd{def:softmax}}
\cfclear
\begin{adef}{def:softmax}[Softmax activation functions]
Let $d \in \N$. 
Then we say that $\Activation$ is the $d$-dimensional \softmaxfun{}
if and only if it holds that
$ 
  \Activation 
  \colon \R^{d} \to \R^{d} 
$ 
is the function from $\R^{d}$ to $\R^{d}$ 
which satisfies 
for all $x = (x_1,\ldots, $ $ x_d) $ $ \in \R^d$ that 
\begin{equation}
\label{def:softmax:eq1}
\begin{split}
	\Activation(x)
& =
	\pr*{
		\tfrac{\exp(x_1)}{\pr*{\sum_{i = 1}^d \exp(x_i) }},
		\tfrac{\exp(x_2)}{\pr*{\sum_{i = 1}^d \exp(x_i) }},
		\ldots,
		\tfrac{\exp(x_d)}{\pr*{\sum_{i = 1}^d \exp(x_i) }}
	}.
\end{split}
\end{equation}
\end{adef}

\cfclear
\begin{lemma}
\label{softmax_distribution}
Let $ d \in \N $
and let 
$\Activation
 =
  (
    \Activation_{ 1 } ,\allowbreak
    \dots ,\allowbreak
    \Activation_{ d } 
  )
$ 
be the $d$-dimensional \softmaxfun{}
\cfload.
Then 
\begin{enumerate}[label=(\roman{*})]
\item \label{softmax_distribution:item1}
 it holds for all $x \in \R^d$, $k \in \{1,2,\ldots,d\}$ that $\Activation_{k}(x) \in (0,1]$ 
 and

\item \label{softmax_distribution:item2}
 it holds for all $x \in \R^d$ that
\begin{equation}
\sum_{k = 1}^d \Activation_{k}(x) = 1.
\end{equation}tum

\end{enumerate}
(cf.\ \cref{def:softmax}).
\end{lemma}

\begin{proof}[Proof of \cref{softmax_distribution}]
Observe 
that
\eqref{def:softmax:eq1}
demonstrates that for all $x = (x_1,\ldots, x_d) \in \R^d$ it holds that
\begin{equation}
	\sum_{k = 1}^d \Activation_{k}(x) 
=
	 \sum_{k = 1}^d \tfrac{\exp(x_k)}{\pr*{\sum_{i = 1}^d \exp(x_i) }}
=
	\tfrac{ \sum_{k = 1}^d \exp(x_k)}{\sum_{i = 1}^d \exp(x_i)}
=
	1.
\end{equation}
The proof of \cref{softmax_distribution} is thus complete.
\end{proof}

\endgroup

\section{Fully-connected feedforward ANNs (structured description)}
\label{subsec:structured_description}

In this section we present an alternative way to describe the fully-connected feedforward \anns\ introduced in \cref{subsec:vectorized_description} above.
Roughly speaking, in \cref{subsec:vectorized_description} above we defined a \emph{vectorized description} of fully-connected feedforward \anns\ in the sense that the trainable parameters of a fully-connected feedforward \ann\ are represented by the components of a single Euclidean vector (cf.\ \cref{def:FFNN} above).
In this section we introduce a \emph{structured description} of fully-connected feedforward \anns\ in which the trainable parameters of a fully-connected feedforward \ann\ are represented by a tuple of matrix-vector pairs corresponding to the weight matrices and bias vectors of the fully-connected feedforward \anns\ (cf.\ \cref{def:ANN,def:ANNrealization} below).

\subsection{Structured description of fully-connected feedforward ANNs}
\label{subsubsec:structured_description_of_DNNs}

\begin{adef}{def:ANN}[Structured description of fully-connected feedforward \anns]
We denote by $ \ANNs $ the set given by 
\begin{equation}
\label{eq:defANN}
\begin{split}
\textstyle
  \ANNs
& 
\textstyle
  =
  \bigcup_{ L \in \N }
  \bigcup_{ l_0, l_1, \dots, l_L \in \N }
  \pr*{
    \bigtimes_{ k = 1 }^L 
    ( \R^{ l_k \times l_{ k - 1 } } \times \R^{ l_k } )
  }
  ,
\end{split}
\end{equation}
for every 
$ L \in \N $, 
$ l_0, l_1, \dots, l_L \in \N $, 
$
  \Phi 
  \in \allowbreak
  \bigl( 
    \bigtimes_{ k = 1 }^L
    \allowbreak( \R^{ l_k \times l_{ k - 1 } } \times \R^{ l_k } )
  \bigr)
  \subseteq \ANNs
$
we denote by
$
	\paramANN(\Phi),\lengthANN(\Phi),\inDimANN(\Phi),\outDimANN(\Phi), \hiddenLength(\Phi) \in \N_0
$
the numbers given by
\begin{equation}
\label{def:ANN:eq1}
\textstyle 
  \paramANN(\Phi)
  =
  \sum_{ k = 1 }^L l_k( l_{ k - 1 } + 1 ) 
  , 
\,\,\,
  \lengthANN( \Phi ) = L ,  
\,\,\,
  \inDimANN( \Phi ) = l_0 ,  
\,\,\,
  \outDimANN( \Phi ) = l_L , 
\,\,\,
\text{and}
\,\,\,
  \hiddenLength( \Phi ) = L - 1 , 
\end{equation}
for every 
$ n \in \N_0 $,
$ L \in \N $, 
$ l_0, l_1, \dots, l_L \in \N $, 
$
  \Phi 
  \in \allowbreak
  \bigl( 
    \bigtimes_{ k = 1 }^L
    \allowbreak( \R^{ l_k \times l_{ k - 1 } } \times \R^{ l_k } )
  \bigr)
  \subseteq \ANNs
$
we denote by
$
  \dimANNlevel_n(\Phi) \in \N_0
$ 
the number given by
\begin{equation}
\label{def:ANN:eq1_B}
\begin{split}
  \dimANNlevel_n (\Phi) 
  =
  \begin{cases}
    l_n & \colon n \leq L 
\\
    0 & \colon n > L ,
  \end{cases}
  \end{split}
\end{equation}
for every
$ \Phi \in \ANNs$
we denote by
$
	\dims(\Phi) \in \N^{\lengthANN( \Phi )+1}
$
the tuple given by
\begin{equation}
\begin{split} 
  \dims( \Phi ) = ( \dimANNlevel_0(\Phi), \dimANNlevel_1(\Phi), \dots, \dimANNlevel_{\lengthANN( \Phi )}(\Phi) ),
\end{split}
\end{equation}
and for every 
$ L \in \N $, 
$ l_0, l_1, \dots, l_L \in \N $, 
$
  \Phi =
  ( ( W_1, B_1 ), \dots, ( W_L, B_L ) )
  \in 
  \allowbreak
  \bigl( 
    \bigtimes_{ k = 1 }^L 
    \allowbreak( \R^{ l_k \times l_{ k - 1 } } \times \R^{ l_k } )
  \bigr)
  \subseteq \ANNs
$,
$ n \in \{ 1, 2, \dots, L \} $
we denote by 
$
	\weightANN{ n }{ \Phi } \in \R^{ l_n \times l_{ n - 1 } }
$,
$
	\biasANN{ n }{ \Phi } \in \R^{ l_n }
$
the matrix and the vector given by
\begin{equation}
	\label{eq:defANNweightsbiases}
  \weightANN{n}{\Phi} = W_n 
\qquad 
  \text{and}
\qquad 
  \biasANN{n}{\Phi} = B_n
  .
\end{equation}
\end{adef}

\cfclear
\begin{adef}{def:neuralnetwork}[Fully-connected feedforward \anns]
  We say that $\Phi$ is a fully-connected feedforward \ann\ 
  (we say that $\Phi$ is an \ann)
  if and only
  if it holds that 
  \begin{equation}
    \Phi \in \ANNs
  \end{equation}
  \cfload.
\end{adef}

\cfclear
\begin{lemma}\label{Lemma:elementaryPropertiesANN}
Let $\Phi\in\ANNs$ \cfload. Then 
\begin{enumerate}[label=(\roman{*})]
\item
\label{elementaryPropertiesANN:ItemOne} 
it holds that 
$ \dims( \Phi ) \in \N^{ \lengthANN( \Phi ) + 1 } $,
\item\label{elementaryPropertiesANN:inout}
it holds that 
\begin{equation}
	\inDimANN(\Phi)=\dimANNlevel_0(\Phi)
	\qquad\text{and}\qquad
	\outDimANN(\Phi)=\dimANNlevel_{\lengthANN(\Phi)}(\Phi),
\end{equation}
and
\item\label{elementaryPropertiesANN:weightsbiases} 
it holds for all $n\in\{1,2,\dots,\lengthANN(\Phi)\}$
that
\begin{equation}
	\weightANN n\Phi\in\R^{\dimANNlevel_{n}(\Phi)\times\dimANNlevel_{n-1}(\Phi)}
	\qquad\text{and}\qquad
	\biasANN n\Phi\in\R^{\dimANNlevel_n(\Phi)}.
\end{equation}
\end{enumerate}
\cfout.
\end{lemma} 
\begin{proof}[Proof of \cref{Lemma:elementaryPropertiesANN}]	
\Nobs that the assumption that 
\begin{equation*}
\textstyle 
  \Phi\in \ANNs
  = \bigcup_{ L \in \N }
  \bigcup_{ (l_0,l_1,\ldots, l_L) \in \N^{L+1} }
  \bpr{
    \bigtimes_{k = 1}^L (\R^{l_k \times l_{k-1}} \times \R^{l_k})
	}
\end{equation*}
ensures that there exist $ L \in \N $, $ l_0, l_1, \dots, l_L \in \N $ 
which satisfy that
\begin{equation}\label{elementaryPropertiesANN:SetForPhi}
  \Phi\in 
  \bpr{
     \textstyle\bigtimes_{ k = 1 }^L 
     ( \R^{ l_k \times l_{k-1} } \times \R^{ l_k } )
  }
  .
\end{equation}
\Nobs that \cref{elementaryPropertiesANN:SetForPhi}, \cref{def:ANN:eq1},
and \cref{def:ANN:eq1_B} imply that 
\begin{equation}
   \lengthANN(\Phi)=L,\qquad \inDimANN(\Phi)=l_0=\dimANNlevel_0(\Phi),\qquad\text{and}\qquad
	  \outDimANN(\Phi)=l_L=\dimANNlevel_{L}(\Phi).
 \end{equation}
This shows that
 \begin{equation}\label{elementaryPropertiesANN:Dims}
  \dims(\Phi)=(l_0,l_1,\dots, l_L)\in \N^{L+1}=\N^{\lengthANN(\Phi)+1}.
\end{equation}
\Moreover 
\cref{elementaryPropertiesANN:SetForPhi},
\cref{def:ANN:eq1_B},
and \cref{eq:defANNweightsbiases}
ensure that for all $n\in\{1,2,\dots,\lengthANN(\Phi)\}$
it holds that
\begin{equation}
	\weightANN n\Phi\in \R^{l_n\times l_{n-1}}=\R^{\dimANNlevel_{n}(\Phi)\times\dimANNlevel_{n-1}(\Phi)}
	\qquad\text{and}\qquad
	\biasANN n\Phi\in\R^{l_n}=\R^{\dimANNlevel_n(\Phi)}.
\end{equation}
 	The proof of \cref{Lemma:elementaryPropertiesANN} is thus complete.
 \end{proof}

\subsection{Realizations of fully-connected feedforward ANNs}
\label{subsubsec:realizations_of_dnns}

\cfclear
\begin{adef}{def:ANNrealization}[Realizations of fully-connected feedforward \anns]
\cfconsiderloaded{def:ANNrealization}
Let 
$\Phi\in\ANNs$
and let
$a \colon \R \to \R$
be a function
\cfload.
Then we denote by
\begin{equation}
	\functionANN{a}(\Phi) 
\colon 
	\R^{\inDimANN(\Phi)} \to \R^{\outDimANN(\Phi)}
\end{equation}
the function which satisfies for all
$x_0\in \R^{\dimANNlevel_0(\Phi)},\,\allowbreak
x_1\in \R^{\dimANNlevel_1(\Phi)},\,\allowbreak
\dots,\,
x_{\lengthANN(\Phi)}\in \R^{\dimANNlevel_{\lengthANN(\Phi)}(\Phi)}$
with 
\begin{equation}
\begin{split} 
\forall \, k \in \{1,2,\dots,\lengthANN(\Phi)\} \colon 
 x_k =\multdim_{a\indicator{(0,\lengthANN(\Phi))}(k)+\id_\R\indicator{\{\lengthANN(\Phi)\}}(k),\dimANNlevel_k(\Phi)}(\weightANN k\Phi x_{k-1} + \biasANN k\Phi)
\end{split}
\end{equation}
that
\begin{equation}
\label{setting_NN:ass2}
( \functionANN{a}(\Phi) ) (x_0) = x_{\lengthANN(\Phi)}
\end{equation}
and
we call $\functionANN{a}(\Phi)$ the realization function of the fully-connected feedforward \ann\ $\Phi$ with activation function $a$
(we call $\functionANN{a}(\Phi)$ the realization of the fully-connected feedforward \ann\ $\Phi$ with activation $a$)
\cfload.
\end{adef}

\todoc{ARNULF: ok?}
\begin{athm}{remark}{remark:anns_different_names}[Different uses of the term \ann\ in the literature]
In \cref{def:neuralnetwork} above, we defined an \ann\ as a structured tuple of real numbers, or in other words, as a structured set of parameters.
However, in the literature and colloquial usage, the term \ann\ sometimes also refers to a different mathematical object. Specifically, for a given architecture and activation function, it may refer to the function that maps parameters and input to the output of the corresponding realization function.

More formally, let $L \in \N$, $l_0, l_1, \dots, l_L \in \N$, let $a \colon \R \to \R$ be a function, and consider the function
\begin{equation}
\label{T_B_D}
\begin{split} 
    \mathscr{f} 
    \colon 
    \textstyle
    \pr*{
    \bigtimes_{ k = 1 }^L 
    ( \R^{ l_k \times l_{ k - 1 } } \times \R^{ l_k } )
    }
    \times
    \R^{l_0} 
    \to 
    \R^{l_L}
\end{split}
\end{equation}
which satisfies for all
$
    \Phi
    \in 
    \pr[\big]{
    \bigtimes_{ k = 1 }^L 
    ( \R^{ l_k \times l_{ k - 1 } } \times \R^{ l_k } )
    }
$,
$x \in \R^{l_0}$
that
\begin{equation}
    \mathscr{f}(\Phi, x) = \functionANN{a}(\Phi)(x)
\end{equation}
\cfload.
In this context, the function $\mathscr{f}$ itself is sometimes referred to as an \ann.
\end{athm}

\cfclear
\begin{exercise}{ANN_calc}
Let
\begin{equation}
\begin{split} 
	&\Phi
=
	(
		(W_1, B_1), (W_2, B_2), (W_3, B_3)
	)
\in
	(\R^{2 \times 1} \times \R^{2})
	\times
	(\R^{3 \times 2} \times \R^{3})
	\times
	(\R^{1 \times 3} \times \R^{1})
\end{split}
\end{equation}
satisfy
\begin{equation}
\begin{split} 
	W_1
=
	\begin{pmatrix}
		1\\2
	\end{pmatrix},
\qquad
	B_1
=
	\begin{pmatrix}
		3\\4
	\end{pmatrix},
\qquad
	W_{2}
=
	\begin{pmatrix}
		-1 & 2 \\ 3 & -4 \\ -5 & 6 
	\end{pmatrix},
\qquad
	B_2
=
	\begin{pmatrix}
		0 \\ 0 \\ 0
	\end{pmatrix},
\end{split}
\end{equation}
\begin{equation}
\begin{split} 
	W_3
=
	\begin{pmatrix}
		-1 & 1 & -1
	\end{pmatrix},
\qandq
	B_3
=
	\begin{pmatrix}
		-4
	\end{pmatrix}.
\end{split}
\end{equation}
Prove or disprove the following statement:
It holds that
\begin{equation}
\begin{split} 
	(\functionANN{\rect}(\Phi))(-1)
=
	0
\end{split}
\end{equation}
\cfout.
\end{exercise}

\cfclear
\begin{exercise}{ex:represent_logistic_with_tanh}
	Let $\activation$ be the \logfunction{}
	\cfload.
  Prove or disprove the following statement: There exists $\Phi\in\ANNs$ such that
  \begin{equation}
    \functionANN{\tanhh}(\Phi)=\activation\ifnocf.
  \end{equation}
  \cfload[.]
\end{exercise}

\filelisting{code:fc-ann-manual}{code/fc-ann-manual.py}{
		{\sc Python} code for implementing a fully-connected feedforward \ann\ in 
		{\sc PyTorch}. 
		The model created here represents the fully-connected feedforward \ann\
		$
			\pr*{\pr*{\spmat{1&0\\0&-1\\-2&2},\spmat{0\\2\\-1}},
			\pr*{\spmat{1&-2&3},\spmat{1}}}\in (\R^{3\times 2}\times \R^3)\times(\R^{1\times 3}\times\R^1)\subseteq \ANNs
		$ using the \ReLU\ activation function after the hidden layer.
	}

\filelisting{code:fc-ann}{code/fc-ann.py}{
	{\sc Python} code for implementing a fully-connected feedforward \ann\ 
	in {\sc PyTorch}. The model implemented here represents a fully-connected
	feedforward \ann\ with two hidden layers, 3 neurons in the input layer,
	20 neurons in the first hidden layer, 30 neurons in the second hidden layer,
	and 1 neuron in the output layer. Unlike \cref{code:fc-ann-manual},
	this code uses the {\tt torch.nn.Linear} class to represent the affine
	transformations.
}

\filelisting{code:fc-ann2}{code/fc-ann2.py}{
		{\sc Python} code for creating a fully-connected feedforward \ann\ in 
		{\sc PyTorch}. This creates the same model as \cref{code:fc-ann} but
		uses the {\tt torch.nn.Sequential} class instead of defining
		a new subclass of {\tt torch.nn.Module}.
	}

\subsection{On the connection to the vectorized description}

\begin{adef}{def:TranslateStructuredIntoVectorizedDescription}[Transformation from the structured to the vectorized description of fully-connected feedforward \anns] 
  We denote by 
  $\MappingStructuralToVectorized\colon \ANNs \to \bigl(\bigcup_{d\in\N} \R^d \bigr)$
  the function which satisfies for all 
	$\Phi\in\ANNs$,
  $k\in\{1,2,\dots,\lengthANN(\Phi)\}$,
  $ d \in \N $, 
  $\theta=(\theta_1,\dots,\theta_d)\in\R^d$
  with $\MappingStructuralToVectorized(\Phi)=\theta$
  that
  \begin{multline}
  \label{eq:translate}
    d
    =
    \paramANN(\Phi),
\qquad
		\biasANN k\Phi
    =
    \begin{pmatrix}
      \theta_{(\sum_{i=1}^{k-1}l_i(l_{i-1}+1))+l_kl_{k-1}+1}\\
      \theta_{(\sum_{i=1}^{k-1}l_i(l_{i-1}+1))+l_kl_{k-1}+2}\\
      \theta_{(\sum_{i=1}^{k-1}l_i(l_{i-1}+1))+l_kl_{k-1}+3}\\
      \vdots\\
      \theta_{ (\sum_{i=1}^{k-1}l_i(l_{i-1}+1))+l_kl_{k-1}+l_k }
    \end{pmatrix}
    ,
 \qquad 
     \text{and}\qquad
		 \weightANN k\Phi
		 =
    \\
    \begin{pmatrix}
      \theta_{(\sum_{i=1}^{k-1}l_i(l_{i-1}+1))+1} & \theta_{(\sum_{i=1}^{k-1}l_i(l_{i-1}+1))+2} & \cdots & \theta_{(\sum_{i=1}^{k-1}l_i(l_{i-1}+1))+l_{k-1}} \\
      \theta_{(\sum_{i=1}^{k-1}l_i(l_{i-1}+1))+l_{k-1}+1} & \theta_{(\sum_{i=1}^{k-1}l_i(l_{i-1}+1))+l_{k-1}+2} & \cdots & \theta_{(\sum_{i=1}^{k-1}l_i(l_{i-1}+1))+2l_{k-1}} \\
      \theta_{(\sum_{i=1}^{k-1}l_i(l_{i-1}+1))+2l_{k-1}+1} & \theta_{(\sum_{i=1}^{k-1}l_i(l_{i-1}+1))+2l_{k-1}+2} & \cdots & \theta_{(\sum_{i=1}^{k-1}l_i(l_{i-1}+1))+3l_{k-1}} \\
      \vdots & \vdots & \ddots & \vdots \\
      \theta_{(\sum_{i=1}^{k-1}l_i(l_{i-1}+1))+(l_k-1)l_{k-1}+1} & \theta_{(\sum_{i=1}^{k-1}l_i(l_{i-1}+1))+(l_k-1)l_{k-1}+2} & \cdots & \theta_{(\sum_{i=1}^{k-1}l_i(l_{i-1}+1))+l_kl_{k-1}}
    \end{pmatrix}
  \end{multline}
  (cf.\ \cref{def:ANN}).
\end{adef}

\cfclear
\begin{athm}{example}{structtovectB_easy}
Let $\Phi \in (\R^{3 \times 3} \times \R^3) \times (\R^{2 \times 3} \times \R^2)$ satisfy
\begin{equation}
\label{structtovectB_easy:concl1}
\begin{split} 
	\Phi
=
	\pr*{
		\pr*{
			\begin{pmatrix}
				1&2&3\\
				4&5&6\\
				7&8&9\\
			\end{pmatrix}
			,
			\begin{pmatrix}
				10\\11\\12
			\end{pmatrix}
		}
		,
		\pr*{
			\begin{pmatrix}
				13&14&15\\
				16&17&18\\
			\end{pmatrix}
			,
			\begin{pmatrix}
				19\\20
			\end{pmatrix}
		}
	}.
\end{split}
\end{equation}
Then 
$
	\MappingStructuralToVectorized(\Phi)
=
	(1, 2, 3, \ldots, 19, 20)
\in 
	\R^{20}
$.
\end{athm}

\begin{aproof}
\Nobs that \eqref{eq:translate} establishes \eqref{structtovectB_easy:concl1}.
\end{aproof}

\cfclear
\begin{lemma}
\label{lem:structtovectB}
	Let 
	  $ a, b \in\N$, 
	$ 
	  W
	  = ( W_{ i, j } )_{ (i,j) \in \{1,2,\dots,a\} \times \{1,2,\dots,b\}}
	  \in\R^{ a \times b }
	$, 
	$ B = ( B_1,\dots,\allowbreak B_a ) \in \R^{a} $. 
	Then
  \begin{multline}
    \label{eq:structtovect2B}
    \MappingStructuralToVectorized\bigl( ((W, B)) \bigr)\\
    =
    \bigl( W_{1,1}, W_{1,2}, \dots, W_{1,b},
    W_{2,1}, W_{2,2}, \dots, W_{2,b}, \dots, 
    W_{a,1}, W_{a,2}, \dots, W_{a,b}, 
    B_1, B_2, \dots, B_a \bigr)
  \end{multline}
\cfout.
\end{lemma}
\begin{proof}[Proof of \cref{lem:structtovectB}]
Observe that \cref{eq:translate} 
establishes \cref{eq:structtovect2B}.
The proof of \cref{lem:structtovectB} is thus complete.
\end{proof}

\cfclear
\begin{lemma}
  \label{lem:structtovect}
	Let 
	  $L\in\N$, 
	  $l_0,l_1,\dots,l_L\in\N$
	and
	for every $k\in\{1,2,\dots,L\}$
	let 
	  $W_k=(W_{k,i,j})_{(i,j)\in\{1,2,\dots,l_k\}\times\{1,2,\dots,l_{k-1}\}}\in\R^{l_k\times l_{k-1}}$, 
	  $B_k=(B_{k,1},\dots,B_{k,l_k})\in\R^{l_k}$.
	Then
  \begin{equation}
    \label{eq:structtovect1}
  \begin{split}
  &
    \MappingStructuralToVectorized\Bigl(
      \bigl(
        (W_1,B_1),(W_2,B_2),\dots,(W_L,B_L)
      \bigr)
    \Bigr)
  \\ &
    =
    \Bigl(
      W_{1,1,1},W_{1,1,2},\dots,W_{1,1,l_0}, 
      \dots ,
      W_{1,l_1,1},W_{1,l_1,2},\dots,W_{1,l_1,l_0},
      B_{1,1},B_{1,2},\dots,B_{1,l_1},
    \\
    &
    \qquad
      W_{2,1,1},W_{2,1,2},\dots,W_{2,1,l_1}, 
      \dots ,
      W_{2,l_2,1},W_{2,l_2,2},\dots,W_{2,l_2,l_1},
      B_{2,1},B_{2,2},\dots,B_{2,l_2},
    \\
    &
    \qquad
      \dots ,
    \\
    &
    \qquad 
      W_{L,1,1},W_{L,1,2},\dots,W_{L,1,l_{L-1}}, 
      \dots,
      W_{L,l_L,1},W_{L,l_L,2},\dots,W_{L,l_L,l_{L-1}},
      B_{L,1},B_{L,2},\dots,B_{L,l_L}
    \Bigr)
  \end{split}
  \end{equation}
\cfout.
\end{lemma}
\begin{proof}[Proof of \cref{lem:structtovect}]
Note that 
\cref{eq:translate} 
implies \eqref{eq:structtovect1}.
The proof of \cref{lem:structtovect} is thus complete.
\end{proof}

\begin{exercise}{ex:translate_injective}
  Prove or disprove the following statement:
  The function $\MappingStructuralToVectorized$
  is injective (cf.\ \cref{def:TranslateStructuredIntoVectorizedDescription}).
\end{exercise}

\begin{exercise}{ex:translate_surjective}
  Prove or disprove the following statement:
  The function $\MappingStructuralToVectorized$
  is surjective (cf.\ \cref{def:TranslateStructuredIntoVectorizedDescription}).
\end{exercise}

\begin{exercise}{ex:translate_bijective}
  Prove or disprove the following statement:
  The function $\MappingStructuralToVectorized$
  is bijective (cf.\ \cref{def:TranslateStructuredIntoVectorizedDescription}).
\end{exercise}

\cfclear
\begin{prop}
  \label{lem:structvsvectgen}
  Let $a \colon \R \to \R$ be a function and let $\Phi\in\ANNs$.
\cfload.
  Then 
  \begin{equation}
  \label{eq:structvsvectgen}
    \functionANN{a}(\Phi)
    =
    \begin{cases}
      \RealV{\MappingStructuralToVectorized(\Phi)}{0}{\inDimANN(\Phi)}{\id_{\R^{\outDimANN(\Phi)}}} &\colon {\hiddenLength(\Phi)}=0\\[0.3cm]
      \RealV{\MappingStructuralToVectorized(\Phi)}{0}{\inDimANN(\Phi)}{\multdim_{a,\dimANNlevel_1(\Phi)},\multdim_{a,\dimANNlevel_2(\Phi)},\ldots,\multdim_{a,\dimANNlevel_{\hiddenLength(\Phi)}(\Phi)},\id_{\R^{\outDimANN(\Phi)}}} & \colon \hiddenLength(\Phi)>0
    \end{cases}
  \end{equation}
  \cfout.
\end{prop}
\begin{proof}[Proof of \cref{lem:structvsvectgen}]
Throughout this proof,
let 
	$L\in \N$,
	$l_0,l_1,\dots,l_L\in\N$
satisfy
\begin{equation}
	\lengthANN(\Phi)=L 
\qandq
	\dims(\Phi)=(l_0,l_1,\dots, l_L).
\end{equation}
  Note that
    \cref{eq:translate}
  shows that for all
    $k\in\{1,2,\dots,L\}$,
    $x\in\R^{l_{k-1}}$ 
  it holds that
  \begin{equation}
  \label{eq:asaff}
    \weightANN{k}{\Phi}x+\biasANN{k}{\Phi}
    = 
    \bigl(\Aff_{l_k,l_{k-1}}^{\MappingStructuralToVectorized(\Phi),\sum_{i=1}^{k-1}l_i(l_{i-1}+1)}\bigr)( x )
  \end{equation}
  \cfload.
  This demonstrates that for all 
    $x_0\in\R^{l_0}$, $x_1\in\R^{l_1}$, $\dots $, $x_{L-1}\in\R^{l_{L-1}}$
    with 
    $\forall \, k \in \{1,2,\dots,L-1\} \colon x_k =\multdim_{a,l_k}(\weightANN{k}{\Phi} x_{k-1}+\biasANN{k}{\Phi})$
  it holds that
  \begin{align}
    x_{L-1}=
    \begin{cases}
      x_0&\colon L=1\\
    \begin{aligned}\bigl(
      &\multdim_{a,l_{L-1}}
      \circ
      \Aff_{l_{L-1},l_{L-2}}^{\MappingStructuralToVectorized(\Phi),\sum_{i=1}^{L-2}l_i(l_{i-1}+1)}
      \\&\quad\circ
      \multdim_{a,l_{L-2}}
      \circ
      \Aff_{l_{L-2},l_{L-3}}^{\MappingStructuralToVectorized(\Phi),\sum_{i=1}^{L-3}l_i(l_{i-1}+1)}
      \circ
      \ldots
      \circ
      \multdim_{a,l_1}
      \circ
      \Aff_{l_1,l_{0}}^{\MappingStructuralToVectorized(\Phi),0}
    \bigr)(x_0)\end{aligned}&\colon L>1
    \end{cases}
  \end{align}
  (cf.\ \cref{def:multidim_version}).
  This, \cref{eq:asaff},
  \cref{eq:FFNN}, and \cref{setting_NN:ass2}
  \prove[ps] that for all 
    $x_0\in\R^{l_0},\,x_1\in\R^{l_1},\dots,\,x_L\in\R^{l_L}$ 
    with $\forall \, k \in \{1,2,\dots,L\} \colon x_k =\multdim_{a\indicator{(0,L)}(k)+\id_\R\indicator{\{L\}}(k),l_k}(\weightANN{k}{\Phi} x_{k-1}+\biasANN{k}{\Phi})$  
  it holds that
  \begin{equation}
  \begin{split}
    \bigl(\functionANN{a}(\Phi)\bigr)(x_0)
    =
    x_L
    &=
    \weightANN{L}{\Phi} x_{L-1}+\biasANN{L}{\Phi}
    =
    \bigl(\Aff_{l_{L},l_{L-1}}^{\MappingStructuralToVectorized(\Phi),\sum_{i=1}^{L-1}l_i(l_{i-1}+1)}\bigr)(x_{L-1})
    \\&=
    \begin{cases}
      \bigl(\RealV{\MappingStructuralToVectorized(\Phi)}{0}{l_0}{\id_{\R^{l_L}}} \bigr)(x_0) & \colon L=1 \\[0.2cm]
      \bigl(\RealV{\MappingStructuralToVectorized(\Phi)}{0}{l_0}{\multdim_{a,l_1},\multdim_{a,l_2},\ldots,\multdim_{a,l_{L-1}},\id_{\R^{l_L}}} \bigr)(x_0) & \colon L>1
    \end{cases}
  \end{split}
  \end{equation}
  (cf.\ \cref{def:ANNrealization,def:FFNN}).
  The proof of \cref{lem:structvsvectgen} is thus complete.
\end{proof}

\section{Convolutional ANNs (CNNs)}
\label{section:cnns}

\todoc{@BENNO: Add an image? (Can be postponed to third version)}

In this section we review \cnns, which are \anns\ designed to process data with a spatial structure.
In a broad sense, \cnns\ can be thought of as any \anns\ involving a convolution operation (cf, \eg, \cref{def:convolution} below).
Roughly speaking, convolutional operations allow \cnns\ to exploit spatial invariance of data by performing the same operations across different regions of an input data point.
In principle, such convolution operations can be employed in combinations with other \ann\ architecture elements, such as
	fully-connected layers (cf., \eg, \cref{subsec:structured_description,subsec:vectorized_description} above),
	residual layers (cf., \eg, \cref{section:resnets} below), and 
	recurrent structures (cf., \eg, \cref{section:rnns} below).
However, for simplicity we introduce in this section in all mathematical details
feedforward \cnns\ 
	only involving convolutional layers
	based on the discrete convolution operation without \emph{padding} (sometimes called \emph{valid padding}) in \cref{def:convolution}
(see \cref{def:CNNrealisation,def:CNN} below).
We refer, \eg, to
	\cite[Section 12.5]{alpaydin2020introduction},
	\cite[Sectino 1.6.1]{berner_grohs_kutyniok_petersen_2022},
	\cite[Chapter 16]{Calin2020},
	\cite[Section 4.2]{Caterini2018},
	\cite[Chapter 9]{Goodfellow2016},
	and
	\cite{Li2022}
for other introductions on \cnns.

\cnns\ were introduced in LeCun et al.~\cite{LeCun1989} for \CV\ applications.
The first successful modern \cnn\ architecture is widely considered to be the \emph{AlexNet} architecture proposed in
Krizhevsky et al.~\cite{Krizhevsky2012}.
A few other very successful early \cnn\ architecures for \CV\ include \cite{Simonyan2014,He2016,Szegedy2015,Girshick2014,Long2015,Huang2017,Mahendran2016,Sermanet2013}.
While \CV\ is by far the most popular domain of application for \cnns, \cnns\ have also been employed successfully in several other areas.
In particular, 
we refer, \eg, to 
	\cite{Santos2014,Zhang2017,Kim2014,Zeng2014,Gehring2017,Zhang2015}
for applications of \cnns\ to \NLP,
we refer, \eg, to 
	\cite{Oord2013,AbdelHamid2014,Sainath2013,Cakir2017,Choi2017}
for applications of \cnns\ to audio processing,
and
we refer, \eg, to 
	\cite{Zheng2014,Wang2017,Borovykh2017,Karim2018,Rajpurkar2017,Ding2015}
for applications of \cnns\ to time series analysis.
Finally, for approximation results for feedforward \cnns\ 
we refer, \eg, to Petersen \& Voigtländer~\cite{petersen2020equivalence} and the references therein. %

\subsection{Discrete convolutions}

\begin{adef}{def:convolution}[Discrete convolutions]
	Let
		$T \in \N$,
		$a_1, a_2, \ldots, a_T,\allowbreak  w_1, w_2, \ldots, w_T, \allowbreak \mathfrak{d}_1,\allowbreak \mathfrak{d}_2, \ldots, \mathfrak{d}_T \in \N$
		and let
		$
			A = 
			(A_{i_1, i_2, \ldots, i_T})_{
				(i_1, i_2, \ldots, i_T) \in 
				(\bigtimes_{t = 1}^T  \{1, 2, \ldots, a_t\} )
			} 
			\in \R^{a_1 \times a_2 \times \ldots \times a_T}$,
			$W =  
			(W_{i_1, i_2, \ldots, i_T})_{
				(i_1, i_2, \ldots, i_T) \in 
				(\bigtimes_{t = 1}^T  \{1, 2, \ldots, w_t\})
			} 
			\in \R^{w_1 \times w_2 \times \ldots \times w_T}
	$
	satisfy for all 
		$ t \in \{1, 2, \ldots, T \}$
	that
	\begin{equation}
	\begin{split} 
	\mathfrak{d}_t = a_t - w_t + 1.
	\end{split}
	\end{equation}
	Then
	we denote by 
	$
		A \convolution W =
			((A \convolution W)_{i_1, i_2, \ldots, i_T})_{
				(i_1, i_2, \ldots, i_T) \in 
				(\bigtimes_{t = 1}^T  \{1, 2, \ldots, \mathfrak{d}_t\})
			} 
		\in \R^{\mathfrak{d}_1 \times \mathfrak{d}_2 \times \ldots \times \mathfrak{d}_T}
	$ 
	the tensor which satisfies for all
		$i_1 \in \{1, 2, \dots, \mathfrak{d}_1\}$,
		$i_2 \in \{1, 2, \dots, \mathfrak{d}_2\}$,
		$\dots$,
		$i_T \in \{1, 2, \dots, \mathfrak{d}_T\}$
	that
	\begin{equation}
	\begin{split} 
		(A \convolution W)_{i_1, i_2, \ldots, i_T}
	=
		\sum_{r_1 = 1}^{w_1} \sum_{r_2 = 1}^{w_2} \dots \sum_{r_T = 1}^{w_T}
			A_{i_1 - 1 + r_1, i_2 - 1 + r_2, \ldots, i_T - 1 + r_T} W_{r_1, r_2, \ldots, r_T}.
	\end{split}
	\end{equation}
\end{adef}

\subsection{Structured description of feedforward CNNs}

\begin{adef}{def:CNN}[Structured description of feedforward \cnns]
  We denote by $\CNNs$ the set given by 
  \begin{multline}
  \label{eq:defCNN}
    \textstyle
    \CNNs
    =\\
    \bigcup_{T, L \in \N}
    \bigcup_{ l_0,l_1,\ldots, l_L \in \N }
    \bigcup_{(c_{{k}, \mathfrak{t}})_{({k}, \mathfrak{t}) \in \{1, 2, \ldots, L\} \times \{1, 2, \ldots, T\}} \subseteq \N}
    \pr*{
      \bigtimes_{{k} = 1}^L 
      \bpr{
      	(\R^{ c_{{k}, 1} \times c_{{k}, 2} \times \ldots \times c_{{k}, T} })^{l_{{k}} \times l_{{k}-1}} \times \R^{l_{{k}}}
      }
    }.
  \end{multline}
\end{adef}

\cfclear
\begin{adef}{def:cnns}[Feedforward \cnns]
  We say that $\Phi$ is a feedforward \cnn\ if and only
  if it holds that 
  \begin{equation}
    \Phi\in\CNNs
  \end{equation}
  \cfload.
\end{adef}

\subsection{Realizations of feedforward CNNs}

\begin{adef}{def:onetensor}[One tensor]
	Let
		$T \in \N$,
		$d_1, d_2, \ldots, d_T \in \N$.
	Then we denote by 
	$
		\onetensor^{d_1, d_2, \ldots, d_T} 
	= 
		(\onetensor^{d_1, d_2, \ldots, d_T}_{i_1, i_2, \ldots, i_T})_{
			(i_1, i_2, \ldots, i_T) 
			\in 
			(\bigtimes_{t = 1}^T \{1, 2, \dots, d_t\})
		} 
		\in \R^{d_1 \times d_2 \times \ldots \times d_T}
	$
	the tensor which satisfies for all
		$i_1 \in \{1, 2, \dots, d_1 \}$,
		$i_2 \in \{1, 2, \dots, d_2 \}$,
		$\dots$,
		$i_T \in \{1, 2, \dots, d_T\}$
	that
	\begin{equation}
	\begin{split} 
		\onetensor^{d_1, d_2, \ldots, d_T}_{i_1, i_2, \ldots, i_T}
	=
		1.
	\end{split}
	\end{equation}
\end{adef}

\begin{adef}{def:CNNrealisation}[Realizations associated to feedforward \cnns]
Let 
	$ T,\allowbreak L \in\N$, 
	$l_0,l_1,\ldots, \allowbreak l_L \in \N$, 
	let $(c_{{k}, \mathfrak{t}})_{({k}, \mathfrak{t}) \in \{1, 2, \ldots, L\} \times \{1, 2, \ldots, T\} } \subseteq \N$,
	let $
		\Phi 
	=
		(
			(
				(W_{k,n,m}\allowbreak )_{(n,m) \in \{1, 2, \ldots, l_k\} \times \{1, 2, \ldots, l_{k-1}\}}
				, \allowbreak 
				(B_{k,n})_{n \in \{1, 2, \ldots, l_k\}}
			)
		)_{k \in \{1, 2, \ldots, L\}}
	\in
			\bigtimes_{{k} = 1}^L ((\R^{c_{{k}, 1} \times c_{{k}, 2} \times \ldots \times c_{{k}, T}})^{l_{{k}} \times l_{{k}-1}} \times \R^{l_{{k}}})
	\subseteq \CNNs
	$,
and let $ a \colon \R \to \R$ be a function.
Then we denote by 
\begin{equation}
\textstyle
\CNNRealisation{a}(\Phi) \colon  
	\pr*{
		\bigcup\limits_{\substack{
			d_1, d_2, \ldots, d_T \in \N \\
			\forall \, t \in \{1, 2, \ldots, T\} \colon d_t - \sum_{{k} = 1}^L (c_{{k}, t}-1) \geq 1
		}}\!\!\!\!
		(\R^{ d_1 \times d_2 \times \ldots \times d_T })^{l_0}
	}
	\to
	\pr*{
		\bigcup\limits_{
					d_1, d_2, \ldots, d_T \in \N
				}
		(\R^{d_1 \times d_2 \times \ldots \times d_T })^{l_L}
	}
\end{equation}
the function which satisfies
for all 
	$(\fd_{{k}, {t}})_{({k}, {t}) \in \{0, 1, \ldots, L\} \times \{1, 2, \ldots, T\}} \subseteq \N$,
	$x_0 = (x_{0,1}, \ldots, x_{0,l_{0}}) \in (\R^{ \fd_{0,1} \times \fd_{0,2} \times \ldots \times \fd_{0, T}})^{l_0}$,
	$x_1 = (x_{1,1}, \ldots, x_{1,l_{1}}) \in (\R^{ \fd_{1,1} \times \fd_{1,2} \times \ldots \times \fd_{1, T}})^{l_1}$,
	$\dots$,
	$x_L = (x_{L,1}, \ldots,\allowbreak  x_{L,l_{L}}) \in (\R^{\fd_{L,1} \times \fd_{L,2} \times \ldots \times \fd_{L, T}})^{l_L}$
with 
\begin{equation}
\begin{split} 
	\forall \, 
		{k} \in \{1,2,\dots,L\}\text{, }
		t \in \{1,2,\dots,T\}  \colon 
		\fd_{{k}, t} = \fd_{{k} - 1,t} - c_{{k},t} + 1
\end{split}
\end{equation}
and
\begin{multline}
\textstyle
\forall \, {k} \in \{1,2,\dots,L\}\text{, }n \in \{1,2,\dots,l_{{k}}\} \colon \\[-2pt]
\textstyle
	x_{{k},n} 
=
	\multdim_{
		a \indicator{(0,L)}({k})
		+
		\id_\R \indicator{\{L\}}({k})
		,
		\fd_{{k},1}, \fd_{{k},2}, \ldots, \fd_{{k},T}
	}
	\pr[Big]{ 
		B_{{k},n} \onetensor^{\fd_{{k},1}, \fd_{{k},2}, \ldots, \fd_{{k},T}} + \sum_{m = 1}^{l_{{k}-1}} x_{k-1,m} \convolution W_{{k},n,m}
	}
\end{multline}
that
\begin{equation}
\label{def:CNNrealisation:eq1}
\begin{split} 
	( \CNNRealisation{a}(\Phi) ) (x_0) = x_L
\end{split}
\end{equation}
and 
we call $\CNNRealisation{a}(\Phi)$ the realization function of the feedforward \cnn\ $\Phi$ with activation function $a$
(we call $\CNNRealisation{a}(\Phi)$ the realization of the feedforward \cnn\ $\Phi$ with activation $a$)
(cf.\ \cref{def:CNN,def:multidim_version,def:onetensor,def:convolution}).
\end{adef}

\filelisting{code:convann}{code/conv-ann.py}{
		{\sc Python} code implementing a feedforward \cnn\
		in {\sc PyTorch}. The implemented model here corresponds to 
		a feedforward \cnn\
		$\Phi\in\CNNs$
		where
		$T=2$,
		$L=2$,
		$l_0=1$,
		$l_1=5$,
		$l_2=5$,
		$(c_{1,1}, c_{1,2}) = (3,3)$,
		$(c_{2,1}, c_{2,2}) = (5,3)$,
		and
		$	\Phi 
		\in 
			\bpr{
				\bigtimes_{{k} = 1}^L 
					\pr[\big]{
						(\R^{ c_{{k}, 1} \times c_{{k}, 2} \times \ldots \times c_{{k}, T} })^{l_{{k}} \times l_{{k}-1}} \times \R^{l_{{k}}}
					}
			}
		=
					((\R^{ 3 \times 3} )^{5 \times 1} \times \R^{5})
			\times
			((\R^{ 3 \times 5})^{5 \times 5} \times \R^{5})
		$.
		The model, given an input of shape $(1, d_1, d_2)$ with
		$d_1\in\N\cap[7,\infty)$, $d_2\in\N\cap[5,\infty)$,
		produces an output of shape $(5, d_1-6, d_2-4)$,
		(corresponding to the realization function
		$\CNNRealisation{a}(\Phi)$ for $a\in C(\R,\R)$ having
		domain
		$\bigcup_{d_1,d_2\in\N,\,d_1\geq 7,\,d_2\geq 5}(\R^{d_1\times d_2})^1$
		and satisfying for all
		$d_1\in \N\cap[7,\infty)$, $d_2\in \N\cap[5,\infty)$,
		$x_0\in(\R^{d_1\times d_2})^{1}$
		that
		$(\CNNRealisation{a}(\Phi))(x_0)\in (\R^{d_1-6, d_2-4})^{5}$).
	}

\cfclear
\begin{athm}{example}{CANN_realization_example}[Example for \cref{def:CNNrealisation}]
Let
	$T = 2$,
	$L= 2$,
	$l_0 = 1$,
	$l_1 = 2$,
	$l_2 = 1$,
	$c_{1, 1} = 2$,
	$c_{1, 2} = 2$,
	$c_{2, 1} = 1$,
	$c_{2, 2} = 1$
and
let
\begin{equation}
\begin{split} 
	\Phi 
\in 
	\pr*{
		\bigtimes_{{k} = 1}^L 
			\pr[\big]{
				(\R^{ c_{{k}, 1} \times c_{{k}, 2} \times \ldots \times c_{{k}, T} })^{l_{{k}} \times l_{{k}-1}} \times \R^{l_{{k}}}
			}
	}
=
		\pr[\big]{
			(\R^{ 2 \times 2} )^{2 \times 1} \times \R^{2}
		}
	\times
		\pr[\big]{
			(\R^{ 1 \times 1})^{1 \times 2} \times \R^{1}
		}
\end{split}
\end{equation}
satisfy
\begin{equation}
\label{CANN_realization_example:ass1}
\begin{split} 
	\Phi
=
	\pr*{
		\pr*{
			\begin{pmatrix}
				\begin{pmatrix}
					0 & 0 \\
					0 & 0 
				\end{pmatrix}	
				\\[10pt]
				\begin{pmatrix}
					1 & 0 \\
					0 & 1
				\end{pmatrix}	
			\end{pmatrix}
			,
			\begin{pmatrix}
				1 \\
				-1
			\end{pmatrix}
		}
		,
		\pr*{
			\begin{pmatrix}
				\begin{pmatrix}
					-2
				\end{pmatrix}	
				&
				\begin{pmatrix}
					2
				\end{pmatrix}	
			\end{pmatrix}
			,
			\begin{pmatrix}
				3
			\end{pmatrix}
		}
	}.
\end{split}
\end{equation}
Then
\begin{equation}
\label{CANN_realization_example:concl1}
\begin{split} 
	\pr[\big]{\CNNRealisation{\rect}(\Phi)}
	\pr*{
		\begin{pmatrix}
			1 & 2 & 3 \\
			4 & 5 & 6 \\
			7 & 8 & 9
		\end{pmatrix}
	}
=
	\begin{pmatrix}
		11 & 15\\
		23 & 27
	\end{pmatrix}
\end{split}
\end{equation}
\cfout.
\end{athm}

\begin{aproof}
Throughout this proof, let
	$x_0 \in \R^{ 3 \times 3}$,
	$x_1 = (x_{1,1}, x_{1,2}) \in (\R^{ 2 \times 2})^{2}$,
	$x_2 \in \R^{2 \times 2}$
with satisfy that
\begin{equation}
\label{CANN_realization_example:setting1}
\begin{split} 
	x_0
=
	\begin{pmatrix}
		1 & 2 & 3 \\
		4 & 5 & 6 \\
		7 & 8 & 9
	\end{pmatrix},
\qquad
	x_{1, 1}
=
	\multdim_{
		\rect
		,
		2, 
		2
	}
	\pr*{
		\onetensor^{2, 2}
		+
		x_0 
		\convolution
		\begin{pmatrix}
			0 & 0\\
			0 & 0
		\end{pmatrix}
	},
\end{split}
\end{equation}
\begin{equation}
\label{CANN_realization_example:setting2}
\begin{split} 
	x_{1, 2}
=
	\multdim_{
		\rect
		,
		2, 
		2
	}
	\pr*{
		(-1)
		\onetensor^{2, 2}
		+
		x_0 
		\convolution
		\begin{pmatrix}
			1 & 0\\
			0 & 1
		\end{pmatrix}
	},
\end{split}
\end{equation}
\begin{equation}
\label{CANN_realization_example:setting3}
\begin{split} 
\andq
	x_{2}
=
	\multdim_{
		\id_\R
		,
		2, 
		2
	}
	\pr*{
		3
		\onetensor^{2, 2}
		+
		x_{1, 1} 
		\convolution
		\begin{pmatrix}
			-2
		\end{pmatrix}
		+
		x_{1, 2} 
		\convolution
		\begin{pmatrix}
			2
		\end{pmatrix}
	}.
\end{split}
\end{equation}
\Nobs that
\enum{
	\eqref{def:CNNrealisation:eq1};
	\eqref{CANN_realization_example:ass1};
	\eqref{CANN_realization_example:setting1};
	\eqref{CANN_realization_example:setting2};
	\eqref{CANN_realization_example:setting3}
}[imply]
that
\begin{equation}
\label{CANN_realization_example:eq1}
\begin{split} 
	\pr[\big]{\CNNRealisation{\rect}(\Phi)}
	\pr*{
		\begin{pmatrix}
			1 & 2 & 3 \\
			4 & 5 & 6 \\
			7 & 8 & 9
		\end{pmatrix}
	}
=
	\pr[\big]{\CNNRealisation{\rect}(\Phi)}
	\pr*{
		x_0
	}
=
	x_2.
\end{split}
\end{equation}
\Moreover
\enum{
	\eqref{CANN_realization_example:setting1}
}[ensure]
that 
\begin{equation}
\label{CANN_realization_example:eq2}
\begin{split} 
	x_{1, 1}
&=
	\multdim_{
		\rect
		,
		2 \times 2
	}
	\pr*{
		\onetensor^{2, 2}
		+
		x_0 
		\convolution
		\begin{pmatrix}
			0 & 0\\
			0 & 0
		\end{pmatrix}
	}
=
	\multdim_{
		\rect
		,
		2 \times 2
	}
	\pr*{
		\begin{pmatrix}
			1 & 1\\
			1 & 1
		\end{pmatrix}
		+
		\begin{pmatrix}
			0 & 0\\
			0 & 0
		\end{pmatrix}
	}\\
&=
	\multdim_{
		\rect
		,
		2 \times 2
	}
	\pr*{
		\begin{pmatrix}
			1 & 1\\
			1 & 1
		\end{pmatrix}
	}
=
	\begin{pmatrix}
		1 & 1\\
		1 & 1
	\end{pmatrix}.
\end{split}
\end{equation}
\Moreover
\enum{
	\eqref{CANN_realization_example:setting2}
}[assure]
that
\begin{equation}
\begin{split} 
	x_{1, 2}
&=
	\multdim_{
		\rect
		,
		2 \times 2
	}
	\pr*{
		(-1)
		\onetensor^{2, 2}
		+
		x_0 
		\convolution
		\begin{pmatrix}
			1 & 0\\
			0 & 1
		\end{pmatrix}
	}
=
	\multdim_{
		\rect
		,
		2 \times 2
	}
	\pr*{
		\begin{pmatrix}
			-1 & -1\\
			-1 & -1
		\end{pmatrix}
		+
		\begin{pmatrix}
			6 & 8\\
			12 & 14
		\end{pmatrix}
	}\\
&=
	\multdim_{
		\rect
		,
		2 \times 2
	}
	\pr*{
		\begin{pmatrix}
			5 & 7\\
			11 & 13
		\end{pmatrix}
	}
=
	\begin{pmatrix}
		5 & 7\\
		11 & 13
	\end{pmatrix}.
\end{split}
\end{equation}
\Moreover
\enum{
	this;
	\eqref{CANN_realization_example:eq2};
	\eqref{CANN_realization_example:setting3}
}[demonstrate]
that
\begin{equation}
\begin{split} 
	x_{2}
&=
	\multdim_{
		\id_\R
		,
		2 \times 2
	}
	\pr*{
		3
		\onetensor^{2, 2}
		+
		x_{1, 1} 
		\convolution
		\begin{pmatrix}
			-2
		\end{pmatrix}
		+
		x_{1, 2} 
		\convolution
		\begin{pmatrix}
			2
		\end{pmatrix}
	}\\
&=
	\multdim_{
		\id_\R
		,
		2 \times 2
	}
	\pr*{
		3
		\onetensor^{2, 2}
		+
		\begin{pmatrix}
			1 & 1\\
			1 & 1
		\end{pmatrix}
		\convolution
		\begin{pmatrix}
			-2
		\end{pmatrix}
		+
		\begin{pmatrix}
			5 & 7\\
			11 & 13
		\end{pmatrix}
		\convolution
		\begin{pmatrix}
			2
		\end{pmatrix}
	}\\
&=
	\multdim_{
		\id_\R
		,
		2 \times 2
	}
	\pr*{
		\begin{pmatrix}
			3 & 3\\
			3 & 3
		\end{pmatrix}
		+
		\begin{pmatrix}
			-2 & -2\\
			-2 & -2
		\end{pmatrix}
		+
		\begin{pmatrix}
			10 & 14\\
			22 & 26
		\end{pmatrix}
	}\\
&=
	\multdim_{
		\id_\R
		,
		2 \times 2
	}
	\pr*{
		\begin{pmatrix}
			11 & 15\\
			23 & 27
		\end{pmatrix}
	}
=
	\begin{pmatrix}
		11 & 15\\
		23 & 27
	\end{pmatrix}.
\end{split}
\end{equation}
This and \eqref{CANN_realization_example:eq1} establish \eqref{CANN_realization_example:concl1}.
\end{aproof}

\filelisting{code:convannex}{code/conv-ann-ex.py}{
		{\sc Python} code implementing the feedforward \cnn\
		$\Phi$ from
		\cref{CANN_realization_example} (see \cref{CANN_realization_example:ass1})
		in {\sc PyTorch}
		and verifying \cref{CANN_realization_example:concl1}.
	}

\cfclear
\begin{exercise}{ex:CNN_realization}
Let
\begin{multline}
\Phi
=
	\bigl(
		((W_{1,n,m})_{(n,m) \in \{1, 2, 3\} \times \{1\}}, (B_{1,n})_{n \in \{1, 2, 3\}}),
		\\
		((W_{2,n,m})_{(n,m) \in \{1\} \times \{1, 2, 3\}}, (B_{2,n})_{n \in \{1\}})
	\bigr)
\in
	((\R^{2})^{3 \times 1} \times \R^{3})
	\times
	((\R^{3})^{1 \times 3} \times \R^{1})
\end{multline}
satisfy
\begin{equation}
\begin{split} 
	W_{1,1,1}
=
	(1,-1),
\;\;
	W_{1,2,1}
=
	(2, -2),
\;\;
	W_{1,3,1}
=
	(-3, 3),
\;\;
	(B_{1,n})_{n \in \{1, 2, 3\}}
=
	(1, 2, 3),
\end{split}
\end{equation}
\begin{equation}
\begin{split} 
	W_{2,1,1}
=
	(1,-1, 1),
\;\;
	W_{2,1,2}
=
	(2, -2, 2),
\;\;
	W_{2,1,3}
=
	(-3, 3, -3),
\;\;
\text{and}
\;\;
	B_{2,1}
=
	-2
\end{split}
\end{equation}
and let $v \in \R^{9}$ satisfy $v = (1, 2, 3, 4, 5, 4, 3, 2, 1)$.
Specify
\begin{equation}
\begin{split} 
	(\CNNRealisation{\rect}(\Phi))(v)
\end{split}
\end{equation}
explicitly and prove that your result is correct
\cfload!
\end{exercise}

\cfclear
\begin{exercise}{CNN_calc}
Let
\begin{multline}
	\Phi
=
	\bigl(
		((W_{1,n,m})_{(n,m) \in \{1, 2, 3\} \times \{1\}}, (B_{1,n})_{n \in \{1, 2, 3\}}),
		\\((W_{2,n,m})_{(n,m) \in \{1\} \times \{1, 2, 3\}}, (B_{2,n})_{n \in \{1\}})
	\bigr)
\in
	((\R^{3})^{3 \times 1} \times \R^{3})
	\times
	((\R^{2})^{1 \times 3} \times \R^{1})
\end{multline}
satisfy
\begin{equation}
\begin{split} 
	W_{1,1,1}
=
	(1,1,1),
\qquad
	W_{1,2,1}
=
	(2, -2, -2),
\end{split}
\end{equation}	
\begin{equation}
\begin{split} 
	W_{1,3,1}
=
	(-3, -3, 3),
\qquad
	(B_{1,n})_{n \in \{1, 2, 3\}}
=
	(3, -2, -1),
\end{split}
\end{equation}
\begin{equation}
\begin{split} 
	W_{2,1,1}
=
	(2, -1),
\quad
	W_{2,1,2}
=
	(-1, 2),
\quad
	W_{2,1,3}
=
	(-1, 0),
\quad
\text{and}
\quad
	B_{2,1}
=
	-2
\end{split}
\end{equation}
and let $v \in \R^{9}$ satisfy $v = (1,-1,1,-1,1,-1,1,-1,1)$.
Specify
\begin{equation}
\begin{split} 
	(\CNNRealisation{\rect}(\Phi))(v)
\end{split}
\end{equation}
explicitly and prove that your result is correct \cfout!
\end{exercise}

\cfclear
\begin{exercise}{ex:express_ANN_as_CNN}
Prove or disprove the following statement:
For every 
	$a \in C(\R, \R)$,
	$\Phi \in \ANNs$ 
there exists 
	$\Psi \in \CNNs$ 
such that for all $x \in \R^{\inDimANN(\Phi)}$ it holds that
	$\R^{\inDimANN(\Phi)} \subseteq \operatorname{Domain}(\CNNRealisation{a}(\Psi))$
and
\begin{equation}
\begin{split} 
	(\CNNRealisation{a}(\Psi))(x) 
=
	(\functionANN{a}(\Phi))(x)
\end{split}
\end{equation}
\cfload.
\end{exercise}

\begin{adef}{def:scalarproduct}[Standard scalar products]
	We denote by $\scp{\cdot,\cdot}\colon\br[\big]{\bigcup_{d\in\N}(\R^d\times\R^d)}\to\R$
	the function which satisfies for all 
	  $d\in\N$,
		$x=(x_1,\dots,x_d),\,y=(y_1,\dots,y_d)\in\R^d$
	that
	\begin{equation}
		\scp{x,y}=\smallsum_{i=1}^d x_iy_i
		.
	\end{equation}
\end{adef}

\cfclear
\begin{exercise}{ex:express_CNN_via_ANN}
For every $d \in \N$ let
	$\mathbf{e}^{(d)}_{1}, \mathbf{e}^{(d)}_{2}, \ldots, \mathbf{e}^{(d)}_{ d} \in \R^d$
satisfy
	$\mathbf{e}^{(d)}_{1} = (1, 0, \ldots, 0)$,
	$\mathbf{e}^{(d)}_{2} = (0, 1, 0, \ldots, 0)$,
	$\dots$,
	$\mathbf{e}^{(d)}_{d} = (0, \ldots, 0, 1)$.
Prove or disprove the following statement:
For all 
	$a \in  C(\R, \R)$,
	$\Phi \in \ANNs$,
	$D \in \N$,
	$x = ((x_{i,j})_{j \in \{1, 2, \ldots, D \}})_{i \in \{1, 2, \ldots, \inDimANN(\Phi)\}} \in (\R^D)^{\inDimANN(\Phi)}$
it holds that
\begin{equation}
\begin{split} 
	(\CNNRealisation{a}(\Phi))(x)
=
	\bpr{
		\bpr{
			\scp{
				\mathbf{e}^{(\outDimANN(\Phi))}_{k},
				(\functionANN{a}(\Phi))((x_{i,j})_{i \in \{1, 2, \ldots, \inDimANN(\Phi)\}})
			}
		}_{j \in \{1, 2, \ldots, D \}}
	}_{k \in \{1, 2, \ldots, \outDimANN(\Phi) \}}
\end{split}
\end{equation}
\cfload.
\end{exercise}

\section{Residual ANNs (ResNets)} 
\label{section:resnets}

\todosecond{Write two definition of Residual ANNs and skip connection ANNs}

In this section we review \resnets.
Roughly speaking, plain-vanilla feedforward \anns\ can be seen as having a computational structure consisting of sequentially chained layers in which each layer feeds information forward to the next layer (cf., \eg, \cref{def:FFNN,def:ANNrealization} above).
\resnets, in turn, are \anns\ involving so-called \emph{skip connections} in their computational structure, which allow information from one layer to be fed not only to the next layer, but also to other layers further down the computational structure.
In principle, such skip connections can be employed in combinations with other \ann\ architecture elements, such as
	fully-connected layers (cf., \eg, \cref{subsec:structured_description,subsec:vectorized_description} above),
	convolutional layers (cf., \eg, \cref{section:cnns} above),
	and
	recurrent structures (cf., \eg, \cref{section:rnns} below).
However, for simplicity we introduce in this section in all mathematical details
feedforward fully-connected \resnets\ 
	in which the skip connection is a learnable linear map
(see \cref{def:ResNetrealization,def:ResNet} below).

\resnets\ were introduced in He et al.~\cite{He2016} as an attempt to improve the performance of deep \anns\ which typically are much harder to train than shallow \anns\
(cf., \eg, \cite{Pascanu2013,Bengio1994,Glorot2010}).
The \resnets\ in He et al.~\cite{He2016} only involve skip connections that are identity mappings without trainable parameters, and are thus a special case of the definition of \resnets\ provided in this section (see \cref{def:ResNet,def:ResNetrealization} below).
The idea of skip connection (sometimes also called \emph{shortcut connections}) has already been introduced before \resnets\ and has been used in earlier \ann\ architecture such as the \emph{highway nets} in Srivastava et al.~\cite{Srivastava2015,Srivastava2015a} (cf.\ also \cite{Szegedy2015,Lee2015,Raiko2012,Vatanen2013,Mao2016}).
In addition, we refer to 
\cite{He2016a,Xie2017,Huang2017,Zagoruyko2016,Wang2017a}
for a few successful \ann\ architecures building on the \resnets\ in He et al.~\cite{He2016}.

\subsection{Structured description of fully-connected ResNets}

\newcommand{\Skipconnections}{S}

\begin{adef}{def:ResNet}[Structured description of fully-connected \resnets]
We denote by $\ResNets$ the set given by 
\begin{multline}
\label{eq:ResNet}
\textstyle
	\ResNets
=
	\bigcup_{L \in \N}
	\bigcup_{l_0,l_1,\ldots, l_L \in \N}
	\bigcup_{\Skipconnections \subseteq \{ (r,k) \in (\N_0)^2 \colon r < k \leq L \}}
		\pr[\Big]{
			\pr[\big]{
				\bigtimes_{k = 1}^L (\R^{l_k \times l_{k-1}} \times \R^{l_k})
			}\\
			\times
			\pr[\big]{
				\bigtimes_{(r, k) \in \Skipconnections} \R^{l_k \times l_{r}}
			}
		}.
\end{multline}
\end{adef}

\cfclear
\begin{adef}{def:resnets}[Fully-connected \resnets]
  We say that $\Phi$ is a fully-connected \resnet\ if and only
  if it holds that 
  \begin{equation}
    \Phi\in\ResNets
  \end{equation}
  \cfload.
\end{adef}

\cfclear
\begin{lemma}[On an empty set of skip connections]
\label{Zero_product}
Let
	$L \in \N$,
	$l_0,l_1,\ldots, l_L \in \N$,
	$\Skipconnections \subseteq \{ (r,k) \in (\N_0)^2 \colon r < k \leq L \}$.
Then
\begin{equation}
\label{Zero_product:concl1}
\begin{split} 
\textstyle
	\#
	\pr[\big]{
		\bigtimes_{(r, k) \in \Skipconnections} \R^{l_k \times l_{r}}
	}
=
	\begin{cases}
		1 &\colon \Skipconnections = \emptyset \\
		\infty &\colon \Skipconnections \neq \emptyset.
	\end{cases}
\end{split}
\end{equation}
\end{lemma}

\begingroup
\newcommand{\setoffunctions}[1]{F\pr*{#1}}
\begin{proof}[Proof of \cref{Zero_product}]
Throughout this proof, for all sets $A$ and $B$ let $\setoffunctions{A, B}$ be the set of all functions from $A$ to $B$.
Note that
\begin{equation}
\label{Zero_product:eq1}
\begin{split} 
\textstyle
	\#
	\pr[\big]{
		\bigtimes_{(r, k) \in \Skipconnections} \R^{l_k \times l_{r}}
	}
&=
	\#\cu[\big]{
		\textstyle
		f \in \setoffunctions{\Skipconnections, {\scriptstyle\bigcup}_{(r, k) \in \Skipconnections} \R^{l_k \times l_{r}}}
	\colon 
		(\forall \, (r, k) \in \Skipconnections \colon f(r, k) \in  \R^{l_k \times l_{r}})
	}.
\end{split}
\end{equation}
\enum{
	This;
	the fact that for all sets $B$ it holds that $\#(\setoffunctions{\emptyset, B}) = 1$
}[ensure]
that
\begin{equation}
\label{Zero_product:eq2}
\begin{split} 
\textstyle
	\#
	\pr[\big]{
		\bigtimes_{(r, k) \in \emptyset} \R^{l_k \times l_{r}}
	}
=
	\#(\setoffunctions{\emptyset, \emptyset})
=
	1.
\end{split}
\end{equation}
\Moreover
\enum{
	\eqref{Zero_product:eq1}
}[assure]
that for all
	$(R, K) \in \Skipconnections$
it holds that
\begin{equation}
\begin{split} 
\textstyle
	\#
	\pr[\big]{
		\bigtimes_{(r, k) \in \Skipconnections} \R^{l_k \times l_{r}}
	}
&\geq
	\#\pr[\big]{
		\setoffunctions{\{(R, K)\},  \R^{l_K \times l_{R}}}
	}
=
	\infty.
\end{split}
\end{equation}
Combining this and \eqref{Zero_product:eq2} establishes \eqref{Zero_product:concl1}.
The proof of \cref{Zero_product} is thus complete.
\end{proof}
\endgroup

\subsection{Realizations of fully-connected ResNets}

\cfclear
\begin{adef}{def:ResNetrealization}[Realizations associated to fully-connected \resnets]
Let 
	$L\in\N$, 
 	$l_0,l_1,\allowbreak\ldots, l_L \in \N$, 
 	$\Skipconnections \subseteq \{ (r,k) \in (\N_0)^2 \colon r < k \leq L \}$,
	$
		\Phi 
	=
		((W_k, B_k)_{k \in \{1, 2, \ldots, L \}}, (V_{r, k})_{(r, k) \in \Skipconnections})
		\in  \allowbreak
		\bigl(
			\bigl( \bigtimes_{k = 1}^L\allowbreak(\R^{l_k \times l_{k-1}} \times \R^{l_k})\bigr)
			\times
			\bigl(
				\bigtimes_{(r, k) \in \Skipconnections} \R^{l_k \times l_{r}}
			\bigr)
		\bigr)
	\subseteq
		\ResNets
	$
and let 
	$a \colon \R \to \R$
be a function.
Then we denote by 
\begin{equation}
\begin{split} \textstyle
	\ResNetRealisation{a}(\Phi) \colon \R^{l_0} \to \R^{l_L}
\end{split}
\end{equation}
the function which satisfies
for all  
	$x_0 \in \R^{l_0}, x_1 \in \R^{l_1}, \ldots, x_L \in \R^{l_{L}}$ 
with 
\begin{multline}
	\forall \, k \in \{1,2,\dots,L\} \colon \\[-2pt]
\textstyle
	x_k 
	=
	\multdim_{a\indicator{(0,L)}(k)+\id_\R\indicator{\{L\}}(k),l_k}
	\pr[Big]{
		W_k x_{k-1} + B_k + \sum_{r \in \N_0, (r, k) \in \Skipconnections} V_{r, k} x_r
	}
\end{multline}
that
\begin{equation}
( \ResNetRealisation{a}(\Phi) ) (x_0) = x_L
\end{equation}
and 
we call $\ResNetRealisation{a}(\Phi)$ the realization function of the fully-connected \resnet\ $\Phi$ with activation function $a$
(we call $\ResNetRealisation{a}(\Phi)$ the realization of the fully-connected \resnet\ $\Phi$ with activation $a$)
(cf.\ \cref{def:ResNet,def:multidim_version}).
\end{adef}

\begin{adef}{def:identityMatrix}[Identity matrices]
Let $ d \in \N $. 
Then we denote by $ \idMatrix_{d} \in \R^{ d \times d } $ 
the identity matrix in $ \R^{ d \times d } $.
\end{adef}

\filelisting{code:resann}{code/res-ann.py}{
		{\sc Python} code implementing a fully-connected \resnet\
		in {\sc PyTorch}. The implemented model here corresponds to 
		a fully-connected \resnet\
		$(\Phi,V)$
		where
		$l_0=3$,
		$l_1=10$,
		$l_2=20$,
		$l_3=10$,
		$l_4=1$,
		$\Phi=((W_1,B_1),(W_2,B_2),(W_3,B_3),(W_4,B_4))\in\bpr{\bigtimes_{k=1}^4(\R^{l_k\times l_{k-1}}\times \R^{l_k})}$,
		$\Skipconnections=\{(1,3)\}$,
		$V=(V_{r,k})_{(r,k)\in S}\in \bpr{\bigtimes_{(r,k)\in S}\R^{l_k\times l_r}}$,
		and $V_{1,3}=\idMatrix_{10}$
		(cf.\ \cref{def:identityMatrix}).
	}

\cfclear
\begin{athm}{example}{ResNets_example}[Example for \cref{def:resnets}]
Let 
	$l_0 = 1$, 
	$l_1 = 1$, 
	$l_2 = 2$, 
	$l_3 = 2$,
	$l_4 = 1$,
	$\Skipconnections = \{ (0, 4) \}$,
let 
\begin{equation}
\begin{split} 
	&\Phi
=
	(
		(W_1, B_1), (W_2, B_2), (W_3, B_3), (W_4, B_4)
	)
\in \textstyle
	\bigl( \bigtimes_{k = 1}^4\allowbreak(\R^{l_k \times l_{k-1}} \times \R^{l_k})\bigr)
\end{split}
\end{equation}
satisfy
\begin{equation}
\begin{split} 
	W_1
=
	\begin{pmatrix}
		1
	\end{pmatrix},
\qquad
	B_1
=
	\begin{pmatrix}
		0
	\end{pmatrix},
\qquad
	W_{2}
=
	\begin{pmatrix}
		1\\2
	\end{pmatrix},
\qquad
	B_2
=
	\begin{pmatrix}
		0\\1
	\end{pmatrix},
\end{split}
\end{equation}
\begin{equation}
\begin{split} 
	W_3
=
	\begin{pmatrix}
		1 & 0 \\ 
		0 & 1
	\end{pmatrix},
\qquad
	B_3
=
	\begin{pmatrix}
		0 \\ 0
	\end{pmatrix},
\qquad
	W_4
=
	\begin{pmatrix}
		2 & 2
	\end{pmatrix},
\qandq
	B_4
=
	\begin{pmatrix}
		1
	\end{pmatrix},
\end{split}
\end{equation}
and
let
	$V = (V_{r, k})_{(r, k) \in \Skipconnections} \in \bigtimes_{(r, k) \in \Skipconnections} \R^{l_k \times l_r}$
satisfy
\begin{equation}
\begin{split} 
	V_{0, 4}
=
	\begin{pmatrix}
		-1
	\end{pmatrix}.
\end{split}
\end{equation}
Then 
\begin{equation}
\label{ResNets_example:concl1}
\begin{split} 
	(\ResNetRealisation{\rect}(\Phi, V))(5)
=
	28
\end{split}
\end{equation}
\cfout.
\end{athm}

\begin{aproof}
Throughout this proof, let
	$x_0 \in \R^{1}$,
	$x_1 \in \R^{1}$,
	$x_2 \in \R^{2}$,
	$x_3 \in \R^{2}$,
	$x_4 \in \R^{1}$
satisfy for all
	$ k \in \{1,2,3,4\}$ 
that $x_0 = 5$ and
\begin{equation}
\label{ResNets_example:setting1}
\begin{split} \textstyle
	x_k 
	=
	\multdim_{\rect \indicator{(0,4)}(k)+\id_\R\indicator{\{4\}}(k),l_k}(
		W_k x_{k-1} + B_k + \sum_{r \in \N_0, (r, k) \in \Skipconnections} V_{r, k} x_r
	).
\end{split}
\end{equation}
\Nobs that
\enum{
	\eqref{ResNets_example:setting1}
}[assure]
that
\begin{equation}
\label{ResNets_example:eq1}
\begin{split} 
	(\ResNetRealisation{\rect}(\Phi, V))(5) 
= 
	x_4.
\end{split}
\end{equation}
\Moreover 
\enum{
	\eqref{ResNets_example:setting1}
}[ensure]
that
\begin{equation}
\begin{split} 
	x_1 
=
	\multdim_{\rect,1}(
		W_1 x_{0} + B_1 
	)
=
	\multdim_{\rect,1}(
		5  
	),
\end{split}
\end{equation}
\begin{equation}
\begin{split} 
	x_2
=
	\multdim_{\rect,2}(
		W_2 x_{1} + B_2 
	)
=
	\multdim_{\rect,1}
	\pr*{
		\begin{pmatrix}
			1\\2
		\end{pmatrix} 
		\begin{pmatrix}
			5
		\end{pmatrix} 
		+
		\begin{pmatrix}
			0\\1
		\end{pmatrix}
	}
=
	\multdim_{\rect,1}
	\pr*{
		\begin{pmatrix}
			5\\11
		\end{pmatrix} 
	}
=
	\begin{pmatrix}
		5\\11
	\end{pmatrix},
\end{split}
\end{equation}
\begin{equation}
\begin{split} 
	x_3
=
	\multdim_{\rect,2}(
		W_3 x_{2} + B_3 
	)
=
	\multdim_{\rect,1}
	\pr*{
		\begin{pmatrix}
			1 & 0 \\
			0 & 1 \\
		\end{pmatrix} 
		\begin{pmatrix}
			5 \\11
		\end{pmatrix} 
		+
		\begin{pmatrix}
			0\\0
		\end{pmatrix}
	}
=
	\multdim_{\rect,1}
	\pr*{
		\begin{pmatrix}
			5 \\ 11
		\end{pmatrix} 
	}
=
	\begin{pmatrix}
		5\\11
	\end{pmatrix},
\end{split}
\end{equation}
\begin{equation}
\begin{split} 
\andq
	x_4
&=
	\multdim_{\rect,1}(
		W_4 x_{3} + B_4 + V_{0, 4} x_0 
	)\\
&=
	\multdim_{\rect,1}
	\pr*{
		\begin{pmatrix}
			2 & 2
		\end{pmatrix} 
		\begin{pmatrix}
			5 \\11
		\end{pmatrix} 
		+
		\begin{pmatrix}
			1
		\end{pmatrix}
		+
		\begin{pmatrix}
			-1
		\end{pmatrix} 
		\begin{pmatrix}
			5
		\end{pmatrix} 
	}
=
	\multdim_{\rect,1}
	\pr*{
		28
	}
=
	28.
\end{split}
\end{equation}
This and \eqref{ResNets_example:eq1} establish \eqref{ResNets_example:concl1}.
\end{aproof}

\cfclear
\begin{exercise}{ResNet_calc}
Let 
	$l_0 = 1$, 
	$l_1 = 2$, 
	$l_2 = 3$, 
	$l_3 = 1$,
	$\Skipconnections = \{ (0, 3), (1, 3)\}$,
let 
\begin{equation}
\begin{split} 
	&\Phi
=
	(
		(W_1, B_1), (W_2, B_2), (W_3, B_3)
	)
\in \textstyle
	\bigl( \bigtimes_{k = 1}^3\allowbreak(\R^{l_k \times l_{k-1}} \times \R^{l_k})\bigr)
\end{split}
\end{equation}
satisfy
\begin{equation}
\begin{split} 
	W_1
=
	\begin{pmatrix}
		1\\2
	\end{pmatrix},
\qquad
	B_1
=
	\begin{pmatrix}
		3\\4
	\end{pmatrix},
\qquad
	W_{2}
=
	\begin{pmatrix}
		-1 & 2 \\ 3 & -4 \\ -5 & 6 
	\end{pmatrix},
\qquad
	B_2
=
	\begin{pmatrix}
		0 \\ 0 \\ 0
	\end{pmatrix},
\end{split}
\end{equation}
\begin{equation}
\begin{split} 
	W_3
=
	\begin{pmatrix}
		-1 & 1 & -1
	\end{pmatrix},
\qandq
	B_3
=
	\begin{pmatrix}
		-4
	\end{pmatrix},
\end{split}
\end{equation}
and
let
	$V = (V_{r, k})_{(r, k) \in \Skipconnections} \in \bigtimes_{(r, k) \in \Skipconnections} \R^{l_k \times l_r}$
satisfy
\begin{equation}
\begin{split} 
	V_{0, 3}
=
	\begin{pmatrix}
		1
	\end{pmatrix}
\qandq
	V_{1, 3}
=
	\begin{pmatrix}
		3 & -2
	\end{pmatrix}.
\end{split}
\end{equation}
Prove or disprove the following statement:
It holds that
\begin{equation}
\begin{split} 
	(\ResNetRealisation{\rect}(\Phi, V))(-1)
=
	0
\end{split}
\end{equation}
\cfout.
\end{exercise}

\section{Recurrent ANNs (RNNs)}
\label{section:rnns}

In this section we review \RNNs, a type of \anns\ designed to take sequences of data points as inputs.
Roughly speaking, unlike in feedforward \anns\ where an input is processed by a successive application of series of \emph{different} parametric functions 
(cf.\ \cref{def:FFNN,def:ANNrealization,def:CNNrealisation,def:ResNetrealization} above), 
in \RNNs\ an input sequence is processed by a repeated application of the \emph{same} parametric function whereby after the first application, each subsequent application of the parametric function takes as input a new element of the input sequence and a partial output from the previous application of the parametric function.
The output of an \RNN\ is then given by a sequence of partial outputs coming from the repeated applications of the parametric function (see \cref{def:RNNs} below for a precise description of \RNNs\
and 
cf., \eg, 
	\cite[Section 12.7]{alpaydin2020introduction},
	\cite[Chapter 17]{Calin2020}
	\cite[Chapter 5]{Caterini2018},
	and
	\cite[Chapter 10]{Goodfellow2016}
for other introductions to \RNNs).

The repeatedly applied parametric function in an \RNN\ is typically called an \emph{\RNN\ node} and
any \RNN\ architecture is determined by specifying the architecture of the corresponding \RNN\ node. 
We review a simple variant of such \RNN\ nodes and the corresponding \RNNs\ in \cref{sec:simple_rnn} in detail
and we briefly address one of the most commonly used \RNN\ nodes, the so-called \LSTM\ node, in \cref{sec:LSTM}.

There is a wide range of application areas where sequential data are considered and \RNN\ based deep learning methods are being employed and developed.
Examples of such applications areas are
\NLP\ including 
	language translation (cf., \eg, \cite{Cho2014,Cho2014a,Bahdanau2014,Sutskever2014} and the references therein),  
	language generation (cf., \eg, \cite{Graves2013a,Karpathy2015,Radford2017,Bowman2016} and the references therein), 
	and
	speech recognition (cf., \eg, \cite{Graves2013,Graves2014,Chorowski2015,Amodei2016,Sak2014} and the references therein),
time series prediction analysis including
	stock market prediction  (cf., \eg, \cite{Fischer2018,Sezer2020,SiamiNamini2018,Fabbri2018} and the references therein) and
	weather prediction (cf., \eg, \cite{Xingjian2015,Wang2021,Reichstein2019} and the references therein)
and
video analysis (cf., \eg, \cite{Donahue2017,Ng2015,Karevan2020,Venugopalan2015} and the references therein).

\newcommand{\initinfo}{\mathbb{I}}

\subsection{Description of RNNs}
\label{sec:descr_rnns}

\begin{adef}{def:unrolling}[Function unrolling]
Let $X, Y, I$ be sets,
let $f \colon X \times I \to Y \times I$ be a function,
and
let  
	$T \in \N$,
	$\initinfo \in I$.
Then we denote by 
$\unrolling{f}{T}{\initinfo} \colon X^T \to Y^T$ 
the function which satisfies for all
	$x_1, x_2, \ldots, x_T \in X$,
	$y_1, y_2, \ldots, y_T \in Y$,
	$i_0, i_1, \ldots, i_T \in I$
with 
$i_0 = \initinfo$ and
$\forall\, t \in \{1, 2, \ldots, T\} \colon (y_t, i_t) \allowbreak = f(x_t, i_{t-1})$ 
that
\begin{equation}
\label{unrolling:eq1}
\begin{split} 
	\unrolling{f}{T}{\initinfo}(x_1, x_2, \ldots, x_T) = (y_1, y_2, \ldots, y_T)
\end{split}
\end{equation}
and 
we call
$\unrolling{f}{T}{\mathfrak{i}}$ 
the $T$-times unrolled function $f$ with initial information $\initinfo$.
\end{adef}

\cfclear
\begin{adef}{def:RNNs}[Description of \RNNs]
	Let $X, Y, I$ be sets,
	let 
		$\defaultParamDim,T \in \N$,
		$\theta \in \R^\defaultParamDim$,
		$\initinfo \in I$, 
	and
	let $\mathfrak{N} = (\mathfrak{N}_{\vartheta})_{\vartheta \in  \R^\defaultParamDim} \colon \R^\defaultParamDim \times X \times I \to Y \times I$ be a function.
	Then we call
		$R$
	the realization function of the $T$-step unrolled \RNN\ 
	with 
		\RNN\ node $\mathfrak{N}$, 
		parameter vector $\theta$,
		and 
		initial information $\initinfo$
		(we call
				$R$
			the realization of the $T$-step unrolled \RNN\ 
			with 
				\RNN\ node $\mathfrak{N}$, 
				parameter vector $\theta$,
				and 
				initial information $\initinfo$)
	if and only if it holds that
	\begin{equation}
	\label{def:RNNs:eq1}
	\begin{split} 
		R 
	= 
		\unrolling{\mathfrak{N}_\theta}{T}{\initinfo}
	\end{split}
	\end{equation}
	\cfload.
\end{adef}

\subsection{Vectorized description of simple fully-connected RNNs}
\label{sec:simple_rnn}

\cfclear
\begin{adef}{def:RNNNode}[Vectorized description of simple fully-connected \RNN\ nodes]
Let 
	$\fx, \fy, \mathfrak{i} \in \N$, 
	$\theta \in \R^{(\fx + \mathfrak{i} + 1)\mathfrak{i} + (\mathfrak{i}+1)\fy}$
and 
let
	$\Psi_1 \colon \R^\mathfrak{i} \to \R^\mathfrak{i}$ and
	$\Psi_2 \colon \R^\fy \to \R^\fy$
be functions.
Then we call
	$r$
the realization function of the simple fully-connected \RNN\ node with 
	parameter vector $\theta$ and 
	activation functions $\Psi_1$ and $\Psi_2$
	(we call
		$r$
	the realization of the simple fully-connected \RNN\ node with 
		parameter vector $\theta$ and 
		activations $\Psi_1$ and $\Psi_2$)
if and only if it holds that
	$r \colon \R^\fx \times \R^\mathfrak{i} \to \R^\fy \times \R^\mathfrak{i}$
is the function from $ \R^\fx \times \R^\mathfrak{i}$ to $\R^\fy \times \R^\mathfrak{i}$ which satisfies
for all
	$x \in \R^\fx$,
	$i \in \R^\mathfrak{i}$
that
\begin{equation}
\begin{split} 
	r(x, i)
=
	\Bigl(
		\bigl(
			\RealV{\theta}{0}{\fx + \mathfrak{i}}{ \Psi_1, \Psi_2}
		\bigr)
		(x, i),
		\bigl(
			\RealV{\theta}{0}{\fx + \mathfrak{i}}{ \Psi_1}
		\bigr)
		(x, i)
	\Bigr)
\end{split}
\end{equation}
\cfload.

\end{adef}

\cfclear
\begin{adef}{def:VanillaRNN}[Vectorized description of simple fully-connected \RNNs]
Let 
	$\fx, \fy, \mathfrak{i}, T \in \N$, 
	$\theta \in \R^{(\fx + \mathfrak{i} + 1)\mathfrak{i} + (\mathfrak{i}+1)\fy}$,
	$\initinfo \in \R^\mathfrak{i}$ and
let
	$\Psi_1 \colon \R^\mathfrak{i} \to \R^\mathfrak{i}$ and
	$\Psi_2 \colon \R^\fy \to \R^\fy$
be functions.
Then we call 
$
	R	
$
the realization function of the $T$-step unrolled simple fully-connected \RNN\ with 
	parameter vector $\theta$, 
	activation functions $\Psi_1$ and $\Psi_2$, and
	initial information $\initinfo$
		(we call 
		$
			R	
		$
		the realization of the $T$-step unrolled simple fully-connected \RNN\ with 
			parameter vector $\theta$, 
			activations $\Psi_1$ and $\Psi_2$, and
			initial information $\initinfo$)
if and only if there exists  $r \colon \R^\fx \times \R^\mathfrak{i} \to \R^\fy \times \R^\mathfrak{i}$ such that
\begin{enumerate}[label=(\roman{*})]
\item 
it holds that $r$ is the realization of the simple fully-connected \RNN\ node with parameter vector $\theta$ and activations $\Psi_1$ and $\Psi_2$ and

\item
it holds that
\begin{equation}
\label{def:VanillaRNN:eq1}
\begin{split} 
	R
=
	\unrolling{r}{T}{\initinfo} 
\end{split}
\end{equation}
\end{enumerate}
(cf.\ \cref{def:RNNNode,def:unrolling}).
\end{adef}

\cfclear
\begin{athm}{lemma}{rnn_equivalent_def}
Let 
	$\fx, \fy, \mathfrak{i}, \defaultParamDim, T \in \N$, 
	$\theta \in \R^{\defaultParamDim}$,
	$\initinfo \in \R^\mathfrak{i}$
satisfy
	$\defaultParamDim = (\fx + \mathfrak{i} + 1)\mathfrak{i} + (\mathfrak{i}+1)\fy$,
let
	$\Psi_1 \colon \R^\mathfrak{i} \to \R^\mathfrak{i}$ and
	$\Psi_2 \colon \R^\fy \to \R^\fy$
be functions,
and let
	$
		\mathfrak{N} 
	= 
		(\mathfrak{N}_{\vartheta})_{\vartheta \in  \R^\defaultParamDim} 
	\colon 
		\R^\defaultParamDim \times \R^\fx \times \R^{\mathfrak{i}}
	\to 
		\R^{\fy} \times  \R^{\mathfrak{i}}
	$
satisfy for all
	$\vartheta \in  \R^\defaultParamDim$
that
	$\mathfrak{N}_\vartheta$ is\cfadd{def:RNNNode}
the realization of the simple fully-connected \RNN\ node with 
	parameter vector $\vartheta$ and 
	activations $\Psi_1$ and $\Psi_2$
\cfload.
Then the following three statements are equivalent:
\begin{enumerate}[label=(\roman *)]
\item 
\label{rnn_equivalent_def:item1}
It holds that 
	$R$
is\cfadd{def:VanillaRNN} the realization of the $T$-step unrolled simple fully-connected \RNN\ with 
	parameter vector $\theta$, 
	activations $\Psi_1$ and $\Psi_2$, and
	initial information $\initinfo$
\cfout.
\item 
\label{rnn_equivalent_def:item2}
It holds that
	$R$
is\cfadd{def:RNNs}
the realization of the $T$-step unrolled \RNN\ 
	with 
		\RNN\ node $\mathfrak{N}$, 
		parameter vector $\theta$,
		and 
		initial information $\initinfo$
\cfout.

\cfclear
\item
\label{rnn_equivalent_def:item3}
It holds that
\begin{equation}
\label{rnn_equivalent_def:eq2}
\begin{split} 
	R
=
	\unrolling{\mathfrak{N}_\theta}{T}{\initinfo}
\end{split}
\end{equation}
\cfload.
\end{enumerate}
\end{athm}

\begin{aproof}
\Nobs that
	\cref{def:RNNs:eq1,def:VanillaRNN:eq1,rnn_equivalent_def:eq2}
\prove that
	(\ref{rnn_equivalent_def:item1} $\leftrightarrow$ \ref{rnn_equivalent_def:item2} $\leftrightarrow$ \ref{rnn_equivalent_def:item3}).
\end{aproof}

\cfclear
\begin{exercise}{ex:VanillaRNN_example}
\cfadd{def:VanillaRNN}
For every 
	$T \in \N$,
	$\alpha \in (0,1)$ 
let
$
	R_{T, \alpha}	
$
be the realization of the $T$-step unrolled simple fully-connected \RNN\ with 
	parameter vector $(1, 0, 0, \alpha, 0, 1-\alpha, 0, 0, -1, 1, 0)$, 
	activations $\multdim_{ \rect, 2 }$ and $\id_{\R}$, and
	initial information $(0, 0)$
\cfload.
For every 
	$T \in \N$,
	$\alpha \in (0,1)$
specify
	$R_{T, \alpha}(1, 1, \ldots, 1)$ 
explicitly and prove that your result is correct
\cfload!
\end{exercise}

\subsection{Long short-term memory (LSTM) RNNs}
\label{sec:LSTM}

In this section we briefly discuss a very popular type of \RNN\ nodes called \emph{\LSTM\ nodes} 
and the corresponding \RNNs\ called \emph{\LSTM\ networks}
which were introduced in Hochreiter \& Schmidhuber~\cite{Hochreiter1997}.
Loosely speaking, \LSTM\ nodes were invented to attempt to the tackle the issue that most \RNNs\ based on simple \RNN\ nodes, such as the simple fully-connected \RNN\ nodes in \cref{sec:simple_rnn} above, struggle to learn to understand long-term dependencies in sequences of data (cf., \eg, \cite{Pascanu2013,Bengio1994}).
Roughly speaking, an \RNN\ processes an input sequence by repeatedly applying an \RNN\ node to a tuple consisting of a new element of the input sequence and a partial output of the previous application of the \RNN\ node
(see \cref{def:RNNs} above for a precise description of \RNNs).
Therefore, the only information on previously processed elements of the input sequence that any application of an \RNN\ node has access to, is the information encoded in the output produced by the last application of the \RNN\ node.
For this reason, \RNNs\ can be seen as only having a \emph{short-term memory}. 
The \LSTM\ architecture, however is designed
with the aim to facilitate the transmission of long-term information within this short-term memory.  
\LSTM\ networks can thus be seen as having a sort of \emph{long short-term memory}.

For a precise definition of \LSTM\ networks we refer to the original article 
Hochreiter \& Schmidhuber~\cite{Hochreiter1997} and, \eg, to the excellent explanations in \cite{Colah,Graves2013a,Fischer2018}.
For a few selected references on \LSTM\ networks in the literature we refer, \eg, to \cite{Fischer2018,Graves2009,Graves2013,Gers2000,Graves2005,Sak2014,Graves2013a,Gers2000,PerezOrtiz2003,Gers2003,Greff2015,Sutskever2014,Luong2015,Bahdanau2014,Cho2014,Schmidhuber2015,Yu2019} and the references therein.

\section{Further types of ANNs}

\todosecond{cite \cite{Hu2018} somewhere. Also check out the attention papers in there.}

In this section we present a selection of references and some rough comments on a couple of further popular types of \anns\ in the literature which were not discussed in the previous sections of this chapter above.

\subsection{ANNs with encoder-decoder architectures: autoencoders}

In this section we discuss the idea of autoencoders which are based on encoder-decoder \ann\ architectures.
Roughly speaking, the goal of autoencoders is to learn a simplified representation of data points and a way to closely reconstruct the original data points from the simplified representation.
The simplified representation of data points is usually called the \emph{encoding} and is obtained by applying an \emph{encoder \ann} to the data points.
The approximate reconstruction of the original data points from the encoded representations is, in turn, called the \emph{decoding} and is obtained by applying a \emph{decoder \ann} to the encoded representations.
The composition of the encoder \ann\ with the decoder \ann\ is called the \emph{autoencoder}.
In the simplest situations the encoder \ann\ and decoder \ann\ are trained to perform their respective desired functions by training the full autoencoder to be as close to the identity mapping on the data points as possible.

A large number of different architectures and training procedures for autoencoders have been proposed in the literature.
In the following we list a selection of a few popular ideas from the scientific literature.
\begin{itemize}
\item 
We refer, \eg, to \cite{Hinton2006,Rumelhart1986,Kramer1991,Hinton1993,Bourlard1988} for foundational references introducing and refining the idea of autoencoders,
\item 
we refer, \eg, to \cite{Vincent2008,Vincent2010,Xie2012} for so-called \emph{denoising autoencoders} which add random pertubation to the input data in the training of autoencoders,
\item 
we refer, \eg, to \cite{Kingma2013,Doersch2016,Bowman2016} for so-called \emph{variational autoencoders} which use techniques from bayesian statistics in the training of autoencoders,
\item 
we refer, \eg, \cite{Masci2011,Ranzato2007} for autoencoders involving convolutions, and
\item 
we refer, \eg, \cite{Makhzani2015,Dumoulin2016} for \emph{adversarial autoencoders} which combine the principles of autoencoders with the paradigm of generative adversarial networks (see Goodfellow et al.~\cite{Goodfellow2014}).
\end{itemize}

\subsection{Transformers and the attention mechanism}
\label{sec:attention}

In \cref{section:rnns} we reviewed \RNNs\ which are a type of \anns\ designed to take sequences of data points as inputs.
Very roughly speaking, \RNNs\ process a sequence of data points by sequentially processing one data point of the sequence after the other and thereby constantly updating an information state encoding previously processed information (see \cref{sec:descr_rnns} above for a precise description of \RNNs).
When processing a data point of the sequence, any information coming from earlier data points is thus only available to the \RNN\ through the information state passed on from the previous processing step of the \RNN. 
Consequently, it can be hard for \RNNs\ to learn to understand long-term dependencies in the input sequence.
In \cref{sec:LSTM} above, we briefly discussed the \LSTM\ architecture for \RNNs\ which is an architecture for \RNNs\ aimed at giving such \RNNs\ the capacity to indeed learn to understand such long-term dependencies.

Another approach in the literature to design \ann\ architectures which process sequential data and are capable to efficiently learn to understand long-term dependencies in data sequences is called the \emph{attention mechanism}.
Very roughly speaking, in the context of sequences of the data, the attention mechanism aims to give \anns\ the capacity
to "\emph{pay attention}" to selected parts of the entire input sequence when they are processing a data point of the sequence.
The idea for using attention mechanisms in \anns\ was first introduced in Bahdanau et al.~\cite{Bahdanau2014} in the context of \RNNs\ trained for machine translation.
In this context the proposed \ann\ architecture still processes the input sequence sequentially, however past information is not only available through the information state from the previous processing step, but also through the attention mechanism, which can directly extract information from data points far away from the data point being processed.

Likely the most famous \anns\ based on the attention mechanism do however not involve any recurrent elements and have been named \emph{Transfomer \anns} by the authors of the seminal paper Vaswani et al.~\cite{Vaswani2017} called "Attention is all you need".
Roughly speaking, Transfomer \anns\ are designed to process sequences of data by considering the entire input sequence at once  and relying only on the attention mechanism to understand dependencies between the data points in the sequence.
Transfomer \anns\ are the basis for many recently very successful \LLMs, such as, 
\GPTs\ in \cite{Brown2020,Radford2018,Radford2019,OpenAI2023} which are the models behind the famous \emph{ChatGPT} application,
\BERT\ models in Devlin et al.~\cite{Devlin2019},
and many others (cf., \eg, \cite{Yang2019,Dai2019,Raffel2020,Xiong2020,Lewis2019} and the references therein).

Beyond the \NLP\ applications for which Transformers and attention mechanisms have been introduced, similar ideas have been employed in several other areas, such as, 
computer vision (cf., \eg,  \cite{Dosovitskiy2020,Liu2021,Khan2021,Wang2017a}),
protein structure prediction  (cf., \eg, \cite{Jumper2021}),
multimodal learning (cf., \eg, \cite{Lu2019}), and
long sequence time-series forecasting (cf., \eg, \cite{Zhou2021}).
Moreover, we refer, \eg, to \cite{Luong2015,Chorowski2015}, \cite[Chapter 17]{Goldberg2017}, and \cite[Section 12.4.5.1]{Goodfellow2016} for explorations and explanations of the attention mechanism in the literature.

\subsection{Graph neural networks (GNNs)}

All \anns\ reviewed in the previous sections of this book are
designed to take real-valued vectors or sequences of real-valued vectors as inputs.
However, there are several learning problems based on data, such as social network data or molecular data, that are not optimally represented by real-valued vectors but are better represented by graphs (see, \eg, West~\cite{West2001} for an introduction on graphs).
As a consequence, many \ann\ architectures which can process graphs as inputs, so-called \GNNs, have been introduced in the literature.
\begin{itemize}
\item 
We refer, \eg, to \cite{SanchezLengeling,Wu2021,Zhou2020,Zhang2022a} for overview articles on \GNNs,
\item 
we refer, \eg, to 
	\cite{Scarselli2009,Gori2005} 
for foundational articles for \GNNs,
\item 
we refer, \eg, to 
	\cite{Velickovic2017,Yun2019} 
for applications of attention mechanisms (cf.\ \cref{sec:attention} above) to \GNNs,
\item 
we refer, \eg, to 
	\cite{Bruna2013,Wu2019,Defferrard2016,Ying2018} 
for \GNNs\ involving convolutions on graphs,
and
\item 
we refer, \eg, to \cite{SanchezGonzalez2020,Wu2018,Schuett2018,Gilmer2017,Battaglia2018} for applications of \GNNs\ to problems from the natural sciences.
\end{itemize}

\subsection{Neural operators}

In this section we review a few popular \ann-type architectures employed in \emph{operator learning}.
Roughly speaking, in operator learning one is not interested in learning a map between finite-dimensional euclidean spaces, but in learning a map from a space of functions to a space of functions.
Such a map between (typically infinite-dimensional) vector spaces is usually called an \emph{operator}.
An example of such a map is the solution operator of an evolutionary \PDE\ which maps the initial condition of the \PDE\ to the corresponding terminal value of the \PDE.
To approximate/learn operators it is necessary to develop parametrized families of operators, objects which we refer to as \emph{neural operators}.
Many different architectures for such neural operators have been proposed in the literature, some of which we now list in the next paragraphs.

One of the most successful neural operator architectures are so-called \FNOs\ introduced in Li et al.~\cite{Li2021fourier} (cf.\ also Kovachki et al.~\cite{Kovachki2023}).
Very roughly speaking, \FNOs\ are parametric maps on function spaces, which involve transformations on function values as well as on Fourier coefficients.
\FNOs\ have been derived based on the neural operators introduced in Li et al.~\cite{LiKovachki2020GraphKernel,li2020multipole} which are based on integral transformations with parametric integration kernels. 
We refer, \eg, to \cite {LiHuang2022GeoFNO,Wen2021,Brandstetter2022,KovachkiLanthaler2021FNO} and the references therein for extensions and theoretical results on \FNOs.

A simple and successful architecture for neural operators, which is based on a universal approximation theorem for neural operators, are the \deepONets\ introduced in Lu et al.~\cite{Lu2021}.
Roughly speaking, a \deepONet\ consists of two \anns\ that take as input the evaluation point of the output space and input function values at predetermined "sensor" points respectively, and that are joined together by a scalar product to produce the output of the \deepONet.
We refer, \eg, to 
\cite{LanthalerMolinaroMishra2022,PhamWarin2022,wang2021learning,meuris2023machine,liu2022deeppropnet,goswami2023learning,tan2022enhanced,kontolati2023learning,wu2023asymptotic,du2023approximation,zhang2023energy} 
for extensions and theoretical results on \deepONets.
For a comparison between \deepONets\ and \FNOs\ we refer, \eg, to Lu et al.~\cite{LuMengCai2022}.
 
A further natural approach is to employ \cnns\ (see \cref{section:cnns}) to develop neural operator architectures.
We refer, \eg, to 
	\cite{Khoo2021,Guo2016,Zhu2018,Raonic2023,Heiss2023} 
for such \cnn-based neural operators.
Finally, we refer, \eg, to \cite{NelsenStuart2021,LiuKutzBrunton2022,ChenWangYang2023,rafiq2022ssno,xiong2023koopman,MR4244919,MR4327361,schwab2023deep,deng2022approximation,de2022generic,Li2021,Jentzen2023a} for further neural operator architectures and theoretical results for neural operators.

%% file: parts/ANN_calculus.tex
\cchapter{ANN calculus}{chapter:ANN_calc}

In this chapter
we review certain operations that can be performed on the set of fully-connected feedforward \anns\ such as 
  compositions (see \cref{subsubsec:compositions_of_dnns}),
  paralellizations (see \cref{subsubsec:parallelizations_of_dnns}),
  scalar multiplications (see \cref{subsec:linear}), and
  sums (see \cref{subsec:sums})
and thereby review an appropriate calculus for fully-connected feedforward \anns.
The operations and the calculus for fully-connected feedforward \anns\ presented in this chapter will be used in \cref{sect:onedApprox,sect:multidApprox} to establish certain \ann\ approximation results.

In the literature such operations on \anns\ and such kind of calculus on \anns\ has been used in many research articles such as
\cite{Elbraechter2022,GrohsHornung2023,Petersen2018,JentzenSalimovaWelti2021,Grohs2022,GononGraeberJentzen2023,Opschoor2020,Guehring2020,Perekrestenko2018}
and the references therein.
The specific presentation of this chapter is based on Grohs et al.~\cite{Grohs2022,GrohsHornung2023}.

\section{Compositions of ANNs}
\label{subsubsec:compositions_of_dnns}

\subsection{Compositions of ANNs}

\begin{adef}{def:ANNcomposition}[Composition of \anns]
  We denote by 
  \begin{equation}
    \compANN{(\cdot)}{(\cdot)}\colon\allowbreak \{(\Phi,\Psi)\allowbreak\in\ANNs\times \ANNs\colon \inDimANN(\Phi)=\outDimANN(\Psi)\}\to\ANNs
  \end{equation}
  the function 
  which satisfies for all 
  $\Phi,\Psi\in\ANNs$,
  $k\in\{1,2,\dots,\lengthANN(\Phi)+\lengthANN(\Psi)-1\}$
  with
    $\inDimANN(\Phi)=\outDimANN(\Psi)$
  that
    $\lengthANN(\compANN\Phi\Psi)=\lengthANN(\Phi)+\lengthANN(\Psi)-1$
  and
  \begin{equation}
    \label{eq:defCompANN}
    (\weightANN k{\compANN\Phi\Psi},\biasANN k{\compANN\Phi\Psi})
    =
    \begin{cases}
      (\weightANN k\Psi,\biasANN k\Psi)&\colon k<\lengthANN(\Psi)\\
      (\weightANN1\Phi\weightANN {\lengthANN(\Psi)}\Psi,\weightANN1\Phi\biasANN {\lengthANN(\Psi)}\Psi+\biasANN1\Phi)&\colon k=\lengthANN(\Psi)\\
      (\weightANN {k-\lengthANN(\Psi)+1}\Phi,\biasANN{k-\lengthANN(\Psi)+1}\Phi)&\colon k>\lengthANN(\Psi)
    \end{cases}
  \end{equation}
  (cf.\ \cref{def:ANN}).
\end{adef}

\subsection{Elementary properties of compositions of ANNs}

\cfclear
\begingroup
\newcommand{\X}[1]{X_{#1}}
\begin{athm}{prop}{Lemma:PropertiesOfCompositions}[Properties of compositions of \anns]
Let 
$ 
  \Phi, \Psi \in \ANNs
$
satisfy 
$
  \inDimANN( \Phi ) = \outDimANN( \Psi ) 
$
\cfload.
Then
\begin{enumerate}[label=(\roman*)]
\item 
\label{PropertiesOfCompositions:Dims} 
it holds that
\begin{equation}
  \dims( \compANN{ \Phi }{ \Psi } ) 
  =
  (
    \dimANNlevel_0( \Psi ), \dimANNlevel_1(\Psi), \ldots, \dimANNlevel_{\hiddenLength(\Psi)}(\Psi), 
    \dimANNlevel_1(\Phi), \dimANNlevel_2(\Phi), \ldots, \dimANNlevel_{\lengthANN(\Phi)}(\Phi)
  ) ,
\end{equation}
\item \label{PropertiesOfCompositions:Length} it holds that
\begin{equation}\label{PropertiesOfCompositions:LengthDisplay}
[\lengthANN(\compANN{\Phi}{\Psi})-1]=[\lengthANN(\Phi)-1]+[\lengthANN(\Psi)-1],
\end{equation}
\item \label{PropertiesOfCompositions:HiddenLength} it holds that
\begin{equation}\label{PropertiesOfCompositions:HiddenLengthDisplay}
\hiddenLength(\compANN{\Phi}{\Psi})=\hiddenLength(\Phi)+\hiddenLength(\Psi),
\end{equation}
\item \label{PropertiesOfCompositions:Params} it holds that
\begin{equation}
\begin{split}
\paramANN(\compANN{\Phi}{\Psi})&
= \paramANN(\Phi)+\paramANN(\Psi)
+\dimANNlevel_1(\Phi)  (\dimANNlevel_{\lengthANN(\Psi)-1}(\Psi)+1)
\\&\quad
-\dimANNlevel_1(\Phi)(\dimANNlevel_0(\Phi) + 1)
- \dimANNlevel_{\lengthANN(\Psi)}(\Psi) (\dimANNlevel_{\lengthANN(\Psi)-1}(\Psi)+1)
\\&\le
\paramANN(\Phi)+\paramANN(\Psi)+\dimANNlevel_1(\Phi)  \dimANNlevel_{\hiddenLength(\Psi)}(\Psi),
\end{split}
\end{equation}  
and
    \item \label{PropertiesOfCompositions:Realization} it holds
for all  $\activation\in C(\R,\R)$
that 
$
  \functionANN{\activation}(\compANN{\Phi}{\Psi})
  \in C(\R^{\inDimANN(\Psi)},\R^{\outDimANN(\Phi)})
$ 
and
\begin{equation}\label{PropertiesOfCompositions:RealizationEquation}
  \functionANN{\activation}(\compANN{\Phi}{\Psi})=[\functionANN{\activation}(\Phi)]\circ [\functionANN{\activation}(\Psi)]
\end{equation}
  \end{enumerate}
\cfout.
\end{athm}

\begin{aproof}
  Throughout this proof,
  let
    $L=\lengthANN(\compANN\Phi\Psi)$
  and for every
    $a\in C(\R,\R)$
  let
  \begin{multline}
    \X a
    =
    \bigl\{
      x=(x_0,x_1,\dots,x_L)\in\R^{\dimANNlevel_0(\compANN\Phi\Psi)}\times\R^{\dimANNlevel_1(\compANN\Phi\Psi)}\times\dots\times\R^{\dimANNlevel_{L}(\compANN\Phi\Psi)}
      \colon\\
      \bpr{\forall\, k\in \{1,2,\dots,L\}\colon
        x_k
        =
        \multdim_{a\ind{(0,L)}(k)+\id_\R\ind{\{L\}}(k),\dimANNlevel_k(\compANN\Phi\Psi)}(\weightANN k{\compANN\Phi\Psi}x_{k-1}+\biasANN k{\compANN\Phi\Psi})
      }
    \bigr\}
    .
  \end{multline}
  \Nobs that
    the fact that
      $\lengthANN(\compANN\Phi\Psi)=\lengthANN(\Phi)+\lengthANN(\Psi)-1$
    and the fact that
      for all
        $\Theta\in\ANNs$
      it holds that
        $\hiddenLength(\Theta)=\lengthANN(\Theta)-1$
  establish
    \cref{PropertiesOfCompositions:Length,PropertiesOfCompositions:HiddenLength}.
  \Nobs that 
    \cref{elementaryPropertiesANN:weightsbiases} in \cref{Lemma:elementaryPropertiesANN}
    and \cref{eq:defCompANN}
  show that for all
    $k\in\{1,2,\dots,L\}$
  it holds that
  \begin{equation}
    \weightANN k{\compANN\Phi\Psi}\in
    \begin{cases}
      \R^{\dimANNlevel_{k}(\Psi)\times \dimANNlevel_{k-1}(\Psi)}&\colon k<\lengthANN(\Psi)\\
      \R^{\dimANNlevel_{1}(\Phi)\times \dimANNlevel_{\lengthANN(\Psi)-1}(\Psi)}&\colon k=\lengthANN(\Psi)\\
      \R^{\dimANNlevel_{k-\lengthANN(\Psi)+1}(\Phi)\times \dimANNlevel_{k-\lengthANN(\Psi)}(\Phi)}&\colon k>\lengthANN(\Psi).
    \end{cases}
  \end{equation}
    This,
    \cref{elementaryPropertiesANN:weightsbiases} in \cref{Lemma:elementaryPropertiesANN},
    and the fact that
      $\hiddenLength(\Psi)=\lengthANN(\Psi)-1$
  ensure that for all
    $k\in \{0,1,\dots,L\}$
  it holds that
  \begin{equation}
    \llabel{eq:levels}
    \dimANNlevel_k(\compANN\Phi\Psi)
    =
    \begin{cases}
      \dimANNlevel_k(\Psi)&\colon k\leq\hiddenLength(\Psi)\\
      \dimANNlevel_{k-\lengthANN(\Psi)+1}(\Phi)&\colon k>\hiddenLength(\Psi).
    \end{cases}
  \end{equation}
    This
  establishes
    \cref{PropertiesOfCompositions:Dims}.
  \Nobs that
    \lref{eq:levels}
  implies that
  \begin{equation}\label{PropertiesOfCompositions:CompositionParameters}
    \begin{split}
    \paramANN(\compANN\Phi\Psi)
    &=
    \smallsum\limits_{j = 1}^{L} \dimANNlevel_j(\compANN\Phi\Psi)(\dimANNlevel_{j-1}(\compANN\Phi\Psi) + 1)
    \\&=
    \br*{\smallsum\limits_{j = 1}^{\hiddenLength(\Psi)} \dimANNlevel_j(\Psi)(\dimANNlevel_{j-1}(\Psi) + 1)}
    +
    \dimANNlevel_1(\Phi)(\dimANNlevel_{\hiddenLength(\Psi)}(\Psi)+1)
    \\&\qquad+
    \br*{\smallsum\limits_{j = \lengthANN(\Psi)+1}^{L} \dimANNlevel_{j-\lengthANN(\Psi)+1}(\Phi)(\dimANNlevel_{j-\lengthANN(\Psi)}(\Phi) + 1)}
    \\&=
    \br*{\smallsum\limits_{j = 1}^{\lengthANN(\Psi)-1} \dimANNlevel_j(\Psi)(\dimANNlevel_{j-1}(\Psi) + 1)}
    +
    \dimANNlevel_1(\Phi)(\dimANNlevel_{\hiddenLength(\Psi)}(\Psi)+1)
    \\&\qquad+
    \br*{\smallsum\limits_{j = 2}^{\lengthANN(\Phi)} \dimANNlevel_{j}(\Phi)(\dimANNlevel_{j-1}(\Phi) + 1)}
    \\&=
    \bbr{\paramANN(\Psi)-\dimANNlevel_{\lengthANN(\Psi)}(\Psi)(\dimANNlevel_{\lengthANN(\Psi)-1}(\Psi) + 1)}
    +
    \dimANNlevel_1(\Phi)(\dimANNlevel_{\hiddenLength(\Psi)}(\Psi)+1)
    \\&\qquad+
    \bbr{\paramANN(\Phi)-\dimANNlevel_{1}(\Phi)(\dimANNlevel_{0}(\Phi) + 1)}
    .
    \end{split}
  \end{equation}
    This
  proves
    \cref{PropertiesOfCompositions:Params}.
  \Nobs that
    \lref{eq:levels}
    and \cref{elementaryPropertiesANN:inout} in \cref{Lemma:elementaryPropertiesANN}
  ensure that
  \begin{equation}
    \begin{split}
    \inDimANN(\compANN\Phi\Psi)
    &=
    \dimANNlevel_0(\compANN\Phi\Psi)
    =
    \dimANNlevel_0(\Psi)
    =
    \inDimANN(\Psi)
    \\\text{and}\qquad
    \outDimANN(\compANN\Phi\Psi)
    &=
    \dimANNlevel_{\lengthANN(\compANN\Phi\Psi)}(\compANN\Phi\Psi)
    =
    \dimANNlevel_{\lengthANN(\compANN\Phi\Psi)-\lengthANN(\Psi)+1}(\Phi)
    =
    \dimANNlevel_{\lengthANN(\Phi)}(\Phi)
    =
    \outDimANN(\Phi)
    .
    \end{split}
  \end{equation}
    This
  demonstrates that for all
    $a\in C(\R,\R)$
  it holds that
  \begin{equation}
    \llabel{eq:real0}
    \functionANN a(\compANN\Phi\Psi)
    \in
    C(\R^{\inDimANN(\compANN\Phi\Psi)},\R^{\outDimANN(\compANN\Phi\Psi)})
    =
    C(\R^{\inDimANN(\Psi)},\R^{\outDimANN(\Phi)})
    .
  \end{equation}
  \Moreover
    \cref{eq:defCompANN}
  implies that for all
    $k\in \N\cap (1,\lengthANN(\Phi)+1)$
  it holds that
  \begin{equation}
    (\weightANN{\lengthANN(\Psi)+k-1}{\compANN\Phi\Psi},\biasANN{\lengthANN(\Psi)+k-1}{\compANN\Phi\Psi})
    =
    (\weightANN{k}{\Phi},\biasANN k\Phi)
    .
  \end{equation}
    This
    and \lref{eq:levels}
  ensure that for all
    $a\in C(\R,\R)$,
    $x=(x_0,x_1,\dots,x_L)\in X_a$,
    $k\in \N\cap (1,\lengthANN(\Phi)+1)$
  it holds that
  \begin{equation}
    \llabel{eq:real1}
    \begin{split}
    x_{\lengthANN(\Psi)+k-1}
    &=
    \multdim_{a\ind{(0,L)}(\lengthANN(\Psi)+k-1)+\id_\R\ind{\{L\}}(\lengthANN(\Psi)+k-1),\dimANNlevel_{k}(\Phi)}(\weightANN {k}{\Phi}x_{\lengthANN(\Psi)+k-2}+\biasANN {k}{\Phi})
    \\&=
    \multdim_{a\ind{(0,\lengthANN(\Phi))}(k)+\id_\R\ind{\{\lengthANN(\Phi)\}}(k),\dimANNlevel_{k}(\Phi)}(\weightANN {k}{\Phi}x_{\lengthANN(\Psi)+k-2}+\biasANN {k}{\Phi})
    .
    \end{split}
  \end{equation}
  \Moreover
    \cref{eq:defCompANN}
    and \lref{eq:levels}
  show that for all
    $a\in C(\R,\R)$,
    $x=(x_0,x_1,\dots,\allowbreak x_L)\in X_a$
  it holds that
  \begin{equation}
    \begin{split}
    x_{\lengthANN(\Psi)}
    &=
    \multdim_{a\ind{(0,L)}(\lengthANN(\Psi))+\id_\R\ind{\{L\}}(\lengthANN(\Psi)),\dimANNlevel_{\lengthANN(\Psi)}(\compANN\Phi\Psi)}(
      \weightANN{\lengthANN(\Psi)}{\compANN\Phi\Psi}x_{\lengthANN(\Psi)-1}+\biasANN{\lengthANN(\Psi)}{\compANN\Phi\Psi}
    )
    \\&=
    \multdim_{a\ind{(0,\lengthANN(\Phi))}(1)+\id_\R\ind{\{\lengthANN(\Phi)\}}(1),\dimANNlevel_{1}(\Phi)}(
      \weightANN1\Phi \weightANN{\lengthANN(\Psi)}\Psi x_{\lengthANN(\Psi)-1}+\weightANN1\Phi\biasANN{\lengthANN(\Psi)}{\Psi}+\biasANN1\Phi
    )
    \\&=
    \multdim_{a\ind{(0,\lengthANN(\Phi))}(1)+\id_\R\ind{\{\lengthANN(\Phi)\}}(1),\dimANNlevel_{1}(\Phi)}(
      \weightANN1\Phi (\weightANN{\lengthANN(\Psi)}\Psi x_{\lengthANN(\Psi)-1}+\biasANN{\lengthANN(\Psi)}{\Psi})+\biasANN1\Phi
    )
    .
    \end{split}
  \end{equation}
  Combining
    this
    and \lref{eq:real1}
  proves that for all
    $a\in C(\R,\R)$,
    $x=(x_0,x_1,\dots,x_L)\in X_a$
  it holds that
  \begin{equation}
    \llabel{eq:real2}
    (\functionANN{a}(\Phi))(\weightANN{\lengthANN(\Psi)}\Psi x_{\lengthANN(\Psi)-1}+\biasANN{\lengthANN(\Psi)}{\Psi})
    =
    x_L
    .
  \end{equation}
  \Moreover[note]%
    \cref{eq:defCompANN}
    and \lref{eq:levels}
  \prove[si] that for all
    $a\in C(\R,\R)$,
    $x=(x_0,x_1,\dots,x_L)\in X_a$,
    $k\in\N\cap(0,\lengthANN(\Psi))$
  it holds that
  \begin{equation}
    x_k=\multdim_{a,\dimANNlevel_k(\Psi)}(\weightANN k{\Psi}x_{k-1}+\biasANN k{\Psi})		
  \end{equation}
    This
  \proves that for all
    $a\in C(\R,\R)$,
    $x=(x_0,x_1,\dots,x_L)\in X_a$
  it holds that
  \begin{equation}
    (\functionANN a(\Psi))(x_0)
    =
    \weightANN{\lengthANN(\Psi)}\Psi x_{\lengthANN(\Psi)-1}+\biasANN{\lengthANN(\Psi)}{\Psi}
    .
  \end{equation}
  Combining
    this
  with
    \lref{eq:real2}
  \proves that for all
    $a\in C(\R,\R)$,
    $x=(x_0,x_1,\dots,x_L)\in X_a$
  it holds that
  \begin{equation}
    \llabel{eq:real4}
    (\functionANN a(\Phi))\bpr{(\functionANN a(\Psi))(x_0) }
    =
    x_L
    =
    \bpr{\functionANN a(\compANN\Phi\Psi)}(x_0)
    .
  \end{equation}
    This
    and \lref{eq:real0}
  prove
    \cref{PropertiesOfCompositions:Realization}.
\end{aproof}
\endgroup

\subsection{Associativity of compositions of ANNs}

\cfclear
\begin{athm}{lemma}{Lemma:CompositionAssociative1}
  Let $ \Phi_1, \Phi_2, \Phi_3 \in \ANNs $
  satisfy
  $ 
    \inDimANN( \Phi_1 ) = \outDimANN( \Phi_2 )
  $,
  $
    \inDimANN( \Phi_2 ) = \outDimANN( \Phi_3 )
  $,
  and $\lengthANN(\Phi_2)=1$
  \cfload.
  Then
  \begin{equation}
    \compANN{(\compANN{\Phi_1}{\Phi_2})}{\Phi_3}=\compANN{\Phi_1}{(\compANN{\Phi_2}{\Phi_3})}
  \end{equation}
  \cfout.
\end{athm}

\begin{aproof}
  \Nobs that
    the fact that
      for all
        $\Psi_1,\Psi_2\in\ANNs$
        with $\inDimANN(\Psi_1)=\outDimANN(\Psi_2)$
      it holds that
        $\lengthANN(\compANN{\Psi_1}{\Psi_2})=\lengthANN(\Psi_1)+\lengthANN(\Psi_2)-1$
    and the assumption that 
      $\lengthANN(\Phi_2)=1$
  ensure that
  \begin{equation}
    \llabel{eq:l1}
    \lengthANN(\compANN{\Phi_1}{\Phi_2})
    =
    \lengthANN(\Phi_1)
    \qquad\text{and}\qquad
    \lengthANN(\compANN{\Phi_2}{\Phi_3})
    =
    \lengthANN(\Phi_3)
  \end{equation}
  \cfload.
  \Hence that
  \begin{equation}
    \llabel{eq:l2}
    \lengthANN(\compANN{(\compANN{\Phi_1}{\Phi_2})}{\Phi_3})
    =
    \lengthANN(\Phi_1)+\lengthANN(\Phi_3)
    =
    \lengthANN(\compANN{\Phi_1}{(\compANN{\Phi_2}{\Phi_3})})
    .
  \end{equation}
  \Moreover
    \lref{eq:l1},
    \cref{eq:defCompANN},
    and the assumption that
      $\lengthANN(\Phi_2)=1$
  imply that for all
    $k\in\{1,2,\dots,\lengthANN(\Phi_1)\}$
  it holds that
  \begin{equation}
    \llabel{eq:1}
    (\weightANN k{\compANN{\Phi_1}{\Phi_2}},\biasANN k{\compANN{\Phi_1}{\Phi_2}})
    =
    \begin{cases}
      (\weightANN1{\Phi_1}\weightANN1{\Phi_2},\weightANN1{\Phi_1}\biasANN1{\Phi_2}+\biasANN1{\Phi_1})&\colon k=1\\
      (\weightANN k{\Phi_1},\biasANN k{\Phi_1})&\colon k>1.
    \end{cases}
  \end{equation}
    This,
    \cref{eq:defCompANN},
    and \lref{eq:l2}
  prove that for all
    $k\in\{1,2,\dots,\lengthANN(\Phi_1)+\lengthANN(\Phi_3)-1\}$
  it holds that
  \begin{equation}
    \llabel{eq:4}
    \begin{split}
      &(\weightANN k{\compANN{(\compANN{\Phi_1}{\Phi_2})}{\Phi_3}},\biasANN k{\compANN{(\compANN{\Phi_1}{\Phi_2})}{\Phi_3}})
      \\&=
      \begin{cases}
        (\weightANN{k}{\Phi_3},\biasANN{k}{\Phi_3})&\colon k<\lengthANN(\Phi_3)\\
        (\weightANN{1}{\compANN{\Phi_1}{\Phi_2}}\weightANN{\lengthANN(\Phi_3)}{\Phi_3},\weightANN1{\compANN{\Phi_1}{\Phi_2}}\biasANN{\lengthANN(\Phi_3)}{\Phi_3}+\biasANN{1}{\compANN{\Phi_1}{\Phi_2}})&\colon k=\lengthANN(\Phi_3)\\
        (\weightANN{k-\lengthANN(\Phi_3)+1}{\compANN{\Phi_1}{\Phi_2}},\biasANN{k-\lengthANN(\Phi_3)+1}{\compANN{\Phi_1}{\Phi_2}})&\colon k>\lengthANN(\Phi_3)
      \end{cases}
      \\&=
      \begin{cases}
        (\weightANN{k}{\Phi_3},\biasANN{k}{\Phi_3})&\colon k<\lengthANN(\Phi_3)\\
        (\weightANN{1}{\compANN{\Phi_1}{\Phi_2}}\weightANN{\lengthANN(\Phi_3)}{\Phi_3},\weightANN1{\compANN{\Phi_1}{\Phi_2}}\biasANN{\lengthANN(\Phi_3)}{\Phi_3}+\biasANN{1}{\compANN{\Phi_1}{\Phi_2}})&\colon k=\lengthANN(\Phi_3)\\
        (\weightANN{k-\lengthANN(\Phi_3)+1}{\Phi_1},\biasANN{k-\lengthANN(\Phi_3)+1}{\Phi_1})&\colon k>\lengthANN(\Phi_3).
      \end{cases}
    \end{split}
  \end{equation}
  \Moreover
    \cref{eq:defCompANN},
    \lref{eq:l1},
    and \lref{eq:l2}
  show that for all
    $k\in \{1,2,\dots,\lengthANN(\Phi_1)+\lengthANN(\Phi_3)-1\}$
  it holds that
  \begin{equation}
    \begin{split}
      &(\weightANN k{\compANN{\Phi_1}{(\compANN{\Phi_2}{\Phi_3})}},\biasANN k{\compANN{\Phi_1}{(\compANN{\Phi_2}{\Phi_3})}})
      \\&=
      \begin{cases}
        (\weightANN k{\compANN{\Phi_2}{\Phi_3}},\biasANN k{\compANN{\Phi_2}{\Phi_3}})&\colon k<\lengthANN({\compANN{\Phi_2}{\Phi_3}})\\
        (\weightANN1{\Phi_1}\weightANN {\lengthANN({\compANN{\Phi_2}{\Phi_3}})}{\compANN{\Phi_2}{\Phi_3}},\weightANN1\Phi\biasANN {\lengthANN({\compANN{\Phi_2}{\Phi_3}})}{\compANN{\Phi_2}{\Phi_3}}+\biasANN1{\Phi_1})&\colon k=\lengthANN({\compANN{\Phi_2}{\Phi_3}})\\
        (\weightANN {k-\lengthANN({\compANN{\Phi_2}{\Phi_3}})+1}{\Phi_1},\biasANN{k-\lengthANN({\compANN{\Phi_2}{\Phi_3}})+1}{\Phi_1})&\colon k>\lengthANN({\compANN{\Phi_2}{\Phi_3}})
      \end{cases}
      \\&=
      \begin{cases}
        (\weightANN k{\Phi_3},\biasANN k{\Phi_3})&\colon k<\lengthANN(\Phi_3)\\
        (\weightANN1{\Phi_1}\weightANN {\lengthANN(\Phi_3)}{\compANN{\Phi_2}{\Phi_3}},\weightANN1\Phi\biasANN {\lengthANN(\Phi_3)}{\compANN{\Phi_2}{\Phi_3}}+\biasANN1{\Phi_1})&\colon k=\lengthANN(\Phi_3)\\
        (\weightANN {k-\lengthANN(\Phi_3)+1}{\Phi_1},\biasANN{k-\lengthANN(\Phi_3)+1}{\Phi_1})&\colon k>\lengthANN(\Phi_3).
      \end{cases}
    \end{split}
  \end{equation}
  Combining
    this
  with
    \lref{eq:4}
  establishes that for all
    $k\in\{1,2,\dots,\lengthANN(\Phi_1)+\lengthANN(\Phi_3)-1\}\backslash\{\lengthANN(\Phi_3)\}$
  it holds that
  \begin{equation}
    \llabel{eq:7}
    (\weightANN k{\compANN{(\compANN{\Phi_1}{\Phi_2})}{\Phi_3}},\biasANN k{\compANN{(\compANN{\Phi_1}{\Phi_2})}{\Phi_3}})
    =
    (\weightANN k{\compANN{\Phi_1}{(\compANN{\Phi_2}{\Phi_3})}},\biasANN k{\compANN{\Phi_1}{(\compANN{\Phi_2}{\Phi_3})}})
    .
  \end{equation}
  \Moreover
    \lref{eq:1}
    and \cref{eq:defCompANN}
  ensure that
  \begin{equation}
    \llabel{eq:6}
    \weightANN{1}{\compANN{\Phi_1}{\Phi_2}}\weightANN{\lengthANN(\Phi_3)}{\Phi_3}
    =
    \weightANN{1}{\Phi_1}\weightANN{1}{\Phi_2}\weightANN{\lengthANN(\Phi_3)}{\Phi_3}
    =
    \weightANN1{\Phi_1}\weightANN {\lengthANN(\Phi_3)}{\compANN{\Phi_2}{\Phi_3}}.
  \end{equation}
  \Moreover
    \lref{eq:1}
    and \cref{eq:defCompANN}
  demonstrate that
  \begin{equation}
    \begin{split}
      \weightANN1{\compANN{\Phi_1}{\Phi_2}}\biasANN{\lengthANN(\Phi_3)}{\Phi_3}+\biasANN{1}{\compANN{\Phi_1}{\Phi_2}}
      &=
      \weightANN1{\Phi_1}\weightANN1{\Phi_2}\biasANN{\lengthANN(\Phi_3)}{\Phi_3}+\weightANN1{\Phi_1}\biasANN1{\Phi_2}+\biasANN1{\Phi_1}
      \\&=
      \weightANN1{\Phi_1}(\weightANN1{\Phi_2}\biasANN{\lengthANN(\Phi_3)}{\Phi_3}+\biasANN1{\Phi_2})+\biasANN1{\Phi_1}
      \\&=
      \weightANN1\Phi\biasANN {\lengthANN(\Phi_3)}{\compANN{\Phi_2}{\Phi_3}}+\biasANN1{\Phi_1}
      .
    \end{split}
  \end{equation}
  Combining
    this
    and \lref{eq:6}
  with
    \lref{eq:7}
  proves that for all
    $k\in\{1,2,\dots,\lengthANN(\Phi_1)+\lengthANN(\Phi_3)-1\}$
  it holds that
  \begin{equation}
    (\weightANN k{\compANN{(\compANN{\Phi_1}{\Phi_2})}{\Phi_3}},\biasANN k{\compANN{(\compANN{\Phi_1}{\Phi_2})}{\Phi_3}})
    =
    (\weightANN k{\compANN{\Phi_1}{(\compANN{\Phi_2}{\Phi_3})}},\biasANN k{\compANN{\Phi_1}{(\compANN{\Phi_2}{\Phi_3})}})
    .
  \end{equation}
    This
    and \lref{eq:l2} 
  imply that
  \begin{equation}
    \compANN{(\compANN{\Phi_1}{\Phi_2})}{\Phi_3}
    =
    \compANN{\Phi_1}{(\compANN{\Phi_2}{\Phi_3})}
    .
  \end{equation}
\end{aproof}

\cfclear
\begin{athm}{lemma}{Lemma:CompositionAssociative2}
  Let $ \Phi_1, \Phi_2, \Phi_3 \in \ANNs $
  satisfy
  $ 
    \inDimANN( \Phi_1 ) = \outDimANN( \Phi_2 )
  $,
  $
    \inDimANN( \Phi_2 ) = \outDimANN( \Phi_3 )
  $,
  and $\lengthANN(\Phi_2)>1$
  \cfload.
  Then
  \begin{equation}
    \compANN{(\compANN{\Phi_1}{\Phi_2})}{\Phi_3}=\compANN{\Phi_1}{(\compANN{\Phi_2}{\Phi_3})}
  \end{equation}
  \cfout.
\end{athm}
\begin{aproof}
  \Nobs that
    the fact that
      for all
        $\Psi,\Theta\in\ANNs$
      it holds that
        $\lengthANN(\compANN\Psi\Theta)=\lengthANN(\Psi)+\lengthANN(\Theta)-1$
  ensures that
  \begin{equation}
    \llabel{eq:l}
  \begin{split}
      \lengthANN(\compANN{(\compANN{\Phi_1}{\Phi_2})}{\Phi_3})
      &=
      \lengthANN(\compANN{\Phi_1}{\Phi_2})+\lengthANN(\Phi_3)-1
      \\&=
      \lengthANN(\Phi_1)+\lengthANN(\Phi_2)+\lengthANN(\Phi_3)-2
      \\&=
      \lengthANN(\Phi_1)+\lengthANN(\compANN{\Phi_2}{\Phi_3})-1
      \\&=
      \lengthANN(\compANN{\Phi_1}{(\compANN{\Phi_2}{\Phi_3})})
  \end{split}
  \end{equation}
  \cfload.
  \Moreover[Furthermore]
    \cref{eq:defCompANN}
  shows that for all
    $k\in\{1,2,\dots,\allowbreak\lengthANN(\compANN{(\compANN{\Phi_1}{\Phi_2})}{\Phi_3})\}$
  it holds that
  \begin{equation}
    \llabel{eq:4}
    \begin{split}
      &(\weightANN k{\compANN{(\compANN{\Phi_1}{\Phi_2})}{\Phi_3}},\biasANN k{\compANN{(\compANN{\Phi_1}{\Phi_2})}{\Phi_3}})
      \\&=
      \begin{cases}
        (\weightANN{k}{\Phi_3},\biasANN{k}{\Phi_3})&\colon k<\lengthANN(\Phi_3)\\
        (\weightANN{1}{\compANN{\Phi_1}{\Phi_2}}\weightANN{\lengthANN(\Phi_3)}{\Phi_3},\weightANN1{\compANN{\Phi_1}{\Phi_2}}\biasANN{\lengthANN(\Phi_3)}{\Phi_3}+\biasANN{1}{\compANN{\Phi_1}{\Phi_2}})&\colon k=\lengthANN(\Phi_3)\\
        (\weightANN{k-\lengthANN(\Phi_3)+1}{\compANN{\Phi_1}{\Phi_2}},\biasANN{k-\lengthANN(\Phi_3)+1}{\compANN{\Phi_1}{\Phi_2}})&\colon k>\lengthANN(\Phi_3).
      \end{cases}
    \end{split}
  \end{equation}
  \Moreover
    \cref{eq:defCompANN}
    and the assumption that
      $\lengthANN(\Phi_2)>1$
  ensure that for all
    $k\in\N\cap(\lengthANN(\Phi_3),\lengthANN(\compANN{(\compANN{\Phi_1}{\Phi_2})}{\Phi_3})]$
  it holds that
  \begin{equation}
    \begin{split}
      &(\weightANN{k-\lengthANN(\Phi_3)+1}{\compANN{\Phi_1}{\Phi_2}},\biasANN{k-\lengthANN(\Phi_3)+1}{\compANN{\Phi_1}{\Phi_2}})
      \\&=
      \begin{cases}
        (\weightANN {k-\lengthANN(\Phi_3)+1}{\Phi_2},\biasANN {k-\lengthANN(\Phi_3)+1}{\Phi_2})&\colon {k-\lengthANN(\Phi_3)+1}<\lengthANN({\Phi_2})\\
        (\weightANN1{\Phi_1}\weightANN {\lengthANN({\Phi_2})}{\Phi_2},\weightANN1{\Phi_1}\biasANN {\lengthANN({\Phi_2})}{\Phi_2}+\biasANN1{\Phi_1})&\colon {k-\lengthANN(\Phi_3)+1}=\lengthANN({\Phi_2})\\
        (\weightANN {{k-\lengthANN(\Phi_3)+1}-\lengthANN({\Phi_2})+1}{\Phi_1},\biasANN{{k-\lengthANN(\Phi_3)+1}-\lengthANN({\Phi_2})+1}{\Phi_1})&\colon {k-\lengthANN(\Phi_3)+1}>\lengthANN({\Phi_2})
      \end{cases}
      \\&=
      \begin{cases}
        (\weightANN {k-\lengthANN(\Phi_3)+1}{\Phi_2},\biasANN {k-\lengthANN(\Phi_3)+1}{\Phi_2})&\colon k<\lengthANN({\Phi_2})+\lengthANN(\Phi_3)-1\\
        (\weightANN1{\Phi_1}\weightANN {\lengthANN({\Phi_2})}{\Phi_2},\weightANN1{\Phi_1}\biasANN {\lengthANN({\Phi_2})}{\Phi_2}+\biasANN1{\Phi_1})&\colon k=\lengthANN({\Phi_2})+\lengthANN(\Phi_3)-1\\
        (\weightANN {{k-\lengthANN(\Phi_3)}-\lengthANN({\Phi_2})+2}{\Phi_1},\biasANN{{k-\lengthANN(\Phi_3)}-\lengthANN({\Phi_2})+2}{\Phi_1})&\colon k>\lengthANN({\Phi_2})+\lengthANN(\Phi_3)-1.
      \end{cases}
    \end{split}
  \end{equation}
  Combining
    this
  with
    \lref{eq:4}
  proves that for all
    $k\in\{1,2,\dots,\lengthANN(\compANN{(\compANN{\Phi_1}{\Phi_2})}{\Phi_3})\}$
  it holds that
  \begin{equation}
    \llabel{eq:c1}
    \begin{split}
      &(\weightANN k{\compANN{(\compANN{\Phi_1}{\Phi_2})}{\Phi_3}},\biasANN k{\compANN{(\compANN{\Phi_1}{\Phi_2})}{\Phi_3}})
      \\&=
      \begin{cases}
        (\weightANN{k}{\Phi_3},\biasANN{k}{\Phi_3})&\colon k<\lengthANN(\Phi_3)\\
        (\weightANN{1}{\Phi_2}\weightANN{\lengthANN(\Phi_3)}{\Phi_3},\weightANN1{\Phi_2}\biasANN{\lengthANN(\Phi_3)}{\Phi_3}+\biasANN{1}{\Phi_2})&\colon k=\lengthANN(\Phi_3)\\
        (\weightANN{k-\lengthANN(\Phi_3)+1}{\Phi_2},\biasANN{k-\lengthANN(\Phi_3)+1}{\Phi_2})&\colon \lengthANN(\Phi_3)<k<\lengthANN(\Phi_2)+\lengthANN(\Phi_3)-1\\
        (\weightANN1{\Phi_1}\weightANN {\lengthANN({\Phi_2})}{\Phi_2},\weightANN1{\Phi_1}\biasANN {\lengthANN({\Phi_2})}{\Phi_2}+\biasANN1{\Phi_1})&\colon k=\lengthANN(\Phi_2)+\lengthANN(\Phi_3)-1\\
        (\weightANN {{k-\lengthANN(\Phi_3)}-\lengthANN({\Phi_2})+2}{\Phi_1},\biasANN{{k-\lengthANN(\Phi_3)}-\lengthANN({\Phi_2})+2}{\Phi_1})&\colon k>\lengthANN(\Phi_2)+\lengthANN(\Phi_3)-1.
      \end{cases}
    \end{split}
  \end{equation}
  \Moreover
    \cref{eq:defCompANN},
    the fact that
      $\lengthANN(\compANN{\Phi_2}{\Phi_3})=\lengthANN(\Phi_2)+\lengthANN(\Phi_3)-1$,
    and the assumption that
      $\lengthANN(\Phi_2)>1$
  demonstrate that for all
    $k\in \{1,2,\dots,\lengthANN(\compANN{\Phi_1}{(\compANN{\Phi_2}{\Phi_3})})\}$
  it holds that
  \begin{equation}
    \begin{split}
      &(\weightANN k{\compANN{\Phi_1}{(\compANN{\Phi_2}{\Phi_3})}},\biasANN k{\compANN{\Phi_1}{(\compANN{\Phi_2}{\Phi_3})}})
      \\&=
      \begin{cases}
      	(\weightANN k{\compANN{\Phi_2}{\Phi_3}},\biasANN k{\compANN{\Phi_2}{\Phi_3}})&\colon k<\lengthANN({\compANN{\Phi_2}{\Phi_3}})\\
      	(\weightANN1{\Phi_1}\weightANN {\lengthANN({\compANN{\Phi_2}{\Phi_3}})}{\compANN{\Phi_2}{\Phi_3}},\weightANN1\Phi\biasANN {\lengthANN({\compANN{\Phi_2}{\Phi_3}})}{\compANN{\Phi_2}{\Phi_3}}+\biasANN1{\Phi_1})&\colon k=\lengthANN({\compANN{\Phi_2}{\Phi_3}})\\
      	(\weightANN {k-\lengthANN({\compANN{\Phi_2}{\Phi_3}})+1}{\Phi_1},\biasANN{k-\lengthANN({\compANN{\Phi_2}{\Phi_3}})+1}{\Phi_1})&\colon k>\lengthANN({\compANN{\Phi_2}{\Phi_3}})
      \end{cases}
      \\&=
      \begin{cases}
        (\weightANN k{\compANN{\Phi_2}{\Phi_3}},\biasANN k{\compANN{\Phi_2}{\Phi_3}})&\colon k<\lengthANN(\Phi_2)+\lengthANN(\Phi_3)-1\\
        \begin{aligned}
          &(\weightANN1{\Phi_1}\weightANN {\lengthANN(\Phi_2)+\lengthANN(\Phi_3)-1}{\compANN{\Phi_2}{\Phi_3}},
          \\&\qquad
          \weightANN1\Phi\biasANN {\lengthANN(\Phi_2)+\lengthANN(\Phi_3)-1}{\compANN{\Phi_2}{\Phi_3}}+\biasANN1{\Phi_1})
        \end{aligned}&\colon k=\lengthANN(\Phi_2)+\lengthANN(\Phi_3)-1\\
        (\weightANN {k-\lengthANN(\Phi_2)-\lengthANN(\Phi_3)+2}{\Phi_1},\biasANN{k-\lengthANN(\Phi_2)-\lengthANN(\Phi_3)+2}{\Phi_1})&\colon k>\lengthANN(\Phi_2)+\lengthANN(\Phi_3)-1
      \end{cases}
      \\&=
      \begin{cases}
        (\weightANN k{\Phi_3},\biasANN k{\Phi_3})&\colon k<\lengthANN({\Phi_3})\\
        (\weightANN1{\Phi_2}\weightANN {\lengthANN(\Phi_3)}{\Phi_3},\weightANN1{\Phi_2}\biasANN {\lengthANN(\Phi_3)}{\Phi_3}+\biasANN1{\Phi_2})&\colon k=\lengthANN(\Phi_3)\\
        (\weightANN {k-\lengthANN(\Phi_3)+1}{\Phi_2},\biasANN{k-\lengthANN(\Phi_3)+1}{\Phi_2})&\colon \lengthANN(\Phi_3)<k<\lengthANN(\Phi_2)+\lengthANN(\Phi_3)-1\\
        (\weightANN1{\Phi_1}\weightANN {\lengthANN(\Phi_2)}{\Phi_2},\weightANN1\Phi\biasANN {\lengthANN(\Phi_2)}{\Phi_2}+\biasANN1{\Phi_1})&\colon k=\lengthANN(\Phi_2)+\lengthANN(\Phi_3)-1\\
        (\weightANN {k-\lengthANN(\Phi_2)-\lengthANN(\Phi_3)+2}{\Phi_1},\biasANN{k-\lengthANN(\Phi_2)-\lengthANN(\Phi_3)+2}{\Phi_1})&\colon k>\lengthANN(\Phi_2)+\lengthANN(\Phi_3)-1.
      \end{cases}
    \end{split}
  \end{equation}
    This,
    \lref{eq:c1},
    and \lref{eq:l}
  establish that for all
    $k\in \{1,2,\dots,\lengthANN(\Phi_1)+\lengthANN(\Phi_2)+\lengthANN(\Phi_3)-2\}$
  it holds that
  \begin{equation}
    (\weightANN k{\compANN{(\compANN{\Phi_1}{\Phi_2})}{\Phi_3}},\biasANN k{\compANN{(\compANN{\Phi_1}{\Phi_2})}{\Phi_3}})
    =
    (\weightANN k{\compANN{\Phi_1}{(\compANN{\Phi_2}{\Phi_3})}},\biasANN k{\compANN{\Phi_1}{(\compANN{\Phi_2}{\Phi_3})}})
    .
  \end{equation}
  \Hence that
  \begin{equation}
    \compANN{(\compANN{\Phi_1}{\Phi_2})}{\Phi_3}
    =
    \compANN{\Phi_1}{(\compANN{\Phi_2}{\Phi_3})}
    .
  \end{equation}
\end{aproof}

\cfclear
\begin{athm}{cor}{Lemma:CompositionAssociative}
  Let $ \Phi_1, \Phi_2, \Phi_3 \in \ANNs $
  satisfy
  $ 
    \inDimANN( \Phi_1 ) = \outDimANN( \Phi_2 )
  $ and
  $
    \inDimANN( \Phi_2 ) = \outDimANN( \Phi_3 )
  $
  \cfload.
  Then
  \begin{equation}
    \llabel{eq:claim}
    \compANN{(\compANN{\Phi_1}{\Phi_2})}{\Phi_3}=\compANN{\Phi_1}{(\compANN{\Phi_2}{\Phi_3})}
  \end{equation}
  \cfout.
\end{athm}
\begin{aproof}
  \Nobs that
    \cref{Lemma:CompositionAssociative1}
    and \cref{Lemma:CompositionAssociative2}
  establish
    \lref{eq:claim}.
\end{aproof}

\subsection{Powers of ANNs}

\todoc{Use A notation for touple: DOES NOT WORK BECAUSE IT IS ONLY INTRODUCED LATER}

\begin{adef}{def:iteratedANNcomposition}[Powers of \anns]
We denote by 
$
  \powANN{ ( \cdot ) }{ n } 
  \colon 
  \{
    \Phi\allowbreak \in \ANNs\colon \inDimANN(\Phi)=\outDimANN(\Phi)
  \}
  \allowbreak \to \ANNs
$, 
$ n \in \N_0 $, 
the functions
which satisfy for all $ n \in \N_0 $, 
$ \Phi \in \ANNs $ 
with $ \inDimANN( \Phi ) = \outDimANN( \Phi ) $ 
that 
\begin{equation}
\label{iteratedANNcomposition:equation}
\begin{split}
  \powANN{ \Phi }{ n }
  =
  \begin{cases} 
    \bpr{ \idMatrix_{ \outDimANN( \Phi ) }, (0, 0, \dots, 0) }
    \in
    \R^{ 
      \outDimANN( \Phi ) \times \outDimANN( \Phi ) } \times \R^{ \outDimANN( \Phi ) 
    }
  & 
    \colon n = 0 
\\[1ex]
  \,
  \compANN{ \Phi }{ ( \powANN{ \Phi }{ (n - 1) } ) } 
  & 
    \colon n \in \N 
\end{cases}
\end{split}
\end{equation}	
(cf.\ \cref{def:ANN,def:ANNcomposition,def:identityMatrix}). 
\end{adef}

\cfclear 
\begin{lemma}[Number of hidden layers of powers of \anns]
\label{lem:number_of_hidden_layers_of_powers_of_ANNs}
Let $ n \in \N_0 $, $ \Phi \in \ANNs $ 
satisfy 
$ 
  \inDimANN( \Phi ) = \outDimANN( \Phi ) 
$
\cfload. 
Then 
\begin{equation}
\label{eq:number_of_hidden_layers_of_powers_of_ANNs}
  \hiddenLength( \powANN{ \Phi }{ n } )
  =
  n \hiddenLength( \Phi )
\end{equation}
\cfload. 
\end{lemma}

\begin{proof}[Proof of \cref{lem:number_of_hidden_layers_of_powers_of_ANNs}]
\Nobs that 
\cref{Lemma:PropertiesOfCompositions}, 
\cref{iteratedANNcomposition:equation}, 
and induction establish 
\cref{eq:number_of_hidden_layers_of_powers_of_ANNs}. 
The proof of \cref{lem:number_of_hidden_layers_of_powers_of_ANNs} 
is thus complete. 
\end{proof}

\section{Parallelizations of ANNs}
\label{subsubsec:parallelizations_of_dnns}

\subsection{Parallelizations of ANNs with the same length}

\cfclear
\begin{adef}{def:simpleParallelization}[Parallelization of \anns]
Let $ n \in \N $. 
Then we denote by 
\begin{equation}
  \parallelizationSpecial_{n} \colon 
  \bcu{
    \Phi = ( \Phi_1, \dots, \Phi_n ) \in \ANNs^n 
    \colon \lengthANN( \Phi_1 ) = 
    \lengthANN( \Phi_2 ) = \ldots = \lengthANN( \Phi_n ) 
  } \to \ANNs
\end{equation}
the function which satisfies 
for all
$ \Phi = ( \Phi_1, \dots, \Phi_n ) \in \ANNs^n $,
$k\in \{1,2,\dots,\lengthANN(\Phi_1)\}$
with 
$ 
  \lengthANN( \Phi_1 ) = \lengthANN( \Phi_2 ) = \dots = \lengthANN( \Phi_n ) 
$ 
  that
\begin{align}
  &\lengthANN(\parallelizationSpecial_n(\Phi))
  =
  \lengthANN(\Phi_1),
  \qquad
  \weightANN k{\parallelizationSpecial_n(\Phi)}
  =
  \begin{pmatrix}
    \weightANN{k}{\Phi_1}& 0& 0& \cdots& 0\\
    0& \weightANN{k}{\Phi_2}& 0&\cdots& 0\\
    0& 0& \weightANN{k}{\Phi_3}&\cdots& 0\\
    \vdots& \vdots&\vdots& \ddots& \vdots\\
    0& 0& 0&\cdots& \weightANN{k}{\Phi_n}
  \end{pmatrix},\nonumber
  \\&\text{and}\qquad\label{parallelisationSameLengthDef}
  \biasANN k{\parallelizationSpecial_n(\Phi)}
  =
  \begin{pmatrix}
    \biasANN k{\Phi_1}\\
    \biasANN k{\Phi_2}\\
    \vdots\\
    \biasANN k{\Phi_n}
  \end{pmatrix}
\end{align}
  (cf.\ \cref{def:ANN}).
\end{adef}

\cfclear
\begin{athm}{lemma}{Lemma:ParallelizationElementary}[Architectures of parallelizations of \anns]
Let $ n, L \in \N $,
$ \Phi = ( \Phi_1, \dots, \Phi_n ) \in \ANNs^n $ 
satisfy 
$
  L = \lengthANN(\Phi_1) = \lengthANN(\Phi_2)=\ldots =\lengthANN(\Phi_n)
$
\cfload.
Then 
\begin{enumerate}[(i)]
  \item\llabel{it:1}
  it holds that
  \begin{equation}\label{ParallelizationElementary:Display}
    \parallelizationSpecial_{n}( \Phi )
    \in 
    \bbbpr{\smalltimes_{k = 1}^L\allowbreak\bpr{\R^{(\sum_{j=1}^n  \dimANNlevel_k(\Phi_j)  ) \times (\sum_{j=1}^n \dimANNlevel_{k-1}(\Phi_{j})  )} \times \R^{(\sum_{j=1}^n  \dimANNlevel_k(\Phi_j)  )}}},
  \end{equation}
  \item \llabel{it:2} 
  it holds for all
  $k\in\N_0$
that
\begin{equation}
  \dimANNlevel_k(\parallelizationSpecial_n(\Phi))
  =
  \dimANNlevel_k(\Phi_1)+\dimANNlevel_k(\Phi_2)+\ldots+\dimANNlevel_k(\Phi_n),
\end{equation}
and
\item \llabel{it:3}
 it holds that
\begin{equation}
  \dims\bpr{\parallelizationSpecial_{n}(\Phi)}
  =
  \dims(\Phi_1)+\dims(\Phi_2)+\ldots+\dims(\Phi_n)
\end{equation}
\end{enumerate}
\cfout.
\end{athm}
\begin{aproof}
Note that 
\cref{elementaryPropertiesANN:weightsbiases} in \cref{Lemma:elementaryPropertiesANN}	
and \cref{parallelisationSameLengthDef} 
imply that for all
  $k\in \{1,2,\dots,L\}$
it holds that
\begin{equation}
  \weightANN k{\parallelizationSpecial_n(\Phi)}
  \in
  \R^{(\sum_{j=1}^n  \dimANNlevel_k(\Phi_j)  ) \times (\sum_{j=1}^n \dimANNlevel_{k-1}(\Phi_{j})  )}
  \qquad\text{and}\qquad
  \biasANN k{\parallelizationSpecial_n(\Phi)}
  \in
  \R^{(\sum_{j=1}^n \dimANNlevel_{k-1}(\Phi_{j}) )}
\end{equation}
\cfload.
\Cref{elementaryPropertiesANN:weightsbiases} in \cref{Lemma:elementaryPropertiesANN}
\hence
establishes
\cref{Lemma:ParallelizationElementary.it:1,Lemma:ParallelizationElementary.it:2}.
\Nobs that
\lref{it:2}
implies
\lref{it:3}.
\end{aproof}

\begingroup
\newcommand{\X}[1]{X^{#1}}
\cfclear
\begin{athm}{prop}{Lemma:PropertiesOfParallelizationEqualLength}[Realizations of parallelizations of \anns]
Let $ \activation \in C(\R,\R) $, $ n \in \N $, 
$ \Phi = ( \Phi_1, \allowbreak\dots, \allowbreak \Phi_n ) \in \ANNs^n $
satisfy 
$
  \lengthANN(\Phi_1)=\lengthANN(\Phi_2)=\dots=\lengthANN(\Phi_n)
$
\cfload.
Then
   \begin{enumerate}[label=(\roman{*})]
    \item\label{PropertiesOfParallelizationEqualLength:ItemOne} it holds that 
    \begin{equation}
\functionANN{\activation}(\parallelizationSpecial_{n}(\Phi))\in C\bpr{\R^{[\sum_{j=1}^n \inDimANN(\Phi_j)]},\R^{[\sum_{j=1}^n \outDimANN(\Phi_j)]}}
    \end{equation}
    and
    \item\label{PropertiesOfParallelizationEqualLength:ItemTwo} it holds for all    $x_1\in\R^{\inDimANN(\Phi_1)},x_2\in\R^{\inDimANN(\Phi_2)},\dots, x_n\in\R^{\inDimANN(\Phi_n)}$ that 
      \begin{equation}\label{PropertiesOfParallelizationEqualLengthFunction}
      \begin{split}
      &\bpr{ \functionANN{\activation}\bpr{\parallelizationSpecial_{n}(\Phi)} } (x_1,x_2,\dots, x_n) 
      \\&=\bpr{(\functionANN{\activation}(\Phi_1))(x_1), (\functionANN{\activation}(\Phi_2))(x_2),\dots,
      (\functionANN{\activation}(\Phi_n))(x_n) }\in \R^{[\sum_{j=1}^n \outDimANN(\Phi_j)]}
      \end{split}
      \end{equation}
  \end{enumerate}
  \cfout.
\end{athm}
\begin{aproof}
  Throughout this proof,
  let $L=\lengthANN(\Phi_1)$,
  for every
    $j\in \{1,2,\dots,\allowbreak n\}$
  let
  \begin{multline}
    \X j
    =
    \bigl\{
      x=(x_0,x_1,\dots,x_L)\in\R^{\dimANNlevel_0(\Phi_j)}\times\R^{\dimANNlevel_1(\Phi_j)}\times\dots\times\R^{\dimANNlevel_{L}(\Phi_j)}
      \colon\\
      \bpr{\forall\, k\in \{1,2,\dots,L\}\colon
        x_k
        =
        \multdim_{a\ind{(0,L)}(k)+\id_\R\ind{\{L\}}(k),\dimANNlevel_k(\Phi_j)}(\weightANN k{\Phi_j}x_{k-1}+\biasANN k{\Phi_j})
      }
    \bigr\},
  \end{multline}
  and let
  \begin{multline}
    \mathfrak X
    =
    \bigl\{
      \mathfrak x=(\mathfrak x_0,\mathfrak x_1,\dots,\mathfrak x_L)\in\R^{\dimANNlevel_0(\parallelizationSpecial_n(\Phi))}\times\R^{\dimANNlevel_1(\parallelizationSpecial_n(\Phi))}\times\dots\times\R^{\dimANNlevel_{L}(\parallelizationSpecial_n(\Phi))}
      \colon\\
      \bpr{\forall\, k\in \{1,2,\dots,L\}\colon
        \mathfrak x_k
        =
        \multdim_{a\ind{(0,L)}(k)+\id_\R\ind{\{L\}}(k),\dimANNlevel_k(\parallelizationSpecial_n(\Phi))}(\weightANN k{\parallelizationSpecial_n(\Phi)}\mathfrak x_{k-1}+\biasANN k{\parallelizationSpecial_n(\Phi)})
      }
    \bigr\}.
  \end{multline}
  \Nobs that
    \cref{Lemma:ParallelizationElementary.it:2} in \cref{Lemma:ParallelizationElementary}
    and \cref{elementaryPropertiesANN:inout} in \cref{Lemma:elementaryPropertiesANN}
  imply that
  \begin{equation}
    \inDimANN(\parallelizationSpecial_n(\Phi))
    =
    \dimANNlevel_0(\parallelizationSpecial_n(\Phi))
    =
    \sum_{j=1}^{n}\dimANNlevel_0(\Phi_n)
    =
    \sum_{j=1}^{n}\inDimANN(\Phi_n).
  \end{equation}
  \Moreover
    \cref{Lemma:ParallelizationElementary.it:2} in \cref{Lemma:ParallelizationElementary}
    and \cref{elementaryPropertiesANN:inout} in \cref{Lemma:elementaryPropertiesANN}
  ensure that
  \begin{equation}
    \outDimANN(\parallelizationSpecial_n(\Phi))
    =
    \dimANNlevel_{\lengthANN(\parallelizationSpecial_n(\Phi))}(\parallelizationSpecial_n(\Phi))
    =
    \sum_{j=1}^{n}\dimANNlevel_{\lengthANN(\Phi_n)}(\Phi_n)
    =
    \sum_{j=1}^{n}\outDimANN(\Phi_n)
    .
  \end{equation}
  \Nobs that
    \cref{parallelisationSameLengthDef}
    and \cref{Lemma:ParallelizationElementary.it:2} in \cref{Lemma:ParallelizationElementary}
  show that for all
    $\mathscr a\in C(\R,\R)$,
    $k\in \{1,2,\dots,L\}$,
    $x^1\in\R^{\dimANNlevel_k(\Phi_1)}$,
    $x^2\in\R^{\dimANNlevel_k(\Phi_2)}$,
    $\dots$,
    $x^n\in\R^{\dimANNlevel_k(\Phi_n)}$,
    $\mathfrak x\in \R^{\br{\sum_{j=1}^n \dimANNlevel_k(\Phi_j)}}$
    with $\mathfrak x =(x^1,x^2,\dots,x^n)$
  it holds that
  \begin{equation}
    \begin{split}
      &\multdim_{\mathscr a,\dimANNlevel_k(\parallelizationSpecial_n(\Phi))}(
      \weightANN k{\parallelizationSpecial_n(\Phi)}\mathfrak x+\biasANN k{\parallelizationSpecial_n(\Phi)}
      )
      \\&=
      \multdim_{\mathscr a,\dimANNlevel_k(\parallelizationSpecial_n(\Phi))}\pr*{
      \begin{pmatrix}
        \weightANN k{\Phi_1}&0&0&\cdots&0\\
        0&\weightANN k{\Phi_2}&0&\cdots&0\\
        0&0&\weightANN k{\Phi_3}&\cdots&0\\
        \vdots&\vdots&\vdots&\ddots&\vdots\\
        0&0&0&\cdots&\weightANN k{\Phi_n}
      \end{pmatrix}
      \begin{pmatrix}
        x_1\\x_2\\x_3\\\vdots\\x_n
      \end{pmatrix}
      +
      \begin{pmatrix}
        \biasANN k{\Phi_1}\\\biasANN k{\Phi_2}\\\biasANN k{\Phi_3}\\\vdots\\\biasANN k{\Phi_n}
      \end{pmatrix}
      }
      \\&=
      \multdim_{\mathscr a,\dimANNlevel_k(\parallelizationSpecial_n(\Phi))}\pr*{
      \begin{pmatrix}
        \weightANN k{\Phi_1}x_1+\biasANN k{\Phi_1}\\
        \weightANN k{\Phi_2}x_2+\biasANN k{\Phi_2}\\
        \weightANN k{\Phi_3}x_3+\biasANN k{\Phi_3}\\
        \vdots\\
        \weightANN k{\Phi_n}x_n+\biasANN k{\Phi_n}
      \end{pmatrix}
      }
      =
      \begin{pmatrix}
        \multdim_{\mathscr a,\dimANNlevel_k(\Phi_1)}(\weightANN k{\Phi_1}x_1+\biasANN k{\Phi_1})\\
        \multdim_{\mathscr a,\dimANNlevel_k(\Phi_2)}(\weightANN k{\Phi_2}x_2+\biasANN k{\Phi_2})\\
        \multdim_{\mathscr a,\dimANNlevel_k(\Phi_3)}(\weightANN k{\Phi_3}x_3+\biasANN k{\Phi_3})\\
        \vdots\\
        \multdim_{\mathscr a,\dimANNlevel_k(\Phi_n)}(\weightANN k{\Phi_n}x_n+\biasANN k{\Phi_n})
      \end{pmatrix}
      .
    \end{split}
  \end{equation}
    This
  proves that for all
    $k\in \{1,2,\dots,L\}$,
    $\mathfrak x=(\mathfrak x_0,\mathfrak x_1,\dots,\mathfrak x_L)\in \mathfrak X$,
    $x^1=(x^1_0,x^1_1,\dots,x^1_L)\in \X1$,
    $x^2=(x^2_0,x^2_1,\dots,x^2_L)\in \X2$,
    $\dots$,
    $x^n=(x^n_0,x^n_1,\dots,x^n_L)\in \X n$
    with $\mathfrak x_{k-1}=(x^1_{k-1},x^2_{k-1},\dots,\allowbreak x^n_{k-1})$
  it holds that
  \begin{equation}
    \mathfrak x_k=(x^1_k,x^2_k,\dots,x^n_k)
    .
  \end{equation}
    Induction,
    and \cref{setting_NN:ass2}
    \hence
  demonstrate that for all
    $k\in \{1,2,\dots,L\}$,
    $\mathfrak x=(\mathfrak x_0,\mathfrak x_1,\dots,\mathfrak x_L)\in \mathfrak X$,
    $x^1=(x^1_0,x^1_1,\dots,x^1_L)\in \X1$,
    $x^2=(x^2_0,x^2_1,\dots,x^2_L)\in \X2$,
    $\dots$,
    $x^n=(x^n_0,x^n_1,\dots,x^n_L)\in \X n$
    with $\mathfrak x_0=(x^1_0,x^2_0,\dots,x^n_0)$
  it holds that
  \begin{equation}
    \begin{split}
      \bpr{\functionANN a(\parallelizationSpecial_n(\Phi))}(\mathfrak x_0)
      &=
      \mathfrak x_L
      =
      (x^1_L,x^2_L,\dots,x^n_L)
      \\&=
      \bpr{(\functionANN a(\Phi_1))(x^1_0),(\functionANN a(\Phi_2))(x^2_0),\dots,(\functionANN a(\Phi_n))(x^n_0)}
      .
    \end{split}
  \end{equation}
  This establishes \cref{PropertiesOfParallelizationEqualLength:ItemTwo}.
\end{aproof}
\endgroup

\begin{prop}[Upper bounds for the numbers of parameters of 
parallelizations of \anns]
\label{Lemma:PropertiesOfParallelizationEqualLengthDims}
  Let 
  $n,L\in\N$, 
  $\Phi_1,\Phi_2,\allowbreak\dots,\allowbreak \Phi_n\in\ANNs$ 
  satisfy 
  $L = \lengthANN(\Phi_1)= \lengthANN(\Phi_2)=\ldots =\lengthANN(\Phi_n)$
  (cf.\ \cref{def:ANN}).
  Then
    \begin{equation}
    \paramANN\bpr{\parallelizationSpecial_{n}(\Phi_1,\Phi_2,\allowbreak\dots,\allowbreak \Phi_n)}
    \le 
    \tfrac{1}{2} \bbr{\smallsum\nolimits_{j=1}^n \paramANN(\Phi_j)}^2
    \end{equation}
  (cf.\   \cref{def:simpleParallelization}).
\end{prop}

\begin{proof}[Proof of \cref{Lemma:PropertiesOfParallelizationEqualLengthDims}]	
Throughout this proof, 
for every
$j\in\{1,2,\dots, n\}$, $k \in \{0, 1,\allowbreak \ldots, \allowbreak L \}$
let  
  $l_{j,k} = \dimANNlevel_k(\Phi_j)$.
\Nobs that 
\cref{Lemma:ParallelizationElementary.it:2} in \cref{Lemma:ParallelizationElementary}
demonstrates that
  \begin{equation}
  \begin{split}
  &\paramANN(\parallelizationSpecial_{n}(\Phi_1,\Phi_2,\allowbreak\dots,\allowbreak \Phi_n))
    =\sum_{k=1}^L \bbbr{\smallsum\nolimits_{i=1}^n l_{i,k}} \bbbr{\bpr{\smallsum\nolimits_{i=1}^n l_{i,k-1}}+1}
\\&=\sum_{k=1}^L \bbbr{\smallsum\nolimits_{i=1}^n l_{i,k}} \bbbr{\bpr{\smallsum\nolimits_{j=1}^n l_{j,k-1}}+1}
  \\&\le \sum_{i=1}^n \sum_{j=1}^n\sum_{k=1}^L l_{i,k} (l_{j,k-1}+1)
  \le \sum_{i=1}^n \sum_{j=1}^n\sum_{k,\ell=1}^L l_{i,k} (l_{j,\ell-1}+1)
  \\&= \sum_{i=1}^n \sum_{j=1}^n\bbbr{\smallsum\nolimits_{k=1}^L  l_{i,k}}
  \bbbr{\smallsum\nolimits_{\ell=1}^L  (l_{j,\ell-1}+1)}
  \\&\le \sum_{i=1}^n \sum_{j=1}^n\bbbr{\smallsum\nolimits_{k=1}^L  \tfrac{1}{2}l_{i,k} (l_{i,k-1}+1)}
  \bbbr{\smallsum\nolimits_{\ell=1}^L  l_{j,\ell}(l_{j,\ell-1}+1)}
  \\&=  \sum_{i=1}^n \sum_{j=1}^n \tfrac{1}{2}\paramANN(\Phi_i) \paramANN(\Phi_j)
  =\tfrac{1}{2}\bbbr{\smallsum\nolimits_{i=1}^n \paramANN(\Phi_i)}^2.
  \end{split}
  \end{equation}
  The proof of \cref{Lemma:PropertiesOfParallelizationEqualLengthDims} is thus complete.
\end{proof}

\cfclear
\begin{cor}[Lower and upper bounds for the numbers of parameters of 
parallelizations of \anns]
\label{Lemma:ParallelizationImprovedBoundsOne}
Let   
$ n \in \N $, 
$ \Phi = ( \Phi_1, \dots, \allowbreak \Phi_n ) \in \ANNs^n $ 
satisfy
$ 
  \dims( \Phi_1 ) = \dims( \Phi_2 ) = \ldots = \dims( \Phi_n ) 
$
\cfload.	
Then 
\begin{equation}
\label{eq:ParallelizationImprovedBoundsOne1}
  \textstyle
    \bbr{\tfrac{n^2}2} \paramANN(\Phi_1)\leq \bbr{\tfrac{n^2+n}2} \paramANN(\Phi_1)\leq \paramANN(\parallelizationSpecial_{n}(\Phi))\leq n^2 \paramANN(\Phi_1)
    \leq \tfrac12\bbr{\smallsum_{i=1}^n \paramANN(\Phi_i)}^2
\end{equation}
\cfout.
\end{cor}

\begin{proof}[Proof of \cref{Lemma:ParallelizationImprovedBoundsOne}]	
Throughout this proof, let 
$ L \in \N $,
$ l_0, l_1, \dots, l_L \in \N $ 
satisfy 
\begin{equation}
\label{eq:in_proof_definition_of_short_notation_l0}
  \dims( \Phi_1 ) = ( l_0, l_1, \dots, l_L )
  .
\end{equation}
\Nobs that 
\cref{eq:in_proof_definition_of_short_notation_l0} 
and the assumption that 
$
  \dims( \Phi_1 )
  =
  \dims( \Phi_2 )
  =
  \ldots 
  =
  \dims( \Phi_n )
$
imply that 
for all 
$
  j \in \{ 1, 2, \dots, n \} 
$
it holds that 
\begin{equation}
  \dims( \Phi_j ) = ( l_0, l_1, \dots, \allowbreak l_L ) 
  .
\end{equation}
Combining this with 
\cref{Lemma:ParallelizationElementary.it:3} in \cref{Lemma:ParallelizationElementary}
demonstrates that
\begin{equation}
\label{eq:ParallelizationImprovedBoundsOne3}
  \paramANN( \parallelizationSpecial_{n}( \Phi ) )
  =
  \smallsum\limits_{j=1}^L  (nl_j)\bpr{(nl_{j-1})+1}
  .
\end{equation}
  Hence, we obtain that
\begin{equation}
  \label{eq:ParallelizationImprovedBoundsOne2}
\begin{split}
  \paramANN(\parallelizationSpecial_{n}( \Phi ) )
  &\le 
  \smallsum\limits_{j=1}^L  (nl_j)\bpr{(nl_{j-1})+n}
  = 
  n^2 \bbbbr{\smallsum\limits_{j=1}^L  l_j(l_{j-1}+1)}
  = 
  n^2 \paramANN(\Phi_1)
  .
\end{split}
\end{equation}
\Moreover 
the assumption that $\dims(\Phi_1)=\dims(\Phi_2)=\ldots=\dims(\Phi_n)$
and the fact that $\paramANN(\Phi_1)\geq l_1(l_0+1)\geq 2$
ensure that
\begin{equation}
  \label{eq:ParallelizationImprovedBoundsOne4}
  n^2\paramANN(\Phi_1)
  \leq
  \tfrac{n^2}2\br{\paramANN(\Phi_1)}^2
  =
  \tfrac12\br{n\paramANN(\Phi_1)}^2
  =
  \tfrac12\bbbbr{\smallsum\limits_{i=1}^n\paramANN(\Phi_1)}^2
  =
  \tfrac12\bbbbr{\smallsum\limits_{i=1}^n\paramANN(\Phi_i)}^2
  .
\end{equation}
\Moreover 
\eqref{eq:ParallelizationImprovedBoundsOne3} and
the fact that for all $a,b\in\N$ it holds that 
\begin{equation}
  2(ab+1)=ab+1+(a-1)(b-1)+a+b\geq ab+a+b+1=(a+1)(b+1)
\end{equation}
show that
\begin{equation}
\begin{split}
  \paramANN(\parallelizationSpecial_{n}( \Phi ))
  &\geq
  \tfrac 12\bbbbr{\smallsum\limits_{j=1}^L  (nl_j)(n+1)(l_{j-1}+1)}
  \\&= 
  \tfrac{n(n+1)}2 \bbbbr{\smallsum\limits_{j=1}^L  l_j(l_{j-1}+1)}
  = 
  \bbr{\tfrac{n^2+n}2} \paramANN(\Phi_1)
  .
\end{split}
\end{equation}
  This,
  \eqref{eq:ParallelizationImprovedBoundsOne2}, and
  \eqref{eq:ParallelizationImprovedBoundsOne4}
establish 
  \eqref{eq:ParallelizationImprovedBoundsOne1}.
The proof of \cref{Lemma:ParallelizationImprovedBoundsOne} is thus complete.
\end{proof}

\begin{exercise}{ex:parallelization_better_bound}
  Prove or disprove the following statement:
  For every $n\in\N$, $\Phi=(\Phi_1,\dots, \allowbreak \Phi_n)\in\ANNs^n$
  with $\lengthANN(\Phi_1)=\lengthANN(\Phi_2)=\ldots=\lengthANN(\Phi_n)$ it holds that
  \begin{equation}
  \textstyle 
    \paramANN(\parallelizationSpecial_{n}(\Phi_1,\Phi_2,\dots,\Phi_n))
    \leq
    n\bbr{\sum_{i=1}^n\paramANN(\Phi_i)}
    .
  \end{equation}
\end{exercise}

\begin{exercise}{ex:parallelization_better_bound2}
  Prove or disprove the following statement:
  For every $n\in\N$, $\Phi=(\Phi_1,\dots,\allowbreak \Phi_n)\in\ANNs^n$
  with $\lengthANN(\Phi_1)=\lengthANN(\Phi_2)=\ldots=\lengthANN(\Phi_n)$ it holds that
  \begin{equation}
    \paramANN(\parallelizationSpecial_{n}(\Phi_1,\Phi_2,\dots,\Phi_n))
    \leq
    n^2\paramANN(\Phi_1)
    .
  \end{equation}
\end{exercise}

\subsection{ANN representations for the identities}
\label{subsec:identity}

In this section, we show how multi-dimensional identities can be represented as realizations of \anns\ with \ReLU\ and softplus activation functions.
In \cref{sect:RePU_identity} we will also show such a representation for the \RePU activation function. For this further \ann\ calculus tools will be needed which are introduced later in this chapter.

\begin{adef}{def:ReLu:identity}[\ReLU\ identity \anns]
  We denote by $\idRelu_d\in \ANNs$, $d\in\N$, the fully-connected feedforward \anns\ which satisfy for all $d \in \N$ that 
  \begin{equation}
  \label{eq:def:id:1}
  \idRelu_1 = \pr*{ \pr*{\begin{pmatrix}
  1\\
  -1
  \end{pmatrix}, \begin{pmatrix}
  0\\
  0
  \end{pmatrix} },
  \bbpr{	\begin{pmatrix}
  1& -1
  \end{pmatrix}, 
  0 }
   }  \in \bpr{(\R^{2 \times 1} \times \R^{2}) \times (\R^{1 \times 2} \times \R^1) }
  \end{equation} 
  and 
  \begin{equation}
  \label{eq:def:id:2}
  \idRelu_d = \paraANN{d} (\idRelu_1, \idRelu_1, \ldots, \idRelu_1)
  \end{equation}
  (cf.\ \cref{def:ANN,def:simpleParallelization}).
\end{adef}

\begin{lemma}[Properties of \ReLU\ identity \anns] 
\label{lem:Relu:identity}
Let $d \in \N$. Then
\begin{enumerate}[label=(\roman{*})]
\item
\label{item:lem:Relu:dims} 
it holds that 
\begin{equation}
  \dims(\idRelu_d) = (d, 2d, d) \in \N^3 
\end{equation}
and 
\item
\label{item:lem:Relu:real}
it holds that 
\begin{equation}
  \functionANN{ \rect }( \idRelu_d )
  = 
  \id_{ 
    \R^d
  }
\end{equation} 
\end{enumerate}
(cf.\ \cref{def:ANN,def:ANNrealization,def:ReLu:identity}). 
\end{lemma}

\begin{proof}[Proof of \cref{lem:Relu:identity}]
Throughout this proof, let $L =2$, $l_0 = 1$, $l_1 = 2$, $l_2 =1$. Note that \cref{eq:def:id:1} 
\proves that
\begin{equation}
\dims (\idRelu_1) = (1, 2, 1) = (l_0, l_1, l_2).
\end{equation}
This,
\eqref{eq:def:id:2},
and
\cref{Lemma:ParallelizationElementary}
prove 
that
\begin{equation}
  \dims(\idRelu_d) = (d, 2d, d) \in \N^3.
\end{equation}
This establishes item~\ref{item:lem:Relu:dims}.
Next note that \cref{eq:def:id:1} assures that for all $ x \in \R $ it holds that
\begin{equation}
  ( \functionANN{ \rect }( \idRelu_1 ) )( x ) 
  = \rect( x ) - \rect( - x ) = \max\{ x, 0 \} - \max\{ - x, 0 \} 
  = x .
\end{equation}
Combining this and
\cref{Lemma:PropertiesOfParallelizationEqualLength}
demonstrates that for all $ x = (x_1, \ldots, x_d) \in \R^d$ it holds that $\functionANN{\rect}(\idRelu_d) \in C(\R^d, \R^d)$ and
\begin{equation}
\begin{split}
(\functionANN{\rect}(\idRelu_d))(x) &= \bpr{ \functionANN{\rect} \bpr{\paraANN{d} (\idRelu_1, \idRelu_1, \ldots, \idRelu_1)}}(x_1, x_2, \ldots, x_d)\\
& =  \bpr{ (\functionANN{\rect}(\idRelu_1))(x_1), (\functionANN{\rect}(\idRelu_1))(x_2), \ldots, (\functionANN{\rect}(\idRelu_1))(x_d)}\\
& = (x_1, x_2, \ldots, x_d) = x
\end{split}
\end{equation}
(cf.\ \cref{def:simpleParallelization}).
This establishes \cref{item:lem:Relu:real}.
The proof of \cref{lem:Relu:identity} is thus complete.
\end{proof}

\cfclear
\begin{athm}{lemma}{lem:softplus:identity}[Softplus identity \anns]
Let $d \in \N$ and
let $a$ be the \softplusfunc{} 
\cfload.
Then
\begin{equation}
\functionANN{a}(\idRelu_d) = \id_{ \R^d }
\end{equation}
\cfout.
\end{athm}

\begin{aproof}
\Nobs that
\enum{
  \cref{eq:softplus1.def};
  \cref{eq:def:id:1}
}\prove that
for all $x \in \R$ it holds that
\begin{equation}
\begin{split}
  (\functionANN{a}(\idRelu_1))(x)
&=
  \log(1 + \exp(x + 0)) - \log(1 + \exp(-x + 0)) + 0\\
&=
  \log(1 + \exp(x)) - \log(1 + \exp(-x))\\
&=
  \log\pr*{
    \tfrac{1 + \exp(x)}{1 + \exp(-x)}
  }\\
&=
  \log\pr*{
    \tfrac{\exp(x)(1 + \exp(-x))}{1 + \exp(-x)}
  }\\
&=
  \log\pr*{
    \exp(x)
  }
= 
  x
\end{split}
\end{equation}
\cfload.
Combining this and
\cref{Lemma:PropertiesOfParallelizationEqualLength}
demonstrates that for all $ x = (x_1, \ldots, x_d) \in \R^d$ it holds that $\functionANN{a}(\idRelu_d) \in C(\R^d, \R^d)$ and
\begin{equation}
\begin{split}
(\functionANN{a}(\idRelu_d))(x) &= \bpr{ \functionANN{a} \bpr{\paraANN{d} (\idRelu_1, \idRelu_1, \ldots, \idRelu_1)}}(x_1, x_2, \ldots, x_d)\\
& =  \bpr{ (\functionANN{a}(\idRelu_1))(x_1), (\functionANN{a}(\idRelu_1))(x_2), \ldots, (\functionANN{a}(\idRelu_1))(x_d)}\\
& = (x_1, x_2, \ldots, x_d) = x
\end{split}
\end{equation}
\cfload.
\end{aproof}

\subsection{Extensions of ANNs}

\begin{adef}{def:ANNenlargement}[Extensions of \anns]
Let $ L \in \N $, 
$ \mathbb{I} \in \ANNs $ satisfy 
$ \inDimANN( \mathbb{I} ) = \outDimANN( \mathbb{I} ) $.
Then
we denote by 
\begin{equation}
  \longerANN{ L, \mathbb{I} } 
  \colon 
  \bcu{ 
    \Phi \in \ANNs \colon 
    \bpr{
      \lengthANN( \Phi ) \le L 
      \andShort 
      \outDimANN( \Phi ) 
      = \inDimANN( \mathbb{I} )
    }
  } \to \ANNs 
\end{equation}
the function which satisfies for all 
$ \Phi \in \ANNs $ with $ \lengthANN( \Phi ) \le L $ 
and $ \outDimANN( \Phi ) = \inDimANN( \mathbb{I} ) $ 
that
\begin{equation}
\label{ANNenlargement:Equation}
  \longerANN{ L, \mathbb{I} }( \Phi ) 
  =	
  \compANN{ 
    ( \mathbb{I}^{ \compANNbullet ( L - \lengthANN( \Phi ) ) } )
  }{ \Phi }
\end{equation}
(cf.\ \cref{def:ANN,def:ANNcomposition,def:iteratedANNcomposition}). 
\end{adef}

\cfclear 
\begin{lemma}[Length of extensions of \anns]
\label{Lemma:PropertiesOfANNenlargementGeometry}
Let 
$ d, \hiddenDimId \in \N $, 
$ \Psi \in \ANNs $ satisfy 
$ \dims( \Psi ) = ( d, \hiddenDimId, d ) $ 
\cfload. 
Then
\begin{enumerate}[label=(\roman *)]
\item 
\label{PropertiesOfANNenlargementGeometry:BulletPower} 
it holds for all $ n \in \N_0 $ 
that
$
  \hiddenLength( \powANN{ \Psi }{ n } ) = n
$,
$
  \lengthANN( \powANN{ \Psi }{ n } ) = n + 1 
$,  
$
  \dims( \powANN{ \Psi }{ n } ) \in \N^{ n + 2 } 
$, 
and 
\begin{equation}
\label{BulletPower:Dimensions}
  \dims( 
    \powANN{ \Psi }{ n }
  ) 
  = 
  \begin{cases}
    (d,d) 
  & 
    \colon n = 0
  \\
    (d,\hiddenDimId,\hiddenDimId,\dots,\hiddenDimId,d) 
  &  
    \colon n \in \N
  \end{cases}
\end{equation}
and
\item 
\label{PropertiesOfANNenlargementGeometry:ItemLonger}
it holds for all $ \Phi \in \ANNs $, 
$ L \in \N \cap [ \lengthANN( \Phi ), \infty ) $ 
with 
$ \outDimANN( \Phi ) = d $
that 
\begin{equation}
  \lengthANN(
    \longerANN{ L, \Psi }( \Phi ) 
  )
  = L
\end{equation}
\end{enumerate}
\cfout. 
\end{lemma}

\begin{proof}[Proof of \cref{Lemma:PropertiesOfANNenlargementGeometry}]
Throughout this proof, let $ \Phi \in \ANNs $ 
satisfy $ \outDimANN( \Phi ) = d $. 
\Nobs[Observe] that \cref{lem:number_of_hidden_layers_of_powers_of_ANNs} 
and the fact that 
$
  \hiddenLength( \Psi ) = 1
$
\prove that 
for all $ n \in \N_0 $ it holds that 
\begin{equation}
\label{eq:power_of_shallow_ANNs_dimension_vector_in_proof0}
  \hiddenLength( 
    \powANN{ \Psi }{ n }
  ) 
  = n \hiddenLength( \Psi ) 
  = n 
\end{equation}
\cfload. 
Combining this with 
\cref{def:ANN:eq1} 
and 
\cref{Lemma:elementaryPropertiesANN}
\proves that
\begin{equation}
\label{eq:power_of_shallow_ANNs_dimension_vector_in_proof}
  \hiddenLength( 
    \powANN{ \Psi }{ n }
  ) = n
  ,
\qquad 
  \lengthANN( 
    \powANN{ \Psi }{ n }
  ) = n + 1 
  ,
\qquad 
  \text{and}
\qquad 
  \dims( 
    \powANN{ \Psi }{ n }
  ) \in \N^{ n + 2 } 
  .
\end{equation}
Next we claim that for all $ n \in \N_0 $ it holds that 
\begin{equation}
\label{BulletPower:DimensionsProof}
  \N^{ n + 2 }
  \ni 
  \dims( 
    \powANN{ \Psi }{ n }
  ) 
  = 
  \begin{cases}
    (d,d) 
  & 
    \colon n = 0
  \\
    (d,\hiddenDimId,\hiddenDimId,\dots,\hiddenDimId,d) 
  & 
    \colon n \in \N .
  \end{cases}
\end{equation}
We now prove \cref{BulletPower:DimensionsProof} 
by induction on $ n \in \N_0 $.
Note that the fact that 
\begin{equation}
  \powANN{ \Psi }{ 0 }
  = ( \idMatrix_d, 0 ) \in \R^{ d \times d } \times \R^d
\end{equation}
establishes \cref{BulletPower:DimensionsProof} in the base case $n=0$
\cfload.
For the induction step assume that 
there exists $ n \in \N_0 $ which satisfies  
\begin{equation}
\label{PropertiesOfANNenlargementGeometry:DimensionsInduction}
  \N^{ n + 2 }
  \ni
  \dims( 
    \powANN{ \Psi }{ n }
  ) 
  = 
\begin{cases}
  (d,d) 
&  
  \colon n = 0 
\\
  (d,\hiddenDimId,\hiddenDimId,\dots,\hiddenDimId,d) 
&
  \colon n \in \N .
\end{cases}
\end{equation}
\Nobs that 
\cref{PropertiesOfANNenlargementGeometry:DimensionsInduction}, 
\cref{iteratedANNcomposition:equation}, 
\cref{eq:power_of_shallow_ANNs_dimension_vector_in_proof}, 
\cref{PropertiesOfCompositions:Dims}
in \cref{Lemma:PropertiesOfCompositions},
and the fact that 
$
  \dims(\Psi) = (d, \hiddenDimId, d) \in \N^3
$ 
imply that 
\begin{equation}\label{BulletPower:DimensionsInduction2}
\begin{split}
  \dims(
    \Psi^{\compANNbullet (n+1)}
  )
  =
  \dims(\compANN{\Psi}{(\Psi^{\compANNbullet n})}) = (d,\hiddenDimId,\hiddenDimId,\dots,\hiddenDimId,d)\in \N^{n+3}
\end{split}
\end{equation}
\cfload. 
Induction \hence proves \cref{BulletPower:DimensionsProof}. 
This and
\cref{eq:power_of_shallow_ANNs_dimension_vector_in_proof}
establish \cref{PropertiesOfANNenlargementGeometry:BulletPower}. 
\Nobs that 
\cref{ANNenlargement:Equation}, 
\cref{PropertiesOfCompositions:HiddenLength} 
in 
\cref{Lemma:PropertiesOfCompositions}, 
\cref{eq:power_of_shallow_ANNs_dimension_vector_in_proof0}, 
and the fact that 
$
  \hiddenLength( \Phi ) 
  =
  \lengthANN( \Phi )
  - 1
$
\prove that
for all $ L \in \N \cap [ \lengthANN( \Phi ), \infty ) $ 
it holds that
\begin{equation}
\begin{split}
  \hiddenLength\bpr{
    \longerANN{ L, \Psi }( \Phi )
  }
&
  =
  \hiddenLength\bpr{
    \compANN{ 
      ( \Psi^{ \compANNbullet ( L - \lengthANN( \Phi ) ) } ) 
    }{ \Phi } 
  }
  =
  \hiddenLength\bpr{
    \Psi^{ \compANNbullet ( L - \lengthANN( \Phi ) ) }  
  }
  +
  \hiddenLength\pr{
    \Phi  
  }
\\
&
  =
  ( L - \lengthANN( \Phi ) )
  +
  \hiddenLength\pr{
    \Phi  
  }
  =
  L - 1
  .
\end{split}
\end{equation}
The fact that 
$
  \hiddenLength\bpr{
    \longerANN{ L, \Psi }( \Phi )
  }
  =
  \lengthANN\bpr{
    \longerANN{ L, \Psi }( \Phi )
  }
  - 1
$
\hence \proves that 
\begin{equation}
\label{PropertiesOfANNenlargementGeometryLengthLonger}
\begin{split}
  \lengthANN\bpr{
    \longerANN{ L, \Psi }( \Phi )
  }
  =
  \hiddenLength\bpr{
    \longerANN{ L, \Psi }( \Phi )
  }
  + 1
  =
  L .
\end{split}
\end{equation}
This establishes \cref{PropertiesOfANNenlargementGeometry:ItemLonger}. 
The proof of \cref{Lemma:PropertiesOfANNenlargementGeometry} is thus complete.
\end{proof}

\begin{lemma}[Realizations of extensions of \anns]
\label{Lemma:PropertiesOfANNenlargementRealization}
Let $ \activation \in C( \R, \R ) $, 
$ \mathbb{I} \in \ANNs $ 
satisfy 
$
  \functionANN\activation( \mathbb{I} )
  =
  \id_{ 
    \R^{
      \inDimANN(\mathbb{I})
    } 
  }
$
(cf.\ \cref{def:ANN,def:ANNrealization}).
Then
\begin{enumerate}[label=(\roman *)]
\item \label{PropertiesOfANNenlargementRealization: itemOne} 
it holds for all $ n \in \N_0 $
that
\begin{equation}\label{BulletPowerRealization:Dimensions}
  \functionANN{ \activation }( \mathbb{I}^{ \bullet n } )
  =
  \id_{ 
    \R^{
      \inDimANN(\mathbb{I})
    } 
  }
\end{equation}
and	
\item 	
\label{PropertiesOfANNenlargementRealization:ItemIdentityLonger}
it holds for all $ \Phi \in \ANNs $,
$
  L \in \N \cap 
  [ \lengthANN( \Phi ), \infty ) 
$
with 
$
  \outDimANN( \Phi ) = { \inDimANN( \mathbb{I} ) }
$
that	
\begin{equation}
  \functionANN{ \activation }( 
    \longerANN{ L, \mathbb{I} }( \Phi )
  )
  =
  \functionANN{ \activation }( \Phi )
\end{equation}
\end{enumerate}
(cf.\ \cref{def:iteratedANNcomposition,def:ANNenlargement}).
\end{lemma}

\begin{proof}[Proof of \cref{Lemma:PropertiesOfANNenlargementRealization}]
  Throughout this proof, let  $\Phi\in\ANNs$, $L,d\in\N$ satisfy  $\lengthANN(\Phi)\le L$ and $\inDimANN(\mathbb{I})=\outDimANN(\Phi)=d$.
  We claim that for all $n\in\N_0$ it holds that 
      \begin{equation}\label{BulletPowerRealization:DimensionsProof}
      \functionANN{\activation}(\mathbb{I}^{\bullet n})\in C(\R^{d},\R^{d}) \qandq \forall\, x\in\R^d\colon\,(\functionANN{\activation}(\mathbb{I}^{\bullet n}))(x)=x.
      \end{equation}
    We now prove \cref{BulletPowerRealization:DimensionsProof}  by induction on $n\in\N_0$.
    Note that \cref{iteratedANNcomposition:equation} and the fact that $\outDimANN(\mathbb{I})=d$  demonstrate that
    $\functionANN{\activation}(\mathbb{I}^{\bullet 0})\in C(\R^d,\R^d)$
    and $\forall\,x\in\R^d\colon(\functionANN{\activation}(\mathbb{I}^{\bullet 0}))(x)=x$.
    This 
    establishes \cref{BulletPowerRealization:DimensionsProof} in the base case $n=0$.
    For the induction step observe that for all $n\in\N_0$ with $\functionANN{\activation}(\mathbb{I}^{\bullet n})\in C(\R^d,\R^d)$
    and $\forall\,x\in\R^d\colon(\functionANN{\activation}(\mathbb{I}^{\bullet n}))(x)=x$ it holds that 
    \begin{equation}
      \functionANN{\activation}(\mathbb{I}^{\bullet (n+1)})=\functionANN{\activation}(\compANN{\mathbb{I}}{(\mathbb{I}^{\bullet n})})
      =(\functionANN{\activation}(\mathbb{I}))\circ (\functionANN{\activation}(\mathbb{I}^{\bullet n}))\in C(\R^d,\R^d)
    \end{equation}
    and
    \begin{equation}
    \begin{split}
        \forall\, x\in\R^d\colon\, \bpr{\functionANN{\activation}(\mathbb{I}^{\bullet (n+1)})}(x)&=\bpr{[\functionANN{\activation}(\mathbb{I})]\circ [\functionANN{\activation}(\mathbb{I}^{\bullet n})]}(x)
        \\&=(\functionANN{\activation}(\mathbb{I}))\bpr{\bpr{\functionANN{\activation}(\mathbb{I}^{\bullet n})}(x)}
        =(\functionANN{\activation}(\mathbb{I}))(x)=x.
    \end{split}
    \end{equation}
    Induction \hence proves \cref{BulletPowerRealization:DimensionsProof}. 
    This establishes 
    \cref{PropertiesOfANNenlargementRealization: itemOne}.
  \Nobs \cref{ANNenlargement:Equation},
   item~\ref{PropertiesOfCompositions:Realization} in \cref{Lemma:PropertiesOfCompositions}, item~\ref{PropertiesOfANNenlargementRealization: itemOne},
   and the fact that $\inDimANN(\mathbb{I})=\outDimANN(\Phi)$
    ensure that 
    \begin{equation}
    \begin{split}
          &\functionANN{\activation}(\longerANN{L,\mathbb{I}}(\Phi))=	\functionANN{\activation}( \compANN{(\mathbb{I}^{\bullet (L-\lengthANN(\Phi))})}{\Phi})
          \\&\in C(\R^{\inDimANN(\Phi)},\R^{\outDimANN(\mathbb{I})})=C(\R^{\inDimANN(\Phi)},\R^{\inDimANN(\mathbb{I})})=C(\R^{\inDimANN(\Phi)},\R^{\outDimANN(\Phi)})
    \end{split}
    \end{equation}
and
  \begin{equation}
  \begin{split}
    \forall\,x\in\R^{\inDimANN(\Phi)}\colon\,	\bpr{\functionANN{\activation}(\longerANN{L,\mathbb{I}}(\Phi))}(x)&=\bpr{\functionANN{\activation}(\mathbb{I}^{\bullet (L-\lengthANN(\Phi))})}\bpr{(\functionANN{\activation}(\Phi))(x)}\\&=(\functionANN{\activation}(\Phi))(x).
  \end{split}
  \end{equation}
  This establishes item~\ref{PropertiesOfANNenlargementRealization:ItemIdentityLonger}.
  The proof of \cref{Lemma:PropertiesOfANNenlargementRealization} is thus complete.
\end{proof}

\cfclear 
\begin{lemma}[Architectures of 
extensions of \anns] 
\label{lem:extension_dims}
Let $ d, \mf i, L, \mf L 
\allowbreak \in \N $,
$ l_0, l_1, \dots, l_{ L - 1 } \in \N $,
$
  \Phi, \Psi \in \ANNs 
$
satisfy
\begin{equation}
  \mf L\geq L ,
  \qquad 
  \dims( \Phi ) = (l_0,l_1,\dots,l_{L-1},d) ,
\qandq
  \dims( \Psi ) = (d, \mf i, d ) 
\end{equation}
\cfload. 
Then 
$
  \dims(\longerANN{\mf L,\Psi}(\Phi))\in\N^{\mf L+1}
$
and
\begin{equation}
\label{eq:extension_dims_claim}
  \dims(\longerANN{\mf L,\Psi}(\Phi))
  =
      \begin{cases}
        (l_0,l_1,\dots,l_{L-1},d) & \colon \mf L=L \\
        (l_0,l_1,\dots,l_{L-1},\mf i,\mf i,\dots,\mf i,d) & \colon \mf L>L
    \end{cases}
\end{equation}
\cfout. 
\end{lemma}
\begin{proof}[Proof of \cref{lem:extension_dims}]
\Nobs that 
\cref{PropertiesOfANNenlargementGeometry:BulletPower} 
in \cref{Lemma:PropertiesOfANNenlargementGeometry} 
\proves that 
\begin{equation}
  \hiddenLength( 
    \powANN{ \Psi }{ ( \mf L - L ) }     
  ) 
  ) 
  = \mf L - L 
  ,
\qquad 
  \dims(
    \powANN{ \Psi }{ ( \mf L - L ) }     
  )
  \in 
  \N^{
    \mf L - L + 2
  } 
  ,
\end{equation}
\begin{equation}
  \text{and}
\qquad 
  \dims(
    \powANN{ \Psi }{ ( \mf L - L ) }     
  )
  =
  \begin{cases}
    (d,d) & \colon \mf L=L \\
    (d,\mf i,\mf i,\dots,\mf i,d) & \colon \mf L>L
  \end{cases}
\end{equation}
\cfload. 
Combining this with \cref{Lemma:PropertiesOfCompositions}
\proves that
\begin{equation}
\begin{split}
  \hiddenLength\bpr{
    \compANN{(\Psi^{\bullet(\mf L-L)})}{\Phi}
  }
  =
  \hiddenLength( \Psi^{ \bullet( \mf L - L ) } )
  +
  \hiddenLength( \Phi )
  =
  (
    \mf L - L
  )
  +
  L - 1
  =
  \mf L - 1 
  ,
\end{split}
\end{equation}
\begin{equation}
  \dims(\compANN{(\Psi^{\bullet(\mf L-L)})}{\Phi})\in\N^{\mf L+1}
  ,
\end{equation}
\begin{equation}
  \text{and}
\qquad 
  \dims(\compANN{(\Psi^{\bullet(\mf L-L)})}{\Phi})
  =
  \begin{cases}
    (l_0,l_1,\dots,l_{L-1},d) 
  & 
    \colon \mf L = L 
  \\
    (l_0,l_1,\dots,l_{L-1},\mf i,\mf i,\dots,\mf i,d) 
  & \colon \mf L > L .
  \end{cases}
\end{equation}
This and \cref{ANNenlargement:Equation} establish \cref{eq:extension_dims_claim}. 
The proof of \cref{lem:extension_dims} is thus complete.
\end{proof}

\subsection{Parallelizations of ANNs with different lengths}

\cfclear
\cfconsiderloaded{def:generalParallelization}
\begin{adef}{def:generalParallelization}[Parallelization of \anns\ with different length]
Let $ n \in\N $, 
$ \Psi=( \Psi_1, \dots, \Psi_n ) \in \ANNs^n $ 
satisfy for all $ j \in \{ 1, 2, \dots, n \} $ 
that 
\begin{equation}
  \hiddenLength(\Psi_j) = 1 
\qquad 
  \text{and}
\qquad 
  \inDimANN( \Psi_j ) = \outDimANN( \Psi_j )
\end{equation}
\cfload. 
Then we denote by 
\begin{equation}
  \parallelization_{ n, \Psi }
  \colon 
  \bigl\{
    \Phi = ( \Phi_1, \dots, \Phi_n ) \in \ANNs^n 
    \colon 
    \, \bigl( \forallDist j \in \{ 1, 2, \dots, n \} \colon \outDimANN( \Phi_j ) = \inDimANN( \Psi_j ) \bigr) 
  \bigr\} 
  \to \ANNs
\end{equation}
the function which satisfies for all 
$
  \Phi = ( \Phi_1, \dots, \Phi_n ) \in \ANNs^n 
$ 
with 
$
  \forallDist j \in \{ 1, 2, \dots, n \} \colon \allowbreak \outDimANN( \Phi_j ) = \inDimANN( \Psi_j ) 
$
that
\begin{equation}
\label{generalParallelization:DefinitionFormula}
  \parallelization_{ n, \Psi }( \Phi ) 
  = \parallelizationSpecial_{ n }\bpr{
    \longerANN{ 
      \max_{ k \in \{ 1, 2, \dots, n \} }
      \lengthANN( \Phi_k ), 
      \Psi_1 }({\Phi_1}),\dots,\longerANN{\max_{k\in\{1,2,\dots,n\}}\lengthANN(\Phi_k),\Psi_n}({\Phi_n})
  }
\end{equation}
\cfadd{Lemma:PropertiesOfANNenlargementGeometry}
\cfout. 
\end{adef}

\cfclear
\begin{lemma}[Realizations for parallelizations of \anns\ with different length] 
\label{Lemma:PropertiesOfParallelizationRealization}
Let $ \activation\in C( \R, \R ) $, $ n \in \N $, 
$
  \mathbb{I} = ( \idANNshort{1}, \dots, \idANNshort{n} ) 
$, 
$
  \Phi = ( \Phi_1, \dots, \allowbreak \Phi_n ) \in \ANNs^n 
$
satisfy for all 
$ j \in \{ 1, 2,\dots, n \} $, 
$ x \in\R^{\outDimANN(\Phi_j)}$ that $\hiddenLength(\idANNshort{j}) =1$, 
$ \inDimANN(\mathbb{I}_j)=\outDimANN(\mathbb{I}_j)=\outDimANN(\Phi_j)$, 
and
$(\functionANN{\activation}(\idANNshort{j}))(x)=x$
(cf.\ \cref{def:ANN,def:ANNrealization}).
Then
   \begin{enumerate}[label=(\roman{*})]
    \item \label{PropertiesOfParallelizationRealization:ItemOne} it holds that
    \begin{equation}
      \functionANN{\activation}\bpr{\parallelization_{n,\mathbb{I}}(\Phi)}\in C\bpr{\R^{[\sum_{j=1}^n \inDimANN(\Phi_j)]},\R^{[\sum_{j=1}^n \outDimANN(\Phi_j)]}}
    \end{equation}
    and
    \item \label{PropertiesOfParallelizationRealization:ItemTwo}
    it holds for all $x_1\in\R^{\inDimANN(\Phi_1)},x_2\in\R^{\inDimANN(\Phi_2)},\dots, x_n\in\R^{\inDimANN(\Phi_n)}$ that 
    \begin{equation}\label{PropertiesOfParallelizationRealizationEqualLengthFunction}
    \begin{split}
    &\bpr{ \functionANN{\activation}(\parallelization_{n,\mathbb{I}}(\Phi)) } (x_1,x_2,\dots, x_n) 
    \\&=\bpr{(\functionANN{\activation}(\Phi_1))(x_1), (\functionANN{\activation}(\Phi_2))(x_2),\dots,
    (\functionANN{\activation}(\Phi_n))(x_n) }\in \R^{[\sum_{j=1}^n \outDimANN(\Phi_j)]}
    \end{split}
    \end{equation}
  \end{enumerate}
(cf.\ \cref{def:generalParallelization}).
\end{lemma}

\begin{proof}[Proof of \cref{Lemma:PropertiesOfParallelizationRealization}]
  Throughout this proof, let $L\in\N$ satisfy  $L=	\allowbreak\max_{j\in\{1,2,\dots,n\}} \allowbreak\lengthANN(\Phi_j)$.
Note that item~\ref{PropertiesOfANNenlargementGeometry:ItemLonger} in \cref{Lemma:PropertiesOfANNenlargementGeometry}, the assumption that
for all $j\in\{1,2,\dots, n\}$ it holds that $\hiddenLength(\idANNshort{j}) =1$, \cref{ANNenlargement:Equation}, \cref{PropertiesOfCompositions:LengthDisplay}, and item~\ref{PropertiesOfANNenlargementRealization:ItemIdentityLonger} in \cref{Lemma:PropertiesOfANNenlargementRealization} demonstrate 
\begin{enumerate}[label=(\Roman{*})]
  \item that for all $j\in\{1,2,\dots, n\}$ it holds that $\lengthANN(\longerANN{L,\mathbb{I}_j}(\Phi_j))=L$ and
   $\functionANN{\activation}(\longerANN{L,\mathbb{I}_j}(\Phi_j))\in C(\R^{\inDimANN(\Phi_j)},\R^{\outDimANN(\Phi_j)})$
   and
   \item that for all
    $j\in\{1,2,\dots, n\}$, $x\in\R^{\inDimANN(\Phi_j)}$	it holds that
   \begin{equation}
   \bpr{\functionANN{\activation}(\longerANN{L,\mathbb{I}_j}(\Phi_j))}(x)=(\functionANN{\activation}(\Phi_j))(x)
   \end{equation}
\end{enumerate}
  (cf.\  \cref{def:ANNenlargement}).
  \Cref{PropertiesOfParallelizationEqualLength:ItemOne,PropertiesOfParallelizationEqualLength:ItemTwo} in	
   \cref{Lemma:PropertiesOfParallelizationEqualLength} therefore imply
   \begin{enumerate}[label=(\Alph{*})]
     \item that 	 
     \begin{equation}
     \functionANN{\activation}\bpr{\parallelizationSpecial_n\bpr{\longerANN{L,\mathbb{I}_1}({\Phi_1}),\longerANN{L,\mathbb{I}_2}({\Phi_2}),\dots,\longerANN{L,\mathbb{I}_n}({\Phi_n})}}
     \in C\bpr{\R^{[\sum_{j=1}^n \inDimANN(\Phi_j)]},\R^{[\sum_{j=1}^n \outDimANN(\Phi_j)]}}
     \end{equation}
     and 
     \item that for all $x_1\in\R^{\inDimANN(\Phi_1)},x_2\in\R^{\inDimANN(\Phi_2)},\dots, x_n\in\R^{\inDimANN(\Phi_n)}$ it holds that
          \begin{equation}
          \begin{split}
          &\bpr{\functionANN{\activation}\bpr{\parallelizationSpecial_n\bpr{\longerANN{L,\mathbb{I}_1}({\Phi_1}),\longerANN{L,\mathbb{I}_2}({\Phi_2}),\dots,\longerANN{L,\mathbb{I}_n}({\Phi_n})}}}(x_1,x_2,\dots, x_n)
          \\&=\bbpr{\bpr{\functionANN{\activation}\bpr{\longerANN{L,\mathbb{I}_1}({\Phi_1})}}(x_1), \bpr{\functionANN{\activation}\bpr{\longerANN{L,\mathbb{I}_2}({\Phi_2})}}(x_2),\dots,
          \bpr{\functionANN{\activation}\bpr{\longerANN{L,\mathbb{I}_n}({\Phi_n})}}(x_n) }
          \\&=\bbpr{(\functionANN{\activation}(\Phi_1))(x_1),(\functionANN{\activation}(\Phi_2))(x_2),\dots, (\functionANN{\activation}(\Phi_n))(x_n)}
          \end{split}
          \end{equation}
   \end{enumerate}
   (cf.\ \cref{def:simpleParallelization}).
   Combining this with  \cref{generalParallelization:DefinitionFormula} and the fact that $L=\allowbreak\max_{j\in\{1,2,\dots,n\}} \allowbreak\lengthANN(\Phi_j)$ ensures
   \begin{enumerate}
     \item[(C)] that 
        \begin{equation}
        \functionANN{\activation}\bpr{\parallelization_{n,\mathbb{I}}(\Phi)}\in C\bpr{\R^{[\sum_{j=1}^n \inDimANN(\Phi_j)]},\R^{[\sum_{j=1}^n \outDimANN(\Phi_j)]}}
        \end{equation}
        and
     \item[(D)] 	 that for all $x_1\in\R^{\inDimANN(\Phi_1)},x_2\in\R^{\inDimANN(\Phi_2)},\dots,\allowbreak x_n\in\R^{\inDimANN(\Phi_n)}$ it holds that
       \begin{equation}
       \begin{split}
       &\bpr{ \functionANN{\activation}\bpr{\parallelization_{n,\mathbb{I}}(\Phi)} } (x_1,x_2,\dots,x_n) 
       \\&=\bpr{\functionANN{\activation}\bpr{\parallelizationSpecial_n\bpr{\longerANN{L,\mathbb{I}_1}({\Phi_1}),\longerANN{L,\mathbb{I}_2}({\Phi_2}),\dots,\longerANN{L,\mathbb{I}_n}({\Phi_n})}}}(x_1,x_2,\dots,x_n)
       \\&=\bbpr{(\functionANN{\activation}(\Phi_1))(x_1),(\functionANN{\activation}(\Phi_2))(x_2),\dots, (\functionANN{\activation}(\Phi_n))(x_n)}.
       \end{split}
       \end{equation}
   \end{enumerate}
This establishes items \cref{PropertiesOfParallelizationRealization:ItemOne,PropertiesOfParallelizationRealization:ItemTwo}.
  The proof of \cref{Lemma:PropertiesOfParallelizationRealization} is thus complete.
\end{proof}

\cfclear
\begin{exercise}{ex:represent_max_function_exactly}
For every
  $d \in \N$
let 
$F_d \colon \R^d \to \R^d$ satisfy for all
  $x = (x_1, \ldots, x_d) \in \R^d$ 
that
\begin{equation}
\begin{split} 
  F_d(x)
=
  (\max\{|x_1|\}, \max\{|x_1|, |x_2|\}, \ldots, \max\{|x_1|, |x_2|, \ldots, |x_d|\}).
\end{split}
\end{equation}
Prove or disprove the following statement:
For all 
  $d \in \N$
there exists $\Phi \in \ANNs$ such that
\begin{equation}
\begin{split} 
  \functionANN{\rect}(\Phi)
=
  F_d
\end{split}
\end{equation}
\cfload.
\end{exercise}

\section{Scalar multiplications of ANNs}
\label{subsec:linear}

\subsection{Affine transformations as ANNs}

\begin{adef}{def:ANN:affine}[Affine transformation \anns]
Let $ m, n \in \N $, $ \weight \in \R^{ m \times n } $, $ \bias \in \R^m $. 
Then we denote by 
\begin{equation}
  \AffineANN_{ \weight, \bias } \in ( \R^{ m \times n } \times \R^m ) \subseteq \ANNs
\end{equation}
the fully-connected feedforward \ann\ given by 
\begin{equation}
  \AffineANN_{ \weight, \bias } = ( \weight, \bias )
\end{equation}
(cf.~\cref{def:ANN,def:neuralnetwork}).
\end{adef}

\begin{athm}{lemma}{lem:ANN:affine}[Realizations of 
affine transformation of \anns]
Let $m,n\in\N$, $\weight\in\R^{m\times n}$, $\bias\in\R^m$. Then
\begin{enumerate}[label=(\roman *)]
\item
\label{lem:ANN:affine:item1}
it holds that $\dims(\AffineANN_{\weight,\bias}) = (n,m) \in \N^2$,
\item
\label{lem:ANN:affine:item2}
it holds 
for all $a \in C(\R,\R)$ 
that $\functionANN{\activation}(\AffineANN_{\weight,\bias}) \in C(\R^n,\R^m)$,
and
\item
\label{lem:ANN:affine:item3}
it holds 
for all 
$a\in C(\R,\R)$, 
$x\in\R^n$ that
\begin{equation}
  (\functionANN{\activation}(\AffineANN_{\weight,\bias}))(x) = \weight\! x + \bias
\end{equation}
\end{enumerate}
(cf.~\cref{def:ANN,def:ANNrealization,def:ANN:affine}).
\end{athm}
\begin{aproof}
\Nobs that 
  the fact that 
    $\AffineANN_{\weight,\bias} \in (\R^{m\times n}\times \R^m)\subseteq \ANNs$ 
\proves that 
\begin{equation}
  \dims(\AffineANN_{\weight,\bias}) = (n,m) \in \N^2
  .
\end{equation}
This \proves[ep] 
  \cref{lem:ANN:affine:item1}.
\Moreover 
  the fact that 
  \begin{equation}
    \AffineANN_{\weight,\bias} = (\weight,\bias) \in (\R^{m\times n}\times \R^m)
  \end{equation}
  and \cref{setting_NN:ass2} 
\prove that for all 
  $a\in C(\R,\R)$, $x\in\R^n$ 
it holds that 
  $\functionANN{\activation}(\AffineANN_{\weight,\bias}) \in C(\R^n,\R^m)$ 
and 
\begin{equation}
  (\functionANN{\activation}(\AffineANN_{\weight,\bias}))(x) 
  = 
  \weight\! x + \bias
  .
\end{equation}
  This 
\proves[ep]
  \cref{lem:ANN:affine:item2,lem:ANN:affine:item3}.
\end{aproof}

\cfclear
\begin{athm}{lemma}{lem:ANN:affine2}[Compositions with 
affine transformation \anns]
Let 
$\Phi\in\ANNs$ \cfload. Then
\begin{enumerate}[label=(\roman *)]
\item
\label{lem:ANN:affine2:item1}
it holds for all 
  $m\in\N$, 
  $\weight \in \R^{m\times\outDimANN(\Phi)}$, 
  $\bias\in\R^{m}$ 
that 
\begin{equation}
  \dims(\AffineANN_{\weight,\bias}\compANNbullet\Phi) 
  = 
  (\dimANNlevel_0(\Phi), \dimANNlevel_1(\Phi), \ldots, \dimANNlevel_{\hiddenLength(\Phi)}(\Phi), m)
  ,
\end{equation}
\item
\label{lem:ANN:affine2:item2}
it holds for all 
  $a\in C(\R,\R)$, 
  $m\in\N$, 
  $\weight \in \R^{m\times\outDimANN(\Phi)}$, 
  $\bias\in\R^{m}$ 
that 
  $\functionANN{\activation}(\AffineANN_{\weight,\bias}\compANNbullet \Phi) \in C(\R^{\inDimANN(\Phi)},\R^m)$,
\item
\label{lem:ANN:affine2:item3}
it holds for all 
  $a\in C(\R,\R)$, 
  $m\in\N$, 
  $\weight \in \R^{m\times\outDimANN(\Phi)}$, 
  $\bias\in\R^{m}$, 
  $x\in \R^{\inDimANN(\Phi)}$ 
that
\begin{equation}
  (\functionANN{\activation}(\AffineANN_{\weight,\bias}\compANNbullet \Phi))(x) 
  = 
  \weight\pr*{(\functionANN{\activation}(\Phi))(x)} + \bias
  ,
\end{equation}
\item
\label{lem:ANN:affine2:item4}
it holds for all 
  $n\in\N$, 
  $\weight\in\R^{\inDimANN(\Phi)\times n}$, 
  $\bias\in\R^{\inDimANN(\Phi)}$ 
that 
\begin{equation}
  \dims(\Phi\compANNbullet \AffineANN_{\weight,\bias}) 
  = 
  (n, \dimANNlevel_1(\Phi), \dimANNlevel_2(\Phi), \ldots, \dimANNlevel_{\lengthANN(\Phi)}(\Phi))
  ,
\end{equation}
\item 
\label{lem:ANN:affine2:item5}
it holds for all 
  $a\in C(\R,\R)$, 
  $n\in\N$, 
  $\weight\in\R^{\inDimANN(\Phi)\times n}$, 
  $\bias\in\R^{\inDimANN(\Phi)}$ 
that 
  $\functionANN{\activation}(\Phi\compANNbullet\AffineANN_{\weight,\bias}) \in C(\R^n,\R^{\outDimANN(\Phi)})$,
and
\item
\label{lem:ANN:affine2:item6}
it holds for all 
  $a\in C(\R,\R)$, 
  $n\in\N$, 
  $\weight\in\R^{\inDimANN(\Phi)\times n}$, 
  $\bias\in\R^{\inDimANN(\Phi)}$, 
  $x\in\R^n$ 
that
\begin{equation}
  (\functionANN{\activation}(\Phi\compANNbullet\AffineANN_{\weight,\bias}))(x) 
  = 
  (\functionANN{\activation}(\Phi))(\weight\! x+\bias)
\end{equation}
\end{enumerate}
\cfout.
\end{athm}

\begin{aproof}
\Nobs that 
  \cref{lem:ANN:affine} 
\proves that for all 
  $m,n\in\N$, 
  $\weight \in \R^{m\times n}$, 
  $\bias\in \R^m$, 
  $a\in C(\R,\R)$, 
  $x\in\R^n$ 
it holds that 
  $\functionANN{\activation}(\AffineANN_{\weight,\bias}) \in C(\R^n,\R^m)$ 
  and
\begin{equation}\label{linear_props2_1}
  (\functionANN{\activation}(\AffineANN_{\weight,\bias}))(x) 
  = 
  \weight\!x + \bias
\end{equation}
\cfload.
Combining 
  this 
and 
  \cref{Lemma:PropertiesOfCompositions}
\proves[ep]
  \cref{lem:ANN:affine2:item1,lem:ANN:affine2:item2,lem:ANN:affine2:item3,lem:ANN:affine2:item4,lem:ANN:affine2:item5,lem:ANN:affine2:item6}.
\end{aproof}

\subsection{Scalar multiplications of ANNs}

\cfclear 
\cfconsiderloaded{def:ANNscalar}
\begin{adef}{def:ANNscalar}[Scalar multiplications of \anns]
We denote by $\scalarMultANN{\pr*{\cdot}}{\pr*{\cdot}} \colon \R \times \ANNs \to \ANNs$ the function which satisfies for all $\lambda \in \R$, $\Phi \in \ANNs$ that
\begin{equation}
\label{def:ANNscalar:eq1}
\scalarMultANN\lambda\Phi = \compANN{ \AffineANN_{\lambda \idMatrix_{\outDimANN(\Phi)},0} }{\Phi}
\end{equation}
\cfout. 
\end{adef}

\cfclear
\begin{athm}{lemma}{lem:ANNscalar}
Let 
  $\lambda \in \R$, 
  $\Phi \in \ANNs$ \cfload. 
Then
\begin{enumerate}[label=(\roman{*})]
  \item\label{item:ANN:scalar:1} 
  it holds that 
    $\dims(\scalarMultANN\lambda\Phi ) = \dims(\Phi)$,
  \item\label{item:ANN:scalar:2} 
  it holds for all 
    $a \in C(\R, \R)$ 
  that 
    $\functionANN{\activation}(\scalarMultANN\lambda\Phi) \in C(\R^{\inDimANN(\Phi)}, \R^{\outDimANN(\Phi)})$,
  and 
  \item\label{item:ANN:scalar:3} 
  it holds for all 
    $a \in C(\R, \R)$, 
    $x \in \R^{\inDimANN(\Phi)}$
  that
  \begin{equation}
      \functionANN{\activation}(\scalarMultANN\lambda\Phi)
    = 
    \lambda 
    \bbr{ 
        \functionANN{\activation}(\Phi)
    }
  \end{equation}
\end{enumerate}
\cfout.
\end{athm}
\begin{aproof}
Throughout this proof, let 
  $L \in \N$, 
  $l_0, l_1, \ldots, l_L \in \N$ 
satisfy  
\begin{equation}
  L = \lengthANN(\Phi)
  \qandq
  (l_0, l_1, \ldots, l_L) = \dims(\Phi)
  .
\end{equation} 
\Nobs that 
  \cref{lem:ANN:affine:item1} in \cref{lem:ANN:affine} 
\proves that
\begin{equation}
  \dims(\AffineANN_{\lambda \idMatrix_{\outDimANN(\Phi)},0}) 
  =  
  (\outDimANN(\Phi), \outDimANN(\Phi))
\end{equation}
\cfload.
Combining 
  this and
  \cref{lem:ANN:affine2:item1} in \cref{lem:ANN:affine2}
\proves that
\begin{equation}
  \dims(\scalarMultANN\lambda\Phi ) 
  = 
  \dims (\compANN{\AffineANN_{\lambda \idMatrix_{\outDimANN(\Phi)},0}}{\Phi}) 
  = 
  (l_0, l_1, \ldots, l_{L-1}, \outDimANN(\Phi)) 
  = 
  \dims (\Phi)
\end{equation}
\cfload.
  This 
\proves[ep]
  \cref{item:ANN:scalar:1}. 
\Nobs that
  \cref{lem:ANN:affine2:item2,lem:ANN:affine2:item3} in \cref{lem:ANN:affine2} 
\prove that for all 
  $a \in C(\R, \R)$, 
  $x \in \R^{\inDimANN(\Phi)}$ 
it holds that 
  $\functionANN{\activation}(\scalarMultANN\lambda\Phi) \in C(\R^{\inDimANN(\Phi)}, \R^{\outDimANN(\Phi)})$ and
\begin{equation}
\begin{split}
  \bpr{\functionANN{\activation}(\scalarMultANN\lambda\Phi)}(x) 
  &= 
  \bpr{\functionANN{\activation} (\compANN{\AffineANN_{\lambda \idMatrix_{\outDimANN(\Phi)},0}}{\Phi})}(x)
  \\&= 
  \lambda  \idMatrix_{\outDimANN(\Phi)}  \bpr{ (\functionANN{\activation} (\Phi))(x) } 
  \\&= 
  \lambda \bpr{ (\functionANN{\activation}(\Phi))(x) }
\end{split}
\end{equation} 
\cfload.
  This 
\proves[ep]
  \cref{item:ANN:scalar:2,item:ANN:scalar:3}.
\end{aproof}

\section{Sums of ANNs with the same length}
\label{subsec:sums}

\todoc{Maybe use another symbol for the identity ANN as the current one is hard to read}

\subsection{Sums of vectors as ANNs}

\cfclear 
\cfconsiderloaded{def:ANN:sum}
\begin{adef}{def:ANN:sum}[Sums of vectors as \anns]
Let 
  $ m, n \in \N $. 
Then we denote by 
\begin{equation}
  \sumANN_{ m, n } \in ( \R^{ m \times ( m n ) } \times \R^m ) \subseteq \ANNs
\end{equation}
the \ann\ \cfadd{def:neuralnetwork}given by
\begin{equation}
\label{eq:ANN:sum}
  \sumANN_{m, n} = 
  \AffineANN_{(\idMatrix_m \,\,\,  \idMatrix_m \,\,\, \ldots \,\,\, \idMatrix_m),0}
\end{equation}
\cfout. 
\end{adef}

\cfclear 
\begin{athm}{lemma}{lem:def:ANNsum}
Let 
  $ m, n \in \N $. 
Then
\begin{enumerate}[label=(\roman{*})]
  \item\label{item:ANNsum:2} 
  it holds that 
    $\dims (\sumANN_{m, n}) = (mn, m) \in \N^2$,
  \item\label{item:ANNsum:3} 
  it holds for all  
    $a \in C(\R, \R)$ 
  that 
    $\functionANN{\activation}(\sumANN_{m, n}) \in C(\R^{mn}, \R^m)$, 
  and
  \item\label{item:ANNsum:4} 
  it holds for all 
    $a \in C(\R, \R)$, 
    $x_1, x_2, \ldots, x_n \in \R^{m}$ 
  that 
  \begin{equation}
    (\functionANN{\activation}(\sumANN_{m, n})) (x_1, x_2, \ldots, x_n) 
    = 
    \smallsum_{k=1}^n x_k
  \end{equation}
\end{enumerate}
\cfout.
\end{athm}
\begin{aproof}
  \Nobs that 
    the fact that 
      $\sumANN_{m, n} \in (\R^{m \times (mn)} \times \R^m)$ 
  \proves that 
  \begin{equation}
    \dims (\sumANN_{m, n}) = (mn, m) \in \N^2
  \end{equation}
  \cfload.
    This 
  \proves[ep]
    \cref{item:ANNsum:2}. 
  \Nobs that
    \cref{lem:ANN:affine:item2,lem:ANN:affine:item3} in \cref{lem:ANN:affine} 
  \prove that for all 
    $a \in C(\R, \R)$, 
    $x_1, x_2, \ldots, x_n \in \R^{m}$  
  it holds that 
    $\functionANN{\activation}(\sumANN_{m, n}) \in C(\R^{mn}, \R^m)$ 
  and
  \begin{equation}
  \begin{split}
    (\functionANN{\activation}(\sumANN_{m, n})) (x_1, x_2, \ldots, x_n) 
    &=
    \bpr{ \functionANN{\activation} \bpr{ \AffineANN_{(\idMatrix_m \,\,\,  \idMatrix_m \,\,\, \ldots \,\,\, \idMatrix_m),0} }}(x_1, x_2, \ldots, x_n)
    \\&= 
    (\idMatrix_m \,\,\,  \idMatrix_m \,\,\, \ldots \,\,\, \idMatrix_m) (x_1, x_2, \ldots, x_n) 
    =
    \smallsum_{k=1}^n x_k
  \end{split}
  \end{equation}
  \cfload.
    This 
  \proves[ep]
    \cref{item:ANNsum:3,item:ANNsum:4}.
\end{aproof}

\cfclear
\begin{athm}{lemma}{lem:def:ANNsum:comp:left}
  Let 
    $m, n \in \N$, 
    $a \in C(\R, \R)$,  
    $\Phi  \in \ANNs$ 
  satisfy 
    $ \outDimANN(\Phi) = mn$ \cfload. 
  Then
  \begin{enumerate}[label=(\roman{*})]
  \item\label{item:ANNsum:comp:left:1} 
  it holds that
    $\functionANN{\activation}(\compANN{\sumANN_{m, n}}{\Phi}) \in C(\R^{\inDimANN(\Phi)}, \R^m) $ 
  and
  \item\label{item:ANNsum:comp:left:2}
  it holds for all  
    $x \in \R^{\inDimANN(\Phi)}$, 
    $y_1, y_2, \ldots, y_n \in \R^{m}$ 
    with 
      $(\functionANN{\activation} (\Phi ))(x) = (y_1, y_2, \ldots, y_n)$ 
  that 
  \begin{equation}
    \bpr{ \functionANN{\activation}(\compANN{\sumANN_{m, n}}{\Phi}) }(x) 
    = 
    \smallsum_{k=1}^n y_k
  \end{equation}
  \end{enumerate}
  \cfout.
\end{athm}
\begin{aproof}
  \Nobs that 
    \cref{lem:def:ANNsum} 
  \proves that for all  
    $x_1, x_2, \ldots, x_n \in \R^{m}$  
  it holds that 
    $\functionANN{\activation}(\sumANN_{m, n}) \in C(\R^{mn}, \R^m)$ and
  \begin{equation}
  \begin{split}
    (\functionANN{\activation}(\sumANN_{m, n})) (x_1, x_2, \ldots, x_n)  
    =  
    \smallsum_{k=1}^n x_k
  \end{split}
  \end{equation}
  \cfload.
  Combining 
    this and
    \cref{PropertiesOfCompositions:Realization}  in \cref{Lemma:PropertiesOfCompositions}
  \proves[ep] 
    \cref{item:ANNsum:comp:left:1,item:ANNsum:comp:left:2}.
\end{aproof}

\cfclear
\begin{athm}{lemma}{lem:def:ANNsum:comp:right}
  Let
    $n \in \N$, 
    $a \in C(\R, \R)$, 
    $\Phi \in \ANNs$ 
  \cfload. 
  Then
  \begin{enumerate}[label=(\roman{*})]
  \item \label{item:ANNsum:comp:right:1} 
  it holds that 
    $\functionANN{\activation}(\compANN{\Phi}{\sumANN_{\inDimANN(\Phi), n}}) \in C(\R^{n \inDimANN(\Phi)}, \R^{\outDimANN(\Phi)}) $ and
  \item \label{item:ANNsum:comp:right:2} 
  it holds for all
    $x_1, x_2, \ldots, x_n \in \R^{\inDimANN(\Phi)}$ 
  that  
  \begin{equation}
    \bpr{\functionANN{\activation}(\compANN{\Phi}{\sumANN_{\inDimANN(\Phi), n}}) }(x_1, x_2, \ldots, x_n)
    = 
    (\functionANN{\activation}(\Phi))\pr*{\smallsum_{k=1}^n x_k}
  \end{equation}
  \end{enumerate}
  \cfout.
\end{athm}
\begin{aproof}
  \Nobs that 
    \cref{lem:def:ANNsum} 
  \proves that for all 
    $m \in \N$, 
    $x_1, x_2, \ldots, x_n \in \R^{m}$  
  it holds that
    $\functionANN{\activation}(\sumANN_{m, n}) \in C(\R^{mn}, \R^m)$ 
  and
  \begin{equation}
  \begin{split}
    (\functionANN{\activation}(\sumANN_{m, n})) (x_1, x_2, \ldots, x_n)
    =  
    \smallsum_{k=1}^n x_k
  \end{split}
  \end{equation}
  \cfload.
  Combining 
    this and
    \cref{PropertiesOfCompositions:Realization} in \cref{Lemma:PropertiesOfCompositions}	
  \proves[ep] 
    \cref{item:ANNsum:comp:right:1,item:ANNsum:comp:right:2}.
\end{aproof}

\subsection{Concatenation of vectors as ANNs}

\begin{adef}{def:Transpose}[Transpose of a matrix]
Let 
  $ m, n \in \N $, 
  $ A \in \R^{ m \times n } $. 
Then we denote by 
  $ \transpose{A} \in \R^{ n \times m } $ 
the transpose of $ A $.
\end{adef}

\cfclear
\begin{adef}{def:ANN:extension}[Concatenation of vectors as \anns]
\cfconsiderloaded{def:ANN:extension}
Let 
  $ m, n \in \N $. 
Then we denote by 
\begin{equation}
  \extensionANN_{m, n} \in 
  ( \R^{ ( m n ) \times m } \times \R^{ m n } ) \subseteq \ANNs 
\end{equation}
the fully-connected feedforward \ann\ \cfadd{def:neuralnetwork}given by
\begin{equation}\label{eq:ANN:extension}
  \extensionANN_{ m, n } = 
  \AffineANN_{\scriptstyle\transpose{(\idMatrix_m \,\,\,  \idMatrix_m \,\,\, \ldots \,\,\, \idMatrix_m)},0}
\end{equation}
\cfload. 
\end{adef}

\cfclear
\begin{athm}{lemma}{lem:ANN:extension}
  Let 
    $m, n \in \N$. 
  Then
  \begin{enumerate}[label=(\roman{*})]
  \item
  \label{item:ANN:extension:2} 
  it holds that 
    $	\dims (\extensionANN_{m, n}) = (m, mn) \in \N^2$,
  \item
  \label{item:ANN:extension:3} 
  it holds for all 
    $a \in C(\R, \R)$ 
  that 
    $\functionANN{\activation}(\extensionANN_{m, n}) \in C(\R^{m}, \R^{mn})$, 
  and
  \item
  \label{item:ANN:extension:4} 
  it holds for all 
    $a \in C(\R, \R)$, 
    $x \in \R^m$ 
  that 
  \begin{equation}
    (\functionANN{\activation}(\extensionANN_{m, n})) (x) 
    = 
    (x, x, \ldots, x)
    \end{equation}
  \end{enumerate}
  \cfout.
\end{athm}
\begin{aproof}
  \Nobs that 
    the fact that 
      $ \extensionANN_{m, n} \in ( \R^{ ( m n ) \times m } \times \R^{ m n } ) $ 
  \proves that 
  \begin{equation}
    \dims (\extensionANN_{m, n}) 
    = 
    (m, mn) \in \N^2
  \end{equation}
  \cfload.
    This 
  \proves[ep]
    \cref{item:ANN:extension:2}. 
  \Nobs that
    \cref{lem:ANN:affine:item3} in \cref{lem:ANN:affine} 
  \proves that for all 
    $a \in C(\R, \R)$, $x  \in \R^m$
  it holds that
    $\functionANN{\activation}(\extensionANN_{m, n}) \in C(\R^{m}, \R^{mn})$ and
  \begin{equation}
  \begin{split}
    (\functionANN{\activation}(\extensionANN_{m, n})) (x) 
    &= 
    \bpr{\functionANN{\activation} \bpr{  \AffineANN_{(\idMatrix_m \,\,\,  \idMatrix_m \,\,\, \ldots \,\,\, \idMatrix_m)^*,0} } }(x)
    \\& = 
    \transpose{(\idMatrix_m \,\,\,  \idMatrix_m \,\,\, \ldots \,\,\, \idMatrix_m)} x 
    =
    (x, x, \ldots, x)
  \end{split}
  \end{equation}
  \cfload.
    This 
  \proves[ep]
    \cref{item:ANN:extension:3,item:ANN:extension:4}.
\end{aproof}

\cfclear
\begin{athm}{lemma}{lem:ANN:extension:comp:left}
  Let 
    $n \in \N$, 
    $a \in C(\R, \R)$, 
    $\Phi \in \ANNs$
    \cfload.
  Then
  \begin{enumerate}[label=(\roman{*})]
  \item\label{item:ANN:extension:comp:left:1} 
  it holds that 
    $\functionANN{\activation}(\compANN{\extensionANN_{\outDimANN(\Phi), n}}{\Phi}) \in C(\R^{\inDimANN(\Phi)}, \R^{n \outDimANN(\Phi)}) $ 
  and
  \item\label{item:ANN:extension:comp:left:2}
  it holds for all 
    $x \in \R^{\inDimANN(\Phi)}$
  that 
  \begin{equation}
    \bpr{ \functionANN{\activation}(\compANN{\extensionANN_{\outDimANN(\Phi), n}}{\Phi}) }(x) 
    =
    \bpr{(\functionANN{\activation} (\Phi))(x), (\functionANN{\activation} (\Phi))(x), \ldots, (\functionANN{\activation} (\Phi))(x) }
  \end{equation}
  \end{enumerate}
  \cfout.
\end{athm}
\begin{aproof}
  \Nobs that 
    \cref{lem:ANN:extension} 
  \proves that for all 
    $m \in \N$,  
    $x  \in \R^m$
  it holds that
    $\functionANN{\activation}(\extensionANN_{m, n}) \in C(\R^{m}, \R^{mn})$ and
  \begin{equation}
  \begin{split}
    (\functionANN{\activation}(\extensionANN_{m, n})) (x)  
    =
    (x, x, \ldots, x)
  \end{split}
  \end{equation}
  \cfload.
  Combining 
    this 
  and
    \cref{PropertiesOfCompositions:Realization}   in \cref{Lemma:PropertiesOfCompositions}
  \proves[ep]
    \cref{item:ANN:extension:comp:left:1,item:ANN:extension:comp:left:2}.
\end{aproof}

\cfclear
\begin{athm}{lemma}{lem:ANN:extension:comp:right}
  Let 
    $m, n \in \N$, 
    $a \in C(\R, \R)$, 
    $\Phi\in\ANNs$ 
  satisfy 
    $\inDimANN(\Phi) = mn$ 
  \cfload. Then
  \begin{enumerate}[label=(\roman{*})]
  \item \label{item:ANN:extension:comp:right:1} 
  it holds that 
    $\functionANN{\activation}(\compANN{\Phi}{\extensionANN_{m, n}}) \in C(\R^{m}, \R^{\outDimANN(\Phi)}) $
  and
  \item \label{item:ANN:extension:comp:right:2} 
  it holds for all
    $x \in \R^{m}$ 
  that  
  \begin{equation}
    \bpr{\functionANN{\activation}(\compANN{\Phi}{\extensionANN_{m, n}}) }(x) 
    = 
    (\functionANN{\activation}(\Phi))(x, x, \ldots, x)
  \end{equation}
  \end{enumerate}
  \cfout.
\end{athm}
\begin{aproof}
  \Nobs that 
    \cref{lem:ANN:extension} 
  \proves that for all
    $x  \in \R^m$ 
  it holds that
    $\functionANN{\activation}(\extensionANN_{m, n}) \in C(\R^{m}, \R^{mn})$ and
  \begin{equation}
  \begin{split}
    (\functionANN{\activation}(\extensionANN_{m, n})) (x)
    =
    (x, x, \ldots, x)
  \end{split}
  \end{equation}
  \cfload.
  Combining 
    this and
    \cref{PropertiesOfCompositions:Realization}   in \cref{Lemma:PropertiesOfCompositions}
  \proves[ep] 
    \cref{item:ANN:extension:comp:right:1,item:ANN:extension:comp:right:2}.
\end{aproof}

\subsection{Sums of ANNs} 

\cfclear
\begin{adef}{def:ANNsum:same}[Sums of \anns\ with the same length]
\cfconsiderloaded{def:ANNsum:same}
  Let 
    $ m \in \Z $, 
    $ n \in \{ m, m+1, \dots \} $, 
    $ \Phi_m, \Phi_{m+1}, \dots, \Phi_n \in \ANNs $ 
  satisfy for all
    $k \in \{m, m+1, \ldots, n\}$ 
  that
  \begin{equation}
    \lengthANN(\Phi_k) = \lengthANN(\Phi_m) ,
    \qquad 
    \inDimANN(\Phi_k) = \inDimANN(\Phi_m) , 
    \qquad 
    \text{and} 
    \qquad 
    \outDimANN(\Phi_k) = \outDimANN(\Phi_m) 
  \end{equation}
  \cfload.
  Then we denote by 
  $
    \bigANNsum_{ k = m }^n \Phi_k 
    \in \ANNs
  $ 
  (we denote by 
  $
    \Phi_m \ANNsum \Phi_{m+1} \ANNsum \allowbreak\ldots\allowbreak \ANNsum \Phi_n 
    \in \ANNs 
  $)
  the \cfadd{def:neuralnetwork}fully-connected feedforward \ann\ given by
  \begin{equation}
    \label{eq:ANNsum:same}
    \textstyle 
    \smallbigANNsum_{k=m}^n \Phi_k 
    = 
    \bpr{ 
      \compANN{\sumANN_{\outDimANN(\Phi_m), n-m+1}}{{\compANN{\bbr{\parallelizationSpecial_{n-m+1}(\Phi_m,\Phi_{m+1},\dots, \Phi_n)}}{\extensionANN_{\inDimANN(\Phi_m), n-m+1}}}} 
    } 
    \in \ANNs
  \end{equation}
  \cfload.
\end{adef}

\cfclear
\begin{athm}{lemma}{lem:sum:ANNs}[Realizations of sums of \anns]
  Let 
    $m \in \Z$, 
    $n \in \{m, m+1, \ldots \}$, 
    $\Phi_m, \Phi_{m+1}, \ldots, \Phi_n \in \ANNs$ 
  satisfy for all 
    $k \in \{m, m+1, \ldots, n\}$ 
  that
  \begin{equation}
    \lengthANN(\Phi_k) = \lengthANN(\Phi_m),
    \qquad	
    \inDimANN(\Phi_k) = \inDimANN(\Phi_m),
    \qquad\text{and}\qquad
    \outDimANN(\Phi_k) = \outDimANN(\Phi_m)
  \end{equation}
  \cfload. 
  Then
  \begin{enumerate}[label=(\roman{*})]
  \item \label{item:sum:ANNs:1} 
  it holds that 
    $\lengthANN\bpr{\bigANNsum_{k = m}^n \Phi_k} = \lengthANN(\Phi_m)$,
  \item\label{item:sum:ANNs:2} 
  it holds that
  \begin{equation}
    \dims\pr*{\smallbigANNsum_{k = m}^n \Phi_k}
    = 
    \bbbpr{\inDimANN(\Phi_m), \smallsum_{k=m}^n \dimANNlevel_1(\Phi_k), 
      \smallsum_{k = m}^n \dimANNlevel_2(\Phi_k), \ldots, \smallsum_{k=m}^n \dimANNlevel_{\hiddenLength(\Phi_m)}(\Phi_k), \outDimANN(\Phi_m)
    }
    ,
  \end{equation}
  and 
  \item\label{item:sum:ANNs:4} 
  it holds for all 
    $a \in C(\R, \R)$ 
  that
  \begin{equation}
    \functionANN{a}\pr*{
      \smallbigANNsum_{k = m}^n \Phi_k 
    } 
    = 
    \sum_{ k = m }^n 
    ( 
      \functionANN{a}( \Phi_k ) 
    )
  \end{equation}
  \end{enumerate}
  \cfout. %
\end{athm}
\begin{aproof}
  First, \nobs that
    \cref{Lemma:ParallelizationElementary}
  \proves that
  \begin{equation}
  \label{eq:lem:sum:par}
  \begin{split}
    & \dims \bpr{ \parallelizationSpecial_{n-m+1}(\Phi_m,\Phi_{m+1},\dots, \Phi_n) }
    \\&= 
    \bbbpr{ \smallsum_{k=m}^n \dimANNlevel_0(\Phi_k), \smallsum_{k = m}^n \dimANNlevel_1(\Phi_k), \ldots,
    \smallsum_{k=m}^n \dimANNlevel_{\lengthANN(\Phi_m)-1}(\Phi_k), \smallsum_{k=m}^n \dimANNlevel_{\lengthANN(\Phi_m)}(\Phi_k)}
    \\& = 
    \begin{multlined}[t]
      \biggl((n-m+1) \inDimANN(\Phi_m), \smallsum_{k=m}^n \dimANNlevel_1(\Phi_k), \smallsum_{k = m}^n \dimANNlevel_2(\Phi_k), \ldots, \smallsum_{k=m}^n \dimANNlevel_{\lengthANN(\Phi_m)-1}(\Phi_k),\\ (n-m+1) \outDimANN(\Phi_m)\biggr)
    \end{multlined}
  \end{split}
  \end{equation}
  \cfload.
  \Moreover 
    \cref{item:ANNsum:2} in \cref{lem:def:ANNsum} 
  \proves that
  \begin{equation}
  \label{eq:lem:sum:dims}
    \dims(\sumANN_{\outDimANN(\Phi_m),n-m+1})
    = 
    ((n-m+1)\outDimANN(\Phi_m), \outDimANN(\Phi_m))
  \end{equation}
  \cfload.
    This, 
    \cref{eq:lem:sum:par}, and
    \cref{PropertiesOfCompositions:Dims} in \cref{Lemma:PropertiesOfCompositions}
  \prove that
  \begin{equation}
  \label{eq:lem:sum:comp}
  \begin{split}
    &\dims \bpr{\compANN{\sumANN_{\outDimANN(\Phi_m), n-m+1}}{\bbr{\parallelizationSpecial_{n-m+1}(\Phi_m,\Phi_{m+1},\dots, \Phi_n)}} }
    \\& = 
    \bbbpr{(n-m+1) \inDimANN(\Phi_m), \smallsum_{k=m}^n \dimANNlevel_1(\Phi_k), \smallsum_{k = m}^n \dimANNlevel_2(\Phi_k), \ldots, \smallsum_{k=m}^n \dimANNlevel_{\lengthANN(\Phi_m)-1}(\Phi_k),  \outDimANN(\Phi_m)}
    .
  \end{split}
  \end{equation}
  \Moreover 
    \cref{item:ANN:extension:2} in \cref{lem:ANN:extension} 
  \proves that
  \begin{equation}
    \dims \bpr{ \extensionANN_{\inDimANN(\Phi_m), n-m+1}} 
    =
    (\inDimANN(\Phi_m), (n-m+1) \inDimANN(\Phi_m))
  \end{equation}
  \cfload.
  Combining 
    this, 
    \cref{eq:lem:sum:comp}, and 
    \cref{PropertiesOfCompositions:Dims} in \cref{Lemma:PropertiesOfCompositions}
  \proves that
  \begin{equation}
  \begin{split}
    &\dims\pr*{\smallbigANNsum_{k = m}^n \Phi_k} 
    \\& = 
    \dims \bpr{ \compANN{\sumANN_{\outDimANN(\Phi_m), (n-m+1)}}{{\compANN{\bbr{\parallelizationSpecial_{n-m+1}(\Phi_m,\Phi_{m+1},\dots, \Phi_n)}}{\extensionANN_{\inDimANN(\Phi_m), (n-m+1)}}}}}
    \\&= 
    \bbbpr{\inDimANN(\Phi_m), \smallsum_{k=m}^n \dimANNlevel_1(\Phi_k), \smallsum_{k = m}^n \dimANNlevel_2(\Phi_k), \ldots, \smallsum_{k=m}^n \dimANNlevel_{\lengthANN(\Phi_m)-1}(\Phi_k), \outDimANN(\Phi_m)}
  \end{split}
  \end{equation}
  \cfload.
    This 
  \proves[ep]
    \cref{item:sum:ANNs:1,item:sum:ANNs:2}.
  \Nobs that 
    \cref{lem:ANN:extension:comp:right} 
    and \cref{eq:lem:sum:par} 
  \prove that for all 
    $a \in C(\R, \R)$, 
    $x \in \R^{\inDimANN(\Phi_m)}$ 
  it holds that  
  \begin{equation}
    \functionANN{\activation}\bpr{\compANN{[\parallelizationSpecial_{n-m+1}(\Phi_m,\Phi_{m+1},\dots, \Phi_n)]}{\extensionANN_{\inDimANN(\Phi_m), n-m+1}}}
    \in 
    C(\R^{\inDimANN(\Phi_m)}, \R^{(n-m+1) \outDimANN(\Phi_m)})	
  \end{equation}
  and
  \begin{equation}
  \begin{split}
    &\bpr{\functionANN{\activation}\bpr{\compANN{[\parallelizationSpecial_{n-m+1}(\Phi_m,\Phi_{m+1},\dots, \Phi_n)]}{\extensionANN_{\inDimANN(\Phi_m), n-m+1}}}} (x)
    \\&= 
    \bpr{\functionANN{\activation}\bpr{\parallelizationSpecial_{n-m+1}(\Phi_m,\Phi_{m+1},\dots, \Phi_n)}}(x, x, \ldots, x)
  \end{split}
  \end{equation}
  \cfload.
  Combining 
    this 
  with
    \cref{PropertiesOfParallelizationEqualLength:ItemTwo} in \cref{Lemma:PropertiesOfParallelizationEqualLength}
  \proves that for all 
    $a \in C(\R, \R)$, 
    $x \in \R^{\inDimANN(\Phi_m)}$ 
  it holds that  
  \begin{equation}
  \begin{split}
    &\bpr{\functionANN{\activation}\bpr{\compANN{[\parallelizationSpecial_{n-m+1}(\Phi_m,\Phi_{m+1},\dots, \Phi_n)]}{\extensionANN_{\inDimANN(\Phi_m), n-m+1}}}} (x)
    \\&= 
    \bpr{ (\functionANN{\activation} (\Phi_m))(x), (\functionANN{\activation} (\Phi_{m+1}))(x), \ldots, (\functionANN{\activation} (\Phi_n))(x) } \in \R^{(n-m+1) \outDimANN(\Phi_m)}
    .
  \end{split}
  \end{equation}
    \cref{lem:def:ANNsum:comp:left}, 
    \cref{eq:lem:sum:dims}, and
    \cref{Lemma:CompositionAssociative}
    \hence
  \prove that for all 
    $a \in C(\R, \R)$, 
    $x \in \R^{\inDimANN(\Phi_m)}$ 
  it holds that 
    $\functionANN{a}\bpr{\bigANNsum_{k = m}^n \Phi_k} \in C(\R^{\inDimANN(\Phi_m)}, \R^{\outDimANN(\Phi_m)})$ and 
  \begin{equation}
  \begin{split}
    &\pr*{\functionANN{a} \pr*{\smallbigANNsum_{k = m}^n \Phi_k } } (x)
    \\ & = 
    \bpr{\functionANN{\activation}\bpr{\compANN{\sumANN_{\outDimANN(\Phi_m), n-m+1}}{{\compANN{[\parallelizationSpecial_{n-m+1}(\Phi_m,\Phi_{m+1},\dots, \Phi_n)]}{\extensionANN_{\inDimANN(\Phi_m), n-m+1}}}}}} (x) 
    \\ & = 
    \sum_{k=m}^n( \functionANN{a}(\Phi_k)) (x)
    .
  \end{split}
  \end{equation}
    This 
  \proves[ep]
    \cref{item:sum:ANNs:4}.
\end{aproof}

%% file: parts/One-dimensional_ANN_approximation_results.tex
\cchapter{One-dimensional ANN approximation results}{sect:onedApprox}

In learning problems \anns\ are heavily used with the aim to approximate certain target functions.
In this chapter we review basic \ReLU\ \ann\ approximation results for a class of one-dimensional target functions (see \cref{sect:ANNapproxoned}). 
\ann\ approximation results for multi-dimensional target functions are treated in \cref{sect:multidApprox} below.

In the scientific literature the capacity of \anns\ to approximate certain classes of target functions has been thoroughly studied;
cf., \eg, \cite{Cybenko1989,Hornik1989,Barron1993,Hornik1991,Blum1991} for early universal \ann\ approximation results, %
cf., \eg, \cite{Boelcskei2019,Petersen2018,Shen2020,Yarotsky2017,Gribonval2022,Beneventano2021} and the references therein for more recent \ann\ approximation results establishing rates in the approximation of different classes of target functions,
and
cf., \eg, \cite{Grohs2023Aproof,Elbraechter2022,SchwabZech2019,Kutyniok2022} and the references therein for approximation capacities of \anns\ related to solutions of \PDEs\ (cf.\ also \cref{subsec:dgm,sect:deepKolmogorov} in \cref{part:MLforPDEs} of \thisbook for machine learning methods for \PDEs).
This chapter is based on Ackermann et al.~\cite[Section 4.2]{Ackermann2023}
(cf., \eg, also Hutzenthaler et al.~\cite[Section 3.4]{HutzenthalerJentzenKruseNguyen2019}).

\section{Linear interpolation of one-dimensional functions}

\subsection{On the modulus of continuity}

\begin{adef}{mod_cont_def}[Modulus of continuity]
Let $ A \subseteq \R$ be a set and let $f\colon A \to \R$ be a function.
Then we denote by
$
\modcont{f} \colon [0,\infty] \to [0,\infty]
$
the function which satisfies for all $h\in [0,\infty]$ that
\begin{equation}
\begin{split}
\label{mod_cont_def:eq1}
  \modcont{f}(h) 
&
  = 
  \sup\bpr{
    \bbr{
      \cup_{x,y\in A, \abs{x-y} \le h}
      \cu{
        \abs{f(x)-f(y)}
      }
    } 
    \cup 
    \{0\}
  }
\\[0.5ex] 
  & = 
  \sup\bpr{
    \bcu{
      r \in \R \colon 
      \pr{
        \exists \, x \in A, y \in A \cap [x-h,x+h]
        \colon r = \abs{ f(x) - f(y) }
      }
    } \cup \{0\}
  }
\end{split}
\end{equation}
and we call $ \modcont{f} $ the modulus of continuity of $f$.
\end{adef}

\cfclear
\begingroup
\begin{athm}{lemma}{lem:mod_continuity_a}[Elementary properties of moduli of continuity]
Let
  $A\subseteq \R$ be a set
and let 
  $ f \colon A \to \R $ be a function. 
Then
\begin{enumerate}[label=(\roman *)]
\item
\llabel{mod_item1}
it holds that $ \modcont{f} $ is non-decreasing,
\item
\llabel{mod_item2}
it holds that
$ f $ is uniformly continuous if and only if 
$
  \lim_{ h \searrow 0 } \modcont{f}( h ) = 0
$,
\item
\llabel{mod_item3}
it holds that
$ f $ is globally bounded if and only if
$ \modcont{f}( \infty ) < \infty $,
and
\item
\llabel{mod_item4}
it holds
for all $ x, y \in A $ that
$
  \abs{f(x) - f(y) }
  \leq
  \modcont{f} (\abs{x- y})
$
\end{enumerate}
\cfout.
\end{athm}
\begin{aproof}
  \Nobs that \cref{mod_cont_def:eq1} \proves[iep]
  \lref{mod_item1,mod_item2,mod_item3,mod_item4}.
\end{aproof}
\endgroup

\cfclear
\begin{athm}{lemma}{lem:mod_continuity_b}[Subadditivity of moduli of continuity]
Let
$a \in [-\infty, \infty]$, $b \in [a,\infty]$,
let $ f \colon ([a, b] \cap \R) \to \R $ 
be a function, 
and let
$h, \fh \in[0,\infty]$. Then
\begin{equation}
  \modcont{f}(h + \fh) \le \modcont{f}(h) + \modcont{f}(\fh)
\end{equation}
\cfout.
\end{athm}
\begin{aproof}
  Throughout this proof, assume without loss of generality that
  $\fh\leq h<\infty$.
\Nobs that 
  the fact that
    for all
      $x,y\in[a,b]\cap \R$
      with $\abs{x-y}\leq h+\fh$
    it holds that
      $[x-h,x+h]\cap [y-\fh,y+\fh]\cap[a,b]\neq\emptyset$
\proves that for all
	$x,y\in [a,b]\cap \R$
	with $\abs{x-y}\leq h+\fh$
there exists
	$z\in [a,b]\cap\R$
such that
\begin{equation}
	\abs{x-z}\leq h \qandq \abs{y-z}\leq\fh.
\end{equation}
	\Itref{lem:mod_continuity_a}{mod_item1,mod_item4} 
	\hence
\prove that for all
	$x,y\in [a,b]\cap\R$
	with $\abs{x-y}\leq h+\fh$
there exists
	$z\in [a,b]\cap\R$
such that
\begin{equation}
\begin{split}
  \abs{f(x)-f(y)}
& \leq
  \abs{f(x)-f(z)}+\abs{f(y)-f(z)}
\\ & 
\leq 
  \modcont{f}( \abs{ x - z } )
  +
  \modcont{f}( \abs{ y - z } )
\leq
  \modcont f(h)+\modcont f(\fh)
  \ifnocf 
  .
\end{split}
\end{equation}
\cfload[.]%
Combining
	this
with
	\cref{mod_cont_def:eq1}
\proves that
\begin{equation}
	\modcont f(h+\fh)
	\leq
	\modcont f(h)+\modcont f(\fh)
	.
\end{equation}
\end{aproof}

\cfclear
\begin{athm}{lemma}{modulus_lipschitz}[Properties of moduli of continuity of Lipschitz continuous functions] 
Let $ A \subseteq \R$ be a set, let $L \in [0,\infty)$, let $f\colon A \to \R$ satisfy for all 
	$x, y \in A$ that
\begin{equation}
  \llabel{eq:Lip}
  \abs{f(x) - f(y)} \leq L \abs{x - y},
\end{equation}
and let $h\in[0,\infty)$.
Then
\begin{equation}
  \modcont{f}( h ) 
  \leq L h
\end{equation}
\cfout.
\end{athm}

\begin{aproof}
\Nobs that 
\cref{mod_cont_def:eq1} and 
\lref{eq:Lip}
\prove that
\begin{equation}
\begin{split}
	\modcont f(h) 
& = 
    \sup\bpr{
    \bbr{
      \cup_{x,y\in A, \abs{x-y} \le h}
      \cu{
        \abs{f(x)-f(y)}
      }
    } 
    \cup 
    \{0\}
  }\\
&\leq
	\sup\bpr{
    \bbr{
      \cup_{x,y\in A, \abs{x-y} \le h}
      \cu{
        L \abs{x-y}
      }
    } 
    \cup 
    \{0\}
  }\\
&\leq
 	\sup(\{Lh, 0 \}) = Lh
\end{split}
\end{equation}
\cfload.
\end{aproof}

\subsection{Linear interpolation of one-dimensional functions}

\begin{adef}{def:lin_interp}[Linear interpolation operator]
Let $ K \in \N $, 
$ \fx_0, \fx_1, \dots, \fx_K, f_0, f_1, \dots,\allowbreak f_K \in\R $ 
satisfy 
$ \fx_0 < \fx_1 < \ldots < \fx_K $. 
Then we denote by 
\begin{equation} 
  \interpol{ \fx_0, \fx_1, \dots, \fx_K }{ f_0, f_1, \dots, f_K } 
  \colon \R \to \R
\end{equation}
the function which satisfies for all 
$ k \in \{ 1, 2, \dots, K \} $, 
$ x \in ( - \infty, \fx_0 ) $, 
$ y \in [ \fx_{ k - 1 }, \fx_k ) $, 
$ z \in [ \fx_K, \infty ) $ 
that 
\begin{equation}
\label{def:lin_interp:eq1A}
  ( \interpol{ \fx_0, \fx_1, \dots, \fx_K }{ f_0, f_1, \dots, f_K } )( x ) 
  = f_0
  ,
  \qquad 
  ( \interpol{ \fx_0, \fx_1, \dots, \fx_K }{ f_0, f_1, \dots, f_K } )( z ) = f_K 
  ,
\end{equation}
\begin{equation}
\label{def:lin_interp:eq1}
  \text{and}
\qquad 
  (
    \interpol{ \fx_0, \fx_1, \dots, \fx_K }{ f_0, f_1, \dots, f_K } 
  )( y ) 
  = f_{ k - 1 } 
  + 
  \bpr{ 
    \tfrac{ y - \fx_{ k - 1 } }{ \fx_k - \fx_{ k - 1 } } 
  }( 
    f_k - f_{ k - 1 } 
  ) 
  .
\end{equation}
\end{adef}

\cfclear
\begin{athm}{lemma}{inerpol_properties}[Elementary properties of the linear interpolation operator]
Let $K\in\N$, $\fx_0,\fx_1,\dots,\fx_K, f_0, f_1, \dots, f_K \in\R$ satisfy  $\fx_0 < \fx_1 < \ldots < \fx_K$.
Then 
\begin{enumerate}[label=(\roman *)]
\item \label{inerpol_properties:item1}
it holds for all 
	$k \in \{ 0, 1, \ldots, K\}$
that
\begin{equation}
\label{inerpol_properties:concl1}
\begin{split} 
(\interpol{\fx_0,\fx_1,\dots,\fx_K}{f_0,f_1,\dots,f_K})(\fx_k) = f_k,
\end{split}
\end{equation}

\item \label{inerpol_properties:item2}
it holds for all 
	$k \in \{ 1, 2, \ldots, K\}$,
	$x \in [\fx_{k-1},\fx_k]$
that
\begin{equation}
\label{inerpol_properties:concl2}
(\interpol{\fx_0,\fx_1,\dots,\fx_K}{f_0,f_1,\dots,f_K})(x) = f_{k-1} + \bpr{\tfrac{x - \fx_{k-1}}{\fx_k - \fx_{k-1}}}(f_k - f_{k-1}),
\end{equation}
and

\item \label{inerpol_properties:item3}
it holds for all 
	$k \in \{ 1, 2, \ldots, K\}$,
	$x \in [\fx_{k-1},\fx_k]$
that
\begin{equation}
\label{inerpol_properties:concl3}
\begin{split} 
	(\interpol{\fx_0,\fx_1,\dots,\fx_K}{f_0,f_1,\dots,f_K})(x)
=
	\bpr{\tfrac{ \fx_{k} - x }{\fx_k - \fx_{k-1}}} f_{k-1} + \bpr{\tfrac{x - \fx_{k-1}}{\fx_k - \fx_{k-1}}}f_k.
\end{split}
\end{equation}
\end{enumerate}
\cfout.

\end{athm}

\begin{aproof}
\Nobs that \cref{def:lin_interp:eq1,def:lin_interp:eq1A} \prove[iep] \cref{inerpol_properties:item1,inerpol_properties:item2}.
\Nobs that \cref{inerpol_properties:item2} \proves that for all 
$ k \in \{ 1, 2, \dots, K \} $,
$ x \in [ \fx_{ k - 1 }, \fx_k ] $
it holds that
\begin{equation}
\label{inerpol_properties:eq1}
\begin{split} 
  (\interpol{\fx_0,\fx_1,\dots,\fx_K}{f_0,f_1,\dots,f_K})(x)
& =
  f_{k-1} + \bpr{\tfrac{x - \fx_{k-1}}{\fx_k - \fx_{k-1}}}(f_k - f_{k-1})\\
& =
  \br*{ \bpr{\tfrac{ \fx_{k} - \fx_{k-1} }{\fx_k - \fx_{k-1}}} - \bpr{\tfrac{x - \fx_{k-1}}{\fx_k - \fx_{k-1}}} }
  f_{k-1} 
  + 
  \bpr{\tfrac{x - \fx_{k-1}}{\fx_k - \fx_{k-1}}}
  f_k
\\ & =
  \bpr{ \tfrac{ \fx_{k} - x }{\fx_k - \fx_{k-1}} } 
  f_{k-1} 
  + 
  \bpr{\tfrac{x - \fx_{k-1}}{\fx_k - \fx_{k-1}}} 
  f_k .
\end{split}
\end{equation}
This \proves[ep] \cref{inerpol_properties:item3}.
\end{aproof}

\begingroup
\cfclear
\begin{athm}{prop}{interpol_nonlipschitz}[Approximation and continuity properties for the linear interpolation operator]
Let $K\in \N$, $\fx_0,\fx_1,\dots, \fx_K \in \R$ 
satisfy $\fx_0 < \fx_1 < \ldots < \fx_K$ and let $f \colon [\fx_0,\fx_K] \to \R$ be a function.
Then 
\begin{enumerate}[label=(\roman *)]
\item
\label{interpol_nonlipschitz:item1}
it holds for all $x,y\in\R$ with $x\neq y$ that 
\begin{equation}
\begin{split}
&
  \abs*{
    (\interpol{
      \fx_0,\fx_1,\dots,\fx_K
		}{
			f(\fx_0), f(\fx_1), \ldots, f(\fx_K)
		})(x)
		- 
		(\interpol{
					\fx_0,\fx_1,\dots,\fx_K
				}{
					f(\fx_0), f(\fx_1), \ldots, f(\fx_K)
				})(y)
	}
\\ & \leq 
	\pr*{\max_{k\in\{1,2,\dots,K\}}\pr*{\frac{\modcont{f}(\fx_{k}-\fx_{k-1})}{\fx_{k}-\fx_{k-1}}}}\abs{x-y}
\end{split}
\end{equation}
and

\item 
\label{interpol_nonlipschitz:item2}
it holds that 
\begin{equation}
\textstyle 
	\sup_{x\in[\fx_0,\fx_K]}
		\babs{
			(\interpol{
					\fx_0,\fx_1,\dots,\fx_K
				}{
					f(\fx_0), f(\fx_1), \ldots, f(\fx_K)
				})(x) 
			- 
			f(x)
		}
\leq 
	\modcont f(\max_{k\in\{1,2,\dots,K\}}\abs{\fx_{k}-\fx_{k-1}})
\end{equation}
\end{enumerate}
\cfout.
\end{athm}

\newcommand{\vl}{\mathfrak l}

\begin{aproof}
Throughout this proof,
let $L\in [0,\infty]$ satisfy 
\begin{equation}
\label{interpol_nonlipschitz:setting1}
L = \max_{k\in\{1,2,\dots,K\}} \pr*{\frac{\modcont{f}(\fx_{k}-\fx_{k-1})}{\fx_{k}-\fx_{k-1}}}
\end{equation}
and let
$\vl \colon \R \to \R$ satisfy for all 
	$x \in \R$
that
\begin{equation}
	\vl (x) 
= 
	(\interpol{\fx_0,\fx_1,\dots,\fx_K}{	f(\fx_0), f(\fx_1), \ldots, f(\fx_K)})(x)  
\end{equation}
\cfload.
\Nobs[Observe] that 
\cref{inerpol_properties:item2} in \cref{inerpol_properties},
\itref{lem:mod_continuity_a}{mod_item4}, and
\eqref{interpol_nonlipschitz:setting1}
\prove that for all $k\in\{1,2,\dots,K\}$, $x,y\in[\fx_{k-1},\fx_{k}]$ with $x \neq y$ it holds that
\begin{equation}
\label{interpol_nonlipschitz:eq1}
\begin{split} 
	\abs{\vl(x) - \vl(y)} 
&= 
	\abs*{
		\bpr{\tfrac{x - \fx_{k-1}}{\fx_k - \fx_{k-1}}}(f(\fx_{k}) - f(\fx_{k-1}))
		-
		\bpr{\tfrac{y - \fx_{k-1}}{\fx_k - \fx_{k-1}}}(f(\fx_{k})- f(\fx_{k-1}))
	}\\
&=
	\abs*{\pr*{\frac{f(\fx_{k}) - f(\fx_{k-1})}{\fx_{k} - \fx_{k-1}}}(x-y)} 
\leq 
	\pr*{\frac{\modcont{f}(\fx_{k}-\fx_{k-1})}{\fx_{k}-\fx_{k-1}}}\abs{x-y}
\leq
	L\abs{x-y}.
\end{split}
\end{equation}
\enum{
  This;
  the triangle inequality;
  \cref{inerpol_properties:item1} in \cref{inerpol_properties};
}\prove that for all
$k,l \in \{1,2,\dots,K\}$, 
$x\in[\fx_{k-1},\fx_{k}]$, 
$y\in[\fx_{l-1},\fx_{l}]$ 
with $k<l$ and $x \neq y$
it holds that
\begin{equation}
  \llabel{eq:12}
\begin{split}
\abs{\vl(x) - \vl(y)}
&  \le 
  \abs{\vl(x) - \vl(\fx_{k})} + \abs{\vl(\fx_{k}) - \vl(\fx_{l-1})} + \abs{\vl(\fx_{l-1}) - \vl(y)}\\
&  \leq 
  \abs{\vl(x) - \vl(\fx_{k})} + \pr*{\sum_{j=k+1}^{l-1} \abs{\vl(\fx_{j-1}) - \vl(\fx_{j})} }+ \abs{\vl(\fx_{l-1}) - \vl(y)} \\
& \leq
  L \pr*{
    \abs{x - \fx_{k}} + 
    \br*{\sum_{j=k+1}^{l-1} \abs{\fx_{j-1} - \fx_{j}}}  
    + \abs{\fx_{l-1} - y}
  } 
  =  L\abs{x-y} 
  .
\end{split}
\end{equation}
Combining this and \cref{interpol_nonlipschitz:eq1} 
\proves that for all $ x, y \in [ \fx_0, \fx_K ] $ 
with $ x \neq y $ 
it holds that 
\begin{equation}
  \abs{\vl(x) - \vl(y)} \le L\abs{x-y}.
\end{equation}
This, the fact that for all $x,y\in(-\infty,\fx_0]$  with $x \neq y$ it holds that 
\begin{equation}
  \abs{\vl(x) - \vl(y)} = 0 \le L\abs{x-y}
  ,
\end{equation}
the fact that for all $x,y\in[\fx_K,\infty)$  with $x \neq y$ it holds that 
\begin{equation}
  \abs{\vl(x) - \vl(y)} = 0 \le L\abs{x-y}
  ,
\end{equation}
and the triangle inequality \hence \prove that for all $x,y\in\R$  with $x \neq y$ it holds that 
\begin{equation}
  \abs{\vl(x) - \vl(y)} \le L\abs{x-y}.
\end{equation}
This \proves[ep]
\cref{interpol_nonlipschitz:item1}.
\Nobs that
\cref{inerpol_properties:item3} in \cref{inerpol_properties}
\proves that for all $k\in\{1,2,\dots,K\}$,
$x\in[\fx_{k-1},\fx_{k}]$ it holds that
\begin{equation}
\begin{split}
	\abs{\vl(x) - f(x)} 
& = 
	\abs*{
		\pr*{\frac{\fx_{k}-x}{\fx_{k}-\fx_{k-1}}} f(\fx_{k-1}) 
		+
		\pr*{\frac{x-\fx_{k-1}}{\fx_{k}-\fx_{k-1}}} f(\fx_k)
		-f(x) 
	} \\
&=
	\abs*{
			\pr*{\frac{\fx_{k}-x}{\fx_{k}-\fx_{k-1}}} (f(\fx_{k-1}) -f(x))
			+
			\pr*{\frac{x-\fx_{k-1}}{\fx_{k}-\fx_{k-1}}} (f(\fx_{k}) - f(x))
	} \\
& \leq 
	 \pr*{\frac{\fx_{k}-x}{\fx_{k}-\fx_{k-1}}} \abs{f(\fx_{k-1}) - f(x)}
	 + 
	 \pr*{\frac{x - \fx_{k-1}}{\fx_{k} - \fx_{k-1}}} \abs{f(\fx_{k}) - f(x)}
   .
\end{split}
\end{equation}
Combining
  this
  with
  \cref{mod_cont_def:eq1}
  and \cref{lem:mod_continuity_a}
\proves that
for all $k\in\{1,2,\dots,K\}$,
$x\in[\fx_{k-1},\fx_{k}]$ it holds that
\begin{equation}
\begin{split}
	\abs{\vl(x) - f(x)} 
& \leq \modcont{f}(\abs{\fx_{k}-\fx_{k-1}})\pr*{\frac{\fx_{k}-x}{\fx_{k}-\fx_{k-1}} + \frac{x - \fx_{k-1}}{\fx_{k} - \fx_{k-1}}}\\
& = \modcont{f}(\abs{\fx_{k} - \fx_{k-1}}) \le \modcont{f}(\textstyle\max_{j\in\{1,2,\dots,K\}}\abs{\fx_{j} - \fx_{j-1}}).
\end{split}
\end{equation}
This \proves[ep] \cref{interpol_nonlipschitz:item2}.
\end{aproof}
\endgroup

\cfclear
\begin{athm}{cor}{interpol_lipschitz}[Approximation and Lipschitz continuity properties for the linear interpolation operator] 
Let $K\in \N$, $L, \fx_0,\fx_1,\dots, \fx_K \in \R$ 
satisfy $\fx_0 < \fx_1 < \ldots < \fx_K$ and let $f \colon [\fx_0,\fx_K] \to \R$ satisfy for all $x,y\in [\fx_0,\fx_K]$ that 
\begin{equation}
  \abs{f(x) - f(y)} \le L\abs{x-y}
  .
\end{equation}
Then 
\begin{enumerate}[label=(\roman *)]
\item
\label{interpol_lipschitz:item1}
it holds for all $x,y\in\R$ that 
\begin{equation}
\label{interpol_lipschitz:concl1}
\begin{split} 
	\abs*{
		(\interpol{
			\fx_0,\fx_1,\dots,\fx_K
		}{
			f(\fx_0), f(\fx_1), \ldots, f(\fx_K)
		})(x)
		- 
		(\interpol{
					\fx_0,\fx_1,\dots,\fx_K
				}{
					f(\fx_0), f(\fx_1), \ldots, f(\fx_K)
				})(y)
	}
\leq 
	L\abs{x-y}
\end{split}
\end{equation}
and
\item 
\label{interpol_lipschitz:item2}
it holds that 
\begin{equation}  
	\sup_{x\in[\fx_0,\fx_K]}
		\babs{
			(\interpol{
					\fx_0,\fx_1,\dots,\fx_K
				}{
					f(\fx_0), f(\fx_1), \ldots, f(\fx_K)
				})(x) 
			- 
			f(x)
		}
\leq 
	L \bbpr{\max_{k\in\{1,2,\dots,K\}}\abs{\fx_{k}-\fx_{k-1}} }
\end{equation}
\end{enumerate}
\cfout.
\end{athm}

\begin{aproof}
\Nobs that the 
assumption that for all $ x, y \in [ \fx_0, \fx_K ] $ 
it holds that $ \abs{ f(x) - f(y) } \le L \abs{ x - y } $
\proves that 
\begin{equation}
  0 
  \leq 
  \frac{ 
    \abs{ f( \fx_K ) - f( \fx_0 ) }
  }{
    ( \fx_K - \fx_0 )
  }
  \leq 
  \frac{ 
    L \abs{ \fx_K - \fx_0 }
  }{
    ( \fx_K - \fx_0 )
  }
  =
  L
  .
\end{equation}
Combining this, \cref{modulus_lipschitz}, 
and the assumption that for all $ x, y \in [ \fx_0, \fx_K ] $ it holds that 
$ 
  \abs{ f(x) - f(y) } \le L \abs{ x - y } 
$
with \cref{interpol_nonlipschitz:item1} in \cref{interpol_nonlipschitz} \proves that for all 
$ x, y \in \R $ 
it holds that 
\begin{equation}
\begin{split} 
	&\abs*{
		(\interpol{
			\fx_0,\fx_1,\dots,\fx_K
		}{
			f(\fx_0), f(\fx_1), \ldots, f(\fx_K)
		})(x)
		- 
		(\interpol{
					\fx_0,\fx_1,\dots,\fx_K
				}{
					f(\fx_0), f(\fx_1), \ldots, f(\fx_K)
				})(y)
	}
\\&\leq 
		\pr*{\max_{k\in\{1,2,\dots,K\}}\pr*{\frac{L\abs{\fx_{k}-\fx_{k-1}}}{\abs{\fx_{k}-\fx_{k-1}}}}}\abs{x-y}
=
	L\abs{x-y}
	.
\end{split}
\end{equation}
This \proves[ep] \cref{interpol_lipschitz:item1}.
\Nobs that 
\enum{the assumption that for all $x,y\in[\fx_0,\fx_K]$ it holds that $\abs{f(x)-f(y)} \le L\abs{x-y}$; \cref{modulus_lipschitz}; \cref{interpol_nonlipschitz:item2} in \cref{interpol_nonlipschitz}} 
\prove that
\begin{equation}
\begin{split} 
	\sup_{x\in[\fx_0,\fx_K]}
		\babs{
			(\interpol{
					\fx_0,\fx_1,\dots,\fx_K
				}{
					f(\fx_0), f(\fx_1), \ldots, f(\fx_K)
				})(x) 
			- 
			f(x)
		}
& \leq 
  \modcont{f}\pr*{
    \max_{ k \in \{ 1, 2, \dots, K \} }
    \abs{ \fx_k - \fx_{ k - 1 } }
  }
\\ & \leq 
  L\pr*{
    \max_{ k \in \{ 1, 2, \dots, K \} }
    \abs{ \fx_k - \fx_{ k - 1 } }
  }
  .
\end{split}
\end{equation}
This \proves[ep] \cref{interpol_lipschitz:item2}.
\end{aproof}

\section{Linear interpolation with ANNs}

\subsection{Activation functions as ANNs}

\cfclear
\begin{adef}{def:padding}[Activation functions as \anns]
  \cfconsiderloaded{def:padding}
Let $n\in\N$.
Then we denote by
\begin{equation}
  \ii_n \in ((\R^{n\times n}\times \R^n)\times (\R^{n\times n}\times \R^n)) \subseteq \ANNs
\end{equation}
the fully-connected feedforward \ann\ given by
\begin{equation}
  \ii_n = ((\idMatrix_n,0),(\idMatrix_n,0)) 
\end{equation}
\cfload.
\end{adef}

\cfclear
\begin{athm}{lemma}{padding_lemma}[Realization functions of activation \anns]
Let $ n \in \N $.
Then
\begin{enumerate}[label=(\roman*)]
\item
\label{padding_lemma:item1}
it holds that 
$
  \dims( \ii_n ) = (n,n,n) \in \N^3 
$
and 
\item
\label{padding_lemma:item3}
it holds for all $a\in C(\R,\R)$ that
\begin{equation}
  \functionANN{a}(\ii_n) = \multdim_{a,n}
\end{equation}
\end{enumerate} 
\cfout.
\end{athm}

\begin{aproof}
\Nobs that the fact that 
$
  \ii_n \in 
  ( ( \R^{ n \times n } \times \R^n ) \times ( \R^{ n \times n } \times \R^n ) ) \subseteq \ANNs
$ 
\proves that 
\begin{equation}
  \dims(\ii_n) = (n,n,n) \in \N^3
\end{equation}
\cfload. 
This \proves[ep] \cref{padding_lemma:item1}. 
\Nobs that 
\cref{setting_NN:ass2}
and the fact that
\begin{equation}
  \ii_n = ((\idMatrix_n,0),(\idMatrix_n,0)) \in ((\R^{n\times n}\times\R^n)\times (\R^{n\times n}\times\R^n))
\end{equation}
\prove that for all $ a \in C(\R,\R) $, 
$ x \in \R^n $ it holds that $\functionANN{a}(\ii_n) \in C(\R^n,\R^n)$ and
\begin{equation}
  (\functionANN{a}(\ii_n))(x) 
  = \idMatrix_n\pr*{\multdim_{a,n}(\idMatrix_n\! x + 0)} + 0 = \multdim_{a,n}(x).
\end{equation}
This \proves[ep] \cref{padding_lemma:item3}. 
\end{aproof}

\cfclear
\begin{athm}{lemma}{padding_lemma2}[Compositions of activation \anns\ with 
general \anns] 
Let $\Phi\in\ANNs$ 
\cfload.
Then
\begin{enumerate}[label=(\roman *)] 
\item
\label{padding_lemma2:item1}
it holds that
\begin{equation}
\begin{split}
&
  \dims(\compANN{\ii_{\outDimANN  (\Phi)}}{\Phi}) 
\\ & = 
  (\dimANNlevel_0(\Phi),\dimANNlevel_1(\Phi),\dimANNlevel_2(\Phi),\dots,\dimANNlevel_{\lengthANN(\Phi)-1}(\Phi),\dimANNlevel_{\lengthANN(\Phi)}(\Phi),\dimANNlevel_{\lengthANN(\Phi)}(\Phi)) \in \N^{\lengthANN(\Phi) + 2}
  ,
\end{split}
\end{equation}
\item
\label{padding_lemma2:item2}
it holds for all $ a \in C( \R, \R ) $ that 
$
  \functionANN{a}(\compANN{\ii_{\outDimANN(\Phi)}}{ \Phi}) \in C(\R^{\inDimANN(\Phi)},\R^{\outDimANN(\Phi)})
$,
\item
\label{padding_lemma2:item3}
it holds for all $a\in C(\R,\R)$ that
$
\functionANN{a}( \compANN{\ii_{\outDimANN(\Phi)} }{ \Phi}) = \multdim_{a,\outDimANN(\Phi)}\circ(\functionANN{a}(\Phi))
$,
\item
\label{padding_lemma2:item4}
it holds that
\begin{equation}
\begin{split}
&
  \dims(\compANN{\Phi}{ \ii_{\inDimANN(\Phi)}}) 
\\ &  
  = (\dimANNlevel_0(\Phi),\dimANNlevel_0(\Phi),\dimANNlevel_1(\Phi),\dimANNlevel_2(\Phi),\dots,\dimANNlevel_{\lengthANN(\Phi)-1}(\Phi),\dimANNlevel_{\lengthANN(\Phi)}(\Phi)) \in \N^{\lengthANN(\Phi) + 2}
  ,
\end{split}
\end{equation} 
\item
\label{padding_lemma2:item5}
it holds for all $ a \in C( \R, \R )$ 
that 
$
  \functionANN{a}( \compANN{\Phi }{ \ii_{ \inDimANN( \Phi ) } } ) 
  \in C( \R^{ \inDimANN( \Phi ) }, \R^{ \outDimANN( \Phi ) } )
$, 
and
\item
\label{padding_lemma2:item6}
it holds for all $a\in C(\R,\R)$ that
$
\functionANN{a}(\compANN{\Phi }{ \ii_{\inDimANN(\Phi)}}) = (\functionANN{a}(\Phi))\circ\multdim_{a,\inDimANN(\Phi)}
$
\end{enumerate}
\cfout.
\end{athm}

\begin{aproof}
\Nobs that \cref{padding_lemma} \proves that 
for all $ n \in \N $, $ a \in C( \R, \R ) $
it holds that 
\begin{equation}
  \functionANN{a}( \ii_n )  
  = \multdim_{ a, n }
\end{equation}
\cfload.
Combining this and 
\cref{Lemma:PropertiesOfCompositions}
\proves[ep]
\cref{padding_lemma2:item1,%
padding_lemma2:item3,%
padding_lemma2:item2,%
padding_lemma2:item4,%
padding_lemma2:item6,padding_lemma2:item5}.
\end{aproof}

\subsection{Representations for ReLU ANNs with one hidden neuron} 

\cfclear
\begin{athm}{lemma}{basicANN}
Let $\alpha, \beta,  h \in \R$, $\basicANN \in \ANNs$ satisfy 
\begin{equation}
  \basicANN = \scalarMultANN h {(\compANN{\ii_1}{ \AffineANN_{\alpha,\beta}})}
\end{equation}
\cfload.
Then 
\begin{enumerate}[label=(\roman *)]
\item
\label{basicANN:item1}
it holds that $\basicANN= ((\alpha,\beta),(h,0))$,
\item
\label{basicANN:item2}
it holds that
$\dims(\basicANN) = (1,1,1) \in \N^3$,
\item 
\label{basicANN:item3}
it holds that
$\Rr(\basicANN) \in C(\R,\R)$,
and
\item 
\label{basicANN:item4}
it holds for all $x\in\R$ that
$
(\Rr(\basicANN))(x) = h\max\{\alpha x + \beta,0\} %
$
\end{enumerate}
\cfout.
\end{athm}

\begin{aproof}
\Nobs that \cref{lem:ANN:affine} \proves that 
\begin{equation}
  \AffineANN_{\alpha,\beta} = (\alpha,\beta)
  ,
  \qquad 
  \dims(\AffineANN_{\alpha,\beta}) = (1,1) \in \N^2
  ,
  \qquad 
  \Rr(\AffineANN_{\alpha,\beta}) \in C(\R,\R)
  ,
\end{equation}
and 
$\forall \, x \in \R \colon (\Rr(\AffineANN_{\alpha,\beta}))(x) = \alpha x + \beta$ \cfload.
\Cref{Lemma:PropertiesOfCompositions},
\cref{padding_lemma},
\cref{padding_lemma2}, 
\cref{def:relu1:eq1}, 
\cref{setting_NN:ass2}, 
and \cref{eq:defCompANN}
\hence \prove that
\begin{gather}
  \compANN{\ii_1 }{ \AffineANN_{\alpha,\beta}} = ((\alpha,\beta),(1,0))
  , 
\;
  \dims(\compANN{\ii_1 }{ \AffineANN_{\alpha,\beta}}) = (1,1,1) \in \N^3
  , 
\;
  \Rr(\compANN{\ii_1 }{ \AffineANN_{\alpha,\beta}}) \in C(\R,\R)
  , 
  \\
  \text{and}
  \qquad 
	\forall\, x \in \R 
\colon 
	(\Rr(\compANN{\ii_1}{ \AffineANN_{\alpha,\beta}}))(x)
=
	\rect(\Rr(\AffineANN_{\alpha,\beta})(x))
= 
	\max\{\alpha x + \beta, 0\}. %
\end{gather}
This, \cref{lem:ANNscalar}, and \cref{def:ANNscalar:eq1}
\prove that 
\begin{gather}
  \basicANN=\scalarMultANN h {( \compANN{\ii_1}{ \AffineANN_{\alpha,\beta}} )} = ((\alpha,\beta),(h,0))
  , 
  \quad 
  \dims(\basicANN) = (1,1,1) 
  , 
  \quad 
  \Rr(\basicANN) \in C(\R,\R)
  ,\\
  \text{and}
\qquad 
(\Rr(\basicANN))(x) = h((\Rr(\compANN{\ii_1 }{ \AffineANN_{\alpha,\beta}}))(x))
= h\max\{\alpha x + \beta,0\}. %
\end{gather}
This \proves[ep]
\cref{basicANN:item1,basicANN:item2,basicANN:item3,basicANN:item4}.
\end{aproof}

\subsection{ReLU ANN representations for linear interpolations}

\cfclear
\begin{athm}{prop}{interpol_ANN_points}[\ReLU\ \ann\ representations for linear interpolations]
Let 
$ 
  K \in \N 
$, 
$ 
  f_0, \allowbreak f_1, \dots, f_K, \fx_0, \fx_1, \dots, \fx_K \in \R 
$
satisfy 
$
  \fx_0 < \fx_1 < \ldots < \fx_K
$
and let $\interpolatingDNN \in \ANNs$ satisfy
\begin{equation} 
\label{interpol_ANN_points:ass1}
	\interpolatingDNN 
= 
	\compANN{
		\AffineANN_{1,f_0}
	}{
	  \pr*{{\smallbigANNsum_{k=0}^K} \pr*{\scalarMultANN{\pr*{\tfrac{(f_{\min\{k+1,K\}}-f_k)}{(\fx_{\min\{k+1,K\}}-\fx_{\min\{k,K-1\}})} - \tfrac{(f_k - f_{\max\{k-1,0\}})}{(\fx_{\max\{k,1\}} - \fx_{\max\{k-1,0\}})}}}{
	  (\compANN{\ii_{1} }{ \AffineANN_{1,-\fx_k}})}}}
	}
\end{equation}
\cfload.
Then 
\begin{enumerate}[label=(\roman *)]
\item 
\label{interpol_ANN_points:item1}
it holds that $\dims(\interpolatingDNN) = (1,K+1,1) \in \N^3$,
\item 
\label{interpol_ANN_points:item3}
it holds that
$
	\Rr(\interpolatingDNN)
=
	\interpol{\fx_0,\fx_1,\dots,\fx_K}{f_0,f_1,\dots,f_K}
$,
and

\item
\label{interpol_ANN_points:item4}
it holds that $\paramANN(\interpolatingDNN) = 3K+4$
\end{enumerate}
\cfout.
\end{athm}

\begin{aproof}
Throughout this proof,
let $ c_0, c_1, \dots, c_K  \in \R$ satisfy for all $k\in\{0,1,\dots,K\}$ that
\begin{equation}\label{interpol_ANN_points:setting1}
  c_k = 
  \frac{(f_{\min\{k+1,K\}}-f_k)}{(\fx_{\min\{k+1,K\}}-\fx_{\min\{k,K-1\}})} - \frac{(f_k - f_{\max\{k-1,0\}})}{(\fx_{\max\{k,1\}} - \fx_{\max\{k-1,0\}})}
\end{equation}
and let $\Phi_0, \Phi_1, \ldots, \Phi_K \in ((\R^{1\times 1}\times \R^1)\times (\R^{1\times 1}\times \R^1)) \subseteq \ANNs$ 
satisfy for all $k\in\{0,1,\dots,K\}$ that
\begin{equation}
  \Phi_k = \scalarMultANN{c_k}{(\compANN{\ii_1 }{ \AffineANN_{1,-\fx_k}})}
  .
\end{equation}
\Nobs that \cref{basicANN} \proves that for all 
$
  k \in \{0,1,\dots,K\}
$
it holds that
\begin{gather}
  \Rr(\Phi_k) \in C(\R,\R)
  ,
  \qquad 
  \dims(\Phi_k) = (1,1,1) \in \N^3 , 
\\
\label{eq:explicit_representation_Phi_k_in_proof}
  \text{and}
\qquad 
  \forall \, x \in \R \colon (\Rr(\Phi_k))(x) = c_k\max\{ x - \fx_k,0\}
\end{gather}
\cfload.
\enum{This; 
\cref{lem:ANN:affine2}; 
\cref{lem:sum:ANNs};
\cref{interpol_ANN_points:ass1}}
\prove that 
\begin{equation}
  \dims(\interpolatingDNN) = (1, K+1, 1) \in \N^3
  \qandq
  \Rr(\interpolatingDNN) \in C(\R,\R)
  .
\end{equation}
This \proves[ep]
\cref{interpol_ANN_points:item1}.
\Nobs that \cref{interpol_ANN_points:item1} and \eqref{def:ANN:eq1} \prove that 
\begin{equation}
\label{interpol_ANN_points:eq0}
\begin{split} 
  \paramANN(\interpolatingDNN)
=
  2(K + 1) + (K+2)
=
  3 K + 4 .
\end{split}
\end{equation}
This \proves \cref{interpol_ANN_points:item4}.
\Nobs that 
\enum{\cref{interpol_ANN_points:setting1};\cref{eq:explicit_representation_Phi_k_in_proof};\cref{lem:ANN:affine2};\cref{lem:sum:ANNs}}
\prove that for all $ x \in \R $ it holds that
\begin{equation}
\label{interpol_ANN_points:eq1}
  (\Rr(\interpolatingDNN))(x) 
  = f_0 + \sum_{k=0}^K (\Rr(\Phi_k))(x) = f_0 + \sum_{k=0}^K c_k \max\{x - \fx_k,0\}.
\end{equation}
This and the fact that 
for all $ k \in \{ 0, 1, \dots, K \} $
it holds that 
$ \fx_0 \le \fx_k $ 
\prove that for all 
$ x \in ( - \infty, \fx_0 ] $ 
it holds that 
\begin{equation}
\label{interpol_ANN_points:eq2}
  (\Rr(\interpolatingDNN))(x) = f_0 + 0 = f_0 .
\end{equation}
Next we claim that for all
$
  k \in \{ 1, 2, \dots, K \} 
$
it holds that
\begin{equation}
\label{interpol_ANN_points:eq1.1}
\begin{split} 
  \sum_{ n = 0 }^{ k - 1 } 
  c_n 
  = 
  \frac{ 
    f_k - f_{ k - 1 } 
  }{
    \fx_k - \fx_{ k - 1 } 
  }
  .
\end{split}
\end{equation}
We now prove \eqref{interpol_ANN_points:eq1.1} by induction on $ k \in \{ 1, 2, \dots, K \} $. 
For the base case $ k = 1 $ 
observe that \cref{interpol_ANN_points:setting1} \proves that 
\begin{equation}
  \sum_{ n = 0 }^0 
  c_n 
  = c_0 
  = \frac{ f_1 - f_0 }{ \fx_1 - \fx_0 }
  .
\end{equation}
This \proves \eqref{interpol_ANN_points:eq1.1} in the base case $ k = 1 $.
For the induction step \nobs that \cref{interpol_ANN_points:setting1} \proves that for all 
$
  k \in \N \cap (1,\infty) \cap (0,K] 
$ 
with 
$
  \sum_{ n = 0 }^{ k - 2 } c_n 
  = 
  \tfrac{ f_{ k - 1 } - f_{ k - 2 } }{ \fx_{ k - 1 } - \fx_{ k - 2 } } 
$ 
it holds that
\begin{equation}
\begin{split} 
  \sum_{ n = 0 }^{ k - 1 } 
  c_n 
= 
  c_{ k - 1 } 
  + 
  \sum_{ n = 0 }^{ k - 2 } 
  c_n 
=
  \frac{ f_k - f_{ k - 1 } }{ \fx_k - \fx_{ k - 1 } }
  - 
  \frac{ f_{ k - 1 } - f_{ k - 2 } }{ \fx_{ k - 1 } - \fx_{ k - 2 } }
  + 
  \frac{ f_{ k - 1 } - f_{ k - 2 } }{ \fx_{ k - 1 } - \fx_{ k - 2 } }
=
  \frac{ f_k - f_{ k - 1 } }{ \fx_k - \fx_{ k - 1 } }
  .
\end{split}
\end{equation}
Induction thus \proves \eqref{interpol_ANN_points:eq1.1}. 
\Moreover 
\eqref{interpol_ANN_points:eq1}, \eqref{interpol_ANN_points:eq1.1}, and
the fact that for all $ k \in \{ 1, 2, \dots, K \} $ it holds that 
$ \fx_{ k - 1 } < \fx_k $
\prove that for all $ k \in \{ 1, 2, \dots, K \} $, $ x \in [ \fx_{ k - 1 }, \fx_k ] $ 
it holds that
\begin{equation}
\label{interpol_ANN_points:eq3}
\begin{split}
  ( \Rr( \interpolatingDNN ) )( x ) 
  - 
  ( \Rr( \interpolatingDNN ) )( \fx_{ k - 1 } ) 
& 
  = 
  \sum_{ n = 0 }^K 
  c_n 
  \pr*{
    \max\{ x - \fx_n, 0 \} - \max\{ \fx_{ k - 1 } - \fx_n ,0 \} 
  }
\\
& 
  = 
  \sum_{ n = 0 }^{ k - 1 } 
  c_n
  [ ( x - \fx_n ) - ( \fx_{ k - 1 } - \fx_n ) ] 
  = 
  \sum_{ n = 0 }^{ k - 1 } 
  c_n
  ( x - \fx_{ k - 1 } )
\\
& 
  = 
  \pr*{ 
    \frac{ f_k - f_{ k - 1 } }{ \fx_k - \fx_{ k - 1 } } 
  } 
  \pr*{ x - \fx_{ k - 1 } }
  .
\end{split}
\end{equation}
Next we claim that for all 
$ k \in \{ 1, 2, \dots, K \} $, 
$ x \in [ \fx_{ k - 1 }, \fx_k ] $ 
it holds that
\begin{equation}
\label{interpol_ANN_points:eq4}
  ( \Rr( \interpolatingDNN ) )(x) 
  = 
  f_{ k - 1 } 
  + 
  \pr*{ 
    \frac{ f_k - f_{ k - 1 } }{ \fx_k - \fx_{ k - 1 } } 
  }
  \pr*{ x - \fx_{ k - 1 } } .
\end{equation}
We now prove \cref{interpol_ANN_points:eq4} by induction on $ k \in \{ 1, 2, \dots, K \} $. 
For the base case $ k = 1 $ 
\nobs that \cref{interpol_ANN_points:eq2} and \cref{interpol_ANN_points:eq3} 
\prove that for all $ x \in [ \fx_0, \fx_1 ] $ 
it holds that
\begin{equation}
\label{interpol_ANN_points:eq5}
  ( \Rr( \interpolatingDNN ) )( x ) 
= 
  ( \Rr( \interpolatingDNN ) )( \fx_0 ) 
  + 
  ( \Rr( \interpolatingDNN ) )( x ) 
  - 
  ( \Rr( \interpolatingDNN ) )( \fx_0 ) 
= 
  f_0 + 
  \pr*{ \frac{ f_1 - f_0 }{ \fx_1 - \fx_0 } } 
  \pr*{ x - \fx_0 } 
  .
\end{equation}
This \proves \cref{interpol_ANN_points:eq4} in the base case $ k = 1 $.
For the induction step \nobs that 
\cref{interpol_ANN_points:eq3} \proves that 
for all $ k \in \N \cap (1,\infty) \cap [ 1, K ] $, 
$ x \in [ \fx_{ k - 1 }, \fx_k ] $ 
with 
$
  \forall \, y \in [ \fx_{ k - 2 }, \fx_{ k - 1 } ] 
  \colon 
  ( \Rr( \interpolatingDNN ) )( y ) 
  = 
  f_{ k - 2 } 
  + 
  \bpr{\frac{ f_{ k - 1 } - f_{ k - 2 } }{ \fx_{ k - 1 } - \fx_{ k - 2 } } } ( y - \fx_{ k - 2 } )
$ 
it holds that
\begin{equation}\label{interpol_ANN_points:eq6}
\begin{split}
  ( \Rr( \interpolatingDNN ) )( x ) 
& 
  = 
  ( \Rr( \interpolatingDNN ) )( \fx_{ k - 1 } ) 
  + 
  ( \Rr( \interpolatingDNN ) )( x ) 
  - 
  ( \Rr( \interpolatingDNN ) )( \fx_{ k - 1 } )
\\
& 
  = 
  f_{ k - 2 } 
  + 
  \pr*{ \frac{ f_{ k - 1 } - f_{ k - 2 } }{ \fx_{ k - 1 } - \fx_{ k - 2 } } }
  \pr*{ \fx_{ k - 1 } - \fx_{ k - 2 } }
  + 
  \pr*{
    \frac{ 
      f_k - f_{ k - 1 } 
    }{
      \fx_k - \fx_{ k - 1 } 
    } 
  }
  \pr*{ 
    x - \fx_{ k - 1 } 
  }
\\ &
  = 
  f_{ k - 1 } 
  + 
  \pr*{ 
    \frac{ f_k - f_{ k - 1 } }{ \fx_k - \fx_{ k - 1 } } 
  }
  \pr*{ x - \fx_{ k - 1 } } 
  .
\end{split}
\end{equation}
Induction thus \proves \cref{interpol_ANN_points:eq4}.
\Moreover \eqref{interpol_ANN_points:setting1} and \eqref{interpol_ANN_points:eq1.1} 
\prove that 
\begin{equation}
\begin{split} 
  \sum_{ n = 0 }^K 
  c_n 
= 
  c_K + \sum_{ n = 0 }^{ K - 1 } c_n 
= 
  - \frac{ f_K - f_{ K - 1 } }{ \fx_K - \fx_{ K - 1 } } 
  + \frac{ f_K - f_{ K - 1 } }{ \fx_K - \fx_{ K - 1 } } 
=
  0 .
\end{split}
\end{equation}
The fact that 
for all $ k \in \{ 0, 1, \dots, K \} $ 
it holds that $ \fx_k \le \fx_K $ 
and \eqref{interpol_ANN_points:eq1} 
\hence \prove that for all $x\in[\fx_K,\infty)$ it holds that
\begin{equation}\label{interpol_ANN_points:eq7}
\begin{split}
(\Rr(\interpolatingDNN))(x) - (\Rr(\interpolatingDNN))(\fx_K) & = \br*{\sum_{n=0}^K c_n \pr*{\max\{x-\fx_n,0\} - \max\{\fx_K - \fx_n,0\}}}\\
& = \sum_{n=0}^K c_n[(x-\fx_n)-(\fx_K-\fx_n)] = \sum_{n=0}^K c_n(x-\fx_K) = 0.
\end{split}
\end{equation}
This and \cref{interpol_ANN_points:eq4} \prove that for all $x\in[\fx_K,\infty)$ it holds that
\begin{equation}\label{interpol_ANN_points:eq8}
(\Rr(\interpolatingDNN))(x) = (\Rr(\interpolatingDNN))(\fx_K) = f_{K-1} + \bpr{\tfrac{f_K - f_{K-1}}{\fx_K - \fx_{K-1}}}(\fx_K - \fx_{K-1}) = f_K.
\end{equation}
Combining this, \cref{interpol_ANN_points:eq2}, \cref{interpol_ANN_points:eq4}, and \cref{def:lin_interp:eq1} \proves[ep]
\cref{interpol_ANN_points:item3}.
\end{aproof}

\cfclear
\begin{exercise}{ex:approximate_1d_function}
	Prove or disprove the following statement: There exists $\Phi\in\ANNs$ such that
	$\paramANN(\Phi)\leq 16$ and
	\begin{equation}
		\sup_{x\in[-2\pi,2\pi]} \bigl\lvert{\cos(x)-(\functionANN{\rect}(\Phi))(x)}\bigr\rvert\leq\tfrac12
	\end{equation}
	\cfload.
\end{exercise}

\cfclear
\begin{exercise}{ex:represent_max4}
  Prove or disprove the following statement: There exists $\Phi\in \ANNs$ such that
  $\inDimANN(\Phi)=4$, $\outDimANN(\Phi)=1$,
  $\paramANN(\Phi)\leq 60$, and 
	$\forall\, x,y,u,v\in\R\colon (\functionANN{\rect}(\Phi))(x,y,u,v)=\max\{x,y,u,v\}$
	\cfload.
\end{exercise}

\cfclear
\begin{exercise}{ex:represent_max2m}
  Prove or disprove the following statement: For every $m\in\N$ there exists $\Phi\in \ANNs$ such that
  $\inDimANN(\Phi)=2^m$, $\outDimANN(\Phi)=1$,
  $\paramANN(\Phi)\leq 3(2^m(2^{m}+1))$, and 
	$\forall\, x=(x_1,x_2,\dots,x_{2^m})\in\R\colon (\functionANN{\rect}(\Phi))(x)=\max\{x_1,x_2,\dots,x_{2^m}\}$
	\cfload.
\end{exercise}

\section{ANN approximations results for one-dimensional functions}
\label{sect:ANNapproxoned}

\subsection{Constructive ANN approximation results}

\cfclear
\begin{athm}{prop}{interpol_ANN_function1}[ANN approximations through linear interpolations]
Let $ K \in \N $, 
$ L, a, \fx_0, \allowbreak \fx_1, \dots, \fx_K \allowbreak \in \R $, 
$ b \in (a,\infty) $
satisfy
for all $ k \in \{ 0, 1,\dots, K \} $ that
$
  \fx_k = a + \frac{ k ( b - a ) }{ K }
$,
let
$
  f \colon [a,b] \to \R
$ 
satisfy for all 
$ x, y \in [a,b] $ that 
\begin{equation}
  \abs{ f( x ) - f(y) }
  \le L \abs{ x - y }
  ,
\end{equation}
and let 
$
  \interpolatingDNN \in \ANNs
$ 
satisfy
\begin{equation}
\interpolatingDNN 
= 
\compANN{
	\AffineANN_{1,f(\fx_0)}
}{
	\pr*{{\smallbigANNsum_{k=0}^K} \pr*{
		\scalarMultANN{
			\pr*{\tfrac{K(f(\fx_{\min\{k+1,K\}})-2f(\fx_k) + f(\fx_{\max\{k-1,0\}}))}{(b-a)}}
		}{
			(\compANN{\ii_{1} }{ \AffineANN_{1,-\fx_k}})
		}
	}}
}
\end{equation}
\cfload.
Then 
\begin{enumerate}[label=(\roman *)]
\item
\label{interpol_ANN_function1:item1}
it holds that $\dims(\interpolatingDNN) = (1,K+1,1)$,
\item
\label{interpol_ANN_function1:item25}
it holds that
$
  \Rr(\interpolatingDNN)
=
  \interpol{
    \fx_0,\fx_1,\dots,\fx_K
  }{
    f(\fx_0), f(\fx_1), \ldots, f(\fx_K)
  }
$,
\item
\label{interpol_ANN_function1:item3}
it holds for all $ x, y \in \R $ that 
$
  \abs{(\Rr(\interpolatingDNN))(x) - (\Rr(\interpolatingDNN))(y)} \le L\abs{x-y}
$,
\item
\label{interpol_ANN_function1:item4}
it holds that 
$
  \sup_{ x \in [a,b] }
  \abs{
    ( \Rr( \interpolatingDNN ) )( x ) - f( x ) 
  } 
  \le 
  L ( b - a ) K^{ - 1 } 
$,
and
\item
\label{interpol_ANN_function1:item5}
it holds that 
$\paramANN(\interpolatingDNN) = 3K + 4 $
\end{enumerate}
\cfout.
\end{athm}

\begin{aproof}
\Nobs that the fact that for all 
$ k \in \{ 0, 1, \dots, K \} $
it holds that
\begin{equation}
  \fx_{ \min\{ k + 1, K \} } - \fx_{ \min\{ k, K - 1 \} } 
  = 
  \fx_{ \max\{ k, 1 \} } - \fx_{ \max\{ k - 1, 0 \} } 
  = (b - a) K^{ - 1 }
\end{equation}
\proves that for all 
$ k \in \{ 0, 1, \dots, K \} $ 
it holds that 
\begin{equation}
\label{interpol_ANN_function1:eq1}
\begin{split}
&
  \frac{
    ( 
      f( \fx_{ \min\{ k + 1, K \} } ) - f( \fx_k ) 
    ) 
  }{
    ( 
      \fx_{ \min\{ k + 1, K \} } - \fx_{ \min\{ k, K - 1 \} } 
    )
  } 
  - 
  \frac{
    ( 
      f( \fx_k ) - f( \fx_{ \max\{ k - 1, 0 \} } ) 
    )
  }{
    ( \fx_{ \max\{ k, 1 \} } - \fx_{ \max\{ k - 1, 0 \} } )
  } 
\\ & 
= 
  \frac{
    K ( f( \fx_{ \min\{ k + 1, K \} } ) - 2 f( \fx_k ) + f( \fx_{ \max\{ k - 1, 0 \} } ) ) 
  }{
    (b - a) 
  }
  .
\end{split}
\end{equation}
This and
\cref{interpol_ANN_points} prove
\cref{interpol_ANN_function1:item1,%
interpol_ANN_function1:item25,%
interpol_ANN_function1:item5}.
\Nobs that 
\cref{interpol_lipschitz:item1} in \cref{interpol_lipschitz},  
\cref{interpol_ANN_function1:item25}, 
and the assumption that 
for all $ x, y \in [a,b] $ it holds that 
\begin{equation}
  \abs{ f(x) - f(y) }
  \le L \abs{ x - y } 
\end{equation}
\prove[ep] \cref{interpol_ANN_function1:item3}. 
\Nobs that 
\cref{interpol_ANN_function1:item25}, 
the assumption that 
for all 
$ x, y \in [a,b] $
it holds that 
\begin{equation}
  \abs{f(x)-f(y)}\le L\abs{x-y}
  ,
\end{equation}
\cref{interpol_lipschitz:item2} in \cref{interpol_lipschitz}, and 
the fact that 
for all 
$ k \in \{ 1, 2, \dots, K \} $ 
it holds that 
\begin{equation}
  \fx_k - \fx_{ k - 1 } = \frac{ ( b - a ) }{ K }
\end{equation}
\prove that for all 
$
  x \in [a,b]
$ 
it holds that
\begin{equation}
  \abs{
    ( \Rr( \interpolatingDNN ) )( x ) - f(x) 
  } 
  \le 
  L 
  \pr*{ 
    \max_{ k \in \{ 1, 2, \dots, K \} }
    \abs{ \fx_k - \fx_{ k - 1 } } 
  } 
  = 
  \frac{ 
    L (b-a) 
  }{
    K
  }.
\end{equation}
This \proves[ep] \cref{interpol_ANN_function1:item4}.
\end{aproof}

\cfclear
\begin{athm}{lemma}{interpol_ANN_function_simple}[Approximations through \anns\ with constant realizations]
Let $L, a \in \R$, $b \in [a,\infty)$, $\xi\in[a,b]$, 
let $f\colon[a,b]\to\R$ satisfy for all $x,y\in[a,b]$ that 
\begin{equation}
  \abs{f(x)-f(y)} \le L\abs{x-y}
  ,
\end{equation}
and let $\interpolatingDNN \in \ANNs$ satisfy
\begin{equation}
  \interpolatingDNN 
  = 
  \compANN{\AffineANN_{1,f(\xi)}}{\pr*{ \scalarMultANN{0}{(\compANN{\ii_{1}}{\AffineANN_{1,-\xi}})} }}
\end{equation}\cfload. 
Then
\begin{enumerate}[label=(\roman *)]
\item
\label{interpol_ANN_function_simple:item1}
it holds that $\dims(\interpolatingDNN) = (1,1,1)$,
\item 
\label{interpol_ANN_function_simple:item2}
it holds that $\Rr(\interpolatingDNN) \in C(\R,\R)$,
\item
\label{interpol_ANN_function_simple:item3}
it holds for all $x\in\R$ that 
$(\Rr(\interpolatingDNN))(x) = f(\xi)$,
\item
\label{interpol_ANN_function_simple:item4}
it holds that $\sup_{x\in[a,b]}\abs{(\Rr(\interpolatingDNN))(x)-f(x)} \le 
L\max\{\xi - a,b - \xi\},$
and
\item
\label{interpol_ANN_function_simple:item5}
it holds that 
$\paramANN(\interpolatingDNN) = 4 $
\end{enumerate}
\cfout.
\end{athm}

\begin{aproof}
\Nobs that %
\cref{lem:ANN:affine2:item1,lem:ANN:affine2:item2} in
\cref{lem:ANN:affine2},
and \cref{basicANN:item3,basicANN:item2} in \cref{basicANN}
establish \cref{interpol_ANN_function_simple:item2,%
interpol_ANN_function_simple:item1}.
\Nobs that
\cref{lem:ANN:affine2:item3} in \cref{lem:ANN:affine2}
and
\cref{item:ANN:scalar:3} in \cref{lem:ANNscalar}
\prove that for all $x\in\R$ it holds that
\begin{equation}\label{non_rep_zero_1}
\begin{split}
	(\Rr(\interpolatingDNN))(x) 
&= 
	(\Rr( \scalarMultANN{0}{(\compANN{\ii_{1} }{ \AffineANN_{1,-\xi}}) }) )(x) + f(\xi) \\
&=
	0  \bpr{ (\Rr(\compANN{\ii_{1} }{ \AffineANN_{1,-\xi}}))(x) } + f(\xi) 
=
	f(\xi)
\end{split}
\end{equation}
\cfload.
This \proves[ep] \cref{interpol_ANN_function_simple:item3}.
\Nobs that \cref{non_rep_zero_1}, the fact that $\xi\in[a,b]$, 
and the assumption that for all $ x, y \in [a,b] $ 
it holds that 
\begin{equation}
  \abs{f(x)-f(y)} \le L\abs{x-y}
\end{equation}
\prove that
for all $x\in[a,b]$ it holds that
\begin{equation}
\abs{(\Rr(\interpolatingDNN))(x) - f(x)} = \abs{f(\xi) - f(x)} \le L\abs{x-\xi} \le L\max\{\xi - a,b - \xi\}.
\end{equation}
This \proves[ep] \cref{interpol_ANN_function_simple:item4}.
\Nobs that \cref{def:ANN:eq1} and \cref{interpol_ANN_function_simple:item1} \prove that 
\begin{equation}
\paramANN(\interpolatingDNN) = 1(1 + 1) + 1(1 + 1)  = 4.
\end{equation}
This \proves[ep] \cref{interpol_ANN_function_simple:item5}.
\end{aproof}

\cfclear
\begin{athm}{cor}{interpol_ANN_function2}[Explicit \ann\ approximations with prescribed error tolerances]
Let 
$\varepsilon \in (0,\infty)$, 
$L, a \in \R$, $b \in (a,\infty)$,
$K \in \N_0 \cap \big[\frac{L (b-a)}{\varepsilon}, \frac{L (b-a)}{\varepsilon} + 1\big)$,
$\fx_0,\fx_1,\dots,\fx_K \in\R$
satisfy
for all $k\in\{0,1,\dots,K\}$ that
$\fx_k = a + \frac{k(b-a)}{\max\{K,1\}}$,
let
$f \colon [a,b] \to \R$ satisfy for all $x,y\in [a,b]$ that 
\begin{equation}
  \abs{f(x)-f(y)}\le L\abs{x-y}
  ,
\end{equation}
and let
$\interpolatingDNN \in \ANNs$ satisfy
\begin{equation}
\interpolatingDNN 
= 
\compANN{
	\AffineANN_{1,f(\fx_0)}
}{
	\pr*{{\smallbigANNsum_{k=0}^K} \pr*{
		\scalarMultANN{
			\pr*{\tfrac{K(f(\fx_{\min\{k+1,K\}})-2f(\fx_k) + f(\fx_{\max\{k-1,0\}}))}{(b-a)}}
		}{
			(\compANN{\ii_{1}}{\AffineANN_{1,-\fx_k}})
		}
	}}
}
\end{equation}
\cfout.
Then 
\begin{enumerate}[label=(\roman *)]
\item
\label{interpol_ANN_function2:item1}
it holds that $\dims(\interpolatingDNN) = (1,K+1,1)$,
\item 
\label{interpol_ANN_function2:item2}
it holds that $\Rr(\interpolatingDNN) \in C(\R,\R)$,
\item
\label{interpol_ANN_function2:item3}
it holds for all $x,y\in\R$ that $\abs{(\Rr(\interpolatingDNN))(x) - (\Rr(\interpolatingDNN))(y)} \le L\abs{x-y}$,
\item
\label{interpol_ANN_function2:item4}
it holds that $\sup_{x\in[a,b]}\abs{(\Rr(\interpolatingDNN))(x)-f(x)} \le \frac{L(b-a)}{\max\{K,1\}} \le \varepsilon$,
and
\item
\label{interpol_ANN_function2:item5}
it holds that 
$\paramANN(\interpolatingDNN) = 3K+4 \leq 3L(b-a)\varepsilon^{-1} + 7 $
\end{enumerate}
\cfout.
\end{athm}

\begin{aproof}
\Nobs that 
the assumption that
$
  K \in \N_0 \cap \bigl[ \tfrac{L (b-a)}{\varepsilon}, \tfrac{L (b-a)}{\varepsilon} + 1\bigr)
$
\proves that 
\begin{equation}
  \frac{L(b-a)}{\max\{K,1\}} \le \varepsilon
  .
\end{equation}
This,
\cref{interpol_ANN_function1:item1,interpol_ANN_function1:item3,interpol_ANN_function1:item4}
in \cref{interpol_ANN_function1},
and
\cref{interpol_ANN_function_simple:item2,%
interpol_ANN_function_simple:item1,%
interpol_ANN_function_simple:item3,%
interpol_ANN_function_simple:item4}
in \cref{interpol_ANN_function_simple}
\prove[ep]
\cref{interpol_ANN_function2:item2,%
interpol_ANN_function2:item1,%
interpol_ANN_function2:item3,%
interpol_ANN_function2:item4}. 
\Nobs that
\cref{interpol_ANN_function1:item5} in \cref{interpol_ANN_function1}, 
\cref{interpol_ANN_function_simple:item5} in \cref{interpol_ANN_function_simple}, 
and the fact that 
\begin{equation}
  K \leq 1 + \frac{L(b-a)}{\varepsilon}
  ,
\end{equation}
\prove that 
\begin{equation}
\paramANN(\interpolatingDNN) = 3K + 4 \leq \frac{3L(b-a)}{\varepsilon} + 7.
\end{equation}
This \proves[ep] \cref{interpol_ANN_function2:item5}. 
\end{aproof}

\subsection{Convergence rates for the approximation error}

\begin{adef}{def:p-norm}[Quasi vector norms]
Let 
$d \in \N$,
$
  p \in ( 0, \infty ] 
$,
$\theta = (\theta_1, \dots, \theta_d) \in \R^d$.
Then we denote by
$ 
  \pnorm p \theta \in \R 
$
the real number given by
\begin{equation}
\textstyle 
  \pnorm p \theta 
  =
  \begin{cases}
    \br[\big]{
      \smallsum_{ i = 1 }^d
        \lvert \theta_i \rvert^p
    }^{ \nicefrac{ 1 }{ p } }
    &
    \colon p \in (0, \infty) \\
    \max_{ i \in \{ 1, 2, \ldots, d \} }
      \lvert \theta_i \rvert
    &
    \colon p = \infty.
  \end{cases}
\end{equation}
\end{adef}

\cfclear
\begin{athm}{cor}{interpol_ANN_function_implicit}[Implicit one-dimensional \ann\ approximations with prescribed error tolerances and explicit parameter bounds]
Let $ \varepsilon \in (0,\infty) $, 
$ L \in [0,\infty) $, 
$ a \in \R $, $ b \in [a,\infty) $
and
let
$f \colon [a,b] \to \R$ satisfy for all $x,y\in [a,b]$ that 
\begin{equation}
  \abs{f(x)-f(y)}\le L\abs{x-y}
  .
\end{equation}
Then there exists $\interpolatingDNN\in\ANNs$ such that
\begin{enumerate}[label=(\roman *)]
\item 
\label{interpol_ANN_function_implicit:item1}
it holds that $\Rr(\interpolatingDNN) \in C(\R,\R)$,
\item
\label{interpol_ANN_function_implicit:item2}
it holds that $\hiddenLength(\interpolatingDNN) = 1$,
\item
\label{interpol_ANN_function_implicit:item3}
it holds that $\dimANNlevel_1(\interpolatingDNN) \le L(b-a)\varepsilon^{-1} + 2$,
\item
\label{interpol_ANN_function_implicit:item4}
it holds for all $x,y\in\R$ that $\abs{(\Rr(\interpolatingDNN))(x) - (\Rr(\interpolatingDNN))(y)} \le L\abs{x-y}$,
\item
\label{interpol_ANN_function_implicit:item5}
it holds that $\sup_{x\in[a,b]}\abs{(\Rr(\interpolatingDNN))(x)-f(x)} \le \varepsilon$,
\item
\label{interpol_ANN_function_implicit:item6}
it holds that 
$\paramANN(\interpolatingDNN) = 3(\dimANNlevel_1(\interpolatingDNN)) + 1 \leq 3L(b-a)\varepsilon^{-1} + 7 $,
and
\item
\label{interpol_ANN_function_implicit:item7}
it holds that 
$\infnorm{\MappingStructuralToVectorized(\interpolatingDNN)}\leq\max\{1,\abs a,\abs b,2L,\abs{f(a)}\}$
\end{enumerate}
\cfout.
\end{athm}

\begin{aproof}
Throughout this proof, assume without loss of generality that $ a < b $,
let 
$ 
  K \in \N_0 \cap \bigl[\frac{L (b-a)}{\varepsilon}, \frac{L (b-a)}{\varepsilon} + 1\bigr) 
$, 
$ 
  \fx_0, \fx_1, \dots, \fx_K \in [ a, b ] 
$, 
$ 
  c_0, c_1, \dots, c_K \in \R 
$
satisfy for all $k\in\{0,1,\dots,K\}$ that
\begin{equation}
\label{interpol_ANN_function_implicit:setting1}
  \fx_k = a + \frac{k(b-a)}{\max\{K,1\}}
\qandq
  c_k
  =
  \frac{ 
    K ( 
      f( \fx_{ \min\{ k + 1, K \} } ) - 2 f( \fx_k ) + f( \fx_{ \max\{ k - 1, 0 \} } ) 
    ) 
  }{ 
    ( b - a ) 
  }
  ,
\end{equation}
and let $\interpolatingDNN\in\ANNs$ satisfy 
\begin{equation}
	\interpolatingDNN 
	= 
	\compANN{
		\AffineANN_{1,f(\fx_0)}
	}{
		\pr*{{\smallbigANNsum_{k=0}^K} \pr*{
			\scalarMultANN{
				c_k
			}{
				(\compANN{\ii_{1} }{ \AffineANN_{1,-\fx_k}})
			}
		}}
	}
\end{equation}
\cfload.
\Nobs that
\cref{interpol_ANN_function2} \proves that 
\begin{enumerate}[label=(\Roman *)]
\item
\label{interpol_ANN_function_implicit:pfitem1}
it holds that $\dims(\interpolatingDNN) = (1,K+1,1)$,
\item 
\label{interpol_ANN_function_implicit:pfitem2}
it holds that $\Rr(\interpolatingDNN) \in C(\R,\R)$,
\item
\label{interpol_ANN_function_implicit:pfitem3}
it holds for all $x,y\in\R$ that $\abs{(\Rr(\interpolatingDNN))(x) - (\Rr(\interpolatingDNN))(y)} \le L\abs{x-y}$,
\item
\label{interpol_ANN_function_implicit:pfitem4}
it holds that $\sup_{x\in[a,b]}\abs{(\Rr(\interpolatingDNN))(x)-f(x)} \le \varepsilon$,
and
\item
\label{interpol_ANN_function_implicit:pfitem5}
it holds that 
$\paramANN(\interpolatingDNN) = 3K+4$
\end{enumerate}
\cfload. 
This \proves[ep]
\cref{interpol_ANN_function_implicit:item1,interpol_ANN_function_implicit:item4,interpol_ANN_function_implicit:item5}.
\Nobs that \cref{interpol_ANN_function_implicit:pfitem1} 
and the fact that 
\begin{equation}
\label{eq:K_bound_for_item_ii}
  K \leq 1 + \frac{L(b-a)}{\varepsilon}
\end{equation}
prove \cref{interpol_ANN_function_implicit:item2,interpol_ANN_function_implicit:item3}.
\Nobs that 
\cref{interpol_ANN_function_implicit:item3}
and 
\cref{interpol_ANN_function_implicit:pfitem1,interpol_ANN_function_implicit:pfitem5}
\prove that
\begin{equation}
\label{interpol_ANN_function_implicit:eq2}
\begin{split} 
  \paramANN(\interpolatingDNN) 
= 
  3K+4
=
  3(K+1) + 1
=
  3 (\dimANNlevel_1(\interpolatingDNN)) + 1 
\leq
  \frac{ 3 L (b-a) }{ \varepsilon } + 7 .
\end{split}
\end{equation}
This \proves[ep] \cref{interpol_ANN_function_implicit:item6}. 
\Nobs that
\cref{basicANN} 
\proves that for all
$ 
  k \in \{ 0, 1, \dots, K \} 
$
it holds that
\begin{equation}
  \scalarMultANN{ c_k }{
    ( \compANN{ \ii_{1} }{ \AffineANN_{ 1, - \fx_k } } ) 
  }
  =
  ( ( 1, - \fx_k ), ( c_k, 0 ) )
  .
\end{equation}
Combining this with \eqref{eq:ANNsum:same}, \eqref{eq:ANN:extension}, \eqref{eq:ANN:sum}, 
and \cref{eq:defCompANN}
\proves that
\begin{multline}
\label{eq:interpol_ANN_function_implicit.1}
	\interpolatingDNN
	=
	\pr*{\!
		\pr*{\!
			\pmat{1\\1\\\vdots\\1}
			,
			\pmat{-\fx_0\\-\fx_1\\\vdots\\-\fx_K}\!
		},
		\pr*{
			\pmat{c_0&c_1&\cdots&c_K},f(\fx_0)
		}
	}\\
	\in
	(\R^{(K+1)\times 1}\times\R^{K+1})\times(\R^{1\times(K+1)}\times\R)
	.
\end{multline}
\Cref{lem:structtovect} 
\hence \proves that
\begin{equation}
	\label{eq:interpol_ANN_function_implicit.2}
	\infnorm{\MappingStructuralToVectorized(\interpolatingDNN)}
	=
	\max\{\abs{\fx_0},\abs{\fx_1},\dots,\abs{\fx_K},\abs{c_0},\abs{c_1},\dots,\abs{c_K},\abs{f(\fx_0)},1\}
	\ifnocf.
\end{equation}
\cfload[.]
\Moreover 
\cref{interpol_ANN_function_implicit:setting1},
the assumption that
for all $ x, y \in [a,b] $ it holds that 
\begin{equation}
  \abs{ f(x) - f(y) } 
  \le L \abs{ x - y },
\end{equation}
and the fact that 
for all $ k \in \N \cap (0,K+1) $
it holds that 
\begin{equation}
  \fx_{k} - \fx_{k-1} = 
  \frac{ ( b - a ) }{ 
    \max\{K,1\}
  }
\end{equation}
\prove that for all
	$k\in\{0,1,\dots,K\}$
it holds that
\begin{equation}
	\begin{split}
	\abs{c_k}
	&\leq
	\frac{K(\abs{f(\fx_{\min\{k+1,K\}})-f(\fx_k)} + \abs{f(\fx_{\max\{k-1,0\}}))-f(\fx_k)}}{(b-a)}
	\\&\leq
	\frac{KL(\abs{\fx_{\min\{k+1,K\}}-\fx_k} + \abs{\fx_{\max\{k-1,0\}}-\fx_k})}{(b-a)}
	\\&\leq
	\frac{2KL(b-a)[\max\{K,1\}]^{-1}}{(b-a)}
	\leq
	2L
	.
	\end{split}
\end{equation}
	This
	and \eqref{eq:interpol_ANN_function_implicit.2}
\prove[ep]
  \cref{interpol_ANN_function_implicit:item7}.
\end{aproof}

\cfclear
\begingroup
\renewcommand{\c}{\fC}
\begin{athm}{cor}{interpol_ANN_function_implicit2}[Implicit one-dimensional \ann\ approximations with prescribed error tolerances and asymptotic parameter bounds]
Let 
$
  L, a \in \R
$, 
$
  b \in [a,\infty)
$
and
let
$
  f \colon [a,b] \to \R
$ 
satisfy for all $ x, y \in [a,b] $ 
that 
\begin{equation}
  \abs{ f(x) - f(y) }
  \le L \abs{ x - y } 
  .
\end{equation}
Then there exists
$ \c \in \R $ 
such that for all
$
	\varepsilon \in (0,1]
$
there exists
$ 
  \interpolatingDNN \in \ANNs 
$ 
such that
\begin{gather}
  \Rr( \interpolatingDNN ) \in C(\R,\R)
  ,
  \qquad
  \sup\nolimits_{ x \in [a,b] } 
  \abs{ ( \Rr(\interpolatingDNN ) )( x ) - f( x ) } 
  \le 
  \varepsilon
  ,
  \qquad
  \hiddenLength( \interpolatingDNN ) = 1
  ,\\
  \infnorm{
    \MappingStructuralToVectorized( \interpolatingDNN ) 
  }
  \leq
  \max\{
    1, \abs a, \abs b, 2 L, \abs{ f(a) }
  \}
  ,
  \qandq
  \paramANN( \interpolatingDNN ) 
  \le 
  \c \varepsilon^{ - 1 }
\end{gather}
\cfout.
\end{athm}

\begin{aproof}
Throughout this proof, 
assume without loss of generality that 
$ a < b $ 
and let 
\begin{equation}
\label{eq:definition_of_constant_in_proof}
  \c = 3 L ( b - a ) + 7 
  .
\end{equation}
\Nobs that the assumption that $ a < b $ 
\proves that 
$ 
  L \geq 0 
$. 
\Moreover 
\cref{eq:definition_of_constant_in_proof} \proves that 
for all
$ \varepsilon \in (0,1] $
it holds that 
\begin{equation}
\begin{split} 
  3L(b-a)\varepsilon^{-1} + 7 \leq 3L(b-a)\varepsilon^{-1} + 7 \varepsilon^{-1} = \c \varepsilon^{-1}.
\end{split}
\end{equation}
This and \cref{interpol_ANN_function_implicit} \prove that 
 for all
$
	\varepsilon \in (0,1]
$
there exists
$ 
  \interpolatingDNN \in \ANNs 
$ 
such that 
\begin{gather}
\textstyle 
  \Rr( \interpolatingDNN ) \in C( \R, \R )
  ,
\qquad 
  \sup_{ x \in [a,b] }
  \abs{
    ( \Rr( \interpolatingDNN ) )( x ) - f( x ) 
  } 
  \le \varepsilon
  ,
\qquad 
  \hiddenLength( \interpolatingDNN ) = 1 ,
\\
  \infnorm{\MappingStructuralToVectorized(\interpolatingDNN)}
  \leq\max\{1,\abs a,\abs b,2L,\abs{f(a)}\}
  ,
\quad\text{and}\quad
  \paramANN( \interpolatingDNN ) 
  \le 3 L (b-a) \varepsilon^{ - 1 } + 7 
  \leq \c \varepsilon^{ - 1 }
\end{gather}
\cfload. 
\end{aproof}
\endgroup

\cfclear
\begingroup
\renewcommand{\c}{\fC}
\begin{athm}{cor}{interpol_ANN_function_implicit3}[Implicit one-dimensional \ann\ approximations with prescribed error tolerances and asymptotic parameter bounds]
Let 
$L, a \in \R$, $b \in [a,\infty)$
and
let
$f \colon [a,b] \to \R$ satisfy for all $x,y\in [a,b]$ that 
\begin{equation}
  \abs{f(x)-f(y)}\le L\abs{x-y}
  .
\end{equation}
Then there exists
	$ \c \in \R $ 
such that for all 
	$ \varepsilon \in (0,1] $ 
there exists
$ 
  \interpolatingDNN \in \ANNs 
$ 
such that
\begin{equation}
	\llabel{claim}
	\begin{gathered}
	\Rr( \interpolatingDNN ) \in C(\R,\R)
	,\quad
	\sup\nolimits_{ x \in [a,b] } \abs{ ( \Rr(\interpolatingDNN ) )( x ) - f( x ) } 
  \le 
	\varepsilon
	,\quad\text{and}\quad
	\paramANN( \interpolatingDNN ) 
  \le 
  \c \varepsilon^{ - 1 }
	\end{gathered}
\end{equation}
\cfout.
\end{athm}

\begin{aproof}
\Nobs that \cref{interpol_ANN_function_implicit2} \proves[ep] \lref{claim}. 
\end{aproof}
\endgroup

\cfclear
\begin{exercise}{ex:approximation}
Let $f \colon [-2,3] \to \R$ satisfy for all 
	$x \in [-2,3]$ 
that
\begin{equation}
\begin{split} 
	f(x)
=
	x^2 + 2 \sin(x).
\end{split}
\end{equation}
Prove or disprove the following statement:
There exist $ c \in \R $ and 
$ 
  \interpolatingDNN = ( \interpolatingDNN_{ \varepsilon } )_{ \varepsilon \in (0,1] } \colon (0,1] \to \ANNs 
$ such that for all $ \varepsilon \in (0,1] $ 
it holds that
\begin{equation}
	\Rr( \interpolatingDNN_{ \varepsilon } ) \in C(\R,\R)
	,\quad
	\sup\nolimits_{ x \in [-2,3] } \abs{ ( \Rr(\interpolatingDNN_{ \varepsilon } ) )( x ) - f( x ) } 
  \le 
	\varepsilon
	,\quad\text{and}\quad
	\paramANN( \interpolatingDNN_{ \varepsilon } ) 
  \le 
  c \varepsilon^{ - 1 }
\end{equation}
\cfload.
\end{exercise}

\cfclear
\begin{exercise}{ex:approximate_1d_function3}
  Prove or disprove the following statement: There exists
  $\Phi\in\ANNs$ such that $\paramANN(\Phi)\leq 10$ and
  \begin{equation}
    \sup_{x\in[0,10]}\bigl\lvert{\sqrt{x}-(\functionANN{\rect}(\Phi))(x)}\bigr\rvert\leq\tfrac14
  \end{equation}
  \cfload.
\end{exercise}

%% file: parts/Multi-dimensional_ANN_approximation_results.tex
\cchapter{Multi-dimensional ANN approximation results}{sect:multidApprox}

In this chapter we review basic deep \ReLU\ \ann\ approximation results for possibly multi-dimensional target functions.
We refer to the beginning of \cref{sect:onedApprox} for a small selection of \ann\ approximation results from the literature.
The specific presentation of this chapter is strongly based on 
\cite[Sections 2.2.6, 2.2.7, 2.2.8, and 3.1]{Beck2019published}, 
\cite[Sections 3 and 4.2]{JentzenRiekert2023}, and 
\cite[Section 3]{JentzenWelti2023}.

\section{Approximations through supremal convolutions}

\cfclear
\cfconsiderloaded{def:metric}
\begin{adef}{def:metric}[Metric] 
We say that $ \delta $ is a metric on $ E $ 
if and only if 
it holds that 
$ \delta \colon E \times E \to [0,\infty) $ 
is a function from 
$ E \times E $
to 
$ [0,\infty) $
which satisfies that
\begin{enumerate}[label=(\roman{*})]
\item 
it holds that 
\begin{equation}
	\textstyle
  \{ 
    (x,y) \in E^2 
    \colon 
    \delta( x, y ) = 0
  \}
  =
  \bigcup_{ 
    x \in E
  }
  \{ ( x, x ) \}
\end{equation}
(positive definiteness), 
\item 
it holds for all $ x, y \in E $ that 
\begin{equation}
  \delta(x,y) = \delta(y,x) 
\end{equation}
(symmetry), 
and 
\item 
it holds for all $ x, y, z \in E $ that 
\begin{equation}
  \delta(x,z) 
  \leq 
  \delta(x,y)
  +
  \delta(y,z)
\end{equation}
(triangle inequality).
\end{enumerate}
\end{adef}

\cfclear
\cfconsiderloaded{def:metric_space}
\begin{adef}{def:metric_space}[Metric space]
We say that $ \cE $ is a metric space if 
and only if there exist a set $ E $ and a \cfadd{def:metric}metric $ \delta $ on $ E $ 
such that 
\begin{equation}
  \cE = ( E, \delta ) 
\end{equation}
\cfload. 
\end{adef}

\cfclear
\begin{prop}[Approximations through supremal convolutions]
\label{lem:lipschitz_extension_prelim}
Let 
$(E,\delta)$ 
be a metric space\cfadd{def:metric_space}, 
let
$L \in[0,\infty)$, 
let 
$D \subseteq E$ and
$\cM \subseteq D$
satisfy
$ \cM \neq \emptyset  $, 
let 
$f\colon D \to \R$
satisfy for all 
$x \in D$,
$y\in \cM$
that 
$\abs{f(x)-f(y)}\leq L\delta(x,y)$, 
and let 
$F\colon E \to \R\cup\{\infty\}$ 
satisfy for all 
$x\in E$ 
that 	
\begin{equation} \label{lipschitz_extension_prelim:definition}
F(x) 
= 
\sup_{y\in\cM} \br*{
f(y) - L \delta(x,y)
}
\end{equation} 	
\cfload.
Then 
\begin{enumerate}[label=(\roman{*})]
\item \label{lipschitz_extension_prelim:item1}
it holds for all 
$x\in\cM$ 
that 
$F(x) = f(x)$,

\item \label{lipschitz_extension_prelim:item2}
it holds for all $x \in D$ that $F(x) \leq f(x)$, 

\item \label{lipschitz_extension_prelim:item3}
it holds for all $x \in E$ that $F(x) < \infty$,

\item \label{lipschitz_extension_prelim:item4}
it holds for all 
$x,y\in E$ 
that 
$\abs{F(x)-F(y)}\leq L \delta(x,y)$,
and

\item  \label{lipschitz_extension_prelim:item5}	
it holds for all 
$x\in D$ 
that 
\begin{equation} 
\abs{F(x)-f(x)} 
\leq 
2L\br*{ \inf_{y\in\cM} \delta(x,y) }.
\end{equation}

\end{enumerate}
\end{prop}

\begin{proof}[Proof of \cref{lem:lipschitz_extension_prelim}]
First, \nobs that the assumption that 
for all $ x \in D $, $ y \in \cM $
it holds that 
$
  \abs{ f(x) - f(y) } 
  \leq L \delta( x, y )
$ 
ensures that for all 
$ x \in D $, $ y \in \cM $
it holds that 
\begin{equation} 
\label{lipschitz_extension_prelim:eq1}
	f(y) + L \delta(x,y) \geq f(x) \geq f(y) - L \delta(x,y). 	
\end{equation} 
Hence, we obtain that for all 
	$x\in D$ 
it holds that 
\begin{equation} 
\label{lipschitz_extension_prelim:eq2}
f(x) \geq \sup_{y\in\cM} \br*{ f(y) - L \delta(x,y) } 
= 
F(x). 
\end{equation} 
This establishes \cref{lipschitz_extension_prelim:item2}.
Moreover, note that \cref{lipschitz_extension_prelim:definition} implies that for all 
	$x\in\cM$ 
it holds that
\begin{equation} 
	F(x) \geq f(x) - L \delta(x,x) = f(x). 
\end{equation} 
This and \eqref{lipschitz_extension_prelim:eq2} 
establish \cref{lipschitz_extension_prelim:item1}.
\Nobs that \eqref{lipschitz_extension_prelim:eq1} 
(applied for every $ y, z \in \mc M $ with $x\is y$, $y\is z$ 
in the notation of \cref{lipschitz_extension_prelim:eq1}) 
and the triangle inequality ensure that for all
	$x \in E$, $y, z \in \cM$
it holds that
\begin{equation}
\begin{split} 
	f(y) - L \delta(x,y) 
\leq
	f(z) + L \delta(y, z) - L \delta(x,y) 
\leq
	f(z) + L \delta(x, z).
\end{split}
\end{equation}	
Hence, we obtain that for all 
	$x \in E$, $z \in \cM$
it holds that
\begin{equation}
\begin{split} 
	F(x) 
= 
	\sup_{y\in\cM} \br*{
	f(y) - L \delta(x,y)
	}
\leq  
	f(z) + L \delta(x, z)
< 
	\infty.
\end{split}
\end{equation}
This and the assumption that $\cM\neq\emptyset$ 
prove \cref{lipschitz_extension_prelim:item3}. 
Note that \cref{lipschitz_extension_prelim:item3}, \cref{lipschitz_extension_prelim:definition}, and the triangle inequality 
show that for all
$x,y\in E$
it holds that 
\begin{equation} 
\begin{split}
F(x)-F(y) 
& = 
\biggl[
\sup_{v\in\cM} ( 
f(v) - L\delta(x,v)
)
\biggr]
- 
\biggl[
\sup_{w\in\cM} 
(
f(w) - L\delta(y,w)
)
\biggr]
\\
& = 
\sup_{v\in\cM} 
\br*{ 
f(v) - L \delta(x,v) 
- 
\sup_{w\in\cM} 
( f(w) - L \delta(y,w) )
} 
\\
& \leq 
\sup_{v\in\cM} 
\bigl[
f(v) - L \delta(x,v) 
- ( f(v) - L\delta(y,v) )
\bigr] 
\\ & = 
\sup_{v\in\cM} 
(L\delta(y,v) - L\delta(x,v) )
\\&\leq 
\sup_{v\in\cM} 
(L\delta(y,x) +L\delta(x,v) - L\delta(x,v) )
=
L \delta(x,y)
. 
\end{split}
\end{equation} 
This and the fact that for all 
$x,y\in E$ 
it holds that 
$\delta(x,y)=\delta(y,x)$
establish \cref{lipschitz_extension_prelim:item4}. 
Observe that
\cref{lipschitz_extension_prelim:item1,lipschitz_extension:item4}, 
the triangle inequality, and 
the assumption that 
$\forall \, x\in D,
y\in\cM
\colon\abs{f(x)-f(y)}\leq L\delta(x,y)$ 
ensure that for all 
$x\in D$ 
it holds that 
\begin{equation} 
\begin{split}
\abs{F(x)-f(x)} 
& = 
\inf_{y\in\cM} 
\abs{F(x)-F(y)+f(y)-f(x)} 
\\ & \leq 
\inf_{y\in\cM} 
\pr*{ \abs{F(x)-F(y)} + \abs{f(y)-f(x)} } 
\\
& \leq 
\inf_{y\in\cM} 
(2L \delta(x,y)) 
= 
2L\biggl[\inf_{y\in\cM} \delta(x,y)\biggr]
.
\end{split}
\end{equation} 
This establishes \cref{lipschitz_extension_prelim:item5}.
The proof of \cref{lem:lipschitz_extension_prelim} is thus complete.
\end{proof}

\begin{cor}[Approximations through supremum convolutions] \label{lem:lipschitz_extension}
		Let 
		$(E,\delta)$ 
    be a metric space\cfadd{def:metric_space}, 
    let
		$L\in [0,\infty)$, 
    let
		$\cM \subseteq E$
		satisfy
		$\cM\neq\emptyset$,
		let 
		$f\colon E \to \R$ 
		satisfy for all 
		$x\in E$,
		$y\in\cM$
		that 
		$\abs{f(x)-f(y)}\leq L\delta(x,y)$, 
		and let 
		$F\colon E \to \R\cup\{\infty\}$ 
		satisfy for all 
		$x\in E$ 
		that 	
		\begin{equation} \label{lipschitz_extension:definition}
		F(x) 
		= 
		\sup_{y\in\cM} \br*{
		f(y) - L \delta(x,y)
		}
		\end{equation}
    \cfload.
		Then 
		\begin{enumerate}[label=(\roman{*})]
			\item \label{lipschitz_extension:item1}
			it holds for all 
			$x\in\cM$ 
			that 
			$F(x) = f(x)$, 
			\item \label{lipschitz_extension:item2}
			it holds for all 
			$x\in E$ 
			that 
			$F(x) \leq f(x)$, 
			\item \label{lipschitz_extension:item3}
			it holds for all 
			$x,y\in E$ 
			that 
			$\abs{F(x)-F(y)}\leq L \delta(x,y)$,    
			and 
			\item  \label{lipschitz_extension:item4}	
			it holds for all 
			$x\in E$ 
			that 
			\begin{equation} 
			\abs{F(x)-f(x)} 
			\leq 
			2L\br*{ \inf_{y\in\cM} \delta(x,y) }. 
			\end{equation} 	
		\end{enumerate}
	\end{cor} 
  \begin{proof}[Proof of \cref{lem:lipschitz_extension}]
  Note that \cref{lem:lipschitz_extension_prelim} establishes \cref{lipschitz_extension:item1,lipschitz_extension:item2,lipschitz_extension:item3,lipschitz_extension:item4}.
  The proof of \cref{lem:lipschitz_extension} is thus complete. 
	\end{proof}

\begin{exercise}{ex:approximate_2d_function}
	Prove or disprove the following statement:
	There exists $\Phi\in\ANNs$ such that
	$\inDimANN(\Phi)=2$, $\outDimANN(\Phi)=1$, $\paramANN(\Phi)\leq 3\,000\,000\,000$, and
	\begin{equation}
		\sup_{x,y\in[0,2\pi]}\abs{\sin(x)\sin(y)-(\functionANN{\rect}(\Phi))(x,y)}\leq \tfrac15.
	\end{equation}
\end{exercise}

\section{ANN representations} \label{sec:dnnrep}

\subsection{ANN representations for the 1-norm} \label{subsec:dnnl1}

\cfclear
\cfconsiderloaded{def:dnn:l1norm}
\begin{adef}{def:dnn:l1norm}[$1$-norm \ann\ representations]
We denote by $(\oneNormANN_d)_{d \in \N} \subseteq \ANNs$ 
the fully-connected feedforward \anns\ which satisfy that
\begin{enumerate}[label=(\roman{*})]
    \item it holds that \begin{equation}
    \label{def:dnn:l1norm:eq1}
 \oneNormANN_1 = \pr*{\!\pr*{\! \begin{pmatrix}
    1 \\ -1
    \end{pmatrix}, \begin{pmatrix}
    0 \\ 0
    \end{pmatrix}\! }, \pr*{ \begin{pmatrix}
    1 & 1
    \end{pmatrix}, \begin{pmatrix}
    0 
    \end{pmatrix} }\!} \in (\R^{2 \times 1} \times \R^{2}) \times (\R^{1 \times 2} \times \R^{1}) 
\end{equation}
and
\item it holds for all $ d \in \{2,3,4, \dots\} $ that 
$ \oneNormANN_d =  \sumANN_{1,d} \compANNbullet \parallelizationSpecial_d(\oneNormANN_1, \oneNormANN_1, \ldots, \oneNormANN_1) $
\end{enumerate}
\cfload.
\end{adef}

\cfclear
\begin{prop}[Properties of 
$ 1 $-norm \anns] 
\label{prop:dnn:l1norm} 
Let $d \in \N$. Then
\begin{enumerate}[label=(\roman{*})]
    \item \label{item:prop:l1:1} it holds that $\dims (\oneNormANN_d) = (d, 2d, 1)$,
    \item \label{item:prop:l1:2} it holds that $\functionANN\rect(\oneNormANN_d) \in C(\R^d, \R)$, and 
    \item \label{item:prop:l1:3} it holds for all $x \in \R^d$ that $(\functionANN\rect(\oneNormANN_d))(x) = \pnorm1{x}$
\end{enumerate} \cfout.
\end{prop}

\begin{proof} [Proof of \cref{prop:dnn:l1norm}]
\Nobs that the fact that $ \dims( \oneNormANN_1 ) = (1,2,1) $ and \cref{Lemma:ParallelizationElementary} show that 
\begin{equation}
  \dims(\parallelizationSpecial_d(\oneNormANN_1, \oneNormANN_1, \dots, \oneNormANN_1) ) = 
  (d, 2 d, d) 
\end{equation} 
\cfload. 
Combining this, \cref{Lemma:PropertiesOfCompositions}, and \cref{lem:ANN:affine} ensures 
that 
\begin{equation}
  \dims(
    \oneNormANN_d
  )
  =
  \dims\bigl( 
    \sumANN_{ 1, d } \compANNbullet \parallelizationSpecial_d( \oneNormANN_1, \oneNormANN_1, \dots, \oneNormANN_1 )
  \bigr) 
  = ( d, 2 d, 1)
\end{equation}
\cfload. 
This establishes item~\ref{item:prop:l1:1}. 
\Nobs that \eqref{def:dnn:l1norm:eq1} assures that for all $x \in \R$ it holds that
\begin{equation}
    (\functionANN\rect(\oneNormANN_1))(x) = \rect(x) + \rect(-x) = \max \{x, 0\} + \max \{-x, 0\} = \abs{x} = \pnorm1{ x }
\end{equation}
\cfload.
Combining this and \cref{Lemma:PropertiesOfParallelizationEqualLength} shows that for all 
$ x = (x_1, \dots, x_d) \in \R^d $ 
it holds that
\begin{equation}
  \bigl( 
    \functionANN\rect(
      \parallelizationSpecial_d( \oneNormANN_1, \oneNormANN_1, \dots, \oneNormANN_1 ) 
    ) 
  \bigr)( x ) 
  = 
  (
    \abs{ x_1 } , \abs{x_2}, \dots, \abs{ x_d } 
  )
  .
\end{equation}
This and \cref{lem:def:ANNsum} demonstrate that for all 
$ 
  x = ( x_1, \dots, x_d ) \in \R^d 
$ 
it holds that
\begin{equation}
\begin{split}
  ( \functionANN\rect( \oneNormANN_d ) )( x ) 
& = 
  \bigl( 
    \functionANN\rect( 
      \sumANN_{ 1, d } \compANNbullet \parallelizationSpecial_d( \oneNormANN_1, \oneNormANN_1, \dots, \oneNormANN_1 ) 
    ) 
  \bigr)( x ) 
\\
&
  = 
  \bigl(
    \functionANN\rect( \sumANN_{ 1, d } ) 
  \bigr)( 
    \abs{ x_1 }, \abs{ x_2 }, \dots, \abs{ x_d } 
  )
=
  \smallsum_{ k = 1 }^d \abs{ x_k } 
= 
  \pnorm1{ x } 
  .
\end{split}
\end{equation}
This establishes \cref{item:prop:l1:2,item:prop:l1:3}. The proof of \cref{prop:dnn:l1norm} is thus complete.
\end{proof}

\cfclear
\begin{lemma} \label{lem:l1norm:param}
Let $d \in \N$. Then
\begin{enumerate}[label=(\roman{*})] 
\item \label{item:lem:l1norm:1} it holds that $\biasANN1{\oneNormANN_d}=0 \in \R^{2d}$,
\item \label{item:lem:l1norm:1a} it holds that $\biasANN2{\oneNormANN_d} = 0 \in \R$,
    \item \label{item:lem:l1norm:2} it holds that $\weightANN1{\oneNormANN_d}  \in \{-1, 0, 1\} ^{(2d) \times d}$,
    \item \label{item:lem:l1norm:3} it holds for all $x \in \R^d$ that $\pnorm\infty{\weightANN1{\oneNormANN_d} x } = \pnorm\infty{ x } $, and
    \item \label{item:lem:l1norm:4} it holds that $\weightANN2{\oneNormANN_d} = \begin{pmatrix} 1 & 1 & \cdots & 1\end{pmatrix} \in \R^{1 \times (2d)}$
\end{enumerate}
\cfout.
\end{lemma}

\begin{proof}[Proof of \cref{lem:l1norm:param}]
Throughout this proof, assume without loss of generality that $d > 1$.
Note that the fact that $\biasANN1{\oneNormANN_1} = 0 \in \R^2$, 
the fact that $\biasANN2{\oneNormANN_1}=0 \in \R$, 
the fact that 
$ \biasANN1{\sumANN_{1,d}} \allowbreak = 0 \in \R$, and 
the fact
that $ \oneNormANN_d =  \sumANN_{1,d} \compANNbullet \parallelizationSpecial_d(\oneNormANN_1, \oneNormANN_1, \ldots, \oneNormANN_1)$
 establish \cref{item:lem:l1norm:1,item:lem:l1norm:1a} \cfload.
In addition, observe that the fact that
\begin{equation}
\label{lem:l1norm:param:eq1}
    \weightANN{1}{\oneNormANN_1} = \begin{pmatrix}
    1 \\ -1
    \end{pmatrix}
    \qandq
    \weightANN{1}{\oneNormANN_d} = \begin{pmatrix}
     \weightANN{1}{\oneNormANN_1} & 0 & \cdots & 0 \\
    0 & \weightANN{1}{\oneNormANN_1} & \cdots & 0 \\
    \vdots & \vdots & \ddots & \vdots \\
    0 & 0 &\cdots & \weightANN{1}{\oneNormANN_1}
    \end{pmatrix} \in \R^{(2d) \times d}
\end{equation}
proves item~\ref{item:lem:l1norm:2}.
Next note that \eqref{lem:l1norm:param:eq1} implies item~\ref{item:lem:l1norm:3}. 
Moreover, note that the fact that $\weightANN2{\oneNormANN_1} = (1\; 1)$ and the fact that 
$  \oneNormANN_d =  \sumANN_{1,d} \compANNbullet \parallelizationSpecial_d(\oneNormANN_1, \oneNormANN_1, \ldots, \oneNormANN_1)$ show that 
\begin{eqsplit}
    \weightANN{2}{\oneNormANN_d} 
		&=
		\weightANN{1}{\sumANN_{1,d}} \weightANN2{\parallelizationSpecial_d(\oneNormANN_1, \oneNormANN_1, \ldots, \oneNormANN_1)}
		\\&=
		\underbrace{\begin{pmatrix} 1 & 1 & \cdots & 1\end{pmatrix}}_{\in \R^{1 \times d}}  
    \underbrace{ \begin{pmatrix}
    \weightANN2{\oneNormANN_1} & 0 & \cdots & 0 \\
    0 & \weightANN2{\oneNormANN_1} & \cdots & 0 \\
    \vdots & \vdots & \ddots & \vdots \\
    0 & 0 &\cdots & \weightANN2{\oneNormANN_1} \end{pmatrix}}_{
    	\in \R^{d \times (2d)}
    }
    \\&=
		\begin{pmatrix} 1 & 1 & \cdots & 1 \end{pmatrix} \in \R^{1 \times (2d)}
		.
\end{eqsplit}
This establishes item~\ref{item:lem:l1norm:4}. The proof of \cref{lem:l1norm:param} is thus complete. 
\end{proof}

\todoc{Maybe we can make a correct exercise out of this, but currently it does not make sense.}

\subsection{RePU ANN representations for the identity}
\label{sect:RePU_identity}

In \cref{subsec:identity} we observed that \ReLU\ and softplus \anns\ can express the real identity function (cf.\ \cref{lem:Relu:identity,lem:softplus:identity}).
In this subsection we show that the $1$-norm \ann\ in \cref{def:dnn:l1norm} can also be used to explicitly specify \RePU\ \ann\ representations for the real identity function.
We note that the remaining parts of this chapter are concerned with \ReLU\ \anns\ and do not use the material presented in this subsection.

\cfclear
\begingroup 
\providecommandordefault{\p}{p}
\begin{athm}{lemma}{lem:Power:identity}%
Let 
$\p \in \N\backslash\{1\}$,
$b_1,b_2,\dots,b_\p \in \R$
satisfy
$b_1 < b_2 < \ldots < b_\p$
\cfload.
Then
\begin{enumerate}[label=(\roman{*})]
\item 
\label{lem:Power:identity:item1}
there exist unique 
$c_0,c_1,\dots,c_\p \in \R$ 
which satisfy for all 
$k \in \{0,1,\dots,\p\}$ 
that
\begin{equation}
\label{T_B_D}
\begin{split} 
    \mathbbm{1}_{\{\p\}}(k) \, c_0 + \SmallSum{i=1}{\p} c_i (b_i)^{k} = \mathbbm{1}_{\{\p-1\}}(k)\,\p^{-1}
\end{split}
\end{equation}
and
\item 
\label{lem:Power:identity:item2}
it holds for all 
  $x \in \R$ 
that 
\begin{equation}
\label{T_B_D}
\begin{split} 
  c_0 
+
  \SmallSum{i=1}{\p} c_i ( x + b_i )^\p
=
  x.
\end{split}
\end{equation}
\end{enumerate}
\end{athm}

\begin{aproof}
\newcommand{\BBB}{{\mathbf{B}}}
\newcommand{\CCC}{{\mathbf{C}}}
\newcommand{\DDD}{{\mathbf{D}}}
Throughout this proof let 
$\BBB = (\BBB_{i,j})_{i,j\in\{1,2,\dots,\p+1\}} \in \R^{(\p+1)\times(\p+1)}$
satisfy for all 
$i,j \in \{1,2,\dots,\p\}$
that
$\BBB_{1,i+1} = 1$,
$\BBB_{i,1} = 0$,
$\BBB_{\p+1,1} = 1$,
and
$\BBB_{i+1,j+1} = (b_j)^{i}$ 
and let
$\DDD = (\DDD_1,\DDD_2,\dots,\DDD_{\p+1}) \in \R^{\p+1}$ 
satisfy for all 
$k \in \{1,2,\dots,\p+1\}$ 
that
$\DDD_k = \mathbbm{1}_{\{\p\}}(k) \, \p^{-1}$
\cfload.
\Nobs that,
\enum{
\eg,
Horn and Johnson \cite[Eq.~(0.9.11.2)]{Horn2013};
the assumption that
$b_1 < b_2 < \ldots < b_\p$
}\prove
that
\begin{equation}
\det(\BBB)
=
(-1)^{\p+1} \det\pr[\big]{ (\BBB_{i,j+1})_{i,j\in\{1,2,\dots,\p\}} }
=
(-1)^{\p}
\br*{
\textstyle
\prod_{\substack{i,j \in \{ 1,2,\dots,\p \} \\ i < j }} \pr[\big]{ b_j - b_i }
}
\neq 
0
.
\end{equation}
\enum{
This}\proves
that there exists a unique
$\CCC = (c_0 , c_1 , \dots , c_\p) \in \R^{\p+1}$
such that
$\BBB \CCC = \DDD$.
This proves \cref{lem:Power:identity:item1}.
\Moreover
\enum{\cref{lem:Power:identity:item1}; 
the binomial theorem}\prove
that for all
$x\in\R$
it holds that
\begin{equation}
\begin{split}
c_0 
+
\SmallSum{i=1}{\p} c_i ( x + b_i )^\p
&=
c_0 
+
\SmallSum{i=1}{\p} c_i \br*{ \SmallSum{j=0}{\p} \textstyle\binom{\p}{j} x^{\p-j} (b_i)^{j} }\\
&=
c_0 
+
\SmallSum{j=0}{\p} \textstyle\binom{\p}{j} \br*{ \SmallSum{i=1}{\p}  c_i (b_i)^{j} } x^{\p-j}\\
&=
c_0 
+
\SmallSum{j=0}{\p} \textstyle\binom{\p}{j} \br[\big]{ \mathbbm{1}_{\{\p-1\}}(j) \, \p^{-1} - \mathbbm{1}_{\{\p\}}(j) \, c_0 } x^{\p-j}\\
&=
  c_0 
+
  \textstyle\binom{\p}{\p-1} \p^{-1}x^{\p - (\p-1)}
-
  \textstyle\binom{\p}{\p} c_0
=
x
.
\end{split}
\end{equation}
This establishes \cref{lem:Power:identity:item2}.
\end{aproof}
\endgroup

\cfclear
\providecommandordefault{\p}{p}
\begin{athm}{lemma}{lem:ReluPower:identity}[Shallow \RePU\ \ann\ representation for the one-dimensional identity function]
Let 
$\p\in\N\backslash\{1\}$, 
$b_1,b_2,\dots,b_\p, c_0,c_1,\dots,c_\p \in \R$
satisfy for all 
$k \in \{0,1,\dots,\p\}$ 
that
$b_1 < b_2 < \ldots < b_\p$ 
and
$\mathbbm{1}_{\{\p\}}(k) \, c_0 + \sum_{i=1}^\p c_i (b_i)^{k} = \mathbbm{1}_{\{\p-1\}}(k)\,\p^{-1}$,\cfadd{lem:Power:identity}
let $\activation$ be the \rePU{} with power $\p$,
let $\I \in \ANNs$ satisfy 
\begin{equation}
\label{T_B_D}
\begin{split} 
  \I 
=
  \begin{cases}
    \idRelu_1 & \colon \p \in \{1, 3, 5, \ldots\}\\
    \oneNormANN_1 & \colon \p \in \{2, 4, 6, \ldots\} ,
  \end{cases}
\end{split}
\end{equation}
and let 
  $\Psi \in \ANNs$ 
satisfy
\begin{equation}
\Psi
=
\compANN{\AffineANN_{1,c_0}}{\pr[\bigg]{ \smallbigANNsum_{i=1}^{\p} \pr[\Big]{ \scalarMultANN{c_i}{ \pr[\big]{ \compANN{\I}{\AffineANN_{1,b_i}} } } } } }
\end{equation}
\cfload.
Then
\begin{enumerate}[label=(\roman{*})]
\item
\label{item:lem:ReluPower:identity}
it holds for all 
$x \in \R$ 
that 
$(\functionANN{\activation} (\I))(x) =x^\p$,
\item
\label{item:lem:ReluPower:dims} 
it holds that 
$\dims(\Psi) = (1,2\p,1) \in \N^3$,
\item
\label{item:lem:ReluPower:cont} 
it holds that 
$ \functionANN{\activation}(\Psi) \in C(\R, \R)$, 
and
\item
\label{item:lem:ReluPower:real}
it holds for all 
$x \in \R$ 
that 
$
(\functionANN{\activation}(\Psi))(x) = x
$
\end{enumerate}
\cfout.
\end{athm}

\begin{aproof}
First,
\nobs that
\enum{
  the fact that 
  \begin{equation}
  \I
  = 
  \left( \left(\begin{pmatrix}
  1 \\
  -1
  \end{pmatrix}
  , 
  \begin{pmatrix}
  0 \\
  0
  \end{pmatrix}
  \right)
  ,
  \Big(	
  \begin{pmatrix}
  1 & (-1)^\p
  \end{pmatrix}
  , 
  0 \Big)
  \right)  
  \in 
  \big((\R^{2 \times 1} \times \R^{2}) \times (\R^{1 \times 2} \times \R^1) \big)
  \end{equation} 
}\proves
that for all 
$x \in \R$ 
it holds that
\begin{equation}
\begin{split}
(\functionANN{\activation}(\I))(x) 
= 
\activation(x) + (-1)^\p \activation(-x) 
&
= 
\pr[\big]{ \max\{x, 0\} }^{\p} + (-1)^\p \pr[\big]{ \max\{-x, 0 \} }^{\p}
\\
&
=
\pr[\big]{ \max\{x, 0\} }^{\p} + \pr[\big]{ \min\{x, 0 \} }^{\p}
= 
x^\p
\end{split}
\end{equation}
\cfload.
\Moreover
\enum{
  \cref{lem:ANN:affine2:item4} in \cref{lem:ANN:affine2};
  \cref{item:ANN:scalar:1} in \cref{lem:ANNscalar};
the fact that $\dims(\I) = (1,2,1)$;
the fact that for all
  $i \in \{1,2,\dots,\p\}$
it holds that
$\dims(\AffineANN_{1,b_i}) = (1,1)$;
}\prove
that for all
  $i \in \{1,2,\dots,\p\}$
it holds that
\begin{equation}
\label{T_B_D}
\begin{split} 
  \dims\pr[\big]{
     \scalarMultANN{c_i}{ \pr[\big]{ \compANN{\I}{\AffineANN_{1,b_i}} } } 
  }
=
  \dims \pr[\big]{ \compANN{\I}{\AffineANN_{1,b_i}} }
=
  (1, 2, 1).
\end{split}
\end{equation}
Combining this with 
\enum{
  \cref{item:sum:ANNs:2} in \cref{lem:sum:ANNs};
}\proves that
\begin{equation}
\label{T_B_D}
\begin{split} 
  \dims\pr*{
    \smallbigANNsum_{i=1}^{\p} \pr[\Big]{ \scalarMultANN{c_i}{ \pr[\big]{ \compANN{\I}{\AffineANN_{1,b_i}} } } } 
  }
=
  (1, 2\p, 1).
\end{split}
\end{equation}
\Cref{lem:ANN:affine2:item1} in \cref{lem:ANN:affine2} \hence \proves that
\begin{equation}
\dims(\Psi)
= 
  \dims\pr*{ \compANN{\AffineANN_{1,c_0}}{\pr[\bigg]{ \smallbigANNsum_{i=1}^{\p} \pr[\Big]{ \scalarMultANN{c_i}{ \pr[\big]{ \compANN{\I}{\AffineANN_{1,b_i}} } } } } } }
=
(1,2\p,1).
\end{equation}
This establishes \cref{item:lem:ReluPower:dims}.
\Moreover
\enum{
\cref{PropertiesOfCompositions:Realization} in \cref{Lemma:PropertiesOfCompositions};
\cref{lem:ANN:affine2:item3} in \cref{lem:ANN:affine2};
\cref{item:sum:ANNs:4} in \cref{lem:sum:ANNs};
\cref{item:ANN:scalar:3} in \cref{lem:ANNscalar};
\cref{lem:ANN:affine:item3} in \cref{lem:ANN:affine};
}\prove
that for all
$x\in\R$
it holds that
\begin{equation}
\begin{split}
\pr[]{ \functionANN{\activation}(\Psi) } (x)
&
=
c_0 
+ 
\pr[\Big]{ \functionANN{\activation}\pr[\Big]{ \ssmallbigANNsum_{i=1}^{\p} \pr[\big]{ \scalarMultANN{c_i}{ \pr[]{ \compANN{\I}{\AffineANN_{1,b_i}} } } } } } (x)
\\
&
=
c_0 
+
\SmallSum{i=1}{\p} c_i \pr[]{ \functionANN{\activation}( \I ) } ( \functionANN{\activation}( \AffineANN_{1,b_i}) (x) ) 
=
c_0 
+
\SmallSum{i=1}{\p} c_i ( x + b_i )^\p.
\end{split}
\end{equation}
\enum{
This;
the hypothesis that
$b_1 < b_2 < \ldots < b_\p$;
the hypothesis that
for all $k \in \{0,1,\dots,\p\}$
it holds that
$\mathbbm{1}_{\{\p\}}(k) \, c_0 + \sum_{i=1}^\p c_i (b_i)^{k} = \mathbbm{1}_{\{\p-1\}}(k)\,\p^{-1}$;
\cref{lem:Power:identity}
}\prove that for all
$x\in\R$
it holds that
\begin{equation}
\begin{split}
\pr[]{ \functionANN{\activation}(\Psi) } (x)
=
x
.
\end{split}
\end{equation}
This establishes \cref{item:lem:ReluPower:cont,item:lem:ReluPower:real}.
\end{aproof}

\subsection{ANN representations for maxima}
\label{subsec:dnnmax}

\cfclear
\begin{lemma}[Unique existence of maxima \anns] 
\label{lem:max_d_welldef}
There exist unique $(\phi_d)_{d \in \N} \subseteq \ANNs$ which satisfy that  
\begin{enumerate}[label=(\roman{*})]
    \item \label{welldef:item1} it holds for all $d \in  \N$ that $\inDimANN(\phi_d) = d$,
    \item \label{welldef:item2} it holds for all $d \in  \N$ that $\outDimANN(\phi_d) = 1$,
    \item \label{welldef:item2.1} it holds that
    $
        \phi_1 = \AffineANN_{1, 0}  \in \R^{1 \times 1} \times \R^1
    $,
    \item it holds that 
    \begin{equation}
        \phi_2 = \pr*{ \! \pr*{ \!
	\begin{pmatrix}
		1 & -1 \\
		0 & 1 \\
		0 & -1
	\end{pmatrix},
	\begin{pmatrix}
		0 \\
		0 \\
		0
	\end{pmatrix} \!  }, \pr*{
	\begin{pmatrix}
		1 & 1 & -1
	\end{pmatrix}, \begin{pmatrix} 0 \end{pmatrix} }  \! } \in (\R^{3 \times 2} \times \R^3) \times (\R^{1 \times 3} \times \R^1 ),
    \end{equation}
    \item \label{welldef:item3} it holds for all $d \in  \{2,3,4, \ldots \}$ that $\phi_{2d} = 	\phi_{d} \compANNbullet \bpr{\parallelizationSpecial_{d}( \phi_2, \phi_2, \ldots, \phi_2) }$, and
    \item it holds for all $d \in \{2,3,4, \ldots \}$ that $\phi_{2d-1} = \phi_d \compANNbullet \bigl( \parallelizationSpecial_d(\phi_2, \phi_2, \dots, \phi_2, \idRelu_1 ) \bigr)$
    \end{enumerate}
\cfout.
\end{lemma}

\begin{proof}[Proof of \cref{lem:max_d_welldef}]
Throughout this proof, let $\psi \in \ANNs$ satisfy
\begin{equation}
\label{eq:definition_of_psi_in_proof_MAX}
    \psi = \pr*{ \! \pr*{ \!
	\begin{pmatrix}
		1 & -1 \\
		0 & 1 \\
		0 & -1
	\end{pmatrix},
	\begin{pmatrix}
		0 \\
		0 \\
		0
	\end{pmatrix} \!  }, \pr*{
	\begin{pmatrix}
		1 & 1 & -1
	\end{pmatrix},
	\begin{pmatrix} 0 \end{pmatrix} } \!  } \in (\R^{3 \times 2} \times \R^3) \times (\R^{1 \times 3} \times \R^1 )
\end{equation} 
\cfload. 
\Nobs that \cref{eq:definition_of_psi_in_proof_MAX} 
and \cref{lem:Relu:identity} \prove that
\begin{equation}
\label{eq:properties_of_psi_in_proof_MAX}
  \inDimANN( \psi ) = 2 ,
\qquad 
  \outDimANN( \psi ) = \inDimANN( \idRelu_1 ) = \outDimANN( \idRelu_1 ) = 1 ,
\qandq
  \cL( \psi ) = \cL( \idRelu_1 ) = 2
  .
\end{equation}
\cref{Lemma:ParallelizationElementary} and \cref{lem:Relu:identity} 
\hence \prove that for all $ d \in \N \cap (1,\infty) $ it holds that
\begin{equation}
  \inDimANN( \parallelizationSpecial_d( \psi, \psi, \dots, \psi ) ) = 2d , 
  \qquad 
  \outDimANN( \parallelizationSpecial_d( \psi, \psi, \dots, \psi ) ) = d , 
\end{equation}
\begin{equation}
  \inDimANN( \parallelizationSpecial_d( \psi, \psi, \dots, \psi, \idRelu_1 ) ) = 2 d - 1 ,
\qandq
  \outDimANN( \parallelizationSpecial_{d} (\psi, \psi, \ldots, \psi, \idRelu_1)) = d
\end{equation}
\cfload. 
Combining \cref{eq:properties_of_psi_in_proof_MAX}, 
\cref{Lemma:PropertiesOfCompositions}, and induction 
\hence \proves that there exists 
unique $\phi_d \in  \ANNs$, $d \in \N$, which satisfy 
for all $d \in \N$ that 
$\inDimANN(\phi_d) = d$, 
$\outDimANN(\phi_d) = 1$, and
\begin{equation}
\begin{split} 
	\phi_d
=
	\begin{cases}
		\AffineANN_{1, 0} & \colon d=1 \\
		\psi & \colon d = 2 \\
		\phi_{d/2} \compANNbullet \bpr{\parallelizationSpecial_{d/2}( \psi, \psi, \ldots, \psi) } & \colon d \in \{4, 6, 8, \ldots \} \\	\phi_{(d+1)/2} \compANNbullet \bpr{\parallelizationSpecial_{(d+1)/2}( \psi, \psi, \ldots, \psi, \idRelu_1) } & \colon d \in \{3, 5, 7, \ldots \}.
	\end{cases}
\end{split}
\end{equation}
The proof of \cref{lem:max_d_welldef} is thus complete.
\end{proof}

\cfclear

\begin{adef}{def:max_d}[Maxima \ann\ representations]
\cfconsiderloaded{def:max_d}
We denote by $(\maxANN_d)_{d \in \N} \subseteq \ANNs$  
the fully-connected feedforward \anns\ which satisfy that
\begin{enumerate}[label=(\roman{*})]
    \item it holds for all $d \in  \N$ that $\inDimANN(\maxANN_d) = d$,
    \item it holds for all $d \in  \N$ that $\outDimANN(\maxANN_d) = 1$,
	\item\label{def:max_d:item1}
	it holds that 
	$
		\maxANN_1 = \AffineANN_{1, 0}  \in \R^{1 \times 1} \times \R^1
	$,
     \item\label{def:max_d:item2}
      it holds that
    \begin{equation} \label{eq:def:m2}
        \maxANN_2 = \pr*{ \! \pr*{ \!
	\begin{pmatrix}
		1 & -1 \\
		0 & 1 \\
		0 & -1
	\end{pmatrix},
	\begin{pmatrix}
		0 \\
		0 \\
		0
	\end{pmatrix} \!  }, \pr*{
	\begin{pmatrix}
		1 & 1 & -1
	\end{pmatrix}, \begin{pmatrix} 0 \end{pmatrix} }  \! } \in (\R^{3 \times 2} \times \R^3) \times (\R^{1 \times 3} \times \R^1 ),
    \end{equation}
    \item\label{def:max_d:item3} it holds for all $d \in  \{2,3,4, \ldots \}$ that $\maxANN_{2d} = 	\maxANN_{d} \compANNbullet \bpr{\parallelizationSpecial_{d}( \maxANN_2, \maxANN_2, \ldots, \maxANN_2) }$, and
    \item\label{def:max_d:item4} it holds for all $d \in \{2,3,4, \ldots \}$ that $\maxANN_{2d-1} = \maxANN_d \compANNbullet \bigl( \parallelizationSpecial_d(\maxANN_2, \maxANN_2, \dots, \maxANN_2, \idRelu_1 ) \bigr)$\cfadd{lem:max_d_welldef}
\end{enumerate}
\cfload[.]
\end{adef}

\begin{adef}{def:ceiling}[Floor and ceiling of real numbers]
We denote by $\ceil{\cdot} \colon \R \to \Z$ and $\floor{\cdot} \colon \R \to \Z$ the functions which satisfy for all $x \in \R$ that 
\begin{equation}
  \ceil{x} = \min(\Z \cap [x, \infty))
\qandq
  \floor{x} = \max(\Z \cap (-\infty, x])
  .
\end{equation}
\end{adef}

\begin{exercise}{ex:ceil}
	Prove or disprove the following statement:
	For all $n\in\{3,5,7,\dots\}$ it holds that
	$\ceil{\logg_2(n+1)}=\ceil{\logg_2(n)}$.
\end{exercise}

\cfclear
\begin{prop}[Properties of maxima \anns]
\label{Prop:max_d}
Let $d \in \N$.
Then
\begin{enumerate}[label=(\roman{*})]
\item \label{max_d:item_3} it holds that
$\hiddenLength(\maxANN_d) = \ceil{\operatorname{log}_2(d)}$,
\item \label{max_d:item_4} it holds for all $i \in \N$ that
$\dimANNlevel_i(\maxANN_d) \leq 3\ceil*{\tfrac{d}{2^{i}}}$,
\item \label{max_d:item_5} it holds that $\functionANN\rect(\maxANN_d)  \in C(\R^d,\R)$, and
\item \label{max_d:item_6} it holds for all $x = (x_1, \dots, x_d) \in \R^d$ that $(\functionANN\rect(\maxANN_d))(x) = \max\{x_1, x_2, \ldots, x_d\}$
\end{enumerate}
\cfout.
\end{prop}

\begin{proof}[Proof of \cref{Prop:max_d}]
Throughout this proof, assume without loss of generality that $d > 1$.
Note that \eqref{eq:def:m2} ensures that 
\begin{equation}
  \hiddenLength(\maxANN_2) = 1
\end{equation}
\cfload.
This and \eqref{parallelisationSameLengthDef} demonstrate that for all $\fd \in \{2,3,4, \ldots\}$ it holds that
\begin{equation}
    \hiddenLength (\parallelizationSpecial_{\fd} (\maxANN_2, \maxANN_2, \ldots, \maxANN_2)) =  \hiddenLength (\parallelizationSpecial_{\fd} (\maxANN_2, \maxANN_2, \ldots, \maxANN_2, \idRelu_1)) = \hiddenLength(\maxANN_2) = 1
\end{equation}
\cfload.
Combining this with \cref{Lemma:PropertiesOfCompositions} establishes that for all $\fd \in \{3,4,5,\ldots\}$ it holds that
\begin{equation}
\label{Prop:max_d:eq1}
\begin{split} 
\hiddenLength (\maxANN_\fd) 
=
\hiddenLength (\maxANN_{\ceil*{\nicefrac{\fd}{2}}}) + 1
\end{split}
\end{equation}
\cfload. 
This assures that for all 
	$\fd \in \{4, 6, 8, \ldots \}$
with
$\hiddenLength(\maxANN_{\nicefrac{\fd}{2}}) = \ceil{\operatorname{log}_2(\nicefrac{\fd}{2})}$
it holds that
\begin{equation}
\label{Prop:max_d:eq2}
\begin{split} 
	\hiddenLength (\maxANN_\fd) 
&=
  \hiddenLength (\maxANN_{\ceil*{\nicefrac{\fd}{2}}}) + 1
=
  \hiddenLength (\maxANN_{\nicefrac{\fd}{2}}) + 1 \\
&=
	\ceil{\operatorname{log}_2(\nicefrac{\fd}{2})} + 1
=
	\ceil{\operatorname{log}_2(\fd) - 1} + 1
=
	\ceil{\operatorname{log}_2(\fd)}. 
\end{split}
\end{equation}
\Moreover 
\enum{
	\eqref{Prop:max_d:eq1};
	the fact that for all
		$\fd \in \{3, 5, 7, \ldots \}$
	it holds that
	$	\ceil{\operatorname{log}_2(\fd+1)} = 	\ceil{\operatorname{log}_2(\fd)}$
}
\prove
that for all
	$\fd \in \{3, 5, 7, \ldots \}$
with
$\hiddenLength(\maxANN_{\ceil{\nicefrac{\fd}{2}}}) = \ceil{\operatorname{log}_2(\ceil{\nicefrac{\fd}{2}})}$
it holds that
\begin{equation}
\label{Prop:max_d:eq3}
\begin{split} 
	\hiddenLength (\maxANN_\fd) 
&=
  \hiddenLength (\maxANN_{\ceil*{\nicefrac{\fd}{2}}}) + 1
=
	\ceil[\big]{\operatorname{log}_2(\ceil{\nicefrac{\fd}{2}})} + 1
=
	\ceil[\big]{\operatorname{log}_2(\nicefrac{(\fd + 1)}{2})} + 1 \\
&=
	\ceil{\operatorname{log}_2(\fd + 1) - 1} + 1
=
	\ceil{\operatorname{log}_2(\fd + 1)} 
=
	\ceil{\operatorname{log}_2(\fd)}.
\end{split}
\end{equation}
Combining 
\enum{
	this;
	\eqref{Prop:max_d:eq2};
}
demonstrates that for all 
$\fd \in \{3, 4, 5, \ldots \}$
with
$
	\forall \, k \in \{2, 3, \ldots,\allowbreak \fd - 1 \} 
\colon 
	\hiddenLength(\maxANN_{k}) = \ceil{\operatorname{log}_2(k)}
$
it holds that
\begin{equation}
\label{Prop:max_d:eq3.0}
	\hiddenLength (\maxANN_\fd) 
=
	\ceil{\operatorname{log}_2(\fd)}
  .
\end{equation}
The fact that $\hiddenLength(\maxANN_2) = 1$ and induction hence establish item~\ref{max_d:item_3}. 
\Nobs that the fact that $\dims (\maxANN_2) = (2,3,1)$ assure that for all
	$i \in \N$
it holds that 
\begin{equation}
\label{Prop:max_d:eq3.1}
\begin{split} 
	\dimANNlevel_i(\maxANN_{2}) \leq 3 =  3 \ceil*{\tfrac{2}{2^{i}}}.
\end{split}
\end{equation}
\Moreover 
\cref{Lemma:PropertiesOfCompositions}  and \cref{Lemma:ParallelizationElementary} imply that for all $\fd \in \{2,3,4, \ldots\}$, $i \in \N$ it holds that
\begin{equation}
\label{Prop:max_d:eq4}
    \dimANNlevel_i(\maxANN_{2 \fd}) 
  = 
    \dimANNlevel_i\pr*{\maxANN_{\fd} \compANNbullet \bpr{\parallelizationSpecial_{\fd}( \maxANN_2, \maxANN_2, \ldots, \maxANN_2) }}
    =
    \begin{cases}
  	3 \fd & \colon i=1 \\
    	\dimANNlevel_{i-1}(\maxANN_\fd) & \colon i \geq 2
    \end{cases}
\end{equation}
and
\begin{equation}
\label{Prop:max_d:eq5}
     \dimANNlevel_i(\maxANN_{2\fd-1}) 
    = 
    \dimANNlevel_i\pr*{\maxANN_{\fd} \compANNbullet \bpr{\parallelizationSpecial_{\fd}( \maxANN_2, \maxANN_2, \ldots, \maxANN_2, \idRelu_1) }}
    =
    \begin{cases}
    3\fd-1 & \colon i=1 \\
    \dimANNlevel_{i-1}(\maxANN_\fd) & \colon i \geq 2.
    \end{cases}
\end{equation}
This and \cref{Prop:max_d:eq3.0} assure that for all 
	$\fd \in \{2, 4, 6, \ldots \}$
it holds that 
\begin{equation}
\label{Prop:max_d:eq6}
\begin{split} 
	\dimANNlevel_1(\maxANN_{\fd})
=
	3 (\tfrac{\fd}{2})
=
	3 \ceil*{\tfrac{\fd}{2}}.
\end{split}
\end{equation}
\Moreover
\enum{
	\eqref{Prop:max_d:eq5}
}[ensure]
that for all 
$
  \fd \in \{3, 5, 7, \dots \}
$
it holds that 
\begin{equation}
\label{Prop:max_d:eq7}
\begin{split} 
	\dimANNlevel_1(\maxANN_{\fd})
=
	3 \ceil*{\tfrac{\fd}{2}} - 1
\leq
	3 \ceil*{\tfrac{\fd}{2}}.
\end{split}
\end{equation}
This and \eqref{Prop:max_d:eq6} show that for all
	$\fd \in \{2, 3, 4, \ldots \}$
it holds that 
\begin{equation}
\label{Prop:max_d:eq7.1}
\begin{split} 
	\dimANNlevel_1(\maxANN_{\fd})
\leq
	3 \ceil*{\tfrac{\fd}{2}}.
\end{split}
\end{equation}
\Moreover 
\enum{
	\eqref{Prop:max_d:eq4}
}[demonstrate]
that for all
	$\fd \in \{4, 6, 8, \ldots \}$,
	$i \in \{2, 3, 4, \ldots \}$
with $\dimANNlevel_{i-1}(\maxANN_{\nicefrac{\fd}{2}}) \leq 3\ceil*{ (\nicefrac{\fd}{2})\tfrac{1}{2^{i-1}}}$ it holds that
\begin{equation}
\label{Prop:max_d:eq8}
\begin{split} 
	\dimANNlevel_i(\maxANN_{\fd})
=
	\dimANNlevel_{i-1}(\maxANN_{\nicefrac{\fd}{2}})
\leq
	3\ceil*{ (\nicefrac{\fd}{2})\tfrac{1}{2^{i-1}}}
=
	3\ceil*{ \tfrac{\fd}{2^{i}}}.
\end{split}
\end{equation}
\Moreover 
\enum{
	\eqref{Prop:max_d:eq5};
	the fact that for all 
		$\fd \in \{3, 5, 7, \ldots \}$,
		$i \in \N$
	it holds that 
	$
		\ceil*{\frac{\fd+1}{2^i}} = \ceil*{\frac{\fd}{2^i}}
	$
}[assure]
that for all
	$\fd \in \{3, 5, 7, \ldots \}$,
	$i \in \{2, 3, 4, \ldots \}$
with $\dimANNlevel_{i-1}(\maxANN_{\ceil*{\nicefrac{\fd}{2}}}) \leq 3\ceil*{ \ceil{\nicefrac{\fd}{2}} \tfrac{1}{2^{i-1}}}$ it holds that
\begin{equation}
\label{Prop:max_d:eq9}
\begin{split} 
	\dimANNlevel_i(\maxANN_{\fd})
=
	\dimANNlevel_{i-1}(\maxANN_{\ceil*{\nicefrac{\fd}{2}}})
\leq
	3\ceil*{ \ceil{\nicefrac{\fd}{2}} \tfrac{1}{2^{i-1}}}
=
	3\ceil*{ \tfrac{\fd + 1}{2^{i}}}
=
	3\ceil*{ \tfrac{\fd}{2^{i}}}.
\end{split}
\end{equation} 
\enum{
	This;
	\eqref{Prop:max_d:eq7.1};
	\eqref{Prop:max_d:eq8}
}[ensure]
that for all
	$\fd \in \{3, 4, 5, \ldots \}$,
	$i \in \N$
with $\forall \, k \in \{2, 3, \ldots, \fd-1 \}, j \in \N \colon \dimANNlevel_{j}(\maxANN_{k}) \leq 3\ceil*{ \tfrac{k}{2^{j}}}$ it holds that
\begin{equation}
\label{Prop:max_d:eq10}
\begin{split} 
	\dimANNlevel_i(\maxANN_{\fd})
\leq
	3\ceil*{ \tfrac{\fd}{2^{i}}}.
\end{split}
\end{equation} 
Combining
\enum{
	this;
	\eqref{Prop:max_d:eq3.1};
}
with induction establishes item~\ref{max_d:item_4}. 
\Nobs that
\eqref{eq:def:m2} ensures that  for all $x=(x_1, x_2) \in \R^2$ it holds that
\begin{equation}
\begin{split}
    (\functionANN\rect(\maxANN_2))(x) &= \max \{x_1-x_2, 0\} + \max \{ x_2 , 0\} - \max\{ -x_2 , 0\} \\
    &= \max \{x_1-x_2, 0\} + x_2 = \max\{x_1, x_2\}
    \end{split}
\end{equation}
\cfload.
\cref{Lemma:PropertiesOfParallelizationEqualLength}, \cref{Lemma:PropertiesOfCompositions}, \cref{lem:Relu:identity}, and induction hence imply that for all $\fd \in  \{2,3, 4, \ldots \}$, $x= ( x_1,x_2,\dots,x_\fd) \in \R^\fd$ it holds that 
\begin{equation}
  \functionANN\rect(\maxANN_\fd) \in C(\R^\fd,\R)
\qandq
  \pr*{\functionANN\rect({\maxANN_\fd})}(x) = \max\{x_1,x_2,\dots,x_\fd\}
  .
\end{equation}
This establishes \cref{max_d:item_5,max_d:item_6}.
The proof of \cref{Prop:max_d} is thus complete.
\end{proof}

 \cfclear
\begin{lemma}
\label{t:max:d} Let $d \in \N$, $i \in \{1,2,  \ldots, \lengthANN(\maxANN_d) \}$ \cfload. 
Then
\begin{enumerate}[label=(\roman{*})]
\item \label{t:max_d:item_1} it holds that $\biasANN{i}{\maxANN_d}= 0 \in \R^{\dimANNlevel_i(\maxANN_d)}$,
\item \label{t:max_d:item_2} it holds that $ \weightANN{i}{\maxANN_d} \in \{-1,0,1\}^{\dimANNlevel_i(\maxANN_d) \times \dimANNlevel_{i-1}(\maxANN_d) }$, and
\item \label{t:max_d:item_3} it holds for all $x \in \R^d$ that $\pnorm\infty{\weightANN{1}{\maxANN_d} x} \leq 2 \pnorm\infty{ x }$
\end{enumerate}
\cfout.
\end{lemma}

\begin{proof}[Proof of \cref{t:max:d}]
Throughout this proof, assume without loss of generality that $d > 2$ (cf.\ \cref{def:max_d:item1,def:max_d:item2} in \cref{def:max_d}) and let $A_1 \in \R^{3 \times 2}$, $A_2 \in \R ^{1 \times 3}$, $C_1 \in \R^{2 \times 1}$, $C_2 \in \R^{1 \times 2}$ satisfy
\begin{equation}
\label{t:max:d:eq1}
\begin{split} 
A_1 = \begin{pmatrix} 1 & -1 \\ 0 & 1 \\ 0 & -1 \end{pmatrix}, 
\qquad
A_2 = \begin{pmatrix}1 & 1 & -1 \end{pmatrix}, 
\qquad
C_1 = \begin{pmatrix}1 \\ -1 \end{pmatrix}, 
\qandq 
C_2 = \begin{pmatrix} 1 & -1 \end{pmatrix}.
\end{split}
\end{equation} 
Note that \cref{def:max_d:item2,def:max_d:item3,def:max_d:item4} in \cref{def:max_d} assure that for all $\fd \in \{2,3,4,\ldots \}$ it holds that
\begin{equation} \label{propmd:eq1}
\begin{split}
    \weightANN1{\maxANN_{2\fd-1}} =  \underbrace{
    \begin{pmatrix}
    A_1 & 0 & \cdots &0 & 0 \\
    0 & A_1 & \cdots &0 & 0 \\
    \vdots & \vdots & \ddots & \vdots & \vdots \\
    0 & 0 & \cdots & A_1& 0 \\
    0 & 0 & \cdots & 0& C_1 
    \end{pmatrix}
    }_{\in \R^{(3\fd-1)\times (2\fd-1)}}, & \qquad 
    \weightANN1{\maxANN_{2\fd }} =  \underbrace{
    \begin{pmatrix}A_1 & 0 & \cdots & 0 \\
    0 & A_1 & \cdots & 0 \\
    \vdots & \vdots & \ddots & \vdots \\
    0 & 0 & \cdots & A_1 \\
    \end{pmatrix}}_{\in \R^{(3 \fd) \times (2 \fd)}}, \\
    \biasANN1{\maxANN_{2\fd-1}} = 0 \in \R^{3 \fd -1}, &\qandq  \biasANN1{\maxANN_{2\fd}} = 0 \in \R^{3 \fd}.
    \end{split}
\end{equation}
This and \eqref{t:max:d:eq1} proves item~\ref{t:max_d:item_3}. 
Furthermore, note that \eqref{propmd:eq1} and item~\ref{def:max_d:item2} in \cref{def:max_d} imply that for all
	$\fd \in \{2,3,4,\ldots \}$ 
it holds that 
$
	\biasANN1{\maxANN_{\fd}} = 0
$.
\Cref{def:max_d:item2,def:max_d:item3,def:max_d:item4} in \cref{def:max_d} hence ensure that
 for all $\fd \in \{2,3,4,\ldots \}$ it holds that
\begin{equation} \label{propmd:eq2}
\begin{split}
    \weightANN{2}{\maxANN_{2\fd-1}} = \weightANN{1}{\maxANN_\fd}   \underbrace{
    \begin{pmatrix}
	    A_2 & 0 & \cdots & 0 & 0 \\
	    0 & A_2 & \cdots & 0 & 0 \\
	    \vdots & \vdots & \ddots & \vdots& \vdots \\
	   	0 & 0 & \cdots & A_2 & 0  \\
	    0 & 0 & \cdots &0 & C_2  
    \end{pmatrix}
    }_{\in\R^{\fd\times (3\fd -1)}}, &\qquad 
      \weightANN{2}{\maxANN_{2\fd}} = \weightANN{1}{\maxANN_\fd} \underbrace{
    \begin{pmatrix}A_2 & 0 & \cdots & 0 \\
    0 & A_2 & \cdots & 0 \\
    \vdots & \vdots & \ddots & \vdots \\
    0 & 0 & \cdots & A_2  \end{pmatrix}}_{\in \R^{\fd \times (3 \fd)}}, \\
    \biasANN2{\maxANN_{2\fd-1}} = \biasANN1{\maxANN_\fd} = 0, &\qandq \biasANN2{\maxANN_{2 \fd}} = \biasANN1{\maxANN_\fd} = 0.
\end{split}
\end{equation}
Combining this and
item~\ref{def:max_d:item2} in \cref{def:max_d} shows that for all
	$\fd \in \{2,3,4,\ldots \}$ 
it holds that 
$
	\biasANN2{\maxANN_{\fd}} = 0
$.
Moreover, note that \eqref{eq:defCompANN} demonstrates that for all 
$ \fd \in \{2,3,4, \ldots, \}$, $i \in \{3,4, \allowbreak \ldots, \lengthANN(\maxANN_\fd)+1 \}$ it holds that
\begin{equation} \label{propmd:eq3}
    \weightANN{i}{\maxANN_{2 \fd -1}} = \weightANN{i}{\maxANN_{2 \fd}} = \weightANN{i-1}{\maxANN_\fd} \qandq   \biasANN{i}{\maxANN_{2 \fd -1}} = \biasANN{i}{\maxANN_{2 \fd}} = \biasANN{i-1}{\maxANN_\fd}.
\end{equation}
This,  \eqref{t:max:d:eq1}, \eqref{propmd:eq1}, \eqref{propmd:eq2}, 
the fact that for all
	$\fd \in \{2,3,4,\ldots \}$ 
it holds that 
$
	\biasANN2{\maxANN_{\fd}} = 0
$,
and induction establish \cref{t:max_d:item_1,t:max_d:item_2}. 
 The proof of \cref{t:max:d} is thus complete. 
\end{proof}

\subsection{ANN representations for maximum convolutions} 
\label{subsec:dnninterpolation}

\cfclear
\begin{exercise}{ex:represent_inf_norm}
	Prove or disprove the following statement: 
	It holds for all $d \in \N$, $x \in \R^d$ that
	\begin{equation}
    \functionANN\rect(\maxANN_d \compANNbullet \parallelizationSpecial_d(\oneNormANN_1, \ldots, \oneNormANN_1))(x) = \pnorm\infty{x}
	\end{equation}
  \cfload.
\end{exercise}

\cfclear
\begin{lemma}\label{dnn:intl1}
Let $d, K \in \N$, $L \in [0, \infty)$, $\fx_1, \fx_2, \ldots, \fx_K \in \R^d$, $\fy = (\fy_1, \ldots, \allowbreak \fy_K) \in \R^K$, $\Phi \in \ANNs$ satisfy
\begin{equation}
\label{dnn:intl1:ass1}
		\Phi = \maxANN_{K} \compANNbullet \AffineANN_{-L \idMatrix_{K}, \fy} \compANNbullet 
		\parallelizationSpecial_{K}  \bigl(  \oneNormANN_d \compANNbullet \AffineANN_{\idMatrix_d, -\fx_1},  \oneNormANN_d \compANNbullet \AffineANN_{\idMatrix_d, -\fx_2}, \ldots,  \oneNormANN_d \compANNbullet \AffineANN_{\idMatrix_d, -\fx_K} \bigr) \compANNbullet \extensionANN_{d, K}
\end{equation}
\cfload. Then 
\begin{enumerate}[label=(\roman{*})]
\item \label{item:dnn:intl1:1}
it holds that $\inDimANN(\Phi) = d$,
\item \label{item:dnn:intl1:1a} it holds that $\outDimANN(\Phi) = 1$,
   \item \label{item:dnn:intl1:2}
    it holds that $\hiddenLength (\Phi) = \ceil{ \operatorname{log}_2( K )} + 1$,
\item \label{item:dnn:intl1:3}
it holds that $\dimANNlevel_1 ( \Phi) = 2 d K$,
\item \label{item:dnn:intl1:4}
it holds for all $i \in \{2,3,4, \ldots \}$ that $\dimANNlevel_i (\Phi) \leq 3 \ceil*{ \frac{K}{2^{i-1}}} $,
    \item \label{item:dnn:intl1:5}
    it holds that $\pnorm\infty{ \MappingStructuralToVectorized(\Phi)} \leq \max \{ 1,L, \max_{k \in \{1, 2, \ldots, K\}} \pnorm\infty{ \fx_k}, 2 \pnorm\infty{\fy}\}$, and
    \item \label{item:dnn:intl1:6}
    it holds for all $x \in \R^d$ that $(\functionANN\rect(\Phi))(x) = \max\nolimits_{k  \in \{1, 2, \ldots, K \} } \pr*{ \fy_k - L \pnorm1{x-\fx_k} }$
\end{enumerate}
\cfout.
\end{lemma}

\begin{proof}[Proof of \cref{dnn:intl1}]
Throughout this proof, let $\Psi_k \in \ANNs$, $k \in \{1,2, \ldots, K\}$, satisfy for all $k \in \{1,2, \ldots, K\}$ that $\Psi_k =  \oneNormANN_d \compANNbullet \AffineANN_{\idMatrix_d, -\fx_k}$, let $\Xi \in \ANNs$ satisfy
\begin{equation}
    \Xi 
		= 
		\AffineANN_{-L \idMatrix_{K}, \fy} \compANNbullet \parallelizationSpecial_{K}  \bigl( \Psi_1, \Psi_2, \ldots, \Psi_{K} \bigr) \compANNbullet \extensionANN_{d, K},
\end{equation}
and let $\normmm{\cdot} \colon \bigcup_{m, n \in \N} \R^{m \times n} \to [0, \infty)$ satisfy for all $m,n \in \N$, $M = (M_{i,j})_{i \in \{1, \ldots, m\}, \,  j \in \{1, \ldots, n \} } \in \R^{m \times n}$ that $\normmm{M} = \max_{i \in \{1, \ldots, m\}, \, j \in \{1, \ldots, n\}} \abs{M_{i,j}}$.
Observe that 
	\eqref{dnn:intl1:ass1} 
	and \cref{Lemma:PropertiesOfCompositions} 
ensure that 
	$\outDimANN(\Phi) = \outDimANN(\maxANN_{K})=1$ 
	and 
	$\inDimANN(\Phi) = \inDimANN( \extensionANN_{d,K}) = d$. 
This proves \cref{item:dnn:intl1:1,item:dnn:intl1:1a}. 
Moreover, observe that
the fact that for all 
	$ m,n \in \N, \, \fW \in \R^{m \times n}, \, \fB \in \R^m $ 
it holds that 
$ \hiddenLength (\AffineANN_{\fW, \fB}) = 0 = \hiddenLength(\extensionANN_{d,K})$,
 the fact that $\hiddenLength (\oneNormANN_d) = 1$, and
\cref{Lemma:PropertiesOfCompositions}
assure that
\begin{equation}
\begin{split} 
	\hiddenLength(\Xi)
=
	\hiddenLength (\AffineANN_{-L \idMatrix_{K}, \fy} ) + \hiddenLength (\parallelizationSpecial_{K}  ( \Psi_1, \Psi_2, \ldots, \Psi_{K} )) + \hiddenLength (\extensionANN_{d,K})
=
	\hiddenLength (\Psi_1) 
=
	\hiddenLength (\oneNormANN_d) 
=
	1.
\end{split}
\end{equation}
\cref{Lemma:PropertiesOfCompositions} and \cref{Prop:max_d} hence 
ensure that 
\begin{equation}
\begin{split}
    \hiddenLength (\Phi) 
&= 
	\hiddenLength (\maxANN_{K} \compANNbullet \Xi) 
= 
	\hiddenLength (\maxANN_{K}) + \hiddenLength (\Xi) 
=
	\ceil{ \operatorname{log} _2 (K)} +1
\end{split}
\end{equation}
\cfload.
This establishes \cref{item:dnn:intl1:2}.
Next observe 
	that the fact that 
		$\hiddenLength(\Xi) = 1$, 
	\cref{Lemma:PropertiesOfCompositions}, 
	and \cref{Prop:max_d} 
assure that
for all 
	$i \in \{2,3,4,\ldots\}$ 
it holds that
\begin{equation}
    \dimANNlevel_i (\Phi) = \dimANNlevel_{i-1}(\maxANN_{K}) \leq 3 \ceil*{ \tfrac{K}{2^{i-1}}}.
\end{equation}
This proves \cref{item:dnn:intl1:4}.
Furthermore, note that \cref{Lemma:PropertiesOfCompositions}, \cref{Lemma:PropertiesOfParallelizationEqualLengthDims}, and \cref{prop:dnn:l1norm} assure that
\begin{equation}
    \dimANNlevel_1(\Phi) 
		= 
		\dimANNlevel_1(\Xi)
		=
		\dimANNlevel_1 \pr*{ \parallelizationSpecial_{K}(\Psi_1, \Psi_2, \ldots, \Psi_{K})}
		=
		\sum_{i=1}^{K} \dimANNlevel_1 \pr*{ \Psi_i}
		=
		\sum_{i=1}^{K} \dimANNlevel_1( \oneNormANN_d) 
		=
		2 d K
		.
\end{equation}
This establishes \cref{item:dnn:intl1:3}. 
Moreover, observe that \eqref{eq:defCompANN} and \cref{t:max:d} imply that
\begin{equation}  \label{proof:eqphi}
\begin{split}
    \Phi 
		=
		\bigl( &(\weightANN1{\Xi}, \biasANN1{\Xi}),(\weightANN1{\maxANN_{K}} \weightANN{2}{\Xi}, \weightANN{1}{ \maxANN_{K}} \biasANN2{\Xi}), \\
    &(\weightANN2{\maxANN_{K}}, 0), \ldots, (\weightANN{\lengthANN(\maxANN_{K})}{\maxANN_{K}}, 0) \bigr).
\end{split}
\end{equation}
Next note that the fact that for all $k \in \{1,2, \ldots, K \}$ it holds that $\weightANN{1}{\Psi_k} =  \weightANN{1}{\AffineANN_{\idMatrix_d, -\fx_k}} \weightANN{1}{\oneNormANN_d} = \weightANN{1}{\oneNormANN_d}$ assures that 
\begin{equation}
\begin{split}
    \weightANN{1}{\Xi} 
&= 
     \weightANN{1}{\parallelizationSpecial_{K}  ( \Psi_1, \Psi_2, \ldots, \Psi_{K} )} \weightANN{1}{\extensionANN_{d,K}}
= 
	\begin{pmatrix}
		\weightANN{1}{\Psi_1} &0 &\cdots &0 \\
		0 &\weightANN{1}{\Psi_2}  &\cdots &0 \\
		\vdots &  \vdots & \ddots &\vdots \\
		0 & 0 & \cdots & \weightANN{1}{\Psi_K}
	\end{pmatrix}
	\begin{pmatrix}
		\idMatrix_d \\
		\idMatrix_d \\
		\vdots \\
		\idMatrix_d 
	\end{pmatrix} \\
&=
    \begin{pmatrix}
    \weightANN{1}{\Psi_1} \\
    \weightANN{1}{\Psi_2} \\
    \vdots \\
    \weightANN{1}{\Psi_K}
    \end{pmatrix}
    = \begin{pmatrix}
    \weightANN{1}{\oneNormANN_d} \\
    \weightANN{1}{\oneNormANN_d} \\
    \vdots \\
    \weightANN{1}{\oneNormANN_d}
    \end{pmatrix}.
\end{split}
\end{equation}
\cref{lem:l1norm:param} hence demonstrates that $\normmm{\weightANN{1}{\Xi}} = 1$. 
In addition, note that \eqref{eq:defCompANN} implies that 
\begin{equation}
\label{dnn:intl1:eq1}
    \biasANN1{\Xi} 
=
	\weightANN{1}{\parallelizationSpecial_{K}  ( \Psi_1, \Psi_2, \ldots, \Psi_{K} )}\biasANN{1}{\extensionANN_{d,K}} 
	+ 
	\biasANN{1}{\parallelizationSpecial_{K}  ( \Psi_1, \Psi_2, \ldots, \Psi_{K} )}
=
	\biasANN{1}{\parallelizationSpecial_{K}  ( \Psi_1, \Psi_2, \ldots, \Psi_{K} )}
=
	\begin{pmatrix}
    \biasANN{1}{\Psi_1} \\
    \biasANN{1}{\Psi_2} \\
    \vdots \\
    \biasANN{1}{\Psi_{K}}
    \end{pmatrix}.
\end{equation}
Furthermore, observe that \cref{lem:l1norm:param} implies that for all $k \in \{1,2, \ldots, K \}$ it holds that 
\begin{equation}
\begin{split} 
	\biasANN1{\Psi_k} 
= 
	\weightANN{1}{\oneNormANN_d}\biasANN{1}{\AffineANN_{\idMatrix_d, -\fx_k}} 
	+ 
	\biasANN{1}{\oneNormANN_d}
= 
	-\weightANN1{\oneNormANN_d} \fx_k.
\end{split}
\end{equation}
This, \eqref{dnn:intl1:eq1}, and \cref{lem:l1norm:param} show that
\begin{equation}
\begin{split} 
	\pnorm\infty{ \biasANN1{\Xi}} 
=
	\max_{k \in \{1, 2, \ldots, K\}} \pnorm\infty{ \biasANN{1}{\Psi_k}}
=
	\max_{k \in \{1, 2, \ldots, K\}} \pnorm\infty{ \weightANN1{\oneNormANN_d} \fx_k}
=
	\max_{k \in \{1, 2, \ldots, K\}} \pnorm\infty{ \fx_k}
\end{split}
\end{equation}
\cfload.
Combining \enum{
this
; 
\eqref{proof:eqphi}
;
\cref{t:max:d} 
;
the fact that $\normmm{\weightANN{1}{\Xi}} = 1$
;
} shows that
\begin{equation} \label{eq:tphiinfty}
	\begin{split}
			\pnorm\infty{\MappingStructuralToVectorized(\Phi)} 
			&=
			\max \cu{ \normmm{\weightANN{1}{\Xi}}, \pnorm\infty{\biasANN1{\Xi}}, \normmm{\weightANN1{\maxANN_{K}} \weightANN{2}{\Xi}}, \pnorm\infty{\weightANN1{\maxANN_{K}} \biasANN2{\Xi}} , 1 } 
			\\&=
			\max \cu*{ 1, \max\nolimits_{k \in \{1, 2, \ldots, K\}} \pnorm\infty{ \fx_k}, \normmm{ \weightANN1{\maxANN_{K}} \weightANN2{\Xi}}, \pnorm\infty{\weightANN1{\maxANN_{K}} \biasANN{2}{\Xi}} }
	\end{split}
\end{equation}
\cfload. 
Next note that \cref{lem:l1norm:param} ensures that for all $k \in \{1,2, \ldots, K \}$ it holds that $\biasANN2{\Psi_k} = \biasANN2{\oneNormANN_d} = 0$.
Hence, we obtain that $\biasANN2 { \parallelizationSpecial_{K}(\Psi_1, \Psi_2, \ldots, \Psi_{K}) } = 0$. This implies that
\begin{equation} \label{proof:b2xi}
    \biasANN2{\Xi} 
= 
	\weightANN{1}{\AffineANN_{-L \idMatrix_{K}, \fy}} \biasANN{2}{\parallelizationSpecial_{K}  ( \Psi_1, \Psi_2, \ldots, \Psi_{K} )} 
	+ 
	\biasANN{1}{\AffineANN_{-L \idMatrix_{K}, \fy}}
=
	\biasANN{1}{\AffineANN_{-L \idMatrix_{K}, \fy}}
=
	\fy.
\end{equation}
In addition, observe that the fact that  for all $k \in \{1,2, \ldots, K \}$ it holds that $\weightANN2{\Psi_k} = \weightANN{2}{\oneNormANN_d} $ assures that 
\begin{equation} \label{proof:w2xi}
\begin{split}
\weightANN{2}{\Xi} 
&= \weightANN{1}{\AffineANN_{-L \idMatrix_{K}, \fy}} \weightANN{2}{\parallelizationSpecial_{K}  ( \Psi_1, \Psi_2, \ldots, \Psi_{K} )} 
= -L \weightANN2{ \parallelizationSpecial_{K}  ( \Psi_1, \Psi_2, \ldots, \Psi_{K} ) } \\
&= -L \begin{pmatrix}
 	\weightANN{2}{\Psi_1} & 0 & \cdots & 0 \\
    0 & \weightANN{2}{\Psi_2} & \cdots & 0 \\
    \vdots & \vdots & \ddots & \vdots \\
    0 & 0 &\cdots & \weightANN{2}{\Psi_K}
\end{pmatrix} 
= \begin{pmatrix}
 -L\weightANN{2}{\oneNormANN_d} & 0 & \cdots & 0 \\
    0 & -L\weightANN{2}{\oneNormANN_d} & \cdots & 0 \\
    \vdots & \vdots & \ddots & \vdots \\
    0 & 0 &\cdots & -L\weightANN{2}{\oneNormANN_d}
\end{pmatrix}.
\end{split}
\end{equation}
\Cref{item:lem:l1norm:4} in \cref{lem:l1norm:param} and \cref{t:max:d} hence imply that 
\begin{equation}
\label{proof:w2xi2}
\begin{split} 
	\normmm{\weightANN1{\maxANN_{K}} \weightANN{2}{\Xi} } = L\normmm{\weightANN1{\maxANN_{K}}} \leq L .
\end{split}
\end{equation}
Moreover, observe that  
\enum{
	\eqref{proof:b2xi};
	\cref{t:max:d}
}[assure]
that
\begin{equation}
\begin{split} 
	\pnorm\infty{ \weightANN1{\maxANN_{K}} \biasANN2{\Xi} } \leq 2 \pnorm\infty{\biasANN2{\Xi}} = 2 \pnorm\infty{\fy}.
\end{split}
\end{equation}
 Combining this with \eqref{eq:tphiinfty} and \eqref{proof:w2xi2} establishes item~\ref{item:dnn:intl1:5}. 
Next observe that \cref{prop:dnn:l1norm} and \cref{lem:ANN:affine2} show that for all $x \in \R^d$, $k \in \{1,2, \ldots, K \}$ it holds that
\begin{equation}
\begin{split} 
	(\functionANN\rect(\Psi_k))(x) =\pr[\big]{ \functionANN\rect(\oneNormANN_d) \circ \functionANN\rect(\AffineANN_{\idMatrix_d, -\fx_k})}(x) = \pnorm1{ x- \fx_k }.
\end{split}
\end{equation}
This, \cref{Lemma:PropertiesOfParallelizationEqualLength}, and \cref{Lemma:PropertiesOfCompositions} imply that for all $x \in \R^d$ it holds that
\begin{equation} 
 \bigl(\functionANN\rect \pr*{ \parallelizationSpecial_{K}(\Psi_1, \Psi_2, \ldots, \Psi_{K}) \compANNbullet \extensionANN_{d,K} } \bigr) (x) 
= \bigl( \pnorm1{ x- \fx_1 }, \pnorm1{ x- \fx_2 }, \ldots, \pnorm1{ x - \fx_K } \bigr).
\end{equation}
\cfload.
Combining this and \cref{lem:ANN:affine2} establishes that for all $x \in \R^d$ it holds that
\begin{equation}
\begin{split}
	(\functionANN\rect(\Xi))(x) 
&= 
	\bigl(\functionANN\rect  \pr*{ \AffineANN_{-L \idMatrix_{K}, \fy}} \circ \functionANN\rect  \pr*{  \parallelizationSpecial_{K}(\Psi_1, \Psi_2, \ldots, \Psi_{K}) \compANNbullet \extensionANN_{d,K} } \bigr) (x) \\
&= \bigl( \fy_1 - L \pnorm1{ x- \fx_1 }, \fy_2 - L \pnorm1{ x - \fx_2 }, \ldots, \fy_K - L \pnorm1{ x- \fx_K } \bigr).
\end{split}
\end{equation}
\cref{Lemma:PropertiesOfCompositions} and \cref{Prop:max_d} hence demonstrate that for all $x \in \R^d$ it holds that
\begin{equation}
\begin{split} 
	(\functionANN\rect(\Phi))(x) 
&= 
	\pr[\big]{\functionANN\rect(\maxANN_{K}) \circ \functionANN\rect(  \Xi )} (x) \\
&= 
	(\functionANN\rect(\maxANN_{K})) \bigl( \fy_1 - L \pnorm1{ x- \fx_1 }, \fy_2 - L \pnorm1{ x - \fx_2 }, \ldots, \fy_K - L \pnorm1{ x- \fx_K } \bigr)\\
&=
	\max\nolimits_{k  \in \{1, 2, \ldots, K \} } \pr*{\fy_k - L \pnorm1{x-\fx_k} }.
\end{split}
\end{equation}
This establishes item~\ref{item:dnn:intl1:6}. The proof of \cref{dnn:intl1} is thus complete.
\end{proof}

\section{ANN approximations results for multi-dimensional functions}

\subsection{Constructive ANN approximation results}
\cfclear
\begin{prop} \label{dnn:intp1}
Let $d,K  \in \N$, $L \in [0, \infty)$, let $E \subseteq \R^d$ be a set, let $\fx_1, \fx_2, \ldots, \fx_K \in E$, let $f \colon E \to \R$ satisfy for all $x, y  \in E$ that $\abs{f(x)-f(y)} \leq L \pnorm1{x-y}$, 
and let $\fy \in \R^{K}$, $\Phi \in \ANNs$ satisfy $\fy = (f(\fx_1), f(\fx_2), \ldots, f(\fx_K))$
and
\begin{equation}
		\Phi = \maxANN_{K} \compANNbullet \AffineANN_{-L \idMatrix_{K}, \fy} \compANNbullet 
		\parallelizationSpecial_{K}  \bigl( \oneNormANN_d \compANNbullet \AffineANN_{\idMatrix_d, -\fx_1} , \oneNormANN_d \compANNbullet \AffineANN_{\idMatrix_d, -\fx_2}, \ldots, \oneNormANN_d \compANNbullet \AffineANN_{\idMatrix_d, -\fx_K} \bigr) \compANNbullet \extensionANN_{d,K}
\end{equation}
\cfload. Then 
    \begin{equation}
    \label{dnn:intp1:concl}
        \sup\nolimits_{x \in E} \abs{(\functionANN\rect(\Phi))(x) - f(x)} \leq 2L \br*{\sup\nolimits_{x \in E} \pr*{ \min\nolimits_{k \in \{1, 2, \ldots, K \} } \pnorm1{x-\fx_k} }}
    \end{equation}
\cfout.
\end{prop}
\begin{proof}[Proof of \cref{dnn:intp1}]
Throughout this proof, let $F \colon \R^d \to \R$ satisfy for all $x \in \R ^d$ that 
\begin{equation}
\label{dnn:intp1:eq0}
       F(x) = \max\nolimits_{k \in \{1, 2, \ldots, K \} } \pr*{ f(\fx_k) - L \pnorm1{x- \fx_k} }.
\end{equation}
Observe that \enum{
	\cref{lem:lipschitz_extension}
	;
	\eqref{dnn:intp1:eq0}
	;
	the assumption that for all $x, y  \in E$ it holds that $\abs{f(x)-f(y)} \leq L \pnorm1{x-y}$
	;
}[assure] that 
\begin{equation}
\label{dnn:intp1:eq1}
    \sup\nolimits_{x \in E} \abs{F(x)- f(x)} \leq 2L \br*{\sup\nolimits_{x \in E} \pr*{\min\nolimits_{k \in \{1, 2, \ldots, K \} } \pnorm1{x-\fx_k} }}.
\end{equation}
Moreover, note that \cref{dnn:intl1} 
ensures that for all $x \in E $ it holds that $F(x)= (\functionANN\rect(\Phi))(x)$. 
Combining this and \eqref{dnn:intp1:eq1} establishes \eqref{dnn:intp1:concl}.
The proof of \cref{dnn:intp1} is thus complete.
\end{proof} 

\begin{exercise}{ex:approx_2d_function2}
	Prove or disprove the following statement: 
	There exists $\Phi\in\ANNs$ such that
	$\inDimANN(\Phi)=2$,
	$\outDimANN(\Phi)=1$,
	$\paramANN(\Phi)<20$,
	and
	\begin{equation}
		\sup_{v=(x,y)\in[0,2]^2}\babs{x^2+y^2-2x-2y+2-(\functionANN{\rect}(\Phi))(v)}
		\leq
		\tfrac38
		.
	\end{equation}
\end{exercise}

\subsection{Covering number estimates}

 \begin{adef}{def:covering_number}[Covering numbers]
 Let
 $ ( E, \delta ) $
 be a metric space
 and
 let
 $ r \in [ 0, \infty ] $.
 Then we denote by
 $ \CovNum{ ( E, \delta ), r } \in \N_0 \cup \{ \infty \} $ 
 (we denote by
 $ \CovNum{ E, r } \in \N_0 \cup \{ \infty \} $)
 the extended real number given by
 \begin{equation}
 \label{def:covering_number:eq1}
 \begin{split}
 \CovNum{ ( E, \delta ), r }
 =
 \min\biggl(
     \biggl\{
         n \in \N_0 \colon
         \biggl[
         \exists \, A \subseteq E \colon
         \biggl(
         \arraycolsep=0pt \begin{array}{c}
             ( \lvert A \rvert \leq n )
             \land
             ( \forall \, x \in E \colon \\
             \exists \, a \in A \colon \delta( a, x ) \leq r )
         \end{array}
         \biggr)
         \biggr]
     \biggr\}
     \cup \{ \infty \}
 \biggr)
 \end{split}
 \end{equation}
 and we call $ \CovNum{ (E,\delta), r }$ the 
$ r $-covering number of $ ( E, \delta ) $ 
(we call $ \CovNum{ E, r } $ the $ r $-covering number of $ E $). 
 \end{adef}

\todoc{ARNULF: I moved the three following lemmas from section 12. ok?}

\begin{lemma}
  \label{lem:covering_radius_aux1}
  Let $(X,d)$ be a metric space,
  let $n\in\N$, $r\in[0,\infty]$,
  assume $X\neq\emptyset$,
  and let $A\subseteq X$ 
    satisfy 
      $\card A\leq n$
      and $\Forall x\in X\colon \Exists a\in A\colon d(a,x)\leq r$.
  Then there exist 
    $x_1,x_2,\dots,x_n\in X$
  such that
  \begin{equation}
    X\subseteq \Biggl[
      \bigcup\limits_{i=1}^n \{v\in X\colon d(x_i,v)\leq r\}
    \Biggr].
  \end{equation}
\end{lemma}
\begin{proof}[Proof of \cref{lem:covering_radius_aux1}]
  Note that 
    the assumption that $X\neq\emptyset$
    and the assumption that $\card A\leq n$
  imply that there exist
    $x_1,x_2,\dots,x_n\in X$
  which satisfy  $A\subseteq\{x_1,x_2,\dots,x_n\}$.
    This
    and the assumption that $\Forall x\in X\colon \Exists a\in A\colon d(a,x)\leq r$
  ensure that
  \begin{equation}
    X
    \subseteq
    \Biggl[\bigcup_{a\in A}\{v\in X\colon d(a,v)\leq r\}\Biggr]
    \subseteq
    \Biggl[\bigcup_{i=1}^n \{v\in X\colon d(x_i,v)\leq r\}\Biggr]
    .
  \end{equation}
  The proof of \cref{lem:covering_radius_aux1} is thus complete.
\end{proof}

\begin{lemma}
  \label{lem:covering_radius_aux2}
  Let $(X,d)$ be a metric space,
  let $n\in\N$, $r\in[0,\infty]$, $x_1,x_2,\dots,x_n\in X$
    satisfy $X\subseteq \bigl[
      \bigcup_{i=1}^n \{v\in X\colon d(x_i,v)\leq r\}
    \bigr]$.
  Then there exists $A\subseteq X$ such that $\card A\leq n$ and
  \begin{equation}
    \Forall x\in X\colon \Exists a\in A\colon d(a,x)\leq r.
  \end{equation}
\end{lemma}
\begin{proof}[Proof of \cref{lem:covering_radius_aux2}]
  Throughout this proof, let $A=\{x_1,x_2,\dots,x_n\}$.
  Note that 
    the assumption that $X\subseteq \bigl[
      \bigcup_{i=1}^n \{v\in X\colon d(x_i,v)\leq r\}
    \bigr]$
  implies that for all
    $v\in X$
  there exists $i\in\{1,2,\dots,n\}$ such that
    $d(x_i,v)\leq r$.
  Hence, we obtain that
  \begin{equation}
    \Forall x\in X\colon \Exists a\in A\colon d(a,x)\leq r.
  \end{equation}
  The proof of \cref{lem:covering_radius_aux2} is thus complete.
\end{proof}

\begin{lemma}
  \label{lem:covering_radius_aux}
  Let $(X,d)$ be a metric space,
  let $n\in\N$, $r\in[0,\infty]$,
  and assume $X\neq\emptyset$.
  Then the following two statements are equivalent:
  \begin{enumerate}[label=(\roman *)]
    \item \label{it:cra.1}
      There exists $A\subseteq X$ such that
      $\card A\leq n$ and
      $
        \Forall x\in X\colon
          \Exists a\in A\colon
          d(a,x)\leq r
      $.
    \item \label{it:cra.2}
      There exist $x_1,x_2,\dots,x_n\in X$ such that
      $
        X\subseteq\bigl[
          \bigcup_{i=1}^n \{v\in X\colon d(x_i,v)\leq r\}
        \bigr]
      $.
  \end{enumerate}
\end{lemma}
\begin{proof}[Proof of \cref{lem:covering_radius_aux}]
  Note that
    \cref{lem:covering_radius_aux1}
    and \cref{lem:covering_radius_aux2}
  prove that (\ref{it:cra.1} $\leftrightarrow$ \ref{it:cra.2}).
  The proof of \cref{lem:covering_radius_aux} is thus complete.
\end{proof}

 \cfclear
 \begin{lemma}
   \label{lem:covering_number_char}
   Let $(E,\delta)$ be a metric space
   and let $r\in[0,\infty]$.
   Then
   \begin{equation}
     \label{eq:cnc.claim}
     \CovNum{(E,\delta),r}
     =
     \begin{cases}
       0 & \colon X=\emptyset \\
       \begin{aligned}
         \inf\biggl(
         \biggl\{ 
           n \in \N \colon
           \biggl(
             &\exists \, x_1, x_2, \dots, x_n \in E \colon
               \\& E \subseteq 
               \br*{ 
                 \textstyle
                 \bigcup\limits_{ m = 1 }^n 
                 \displaystyle
                 \cu*{ 
                   v \in E \colon
                   d( x_m , v ) \leq r
                 }
               }
           \biggr)
         \biggr\}
         \cup \{ \infty \}
         \biggr)
       \end{aligned}
       & \colon X\neq\emptyset
     \end{cases}
   \end{equation}
   \cfout.
 \end{lemma}
 \begin{proof}[Proof of \cref{lem:covering_number_char}]
   Throughout this proof,
     assume without loss of generality that $E\neq\emptyset$.
   Observe that
     \cref{lem:covering_radius_aux}
   establishes
     \cref{eq:cnc.claim}.
   The proof of \cref{lem:covering_number_char} is thus complete.
 \end{proof}

\begin{exercise}{ex:covering_number_subspace}
	Prove or disprove the following statement: For every
	metric space $(X,d)$, every $Y\subseteq X$,
	and every $r\in[0,\infty]$ it holds that
	$\CovNum{(Y,d|_{Y\times Y}),r}\leq \CovNum{(X,d),r}$.
\end{exercise}

\begin{exercise}{ex:covering_number_infinity}
	Prove or disprove the following statement:
	For every metric space $(E,\delta)$ it holds that
	$\CovNum{(E,\delta),\infty}=1$.
\end{exercise}

\begin{exercise}{ex:covering_number_bounded}
	Prove or disprove the following statement:
	For every metric space $(E,\delta)$ and every
	$r\in[0,\infty)$ with $\CovNum{(E,\delta),r}<\infty$
	it holds that $E$ is bounded. (\emph{Note:} A metric space
	$(E,\delta)$ is bounded if and only if there exists
	$r\in[0,\infty)$ such that it holds for all
	$x,y\in E$ that $\delta(x,y)\leq r$.)
\end{exercise}

\begin{exercise}{ex:covering_number_bounded2}
	Prove or disprove the following statement:
	For every bounded metric space $(E,\delta)$ and every
	$r\in[0,\infty]$ it holds that $\CovNum{(E,\delta),r}<\infty$.
\end{exercise}

\cfclear
\begin{athm}{lemma}{lem:covering_number_cube}
Let
$ d \in \N $,
$ a \in \R $,
$ b \in ( a, \infty ) $,
$ r \in ( 0, \infty ) $
and
for every
$ p \in [ 1, \infty ) $
let
$ \delta_p \colon ( [ a, b ]^d ) \times ( [ a, b ]^d ) \to [ 0, \infty ) $
satisfy for all
$ x, y \in [ a, b ]^d $
that
$ \delta_p( x, y ) = \pnorm{p}{ x - y } $
\cfload.
Then
it holds for all
$ p \in [ 1, \infty ) $
that
\begin{equation}
\CovNum{ ( [ a, b ]^d, \delta_p ), r }
\leq
\Bigl(
 \ceil*{
    \tfrac{ d^{ \nicefrac{1}{p} } ( b - a ) }{2r} 
 }
\Bigr)^{ \! d }
\leq
\begin{cases}
1
& \colon
r \geq \nicefrac{ d ( b - a ) }{2}
\\
\bigl(
    \tfrac{ d ( b - a ) }{r}
\bigr)^d
& \colon
r < \nicefrac{ d ( b - a ) }{2}.
\end{cases}
\end{equation}
\cfout.
\end{athm}

\begin{aproof}
Throughout this proof,
let
$ ( \fN_p )_{ p \in [ 1, \infty ) } \subseteq \N $
satisfy for all
$ p \in [ 1, \infty ) $
that
\begin{equation}
\label{eq:def_fNp}
\fN_p =
\ceil*{
    \tfrac{ d^{ \nicefrac{1}{p} } ( b - a ) }{2r}
}
,
\end{equation}
for every
$ N \in \N $,
$ i \in \{ 1, 2, \ldots, N \} $
let
$ g_{ N, i } \in [ a, b ] $
be given by
\begin{equation}
\label{covering_number_cube:eq1}
\begin{split} 
g_{ N, i } = a + \nicefrac{ ( i - \nicefrac{1}{2} ) ( b - a ) }{N} 
\end{split}
\end{equation}
and
for every
$ p \in [ 1, \infty ) $
let
$ A_p \subseteq [ a, b ]^d $
be given by
\begin{equation}
\label{covering_number_cube:eq2}
\begin{split} 
A_p =
\{ g_{ \fN_p, 1 }, g_{ \fN_p, 2 }, \ldots, g_{ \fN_p, \fN_p } \}^d 
\end{split}
\end{equation}
\cfload.
Observe that it holds for all
$ N \in \N $,
$ i \in \{ 1, 2, \ldots, N \} $,
$ x \in [ a + \nicefrac{ ( i - 1 )( b - a ) }{N}, g_{ N, i } ] $
that
\begin{equation}
\label{eq:grid_estimate1}
\lvert x - g_{ N, i } \rvert
=
a + \tfrac{ ( i - \nicefrac{1}{2} ) ( b - a ) }{N}
-
x
\leq
a + \tfrac{ ( i - \nicefrac{1}{2} ) ( b - a ) }{N}
-
\bigl( a + \tfrac{ ( i - 1 )( b - a ) }{N} \bigr)
=
\tfrac{ b - a }{2N}
.
\end{equation}
In addition,
note that it holds for all
$ N \in \N $,
$ i \in \{ 1, 2, \ldots, N \} $,
$ x \in [ g_{ N, i }, a + \nicefrac{ i ( b - a ) }{N} ] $
that%
\begin{equation}
\label{eq:grid_estimate2}
\lvert x - g_{ N, i } \rvert
=
x
-
\bigl( a + \tfrac{ ( i - \nicefrac{1}{2} ) ( b - a ) }{N} \bigr)
\leq
a + \tfrac{ i ( b - a ) }{N}
-
\bigl( a + \tfrac{ ( i - \nicefrac{1}{2} ) ( b - a ) }{N} \bigr)
=
\tfrac{ b - a }{2N}
.
\end{equation}
Combining this with \cref{eq:grid_estimate1}
implies for all
$ N \in \N $,
$ i \in \{ 1, 2, \ldots, N \} $,
$ x \in [ a + \nicefrac{ ( i - 1 )( b - a ) }{N}, a + \nicefrac{ i ( b - a ) }{N} ] $
that
$ \lvert x - g_{ N, i } \rvert
\leq
\nicefrac{ ( b - a ) }{(2N)} $.
This proves that
for every
$ N \in \N $,
$ x \in [ a, b ] $
there exists
$ y \in \{ g_{ N, 1 }, g_{ N, 2 }, \ldots, g_{ N, N } \} $
such that
\begin{equation}
\label{eq:grid_estimate_final}
\lvert x - y \rvert
\leq
\tfrac{ b - a }{2N}
.
\end{equation}
This
establishes that
for every
$ p \in [ 1, \infty ) $,
$ x = ( x_1, \ldots, x_d )
\in [ a, b ]^d $
there exists
$ y = ( y_1, \ldots, y_d )
\in A_p $
such that
\begin{equation}
\label{eq:covering_Lp}
\delta_p( x, y )
=
\pnorm p{x - y}
=
\biggl(
    \smallsum_{i=1}^d
        \lvert x_i - y_i \rvert^p
\biggr)^{ \!\! \nicefrac{1}{p} }
\leq
\biggl(
    \smallsum_{i=1}^d
        \tfrac{ ( b - a )^p }{ ( 2\fN_p )^p }
\biggr)^{ \!\! \nicefrac{1}{p} }
=
\tfrac{ d^{ \nicefrac{1}{p} } ( b - a ) }{2\fN_p}
\leq
\tfrac{ d^{ \nicefrac{1}{p} } ( b - a ) 2r }{ 2 d^{ \nicefrac{1}{p} } ( b - a ) }
=
r
.
\end{equation}
Combining
	this
with
\cref{def:covering_number:eq1},
\cref{covering_number_cube:eq2},
\cref{eq:def_fNp},
and
the fact that
$ \forall \, x \in [ 0, \infty ) \colon
\lceil x \rceil
\leq
\ind{ ( 0, 1 ] }( x )
+
2 x \ind{ ( 1, \infty ) }( x )
=
\ind{ ( 0, r ] }( r x )
+
2 x \ind{ ( r, \infty ) }( r x ) $
yields that for all
$ p \in [ 1, \infty ) $
it holds that
\begin{equation}
\begin{split}
\CovNum{ ( [ a, b ]^d, \delta_p ), r }
& \leq
\lvert A_p \rvert = ( \fN_p )^d
=
\Bigl(
\Bigl\lceil
    \tfrac{ d^{ \nicefrac{1}{p} } ( b - a ) }{2r}
\Bigr\rceil
\Bigr)^{ \! d }
\leq
\bigl(
\bigl\lceil
    \tfrac{ d ( b - a ) }{2r}
\bigr\rceil
\bigr)^d
\\ &
\leq
\bigl(
\ind{ ( 0, r ] }\bigl( \tfrac{ d ( b - a ) }{2} \bigr)
+
\tfrac{ 2 d ( b - a ) }{2r}
\ind{ ( r, \infty ) }\bigl( \tfrac{ d ( b - a ) }{2} \bigr)
\bigr)^d
\\ &
=
\ind{ ( 0, r ] }\bigl( \tfrac{ d ( b - a ) }{2} \bigr)
+
\bigl(
\tfrac{ d ( b - a ) }{r}
\bigr)^d
\ind{ ( r, \infty ) }\bigl( \tfrac{ d ( b - a ) }{2} \bigr)
\end{split}
\end{equation}
\cfload.
\end{aproof}

\subsection{Convergence rates for the approximation error}

\cfclear
\begin{athm}{lemma}{lem:approximation_error_structured}
Let
$ d  \in \N $,
$ L, a \in \R $,
$ b \in ( a, \infty ) $,
let
$ f \colon [ a, b ]^d \to  \R $
satisfy for all
$ x, y \in [ a, b ]^d $
that
$ \lvert f( x ) - f( y ) \rvert \leq L \pnorm1{ x - y } $,
and let 
$
\interpolatingDNN 
= 
\AffineANN_{0, f(\nicefrac{ ( a + b ) }{2}, \nicefrac{ ( a + b ) }{2}, \ldots, \nicefrac{ ( a + b ) }{2})}
	 \in \R^{1 \times d} \times \R^1
$
\cfload.
Then
\begin{enumerate}[label=(\roman *)]
\item \label{approximation_error_structured:item1}
it holds that $\inDimANN(\interpolatingDNN) = d$,

\item \label{approximation_error_structured:item2}
it holds that $\outDimANN(\interpolatingDNN) = 1$,

\item \label{approximation_error_structured:item3}
it holds that 
$
	\hiddenLength (\interpolatingDNN) 
=
	0
$,

\item \label{approximation_error_structured:item4}
it holds that
$\paramANN(\interpolatingDNN) = d + 1$,

\item \label{approximation_error_structured:item5}
it holds that 
$ \infnorm{ \MappingStructuralToVectorized(\interpolatingDNN) }
\leq \sup_{ x \in [ a, b ]^d } \lvert f( x ) \rvert $,
and

\item \label{approximation_error_structured:item6}
it holds that 
$
\sup\nolimits_{ x \in [ a, b ]^d }
    \lvert (\functionANN\rect(\interpolatingDNN))( x ) - f( x ) \rvert
\leq
\frac{ d L ( b - a ) }{ 2 }
$
\end{enumerate}
\cfout.
\end{athm}

\begin{aproof}
Note that the assumption that for all
$ x, y \in [ a, b ]^d $
it holds that
$ \lvert f( x ) - f( y ) \rvert \leq L \pnorm1{ x - y } $ assures that $L \geq 0$.
Next observe that \cref{lem:ANN:affine} assures that for all
	$ x \in \R^d $
it holds that
\begin{equation}
	(\functionANN\rect(\interpolatingDNN))( x )
=
	f\bpr{\nicefrac{ ( a + b ) }{2}, \nicefrac{ ( a + b ) }{2}, \ldots, \nicefrac{ ( a + b ) }{2}}
.
\end{equation}
\enum{
	The fact that for all 
		$x \in [a,b]$
	it holds that 
	$\abs{x -  \nicefrac{ ( a + b ) }{2}} \leq \nicefrac{ ( b - a ) }{2}$;
	the assumption that for all 
	$ x, y \in [ a, b ]^d $ it holds that 
	$\lvert f( x ) - f( y ) \rvert \leq L \pnorm1{ x - y } $
}
hence ensure 
that for all
$ x = ( x_1, \ldots, x_d ) \in [ a, b ]^d $
it holds that
\begin{equation}
\begin{split}
\lvert (\functionANN\rect(\interpolatingDNN))( x ) - f( x ) \rvert
& =
\lvert f\bpr{\nicefrac{ ( a + b ) }{2}, \nicefrac{ ( a + b ) }{2}, \ldots, \nicefrac{ ( a + b ) }{2}} - f( x ) \rvert \\
&\leq
L \bpnorm1{ \bpr{\nicefrac{ ( a + b ) }{2}, \nicefrac{ ( a + b ) }{2}, \ldots, \nicefrac{ ( a + b ) }{2}} - x } \\
&=
L
\smallsum_{i=1}^d
    \lvert \nicefrac{ ( a + b ) }{2} - x_i \rvert
\leq
\smallsum_{i=1}^d
    \tfrac{ L ( b - a ) }{2}
=
	\tfrac{ d L ( b - a ) }{2}
.
\end{split}
\end{equation}
This and the fact that
$ \infnorm{ \MappingStructuralToVectorized(\interpolatingDNN) }
= \lvert f(\nicefrac{ ( a + b ) }{2}, \nicefrac{ ( a + b ) }{2}, \ldots, \nicefrac{ ( a + b ) }{2}) \rvert
\leq \sup_{ x \in [ a, b ]^d } \lvert f( x ) \rvert $
complete
\end{aproof}

\cfclear
\begin{prop}%
\label{prop:approximation_error_structured_rsmall}
Let
$ d \in \N $,
$ L, a \in \R $,
$ b \in ( a, \infty ) $,
$r \in (0, \nicefrac{d}{4})$, 
let 
$ f \colon [ a, b ]^d \to \R $
and
$\delta \colon [a,b]^d \times [a,b]^d \to \R$ 
satisfy for all
$ x, y \in [ a, b ]^d $
that
$ \abs{ f( x ) - f( y ) } \leq L \pnorm1{ x - y } $
and $\delta(x,y) = \pnorm1{x-y}$,
and
let 
	$K \in \N$,
	$\fx_1, \fx_2, \ldots, \fx_K \in [a,b]^d$,
	$\fy \in \R^{K}$,
	$\interpolatingDNN \in \ANNs$ 
satisfy
	$K = \CovNum{ ( [ a, b ]^d, \delta ), (b-a)r } $,
	$\sup\nolimits_{x \in [a,b]^d} \br*{ \min\nolimits_{k  \in \{1,2, \ldots, K \}} \delta (x,  \fx_k) } \leq (b-a)r$,
	$\fy = (f(\fx_1), f(\fx_2), \ldots, f(\fx_K))$,
and
\begin{equation}
\label{prop:approximation_error_structured_rsmall:ass1}
\begin{split} 
	\interpolatingDNN
=
	\maxANN_{K} \compANNbullet \AffineANN_{-L \idMatrix_{K}, \fy} \compANNbullet 
	\parallelizationSpecial_{K}  \bigl( \oneNormANN_d \compANNbullet \AffineANN_{\idMatrix_d, -\fx_1} , \oneNormANN_d \compANNbullet \AffineANN_{\idMatrix_d, -\fx_2}, \ldots, \oneNormANN_d \compANNbullet \AffineANN_{\idMatrix_d, -\fx_K} \bigr) \compANNbullet \extensionANN_{d,K}	
\end{split}
\end{equation}
\cfload.
Then 
\begin{enumerate}[label=(\roman *)]
\item \label{prop:approximation_error_structured_rsmall:item1}
it holds that $\inDimANN(\interpolatingDNN) = d$,

\item \label{prop:approximation_error_structured_rsmall:item2}
it holds that $\outDimANN(\interpolatingDNN) = 1$,

\item \label{prop:approximation_error_structured_rsmall:item3}
it holds that 
$
	\hiddenLength (\interpolatingDNN) 
\leq
	\ceil*{ d \operatorname{log}_2 \bpr{  \tfrac{3d}{4r} }} + 1
$,
    
\item \label{prop:approximation_error_structured_rsmall:item4}
it holds that 
$
	\dimANNlevel_1 ( \interpolatingDNN) 
\leq 
	2d \bpr{  \tfrac{3d}{4r}  }^{ d }
$,

\item \label{prop:approximation_error_structured_rsmall:item5}
it holds for all $i \in \{2,3,4,\ldots\}$ that 
$
	\dimANNlevel_i (\interpolatingDNN)
\leq 
	3 \ceil[\big]{ \pr[\big]{ \tfrac{3d}{4r} }^{ d } \tfrac{1}{2^{i-1}} } 
$,

\item \label{prop:approximation_error_structured_rsmall:item6}
it holds that
$
	\paramANN(\interpolatingDNN)
\leq
	35 \bpr{  \tfrac{3d}{4r}  }^{ 2d }d^2
$,

\item \label{prop:approximation_error_structured_rsmall:item7}
it holds that
$ \infnorm{ \MappingStructuralToVectorized(\interpolatingDNN) }
\leq \max\{ 1, L, \abs{a}, \abs{b}, \allowbreak 2[ \sup_{ x \in [ a, b ]^d } \abs{ f( x ) } ] \} $,
and

\item \label{prop:approximation_error_structured_rsmall:item8}
it holds that
$
\sup\nolimits_{ x \in [ a, b ]^d }
    \abs{ (\functionANN{\rect}(\interpolatingDNN))( x ) - f( x ) }
\leq
	2L(b-a)r
$
\end{enumerate}
\cfout.
\end{prop}

\begin{proof}[Proof of \cref{prop:approximation_error_structured_rsmall}]
Note that the assumption that for all
$ x, y \in [ a, b ]^d $
it holds that
$ \lvert f( x ) - f( y ) \rvert \leq L \pnorm1{ x - y } $ assures that $L \geq 0$.
Next observe that \eqref{prop:approximation_error_structured_rsmall:ass1}, \cref{dnn:intl1}, and \cref{dnn:intp1} demonstrate that 
\begin{enumerate}[label=(\Roman{*})]
\item\label{approximation_error_structured_rsmall:it1} 
	it holds that $\inDimANN(\interpolatingDNN)=d$,
\item\label{approximation_error_structured_rsmall:it2} 
    it holds that $\outDimANN(\interpolatingDNN)=1$,    
\item\label{approximation_error_structured_rsmall:it3} 
     it holds that $\hiddenLength (\interpolatingDNN) = \ceil{ \operatorname{log}_2 (K)} + 1$,
\item\label{approximation_error_structured_rsmall:it4} 
     it holds that $\dimANNlevel_1(\interpolatingDNN) =2d K$,
\item\label{approximation_error_structured_rsmall:it5} 
     it holds for all $i \in \{2,3,4,\ldots\}$ that $\dimANNlevel_i(\interpolatingDNN) \leq 3 \ceil*{ \frac{K}{2^{i-1}} } $,
\item\label{approximation_error_structured_rsmall:it6} 
     it holds that $\pnorm\infty{ \MappingStructuralToVectorized(\interpolatingDNN)} \leq \max \{ 1,L, \max_{k \in \{1, 2, \ldots, K\}} \infnorm{ \fx_k }, 2 [\max_{k \in \{1, 2, \ldots, K\}} \abs{f(\fx_k)}]\}$, and
\item\label{approximation_error_structured_rsmall:it7} 
it holds that
$
	\sup\nolimits_{x \in [a,b]^d} \abs{(\functionANN\rect(\interpolatingDNN))(x) - f(x)} 
\leq 
	2L \br*{\sup\nolimits_{x \in [a,b]^d} \pr*{ \min\nolimits_{k \in \{1,2, \ldots, K\}} \delta (x,  \fx_k)}}
$
\end{enumerate}
\cfload. 
Note that items \ref{approximation_error_structured_rsmall:it1} and \ref{approximation_error_structured_rsmall:it2} establish items \ref{prop:approximation_error_structured_rsmall:item1} and \ref{prop:approximation_error_structured_rsmall:item2}.
Next observe that 
\enum{
	\cref{lem:covering_number_cube};
	the fact that $\tfrac{ d  }{2r} \geq 2$
}[imply] 
that
\begin{equation}
\label{approximation_error_structured_rsmall:eq1}
\begin{split}
	K 
=  
	\CovNum{ ( [ a, b ]^d, \delta ), (b-a)r }
\leq
	 \pr[\Big]{ \ceil*{ \tfrac{ d ( b - a ) }{2(b-a)r} } }^{ d }
=
	 \pr[\big]{ \ceil*{ \tfrac{ d  }{2r} } }^{ d }
\leq
	 \pr[\big]{ \tfrac{3}{2}  (\tfrac{ d  }{2r} )  }^{ d }
=
	 \pr[\big]{ \tfrac{3d}{4r} }^{ d }.
\end{split}
\end{equation}
Combining this with item~\ref{approximation_error_structured_rsmall:it3} assures that
\begin{equation}
\label{approximation_error_structured_rsmall:eq2}
\begin{split} 
	\hiddenLength (\interpolatingDNN)
= 
	\ceil{ \operatorname{log}_2 (K)} + 1
\leq
	\ceil*{ \operatorname{log}_2 \pr[\Big]{\pr[\big]{ \tfrac{3d}{4r} }^{ d } }} + 1
=
	\ceil{ d \operatorname{log}_2 \pr[\big]{ \tfrac{3d}{4r} }} + 1.
\end{split}
\end{equation}
This establishes item~\ref{prop:approximation_error_structured_rsmall:item3}.
Moreover, note that \eqref{approximation_error_structured_rsmall:eq1} and item~\ref{approximation_error_structured_rsmall:it4}
imply that
\begin{equation}
\label{approximation_error_structured_rsmall:eq3}
\begin{split} 
	\dimANNlevel_1(\interpolatingDNN) 
=
	2d K
\leq
	2d \pr[\big]{  \tfrac{3d}{4r}  }^{ d }.
\end{split}
\end{equation}
This establishes item~\ref{prop:approximation_error_structured_rsmall:item4}.
In addition, observe that
	\cref{approximation_error_structured_rsmall:it5}
	and \eqref{approximation_error_structured_rsmall:eq1}
establish \cref{prop:approximation_error_structured_rsmall:item5}.
Next note that
  \cref{approximation_error_structured_rsmall:it3}
ensures that for all
	$i\in\N\cap(1,\hiddenLength(\interpolatingDNN)]$
it holds that
\begin{equation}
	\tfrac K{2^{i-1}}
	\geq
	\tfrac K{2^{\hiddenLength(\interpolatingDNN)-1}}
	=
	\tfrac K{2^{\ceil{\logg_2(K)}}}
	\geq
	\tfrac K{2^{\logg_2(K)+1}}
	=
	\tfrac K{2K}
	=
	\tfrac 12
	.
\end{equation}
	\Cref{approximation_error_structured_rsmall:it5}
	and \eqref{approximation_error_structured_rsmall:eq1}
  hence 
show that for all
	$i\in\N\cap(1,\hiddenLength(\interpolatingDNN)]$
it holds that
\begin{equation}
\label{approximation_error_structured_rsmall:eq4}
\begin{split} 
	\dimANNlevel_i(\interpolatingDNN) 
\leq 
	3 \ceil*{ \tfrac{K}{2^{i-1}} } 
\leq 
	\tfrac{3K}{2^{i-2}}
\leq
	\pr[\big]{ \tfrac{3d}{4r} }^{ d } \tfrac{3}{2^{i-2}}.
\end{split}
\end{equation}
Furthermore, note that 
\enum{
	the fact that for all 
		$x \in [a,b]^d $
	it holds that
	$\infnorm{x} \leq \max\{\abs{a},\abs{b}\}$;
	item~\ref{approximation_error_structured_rsmall:it6};
}[imply]
that
\begin{equation}
\label{approximation_error_structured_rsmall:eq5}
\begin{split} 
	\pnorm\infty{ \MappingStructuralToVectorized(\interpolatingDNN)} 
&\leq 
	\max \{ 1,L, \max\nolimits_{k \in \{1, 2, \ldots, K\}} \infnorm{ \fx_k }, 2 [\max\nolimits_{k \in \{1, 2, \ldots, K\}} \abs{f(\fx_k)}]\} \\
&\leq
	\max \{ 1,L, \abs{a}, \abs{b}, 2 [\sup\nolimits_{x \in [a,b]^d} \abs{f(x)}] \}.
\end{split}
\end{equation}
This establishes item~\ref{prop:approximation_error_structured_rsmall:item7}.
Moreover, observe that
	the assumption that 
	\begin{equation}
		\textstyle
		\sup\nolimits_{x \in [a,b]^d} \br*{ \min\nolimits_{k  \in \{1,2, \ldots, K \}} \delta (x,  \fx_k) } \leq (b-a)r
	\end{equation}
	and
	\cref{approximation_error_structured_rsmall:it7}
demonstrate
that
\begin{equation}
\label{approximation_error_structured_rsmall:eq6}
\begin{split} 
	\sup\nolimits_{x \in [a,b]^d} \abs{ (\functionANN\rect(\interpolatingDNN))(x) - f(x)} 
\leq 
	2L \br*{\sup\nolimits_{x \in [a,b]^d} \pr*{ \min\nolimits_{k \in \{1,2, \ldots, K\}} \delta (x,  \fx_k)}}
\leq
	2L (b-a) r.
\end{split}
\end{equation}
This establishes item~\ref{prop:approximation_error_structured_rsmall:item8}.
It thus remains to prove item~\ref{prop:approximation_error_structured_rsmall:item6}.
For this note that 
\enum{
	items~\ref{approximation_error_structured_rsmall:it1} and \ref{approximation_error_structured_rsmall:it2};
	\eqref{approximation_error_structured_rsmall:eq3};
	\eqref{approximation_error_structured_rsmall:eq4}
}[assure]
that
\begin{equation}
\label{approximation_error_structured_rsmall:eq7}
\begin{split} 
	\paramANN(\interpolatingDNN)
&=
	\sum_{i = 1}^{\lengthANN(\interpolatingDNN)} \dimANNlevel_i(\interpolatingDNN) (\dimANNlevel_{i-1}(\interpolatingDNN) + 1) \\
&\leq
	2d \pr[\big]{  \tfrac{3d}{4r}  }^{ d } ( d + 1)
	+
	\pr[\big]{ \tfrac{3d}{4r} }^{ d } 3 \pr[\big]{ 2d \pr[\big]{  \tfrac{3d}{4r}  }^{ d } + 1} \\
&\quad
	+
	\br*{
		\sum_{i = 3}^{\lengthANN(\interpolatingDNN)-1}
			\pr[\big]{ \tfrac{3d}{4r} }^{ d } \tfrac{3}{2^{i-2}} \pr[\big]{\pr[\big]{ \tfrac{3d}{4r} }^{ d } \tfrac{3}{2^{i-3}} + 1}
	}
	+
	\pr[\big]{ \tfrac{3d}{4r} }^{ d } \tfrac{3}{2^{\lengthANN(\interpolatingDNN)-3}} + 1.
\end{split}
\end{equation}
Next note that the fact that $ \tfrac{3d}{4r} \geq 3$ ensures that
\begin{equation}
\label{approximation_error_structured_rsmall:eq8}
\begin{split} 
	&2d \pr[\big]{  \tfrac{3d}{4r}  }^{ d } ( d + 1)
	+
	\pr[\big]{ \tfrac{3d}{4r} }^{ d } {3} \pr[\big]{ 2d \pr[\big]{  \tfrac{3d}{4r}  }^{ d } + 1}
	+
	\pr[\big]{ \tfrac{3d}{4r} }^{ d } \tfrac{3}{2^{\lengthANN(\interpolatingDNN)-3}} + 1\\
&\leq
	\pr[\big]{  \tfrac{3d}{4r}  }^{ 2d }
	\pr[\big]{
		2d  ( d + 1)
		+
		 3 ( 2d  + 1)
		+
		\tfrac{3}{2^{1-3}} 
		+ 
		1
	}\\
&\leq
	\pr[\big]{  \tfrac{3d}{4r}  }^{ 2d }d^2
		(
			4
			+
			9
			+
			12
			+ 
			1
		)
=
	26 \pr[\big]{  \tfrac{3d}{4r}  }^{ 2d }d^2.
\end{split}
\end{equation}
Moreover, observe that the fact that $ \tfrac{3d}{4r} \geq 3$ implies that
\begin{equation}
\label{approximation_error_structured_rsmall:eq9}
\begin{split} 
	\sum_{i = 3}^{\lengthANN(\interpolatingDNN)-1}
		\pr[\big]{ \tfrac{3d}{4r} }^{ d } \tfrac{3}{2^{i-2}} \pr[\big]{\pr[\big]{ \tfrac{3d}{4r} }^{ d } \tfrac{3}{2^{i-3}} + 1}
&\leq
	\pr[\big]{ \tfrac{3d}{4r} }^{ 2d }
	\sum_{i = 3}^{\lengthANN(\interpolatingDNN)-1}
		 \tfrac{3}{2^{i-2}} \pr[\big]{ \tfrac{3}{2^{i-3}} + 1}\\
&=
	\pr[\big]{ \tfrac{3d}{4r} }^{ 2d }
		\sum_{i = 3}^{\lengthANN(\interpolatingDNN)-1}
			 \bbbr{ \tfrac{9}{2^{2i - 5}} + \tfrac{3}{2^{i - 2}} }\\
&=
	\pr[\big]{ \tfrac{3d}{4r} }^{ 2d }
		\sum_{i = 0}^{\lengthANN(\interpolatingDNN)-4}
			 \bbbr{ \tfrac{9}{2} (4^{-i})  + \tfrac{3}{2} (2^{-i}) }\\
&\leq
	\pr[\big]{ \tfrac{3d}{4r} }^{ 2d }
	\pr[\big]{
		\tfrac{9}{2} \pr[\big]{\tfrac{1}{1-4^{-1}}} + \tfrac{3}{2} \pr[\big]{\tfrac{1}{1-2^{-1}}}
	}
=
	9 \pr[\big]{ \tfrac{3d}{4r} }^{ 2d }.
\end{split}
\end{equation}
Combining this, \eqref{approximation_error_structured_rsmall:eq7}, and \eqref{approximation_error_structured_rsmall:eq8} demonstrates that
\begin{equation}
\label{approximation_error_structured_rsmall:eq10}
\begin{split} 
	\paramANN(\interpolatingDNN)
\leq
	26 \pr[\big]{  \tfrac{3d}{4r}  }^{ 2d }d^2 
	+ 
	9 \pr[\big]{ \tfrac{3d}{4r} }^{ 2d }
\leq
	35 \pr[\big]{  \tfrac{3d}{4r}  }^{ 2d }d^2.
\end{split}
\end{equation}
This establishes item~\ref{prop:approximation_error_structured_rsmall:item6}.
The proof of \cref{prop:approximation_error_structured_rsmall} is thus complete.
\end{proof}

\cfclear
\begin{prop}%
\label{prop:approximation_error_structured_r}
Let
$ d \in \N $,
$ L, a \in \R $,
$ b \in ( a, \infty ) $,
$r \in (0, \infty )$
and
let 
$ f \colon [ a, b ]^d \to \R $
satisfy for all
$ x, y \in [ a, b ]^d $
that
$ \abs{ f( x ) - f( y ) } \leq L \pnorm1{ x - y } $
\cfload.
Then there exists $\interpolatingDNN \in \ANNs$ such that
\begin{enumerate}[label=(\roman *)]
\item \label{prop:approximation_error_structured_r:item1}
it holds that $\inDimANN(\interpolatingDNN) = d$,

\item \label{prop:approximation_error_structured_r:item2}
it holds that $\outDimANN(\interpolatingDNN) = 1$,

\item \label{prop:approximation_error_structured_r:item3}
it holds that 
$
	\hiddenLength (\interpolatingDNN) 
\leq
	\pr[\big]{\ceil*{ d \operatorname{log}_2 \pr[\big]{  \tfrac{3d}{4r} }} + 1} \indicator{(0,\nicefrac{d}{4})}(r) 
$,
    
\item \label{prop:approximation_error_structured_r:item4}
it holds that 
$
	\dimANNlevel_1 ( \interpolatingDNN) 
\leq 
	2d \pr[\big]{  \tfrac{3d}{4r}  }^{ d } \indicator{(0,\nicefrac{d}{4})}(r) 
	+
	\indicator{[\nicefrac{d}{4}, \infty)}(r) 
$,

\item \label{prop:approximation_error_structured_r:item5}
it holds for all $i \in \{2,3,4,\ldots\}$ that 
$
	\dimANNlevel_i (\interpolatingDNN)
\leq 
	3 \ceil[\big]{ \pr[\big]{ \tfrac{3d}{4r} }^{ d } \tfrac{1}{2^{i-1}} } 
$,

\item \label{prop:approximation_error_structured_r:item6}
it holds that
$
	\paramANN(\interpolatingDNN)
\leq
	35 \pr[\big]{  \tfrac{3d}{4r}  }^{ 2d }d^2 \indicator{(0,\nicefrac{d}{4})}(r)
	+
	(d+1)\indicator{[\nicefrac{d}{4}, \infty)}(r)
$,

\item \label{prop:approximation_error_structured_r:item7}
it holds that
$ \infnorm{ \MappingStructuralToVectorized(\interpolatingDNN) }
\leq \max\{ 1, L, \abs{a}, \abs{b}, \allowbreak 2[ \sup_{ x \in [ a, b ]^d } \abs{ f( x ) } ] \} $,
and

\item \label{prop:approximation_error_structured_r:item8}
it holds that
$
\sup\nolimits_{ x \in [ a, b ]^d }
    \abs{ (\functionANN{\rect}(\interpolatingDNN))( x ) - f( x ) }
\leq
	2L(b-a)r
$
\end{enumerate}
\cfout.
\end{prop}

\begin{proof}[Proof of \cref{prop:approximation_error_structured_r}]
Throughout this proof, assume without loss of generality that $r < \nicefrac{d}{4}$
(cf.\ \cref{lem:approximation_error_structured}),
let $\delta \colon [a,b]^d \times [a,b]^d \to \R$ satisfy for all $x,y \in [a,b]^d$ that 
\begin{equation}
  \delta(x,y) = \pnorm1{x-y}
  ,
\end{equation}
and 
let $K \in \N \cup \{ \infty\}$ satisfy 
\begin{equation}
  K = \CovNum{ ( [ a, b ]^d, \delta ), (b-a)r } 
  .
\end{equation}
Note that \cref{lem:covering_number_cube} assures that $K < \infty$.
This and \eqref{def:covering_number:eq1} ensure that there exist $\fx_1, \fx_2, \ldots, \fx_K \in [a,b]^d$ 
such that
\begin{equation}
  \sup\nolimits_{ x \in [a,b]^d } 
  \br*{ 
    \min\nolimits_{k  \in \{1,2, \ldots, K \}} \delta (x,  \fx_k) 
  } 
  \leq (b - a) r
  .
\end{equation}
Combining this with \cref{prop:approximation_error_structured_rsmall} establishes 
\cref{prop:approximation_error_structured_r:item1,prop:approximation_error_structured_r:item2,prop:approximation_error_structured_r:item3,prop:approximation_error_structured_r:item4,prop:approximation_error_structured_r:item5,prop:approximation_error_structured_r:item6,prop:approximation_error_structured_r:item7,prop:approximation_error_structured_r:item8}.
The proof of \cref{prop:approximation_error_structured_r} is thus complete.
\end{proof}

\cfclear
\begingroup 
\begin{prop}[Implicit multi-dimensional \ann\ approximations with 
prescribed error tolerances and explicit parameter bounds]
\label{prop:approximation_error_structured_eps}
Let
$ d \in \N $,
$ L, a \in \R $,
$ b \in [ a, \infty ) $,
$
  \varepsilon \in (0, 1]
$
and
let 
$ f \colon [ a, b ]^d \to \R $
satisfy for all
$ x, y \in [ a, b ]^d $
that
\begin{equation}
  \abs{ f( x ) - f( y ) } \leq L \pnorm1{ x - y } 
\end{equation}
\cfload.
Then there exists $\interpolatingDNN \in \ANNs$ such that
\begin{enumerate}[label=(\roman *)]
\item \label{prop:approximation_error_structured_eps:item1}
it holds that $\inDimANN(\interpolatingDNN) = d$,

\item \label{prop:approximation_error_structured_eps:item2}
it holds that $\outDimANN(\interpolatingDNN) = 1$,

\item \label{prop:approximation_error_structured_eps:item3}
it holds that 
$
	\hiddenLength (\interpolatingDNN) 
\leq
	d 
	\pr[\big]{
	  \operatorname{log}_2\pr[\big]{
        \max\cu[\big]{
            \tfrac{ 3 d L ( b - a ) }{ 2 } 
        , 1} 
      }
      +  
      \operatorname{log}_2(
        \varepsilon^{ - 1 }
      )
    } 
    + 2
$,
    
\item \label{prop:approximation_error_structured_eps:item4}
it holds that 
$
	\dimANNlevel_1 ( \interpolatingDNN) 
\leq 
	\varepsilon^{-d} d (  3d\max\{L(b-a), 1\}  )^{ d }
$,

\item \label{prop:approximation_error_structured_eps:item5}
it holds for all $i \in \{2,3,4,\ldots\}$ that 
$
	\dimANNlevel_i (\interpolatingDNN)
\leq 
	\varepsilon^{-d} 3 \pr[\big]{ \tfrac{(3dL(b-a))^d}{2^{i}} + 1} 
$,

\item \label{prop:approximation_error_structured_eps:item6}
it holds that
$
	\paramANN(\interpolatingDNN)
\leq
	\varepsilon^{-2d} 9 \pr[\big]{ 3d\max\{L(b-a), 1\}  }^{ 2d }d^2
$,

\item \label{prop:approximation_error_structured_eps:item7}
it holds that
$ \infnorm{ \MappingStructuralToVectorized(\interpolatingDNN) }
\leq \max\{ 1, L, \abs{a}, \abs{b}, \allowbreak 2[ \sup_{ x \in [ a, b ]^d } \abs{ f( x ) } ] \} $,
and

\item \label{prop:approximation_error_structured_eps:item8}
it holds that
$
\sup\nolimits_{ x \in [ a, b ]^d }
    \abs{ (\functionANN{\rect}(\interpolatingDNN))( x ) - f( x ) }
\leq
	\eps
$
\end{enumerate}
\cfout.
\end{prop}

\begin{proof}[Proof of \cref{prop:approximation_error_structured_eps}]
Throughout this proof, assume without loss of generality that 
\begin{equation}
\label{eq:assume_wlog}
  L ( b - a ) \neq 0 
  .
\end{equation}
\Nobs that \cref{eq:assume_wlog} 
ensures that 
$ L \neq 0 $
and 
$ a < b $. 
Combining this with 
the assumption that
	for all
	$ x, y \in [ a, b ]^d $
	it holds that
\begin{equation}
  \abs{ f( x ) - f( y ) } \leq L \pnorm1{ x - y } ,
\end{equation}
ensures that $ L > 0 $.
\cref{prop:approximation_error_structured_r} \hence \proves that 
there exists $\interpolatingDNN \in \ANNs$ which satisfies that 
\begin{enumerate}[label=(\Roman *)]
\item \label{approximation_error_structured_eps:pfitem1}
it holds that $\inDimANN(\interpolatingDNN) = d$,

\item \label{approximation_error_structured_eps:pfitem2}
it holds that $\outDimANN(\interpolatingDNN) = 1$,

\item \label{approximation_error_structured_eps:pfitem3}
it holds that 
$
	\hiddenLength (\interpolatingDNN) 
\leq
	\pr[\big]{\ceil[\big]{ d \operatorname{log}_2 \pr[\big]{\tfrac{3dL(b-a)}{2\varepsilon} }} + 1} \indicator{(0,\nicefrac{d}{4})}\pr[\big]{\frac{\varepsilon}{2L(b-a)}} 
$,
    
\item \label{approximation_error_structured_eps:pfitem4}
it holds that 
$
	\dimANNlevel_1 ( \interpolatingDNN) 
\leq 
	2d \pr[\big]{  \tfrac{3dL(b-a)}{2\varepsilon}  }^{ d } \indicator{(0,\nicefrac{d}{4})}\pr[\big]{\frac{\varepsilon}{2L(b-a)}} 
	+
	\indicator{[\nicefrac{d}{4}, \infty)}\pr[\big]{\frac{\varepsilon}{2L(b-a)}}  
$,

\item \label{approximation_error_structured_eps:pfitem5}
it holds for all $i \in \{2,3,4,\ldots\}$ that 
$
	\dimANNlevel_i (\interpolatingDNN)
\leq 
	3 \ceil[\big]{ \pr[\big]{ \tfrac{3dL(b-a)}{2\varepsilon} }^{ d } \tfrac{1}{2^{i-1}} } 
$,

\item \label{approximation_error_structured_eps:pfitem6}
it holds that
$
	\paramANN(\interpolatingDNN)
\leq
	35 \pr[\big]{  \tfrac{3dL(b-a)}{2\varepsilon}  }^{ 2d }d^2 \indicator{(0,\nicefrac{d}{4})}\pr[\big]{\tfrac{\varepsilon}{2L(b-a)}}
	+
	(d+1)  \indicator{[\nicefrac{d}{4}, \infty)}\pr[\big]{\tfrac{\varepsilon}{2L(b-a)}}
$,

\item \label{approximation_error_structured_eps:pfitem7}
it holds that
$ \infnorm{ \MappingStructuralToVectorized(\interpolatingDNN) }
\leq \max\{ 1, L, \abs{a}, \abs{b}, \allowbreak 2[ \sup_{ x \in [ a, b ]^d } \abs{ f( x ) } ] \} $,
and

\item \label{approximation_error_structured_eps:pfitem8}
it holds that
$
\sup\nolimits_{ x \in [ a, b ]^d }
    \abs{ (\functionANN{\rect}(\interpolatingDNN))( x ) - f( x ) }
\leq
	\varepsilon
$
\end{enumerate}
\cfload.
\Nobs that \cref{approximation_error_structured_eps:pfitem3} assures that
\begin{equation}
\label{approximation_error_structured_eps:eq1}
\begin{split} 
	\hiddenLength (\interpolatingDNN) 
&\leq
	\pr[\big]{ d \pr[\big]{\operatorname{log}_2 \pr[\big]{\tfrac{3dL(b-a)}{2} } +  \operatorname{log}_2(\varepsilon^{-1})}+ 2} \indicator{(0,\nicefrac{d}{4})}\pr[\big]{\tfrac{\varepsilon}{2L(b-a)}} 
	\\
&\leq
	d \pr[\big]{\max\cu[\big]{\operatorname{log}_2 \pr[\big]{\tfrac{3dL(b-a)}{2} }, 0} +  \operatorname{log}_2(\varepsilon^{-1})}+2
  .
\end{split}
\end{equation}
\Moreover \cref{approximation_error_structured_eps:pfitem4} \proves that
\begin{equation}
\label{approximation_error_structured_eps:eq2}
\begin{split} 
	\dimANNlevel_1 ( \interpolatingDNN) 
&\leq
	d \pr[\big]{  \tfrac{3d\max\{L(b-a), 1\}}{\varepsilon}  }^{ d } \indicator{(0,\nicefrac{d}{4})} \pr[\big]{\tfrac{\varepsilon}{2L(b-a)}} 
	+
	\indicator{[\nicefrac{d}{4}, \infty)}\pr[\big]{\tfrac{\varepsilon}{2L(b-a)}} 
\\ 
&
  \leq
	\varepsilon^{-d} d (  3d\max\{L(b-a), 1\}  )^{ d } .
\end{split}
\end{equation}
\Moreover \cref{approximation_error_structured_eps:pfitem5} \proves that for all 
$ i \in \{ 2, 3, 4, \dots \} $ 
it holds that
\begin{equation}
\label{approximation_error_structured_eps:eq3}
\begin{split} 
	\dimANNlevel_i (\interpolatingDNN)
\leq 
	3 \pr[\big]{ \pr[\big]{ \tfrac{3dL(b-a)}{2\varepsilon} }^{ d } \tfrac{1}{2^{i-1}} + 1}
\leq
	\varepsilon^{-d} 3 \pr[\big]{ \tfrac{(3dL(b-a))^d}{2^{i}} + 1}.
\end{split}
\end{equation}
\Moreover \cref{approximation_error_structured_eps:pfitem6} ensures that
\begin{equation}
\label{approximation_error_structured_eps:eq4}
\begin{split} 
	\paramANN(\interpolatingDNN)
&\leq
	9 \pr[\big]{  \tfrac{3d\max\{L(b-a), 1\}}{\varepsilon}  }^{ 2d }d^2 \indicator{(0,\nicefrac{d}{4})}\pr[\big]{\tfrac{\varepsilon}{2L(b-a)}}
	+
	(d+1)  \indicator{[\nicefrac{d}{4}, \infty)}\pr[\big]{\tfrac{\varepsilon}{2L(b-a)}}\\
&\leq	
	\varepsilon^{-2d} 9 \pr[\big]{ 3d\max\{L(b-a), 1\}  }^{ 2d }d^2 .
\end{split}
\end{equation}
Combining
\enum{
	this;
	\eqref{approximation_error_structured_eps:eq1};
	\eqref{approximation_error_structured_eps:eq2};
	\eqref{approximation_error_structured_eps:eq3}}
with
\cref{approximation_error_structured_eps:pfitem1,approximation_error_structured_eps:pfitem2,approximation_error_structured_eps:pfitem7,approximation_error_structured_eps:pfitem8}
establishes \cref{prop:approximation_error_structured_eps:item1,%
prop:approximation_error_structured_eps:item2,%
prop:approximation_error_structured_eps:item3,%
prop:approximation_error_structured_eps:item4,%
prop:approximation_error_structured_eps:item5,%
prop:approximation_error_structured_eps:item6,%
prop:approximation_error_structured_eps:item7,%
prop:approximation_error_structured_eps:item8}.
The proof of \cref{prop:approximation_error_structured_eps} is thus complete.
\end{proof}
\endgroup

\cfclear
\begingroup
\renewcommand{\c}{\fC}
\begin{athm}{cor}{approx_error_rate2}[Implicit multi-dimensional \ann\ approximations with 
prescribed error tolerances and asymptotic parameter bounds]
Let
$ d  \in \N $,
$ L, a \in \R $,
$ b \in [ a, \infty ) $
and let
$ f \colon [ a, b ]^d \to  \R $
satisfy for all
$ x, y \in [ a, b ]^d $
that
\begin{equation}
  \abs{ f( x ) - f( y ) } \leq L \pnorm1{ x - y } 
\end{equation}
\cfload. 
Then there exist 
$ \c \in \R $ 
such that  for all 
	$ \varepsilon \in (0,1] $ 
there exists
$
  \interpolatingDNN \in \ANNs
$ 
such that 
\begin{equation}
		\hiddenLength (\interpolatingDNN) 
		\leq
			\c ( \operatorname{log}_2(\varepsilon^{-1}) + 1 )
	,\qquad
	\infnorm{ \MappingStructuralToVectorized(\interpolatingDNN) }
	\textstyle
	\leq \max\bcu{ 1, L, \abs{a}, \abs{b}, \allowbreak 2\bbr{ \sup_{ x \in [ a, b ]^d } \abs{ f( x ) } } } 
	,
\end{equation}
\begin{equation}
	\functionANN\rect(\interpolatingDNN) \in C(\R^d, \R)
	,\quad
	\sup\nolimits_{ x \in [ a, b ]^d }
    \abs{ (\functionANN\rect(\interpolatingDNN))( x ) - f( x ) }
\leq \varepsilon
,\quad\text{and}\quad
\paramANN(\interpolatingDNN) \leq \c \varepsilon^{-2d}
\end{equation}
\cfout.
\end{athm}

\begin{aproof}
Throughout this proof, 
let $ \c \in \R $ satisfy 
\begin{equation}
\llabel{eq:definition_of_C}
  \c = 9 \pr[\big]{ 3d\max\{L(b-a), 1\}  }^{ 2d }d^2
  .
\end{equation}
\Nobs that 
\cref{prop:approximation_error_structured_eps:item1,%
prop:approximation_error_structured_eps:item2,%
prop:approximation_error_structured_eps:item3,%
prop:approximation_error_structured_eps:item6,%
prop:approximation_error_structured_eps:item7,%
prop:approximation_error_structured_eps:item8} 
in \cref{prop:approximation_error_structured_eps}
and the fact that 
for all $ \eps \in (0,1] $
it holds that
\begin{equation}
\begin{split}
	d 
	\pr[\big]{
        \operatorname{log}_2\pr[\big]{ 
            \max\cu[\big]{
                \tfrac{3dL(b-a)}{2} 
            , 1 } 
        }
        +  
        \operatorname{log}_2( \varepsilon^{ - 1 } )
    } 
    + 2
&
\leq
    d \pr[\big]{
            \max\cu[\big]{
                \tfrac{ 3 d L ( b - a ) }{ 2 } 
            , 1 } 
      + 
      \operatorname{log}_2( \varepsilon^{ - 1 } )
    }
    + 2
\\ &
\leq
    d 
            \max\cu[\big]{
                3 d L ( b - a ) 
            , 1 } 
    + 2 
    + d \operatorname{log}_2( \varepsilon^{ - 1 } ) 
\\ & 
\leq
    \c (\operatorname{log}_2(\varepsilon^{-1})+1)
\end{split}
\end{equation}
imply that for every $\varepsilon \in (0,1]$ there exists
$\interpolatingDNN \in \ANNs$ 
such that
\begin{equation}
  \hiddenLength (\interpolatingDNN) 
\leq
	\c(\operatorname{log}_2(\varepsilon^{-1})+1)
	,
\qquad 
	\infnorm{ \MappingStructuralToVectorized(\interpolatingDNN) }
	\textstyle
	\leq \max\bcu{ 1, L, \abs{a}, \abs{b}, \allowbreak 2\bbr{ \sup_{ x \in [ a, b ]^d } \abs{ f( x ) } } } 
  ,
\end{equation}
\begin{equation}
  \functionANN\rect(\interpolatingDNN) \in C(\R^d, \R) 
  ,
\qquad 
  \sup\nolimits_{ x \in [ a, b ]^d }
    \abs{ (\functionANN\rect(\interpolatingDNN))( x ) - f( x ) }
\leq \varepsilon 
  ,
  \quad\text{and}\quad
  \paramANN(\interpolatingDNN) \leq \c \varepsilon^{-2d}
\end{equation}
\cfload. 
\end{aproof}
\endgroup

\cfclear
\begingroup
\begin{athm}{lemma}{lp_norm_estimate}[Explicit estimates for vector norms]
Let 
$ d \in \N $, 
$ p, q \in (0,\infty] $ satisfy 
$
  p \leq q 
$. 
Then it holds for all $ x \in \R^d $ that 
\begin{equation}
\llabel{claim}
  \pnorm{p}{x}
  \geq 
  \pnorm{q}{x}
\end{equation}
\cfout.
\end{athm}
\begin{aproof}
Throughout this proof, assume without loss of generality that $ q < \infty $, 
let $ e_1, e_2, \dots, e_d \in \R^d $ 
satisfy 
$
  e_1 = ( 1, 0, \dots, 0 )
$, 
$
  e_2 = ( 0, 1, 0, \dots, 0 )
$, 
$ \dots $, 
$ 
  e_d = ( 0, \dots, 0, 1 )
$,
let $ r \in \R $ satisfy 
\begin{equation}
  r 
  = p^{ - 1 } q ,
\end{equation}
and let $ x = ( x_1, \dots, x_d ) $, $ y = ( y_1, \dots, y_d ) \in \R^d $ 
satisfy for all $ i \in \{ 1, 2, \dots, d \} $ that 
\begin{equation}
\llabel{eq:definition_of_components_of_y}
  y_i = | x_i |^p
  .
\end{equation}
\Nobs that 
\lref{eq:definition_of_components_of_y}, 
the fact that 
\begin{equation}
  y = \sum_{ i = 1 }^d y_i e_i ,
\end{equation}
and the fact that for all 
$ v, w \in \R^d $ 
it holds that 
\begin{equation}
  \pnorm{r}{v + w}
  \leq 
  \pnorm{r}{v}
  +
  \pnorm{r}{w}
\end{equation}
\cfload\ 
ensures that
\begin{equation}
\begin{split}
  \pnorm{q}{ x }
& =
  \br*{
    \sum_{ i = 1 }^d | x_i |^q 
  }^{ \nicefrac{ 1 }{ q } }
=
  \br*{
    \sum_{ i = 1 }^d | x_i |^{ p r } 
  }^{ \nicefrac{ 1 }{ q } }
=
  \br*{
    \sum_{ i = 1 }^d | y_i |^r 
  }^{ \nicefrac{ 1 }{ q } }
=
  \br*{
    \sum_{ i = 1 }^d | y_i |^r 
  }^{ \nicefrac{ 1 }{ ( p r ) } }
=
  \pnorm{r}{ y }^{ \nicefrac{ 1 }{ p } }
\\ &
= 
  \bbbbpnorm{r}{ 
    \sum_{ i = 1 }^d y_i e_i
  }^{ \nicefrac{ 1 }{ p } }
\leq 
  \br*{
    \sum_{ i = 1 }^d 
    \pnorm{r}{ 
      y_i e_i
    }
  }^{ \nicefrac{ 1 }{ p } }
=
  \br*{
    \sum_{ i = 1 }^d 
    | y_i |
    \pnorm{r}{ 
      e_i
    }
  }^{ \nicefrac{ 1 }{ p } }
=
  \br*{
    \sum_{ i = 1 }^d 
    | y_i |
  }^{ \nicefrac{ 1 }{ p } }
\\ &
=
  \pnorm{1}{y}^{ \nicefrac{ 1 }{ p } }
=
  \pnorm{p}{x}
  .
\end{split}
\end{equation}
This establishes \lref{claim}. 
\end{aproof}
\endgroup

\cfclear
\begingroup
\renewcommand{\c}{\fC}
\begin{athm}{cor}{approx_error_rate}[Implicit multi-dimensional \ann\ approximations with 
prescribed error tolerances and asymptotic parameter bounds]
Let
$ d  \in \N $,
$ L, a \in \R $,
$ b \in [ a, \infty ) $
and let
$ f \colon [ a, b ]^d \to  \R $
satisfy for all
$ x, y \in [ a, b ]^d $
that
\begin{equation}
  \abs{ f( x ) - f( y ) } 
  \leq L \pnorm1{ x - y } 
\end{equation}
\cfload.
Then there exists
$ \c \in \R $ 
such that for all 
$
  \varepsilon \in (0,1] 
$ 
there exists 
$
  \interpolatingDNN \in \ANNs
$ 
such that 
\begin{equation}
\llabel{claim}
  \functionANN\rect(\interpolatingDNN) \in C(\R^d, \R)
,\quad
  \sup\nolimits_{ x \in [ a, b ]^d }
  \abs{ (\functionANN\rect(\interpolatingDNN))( x ) - f( x ) }
\leq \varepsilon
,
  \quad\text{and}\quad
  \paramANN(\interpolatingDNN) 
  \leq \c \varepsilon^{-2d}
\end{equation}
\cfout.
\end{athm}

\begin{aproof}
\Nobs that \cref{approx_error_rate2} establishes \lref{claim}.
\end{aproof}
\endgroup

\section{Refined ANN approximations results for multi-di\-men\-sio\-nal functions}
\label{sect:implicit_approx}

In \cref{sec:composed_error} below we 
establish estimates for the \emph{overall error} in the training of suitable rectified clipped \anns\ (see \cref{sec:rectified_clipped_anns} below) in the specific situation of \GD-type optimization methods with many independent random initializations.
Besides
\emph{optimization error} estimates from \cref{part:opt} and
\emph{generalization error} estimates from \cref{part:generalization},
for this overall error analysis we also employ suitable \emph{approximation error} estimates with a somewhat more refined control on the architecture of the approximating \anns\ than the approximation error estimates established in the previous sections of this chapter (cf., \eg, \cref{approx_error_rate2,approx_error_rate} above).
It is exactly the subject of this section to establish such refined approximation error estimates (see \cref{prop:approximation_error} below).

This section is specifically tailored to the requirements of the overall error analysis presented in \cref{sec:composed_error} and does not offer much more significant insights into the approximation error analyses of \anns\ than the content of the previous sections in this chapter. It can therefore be skipped at the first reading of this book and only needs to be considered when the reader is studying \cref{sec:composed_error} in detail.

\begingroup
\newcommand{\inputDim}{{d}}
\newcommand{\netDim}{{\mathfrak d}}

\subsection{Rectified clipped ANNs}
\label{sec:rectified_clipped_anns}

\cfclear
\begin{adef}{def:rectclippedFFANN}[Rectified clipped \anns]
  \cfconsiderloaded{def:rectclippedFFANN}
  Let 
    $L,\netDim\in\N$,
	$u\in[-\infty,\infty)$,
	$v\in(u,\infty]$,
    $\mathbf l=(l_0,l_1,\dots,l_L)\in\N^{L+1}$,
    $\theta\in\R^\netDim$
  satisfy
  \begin{equation}
    \netDim\geq \sum_{k=1}^Ll_k(l_{k-1}+1).
  \end{equation}
  Then we denote by 
    $\ClippedRealV{\theta}{\mathbf l} uv\colon \R^{l_0}\to\R^{l_L}$
  the function which satisfies for all 
    $x\in\R^{l_0}$
  that
  \begin{equation}
    \label{eq:rectclippedFFANN}
    \ClippedRealV{\theta}{\mathbf l}uv(x)
    =
    \begin{cases}
      \bigl( 
        \RealV{ \theta}{ 0}{ l_0 }{ \Clip uv{l_L} }
      \bigr)(x)
    &
      \colon 
      L = 1
    \\
      \bigl(\RealV{\theta}{0}{l_0}{\Rect_{l_1},\Rect_{l_2},\dots,\Rect_{l_{L-1}},\Clip uv{l_L}}\bigr)(x)
    &
      \colon 
      L > 1
    \end{cases}
  \end{equation}
  \cfload.
\end{adef}

\cfclear
\begin{athm}{lemma}{cor:structvsvect}
  Let $\Phi\in\ANNs
$ \cfload.
  Then it holds for all 
    $x\in\R^{\inDimANN(\Phi)}$ 
  that
  \begin{equation}
  \label{eq:structvsvect}
    \bigl(\UnclippedRealV{\MappingStructuralToVectorized(\Phi)}{\dims(\Phi)}\bigr)(x)
    =
    (\functionANN{\rect}(\Phi))(x)
  \end{equation}
  \cfout.
\end{athm}
\begin{aproof}
  \Nobs that
    \cref{lem:structvsvectgen},
    \cref{eq:rectclippedFFANN},
    \cref{eq:relu},
    and the fact that for all
      $d\in\N$
    it holds that 
      $\Clip{-\infty}\infty d=\id_{\R^d}$
  \prove
    \cref{eq:structvsvect}
  \cfload.
\end{aproof}

\endgroup

\subsection{Embedding ANNs in larger architectures}

\cfclear
\begin{athm}{lemma}{lem:embednet}
  Let 
    $a\in C(\R,\R)$,
    $L\in\N$, 
    $l_0,l_1,\dots,l_L,\mf l_0,\mf l_1,\dots,\mf l_L\in\N$
  satisfy for all 
    $k\in\{1,2,\dots,L\}$
  that
    $\mf l_0=l_0$,
    $\mf l_L=l_L$,
    and $\mf l_k\geq l_k$,
  for every $k\in\{1,2,\dots,L\}$ let
    $W_k=(W_{k,i,j})_{(i,j)\in\{1,2,\dots,l_k\}\times\{1,2,\dots,l_{k-1}\}}\in\R^{l_k\times l_{k-1}}$,
    $\mathscr{W}_k=(\mathscr{W}_{k,i,j})_{(i,j)\in\{1,2,\dots,\mf l_k\}\times\{1,2,\dots,\mf l_{k-1}\}}\in\R^{\mf l_k\times \mf l_{k-1}}$,
    $B_k=(B_{k,i})_{i\in\{1,2,\dots,l_k\}}\in\R^{l_k}$,
    $\mathscr{B}_k=(\mathscr{B}_{k,i})_{i\in\{1,2,\dots,\mf l_k\}}\in\R^{\mf l_k}$,
  assume for all
    $k\in\{1,2,\dots,L\}$,
    $i\in\{1,2,\dots,l_k\}$,
    $j\in\N\cap(0,l_{k-1}]$
  that 
  \begin{equation}
    \mathscr{W}_{k,i,j}=W_{k,i,j}
    \qandq
    \mathscr{B}_{k,i}=B_{k,i},
  \end{equation}
  and assume for all 
    $k\in\{1,2,\dots,L\}$,
    $i\in\{1,2,\dots,l_k\}$,
    $j\in\N\cap(l_{k-1},\mf l_{k-1}+1)$
  that
    $\mathscr{W}_{k,i,j}=0$.
  Then
  \begin{equation}
    \functionANN{a}\bigl(((W_1,B_1),(W_2,B_2),\dots,(W_L,B_L))\bigr)=\functionANN{a}\bigl(((\mathscr{W}_1,\mathscr{B}_1),(\mathscr{W}_2,\mathscr{B}_2),\dots,(\mathscr{W}_L,\mathscr{B}_L))\bigr)
  \end{equation}
  \cfout.
\end{athm}
\begin{aproof}
  Throughout this proof, let
    $\pi_k\colon\R^{\mf l_k}\to\R^{l_k}$, $k\in\{0,1,\dots,L\}$,
  satisfy for all
    $k\in\{0,1,\dots,L\}$,
    $x=(x_1,x_2,\dots,x_{\mf l_k})$
  that
  \begin{equation}
    \pi_k(x)
    =
    (x_1,x_2,\dots,x_{l_k})
    .
  \end{equation}
  \Nobs that
    the assumption that $\mf l_0=l_0$ and $\mf l_L=l_L$
  \proves that
  \begin{equation}
    \label{eq:embednet2}
    \functionANN{a}\bigl(((W_1,B_1),(W_2,B_2),\dots,(W_L,B_L))\bigr)\in C(\R^{\mf l_0},\R^{\mf l_L})
  \end{equation}
  \cfload.
  \Moreover
    the assumption that for all
      $k\in\{1,2,\dots,l\}$,
      $i\in\{1,2,\dots,l_k\}$,
      $j\in\N\cap(l_{k-1},\mf l_{k-1}+1)$
    it holds that
      $\mathscr{W}_{k,i,j}=0$
  \proves that for all 
    $k\in\{1,2,\dots,L\}$,
    $x=(x_1,\dots,x_{\mf l_{k-1}})\in\R^{\mf l_{k-1}}$
  it holds that
  \begin{equation}
  \begin{split}
    &\pi_k(\mathscr{W}_k x+\mathscr{B}_k)
    \\&=
    \pr*{
      \br*{\sum_{i=1}^{\mf l_{k-1}} \mathscr{W}_{k,1,i}x_{i}}+\mathscr{B}_{k,1},
      \br*{\sum_{i=1}^{\mf l_{k-1}} \mathscr{W}_{k,2,i}x_{i}}+\mathscr{B}_{k,2},
      \dots,
      \br*{\sum_{i=1}^{\mf l_{k-1}} \mathscr{W}_{k,l_k,i}x_{i}}+\mathscr{B}_{k,l_k}
    }
    \\&=
    \pr*{
      \br*{\sum_{i=1}^{l_{k-1}} \mathscr{W}_{k,1,i}x_{i}}+\mathscr{B}_{k,1},
      \br*{\sum_{i=1}^{l_{k-1}} \mathscr{W}_{k,2,i}x_{i}}+\mathscr{B}_{k,2},
      \dots,
      \br*{\sum_{i=1}^{l_{k-1}} \mathscr{W}_{k,l_k,i}x_{i}}+\mathscr{B}_{k,l_k}
    }
    .
  \end{split}
  \end{equation}
  Combining 
    this
  with 
    the assumption that for all
      $k\in\{1,2,\dots,L\}$,
      $i\in\{1,2,\dots,l_k\}$,
      $j\in\N\cap(0,l_{k-1}]$
    it holds that 
      $\mathscr{W}_{k,i,j}=W_{k,i,j}$ and $\mathscr{B}_{k,i}=B_{k,i}$
  \proves that for all
    $k\in\{1,2,\dots,L\}$,
    $x=(x_1,\dots,x_{\mf l_{k-1}})\in\R^{\mf l_{k-1}}$
  it holds that
  \begin{equation}
    \label{eq:embednet1}
  \begin{split}
    &\pi_k(\mathscr{W}_kx+\mathscr{B}_k)
    \\&=
    \Biggl(
      \br*{\sum_{i=1}^{l_{k-1}} W_{k,1,i}x_{i}}+B_{k,1},
      \br*{\sum_{i=1}^{l_{k-1}} W_{k,2,i}x_{i}}+B_{k,2},
      \dots,
      \br*{\sum_{i=1}^{l_{k-1}} W_{k,l_k,i}x_{i}}+B_{k,l_k}
    \Biggr)
    \\&=
    W_k \pi_{k-1}(x)+B_k
    .
  \end{split}
  \end{equation}
  \Hence that for all
    $x_0%
    \in \R^{\mf l_0}
    ,\, 
    x_1%
    \in \R^{\mf l_1}
    , 
    \ldots,\, 
    x_{L-1}%
    \in \R^{\mf l_{L-1}}
    $,
    $k\in\N\cap(0,L)$
    with $\forall \, m \in \N \cap (0,L) \colon x_m =\activationDim{\mf l_m}(\mathscr{W}_m x_{m-1} + \mathscr{B}_m)$ 
  it holds that
  \begin{equation}
    \pi_k(x_k)
    =
    \activationDim{l_k}(\pi_k(\mathscr{W}_kx_{k-1}+\mathscr{B}_k))
    =
    \activationDim{l_k}(W_k\pi_{k-1}(x_{k-1})+B_k)
  \end{equation}
  \cfload.
    Induction,
    the assumption that $l_0=\mf l_0$ and $l_L=\mf l_L$,
    and \cref{eq:embednet1}
    \hence
  \prove that for all
    $x_0\in\R^{\mf l_0},\, 
    x_1\in\R^{\mf l_1}, 
    \ldots,\, 
    x_{L-1}\in \R^{\mf l_{L-1}}$
    with $\forall \, k \in \N \cap (0,L) \colon x_k =\activationDim{\mf l_k}(\mathscr{W}_k x_{k-1} + \mathscr{B}_k)$ 
  it holds that
  \begin{equation}
  \begin{split}
    &
    \bigl(\functionANN{a}\bigl(((W_1,B_1),(W_2,B_2),\dots,(W_L,B_L))\bigr)\bigr)(x_0)
    \\&=
    \bigl(\functionANN{a}\bigl(((W_1,B_1),(W_2,B_2),\dots,(W_L,B_L))\bigr)\bigr)(\pi_0(x_0))
    \\&=
    W_L\pi_{L-1}(x_{L-1})+B_L
    \\&=
    \pi_L(\mathscr{W}_Lx_{L-1}+\mathscr{B}_L)
    =
    \mathscr{W}_Lx_{L-1}+\mathscr{B}_L
    \\&=
    \bigl(\functionANN{a}\bigl(((\mathscr{W}_1,\mathscr{B}_1),(\mathscr{W}_2,\mathscr{B}_2),\dots,(\mathscr{W}_L,\mathscr{B}_L))\bigr)\bigr)(x_0)
    .
  \end{split}
  \end{equation}
\end{aproof}

\cfclear
\begin{athm}{lemma}{lem:embednet_exist}
  Let $a\in C(\R,\R)$,
  $L\in\N$, 
  $l_0,l_1,\dots,l_L,\mf l_0,\mf l_1,\dots,\mf l_L\in\N$
  satisfy for all 
    $k\in\{1,2,\dots,L\}$
  that
  \begin{equation}
    \mf l_0=l_0,
    \qquad
    \mf l_L=l_L,
    \qandq 
    \mf l_k\geq l_k
  \end{equation}
  and let $\Phi\in\ANNs$ satisfy
    $\dims(\Phi)=(l_0,l_1,\dots,l_L)$
  \cfload.
  Then there exists $\Psi\in\ANNs$ such that
  \begin{equation}
    \dims(\Psi)=(\mf l_0,\mf l_1,\dots,\mf l_L),
    \qquad
    \infnorm{\MappingStructuralToVectorized(\Psi)}
    =
    \infnorm{\MappingStructuralToVectorized(\Phi)},
    \qandq
    \functionANN{a}(\Psi)
    =
    \functionANN{a}(\Phi)
  \end{equation}
  \cfout.
\end{athm}
\begin{aproof}
  Throughout this proof, let
    $B_k=(B_{k,i})_{i\in\{1,2,\dots,l_k\}}\in\R^{l_k}$, $k\in\{1,2,\allowbreak\dots,\allowbreak L\}$,
    and $W_k=(W_{k,i,j})_{(i,j)\in\{1,2,\dots,l_k\}\times\{1,2,\dots,l_{k-1}\}}\in\R^{l_k\times l_{k-1}}$, $k\in\{1,2,\allowbreak\dots,L\}$,
  satisfy
  \begin{equation}
    \Phi=((W_1,B_1),(W_2,B_2),\dots,(W_L,\allowbreak B_L))
  \end{equation}
  and let
    $\mf W_k=(\mf W_{k,i,j})_{(i,j)\in\{1,2,\dots,\mf l_k\}\times\{1,2,\dots,\mf l_{k-1}\}}\in\R^{\mf l_k\times \mf l_{k-1}}$, $k\in\{1,2,\dots,L\}$,
    and $\mf B_k=(\mf B_{k,i})_{i\in\{1,2,\dots,\mf l_k\}}\in\R^{\mf l_k}$, $k\in\{1,2,\dots,L\}$,
  satisfy for all
    $k\in\{1,2,\dots,L\}$,
    $i\in\{1,2,\dots,\mf l_k\}$,
    $j\in\{1,2,\dots,\mf l_{k-1}\}$
  that
  \begin{equation}
  \label{eq:embednet2.WB}
    \mf W_{k,i,j}
    =
    \begin{cases}
      W_{k,i,j}&\colon (i\leq l_k)\land(j\leq l_{k-1})\\
      0 & \colon (i>l_k)\lor(j>l_{k-1})
    \end{cases}
    \qquad\text{and}\qquad
    \mf B_{k,i}
    =
    \begin{cases}
      B_{k,i} &\colon i\leq l_k\\
      0 & \colon i>l_k.
    \end{cases}
  \end{equation}
  \Nobs that
    \cref{eq:defANN}
  \proves that
  $
    ((\mf W_1,\mf B_1),(\mf W_2,\mf B_2),\dots,(\mf W_L,\mf B_L))
    \in
    \bigl(\textstyle\bigtimes_{i=1}^{L}(\R^{\mf l_i\times \mf l_{i-1}}\times\R^{\mf l_i})\bigr)
    \subseteq
    \ANNs
  $ and
  \begin{equation}
    \dims\bigl(((\mf W_1,\mf B_1),(\mf W_2,\mf B_2),\dots,(\mf W_L,\mf B_L))\bigr)
    =
    (\mf l_0,\mf l_1,\dots,\mf l_{L}).
  \end{equation}
  \Moreover
    \cref{lem:structtovect}
    and \cref{eq:embednet2.WB}
  \prove that
  \begin{equation}
    \infnorm{\MappingStructuralToVectorized\bigl(((\mf W_1,\mf B_1),(\mf W_2,\mf B_2),\dots,(\mf W_L,\mf B_L))\bigr)}
    =
    \infnorm{\MappingStructuralToVectorized(\Phi)}
  \end{equation}
  \cfload.
  \Moreover
    \cref{lem:embednet}
  \proves that
  \begin{eqsplit}
    \functionANN{a}(\Phi)
    &=
    \functionANN{a}\bigl(((W_1,B_1),(W_2,B_2),\dots,(W_L,\allowbreak B_L))\bigr)
    \\&=
    \functionANN{a}\bigl(((\mf W_1,\mf B_1),(\mf W_2,\mf B_2),\dots,(\mf W_L,\mf B_L))\bigr)
  \end{eqsplit}
  \cfload.
\end{aproof}

\cfclear
\begin{athm}{lemma}{lem:composition_infnorm}
  Let
  $ L,\mathfrak{L}\in\N$, $l_0,l_1,\ldots, l_L, \mathfrak{l}_0,\mathfrak{l}_1,\ldots, \mathfrak{l}_\mathfrak{L} \in \N$, 
	$
	\Phi_1
	=
	((W_1, B_1),(W_2, B_2),\allowbreak \ldots, (W_L,\allowbreak B_L))
	\in  \allowbreak
	\bigl( \bigtimes_{k = 1}^L\allowbreak(\R^{l_k \times l_{k-1}} \times \R^{l_k})\bigr)
	$,
	$
	\Phi_2
	=
	((\mathfrak{W}_1, \mathfrak{B}_1),\allowbreak(\mathfrak{W}_2, \mathfrak{B}_2),\allowbreak \ldots, (\mathfrak{W}_\mathfrak{L},\allowbreak \mathfrak{B}_\mathfrak{L}))
	\in  \allowbreak
	\bigl( \bigtimes_{k = 1}^\mathfrak{L}\allowbreak(\R^{\mathfrak{l}_k \times \mathfrak{l}_{k-1}} \times \R^{\mathfrak{l}_k})\bigr)
	$.
  Then
  \begin{equation}
    \label{eq:composition_infnorm}
    \infnorm{\MappingStructuralToVectorized(\compANN{\Phi_1}{\Phi_2})}
    \leq
    \max\bigl\{
      \infnorm{\MappingStructuralToVectorized(\Phi_1)},
      \infnorm{\MappingStructuralToVectorized(\Phi_2)},
      \asinfnorm{\MappingStructuralToVectorized\bigl(((W_1\mf W_{\mf L},W_1\mf B_{\mf L}+B_1))\bigr)}
    \bigr\}
  \end{equation}
  \cfout.
\end{athm}
\begin{aproof}
  \Nobs that
    \cref{eq:defCompANN}
    and \cref{lem:structtovect}
  \prove[ep] \cref{eq:composition_infnorm}.
\end{aproof}

\cfclear
\begin{athm}{lemma}{lem:extension}
  Let $d,L\in\N$,
  $\Phi\in\ANNs$
  satisfy
    $L\geq\lengthANN(\Phi)$
    and $d=\outDimANN(\Phi)$
  \cfload.
  Then
  \begin{equation}
    \infnorm{\MappingStructuralToVectorized(\longerANN{L,\idRelu_d}(\Phi))}
    \leq
    \max\{1,\infnorm{\MappingStructuralToVectorized(\Phi)}\}
  \end{equation}
  \cfout.
\end{athm}
\begin{aproof}
  Throughout this proof, 
  assume without loss of generality that $L>\lengthANN(\Phi)$
  and let $l_0,l_1,\dots,l_{L-\lengthANN(\Phi)+1}\in\N$
  satisfy 
  \begin{equation}
    (l_0,l_1,\dots,l_{L-\lengthANN(\Phi)+1})=(d,2d,2d,\dots,2d,d)
    .
  \end{equation}
  \Nobs that
    \cref{lem:Relu:identity}
  \proves that $\dims(\idRelu_d)=(d,2d,d)\in\N^3$
  \cfload.
    \Cref{PropertiesOfANNenlargementGeometry:BulletPower} in \cref{Lemma:PropertiesOfANNenlargementGeometry}
  \hence \proves that
  \begin{eqsplit}
    \lengthANN((\idRelu_d)^{\compANNbullet(L-\lengthANN(\Phi))})
    &=
    L-\lengthANN(\Phi)+1\\
    \text{and}\qquad
    \dims((\idRelu_d)^{\compANNbullet(L-\lengthANN(\Phi))})
    &=
    (l_0,l_1,\dots,l_{L-\lengthANN(\Phi)+1})\in\N^{L-\lengthANN(\Phi)+2}
  \end{eqsplit}
  \cfload.
    This
  \proves that there exist
    $W_k\in\R^{l_k\times l_{k-1}}$, $k\in\{1,2,\dots,L-\lengthANN(\Phi)+1\}$,
    and $B_k\in\R^{l_k}$, $k\in\{1,2,\dots,L-\lengthANN(\Phi)+1\}$,
    which satisfy
  \begin{equation}
    (\idRelu_d)^{\compANNbullet ( L-\lengthANN(\Phi))}=((W_1,B_1),(W_2,B_2),\dots,(W_{L-\lengthANN(\Phi)+1},B_{L-\lengthANN(\Phi)+1})).
  \end{equation}
  \Moreover
    \cref{parallelisationSameLengthDef},
    \cref{eq:def:id:1},
    \cref{eq:def:id:2},
    \cref{eq:defCompANN},
    and \cref{iteratedANNcomposition:equation}
  \prove that
  \begin{equation}
    \label{eq:extension1}
    \begin{split}
    W_1
    &=
    \pmat{
      1&0&\cdots&0\\
      -1&0&\cdots&0\\
      0&1&\cdots&0\\
      0&-1&\cdots&0\\
      \vdots&\vdots&\ddots&\vdots\\
      0&0&\cdots&1\\
      0&0&\cdots&-1
    }\in\R^{(2d)\times d}
    \\\text{and}\qquad
    W_{L-\lengthANN(\Phi)+1}
    &=
    \pmat{
      1&-1&0&0&\cdots&0&0\\
      0&0&1&-1&\cdots&0&0\\
      \vdots&\vdots&\vdots&\vdots&\ddots&\vdots&\vdots\\
      0&0&0&0&\cdots&1&-1
    }\in\R^{d\times (2d)}
    .
    \end{split}
  \end{equation}
  \Moreover
    \cref{parallelisationSameLengthDef},
    \cref{eq:def:id:1},
    \cref{eq:def:id:2},
    \cref{eq:defCompANN},
    and \cref{iteratedANNcomposition:equation}
  \prove that for all
    $k\in\N\cap(1,L-\lengthANN(\Phi)+1)$
  it holds that
  \begin{equation}
    \label{eq:extension4}
    \begin{split}
    W_k
    &=
    \underbrace{
    \pmat{
      1&0&\cdots&0\\
      -1&0&\cdots&0\\
      0&1&\cdots&0\\
      0&-1&\cdots&0\\
      \vdots&\vdots&\ddots&\vdots\\
      0&0&\cdots&1\\
      0&0&\cdots&-1
    }
    }_{\in\R^{(2d)\times d}}
    \underbrace{
    \pmat{
      1&-1&0&0&\cdots&0&0\\
      0&0&1&-1&\cdots&0&0\\
      \vdots&\vdots&\vdots&\vdots&\ddots&\vdots&\vdots\\
      0&0&0&0&\cdots&1&-1
    }
    }_{\in\R^{d\times(2d)}}
    \\&=
    \pmat{
      1&-1&0&0&\cdots&0&0\\
      -1&1&0&0&\cdots&0&0\\
      0&0&1&-1&\cdots&0&0\\
      0&0&-1&1&\cdots&0&0\\
      \vdots&\vdots&\vdots&\vdots&\ddots&\vdots&\vdots\\
      0&0&0&0&\cdots&1&-1\\
      0&0&0&0&\cdots&-1&1
    }
    \in\R^{(2d)\times (2d)}
    .
    \end{split}
  \end{equation}
  \Moreover
    \cref{eq:def:id:1},
    \cref{eq:def:id:2},
    \cref{parallelisationSameLengthDef},
    \cref{iteratedANNcomposition:equation},
    and \cref{eq:defCompANN}
  \prove that for all
    $k\in\N\cap[1,L-\lengthANN(\Phi)]$
  it holds that
  \begin{equation}
    \label{eq:extension5}
    B_k=0\in\R^{2d}
    \qandq
    B_{L-\lengthANN(\Phi)+1}=0\in\R^d
    .
  \end{equation}
  Combining
    this,
    \cref{eq:extension1},
    and \cref{eq:extension4}
  \proves that
  \begin{equation}
    \label{eq:extension3}
    \asinfnorm{\MappingStructuralToVectorized\bigl((\idRelu_d)^{\compANNbullet ( L-\lengthANN(\Phi))}\bigr)}
    =
    1
  \end{equation}
  \cfload.
  \Moreover
    \cref{eq:extension1}
  \proves that for all
    $k\in\N$,
    $\mf W=(w_{i,j})_{(i,j)\in\{1,2,\dots,d\}\times\{1,2,\dots,k\}}\in\R^{d\times k}$
  it holds that
  \begin{equation}
    \label{eq:extension2}
    W_1\mf W
    =
    \pmat{
      w_{1,1} & w_{1,2} & \cdots & w_{1,k}\\
      -w_{1,1} & -w_{1,2} & \cdots & -w_{1,k}\\
      w_{2,1} & w_{2,2} & \cdots & w_{2,k}\\
      -w_{2,1} & -w_{2,2} & \cdots & -w_{2,k}\\
      \vdots & \vdots & \ddots &\vdots\\
      w_{d,1} & w_{d,2} & \cdots & w_{d,k}\\
      -w_{d,1} & -w_{d,2} & \cdots & -w_{d,k}
    }\in\R^{(2d)\times k}
    .
  \end{equation}
  \Moreover
    \cref{eq:extension1}
    and \cref{eq:extension5}
  \prove that for all
    $\mf B=(b_1,b_2,\dots,b_d)\in\R^{d}$
  it holds that
  \begin{equation}
    W_1 \mf B+B_1
    =
    \pmat{
      1&0&\cdots&0\\
      -1&0&\cdots&0\\
      0&1&\cdots&0\\
      0&-1&\cdots&0\\
      \vdots&\vdots&\ddots&\vdots\\
      0&0&\cdots&1\\
      0&0&\cdots&-1
    }\pmat{b_1\\b_2\\\vdots\\b_d}
    =
    \pmat{b_1\\-b_1\\b_2\\-b_2\\\vdots\\b_d\\-b_d}
    \in\R^{2d}
    .
  \end{equation}
  Combining
    this
  with
    \cref{eq:extension2}
  \proves that for all
    $k\in\N$,
    $\mf W\in\R^{d\times k}$,
    $\mf B\in\R^{d}$
  it holds that
  \begin{equation}
    \asinfnorm{\MappingStructuralToVectorized\bigl(((W_1\mf W,W_1\mf B+B_1))\bigr)}
    =
    \asinfnorm{\MappingStructuralToVectorized\bigl(((\mf W,\mf B))\bigr)}.
  \end{equation}
    This,
    \cref{lem:composition_infnorm},
    and \cref{eq:extension3}
  \prove that
  \begin{equation}
    \begin{split}
    &\infnorm{\MappingStructuralToVectorized(\longerANN{L,\idRelu_d}(\Phi))}
    =
    \asinfnorm{\MappingStructuralToVectorized\bigl(\compANN{((\idRelu_d)^{\compANNbullet(L-\lengthANN(\Phi))})}{\Phi}\bigr)}
    \\&\leq
    \max\bigl\{
      \asinfnorm{\MappingStructuralToVectorized\bigl((\idRelu_d)^{\compANNbullet(L-\lengthANN(\Phi))}\bigr)},
      \asinfnorm{\MappingStructuralToVectorized(\Phi)}
    \bigr\}
    =
    \max\{
      1,
      \infnorm{\MappingStructuralToVectorized(\Phi)}
    \}
    \end{split}
  \end{equation}
  \cfload.
\end{aproof}

\cfclear
\begin{athm}{lemma}{lem:embednet3}
  Let $L,\mf L\in\N$, 
  $l_0,l_1,\dots,l_L,\mf l_0,\mf l_1,\dots,\mf l_{\mf L}\in\N$
  satisfy
  \begin{equation}
    \mf L\geq L,
    \qquad
    \mf l_0=l_0,
    \qquad\text{and}\qquad
    \mf l_{\mf L}=l_L,
  \end{equation}
  assume for all
    $i\in\N\cap[0,L)$
  that
    $\mf l_i\geq l_i$,
  assume for all 
    $i\in \N\cap(L-1,\mf L)$
  that
    $\mf l_i\geq 2l_L$,
  and let $\Phi\in\ANNs$
  satisfy $\dims(\Phi)=(l_0,l_1,\dots,l_L)$
  \cfload.
  Then there exists
    $\Psi\in\ANNs$
  such that
  \begin{equation}
    \label{eq:embednet3.claim}
    \dims(\Psi)=(\mf l_0,\mf l_1,\dots,\mf l_{\mf L}),
    \;\;
    \infnorm{\MappingStructuralToVectorized(\Psi)}
    \leq
    \max\{
      1,\infnorm{\MappingStructuralToVectorized(\Phi)}
    \},
    \;\;\text{and}\;\;
    \functionANN{\rect}(\Psi)=\functionANN{\rect}(\Phi)
  \end{equation}
  \cfout.
\end{athm}  
\begin{aproof}
  Throughout this proof,
  let $\Xi\in\ANNs$ satisfy
    $\Xi=\longerANN{\mf L,\idRelu_{l_L}}(\Phi)$
  \cfload.
  \Nobs that
    \cref{item:lem:Relu:dims} in \cref{lem:Relu:identity}
  \proves that
  $\dims(\idRelu_{l_L})=(l_L,2l_L,l_L)\in\N^3$.
  Combining
    this
  with
    \cref{lem:extension_dims}
  \proves that
    $\dims(\Xi)\in\N^{\mf L+1}$
    and
    \begin{equation}
      \label{eq:embednet3.dimsXi}
      \dims(\Xi)
      =
      \begin{cases}
        (l_0,l_1,\dots,l_L) & \colon \mf L=L \\
        (l_0,l_1,\dots,l_{L-1},2l_L,2l_L,\dots,2l_L,l_L) & \colon \mf L>L.
      \end{cases}
    \end{equation}
  \Moreover
    \cref{lem:extension}
    (applied with
      $d\is l_L$,
      $L\is\mf L$,
      $\Phi\is\Phi$
    in the notation of \cref{lem:extension})
  \proves that
  \begin{equation}
    \label{eq:embednet3.norm}
    \infnorm{\MappingStructuralToVectorized(\Xi)}
    \leq
    \max\{
      1,
      \infnorm{\MappingStructuralToVectorized(\Phi)}
    \}
  \end{equation}
  \cfload.
  \Moreover
    \cref{item:lem:Relu:real} in \cref{lem:Relu:identity}
  \proves that for all
    $x\in\R^{l_L}$
  it holds that
  \begin{equation}
    (\functionANN{\rect}(\idRelu_{l_L}))(x)=x
  \end{equation}
  \cfload.
    This
    and \cref{PropertiesOfANNenlargementRealization:ItemIdentityLonger} in \cref{Lemma:PropertiesOfANNenlargementRealization}
  prove that
  \begin{equation}
    \label{eq:embednet3.real}
    \functionANN{\rect}(\Xi)
    =
    \functionANN{\rect}(\Phi).
  \end{equation}
  \Moreover
    \cref{eq:embednet3.dimsXi},
    the assumption that for all 
      $i\in[0,L)$
    it holds that 
      $\mf l_0=l_0$,
      $\mf l_{\mf L}=l_L$,
      and $\mf l_i\leq l_i$,
    the assumption that for all 
      $i\in\N\cap(L-1,\mf L)$
    it holds that
      $\mf l_i\geq 2l_L$,
    and \cref{lem:embednet_exist}
    (applied with
      $a\is\rect$,
      $L\is\mf L$,
      $(l_0,l_1,\dots,l_L)\is\dims(\Xi)$,
      $(\mf l_0,\mf l_1,\dots,\mf l_{\mf L})\is(\mf l_0,\mf l_1,\dots,\mf l_{\mf L})$,
      $\Phi\is\Xi$
    in the notation of \cref{lem:embednet_exist})
  \prove that there exists
    $\Psi\in\ANNs$
  such that
  \begin{equation}
    \dims(\Psi)
    =
    (\mf l_0,\mf l_1,\dots,\mf l_{\mf L}),
    \qquad
    \infnorm{\MappingStructuralToVectorized(\Psi)}
    =
    \infnorm{\MappingStructuralToVectorized(\Xi)},
    \qandq
    \functionANN{\rect}(\Psi)
    =
    \functionANN{\rect}(\Xi)
    .
  \end{equation}
  Combining
    this
  with 
    \cref{eq:embednet3.norm}
    and \cref{eq:embednet3.real}
  \proves[ep] \cref{eq:embednet3.claim}.
\end{aproof}

\cfclear
\begin{athm}{lemma}{lem:embednet_vectorized}
  Let 
  $u\in[-\infty,\infty)$,
  $v\in(u,\infty]$,
  $L,\mf L,d,\mf d\in\N$, 
  $\theta\in\R^d$,
  $l_0,l_1,\dots,l_L,\mf l_0,\mf l_1,\allowbreak \dots,\allowbreak \mf l_{\mf L}\in\N$
  satisfy that
  \begin{equation}
    \textstyle
    d\geq\sum_{i=1}^L l_i(l_{i-1}+1),
    \quad
    \mf d\geq\sum_{i=1}^{\mf L} \mf l_i(\mf l_{i-1}+1),
    \quad
    \mf L\geq L,
    \quad
    \mf l_0=l_0,
    \quad\text{and}\quad
    \mf l_{\mf L}=l_L,
  \end{equation}
  assume for all
  $i\in\N\cap[0,L)$
  that
  $\mf l_i\geq l_i$,
  and assume for all 
    $i\in \N\cap(L-1,\mf L)$
  that
    $\mf l_i\geq 2l_L$.
  Then there exists $\vartheta\in\R^{\mf d}$ such that
  \begin{equation}
    \label{eq:embednet_vectorized:claim}
    \infnorm{\vartheta}\leq\max\{1,\infnorm{\theta}\}
    \qandq
    \ClippedRealV{\vartheta}{(\mf l_0,\mf l_1,\dots,\mf l_{\mf L})}uv
    =
    \ClippedRealV{\theta}{(l_0,l_1,\dots,l_L)}uv
  \end{equation}
  \cfout.
\end{athm}
\begin{aproof}
  Throughout this proof, let
    $\eta_1,\eta_2,\dots,\eta_d\in\R$
  satisfy
  \begin{equation}
    \label{eq:embednet_vectorized.4}
    \theta=(\eta_1,\eta_2,\dots,\eta_d)
  \end{equation}
  and let
    $\Phi\in\bigl(\bigtimes_{i=1}^L\R^{l_i\times l_{i-1}}\times\R^{l_i}\bigr)$
  satisfy
  \begin{equation}
    \label{eq:embednet_vectorized.3}
    \MappingStructuralToVectorized(\Phi)=(\eta_1,\eta_2,\dots,\eta_{\paramANN(\Phi)})
  \end{equation}
  \cfload.
  \Nobs that
    \cref{lem:embednet3}
  \proves that there exists $\Psi\in\ANNs$ which satisfies
  \begin{equation}
    \label{eq:embednet_vectorized.2}
    \dims(\Psi)=(\mf l_0,\mf l_1,\dots,\mf l_{\mf L}),
    \;\;
    \infnorm{\MappingStructuralToVectorized(\Psi)}
    \leq
    \max\{
      1,\infnorm{\MappingStructuralToVectorized(\Phi)}
    \},
    \;\;\text{and}\;\;
    \functionANN{\rect}(\Psi)=\functionANN{\rect}(\Phi)
  \end{equation}
  \cfload.
  Next let
    $\vartheta=(\vartheta_1,\dots,\vartheta_{\mf d})\in\R^{\mf d}$
  satisfy 
  \begin{equation}
    \label{eq:embednet_vectorized.1}
    (\vartheta_1,\vartheta_2,\dots,\vartheta_{\paramANN(\Psi)})
    =
    \MappingStructuralToVectorized(\Psi)
    \qandq
    \Forall i\in\N\cap(\paramANN(\Psi),\mf d+1)\colon \vartheta_i=0
    .
  \end{equation}
  \Nobs that
    \cref{eq:embednet_vectorized.4},
    \cref{eq:embednet_vectorized.3},
    \cref{eq:embednet_vectorized.2},
    and \cref{eq:embednet_vectorized.1}
  \prove that
  \begin{equation}
    \label{eq:embednet_vectorized:normclaim}
    \infnorm{\vartheta}
    =
    \infnorm{\MappingStructuralToVectorized(\Psi)}
    \leq
    \max\{1,\infnorm{\MappingStructuralToVectorized(\Phi)}\}
    \leq
    \max\{1,\infnorm\theta\}
    .
  \end{equation}
  \Moreover
    \cref{cor:structvsvect}
    and \cref{eq:embednet_vectorized.3}
  \prove that for all
    $x\in\R^{l_0}$
  it holds that
  \begin{equation}
    \label{eq:embednet_vectorized.realPhi}
    \bigl(\UnclippedRealV{\theta}{(l_0,l_1,\dots,l_L)}\bigr)(x)
    =
    \bigl(\UnclippedRealV{\MappingStructuralToVectorized(\Phi)}{\dims(\Phi)}\bigr)(x)
    =
    (\functionANN{\rect}(\Phi))(x)
  \end{equation}
  \cfload.
  \Moreover
    \cref{cor:structvsvect},
    \cref{eq:embednet_vectorized.2},
    and \cref{eq:embednet_vectorized.1}
  \prove that for all
    $x\in\R^{\mf l_0}$
  it holds that
  \begin{equation}
    \bigl(\UnclippedRealV{\vartheta}{(\mf l_0,\mf l_1,\dots,\mf l_{\mf L})}\bigr)(x)
    =
    \bigl(\UnclippedRealV{\MappingStructuralToVectorized(\Psi)}{\dims(\Psi)}\bigr)(x)
    =
    (\functionANN{\rect}(\Psi))(x)
    .
  \end{equation}
  Combining
    this
    and \cref{eq:embednet_vectorized.realPhi}
  with
    \cref{eq:embednet_vectorized.2}
    and the assumption that $\mf l_0=l_0$ and $\mf l_{\mf L}=l_L$
  \proves that
  \begin{equation}
    \UnclippedRealV{\theta}{(l_0,l_1,\dots,l_L)}
    =
    \UnclippedRealV{\vartheta}{(\mf l_0,\mf l_1,\dots,\mf l_{\mf L})}
    .
  \end{equation}
  \Hence that
  \begin{equation}
    \ClippedRealV{\theta}{(l_0,l_1,\dots,l_L)}uv
    =
    \Clip uv{l_L}\circ\UnclippedRealV{\theta}{(l_0,l_1,\dots,l_L)}
    =
    \Clip uv{\mf l_{\mf L}}\circ\UnclippedRealV{\vartheta}{(\mf l_0,\mf l_1,\dots,\mf l_{\mf L})}
    =
    \ClippedRealV{\vartheta}{(\mf l_0,\mf l_1,\dots,\mf l_{\mf L})}uv
  \end{equation}
  \cfload.
    This
    and \cref{eq:embednet_vectorized:normclaim}
  \prove[ep] \cref{eq:embednet_vectorized:claim}.
\end{aproof}

\subsection{Approximation through ANNs with variable architectures}

\cfclear
\begin{athm}{cor}{approx:lip} 
Let 
  $d, K, \bfd, \bfL \in \N$, 
  $\bfl=(\bfl_0,\bfl_1, \ldots, \bfl_\bfL) \in \N^{\bfL+1}$, 
  $L \in [0, \infty)$ 
satisfy that 
\begin{equation}
  \bfL \geq \ceil{ \logg_2 (K)} +2, 
  \quad
  \bfl_0=d, 
  \quad
  \bfl_\bfL=1, 
  \quad
  \bfl_1 \geq 2d K, 
  \quad\text{and}\quad 
  \textstyle
  \bfd \geq \sum_{i=1}^\bfL \bfl_i(\bfl_{i-1}+1), 
\end{equation}
assume for all 
  $i \in \N\cap(1,\bfL)$ 
that
$\bfl_i \geq 3 \ceil[\big]{ \tfrac{ K}{2^{i-1}} }$, 
let $E \subseteq \R^d$ be a set, 
let $\fx_1, \fx_2, \ldots, \fx_K \in E$, 
and let $f \colon E \to \R$ 
satisfy for all 
  $x, y \in E$ 
that
  $\abs{f(x)-f(y)} \leq L \pnorm1{x-y}$ \cfload. 
Then there exists $\theta \in \R^{\bfd}$ such that 
\begin{equation}
    \pnorm\infty{ \theta } \leq \max \{ 1,L, \max\nolimits_{k \in \{1, 2, \ldots, K\}} \pnorm\infty{ \fx_k }, 2 \max\nolimits_{k \in \{1, 2, \ldots, K\}} \abs{f(\fx_k)}\}
\end{equation}
and 
\begin{equation}
    \sup\nolimits_{x \in E}  \babs{ f(x)- \ClippedRealV{\theta}{\bfl}{-\infty}{ \infty}(x) } \leq 2L \br*{\sup\nolimits_{x \in E} \pr*{ \inf\nolimits_{k  \in \{1,2, \ldots, K \}} \pnorm1{x- \fx_k}}}
\end{equation}
\cfout.
\end{athm}
\begin{aproof}
Throughout this proof, let $\fy \in \R^{K}$, $\Phi \in \ANNs$ satisfy 
$\fy = (f(\fx_1),\allowbreak f(\fx_2),\allowbreak \ldots, \allowbreak f(\fx_K))$
and
\begin{equation}
    \Phi = \maxANN_{K} \compANNbullet \AffineANN_{-L \idMatrix_{K}, \fy} \compANNbullet \parallelizationSpecial_{K}  \bigl( \oneNormANN_d \compANNbullet \AffineANN_{\idMatrix_d, -\fx_1}, \oneNormANN_d \compANNbullet \AffineANN_{\idMatrix_d, -\fx_2}, \ldots, \oneNormANN_d \compANNbullet \AffineANN_{\idMatrix_d, -\fx_K} \bigr) \compANNbullet \extensionANN_{d,K}
\end{equation}
\cfload.
\Nobs that \cref{dnn:intl1} and \cref{dnn:intp1} \prove that 
\begin{enumerate}[label=(\Roman{*})]
    \item it holds that $\lengthANN (\Phi) = \ceil{ \logg_2 (K)} + 2$,
    \item it holds that $\inDimANN(\Phi)=d$,
 	\item it holds that $\outDimANN(\Phi)=1$,    
 	\item it holds that $\dimANNlevel_1(\Phi) =2d K$,
    \item it holds for all $i \in \{2,3, \ldots, \lengthANN(\Phi)-1\}$ that $\dimANNlevel_i(\Phi) \leq 3 \ceil{ \frac{K}{2^{i-1}} } $,
    \item \label{approx:lip:proofitem1} it holds that $\pnorm\infty{ \MappingStructuralToVectorized(\Phi)} \leq \max \{ 1,L, \max_{k \in \{1, 2, \ldots, K\}} \infnorm{ \fx_k }, 2 \max_{k \in \{1, 2, \ldots, K\}} \abs{f(\fx_k)}\}$, and
    \item \label{approx:lip:proofitem2} it holds that
$\sup\nolimits_{x \in E} \abs{f(x)- (\functionANN\rect(\Phi))(x)} \leq 2L \br*{\sup\nolimits_{x \in E} \pr*{ \inf\nolimits_{k \in \{1,2, \ldots, K\}} \pnorm1{x - \fx_k} }}$
\end{enumerate}
\cfload. 
\Moreover
\enum{
	the fact that $\bfL \geq \ceil{ \logg_2(K)} +2 = \lengthANN(\Phi)$;
	the fact that $\bfl_0=d = \dimANNlevel_0(\Phi)$;
	the fact that $\bfl_1 \geq 2dK = \dimANNlevel_1(\Phi)$;
	the fact that for all $i \in \{1,2, \ldots, \lengthANN(\Phi)-1\}\backslash\{1\}$ 
		it holds that $\bfl_i \geq 3 \ceil{ \tfrac{ K }{2^{i-1}} } \geq  \dimANNlevel_i(\Phi)$;
	the fact that for all $i \in \N\cap(\lengthANN(\Phi)-1, \bfL )$ 
			it holds that $\bfl_i \geq 3 \ceil{ \tfrac{ K }{2^{i-1}} } \geq  2 = 2\dimANNlevel_{\lengthANN(\Phi)}(\Phi)$;
	the fact that $\bfl_\bfL = 1 = \dimANNlevel_{\lengthANN(\Phi)}(\Phi)$;
  \cref{lem:embednet_vectorized}
}
\prove that there exists $\theta \in \R^{\bfd}$ which satisfies that
\begin{equation}
\label{approx:lip:eq2}
\begin{split} 
	\infnorm{\theta} 
\leq
	\max\{1,\infnorm{\MappingStructuralToVectorized(\Phi)}\}
 \qandq
	 \ClippedRealV{\theta}{(\bf l_0,\bf l_1,\dots,\bf l_{\bf L})}{-\infty}{ \infty}
=
	\ClippedRealV{\MappingStructuralToVectorized(\Phi)}
	{\dims(\Phi)}
	{-\infty}{ \infty}.
\end{split}
\end{equation}
This and \cref{approx:lip:proofitem1} \prove that 
\begin{equation}
\label{approx:lip:eq3}
\begin{split} 
	\pnorm\infty{ \theta} 
\leq 
\textstyle
	\max \{ 1,L, \max_{k \in \{1, 2, \ldots, K\}} \infnorm{ \fx_k }, 2 \max_{k \in \{1, 2, \ldots, K\}} \abs{f(\fx_k)}\}.
\end{split}
\end{equation}
\Moreover
\enum{
	\eqref{approx:lip:eq2};
	\cref{cor:structvsvect};
	item~\ref{approx:lip:proofitem2};
}
\prove
that
\begin{equation}
\label{approx:lip:eq1}
\begin{split} 
	\sup\nolimits_{x \in E}  
	\babs{
		f(x)- \ClippedRealV{\theta}{(\bf l_0,\bf l_1,\dots,\bf l_{\bf L})}{-\infty}{ \infty}(x)
	} 
&=
	\sup\nolimits_{x \in E}  
	\babs{
		f(x)- 
		\ClippedRealV{\MappingStructuralToVectorized(\Phi)}
		{\dims(\Phi)}
		{-\infty}{ \infty}(x) 
	} \\
&=
	\sup\nolimits_{x \in E}  \babs{ f(x)- (\functionANN\rect(\Phi))(x) } \\
&\leq 
	2L \br*{\sup\nolimits_{x \in E} \pr*{ \inf\nolimits_{k  \in \{1,2, \ldots, K \}} \pnorm1{x- \fx_k}}}
\end{split}
\end{equation}
\cfload.
\end{aproof}

\cfclear 
\begin{athm}{cor}{approx:lipuv} 
Let $d, K, \bfd, \bfL \in \N$, 
$\bfl=(\bfl_0,\bfl_1, \ldots, \bfl_\bfL) \in \N^{\bfL+1}$, 
$L \in [0, \infty)$, 
$u \in [-\infty, \infty)$, 
$v \in (u, \infty]$ 
satisfy that 
\begin{equation}
  \bfL \geq \ceil{ \logg_2 K} +2, 
  \quad
\bfl_0=d, 
\quad
\bfl_\bfL=1, 
\quad
\bfl_1 \geq 2d K, 
\quad\text{and}\quad
\textstyle
\bfd \geq \sum_{i=1}^\bfL \bfl_i(\bfl_{i-1}+1), 
\end{equation}
assume for all 
$i \in \N\cap(1,\bfL)$ 
that
$\bfl_i \geq 3 \ceil[\big]{ \tfrac{ K}{2^{i-1}} }$, 
let $E \subseteq \R^d$ be a set, let $\fx_1, \fx_2, \ldots, \fx_K \in E$, and let $f \colon E \to ([u,v] \cap \R)$ satisfy for all $x , y \in E$ that $\abs{f(x)-f(y)} \leq L \pnorm1{x-y}$ \cfload. Then there exists $\theta \in \R^{\bfd}$ such that 
\begin{equation}
    \pnorm\infty{ \theta }\leq \max \{ 1,L, \max\nolimits_{k \in \{1, 2, \ldots, K\}} \infnorm{ \fx_k }, 2 \max\nolimits_{k \in \{1, 2, \ldots, K\}} \abs{f(\fx_k)}\}
\end{equation}
and 
\begin{equation}
    \sup\nolimits_{x \in E}  \babs{ f(x)- \ClippedRealV{\theta}{\bfl}{u}{v}(x) } \leq 2L \br*{\sup\nolimits_{x \in E} \pr*{ \inf\nolimits_{k  \in \{1,2, \ldots, K \}} \pnorm1{x- \fx _k}}}.
\end{equation}
\cfout.
\end{athm}
\begin{aproof}
\Nobs that \cref{approx:lip} \proves that there exists $\theta \in \R^{\bfd}$ such that
\begin{equation}
    \infnorm{ \theta } \leq \max \{ 1,L, \max\nolimits_{k \in \{1, 2, \ldots, K\}} \infnorm{ \fx_k }, 2 \max\nolimits_{k \in \{1, 2, \ldots, K\}} \abs{f(\fx_k)}\}
\end{equation}
and
\begin{equation}
\label{approx:lipuv:eq1}
      \sup\nolimits_{x \in E}  \babs{ f(x)- \ClippedRealV{\theta}{\bfl}{-\infty}{\infty}(x) } \leq 2L \br*{\sup\nolimits_{x \in E} \pr*{ \inf\nolimits_{k  \in \{1,2, \ldots, K \}} \pnorm1{x- \fx_k}}}.
\end{equation}
\Moreover
the assumption that $f(E) \subseteq [u,v]$ 
\proves that for all $x \in E$ it holds that 
\begin{equation}
  f(x)= \clip{u}{v}(f(x))
\end{equation}
\cfout. 
The fact that for all $x,y \in \R$ it holds that 
$\abs{\clip{u}{v}(x)-\clip{u}{v}(y)} \leq \abs{x-y}$ 
and \eqref{approx:lipuv:eq1} \hence \prove that
\begin{equation}
    \begin{split}
          &\sup\nolimits_{x \in E} \babs{ f(x)- \ClippedRealV{\theta}{\bfl}{u}{v}(x) } =   \sup\nolimits_{x \in E} \abs{ \clip{u}{v}(f(x))- \fc_{u,v}(\ClippedRealV{\theta}{\bfl}{-\infty}{\infty}(x)) } \\ 
          &\leq  \sup\nolimits_{x \in E} \babs{ f(x)- \ClippedRealV{\theta}{\bfl}{-\infty}{\infty}(x) } \leq 2L \br*{\sup\nolimits_{x \in E} \pr*{ \inf\nolimits_{k  \in \{1,2, \ldots, K \}} \pnorm1{x- \fx_k}}}.  
    \end{split}
\end{equation}
\end{aproof}

\subsection{Refined convergence rates for the approximation error}

\cfclear
\begin{athm}{lemma}{lem:approximation_error}
Let
$ d, \bfd, \bfL \in \N $,
$ L, a \in \R $,
$ b \in ( a, \infty ) $,
$ u \in [ -\infty, \infty ) $,
$ v \in ( u, \infty ] $,
$ \bfl = ( \bfl_0, \bfl_1, \ldots, \bfl_\bfL ) \in \N^{ \bfL + 1 } $,
assume
$ \bfl_0 = d $,
$ \bfl_\bfL = 1 $,
and
$ \bfd \geq \sum_{i=1}^{\bfL} \bfl_i( \bfl_{ i - 1 } + 1 ) $,
and
let
$ f \colon [ a, b ]^d \to ( [ u, v ] \cap \R ) $
satisfy for all
$ x, y \in [ a, b ]^d $
that
$ \lvert f( x ) - f( y ) \rvert \leq L \pnorm1{ x - y } $
\cfload.
Then
there exists $ \vartheta \in \R^\bfd $
such that
$ \infnorm{ \vartheta }
\leq \sup_{ x \in [ a, b ]^d } \lvert f( x ) \rvert $
and
\begin{equation}
\llabel{eq.claim}
\sup\nolimits_{ x \in [ a, b ]^d }
    \lvert \clippedNN{\vartheta}{\bfl}{u}{v}( x ) - f( x ) \rvert
\leq
\frac{ d L ( b - a ) }{ 2 }
\end{equation}
\cfout.
\end{athm}
\begin{aproof}
Throughout this proof,
let
$ \fd = \sum_{i=1}^{\bfL} \bfl_i( \bfl_{ i - 1 } + 1 ) $,
let
$ \bfm = ( \bfm_1, \allowbreak\ldots, \bfm_d ) \in [ a, b ]^d $
satisfy for all
$ i \in \{ 1, 2, \ldots, d \} $
that
\begin{equation}
  \llabel{eq:mi}
  \bfm_i = \frac{  a + b  }{2},
\end{equation}
and
let
$ \vartheta = ( \vartheta_1, \ldots,\allowbreak \vartheta_\bfd )\in \R^\bfd $
satisfy for all
$ i \in \{ 1, 2, \ldots, \bfd \} \backslash \{ \fd \} $
that
$ \vartheta_i = 0 $
and
$ \vartheta_\fd = f( \bfm ) $.
\Nobs that
the assumption that
$ \bfl_\bfL = 1 $
and
the fact that
$ \forall \, i \in \{ 1, 2, \ldots, \fd - 1 \} \colon
\vartheta_i = 0 $
\prove that for all
$ x = ( x_1, \ldots, x_{ \bfl_{ \bfL - 1 } } ) \in \R^{ \bfl_{ \bfL - 1 } } $
it holds that
\begin{equation}
\begin{split}
\Aff_{ 1, \bfl_{ \bfL - 1 } }^{ \vartheta, \sum_{i = 1}^{\bfL-1} \bfl_i ( \bfl_{ i - 1 } + 1) }( x )
&
=
\biggl[
    \smallsum_{ i = 1 }^{ \bfl_{ \bfL - 1 } } \vartheta_{ \br*{ \sum_{i = 1}^{\bfL-1} \bfl_i ( \bfl_{ i - 1 } + 1) } + i } x_i
\biggr]
+ \vartheta_{ \br*{ \sum_{i = 1}^{\bfL-1} \bfl_i ( \bfl_{ i - 1 } + 1) } + \bfl_{ \bfL - 1 } + 1 }
\\ &
=
\biggl[
    \smallsum_{ i = 1 }^{ \bfl_{ \bfL - 1 } } \vartheta_{ \br*{ \sum_{i = 1}^{\bfL} \bfl_i ( \bfl_{ i - 1 } + 1) } - ( \bfl_{ \bfL - 1 } - i + 1 ) } x_i
\biggr]
+ \vartheta_{ \sum_{i = 1}^{\bfL} \bfl_i ( \bfl_{ i - 1 } + 1) }
\\ &
=
\biggl[
    \smallsum_{ i = 1 }^{ \bfl_{ \bfL - 1 } } \vartheta_{ \fd - ( \bfl_{ \bfL - 1 } - i + 1 ) } x_i
\biggr]
+ \vartheta_{ \fd }
=
\vartheta_{ \fd }
=
f( \bfm )
\end{split}
\end{equation}
\cfload.
Combining this with
the fact that
$ f( \bfm ) \in [ u, v ] $
\proves that for all
$ x \in \R^{ \bfl_{ \bfL - 1 } } $
it holds that
\begin{equation}
\begin{split}
\bigl(
\Clip uv { \bfl_\bfL }
\circ
\Aff_{ \bfl_\bfL, \bfl_{ \bfL - 1 } }^{ \vartheta, \sum_{i = 1}^{\bfL-1} \bfl_i ( \bfl_{ i - 1 } + 1) }
\bigr)( x )
& =
\bigl(
\Clip uv1
\circ
\Aff_{ 1, \bfl_{ \bfL - 1 } }^{ \vartheta, \sum_{i = 1}^{\bfL-1} \bfl_i ( \bfl_{ i - 1 } + 1) }
\bigr)( x )
\\& =
\clip uv( f( \bfm ) )
=
\max\{ u, \min\{ f( \bfm ), v \} \}
\\&=
\max\{ u, f( \bfm ) \}
=
f( \bfm )
\end{split}
\end{equation}
\cfload.
This \proves for all
$ x \in \R^d $
that
\begin{equation}
\label{eq:NN_constant}
\clippedNN{\vartheta}{\bfl}{u}{v}( x )
=
f( \bfm )
.
\end{equation}
\Moreover
  \lref{eq:mi}
\proves that for all
$ x \in [ a, \bfm_1 ] $,
$ \fx \in [ \bfm_1, b ] $
it holds that
\begin{eqsplit}
  \lvert \bfm_1 - x \rvert
&= \bfm_1 - x
= \nicefrac{ ( a + b ) }{2} - x
\leq \nicefrac{ ( a + b ) }{2} - a
= \nicefrac{ ( b - a ) }{2}
\\ \text{and}\qquad
\lvert \bfm_1 - \fx \rvert
&= \fx - \bfm_1
= \fx - \nicefrac{ ( a + b ) }{2}
\leq b - \nicefrac{ ( a + b ) }{2}
= \nicefrac{ ( b - a ) }{2}
.
\end{eqsplit}
The assumption that
$ \forall \, x, y \in [ a, b ]^d \colon
\lvert f( x ) - f( y ) \rvert \leq L \pnorm1 { x - y } $
and
\cref{eq:NN_constant}
\hence
\prove that for all
$ x = ( x_1, \ldots, x_d ) \in [ a, b ]^d $
it holds that
\begin{equation}
\begin{split}
\lvert \clippedNN{\vartheta}{\bfl}{u}{v}( x ) - f( x ) \rvert
& =
\lvert f( \bfm ) - f( x ) \rvert
\leq
L \pnorm1 { \bfm - x }
=
L
\smallsum_{i=1}^d
    \lvert \bfm_i - x_i \rvert
\\ &
=
L
\smallsum_{i=1}^d
    \lvert \bfm_1 - x_i \rvert
\leq
\smallsum_{i=1}^d
    \frac{ L ( b - a ) }{2}
=
\frac{ d L ( b - a ) }{2}
.
\end{split}
\end{equation}
This and the fact that
$ \infnorm{ \vartheta }
=
\max_{ i \in \{ 1, 2, \ldots, \bfd \} }
    \lvert \vartheta_i \rvert
= \lvert f( \bfm ) \rvert
\leq \sup_{ x \in [ a, b ]^d } \lvert f( x ) \rvert $
\prove \lref{eq.claim}.
\end{aproof}

\cfclear
\begin{athm}{prop}{prop:approximation_error}
Let
$ d, \bfd, \bfL \in \N $,
$ A \in ( 0, \infty ) $, $L, a \in \R$,
$ b \in ( a, \infty ) $,
$ u \in [ -\infty, \infty ) $,
$ v \in ( u, \infty ] $,
$ \bfl = ( \bfl_0, \bfl_1, \ldots, \bfl_\bfL ) \in \N^{ \bfL + 1 } $,
assume
\begin{equation}
  \bfL \geq 1 + (\ceil{ \logg_2 \pr*{\nicefrac{ A  }{ ( 2d ) } }} + 1) \indicator{ ( 6^d, \infty )}( A ),
  \quad
  \bfl_0 = d ,
  \quad
  \bfl_1 \geq A \indicator{ ( 6^d, \infty )}( A ),
  \quad
  \bfl_\bfL = 1,
\end{equation}
and
$ \bfd \geq \sum_{i=1}^{\bfL} \bfl_i( \bfl_{ i - 1 } + 1 ) $,
assume for all
$ i \in \{ 1, 2, \ldots , \bfL\}\backslash\{1,\bfL\}$
that
\begin{equation}
  \bfl_i \geq   3 \ceil{\nicefrac{A}{(2^id)}} \indicator{ ( 6^d, \infty )}( A )
  ,
\end{equation}
and
let
$ f \colon [ a, b ]^d \to ( [ u, v ] \cap \R ) $
satisfy for all
$ x, y \in [ a, b ]^d $
that
\begin{equation}
\begin{split} 
\abs{ f( x ) - f( y ) } \leq L \pnorm1{ x - y }
\end{split}
\end{equation}
\cfload.
Then
there exists $ \vartheta \in \R^\bfd $
such that
$ \infnorm{ \vartheta }
\leq \max\{ 1, L,\allowbreak \abs{a}, \allowbreak \abs{b}, \allowbreak 2[ \sup_{ x \in [ a, b ]^d } \abs{ f( x ) } ] \} $
and
\begin{equation}
  \llabel{claim}
\sup\nolimits_{ x \in [ a, b ]^d }
    \abs{ \ClippedRealV{\vartheta}{\bfl}{u}{v}( x ) - f( x ) }
\leq
\frac{ 3 d L ( b - a ) }{ A^{ \nicefrac{1}{d} } }
\end{equation}
\cfout.
\end{athm}
\begin{aproof}
Throughout this proof, assume without loss of generality that $A > 6^d$\cfadd{lem:approximation_error}
\cfload,
let $\fZ = \floor{\bigl(\tfrac{ A }{ 2d } \bigr)^{ \nicefrac{1}{d} }}\in\Z$.
\Nobs that the fact that for all
$ k \in \N $
it holds that
$
2 k \leq 2 ( 2^{ k - 1 }) = 2^k
$
\proves that
$ 3^d
= \nicefrac{6^d}{2^d}
\leq \nicefrac{A}{(2d)} $.
\Hence that
\begin{equation} \label{eq:fnestimate}
    2 \leq \tfrac{2}{3}\bigl(\tfrac{ A }{ 2d } \bigr)^{ \nicefrac{1}{d} } \leq \bigl(\tfrac{ A }{ 2d } \bigr)^{ \nicefrac{1}{d} }-1 < \fZ. 
\end{equation}
In the next step let $r=\nicefrac{d(b-a)}{2\fZ}\in(0,\infty)$, 
let $\delta \colon [a,b]^d \times [a,b]^d \to \R$ satisfy for all $x,y \in [a,b]^d$ that 
$\delta(x,y) = \pnorm1{x-y}$, 
and let 
$K = \max(2, \CovNum{ ( [ a, b ]^d, \delta ), r })\in \N \cup \{ \infty \}$ \cfload.
\Nobs that \eqref{eq:fnestimate} and 
\cref{lem:covering_number_cube}
\prove that
\begin{equation}
\begin{split}
K = \max\{ 2, \CovNum{ ( [ a, b ]^d, \delta ), r } \}
\leq\max\Bigl\{ 2, \Bigl( \ceil{ \tfrac{ d ( b - a ) }{2r} } \Bigr)^{ \! d }
\Bigr\}
=\max\{ 2, ( \lceil \fZ \rceil )^d \}
=\fZ^d < \infty.
\end{split}
\end{equation}
This \proves that
\begin{equation}
\label{eq:fN_upper_bound}
4 \leq 2 d K \leq 2 d \fZ^d
\leq\tfrac{ 2 d A }{2d} = A.
\end{equation}
Combining this and the fact that
$ \bfL \geq 1+ (\ceil{ \logg_2 \pr*{\nicefrac{ A  }{ ( 2d ) } }} + 1) \indicator{ ( 6^d, \infty )}( A ) = \ceil{ \logg_2 \pr*{\nicefrac{ A  }{ ( 2d ) } }} + 2$
\hence \proves that
$ \ceil{\logg_2(K)}
\leq \ceil{\operatorname{log}_2 \pr*{\nicefrac{A}{(2d)}}}
\leq \bfL - 2 $.
This,
\cref{eq:fN_upper_bound},
the assumption that
$ \bfl_1 \geq A \indicator{ ( 6^d, \infty )}( A ) = A $,
and the assumption that
$ \forall \, i \in  \{ 2, 3, \ldots, \bfL - 1 \} \colon
 \bfl_i \geq  3 \ceil{\nicefrac{A}{(2^id)}}  \indicator{ ( 6^d, \infty )}( A ) =   3 \ceil{\nicefrac{A}{(2^id)}} $
\prove that for all
$ i \in \{ 2, 3, \ldots, \bfL-1 \} $ it holds that
\begin{equation}
\label{eq:bfl_assumptions1}
\bfL
\geq \ceil{\operatorname{log}_2(K)} + 2,
\qquad
\bfl_1 \geq A \geq 2 d K,
\qandq 
\bfl_i \geq 3\ceil{\tfrac{A}{2^i d}}
\geq  3\ceil{\tfrac{ K }{ 2^{i-1} }}.
\end{equation}
Let $\fx_1, \fx_2, \ldots, \fx_K \in [a,b]^d$ satisfy
\begin{equation} \label{eq:maxdist}
    \sup\nolimits_{x \in [a,b]^d} \br*{ \inf\nolimits_{k  \in \{1,2, \ldots, K \}} \delta (x,  \fx_k) } \leq r.
\end{equation}
\Nobs that \eqref{eq:bfl_assumptions1}, the assumptions that
$ \bfl_0 = d $,
$ \bfl_\bfL = 1 $,
$ \bfd \geq \sum_{i=1}^{\bfL} \bfl_i( \bfl_{ i - 1 } + 1 ) $, 
and $\forall \, x,y \in [a,b]^d \colon \abs{f(x)-f(y)} \leq L \pnorm1{x-y}$, 
and \cref{approx:lipuv} 
\prove that there exists
$ \vartheta \in \R^\bfd $
such that
\begin{equation} \label{eq:prop:approx:normtheta}
     \infnorm{ \vartheta } \leq
\max \{ 1,L, \max\nolimits_{k \in \{1, 2, \ldots, K\}} \infnorm{ \fx_k }, 2 \max\nolimits_{k \in \{1, 2, \ldots, K\}} \abs{f(\fx_k)}\}
\end{equation}
and
\begin{equation}
\label{eq:NN_approximation}
\begin{split}
\sup\nolimits_{ x \in [ a, b ]^d }
    \abs{ \ClippedRealV{\vartheta}{\bfl}{u}{v}( x ) - f( x ) }
&\leq
2 L\br*{
\sup\nolimits_{ x \in [ a, b ]^d }
 \pr*{\inf\nolimits_{k  \in \{1, 2, \ldots, K \}}
    \pnorm1{x- \fx_k} }} \\
&=2 L\br*{\sup\nolimits_{ x \in [ a, b ]^d }
\pr*{ \inf\nolimits_{k  \in \{1, 2, \ldots, K \}}
    \delta ( x, \fx_k ) } }.
    \end{split}
\end{equation}
\Nobs that \eqref{eq:prop:approx:normtheta} \proves that
\begin{equation}
\label{eq:norm_estimate}
\infnorm{\vartheta }
\leq
\max\{ 1, L, \abs{a}, \abs{b}, 2  \sup\nolimits_{ x \in [ a, b ]^d } \abs{f(x)}  \}.
\end{equation}
\Moreover
\cref{eq:NN_approximation}, \eqref{eq:fnestimate}, \eqref{eq:maxdist},
and
 the fact that for all
$ k \in \N $
it holds that
$
2 k \leq 2 ( 2^{ k - 1 }) = 2^k
$
\prove that
\begin{equation}
\begin{split}
&\sup\nolimits_{ x \in [ a, b ]^d }
    \abs{ \ClippedRealV{\vartheta}{\bfl}{u}{v}( x ) - f( x ) }
\leq 2 L
\bigl[
\sup\nolimits_{ x \in [ a, b ]^d }
\pr*{  \inf\nolimits_{k  \in \{1, 2, \ldots, K \}}
    \delta ( x, \fx_k )}
\bigr]
\\&\leq
2 L r
=
\frac{ d L ( b - a ) }{ \fZ }
\leq
\frac{ d L ( b - a ) }{ \frac{2}{3} \bigl( \frac{ A }{ 2d } \bigr)^{ \nicefrac{1}{d} } }
= \frac{ ( 2 d )^{ \nicefrac{1}{d} } 3 d L ( b - a ) }{ 2 A^{ \nicefrac{1}{d} } }
\leq \frac{ 3 d L ( b - a ) }{ A^{ \nicefrac{1}{d} } }.
\end{split}
\end{equation}
Combining this with
\cref{eq:norm_estimate} \proves \lref{claim}.
\end{aproof}

\begingroup
\providecommandordefault{\c}{\mathfrak{c}}
\providecommandordefault{\C}{\mathfrak{C}}
\providecommandordefault{\F}{\interpolatingDNN}

\providecommandordefault{\Ae}{{A_\varepsilon}}
\providecommandordefault{\Le}{{\bfL_\varepsilon}}
\providecommandordefault{\le}{\bfl^{(\varepsilon)}}
\providecommandordefault{\Fe}{\mathbf{F}_\varepsilon}

\cfclear
\begin{athm}{cor}{cor:approxerror:eps}
Let
$ d  \in \N $,
$a \in \R$,
$ b \in ( a, \infty ) $,
$L \in (0,\infty)$
and let
$ f \colon [ a, b ]^d \to  \R $
satisfy for all
$ x, y \in [ a, b ]^d $
that
\begin{equation}
\begin{split} 
\abs{ f( x ) - f( y ) } \leq L \pnorm1{ x - y }
\end{split}
\end{equation}
\cfload.
Then 
there exist $\C \in \R$ 
such that for all
	$\varepsilon \in (0,1]$
there exists  
$\F \in \ANNs$ such that 
\begin{equation}
\label{cor:approxerror:eps:concl1}
\begin{split} 
	\hiddenLength(\F) 
\leq
	\max\cu[\big]{
		0 , 
		d( \operatorname{log}_2(\varepsilon^{-1}) + \operatorname{log}_2 (d) + \operatorname{log}_2 ( 3L(b-a)) + 1 )
	},
\end{split}
\end{equation}
\begin{equation}
\label{cor:approxerror:eps:concl2}
\begin{split} 
	\infnorm{ \MappingStructuralToVectorized(\F ) }
	\leq \max\cu[\big]{ 1, L, \abs{a}, \abs{b}, \allowbreak 2[ \sup\nolimits_{ x \in [ a, b ]^d } \abs{ f( x ) } ] },
\qquad
	\functionANN\rect(\F) \in C(\R^d, \R),
\end{split}
\end{equation}
\begin{equation}
\label{cor:approxerror:eps:concl3}
\begin{split} 
	\sup\nolimits_{ x \in [ a, b ]^d }
	    \abs{ (\functionANN\rect(\F))( x ) - f( x ) }
	\leq \varepsilon,
\qandq
	\paramANN(\F) \leq \C \varepsilon^{-2d}
\end{split}
\end{equation}
\cfout.
\end{athm}

\begin{aproof}
Throughout this proof, %
let $\C \in \R$ satisfy
\begin{equation}
\label{eq:approxeps:defc}
	\C 
= 
	\tfrac{9}{8}
	\bigl(3dL(b - a ) \bigr) ^{2d}
	+
	\pr{
		d+22
	}
	\bigl( 3 d L ( b-a) \bigr) ^d
	+
	d + 11,
\end{equation}
for every 
	$\varepsilon \in (0, 1 ]$
let
	$\Ae \in (0,\infty)$,
	$\bfL_\varepsilon \in \N$,
	$\bfl^{(\varepsilon)} = (\bfl^{(\varepsilon)}_0, \bfl^{(\varepsilon)}_1, \ldots, \bfl^{(\varepsilon)}_\Le) \in \N^{\Le + 1}$
satisfy 
\begin{equation}
\label{eq:approxeps:defA}
\begin{split} 
	\Ae = \pr*{ \frac{3d L (b-a)}{\varepsilon} } ^d,
\qquad
	\bfL_\varepsilon = 1 + \pr*{\ceil[\big]{ \logg_2 \pr[\big]{\tfrac{ \Ae }{ 2d }}} + 1} \indicator{ ( 6^d, \infty )}( \Ae ),
\end{split}
\end{equation}
\begin{equation}
\label{eq:approxeps:defL1}
\begin{split} 
	\bfl^{(\varepsilon)}_0=d, 
\qquad
	\bfl^{(\varepsilon)}_1 = \floor{\Ae} \indicator{ ( 6^d, \infty )}( \Ae ) + 1,	
\qandq
	\bfl^{(\varepsilon)}_\Le = 1, 
\end{split}
\end{equation}
and
assume for all 
	$\varepsilon \in (0,1]$,
	$i \in \{2,3, \ldots, \Le - 1 \}$ 
that
\begin{equation}
\label{eq:approxeps:defL2}
\begin{split} 
	\bfl^{(\varepsilon)}_i =  3 \ceil[\big]{\tfrac{\Ae}{2^id}} \indicator{ ( 6^d, \infty )}( \Ae )
\end{split}
\end{equation}
\cfload. 
Observe that 
the fact that for all 
	$\varepsilon \in (0,1]$
it holds that
$\Le \geq 1 + \pr*{ \bceil{ \operatorname{log}_2 \bpr{\tfrac{ \Ae  }{  2d  } }} + 1 } \indicator{ ( 6^d, \infty )}( \Ae )$,
the fact that for all 
	$\varepsilon \in (0,1]$
it holds that
	$\le_0 = d$, 
the fact that for all 
	$\varepsilon \in (0,1]$
it holds that
$\le_1 \geq \Ae \indicator{ ( 6^d, \infty )}( \Ae )$, 
the fact that for all 
	$\varepsilon \in (0,1]$
it holds that
$\le_\Le = 1$,
the fact that for all 
	$\varepsilon \in (0,1]$,
	$i \in \{2,3, \ldots, \Le - 1 \}$ 
it holds that $\le_i \geq  3 \ceil{\tfrac{\Ae}{2^i d}} \indicator{ ( 6^d, \infty )}( \Ae )$,
\cref{prop:approximation_error}, 
and
\cref{cor:structvsvect}
\prove that for all
	$\varepsilon \in (0,1]$
there exists 
$\Fe \in \bpr{ \bigtimes_{i=1}^\Le  \pr*{\R^{\le_i \times \le_{i-1}} \times \R^{\le_i} } } \subseteq \ANNs$ which satisfies $\infnorm{\MappingStructuralToVectorized(\Fe)} \leq \max\{ 1, L, \abs{a}, \abs{b}, \allowbreak 2[ \sup_{ x \in [ a, b ]^d } \abs{ f( x ) } ] \} $ and
\begin{equation}
  \llabel{eq.approx}
    \sup \nolimits_{x \in [a,b]^d }   \abs{ (\functionANN\rect(\Fe ))( x ) - f( x ) } \leq \frac{ 3 d L ( b - a ) }{ (\Ae)^{ \nicefrac{1}{d} } } = \varepsilon.
\end{equation}
\cfload.
\Moreover
the fact that $d \geq 1$ \proves 
that for all	
	$\varepsilon \in (0,1]$ 
it holds that
\begin{equation}
\begin{split}
	\hiddenLength(\Fe) 
	&= \Le - 1 
	= (\ceil[\big]{ \logg_2 \pr[\big]{\tfrac{ \Ae }{  2d  } }} + 1) \indicator{ ( 6^d, \infty )}( \Ae )\\
&= \ceil{\logg_2 ( \tfrac{\Ae}{d})} \indicator{ ( 6^d, \infty )}( \Ae )
	\leq
	\max\{0, \logg_2(\Ae) + 1 \}.
\end{split}
\end{equation}
Combining this and the fact that 
for all	
	$\varepsilon \in (0,1]$ 
it holds that
\begin{equation}
\begin{split} 
	\logg_2(\Ae)
=
	d \logg_2 \pr*{ \tfrac{3dL(b-a)}{\varepsilon} }
=
	d \bpr{ \logg_2(\varepsilon^{-1}) + \logg_2 (d) + \logg_2(3L(b-a)) }
\end{split}
\end{equation}
\proves
that for all	
	$\varepsilon \in (0,1]$ 
it holds that
\begin{equation}
\label{cor:approxerror:eps:eq1}
\begin{split} 
	\hiddenLength(\Fe)
\leq
	\max\cu*{0, d \bpr{ \logg_2(\varepsilon^{-1}) + \logg_2 (d) + \logg_2(3L(b-a)) } + 1 }.
\end{split}
\end{equation}
\Moreover
\eqref{eq:approxeps:defL1}
and
\eqref{eq:approxeps:defL2}
\prove
that for all	
	$\varepsilon \in (0,1]$ 
it holds that
\begin{equation}
\begin{split} 
\label{eq:approxeps1}
	&\paramANN(\Fe) 
= 
	\sum_{i=1}^\Le \le_i (\le_{i-1} + 1) \\
&\leq
	\pr[\big]{\floor{\Ae}  + 1} (d + 1) 
	+
	3 \ceil[\big]{\tfrac{\Ae}{4d}}\pr[\big]{\floor{\Ae} + 2} \\
&\quad
	+
	\max\cu[\big]{
		\floor{\Ae} + 1 , 
		3 \ceil[\big]{\tfrac{\Ae}{2^{\Le-1}d}}
	} 
	+ 1
	+
	\sum_{i=3}^{\Le-1} 3 \ceil[\big]{\tfrac{\Ae}{2^id}}  (3 \ceil[\big]{\tfrac{\Ae}{2^{i-1}d}} + 1) 
\\&\leq
	(\Ae + 1)(d + 1) 
	+
	3 \pr[\big]{\tfrac{\Ae}{4} + 1} \pr[\big]{\Ae + 2}
	+
	3 \Ae 
	+ 
	4
	+
	\sum_{i=3}^{\Le-1} 3 \pr[\big]{ \tfrac{\Ae}{2^i} + 1 }  \pr[\big]{\tfrac{3\Ae}{2^{i-1}} + 4}.
\end{split}
\end{equation}
\Moreover
the fact that $\forall \, x \in (0, \infty) \colon \logg_2 (x) = \logg_2 ( \nicefrac{x}{2}) + 1 \leq \nicefrac{x}{2} + 1$ 
\proves 
that for all	
	$\varepsilon \in (0,1]$ 
it holds that
\begin{equation}
    \Le \leq 2 + \logg_2 ( \tfrac{\Ae}{d}) \leq 3 + \tfrac{\Ae}{2d} \leq 3 + \tfrac{\Ae}{2}.
\end{equation}
This \proves 
that for all	
	$\varepsilon \in (0,1]$ 
it holds that
\begin{equation} 
\label{eq:approxeps3}
\begin{split}
	&\sum_{i=3}^{\Le-1} 3 \pr[\big]{ \tfrac{\Ae}{2^i} + 1 }  \pr[\big]{\tfrac{3\Ae}{2^{i-1}} + 4}\\
&\leq
	9(\Ae)^2
	\br*{ 
		\sum_{i=3}^{\Le - 1} 2^{1-2i}
	} 
	+ 
	12 \Ae 
	\br*{ 
		\sum_{i=3}^{\Le - 1} 2^{-i}
	} 
	+ 
	9\Ae
	\br*{
		\sum_{i=3}^{\Le - 1} 2^{1-i}
	} 
	+ 
	12(\Le - 3)  \\
&\leq
	\tfrac{9(\Ae)^2}{8} 
	\br*{ 
		\sum_{i=1}^{\infty} 4^{-i}
	} 
	+ 
	3 \Ae
	\br*{ 
		\sum_{i=1}^{\infty} 2^{-i}
	} 
	+ 
	\tfrac{9\Ae}{2}
	\br*{
		\sum_{i=1}^{\infty} 2^{-i}
	} 
	+ 
	6 \Ae   \\
&= 
	\tfrac{3}{8} (\Ae)^2 + 3\Ae + \tfrac{9}{2} \Ae + 6\Ae 
= 
	\tfrac{3}{8} (\Ae)^2 + \tfrac{27}{2} \Ae.
\end{split}
\end{equation}
This and \eqref{eq:approxeps1}
\prove
that for all	
	$\varepsilon \in (0,1]$ 
it holds that
\begin{equation}
\begin{split} 
	\paramANN(\Fe) 
&\leq
	\pr{
		\tfrac{3}{4}
		+
		\tfrac{3}{8}
	}
	(\Ae)^2
	+
	\pr{
		d+1
		+
		\tfrac{9}{2}
		+
		3
		+
		\tfrac{27}{2}
	}
	\Ae
	+
	d + 1
	+
	6
	+
	4 \\
&=
	\tfrac{9}{8}
	(\Ae)^2
	+
	\pr{
		d+22
	}
	\Ae
	+
	d + 11.
\end{split}
\end{equation}
Combining 
this,
\eqref{eq:approxeps:defc}, and 
\eqref{eq:approxeps:defA}
\proves that
\begin{equation}
\label{cor:approxerror:eps:eq2}
\begin{split}
	\paramANN(\Fe) 
&\leq 
	\tfrac{9}{8}
	\bigl(3dL(b - a ) \bigr) ^{2d} \varepsilon ^{-2d} 
	+
	\pr{
		d+22
	}
	\bigl( 3 d L ( b-a) \bigr) ^d \varepsilon ^{- d}
	+
	d + 11\\
&\leq
	\br*{
		\tfrac{9}{8}
		\bigl(3dL(b - a ) \bigr) ^{2d}
		+
		\pr{
			d+22
		}
		\bigl( 3 d L ( b-a) \bigr) ^d
		+
		d + 11
	} 
	\varepsilon ^{-2d} 
= 
	\C \varepsilon^{-2d}.
\end{split}
\end{equation}
Combining 
this
with
\lref{eq.approx} and
\eqref{cor:approxerror:eps:eq1}
\proves[ep]
\cref{cor:approxerror:eps:concl1,cor:approxerror:eps:concl2,cor:approxerror:eps:concl3}.
\end{aproof}

\begin{athm}{remark}{high_dim_approx_remark}[High-dimensional \ann\ approximation results]
\cref{cor:approxerror:eps} above is a multi-dimensional \ann\ approximation result in the sense that the input dimension $d \in \N$ of the domain of definition $[a,b]^d$ of the considered target function $f$ that we intend to approximate can be any natural number.
However, 
we note that
\cref{cor:approxerror:eps} does not provide a useful contribution in the case when the dimension $d$ is large, say $d \geq 5$, as
\cref{cor:approxerror:eps} does not provide any information on how the constant $\C$ in \cref{cor:approxerror:eps:concl3} grows in $d$ and as the dimension $d$ appears in the exponent of the reciprocal $\varepsilon^{-1}$ of the prescribed approximation accuracy $\varepsilon$ in the bound for the number of \ann\ parameters in \cref{cor:approxerror:eps:concl3}.

In the literature there are also a number of suitable high-dimensional \ann\ approximation results which assure that the constant in the parameter bound grows at most polynomially in the dimension $d$ and which assure that the exponent of the reciprocal $\varepsilon^{-1}$ of the prescribed approximation accuracy $\varepsilon$ in the \ann\ parameter bound
is completely independent of the dimension $d$.
Such results do have the potential to provide a useful practical conclusion for \ann\ approximations even when the dimension $d$ is large.
We refer, \eg, to \cite{Barron1994,Barron1993,Weinan2022,Gonon2023a,Beneventano2021,Cheridito2022a} and the references therein  for such
high-dimensional \ann\ approximation results in the context of general classes of target functions and 
we refer, \eg, to \cite{Grohs2023Aproof,Grohs2021,HutzenthalerJentzenKruseNguyen2019,Weinan2022a,Gonon2021,Gonon2022,Gonon2023b,Beneventano2020,Ackermann2023,Berner2020,Elbraechter2022,HornungJentzenSalimova2020arXiv,JentzenSalimovaWelti2021,Kutyniok2022,Reisinger2020} and the references therein for such high-dimensional \ann\ approximation results where the target functions are solutions of \PDEs\ (cf.\ also \cref{sec:ANN_approx_PDEs} below).
\end{athm}
\endgroup

\begin{athm}{remark}{T_B_D}[Infinite-dimensional \ann\ approximation results]
In the literature there are now also results where the target function that we intend to approximate is defined on an infinite-dimensional vector space and where the dimension of the domain of definition of the target function is thus infinity
(see, \eg, \cite{Benth2023,Sandberg1991,Chen1993,Chen1995,Hornik1993,Kratsios2021} and the references therein).
This perspective seems to be very reasonable as in many applications, input data, such as images and videos, that should be processed through the target function are more naturally represented by elements of infinite-dimensional spaces instead of elements of finite-dimensional spaces.
\end{athm}

%% file: parts/Optimization_through_flows_of_ODEs.tex
\cchapter{Optimization through gradient flow (GF) trajectories}{chapter:flow}
\chaptermark{Optimization through ODEs}

\begin{introductions}
	In \cref{chapter:deterministic,chapter:stochastic} below 
	we study deterministic and stochastic \GD-type optimization methods from the literature. 
	Such methods are widely used in machine learning problems 
	to approximately minimize suitable objective functions. 
	The \SGD-type optimization methods 
	in Chapter~\ref{chapter:stochastic} can be 
	viewed as suitable Monte Carlo approximations of 
	the deterministic \GD-type optimization methods 
	in Chapter~\ref{chapter:deterministic} and the deterministic 
	\GD-type optimization methods in Chapter~\ref{chapter:deterministic} 
	can, roughly speaking, be viewed as time-discrete approximations of 
	solutions of suitable \GF\ \ODEs. 
	To develop intuitions for \GD-type optimization methods
	and for some of the tools which we employ to analyze such methods, 
	we study in this chapter such \GF\ \ODEs. 
	In particular, we show in this chapter how such \GF\ 
	 \ODEs\ can be used to approximately 
	solve appropriate optimization problems.
	
	Further investigations on optimization through \GF\ \ODEs\ can, for example, be found in 
	\cite{Jentzen2023,JentzenRiekert2022Existence,Eberle2023,Bolte2006,Kurdyka2000,Absil2005,Ibragimov2022}
	 and the references therein.
\end{introductions}

\section{Introductory comments for the training of ANNs}
\label{sec:intro_training_anns}

\cfclear
\begingroup
\providecommandordefault{\inputDim}{\defaultInputDim}
\providecommandordefault{\netDim}{{\defaultNetDim}}
\providecommandordefault{\LossFunction}{\defaultLossFunction}
\providecommandordefault{\targetFunction}{\cE}
\providecommandordefault{\empRiskInifite}{\mathfrak{L}}
\providecommandordefault{\x}{\defaultx}
\providecommandordefault{\y}{\defaulty}

\renewcommand{\d}{\mathfrak d}
\newcommand{\m}[2]{{\mathsf m}_{#1,#2}}
\renewcommand{\th}[1]{\theta_{#1}}

Key components of deep supervised learning algorithms are typically
deep \anns\ and also suitable \emph{gradient based optimization methods}.
In \cref{part:ANNs,part:approx} we have introduced and studied different types of \anns\ while in \cref{part:opt} we introduce and study gradient based optimization methods.
In this section we briefly outline the main ideas behind gradient based optimization methods and
sketch how such gradient based optimization methods arise within deep supervised learning algorithms.
To do this, we now recall the deep supervised learning framework
from the \hyperref[sec:intro]{introduction}.

Specifically, 
let $ \defaultInputDim, M \in \N $, 
$ \targetFunction \in C( \R^\defaultInputDim, \R ) $,
$ \x_1, \x_2, \dots, \x_{ M + 1 } \in \R^\defaultInputDim $, 
$ \y_1, \y_2, \dots, \y_M \in \R $ 
satisfy for all $ m \in \{ 1, 2, \dots, M \} $ that
\begin{equation}
	\label{eq:gdintro.1}
  \y_m = \targetFunction( \x_m )
\end{equation}
and let 
$ \empRiskInifite \colon C( \R^\defaultInputDim, \R ) \to [0,\infty) $
satisfy for all 
$
  \phi 
  \in C( \R^\defaultInputDim, \R )
$
that
\begin{equation}
	\label{eq:gdintro.2}
  \empRiskInifite( \phi )
  =
	\frac1M\br*{
  \sum_{ m = 1 }^M 
  \abs*{ \phi( \x_m ) - \y_m }^2
	}
  .
\end{equation}
As in the  \hyperref[sec:intro]{introduction} we think of $ M \in \N $ as the number of available known input-output data pairs, 
we think of $ \defaultInputDim \in \N $ as the dimension of the input data, 
we think of $ \targetFunction \colon \R^\defaultInputDim \to \R $ as an unknown function 
which relates input and output data through \cref{eq:gdintro.1}, 
we think of $ \x_1, \x_2, \dots, \x_{ M + 1 } \in \R^\defaultInputDim $ as the available known 
input data, we think of $ \y_1, \y_2, \dots, \y_M \in \R $ as the available 
known output data, and we have that the function
$\empRiskInifite \colon C(\R^\defaultInputDim,\R)\to[0,\infty)$ 
in \eqref{eq:gdintro.2}
is the objective function (the function we want to minimize) in the optimization problem associated to
the considered learning problem 
(cf.\  \eqref{eq:intro.2} in
the  \hyperref[sec:intro]{introduction}).
In particular, observe that
\begin{equation}
\begin{split} 
\empRiskInifite(\targetFunction)=0
\end{split}
\end{equation}
and we are trying
to approximate the function $\targetFunction$ by
computing an approximate minimizer
of the function $\empRiskInifite\colon C(\R^\defaultInputDim,\R)\to[0,\infty)$.
In order to make this optimization problem amenable to
numerical computations, we consider a
spatially discretized version of the optimization problem associated to \cref{eq:gdintro.2} by employing parametrizations of \anns\ (cf.\ \cref{intro:eq1} in the \hyperref[sec:intro]{introduction}).

More formally, let 
$\activation \colon \R \to \R$ be differentiable,
let
$h\in\N$,
$l_1,l_2,\dots,l_h,\defaultParamDim\in\N$ satisfy
$\defaultParamDim=l_1(d+1)+\br[\big]{\sum_{k=2}^h l_k(l_{k-1}+1)}+l_h+1$, and
consider 
the parametrization function
\begin{equation}
\label{trainingintro:eq1}
	\R^\defaultParamDim \ni \theta
\mapsto
	\RealV{ \theta}{0}{\defaultInputDim}{ \multdim_{\activation, l_1}, \multdim_{\activation, l_2}, \dots, \multdim_{\activation, l_h} , \id_{ \R } }
	\in
	C(\R^\defaultInputDim,\R)
\end{equation}
\cfload.
Note that $h$ is the number of hidden layers of the \anns\ in \cref{trainingintro:eq1},
note for every $i\in\{1,2,\dots,h\}$ that $l_i\in\N$
is the number of neurons in the $i$-th hidden layer of
the \anns\ in \cref{trainingintro:eq1},
and
note that $\defaultParamDim$ is the number of real parameters used to describe
the \anns\ in \cref{trainingintro:eq1}.
Observe that for every $\theta \in \R^\defaultParamDim$ we have that the function 
\begin{equation}
\begin{split} 
	\R^{\defaultInputDim} \ni x 
\mapsto
	\RealV{ \theta}{0}{\defaultInputDim}{ \multdim_{\activation, l_1}, \multdim_{\activation, l_2}, \dots, \multdim_{\activation, l_h} , \id_{ \R } }
	\in \R
\end{split}
\end{equation}
in  \cref{trainingintro:eq1}  is nothing else than the realization function associated to a \emph{fully-connected feedforward} \ann\ %
where before each hidden layer a multi-dimensional version of the activation function $\activation \colon \R \to \R$ is applied.
We restrict ourselves in this section to a differentiable activation function
as this differentiability property allows us to consider gradients
(cf.\ \eqref{eq:gdintro.4}, \eqref{eq:gdintro.5}, and \cref{sec:diff_anns} below for details).

We now discretize the optimization problem in \cref{eq:gdintro.2}
as the 
problem of computing approximate minimizers of the function $\defaultLossFunction \colon\R^\defaultParamDim\to[0,\infty)$ which satisfies
for all $\theta\in\R^\defaultParamDim$ that
\begin{equation}
\label{eq:gdintro.3}
  \defaultLossFunction( \theta )
  = 
	\frac1M
  \br*{
    \sum_{ m = 1 }^M 
    \abs[\big]{ \bpr{ \RealV{ \theta}{0}{\defaultInputDim}{ \multdim_{\activation, l_1}, \multdim_{\activation, l_2}, \dots, \multdim_{\activation, l_h} , \id_{ \R } } } ( \x_m ) - \y_m }^2
  }
\end{equation}
and this resulting optimization problem is now accessible to numerical computations.
Specifically, deep learning algorithms solve optimization problems of the type \cref{eq:gdintro.3} by means of \emph{gradient based optimization methods}.
Loosely speaking, gradient based optimization methods aim to minimize the considered objective function (such as \cref{eq:gdintro.3} above) by performing successive steps based on the direction of the negative gradient of the objective function. 
One of the simplest gradient based optimization method is the plain-vanilla \GD\ optimization method 
which performs successive steps in the direction of the negative gradient
and we now sketch the \GD\ optimization method applied to \cref{eq:gdintro.3}.
Let
$\xi\in\R^\defaultParamDim$,
let
$(\gamma_n)_{n\in\N}\subseteq[0,\infty)$, 
and let $\theta = (\theta_n)_{n \in \N_0} \colon  \N_0 \to \R^\defaultParamDim$ satisfy for all 
	$n \in \N$
that
\begin{equation}
	\label{eq:gdintro.4}
   \theta_0 = \xi \qandq \theta_n = \theta_{n-1} - \gamma_n (\nabla \defaultLossFunction)(\theta_{n-1}).
\end{equation}
The process $(\theta_n)_{n \in \N_0}$ is the \GD\ process for the minimization problem associated to \cref{eq:gdintro.3} with learning rates $(\gamma_n)_{n\in\N}$  and initial value $\xi$ (see \cref{def:GD} below for the precise definition).

This plain-vanilla \GD\ optimization method and related \GD-type optimization methods can be regarded as discretizations of solutions of \GF\ \ODEs.
In the context of the minimization problem in  \cref{eq:gdintro.3} such solutions of \GF\ \ODEs\ can be described as follows.
Let $\Theta = (\Theta_t)_{t \in [0,\infty)} \colon  [0,\infty) \to \R^\defaultParamDim$ be a continuously differentiable function which satisfies for all
	$t \in [0,\infty)$ that
\begin{equation}
\label{eq:gdintro.5}
\begin{split} 
	\Theta_0 = \xi 
\qandq 
	\dot{\Theta}_t 
	= 
	\tfrac{\partial}{\partial t}\Theta_t
	=
	-(\nabla \defaultLossFunction)(\Theta_t).
\end{split}
\end{equation}
The process $(\Theta_t)_{t \in [0,\infty)}$ is the solution of the \GF\ \ODE\ corresponding to the minimization problem associated to \cref{eq:gdintro.3} with initial value $\xi$.

In \cref{chapter:deterministic} below we introduce and study deterministic \GD-type optimization methods such as the \GD\ optimization method in \cref{eq:gdintro.4}.
To develop intuitions for \GD-type optimization methods
and for some of the tools which we employ to analyze such \GD-type optimization methods, 
we study in the remainder of this chapter \GF\ \ODEs\ such as \cref{eq:gdintro.5} above. 
In deep learning algorithms usually not \GD-type optimization methods but stochastic variants of \GD-type optimization methods are employed to solve optimization problems of the form \cref{eq:gdintro.3}.
Such \SGD-type optimization methods can be viewed as suitable Monte Carlo approximations of deterministic \GD-type methods and in \cref{chapter:stochastic} below we treat
such \SGD-type optimization methods.

\endgroup

\section{Basics for GFs}

\subsection{GF ordinary differential equations (ODEs)}

\cfclear
\begingroup
\renewcommand{\d}{\mathfrak d}
\newcommand{\s}{\xi}
\renewcommand{\loss}{\mathscr{L}}
\newcommand{\gradient}{\mathscr{G}}
\newcommand{\flow}{\Theta}
\begin{athm}{definition}{def:gradientflow}[\GF\ trajectories]
	Let 
		$\d\in\N$,
		$\s\in\R^\d$,
	let
		$\loss\colon \R^\d\to\R$
		be a function,
	and let 
		$\gradient\colon\R^\d\to\R^\d$
		be a $\Borel(\R^\d)$/$\Borel(\R^\d)$-measurable function
	which satisfies for all
		$U\in\{V\subseteq \R^\d\colon \text{V is open}\}$,
		$\theta\in U$
		with $\loss|_U\in C^1(U,\R^\d)$
	that
	\begin{equation}
		\llabel{eq:grad}
		\gradient(\theta)
		=
		(\nabla \loss)(\theta)
		.
	\end{equation}
	Then we say that
		$\Theta$
	is a \GF\ trajectory
		for the objective function $\loss$
		with generalized gradient $\gradient$
		and initial value $\s$
	(we say that 
		$\Theta$
	is a \GF\ trajectory
		for the objective function $\loss$
		with initial value $\s$,
	we say that
			$\Theta$
	is a solution of the \GF\ \ODE\
		for the objective function $\loss$
		with generalized gradient $\gradient$
		and initial value $\s$,
	we say that
			$\Theta$
	is a solution of the \GF\ \ODE\
		for the objective function $\loss$
		with initial value $\s$)
	if and only if it holds that
		$\Theta\colon [0,\infty)\to\R^\d$
		is a continuous function from $[0,\infty)$ to $\R^\d$
	which satisfies for all
		$t\in[0,\infty)$
	that
	$
		\int_0^t 
			\Pnorm2{\gradient(\Theta_s)}
		\,\diff s
	<
		\infty
	$
	and
	\begin{equation}
		\llabel{eq:gf}
		\Theta_t 
		= 
		\xi-\int_0^t \gradient(\Theta_s)\,\diff s
	\end{equation}
	\cfload.
\end{athm}
\endgroup

\subsection{Direction of negative gradients}

\cfclear
\begingroup
\providecommand{\d}{}
\renewcommand{\d}{\defaultParamDim}
\providecommand{\f}{}
\renewcommand{\f}{\defaultLossFunction}
\providecommand{\g}{}
\renewcommand{\g}{\defaultGradientFunction}
\begin{athm}{lemma}{lem:steepest_descent}
	Let 
		$\d\in\N$,
		$\f\in C^1(\R^\d,\R)$,
		$\altpoint\in\R^\d$,
		$r\in(0,\infty)$
	and let
		$\g\colon\R^\d\to\R$
	satisfy for all
		$\altpointTwo\in\R^\d$
	that
	\begin{equation}
		\llabel{gdef}
		\g(\altpointTwo)
		=
		\lim_{h\to 0}\pr*{\frac{\f(\altpoint+h\altpointTwo)-\f(\altpoint)}{h}}
		=
		[\f'(\altpoint)](\altpointTwo)
		.
	\end{equation}
	Then
	\begin{enumerate}[(i)]
		\item \llabel{it:1}
		it holds that
		\begin{equation}
			\sup_{\altpointTwo\in\{\altpointThree\in \R^\d\colon \Pnorm2\altpointThree=r\}} \g(\altpointTwo)
			=
			r\Pnorm2{(\nabla\f)(\altpoint)}
			=
			\begin{cases}
				0 & \colon (\nabla\f)(\altpoint)=0 \\
				\g\bpr{\tfrac{r(\nabla\f)(\altpoint)}{\Pnorm2{(\nabla\f)(\altpoint)}}} & \colon (\nabla\f)(\altpoint)\neq 0
			\end{cases}
		\end{equation}
		and
		\item \llabel{it:2}
		it holds that
		\begin{equation}
			\inf_{\altpointTwo\in\{\altpointThree\in \R^\d\colon \Pnorm2\altpointThree=r\}} \g(\altpointTwo)
			=
			-r\Pnorm2{(\nabla\f)(\altpoint)}
			=
			\begin{cases}
				0 & \colon (\nabla\f)(\altpoint)=0 \\
				\g\bpr{\tfrac{-r(\nabla\f)(\altpoint)}{\Pnorm2{(\nabla\f)(\altpoint)}}} & \colon (\nabla\f)(\altpoint)\neq 0\ifnocf.
			\end{cases}
		\end{equation}
	\end{enumerate}
	\cfout[.]
\end{athm}
\begin{aproof}
	Note that 
		\lref{gdef}
	implies that for all
  	$\altpointTwo\in\R^\d$
	it holds that
	\begin{equation}
		\llabel{eq:1}
		\g(\altpointTwo)
		=
		\scp{(\nabla\f)(\altpoint),\altpointTwo}
	\end{equation}
	\cfload.
		The Cauchy--Schwarz inequality
		\hence
	ensures that for all
		$\altpointTwo\in\R^\d$
		with $\Pnorm2{\altpointTwo}=r$
	it holds that
	\begin{eqsplit}
		\llabel{eq:2}
		&
		{-}r\Pnorm2{(\nabla\f)(\altpoint)}
		=
		-\Pnorm2{(\nabla\f)(\altpoint)}\Pnorm2{\altpointTwo}
		\leq
		-\scp{-(\nabla\f)(\altpoint),\altpointTwo}
		\\&=
		\g(\altpointTwo)
		\leq
		\Pnorm2{(\nabla\f)(\altpoint)}\Pnorm2{\altpointTwo}
		=
		r\Pnorm2{(\nabla\f)(\altpoint)}
	\end{eqsplit}
	\cfload.
	\Moreover
		\lref{eq:1}
	shows that for all
		$c\in\R$
	it holds that
	\begin{equation}
		\g(c(\nabla\f)(\altpoint))
		=
		\scp{(\nabla\f)(\altpoint),c(\nabla\f)(\altpoint)}
		=
		c\Pnorm2{(\nabla\f)(\altpoint)}^2
		.
	\end{equation}
	Combining
		this
		and \lref{eq:2}
	proves
		\lref{it:1} and \lref{it:2}.
\end{aproof}
\endgroup

\cfclear
\begingroup
\providecommand{\d}{}
\renewcommand{\d}{\defaultParamDim}
\providecommand{\f}{}
\renewcommand{\f}{\defaultLossFunction}
\begin{athm}{lemma}{prop:gf:chainrule}
	Let 
	$ \d \in \N $, 
	$ \Theta \in C( [0,\infty), \R^{ \d } ) $, 
	$ \f \in C^1( \R^{ \d }, \R ) $
	and assume
	for all $ t \in [0,\infty) $ 
	that
	$
  	\Theta_t = \Theta_0 - \int_0^t (\nabla\f)( \Theta_s ) \, \diff s
	$
	\cfload. 
	Then 
	\begin{enumerate}[(i)]
		\item \llabel{it:0}
		it holds that 
			$\Theta \in C^1( [0,\infty), \R^{ \d } ) $,
		\item \llabel{it:1}
		it holds for all
			$t\in[0,\infty)$
		that 
		$
			\dot\Theta_t
			=
			- (\nabla\f)( \Theta_t )
		$,
		and
		\item \llabel{it:2}
		it holds
		for all $ t \in [0,\infty) $ 
		that
		\begin{equation}
			\f( \Theta_t ) = \f( \Theta_0 ) - \int_0^t \pnorm2{ (\nabla\f)( \Theta_s ) }^2 \, \diff s
		\end{equation}
	\end{enumerate}
	\cfout.
\end{athm}

\begin{aproof}
\Nobs that 
	the fundamental theorem of calculus 
implies
	\lref{it:0} and \lref{it:1}.
Combining 
	\lref{it:1}
with 
	the fundamental theorem of calculus
	and the chain rule
ensures that for all
	$t \in [0,\infty)$
it holds that
\begin{equation}
\label{prop:gf:chainrule:eq2}
\begin{split} 
	\f( \Theta_t )
=
	\f( \Theta_0 ) + \int_0^t \scp{(\nabla \f)(\Theta_s), \dot\Theta_s} \, \diff s
=
	\f( \Theta_0 ) - \int_0^t \pnorm2{ (\nabla\f)( \Theta_s ) }^2 \, \diff s
\end{split}
\end{equation}
\cfload.
This establishes \lref{it:2}.
\end{aproof}
\endgroup

\cfclear
\begingroup
\providecommand{\d}{}
\renewcommand{\d}{\defaultParamDim}
\providecommand{\f}{}
\renewcommand{\f}{\defaultLossFunction}
\providecommand{\g}{}
\renewcommand{\g}{\defaultGradientFunction}
\begin{athm}{cor}{cor:steepest_descent2}[Illustration for the negative \GF]
	Let 
		$ \d \in \N $, 
		$ \Theta \in C( [0,\infty), \R^{ \d } ) $, 
		$ \f \in C^1( \R^{ \d }, \R ) $
	and assume
		for all 
			$ t \in [0,\infty) $ 
		that
		$\Theta(t) = \Theta(0) - \int_0^t (\nabla\f)( \Theta(s) ) \, \diff s$
	\cfload. 
	Then 
	\begin{enumerate}[(i)]
		\item \llabel{it:0}
		it holds that
		$\Theta \in C^1( [0,\infty), \R^{ \d } ) $,
		\item \llabel{it:1}
		it holds for all
			$t\in(0,\infty)$
		that
		\begin{equation}
			(\f\circ\Theta)'(t)
			=
			-\Pnorm2{(\nabla\f)(\Theta(t))}^2,
		\end{equation}
		and
		\item\llabel{it:2}
		it holds for all
			$\Xi\in C^1( [0,\infty), \R^{ \d } )$,
			$\tau\in(0,\infty)$
			with
			  $\Xi(\tau)=\Theta(\tau)$
				and $\Pnorm2{\Xi'(\tau)}=\Pnorm2{\Theta'(\tau)}$
		that
		\begin{equation}
			(\f\circ \Theta)'(\tau)
			\leq
			(\f\circ \Xi)'(\tau)
		\end{equation}
	\end{enumerate}
	\cfout.
\end{athm}
\begin{aproof}
	\Nobs that
		\cref{prop:gf:chainrule}
		and the fundamental theorem of calculus
	imply \lref{it:0,it:1}.
	\Nobs that
		\cref{lem:steepest_descent}
	shows for all
		$\Xi\in C^1( [0,\infty), \R^{ \d } )$,
		$t\in(0,\infty)$
	it holds that
	\begin{eqsplit}
		(\f\circ\Xi)'(t)
		&=
		[\f'(\Xi(t))](\Xi'(t))
		\\&\geq
		\inf_{v\in\{w\in\R^\d\colon \Pnorm2{w}=\Pnorm2{\Xi'(t)}\}} [\f'(\Xi(t))](v)
		\\&=
		-\Pnorm2{\Xi'(t)}\Pnorm2{(\nabla\f)(\Xi(t))}
	\end{eqsplit}
	\cfload.
		\Cref{prop:gf:chainrule}
		\hence
	ensures that for all
		$\Xi\in C^1( [0,\infty), \R^{ \d } )$,
		$\tau\in(0,\infty)$
		with
			$\Xi(\tau)=\Theta(\tau)$
			and $\Pnorm2{\Xi'(\tau)}=\Pnorm2{\Theta'(\tau)}$
	it holds that
	\begin{equation}
	\begin{split}
		(\f\circ\Xi)'(\tau)
		&\geq
		-\Pnorm2{\Xi'(\tau)}\Pnorm2{(\nabla\f)(\Xi(\tau))}
		\geq
		-\Pnorm2{\Theta'(\tau)}\Pnorm2{(\nabla\f)(\Theta(\tau))}\\
		&=
		-\Pnorm2{(\nabla\f)(\Theta(\tau))}^2
		=
		(\f \circ \Theta)'(\tau).
	\end{split}
	\end{equation}
		This
	establishes
		\lref{it:2}.
\end{aproof}
\endgroup

\begin{minipage}{0.93\linewidth}
		\begin{center}
			\IfFileExists{plots/gradient_plot1.pdf}{\includegraphics[width=\linewidth]{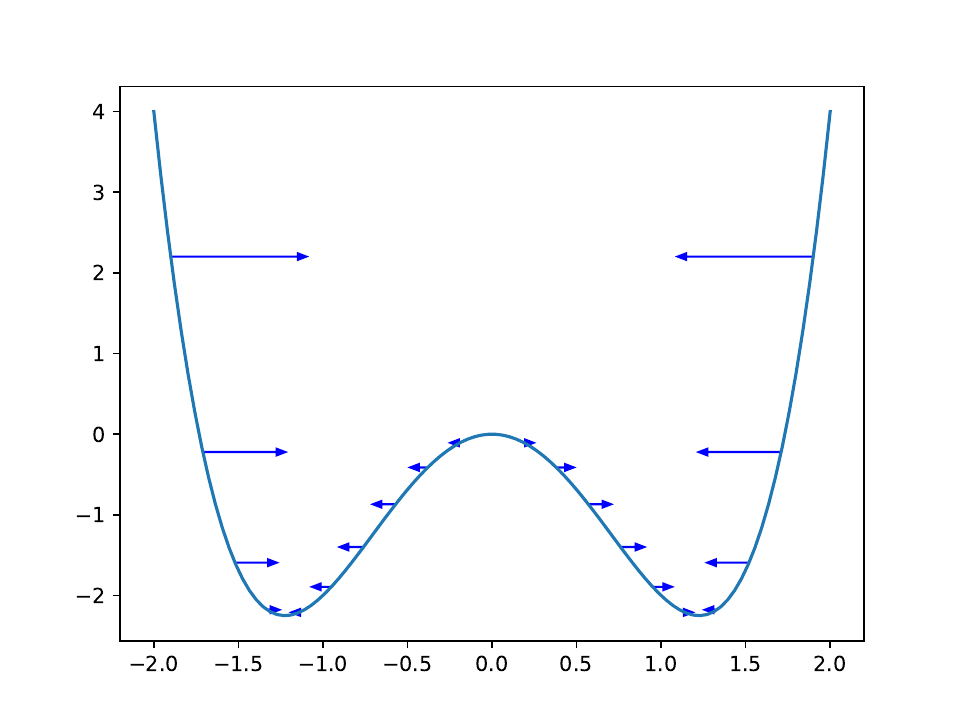}}{}
		\captionof{figure}{\label{fig:gradientgraph1}
		Illustration of negative gradients in a one-dimensional example.
		The plot shows the graph of the function $[-2,2]\ni x\mapsto x^4-3x^2\in\R$
		with the value of the negative gradient, scaled by $\tfrac{1}{20}$,
		indicated by horizontal arrows at several points.
		The {\sc Python} code used to produce this plot is given in 
		\cref{code:gradient_graph1}.
		}
		\end{center}
	\end{minipage}
	
	\begin{minipage}{0.93\linewidth}
		\begin{center}
			\IfFileExists{plots/gradient_plot2.pdf}{\includegraphics[width=\linewidth]{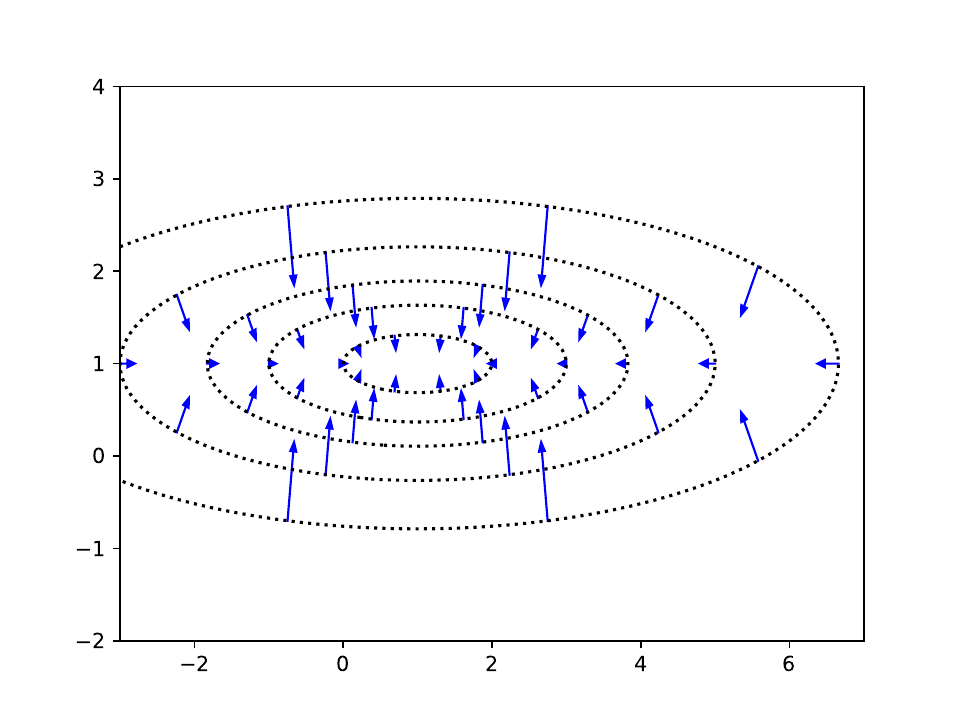}}{}
		\captionof{figure}{\label{fig:gradientgraph2}
		Illustration of negative gradients in a two-dimensional example.
		The plot shows contour lines of the function 
		$\R^2\ni(x,y)\mapsto \tfrac12\abs{x-1}^2+5\abs{y-1}^2\in\R$
		with arrows indicating the direction and magnitude, scaled by $\tfrac{1}{20}$,
		of the negative gradient at several points along these
		contour lines.
		The {\sc Python} code used to produce this plot is given in 
		\cref{code:gradient_graph2}.
		}
		\end{center}
	\end{minipage}
	\bigskip
	
\filelisting{code:gradient_graph1}{code/gradient_plot1.py}{{\sc Python} code used to create \cref{fig:gradientgraph1}}

\filelisting{code:gradient_graph2}{code/gradient_plot2.py}{{\sc Python} code used to create \cref{fig:gradientgraph2}}

\section{Regularity properties for ANNs}

\subsection{On the differentiability of compositions of parametric functions}

\cfclear
\begingroup 
\providecommand{\d}{}
\renewcommand{\d}{\defaultParamDim}
\providecommand{\H}{}
\renewcommand{\H}{f}
\providecommand{\F}{}
\renewcommand{\F}{F}
\providecommand{\A}{}
\renewcommand{\A}{A}
\providecommand{\B}{}
\renewcommand{\B}{B}
\begin{athm}{lemma}{stacking_diff}
Let
	$\d_1, \d_2, l_1, l_2 \in \N$,
let 
$\A_1 \colon \R^{l_1} \to \R^{l_1} \times \R^{l_2}$ and 
$\A_2 \colon \R^{l_2} \to \R^{l_1} \times \R^{l_2}$
satisfy for all 
	$x_1 \in \R^{l_1}$,
	$x_2 \in \R^{l_2}$
that
$
	\A_1(x_1) = (x_1, 0)
$
and
$
	\A_2(x_2) = (0, x_2)
$,
for every 
	$k \in \{1, 2\}$ 
let $\B_k \colon \R^{l_1} \times \R^{l_2} \to \R^{l_k}$ satisfy for all 
	$x_1 \in \R^{l_1}$,
	$x_2 \in \R^{l_2}$
that
$\B_k(x_1, x_2) = x_k$,
for every 
	$k \in \{1, 2\}$
let
$
	\F_k
\colon 
	\R^{\d_k} \to \R^{l_k}
$
be  differentiable,
and let $\H \colon \R^{\d_1} \times \R^{\d_2} \to \R^{l_1} \times \R^{l_2}$ satisfy for all
	$x_1 \in \R^{\d_1}$,
	$x_2 \in \R^{\d_2}$
that
\begin{equation}
\llabel{ass1}
\begin{split} 
	\H(x_1, x_2)
=
	(
		\F_1(x_1),
		\F_2(x_2)
	).
\end{split}
\end{equation}
Then
\begin{enumerate}[(i)]
	\item \llabel{it:repr}
	it holds that
	$\H 
	=
		\A_1 \circ \F_1 \circ \B_1 + \A_2 \circ \F_2 \circ \B_2$
	and
	\item \llabel{it:diff}
	it holds that
	$\H%
	$ is differentiable.
\end{enumerate}
\end{athm}

\begin{aproof}
	\Nobs that
		\lref{ass1}
	implies that for all
		$x_1\in\R^{\d_1}$,
		$x_2\in\R^{\d_2}$
	it holds that
	\begin{eqsplit}
		(\A_1 \circ \F_1 \circ \B_1 + \A_2 \circ \F_2 \circ \B_2)(x_1,x_2)
		&=
		(\A_1\circ \F_1)(x_1) + (\A_2\circ \F_2)(x_2)
		\\&=
		(\F_1(x_1),0) + (0,\F_2(x_2))
		\\&=
		(\F_1(x_1),\F_2(x_2))
		.
	\end{eqsplit} 
Combining 
	this and
	the fact that $\A_1$, $\A_2$, $\F_1$, $\F_2$, $\B_1$, and $\B_2$ are differentiable
with the chain rule establishes that $\H$ is differentiable.
\end{aproof}
\endgroup

\begingroup
\providecommand{\d}{}
\renewcommand{\d}{\defaultParamDim}
\providecommand{\H}{}
\renewcommand{\H}{f}
\providecommand{\F}{}
\renewcommand{\F}{F}
\providecommand{\A}{}
\renewcommand{\A}{A}
\providecommand{\B}{}
\renewcommand{\B}{B}
\begin{athm}{lemma}{param_comp_diff}
Let
	$\d_1, \d_2, l_0, l_1, l_2 \in \N$,
	let
	$\A \colon \R^{\d_1} \times \R^{\d_{2}} \times \R^{l_{0}} \to \R^{\d_{2}} \times \R^{\d_1 + l_0} $ and
	$\B \colon \R^{\d_{2}} \times \R^{\d_1 + l_0} \to \R^{\d_2} \times \R^{l_1}$
satisfy for all
	$\theta_1 \in \R^{\d_1}$,
	$\theta_2 \in \R^{\d_2}$,
	$x \in \R^{l_0}$
that
\begin{equation}
\label{param_comp_diff:setting1}
\begin{split} 
	\A(\theta_1, \theta_2, x)
=
	(\theta_2, (\theta_1, x))
\qandq
	\B(\theta_2, (\theta_1, x))
=
	(
		\theta_2,
		\F_1(\theta_1, x)
	),
\end{split}
\end{equation}
for every 
	$k \in \{1, 2\}$
let
$
	\F_k 
\colon 
	\R^{\d_k} \times \R^{l_{k-1}} \to \R^{l_k}
$
be  differentiable,
and
let
$
	\H
\colon 
	\R^{\d_1} \times \R^{\d_{2}} \times \R^{l_{0}} 
\to 
	\R^{l_2}
$
satisfy for all
	$\theta_1 \in \R^{\d_1}$,
	$\theta_2 \in \R^{\d_2}$,
	$x \in \R^{l_0}$
that
\begin{equation}
\label{param_comp_diff:ass1}
\begin{split} 
	\H(\theta_1, \theta_2, x)
=
	\bpr{
		\F_2(\theta_2, \cdot) \circ 
		\F_{1}(\theta_{1}, \cdot)
	} 
	(x).
\end{split}
\end{equation}
Then 
\begin{enumerate}[(i)]
	\item \llabel{it:repr}
	it holds that
	$\H=\F_2\circ\B\circ\A$
	and
	\item
	it holds that
	$\H%
	$ is differentiable.
\end{enumerate}
\end{athm}

\begin{aproof}
\Nobs that 
\enum{
	\eqref{param_comp_diff:setting1};
	\eqref{param_comp_diff:ass1};
}[ensure]
that for all
	$\theta_1 \in \R^{\d_1}$,
	$\theta_2 \in \R^{\d_2}$,
	$x \in \R^{l_0}$
it holds that 
\begin{equation}
\label{param_comp_diff:eq1}
\begin{split} 
	\H(\theta_1, \theta_2, x)
=
	\F_2(\theta_2, \F_{1}(\theta_{1}, x)) 
=
	\F_2(\B(\theta_2, (\theta_{1}, x)))
=
	\F_2(\B(\A(\theta_{1}, \theta_2, x))).
\end{split}
\end{equation}
\Nobs that \cref{stacking_diff} 
(applied with 
$\d_1 \is \d_2$,
$\d_2 \is \d_1 + l_1$,
$l_1 \is \d_2$,
$l_2 \is l_1$,
$\F_1 \is (\R^{\d_2} \ni \theta_2 \mapsto \theta_2 \in \R^{\d_2})$,
$\F_2 \is (\R^{\d_1 + l_1} \ni (\theta_1, x) \mapsto \F_1(\theta_1, x) \in \R^{l_1})$
in the notation of \cref{stacking_diff})
implies that $\B$ is differentiable.
Combining 
\enum{
	this;
	the fact that $\A$ is differentiable;
	the fact that $\F_2$ is differentiable;
	\eqref{param_comp_diff:eq1}
}
with
the chain rule
assures that $\H$ is differentiable.
\end{aproof}
\endgroup

\subsection{On the differentiability of realizations of ANNs}
\label{sec:diff_anns}

\cfclear
\begingroup
\renewcommand{\d}{\mathfrak d}
\begin{athm}{lemma}{lem:differentiability_ANNs}[Differentiability of realization functions of \anns]
	Let 
		$L \in \N$, 
		$l_0,l_1,\ldots, \allowbreak l_L \in \N$,
	for every
		$k\in \{1,2,\dots,L\}$
		let
			$\d_k=l_k(l_{k-1}+1)$,
	for every
		$k\in \{1,2,\dots,L\}$
		let 
			$\Psi_k \colon \R^{l_k} \to \R^{l_k}$
			be differentiable,
	and for every
		$k\in \{1,2,\dots,L\}$
		let 
			$F_k\colon \R^{\d_k}\times\R^{l_{k-1}}\to\R^{l_k}$
		satisfy for all
			$\theta\in\R^{\d_k}$,
			$x\in \R^{l_{k-1}}$
		that
		\begin{equation}
			F_k(\theta,x)
			=
			\Psi_k\bpr{\Aff_{l_k,l_{k-1}}^{\theta,0}(x)}
		\end{equation}
	\cfload.
	Then
	\begin{enumerate}[(i)]
		\item \llabel{it:repr}
		it holds for all
			$\theta_1\in\R^{\d_1}$,
			$\theta_2\in\R^{\d_2}$,
			$\ldots$,
			$\theta_L\in\R^{\d_L}$,
			$x\in\R^{l_0}$
		that
		\begin{equation}
			\bpr{\RealV{(\theta_1,\theta_2,\dots,\theta_L)}{0}{l_0}{\Psi_1,\Psi_2,\dots,\Psi_L}}(x)
			=
			(F_L(\theta_L,\cdot)\circ F_{L-1}(\theta_{L-1},\cdot)\circ\ldots\circ F_1(\theta_1,\cdot))(x)
		\end{equation}
		and
		\item \llabel{it:diff}
		it holds that
		\begin{equation}
			\R^{\d_1+\d_2+\ldots+\d_L}\times\R^{l_0}\ni(\theta,x)\mapsto 
			\bpr{\RealV \theta0{l_0}{\Psi_1,\Psi_2,\dots,\Psi_L}}(x)\in\R^{l_L}
		\end{equation}
		is differentiable
	\end{enumerate}
	\cfout.
\end{athm}
\begin{aproof}
	\Nobs that
		\cref{def:affine:eq1}
	shows that for all
		$\theta_1\in\R^{\d_1}$,
		$\theta_2\in\R^{\d_2}$,
		$\ldots$,
		$\theta_L\in\R^{\d_L}$,
		$k\in \{1,2,\dots,L\}$
	it holds that
	\begin{equation}
		\Aff^{(\theta_1,\theta_2,\dots,\theta_L),\sum_{j=1}^{k-1}\d_j}_{l_k,l_{k-1}}
		=
		\Aff^{\theta_k,0}_{l_k,l_{k-1}}
		.
	\end{equation}
	\Hence that for all
		$\theta_1\in\R^{\d_1}$,
		$\theta_2\in\R^{\d_2}$,
		$\ldots$,
		$\theta_L\in\R^{\d_L}$,
		$k\in \{1,2,\dots,L\}$
	it holds that
	\begin{equation}
		F_k(\theta_k,x)
		=
		\bpr{\Psi_k\circ\Aff^{(\theta_1,\theta_2,\dots,\theta_L),\sum_{j=1}^{k-1}\d_j}_{l_k,l_{k-1}}}(x)
		.
	\end{equation}
	Combining
		this
	with
		\cref{eq:FFNN}
	establishes
		\lref{it:repr}.
	\Nobs that
		the assumption that
			for all
				$k\in \{1,2,\dots,L\}$
			it holds that
				$\Psi_k$ is differentiable,
		the fact that
			for all
				$m,n\in\N$,
				$\theta\in\R^{m(n+1)}$
			it holds that
				$\R^{m(n+1)}\times\R^n\ni (\theta,x)\mapsto \Aff^{\theta,0}_{m,n}(x)\in\R^m$
				is differentiable,
		and the chain rule
	ensure that for all
		$k\in \{1,2,\dots,L\}$
	it holds that
		$F_k$ is differentiable.
		\Cref{param_comp_diff}
		and induction
		hence
	prove that
	\begin{multline}
		\R^{\d_1}\times\R^{\d_2}\times\ldots\times\R^{\d_L}\times\R^{l_0}
		\ni
		(\theta_1,\theta_2,\dots,\theta_L,x)
		\\\mapsto
		(F_L(\theta_L,\cdot)\circ F_{L-1}(\theta_{L-1},\cdot)\circ\ldots\circ F_1(\theta_1,\cdot))(x)
		\in\R^{l_L}
	\end{multline}
	is differentiable.
		This
		and \lref{it:repr}
	prove
		\lref{it:diff}.
\end{aproof}
\endgroup

\cfclear
\begingroup
\renewcommand{\d}{\mathfrak d}
\renewcommand{\loss}{\mathbf{L}}
\newcommand{\Loss}{\mathscr{L}}
\newcommand{\x}[1]{x_{#1}}
\newcommand{\y}[1]{y_{#1}}
\begin{athm}{lemma}{lem:differentiability_loss}[Differentiability of the empirical risk function]
	Let
		$L,\d\in\N\backslash\{1\}$,
		$M,l_0,\allowbreak l_1,\dots,\allowbreak l_L\in\N$,
		$\x 1,\x 2,\dots,\x M\in \R^{l_0}$,
		$\y 1,\y 2,\dots,\allowbreak \y M\in \R^{l_L}$
	satisfy
		$\d=\sum_{k=1}^{L}l_k(l_{k-1}+1)$,
	for every
		$k\in \{1,2,\dots,L\}$
		let 
			$\Psi_k \colon \R^{l_k} \to \R^{l_k}$
			be differentiable,
	and let
		$\Loss\colon\R^\d\to\R$
	satisfy for all
		$\theta\in\R^\d$ that
	\begin{equation}
		\llabel{eq:defLoss}
		\Loss(\theta)
		=
		\frac1M\br*{\sum_{m=1}^{M}\loss\bpr{\bpr{\RealV \theta0{l_0}{
			\Psi_1 , 
			\Psi_2 , 
			\dots, 
			\Psi_L
			}
		}(\x m),\y m}}
	\end{equation}
	\cfload.
	Then $\Loss$ is differentiable.
\end{athm}
\begin{aproof}
	\Nobs that
		\cref{lem:differentiability_ANNs}
		and \cref{stacking_diff}
		(applied with
			$\mathfrak d_1\is \d+l_0$,
			$\mathfrak d_2\is l_L$,
			$l_1\is l_L$,
			$l_2\is l_L$,
			$F_1\is(\R^{\d}\times\R^{l_0}\ni (\theta,x)\mapsto \bpr{\RealV \theta0{l_0}{\Psi_1 , 
			\Psi_2 , 
			\dots, 
			\Psi_L}}(x)\in\R^{l_L})$,
			$F_2\is\id_{\R^{l_L}}$
		in the notation of \cref{stacking_diff})
	\prove that
	\begin{equation}
		\R^\d\times\R^{l_0}\times\R^{l_L}
		\ni
		(\theta,x,y)
		\mapsto
		\bpr{\bpr{\RealV \theta0{l_0}{\Psi_1 , 
		\Psi_2 , 
		\dots, 
		\Psi_L}}(x),y}
		\in
		\R^{l_L}\times \R^{l_L}
	\end{equation}
	is differentiable.
		The assumption that
			$\loss$ is differentiable
		and the chain rule
		\hence
	\prove that for all
		$x\in\R^{l_0}$,
		$y\in\R^{l_L}$
	it holds that
	\begin{equation}
		\R^\d\ni\theta\mapsto
		\loss\bpr{\bpr{\RealV \theta0{l_0}{\Psi_1 , 
		\Psi_2 , 
		\dots, 
		\Psi_L}}(\x m),\y m}
		\in\R
	\end{equation}
	is differentiable.
		This
	\proves that
		$\mathscr L$ is differentiable.
\end{aproof}
\endgroup

\cfclear
\begingroup
\begin{athm}{lemma}{lem:multdim_differentiable}
	Let $a\colon\R\to\R$ be differentiable and let
	$d\in \N$.
	Then $\multdim_{a,d}$ is differentiable
	\cfload.
\end{athm}
\begin{aproof}
	\Nobs that
		the assumption that 
			$a$ is differentiable,
		\cref{stacking_diff},
		and induction
	\prove that for all
		$m\in\N$
	it holds that
		$\multdim_{a,m}$
	is differentiable.
\end{aproof}
\endgroup

\cfclear
\begingroup
\renewcommand{\d}{\mathfrak d}
\renewcommand{\loss}{\mathbf{L}}
\newcommand{\Loss}{\mathscr{L}}
\newcommand{\x}[1]{x_{#1}}
\newcommand{\y}[1]{y_{#1}}
\begin{athm}{cor}{lem:differentiability_loss2}
	Let
		$L,\d\in\N\backslash\{1\}$,
		$M,l_0,\allowbreak l_1,\dots,\allowbreak l_L\in\N$,
		$\x 1,\x 2,\dots,\x M\in \R^{l_0}$,
		$\y 1,\y 2,\allowbreak\dots,\allowbreak \y M\in \R^{l_L}$
	satisfy
		$\d=\sum_{k=1}^{L}l_k(l_{k-1}+1)$,
	let
		$a\colon\R\to\R$
		and $\loss\colon\R^{l_L}\times\R^{l_L}\to\R$
		be differentiable,
	and let
		$\Loss\colon\R^\d\to\R$
	satisfy for all
		$\theta\in\R^\d$ that
	\begin{equation}
		\llabel{eq:defLoss}
		\Loss(\theta)
		=
		\frac1M\br*{\sum_{m=1}^{M}\loss\bpr{\bpr{\RealV \theta0{l_0}{\multdim_{ a, l_1 } , 
		\multdim_{ a, l_2 } , 
		\dots, 
		\multdim_{ a, l_{ L - 1 } } , 
		\operatorname{id}_{ \R^{ l_L }}}\!}(\x m),\y m}}
	\end{equation}
	\cfload.
	Then $\Loss$ is differentiable.
\end{athm}
\begin{aproof}
	\Nobs that
		\cref{lem:multdim_differentiable},
		and \cref{lem:differentiability_loss}
	\prove that
		$\mathscr L$ is differentiable.
\end{aproof}
\endgroup

\cfclear
\begingroup
\renewcommand{\d}{\mathfrak d}
\renewcommand{\loss}{\mathbf{L}}
\newcommand{\Loss}{\mathscr{L}}
\newcommand{\x}[1]{x_{#1}}
\newcommand{\y}[1]{y_{#1}}
\begin{athm}{cor}{lem:differentiability_loss3}
	Let
		$L,\d\in\N\backslash\{1\}$,
		$M,l_0,\allowbreak l_1,\dots,\allowbreak l_L\in\N$,
		$\x 1,\x 2,\dots,\x M\in \R^{l_0}$,
		$\y 1,\y 2,\allowbreak\dots,\allowbreak \y M\in (0,\infty)^{l_L}$
	satisfy
		$\d=\sum_{k=1}^{L}l_k(l_{k-1}+1)$,
	let $A$ be the $l_L$-dimensional softmax activation function,\cfadd{def:softmax}
	let
		$a\colon\R\to\R$
		and $\loss\colon(0,\infty)^{l_L}\times(0,\infty)^{l_L}\to\R$
		be differentiable,
	and let
		$\Loss\colon\R^\d\to\R$
	satisfy for all
		$\theta\in\R^\d$ that
	\begin{equation}
		\llabel{eq:defLoss}
		\cfadd{softmax_distribution}
		\Loss(\theta)
		=
		\frac1M\br*{\sum_{m=1}^{M}\loss\bpr{\bpr{\RealV \theta0{l_0}{\multdim_{ a, l_1 } , 
		\multdim_{ a, l_2 } , 
		\dots, 
		\multdim_{ a, l_{ L - 1 } } , 
		A}}(\x m),\y m}}
	\end{equation}
	\cfload.
	Then $\Loss$ is differentiable.
\end{athm}
\begin{aproof}
	\Nobs that
		\cref{lem:multdim_differentiable},
		the fact that
			$A$ is differentiable,
		and \cref{lem:differentiability_loss}
	\prove that
		$\mathscr L$ is differentiable.
\end{aproof}
\endgroup

\section{Loss functions}
\begingroup
\renewcommand{\loss}{\mathbf L}

\subsection{Absolute error loss}

\begin{athm}{definition}{def:l1loss}
	Let $d\in\N$ and let $\opnorm\cdot\colon \R^d\to[0,\infty)$ be a norm. 
	Then we say that $\loss$ is the
	$\Lone$-error loss function based on $\opnorm{\cdot}$ 
	(we say that $\loss$ is the absolute error loss function based on
	$\opnorm{\cdot}$) if and only if it holds that
	$\loss\colon \R^d\times\R^d\to\R$ is the function from
	$\R^d\times \R^d$ to $\R$ which satisfies for all
	$x,y\in\R^d$ that
	\begin{equation}
		\loss(x,y) = \opnorm{x-y}
		.
	\end{equation}
\end{athm}

\begin{figure}[!ht]
	\centering
	\includegraphics[width=0.5\linewidth]{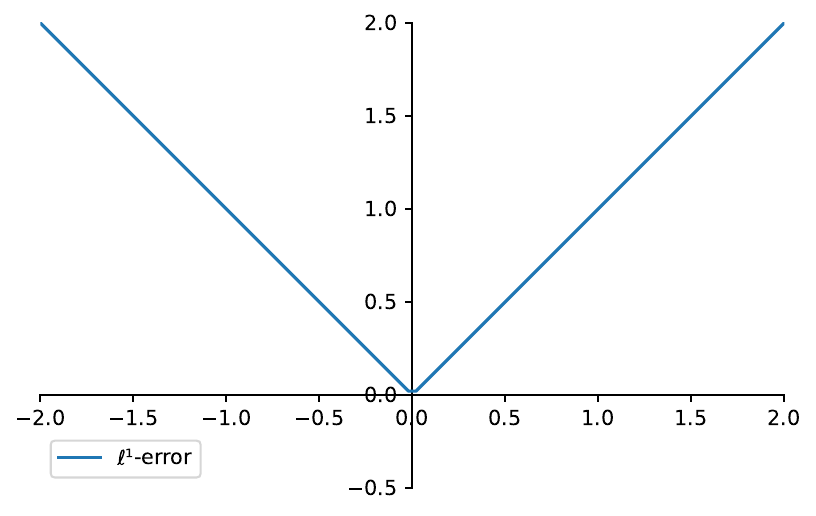}
	\caption{\cfclear\label{fig:l1loss_plot}A plot of the function
	$\R\ni x\mapsto \loss(x,0)\in[0,\infty)$ where
	$\loss$ is the $\Lone$-error loss function\cfadd{def:l1loss} based on
	$\R\ni x\mapsto \abs x\in[0,\infty)$
	\cfload.}
\end{figure}

\filelisting{code:l1loss_plot}{code/loss_functions/l1loss_plot.py}{{\sc Python} code used to create \cref{fig:l1loss_plot}}

\subsection{Mean squared error loss}
\label{sect:MSE}

\begin{athm}{definition}{def:mseloss}
	Let $d\in\N$ and let $\opnorm\cdot\colon \R^d\to[0,\infty)$ be a norm. 
	Then we say that $\loss$ is the 
	mean squared error loss function based on $\opnorm\cdot$ if and only if it holds that
	$\loss\colon \R^d\times\R^d\to\R$ is the function from
	$\R^d\times\R^d$ to $\R$ which satisfies for all
	$x,y\in\R^d$ that
	\begin{equation}
		\llabel{eq}
		\loss(x,y) = \opnorm{x-y}^2
		.
	\end{equation}
\end{athm}

\begin{figure}[!ht]
	\centering
	\includegraphics[width=0.5\linewidth]{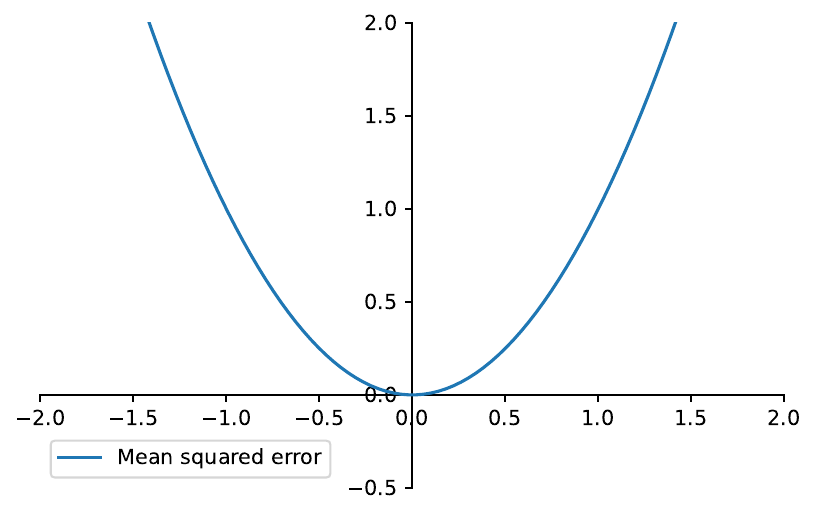}
	\caption{\cfclear\label{fig:mseloss_plot}A plot of the function
	$\R\ni x\mapsto \loss(x,0)\in[0,\infty)$ where
	$\loss$ is the mean squared error\cfadd{def:mseloss} loss function based on
	$\R\ni x\mapsto \abs x\in[0,\infty)$
	\cfload.}
\end{figure}

\filelisting{code:mseloss_plot}{code/loss_functions/mseloss_plot.py}{{\sc Python} code used to create \cref{fig:mseloss_plot}}

\cfclear
\begin{athm}{lemma}{lem:mseloss_analytic}
	Let $d\in\N$ and let $\loss$ be the
	mean squared error\cfadd{def:mseloss} loss function
	based on $\R^d\ni x\mapsto\Pnorm2 x\in[0,\infty)$
	\cfload.
	Then
	\begin{enumerate}[(i)]
		\item \llabel{it:1}
		it holds that 
			$\loss\in C^\infty(\R^d\times\R^d,\R)$
		\item \llabel{it:2}
		it holds for all
			$x,y,u,v\in\R^d$
		that
		\begin{equation}
			\llabel{claim}
			\loss(u,v)
			=
			\loss(x,y) + \loss'(x,y)(u-x,v-y) + \tfrac12\loss^{(2)}(x,y)\bpr{(u-x,v-y),(u-x,v-y)}
			.
		\end{equation}
	\end{enumerate}
\end{athm}
\begin{aproof}
	\Nobs that
		\cref{def:mseloss.eq}
	implies that for all
		$x=(x_1,\dots,x_d),\,\allowbreak y=(y_1,\dots,\allowbreak y_d)\in\R^d$
	it holds that
	\begin{equation}
		\loss(x,y)
		=
		\Pnorm2{x-y}^2
		=
		\scp{x-y,x-y}
		=
		\sum_{i=1}^{d}(x_i-y_i)^2
		.
	\end{equation}
	\Hence that	for all
		$x,y\in\R^d$
	it holds that
		$\loss\in C^1(\R^d\times\R^d,\R)$
		and
	\begin{equation}
		\llabel{eq:grad}
		(\nabla\loss)(x,y)
		=
		(2(x-y),-2(x-y))\in\R^{2d}
		.
	\end{equation}
		This
	implies that for all
		$x,y,h,k\in\R^d$
	it holds that
	\begin{equation}
		\loss'(x,y)(h,k)
		=
		\scp{2(x-y),h} + \scp{-2(x,y),k}
		=
		2\scp{x-y,h-k}
		.
	\end{equation}
	\Moreover
		\lref{eq:grad}
	implies that for all
		$x,y\in\R^d$
	it holds that
		$\loss\in C^2(\R^d\times\R^d,\R)$
	and
	\begin{equation}
		(\Hess_{(x,y)}\loss)
		=
		\begin{pmatrix}
			2\idMatrix_d&-2\idMatrix_d\\
			-2\idMatrix_d&2\idMatrix_d
		\end{pmatrix}
		.
	\end{equation}
	\Hence that for all
		$x,y,h,k\in\R^d$
	it holds that
	\begin{equation}
		\loss^{(2)}(x,y)\bpr{(h,k),(h,k)}
		=
		2\scp{h,h}-2\scp{h,k}-2\scp{k,h}+2\scp{k,k}
		=
		2\Pnorm2{h-k}^2
		.
	\end{equation}
	Combining
		this
	with
		\lref{eq:grad}
	shows that for all
		$x,y\in\R^d$,
		$h,k\in\R^d$
	it holds that
		$\loss\in C^\infty(\R^d\times\R^d,\R)$
	and
	\begin{eqsplit}
		&\loss(x,y) + \loss'(x,y)(h,k) + \tfrac12\loss^{(2)}(x,y)\bpr{(h,k),(h,k)}
		\\&=
		\Pnorm2{x-y}^2 + 2\scp{x-y,h-k} + \Pnorm2{h-k}^2
		\\&=
		\Pnorm2{x-y+(h-k)}^2
		\\&=
		\loss(x+h,y+k)
		.
	\end{eqsplit}
		This
	implies
		\cref{lem:mseloss_analytic.it:1,lem:mseloss_analytic.it:2}.
\end{aproof}

\subsection{Huber error loss}

\cfclear
\begin{athm}{definition}{def:huberloss}
	Let $d\in\N$, $\delta\in[0,\infty)$
	and let $\opnorm\cdot\colon \R^d\to[0,\infty)$ be a norm. 
	Then we say that $\loss$ is the 
	$\delta$-Huber-error loss function based on $\opnorm\cdot$ if and only if it holds that
	$\loss\colon \R^d\times\R^d\to\R$ is the function from
	$\R^d\times\R^d$ to $\R$ which satisfies for all
	$x,y\in\R^d$ that
	\begin{equation}
		\label{def:huberloss:eq1}
		\loss(x,y)
		= 
		\begin{cases}
			\frac12\opnorm{x-y}^2 &\colon \opnorm{x-y}\leq\delta\\
			\delta(\opnorm{x-y}-\frac\delta2) & \colon \opnorm{x-y}>\delta.
		\end{cases}
	\end{equation}
\end{athm}

\begingroup
\newcommand{\Huber}{\mathbf{H}}
\begin{athm}{lemma}{huber_loss_cont_I}
	Let $\delta\in[0,\infty)$ and
	let $\Huber \colon \R \to [0,\infty)$ 
	satisfy
	for all
		$z\in\R$ 
	that
	\begin{equation}
		\Huber(z) = 
		\begin{cases}
			\tfrac12z^2 &\colon z\leq\delta\\
			\delta(z-\tfrac\delta2) & \colon z>\delta.
		\end{cases}
	\end{equation}
	Then $\Huber$ is continuous.
\end{athm}
\begin{aproof}
	Throughout this proof, let  
	$f,g \in C(\R, \R)$ satisfy for all
		$z\in\R$ that
	\begin{equation}
		\llabel{eq1}
		f(z) = \tfrac12z^2
		\qandq
		g(z) = \delta(z-\tfrac\delta2)
		.
	\end{equation} 
	\Nobs that \lref{eq1} \proves that
	\begin{equation}
	\label{T_B_D}
	\begin{split}
		g(\delta) &= \delta(\delta-\tfrac\delta2)
		=
		\tfrac12\delta^2
		=
		f(\delta)
		.
	\end{split}
	\end{equation}
	Combining this with the fact that for all 
		$z \in \R$
	it holds that 
	\begin{equation}
	\label{T_B_D}
	\begin{split}
		\Huber(z) = 
		\begin{cases}
			f(z) &\colon z\leq\delta\\
			g(z) & \colon z>\delta
		\end{cases}
	\end{split}
	\end{equation}
	\proves[ep] that
		$\Huber$ is continuous.
\end{aproof}
\endgroup

\begingroup
\newcommand{\Huber}{\mathbf{H}}
\begin{athm}{cor}{huber_loss_cont_II}
	Let $d\in\N$, $\delta\in[0,\infty)$,
	let $\opnorm\cdot\colon \R^d\to[0,\infty)$ be a norm,
	and let 
	$\loss$ be the 
	$\delta$-Huber-error loss function based on $\opnorm\cdot$\cfadd{def:huberloss}
	\cfload.
	Then $\loss$ is continuous.
\end{athm}
\begin{aproof}
	Throughout this proof, let $\Huber \colon \R \to [0,\infty)$ satisfy for all
		$z\in\R$
	that
	\begin{equation}
		\Huber(z) = 
		\begin{cases}
			\tfrac12z^2 &\colon z\leq\delta\\
			\delta(z-\tfrac\delta2) & \colon z>\delta.
		\end{cases}
	\end{equation}
	\Nobs that \cref{def:huberloss:eq1} \proves that
	for all 
		$x,y\in\R^d$
	it holds that
	\begin{equation}
		\llabel{eq1}
		\loss(x,y) = \Huber(\opnorm{x-y})
		.
	\end{equation}
	\Moreover \cref{huber_loss_cont_I} \proves that
	$\Huber$ is continuous.
	Combining this 
	and the fact that 
	$(\R^d\times\R^d\ni(x,y)\mapsto\opnorm{x-y}\in\R)$
	is continuous
	with \lref{eq1} \proves that
	$\loss$ is continuous.
\end{aproof}
\endgroup

\begin{figure}[!ht]
	\centering
	\includegraphics[width=0.5\linewidth]{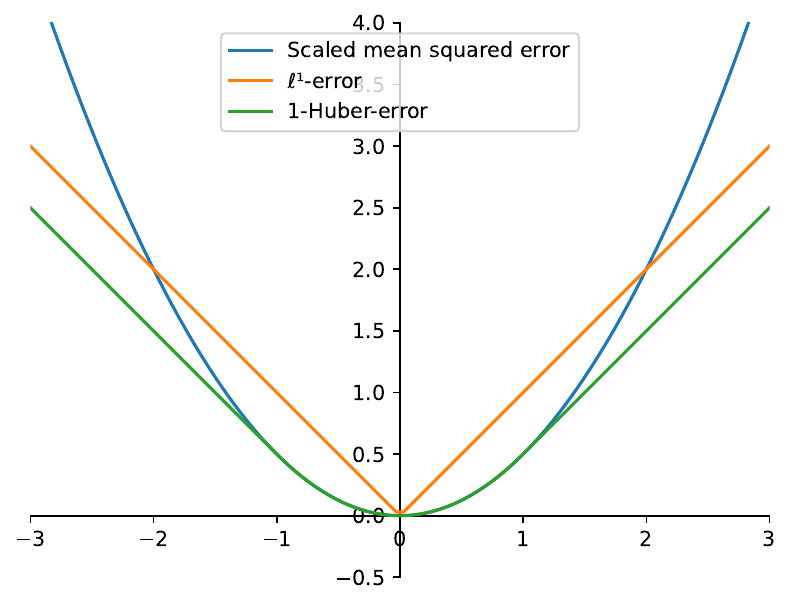}
	\caption{\cfclear\label{fig:huberloss_plot}A plot of the functions
	$\R\ni x\mapsto \loss_i(x,0)\in[0,\infty)$, $i\in\{1,2,3\}$, 
	where
	$\loss_0$ is the \cfadd{def:mseloss}mean squared error loss function based on
	$\R\ni x\mapsto \abs x\in[0,\infty)$,
	where 
	$\loss_1\colon\R^d\times\R^d\to[0,\infty)$ satisfies for all
	$x,y\in\R^d$ that $\loss_1(x,y)=\tfrac12\loss_0(x,y)$,
	where 
	$\loss_2$ is the $\Lone$-error \cfadd{def:l1loss}loss function based on
	$\R\ni x\mapsto \abs x\in[0,\infty)$, 
	and where 
	$\loss_3$ is the $1$-Huber\cfadd{def:huberloss} loss function based on
	$\R\ni x\mapsto \abs x\in[0,\infty)$.}
\end{figure}
\filelisting{code:huberloss_plot}{code/loss_functions/huberloss_plot.py}{{\sc Python} code used to create \cref{fig:huberloss_plot}}

\subsection{Cross-entropy loss}

\newcommand{\limitVariable}[1]{\mathbf{#1}}

\cfclear
\begin{athm}{definition}{def:crossentropyloss}
	Let $d\in\N$.
	Then we say that $\loss$ is the $d$-dimensional
	cross-entropy loss function if and only if it holds that
	$\loss\colon [0,\infty)^d\times[0,\infty)^d\to(-\infty,\infty]$ is the function from
	$[0,\infty)^d\times[0,\infty)^d$ to $(-\infty,\infty]$ which satisfies for all
	$x=(x_1,\dots,x_d),\,\allowbreak y=(y_1,\dots,y_d)\in[0,\infty)^d$ that
	\begin{equation}
		\llabel{eq}
		\loss(x,y)
		=
    -\sum_{i=1}^{d} \textstyle\lim_{\limitVariable{x}\searrow x_i}\bbr{ \ln(\limitVariable{x})y_i}
		.
	\end{equation}
\end{athm}

\begin{figure}[!ht]
	\centering
	\includegraphics[width=0.5\linewidth]{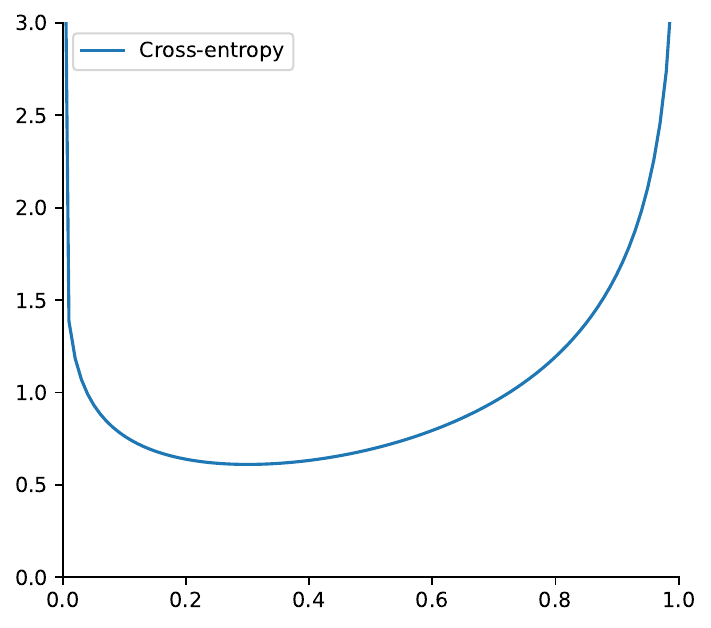}
	\caption{\cfclear\label{fig:crossentropyloss_plot}A plot of the function
	$(0,1)\ni x\mapsto \loss\bpr{(x,1-x),\bpr{\tfrac3{10},\tfrac7{10}}}\in\R$ where
	$\loss$ is the $2$-dimensional \cfadd{def:crossentropyloss}cross-entropy loss function
	\cfload.}
\end{figure}

\filelisting{code:crossentropyloss_plot}{code/loss_functions/crossentropyloss_plot.py}{{\sc Python} code used to create \cref{fig:crossentropyloss_plot}}

\cfclear
\begingroup
\begin{athm}{lemma}{lem:cel_basic2}
  Let
    $d\in\N$
  and let
    $\loss$ be the $d$-dimensional cross-entropy\cfadd{def:crossentropyloss} loss function
  \cfload.
  Then
  \begin{enumerate}[(i)]
    \item \llabel{it.1}
    it holds for all 
      $x=(x_1,\dots,x_d),\,\allowbreak y=(y_1,\dots,y_d)\in[0,\infty)^d$
    that
    \begin{equation}
      (\loss(x,y) = \infty)
			\;\leftrightarrow\;
			\bpr{\exists\,i\in\{1,2,\dots,d\}\colon [(x_i=0)\land (y_i\neq 0)]}
			,
    \end{equation}
    \item \llabel{it.2}
    it holds for all
      $x=(x_1,\dots,x_d),\,\allowbreak y=(y_1,\dots,y_d)\in[0,\infty)^d$
      with
        $\forall\, i\in \{1,2,\dots,d\}\colon\allowbreak\br{ (x_i\neq 0)\lor (y_i=0)}$
    that
    \begin{equation}
      \loss(x,y)
      =
      -\!\!\sum_{\substack{i\in \{1,2,\dots,d\},\\y_i\neq 0}} \!\!\ln(x_i)y_i
      \in\R
      ,
    \end{equation}
    and
    \item \llabel{it.3}
    it holds for all
      $x=(x_1,\dots,x_d)\in(0,\infty)^d$,
      $y=(y_1,\dots,y_d)\in[0,\infty)^d$
    that
    \begin{equation}
      \loss(x,y)
      =
      -\sum_{i=1}^d \ln(x_i)y_i
      \in\R
      .
    \end{equation}
  \end{enumerate}
\end{athm}
\begin{aproof}
  \Nobs that
    \cref{def:crossentropyloss.eq}
    and the fact that 
      for all
        $a,b\in [0,\infty)$
      it holds that
      \begin{equation}
        \lim_{\limitVariable{a}\searrow a}\bbr{\ln(\limitVariable{a})b}
        =
        \begin{cases}
          0 & \colon b=0\\
          \ln(a)b & \colon (a\neq 0)\land (b\neq 0)\\
          -\infty & \colon (a=0)\land (b\neq 0)
        \end{cases}
      \end{equation}
  \prove[ep]
    \cref{lem:cel_basic2.it.1,lem:cel_basic2.it.2,lem:cel_basic2.it.3}.
\end{aproof}
\endgroup

\cfclear
\begingroup
\providecommandordefault{\f}{f}
\providecommandordefault{\g}{g}
\begin{athm}{lemma}{lem:cel}
	Let $d\in\N$, 
	let $\loss$ be the $d$-dimensional \cfadd{def:crossentropyloss}cross-entropy loss	function, 
	let $x=(x_1,\dots,x_d),\,\allowbreak y=(y_1,\dots,y_d)\in[0,\infty)^d$ 
	satisfy $\sum_{i=1}^{d}x_i=\sum_{i=1}^{d}y_i$
	and $x\neq y$, 
	and let $\f\colon [0,1]\to(-\infty,\infty]$ satisfy for all
	$h\in[0,1]$ that 
	\begin{equation}
    \llabel{eq.deff}
		\f(h) = \loss(x+h(y-x),y)
	\end{equation}
	\cfload.
	Then $\f$ is strictly decreasing.
\end{athm}
\begin{aproof}
	Throughout this proof, let
		$\g\colon [0,1)\to(-\infty,\infty]$
	satisfy for all 
		$h\in[0,1)$
	that
	\begin{equation}
		\llabel{eq.defg}
		\g(h) = \f(1-h)
	\end{equation}
  and let
    $J=\{i\in\{1,2,\dots,d\}\colon y_i\neq 0\}$.
	\Nobs that
		\lref{eq.deff}
	\proves that for all
		$h\in[0,1)$
	it holds that
	\begin{equation}
    \llabel{eq.b1}
		\g(h)
		=
    \loss(x+(1-h)(y-x), y)
    =
    \loss(y+h(x-y), y)
		.
	\end{equation}
  \Moreover
    the fact that
      for all $i\in J$ it holds that $x_i\in[0,\infty)$ and $y_i\in(0,\infty)$
  \proves that for all
    $i\in J$,
    $h\in [0,1)$
  it holds that
  \begin{equation}
    y_i+h(x_i-y_i)
    =
    (1-h)y_i + hx_i
    \geq
    (1-h)y_i
    >
    0
    .
  \end{equation}
    This,
    \lref{eq.b1},
    and \cref{lem:cel_basic2.it.2} in \cref{lem:cel_basic2}
  \prove that for all
    $h\in[0,1)$
  it holds that
  \begin{equation}
    \llabel{eq.aa}
    \g(h)
    =
    -\sum_{i\in J}\ln(y_i+h(x_i-y_i)) y_i
    \in\R
    .
  \end{equation}
		The chain rule
		\hence 
	\proves that for all
		$h\in[0,1)$
	it holds that
		$([0,1)\ni z\mapsto g(z)\in\R)\in C^\infty([0,1),\R)$
	and
	\begin{equation}
		\llabel{eq.diff}
		\g'(h)
		=
		-\sum_{i\in J}\frac{y_i(x_i-y_i)}{y_i+h(x_i-y_i)}
		.
	\end{equation}
		This
		and the chain rule
	\prove that for all
		$h\in[0,1)$
	it holds that
	\begin{equation}
		\llabel{eq.1}
		\g''(h)
		=
		\sum_{i\in J}\frac{y_i(x_i-y_i)^2}{(y_i+h(x_i-y_i))^2}
		.
	\end{equation}
	\Moreover
		the fact that
			for all 
				$z=(z_1,\dots,z_d)\in[0,\infty)^d$
				with 
					$\sum_{i=1}^{d} z_i = \sum_{i=1}^{d} y_i$
					and $\forall\, i\in J\colon z_i=y_i$
			it holds that
			\begin{eqsplit}
				\sum_{i\in\{1,2,\dots,d\}\backslash J} z_i
				&=
        \bbbbbr{\sum_{i\in\{1,2,\dots,d\}} \!\!z_i} - \bbbbbr{\sum_{i\in J} z_i}
				\\&=
				\bbbbbr{\sum_{i\in \{1,2,\dots,d\}} \!\! y_i }- \bbbbbr{\sum_{i\in J} z_i}
				\\&=
				\sum_{i\in J} (y_i - z_i)
				=
				0
			\end{eqsplit}
	\proves that for all
		$z=(z_1,\dots,z_d)\in[0,\infty)^d$
		with 
			$\sum_{i=1}^{d} z_i = \sum_{i=1}^{d} y_i$
			and $\forall\,i\in J\colon z_i=y_i$
	it holds that
		$z=y$.
		The assumption that
      $\sum_{i=1}^{d} x_i = \sum_{i=1}^{d} y_i$
      and $x\neq y$
    \hence
	\proves[ni] that there exists
		$i\in J$
	such that
		$x_i\neq y_i>0$.
	Combining
		this
	with
		\lref{eq.1}
	\proves that for all 
		$h\in[0,1)$
	it holds that
	\begin{equation}
		\g''(h) > 0
		.
	\end{equation}
		The fundamental theorem of calculus
		\hence
	\proves that for all
		$h\in(0,1)$
	it holds that
	\begin{equation}
		\llabel{eq.2}
		\g'(h)
		=
		\g'(0) + \int_{0}^h \g''(\fh)\,\diff\fh
		>
		\g'(0)
		.
	\end{equation}
	\Moreover
    \lref{eq.diff} and
		the assumption that
			$\sum_{i=1}^{d}x_i = \sum_{i=1}^{d}y_i$
	\prove that
	\begin{eqsplit}
		&\g'(0)
    =
    -\sum_{i\in J}\frac{y_i(x_i-y_i)}{y_i}
		=
		\sum_{i\in J}(y_i-x_i)
		=
		\br*{\sum_{i\in J}y_i}- \br*{\sum_{i\in J}x_i}
		\\&=
		\bbbbbr{\sum_{i\in \{1,2,\dots,d\}}y_i}- \br*{\sum_{i\in J}x_i}
		=
		\bbbbbr{\sum_{i\in \{1,2,\dots,d\}}x_i}- \br*{\sum_{i\in J}x_i}
		=
		\bbbbbr{\sum_{i\in \{1,2,\dots,d\}\setminus J}x_i}
		\geq
		0
		.		
	\end{eqsplit}
	Combining
		this
	and
		\lref{eq.2}
	\proves that for all
		$h\in(0,1)$
	it holds that
	\begin{equation}
		\g'(h)
		>
		0
		.
	\end{equation}
	\Hence that
		$\g$ is strictly increasing.
		This
		and \lref{eq.defg}
	\prove that
		$\f|_{(0,1]}$ is strictly decreasing.
	\Moreover
		\lref{eq.defg}
		and \lref{eq.aa}
	\prove that for all
		$h\in(0,1]$
	it holds that
	\begin{equation}
		\llabel{eq:re}
		f(h)
		=
		-\sum_{i\in J}\ln(y_i+(1-h)(x_i-y_i)) y_i
		=
		-\sum_{i\in J}\ln(x_i+h(y_i-x_i)) y_i
    \in\R
		.
	\end{equation}
	In the remainder of our proof that $f$ is strictly decreasing we distinguish between the case $f(0) = \infty$ and the case $f(0) < \infty$.
	We first prove that $f$ is strictly  decreasing in the case 
	\begin{equation}
		\llabel{eq:f0inf}
		f(0) = \infty.
	\end{equation}
	\Nobs that 
	\enum{
		\lref{eq:f0inf};
		the fact that $\f|_{(0,1]}$ is strictly decreasing;
		\lref{eq:re}
	}  
	\prove that $f$ is strictly  decreasing.
	This establishes that $f$ is strictly  decreasing in the case $f(0) = \infty$.
	In the next step we prove that $f$ is strictly  decreasing in the case 
	\begin{equation}
		\llabel{eq:f0fin}
		f(0) < \infty.
	\end{equation}
	\Nobs that 
		\enum{
			\lref{eq:f0fin};
			\cref{lem:cel_basic2.it.1,lem:cel_basic2.it.2} in \cref{lem:cel_basic2}
		}
	\prove that
	\begin{equation}
		0 \notin \cup_{ i \in J } \{ x_i \}
		\qandq
		f(0)
		=
		-\sum_{i\in J}\ln(x_i+0(y_i-x_i)) y_i
		\in\R
		.
	\end{equation}
	Combining
		this
	with
		\enum{
			\lref{eq:re};
		}
	\proves that $f([0,1])\subseteq\R$ and
	\begin{equation}
		([0,1]\ni h\mapsto f(h)\in\R)\in C([0,1],\R)
		.
	\end{equation}
		This
		and the fact that
			$f|_{(0,1]}$ is strictly decreasing
	\prove that
		$f$ is strictly decreasing.
	This establishes that $f$ is strictly decreasing in the case $f(0) < \infty$.
\end{aproof}
\endgroup

\cfclear
\begingroup
\begin{athm}{cor}{cor:ce_min}
	Let $d\in\N$,
	let $A = \{x=(x_1,\dots,x_d)\in [0,1]^d\colon \sum_{i=1}^{d}x_i=1\}$,
	let $\loss$ be the $d$-dimensional \cfadd{def:crossentropyloss}cross-entropy loss	function, 
	and let $y\in A$
	\cfload.
	Then
	\begin{enumerate}[(i)]
		\item \llabel{it:1}
		it holds that
		\begin{equation}
			\textstyle
			\bcu{x\in A\colon \loss(x,y) = \inf_{z\in A} \loss(z,y)}
			=
			\{y\}
		\end{equation}
		and
		\item \llabel{it:2}
		it holds that
		\begin{equation}
      \inf_{z\in A} \loss(z,y)
      =
			\loss(y,y)
			=
			-\!\!\sum_{\substack{i\in\{1,2,\dots,d\},\\y_i\neq 0}}\!\!\ln(y_i)y_i
		  .
    \end{equation}
	\end{enumerate}
\end{athm}
\begin{aproof}
	\Nobs that
		\cref{lem:cel}
	shows that for all
		$x\in A\backslash\{y\}$
	it holds that
	\begin{equation}
		\loss(x,y)
		=
		\loss(x+0(y-x),y)
		>
		\loss(x+1(y-x),y)
		=
		\loss(y,y)
		.
	\end{equation}
		This
		and  
		\itref{lem:cel_basic2}{it.2}
	\prove
		\cref{cor:ce_min.it:1,cor:ce_min.it:2}.
\end{aproof}
\endgroup

\subsection{Kullback--Leibler divergence loss}

\todoc{Hopital is a bit overkill here}

\cfclear
\begingroup
\begin{athm}{lemma}{lem:kldaux}
	Let $z\in(0,\infty)$. Then
	\begin{enumerate}[(i)]
		\item \llabel{it:1}
		it holds that
		\begin{equation}
			\liminf_{x\searrow 0}\abs{\ln(x)x} = 0 
		\end{equation}
		and
		\item \llabel{it:2}
		it holds for all
			$y\in[0,\infty)$
		that
		\begin{equation}
			\liminf_{\limitVariable{y}\searrow y}\bbr{\ln\bpr{\tfrac z{\limitVariable{y}}}\limitVariable{y}}
			=
			\limsup_{\limitVariable{y}\searrow y}\bbr{\ln\bpr{\tfrac z{\limitVariable{y}}}\limitVariable{y}}
			=
			\begin{cases}
				0 &\colon y=0\\
				\ln\bpr{\tfrac zy}y&\colon y>0
			\end{cases}
		\end{equation}
	\end{enumerate}
\end{athm}
\begin{aproof}
	Throughout this proof, let
		$f\colon (0,\infty)\to \R$
		and $g\colon (0,\infty)\to\R$
	satisfy for all
		$x\in(0,\infty)$
	that
	\begin{equation}
		f(x) = \ln(x^{-1})
		\qandq
		g(x) = x
		.
	\end{equation}
	\Nobs that
		the chain rule
	\proves that for all
		$x\in(0,\infty)$
	it holds that
		$f$ is differentiable
	and
	\begin{equation}
		f'(x)
		=
		-x^{-2}(x^{-1})^{-1}
		=
		-x^{-1}
		.
	\end{equation}
	Combining
		this,
		the fact that $\lim_{x\to\infty}\abs{f(x)} = \infty = \lim_{x\to\infty}\abs{g(x)}$,
		the fact that $g$ is differentiable,
		the fact that for all $x\in(0,\infty)$ it holds that $g'(x)=1\neq 0$,
		and the fact that $\lim_{x\to\infty} \tfrac{-x^{-1}}{1} = 0$
	with
		l'Hôpital's rule
	\proves that
	\begin{equation}
		\liminf_{x\to\infty}\tfrac{f(x)}{g(x)}
		=
		0
		=
		\limsup_{x\to\infty}\tfrac{f(x)}{g(x)}
		.
	\end{equation}
		This
	\proves that
	\begin{equation}
		\liminf_{x\searrow 0}\tfrac{f(x^{-1})}{g(x^{-1})}
		=
		0
		=
		\limsup_{x\searrow 0}\tfrac{f(x^{-1})}{g(x^{-1})}
		.
	\end{equation}
		The fact that
			for all
				$x\in(0,\infty)$
			it holds that
				$\tfrac{f(x^{-1})}{g(x^{-1})} = \ln(x)x$
		\hence
	\proves[ep]
		\lref{it:1}.
	\Nobs that
		\lref{it:1}
		and the fact that
			for all 
				$x\in(0,\infty)$
			it holds that
				$\ln\bpr{\tfrac zx}x = \ln(z)x  - \ln(x)x$
	\prove[ep]
		\lref{it:2}.
\end{aproof}
\endgroup

\begin{athm}{definition}{def:kldloss}
	Let $d\in\N$.
	Then we say that $\loss$ is the $d$-dimensional
	Kullback--Leibler divergence loss function if and only if it holds that
	$\loss\colon [0,\infty)^d\times[0,\infty)^d\to(-\infty,\infty]$ is the function from
	$[0,\infty)^d\times[0,\infty)^d$ to $(-\infty,\infty]$ which satisfies for all
	$x=(x_1,\dots,x_d),\,\allowbreak y=(y_1,\dots,y_d)\in[0,\infty)^d$ that\cfadd{lem:kldaux}
	\begin{equation}
		\llabel{eq}
		\loss(x,y)
		=
		-\sum_{i=1}^{d} \lim_{\limitVariable{x}\searrow x_i}\lim_{\limitVariable{y}\searrow y_i} 
		\bbr{\ln\bpr{\tfrac{\limitVariable{x}}{\limitVariable{y}}}\limitVariable{y}}
	\end{equation}
	\cfload.
\end{athm}

\begin{figure}[!ht]
	\centering
	\includegraphics[width=0.5\linewidth]{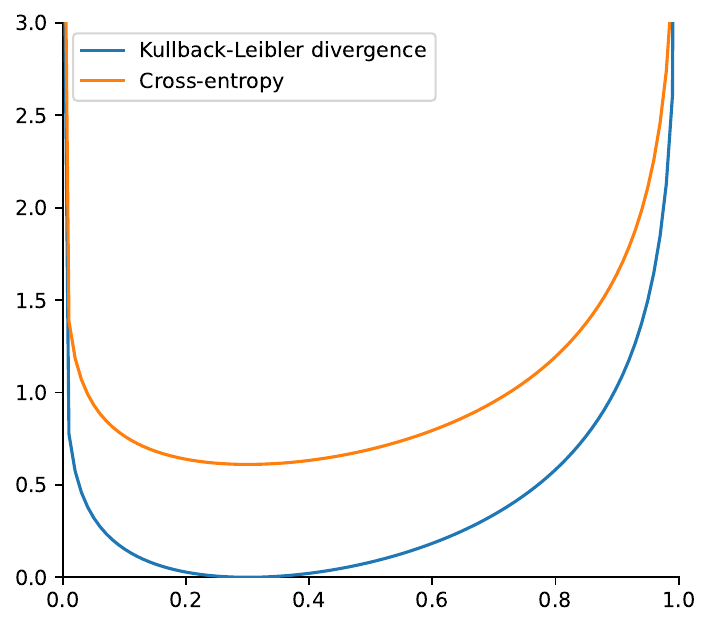}
	\caption{\cfclear\label{fig:kldloss_plot}A plot of the functions
	$(0,1)\ni x\mapsto \loss_i\bpr{(x,1-x),\bpr{\tfrac3{10},\tfrac7{10}}}\in\R$, $i\in\{1,2\}$, where
	$\loss_1$ is the $2$-dimensional Kullback--Leibler\cfadd{def:kldloss} divergence loss function
	and where
	$\loss_2$ is the $2$-dimensional \cfadd{def:crossentropyloss}cross-entropy loss function
	\cfload.}
\end{figure}

\filelisting{code:kldloss_plot}{code/loss_functions/kldloss_plot.py}{{\sc Python} code used to create \cref{fig:kldloss_plot}}

\cfclear
\begingroup
\newcommand{\losskd}{\loss_{\operatorname{KLD}}}
\newcommand{\lossce}{\loss_{\operatorname{CE}}}
\begin{athm}{lemma}{lem:kldcel}
	Let $d\in\N$,
	let $\lossce$ be the $d$-dimensional \cfadd{def:crossentropyloss}cross-entropy loss function,
	and let $\losskd$ be the $d$-dimensional \cfadd{def:kldloss}Kullback--Leibler divergence loss function
	\cfload.
	Then it holds for all
		$x,y\in[0,\infty)^d$
	that
	\begin{equation}
		\llabel{claim}
		\lossce(x,y)
		=
		\losskd(x,y)+\lossce(y,y)
		.
	\end{equation}
\end{athm}
\begin{aproof}
	\Nobs that 
		\cref{lem:kldaux}
	\proves that for all 
		$a,b\in[0,\infty)$
	it holds that
	\begin{eqsplit}
		\lim_{\limitVariable{a}\searrow a}\lim_{\limitVariable{b}\searrow b} \bbr{\ln\bpr{\tfrac{\limitVariable{a}}{\limitVariable{b}}}\limitVariable{b}}
		&=
		\lim_{\limitVariable{a}\searrow a}\lim_{\limitVariable{b}\searrow b} \bbr{\ln(\limitVariable{a})\limitVariable{b}-\ln(\limitVariable{b})\limitVariable{b}}
		\\&=
		\lim_{\limitVariable{a}\searrow a}\bbbr{\ln(\limitVariable{a})b - \lim_{\limitVariable{b}\searrow b}\br{\ln(\limitVariable{b})\limitVariable{b}}}
		\\&=
		\bbpr{\lim_{\limitVariable{a}\searrow a}\br{\ln(\limitVariable{a})b}} - \bbpr{\lim_{\limitVariable{b}\searrow b}\br{\ln(\limitVariable{b})\limitVariable{b}}}
		.
	\end{eqsplit}
		This
		and	\cref{def:kldloss.eq}
	\prove that for all
		$x=(x_1,\dots,x_d),\,\allowbreak y=(y_1,\dots,y_d)\in[0,\infty)^d$
	it holds that
	\begin{eqsplit}
		\llabel{eq:er}
		\losskd(x,y)
		&=
		-\sum_{i=1}^{d} \lim_{\limitVariable{x}\searrow x_i}\lim_{\limitVariable{y}\searrow y_i} \bbr{\ln\bpr{\tfrac{\limitVariable{x}}{\limitVariable{y}}}\limitVariable{y}}
		\\&=
		-\bbbbpr{\sum_{i=1}^{d}\lim_{\limitVariable{x}\searrow x_i}\br{\ln(\limitVariable{x})y_i}} + \bbbbpr{\sum_{i=1}^{d}\lim_{\limitVariable{y}\searrow y_i}\br{\ln(\limitVariable{y})\limitVariable{y}}}
		.
	\end{eqsplit}
	\Moreover
		\cref{lem:kldaux}
	\proves that for all
		$b\in[0,\infty)$
	it holds that
	\begin{equation}
		\lim_{\limitVariable{b}\searrow b}\bbr{\ln(\limitVariable{b})\limitVariable{b}}
		=
		\begin{cases}
			0	&\colon b=0\\
			\ln(b)b &\colon b>0
		\end{cases}
		=
		\lim_{\limitVariable{b}\searrow b}\bbr{\ln(\limitVariable{b})b}
		.
	\end{equation}
	Combining
		this
	with
		\lref{eq:er}
	\proves that for all
		$x=(x_1,\dots,x_d),\,\allowbreak y=(y_1,\dots,y_d)\in[0,\infty)^d$
	it holds that
	\begin{equation}
		\losskd(x,y)
		=
		-\bbbbpr{\sum_{i=1}^{d}\lim_{\limitVariable{x}\searrow x_i}\br{\ln(\limitVariable{x})y_i}} + \bbbbpr{\sum_{i=1}^{d}\lim_{\limitVariable{y}\searrow y_i}\br{\ln(\limitVariable{y})\limitVariable{y}}}
		=
		\lossce(x,y) - \lossce(y,y)
		.
	\end{equation}
	\Hence
		\lref{claim}.
\end{aproof}
\endgroup

\cfclear
\begingroup
\providecommandordefault{\f}{f}
\providecommandordefault{\g}{g}
\begin{athm}{lemma}{lem:kld}
	Let 
		$d\in\N$, 
	let 
		$\loss$ be the $d$-dimensional Kullback--Leibler\cfadd{def:kldloss} divergence loss function, 
	let 
		$x=(x_1,\dots,x_d),\,\allowbreak y=(y_1,\dots,y_d)\in[0,\infty)^d$ 
	satisfy 
		$\sum_{i=1}^{d}x_i=\sum_{i=1}^{d}y_i$
		and $x\neq y$, 
	and let 
		$\f\colon [0,1]\to(-\infty,\infty]$ 
	satisfy for all
		$h\in[0,1]$ 
	that 
	\begin{equation}
		\llabel{claim}
		\f(h) = \loss(x+h(y-x),y)
	\end{equation}
	\cfload.
	Then $\f$ is strictly decreasing.
\end{athm}
\begin{aproof}
	\Nobs that
		\cref{lem:cel}
		and \cref{lem:kldcel}
	\prove
		that $\f$ is strictly decreasing.
\end{aproof}
\endgroup

\cfclear
\begingroup
\begin{athm}{cor}{cor:kld_min}
	Let $d\in\N$,
	let $A = \{x=(x_1,\dots,x_d)\in [0,1]^d\colon \sum_{i=1}^{d}x_i=1\}$,
	let $\loss$ be the $d$-dimensional \cfadd{def:kldloss}Kullback--Leibler divergence loss function,
	and let $y\in A$
	\cfload.
	Then
	\begin{enumerate}[(i)]
		\item \llabel{it:1}
		it holds that
		\begin{equation}
			\textstyle
			\bcu{x\in A\colon \loss(x,y) = \inf_{z\in A} \loss(z,y)}
			=
			\{y\}
		\end{equation}
		and
		\item \llabel{it:2}
		it holds that
		$
    \inf_{z\in A} \loss(z,y) = \loss(y,y)
			=
			0
			$.
	\end{enumerate}
\end{athm}
\begin{aproof}
	\Nobs that
		\cref{lem:kldcel}
		and \cref{lem:kldcel}
	\prove
		\cref{cor:kld_min.it:1,cor:kld_min.it:2}.
\end{aproof}
\endgroup

\endgroup

\section{GF optimization in the training of ANNs}

\todoc{Resultat dass zeigt, dass das integral wohldefiniert ist.}

\cfclear
\begingroup
\renewcommand{\d}{\mathfrak d}
\renewcommand{\th}[1]{\Theta_{#1}}
\renewcommand{\loss}{\mathbf{L}}
\providecommandordefault{\x}{\defaultx}
\providecommandordefault{\y}{\defaulty}

\begin{athm}{example}{lem:gf_example}
	\cfconsiderloaded{def:mseloss}%
	\cfconsiderloaded{def:p-norm}%
	\cfconsiderloaded{def:FFNN}%
	\cfconsiderloaded{def:multidim_version}%
	Let 
		$d,L,\mathfrak d\in\N$,
		$l_1,l_2,\dots,l_L\in\N$
	satisfy
	\begin{equation}
	\label{T_B_D}
	\begin{split} 
		\mathfrak d=l_1(d+1)+\br[\big]{\sum_{k=2}^L l_k(l_{k-1}+1)},
	\end{split}
	\end{equation}
	let
		$a\colon \R\to\R$ be continuously differentiable,
	let
		$M\in\N$,
		$\x_1,\x_2,\dots,\x_M\in\R^d$,
		$\y_1,\y_2,\allowbreak\dots,\allowbreak\y_M\in\R^{l_L}$,
	let
		$\loss\colon\R^{l_L}\times\R^{l_L}\to\R$
		be the mean squared error loss function based on
		$\R^d\ni x\mapsto \Pnorm2 x\in[0,\infty)$,
	let
		$\mathscr L\colon \R^{\mathfrak d}\to[0,\infty)$
	satisfy for all
		$\theta\in\R^{\mathfrak d}$
	that
	\begin{equation}
		\llabel{eq:defL}
		\mathscr L(\theta)
		=
		\frac1M\br*{\sum_{m=1}^{M}
		\loss\bpr{\bpr{ \RealV{ \theta}{0}{\defaultInputDim}{ \multdim_{a, l_1}, \multdim_{a, l_2}, \dots, \multdim_{a, l_{L-1}} , \id_{ \R^{l_L} } } } ( \x_m ), \y_m }
		}
		,
	\end{equation}
	let 	
		$\xi\in\R^\d$,
	and let
		$\th{} \in C([0,\infty),\R^\d)$
	satisfy for all
		$t\in[0,\infty)$
	that
	\begin{equation}
		\llabel{eq:gf}
		\th t
		=
		\xi-\int_0^t(\nabla \mathscr L)(\th{s})
		\,\diff s
	\end{equation}
	(cf.\ \cref{def:mseloss,def:p-norm,def:FFNN,def:multidim_version}, \cref{lem:differentiability_loss2}, and \cref{lem:mseloss_analytic}).
	Then 
		$\th{}$ is a \GF\ trajectory\cfadd{def:gradientflow} for the objective function
		$\mathscr L$ with
		initial value $\xi$
	\cfout.
\end{athm}
\begin{aproof}
	\Nobs that
		\cref{def:gradientflow.eq:grad},
		\cref{def:gradientflow.eq:gf},
		and \lref{eq:gf}
	demonstrate that
	$\th{}$ is a \GF\ trajectory\cfadd{def:gradientflow} for the objective function
	$\mathscr L$ with
	initial value $\xi$
	\cfload.
\end{aproof}
\endgroup

\cfclear
\begingroup
\renewcommand{\d}{\mathfrak d}
\renewcommand{\th}[2]{\Theta^{#1}_{#2}}
\renewcommand{\loss}{\mathbf{L}}
\newcommand{\Loss}{\mathscr{L}}
\providecommandordefault{\x}{\defaultx}
\providecommandordefault{\y}{\defaulty}
\begin{athm}{example}{lem:gf_example2}
	Let 
		$d,L,\mathfrak d\in\N$,
		$l_1,l_2,\dots,l_L\in\N$
	satisfy
	\begin{equation}
	\label{T_B_D}
	\begin{split} 
		\mathfrak d=l_1(d+1)+\br[\big]{\sum_{k=2}^L l_k(l_{k-1}+1)},
	\end{split}
	\end{equation}
	let
		$a\colon \R\to\R$ be continuously differentiable,
	let
		$A\colon \R^{l_L}\to\R^{l_L}$
		be the $l_L$-dimensional softmax activation function,\cfadd{def:softmax}
	let
		$M\in\N$,
		$\x_1,\x_2,\dots,\x_M\in\R^d$,
		$\y_1,\y_2,\dots,\y_M\in[0,\infty)^{l_L}$,
	let
		$\loss_1$
		be the $l_L$-dimensional cross-entropy loss function\cfadd{def:crossentropyloss},
	let
		$\loss_2$
		be the $l_L$-dimensional Kullback--Leibler divergence loss function\cfadd{def:kldloss},
	for every
		$i\in\{1,2\}$
	let
		$\Loss_i\colon \R^{\mathfrak d}\to[0,\infty)$
	satisfy for all
		$\theta\in\R^{\mathfrak d}$
	that
	\begin{equation}
		\llabel{eq:defL}
		\Loss_i(\theta)
		=
		\frac1M\br*{\sum_{m=1}^{M}
		\loss_i\bpr{\bpr{ \RealV{ \theta}{0}{\defaultInputDim}{ \multdim_{a, l_1}, \multdim_{a, l_2}, \dots, \multdim_{a, l_{L-1}} , A } } ( \x_m ), \y_m }
		}
		,
	\end{equation}
	let 	
		$\xi\in\R^{\d}$,
	and for every
		$i\in\{1,2\}$
	let
		$\th i{} \in C([0,\infty),\R^\d)$
	satisfy for all
		$t\in[0,\infty)$
	that
	\begin{equation}
		\llabel{eq:gf}
		\th it
		=
		\xi-\int_0^t(\nabla \mathscr L_i)(\th is) \,\diff s
		\cfadd{lem:differentiability_loss3}
	\end{equation}
	\cfload.
	Then 
		it holds for all
			$i,j\in\{1,2\}$
		that
			$\th i{}$ is a \GF\ trajectory\cfadd{def:gradientflow} for the objective function
			$\mathscr L_j$ with
			initial value $\xi$
	\cfout.
\end{athm}
\begin{aproof}
	\Nobs that
		\cref{lem:kldcel}
	\proves that for all
		$x,y\in (0,\infty)^{l_L}$
	it holds that
	\begin{equation}
		(\nabla_x \loss_1)(x,y)
		=
		(\nabla_x\loss_2)(x,y)
		.
	\end{equation}
	\Hence that for all
		$x\in\R^d$
	it holds that
	\begin{equation}
		(\nabla\Loss_1)(x)
		=
		(\nabla\Loss_2)(x)
		.
	\end{equation}
		This,
		\cref{def:gradientflow.eq:grad},
		\cref{def:gradientflow.eq:gf},
		and \lref{eq:gf}
	\prove that for all
		$i\in\{1,2\}$
	it holds that
	$\th i{}$ is a \GF\ trajectory\cfadd{def:gradientflow} for the objective function
	$\mathscr L_j$ with
	initial value $\xi$
	\cfout.
\end{aproof}
\endgroup

\section{Critical points in optimization problems}

\subsection{Local and global minimizers}

\cfclear
\begin{adef}{def:local_minimum}[Local minimum point]
	Let $\defaultParamDim \in \N$, 
	let $O \subseteq \R^\defaultParamDim$ be a set, 
	let $\vartheta \in O$,
	and let 
	$\defaultLossFunction \colon O \to \R$
	be a function.
	Then we say that $\vartheta$ is a local minimum point of $\defaultLossFunction$ 
	(we say that $\vartheta$ is a local minimizer of $\defaultLossFunction$)
	if and only if there exists $\varepsilon \in (0,\infty)$ such that for all $\theta \in O$ with $\pnorm2{\theta - \vartheta} < \varepsilon$ it holds that
	\begin{equation}
	\label{def:local_minimum:eq1}
	\defaultLossFunction(\vartheta) \leq \defaultLossFunction(\theta)
	\end{equation}
	\cfload.
\end{adef}

\begin{adef}{def:global_minimum}[Global minimum point]
	Let $\defaultParamDim \in \N$,
	let $O \subseteq \R^\defaultParamDim$ be a set,
	let $\vartheta \in O$,
	and let 
	$\defaultLossFunction \colon O \to \R$
	be a function.
	Then we say that $\vartheta$ is a global minimum point of $\defaultLossFunction$ 
	(we say that $\vartheta$ is a global minimizer of $\defaultLossFunction$)
	if and only if it holds for all $\theta \in O$ that
	\begin{equation}
	\label{def:global_minimum:eq1}
	\defaultLossFunction(\vartheta) \leq \defaultLossFunction(\theta).
	\end{equation}
	\cfload.
\end{adef}

\subsection{Local and global maximizers}

\cfclear
\begin{adef}{def:local_maximum}[Local maximum point]
	Let $\defaultParamDim \in \N$, 
	let $O \subseteq \R^\defaultParamDim$ be a set, 
	let $\vartheta \in O$,
	and let 
	$\defaultLossFunction \colon O \to \R$
	be a function.
	Then we say that $\vartheta$ is a local maximum point of $\defaultLossFunction$ 
	(we say that $\vartheta$ is a local maximizer of $\defaultLossFunction$)
	if and only if there exists $\varepsilon \in (0,\infty)$ such that for all $\theta \in O$ with $\pnorm2{\theta - \vartheta} < \varepsilon$ it holds that
	\begin{equation}
	\label{def:local_maximum:eq1}
	\defaultLossFunction(\vartheta) \geq \defaultLossFunction(\theta)
	\end{equation}
	\cfload.
\end{adef}

\begin{adef}{def:global_maximum}[Global maximum point]
	Let $\defaultParamDim \in \N$,
	let $O \subseteq \R^\defaultParamDim$ be a set,
	let $\vartheta \in O$,
	and let 
	$\defaultLossFunction \colon O \to \R$
	be a function.
	Then we say that $\vartheta$ is a global maximum point of $\defaultLossFunction$ 
	(we say that $\vartheta$ is a global maximizer of $\defaultLossFunction$)
	if and only if it holds for all $\theta \in O$ that
	\begin{equation}
	\label{def:global_maximum:eq1}
	\defaultLossFunction(\vartheta) \geq \defaultLossFunction(\theta).
	\end{equation}
	\cfload.
\end{adef}

\subsection{Critical points}
\label{sec:critical_points}

\begin{adef}{def:critical_point}[Critical point]
	Let $\defaultParamDim \in \N$, 
	let $\vartheta \in \R^\defaultParamDim$,
	let $O \subseteq \R^\defaultParamDim$ be an environment of $\vartheta$,
	and
	let $\defaultLossFunction \colon O \to \R$ be differentiable at $\vartheta$.
	Then we say that $\vartheta$ is a critical point of $\defaultLossFunction$ if and only if it holds that
	\begin{equation}
	\label{def:critical_point:eq1}
	(\nabla \defaultLossFunction)(\vartheta) = 0.
	\end{equation}
\end{adef}

\cfclear
\begingroup
\providecommand{\d}{}
\renewcommand{\d}{\defaultParamDim}
\providecommand{\f}{}
\renewcommand{\f}{\defaultLossFunction}
\providecommand{\g}{}
\renewcommand{\g}{L}
\begin{lemma}
\label{minimum1}
Let $ \d \in \N $, 
let $ O \subseteq \R^\d $ be open, 
let $ \vartheta \in O $, 
let $ \f \colon O \to \R $ be a function,
assume that $ \f $ is differentiable at $ \vartheta $, 
and
assume that $ ( \nabla \f )( \vartheta ) \neq 0 $. 
Then there exists $\theta \in O$ such that $\f(\theta) < \f(\vartheta)$.
\end{lemma}

\begin{proof}[Proof of \cref{minimum1}]
Throughout this proof, 
let $ v \in \R^\d \backslash \{ 0 \} $ satisfy
$ 
  v = - (\nabla \f)(\vartheta)
$,
let $ \delta \in (0,\infty) $ satisfy 
for all $ t \in ( - \delta, \delta ) $ that
\begin{equation}
  \vartheta + t v = \vartheta - t (\nabla \f)(\vartheta) \in O,
\end{equation}
and let $ \g \colon (-\delta,\delta) \to \R$ satisfy for all $t \in (-\delta, \delta)$ that 
\begin{equation}
  \g(t) = \f( \vartheta + t v ).
\end{equation}
Note that for all $ t \in (0,\delta) $ 
it holds that
\begin{equation}
\begin{split}
&
  \abs*{
  \br*{
    \frac{ 
      \g(t) - \g(0)
    }{
      t
    }
  }
  +
    \pnorm2{v}^2
  }
=
  \abs*{
  \br*{
    \frac{ 
      \f( \vartheta + t v ) - \f( \vartheta )
    }{
      t
    }
  }
    +
    \pnorm2{
      ( \nabla \f )( \vartheta )
    }^2
  }
\\ & =
  \abs*{
  \br*{
    \frac{ 
      \f( \vartheta + t v ) - \f( \vartheta )
    }{
      t
    }
  }
  +
    \scp{ (\nabla \f)(\vartheta ), (\nabla \f)(\vartheta) }
  }
\\ & =
  \abs*{
  \br*{
    \frac{ 
      \f( \vartheta + t v ) - \f( \vartheta )
    }{
      t
    }
  }
  -
    \scp{ (\nabla \f)(\vartheta ), v }
  }
  .
\end{split}
\end{equation}
Therefore, we obtain that
for all $ t \in (0,\delta) $ 
it holds that
\begin{equation}
\begin{split}
&
  \abs*{
  \br*{
    \frac{ 
      \g(t) - \g(0)
    }{
      t
    }
  }
  +
    \pnorm2v^2
  }
=
  \abs*{
  \br*{
    \frac{ 
      \f( \vartheta + t v ) - \f( \vartheta )
    }{
      t
    }
  }
  -
    \f'(\vartheta ) v 
  }
\\ & =
  \abs*{
    \frac{ 
      \f( \vartheta + t v ) 
      - 
      \f( \vartheta )
      -
      \f'(\vartheta ) t v 
    }{
      t
    }
  }
=
  \frac{ 
    \abs*{
      \f( \vartheta + t v ) 
      - 
      \f( \vartheta )
      -
      \f'(\vartheta ) t v 
    }
  }{
    t
  }
  .
\end{split}
\end{equation}
The assumption that $ \f $ is differentiable at $ \vartheta $ hence 
demonstrates that
\begin{equation}
  \limsup_{ t \searrow 0 }
  \abs*{
  \br*{
    \frac{ 
      \g(t) - \g(0)
    }{
      t
    }
  }
  +
    \pnorm2v^2
  }
  = 0 .
\end{equation}
The fact that 
$
  \pnorm2{ v}^2
  > 0
$
therefore demonstrates that there exists $ t \in (0,\delta) $ 
which satisfies 
\begin{equation}
\label{minimum1:eq1}
  \abs*{
  \br*{
    \frac{ 
      \g(t) - \g(0)
    }{
      t
    }
  }
  +
	\pnorm2v^2
  }
  < 
  \frac{ \pnorm2v^2 }{ 2 } 
  .
\end{equation}
\Nobs that
the triangle inequality,
the fact that $ \pnorm2{v}^2 > 0 $,
and
\cref{minimum1:eq1}
prove that
\begin{equation}
\begin{split}
    \frac{ 
      \g(t) - \g(0)
    }{
      t
    }
& =
  \br*{
    \frac{ 
      \g(t) - \g(0)
    }{
      t
    }
    +
    \pnorm2v^2
  }
  -
  \pnorm2v^2
  \leq
  \abs*{
    \br*{
    \frac{ 
      \g(t) - \g(0)
    }{
      t
    }
    }
    +
    \pnorm2v^2
  }
  -
  \pnorm2v^2
\\ & <
  \frac{
    \pnorm2v^2
  }{ 2 }
  -
  \pnorm2v^2
  =
  -
  \frac{
    \pnorm2v^2
  }{ 2 }
  < 0 
  .
\end{split}
\end{equation}
This ensures that
\begin{equation}
  \f( \vartheta + t v )
  =
  \g(t)
  < \g(0)
  =
  \f( \vartheta )
  .
\end{equation}
The proof of \cref{minimum1} is thus complete.
\end{proof}
\endgroup

\cfclear
\begingroup
\providecommand{\d}{}
\renewcommand{\d}{\defaultParamDim}
\providecommand{\f}{}
\renewcommand{\f}{\defaultLossFunction}
\providecommand{\g}{}
\renewcommand{\g}{\defaultGradientFunction}
\begin{lemma}[A necessary condition for a local minimum point]
\label{minimum2}
Let $\d \in \N$, 
let $O \subseteq \R^\d$ be open, 
let $\vartheta \in O$, 
let $\f \colon O \to \R$ be a function,
assume that $\f$ is differentiable at $\vartheta$, and
assume
\begin{equation}
\label{minimum2:ass1}
\f(\vartheta)= \inf\nolimits_{\theta \in O} \f(\theta).
\end{equation}
Then $(\nabla \f)(\vartheta) = 0$.
\end{lemma}

\begin{proof}[Proof of \cref{minimum2}]
We prove \cref{minimum2} by contradiction. 
We thus assume that $(\nabla \f)(\vartheta) \neq 0 $. 
\cref{minimum1} then implies that there exists $\theta \in O$ such that $\f(\theta) < \f(\vartheta)$. 
Combining this with \eqref{minimum2:ass1} shows that
\begin{equation}
  \f( \theta ) < \f( \vartheta ) 
  =
  \inf_{ w \in O } \f( w )
  \leq 
  \f( \theta )
  .
\end{equation}
The proof of \cref{minimum2} is thus complete.
\end{proof}
\endgroup

\cfclear
\begingroup
\providecommand{\d}{}
\renewcommand{\d}{\defaultParamDim}
\providecommand{\f}{}
\renewcommand{\f}{\defaultLossFunction}
\begin{cor}[Necessary condition for local minimum points]
\label{cor:necessary_condition_local_minimum}
Let $\d \in \N$, 
let $O \subseteq \R^\d$ be open, 
let $\vartheta \in O$, 
let $\f \colon O \to \R$ be differentiable at $\vartheta$,
and assume that $\vartheta$ is a local minimum point of $\f$.
Then\cfadd{def:critical_point} $\vartheta$ is a critical point of $\f$ \cfload.
\end{cor}

\begin{proof}[Proof of \cref{cor:necessary_condition_local_minimum}]
\Nobs that	
\cref{minimum2}
\proves that
$(\nabla \f)(\vartheta) = 0$.
The proof of \cref{cor:necessary_condition_local_minimum} is thus complete.
\end{proof}
\endgroup

\section{Conditions on objective functions in optimization problems}

In this section we discuss different common  assumptions from the scientific literature on the objective function (the function one intends to minimize) of optimization problems. For further reading we refer, \eg, to \cite{Garrigos2023}.

\subsection{Convexity}

\cfclear
\begingroup
\providecommand{\d}{}
\renewcommand{\d}{\defaultParamDim}
\providecommand{\f}{}
\renewcommand{\f}{\defaultLossFunction}
\providecommand{\g}{}
\renewcommand{\g}{\defaultGradientFunction}
\begin{athm}{definition}{def:convex}[Convex functions]
	Let 
		$\d \in \N$
	and let 
		$\f \colon \R^{\d} \to \R$
	be a function. 
	Then we say that $\f$ is a convex function
	(we say that $\f$ is convex)
	if and only if
	it holds for all 
		$\altpointTwo, \altpointThree \in \R^{\d}$,
		$t \in (0,1)$
	that
	\begin{equation}
	\label{def:convex:eq1}
		\f(t\altpointTwo + (1-t)\altpointThree) \leq t\f(\altpointTwo) + (1-t)\f(\altpointThree).
	\end{equation}
\end{athm}
\endgroup

\cfclear
\begingroup
\providecommand{\d}{}
\renewcommand{\d}{\defaultParamDim}
\providecommand{\f}{}
\renewcommand{\f}{\defaultLossFunction}
\providecommand{\g}{}
\renewcommand{\g}{\defaultGradientFunction}
\begin{athm}{lemma}{lemma:convex_equivalence_1}[Equivalence for convex functions]
	Let 
		$\d \in \N$
	and let 
		$\f \colon \R^{\d} \to \R$
	be a function.
	Then the following three statements are equivalent:
	\begin{enumerate}[label=(\roman*)]
		\item \label{lemma:convex_equivalence_1:item1} It holds that $\f$ is convex\cfadd{def:convex} \cfout.
		\cfclear
		\item \label{lemma:convex_equivalence_1:item2} It holds for all 
			$\altpoint, \altpointTwo \in \R^{\d}$,
			$t \in (0,1)$
		that
		\begin{equation}
			\f(\altpoint + t \altpointTwo) 
		\leq 
			\f(\altpoint) + t (\f(\altpoint + \altpointTwo) - \f(\altpoint)).
		\end{equation}
		\item \label{lemma:convex_equivalence_1:item3} It holds for all 
			$\altpoint, \altpointTwo \in \R^{\d}$,
			$t \in (0,1)$
		that
		\begin{equation}
			t\pr[\big]{\f(\altpoint + \altpointTwo) - \f(\altpoint + t \altpointTwo)}
			-
			(1-t)\pr[\big]{\f(\altpoint + t \altpointTwo) - \f(\altpoint)}
		\geq
			0.
		\end{equation}
	\end{enumerate}
\end{athm}
\begin{aproof}
\Nobs that \cref{def:convex:eq1} establishes that 
(\ref{lemma:convex_equivalence_1:item1} $\leftrightarrow$ \ref{lemma:convex_equivalence_1:item2})
and 
(\ref{lemma:convex_equivalence_1:item1} $\leftrightarrow$ \ref{lemma:convex_equivalence_1:item3}).
\end{aproof}

\todoc{Do we need $C^1$? (see arnulf notes -> Third version)}

\cfclear
\begingroup
\providecommand{\d}{}
\renewcommand{\d}{\defaultParamDim}
\providecommand{\f}{}
\renewcommand{\f}{\defaultLossFunction}
\providecommand{\g}{}
\renewcommand{\g}{\defaultGradientFunction}
\providecommandordefault{\v}{\altpointTwo}
\providecommandordefault{\w}{\altpointThree}
\begin{athm}{lemma}{lemma:convex_equivalence}[Equivalence for differentiable convex functions]
	Let 
		$\d \in \N$,
		$\f \in C^1(\R^{\d}, \R)$.
	Then the following three statements are equivalent:
	\begin{enumerate}[label=(\roman*)]
		\item It holds that $\f$ is convex\cfadd{def:convex} \cfout.
		
		\cfclear
		\item It holds for all 
			$\v, \w \in \R^{\d}$
		that
		\begin{equation}
			\f(\v)
		\geq	
			\f(\w) + \scp{
				(\nabla \f)(\w), \v - \w
			}
		\end{equation}
		\cfout.
		
		\cfclear
		\item  
		It holds for all
			$\v, \altpointThree \in \R^{\d}$
		that
		\begin{equation}
			\scp{
				(\nabla \f)(\v) - (\nabla \f)(\altpointThree), \v - \altpointThree
			}
		\geq 
			0
		\end{equation}
		\cfout.
	\end{enumerate}
\end{athm}
\begin{aproof}
We first prove that (\ref{lemma:convex_equivalence_1:item1} $\rightarrow$ \ref{lemma:convex_equivalence_1:item2}).
For this assume that $\f$ is convex\cfadd{def:convex} \cfload.
\argument{
	the assumption that $\f$ is convex;
	\cref{lemma:convex_equivalence_1:item2} in \cref{lemma:convex_equivalence_1}
}
{
	that for all
		$\v, \w \in \R^{\d}$,
		$t \in (0,1)$
	it holds that
	\begin{equation}
		\llabel{eq0}
			\f(\w + t (\v - \w)) 
		\leq 
			\f(\w) + t (\f(\v) - \f(\w)).
	\end{equation}
}
\argument{
	\lref{eq0};
}
{
	that for all
		$\v, \w \in \R^{\d}$,
		$t \in (0,1)$
	it holds that
	\begin{equation}
		\llabel{eq1}
		\f(\v)
	\geq
		\f(\w)
		+
		\frac{
			\f(\w + t(\v - \w)) - \f(\w)
		}{t}.
	\end{equation}	
}
\argument{
	\lref{eq1};
	the assumption that $\f$ is differentiable
}
{
	that for all
		$\v, \w \in \R^{\d}$
	it holds that
	\begin{equation}
		\llabel{eq2}
		\f(\v) 
	\geq
		\f(\w)
		+
		\limsup_{t \to 0}
		\frac{
			\f(\w + t (\v - \w)) - \f(\w)
		}{t}
	=
		\f(\w)
		+
		\scp{
			(\nabla \f)(\w), \v - \w
		}
	\end{equation}
}
This proves that (\ref{lemma:convex_equivalence_1:item1} $\rightarrow$ \ref{lemma:convex_equivalence_1:item2}).

In the next step we prove that (\ref{lemma:convex_equivalence_1:item2} $\rightarrow$ \ref{lemma:convex_equivalence_1:item3}).
For this assume that for all 
	$\v, \w \in \R^{\d}$
it holds that
\begin{equation}
	\llabel{eq3}
	\f(\v)
\geq	
	\f(\w) + \scp{
		(\nabla \f)(\w), \v - \w
	}.
\end{equation}
\startnewargseq
\argument{
	\lref{eq3};
}
{
	that for all
		$\v, \w \in \R^{\d}$
	it holds that
	\begin{equation}
	\begin{split}
		\llabel{eq5}
		\f(\v) + \f(\w)
	&\geq	
		\f(\w) + \scp{
			(\nabla \f)(\w), \v - \w
		}
		+ 
		\f(\v) + \scp{
			(\nabla \f)(\v), \w - \v
		}\\
	&=
		\f(\v) + \f(\w)
		-
		\scp{
			(\nabla \f)(\v) - (\nabla \f)(\w), \v - \w
		}
	\end{split}
	\end{equation}
}
\argument{
	\lref{eq5};
}
{
	that for all
		$\v, \w \in \R^{\d}$
	it holds that
	\begin{equation}
		\llabel{eq6}
		\scp{
			(\nabla \f)(\v) - (\nabla \f)(\w), \v - \w
		}
	\geq 
		0.
	\end{equation}
}
This proves that (\ref{lemma:convex_equivalence_1:item2} $\rightarrow$ \ref{lemma:convex_equivalence_1:item3}).

In the next step we prove that (\ref{lemma:convex_equivalence_1:item3} $\rightarrow$ \ref{lemma:convex_equivalence_1:item1}).
For this assume that for all 
	$\v, \w \in \R^{\d}$
it holds that
\begin{equation}
	\llabel{eq6.1}
	\scp{
		(\nabla \f)(\v) - (\nabla \f)(\w), \v - \w
	}
	\geq 
		0.
\end{equation}
\startnewargseq
\argument{
	\lref{eq6.1};
}
{
	that for all 
		$\altpoint, \altpointTwo \in \R^{\d}$,
		$\alpha, \beta \in \R$
	with 
		$\alpha > \beta$
	it holds that
	\begin{equation}
	\begin{split}
		\llabel{eq7}
		&\scp{
			(\nabla \f)(\altpoint + \alpha \altpointTwo) - (\nabla \f)(\altpoint + \beta \altpointTwo)
			, 
			\altpointTwo
		} \\
	&=
		(\alpha - \beta)^{-1}
		\scp{
			(\nabla \f)(\altpoint + \alpha \altpointTwo) - (\nabla \f)(\altpoint + \beta \altpointTwo)
			, 
			(\alpha - \beta) \altpointTwo
		} 
	\geq
		0.
	\end{split}
	\end{equation}
}
\argument{
	\lref{eq7};
	the fundamental theorem of calculus
}
{
	that for all 
		$\altpoint, \altpointTwo \in \R^{\d}$,
		$t \in (0,1)$
	it holds that
	\begin{equation}
	\llabel{eq8}
	\begin{split}
		&t\pr[\big]{\f(\altpoint + \altpointTwo) - \f(\altpoint + t \altpointTwo)}
		-
		(1-t)\pr[\big]{\f(\altpoint + t \altpointTwo) - \f(\altpoint)}\\
	&=
		t
		\pr*{
			\int_t^1
				\scp{
					(\nabla \f)(\altpoint + s \altpointTwo), \altpointTwo
				}
			\,ds
		}
		-
		(1-t)
		\pr*{
			\int_0^t
				\scp{
					(\nabla \f)(\altpoint + s \altpointTwo), \altpointTwo
				}
			\,ds
		}\\
	&=
		t (1-t)
		\pr*{
			\int_0^1
				\scp{
					(\nabla \f)(\altpoint + (t + s(1-t)) \altpointTwo), \altpointTwo
				}
			\,ds
		} \\
	&\quad
		-
		(1-t) t
		\pr*{
			\int_0^1
				\scp{
					(\nabla \f)(\altpoint + s t \altpointTwo), \altpointTwo
				}
			\,ds
		}\\
	&=
		t (1-t)
		\pr*{
			\int_0^1
				\scp{
					(\nabla \f)(\altpoint + (t + s(1-t)) \altpointTwo) - (\nabla \f)(\altpoint + s t \altpointTwo)
					, 
					\altpointTwo
				}
			\,ds
		} \\
	&\geq
		0.
	\end{split}
	\end{equation}	
}
\argument{
	\lref{eq8};
	\cref{lemma:convex_equivalence_1:item3} in \cref{lemma:convex_equivalence_1}
}
{
	that $\f$ is convex.
}
This proves that (\ref{lemma:convex_equivalence_1:item3} $\rightarrow$ \ref{lemma:convex_equivalence_1:item1}).
\end{aproof}
\endgroup

\subsection{Strict convexity}

\newcommand{\strictlyConvex}{strictly convex\cfadd{def:strictly_convex}}
\cfclear
\begingroup
\providecommand{\d}{}
\renewcommand{\d}{\defaultParamDim}
\providecommand{\f}{}
\renewcommand{\f}{\defaultLossFunction}
\providecommand{\g}{}
\renewcommand{\g}{\defaultGradientFunction}
\begin{athm}{definition}{def:strictly_convex}[Strictly convex functions]
	Let 
		$\d \in \N$
	and let 
		$\f \colon \R^{\d} \to \R$
	be a function. 
	Then we say that $\f$ is a \strictlyConvex{} function
	(we say that $\f$ is \strictlyConvex{})
	if and only if
	it holds for all 
		$\altpointTwo, \altpointThree \in \R^{\d}$,
		$t \in (0,1)$
	with 
		$\altpointTwo \neq \altpointThree$
	that
	\begin{equation}
	\label{def:strictly_convex:eq1}
		\f(t\altpointTwo + (1-t)\altpointThree) < t\f(\altpointTwo) + (1-t)\f(\altpointThree).
	\end{equation}
\end{athm}
\endgroup

\cfclear
\begingroup
\providecommand{\d}{}
\renewcommand{\d}{\defaultParamDim}
\providecommand{\f}{}
\renewcommand{\f}{\defaultLossFunction}
\providecommand{\g}{}
\renewcommand{\g}{\defaultGradientFunction}
\begin{athm}{lemma}{strictly_convex_functions_are_convex}[Strictly convex functions are convex]
	Let 
		$\d \in \N$,
		$\vartheta \in \R^{\d}$
	and
	let 
		$\f \colon \R^{\d} \to \R$
	be \strictlyConvex{} \cfload.
	Then it holds that $\f$ is convex\cfadd{def:convex} \cfout.
\end{athm}
\begin{aproof}
\argument{
	\cref{def:convex:eq1};
	\cref{def:strictly_convex:eq1};
}
{
	that $\f$ is convex\cfadd{def:convex}
}
\end{aproof}
\endgroup

\cfclear
\begingroup
\providecommand{\d}{}
\renewcommand{\d}{\defaultParamDim}
\providecommand{\f}{}
\renewcommand{\f}{\defaultLossFunction}
\providecommand{\g}{}
\renewcommand{\g}{\defaultGradientFunction}
\begin{athm}{lemma}{stricly_convex_minima}[Global minima of strictly convex functions are unique]
	Let 
		$\d \in \N$,
		$\vartheta \in \R^{\d}$, 
	let 
		$\f \colon \R^{\d} \to \R$
	be \strictlyConvex{},
	and assume that $\vartheta$ is a global minimum point of $\f$\cfadd{def:global_minimum} \cfload.
	Then it holds for all global minimum points $\altpointTwo \in \R^\d$ of $\f$ that \begin{equation}
	\label{T_B_D}
	\begin{split} 
		\altpointTwo = \vartheta.
	\end{split}
	\end{equation}
\end{athm}
\begin{aproof}
\argument{
	\cref{def:strictly_convex:eq1};
	the assumption that $\vartheta$ is a global minimum point of $\f$;
	the assumption that $\altpointTwo$ is a global minimum point of $\f$;
}
{
	that for all 
		$\altpointTwo \in \R^\d \backslash \{\vartheta\}$ 
	with $\f(\altpointTwo) = \inf_{\altpoint \in \R^\d} \f(\altpoint)$ 
	it holds that
	\begin{equation}
	\begin{split}
		\llabel{eq0}
		\f(\vartheta)
	&\leq
		\f\pr*{
			\tfrac{\altpointTwo + \vartheta}{2}
		}
	<
		\tfrac{1}{2} \br*{
			\f(\altpointTwo)
			+
			\f(\vartheta)
		}
	=
		\f(\vartheta)
	\end{split}
	\end{equation}
}
\argument{
	\lref{eq0};
}
{
	that 
	\begin{equation}
	\label{T_B_D}
	\begin{split} 
		\cu*{
			\altpointTwo \in \R^\d \backslash \{\vartheta\} \colon 
			\f(\altpointTwo)
			=
			\inf_{\altpoint \in \R^\d} \f(\altpoint)
		}
		=
		\{\}.
	\end{split}
	\end{equation}
}
\end{aproof}
\endgroup

\cfclear
\begingroup
\providecommand{\d}{}
\renewcommand{\d}{\defaultParamDim}
\providecommand{\f}{}
\renewcommand{\f}{\defaultLossFunction}
\providecommand{\g}{}
\renewcommand{\g}{\defaultGradientFunction}
\begin{athm}{example}{example_strict_convexity}
Let $f \colon \R \to \R$ and $g \colon \R \to \R$ satisfy for all $x \in \R$ that
\begin{equation}
\label{T_B_D}
\begin{split} 
	f(x) = \exp(x) \qandq g(x) = 0.
\end{split}
\end{equation}
Then
\begin{enumerate}[label=(\roman*)]
	\item it holds that $f$ is \strictlyConvex{},
	\item it holds that $g$ is convex\cfadd{def:convex}, and
	\item it holds that $g$ is not \strictlyConvex{}
\end{enumerate}
\cfout.
\end{athm}

\subsection{Monotonicity}

\todoc{Decide on order in scalar product and try to do it consistently over most of document.}

\cfclear
\begingroup
\providecommand{\d}{}
\renewcommand{\d}{\defaultParamDim}
\providecommand{\f}{}
\renewcommand{\f}{\defaultLossFunction}
\providecommand{\g}{}
\renewcommand{\g}{\defaultGradientFunction}
\begin{athm}{definition}{def:monotonically_increasing_1}[Monotonically increasing functions]
	Let 
		$\d \in \N$
	and let 
		$\g \colon \R^{\d} \to \R^{\d}$
	be a function. 
	Then we say that $\g$ is a monotonically increasing function
	(we say that $\g$ is monotonically increasing)
	if and only if
	it holds for all $\altpointTwo, \altpointThree \in \R^{\d}$ that
	\begin{equation}
	\label{def:monotonically_increasing_1:eq1}
		\scp{
			\g (\altpointTwo) - \g (\altpointThree), \altpointTwo - \altpointThree
		} 
		\geq 
			0
	\end{equation}
	\cfload.
\end{athm}
\endgroup

\cfclear
\begingroup
\providecommand{\d}{}
\renewcommand{\d}{\defaultParamDim}
\providecommand{\f}{}
\renewcommand{\f}{\defaultLossFunction}
\providecommand{\g}{}
\renewcommand{\g}{\defaultGradientFunction}
\begin{athm}{definition}{def:monotonically_decreasing_1}[Monotonically decreasing functions]
	Let 
		$\d \in \N$
	and let 
		$\g \colon \R^{\d} \to \R^{\d}$
	be a function. 
	Then we say that $\g$ is a monotonically decreasing function
	(we say that $\g$ is monotonically decreasing)
	if and only if
	it holds for all $\altpointTwo, \altpointThree \in \R^{\d}$ that
	\begin{equation}
	\label{def:monotonically_decreasing_1:eq1}
		\scp{
			\g (\altpointTwo) - \g (\altpointThree), \altpointTwo - \altpointThree
		} 
		\leq 
			0
	\end{equation}
	\cfload.
\end{athm}
\endgroup

\cfclear
\begingroup
\providecommand{\d}{}
\renewcommand{\d}{\defaultParamDim}
\providecommand{\f}{}
\renewcommand{\f}{\defaultLossFunction}
\providecommand{\g}{}
\renewcommand{\g}{\defaultGradientFunction}
\begin{athm}{lemma}{lemma:increasing_decreasing_equivalence_1}[Equivalence for monotonically increasing and decreasing functions]
	Let 
		$\d \in \N$
	and let 
		$\g \colon \R^{\d} \to \R^{\d}$
	be a function. 
	Then the following two statements are equivalent:
	\begin{enumerate}[label=(\roman*)]
		\item 
		\label{increasing_decreasing_equivalence_1:item1}
		It holds that $\g$ is monotonically increasing\cfadd{def:monotonically_increasing_1} \cfout.
		\cfclear
		\item 
		\label{increasing_decreasing_equivalence_1:item2}
		It holds that $-\g$ is monotonically decreasing\cfadd{def:monotonically_decreasing_1} \cfout.
	\end{enumerate}
\end{athm}

\begin{aproof}
\Nobs that
\enum{
	\cref{def:monotonically_increasing_1:eq1};
	\cref{def:monotonically_decreasing_1:eq1};
}
\prove
that 
(\ref{increasing_decreasing_equivalence_1:item1} $\leftrightarrow$ \ref{increasing_decreasing_equivalence_1:item2}).
\end{aproof}

\cfclear
\begingroup
\providecommand{\d}{}
\renewcommand{\d}{\defaultParamDim}
\providecommand{\f}{}
\renewcommand{\f}{\defaultLossFunction}
\providecommand{\g}{}
\renewcommand{\g}{\defaultGradientFunction}
\begin{athm}{lemma}{lemma:convex_monotonicity}[Convexity and monotonicity]
	Let 
		$\d \in \N$,
		$\f \in C^1(\R^{\d}, \R)$.
	Then the following three statements are equivalent:
	\begin{enumerate}[label=(\roman*)]
		\item \label{lemma:convex_monotonicity:item1} It holds that $\f$ is convex\cfadd{def:convex} \cfout.
		\item \label{lemma:convex_monotonicity:item2} It holds that $\nabla \f$ is monotonically increasing\cfadd{def:monotonically_increasing_1} \cfout.
		\item \label{lemma:convex_monotonicity:item3} It holds that $-(\nabla \f)$ is monotonically decreasing\cfadd{def:monotonically_decreasing_1} \cfout.
	\end{enumerate}
\end{athm}
\begin{aproof}
\argument{
	\cref{lemma:convex_equivalence};
	\cref{lemma:increasing_decreasing_equivalence_1};
}[verbs=ep]
{
	that 
	(\ref{lemma:convex_monotonicity:item1} $\leftrightarrow$ \ref{lemma:convex_monotonicity:item2})
	and that
	(\ref{lemma:convex_monotonicity:item1} $\leftrightarrow$ \ref{lemma:convex_monotonicity:item3}).
}
\end{aproof}

\newcommand{\cMonotonicallyIncreasing}[1]{#1-generalized monotonically increasing\cfadd{def:monotonically_increasing_2}}

\cfclear
\begingroup
\providecommand{\d}{}
\renewcommand{\d}{\defaultParamDim}
\providecommand{\f}{}
\renewcommand{\f}{\defaultLossFunction}
\providecommand{\g}{}
\renewcommand{\g}{\defaultGradientFunction}
\begin{athm}{definition}{def:monotonically_increasing_2}[Generalized monotonically increasing functions]
	Let 
		$\d \in \N$,
		$c \in \R$
	and let 
		$\g \colon \R^{\d} \to \R^{\d}$
	be a function. 
	Then we say that $\g$ is a \cMonotonicallyIncreasing{c} function
	(we say that $\g$ is \cMonotonicallyIncreasing{c})
	if and only if
	it holds for all $\altpointTwo, \altpointThree \in \R^{\d}$ that\cfclear
	\begin{equation}
	\label{def:monotonically_increasing_2:eq1}
		\scp{
			\g (\altpointTwo) - \g (\altpointThree), \altpointTwo - \altpointThree
		} 
		\geq 
		c \pnorm2{\altpointTwo - \altpointThree}^2
	\end{equation}
	\cfload.
\end{athm}
\endgroup

\newcommand{\cMonotonicallyDecreasing}[1]{#1-generalized monotonically decreasing\cfadd{def:monotonically_decreasing_2}}

\cfclear
\begingroup
\providecommand{\d}{}
\renewcommand{\d}{\defaultParamDim}
\providecommand{\f}{}
\renewcommand{\f}{\defaultLossFunction}
\providecommand{\g}{}
\renewcommand{\g}{\defaultGradientFunction}
\begin{athm}{definition}{def:monotonically_decreasing_2}[Generalized monotonically decreasing functions]
	Let 
		$\d \in \N$,
		$c \in \R$
	and let 
		$\g \colon \R^{\d} \to \R^{\d}$
	be a function. 
	Then we say that $\g$ is a \cMonotonicallyDecreasing{c} function
	(we say that $\g$ is \cMonotonicallyDecreasing{c})
	if and only if
	it holds for all $\altpointTwo, \altpointThree \in \R^{\d}$ that\cfclear
	\begin{equation}
	\label{def:monotonically_decreasing_2:eq1}
		\scp{
			\g (\altpointTwo) - \g (\altpointThree), \altpointTwo - \altpointThree
		} 
		\leq 
		-c \pnorm2{\altpointTwo - \altpointThree}^2.
	\end{equation}
	\cfload.
\end{athm}
\endgroup

\cfclear
\begingroup
\providecommand{\d}{}
\renewcommand{\d}{\defaultParamDim}
\providecommand{\f}{}
\renewcommand{\f}{\defaultLossFunction}
\providecommand{\g}{}
\renewcommand{\g}{\defaultGradientFunction}
\begin{athm}{lemma}{lemma:increasing_decreasing_equivalence_2}[Equivalence for monotonically increasing and decreasing functions]
	Let 
		$\d \in \N$,
		$c \in \R$
	and let 
		$\g \colon \R^{\d} \to \R^{\d}$
	be a function. 
	Then the following two statements are equivalent:
	\begin{enumerate}[label=(\roman*)]
		\item 
		\label{increasing_decreasing_equivalence_2:item1}
		It holds that $\g$ is \cMonotonicallyIncreasing{c} \cfout.
		\cfclear
		\item 
		\label{increasing_decreasing_equivalence_2:item2}
		It holds that $-\g$ is \cMonotonicallyDecreasing{c} \cfout.
	\end{enumerate}
\end{athm}

\begin{aproof}
\Nobs that
\enum{
	\cref{def:monotonically_increasing_2:eq1};
	\cref{def:monotonically_decreasing_2:eq1};
}
\prove
that 
(\ref{increasing_decreasing_equivalence_2:item1} $\leftrightarrow$ \ref{increasing_decreasing_equivalence_2:item2}).
\end{aproof}

\endgroup

\cfclear
\begingroup
\providecommand{\d}{}
\renewcommand{\d}{\defaultParamDim}
\providecommand{\f}{}
\renewcommand{\f}{\defaultLossFunction}
\providecommand{\g}{}
\renewcommand{\g}{\defaultGradientFunction}
\begin{athm}{lemma}{lemma:increasing_jacobian_equivalence}[Equivalence for differentiable monotonically increasing functions]
	Let 
		$\d \in \N$,
		$c \in \R$
	and let 
		$\g \in C^1(\R^{\d}, \R^{\d})$.
	Then the following two statements are equivalent:
	\begin{enumerate}[label=(\roman*)]
		\item 
		\label{increasing_jacobian_equivalence:item1}
		It holds that $\g$ is \cMonotonicallyIncreasing{c} \cfout.
		\cfclear
		\item 
		\label{increasing_jacobian_equivalence:item2}
		It holds for all
			$\altpointTwo, \altpointThree \in \R^\d$
		that
		\begin{equation}
		\label{T_B_D}
		\begin{split} 
			\scp{
				\g' (\altpointTwo) \altpointThree, \altpointThree
			} 
		\geq
			c \pnorm2{w}^2
		\end{split}
		\end{equation}
		\cfout.
	\end{enumerate}
\end{athm}

\begin{aproof}
We first prove that (\ref{increasing_jacobian_equivalence:item1} $\rightarrow$ \ref{increasing_jacobian_equivalence:item2}).
For this assume that $\g$ is \cMonotonicallyIncreasing{c} \cfload.
\argument{
	\cref{def:monotonically_increasing_2:eq1};
	the fact that $\g$ is differentiable;
}
{
	that for all
		$\altpointTwo, \altpointThree \in \R^\d$
	it holds that
	\begin{equation}
	\label{increasing_jacobian_equivalence:eq1}
	\begin{split} 
		\scp{
			\g' (\altpointTwo) \altpointThree, \altpointThree
		} 
	&=
		\limsup_{t \to 0}
			\pr*{
				\scp*{
					\frac{\g (\altpointTwo + t \altpointThree) - \g (\altpointTwo)}{t}, \altpointThree
				}
			}\\
	&=
		\limsup_{t \to 0}
			\pr*{
				\frac{1}{t^2}
				\scp*{
					\g (\altpointTwo + t \altpointThree) - \g (\altpointTwo), t \altpointThree
				}
			} \\
	&\geq
		\limsup_{t \to 0}
			\pr*{
				\frac{c}{t^2}
				\pnorm2{t \altpointThree}^2
			}
	=
		c \pnorm2{\altpointThree}^2
	\end{split}
	\end{equation}
\cfload.
}
This proves that (\ref{increasing_jacobian_equivalence:item1} $\rightarrow$ \ref{increasing_jacobian_equivalence:item2}).

In the next step, we prove that (\ref{increasing_jacobian_equivalence:item2} $\rightarrow$ \ref{increasing_jacobian_equivalence:item1}).
For this assume that for all
	$\altpointTwo, \altpointThree \in \R^\d$
it holds that
\begin{equation}
\label{increasing_jacobian_equivalence:eq2}
\begin{split} 
	\scp{
		\g' (\altpointTwo) \altpointThree, \altpointThree
	} 
\geq
	c \pnorm2{w}^2.
\end{split}
\end{equation}
\startnewargseq
\argument{
	\cref{increasing_jacobian_equivalence:eq2};
	the fundamental theorem of calculus;
}
{
	that for all
		$\altpointTwo, \altpointThree \in \R^\d$
	it holds that
	\begin{equation}
	\label{increasing_jacobian_equivalence:eq3}
	\begin{split} 
		\scp*{
			\g (\altpointTwo) - \g (\altpointThree), \altpointTwo - \altpointThree
		} 
	&=
		\scp*{
			\int_0^1 \g' (\altpointThree + t (\altpointTwo - \altpointThree)) (\altpointTwo - \altpointThree) \, dt
			, 
			\altpointTwo - \altpointThree
		} \\
	&=
		\int_0^1 \scp*{
			\g' (\altpointThree + t (\altpointTwo - \altpointThree)) (\altpointTwo - \altpointThree)
			, 
			\altpointTwo - \altpointThree
		} \, dt \\
	&\geq
		\int_0^1
			c \pnorm2{\altpointTwo - \altpointThree}^2 \, 
		dt 
	=
		c \pnorm2{\altpointTwo - \altpointThree}^2.
	\end{split}
	\end{equation}
}
This proves that (\ref{increasing_jacobian_equivalence:item2} $\rightarrow$ \ref{increasing_jacobian_equivalence:item1}).
\end{aproof}

\endgroup

\subsection{Subgradients}

\cfclear
\begingroup
\providecommand{\d}{}
\renewcommand{\d}{\defaultParamDim}
\providecommand{\f}{}
\renewcommand{\f}{\defaultLossFunction}
\providecommand{\g}{}
\renewcommand{\g}{\defaultGradientFunction}
\begin{athm}{definition}{def:subgradient}[Subgradients]
	Let 
		$\d \in \N$,
		$g, \altpoint \in \R^{\d}$
	and let 
		$\f \colon \R^{\d} \to \R$
	be a function. 
	Then we say that $g$ is a subgradient of $\f$ at $\altpoint$ 
	if and only if
	it holds for all $\altpointTwo \in \R^{\d}$ that
	\begin{equation}
	\label{def:subgradient:eq1}
		\f(\altpointTwo) \geq \f(\altpoint) + \scp{g, \altpointTwo - \altpoint}
	\end{equation}
	\cfload.
\end{athm}
\endgroup

\cfclear
\begingroup
\providecommand{\d}{}
\renewcommand{\d}{\defaultParamDim}
\providecommand{\f}{}
\renewcommand{\f}{\defaultLossFunction}
\providecommand{\g}{}
\renewcommand{\g}{\defaultGradientFunction}
\begin{athm}{lemma}{lemma:convex_subgradients}[Convexity and subgradients]
	Let 
		$\d \in \N$,
		$\f \in C^1(\R^{\d}, \R)$.
	Then the following two statements are equivalent:
	\begin{enumerate}[label=(\roman*)]
		\item \label{lemma:convex_subgradients:item1} It holds that $\f$ is convex\cfadd{def:convex} \cfout.
		\cfclear
		\item \label{lemma:convex_subgradients:item2} It holds for all $\altpoint \in \R^{\d}$ that $(\nabla \f)(\altpoint)$ is a subgradient of $\f$ at $\altpoint$\cfadd{def:subgradient} \cfout.
	\end{enumerate}
\end{athm}
\begin{aproof}
\argument{
	\cref{lemma:convex_equivalence}
}[verbs=ep]
{
	that 
	(\ref{lemma:convex_subgradients:item1} $\leftrightarrow$ \ref{lemma:convex_subgradients:item2}).
}
\end{aproof}

\subsection{Strong convexity}

\newcommand{\generalizedConvex}[1]{$#1$-generalized convex\cfadd{def:generalized_convex}}

\cfclear
\begingroup
\providecommand{\d}{}
\renewcommand{\d}{\defaultParamDim}
\providecommand{\f}{}
\renewcommand{\f}{\defaultLossFunction}
\providecommand{\g}{}
\renewcommand{\g}{\defaultGradientFunction}
\begin{athm}{definition}{def:generalized_convex}[Generalized convex functions]
	Let 
		$\d \in \N$,
		$c \in \R$
	and let 
		$\f \colon \R^{\d} \to \R$
	be a function. 
	Then we say that $\f$ is a \generalizedConvex{c} function (we say that $\f$ is \generalizedConvex{c}) 
	if and only if it holds that\cfclear
	\begin{equation}
	\label{def:generalized_convex:eq1}
	\begin{split} 
		\R^{\d} \ni \theta \mapsto \f(\theta) - \tfrac{c}{2} \pnorm2{\theta}^2 \in \R
	\end{split}
	\end{equation}
	is convex\cfadd{def:convex}
	\cfload.
\end{athm}
\endgroup

\newcommand{\stronglyConvex}{strongly convex\cfadd{def:strongly_convex}}

\cfclear
\begingroup
\providecommand{\d}{}
\renewcommand{\d}{\defaultParamDim}
\providecommand{\f}{}
\renewcommand{\f}{\defaultLossFunction}
\providecommand{\g}{}
\renewcommand{\g}{\defaultGradientFunction}
\begin{athm}{definition}{def:strongly_convex}[Strongly convex functions]
	Let 
		$\d \in \N$
	and let 
		$\f \colon \R^{\d} \to \R$
	be a function. 
	Then we say that $\f$ is a \stronglyConvex{} function (we say that $\f$ is \stronglyConvex{}) 
	if and only if there exists $c \in (0,\infty)$ such that $\f$ is \generalizedConvex{c} \cfload.
\end{athm}
\endgroup

\cfclear
\begingroup
\providecommand{\d}{}
\renewcommand{\d}{\defaultParamDim}
\providecommand{\f}{}
\renewcommand{\f}{\defaultLossFunction}
\providecommand{\g}{}
\renewcommand{\g}{\defaultGradientFunction}
\begin{athm}{lemma}{lemma:generalized_convex_equivalence}[Equivalence for generalized convex functions]
	Let 
		$\d \in \N$,
		$c \in \R$
	and let 
		$\f \colon \R^{\d} \to \R$
	be a function. 
	Then the following four statements are equivalent:
	\begin{enumerate}[label=(\roman*)]
		\item \label{lemma:generalized_convex_equivalence:item1} It holds that $\f$ is \generalizedConvex{c} \cfout.
		\cfclear
		\item \label{lemma:generalized_convex_equivalence:item2} It holds for all $\altpointTwo, \altpointThree \in \R^{\d}$, $t \in (0,1)$ that
		\begin{equation}
			\llabel{concl1}
			\f(t\altpointTwo + (1-t)\altpointThree) \leq t\f(\altpointTwo) + (1-t)\f(\altpointThree) - \tfrac{c}{2} \br*{ t (1-t) \pnorm2{\altpointTwo - \altpointThree}^2}
		\end{equation}
		\cfout.

		\cfclear
		\item \label{lemma:generalized_convex_equivalence:item3} 
		It holds for all 
			$\altpoint, \altpointTwo \in \R^{\d}$,
			$t \in (0,1)$
		that
		\begin{equation}
			\f(\altpoint + t \altpointTwo) 
		\leq 
			\f(\altpoint) + t (\f(\altpoint + \altpointTwo) - \f(\altpoint))
			-
			\tfrac{c}{2} \br*{ t (1-t) \pnorm2{\altpointTwo}^2}
		\end{equation}
		\cfout.

		\cfclear
		\item \label{lemma:generalized_convex_equivalence:item4} 
		It holds for all 
		$\altpoint, \altpointTwo \in \R^{\d}$,
		$t \in (0,1)$
		that
		\begin{equation}
			t\pr[\big]{\f(\altpoint + \altpointTwo) - \f(\altpoint + t \altpointTwo)}
			-
			(1-t)\pr[\big]{ \f(\altpoint + t \altpointTwo) - \f(\altpoint)}
		\geq
			\tfrac{c}{2} \br*{ t (1-t) \pnorm2{\altpointTwo - \altpointThree}^2}
		\end{equation}
		\cfout.
	\end{enumerate}
\end{athm}
\begin{aproof}

\argument{
	\cref{def:convex:eq1};
	\cref{def:generalized_convex:eq1};
}
{
	that $\f$ is \generalizedConvex{c} if and only if it holds for all
		$\altpointTwo, \altpointThree \in \R^{\d}$, $t \in (0,1)$ that
	\begin{equation}
		\llabel{eq1}
		\f(t\altpointTwo + (1-t)\altpointThree) 
		-
		\tfrac{c}{2} \pnorm2{t\altpointTwo + (1-t)\altpointThree}^2
	\leq 
		t \pr*{
			\f(\altpointTwo)
			-
			\tfrac{c}{2} \pnorm2{\altpointTwo}^2
		}
		+
		(1-t) \pr*{
			\f(\altpointThree)
			-
			\tfrac{c}{2} \pnorm2{\altpointThree}^2
		}
	\end{equation}
}
\argument{
	\lref{eq1};
}
{
	that $\f$ is \generalizedConvex{c} if and only if it holds for all
		$\altpointTwo, \altpointThree \in \R^{\d}$, $t \in (0,1)$ 
	that
	\begin{equation}
	\begin{split}
		\llabel{eq2}
		\f(t\altpointTwo + (1-t)\altpointThree) 
	&\leq 
		t \f(\altpointTwo)
		+
		(1-t) \f(\altpointThree)\\
	&\quad
		-
		\tfrac{c}{2} 
		\pr*{ 
			t \pnorm2{\altpointTwo}^2
			+
			(1-t) \pnorm2{\altpointThree}^2
			-
			\pnorm2{t\altpointTwo + (1-t)\altpointThree}^2
		}.
	\end{split}
	\end{equation}
}
\argument{
	the fact that for all 
		$t \in (0,1)$
	it holds that
	\begin{equation}
		(1-t) - (1-t)^2 
	=
		1 - t - t^2 +2t - 1 
	=
		t(1-t)
	\end{equation}
}
{
	that for all
		$\altpointTwo, \altpointThree \in \R^{\d}$, 
		$t \in (0,1)$
	it holds that 
	\begin{equation}
	\llabel{eq3}
	\begin{split} 
		&t \pnorm2{\altpointTwo}^2
		+
		(1-t) \pnorm2{\altpointThree}^2
		-
		\pnorm2{t\altpointTwo + (1-t)\altpointThree}^2\\
	&=
		t \pnorm2{\altpointTwo}^2
		+
		(1-t) \pnorm2{\altpointThree}^2
		-
		\pr*{
			t^2 \pnorm2{\altpointTwo}^2
			+
			(1-t)^2 \pnorm2{\altpointThree}^2
			+
			2t(1-t) \scp{\altpointTwo, \altpointThree}
		}
		\\
	&=
		(t - t^2) \pnorm2{\altpointTwo}^2
		+
		(1-t - (1-t)^2) \pnorm2{\altpointThree}^2
		-
		2t(1-t) \scp{\altpointTwo, \altpointThree}\\
	&=
		t(1-t) \pr*{
			\pnorm2{\altpointTwo}^2
			+
			\pnorm2{\altpointThree}^2
			-
			2 \scp{\altpointTwo, \altpointThree}
		}
		\\
	&=
		t(1-t) \pnorm2{\altpointTwo - \altpointThree}^2.
	\end{split}
	\end{equation}
}
\argument{
	\lref{eq2};
	\lref{eq3};
}
{
	that $\f$ is \generalizedConvex{c} if and only if it holds for all
		$\altpointTwo, \altpointThree \in \R^{\d}$, 
		$t \in (0,1)$ 
	that
	\begin{equation}
	\begin{split}
		\llabel{eq4}
		\f(t\altpointTwo + (1-t)\altpointThree) 
	&\leq 
		t \f(\altpointTwo)
		+
		(1-t) \f(\altpointThree)
		-
		\tfrac{c}{2} \br*{ t(1-t) \pnorm2{\altpointTwo - \altpointThree}^2}.
	\end{split}
	\end{equation} 
}
\argument{
	\lref{eq4};
}[verbs=ep]
{
	that (\ref{lemma:generalized_convex_equivalence:item1} $\leftrightarrow$ \ref{lemma:generalized_convex_equivalence:item2}).
}
\argument{
	\lref{concl1};
}[verbs=ep]
{
	that (\ref{lemma:generalized_convex_equivalence:item2} $\leftrightarrow$ \ref{lemma:generalized_convex_equivalence:item3})
	and that (\ref{lemma:generalized_convex_equivalence:item3} $\leftrightarrow$ \ref{lemma:generalized_convex_equivalence:item4}).
}
\end{aproof}
\endgroup

\cfclear
\begingroup
\providecommand{\d}{}
\renewcommand{\d}{\defaultParamDim}
\providecommand{\f}{}
\renewcommand{\f}{\defaultLossFunction}
\providecommand{\g}{}
\renewcommand{\g}{\defaultGradientFunction}
\begin{athm}{lemma}{strongly_convex_functions_are_stricly_convex}[Strongly convex functions are strictly convex]
	Let 
		$\d \in \N$	
	and
	let 
		$\f \colon \R^{\d} \to \R$
	be \stronglyConvex{} \cfload.
	Then it holds that $\f$ is \strictlyConvex{} \cfout.
\end{athm}
\begin{aproof}
\argument{
	\cref{def:strictly_convex:eq1};
	\cref{lemma:generalized_convex_equivalence:item2} in \cref{lemma:generalized_convex_equivalence};
}
{
	that $\f$ is \strictlyConvex{}
}
\end{aproof}
\endgroup

\cfclear
\begingroup
\providecommand{\d}{}
\renewcommand{\d}{\defaultParamDim}
\providecommand{\f}{}
\renewcommand{\f}{\defaultLossFunction}
\providecommand{\g}{}
\renewcommand{\g}{\defaultGradientFunction}
\providecommand{\C}{}
\renewcommand{\C}{\mathfrak{C}} 
\providecommandordefault{\v}{\altpointTwo}
\providecommandordefault{\w}{\altpointThree}
\begin{athm}{cor}{lemma:strongly_convex_minimum_point}[Strongly convex functions have a unique minimum point]
	Let 
		$\d \in \N$	
	and
	let 
		$\f \in C(\R^{\d} , \R)$
	be \stronglyConvex{} 
	\cfload.
	Then there exists a unique $\vartheta \in \R^{\d}$ such that $\vartheta$ is a global minimum point of $\f$\cfadd{def:global_minimum}
	\cfout.
\end{athm}

\begin{aproof}
Throughout this proof, for every $r \in (0,\infty)$ let 
\begin{equation}
\label{T_B_D}
\begin{split} 
	B_r 
=
	\{ \altpointTwo \in \R^{\d} \colon \pnorm2{\altpointTwo} \leq r \},
\end{split}
\end{equation}
let 
\begin{equation}
\label{T_B_D}
\begin{split} 
	S 
=
	\{ \altpointTwo \in \R^{\d} \colon \pnorm2{\altpointTwo} = 1 \}
\end{split}
\end{equation}
let $c \in (0,\infty)$ satisfy that $\f$ is \generalizedConvex{c},
and let $C \in \R$ satisfy
\begin{equation}
\label{T_B_D}
\begin{split} 
		C
	=
		\br*{ \inf_{\altpoint \in S} \f\pr*{\altpoint} }
		- \f(0) 
		-\tfrac{c}{2} 
\end{split}
\end{equation}
\cfload.
\argument{
	\cref{lemma:generalized_convex_equivalence:item3} in \cref{lemma:generalized_convex_equivalence};
	the fact that $\f$ is \generalizedConvex{c}
}
{
	that for all
		$\altpointTwo \in \R^{\d} \backslash B_1$
	it holds that 
	\begin{equation}
	\llabel{eq0}
		\f\pr*{0 + \tfrac{1}{\pnorm2{\altpointTwo}} \altpointTwo} 
	\leq 
		\f(0) + \tfrac{1}{\pnorm2{\altpointTwo}} (\f(\altpointTwo) - \f(0))
		-
		\tfrac{c}{2} \br*{ \tfrac{1}{\pnorm2{\altpointTwo}} \pr*{1-\tfrac{1}{\pnorm2{\altpointTwo}}} \pnorm2{\altpointTwo}^2}
	\end{equation}
}
\argument{
	\lref{eq0};
}
{
	that for all
		$\altpointTwo \in \R^{\d} \backslash B_1$
	it holds that 
	\begin{equation}
	\begin{split}
	\llabel{eq1}
		\f(\altpointTwo) 
	&\geq 
		\f(0)
		+
		\pnorm2{\altpointTwo}\pr*{
			\f\pr*{\tfrac{1}{\pnorm2{\altpointTwo}} \altpointTwo} - \f(0) 
		}
		+
		\tfrac{c}{2} \pr*{1-\tfrac{1}{\pnorm2{\altpointTwo}}} \pnorm2{\altpointTwo}^2\\
	&\geq
		\f(0)
		+
		\pr*{
			\br*{ \inf_{\altpoint \in S} \f\pr*{\altpoint} }
			- \f(0) 
			-\tfrac{c}{2} 
		}		
		\pnorm2{\altpointTwo}
		+
		\tfrac{c}{2} 
		\pnorm2{\altpointTwo}^2\\
	&=
		\f(0)
		+
		C
		\pnorm2{\altpointTwo}
		+
		\tfrac{c}{2} 
		\pnorm2{\altpointTwo}^2
		\dott
	\end{split}
	\end{equation}
}
\argument{
	\lref{eq1};
}
{
	that 
	\begin{equation}
	\llabel{eq11}
	\begin{split} 
		\limsup_{ r \to \infty } \pr*{\inf_{ \altpointTwo \in \R^{\d} \backslash B_r} 
			\f(\altpointTwo) 
		}
	&\geq 
		\limsup_{ r \to \infty } \pr*{\inf_{ \altpointTwo \in \R^{\d} \backslash B_r} 
			\f(0)
			+
			C
			\pnorm2{\altpointTwo}
			+
			\tfrac{c}{2} 
			\pnorm2{\altpointTwo}^2
		} \\
	&=
		\limsup_{ r \to \infty } \pr*{\inf_{ s \in (r, \infty) } 
			\f(0)
			+
			C
			s
			+
			\tfrac{cs^2}{2} 
		} \\
	&=
		\infty.
	\end{split}
	\end{equation}
}
\argument{
	\lref{eq11}
}
{
	that there exists $r \in (0,\infty)$ which satisfies that
	\begin{equation}
	\llabel{eq2}
	\begin{split} 
		\inf_{ \altpointTwo \in \R^{\d} } \f(\altpointTwo)
	=
		\inf_{ \altpointTwo \in B_r } \f(\altpointTwo) 
	.
	\end{split}
	\end{equation}
}
\argument{
	\lref{eq2};
	the fact that $B_r$ is compact;
	the assumption that $\f$ is continuous;
}
{
	that there exists $\vartheta \in B_r$ which satisfies that
	\begin{equation}
	\llabel{eq3}
	\begin{split} 
		\f(\vartheta)
	=
		\inf_{ \altpointTwo \in B_r } \f(\altpointTwo) 
	=
		\inf_{ \altpointTwo \in \R^{\d} } \f(\altpointTwo)
	.
	\end{split}
	\end{equation}
}
\argument{
	\lref{eq3};
	\cref{stricly_convex_minima};
	\cref{strongly_convex_functions_are_stricly_convex};
}
{
	that $\vartheta$ is the unique global minimum point of $\f$\cfadd{def:global_minimum}
}
\end{aproof}
\endgroup

\cfclear
\begingroup
\providecommand{\d}{}
\renewcommand{\d}{\defaultParamDim}
\providecommand{\f}{}
\renewcommand{\f}{\defaultLossFunction}
\providecommand{\g}{}
\renewcommand{\g}{\defaultGradientFunction}
\providecommandordefault{\v}{\altpointTwo}
\providecommandordefault{\w}{\altpointThree}
\begin{athm}{prop}{prop:generalized_convex_equivalence_1}[Equivalence for differentiable generalized-convex functions]
	Let 
		$\d \in \N$,
		$c \in \R$,
		$\f \in C^1(\R^{\d}, \R)$.
	Then the following six statements are equivalent:
	\begin{enumerate}[label=(\roman*)]
		\item \label{prop:generalized_convex_equivalence_1:item1} It holds that $\f$ is $c$-generalized-convex\cfadd{def:generalized_convex} \cfout.
		\cfclear
		\item \label{prop:generalized_convex_equivalence_1:item2} It holds for all $\altpointTwo, \altpointThree \in \R^{\d}$ that
		\begin{equation}
			\f(\altpointTwo) 
		\geq 
			\f(\altpointThree) 
		+
			\scp{(\nabla\f)(\altpointThree), \altpointTwo - \altpointThree}
		+
			\tfrac{c}{2} \pnorm2{\altpointTwo - \altpointThree}^2
		\end{equation}
		\cfout.
		\cfclear
		\item \label{prop:generalized_convex_equivalence_1:item3} It holds for all $\altpointTwo, \altpointThree \in \R^{\d}$ that
		\begin{equation}
			\scp{
				(\nabla \f)(\altpointTwo) - (\nabla \f)(\altpointThree), \altpointTwo - \altpointThree
			} 
			\geq 
			c \pnorm2{\altpointTwo - \altpointThree}^2
		\end{equation}
		\cfout.
		\cfclear
		\item \label{prop:generalized_convex_equivalence_1:item4} It holds for all $\altpoint \in \R^{\d}$ that $(\nabla \f)(\altpoint) - c \altpoint$ is a subgradient of $\R^{\d} \ni \altpointTwo \mapsto \f(\altpointTwo) - \tfrac{c}{2} \pnorm2{\altpointTwo}^2 \in \R$ at $\altpoint$\cfadd{def:subgradient} \cfout.
		\cfclear
		\item \label{prop:generalized_convex_equivalence_1:item5} It holds that $\nabla \f$ is $c$-monotonically increasing\cfadd{def:monotonically_increasing_2} \cfout.
		\cfclear
		\item \label{prop:generalized_convex_equivalence_1:item6} It holds that $-\nabla \f$ is $c$-monotonically decreasing\cfadd{def:monotonically_decreasing_2} \cfout.
	\end{enumerate}
\end{athm}
\begin{aproof}
We first prove that (\ref{prop:generalized_convex_equivalence_1:item1} $\rightarrow$ \ref{prop:generalized_convex_equivalence_1:item2}).
For this assume that $\f$ is \generalizedConvex{c}.
\argument{
	the assumption that $\f$ is \generalizedConvex{c};
	\cref{lemma:generalized_convex_equivalence}
}
{
	that for all
		$\v, \w \in \R^{\d}$,
		$t \in (0,1)$
	it holds that
	\begin{equation}
		\llabel{eq0}
		\f(\w + t(\v - \w)) 
	\leq 
		\f(\w) + t(\f(\v) -\f(\w))- \tfrac{c}{2} \br*{ t (1-t) \pnorm2{\w - \v}^2}.
	\end{equation}
}
\argument{
	\lref{eq0};
}
{
	that for all
		$\v, \w \in \R^{\d}$,
		$t \in (0,1)$
	it holds that
	\begin{equation}
		\llabel{eq1}
		\f(\v)
	\geq
		\f(\w)
		+
		\frac{
			\f(\w + t(\v - \w)) 
			- 
			\f(\w)
		}{t} 
		+ 
		\tfrac{c}{2} \br*{(1-t) \pnorm2{\v - \w}^2}
	\end{equation}
}
\argument{
	\lref{eq1};
	the assumption that $\f$ is differentiable
}
{
	that for all
		$\v, \w \in \R^{\d}$
	it holds that
	\begin{equation}
	\begin{split}
		\llabel{eq2}
		\f(\v) 
	&\geq
		\f(\w)
		+
		\limsup_{t \to 0} \pr*{
			\frac{
				\f(\w + t(\v - \w)) 
				- 
				\f(\w)
			}{t} 
			+ 
			\tfrac{c}{2} \br*{(1-t) \pnorm2{\v - \w}^2}
		}\\
	&=
		\f(\w)
		+
		\scp{
			(\nabla \f)(\w), \v - \w
		}
		+
		\tfrac{c}{2} \pnorm2{\v - \w}^2.
	\end{split}
	\end{equation}
}
This proves that (\ref{prop:generalized_convex_equivalence_1:item1} $\rightarrow$ \ref{prop:generalized_convex_equivalence_1:item2}).

In the next step we prove that (\ref{prop:generalized_convex_equivalence_1:item2} $\rightarrow$ \ref{prop:generalized_convex_equivalence_1:item3}).
For this assume that for all 
	$\v, \w \in \R^{\d}$ 
it holds that
\begin{equation}
	\llabel{eq3}
	\f(\altpointTwo) 
\geq 
	\f(\altpointThree) 
+
	\scp{(\nabla\f)(\altpointThree), \altpointTwo - \altpointThree}
+
	\tfrac{c}{2} \pnorm2{\altpointTwo - \altpointThree}^2.
\end{equation}
\startnewargseq
\argument{
	\lref{eq3};
}
{
	that for all
		$\altpointTwo, \altpointThree \in \R^{\d}$
	it holds that
	\begin{equation}
	\begin{split}
		\llabel{eq5}
		\f(\altpointTwo) + \f(\altpointThree)
	&\geq	
		\f(\altpointThree) 
		+ 
		\scp{
			(\nabla \f)(\altpointThree), \altpointTwo - \altpointThree
		}
		+
		\tfrac{c}{2} \pnorm2{\altpointTwo - \altpointThree}^2\\
	&\quad
		+ 
		\f(\altpointTwo) 
		+ 
		\scp{
			(\nabla \f)(\altpointTwo), \altpointThree - \altpointTwo
		}
		+
		\tfrac{c}{2} \pnorm2{\altpointThree - \altpointTwo}^2
		\\
	&=
		\f(\altpointTwo) + \f(\altpointThree)
		-
		\scp{
			(\nabla \f)(\altpointTwo) - (\nabla \f)(\altpointThree), \altpointTwo - \altpointThree
		}
		+
		c \pnorm2{\altpointThree - \altpointTwo}^2.
	\end{split}
	\end{equation}
}
\argument{
	\lref{eq5};
}
{
	that for all
		$\altpointTwo, \altpointThree \in \R^{\d}$
	it holds that
	\begin{equation}
		\llabel{eq6}
		\scp{
			(\nabla \f)(\altpointTwo) - (\nabla \f)(\altpointThree), \altpointTwo - \altpointThree
		}
	\geq 
		c \pnorm2{\altpointTwo - \altpointThree}^2.
	\end{equation}
}
This proves that (\ref{prop:generalized_convex_equivalence_1:item2} $\rightarrow$ \ref{prop:generalized_convex_equivalence_1:item3}).

In the next step we prove that (\ref{prop:generalized_convex_equivalence_1:item3} $\rightarrow$ \ref{prop:generalized_convex_equivalence_1:item1}).
For this assume that for all 
	$\v, \w \in \R^{\d}$ 
it holds that
\begin{equation}
	\llabel{eq6.1}
	\scp{
		(\nabla \f)(\altpointTwo) - (\nabla \f)(\altpointThree), \altpointTwo - \altpointThree
	}
\geq 
	c \pnorm2{\altpointTwo - \altpointThree}^2.
\end{equation}
\startnewargseq
\argument{
	\lref{eq6.1};
}
{
	that for all 
		$\altpoint, \altpointTwo \in \R^{\d}$,
		$\alpha, \beta \in \R$
	with 
		$\alpha > \beta$
	it holds that
	\begin{equation}
	\begin{split}
		\llabel{eq7}
		&\scp{
			(\nabla \f)(\altpoint + \alpha \altpointTwo) - (\nabla \f)(\altpoint + \beta \altpointTwo)
			, 
			\altpointTwo
		} \\
	&=
		(\alpha - \beta)^{-1}
		\scp{
			(\nabla \f)(\altpoint + \alpha \altpointTwo) - (\nabla \f)(\altpoint + \beta \altpointTwo)
			, 
			(\alpha - \beta) \altpointTwo
		}  \\
	&\geq
		(\alpha - \beta)^{-1}
		c \pnorm2{(\alpha - \beta) \altpointTwo}^2
	=
		(\alpha - \beta) c \pnorm2{\altpointTwo}^2.
	\end{split}
	\end{equation}
}
\argument{
	\lref{eq7};
	the fundamental theorem of calculus
}
{
	that for all 
		$\altpoint, \altpointTwo \in \R^{\d}$,
		$t \in (0,1)$
	it holds that
	\begin{equation}
	\llabel{eq8}
	\begin{split}
		&t\pr[\big]{\f(\altpoint + \altpointTwo) - \f(\altpoint + t \altpointTwo)}
		-
		(1-t)\pr[\big]{\f(\altpoint + t \altpointTwo) - \f(\altpoint)}\\
	&=
		t
		\pr*{
			\int_t^1
				\scp{
					(\nabla \f)(\altpoint + s \altpointTwo), \altpointTwo
				}
			\,ds
		}
		-
		(1-t)
		\pr*{
			\int_0^t
				\scp{
					(\nabla \f)(\altpoint + s \altpointTwo), \altpointTwo
				}
			\,ds
		}\\
	&=
		t (1-t)
		\pr*{
			\int_0^1
				\scp{
					(\nabla \f)(\altpoint + (t + s(1-t)) \altpointTwo), \altpointTwo
				}
			\,ds
		} \\
	&\quad
		-
		(1-t) t
		\pr*{
			\int_0^1
				\scp{
					(\nabla \f)(\altpoint + s t \altpointTwo), \altpointTwo
				}
			\,ds
		}\\
	&=
		t (1-t)
		\pr*{
			\int_0^1
				\scp{
					(\nabla \f)(\altpoint + (t + s(1-t)) \altpointTwo) - (\nabla \f)(\altpoint + s t \altpointTwo)
					, 
					\altpointTwo
				}
			\,ds
		} \\
	&\geq
		t (1-t)
		\pr*{
			\int_0^1
				(t+s - 2st) c \pnorm2{\altpointTwo}^2
			\,ds
		} \\
	&=
		t (1-t) (t + \tfrac{1}{2} - t)
		c \pnorm2{\altpointTwo}^2
	=
		\tfrac{c}{2} \br*{ t (1-t) \pnorm2{\altpointTwo}^2 }
	\end{split}
	\end{equation}	
}
\argument{
	\lref{eq8};
	\cref{lemma:generalized_convex_equivalence}
}
{
	that $\f$ is \generalizedConvex{c} \cfload.
}
This proves that (\ref{prop:generalized_convex_equivalence_1:item3} $\rightarrow$ \ref{prop:generalized_convex_equivalence_1:item1}).

\startnewargseq
\argument{
	\cref{lemma:convex_subgradients};
}
{
	that
	(\ref{prop:generalized_convex_equivalence_1:item1} $\leftrightarrow$ \ref{prop:generalized_convex_equivalence_1:item4}).
}
\argument{
	\cref{def:monotonically_increasing_2:eq1};
}
{
	that
	(\ref{prop:generalized_convex_equivalence_1:item3} $\leftrightarrow$ \ref{prop:generalized_convex_equivalence_1:item5}).
}
\argument{
	\cref{lemma:increasing_decreasing_equivalence_2};
}
{
	that
	(\ref{prop:generalized_convex_equivalence_1:item5} $\leftrightarrow$ \ref{prop:generalized_convex_equivalence_1:item6}).
}
\end{aproof}
\endgroup

\cfclear
\begingroup
\providecommand{\d}{}
\renewcommand{\d}{\defaultParamDim}
\providecommand{\f}{}
\renewcommand{\f}{\defaultLossFunction}
\providecommand{\g}{}
\renewcommand{\g}{\defaultGradientFunction}
\providecommandordefault{\v}{\altpointTwo}
\providecommandordefault{\w}{\altpointThree}
\begin{athm}{prop}{prop:generalized_convex_equivalence_3}[Equivalence for two times differentiable generalized-convex functions]
	Let 
		$\d \in \N$,
		$c \in \R$,
		$\f \in \C^2(\R^{\d}, \R)$.
	Then the following two statements are equivalent:
	\begin{enumerate}[label=(\roman*)]
		\item \label{prop:generalized_convex_equivalence_3:item1} It holds that $\f$ is $c$-generalized-convex\cfadd{def:generalized_convex} \cfout.
		\cfclear 
		\item \label{prop:generalized_convex_equivalence_3:item2} It holds for all $\altpointTwo, \altpointThree \in \R^{\d}$ that
		\begin{equation}
			\scp{
				(\Hess \f)(\altpointTwo) \altpointThree, \altpointThree
			}
			\geq
				c \pnorm2{\altpointThree}^2
		\end{equation}
		\cfout.
	\end{enumerate}
\end{athm}
\begin{aproof}
\argument{
	\cref{lemma:increasing_jacobian_equivalence,prop:generalized_convex_equivalence_1};
	the fact that $\nabla \f$ is continuously differentiable;
}[verbs=ep]
{
	that
	(\ref{prop:generalized_convex_equivalence_3:item1} $\leftrightarrow$ \ref{prop:generalized_convex_equivalence_3:item2}).
}
\end{aproof}
\endgroup

\subsection{Coercivity}

\cfclear
\begingroup
\providecommand{\d}{}
\renewcommand{\d}{\defaultParamDim}
\providecommand{\f}{}
\renewcommand{\f}{\defaultLossFunction}
\providecommand{\g}{}
\renewcommand{\g}{\defaultGradientFunction}
\begin{athm}{definition}{def:coercivity}[Coercivity-type conditions]
	Let 
		$\d \in \N$,
		$\vartheta \in \R^{\d}$,
		$c \in (0,\infty)$,
	let 
		$O \subseteq \R^\d$ 
	be open, 
	and let
		$\f \colon O \to \R$
	be a function.
	Then we say that 
		$\f$
	satisfies a coercivity-type condition 
	with coercivity constant
		$c$
	at 
		$\vartheta$
	if and only if 
	\begin{enumerate}[(i)]
	\item 
	it holds that $ \f $ is differentiable and 
	\item  
	it holds for all 
		$\theta \in O$
	that
	\begin{equation}
	\label{def:coercivity:eq1}
	\begin{split}
		\scp{\theta-\vartheta, (\nabla \f)(\theta) }\geq  c\pnorm2{\theta-\vartheta}^2
	\end{split}
	\end{equation}
	\end{enumerate}
	\cfload.
\end{athm}
\endgroup

\cfclear
\begingroup
\providecommand{\d}{}
\renewcommand{\d}{\defaultParamDim}
\providecommand{\f}{}
\renewcommand{\f}{\defaultLossFunction}
\providecommand{\g}{}
\renewcommand{\g}{\defaultGradientFunction}
\begin{athm}{definition}{def:coercivityI}[Coercive-type functions]
	Let 
		$\d \in \N$
	and let 
		$\f \colon \R^{\d} \to \R$
	be a function. 
	Then we say that $\f$ is a coercive-type function 
	if and only if
	there exist
		$\vartheta \in \R^{\d}$,
		$c \in (0,\infty)$
	such that it holds that\cfadd{def:coercivity} $\f$ satisfies a coercivity-type condition at $\vartheta$ with coercivity constant $c$
	\cfload.
\end{athm}
\endgroup

\cfclear
\begingroup
\providecommand{\d}{}
\renewcommand{\d}{\defaultParamDim}
\providecommand{\f}{}
\renewcommand{\f}{\defaultLossFunction}
\providecommand{\g}{}
\renewcommand{\g}{\defaultGradientFunction}
\providecommandordefault{\v}{\altpointTwo}
\providecommandordefault{\w}{\altpointThree}
\begin{athm}{cor}{lemma:strongly_convex_coercive_exact}[Strongly convex functions are coercive]
	Let 
		$\d \in \N$,
		$c \in (0,\infty)$
	and let
		$\f \in \C^1(\R^{\d}, \R)$
	be \generalizedConvex{c}
	\cfload.
	Then 
	\begin{enumerate}[label=(\roman*)]
		\item 
		\label{prop:strongly_convex_coercive_exact:item1}
		there exists a unique $\vartheta \in \R^{\d}$ such that $\vartheta$ is a global minimum point of $\f$\cfadd{def:global_minimum} and
		
		\item 
		\label{prop:strongly_convex_coercive_exact:item2}
		it holds that $\f$ satisfies a coercivity-type condition at $\vartheta$ with coercivity constant $c$\cfadd{def:coercivity}
	\end{enumerate}
	\cfout.
\end{athm}

\begin{aproof}
\argument{
	\cref{lemma:strongly_convex_minimum_point};
	the assumption that $\f$ is \generalizedConvex{c} and continuous;
}
{
	that there exists a unique $\vartheta \in \R^{\d}$ which satisfies that $\vartheta$ is a global minimum point of $\f$\cfadd{def:global_minimum} 
}
This establishes \cref{prop:strongly_convex_coercive_exact:item1}.
\argument{
	\cref{prop:generalized_convex_equivalence_1:item3} in \cref{prop:generalized_convex_equivalence_1};
	the assumption that $\f$ is \generalizedConvex{c};
}
{
	that for all $\altpointTwo, \altpointThree \in \R^{\d}$ it holds that
	\begin{equation}
		\llabel{eq1}
		\scp{
			(\nabla \f)(\altpointTwo) - (\nabla \f)(\altpointThree), \altpointTwo - \altpointThree
		} 
		\geq 
		c \pnorm2{\altpointTwo - \altpointThree}^2
	\end{equation}
}
\argument{
	\cref{cor:necessary_condition_local_minimum};
}
{
	that 
	\begin{equation}
	\llabel{eq0}
	\begin{split} 
		(\nabla \f)(\vartheta) = 0.
	\end{split}
	\end{equation}
}
\argument{
	\lref{eq0};
	\lref{eq1};
}
{
	that it holds for all 
		$\theta \in \R^{\d}$
	that
	\begin{equation}
	\llabel{eq2}
	\begin{split}
			\scp{\theta-\vartheta, (\nabla \f)(\theta) }
		=
			\scp{\theta-\vartheta, (\nabla \f)(\theta) - (\nabla \f)(\vartheta)}
		\geq 
			c \pnorm2{\theta-\vartheta}^2.
	\end{split}
	\end{equation}
}
\argument{
	\lref{eq2};
	\cref{def:coercivity:eq1}
}
{
	that $\f$ satisfies a coercivity-type condition at $\vartheta$ with coercivity constant $c$\cfadd{def:coercivity}
}
\end{aproof}
\endgroup

\cfclear
\begingroup
\providecommand{\d}{}
\renewcommand{\d}{\defaultParamDim}
\providecommand{\f}{}
\renewcommand{\f}{\defaultLossFunction}
\providecommand{\g}{}
\renewcommand{\g}{\defaultGradientFunction}
\providecommandordefault{\v}{\altpointTwo}
\providecommandordefault{\w}{\altpointThree}
\begin{athm}{cor}{lemma:strongly_convex_coercive}
	Let 
		$\d \in \N$,
		$\vartheta \in \R^{\d}$ 
	and
	let 
		$\f \in \C^1(\R^{\d}, \R)$
	be \stronglyConvex{} \cfload.
	Then it holds that $\f$ is a coercive-type function\cfadd{def:coercivityI} \cfout.
\end{athm}
\begin{aproof}
\argument{
	\cref{lemma:strongly_convex_coercive_exact}
}
{
	that $\f$ is a coercive-type function\cfadd{def:coercivityI}
}
\end{aproof}
\endgroup

\cfclear
\begingroup
\providecommand{\d}{}
\renewcommand{\d}{\defaultParamDim}
\providecommand{\f}{}
\renewcommand{\f}{\defaultLossFunction}
\providecommand{\g}{}
\renewcommand{\g}{\defaultGradientFunction}
\begin{lemma}[A sufficient condition for a local minimum point]
\label{fcond1}
Let $\d \in \N$, 
$ c \in (0,\infty) $, 
$ r \in (0,\infty] $, 
$ \vartheta \in \R^{ \d } $, 
$ \mathbb{B} = \{ w \in \R^\d \colon \pnorm2{w-\vartheta} \leq r \} $, 
$ \f \in C^1( \R^\d, \R ) $ 
satisfy for all $ \theta \in \mathbb{B} $ that
\begin{equation}
\label{fcond1:assumption1}
\scp{\theta-\vartheta, (\nabla \f)(\theta) } \geq  c\pnorm2{\theta-\vartheta}^2\ifnocf.
\end{equation}
\cfload[.] 
Then 
\begin{enumerate}[label=(\roman *)]
\item \label{fcond1:item1}
it holds 
for all $ \theta \in \mathbb{B} $ 
that 
$ 
  \f(\theta)-\f(\vartheta) \geq \tfrac{c}{2}\pnorm2{\theta-\vartheta}^2
$,
\item \label{fcond1:item2}
it holds that
$
  \{\theta \in \mathbb{B}  \colon  \f(\theta)= \inf\nolimits_{w \in \mathbb{B}} \f(w)  \} = \{\vartheta \}
$,
and
\item \label{fcond1:item3}
it holds that $(\nabla \f)(\vartheta) = 0$.
\end{enumerate}
\end{lemma}

\begin{proof}[Proof of \cref{fcond1}]
Throughout this proof, let $B$ be the set given by
\begin{equation}
B = \{w \in \R^\d \colon \pnorm2{w-\vartheta} < r \}.
\end{equation}  
Note that \eqref{fcond1:assumption1}
implies that for all $v \in \R^\d$ with $\pnorm2{v} \leq r$ it holds that
\begin{equation}
\scp{(\nabla \f)(\vartheta + v), v } \geq  c\pnorm2{v}^2.
\end{equation}
The fundamental theorem of calculus hence demonstrates that for all $\theta \in \mathbb{B}$ it holds that
\begin{equation}
\label{eq:f_estimate_from_below}
\begin{split}
\f(\theta)-\f(\vartheta) 
&= 
\br[\big]{ \f(\vartheta + t(\theta - \vartheta)) }_{t = 0}^{t = 1}  
\\&=
\int_0^1 \f'(\vartheta + t(\theta-\vartheta))(\theta-\vartheta) \, \diff t  
\\&= 
\int_0^1 \scp{(\nabla \f)(\vartheta + t(\theta-\vartheta)),t(\theta-\vartheta) } \frac{1}{t} \,\diff t 
\\& \geq 
\int_0^1c \pnorm2{t(\theta-\vartheta)}^2\frac{1}{t} \, \diff t 
= 
c\pnorm2{\theta-\vartheta}^2\br*{\int_0^1t\, \diff t}
= 
\tfrac{c}{2}\pnorm2{\theta-\vartheta}^2
.
\end{split}
\end{equation}
This proves  \cref{fcond1:item1}. 
Next observe that 
\eqref{eq:f_estimate_from_below}
ensures that for all $\theta \in \mathbb{B} \backslash \{\vartheta\}$ it holds that
\begin{equation}
\f(\theta) \geq \f(\vartheta) + \tfrac{c}{2}\pnorm2{\theta-\vartheta}^2 > \f(\vartheta).
\end{equation}
Hence, we obtain for all $\theta \in \mathbb{B} \backslash \{\vartheta\}$ that
\begin{equation}
\inf_{w \in \mathbb{B}} \f(w)  = \f(\vartheta) < \f(\theta).
\end{equation}
This establishes \cref{fcond1:item2}. 
It thus remains thus remains to prove \cref{fcond1:item3}.
For this observe that \cref{fcond1:item2} ensures that
\begin{equation}
\{\theta \in B \colon  \f(\theta)= \inf\nolimits_{w \in B} \f(w)  \} = \{\vartheta \}.
\end{equation}
Combining this, the fact that $B$ is open, and \cref{minimum2} 
(applied with $\d \is \d$, $O \is B$, $\vartheta \is \vartheta$, $\f\is \f|_B $ in the notation of \cref{minimum2}) assures that $(\nabla \f)(\vartheta) = 0$.
This establishes \cref{fcond1:item3}.
The proof of \cref{fcond1} is thus complete.
\end{proof}
\endgroup

\cfclear
\begingroup
\newcommand{\lrnorm}[1]{\apnorm2{#1}}
\newcommand{\norm}[1]{\pnorm2{#1}}
\providecommand{\d}{}
\renewcommand{\d}{\defaultParamDim}
\providecommand{\f}{}
\renewcommand{\f}{\defaultLossFunction}
\providecommand{\g}{}
\renewcommand{\g}{\defaultGradientFunction}
\begin{athm}{example}{lemma:example_coercivity}
Let $\d \in \N$, 
$\xi\in\R^\d$, $\vartheta = (\vartheta_1,\dots,\vartheta_\d) \in \R^\d$, 
 $\kappa, \lambda_1,\lambda_2,\ldots, \lambda_\d \in (0,\infty)$ satisfy
$\kappa = \min\{\lambda_1,\lambda_2,\ldots, \lambda_\d \}$
and let $\f \colon \R^\d \to \R$ satisfy for all $\theta = (\theta_1,\dots,\theta_\d) \in \R^\d$ that 
\begin{equation}
\llabel{assumption0}
\f(\theta) = \tfrac{1}{2} \br*{\sum_{i = 1}^\d\lambda_i \abs{\theta_i - \vartheta_i}^2 }.
\end{equation}
Then 
\begin{enumerate}[label=(\roman*)]
	\item \label{lemma:example_coercivity:item1} it holds that $\f$ is \generalizedConvex{\kappa},
	\item \label{lemma:example_coercivity:item2} it holds that $\f$ is strongly convex\cfadd{def:strongly_convex},
	\item \label{lemma:example_coercivity:item3} it holds that $\f$ satisfies a coercivity-type condition at $\vartheta$ with coercivity constant $\kappa$\cfadd{def:coercivity}, and
	\item \label{lemma:example_coercivity:item4} it holds that $\f$ is a coercive-type function\cfadd{def:coercivityI}
\end{enumerate}
\cfout.
\end{athm}
\begin{aproof}
\argument{
	\lref{assumption0}
}
{
	that for all $\theta = (\theta_1,\dots,\theta_\d) \in \R^\d$ it holds that
	\begin{equation}
		\llabel{eq1}
		(\nabla \f)(\theta) = (\lambda_1(\theta_1-\vartheta_1),\dots,\lambda_\d(\theta_\d-\vartheta_\d)).
	\end{equation}
}
\argument{
	\lref{eq1}
}
{
	that for all $\altpointTwo = (\altpointTwo_1,\dots,\altpointTwo_\d), \altpointThree = (\altpointThree_1,\dots,\altpointThree_\d) \in \R^{\d}$ it holds that
	\begin{equation}
	\llabel{eq2}
	\begin{split}
		\scp{
			(\nabla \f)(\altpointTwo) - (\nabla \f)(\altpointThree), \altpointTwo - \altpointThree
		} 
		&=
		\sum_{i = 1}^\d \lambda_i (\altpointTwo_i-\altpointThree_i)(\altpointTwo_i-\altpointThree_i)\\
		&\geq 
		\kappa \sum_{i = 1}^\d (\altpointTwo_i-\altpointThree_i)^2
		=
		\kappa \pnorm2{\altpointTwo - \altpointThree}^2
	\end{split}
	\end{equation}
}
\argument{
	\lref{eq2};
	\cref{lemma:generalized_convex_equivalence}
}
{
	that\llabel{eq3} $\f$ is \generalizedConvex{\kappa}
}
This establishes \cref{lemma:example_coercivity:item1}.
\startnewargseq
\argument{
	\cref{lemma:example_coercivity:item1};
	the fact that $(\nabla \f)(\vartheta) = 0$
}[verbs=ep]
{
	\cref{lemma:example_coercivity:item2,lemma:example_coercivity:item3,lemma:example_coercivity:item4}.
}
\end{aproof}
\endgroup

\section{Lyapunov-type functions for GFs}
\label{sect:aux_results_flow}

\subsection{Gronwall differential inequalities}

The following lemma, \cref{lem:gronwall_general} below, 
is referred to as a Gronwall inequality in the literature (cf., \eg, Henry~\cite[Chapter~7]{h81}). 
Gronwall inequalities are powerful tools to study dynamical systems and, especially, 
solutions of \ODEs.

\cfclear
\begingroup
\begin{athm}{lemma}{lem:gronwall_general}[Gronwall inequality]
	Let
		$T\in(0,\infty)$,
		$\alpha \in \R$,
		$\epsilon\in C^1([0,T],\R)$,
		$\beta\in C([0,T],\R)$
	satisfy for all
		$t\in[0,T]$
	that
	\begin{equation}
		\epsilon'(t)
		\leq 
		\alpha\epsilon(t)+\beta(t)
		.
	\end{equation}
	Then it holds for all
		$t\in[0,T]$
	that
	\begin{equation}
		\llabel{eq:claim}
		\epsilon(t)
		\leq
		e^{\alpha t} \epsilon(0) 
		+
		\int_0^t e^{\alpha(t-s)}\beta(s)\,\diff s
		.
	\end{equation}
\end{athm}
\begin{aproof}
	Throughout this proof,
		let
			$v\colon [0,T]\to\R$
		satisfy for all
			$t\in[0,T]$
		that
		\begin{equation}
			\llabel{eq:defv}
			v(t)
			=
			e^{\alpha t}\br*{\int_0^t e^{-\alpha s}\beta(s)\,\diff s}
		\end{equation}
		and let
			$u\colon[0,T]\to\R$
		satisfy for all
			$t\in[0,T]$
		that
		\begin{equation}
			\llabel{eq:defu}
			u(t)
			=
			\br*{\epsilon(t)-v(t)}e^{-\alpha t}
			.
		\end{equation}
	\Nobs that
		the product rule and the fundamental theorem of calculus
	demonstrate that
	for all
		$t\in[0,T]$
	it holds that
	$v\in C^1([0,T],\R)$ and
	\begin{equation}
		v'(t)
		=
		\alpha e^{\alpha t}\br*{\int_0^t e^{-\alpha s}\beta(s)\,\diff s}
		+
		e^{\alpha t}\br*{e^{-\alpha t}\beta(t)}
		=
		\alpha v(t)+\beta(t)
		.
	\end{equation}
		The assumption that 
			$\epsilon\in C^1([0,T],\R)$
		and the product rule
		\hence
	ensure that for all
		$t\in[0,T]$
	it holds that
		$u\in C^1([0,T],\R)$
	and
	\begin{eqsplit}
		u'(t)
		&=
		\br*{\epsilon'(t)-v'(t)}e^{-\alpha t}
		-\br*{\epsilon(t)-v(t)}\alpha e^{-\alpha t}
		\\&=
		[\epsilon'(t)-v'(t)-\alpha \epsilon(t)+\alpha v(t)]e^{-\alpha t}
		\\&=
		[\epsilon'(t)-\alpha v(t)-\beta(t)-\alpha \epsilon(t)+\alpha v(t)]e^{-\alpha t}
		\\&=
		[\epsilon'(t)-\beta(t)-\alpha \epsilon(t)]e^{-\alpha t}
		.
	\end{eqsplit}
	Combining
		this
	with 
		the assumption that
			for all
				$t\in[0,T]$
			it holds that
				$\epsilon'(t)\leq \alpha \epsilon(t)+\beta(t)$
	proves that for all
		$t\in[0,T]$
	it holds that
	\begin{equation}
		u'(t)
		\leq
		[\alpha \epsilon(t)+\beta(t)-\beta(t)-\alpha \epsilon(t)]e^{-\alpha t}
		=
		0
		.
	\end{equation}
		This
		and the fundamental theorem of calculus
	imply that for all
		$t\in[0,T]$
	it holds that
	\begin{equation}
		u(t)
		=
		u(0)+\int_0^t u'(s)\,\diff s
		\leq
		u(0)+\int_0^t 0\,\diff s
		=
		u(0)
		=
		\epsilon(0)
		.
	\end{equation}
	Combining
		\enum{
			this;
			\lref{eq:defv};
			\lref{eq:defu}
		}
	shows that for all
		$t\in[0,T]$
	it holds that
	\begin{equation}
		\epsilon(t)
		=
		e^{\alpha t} u(t) + v(t)
		\leq 
		e^{\alpha t} \epsilon(0)+v(t)
		=
		e^{\alpha t} \epsilon(0)+\int_0^t e^{\alpha(t-s)}\beta(s)\,\diff s
		.
	\end{equation}
\end{aproof}
\endgroup

\subsection{Lyapunov-type functions for ODEs}
\label{subsec:Lyapunov_flow}

\cfclear
\begingroup
\providecommand{\d}{}
\renewcommand{\d}{\defaultParamDim}
\providecommand{\g}{}
\renewcommand{\g}{\defaultGradientFunction}
\providecommand{\a}{}
\renewcommand{\a}{\alpha}
\providecommand{\b}{}
\renewcommand{\b}{\beta}
\begin{athm}{prop}{lem:lyapunov_general}[Lyapunov-type functions for \ODEs]
	Let
		$\d\in\N$,
		$T\in(0,\infty)$,
		$\a \in \R$,
	let
		$O\subseteq\R^\d$
		be open,
	let 
		$\b\in C(O,\R)$,
		$\g\in C(O,\R^\d)$,
		$V\in C^1(O,\R)$
	satisfy for all
		$\theta\in O$
	that
	\begin{equation}
		\llabel{eq:assum}
		V'(\theta)\g(\theta)
		=
		\scp{(\nabla V)(\theta),\g(\theta)}
		\leq
		\a V(\theta)+\b(\theta),
	\end{equation}
	and let
		$\Theta\in C([0,T],O)$
	satisfy for all
		$t\in[0,T]$
	that
		$\Theta_t=\Theta_0+\int_0^t \g(\Theta_s)\,\diff s$
	\cfout.
	Then it holds for all
		$t\in[0,T]$
	that
	\begin{equation}
		V(\Theta_t)
		\leq
		e^{\a t}
		V(\Theta_0)
		+
		\int_0^t e^{\a (t-s)} \b(\Theta_s)\,\diff s
		.
	\end{equation}
\end{athm}
\begin{aproof}
	\providecommand{\aa}{}
	\renewcommand{\aa}{a}
	\providecommand{\bb}{}
	\renewcommand{\bb}{b}
	Throughout this proof,
		let
			$\epsilon,\bb\in C([0,T],\R)$
		satisfy for all
			$t\in [0,T]$
		that 
		\begin{equation}
			\llabel{eq:defs}
			\epsilon(t)=V(\Theta_t)
			\qandq
			\bb(t)=\b(\Theta_t)
			.
		\end{equation}
	\Nobs that 
		\lref{eq:assum},
		\lref{eq:defs},
		the fundamental theorem of calculus,
		and the chain rule
	ensure that	for all
		$t\in[0,T]$
	it holds that
	\begin{equation}
		\epsilon'(t)
		=
		\tfrac{\diff}{\diff t}(V(\Theta_t))
		=
		V'(\Theta_t){\dot \Theta_t}
		=
		V'(\Theta_t)\g(\Theta_t)
		\leq
		\a V(\Theta_t)+\b(\Theta_t)
		=
		\a \epsilon(t)+\bb(t)
		.
	\end{equation}
		\Cref{lem:gronwall_general} and
		\lref{eq:defs}
		\hence
	demonstrate that for all
		$t\in[0,T]$
	it holds that
	\begin{eqsplit}
		V(\Theta_t)
		&=
		\epsilon(t)
		\leq
		e^{\alpha t}\epsilon(0)+\int_0^t e^{\alpha (t-s)}\bb(s)\,\diff s
		=
		e^{\alpha t}V(\Theta_0)+\int_0^t e^{\alpha (t-s)} \b(\Theta_s)\,\diff s
		.
	\end{eqsplit}
\end{aproof}
\endgroup

\begingroup
\cfclear
\providecommand{\d}{}
\renewcommand{\d}{\defaultParamDim}
\providecommand{\g}{}
\renewcommand{\g}{\defaultGradientFunction}
\begin{cor}
\label{Lyapunov}
Let $\d \in \N$, $T \in (0,\infty)$, $\alpha \in \R$, 
let $O \subseteq \R^\d$ be open, 
let $\g \in C(O,\R^\d)$, $V \in C^1(O,\R)$ satisfy for all $\theta \in O$ that
\begin{equation}
\label{Lyapunov:ass1}
V'(\theta) \g(\theta) = \scp{(\nabla V)(\theta) , \g(\theta) } \leq \alpha V(\theta),
 \end{equation}
and let $\Theta \in C([0,T], O)$ satisfy for all $t \in [0,T]$ that $ \Theta_t = \Theta_0 + \int_0^t \g(\Theta_s) \, \diff s$
\cfload.
Then it holds for all $t \in [0,T]$ that
\begin{equation}
\label{Lyapunov:concl1}
V(\Theta_t) \leq e^{\alpha t} V(\Theta_0).
\end{equation}
\end{cor}

\begin{proof}[Proof of \cref{Lyapunov}]
\Nobs that
\enum{
	\cref{lem:lyapunov_general};
	\eqref{Lyapunov:ass1}
}[establish]
\eqref{Lyapunov:concl1}.
The proof of \cref{Lyapunov} is thus complete.
\end{proof}
\endgroup

\subsection{On Lyapunov-type functions and coercivity-type conditions}
\label{subsec:Lyapunov_coercivity}

\cfclear
\begingroup
\providecommand{\d}{}
\renewcommand{\d}{\defaultParamDim}
\providecommand{\f}{}
\renewcommand{\f}{V}
\begin{lemma}[Derivative of the standard norm]
\label{der_of_norm}
Let $\d \in \N$, $\vartheta \in \R^\d$
and let $\f \colon \R^\d \to \R$ satisfy for all $\theta \in \R^\d$ that
\begin{equation}
\f(\theta) = \pnorm2{\theta- \vartheta}^2\ifnocf.
\end{equation}
\cfload[.]
Then it holds for all $\theta \in \R^\d$ that $\f \in C^{\infty}(\R^\d, \R)$  and
\begin{equation}
(\nabla \f)(\theta) = 2 (\theta - \vartheta).
\end{equation}
\end{lemma}

\begin{proof}[Proof of \cref{der_of_norm}]
Throughout this proof, let $\vartheta_1,\ldots, \allowbreak\vartheta_\d \in \R$ satisfy  
$\vartheta = (\vartheta_1,\ldots,\allowbreak \vartheta_\d)$.
Note that the fact that for all $\theta = (\theta_1,\ldots,\theta_\d) \in \R^\d$ it holds that
\begin{equation}
\f(\theta) = \sum_{i = 1}^\d \pr{\theta_i - \vartheta_i}^2
\end{equation}
implies that for all $\theta = (\theta_1,\ldots,\theta_\d) \in \R^\d$ it holds that $\f \in C^{\infty}(\R^\d, \R)$ and
\begin{equation}
(\nabla \f)(\theta) =
\begin{pmatrix}
\bpr{\tfrac{\partial \f}{ \partial \theta_1}}(\theta) \\
\vdots \\
\bpr{\tfrac{\partial \f}{ \partial \theta_\d}}(\theta)
\end{pmatrix} 
=
\begin{pmatrix}
2(\theta_1- \vartheta_1)\\
\vdots \\
2(\theta_\d- \vartheta_\d)
\end{pmatrix} 
=
2(\theta-\vartheta).
\end{equation}
The proof of \cref{der_of_norm} is thus complete.
\end{proof}
\endgroup

In the next result, \cref{Lyapunov_cor} below, we establish an error analysis for \GFs\ in which the objective function satisfies a coercivity-type condition in the sense of \cref{def:coercivity}.

\cfclear
\begingroup
\providecommand{\d}{}
\renewcommand{\d}{\defaultParamDim}
\providecommand{\f}{}
\renewcommand{\f}{\defaultLossFunction}
\providecommand{\g}{}
\renewcommand{\g}{\defaultGradientFunction}
\begin{cor}[On quadratic Lyapunov-type functions and coercivity-type conditions]
\label{Lyapunov_cor}
Let $\d \in \N$, $c \in \R$, $T \in (0,\infty)$, $\vartheta \in \R^\d$, 
let $O \subseteq \R^\d$ be open, 
let $\f \in C^1(O,\R)$ satisfy for all $\theta \in O$ that 
\begin{equation}
\label{Lyapunov_cor:ass1}
\scp{\theta-\vartheta, (\nabla \f)(\theta) }\geq  c\pnorm2{\theta-\vartheta}^2,
\end{equation}
and let $\Theta\in C([0,T], O)$ satisfy for all $t \in [0,T]$ that $ \Theta_t = \Theta_0 - \int_0^t (\nabla \f)(\Theta_s) \, \diff s$
\cfload.
Then it holds for all $t \in [0,T]$ that
\begin{equation}
\pnorm2{\Theta_t-\vartheta} \leq e^{-c t} \pnorm2{\Theta_0-\vartheta}.
\end{equation}
\end{cor}

\begin{proof}[Proof of \cref{Lyapunov_cor}]
Throughout this proof, let $\g \colon O \to \R^\d$ satisfy 
for all $ \theta \in O $ that 
\begin{equation}
\g(\theta) = - (\nabla \f)(\theta)
\end{equation}
and let $ V \colon O \to \R $ satisfy for all $ \theta \in O $ that 
\begin{equation}
  V(\theta) = \pnorm2{\theta - \vartheta}^2 .
\end{equation}
Observe that \cref{der_of_norm} and \eqref{Lyapunov_cor:ass1} 
ensure that 
for all $ \theta \in O $ it holds that $ V \in C^1(O,\R) $ 
and 
\begin{equation}
\begin{split}
V'(\theta) \g(\theta) 
&= 
\scp{(\nabla V)(\theta) , \g(\theta)}
= 
\scp{2(\theta - \vartheta) , \g(\theta) }\\
&=
-2\scp{\theta - \vartheta , (\nabla \f)(\theta) }
\leq 
-2c \pnorm2{\theta-\vartheta}^2 
= 
-2c V(\theta).
\end{split}
\end{equation}
\cref{Lyapunov} hence proves that for all $t \in [0,T]$ it holds that
\begin{equation}
\pnorm2{\Theta_t-\vartheta}^2 
= 
V(\Theta_t) \leq e^{-2c t} \, V(\Theta_0) 
=  
e^{-2c t}\,\pnorm2{\Theta_0-\vartheta}^2
.
\end{equation}
The proof of \cref{Lyapunov_cor} is thus complete.
\end{proof}
\endgroup

\subsection{On a linear growth condition}

\todoc{Introduce $L$-smooth}

\cfclear
\begingroup
\providecommand{\d}{}
\renewcommand{\d}{\defaultParamDim}
\providecommand{\f}{}
\renewcommand{\f}{\defaultLossFunction}
\providecommand{\g}{}
\renewcommand{\g}{\defaultGradientFunction}
\begin{lemma}[On a linear growth condition]
\label{fcond2}
Let $\d \in \N$,
$L \in \R$, 
$r \in (0,\infty]$, 
$\vartheta \in \R^\d$, 
$\mathbb{B} = \{w \in \R^\d \colon \pnorm2{w-\vartheta} \leq r \}$, $\f\in C^1(\R^\d,\R)$ satisfy for all $\theta \in \mathbb{B}$ that
\begin{equation}
\label{fcond2:assumption1}
\pnorm2{(\nabla \f)(\theta)} \leq L \pnorm2{\theta-\vartheta}\ifnocf.
\end{equation}
\cfload[.]
Then it holds for all $\theta \in \mathbb{B}$ that 
\begin{equation}
\f(\theta)-\f(\vartheta) \leq \tfrac{L}{2} \pnorm2{\theta-\vartheta}^2.
\end{equation}
\end{lemma}

\begin{proof}[Proof of \cref{fcond2}]
Observe that (\ref{fcond2:assumption1}), 
the Cauchy-Schwarz inequality, 
and the fundamental theorem of calculus 
ensure that for all $\theta \in \mathbb{B}$ it holds that
\begin{equation}
\begin{split}
\f(\theta)-\f(\vartheta) 
&= 
\bbr{ \f(\vartheta + t(\theta - \vartheta)) }_{t = 0}^{t = 1}
\\&= 
\int_0^1 \f'(\vartheta + t(\theta-\vartheta))(\theta-\vartheta) \, \diff t 
\\&= 
\int_0^1\scp{(\nabla \f)(\vartheta + t(\theta-\vartheta)),\theta-\vartheta } \,\diff t 
\\& \leq  
\int_0^1 \pnorm2{(\nabla \f)(\vartheta + t(\theta-\vartheta))}\pnorm2{\theta-\vartheta} \,\diff t 
\\&\leq 
\int_0^1 L\pnorm2{\vartheta + t(\theta-\vartheta) - \vartheta}\pnorm2{\theta-\vartheta} \, \diff t 
\\&= 
L\pnorm2{\theta-\vartheta}^2  \br*{\int_0^1t \, \diff t }
= 
\tfrac{L}{2}\pnorm2{\theta-\vartheta}^2 
\ifnocf.
\end{split}
\end{equation}
\cfload[.]
The proof of \cref{fcond2} is thus complete.
\end{proof}
\endgroup

\section{Optimization through flows of ODEs}

\subsection{Approximation of local minimum points through GFs}
\label{sect:flow}

\cfclear
\begingroup
\providecommand{\d}{}
\renewcommand{\d}{\defaultParamDim}
\providecommand{\f}{}
\renewcommand{\f}{\defaultLossFunction}
\providecommand{\g}{}
\renewcommand{\g}{\defaultGradientFunction}
\begin{prop}[Approximation of local minimum points through \GFs]
\label{flow_finite_horizon}
Let $\d \in \N$,
$c,T \in (0,\infty)$, $r \in (0,\infty]$, $\vartheta \in \R^\d$, 
$\mathbb{B} = \{w \in \R^\d \colon \pnorm2{w-\vartheta} \leq r \}$, $\xi \in \mathbb{B}$, $\f \in C^1(\R^\d,\R)$ satisfy for all $\theta \in \mathbb{B}$ that
\begin{equation}
\label{flow_finite_horizon:assumption1}
\scp{\theta-\vartheta, (\nabla \f)(\theta) } \geq  c \pnorm2{\theta-\vartheta}^2 
  ,
\end{equation} 
and let $\Theta \in C([0,T], \R^\d)$ satisfy for all $t \in [0,T]$ that $ \Theta_t = \xi - \int_0^t (\nabla \f)(\Theta_s) \, \diff s$ \cfload.
Then
\begin{enumerate}[label=(\roman *)]
\item \label{flow_finite_horizon:item1}
it holds that
$
  \{\theta \in \mathbb{B}  \colon  \f(\theta)= \inf\nolimits_{w \in \mathbb{B}} \f(w)  \} = \{\vartheta \}
$,

\item \label{flow_finite_horizon:item2}
it holds for all $t \in [0,T]$ that
$
  \pnorm2{\Theta_t -\vartheta} \leq e^{-ct} \pnorm2{\xi - \vartheta}
$,
and
\item \label{flow_finite_horizon:item3}
it holds for all $t \in [0,T]$ that
\begin{equation}
0 \leq \tfrac{c}{2} \pnorm2{\Theta_t - \vartheta}^2 \leq \f(\Theta_t) -\f(\vartheta) .
\end{equation}
\end{enumerate}
\end{prop}

\begin{proof}[Proof of \cref{flow_finite_horizon}]
Throughout this proof,  let $V \colon \R^\d \to [0,\infty)$ satisfy for all $\theta \in \R^\d$ that $V(\theta) = \pnorm2{\theta-\vartheta}^2$, 
let $\epsilon \colon [0,T] \to [0,\infty) $ satisfy for all $t \in [0,T]$ that $ \epsilon(t) = \pnorm2{\Theta_t-\vartheta}^2 = V(\Theta_t)$, and 
let $\tau \in [0,T]$ be the real number given by 
\begin{equation}
\label{eq:in_proof_definition_of_tau}
\tau 
=
 \inf \pr*{\{t \in [0,T] \colon \Theta_t \notin \mathbb{B} \} \cup \{T\}} = 
\inf\pr*{\{t \in [0,T] \colon \epsilon(t) > r^2 \}\cup \{T\}}.
\end{equation}
Note that (\ref{flow_finite_horizon:assumption1}) and \cref{fcond1:item2} in \cref{fcond1} establish \cref{flow_finite_horizon:item1}.
Next observe that \cref{der_of_norm} implies that for all $\theta \in \R^\d$ it holds that $V \in C^1(\R^\d, [0,\infty))$ and
\begin{equation}
\label{flow_finite_horizon:eq2}
(\nabla V)(\theta) 
=
2(\theta-\vartheta).
\end{equation}
Moreover, observe that the fundamental theorem of calculus (see, \eg, Coleman \cite[Theorem 3.9]{Coleman12}) and the fact that 
$ \R^\d \ni v \mapsto ( \nabla \f )( v ) \in \R^\d $ and $\Theta\colon [0,T] \to \R^\d$ are continuous functions ensure that for all $t \in [0,T]$ it holds that $\Theta \in C^1([0,T], \R^\d)$ and
\begin{equation}
\tfrac{ \mathrm{d}}{\mathrm{d}t} (\Theta_t) = -(\nabla \f)(\Theta_t).
\end{equation}
Combining (\ref{flow_finite_horizon:assumption1}) and (\ref{flow_finite_horizon:eq2}) hence demonstrates that for all  $t \in [0,\tau]$ it holds that $\epsilon \in C^1([0,T],[0,\infty))$ and 
\begin{equation}
\label{flow_finite_horizon:eq3}
\begin{split}
 \epsilon'(t)& = 
 \tfrac{ \mathrm{d} }{ \mathrm{dt} } \pr[\big]{ V( \Theta_t ) }
 = V'(\Theta_t) \pr*{\tfrac{\mathrm{d}}{\mathrm{d} t}(\Theta_t)} \\
&= \scp{(\nabla V)(\Theta_t) , \tfrac{\mathrm{d}}{\mathrm{d} t}(\Theta_t)
} \\
&= \scp{ 2(\Theta_t-\vartheta) , -(\nabla \f)(\Theta_t)} \\
&= -2\scp{ (\Theta_t-\vartheta) , (\nabla \f)(\Theta_t)} \\
&\leq -2 c\pnorm2{\Theta_t-\vartheta}^2 = -2c \epsilon(t).
\end{split}
\end{equation}
The Gronwall inequality, \eg, in \cref{lem:gronwall_general} therefore implies that for all $t \in [0,\tau]$ it holds that
\begin{equation}
\epsilon(t) \leq \epsilon(0)e^{-2ct}.
\end{equation}
Hence, we obtain for all $t \in [0,\tau]$ that
\begin{equation}
\label{flow_finite_horizon:eq4}
\pnorm2{\Theta_t-\vartheta} 
= 
\sqrt{\epsilon(t)} \leq \sqrt{\epsilon(0)}e^{-ct} 
=
\pnorm2{\Theta_0-\vartheta} e^{-ct} 
=
\pnorm2{\xi-\vartheta} e^{-ct}
.
\end{equation}
In the next step we prove that
\begin{eqsplit}
\label{eq:in_proof_tau_strictly_bigger_than_zero}
  \tau > 0 .
\end{eqsplit}
In our proof of \cref{eq:in_proof_tau_strictly_bigger_than_zero} 
we distinguish between the case $ \varepsilon( 0 ) = 0 $ and the case $ \varepsilon( 0 ) > 0 $. 
We first prove \cref{eq:in_proof_tau_strictly_bigger_than_zero} 
in the case 
\begin{eqsplit}
\label{eq:in_proof_case_varepsilon_equal_to_zero}
  \varepsilon( 0 ) = 0
  .
\end{eqsplit}
\Nobs that \cref{eq:in_proof_case_varepsilon_equal_to_zero}, 
the assumption that $ r \in (0,\infty] $, 
and the fact that $\epsilon \colon [0,T] \to [0,\infty) $ 
is a continuous function show that 
\begin{equation}
\label{flow_finite_horizon:eq4.1}
  \tau = 
  \inf\pr*{
    \{ t \in [0,T] \colon 
    \epsilon(t) > r^2 \}
    \cup \{ T \}
  } > 0 .
\end{equation}
This establishes \cref{eq:in_proof_tau_strictly_bigger_than_zero} 
in the case $ \varepsilon(0) = 0 $. 
In the next step we prove \cref{eq:in_proof_tau_strictly_bigger_than_zero} 
in the case 
\begin{eqsplit}
\label{eq:in_proof_case_varepsilon_bigger_than_zero}
  \varepsilon( 0 ) > 0 .
\end{eqsplit}
\Nobs that 
\eqref{flow_finite_horizon:eq3} 
and the assumption that $c \in (0,\infty)$ assure that for all $t \in [0,\tau]$ with $\epsilon(t) > 0$ it holds that
\begin{equation}
  \epsilon'(t) 
  \leq - 2 c \epsilon(t) < 0.
\end{equation} 
Combining this with \cref{eq:in_proof_case_varepsilon_bigger_than_zero} 
shows that 
\begin{equation}
  \epsilon'(0) 
  < 0
  .
\end{equation} 
The fact that $ \epsilon' \colon [0,T] \to [0,\infty) $ is a continuous function 
and the assumption that $ T \in (0,\infty) $
therefore demonstrate that 
\begin{equation}
\label{flow_finite_horizon:eq4.2}
  \inf\pr*{  
    \{ 
      t \in [0,T] \colon \epsilon'( t ) > 0 
    \} \cup \{ T \} 
  } > 0 .
\end{equation}
Next note that the fundamental theorem of calculus and 
the assumption that $ \xi \in \mathbb{B} $ imply 
that for all $ s \in [0,T] $ with
$
  s < \inf\pr*{ \{ t \in [0,T] \colon \epsilon'(t) > 0 \} \cup \{ T \} }
$
it holds that
\begin{equation}
\epsilon(s) = \epsilon(0) + \int_0^s \epsilon'(u) \, \diff u 
\leq  \epsilon(0)  = \pnorm2{\xi-\vartheta}^2 \leq r^2.
\end{equation}
Combining this with \eqref{flow_finite_horizon:eq4.2} proves that 
\begin{eqsplit}
  \tau = 
  \inf\pr*{
    \{ s \in [0,T] \colon 
    \epsilon(s) > r^2 \}
    \cup \{ T \}
  } > 0 .
\end{eqsplit}
This establishes 
\cref{eq:in_proof_tau_strictly_bigger_than_zero} 
in the case $ \varepsilon(0) > 0 $. 
\Nobs that 
\eqref{flow_finite_horizon:eq4}, 
\cref{eq:in_proof_tau_strictly_bigger_than_zero}, 
and the assumption that $ c \in (0,\infty) $
demonstrate that
\begin{equation}
\begin{split}
\pnorm2{\Theta_{\tau} -\vartheta}
\leq \pnorm2{\xi-\vartheta} e^{-c\tau} < r. %
\end{split}
\end{equation}
The fact that $\epsilon \colon [0,T] \to [0,\infty)$ is a continuous function, 
\cref{eq:in_proof_definition_of_tau}, 
and 
\cref{eq:in_proof_tau_strictly_bigger_than_zero} 
hence assure that $\tau = T$.
Combining this with \eqref{flow_finite_horizon:eq4} proves that for all $t \in [0,T]$ it holds that
\begin{equation}
\label{flow_finite_horizon:eq5}
\pnorm2{\Theta_t-\vartheta} \leq  \pnorm2{\xi-\vartheta} e^{-ct}.
\end{equation}
This establishes \cref{flow_finite_horizon:item2}. 
It thus remains to prove \cref{flow_finite_horizon:item3}. 
For this observe that \eqref{flow_finite_horizon:assumption1} 
and \cref{fcond1:item1} in \cref{fcond1} 
demonstrate that for all $\theta \in \mathbb{B}$ it holds that
\begin{equation}
\label{flow_finite_horizon:eq6}
0 \leq \tfrac{c}{2} \pnorm2{\theta - \vartheta}^2 \leq  \f(\theta)-\f(\vartheta) .
\end{equation}
Combining this %
and \cref{flow_finite_horizon:item2} implies that for all $t \in [0,T]$ it holds that
\begin{equation}
  0 \leq \tfrac{c}{2} \pnorm2{\Theta_t - \vartheta}^2 \leq \f(\Theta_t)-\f(\vartheta) 
\end{equation}
This establishes \cref{flow_finite_horizon:item3}. 
The proof of \cref{flow_finite_horizon} is thus complete.
\end{proof}
\endgroup

\subsection{Existence and uniqueness of solutions of ODEs}

\begingroup
\providecommand{\d}{}
\renewcommand{\d}{\defaultParamDim}
\providecommand{\f}{}
\renewcommand{\f}{\defaultLossFunction}
\providecommand{\g}{}
\renewcommand{\g}{\defaultGradientFunction}
\begin{athm}{lemma}{ODE_finite}[Local existence of maximal solution of \ODEs]
Let $\d \in \N$, $\xi \in \R^\d$, $T\in (0,\infty)$,
let $\normmm{\cdot} \colon \R^\d \to [0,\infty)$ be a norm, and 
let $\g \colon \R^\d \to \R^\d$ be locally Lipschitz continuous. 
Then there exist a unique real number $\tau \in (0,T]$  and a unique continuous function $\Theta \colon [0,\tau) \to \R^\d$ such that for all $t \in [0,\tau)$ it holds that 
\begin{equation}
\label{ODE_finite:conclusion1}
\liminf_{s \nearrow \tau} \bbr{ 
  \normmm{\Theta_s} + \tfrac{1}{ (T-s) }
} = \infty 
\qandq
\Theta_t = \xi + \int_0^t \g(\Theta_s) \, \diff s.
\end{equation}
\end{athm}

\begin{aproof}
Note that, \eg, 
Teschl~\cite[Theorem~2.2 and Corollary~2.16]{teschl2012ordinary}
implies \cref{ODE_finite:conclusion1} (cf., \eg, 
\cite[Theorem~7.6]{amann1990ordinary} 
and
\cite[Theorem 1.1]{Jentzen2018}).
\end{aproof}
\endgroup

\begingroup
\providecommand{\d}{}
\renewcommand{\d}{\defaultParamDim}
\providecommand{\f}{}
\renewcommand{\f}{\defaultLossFunction}
\providecommand{\g}{}
\renewcommand{\g}{\defaultGradientFunction}
\begin{lemma}[Local existence of maximal solution of \ODEs\ on an infinite time interval]
\label{ODE_infinite}
Let $\d \in \N$, $\xi \in \R^\d$,
let $\normmm{\cdot} \colon \R^\d \to [0,\infty)$ be a norm, and 
let $\g \colon \R^\d \to \R^\d$ be locally Lipschitz continuous. 
Then there exist a unique extended real number $\tau \in (0,\infty]$  and a unique continuous function $\Theta \colon [0,\tau) \to \R^\d$ such that for all $t \in [0,\tau)$ it holds that 
\begin{equation}
\label{ODE_infinite:conclusion1}
\liminf_{s \nearrow \tau} \bbr{ \normmm{\Theta_s} + s } = \infty 
\qandq
\Theta_t = \xi + \int_0^t \g(\Theta_s) \, \diff s.
\end{equation}
\end{lemma}

\begin{proof}[Proof of \cref{ODE_infinite}]
First, observe that \cref{ODE_finite} implies that there exist unique real numbers $\tau_n \in (0,n]$, $n \in \N$,  and unique continuous functions $\Theta^{(n)} \colon [0,\tau_n) \to \R^\d$, $n \in \N$, such that for all $n \in \N$, $t \in [0,\tau_n)$ it holds that 
\begin{equation}
\label{ODE_infinite:eq1}
\liminf_{s \nearrow \tau_n} \br*{\normmm{\Theta^{(n)}_s} + \tfrac{1}{(n-s)} } = \infty 
\qandq
\Theta^{(n)}_t = \xi + \int_0^t \g(\Theta^{(n)}_s) \, \diff s.
\end{equation}
This shows that for all $ n \in \N $, $ t \in [0,\min\{ \tau_{ n + 1 }, n \} ) $
it holds that 
\begin{equation}
  \liminf_{ s \nearrow \tau_{ n + 1 } }
  \br*{ 
    \normmm{ \Theta^{ (n+1) }_s }
    +
    \tfrac{ 1 }{ ( n + 1 - s ) }
  }
  = \infty 
\qandq
  \Theta^{(n+1)}_t = \xi + \int_0^t \g(\Theta^{(n+1)}_s) \, \diff s
  .
\end{equation}
Hence, we obtain that for all $n \in \N$, $t \in [0,\min\{\tau_{n+1},n\})$ it holds that 
\begin{equation}
\liminf_{s \nearrow \min\{\tau_{n+1},n\}} 
\br*{ 
  \normmm{\Theta^{(n+1)}_s}
  + 
  \tfrac{1}{(n-s)}
} = \infty 
\end{equation}
\begin{equation}
\andq
\Theta^{(n+1)}_t = \xi + \int_0^t \g(\Theta^{(n+1)}_s) \, \diff s.
\end{equation}
Combining this with \eqref{ODE_infinite:eq1} demonstrates that for all $n \in \N$ it holds that
\begin{equation}
\label{ODE_infinite:eq2}
\tau_{n} = \min\{\tau_{n+1},n\} 
\qandq 
\Theta^{(n)} 
= 
\Theta^{(n+1)}|_{ [0, \min\{ \tau_{ n + 1 }, n \} ) }
.
\end{equation}
Therefore, we obtain that for all $ n \in \N $ it holds that
\begin{equation}
\label{ODE_infinite:eq2b}
  \tau_n \leq \tau_{ n + 1 }
\qandq
  \Theta^{(n)} 
  = 
  \Theta^{(n+1)}|_{[0,\tau_n)}
  .
\end{equation}
Next let $\mathfrak{t} \in (0,\infty]$ be the extended real number given by 
\begin{equation}
\label{mathfrak_limit:eq}
  \mathfrak{t} 
  = 
  \lim_{n \to \infty} \tau_n
\end{equation} 
and let $\varTheta \colon [0,\mathfrak{t}) \to \R^\d$ satisfy for all $n \in \N$, $t \in [0,\tau_n)$ that
\begin{equation}
\label{ODE_infinite:eq3}
\varTheta_t = \Theta^{(n)}_t.
\end{equation}
Observe that for all $ t \in [0,\mathfrak{t}) $ there exists 
$ n \in \N $ such that
$
  t \in [0,\tau_n)
$.
This, \eqref{ODE_infinite:eq1}, and \eqref{ODE_infinite:eq2b} 
assure that for all $t \in [0,\mathfrak{t})$ it holds that 
$ \varTheta \in C( [0,\mathfrak{t}) , \R^\d ) $ 
and 
\begin{equation}
\label{ODE_infinite:eq4}
  \varTheta_t = \xi + \int_0^t \g(\varTheta_s) \, \diff s .
\end{equation}
In addition, note that \eqref{ODE_infinite:eq2} ensures that for all $n \in \N$, $k \in \N\cap [n,\infty)$ it holds that
\begin{equation}
\min \{\tau_{k+1},n \} 
=
\min\{ \tau_{ k + 1 }, k, n \}
= \min \{ \min \{\tau_{k+1},k \} ,n \} = \min\{ \tau_k, n \} .
\end{equation}
This shows that for all $ n \in \N $, $ k \in \N\cap (n,\infty) $
it holds that
$
  \min\{ \tau_k , n \} = \min\{ \tau_{ k - 1 }, n \}
$.
Hence, we obtain that for all $n \in \N$, $k \in\N\cap (n,\infty)$ it holds that
\begin{equation}
  \min\{ \tau_k, n \} 
  = 
  \min\{ \tau_{ k - 1 }, n \}
  =
  \ldots 
  = 
  \min\{ \tau_{ n + 1 }, n \}
  =
  \min\{ \tau_n, n \}
  = \tau_n.
\end{equation}
Combining this with the fact that $(\tau_n)_{n \in \N} \subseteq [0,\infty) $ is a non-decreasing sequence implies that for all $n \in \N$ it holds that
\begin{equation}
\min \{\mathfrak{t},n\} 
= 
\min\cu*{
  \lim_{k \to \infty} \tau_k,n
}
=
\lim_{ k \to \infty} 
  \bpr{ 
    \min\{ \tau_k, n \} 
  }
=
\lim_{ k \to \infty } \tau_n
=
  \tau_n .
\end{equation}
Therefore, we obtain that for all $ n \in \N $ with $ \mathfrak{t} < n $ it holds that 
\begin{equation}
\label{ODE_infinite:eq5}
  \tau_n = \min\{ \mathfrak{t}, n \}  = \mathfrak{t} .
\end{equation}
This, \eqref{ODE_infinite:eq1}, and \eqref{ODE_infinite:eq3} demonstrate that for all $n \in \N$ with $ \mathfrak{t} < n $ it holds that 
\begin{equation}
\begin{split}
\liminf_{ s \nearrow \mathfrak{t} } \normmm{\varTheta_s} 
&= 
\liminf_{s \nearrow \tau_n} \normmm{\varTheta_s} 
= 
\liminf_{s \nearrow \tau_n} \normmm{\Theta^{(n)}_s} 
\\
&= 
- \tfrac{1}{ ( n - \mathfrak{t} ) } + \liminf_{s \nearrow \tau_n} \br*{\normmm{\Theta^{(n)}_s} + \tfrac{1}{(n-\mathfrak{t})} }
\\
&= 
- \tfrac{1}{ ( n - \mathfrak{t} ) } + \liminf_{s \nearrow \tau_n} \br*{\normmm{\Theta^{(n)}_s} + \tfrac{1}{(n-s)} }
= \infty.
\end{split}
\end{equation}
Therefore, we obtain that
\begin{equation}
\label{ODE_infinite:eq6}
\liminf_{ s \nearrow \mathfrak{t} }
\bbr{ 
  \normmm{\varTheta_s}
+ 
s
} 
  = \infty 
  .
\end{equation}
Next note that for all 
$ \hat{\mathfrak{t}} \in (0,\infty] $, $ \hat{\varTheta} \in C([0,\hat{\mathfrak{t}}), \R^\d) $, 
$ n \in \N $, $ t \in [ 0, \min\{ \hat{\mathfrak{t}}, n \} ) $ 
with 
$
  \liminf_{ s \nearrow \hat{\mathfrak{t}}} 
  [ 
    \opnorm{\hat{\varTheta}_s} + s 
  ] = \infty 
$ 
and 
$ 
  \forall \, s \in [0,\hat{\mathfrak{t}}) \colon 
  \hat{\varTheta}_s 
  = \xi + \int_0^s \g( \hat{\varTheta}_u ) \, \diff u
$ 
it holds that
\begin{equation}
\liminf_{
  s \nearrow \min\{ \hat{\mathfrak{t}},n \} } \br*{
  \opnorm{\hat{\varTheta}_s}
+ 
  \tfrac{1}{(n-s)}
} = \infty 
\qandq
  \hat{\varTheta}_t = \xi + \int_0^t \g(\hat{\varTheta}_s) \, \diff s .
\end{equation}
This and \eqref{ODE_infinite:eq1} prove that for all 
$ \hat{\mathfrak{t}} \in (0,\infty] $, 
$ \hat{\varTheta} \in C( [0,\hat{\mathfrak{t}} ), \R^\d) $, 
$ n \in \N $ 
with 
$
  \liminf_{ t \nearrow \hat{\mathfrak{t}} } 
  [
    \opnorm{ \hat{\varTheta}_t } + t 
  ] = \infty 
$ 
and 
$
  \forall \, t \in [0,\hat{\mathfrak{t}}) \colon \hat{\varTheta}_t = \xi + \int_0^t \g(\hat{\varTheta}_s) \, \diff s
$
it holds that
\begin{equation}
  \min\{ \hat{\mathfrak{t}}, n \} 
  = \tau_n 
\qandq
  \hat{\varTheta}|_{ [0,\tau_n) } 
  = 
  \Theta^{ (n) } 
  .
\end{equation}
Combining 
\eqref{ODE_infinite:eq4} and \eqref{ODE_infinite:eq6}
hence assures that for all  
$ \hat{\mathfrak{t}} \in (0,\infty] $, 
$ \hat{\varTheta} \in C( [0,\hat{\mathfrak{t}} ), \R^\d) $, 
$ n \in \N $ 
with 
$
  \liminf_{ t \nearrow \hat{\mathfrak{t}} } 
  [
    \opnorm{ \hat{\varTheta}_t } + t 
  ] = \infty 
$ 
and 
$
  \forall \, t \in [0,\hat{\mathfrak{t}}) \colon \hat{\varTheta}_t = \xi + \int_0^t \g(\hat{\varTheta}_s) \, \diff s
$
it holds that
\begin{equation}
  \min\{ \hat{\mathfrak{t}}, n \} 
  = \tau_n 
  = 
  \min\{ \mathfrak{t}, n \}
\qandq
  \hat{\varTheta}|_{ [0,\tau_n) } 
  = 
  \Theta^{ (n) } 
  = \varTheta |_{ [0, \tau_n) }
  .
\end{equation}
This and \eqref{mathfrak_limit:eq}
show that for all  
$ \hat{\mathfrak{t}} \in (0,\infty] $, 
$
  \hat{\varTheta} \in C( [0,\hat{\mathfrak{t}}), \R^\d )
$ 
with
$
  \liminf_{
    t \nearrow \hat{\mathfrak{t}}
  } 
  [ 
    \normmm{ \shat{\varTheta}_t } + t 
  ] 
  = \infty 
$ 
and 
$ 
  \forall \, t \in [0,\hat{\mathfrak{t}}) \colon 
  \hat{\varTheta}_t = \xi + \int_0^t \g(\hat{\varTheta}_s) \, \diff s
$
it holds that
\begin{equation}
  \hat{\mathfrak{t}} = \mathfrak{t} 
\qandq 
  \hat{\varTheta} = \varTheta
  .
\end{equation}
Combining this, \eqref{ODE_infinite:eq4}, and \eqref{ODE_infinite:eq6} 
completes the proof of \cref{ODE_infinite}.
\end{proof}
\endgroup

\subsection{Approximation of local minimum points through GFs revisited}

\cfclear
\begingroup
\providecommand{\d}{}
\renewcommand{\d}{\defaultParamDim}
\providecommand{\f}{}
\renewcommand{\f}{\defaultLossFunction}
\providecommand{\g}{}
\renewcommand{\g}{\defaultGradientFunction}
\begin{theorem}[Approximation of local minimum points through \GFs\ revisited]
\label{flow}
Let $ \d \in \N $,
$ c \in (0,\infty)$, $r \in (0,\infty]$, $\vartheta \in \R^\d$, 
$\mathbb{B} = \{w \in \R^\d \colon \pnorm2{w-\vartheta} \leq r \}$, $\xi \in \mathbb{B}$, $\f \in C^2(\R^\d,\R)$ satisfy for all $\theta \in \mathbb{B}$ that
\begin{equation}
\label{flow:assumption1}
\scp{\theta-\vartheta, (\nabla \f)(\theta) } \geq  c \pnorm2{\theta-\vartheta}^2 \ifnocf.
\end{equation} 
\cfload[.]%
Then
\begin{enumerate}[label=(\roman *)]
\item \label{flow:item0}
there exists a unique continuous function $\Theta \colon [0,\infty) \to \R^\d$ such that for all $t \in [0,\infty)$ it holds that 
\begin{equation}
\Theta_t = \xi - \int_0^t (\nabla \f)(\Theta_s) \, \diff s,
\end{equation} 

\item \label{flow:item1}
it holds that
$ 
  \{\theta \in \mathbb{B}  \colon  \f(\theta)= \inf\nolimits_{w \in \mathbb{B}} \f(w)  \} = \{\vartheta \}
$,
\item \label{flow:item2}
it holds for all $t \in [0,\infty)$ that
$  
  \pnorm2{\Theta_t -\vartheta} \leq e^{-ct} \pnorm2{\xi - \vartheta}
$,
and
\item \label{flow:item3}
it holds for all $t \in [0,\infty)$ that
\begin{equation}
0 \leq \tfrac{c}{2} \pnorm2{\Theta_t - \vartheta}^2 \leq \f(\Theta_t) -\f(\vartheta) .
\end{equation}
\end{enumerate}
\end{theorem}

\begin{proof}[Proof of \cref{flow}]
First, observe that the assumption that 
$ \f \in C^2( \R^\d, \R ) $
ensures that
\begin{equation}
  \R^\d \ni \theta \mapsto - ( \nabla \f )( \theta ) \in \R^\d
\end{equation}
is continuously differentiable. 
The fundamental theorem of calculus hence implies that
\begin{equation}
  \R^\d \ni \theta \mapsto - ( \nabla \f )( \theta ) \in \R^\d
\end{equation}
is locally Lipschitz continuous.
Combining this with 
\cref{ODE_infinite} 
(applied with $ \g\is (\R^\d\ni \theta\mapsto - ( \nabla \f )( \theta )\in\R^\d) $ 
in the notation of 
\cref{ODE_infinite})
proves that there exists a unique extended real number $\tau \in (0,\infty]$  and a unique continuous function $\Theta \colon [0,\tau) \to \R^\d$ such that for all $t \in [0,\tau)$ it holds that 
\begin{equation}
\label{flow:eq1}
\liminf_{s \nearrow \tau} \bbr{
  \pnorm2{\Theta_s} + s 
} = \infty 
\qandq
\Theta_t = \xi - \int_0^t (\nabla \f)(\Theta_s) \, \diff s.
\end{equation}
Next observe that \cref{flow_finite_horizon} proves that for all $t \in [0,\tau)$ it holds that
\begin{equation}
\pnorm2{\Theta_t -\vartheta} \leq e^{-ct} \pnorm2{\xi - \vartheta}.
\end{equation}
This implies that 
\begin{equation}
\begin{split}
&\liminf_{s \nearrow \tau} \pnorm2{\Theta_s} 
\leq 
\br*{
\liminf_{s \nearrow \tau} \pnorm2{\Theta_s-\vartheta} 
} 
+ \pnorm2{\vartheta} 
\\
&\leq 
\br*{
\liminf_{s \nearrow \tau} e^{-cs} \pnorm2{\xi - \vartheta} 
} 
+ \pnorm2{\vartheta}
\leq 
\pnorm2{\xi - \vartheta} +\pnorm2{\vartheta} < \infty.
\end{split}
\end{equation}
This and \eqref{flow:eq1} demonstrate that 
\begin{equation}
  \tau = \infty
  .
\end{equation}
This and \eqref{flow:eq1} prove \cref{flow:item0}. 
Moreover, note that 
\cref{flow_finite_horizon}
and \cref{flow:item0} 
establish \cref{flow:item1,flow:item2,flow:item3}.
The proof of \cref{flow} is thus complete.
\end{proof}
\endgroup

\subsection{Approximation error with respect to the objective function}

\cfclear
\begingroup
\providecommand{\d}{}
\renewcommand{\d}{\defaultParamDim}
\providecommand{\f}{}
\renewcommand{\f}{\defaultLossFunction}
\providecommand{\g}{}
\renewcommand{\g}{\defaultGradientFunction}
\begin{cor}[Approximation error with respect to the objective function]
\label{cor_flow}
Let $ \d \in \N $, 
$ c, L \in (0,\infty)$, $r \in (0,\infty]$, 
$\vartheta \in \R^\d$, $\mathbb{B} = \{w \in \R^\d \colon \pnorm2{w-\vartheta} \leq r \}$, $\xi \in \mathbb{B}$, $\f \in C^2(\R^\d,\R)$ satisfy for all $\theta \in \mathbb{B}$ that
\begin{equation}
\label{cor_flow:assumption1}
\scp{\theta-\vartheta, (\nabla \f)(\theta) } \geq  c \pnorm2{\theta-\vartheta}^2 \qandq \pnorm2{(\nabla \f)(\theta)} \leq L \pnorm2{\theta-\vartheta}\ifnocf.
\end{equation} 
\cfload[.]%
Then
\begin{enumerate}[label=(\roman *)]
\item \label{cor_flow:item0}
there exists a unique continuous function $\Theta \colon [0,\infty) \to \R^\d$ such that for all $t \in [0,\infty)$ it holds that 
\begin{equation}
\Theta_t = \xi - \int_0^t (\nabla \f)(\Theta_s) \, \diff s,
\end{equation} 

\item \label{cor_flow:item1}
it holds that
$ 
  \{\theta \in \mathbb{B}  \colon  \f(\theta)= \inf\nolimits_{w \in \mathbb{B}} \f(w)  \} = \{\vartheta \}
$,
\item \label{cor_flow:item2}
it holds for all $t \in [0,\infty)$ that
$  
  \pnorm2{\Theta_t -\vartheta} \leq e^{-ct} \pnorm2{\xi - \vartheta}
$,
and
\item \label{cor_flow:item3}
it holds for all $t \in [0,\infty)$ that
\begin{equation}
0 \leq \tfrac{c}{2} \pnorm2{\Theta_t - \vartheta}^2 \leq \f(\Theta_t) -\f(\vartheta) 
\leq
\tfrac{ L }{ 2 } 
\pnorm2{ \Theta_t - \vartheta }^2
\leq
\tfrac{L}{2}e^{-2ct} \pnorm2{\xi - \vartheta}^2.
\end{equation}
\end{enumerate}
\end{cor}

\begin{proof}[Proof of 
\cref{cor_flow}]
\Cref{flow} and \cref{fcond2} establish \cref{cor_flow:item0,cor_flow:item1,cor_flow:item2,cor_flow:item3}.
The proof of \cref{cor_flow} is thus complete.
\end{proof}
\endgroup

%% file: parts/Optimization_methods_definitions.tex
\NewDocumentCommand{\deterministicGDdef}{ O{true} O{} O{false} O{} mmmmm }{%
\cfclear 
\begingroup 
\ifthenelse{\equal{#1}{true}}{}{
\providecommandordefault{\defaultGradientFunction}{(\nabla \defaultLossFunction)}
}
\providecommandordefault{\grad}{\defaultGradientFunction}
\begin{adef}{#5}[#6]
	\ifthenelse{\equal{#3}{false}}{
		Let 
		$\defaultParamDim \in \N$,
		let $ \defaultLossFunction \colon \R^\defaultParamDim \to \R $  be differentiable, \ignorespaces
	}{\ignorespaces}
	\ifthenelse{\equal{#3}{true}}{
		Let 
		$\defaultParamDim_1, \defaultParamDim_2 \in \N$,
		let $ \defaultLossFunction \colon \R^{\defaultParamDim_1 \times \defaultParamDim_2} \to \R $ be differentiable, \ignorespaces
	}{\ignorespaces}
	\ifthenelse{\equal{#3}{muon}}{
		Let 
		$\defaultParamDim_1, \defaultParamDim_2, K \in \N$,
		let $ \defaultLossFunction \colon \R^{\defaultParamDim_1 \times \defaultParamDim_2} \to \R $ be differentiable, \ignorespaces 
	}{\ignorespaces}
    \ifthenelse{\equal{#1}{true}}{let
        $\defaultGradientFunction=(\defaultGradientFunction_1,\dots,\defaultGradientFunction_{\defaultParamDim}) \colon \R^\defaultParamDim \to \R^\defaultParamDim$ 
        satisfy
        for all 
        $\theta\in \R^\defaultParamDim$
        that
        \begin{equation}
        \defaultGradientFunction( \theta ) = ( \nabla \defaultLossFunction )( \theta ),
        \end{equation}
    }{ }\unskip{}\ignorespaces
    let
        \ignorespaces#7\unskip{}
	\ifthenelse{\equal{#3}{false}}{
		\unskip{}\ignorespaces
		$\xi \in \R^\defaultParamDim$,
		\ignorespaces
	}{
		\unskip{}\ignorespaces
		$\xi \in \R^{\defaultParamDim_1 \times \defaultParamDim_2}$,
		\ignorespaces
	}
	\ignorespaces#4\ignorespaces
    and let\ifthenelse{\equal{#1}{true}}{
        $ 
        \Theta = 
        ( \Theta^{ ( 1 ) }, \dots,\allowbreak \Theta^{ ( \defaultParamDim ) } ) 
        \colon \N_0 \to \R^\defaultParamDim 
        $
		\ignorespaces
    }{\ifthenelse{\equal{#3}{false}}{
			$ \Theta \colon \N_0 \to \R^\defaultParamDim $
			\ignorespaces
		}
		{
			$\Theta \colon \N_0 \to \R^{\defaultParamDim_1 \times \defaultParamDim_2}$
			\ignorespaces
		}
	} 
	be a function
	\cfload
	.
    Then we say that $ \Theta $ is the 
    \ignorespaces#8\unskip{}
    \ifthenelse{\equal{#2}{}}{}{\ignorespaces(we say that $ \Theta $ is the \ignorespaces#2\unskip{})
    }\unskip{}
    if and only if 
    \ignorespaces#9
\end{adef}
\endgroup
}

\NewDocumentCommand{\SGDdef}{ O{true} O{} O{false} O{} mmmmm }{%
\cfclear
\begin{adef}{#5}[#6]
	\ifthenelse{\equal{#3}{false}}{
		Let 
		$\defaultParamDim \in \N$, 
		let $(S, \mathcal{S})$ be a measurable space,
		let 
			$\defaultStochLoss = \allowbreak (\defaultStochLoss(\theta,x))_{(\theta,x) \in \R^\defaultParamDim \times S} \colon \allowbreak \R^\defaultParamDim \times S \to \R$
		be measurable, 
		assume for all 
			$x \in S$ that
		$\defaultStochLoss(\cdot,x)$ is differentiable,
	}{\ignorespaces}
	\ifthenelse{\equal{#3}{true}}{
		Let 
		$\defaultParamDim_1, \defaultParamDim_2 \in \N$, 
		let $(S, \mathcal{S})$ be a measurable space,
		let 
			$\defaultStochLoss = \allowbreak (\defaultStochLoss(\theta,x))_{(\theta,x) \in \R^{\defaultParamDim_1 \times \defaultParamDim_2} \times S} \colon \allowbreak \R^{\defaultParamDim_1 \times \defaultParamDim_2} \times S \to \R$
		be measurable, 
		assume for all 
			$x \in S$ that
		$\defaultStochLoss(\cdot,x)$ is differentiable,
	}{\ignorespaces}
	\ifthenelse{\equal{#3}{muon}}{
		Let 
		$\defaultParamDim_1, \defaultParamDim_2, K \in \N$, 
		let $(S, \mathcal{S})$ be a measurable space,
		let 
			$\defaultStochLoss = \allowbreak (\defaultStochLoss(\theta,x))_{(\theta,x) \in \R^{\defaultParamDim_1 \times \defaultParamDim_2} \times S} \colon \allowbreak \R^{\defaultParamDim_1 \times \defaultParamDim_2} \times S \to \R$
		be measurable, 
		assume for all 
			$x \in S$ that
		$\defaultStochLoss(\cdot,x)$ is differentiable,
	}{\ignorespaces}
	\ignorespaces\unskip{}
	let\ifthenelse{\equal{#1}{true}}{
		$\defaultStochGradient = (\defaultStochGradient_1,\dots,\defaultStochGradient_{\defaultParamDim}) \colon \R^\defaultParamDim \times S \to \R^\defaultParamDim$
	}{\ifthenelse{\equal{#3}{false}}{
		$\defaultStochGradient \colon \R^\defaultParamDim \times S \to \R^\defaultParamDim$
	}{
		$\defaultStochGradient \colon \R^{\defaultParamDim_1 \times \defaultParamDim_2} \times S \to \R^{\defaultParamDim_1 \times \defaultParamDim_2}$
	}}satisfy for all\ifthenelse{\equal{#3}{false}}{
			\ignorespaces
			$\theta\in \R^{\defaultParamDim}$,
			$x \in S$
			that
		}{
			\ignorespaces
			$\theta\in \R^{\defaultParamDim_1 \times \defaultParamDim_2}$,
			$x \in S$
			that
		}
	\begin{equation}
	\defaultStochGradient(\theta,x) = (\nabla_\theta \defaultStochLoss)(\theta, x),
	\end{equation}
	let
		\ignorespaces#7\unskip{}
	$(J_n)_{n \in \N} \subseteq \N$, 
	\ignorespaces#4
	let
	$(\Omega, \mathcal{F}, \P)$ be a probability space,\ifthenelse{\equal{#3}{false}}{
		let $\xi \colon \Omega \to \R^\defaultParamDim$ be a random variable, 
	}{
		let $\xi \colon \Omega \to \R^{\defaultParamDim_1 \times \defaultParamDim_2}$ be a random variable, 
	}for every $n,j \in \N$
	let 
		$X_{n,j}\colon \Omega \to S$
	be a random variable,
	and let\ifthenelse{\equal{#1}{true}}{
		$ 
		\Theta = 
		( \Theta^{ ( 1 ) }, \dots, \Theta^{ ( \defaultParamDim ) } ) 
		\colon \N_0 \times \Omega \to \R^\defaultParamDim 
		$ 
	}{\ifthenelse{\equal{#3}{false}}{
		$ 
		\Theta \colon \N_0 \times \Omega \to \R^\defaultParamDim 
		$ 
	}{
		$ 
		\Theta \colon \N_0 \times \Omega \to \R^{\defaultParamDim_1 \times \defaultParamDim_2} 
		$ 
	}}be a function
	\cfload
	.
	Then we say that $ \Theta $ is the 
	\ignorespaces#8\unskip{}
	batch sizes $(J_n)_{n \in \N}$,
	and data $(X_{n,j})_{(n,j)\in \N^2}$
	\ifthenelse{\equal{#2}{}}{}{\ignorespaces(we say that $ \Theta $ is the \ignorespaces#2\unskip{}
	batch sizes $(J_n)_{n \in \N}$,
	and data $(X_{n,j})_{(n,j)\in \N^2}$)
	}\unskip{}
	if and only if 
	#9
\end{adef}
}

\newcommand{\assign}{\leftarrow}
\newcommand{\separator}{;\,}
\NewDocumentCommand{\algorithmicDescription}{O{true} O{false} O{false} m m m m m m}{
	
	\begin{myalgorithm}{#4}{#5:algorithm}
		\textbf{Input:}\unskip{}
		\ifthenelse{\equal{#2}{false}}{\ignorespaces
			$\defaultParamDim, \maxsteps \in \N$, 
			$\defaultLossFunction \in C^1(\R^\defaultParamDim, \R)$,
			\ignorespaces#6\unskip{}
			$\xi \in \R^\defaultParamDim$\unskip{}
		}{\ignorespaces}\ifthenelse{\equal{#2}{true}}{\ignorespaces
			$\defaultParamDim_1, \defaultParamDim_2, \maxsteps \in \N$, 
			$\defaultLossFunction \in C^1(\R^{\defaultParamDim_1} \times \R^{\defaultParamDim_2}, \R)$,
			\ignorespaces#6\unskip{}
			$\xi \in \R^{\defaultParamDim_1 \times \defaultParamDim_2}$}{}\ifthenelse{\equal{#2}{muon}}{\ignorespaces
			$\defaultParamDim_1, \defaultParamDim_2, \maxsteps, K \in \N$, 
			$\defaultLossFunction \in C^1(\R^{\defaultParamDim_1} \times \R^{\defaultParamDim_2}, \R)$,
			\ignorespaces#6\unskip{}
			$\xi \in \R^{\defaultParamDim_1 \times \defaultParamDim_2}$}{}\ifthenelse{\equal{#3}{false}}{}{, #3}

			\textbf{Output:}\cfclear\cfadd{#5} $\maxsteps$-th step of the \ignorespaces#7\unskip{} \cfload

			\myalgorithmLine

		\begin{algorithmic}[1] %
			\State \textbf{Initialization:} \ignorespaces#8\unskip{}

			\For{$n = 1, \ldots, \maxsteps$} 
			\ifthenelse{\equal{#1}{true}}{
				\mycomment{(cf.\ \cref{def:componentwise_operations})}
			}{}
				#9
			\EndFor

			\State \Return $\Theta$
		\end{algorithmic}
	\end{myalgorithm}
}

\newcommand{\pseudoCodeGradient}{g}
\newcommand{\pseudoCodeGradientAssign}[1][\Theta]{
	\State $\pseudoCodeGradient \assign \frac{1}{J_n}\sum_{j = 1}^{J_n} (\nabla_\theta \defaultStochLoss)(#1, X_{n,j})$
}
\NewDocumentCommand{\algorithmicDescriptionStochastic}{O{true} O{false} O{false} m m m m m m}{
	
	\cfclear\cfadd{#5} 
	\begin{myalgorithm}{#4}{#5:algorithm}
			\textbf{Input:}\unskip{}
			\ifthenelse{\equal{#2}{false}}{\ignorespaces
				$\defaultParamDim, \mathbf{d}, \maxsteps \in \N$,
				$\defaultStochLoss = (\defaultStochLoss(\theta,x))_{(\theta,x) \in \R^\defaultParamDim \times \R^{\mathbf{d}}} \in C^1(\R^\defaultParamDim \times \R^{\mathbf{d}}, \R)$,}{}\ifthenelse{\equal{#2}{true}}{\ignorespaces
				$\defaultParamDim_1, \defaultParamDim_2, \mathbf{d}, \maxsteps \in \N$,
				$\defaultStochLoss = (\defaultStochLoss(\theta,x))_{(\theta,x) \in \R^{\defaultParamDim_1 \times \defaultParamDim_2} \times \R^{\mathbf{d}}} \in C^1(\R^{\defaultParamDim_1 \times \defaultParamDim_2} \times \R^{\mathbf{d}}, \R)$,}{}\ifthenelse{\equal{#2}{muon}}{\ignorespaces
				$\defaultParamDim_1, \defaultParamDim_2, \mathbf{d}, \maxsteps, K \in \N$,
				$\defaultStochLoss = (\defaultStochLoss(\theta,x))_{(\theta,x) \in \R^{\defaultParamDim_1 \times \defaultParamDim_2} \times \R^{\mathbf{d}}} \in C^1(\R^{\defaultParamDim_1 \times \defaultParamDim_2} \times \R^{\mathbf{d}}, \R)$,}{}
			\ignorespaces#6\unskip{}
			$(J_n)_{n \in \N} \subseteq \N$,
			probability space $(\Omega, \mathcal{F}, \P)$,\unskip{}
			\ifthenelse{\equal{#2}{false}}{
				\ignorespaces
				random variable $\xi \colon \Omega \to \R^\defaultParamDim$,}
			{random variable $\xi \colon \Omega \to \R^{\defaultParamDim_1 \times \defaultParamDim_2}$,}
			random variables $X_{n,j}\colon \Omega \to \R^{\mathbf{d}}$ for $n,j \in \N$\ifthenelse{\equal{#3}{false}}{}{, #3}

			\textbf{Output:} $\maxsteps$-th step of the \ignorespaces#7\unskip{} batch sizes $(J_n)_{n \in \N}$, and data $(X_{n,j})_{(n,j)\in \N^2}$ \cfload

			\myalgorithmLine

			\begin{algorithmic}[1] %
				\State \textbf{Initialization:}#8

			\For{$n = 1, \ldots, \maxsteps$}
			\ifthenelse{\equal{#1}{true}}{
				\mycomment{(cf.\ \cref{def:componentwise_operations})}
			}{}
				#9
			\EndFor

			\State \Return $\Theta$
		\end{algorithmic}
	\end{myalgorithm}
}

\newtcolorbox{summaryStyle}[2][]{
    enhanced,
    title=#2,
    #1,
    colframe=black,
    colback=white,
    coltitle=black,
    colbacktitle=white,
    sharp corners,
    boxrule=0.5pt,
    boxsep=1mm,
    top=1mm,
    bottom=1mm,
    left=0mm,
    right=0mm
}

\newcommand{\algorithmSummaryDescription}[3]{
\begin{summaryStyle}{{\textbf{#1}} (\cref{#2})}
	\begin{algorithmic}[0] %
		#3
	\end{algorithmic}
\end{summaryStyle}
}

\providecommandordefault{\maxsteps}{N}
\providecommandordefault{\grad}{(\nabla \defaultLossFunction)}
\newcommand{\compMulti}{}

\newcommand{\defGD}{
\deterministicGDdef[false]
	{def:GD}
	{\GD\ optimization method}
	{ 
	$(\gamma_n)_{n \in \N} \subseteq [0,\infty)$, 
	}
	{\GD\ process for the objective function $\defaultLossFunction$ with learning rates $(\gamma_n)_{n \in \N}$ and initial value $\xi$}
	{it holds for all $n \in \N$ that 
	\begin{equation}
	\label{def:GD:eq1}
	\Theta_0 = \xi \qandq \Theta_n = \Theta_{n-1} - \gamma_n \defaultGradientFunction(\Theta_{n-1}).
	\end{equation}
	}
}

\newcommand{\iterationStepDetermGD}{
	\State $\Theta \assign \Theta - \gamma_n \grad(\Theta)$
}
\newcommand{\algDescrDetermGD}{\algorithmicDescription[false]
	{\GD\ optimization method}
	{def:GD}
	{ 
   	$(\gamma_n)_{n \in \N} \subseteq [0,\infty)$,}
	{
		 \GD\ process for the objective function $\defaultLossFunction$ with learning rates $(\gamma_n)_{n \in \N}$ and initial value $\xi$
	}
	{$\Theta \assign \xi$}
	{
		\iterationStepDetermGD
	}
}

\newcommand{\iterationStepSGD}{
	\pseudoCodeGradientAssign
	\State $\Theta \assign \Theta - \gamma_n \pseudoCodeGradient$
}
\newcommand{\algDescrSGD}{\algorithmicDescriptionStochastic[false]
	{\SGD\ optimization method}
	{def:SGD}
	{ 
   	$(\gamma_n)_{n \in \N} \subseteq [0,\infty)$,
	}
	{
	 	\SGD\ process for the loss function $\defaultStochLoss$ with learning rates $(\gamma_n)_{n \in \N}$, initial value $\xi$,
	}
	{$\Theta \assign \xi$
	}{
		\iterationStepSGD
	}
}

\newcommand{\defSGD}{
\SGDdef[false]
	{def:SGD}
	{\SGD\ optimization method}
	{ 
	$(\gamma_n)_{n \in \N} \subseteq [0,\infty)$,}
	{\SGD\ process for the loss function $\defaultStochLoss$ with learning rates $(\gamma_n)_{n \in \N}$, 
	initial value $\xi$,}
	{it holds for all $n \in \N$ that 
	\begin{equation}
	\label{def:SGD:eq1}
	\Theta_0 = \xi \qandq \Theta_n = \Theta_{n-1} - \gamma_n \br*{\frac{1}{J_n}\sum_{j = 1}^{J_n}\defaultStochGradient(\Theta_{n-1}, X_{n,j})}. 
	\end{equation}
	}
}

\newcommand{\defmidpointGD}{
\deterministicGDdef[false]
	{def:midpointGD}
	{Explicit midpoint \GD\ optimization method}
	{ 
	$(\gamma_n)_{n \in \N} \subseteq [0,\infty)$, }
	{
		explicit midpoint \GD\ process for the objective function $\defaultLossFunction$ with learning rates $(\gamma_n)_{n \in \N}$ and initial value $\xi$
	}
	{it holds for all $n \in \N$ that 
	\begin{equation}
	\label{def:midpointGD:eq1}
	  \Theta_0 = 
	  \xi 
	\qandq 
	  \Theta_n 
	  = 
	  \Theta_{n-1} - \gamma_n 
	  \defaultGradientFunction\pr[\big]{
	    \Theta_{n-1} - \tfrac{ \gamma_n }{ 2 } 
	    \defaultGradientFunction( 
	      \Theta_{n-1}
	    )
	  }
	  .
	\end{equation}
	}
}

\newcommand{\iterationStepDetermmidpointGD}{
	\State 
	$
		\Theta 
	\assign 
		\Theta - \gamma_n 
		\grad\pr[\big]{
			\Theta - \tfrac{ \gamma_n }{ 2 } 
			\grad( 
			\Theta
		)
		}
	$
}
\newcommand{\algDescrDetermmidpointGD}{\algorithmicDescription[false]
	{Explicit midpoint \GD\ optimization method}
	{def:midpointGD}
	{ 
   	$(\gamma_n)_{n \in \N} \subseteq [0,\infty)$,}
	{
		explicit midpoint \GD\ process for the objective function $\defaultLossFunction$ with learning rates $(\gamma_n)_{n \in \N}$ and initial value $\xi$
	}
	{$\Theta \assign \xi$}
	{
		\iterationStepDetermmidpointGD
	}
}

\newcommand{\iterationStepmidpointSGD}{
	\State 
		$\pseudoCodeGradient_1 \assign \frac{1}{J_n}\sum_{j = 1}^{J_n} (\nabla_\theta \defaultStochLoss)(\Theta_{n-1}, X_{n,j})$
	\State 
		$\pseudoCodeGradient_2 \assign \frac{1}{J_n}\sum_{j = 1}^{J_n} (\nabla_\theta \defaultStochLoss)(\Theta_{n-1} - \tfrac{ \gamma_n }{ 2 } \pseudoCodeGradient_1, X_{n,j})$
	\State 
	$
		\Theta 
	\assign 
		\Theta - \gamma_n 
		\pseudoCodeGradient_2
	$
}
\newcommand{\algDescrmidpointSGD}{\algorithmicDescriptionStochastic[false]
	{Explicit midpoint \SGD\ optimization method}
	{def:midpointSGD}
	{
   	$(\gamma_n)_{n \in \N} \subseteq [0,\infty)$,
	}
	{
		explicit midpoint \SGD\ process for the loss function $\defaultStochLoss$ with learning rates $(\gamma_n)_{n \in \N}$, initial value $\xi$,
	}
	{$\Theta \assign \xi$}
	{
		\iterationStepmidpointSGD
	}
}

\newcommand{\defmidpointSGD}{
\SGDdef[false]
	{def:midpointSGD}{Explicit midpoint \SGD\ optimization method}
	{ 
	$(\gamma_n)_{n \in \N} \subseteq [0,\infty)$,}
	{explicit midpoint \SGD\ process for the loss function $\defaultStochLoss$ with learning rates $(\gamma_n)_{n \in \N}$, initial value $\xi$,}
	{it holds for all $n \in \N$ that
	\begin{equation}
	\label{def:midpointSGD:eq1}
		\Theta_0 
	= 
		\xi 
	\quad\text{and}\quad 
		\Theta_n 
	= 
		\Theta_{n-1} - \gamma_n 
		\br*{
			\frac{1}{J_n}
			\sum_{j = 1}^{J_n}
				\defaultStochGradient\pr*{
					\Theta_{n-1} - \frac{\gamma_n}{2} 
						\br*{
							\tfrac{1}{J_n}
							\textstyle{\sum_{j = 1}^{J_n}}
								\defaultStochGradient(
									\Theta_{n-1}, 
									X_{n,j}
								)
						}
					,
					X_{n,j}
				}
		}. 
	\end{equation} 
	}
}

\newcommand{\defdetermMomentum}{
\deterministicGDdef[false]
	[momentum \GD\ process (\first version) for the objective function $\defaultLossFunction$ with learning rates $(\gamma_n)_{n \in \N}$, 
	momentum decay factors $(\alpha_n)_{n \in \N}$, 
	and initial value $\xi$]
	{def:determ_momentum}
	{Momentum \GD\ optimization method}
	{ 
	$(\gamma_n)_{n \in \N} \subseteq [0,\infty)$, 
	$(\alpha_n)_{n \in \N} \subseteq [0,1]$, }
	{momentum \GD\ process for the objective function $\defaultLossFunction$ with learning rates $(\gamma_n)_{n \in \N}$, 
	momentum decay factors $(\alpha_n)_{n \in \N}$, 
	and initial value $\xi$}
	{there exists $\mathbf{m} \colon \N_0 \to \R^\defaultParamDim$ such that for all $n \in \N$ it holds that
	\begin{equation}
	\label{eq:def:momentum_1_1}
	\Theta_0 = \xi, \qquad \mathbf{m}_0 = 0,
	\end{equation}
	\begin{equation}
	\label{eq:def:momentum_1_2}
	\mathbf{m}_n = \alpha_n \mathbf{m}_{n-1} + (1-\alpha_n)\defaultGradientFunction(\Theta_{n-1}), 
	\end{equation}
	\begin{equation}
	\label{eq:def:momentum_1_3}
	\andq \Theta_n = \Theta_{n-1} - \gamma_n  \mathbf{m}_n. 
	\end{equation}
	}
}

\newcommand{\iterationStepDetermMomentum}{
	\State 
	$
		\mathbf{m} \assign \alpha_n \mathbf{m} + (1-\alpha_n)\grad(\Theta)
	$
	\State
	$
		\Theta \assign \Theta - \gamma_n  \mathbf{m}
	$
}
\newcommand{\algDescrDetermMomentum}{\algorithmicDescription[false]{Momentum \GD\ optimization method}{def:determ_momentum}
	{ 
	$(\gamma_n)_{n \in \N} \subseteq [0,\infty)$,
	$(\alpha_n)_{n \in \N} \subseteq [0,1]$,
   	}
{
	momentum \GD\ process
	for the objective function $\defaultLossFunction$ with learning rates $(\gamma_n)_{n \in \N}$, 
	momentum decay factors $(\alpha_n)_{n \in \N}$, 
	and initial value $\xi$
}{
	$\Theta \assign \xi$\separator $\mathbf{m} \assign 0 \in \R^\defaultParamDim$
}{
	\iterationStepDetermMomentum
}
}

\newcommand{\iterationStepMomentum}{
	\pseudoCodeGradientAssign
	\State 
	$
		\mathbf{m} \assign \alpha_n \mathbf{m} + (1-\alpha_n) \pseudoCodeGradient
	$
	\State
	$
		\Theta \assign \Theta - \gamma_n  \mathbf{m}
	$
}
\newcommand{\algDescrMomentum}{\algorithmicDescriptionStochastic[false]
	{Momentum \SGD\ optimization method}
	{def:momentum}
	{ 
	$(\gamma_n)_{n \in \N} \subseteq [0,\infty)$,
	$(\alpha_n)_{n \in \N} \subseteq [0,1]$,}
	{
	momentum \SGD\ process for the loss function $\defaultStochLoss$ with learning rates $(\gamma_n)_{n \in \N}$, 
	momentum decay factors $(\alpha_n)_{n \in \N}$, 
	initial value $\xi$,
}{
	$\Theta \assign \xi$\separator $\mathbf{m} \assign 0 \in \R^\defaultParamDim$
}{
	\iterationStepMomentum
}
}

\newcommand{\defmomentum}{
\SGDdef[false]
	[momentum \SGD\ process (\first version) for the loss function $\defaultStochLoss$ with learning rates $(\gamma_n)_{n \in \N}$, 
	momentum decay factors $(\alpha_n)_{n \in \N}$, 
	initial value $\xi$,]
	{def:momentum}
	{Momentum \SGD\ optimization method}
	{ 
	$(\gamma_n)_{n \in \N} \subseteq [0,\infty)$, 
	$(\alpha_n)_{n \in \N} \subseteq [0,1] $,}
	{momentum \SGD\ process for the loss function $\defaultStochLoss$ with learning rates $(\gamma_n)_{n \in \N}$, 
	momentum decay factors $(\alpha_n)_{n \in \N}$, 
	initial value $\xi$,}
	{there exists $\mathbf{m} \colon \N_0 \times \Omega \to \R^\defaultParamDim$ such that for all $n \in \N$ it holds that
	\begin{equation}
	\Theta_0 = \xi, \qquad \mathbf{m}_0 = 0,
	\end{equation}
	\begin{equation}
	\label{def:momentum:eq1}
	  \mathbf{m}_n = \alpha_n \mathbf{m}_{n-1} + (1-\alpha_n)\br*{\frac{1}{J_n}\sum_{j = 1}^{J_n}\defaultStochGradient(\Theta_{n-1}, X_{n,j}) },
	\end{equation}
	\begin{equation}
	  \andq 
	  \Theta_n = \Theta_{n-1} - \gamma_n  \mathbf{m}_n .
	\end{equation}
	}
}

\newcommand{\defdetermMomentumTwo}{
\deterministicGDdef[false]
	{def:determ_momentum_two}
	{Momentum \GD\ optimization method (\second version)}
	{
	$(\gamma_n)_{n \in \N} \subseteq [0,\infty)$,
	$(\alpha_n)_{n \in \N} \subseteq [0,\infty)$,}
	{momentum \GD\ process (\second version) for the objective function $\defaultLossFunction$ with learning rates $(\gamma_n)_{n \in \N}$,
	momentum decay factors $(\alpha_n)_{n \in \N}$,
	and initial value $\xi$}
	{there exists $\mathbf{m} \colon \N_0 \to \R^\defaultParamDim$ such that for all $n \in \N$ it holds that
	\begin{equation}
	\label{eq:def:momentum_2_1}
	\Theta_0 = \xi, \qquad \mathbf{m}_0 = 0,
	\end{equation}
	\begin{equation}
	\label{eq:def:momentum_2_2}
	\mathbf{m}_n = \alpha_n \mathbf{m}_{n-1} + \defaultGradientFunction(\Theta_{n-1}),
	\end{equation}
	\begin{equation}
	\label{eq:def:momentum_2_3}
	\andq \Theta_n = \Theta_{n-1} - \gamma_n  \mathbf{m}_n.
	\end{equation}
	}
}

\newcommand{\iterationStepDetermMomentumTwo}{
	\State 
	$
		\mathbf{m} \assign \alpha_n \mathbf{m} + \grad(\Theta)
	$
	\State
	$
		\Theta \assign \Theta - \gamma_n  \mathbf{m}
	$
}
\newcommand{\algDescrDetermMomentumTwo}{\algorithmicDescription[false]{Momentum \GD\ optimization method (\second version)}{def:determ_momentum_two}{
	$(\gamma_n)_{n \in \N} \subseteq [0,\infty)$,
	$(\alpha_n)_{n \in \N} \subseteq [0,\infty)$,
}{
	momentum \GD\ process (\second version)
	for the objective function $\defaultLossFunction$ with learning rates $(\gamma_n)_{n \in \N}$, 
	momentum decay factors $(\alpha_n)_{n \in \N}$, 
	and initial value $\xi$
}{
	$\Theta \assign \xi$\separator $\mathbf{m} \assign 0 \in \R^\defaultParamDim$
}{
	\iterationStepDetermMomentumTwo
}
}

\newcommand{\iterationStepMomentumTwo}{
	\pseudoCodeGradientAssign
	\State 
	$
		\mathbf{m} \assign \alpha_n \mathbf{m} + \pseudoCodeGradient
	$
	\State
	$
		\Theta \assign \Theta - \gamma_n  \mathbf{m}
	$
}
\newcommand{\algDescrMomentumTwo}{\algorithmicDescriptionStochastic[false]{Momentum \SGD\ optimization method (\second version)}{def:momentum_two}{ 
	$(\gamma_n)_{n \in \N} \subseteq [0,\infty)$,
	$(\alpha_n)_{n \in \N} \subseteq [0,\infty)$,
}{
	momentum \SGD\ process (\second version)
	for the loss function $\defaultStochLoss$ with learning rates $(\gamma_n)_{n \in \N}$, 
	momentum decay factors $(\alpha_n)_{n \in \N}$, 
	initial value $\xi$,
}{
	$\Theta \assign \xi$\separator $\mathbf{m} \assign 0 \in \R^\defaultParamDim$
}{
	\iterationStepMomentumTwo
}
}

\newcommand{\defmomentumTwo}{
\SGDdef[false]
	{def:momentum_two}
	{Momentum \SGD\ optimization method (\second version)}
	{
	$(\gamma_n)_{n \in \N} \subseteq [0,\infty)$,
	$(\alpha_n)_{n \in \N} \subseteq [0,\infty)$,}
	{momentum \SGD\ process (\second version) for the loss function $\defaultStochLoss$ with learning rates $(\gamma_n)_{n \in \N}$,
	momentum decay factors $(\alpha_n)_{n \in \N}$,
	initial value $\xi$,}
	{there exists $\mathbf{m} \colon \N_0 \to \R^\defaultParamDim$ such that for all $n \in \N$ it holds that
		it holds for all $n \in \N$ that 
		\begin{equation}
		\Theta_0 = \xi, \qquad \mathbf{m}_0 = 0,
		\end{equation}
		\begin{equation}
		\label{def:momentum_two:eq1}
		  \mathbf{m}_n = \alpha_n \mathbf{m}_{n-1} + \br*{\frac{1}{J_n}\sum_{j = 1}^{J_n}\defaultStochGradient(\Theta_{n-1}, X_{n,j}) }, 
		\end{equation}
		\begin{equation}
		  \andq 
		  \Theta_n = \Theta_{n-1} - \gamma_n  \mathbf{m}_n . 
		\end{equation}
	}
}

\newcommand{\defdetermMomentumThree}{
\deterministicGDdef[false]
	{def:determ_momentum_three}
	{Momentum \GD\ optimization method (\third version)}
	{ 
	$(\gamma_n)_{n \in \N} \subseteq [0,\infty)$,
	$(\alpha_n)_{n \in \N} \subseteq [0,\infty)$,}
	{momentum \GD\ process (\third version) for the objective function $\defaultLossFunction$ with learning rates $(\gamma_n)_{n \in \N}$,
	momentum decay factors $(\alpha_n)_{n \in \N}$,
	and initial value $\xi$}
	{there exists $\mathbf{m} \colon \N_0 \to \R^\defaultParamDim$ such that for all $n \in \N$ it holds that
	\begin{equation}
	\label{eq:def:momentum_3_1}
	\Theta_0 = \xi, \qquad \mathbf{m}_0 = 0,
	\end{equation}
	\begin{equation}
	\label{eq:def:momentum_3_2}
	\mathbf{m}_n = \alpha_n \mathbf{m}_{n-1} + (1-\alpha_n)\gamma_n \grad(\Theta_{n-1}),
	\end{equation}
	\begin{equation}
	\label{eq:def:momentum_3_3}
	\andq \Theta_n = \Theta_{n-1} -  \mathbf{m}_n.
	\end{equation}
	}
}

\newcommand{\iterationStepDetermMomentumThree}{
	\State 
	$
		\mathbf{m} \assign \alpha_n \mathbf{m} + (1-\alpha_n)\gamma_n \grad(\Theta)
	$
	\State
	$
		\Theta \assign \Theta -  \mathbf{m}
	$
}
\newcommand{\algDescrDetermMomentumThree}{\algorithmicDescription[false]{Momentum \GD\ optimization method (\third version)}{def:determ_momentum_three}{ 
	$(\gamma_n)_{n \in \N} \subseteq [0,\infty)$,
	$(\alpha_n)_{n \in \N} \subseteq [0,1]$,
}{
	momentum \GD\ process (\third version)
	for the objective function $\defaultLossFunction$ with learning rates $(\gamma_n)_{n \in \N}$, 
	momentum decay factors $(\alpha_n)_{n \in \N}$, 
	and initial value $\xi$
}{
	$\Theta \assign \xi$\separator $\mathbf{m} \assign 0 \in \R^\defaultParamDim$
}{
	\iterationStepDetermMomentumThree
}
}

\newcommand{\iterationStepMomentumThree}{
	\pseudoCodeGradientAssign
	\State 
	$
		\mathbf{m} \assign \alpha_n \mathbf{m} + (1-\alpha_n)\gamma_n \pseudoCodeGradient
	$
	\State
	$
		\Theta \assign \Theta -  \mathbf{m}
	$
}
\newcommand{\algDescrMomentumThree}{\algorithmicDescriptionStochastic[false]{Momentum \SGD\ optimization method (\third version)}{def:momentum_three}{ 
	$(\gamma_n)_{n \in \N} \subseteq [0,\infty)$,
	$(\alpha_n)_{n \in \N} \subseteq [0,1]$,
}{
	momentum \SGD\ process (\third version)
	for the loss function $\defaultStochLoss$ with learning rates $(\gamma_n)_{n \in \N}$, 
	momentum decay factors $(\alpha_n)_{n \in \N}$, 
	initial value $\xi$,
}
{
	$\Theta \assign \xi$\separator $\mathbf{m} \assign 0 \in \R^\defaultParamDim$
}{
	\iterationStepMomentumThree
}
}

\newcommand{\defmomentumThree}{
\SGDdef[false]
	{def:momentum_three}
	{Momentum \SGD\ optimization method (\third version)}
	{
	$(\gamma_n)_{n \in \N} \subseteq [0,\infty)$,
	$(\alpha_n)_{n \in \N} \subseteq [0,1]$,}
	{momentum \SGD\ process (\third version) for the loss function $\defaultStochLoss$ with learning rates $(\gamma_n)_{n \in \N}$,
	momentum decay factors $(\alpha_n)_{n \in \N}$,
	initial value $\xi$,}
	{there exists $\mathbf{m} \colon \N_0 \to \R^\defaultParamDim$ such that for all $n \in \N$ it holds that
		it holds for all $n \in \N$ that 
		\begin{equation}
		\Theta_0 = \xi, \qquad \mathbf{m}_0 = 0,
		\end{equation}
		\begin{equation}
		\label{def:momentum_three:eq1}
		  \mathbf{m}_n = \alpha_n \mathbf{m}_{n-1} + (1-\alpha_n)\gamma_n \br*{\frac{1}{J_n}\sum_{j = 1}^{J_n}\defaultStochGradient(\Theta_{n-1}, X_{n,j}) },
		\end{equation}
		\begin{equation}
		  \andq 
		  \Theta_n = \Theta_{n-1} -  \mathbf{m}_n . 
		\end{equation}
	}
}

\newcommand{\defdetermMomentumFour}{
\deterministicGDdef[false]
	{def:determ_momentum_four}
	{Momentum \GD\ optimization method (\fourth version)}
	{ 
	$(\gamma_n)_{n \in \N} \subseteq [0,\infty)$,
	$(\alpha_n)_{n \in \N} \subseteq [0,1]$,}
	{momentum \GD\ process (\fourth version) for the objective function $\defaultLossFunction$ with learning rates $(\gamma_n)_{n \in \N}$,
	momentum decay factors $(\alpha_n)_{n \in \N}$,
	and initial value $\xi$}
	{there exists $\mathbf{m} \colon \N_0 \to \R^\defaultParamDim$ such that for all $n \in \N$ it holds that
	\begin{equation}
	\label{eq:def:momentum_4_1}
	\Theta_0 = \xi, \qquad \mathbf{m}_0 = 0,
	\end{equation}
	\begin{equation}
	\label{eq:def:momentum_4_2}
	\mathbf{m}_n = \alpha_n \mathbf{m}_{n-1} + \gamma_n \grad(\Theta_{n-1}),
	\end{equation}
	\begin{equation}
	\label{eq:def:momentum_4_3}
	\andq \Theta_n = \Theta_{n-1} -  \mathbf{m}_n.
	\end{equation}
	}
}

\newcommand{\iterationStepDetermMomentumFour}{
	\State 
	$
		\mathbf{m} \assign \alpha_n \mathbf{m} + \gamma_n \grad(\Theta)
	$
	\State
	$
		\Theta \assign \Theta -  \mathbf{m}
	$
}
\newcommand{\algDescrDetermMomentumFour}{\algorithmicDescription[false]{Momentum \GD\ optimization method (\fourth version)}{def:determ_momentum_four}{ 
	$(\gamma_n)_{n \in \N} \subseteq [0,\infty)$,
	$(\alpha_n)_{n \in \N} \subseteq [0,\infty)$,
}{
	momentum \GD\ process (\fourth version)
	for the objective function $\defaultLossFunction$ with learning rates $(\gamma_n)_{n \in \N}$, 
	momentum decay factors $(\alpha_n)_{n \in \N}$, 
	and initial value $\xi$
}{
	$\Theta \assign \xi$\separator $\mathbf{m} \assign 0 \in \R^\defaultParamDim$
}{
	\iterationStepDetermMomentumFour
}
}

\newcommand{\iterationStepMomentumFour}{
	\pseudoCodeGradientAssign
	\State 
	$
		\mathbf{m} \assign \alpha_n \mathbf{m} + \gamma_n \pseudoCodeGradient
	$
	\State
	$
		\Theta \assign \Theta -  \mathbf{m}
	$
}
\newcommand{\algDescrMomentumFour}{\algorithmicDescriptionStochastic[false]{Momentum \SGD\ optimization method (\fourth version)}{def:momentum_four}{ 
	$(\gamma_n)_{n \in \N} \subseteq [0,\infty)$,
	$(\alpha_n)_{n \in \N} \subseteq [0,\infty)$,
}{
	momentum \SGD\ process (\fourth version)
	for the loss function $\defaultStochLoss$ with learning rates $(\gamma_n)_{n \in \N}$, 
	momentum decay factors $(\alpha_n)_{n \in \N}$, 
	initial value $\xi$,
}
{
	$\Theta \assign \xi$\separator $\mathbf{m} \assign 0 \in \R^\defaultParamDim$
}{
	\iterationStepMomentumFour
}
}

\newcommand{\defmomentumFour}{
\SGDdef[false]
	{def:momentum_four}
	{Momentum \SGD\ optimization method (\fourth version)}
	{
	$(\gamma_n)_{n \in \N} \subseteq [0,\infty)$,
	$(\alpha_n)_{n \in \N} \subseteq [0,\infty)$,}
	{momentum \SGD\ process (\fourth version) for the loss function $\defaultStochLoss$ with learning rates $(\gamma_n)_{n \in \N}$,
	momentum decay factors $(\alpha_n)_{n \in \N}$,
	initial value $\xi$,}
	{there exists $\mathbf{m} \colon \N_0 \to \R^\defaultParamDim$ such that for all $n \in \N$ it holds that
		it holds for all $n \in \N$ that 
		\begin{equation}
		\Theta_0 = \xi, \qquad \mathbf{m}_0 = 0,
		\end{equation}
		\begin{equation}
		\label{def:momentum_four:eq1}
		  \mathbf{m}_n = \alpha_n \mathbf{m}_{n-1} + \gamma_n \br*{\frac{1}{J_n}\sum_{j = 1}^{J_n}\defaultStochGradient(\Theta_{n-1}, X_{n,j}) },
		\end{equation}
		\begin{equation}
		  \andq 
		  \Theta_n = \Theta_{n-1} -  \mathbf{m}_n . 
		\end{equation}
	}
}

\newcommand{\defdetermMomentumBias}{
\deterministicGDdef[false]
	{def:determ_momentum_bias}
	{Bias-adjusted momentum \GD\ optimization method}
	{ 
	$(\gamma_n)_{n \in \N} \subseteq [0,\infty)$, 
	$(\alpha_n)_{n \in \N} \subseteq [0,1)$, }
	{bias-adjusted momentum \GD\ process for the objective function $\defaultLossFunction$ with learning rates $(\gamma_n)_{n \in \N}$, 
	momentum decay factors $(\alpha_n)_{n \in \N}$, 
	and initial value $\xi$}
	{there exists $\mathbf{m} \colon \N_0 \to \R^\defaultParamDim$ such that for all $n \in \N$ it holds that
	\begin{equation}
	\label{determ_momentum_bias:eq1}
	\Theta_0 = \xi, \qquad \mathbf{m}_0 = 0,
	\end{equation}
	\begin{equation}
	\label{determ_momentum_bias:eq2}
	\mathbf{m}_n = \alpha_n \mathbf{m}_{n-1} + (1-\alpha_n)\defaultGradientFunction(\Theta_{n-1}), 
	\end{equation}
	\begin{equation}
	\label{determ_momentum_bias:eq3}
	\andq \Theta_n = \Theta_{n-1} - \frac{\gamma_n  \mathbf{m}_n}{1- \prod_{l=1}^n\alpha_l}. 
	\end{equation}
	}
}

\newcommand{\iterationStepDetermMomentumBias}{
	\State 
	$
		\mathbf{m} \assign \alpha_n \mathbf{m} + (1-\alpha_n)\grad(\Theta)
	$
	\vspace{2mm}
	\State
	$
		\Theta \assign \Theta - \frac{\gamma_n  \mathbf{m}}{1- \prod_{l=1}^n\alpha_l}
	$
}
\newcommand{\algDescrDetermMomentumBias}{\algorithmicDescription[false]{Bias-adjusted momentum \GD\ optimization method}{def:determ_momentum_bias}{ 
	$(\gamma_n)_{n \in \N} \subseteq [0,\infty)$,
	$(\alpha_n)_{n \in \N} \subseteq [0,1)$,
}{
	bias-adjusted momentum \GD\ process
	for the objective function $\defaultLossFunction$ with learning rates $(\gamma_n)_{n \in \N}$, 
	momentum decay factors $(\alpha_n)_{n \in \N}$, 
	and initial value $\xi$
}{
	$\Theta \assign \xi$\separator $\mathbf{m} \assign 0 \in \R^\defaultParamDim$
}{
	\iterationStepDetermMomentumBias
}
}

\newcommand{\iterationStepMomentumBias}{
	\pseudoCodeGradientAssign
	\State 
	$
		\mathbf{m} \assign \alpha_n \mathbf{m} + (1-\alpha_n)\pseudoCodeGradient
	$
	\vspace{2mm}
	\State
	$
		\Theta \assign \Theta - \frac{\gamma_n  \mathbf{m}}{1- \prod_{l=1}^n\alpha_l}
	$
}
\newcommand{\algDescrMomentumBias}{\algorithmicDescriptionStochastic[false]{Bias-adjusted momentum \SGD\ optimization method}{def:momentum_bias}
	{ 
	$(\gamma_n)_{n \in \N} \subseteq [0,\infty)$,
	$(\alpha_n)_{n \in \N} \subseteq [0,1)$,}
{
	bias-adjusted momentum \SGD\ process for the loss function $\defaultStochLoss$ with learning rates $(\gamma_n)_{n \in \N}$, 
	momentum decay factors $(\alpha_n)_{n \in \N}$, 
	initial value $\xi$,
}{
	$\Theta \assign \xi$\separator $\mathbf{m} \assign 0 \in \R^\defaultParamDim$
}{
	\iterationStepMomentumBias
}
}

\newcommand{\defStochMomentumBias}{
\SGDdef[false]
	{def:momentum_bias}
	{Bias-adjusted momentum \SGD\ optimization method}
	{
	$(\gamma_n)_{n \in \N} \subseteq [0,\infty)$,
	$(\alpha_n)_{n \in \N} \subseteq [0,1)$,}
	{bias-adjusted momentum \SGD\ process for the loss function $\defaultStochLoss$ with learning rates $(\gamma_n)_{n \in \N}$,
	momentum decay factors $(\alpha_n)_{n \in \N}$,
	initial value $\xi$,}
	{there exists $\mathbf{m} \colon \N_0 \to \R^\defaultParamDim$ such that for all $n \in \N$ it holds that
	\begin{equation}
	\label{def:momentum_bias:eq1}
	\Theta_0 = \xi, \qquad \mathbf{m}_0 = 0,
	\end{equation}
	\begin{equation}
	\label{def:momentum_bias:eq2}
	\mathbf{m}_n = \alpha_n \mathbf{m}_{n-1} + (1-\alpha_n)\br*{\frac{1}{J_n}\sum_{j = 1}^{J_n}\defaultStochGradient(\Theta_{n-1}, X_{n,j}) },
	\end{equation}
	\begin{equation}
	\label{def:momentum_bias:eq3}
	\andq \Theta_n = \Theta_{n-1} -  \frac{\gamma_n  \mathbf{m}_n}{1- \prod_{l=1}^n\alpha_l}.
	\end{equation}
	}
}

\newcommand{\defdetermNesterov}{
\deterministicGDdef[false]
	[Nesterov accelerated \GD\ process (\first version) for the objective function $\defaultLossFunction$ with learning rates $(\gamma_n)_{n \in \N}$, 
	momentum decay factors $(\alpha_n)_{n \in \N}$, 
	and initial value $\xi$]
	{def:determ_nesterov}
	{Nesterov accelerated \GD\ optimization method}
	{ 
	$(\gamma_n)_{n \in \N} \subseteq [0,\infty)$, 
	$(\alpha_n)_{n \in \N} \subseteq [0,1]$, }
	{Nesterov accelerated \GD\ process for the objective function $\defaultLossFunction$ with learning rates $(\gamma_n)_{n \in \N}$, 
	momentum decay factors $(\alpha_n)_{n \in \N}$, 
	and initial value $\xi$}
	{there exists $\mathbf{m} \colon \N_0 \to \R^\defaultParamDim$ such that for all $n \in \N$ it holds that
	\begin{equation}
	\label{eq:def:nesterov_1_1}
	\Theta_0 = \xi, \qquad \mathbf{m}_0 = 0,
	\end{equation}
	\begin{equation}
	\label{eq:def:nesterov_1_2}
	\mathbf{m}_n = \alpha_n \mathbf{m}_{n-1} + (1-\alpha_n)  \defaultGradientFunction(\Theta_{n-1} - \gamma_n\alpha_n \mathbf{m}_{n-1}), 
	\end{equation}
	\begin{equation}
	\label{eq:def:nesterov_1_3}
	\andq \Theta_n = \Theta_{n-1} - \gamma_n  \mathbf{m}_n. 
	\end{equation}
	}
}

\newcommand{\iterationStepDetermNesterov}{
	\State 
	$
		\mathbf{m} \assign \alpha_n \mathbf{m} + (1-\alpha_n)  \grad(\Theta - \gamma_n\alpha_n \mathbf{m})
	$
	\State
	$
		\Theta \assign \Theta - \gamma_n  \mathbf{m}
	$
}
\newcommand{\algDescrDetermNesterov}{\algorithmicDescription[false]{Nesterov accelerated \GD\ optimization method}{def:determ_nesterov}{ 
	$(\gamma_n)_{n \in \N} \subseteq [0,\infty)$,
	$(\alpha_n)_{n \in \N} \subseteq [0,1]$,
}{
	Nesterov accelerated \GD\ process for the objective function $\defaultLossFunction$ 
	with learning rates $(\gamma_n)_{n \in \N}$, 
	momentum decay factors $(\alpha_n)_{n \in \N}$, 
	and initial value $\xi$
}{
	$\Theta \assign \xi$\separator $\mathbf{m} \assign 0 \in \R^\defaultParamDim$
}{
	\iterationStepDetermNesterov
}
}

\newcommand{\iterationStepNesterov}{
	\pseudoCodeGradientAssign[\Theta - \gamma_n\alpha_n \mathbf{m}]
	\State 
	$
		\mathbf{m} \assign \alpha_n \mathbf{m} + (1-\alpha_n)  \pseudoCodeGradient
	$
	\State
	$
		\Theta \assign \Theta - \gamma_n  \mathbf{m}
	$
}
\newcommand{\algDescrNesterov}{\algorithmicDescriptionStochastic[false]{Nesterov accelerated \SGD\ optimization method}{def:nesterov}{ 
	$(\gamma_n)_{n \in \N} \subseteq [0,\infty)$,
	$(\alpha_n)_{n \in \N} \subseteq [0,1]$,
}{
	Nesterov accelerated \SGD\ process for the loss function $\defaultStochLoss$ 
	with learning rates $(\gamma_n)_{n \in \N}$, 
	momentum decay factors $(\alpha_n)_{n \in \N}$, 
	initial value $\xi$,
}{
	$\Theta \assign \xi$\separator $\mathbf{m} \assign 0 \in \R^\defaultParamDim$
}{
	\iterationStepNesterov
}
}

\newcommand{\defStochNesterov}{
\SGDdef[false]
	[Nesterov accelerated \SGD\ process (\first version) for the loss function $\defaultStochLoss$ with learning rates $(\gamma_n)_{n \in \N}$,
	momentum decay factors $(\alpha_n)_{n \in \N}$,
	initial value $\xi$,]
	{def:nesterov}
	{Nesterov accelerated \SGD\ optimization method}
	{
	$(\gamma_n)_{n \in \N} \subseteq [0,\infty)$,
	$(\alpha_n)_{n \in \N} \subseteq [0,1]$,}
	{Nesterov accelerated \SGD\ process for the loss function $\defaultStochLoss$ with learning rates $(\gamma_n)_{n \in \N}$,
	momentum decay factors $(\alpha_n)_{n \in \N}$,
	initial value $\xi$,}
	{there exists $\mathbf{m} \colon \N_0 \to \R^\defaultParamDim$ such that for all $n \in \N$ it holds that
	\begin{equation}
	\Theta_0 = \xi , 
	\qquad \mathbf{m}_0 = 0,
	\end{equation}
	\begin{equation}
	\mathbf{m}_n 
	= 
	\alpha_n \mathbf{m}_{n-1} 
	+ (1-\alpha_n)
	\br*{ \frac{ 1 }{ J_n } 
	\sum_{ j = 1 }^{ J_n }
	\defaultStochGradient\bpr{
		\Theta_{n-1} - \gamma_n \alpha_n \mathbf{m}_{n-1}, \, X_{n,j}} 
	} ,
	\end{equation}
	\begin{equation}
	\andq 
	\Theta_n = \Theta_{n-1} - \gamma_n  \mathbf{m}_n . 
	\end{equation}
	}
}

\newcommand{\defdetermNesterovTwo}{
\deterministicGDdef[false]
	{def:determ_nesterov_two}
	{Nesterov accelerated \GD\ optimization method (\second version)}
	{ 
	$(\gamma_n)_{n \in \N} \subseteq [0,\infty)$, 
	$(\alpha_n)_{n \in \N} \subseteq [0,\infty)$, }
	{Nesterov accelerated \GD\ process (\second version) for the objective function $\defaultLossFunction$ with learning rates $(\gamma_n)_{n \in \N}$, 
	momentum decay factors $(\alpha_n)_{n \in \N}$, 
	and initial value $\xi$}
	{there exists $\mathbf{m} \colon \N_0 \to \R^\defaultParamDim$ such that for all $n \in \N$ it holds that
	\begin{equation}
	\label{eq:def:nesterov_2_1}
	\Theta_0 = \xi, \qquad \mathbf{m}_0 = 0,
	\end{equation}
	\begin{equation}
	\label{eq:def:nesterov_2_2}
	\mathbf{m}_n = \alpha_n \mathbf{m}_{n-1} + \grad(\Theta_{n-1} - \gamma_n\alpha_n \mathbf{m}_{n-1}),
	\end{equation}
	\begin{equation}
	\label{eq:def:nesterov_2_3}
	\andq \Theta_n = \Theta_{n-1} - \gamma_n  \mathbf{m}_n.
	\end{equation}
	}
}

\newcommand{\iterationStepDetermNesterovTwo}{
	\State 
	$
		\mathbf{m} \assign \alpha_n \mathbf{m} + \grad(\Theta - \gamma_n\alpha_n \mathbf{m})
	$
	\State
	$
		\Theta \assign \Theta - \gamma_n  \mathbf{m}
	$
}
\newcommand{\algDescrDetermNesterovTwo}{\algorithmicDescription[false]{Nesterov accelerated \GD\ optimization method (\second version)}{def:determ_nesterov_two}{ 
	$(\gamma_n)_{n \in \N} \subseteq [0,\infty)$,
	$(\alpha_n)_{n \in \N} \subseteq [0,\infty)$,
}{
	Nesterov accelerated \GD\ process (\second version) for the objective function $\defaultLossFunction$ 
	with learning rates $(\gamma_n)_{n \in \N}$, 
	momentum decay factors $(\alpha_n)_{n \in \N}$, 
	and initial value $\xi$
}{
	$\Theta \assign \xi$\separator $\mathbf{m} \assign 0 \in \R^\defaultParamDim$
}{
	\iterationStepDetermNesterovTwo
}
}

\newcommand{\iterationStepNesterovTwo}{
	\pseudoCodeGradientAssign[\Theta - \gamma_n\alpha_n \mathbf{m}]
	\State 
	$
		\mathbf{m} \assign \alpha_n \mathbf{m} + \pseudoCodeGradient
	$
	\State
	$
		\Theta \assign \Theta - \gamma_n  \mathbf{m}
	$
}
\newcommand{\algDescrNesterovTwo}{\algorithmicDescriptionStochastic[false]{Nesterov accelerated \SGD\ optimization method (\second version)}{def:nesterov_two}{ 
	$(\gamma_n)_{n \in \N} \subseteq [0,\infty)$,
	$(\alpha_n)_{n \in \N} \subseteq [0,\infty)$,
}{
	Nesterov accelerated \SGD\ process (\second version) for the loss function $\defaultStochLoss$ 
	with learning rates $(\gamma_n)_{n \in \N}$, 
	momentum decay factors $(\alpha_n)_{n \in \N}$, 
	initial value $\xi$,
}{
	$\Theta \assign \xi$\separator $\mathbf{m} \assign 0 \in \R^\defaultParamDim$
}{
	\iterationStepNesterovTwo
}
}

\newcommand{\defStochNesterovTwo}{
\SGDdef[false]
	{def:nesterov_two}
	{Nesterov accelerated \SGD\ optimization method (\second version)}
	{
	$(\gamma_n)_{n \in \N} \subseteq [0,\infty)$,
	$(\alpha_n)_{n \in \N} \subseteq [0,\infty]$,}
	{Nesterov accelerated \SGD\ process (\second version) for the loss function $\defaultStochLoss$ with learning rates $(\gamma_n)_{n \in \N}$,
	momentum decay factors $(\alpha_n)_{n \in \N}$,
	initial value $\xi$,}
	{there exists $\mathbf{m} \colon \N_0 \to \R^\defaultParamDim$ such that for all $n \in \N$ it holds that
	\begin{equation}
	\Theta_0 = \xi , 
	\qquad \mathbf{m}_0 = 0,
	\end{equation}
	\begin{equation}
	\mathbf{m}_n 
	= 
	\alpha_n \mathbf{m}_{n-1} 
	+ \br*{ \frac{ 1 }{ J_n } 
	\sum_{ j = 1 }^{ J_n }
	\defaultStochGradient\bpr{
		\Theta_{n-1} - \gamma_n\alpha_n \mathbf{m}_{n-1}, \, X_{n,j}} 
	} ,
	\end{equation}
	\begin{equation}
	\andq 
	\Theta_n = \Theta_{n-1} - \gamma_n  \mathbf{m}_n . 
	\end{equation}
	}
}

\newcommand{\defdetermNesterovThree}{
\deterministicGDdef[false]
	{def:determ_nesterov_three}
	{Nesterov accelerated \GD\ optimization method (\third version)}
	{ 
	$(\gamma_n)_{n \in \N} \subseteq [0,\infty)$, 
	$(\alpha_n)_{n \in \N} \subseteq [0,1]$, }
	{Nesterov accelerated \GD\ process (\third version) for the objective function $\defaultLossFunction$ with learning rates $(\gamma_n)_{n \in \N}$, 
	momentum decay factors $(\alpha_n)_{n \in \N}$, 
	and initial value $\xi$}
	{there exists $\mathbf{m} \colon \N_0 \to \R^\defaultParamDim$ such that for all $n \in \N$ it holds that
	\begin{equation}
	\label{eq:def:nesterov_3_1}
	\Theta_0 = \xi, \qquad \mathbf{m}_0 = 0,
	\end{equation}
	\begin{equation}
	\label{eq:def:nesterov_3_2}
	\mathbf{m}_n = \alpha_n \mathbf{m}_{n-1} + (1-\alpha_n)\gamma_n \grad(\Theta_{n-1} - \alpha_n \mathbf{m}_{n-1}),
	\end{equation}
	\begin{equation}
	\label{eq:def:nesterov_3_3}
	\andq \Theta_n = \Theta_{n-1} - \mathbf{m}_n.
	\end{equation}
	}
}

\newcommand{\iterationStepDetermNesterovThree}{
	\State 
	$
		\mathbf{m} \assign \alpha_n \mathbf{m} + (1-\alpha_n)\gamma_n \grad(\Theta - \alpha_n \mathbf{m})
	$
	\State
	$
		\Theta \assign \Theta - \mathbf{m}
	$
}
\newcommand{\algDescrDetermNesterovThree}{\algorithmicDescription[false]{Nesterov accelerated \GD\ optimization method (\third version)}{def:determ_nesterov_three}{ 
	$(\gamma_n)_{n \in \N} \subseteq [0,\infty)$,
	$(\alpha_n)_{n \in \N} \subseteq [0,1]$,
}{
	Nesterov accelerated \GD\ process (\third version) for the objective function $\defaultLossFunction$ 
	with learning rates $(\gamma_n)_{n \in \N}$, 
	momentum decay factors $(\alpha_n)_{n \in \N}$, 
	and initial value $\xi$
}{
	$\Theta \assign \xi$\separator $\mathbf{m} \assign 0 \in \R^\defaultParamDim$
}{
	\iterationStepDetermNesterovThree
}
}

\newcommand{\iterationStepNesterovThree}{
	\pseudoCodeGradientAssign[\Theta - \alpha_n \mathbf{m}]
	\State 
	$
		\mathbf{m} \assign \alpha_n \mathbf{m} + (1-\alpha_n)\gamma_n \pseudoCodeGradient
	$
	\State
	$
		\Theta \assign \Theta - \mathbf{m}
	$
}
\newcommand{\algDescrNesterovThree}{\algorithmicDescriptionStochastic[false]{Nesterov accelerated \SGD\ optimization method (\third version)}{def:nesterov_three}{ 
	$(\gamma_n)_{n \in \N} \subseteq [0,\infty)$,
	$(\alpha_n)_{n \in \N} \subseteq [0,1]$,
}{
	Nesterov accelerated \SGD\ process (\third version) for the loss function $\defaultStochLoss$ 
	with learning rates $(\gamma_n)_{n \in \N}$, 
	momentum decay factors $(\alpha_n)_{n \in \N}$, 
	initial value $\xi$,
}{
	$\Theta \assign \xi$\separator $\mathbf{m} \assign 0 \in \R^\defaultParamDim$
}{
	\iterationStepNesterovThree
}
}

\newcommand{\defStochNesterovThree}{
\SGDdef[false]
	{def:nesterov_three}
	{Nesterov accelerated \SGD\ optimization method (\third version)}
	{
	$(\gamma_n)_{n \in \N} \subseteq [0,\infty)$,
	$(\alpha_n)_{n \in \N} \subseteq [0,1]$,}
	{Nesterov accelerated \SGD\ process (\third version) for the loss function $\defaultStochLoss$ with learning rates $(\gamma_n)_{n \in \N}$,
	momentum decay factors $(\alpha_n)_{n \in \N}$,
	initial value $\xi$,}
	{there exists $\mathbf{m} \colon \N_0 \to \R^\defaultParamDim$ such that for all $n \in \N$ it holds that
	\begin{equation}
	\Theta_0 = \xi , 
	\qquad \mathbf{m}_0 = 0,
	\end{equation}
	\begin{equation}
	\mathbf{m}_n 
	= 
	\alpha_n \mathbf{m}_{n-1} 
	+ 
	(1-\alpha_n)\gamma_n
	\br*{
		\frac{ 1 }{ J_n }
		\sum_{ j = 1 }^{ J_n }
		\defaultStochGradient\bpr{
			\Theta_{n-1} - \alpha_n \mathbf{m}_{n-1}, \, X_{n,j}}
	} ,
	\end{equation}
	\begin{equation}
	\andq 
	\Theta_n = \Theta_{n-1} - \mathbf{m}_n . 
	\end{equation}
	}
}

\newcommand{\defdetermNesterovFour}{
\deterministicGDdef[false]
	{def:determ_nesterov_four}
	{Nesterov accelerated \GD\ optimization method (\fourth version)}
	{ 
	$(\gamma_n)_{n \in \N} \subseteq [0,\infty)$, 
	$(\alpha_n)_{n \in \N} \subseteq [0,\infty)$, }
	{Nesterov accelerated \GD\ process (\fourth version) for the objective function $\defaultLossFunction$ with learning rates $(\gamma_n)_{n \in \N}$, 
	momentum decay factors $(\alpha_n)_{n \in \N}$, 
	and initial value $\xi$}
	{there exists $\mathbf{m} \colon \N_0 \to \R^\defaultParamDim$ such that for all $n \in \N$ it holds that
	\begin{equation}
	\label{eq:def:nesterov_4_1}
	\Theta_0 = \xi, \qquad \mathbf{m}_0 = 0,
	\end{equation}
	\begin{equation}
	\label{eq:def:nesterov_4_2}
	\mathbf{m}_n = \alpha_n \mathbf{m}_{n-1} + \gamma_n \grad(\Theta_{n-1} - \alpha_n \mathbf{m}_{n-1}),
	\end{equation}
	\begin{equation}
	\label{eq:def:nesterov_4_3}
	\andq \Theta_n = \Theta_{n-1} - \mathbf{m}_n.
	\end{equation}
	}
}

\newcommand{\iterationStepDetermNesterovFour}{
	\State 
	$
		\mathbf{m} \assign \alpha_n \mathbf{m} + \gamma_n \grad(\Theta - \alpha_n \mathbf{m})
	$
	\State
	$
		\Theta \assign \Theta - \mathbf{m}
	$
}
\newcommand{\algDescrDetermNesterovFour}{\algorithmicDescription[false]{Nesterov accelerated \GD\ optimization method (\fourth version)}{def:determ_nesterov_four}{ 
	$(\gamma_n)_{n \in \N} \subseteq [0,\infty)$,
	$(\alpha_n)_{n \in \N} \subseteq [0,\infty)$,
}{
	Nesterov accelerated \GD\ process (\fourth version) for the objective function $\defaultLossFunction$ 
	with learning rates $(\gamma_n)_{n \in \N}$, 
	momentum decay factors $(\alpha_n)_{n \in \N}$, 
	and initial value $\xi$
}{
	$\Theta \assign \xi$\separator $\mathbf{m} \assign 0 \in \R^\defaultParamDim$
}{
	\iterationStepDetermNesterovFour
}
}

\newcommand{\iterationStepNesterovFour}{
	\pseudoCodeGradientAssign[\Theta - \alpha_n \mathbf{m}]
	\State 
	$
		\mathbf{m} \assign \alpha_n \mathbf{m} + \gamma_n \pseudoCodeGradient
	$
	\State
	$
		\Theta \assign \Theta - \mathbf{m}
	$
}
\newcommand{\algDescrNesterovFour}{\algorithmicDescriptionStochastic[false]{Nesterov accelerated \SGD\ optimization method (\fourth version)}{def:nesterov_four}{ 
	$(\gamma_n)_{n \in \N} \subseteq [0,\infty)$,
	$(\alpha_n)_{n \in \N} \subseteq [0,\infty)$,
}{
	Nesterov accelerated \SGD\ process (\fourth version) for the loss function $\defaultStochLoss$ 
	with learning rates $(\gamma_n)_{n \in \N}$, 
	momentum decay factors $(\alpha_n)_{n \in \N}$, 
	initial value $\xi$,
}{
	$\Theta \assign \xi$\separator $\mathbf{m} \assign 0 \in \R^\defaultParamDim$
}{
	\iterationStepNesterovFour
}
}

\newcommand{\defStochNesterovFour}{
\SGDdef[false]
	{def:nesterov_four}
	{Nesterov accelerated \SGD\ optimization method (\fourth version)}
	{
	$(\gamma_n)_{n \in \N} \subseteq [0,\infty)$,
	$(\alpha_n)_{n \in \N} \subseteq [0,\infty]$,}
	{Nesterov accelerated \SGD\ process (\fourth version) for the loss function $\defaultStochLoss$ with learning rates $(\gamma_n)_{n \in \N}$,
	momentum decay factors $(\alpha_n)_{n \in \N}$,
	initial value $\xi$,}
	{there exists $\mathbf{m} \colon \N_0 \to \R^\defaultParamDim$ such that for all $n \in \N$ it holds that
	\begin{equation}
	\Theta_0 = \xi , 
	\qquad \mathbf{m}_0 = 0,
	\end{equation}
	\begin{equation}
	\mathbf{m}_n 
	= 
	\alpha_n \mathbf{m}_{n-1} 
	+ 
	\gamma_n
	\br*{
		\frac{ 1 }{ J_n }
		\sum_{ j = 1 }^{ J_n }
		\defaultStochGradient\bpr{
			\Theta_{n-1} - \alpha_n \mathbf{m}_{n-1}, \, X_{n,j}}
	} ,
	\end{equation}
	\begin{equation}
	\andq 
	\Theta_n = \Theta_{n-1} - \mathbf{m}_n . 
	\end{equation}
	}
}

\newcommand{\defdetermNesterovAlt}{
\deterministicGDdef[false]
	{def:determ_nesterov_shifted}
	{Shifted Nesterov accelerated \GD\ optimization method}
	{ 
	$(\gamma_n)_{n \in \N} \subseteq [0,\infty)$, 
	$(\alpha_n)_{n \in \N} \subseteq [0,1]$, }
	{shifted Nesterov accelerated \GD\ process for the objective function $\defaultLossFunction$ with learning rates $(\gamma_n)_{n \in \N}$, 
	momentum decay factors $(\alpha_n)_{n \in \N}$, 
	and initial value $\xi$}
	{there exists $\mathbf{m} \colon \N_0 \to \R^\defaultParamDim$ such that for all $n \in \N$ it holds that
	\begin{equation}
	\Theta_0 = \xi, \qquad \mathbf{m}_0 = 0,
	\end{equation}
	\begin{equation}
	\mathbf{m}_n = \alpha_n \mathbf{m}_{n-1} + (1-\alpha_n)  \grad(\Theta_{n-1}),
	\qand
	\end{equation}
	\begin{equation}
	\Theta_n = \Theta_{n-1} - \gamma_{n+1} \alpha_{n+1} \mathbf{m}_n - \gamma_n (1-\alpha_n) \grad(\Theta_{n-1}). 
	\end{equation}
	}
}

\newcommand{\iterationStepDetermNesterovAlt}{
	\State 
	$
		\mathbf{m} \assign \alpha_n \mathbf{m} + (1-\alpha_n)  \grad(\Theta)
	$
	\State
	$
		\Theta \assign \Theta - \gamma_{n+1} \alpha_{n+1} \mathbf{m} - \gamma_n (1-\alpha_n) \grad(\Theta)
	$
}
\newcommand{\algDescrDetermNesterovAlt}{\algorithmicDescription[false]{Shifted Nesterov accelerated \GD\ optimization method}{def:determ_nesterov_shifted}{ 
	$(\gamma_n)_{n \in \N} \subseteq [0,\infty)$,
	$(\alpha_n)_{n \in \N} \subseteq [0,1]$,
}{
	shifted Nesterov accelerated \GD\ process for the objective function $\defaultLossFunction$ 
	with learning rates $(\gamma_n)_{n \in \N}$, 
	momentum decay factors $(\alpha_n)_{n \in \N}$, 
	and initial value $\xi$
}{
	$\Theta \assign \xi$\separator $\mathbf{m} \assign 0 \in \R^\defaultParamDim$
}{
	\iterationStepDetermNesterovAlt
}
}

\newcommand{\iterationStepNesterovAlt}{
	\pseudoCodeGradientAssign
	\State 
	$
		\mathbf{m} \assign \alpha_n \mathbf{m} + (1-\alpha_n)  \pseudoCodeGradient
	$
	\State
	$
		\Theta \assign \Theta - \gamma_{n+1} \alpha_{n+1} \mathbf{m} - \gamma_n (1-\alpha_n) \pseudoCodeGradient
	$
}
\newcommand{\algDescrNesterovAlt}{\algorithmicDescriptionStochastic[false]{Shifted Nesterov accelerated \SGD\ optimization method}{def:nesterov_shifted}{ 
	$(\gamma_n)_{n \in \N} \subseteq [0,\infty)$,
	$(\alpha_n)_{n \in \N} \subseteq [0,1]$,
}{
	shifted Nesterov accelerated \SGD\ process for the loss function $\defaultStochLoss$ 
	with learning rates $(\gamma_n)_{n \in \N}$, 
	momentum decay factors $(\alpha_n)_{n \in \N}$, 
	initial value $\xi$,
}{
	$\Theta \assign \xi$\separator $\mathbf{m} \assign 0 \in \R^\defaultParamDim$
}{
	\iterationStepNesterovAlt
}
}

\newcommand{\defNesterovAlt}{
\SGDdef[false]
	{def:nesterov_shifted}
	{Shifted Nesterov accelerated \SGD\ optimization method}
	{ 
	$(\gamma_n)_{n \in \N} \subseteq [0,\infty)$, 
	$(\alpha_n)_{n \in \N} \subseteq [0,1]$, }
	{shifted Nesterov accelerated \SGD\ process for the loss function $\defaultStochLoss$ with learning rates $(\gamma_n)_{n \in \N}$, momentum decay factors $(\alpha_n)_{n \in \N}$, initial value $\xi$,}
	{there exists $\mathbf{m} \colon \N_0 \to \R^\defaultParamDim$ such that for all $n \in \N$ it holds that
	\begin{equation}
	\Theta_0 = \xi, \qquad \mathbf{m}_0 = 0,
	\end{equation}
	\begin{equation}
	\mathbf{m}_n = \alpha_n \mathbf{m}_{n-1} + (1-\alpha_n) 
	\br*{
		\frac{1}{J_n} \sum_{j=1}^{J_n} \defaultStochGradient\bpr{\Theta_{n-1}, X_{n,j}}
	}
	\qand
	\end{equation}
	\begin{equation}
	\Theta_n = \Theta_{n-1} - \gamma_{n+1} \alpha_{n+1} \mathbf{m}_n - \gamma_n (1-\alpha_n) 
	\br*{
		\frac{1}{J_n} \sum_{j=1}^{J_n} \defaultStochGradient\bpr{\Theta_{n-1}, X_{n,j}}
	}.
	\end{equation}
	}
}

\newcommand{\defdetermNesterovAltTwo}{
\deterministicGDdef[false]
	{def:determ_nesterov_shifted_two}
	{Shifted Nesterov accelerated \GD\ optimization method (\second version)}
	{ 
	$(\gamma_n)_{n \in \N} \subseteq [0,\infty)$, 
	$(\alpha_n)_{n \in \N} \subseteq [0,\infty)$, }
	{shifted Nesterov accelerated \GD\ process (\second version) for the objective function $\defaultLossFunction$ with learning rates $(\gamma_n)_{n \in \N}$, 
	momentum decay factors $(\alpha_n)_{n \in \N}$, 
	and initial value $\xi$}
	{there exists $\mathbf{m} \colon \N_0 \to \R^\defaultParamDim$ such that for all $n \in \N$ it holds that
	\begin{equation}
	\label{eq:def:nesterov_shifted_two_1}
	\Theta_0 = \xi, \qquad \mathbf{m}_0 = 0,
	\end{equation}
	\begin{equation}
	\label{eq:def:nesterov_shifted_two_2}
	\mathbf{m}_n = \alpha_n \mathbf{m}_{n-1} + \grad(\Theta_{n-1}),
	\qand
	\end{equation}
	\begin{equation}
	\label{eq:def:nesterov_shifted_two_3}
	\Theta_n = \Theta_{n-1} - \gamma_{n+1} \alpha_{n+1} \mathbf{m}_n - \gamma_n \grad(\Theta_{n-1}). 
	\end{equation}
	}
}

\newcommand{\iterationStepDetermNesterovAltTwo}{
	\State 
	$
		\mathbf{m} \assign \alpha_n \mathbf{m} + \grad(\Theta)
	$
	\State
	$
		\Theta \assign \Theta - \gamma_{n+1} \alpha_{n+1} \mathbf{m} - \gamma_n \grad(\Theta)
	$
}
\newcommand{\algDescrDetermNesterovAltTwo}{\algorithmicDescription[false]{Shifted Nesterov accelerated \GD\ optimization method (\second version)}{def:determ_nesterov_shifted_two}{ 
	$(\gamma_n)_{n \in \N} \subseteq [0,\infty)$,
	$(\alpha_n)_{n \in \N} \subseteq [0,\infty)$,
}{
	shifted Nesterov accelerated \GD\ process (\second version) for the objective function $\defaultLossFunction$ 
	with learning rates $(\gamma_n)_{n \in \N}$, 
	momentum decay factors $(\alpha_n)_{n \in \N}$, 
	and initial value $\xi$
}{
	$\Theta \assign \xi$\separator $\mathbf{m} \assign 0 \in \R^\defaultParamDim$
}{
	\iterationStepDetermNesterovAltTwo
}
}

\newcommand{\iterationStepNesterovAltTwo}{
	\pseudoCodeGradientAssign
	\State 
	$
		\mathbf{m} \assign \alpha_n \mathbf{m} + \pseudoCodeGradient
	$
	\State
	$
		\Theta \assign \Theta - \gamma_{n+1} \alpha_{n+1} \mathbf{m} - \gamma_n \pseudoCodeGradient
	$
}
\newcommand{\algDescrNesterovAltTwo}{\algorithmicDescriptionStochastic[false]{Shifted Nesterov accelerated \SGD\ optimization method (\second version)}{def:nesterov_shifted_two}{ 
	$(\gamma_n)_{n \in \N} \subseteq [0,\infty)$,
	$(\alpha_n)_{n \in \N} \subseteq [0,\infty)$,
}{
	shifted Nesterov accelerated \SGD\ process (\second version) for the loss function $\defaultStochLoss$ 
	with learning rates $(\gamma_n)_{n \in \N}$, 
	momentum decay factors $(\alpha_n)_{n \in \N}$, 
	initial value $\xi$,
}{
	$\Theta \assign \xi$\separator $\mathbf{m} \assign 0 \in \R^\defaultParamDim$
}{
	\iterationStepNesterovAltTwo
}
}

\newcommand{\defNesterovAltTwo}{
\SGDdef[false]
	{def:nesterov_shifted_two}
	{Shifted Nesterov accelerated \SGD\ optimization method (\second version)}
	{ 
	$(\gamma_n)_{n \in \N} \subseteq [0,\infty)$, 
	$(\alpha_n)_{n \in \N} \subseteq [0,\infty)$, }
	{shifted Nesterov accelerated \SGD\ process (\second version) for the loss function $\defaultStochLoss$ with learning rates $(\gamma_n)_{n \in \N}$, momentum decay factors $(\alpha_n)_{n \in \N}$, initial value $\xi$,}
	{there exists $\mathbf{m} \colon \N_0 \to \R^\defaultParamDim$ such that for all $n \in \N$ it holds that
	\begin{equation}
	\Theta_0 = \xi, \qquad \mathbf{m}_0 = 0,
	\end{equation}
	\begin{equation}
	\mathbf{m}_n = \alpha_n \mathbf{m}_{n-1} + 
	\frac{1}{J_n} \sum_{j=1}^{J_n} \defaultStochGradient(\Theta_{n-1}, X_{n,j}),
	\qand
	\end{equation}
	\begin{equation}
	\Theta_n = \Theta_{n-1} - \gamma_{n+1} \alpha_{n+1} \mathbf{m}_n -
	\gamma_n \br*{
		\frac{ 1 }{J_n} \sum_{j=1}^{J_n} \defaultStochGradient(\Theta_{n-1}, X_{n,j})
	}.
	\end{equation}
	}
}

\newcommand{\defdetermNesterovAltThree}{
\deterministicGDdef[false]
	{def:determ_nesterov_shifted_three}
	{Shifted Nesterov accelerated \GD\ optimization method (\third version)}
	{ 
	$(\gamma_n)_{n \in \N} \subseteq [0,\infty)$, 
	$(\alpha_n)_{n \in \N} \subseteq [0,\infty)$, }
	{shifted Nesterov accelerated \GD\ process (\third version) for the objective function $\defaultLossFunction$ with learning rates $(\gamma_n)_{n \in \N}$, 
	momentum decay factors $(\alpha_n)_{n \in \N}$, 
	and initial value $\xi$}
	{there exists $\mathbf{m} \colon \N_0 \to \R^\defaultParamDim$ such that for all $n \in \N$ it holds that
	\begin{equation}
	\label{eq:def:nesterov_shifted_three_1}
	\Theta_0 = \xi, \qquad \mathbf{m}_0 = 0,
	\end{equation}
	\begin{equation}
	\label{eq:def:nesterov_shifted_three_2}
	\mathbf{m}_n = \alpha_n \mathbf{m}_{n-1} + (1 - \alpha_n) \gamma_n \grad(\Theta_{n-1}),
	\end{equation}
	\begin{equation}
	\label{eq:def:nesterov_shifted_three_3}
	\andq \Theta_n = \Theta_{n-1} - \alpha_{n+1} \mathbf{m}_n - (1 - \alpha_n) \gamma_n \grad(\Theta_{n-1}).
	\end{equation}
	}
}

\newcommand{\iterationStepDetermNesterovAltThree}{
	\State 
	$
		\mathbf{m} \assign \alpha_n \mathbf{m} + (1 - \alpha_n) \gamma_n \grad(\Theta)
	$
	\State
	$
		\Theta \assign \Theta - \alpha_{n+1} \mathbf{m} - (1 - \alpha_n) \gamma_n \grad(\Theta)
	$
}
\newcommand{\algDescrDetermNesterovAltThree}{\algorithmicDescription[false]{Shifted Nesterov accelerated \GD\ optimization method (\third version)}{def:determ_nesterov_shifted_three}{ 
	$(\gamma_n)_{n \in \N} \subseteq [0,\infty)$,
	$(\alpha_n)_{n \in \N} \subseteq [0,\infty)$,
}{
	shifted Nesterov accelerated \GD\ process (\third version) for the objective function $\defaultLossFunction$ 
	with learning rates $(\gamma_n)_{n \in \N}$, 
	momentum decay factors $(\alpha_n)_{n \in \N}$, 
	and initial value $\xi$
}{
	$\Theta \assign \xi$\separator $\mathbf{m} \assign 0 \in \R^\defaultParamDim$
}{
	\iterationStepDetermNesterovAltThree
}
}

\newcommand{\iterationStepNesterovAltThree}{
	\pseudoCodeGradientAssign
	\State 
	$
		\mathbf{m} \assign \alpha_n \mathbf{m} + (1 - \alpha_n) \gamma_n \pseudoCodeGradient
	$
	\State
	$
		\Theta \assign \Theta - \alpha_{n+1} \mathbf{m} - (1 - \alpha_n) \gamma_n \pseudoCodeGradient
	$
}
\newcommand{\algDescrNesterovAltThree}{\algorithmicDescriptionStochastic[false]{Shifted Nesterov accelerated \SGD\ optimization method (\third version)}{def:nesterov_shifted_three}{ 
	$(\gamma_n)_{n \in \N} \subseteq [0,\infty)$,
	$(\alpha_n)_{n \in \N} \subseteq [0,\infty)$,
}{
	shifted Nesterov accelerated \SGD\ process (\third version) for the loss function $\defaultStochLoss$ 
	with learning rates $(\gamma_n)_{n \in \N}$, 
	momentum decay factors $(\alpha_n)_{n \in \N}$, 
	initial value $\xi$,
}{
	$\Theta \assign \xi$\separator $\mathbf{m} \assign 0 \in \R^\defaultParamDim$
}{
	\iterationStepNesterovAltThree
}
}

\newcommand{\defNesterovAltThree}{
\SGDdef[false]
	{def:nesterov_shifted_three}
	{Shifted Nesterov accelerated \SGD\ optimization method (\third version)}
	{ 
	$(\gamma_n)_{n \in \N} \subseteq [0,\infty)$, 
	$(\alpha_n)_{n \in \N} \subseteq [0,\infty)$, }
	{shifted Nesterov accelerated \SGD\ process (\third version) for the loss function $\defaultStochLoss$ with learning rates $(\gamma_n)_{n \in \N}$, momentum decay factors $(\alpha_n)_{n \in \N}$, initial value $\xi$,}
	{there exists $\mathbf{m} \colon \N_0 \to \R^\defaultParamDim$ such that for all $n \in \N$ it holds that
	\begin{equation}
	\Theta_0 = \xi, \qquad \mathbf{m}_0 = 0,
	\end{equation}
	\begin{equation}
	\mathbf{m}_n = \alpha_n \mathbf{m}_{n-1} + (1 - \alpha_n) \gamma_n 
	\br*{
		\frac{1}{J_n} \sum_{j=1}^{J_n} \defaultStochGradient(\Theta_{n-1}, X_{n,j})
	},
	\end{equation}
	\begin{equation}
	\andq \Theta_n = \Theta_{n-1} - \alpha_{n+1} \mathbf{m}_n - (1 - \alpha_n) \gamma_n 
	\br*{
		\frac{ 1 }{J_n} \sum_{j=1}^{J_n} \defaultStochGradient(\Theta_{n-1}, X_{n,j})
	}.
	\end{equation}
	}
}

\newcommand{\defdetermNesterovAltFour}{
\deterministicGDdef[false]
	{def:determ_nesterov_shifted_four}
	{Shifted Nesterov accelerated \GD\ optimization method (\fourth version)}
	{ 
	$(\gamma_n)_{n \in \N} \subseteq [0,\infty)$, 
	$(\alpha_n)_{n \in \N} \subseteq [0,\infty)$, }
	{shifted Nesterov accelerated \GD\ process (\third version) for the objective function $\defaultLossFunction$ with learning rates $(\gamma_n)_{n \in \N}$, 
	momentum decay factors $(\alpha_n)_{n \in \N}$, 
	and initial value $\xi$}
	{there exists $\mathbf{m} \colon \N_0 \to \R^\defaultParamDim$ such that for all $n \in \N$ it holds that
	\begin{equation}
	\label{eq:def:nesterov_shifted_four_1}
	\Theta_0 = \xi, \qquad \mathbf{m}_0 = 0,
	\end{equation}
	\begin{equation}
	\label{eq:def:nesterov_shifted_four_2}
	\mathbf{m}_n = \alpha_n \mathbf{m}_{n-1} + \gamma_n \grad(\Theta_{n-1}),
	\end{equation}
	\begin{equation}
	\label{eq:def:nesterov_shifted_four_3}
	\andq \Theta_n = \Theta_{n-1} - \alpha_{n+1} \mathbf{m}_n - \gamma_n \grad(\Theta_{n-1}).
	\end{equation}
	}
}

\newcommand{\iterationStepDetermNesterovAltFour}{
	\State 
	$
		\mathbf{m} \assign \alpha_n \mathbf{m} + \gamma_n \grad(\Theta)
	$
	\State
	$
		\Theta \assign \Theta - \alpha_{n+1} \mathbf{m} - \gamma_n \grad(\Theta)
	$
}
\newcommand{\algDescrDetermNesterovAltFour}{\algorithmicDescription[false]{Shifted Nesterov accelerated \GD\ optimization method (\fourth version)}{def:determ_nesterov_shifted_four}{ 
	$(\gamma_n)_{n \in \N} \subseteq [0,\infty)$,
	$(\alpha_n)_{n \in \N} \subseteq [0,\infty)$,
}{
	shifted Nesterov accelerated \GD\ process (\fourth version) for the objective function $\defaultLossFunction$ 
	with learning rates $(\gamma_n)_{n \in \N}$, 
	momentum decay factors $(\alpha_n)_{n \in \N}$, 
	and initial value $\xi$
}{
	$\Theta \assign \xi$\separator $\mathbf{m} \assign 0 \in \R^\defaultParamDim$
}{
	\iterationStepDetermNesterovAltFour
}
}

\newcommand{\iterationStepNesterovAltFour}{
	\pseudoCodeGradientAssign
	\State 
	$
		\mathbf{m} \assign \alpha_n \mathbf{m} + \gamma_n \pseudoCodeGradient
	$
	\State
	$
		\Theta \assign \Theta - \alpha_{n+1} \mathbf{m} - \gamma_n \pseudoCodeGradient
	$
}
\newcommand{\algDescrNesterovAltFour}{\algorithmicDescriptionStochastic[false]{Shifted Nesterov accelerated \SGD\ optimization method (\fourth version)}{def:nesterov_shifted_four}{ 
	$(\gamma_n)_{n \in \N} \subseteq [0,\infty)$,
	$(\alpha_n)_{n \in \N} \subseteq [0,\infty)$,
}{
	shifted Nesterov accelerated \SGD\ process (\fourth version) for the loss function $\defaultStochLoss$ 
	with learning rates $(\gamma_n)_{n \in \N}$, 
	momentum decay factors $(\alpha_n)_{n \in \N}$, 
	initial value $\xi$,
}{
	$\Theta \assign \xi$\separator $\mathbf{m} \assign 0 \in \R^\defaultParamDim$
}{
	\iterationStepNesterovAltFour
}
}

\newcommand{\defNesterovAltFour}{
\SGDdef[false]
	{def:nesterov_shifted_four}
	{Shifted Nesterov accelerated \SGD\ optimization method (\fourth version)}
	{ 
	$(\gamma_n)_{n \in \N} \subseteq [0,\infty)$, 
	$(\alpha_n)_{n \in \N} \subseteq [0,\infty)$, }
	{shifted Nesterov accelerated \SGD\ process (\fourth version) for the loss function $\defaultStochLoss$ with learning rates $(\gamma_n)_{n \in \N}$, momentum decay factors $(\alpha_n)_{n \in \N}$, initial value $\xi$,}
	{there exists $\mathbf{m} \colon \N_0 \to \R^\defaultParamDim$ such that for all $n \in \N$ it holds that
	\begin{equation}
	\Theta_0 = \xi, \qquad \mathbf{m}_0 = 0,
	\end{equation}
	\begin{equation}
	\mathbf{m}_n = \alpha_n \mathbf{m}_{n-1} + 
	\gamma_n 
	\br*{
		\frac{1}{J_n} \sum_{j=1}^{J_n} \defaultStochGradient(\Theta_{n-1}, X_{n,j})
	},
	\end{equation}
	\begin{equation}
	\andq \Theta_n = \Theta_{n-1} - \alpha_{n+1} \mathbf{m}_n -
	\gamma_n \br*{
		\frac{ 1 }{J_n} \sum_{j=1}^{J_n} \defaultStochGradient(\Theta_{n-1}, X_{n,j})
	}.
	\end{equation}
	}
}

\newcommand{\defdetermNesterovBias}{
\deterministicGDdef[false]
	{def:determ_nesterov_bias}
	{Bias-adjusted Nesterov accelerated \GD\ optimization method}
	{ 
	$(\gamma_n)_{n \in \N} \subseteq [0,\infty)$, 
	$(\alpha_n)_{n \in \N} \subseteq [0,1)$, }
	{bias-adjusted Nesterov accelerated \GD\ process for the objective function $\defaultLossFunction$ with learning rates $(\gamma_n)_{n \in \N}$, 
	momentum decay factors $(\alpha_n)_{n \in \N}$, 
	and initial value $\xi$}
	{there exists $\mathbf{m} \colon \N_0 \to \R^\defaultParamDim$ such that for all $n \in \N$ it holds that
	\begin{equation}
	\label{eq:def:nesterov_bias_1}
	\Theta_0 = \xi, \qquad \mathbf{m}_0 = 0,
	\end{equation}
	\begin{equation}
	\label{eq:def:nesterov_bias_2}
	\mathbf{m}_n = \alpha_n \mathbf{m}_{n-1} + (1-\alpha_n)  \grad\pr*{\Theta_{n-1} - \frac{\gamma_n\alpha_n \mathbf{m}_{n-1}}{1- \prod_{l=1}^n\alpha_l}}, 
	\end{equation}
	\begin{equation}
	\label{eq:def:nesterov_bias_3}
	\andq \Theta_n = \Theta_{n-1} - \frac{\gamma_{n} \mathbf{m}_n}{1- \prod_{l=1}^{n}\alpha_l}.
	\end{equation}
	}
}

\newcommand{\iterationStepDetermNesterovBias}{
	\State 
	$
		\mathbf{m} 
	\assign 
		\alpha_n \mathbf{m} 
		+ 
		(1-\alpha_n) \grad\pr[\big]{\Theta - \frac{\gamma_n\alpha_n \mathbf{m}}{1- \prod_{l=1}^n\alpha_l}}
	$
	\State
	$
		\Theta \assign \Theta - \frac{\gamma_{n} \mathbf{m}}{1- \prod_{l=1}^{n}\alpha_l}
	$
}
\newcommand{\algDescrDetermNesterovBias}{\algorithmicDescription[false]{Bias-adjusted Nesterov accelerated \GD\ optimization method}{def:determ_nesterov_bias}{ 
	$(\gamma_n)_{n \in \N} \subseteq [0,\infty)$,
	$(\alpha_n)_{n \in \N} \subseteq [0,1)$,
}{
	bias-adjusted Nesterov accelerated \GD\ process for the objective function $\defaultLossFunction$ 
	with learning rates $(\gamma_n)_{n \in \N}$, 
	momentum decay factors $(\alpha_n)_{n \in \N}$, 
	and initial value $\xi$
}{
	$\Theta \assign \xi$\separator $\mathbf{m} \assign 0 \in \R^\defaultParamDim$
}{
	\iterationStepDetermNesterovBias
}
}

\newcommand{\iterationStepNesterovBias}{
	\pseudoCodeGradientAssign[\Theta - \frac{\gamma_n\alpha_n \mathbf{m}}{1- \prod_{l=1}^n\alpha_l}]
	\State 
	$
		\mathbf{m} 
	\assign 
		\alpha_n \mathbf{m} 
		+ 
		(1-\alpha_n) \pseudoCodeGradient
	$
	\State
	$
		\Theta \assign \Theta - \frac{\gamma_{n} \mathbf{m}}{1- \prod_{l=1}^{n}\alpha_l}
	$
}
\newcommand{\algDescrNesterovBias}{\algorithmicDescriptionStochastic[false]{Bias-adjusted Nesterov accelerated \SGD\ optimization method}{def:nesterov_bias}{ 
	$(\gamma_n)_{n \in \N} \subseteq [0,\infty)$,
	$(\alpha_n)_{n \in \N} \subseteq [0,1)$,
}{
	bias-adjusted Nesterov accelerated \SGD\ process for the loss function $\defaultStochLoss$ 
	with learning rates $(\gamma_n)_{n \in \N}$, 
	momentum decay factors $(\alpha_n)_{n \in \N}$, 
	initial value $\xi$,
}{
	$\Theta \assign \xi$\separator $\mathbf{m} \assign 0 \in \R^\defaultParamDim$
}{
	\iterationStepNesterovBias
}
}

\newcommand{\defNesterovBias}{
\SGDdef[false]
	{def:nesterov_bias}
	{Bias-adjusted Nesterov accelerated \SGD\ optimization method}
	{ 
	$(\gamma_n)_{n \in \N} \subseteq [0,\infty)$, 
	$(\alpha_n)_{n \in \N} \subseteq [0,1)$, }
	{bias-adjusted Nesterov accelerated \SGD\ process for the loss function $\defaultStochLoss$ with learning rates $(\gamma_n)_{n \in \N}$, momentum decay factors $(\alpha_n)_{n \in \N}$, initial value $\xi$,}
	{there exists $\mathbf{m} \colon \N_0 \to \R^\defaultParamDim$ such that for all $n \in \N$ it holds that
	\begin{equation}
	\label{eq:nesterov_bias_1}
	\Theta_0 = \xi, \qquad \mathbf{m}_0 = 0,
	\end{equation}
	\begin{equation}
	\label{eq:nesterov_bias_2}
	\mathbf{m}_n = \alpha_n \mathbf{m}_{n-1} + (1-\alpha_n) 
	\br*{
		\frac{1}{J_n} \sum_{j=1}^{J_n} \defaultStochGradient\pr*{\Theta_{n-1} - \frac{\gamma_n\alpha_n \mathbf{m}_{n-1}}{1- \prod_{l=1}^n\alpha_l}, X_{n,j}},
	}
	\end{equation}
	\begin{equation}
	\label{eq:nesterov_bias_3}
	\andq \Theta_n = \Theta_{n-1} - \frac{\gamma_{n} \mathbf{m}_n}{1- \prod_{l=1}^{n}\alpha_l}.
	\end{equation}
	}
}

\newcommand{\defdetermNesterovAltBias}{
\deterministicGDdef[false]
	{def:determ_nesterov_shifted_bias}
	{Shifted bias-adjusted Nesterov accelerated \GD\ optimization method}
	{ 
	$(\gamma_n)_{n \in \N} \subseteq [0,\infty)$, 
	$(\alpha_n)_{n \in \N} \subseteq [0,1)$, }
	{shifted bias-adjusted Nesterov accelerated \GD\ process for the objective function $\defaultLossFunction$ with learning rates $(\gamma_n)_{n \in \N}$, 
	momentum decay factors $(\alpha_n)_{n \in \N}$, 
	and initial value $\xi$}
	{there exists $\mathbf{m} \colon \N_0 \to \R^\defaultParamDim$ such that for all $n \in \N$ it holds that
	\begin{equation}
	\Theta_0 = \xi, \qquad \mathbf{m}_0 = 0,
	\end{equation}
	\begin{equation}
	\mathbf{m}_n = \alpha_n \mathbf{m}_{n-1} + (1-\alpha_n)  \grad\pr*{\Theta_{n-1}}, 
	\qand
	\end{equation}
	\begin{equation}
	\Theta_n = \Theta_{n-1} - \frac{\gamma_n(1-\alpha_n) \grad(\Theta_{n-1})}{1- \prod_{l=1}^n\alpha_l} - \frac{\gamma_{n+1}\alpha_{n+1} \mathbf{m}_n}{1- \prod_{l=1}^{n+1}\alpha_l}. 
	\end{equation}
	}
}

\newcommand{\iterationStepDetermNesterovAltBias}{
	\State 
	$
		\mathbf{m} 
	\assign 
		\alpha_n \mathbf{m} 
		+ 
		(1-\alpha_n) \grad\pr*{\Theta}
	$
	\State
	$
		\Theta 
	\assign 
		\Theta 
		-  
		\frac{\gamma_{n}(1-\alpha_{n}) \grad(\Theta)}{1- \prod_{l=1}^{n}\alpha_l}
		-
		\frac{\gamma_{n+1}\alpha_{n+1} \mathbf{m}}{1- \prod_{l=1}^{n+1}\alpha_l}
	$
}
\newcommand{\algDescrDetermNesterovAltBias}{\algorithmicDescription[false]{Shifted bias-adjusted Nesterov accelerated \GD\ optimization method}{def:determ_nesterov_shifted_bias}{ 
	$(\gamma_n)_{n \in \N} \subseteq [0,\infty)$,
	$(\alpha_n)_{n \in \N} \subseteq [0,1)$, 
}{
	Nesterov accelerated \GD\ process for the objective function $\defaultLossFunction$ 
	with learning rates $(\gamma_n)_{n \in \N}$, 
	momentum decay factors $(\alpha_n)_{n \in \N}$, 
	and initial value $\xi$
}{
	$\Theta \assign \xi$\separator $\mathbf{m} \assign 0 \in \R^\defaultParamDim$
}{
	\iterationStepDetermNesterovAltBias
}
}

\newcommand{\iterationStepNesterovAltBias}{
	\pseudoCodeGradientAssign
	\State 
	$
		\mathbf{m} 
	\assign 
		\alpha_n \mathbf{m} 
		+ 
		(1-\alpha_n) \pseudoCodeGradient
	$
	\State
	$
		\Theta 
	\assign 
		\Theta 
		-  
		\frac{\gamma_{n}(1-\alpha_{n}) \pseudoCodeGradient}{1- \prod_{l=1}^{n}\alpha_l}
		-
		\frac{\gamma_{n+1}\alpha_{n+1} \mathbf{m}}{1- \prod_{l=1}^{n+1}\alpha_l}
	$
}
\newcommand{\algDescrNesterovAltBias}{\algorithmicDescriptionStochastic[false]{Shifted bias-adjusted Nesterov accelerated \SGD\ optimization method}{def:nesterov_shifted_bias}{ 
	$(\gamma_n)_{n \in \N} \subseteq [0,\infty)$,
	$(\alpha_n)_{n \in \N} \subseteq [0,1)$, 
}{
	Nesterov accelerated \SGD\ process for the loss function $\defaultStochLoss$ 
	with learning rates $(\gamma_n)_{n \in \N}$, 
	momentum decay factors $(\alpha_n)_{n \in \N}$, 
	initial value $\xi$,
}{
	$\Theta \assign \xi$\separator $\mathbf{m} \assign 0 \in \R^\defaultParamDim$
}{
	\iterationStepNesterovAltBias
}
}

\newcommand{\defNesterovAltBias}{
\SGDdef[false]
	{def:nesterov_shifted_bias}
	{Shifted bias-adjusted Nesterov accelerated \SGD\ optimization method}
	{ 
	$(\gamma_n)_{n \in \N} \subseteq [0,\infty)$, 
	$(\alpha_n)_{n \in \N} \subseteq [0,1)$, }
	{shifted bias-adjusted Nesterov accelerated \SGD\ process for the loss function $\defaultStochLoss$ with learning rates $(\gamma_n)_{n \in \N}$, momentum decay factors $(\alpha_n)_{n \in \N}$, initial value $\xi$,}
	{there exists $\mathbf{m} \colon \N_0 \to \R^\defaultParamDim$ such that for all $n \in \N$ it holds that
	\begin{equation}
	\label{eq:nesterov_shifted_bias_1}
	\Theta_0 = \xi, \qquad \mathbf{m}_0 = 0,
	\end{equation}
	\begin{equation}
	\label{eq:nesterov_shifted_bias_2}
	\mathbf{m}_n = \alpha_n \mathbf{m}_{n-1} + (1-\alpha_n) 
	\br*{
		\frac{1}{J_n} \sum_{j=1}^{J_n} \defaultStochGradient\pr*{\Theta_{n-1}, X_{n,j}}
	},
	\qand
	\end{equation}
	\begin{equation}
	\label{eq:nesterov_shifted_bias_3}
		\Theta_n 
	= 
		\Theta_{n-1} - \frac{\gamma_n(1-\alpha_n)}{1- \prod_{l=1}^n\alpha_l} 
		\br*{
			\frac{1}{J_n} \sum_{j=1}^{J_n} \defaultStochGradient\pr*{\Theta_{n-1}, X_{n,j}}
		}
		- \frac{\gamma_{n+1}\alpha_{n+1} \mathbf{m}_n}{1- \prod_{l=1}^{n+1}\alpha_l}.
	\end{equation}
	}
}

\newcommand{\defdetermNesterovSimple}{
\deterministicGDdef[false]
	{def:determ_nesterov_simple}
	{Simplified Nesterov accelerated \GD\ optimization method}
	{ 
	$(\gamma_n)_{n \in \N} \subseteq [0,\infty)$, 
	$(\alpha_n)_{n \in \N} \subseteq [0,\infty)$, }
	{simplified Nesterov accelerated \GD\ process for the objective function $\defaultLossFunction$ with learning rates $(\gamma_n)_{n \in \N}$, 
	momentum decay factors $(\alpha_n)_{n \in \N}$, 
	and initial value $\xi$}
	{there exists $\mathbf{m} \colon \N_0 \to \R^\defaultParamDim$ such that for all $n \in \N$ it holds that
	\begin{equation}
	\label{eq:def:nesterov_simple_1}
	\Theta_0 = \xi, \qquad \mathbf{m}_0 = 0,
	\end{equation}
	\begin{equation}
	\label{eq:def:nesterov_simple_2}
	\mathbf{m}_n = \alpha_n \mathbf{m}_{n-1} + \grad(\Theta_{n-1}),
	\qand
	\end{equation}
	\begin{equation}
	\label{eq:def:nesterov_simple_3}
	\Theta_n = \Theta_{n-1} - \gamma_{n} \alpha_{n} \mathbf{m}_n - \gamma_n \grad(\Theta_{n-1}). 
	\end{equation}
	}
}

\newcommand{\iterationStepDetermNesterovSimple}{
	\State 
	$
		\mathbf{m} \assign \alpha_n \mathbf{m} + \grad(\Theta)
	$
	\State
	$
		\Theta \assign \Theta - \gamma_{n} \alpha_{n} \mathbf{m} - \gamma_n \grad(\Theta)
	$
}
\newcommand{\algDescrDetermNesterovSimple}{\algorithmicDescription[false]{Simplified Nesterov accelerated \GD\ optimization method}{def:determ_nesterov_simple}{ 
	$(\gamma_n)_{n \in \N} \subseteq [0,\infty)$,
	$(\alpha_n)_{n \in \N} \subseteq [0,\infty)$,
}{
	simplified Nesterov accelerated \GD\ process for the objective function $\defaultLossFunction$ 
	with learning rates $(\gamma_n)_{n \in \N}$, 
	momentum decay factors $(\alpha_n)_{n \in \N}$, 
	and initial value $\xi$
}{
	$\Theta \assign \xi$\separator $\mathbf{m} \assign 0 \in \R^\defaultParamDim$
}{
	\iterationStepDetermNesterovSimple
}
}

\newcommand{\iterationStepNesterovSimple}{
	\pseudoCodeGradientAssign
	\State 
	$
		\mathbf{m} \assign \alpha_n \mathbf{m} + \pseudoCodeGradient
	$
	\State
	$
		\Theta \assign \Theta - \gamma_{n} \alpha_{n} \mathbf{m} - \gamma_n \pseudoCodeGradient
	$
}
\newcommand{\algDescrNesterovSimple}{\algorithmicDescriptionStochastic[false]{Simplified Nesterov accelerated \SGD\ optimization method}{def:nesterov_simple}{ 
	$(\gamma_n)_{n \in \N} \subseteq [0,\infty)$,
	$(\alpha_n)_{n \in \N} \subseteq [0,\infty)$,
}{
	simplified Nesterov accelerated \SGD\ process for the loss function $\defaultStochLoss$ 
	with learning rates $(\gamma_n)_{n \in \N}$, 
	momentum decay factors $(\alpha_n)_{n \in \N}$, 
	initial value $\xi$,
}{
	$\Theta \assign \xi$\separator $\mathbf{m} \assign 0 \in \R^\defaultParamDim$
}{
	\iterationStepNesterovSimple
}
}

\newcommand{\defNesterovSimple}{
\SGDdef[false]
	{def:nesterov_simple}
	{Simplified Nesterov accelerated \SGD\ optimization method}
	{ 
	$(\gamma_n)_{n \in \N} \subseteq [0,\infty)$, 
	$(\alpha_n)_{n \in \N} \subseteq [0,\infty)$, }
	{simplified Nesterov accelerated \SGD\ process for the loss function $\defaultStochLoss$ with learning rates $(\gamma_n)_{n \in \N}$, momentum decay factors $(\alpha_n)_{n \in \N}$, initial value $\xi$,}
	{there exists $\mathbf{m} \colon \N_0 \to \R^\defaultParamDim$ such that for all $n \in \N$ it holds that
	\begin{equation}
	\Theta_0 = \xi, \qquad \mathbf{m}_0 = 0,
	\end{equation}
	\begin{equation}
	\mathbf{m}_n = \alpha_n \mathbf{m}_{n-1} + 
	\frac{1}{J_n} \sum_{j=1}^{J_n} \defaultStochGradient(\Theta_{n-1}, X_{n,j}),
	\qand
	\end{equation}
	\begin{equation}
	\Theta_n = \Theta_{n-1} - \gamma_{n} \alpha_{n} \mathbf{m}_n -
	\gamma_n \br*{
		\frac{ 1 }{J_n} \sum_{j=1}^{J_n} \defaultStochGradient(\Theta_{n-1}, X_{n,j})
	}.
	\end{equation}
	}
}

\newcommand{\defdetermAdagrad}{
\deterministicGDdef[true]
	{def:determ_adagrad}
	{\Adagrad\ \GD\ optimization method}
	{ 
	$(\gamma_n)_{n \in \N} \subseteq [0,\infty)$, 
	$\eps\in(0,\infty)$, }
	{\Adagrad\ \GD\ process for the objective function $\defaultLossFunction$ with learning rates $(\gamma_n)_{n \in \N}$, 
	regularizing factor $\varepsilon$, 
	and initial value $\xi$}
	{it holds for all $n \in \N$, $i \in \{1,2,\ldots,\defaultParamDim\}$ that 
	\begin{equation}
	\label{def:determ_adagrad:eq1}
	\Theta_0 = \xi \qandq
	\Theta_n^{(i)}
	= 
	\Theta_{n-1}^{(i)} - \gamma_n  
	\br*{ 
	  \varepsilon 
	  + 
		\br*{ 
		\textstyle
		\sum\limits_{ k = 0 }^{ n - 1 }  
		\abs*{ \defaultGradientFunction_i( \Theta_k ) }^2
		}^{\nicefrac{ 1 }{ 2 } }
	}^{-1}  
	\defaultGradientFunction_i(\Theta_{n-1}).
	\end{equation}
	}
}

\newcommand{\iterationStepDetermAdagrad}{
	\State 
		$
			\mathbb{M}
		\assign
			\mathbb{M}
			+
			\br[\big]{ \grad( \Theta ) }^2
		$
	\State
		$
			\Theta
		\assign 
			\Theta - \gamma_n  
			\br[big]{ 
				 \varepsilon 
				 + 
				 \textstyle
				 \mathbb{M}^{\nicefrac{ 1 }{ 2 }}
			}^{-1} 
			\compMulti
			\grad(\Theta)
		$
}

\newcommand{\algDescrDetermAdagrad}{\algorithmicDescription{\Adagrad\ \GD\ optimization method}{def:determ_adagrad}{ 
	$(\gamma_n)_{n \in \N} \subseteq [0,\infty)$,
	$\eps\in(0,\infty)$,
}{
	\Adagrad\ \GD\ process for the objective function $\defaultLossFunction$ with learning rates $(\gamma_n)_{n \in \N}$, regularizing factor $\varepsilon$, and initial value $\xi$
}{
	$\Theta \assign \xi$\separator $\mathbb{M} \assign 0 \in \R^\defaultParamDim$
}{
	\iterationStepDetermAdagrad
}
}

\newcommand{\iterationStepAdagrad}{
	\pseudoCodeGradientAssign
	\State 
		$
			\mathbb{M} 
		\assign
			\mathbb{M}
			+
			\pseudoCodeGradient^2
		$
	\State
		$
			\Theta
		\assign 
			\Theta - \gamma_n  
			\br[big]{ 
				 \varepsilon 
				 + 
				 \textstyle
				 \mathbb{M}^{\nicefrac{ 1 }{ 2 }}
			}^{-1} 
			\compMulti
			\pseudoCodeGradient
		$
}

\newcommand{\algDescrAdagrad}{\algorithmicDescriptionStochastic{\Adagrad\ \SGD\ optimization method}{def:adagrad}{ 
	$(\gamma_n)_{n \in \N} \subseteq [0,\infty)$,
	$\eps\in(0,\infty)$,
}{
	\Adagrad\ \SGD\ process for the loss function $\defaultStochLoss$ with learning rates $(\gamma_n)_{n \in \N}$, regularizing factor $\varepsilon$, initial value $\xi$,
}{
	$\Theta \assign \xi$\separator $\mathbb{M} \assign 0 \in \R^\defaultParamDim$
}{
	\iterationStepAdagrad
}
}

\newcommand{\defAdagrad}{
\SGDdef[false]
	{def:adagrad}
	{\Adagrad\ \SGD\ optimization method}
	{ 
	$(\gamma_n)_{n \in \N} \subseteq [0,\infty)$, 
	$\varepsilon \in (0,\infty)$,}
	{\Adagrad\ \SGD\ process for the loss function $\defaultStochLoss$ with learning rates $(\gamma_n)_{n \in \N}$, regularizing factor $\varepsilon$, initial value $\xi$,}
	{it holds for all $n \in \N$, $i \in \{1,2,\ldots,\defaultParamDim\}$ that 	
	\begin{equation}
	\label{T_B_D}
	\begin{split}
		\Theta_n^{(i)}= \Theta_{n-1}^{(i)} 
		-\gamma_n  
		\br*{
			\varepsilon + \br*{\textstyle\sum\limits_{k = 1}^n \bbpr{ \tfrac{1}{J_k} {\textstyle \sum_{j=1}^{J_k}}\defaultStochGradient_{i}(\Theta_{k-1}, X_{k,j})}^2}^{\nicefrac12}
		}^{-1}  \br*{\frac{1}{J_n}\sum_{j = 1}^{J_n}\defaultStochGradient_{i}(\Theta_{n-1}, X_{n,j})}.
	\end{split}
	\end{equation}
	}
}

\newcommand{\defdetermRMSprop}{
\deterministicGDdef[true]
	{def:determ_RMSprop}
	{\RMSprop\ \GD\ optimization method}
	{ 
	$(\gamma_n)_{n \in \N} \subseteq [0,\infty)$, 
	$(\beta_n)_{n \in \N}\subseteq [0,1]$, 
	$\varepsilon \in (0,\infty)$, }
	{\RMSprop\ \GD\ process for the objective function $\defaultLossFunction$ with learning rates $(\gamma_n)_{n \in \N}$, 
	second moment decay factors $(\beta_n)_{n \in \N}$, 
	regularizing factor $\varepsilon$, 
	and initial value $\xi$}
	{there exists $\mathbb{M} = (\mathbb{M}^{(1)},\ldots,\mathbb{M}^{(\defaultParamDim)}) \colon \N_0 \to \R^\defaultParamDim$ such that for all $n \in \N$, $i \in \{1,2,\ldots,\defaultParamDim\}$ it holds that
	\begin{equation}
	\label{def:determ_RMSprop:eq1}
	\Theta_0 = \xi, \qquad \mathbb{M}_0 = 0, 
	\qquad
	  \mathbb{M}_n^{ (i) } 
	  = \beta_n  \mathbb{M}_{ n - 1 }^{ (i) } 
	  + ( 1 - \beta_n ) 
	  \abs*{ \defaultGradientFunction_i( \Theta_{ n - 1 } ) }^2 , 
	\end{equation}
	\begin{equation}
	\label{def:determ_RMSprop:eq2}
	\andq
	  \Theta_n^{(i)}= \Theta_{n-1}^{(i)} 
	  - 
	      \gamma_n 
	    \br*{ 
	      \varepsilon + \br{\mathbb{M}_n^{ (i) }}^{ 1/2 }
	    }^{ - 1 } 
	  \defaultGradientFunction_i( \Theta_{ n - 1 } ) 
	  . 
	\end{equation}
	}
}

\newcommand{\iterationStepDetermRMSprop}{
	\State
		$
			\mathbb{M}
		\assign
			\beta_n  \mathbb{M}
			+ 
			( 1 - \beta_n ) 
			\br*{ \grad( \Theta ) }^2
		$
		\State
		$
			\Theta
		\assign
			\Theta
			- 
			\gamma_n 
			\br[\big]{ 
			\varepsilon + \mathbb{M}^{ 1/2 }
			}^{ - 1  } 
			\compMulti
			\grad( \Theta ) 
		$
}
\newcommand{\algDescrDetermRMSprop}{\algorithmicDescription{\RMSprop\ \GD\ optimization method}{def:determ_RMSprop}{ 
	$(\gamma_n)_{n \in \N} \subseteq [0,\infty)$,
	$(\beta_n)_{n \in \N}\subseteq [0,1]$,
	$\varepsilon \in (0,\infty)$,
}{
	\RMSprop\ \GD\ process for the objective function $\defaultLossFunction$ with learning rates $(\gamma_n)_{n \in \N}$, second moment decay factors $(\beta_n)_{n \in \N}$, regularizing factor $\varepsilon$, and initial value $\xi$
}{
	$\Theta \assign \xi$\separator 
	$\mathbb{M} \assign 0 \in \R^\defaultParamDim$
}{
	\iterationStepDetermRMSprop
}
}

\newcommand{\iterationStepRMSprop}{
	\pseudoCodeGradientAssign
	\State
		$
			\mathbb{M}
		\assign
			\beta_n  \mathbb{M}
			+ 
			( 1 - \beta_n ) 
			\pseudoCodeGradient^2
		$
		\State
		$
			\Theta
		\assign
			\Theta
			- 
			\gamma_n 
			\br[\big]{ 
			\varepsilon + \mathbb{M}^{ 1/2 }
			}^{ - 1 } 
			\compMulti
			\pseudoCodeGradient 
		$
}
\newcommand{\algDescrRMSprop}{\algorithmicDescriptionStochastic{\RMSprop\ \SGD\ optimization method}{def:rmsprop}{ 
	$(\gamma_n)_{n \in \N} \subseteq [0,\infty)$,
	$(\beta_n)_{n \in \N}\subseteq [0,1]$,
	$\varepsilon \in (0,\infty)$,
}{
	\RMSprop\ \SGD\ process for the loss function $\defaultStochLoss$ with learning rates $(\gamma_n)_{n \in \N}$, second moment decay factors $(\beta_n)_{n \in \N}$, regularizing factor $\varepsilon$, initial value $\xi$,
}{
	$\Theta \assign \xi$\separator 
	$\mathbb{M} \assign 0 \in \R^\defaultParamDim$
}{
	\iterationStepRMSprop
}
}

\newcommand{\defRMSprop}{
\SGDdef[true]
	{def:rmsprop}
	{\RMSprop\ \SGD\ optimization method}
	{ 
	$(\gamma_n)_{n \in \N} \subseteq [0,\infty)$, 
	$(\beta_n)_{n \in \N}\subseteq [0,1]$, 
	$\varepsilon \in (0,\infty)$,}
	{\RMSprop\ \SGD\ process for the loss function $\defaultStochLoss$ with learning rates $(\gamma_n)_{n \in \N}$, second moment decay factors $(\beta_n)_{n \in \N}$, regularizing factor $\varepsilon$, initial value $\xi$,}
	{there exist $\mathbb{M} = (\mathbb{M}^{(1)},\ldots,\mathbb{M}^{(\defaultParamDim)}) \colon \N_0 \times \Omega \to \R^\defaultParamDim$ such that for all $n \in \N$, $i \in \{1,2,\ldots,\defaultParamDim\}$ it holds that
	\begin{equation}
	\label{eq:rmsprop_1}
	\Theta_0 = \xi, \qquad \mathbb{M}_0 = 0,
	\end{equation}
	\begin{equation}
	\label{eq:rmsprop_2}
	\mathbb{M}_n^{(i)} 
	= 
	\beta_n\,\mathbb{M}_{n-1}^{(i)} + (1-\beta_n)\br*{\frac{1}{J_n} { \sum_{j = 1}^{J_n}} \defaultStochGradient_{i}(\Theta_{n-1}, X_{n,j})}^2, 
	\end{equation}
	\begin{equation}
	\label{eq:rmsprop_3}
	\andq
	\Theta_n^{(i)}
	= 
	\Theta_{n-1}^{(i)} - \gamma_n\br*{ \varepsilon + \br*{\mathbb{M}_n^{(i)}}^{1/2} }^{-1} \br*{\frac{1}{J_n}\sum_{j = 1}^{J_n}\defaultStochGradient_{i}(\Theta_{n-1}, X_{n,j})}.
	\end{equation}
	}
}

\newcommand{\defdetermRMSpropBias}{
\deterministicGDdef[true]
	{def:determ_RMSprop_bias}
	{Bias-adjusted \RMSprop\ \GD\ optimization method}
	{ 
	$(\gamma_n)_{n \in \N} \subseteq [0,\infty)$, 
	$(\beta_n)_{n \in \N} \subseteq [0,1)$, 
	$\varepsilon \in (0,\infty)$, }
	{bias-adjusted \RMSprop\ \GD\ process for the objective function $\defaultLossFunction$ with learning rates $(\gamma_n)_{n \in \N}$, 
	second moment decay factors $(\beta_n)_{n \in \N}$, 
	regularizing factor $\varepsilon$, 
	and initial value $\xi$}
	{there exists $\mathbb{M} = (\mathbb{M}^{(1)},\ldots,\mathbb{M}^{(\defaultParamDim)}) \colon \N_0 \to \R^\defaultParamDim$ such that for all $n \in \N$, $i \in \{1,2,\ldots,\defaultParamDim\}$ it holds that
	\begin{equation}
	\label{determ_RMSprop_bias:eq1}
	\Theta_0 = \xi, \qquad \mathbb{M}_0 = 0, 
	\qquad
	  \mathbb{M}_n^{ (i) } 
	  = \beta_n  \mathbb{M}_{ n - 1 }^{ (i) } 
	  + ( 1 - \beta_n ) 
	  \abs*{ \defaultGradientFunction_i( \Theta_{ n - 1 } ) }^2 , 
	\end{equation}
	\begin{equation}
	\label{determ_RMSprop_bias:eq2}
	\andq
	  \Theta_n^{(i)}= \Theta_{n-1}^{(i)} 
	  - 
	      \gamma_n 
	  {\textstyle 
	    \br*{\varepsilon + \br*{ \frac{ \mathbb{M}_n^{ (i) } }{ 1 - \prod_{k = 1}^n \beta_k } }^{ \nicefrac{1}{2} } 
	    }^{ - 1 } 
	  } 
	  \defaultGradientFunction_i( \Theta_{ n - 1 } ) 
	  .
	\end{equation}
	}
}

\newcommand{\iterationStepDetermRMSpropBias}{
	\State
		$
			\mathbb{M}
		\assign
			\beta_n  \mathbb{M}
			+ 
			( 1 - \beta_n ) 
			\br*{ \grad( \Theta ) }^2
		$
		\State
		$
			\Theta
		\assign
			\Theta
			- 
			\gamma_n 
			\br*{ 
			\varepsilon + \br*{ \frac{ \mathbb{M} }{ 1 - \prod_{k = 1}^n \beta_k } }^{ \nicefrac{1}{2} } 
			}^{ - 1 } 
			\compMulti
			\grad( \Theta ) 
		$
}
\newcommand{\algDescrDetermRMSpropBias}{\algorithmicDescription{Bias-adjusted \RMSprop\ \GD\ optimization method}{def:determ_RMSprop_bias}{ 
	$(\gamma_n)_{n \in \N} \subseteq [0,\infty)$,
	$(\beta_n)_{n \in \N}\subseteq [0,1)$,
	$\varepsilon \in (0,\infty)$,
}{
	bias-adjusted \RMSprop\ \GD\ process for the objective function $\defaultLossFunction$ with learning rates $(\gamma_n)_{n \in \N}$, second moment decay factors $(\beta_n)_{n \in \N}$, regularizing factor $\varepsilon$, and initial value $\xi$
}{
	$\Theta \assign \xi$\separator 
	$\mathbb{M} \assign 0 \in \R^\defaultParamDim$
}{
	\iterationStepDetermRMSpropBias
}
}

\newcommand{\iterationStepRMSpropBias}{
	\pseudoCodeGradientAssign
	\State
		$
			\mathbb{M}
		\assign
			\beta_n  \mathbb{M}
			+ 
			( 1 - \beta_n ) 
			\pseudoCodeGradient^2
		$
		\State
		$
			\Theta
		\assign
			\Theta
			- 
			\gamma_n 
			\br*{ 
			\varepsilon + \br*{ \frac{ \mathbb{M} }{ 1 - \prod_{k = 1}^n \beta_k } }^{ \nicefrac{1}{2} } 
			}^{ - 1 } 
			\compMulti
			\pseudoCodeGradient 
		$
}
\newcommand{\algDescrRMSpropBias}{\algorithmicDescriptionStochastic{Bias-adjusted \RMSprop\ \SGD\ optimization method}{def:RMSprop_bias}{ 
	$(\gamma_n)_{n \in \N} \subseteq [0,\infty)$,
	$(\beta_n)_{n \in \N}\subseteq [0,1)$,
	$\varepsilon \in (0,\infty)$,
}{
	bias-adjusted \RMSprop\ \SGD\ process for the loss function $\defaultStochLoss$ with learning rates $(\gamma_n)_{n \in \N}$, second moment decay factors $(\beta_n)_{n \in \N}$, regularizing factor $\varepsilon$, initial value $\xi$,
}{
	$\Theta \assign \xi$\separator 
	$\mathbb{M} \assign 0 \in \R^\defaultParamDim$
}{
	\iterationStepRMSpropBias
}
}

\newcommand{\defRMSpropBias}{
\SGDdef[true]
	{def:RMSprop_bias}
	{Bias-adjusted \RMSprop\ \SGD\ optimization method}
	{ 
	$(\gamma_n)_{n \in \N} \subseteq [0,\infty)$, 
	$(\beta_n)_{n \in \N} \subseteq [0,1)$, 
	$\varepsilon \in (0,\infty)$,}
	{bias-adjusted \RMSprop\ \SGD\ process for the loss function $\defaultStochLoss$ with learning rates $(\gamma_n)_{n \in \N}$, second moment decay factors $(\beta_n)_{n \in \N}$, regularizing factor $\varepsilon$, initial value $\xi$,}
	{there exists $\mathbb{M} = (\mathbb{M}^{(1)},\ldots,\mathbb{M}^{(\defaultParamDim)}) \colon \N_0 \to \R^\defaultParamDim$ such that for all $n \in \N$, $i \in \{1,2,\ldots,\defaultParamDim\}$ it holds that
	\begin{equation}
	\label{eq:RMSprop_bias_1}
	\Theta_0 = \xi, \qquad \mathbb{M}_0 = 0,
	\end{equation}
	\begin{equation}
	\label{eq:RMSprop_bias_2}
	\mathbb{M}_n^{(i)} 
	= 
	\beta_n\,\mathbb{M}_{n-1}^{(i)} + (1-\beta_n)\br*{\frac{1}{J_n} { \sum_{j = 1}^{J_n}} \defaultStochGradient_{i}(\Theta_{n-1}, X_{n,j})}^2, 
	\qand
	\end{equation}
	\begin{equation}
	\label{eq:RMSprop_bias_3}
	\Theta_n^{(i)}
	= 
	\Theta_{n-1}^{(i)} 
	- 
	\gamma_n
	{\textstyle 
	    \br*{\varepsilon + \br*{ \frac{ \mathbb{M}_n^{ (i) } }{ ( 1 - \prod_{l = 1}^n \beta_l ) } }^{ \nicefrac{1}{2} } 
	    }^{ - 1 } 
	  }
	\br*{\frac{1}{J_n}\sum_{j = 1}^{J_n}\defaultStochGradient_{i}(\Theta_{n-1}, X_{n,j})}.
	\end{equation}
	}
}

\newcommand{\defdetermAdadelta}{
\deterministicGDdef[true]
	{def:determ_adadelta}
	{Adadelta \GD\ optimization method}
	{ 
	$(\beta_n)_{n \in \N} \subseteq [0,1]$, 
	$ (\delta_n)_{n \in \N} \subseteq [0,1]$, 
	$\varepsilon \in (0,\infty)$, }
	{Adadelta \GD\ process for the objective function $\defaultLossFunction$ with second moment decay factors $(\beta_n)_{n \in \N}$, 
	delta decay factors $(\delta_n)_{n \in \N}$, 
	regularizing factor $\varepsilon$, 
	and initial value $\xi$}
	{there exist $\mathbb{M} = (\mathbb{M}^{(1)},\ldots,\mathbb{M}^{(\defaultParamDim)}) \colon \N_0 \to \R^\defaultParamDim$ and $ \Delta = (\Delta^{(1)}, \ldots, \Delta^{(\defaultParamDim)}) \colon \N_0 \to \R^\defaultParamDim $ such that for all $n \in \N$, $i \in \{1,2,\ldots,\defaultParamDim\}$ it holds that
	\begin{equation}
	\label{def:determ_adadelta:eq1}
	\Theta_0 = \xi, \qquad \mathbb{M}_0 = 0, \qquad \Delta_0 = 0, 
	\end{equation}
	\begin{equation}
	\label{def:determ_adadelta:eq2}
	\mathbb{M}_n^{(i)}= \beta_n \mathbb{M}_{n-1}^{(i)} + (1-\beta_n) \abs*{ \defaultGradientFunction_i( \Theta_{n-1}) }^2, 
	\end{equation}
	\begin{equation}
	\label{def:determ_adadelta:eq3}
	\Theta_n^{(i)} = \Theta_{n-1}^{(i)} - \bbbbr{\frac{\varepsilon +\Delta_{n-1}^{(i)} }{\varepsilon +\mathbb{M}_n^{(i)} }}^{\!\nicefrac{1}{2}}  \defaultGradientFunction_i(\Theta_{n-1}), 
	\end{equation}
	\begin{equation}
	\label{def:determ_adadelta:eq4}
	\andq 
	  \Delta_n^{ (i) }
	  =
	  \delta_n
	  \Delta_{ n - 1 }^{ (i) }
	  +
	  ( 1 - \delta_n )
	  \abs{
	    \Theta_n^{(i)} - \Theta_{n-1}^{(i)}
	  }^2 .
	\end{equation}
	}
}

\newcommand{\iterationStepDetermAdadelta}{
	\State
	$
		\mathbb{M}
	\assign
		\beta_n \mathbb{M} 
		+ 
		(1-\beta_n) 
		\br*{ \grad( \Theta) }^2
	$
	\State
	$
		\Theta 
	\assign
		\Theta 
		- 
		\br*{\frac{\varepsilon +\Delta }{\varepsilon +\mathbb{M} }}^{1/2}  
		\compMulti
		\grad(\Theta)
	$
	\State
	$
		\Delta
	\assign
		\delta_n
		\Delta
		+ 
		( 1 - \delta_n )
		\br{ 
			\Theta - \Xi
		}^2 
	$
	\State 
	$
		\Xi
	\assign
		\Theta
	$
}
\newcommand{\algDescrDetermAdadelta}{\algorithmicDescription{Adadelta \GD\ optimization method}{def:determ_adadelta}{ 
	$(\beta_n)_{n \in \N} \subseteq [0,1]$, 
	$ (\delta_n)_{n \in \N} \subseteq [0,1]$, 
	$\varepsilon \in (0,\infty)$, 
}{
	Adadelta \GD\ process for the objective function $\defaultLossFunction$ with second moment decay factors $(\beta_n)_{n \in \N}$, delta decay factors $(\delta_n)_{n \in \N}$, regularizing factor $\varepsilon$, and initial value $\xi$
}{
	$\Theta \assign \xi$\separator 
	$\Xi \assign \xi$\separator
	$\mathbb{M} \assign 0 \in \R^\defaultParamDim$\separator
	$\Delta \assign 0 \in \R^\defaultParamDim$
}{
	\iterationStepDetermAdadelta
}
}

\newcommand{\iterationStepAdadelta}{
	\pseudoCodeGradientAssign
	\State
	$
		\mathbb{M}
	\assign
		\beta_n \mathbb{M} 
		+ 
		(1-\beta_n) 
		\pseudoCodeGradient^2
	$
	\State
	$
		\Theta 
	\assign
		\Theta 
		- 
		\br*{\frac{\varepsilon +\Delta }{\varepsilon +\mathbb{M} }}^{1/2}  
		\compMulti
		\pseudoCodeGradient
	$
	\State
	$
		\Delta
	\assign
		\delta_n
		\Delta
		+ 
		( 1 - \delta_n )
		\br{ 
			\Theta - \Xi
		}^2 
	$
	\State 
	$
		\Xi
	\assign
		\Theta
	$
}
\newcommand{\algDescrAdadelta}{\algorithmicDescriptionStochastic{Adadelta \SGD\ optimization method}{def:adadelta}{ 
	$(\beta_n)_{n \in \N} \subseteq [0,1]$, 
	$ (\delta_n)_{n \in \N} \subseteq [0,1]$, 
	$\varepsilon \in (0,\infty)$, 
}{
	Adadelta \SGD\ process for the loss function $\defaultStochLoss$ with second moment decay factors $(\beta_n)_{n \in \N}$, delta decay factors $(\delta_n)_{n \in \N}$, regularizing factor $\varepsilon$, initial value $\xi$,
}{
	$\Theta \assign \xi$\separator 
	$\Xi \assign \xi$\separator
	$\mathbb{M} \assign 0 \in \R^\defaultParamDim$\separator
	$\Delta \assign 0 \in \R^\defaultParamDim$
}{
	\iterationStepAdadelta
}
}

\newcommand{\defAdadelta}{
\SGDdef[true]
	{def:adadelta}
	{Adadelta \SGD\ optimization method}
	{ 
	$(\beta_n)_{n \in \N} \subseteq [0,1]$, 
	$(\delta_n)_{n \in \N} \subseteq [0,1]$, 
	$\varepsilon \in (0,\infty)$,}
	{Adadelta \SGD\ process for the loss function $\defaultStochLoss$ with 
	second moment decay factors $(\beta_n)_{n \in \N}$, 
	delta decay factors $(\delta_n)_{n \in \N}$, 
	regularizing factor $\varepsilon$, 
	initial value $\xi$,}
	{there exist $\mathbb{M} = (\mathbb{M}^{(1)},\ldots,\mathbb{M}^{(\defaultParamDim)})\colon \N_0\times \Omega \to \R^\defaultParamDim$ and $\Delta = (\Delta^{(1)}, \ldots, \Delta^{(\defaultParamDim)}) \colon \N_0\times \Omega \to \R^\defaultParamDim$ such that for all $n \in \N$, $i \in \{1,2,\ldots,\defaultParamDim\}$ it holds that
	\begin{equation}
	\label{eq:adadelta_1}
	\Theta_0 = \xi, \qquad \mathbb{M}_0 = 0, \qquad \Delta_0 = 0, 
	\end{equation}
	\begin{equation}
	\label{eq:adadelta_2}
	\mathbb{M}_n^{(i)}
	= 
	\beta_n \,\mathbb{M}_{n-1}^{(i)} + (1-\beta_n)\br*{\frac{1}{J_n}\sum_{j = 1}^{J_n} \defaultStochGradient_{i}(\Theta_{n-1}, X_{n,j})}^2,
	\end{equation}
	\begin{equation}
	\label{eq:adadelta_3}
	\Theta_n^{(i)}
	=
	\Theta_{n-1}^{(i)} - \bbbpr{\frac{\varepsilon +\Delta_{n-1}^{(i)} }{\varepsilon +\mathbb{M}_n^{(i)} }}^{\!\nicefrac{1}{2}}
	\br*{\frac{1}{J_n}\sum_{j = 1}^{J_n}\defaultStochGradient_{i}(\Theta_{n-1}, X_{n,j})},
	\end{equation}
	\begin{equation}
	\label{eq:adadelta_4}
	\andq \Delta_n^{(i)} = \delta_n\Delta_{n-1}^{(i)} + (1-\delta_n)\babs{\Theta_n^{(i)}-\Theta_{n-1}^{(i)}}^2.
	\end{equation}
	}
}

\newcommand{\defdetermAdam}{
\deterministicGDdef[true]
	{def:determ_adam}
	{\Adam\ \GD\ optimization method}
	{ 
	$(\gamma_n)_{n \in \N} \subseteq [0,\infty)$, 
	$(\alpha_n)_{n \in \N} \subseteq [0,1)$, 
	$(\beta_n)_{n \in \N} \subseteq [0,1)$, 
	$\varepsilon \in (0,\infty)$, }
	{\Adam\ \GD\ process for the objective function $\defaultLossFunction$ with learning rates $(\gamma_n)_{n \in \N}$, 
	momentum decay factors $(\alpha_n)_{n \in \N}$, 
	second moment decay factors $(\beta_n)_{n \in \N}$, 
	regularizing factor $\varepsilon$, 
	and initial value $\xi$}
	{there exist $\mathbf{m} = (\mathbf{m}^{(1)},\ldots,\mathbf{m}^{(\defaultParamDim)}) \colon \N_0 \to \R^\defaultParamDim$ and $\mathbb{M} = (\mathbb{M}^{(1)}, \ldots, \mathbb{M}^{(\defaultParamDim)}) \colon \N_0 \to \R^\defaultParamDim$ such that for all $n \in \N$, $i \in \{1,2,\ldots,\defaultParamDim\}$ it holds that
	\begin{equation}
	\Theta_0 = \xi, 
	\qquad 
	\mathbf{m}_0 = 0, 
	\qquad 
	\mathbb{M}_0 = 0, 
	\end{equation}
	\begin{equation}
	\mathbf{m}_n = \alpha_n \mathbf{m}_{ n - 1 } 
	+ ( 1 - \alpha_n ) \defaultGradientFunction( \Theta_{ n - 1 } ) ,
	\end{equation}
	\begin{equation}
	\mathbb{M}_n^{(i)} 
	= \beta_n\mathbb{M}_{n-1}^{(i)} 
	+ (1-\beta_n)
	\abs*{ \defaultGradientFunction_i( \Theta_{n-1} ) }^2
	, 
	\end{equation}
	\begin{equation}
	\andq 
	\Theta_n^{ (i) } 
	= 
	\Theta_{n-1}^{ (i) } - 
	\gamma_n 
	{
	\textstyle
	\br*{
	\varepsilon
	+
	\br*{
	\frac{
	\mathbb{M}_n^{ (i) }
	}{
	( 1 - \prod_{ l = 1 }^n \beta_l )
	}
	}^{ \nicefrac{ 1 }{ 2 } }
	}^{ - 1 }
	}
	\br*{
	\frac{
	\mathbf{m}_n^{ (i) }
	}{
	( 1 - \prod_{ l = 1 }^n \alpha_l )
	}
	}
	.
	\end{equation}
	}
}

\newcommand{\iterationStepDetermAdam}{
	\State
 	$
		\mathbf{m} 
	\assign
		\alpha_n  \mathbf{m} 
		+ 
		( 1 - \alpha_n ) \grad( \Theta )
	$
	\State
	$
		\mathbb{M}
	\assign
		\beta_n\mathbb{M}
		+ 
		(1-\beta_n)
		\br*{ \grad( \Theta ) }^2
	$
	\State
	$
		\Theta
	\assign
		\Theta 
		-  
		\br*{
		\varepsilon 
		+ 
		\br*{ 
		\frac{
		\mathbb{M}
		}{ 
			1 - \prod_{ k = 1 }^n \beta_k 
		}
		}^{ \nicefrac{ 1 }{ 2 } } 
		}^{ - 1 }
		\br*{
		\frac{
			\gamma_n \mathbf{m} 
		}{
			1 - \prod_{ k = 1 }^n \alpha_k
		}
		}
	$
}
\newcommand{\algDescrDetermAdam}{\algorithmicDescription{\Adam\ \GD\ optimization method}{def:determ_adam}{ 
	$ ( \gamma_n )_{ n \in \N } \subseteq [0,\infty) $, 
	$ ( \alpha_n )_{ n \in \N } \subseteq [0,1) $, 
	$ ( \beta_n )_{ n \in \N } \subseteq [0,1) $,
	$ \varepsilon \in (0,\infty) $, 
}{
	\Adam\ \GD\ process 
	for the objective function $ \defaultLossFunction $ 
	with learning rates $ ( \gamma_n )_{ n \in \N } $, 
	momentum decay factors $ ( \alpha_n )_{ n \in \N } $, 
	second moment decay factors $ ( \beta_n )_{ n \in \N } $, 
	regularizing factor $ \varepsilon $, 
	and initial value $ \xi $
}{
	$\Theta \assign \xi$\separator 
	$\mathbf{m} \assign 0 \in \R^\defaultParamDim$\separator
	$\mathbb{M} \assign 0 \in \R^\defaultParamDim$
}{
 	\iterationStepDetermAdam
}
}

\newcommand{\iterationStepAdam}{
	\pseudoCodeGradientAssign
	\State
 	$
		\mathbf{m} 
	\assign
		\alpha_n  \mathbf{m} 
		+ 
		( 1 - \alpha_n ) \pseudoCodeGradient
	$
	\State
	$
		\mathbb{M}
	\assign
		\beta_n\mathbb{M}
		+ 
		(1-\beta_n)
		\pseudoCodeGradient^2
	$
	\State
	$
		\Theta
	\assign
		\Theta 
		-  
		\br*{
		\varepsilon 
		+ 
		\br*{ 
		\frac{
		\mathbb{M}
		}{ 
			1 - \prod_{ k = 1 }^n \beta_k 
		}
		}^{ \nicefrac{ 1 }{ 2 } } 
		}^{ - 1 }
		\br*{
		\frac{
			\gamma_n \mathbf{m} 
		}{
			1 - \prod_{ k = 1 }^n \alpha_k
		}
		}
	$
}
\newcommand{\algDescrAdam}{\algorithmicDescriptionStochastic{\Adam\ \SGD\ optimization method}{def:adam}{ 
	$ ( \gamma_n )_{ n \in \N } \subseteq [0,\infty) $, 
	$ ( \alpha_n )_{ n \in \N } \subseteq [0,1) $, 
	$ ( \beta_n )_{ n \in \N } \subseteq [0,1) $,
	$ \varepsilon \in (0,\infty) $, 
}{
	\Adam\ \SGD\ process 
	for the loss function $\defaultStochLoss$ 
	with learning rates $ ( \gamma_n )_{ n \in \N } $, 
	momentum decay factors $ ( \alpha_n )_{ n \in \N } $, 
	second moment decay factors $ ( \beta_n )_{ n \in \N } $, 
	regularizing factor $ \varepsilon $, 
	initial value $ \xi $,
}{
	$\Theta \assign \xi$\separator 
	$\mathbf{m} \assign 0 \in \R^\defaultParamDim$\separator
	$\mathbb{M} \assign 0 \in \R^\defaultParamDim$
}{
 	\iterationStepAdam
}
}

\newcommand{\defAdam}{
\SGDdef[true]
	{def:adam}
	{\Adam\ \SGD\ optimization method}
	{ 
	$(\gamma_n)_{n \in \N} \subseteq [0,\infty)$, 
	$(\alpha_n)_{n \in \N} \subseteq [0,1]$, 
	$(\beta_n)_{n \in \N} \subseteq [0,1]$, 
	$\varepsilon \in (0,\infty)$,}
	{\Adam\ \SGD\ process for the loss function $\defaultStochLoss$ with 
	learning rates $(\gamma_n)_{n \in \N}$, 
	momentum decay factors $(\alpha_n)_{n \in \N}$, 
	second moment decay factors $(\beta_n)_{n \in \N}$, 
	regularizing factor $\varepsilon$, 
	initial value $\xi$,}
	{there exist $\mathbf{m} = (\mathbf{m}^{(1)},\ldots,\mathbf{m}^{(\defaultParamDim)}) \colon \N_0\times \Omega \to \R^\defaultParamDim$ and $\mathbb{M} = (\mathbb{M}^{(1)},\ldots,\mathbb{M}^{(\defaultParamDim)}) \colon \N_0\times \Omega \to \R^\defaultParamDim$ such that for all $n \in \N$, $i \in \{1,2,\ldots,\defaultParamDim\}$ it holds that
	\begin{equation}
	\Theta_0 = \xi, \qquad \mathbf{m}_0 = 0, \qquad \mathbb{M}_0 = 0, 
	\end{equation}
	\begin{equation}
	\mathbf{m}_n = \alpha_n \, \mathbf{m}_{n-1} + (1-\alpha_n)\br*{\frac{1}{J_n}\sum_{j = 1}^{J_n}\defaultStochGradient(\Theta_{n-1}, X_{n,j})},
	\end{equation}
	\begin{equation}
	\mathbb{M}_n^{(i)} = \beta_n\,\mathbb{M}_{n-1}^{(i)} + (1-\beta_n)\br*{\frac{1}{J_n} \sum_{j = 1}^{J_n} \defaultStochGradient_{i}(\Theta_{n-1}, X_{n,j})}^2,
	\end{equation}
	\begin{equation}
	\andq
	\Theta_n^{(i)} = \Theta_{n-1}^{(i)}
	-
	\gamma_n
	\,
	{\textstyle
	\br*{\varepsilon + \br*{ \frac{ \mathbb{M}_n^{ (i) } }{ ( 1 - \prod_{l = 1}^n \beta_l ) } }^{ \nicefrac{1}{2} }
	}^{ - 1 }
	}
	\br*{
	\frac{ \mathbf{m}_n^{ (i) } }{ ( 1 - \prod_{l = 1}^n \alpha_l ) }
	}.
	\end{equation}
	}
}

\newcommand{\defdetermAdamax}{
\deterministicGDdef[true]
	{def:determ_adamax}
	{Adamax \GD\ optimization method}
	{ 
	$(\gamma_n)_{n \in \N} \subseteq [0,\infty)$, 
	$(\alpha_n)_{n \in \N} \subseteq [0,1)$, 
	$(\beta_n)_{n \in \N} \subseteq [0,1]$, 
	$\varepsilon \in (0,\infty)$, }
	{Adamax \GD\ process for the objective function $\defaultLossFunction$ with learning rates $(\gamma_n)_{n \in \N}$, 
	momentum decay factors $(\alpha_n)_{n \in \N}$, 
	second moment decay factors $(\beta_n)_{n \in \N}$, 
	regularizing factor $\varepsilon$, 
	and initial value $\xi$}
	{there exist $\mathbf{m} = (\mathbf{m}^{(1)},\ldots,\mathbf{m}^{(\defaultParamDim)}) \colon \N_0 \to \R^\defaultParamDim$ and $\mathbb{M} = (\mathbb{M}^{(1)}, \ldots, \mathbb{M}^{(\defaultParamDim)}) \colon \N_0 \to \R^\defaultParamDim$ such that for all $n \in \N$, $i \in \{1,2,\ldots,\defaultParamDim\}$ it holds that
	\begin{equation}
	\Theta_0 = \xi, 
	\qquad 
	\mathbf{m}_0 = 0, 
	\qquad 
	\mathbb{M}_0 = 0, 
	\end{equation}
	\begin{equation}
	\mathbf{m}_n = \alpha_n  \mathbf{m}_{ n - 1 } 
	+ ( 1 - \alpha_n )  \defaultGradientFunction( \Theta_{ n - 1 } ) ,
	\end{equation}
	\begin{equation}
	\mathbb{M}_n^{(i)} 
	= 
	\max\cu[\big]{
		\beta_n\mathbb{M}_{n-1}^{(i)} 
		,
		\abs*{ \defaultGradientFunction_i( \Theta_{n-1} ) }
	},
	\end{equation}
	\begin{equation}
	\andq 
	\Theta_n^{ (i) } 
	= 
	\Theta_{n-1}^{ (i) } - 
	\gamma_n 
	{
	\textstyle
	\br*{
	\varepsilon
	+
	\mathbb{M}_n^{ (i)} 
	}^{ - 1 }
	}
	\br*{
	\frac{
	\mathbf{m}_n^{ (i) } 
	}{
	( 1 - \prod_{ l = 1 }^n \alpha_l )
	}
	}
	.
	\end{equation}
	}
}

\newcommand{\iterationStepDetermAdamax}{
	\State
 	$
		\mathbf{m} 
	\assign
		\alpha_n  \mathbf{m} 
		+ 
		( 1 - \alpha_n ) \grad( \Theta )
	$
	\State
	$
		\mathbb{M}
	\assign
		\max\cu[\big]{
			\beta_n\mathbb{M} 
			,
			\abs*{ \grad( \Theta ) }
		}
	$
	\State
	$
		\Theta
	\assign
		\Theta
		- 
		\gamma_n 
		\br*{
			\varepsilon 
			+ 
			\mathbb{M}
		}^{ - 1 }
		\br*{
			\frac{
			\mathbf{m}
			}{
				1 - \prod_{ k = 1 }^n \alpha_k
			}
		}
	$
}
\newcommand{\algDescrDetermAdamax}{\algorithmicDescription{Adamax \GD\ optimization method}{def:determ_adamax}{ 
	$ ( \gamma_n )_{ n \in \N } \subseteq [0,\infty) $, 
	$ ( \alpha_n )_{ n \in \N } \subseteq [0,1) $, 
	$ ( \beta_n )_{ n \in \N } \subseteq [0,1] $,
	$ \varepsilon \in (0,\infty) $, 
}{
	Adamax \GD\ process 
	for the objective function $ \defaultLossFunction $ 
	with learning rates $ ( \gamma_n )_{ n \in \N } $, 
	momentum decay factors $ ( \alpha_n )_{ n \in \N } $, 
	second moment decay factors $ ( \beta_n )_{ n \in \N } $, 
	regularizing factor $ \varepsilon $, 
	and initial value $ \xi $
}{
	$\Theta \assign \xi$\separator 
	$\mathbf{m} \assign 0 \in \R^\defaultParamDim$\separator
	$\mathbb{M} \assign 0 \in \R^\defaultParamDim$
}{
	\iterationStepDetermAdamax
}
}

\newcommand{\iterationStepAdamax}{
	\pseudoCodeGradientAssign
	\State
 	$
		\mathbf{m} 
	\assign
		\alpha_n  \mathbf{m} 
		+ 
		( 1 - \alpha_n ) \pseudoCodeGradient
	$
	\State
	$
		\mathbb{M}
	\assign
		\max\cu[\big]{
			\beta_n\mathbb{M} 
			,
			\abs*{ \pseudoCodeGradient }
		}
	$
	\State
	$
		\Theta
	\assign
		\Theta
		- 
		\gamma_n 
		\br*{
			\varepsilon 
			+ 
			\mathbb{M}
		}^{ - 1 }
		\br*{
			\frac{
			\mathbf{m}
			}{
				1 - \prod_{ k = 1 }^n \alpha_k
			}
		}
	$
}
\newcommand{\algDescrAdamax}{\algorithmicDescriptionStochastic{Adamax \SGD\ optimization method}{def:adamax}{ 
	$ ( \gamma_n )_{ n \in \N } \subseteq [0,\infty) $, 
	$ ( \alpha_n )_{ n \in \N } \subseteq [0,1) $, 
	$ ( \beta_n )_{ n \in \N } \subseteq [0,1] $,
	$ \varepsilon \in (0,\infty) $, 
}{
	Adamax \SGD\ process 
	for the loss function $\defaultStochLoss$ 
	with learning rates $ ( \gamma_n )_{ n \in \N } $, 
	momentum decay factors $ ( \alpha_n )_{ n \in \N } $, 
	second moment decay factors $ ( \beta_n )_{ n \in \N } $, 
	regularizing factor $ \varepsilon $, 
	initial value $ \xi $,
}{
	$\Theta \assign \xi$\separator 
	$\mathbf{m} \assign 0 \in \R^\defaultParamDim$\separator
	$\mathbb{M} \assign 0 \in \R^\defaultParamDim$
}{
	\iterationStepAdamax
}
}

\newcommand{\defAdamax}{
\SGDdef[true]
	{def:adamax}
	{Adamax \SGD\ optimization method}
	{ 
	$(\gamma_n)_{n \in \N} \subseteq [0,\infty)$, 
	$(\alpha_n)_{n \in \N} \subseteq [0,1)$, 
	$(\beta_n)_{n \in \N} \subseteq [0,1]$, 
	$\varepsilon \in (0,\infty)$,}
	{Adamax \SGD\ process for the loss function $\defaultStochLoss$ with 
	learning rates $(\gamma_n)_{n \in \N}$, 
	momentum decay factors $(\alpha_n)_{n \in \N}$, 
	second moment decay factors $(\beta_n)_{n \in \N}$, 
	regularizing factor $\varepsilon$, 
	initial value $\xi$,}
	{there exist $\mathbf{m} = (\mathbf{m}^{(1)},\ldots,\mathbf{m}^{(\defaultParamDim)}) \colon \N_0\times \Omega \to \R^\defaultParamDim$ and $\mathbb{M} = (\mathbb{M}^{(1)},\ldots,\mathbb{M}^{(\defaultParamDim)}) \colon \N_0\times \Omega \to \R^\defaultParamDim$ such that for all $n \in \N$, $i \in \{1,2,\ldots,\defaultParamDim\}$ it holds that
	\begin{equation}
	\Theta_0 = \xi, 
	\qquad 
	\mathbf{m}_0 = 0, 
	\qquad 
	\mathbb{M}_0 = 0, 
	\end{equation}
	\begin{equation}
	\mathbf{m}_n = \alpha_n  \mathbf{m}_{ n - 1 } 
	+ ( 1 - \alpha_n )  \br*{ \frac{1}{J_n}\sum_{j = 1}^{J_n}\defaultStochGradient(\Theta_{ n - 1 }, X_{ n, j }) },
	\end{equation}
	\begin{equation}
	\mathbb{M}_n^{(i)} 
	= 
	\max\cu*{
		\beta_n\mathbb{M}_{n-1}^{(i)} 
		,
		\abs*{ \frac{1}{J_n}\sum_{j = 1}^{J_n}\defaultStochGradient_i(\Theta_{n-1}, X_{n,j}) }
	},
	\end{equation}
	\begin{equation}
	\andq
	\Theta_n^{ (i) }
	=
	\Theta_{n-1}^{ (i) }
	-
	\gamma_n 
	{
	\textstyle
	\br*{
	\varepsilon
	+
	\mathbb{M}_n^{ (i) }
	}^{ - 1 }
	}
	\br*{
	\frac{
	\mathbf{m}_n^{ (i) }
	}{
	( 1 - \prod_{ l = 1 }^n \alpha_l )
	}
	}
	.
	\end{equation}
	}
}

\newcommand{\defdetermNadam}{
\deterministicGDdef[true]
	{def:determ_nadam}
	{\Nadam\ \GD\ optimization method}
	{ 
	$(\gamma_n)_{n \in \N} \subseteq [0,\infty)$, 
	$(\alpha_n)_{n \in \N} \subseteq [0,1)$, 
	$(\beta_n)_{n \in \N} \subseteq [0,1)$, 
	$\varepsilon \in (0,\infty)$, }
	{\Nadam\ \GD\ process for the objective function $\defaultLossFunction$ with learning rates $(\gamma_n)_{n \in \N}$, 
	momentum decay factors $(\alpha_n)_{n \in \N}$, 
	second moment decay factors $(\beta_n)_{n \in \N}$, 
	regularizing factor $\varepsilon$, 
	and initial value $\xi$}
	{there exist 
		$\mathbf{m} = (\mathbf{m}^{(1)},\ldots,\mathbf{m}^{(\defaultParamDim)}) \colon \N_0 \to \R^\defaultParamDim$
		and
		$\mathbb{M} = (\mathbb{M}^{(1)}, \ldots, \mathbb{M}^{(\defaultParamDim)}) \colon \N_0 \to \R^\defaultParamDim$
	such that for all $n \in \N$, $i \in \{1,2,\ldots,\defaultParamDim\}$ it holds that
	\begin{equation}
	\Theta_0 = \xi, 
	\qquad 
	\mathbf{m}_0 = 0, 
	\qquad 
	\mathbb{M}_0 = 0, 
	\end{equation}
	\begin{equation}
	\mathbf{m}_n = \alpha_n  \mathbf{m}_{ n - 1 } 
	+ ( 1 - \alpha_n ) \defaultGradientFunction( \Theta_{ n - 1 } ) ,
	\end{equation}
	\begin{equation}
	\mathbb{M}_n^{(i)}
	= \beta_n\mathbb{M}_{n-1}^{(i)}
	+ (1-\beta_n)
	\abs*{ \defaultGradientFunction_i( \Theta_{n-1} ) }^2
	,
	\qand
	\end{equation}
	\begin{equation}
	\Theta_n^{ (i) }
	=
	\Theta_{n-1}^{ (i) } 
	-
	{
	\textstyle
	\br*{
	\varepsilon
	+
	\br*{
	\frac{
	\mathbb{M}_n^{ (i) }
	}{
	( 1 - \prod_{ l = 1 }^n \beta_l )
	}
	}^{ \nicefrac{ 1 }{ 2 } }
	}^{ - 1 }
	}
	\br*{
		\br*{
			\tfrac{
				\gamma_n
				(1 - \alpha_n)
			}{
				1 - \prod_{ l = 1 }^n \alpha_l
			}
		}
		\defaultGradientFunction_i( \Theta_{ n - 1 } )
		+
		\br*{
			\tfrac{
				\gamma_{n+1}
				\alpha_{n+1}
			}{
				1 - \prod_{ l = 1 }^{n+1} \alpha_l
			}
		}
		\mathbf{m}_n^{(i)}
	}
	.
	\end{equation}
	}
}

\newcommand{\iterationStepDetermNadam}{
	\State
 	$
		\mathbf{m} 
	\assign
		\alpha_n  \mathbf{m} 
		+ 
		( 1 - \alpha_n ) \grad( \Theta )
	$
	\State
	$
		\mathbb{M}
	\assign
		\beta_n\mathbb{M}
		+ 
		(1-\beta_n)
		\br*{ \grad( \Theta ) }^2
	$
	\State
	$
		\Theta
	\assign
		\Theta 
		-  
		\br*{
		\varepsilon 
		+ 
		\br*{ 
		\frac{
		\mathbb{M}
		}{ 
			1 - \prod_{ k = 1 }^n \beta_k 
		}
		}^{ \nicefrac{ 1 }{ 2 } } 
		}^{ - 1 }
		\br[\bigg]{
			\br*{
					\tfrac{
						\gamma_n
						(1 - \alpha_n)
					}{
						1 - \prod_{ k = 1 }^n \alpha_k
					}
				}
				\grad( \Theta )
				+
				\br*{
					\tfrac{
						\gamma_{n+1}
						\alpha_{n+1}
					}{
						1 - \prod_{ k = 1 }^{n+1} \alpha_k
					}
				}
				\mathbf{m}
		}
	$
}
\newcommand{\algDescrDetermNadam}{\algorithmicDescription{\Nadam\ \GD\ optimization method}{def:determ_nadam}{ 
	$ ( \gamma_n )_{ n \in \N } \subseteq [0,\infty) $, 
	$ ( \alpha_n )_{ n \in \N } \subseteq [0,1) $, 
	$ ( \beta_n )_{ n \in \N } \subseteq [0,1) $,
	$ \varepsilon \in (0,\infty) $, 
}{
	\Nadam\ \GD\ process 
	for the objective function $ \defaultLossFunction $ 
	with learning rates $ ( \gamma_n )_{ n \in \N } $, 
	momentum decay factors $ ( \alpha_n )_{ n \in \N } $, 
	second moment decay factors $ ( \beta_n )_{ n \in \N } $, 
	regularizing factor $ \varepsilon $, 
	and initial value $ \xi $
}{
	$\Theta \assign \xi$\separator 
	$\mathbf{m} \assign 0 \in \R^\defaultParamDim$\separator
	$\mathbb{M} \assign 0 \in \R^\defaultParamDim$
}{
 	\iterationStepDetermNadam
}
}

\newcommand{\iterationStepNadam}{
	\pseudoCodeGradientAssign
	\State
 	$
		\mathbf{m} 
	\assign
		\alpha_n  \mathbf{m} 
		+ 
		( 1 - \alpha_n ) \pseudoCodeGradient
	$
	\State
	$
		\mathbb{M}
	\assign
		\beta_n\mathbb{M}
		+ 
		(1-\beta_n)
		\pseudoCodeGradient^2
	$
	\State
	$
		\Theta
	\assign
		\Theta 
		-  
		\br*{
		\varepsilon 
		+ 
		\br*{ 
		\frac{
		\mathbb{M}
		}{ 
			1 - \prod_{ k = 1 }^n \beta_k 
		}
		}^{ \nicefrac{ 1 }{ 2 } } 
		}^{ - 1 }
		\br[\bigg]{
			\br*{
					\tfrac{
						\gamma_n
						(1 - \alpha_n)
					}{
						1 - \prod_{ k = 1 }^n \alpha_k
					}
				}
				\pseudoCodeGradient
				+
				\br*{
					\tfrac{
						\gamma_{n+1}
						\alpha_{n+1}
					}{
						1 - \prod_{ k = 1 }^{n+1} \alpha_k
					}
				}
				\mathbf{m}
		}
	$
}
\newcommand{\algDescrNadam}{\algorithmicDescriptionStochastic{\Nadam\ \SGD\ optimization method}{def:nadam}{ 
	$ ( \gamma_n )_{ n \in \N } \subseteq [0,\infty) $, 
	$ ( \alpha_n )_{ n \in \N } \subseteq [0,1) $, 
	$ ( \beta_n )_{ n \in \N } \subseteq [0,1) $,
	$ \varepsilon \in (0,\infty) $, 
}{
	\Nadam\ \SGD\ process 
	for the loss function $\defaultStochLoss$ 
	with learning rates $ ( \gamma_n )_{ n \in \N } $, 
	momentum decay factors $ ( \alpha_n )_{ n \in \N } $, 
	second moment decay factors $ ( \beta_n )_{ n \in \N } $, 
	regularizing factor $ \varepsilon $, 
	initial value $ \xi $,
}{
	$\Theta \assign \xi$\separator 
	$\mathbf{m} \assign 0 \in \R^\defaultParamDim$\separator
	$\mathbb{M} \assign 0 \in \R^\defaultParamDim$
}{
 	\iterationStepNadam
}
}

\newcommand{\defNadam}{
\SGDdef[true]
	{def:nadam}
	{\Nadam\ \SGD\ optimization method}
	{ 
	$(\gamma_n)_{n \in \N} \subseteq [0,\infty)$, 
	$(\alpha_n)_{n \in \N} \subseteq [0,1)$, 
	$(\beta_n)_{n \in \N} \subseteq [0,1)$, 
	$\varepsilon \in (0,\infty)$,}
	{\Nadam\ \SGD\ process for the loss function $\defaultStochLoss$ with 
	learning rates $(\gamma_n)_{n \in \N}$, 
	momentum decay factors $(\alpha_n)_{n \in \N}$, 
	second moment decay factors $(\beta_n)_{n \in \N}$, 
	regularizing factor $\varepsilon$, 
	initial value $\xi$,}
	{there exist 
		$\mathbf{m} = (\mathbf{m}^{(1)},\ldots,\mathbf{m}^{(\defaultParamDim)}) \colon \N_0\times \Omega \to \R^\defaultParamDim$
		and
		$\mathbb{M} = (\mathbb{M}^{(1)},\ldots,\mathbb{M}^{(\defaultParamDim)}) \colon \N_0\times \Omega \to \R^\defaultParamDim$
	such that for all $n \in \N$, $i \in \{1,2,\ldots,\defaultParamDim\}$ it holds that
	\begin{equation}
	\Theta_0 = \xi, 
	\qquad 
	\mathbf{m}_0 = 0, 
	\qquad 
	\mathbb{M}_0 = 0, 
	\end{equation}
	\begin{equation}
	\mathbf{m}_n = \alpha_n  \mathbf{m}_{ n - 1 } 
	+ ( 1 - \alpha_n ) \br*{ \frac{1}{J_n}\sum_{j = 1}^{J_n}\defaultStochGradient(\Theta_{ n - 1 }, X_{ n, j }) },
	\end{equation}
	\begin{equation}
	\mathbb{M}_n^{(i)}
	= \beta_n\mathbb{M}_{n-1}^{(i)}
	+ (1-\beta_n)
	\br*{ \frac{1}{J_n}\sum_{j = 1}^{J_n}\defaultStochGradient_i(\Theta_{n-1}, X_{n,j}) }^2
	,
	\qand
	\end{equation}
	\begin{multline}
	\Theta_n^{ (i) }
	=
	\Theta_{n-1}^{ (i) }
	-
	{
	\textstyle
	\br*{
	\varepsilon
	+
	\br*{
	\frac{
	\mathbb{M}_n^{ (i) }
	}{
	( 1 - \prod_{ l = 1 }^n \beta_l )
	}
	}^{ \nicefrac{ 1 }{ 2 } }
	}^{ - 1 }
	}
	\Biggl[
		\br*{
			\tfrac{
				\gamma_n
				(1 - \alpha_n)
			}{
				1 - \prod_{ l = 1 }^n \alpha_l
			}
		}
		\br*{\frac{1}{J_n}\sum_{j = 1}^{J_n}\defaultStochGradient_i(\Theta_{ n - 1 }, X_{ n, j })}
		\\+
		\br*{
			\tfrac{
				\gamma_{n+1}
				\alpha_{n+1}
			}{
				1 - \prod_{ l = 1 }^{n+1} \alpha_l
			}
		}
		\mathbf{m}_n^{(i)}
	\Biggr]
	.
	\end{multline}
	}
}

\newcommand{\defdetermSimplifiedNadam}{
\deterministicGDdef[true]
	{def:determ_simplified_nadam}
	{Simplified \Nadam\ \GD\ optimization method}
	{ 
	$(\gamma_n)_{n \in \N} \subseteq [0,\infty)$, 
	$(\alpha_n)_{n \in \N} \subseteq [0,1)$, 
	$(\beta_n)_{n \in \N} \subseteq [0,1)$, 
	$\varepsilon \in (0,\infty)$, }
	{simplified \Nadam\ \GD\ process for the objective function $\defaultLossFunction$ with learning rates $(\gamma_n)_{n \in \N}$, 
	momentum decay factors $(\alpha_n)_{n \in \N}$, 
	second moment decay factors $(\beta_n)_{n \in \N}$, 
	regularizing factor $\varepsilon$, 
	and initial value $\xi$}
	{there exist 
		$\mathbf{m} = (\mathbf{m}^{(1)},\ldots,\mathbf{m}^{(\defaultParamDim)}) \colon \N_0 \to \R^\defaultParamDim$
		and
		$\mathbb{M} = (\mathbb{M}^{(1)}, \ldots, \mathbb{M}^{(\defaultParamDim)}) \colon \N_0 \to \R^\defaultParamDim$
	such that for all $n \in \N$, $i \in \{1,2,\ldots,\defaultParamDim\}$ it holds that
	\begin{equation}
	\Theta_0 = \xi, 
	\qquad 
	\mathbf{m}_0 = 0, 
	\qquad 
	\mathbb{M}_0 = 0, 
	\end{equation}
	\begin{equation}
	\mathbf{m}_n = \alpha_n  \mathbf{m}_{ n - 1 } 
	+ ( 1 - \alpha_n ) \defaultGradientFunction( \Theta_{ n - 1 } ) ,
	\end{equation}
	\begin{equation}
	\mathbb{M}_n^{(i)}
	= \beta_n\mathbb{M}_{n-1}^{(i)}
	+ (1-\beta_n)
	\abs*{ \defaultGradientFunction_i( \Theta_{n-1} ) }^2
	,
	\qand
	\end{equation}
	\begin{equation}
	\Theta_n^{ (i) }
	=
	\Theta_{n-1}^{ (i) } 
	-
	\gamma_n
	{
	\textstyle
	\br*{
	\varepsilon
	+
	\br*{
	\frac{
	\mathbb{M}_n^{ (i) }
	}{
	( 1 - \prod_{ l = 1 }^n \beta_l )
	}
	}^{ \nicefrac{ 1 }{ 2 } }
	}^{ - 1 }
	}
	\br*{
		\br*{
			\tfrac{
				1 - \alpha_n
			}{
				1 - \prod_{ l = 1 }^n \alpha_l
			}
		}
		\defaultGradientFunction_i( \Theta_{ n - 1 } )
		+
		\br*{
			\tfrac{
				\alpha_{n+1}
			}{
				1 - \prod_{ l = 1 }^{n+1} \alpha_l
			}
		}
		\mathbf{m}_n^{(i)}
	}
	.
	\end{equation}
	}
}

\newcommand{\iterationStepDetermSimplifiedNadam}{
	\State
 	$
		\mathbf{m} 
	\assign
		\alpha_n  \mathbf{m} 
		+ 
		( 1 - \alpha_n ) \grad( \Theta )
	$
	\State
	$
		\mathbb{M}
	\assign
		\beta_n\mathbb{M}
		+ 
		(1-\beta_n)
		\br*{ \grad( \Theta ) }^2
	$
	\State
	$
		\Theta
	\assign
		\Theta 
		-  
		\gamma_n
		\br*{
		\varepsilon 
		+ 
		\br*{ 
		\frac{
		\mathbb{M}
		}{ 
			1 - \prod_{ k = 1 }^n \beta_k 
		}
		}^{ \nicefrac{ 1 }{ 2 } } 
		}^{ - 1 }
		\br[\bigg]{
			\br*{
					\tfrac{
						1 - \alpha_n
					}{
						1 - \prod_{ k = 1 }^n \alpha_k
					}
				}
				\grad( \Theta )
				+
				\br*{
					\tfrac{
						\alpha_{n+1}
					}{
						1 - \prod_{ k = 1 }^{n+1} \alpha_k
					}
				}
				\mathbf{m}
		}
	$
}
\newcommand{\algDescrDetermSimplifiedNadam}{\algorithmicDescription{Simplified \Nadam\ \GD\ optimization method}{def:determ_simplified_nadam}{ 
	$ ( \gamma_n )_{ n \in \N } \subseteq [0,\infty) $, 
	$ ( \alpha_n )_{ n \in \N } \subseteq [0,1) $, 
	$ ( \beta_n )_{ n \in \N } \subseteq [0,1) $,
	$ \varepsilon \in (0,\infty) $, 
}{
	simplified Simplified \Nadam\ \GD\ process 
	for the objective function $ \defaultLossFunction $ 
	with learning rates $ ( \gamma_n )_{ n \in \N } $, 
	momentum decay factors $ ( \alpha_n )_{ n \in \N } $, 
	second moment decay factors $ ( \beta_n )_{ n \in \N } $, 
	regularizing factor $ \varepsilon $, 
	and initial value $ \xi $
}{
	$\Theta \assign \xi$\separator 
	$\mathbf{m} \assign 0 \in \R^\defaultParamDim$\separator
	$\mathbb{M} \assign 0 \in \R^\defaultParamDim$
}{
 	\iterationStepDetermSimplifiedNadam
}
}

\newcommand{\iterationStepSimplifiedNadam}{
	\pseudoCodeGradientAssign
	\State
 	$
		\mathbf{m} 
	\assign
		\alpha_n  \mathbf{m} 
		+ 
		( 1 - \alpha_n ) \pseudoCodeGradient
	$
	\State
	$
		\mathbb{M}
	\assign
		\beta_n\mathbb{M}
		+ 
		(1-\beta_n)
		\pseudoCodeGradient^2
	$
	\State
	$
		\Theta
	\assign
		\Theta 
		-  
		\gamma_n
		\br*{
		\varepsilon 
		+ 
		\br*{ 
		\frac{
		\mathbb{M}
		}{ 
			1 - \prod_{ k = 1 }^n \beta_k 
		}
		}^{ \nicefrac{ 1 }{ 2 } } 
		}^{ - 1 }
		\br[\bigg]{
			\br*{
					\tfrac{
						1 - \alpha_n
					}{
						1 - \prod_{ k = 1 }^n \alpha_k
					}
				}
				\pseudoCodeGradient
				+
				\br*{
					\tfrac{
						\alpha_{n+1}
					}{
						1 - \prod_{ k = 1 }^{n+1} \alpha_k
					}
				}
				\mathbf{m}
		}
	$
}
\newcommand{\algDescrSimplifiedNadam}{\algorithmicDescriptionStochastic{Simplified \Nadam\ \SGD\ optimization method}{def:simplified_nadam}{ 
	$ ( \gamma_n )_{ n \in \N } \subseteq [0,\infty) $, 
	$ ( \alpha_n )_{ n \in \N } \subseteq [0,1) $, 
	$ ( \beta_n )_{ n \in \N } \subseteq [0,1) $,
	$ \varepsilon \in (0,\infty) $, 
}{
	simplified \Nadam\ \SGD\ process 
	for the loss function $\defaultStochLoss$ 
	with learning rates $ ( \gamma_n )_{ n \in \N } $, 
	momentum decay factors $ ( \alpha_n )_{ n \in \N } $, 
	second moment decay factors $ ( \beta_n )_{ n \in \N } $, 
	regularizing factor $ \varepsilon $, 
	initial value $ \xi $,
}{
	$\Theta \assign \xi$\separator 
	$\mathbf{m} \assign 0 \in \R^\defaultParamDim$\separator
	$\mathbb{M} \assign 0 \in \R^\defaultParamDim$
}{
 	\iterationStepSimplifiedNadam
}
}

\newcommand{\defSimplifiedNadam}{
\SGDdef[true]
	{def:simplified_nadam}
	{Simplified \Nadam\ \SGD\ optimization method}
	{ 
	$(\gamma_n)_{n \in \N} \subseteq [0,\infty)$, 
	$(\alpha_n)_{n \in \N} \subseteq [0,1)$, 
	$(\beta_n)_{n \in \N} \subseteq [0,1)$, 
	$\varepsilon \in (0,\infty)$,}
	{simplified Simplified \Nadam\ \SGD\ process for the loss function $\defaultStochLoss$ with 
	learning rates $(\gamma_n)_{n \in \N}$, 
	momentum decay factors $(\alpha_n)_{n \in \N}$, 
	second moment decay factors $(\beta_n)_{n \in \N}$, 
	regularizing factor $\varepsilon$, 
	initial value $\xi$,}
	{there exist 
		$\mathbf{m} = (\mathbf{m}^{(1)},\ldots,\mathbf{m}^{(\defaultParamDim)}) \colon \N_0\times \Omega \to \R^\defaultParamDim$
		and
		$\mathbb{M} = (\mathbb{M}^{(1)},\ldots,\mathbb{M}^{(\defaultParamDim)}) \colon \N_0\times \Omega \to \R^\defaultParamDim$
	such that for all $n \in \N$, $i \in \{1,2,\ldots,\defaultParamDim\}$ it holds that
	\begin{equation}
	\Theta_0 = \xi, 
	\qquad 
	\mathbf{m}_0 = 0, 
	\qquad 
	\mathbb{M}_0 = 0, 
	\end{equation}
	\begin{equation}
	\mathbf{m}_n = \alpha_n  \mathbf{m}_{ n - 1 } 
	+ ( 1 - \alpha_n ) \br*{ \frac{1}{J_n}\sum_{j = 1}^{J_n}\defaultStochGradient(\Theta_{ n - 1 }, X_{ n, j }) },
	\end{equation}
	\begin{equation}
	\mathbb{M}_n^{(i)}
	= \beta_n\mathbb{M}_{n-1}^{(i)}
	+ (1-\beta_n)
	\br*{ \frac{1}{J_n}\sum_{j = 1}^{J_n}\defaultStochGradient_i(\Theta_{n-1}, X_{n,j}) }^2
	,
	\qand
	\end{equation}
	\begin{multline}
	\Theta_n^{ (i) }
	=
	\Theta_{n-1}^{ (i) }
	-
	\gamma_n
	{
	\textstyle
	\br*{
	\varepsilon
	+
	\br*{
	\frac{
	\mathbb{M}_n^{ (i) }
	}{
	( 1 - \prod_{ l = 1 }^n \beta_l )
	}
	}^{ \nicefrac{ 1 }{ 2 } }
	}^{ - 1 }
	}
	\Biggl[
		\br*{
			\tfrac{
				1 - \alpha_n
			}{
				1 - \prod_{ l = 1 }^n \alpha_l
			}
		}
		\br*{\frac{1}{J_n}\sum_{j = 1}^{J_n}\defaultStochGradient_i(\Theta_{ n - 1 }, X_{ n, j })}
		\\+
		\br*{
			\tfrac{
				\alpha_{n+1}
			}{
				1 - \prod_{ l = 1 }^{n+1} \alpha_l
			}
		}
		\mathbf{m}_n^{(i)}
	\Biggr]
	.
	\end{multline}
	}
}

\newcommand{\defdetermNadamax}{
\deterministicGDdef[true]
	{def:determ_nadamax}
	{Nadamax \GD\ optimization method}
	{ 
	$(\gamma_n)_{n \in \N} \subseteq [0,\infty)$, 
	$(\alpha_n)_{n \in \N} \subseteq [0,1)$, 
	$(\beta_n)_{n \in \N} \subseteq [0,1]$, 
	$\varepsilon \in (0,\infty)$, }
	{Nadamax \GD\ process for the objective function $\defaultLossFunction$ with learning rates $(\gamma_n)_{n \in \N}$, 
	momentum decay factors $(\alpha_n)_{n \in \N}$, 
	second moment decay factors $(\beta_n)_{n \in \N}$, 
	regularizing factor $\varepsilon$, 
	and initial value $\xi$}
	{there exist $\mathbf{m} \colon \N_0 \to \R^\defaultParamDim$
	and $\mathbb{M} \colon \N_0 \to \R^\defaultParamDim$ such that for all $n \in \N$, $i \in \{1,2,\ldots,\defaultParamDim\}$ it holds that
	\begin{equation}
	\Theta_0 = \xi, 
	\qquad 
	\mathbf{m}_0 = 0, 
	\qquad 
	\mathbb{M}_0 = 0, 
	\end{equation}
	\begin{equation}
	\mathbf{m}_n = \alpha_n \mathbf{m}_{ n - 1 } 
	+ ( 1 - \alpha_n ) \defaultGradientFunction( \Theta_{ n - 1 } ) ,
	\end{equation}
	\begin{equation}
	\mathbb{M}_n^{(i)}
	= \max\cu[\big]{
		\beta_n\mathbb{M}_{n-1}^{(i)} 
		,
		\abs*{ \defaultGradientFunction_i( \Theta_{n-1} ) }
	},
	\qand
	\end{equation}
	\begin{equation}
	\Theta_n^{ (i) }
	=
	\Theta_{n-1}^{ (i) }
	-
	\br*{
		\varepsilon 
		+
		\mathbb{M}_n^{ (i)} 
	}^{ - 1 }
	\br*{
		\br*{
			\tfrac{
				\gamma_n 
				( 1 - \alpha_n )
			}{
				1 - \prod_{ l = 1 }^n \alpha_l
			}
		}
		\defaultGradientFunction_i( \Theta_{ n - 1 } )
		+
		\br*{
			\tfrac{
				\gamma_{n+1}
				\alpha_{n+1}
			}{
				1 - \prod_{ l = 1 }^{n+1} \alpha_l
			}
		}
		\mathbf{m}_n^{ (i) }
	}
	.
	\end{equation}
	}
}

\newcommand{\iterationStepDetermNadamax}{
	\State
 	$
		\mathbf{m} 
	\assign
		\alpha_n  \mathbf{m} 
		+ 
		( 1 - \alpha_n ) \grad( \Theta )
	$
	\State
	$
		\mathbb{M}
	\assign
		\max\cu[\big]{
			\beta_n\mathbb{M} 
			,
			\abs*{ \grad( \Theta ) }
		}
	$
	\State
	$
		\Theta
	\assign
		\Theta
		- 
		\br*{
			\varepsilon 
			+ 
			\mathbb{M}
		}^{ - 1 }
		\br*{
			\br*{
					\tfrac{
						\gamma_n
						( 1 - \alpha_n )
					}{
						1 - \prod_{ l = 1 }^n \alpha_l
					}
				}
				\grad( \Theta )
				+
				\br*{
					\tfrac{
						\gamma_{n+1}
						\alpha_{n+1}
					}{
						1 - \prod_{ l = 1 }^{n+1} \alpha_l
					}
				}
				\mathbf{m}
		}
	$
}
\newcommand{\algDescrDetermNadamax}{\algorithmicDescription{Nadamax \GD\ optimization method}{def:determ_nadamax}{ 
	$ ( \gamma_n )_{ n \in \N } \subseteq [0,\infty) $, 
	$ ( \alpha_n )_{ n \in \N } \subseteq [0,1) $, 
	$ ( \beta_n )_{ n \in \N } \subseteq [0,1] $,
	$ \varepsilon \in (0,\infty) $, 
}{
	Nadamax \GD\ process 
	for the objective function $ \defaultLossFunction $ 
	with learning rates $ ( \gamma_n )_{ n \in \N } $, 
	momentum decay factors $ ( \alpha_n )_{ n \in \N } $, 
	second moment decay factors $ ( \beta_n )_{ n \in \N } $, 
	regularizing factor $ \varepsilon $, 
	and initial value $ \xi $
}{
	$\Theta \assign \xi$\separator 
	$\mathbf{m} \assign 0 \in \R^\defaultParamDim$\separator
	$\mathbb{M} \assign 0 \in \R^\defaultParamDim$
}{
 	\iterationStepDetermNadamax
}
}

\newcommand{\iterationStepNadamax}{
	\pseudoCodeGradientAssign
	\State
 	$
		\mathbf{m} 
	\assign
		\alpha_n  \mathbf{m} 
		+ 
		( 1 - \alpha_n ) \pseudoCodeGradient
	$
	\State
	$
		\mathbb{M}
	\assign
		\max\cu[\big]{
			\beta_n\mathbb{M} 
			,
			\abs*{ \pseudoCodeGradient }
		}
	$
	\State
	$
		\Theta
	\assign
		\Theta
		- 
		\br*{
			\varepsilon 
			+ 
			\mathbb{M}
		}^{ - 1 }
		\br*{
			\br*{
					\tfrac{
						\gamma_n
						( 1 - \alpha_n )
					}{
						1 - \prod_{ l = 1 }^n \alpha_l
					}
				}
				\pseudoCodeGradient
				+
				\br*{
					\tfrac{
						\gamma_{n+1}
						\alpha_{n+1}
					}{
						1 - \prod_{ l = 1 }^{n+1} \alpha_l
					}
				}
				\mathbf{m}
		}
	$
}
\newcommand{\algDescrNadamax}{\algorithmicDescriptionStochastic{Nadamax \SGD\ optimization method}{def:nadamax}{ 
	$ ( \gamma_n )_{ n \in \N } \subseteq [0,\infty) $, 
	$ ( \alpha_n )_{ n \in \N } \subseteq [0,1) $, 
	$ ( \beta_n )_{ n \in \N } \subseteq [0,1] $,
	$ \varepsilon \in (0,\infty) $, 
}{
	Nadamax \SGD\ process 
	for the loss function $\defaultStochLoss$ 
	with learning rates $ ( \gamma_n )_{ n \in \N } $, 
	momentum decay factors $ ( \alpha_n )_{ n \in \N } $, 
	second moment decay factors $ ( \beta_n )_{ n \in \N } $, 
	regularizing factor $ \varepsilon $, 
	initial value $ \xi $,
}{
	$\Theta \assign \xi$\separator 
	$\mathbf{m} \assign 0 \in \R^\defaultParamDim$\separator
	$\mathbb{M} \assign 0 \in \R^\defaultParamDim$
}{
 	\iterationStepNadamax
}
}

\newcommand{\defNadamax}{
\SGDdef[true]
	{def:nadamax}
	{Nadamax \SGD\ optimization method}
	{ 
	$(\gamma_n)_{n \in \N} \subseteq [0,\infty)$, 
	$(\alpha_n)_{n \in \N} \subseteq [0,1)$, 
	$(\beta_n)_{n \in \N} \subseteq [0,1]$, 
	$\varepsilon \in (0,\infty)$,}
	{Nadamax \SGD\ process for the loss function $\defaultStochLoss$ with 
	learning rates $(\gamma_n)_{n \in \N}$, 
	momentum decay factors $(\alpha_n)_{n \in \N}$, 
	second moment decay factors $(\beta_n)_{n \in \N}$, 
	regularizing factor $\varepsilon$, 
	initial value $\xi$,}
	{there exist 
		$\mathbf{m} = (\mathbf{m}^{(1)},\ldots,\mathbf{m}^{(\defaultParamDim)}) \colon \N_0\times \Omega \to \R^\defaultParamDim$
		and
		$\mathbb{M} = (\mathbb{M}^{(1)},\ldots,\mathbb{M}^{(\defaultParamDim)}) \colon \N_0\times \Omega \to \R^\defaultParamDim$
	such that for all $n \in \N$, $i \in \{1,2,\ldots,\defaultParamDim\}$ it holds that
	\begin{equation}
	\Theta_0 = \xi, 
	\qquad 
	\mathbf{m}_0 = 0, 
	\qquad 
	\mathbb{M}_0 = 0, 
	\end{equation}
	\begin{equation}
	\mathbf{m}_n = \alpha_n \mathbf{m}_{ n - 1 } 
	+ ( 1 - \alpha_n ) \br*{ \frac{1}{J_n}\sum_{j = 1}^{J_n}\defaultStochGradient(\Theta_{ n - 1 }, X_{ n, j }) },
	\end{equation}
	\begin{equation}
	\mathbb{M}_n^{(i)}
	= \max\cu*{
		\beta_n\mathbb{M}_{n-1}^{(i)} 
		,
		\abs*{ \frac{1}{J_n}\sum_{j = 1}^{J_n}\defaultStochGradient_i(\Theta_{n-1}, X_{n,j}) }
	},
	\qand
	\end{equation}
	\begin{equation}
	\Theta_n^{ (i) }
	=
	\Theta_{n-1}^{ (i) }
	-
	\br*{
		\varepsilon 
		+
		\mathbb{M}_n^{ (i)} 
	}^{ - 1 }
	\br*{
		\br*{
			\tfrac{
				\gamma_n 
				( 1 - \alpha_n )
			}{
				1 - \prod_{ l = 1 }^n \alpha_l
			}
		}
		\br*{\frac{1}{J_n}\sum_{j = 1}^{J_n}\defaultStochGradient_i(\Theta_{ n - 1 }, X_{ n, j })}
		+
		\br*{
			\tfrac{
				\gamma_{n+1}
				\alpha_{n+1}
			}{
				1 - \prod_{ l = 1 }^{n+1} \alpha_l
			}
		}
		\mathbf{m}_n^{(i)}
	}
	.
	\end{equation}
	}
}

\newcommand{\defdetermAdamLtwo}{
\deterministicGDdef[true]
	{def:determ_adamLtwo}
	{\Adam\ \GD\ optimization method with $L^2$-regularization}
	{ 
	$(\gamma_n)_{n \in \N} \subseteq [0,\infty)$, 
	$(\alpha_n)_{n \in \N} \subseteq [0,1)$, 
	$(\beta_n)_{n \in \N} \subseteq [0,1)$, 
	$\lambda \in \R$, 
	$\varepsilon \in (0,\infty)$, }
	{\Adam\ \GD\ process for the objective function $\defaultLossFunction$ with learning rates $(\gamma_n)_{n \in \N}$, 
	momentum decay factors $(\alpha_n)_{n \in \N}$, 
	second moment decay factors $(\beta_n)_{n \in \N}$, 
	$L^2$-regularization factor $\lambda$,
	regularizing factor $\varepsilon$, 
	and initial value $\xi$}
	{there exist $\mathbf{m}= ( \mathbf{m}^{(1)},\ldots,\mathbf{m}^{(\defaultParamDim)}) \colon \N_0 \to \R^\defaultParamDim$ and $\mathbb{M} = (\mathbb{M}^{(1)}, \ldots, \mathbb{M}^{(\defaultParamDim)}) \colon \N_0 \to \R^\defaultParamDim$ such that for all $n \in \N$, $i \in \{1,2,\ldots,\defaultParamDim\}$ it holds that
	\begin{equation}
	\Theta_0 = \xi, 
	\qquad 
	\mathbf{m}_0 = 0, 
	\qquad 
	\mathbb{M}_0 = 0, 
	\end{equation}
	\begin{equation}
	\mathbf{m}_n = \alpha_n  \mathbf{m}_{ n - 1 } 
	+ ( 1 - \alpha_n ) (\defaultGradientFunction( \Theta_{ n - 1 } ) + \lambda \Theta_{ n - 1 }),
	\end{equation}
	\begin{equation}
	\mathbb{M}_n^{(i)} 
	= \beta_n\mathbb{M}_{n-1}^{(i)} 
	+ (1-\beta_n)
	\abs[\big]{ 
		\defaultGradientFunction_i( \Theta_{n-1} ) 
		+
		\lambda \Theta_{ n - 1 }
	}^2
	,
	\end{equation}
	\begin{equation}
	\andq
	\Theta_n^{ (i) }
	=
	\Theta_{n-1}^{ (i) } -
	\gamma_n 
	{
	\textstyle
	\br*{
	\varepsilon
	+
	\br*{
	\frac{
	\mathbb{M}_n^{ (i) }
	}{
	( 1 - \prod_{ l = 1 }^n \beta_l ) 
	}
	}^{ \nicefrac{ 1 }{ 2 } }
	}^{ - 1 }
	}
	\br*{
	\frac{
	\mathbf{m}_n^{ (i) } 
	}{
	( 1 - \prod_{ l = 1 }^n \alpha_l )
	}
	}
	.
	\end{equation}
	}
}

\newcommand{\iterationStepDetermAdamLtwo}{
	\State
 	$
		\mathbf{m} 
	\assign
		\alpha_n  \mathbf{m} 
		+ 
		( 1 - \alpha_n ) \pr{ \grad( \Theta ) + \lambda \Theta }
	$
	\State
	$
		\mathbb{M}
	\assign
		\beta_n\mathbb{M}
		+ 
		(1-\beta_n)
		\br*{ \grad( \Theta )  + \lambda \Theta }^2
	$
	\State
	$
		\Theta
	\assign
		\Theta 
		-  
		\br*{
		\varepsilon 
		+ 
		\br*{ 
		\frac{
		\mathbb{M}
		}{ 
			1 - \prod_{ k = 1 }^n \beta_k 
		}
		}^{ \nicefrac{ 1 }{ 2 } } 
		}^{ - 1 }
		\br*{
		\frac{
			\gamma_n \mathbf{m} 
		}{
			1 - \prod_{ k = 1 }^n \alpha_k
		}
		}
	$
}
\newcommand{\algDescrDetermAdamLtwo}{\algorithmicDescription{\Adam\ \GD\ optimization method with $L^2$-regularization}{def:determ_adamLtwo}{ 
	$ ( \gamma_n )_{ n \in \N } \subseteq [0,\infty) $, 
	$ ( \alpha_n )_{ n \in \N } \subseteq [0,1) $, 
	$ ( \beta_n )_{ n \in \N } \subseteq [0,1) $,
	$ \lambda \in \R $,
	$ \varepsilon \in (0,\infty) $, 
}{
	\Adam\ \GD\ process 
	for the objective function $ \defaultLossFunction $ 
	with learning rates $ ( \gamma_n )_{ n \in \N } $, 
	momentum decay factors $ ( \alpha_n )_{ n \in \N } $, 
	second moment decay factors $ ( \beta_n )_{ n \in \N } $, 
	$L^2$-regularization factor $\lambda$,
	regularizing factor $ \varepsilon $, 
	and initial value $ \xi $
}{
	$\Theta \assign \xi$\separator 
	$\mathbf{m} \assign 0 \in \R^\defaultParamDim$\separator
	$\mathbb{M} \assign 0 \in \R^\defaultParamDim$
}{
 	\iterationStepDetermAdamLtwo
}
}

\newcommand{\iterationStepAdamLtwo}{
	\pseudoCodeGradientAssign
	\State
 	$
		\mathbf{m} 
	\assign
		\alpha_n  \mathbf{m} 
		+ 
		( 1 - \alpha_n ) \pr{ \pseudoCodeGradient + \lambda \Theta }
	$
	\State
	$
		\mathbb{M}
	\assign
		\beta_n\mathbb{M}
		+ 
		(1-\beta_n)
		\br*{ \pseudoCodeGradient  + \lambda \Theta }^2
	$
	\State
	$
		\Theta
	\assign
		\Theta 
		-  
		\br*{
		\varepsilon 
		+ 
		\br*{ 
		\frac{
		\mathbb{M}
		}{ 
			1 - \prod_{ k = 1 }^n \beta_k 
		}
		}^{ \nicefrac{ 1 }{ 2 } } 
		}^{ - 1 }
		\br*{
		\frac{
			\gamma_n \mathbf{m} 
		}{
			1 - \prod_{ k = 1 }^n \alpha_k
		}
		}
	$
}
\newcommand{\algDescrAdamLtwo}{\algorithmicDescriptionStochastic{\Adam\ \SGD\ optimization method with $L^2$-regularization}{def:adamLtwo}{ 
	$ ( \gamma_n )_{ n \in \N } \subseteq [0,\infty) $, 
	$ ( \alpha_n )_{ n \in \N } \subseteq [0,1) $, 
	$ ( \beta_n )_{ n \in \N } \subseteq [0,1) $,
	$ \lambda \in \R $,
	$ \varepsilon \in (0,\infty) $, 
}{
	\Adam\ \SGD\ process 
	for the loss function $\defaultStochLoss$ 
	with learning rates $ ( \gamma_n )_{ n \in \N } $, 
	momentum decay factors $ ( \alpha_n )_{ n \in \N } $, 
	second moment decay factors $ ( \beta_n )_{ n \in \N } $, 
	$L^2$-regularization factor $\lambda$,
	regularizing factor $ \varepsilon $, 
	initial value $ \xi $,
}{
	$\Theta \assign \xi$\separator 
	$\mathbf{m} \assign 0 \in \R^\defaultParamDim$\separator
	$\mathbb{M} \assign 0 \in \R^\defaultParamDim$
}{
 	\iterationStepAdamLtwo
}
}

\newcommand{\defAdamLtwo}{
\SGDdef[true]
	{def:adamLtwo}
	{\Adam\ \SGD\ optimization method with $L^2$-regularization}
	{ 
	$(\gamma_n)_{n \in \N} \subseteq [0,\infty)$, 
	$(\alpha_n)_{n \in \N} \subseteq [0,1)$, 
	$(\beta_n)_{n \in \N} \subseteq [0,1)$, 
	$\lambda \in \R$, 
	$\varepsilon \in (0,\infty)$,}
	{\Adam\ \SGD\ process for the loss function $\defaultStochLoss$ with 
	learning rates $(\gamma_n)_{n \in \N}$, 
	momentum decay factors $(\alpha_n)_{n \in \N}$, 
	second moment decay factors $(\beta_n)_{n \in \N}$, 
	$L^2$-regularization factor $\lambda$,
	regularizing factor $\varepsilon$, 
	initial value $\xi$,}
	{there exist 
		$\mathbf{m} = (\mathbf{m}^{(1)},\ldots,\mathbf{m}^{(\defaultParamDim)}) \colon \N_0\times \Omega \to \R^\defaultParamDim$
		and
		$\mathbb{M} = (\mathbb{M}^{(1)},\ldots,\mathbb{M}^{(\defaultParamDim)}) \colon \N_0\times \Omega \to \R^\defaultParamDim$
	such that for all $n \in \N$, $i \in \{1,2,\ldots,\defaultParamDim\}$ it holds that
	\begin{equation}
	\Theta_0 = \xi, 
	\qquad 
	\mathbf{m}_0 = 0, 
	\qquad 
	\mathbb{M}_0 = 0, 
	\end{equation}
	\begin{equation}
	\mathbf{m}_n = \alpha_n \mathbf{m}_{ n - 1 } 
	+ ( 1 - \alpha_n ) \br*{\lambda \Theta_{ n - 1 } + \frac{1}{J_n}\sum_{j = 1}^{J_n}\defaultStochGradient(\Theta_{ n - 1 }, X_{ n, j }) },
	\end{equation}
	\begin{equation}
	\mathbb{M}_n^{(i)}
	= \beta_n\mathbb{M}_{n-1}^{(i)}
	+ (1-\beta_n)
	\br*{ 
		\lambda \Theta_{ n - 1 }^{ (i) }
		+
		\frac{1}{J_n}\sum_{j = 1}^{J_n}\defaultStochGradient_i(\Theta_{n-1}, X_{n,j})
	}^2
	,
	\end{equation}
	\begin{equation}
	\Theta_n^{ (i) }
	=
	\Theta_{n-1}^{ (i) }
	-
	\gamma_n 
	{
	\textstyle
	\br*{
	\varepsilon
	+
	\br*{
	\frac{
	\mathbb{M}_n^{ (i) }
	}{
	( 1 - \prod_{ l = 1 }^n \beta_l ) 
	}
	}^{ \nicefrac{ 1 }{ 2 } }
	}^{ - 1 }
	}
	\br*{
	\frac{
	\mathbf{m}_n^{ (i) } 
	}{
	( 1 - \prod_{ l = 1 }^n \alpha_l )
	}
	}
	.
	\end{equation}
	}
}

\newcommand{\defdetermAdamW}{
\deterministicGDdef[true]
	{def:determ_adamW}
	{\AdamW\ \GD\ optimization method}
	{ 
	$(\gamma_n)_{n \in \N} \subseteq [0,\infty)$, 
	$(\alpha_n)_{n \in \N} \subseteq [0,1)$, 
	$(\beta_n)_{n \in \N} \subseteq [0,1)$, 
	$\lambda \in \R$, 
	$\varepsilon \in (0,\infty)$, }
	{\AdamW\ \GD\ process for the objective function $\defaultLossFunction$ with learning rates $(\gamma_n)_{n \in \N}$, 
	momentum decay factors $(\alpha_n)_{n \in \N}$, 
	second moment decay factors $(\beta_n)_{n \in \N}$, 
	weight decay factor $\lambda$,
	regularizing factor $\varepsilon$, 
	and initial value $\xi$}
	{there exist $\mathbf{m}= ( \mathbf{m}^{(1)},\ldots,\mathbf{m}^{(\defaultParamDim)}) \colon \N_0 \to \R^\defaultParamDim$ and $\mathbb{M} = (\mathbb{M}^{(1)}, \ldots, \mathbb{M}^{(\defaultParamDim)}) \colon \N_0 \to \R^\defaultParamDim$ such that for all $n \in \N$, $i \in \{1,2,\ldots,\defaultParamDim\}$ it holds that
	\begin{equation}
	\Theta_0 = \xi, 
	\qquad 
	\mathbf{m}_0 = 0, 
	\qquad 
	\mathbb{M}_0 = 0, 
	\end{equation}
	\begin{equation}
	\mathbf{m}_n = \alpha_n  \mathbf{m}_{ n - 1 } 
	+ ( 1 - \alpha_n ) \defaultGradientFunction( \Theta_{ n - 1 } ),
	\end{equation}
	\begin{equation}
	\mathbb{M}_n^{(i)} 
	= \beta_n\mathbb{M}_{n-1}^{(i)} 
	+ (1-\beta_n)
	\abs[\big]{ 
		\defaultGradientFunction_i( \Theta_{n-1} ) 
	}^2
	, \qand
	\end{equation}
	\begin{equation}
		\Theta_n^{ (i) } 
		= 
		\Theta_{n-1}^{ (i) } - 
		\gamma_n 
		\pr*{
		  \textstyle 
		  \br*{
			\varepsilon 
			+ 
			\br*{ 
			  \frac{
				\mathbb{M}_n^{ (i)} 
			  }{ 
				( 1 - \prod_{ l = 1 }^n \beta_l ) 
			  }
			}^{ \nicefrac{ 1 }{ 2 } } 
		  }^{ - 1 }
		\br*{
		  \frac{
			\mathbf{m}_n^{ (i) } 
		  }{
			( 1 - \prod_{ l = 1 }^n \alpha_l )
		  }
		}
		+
		\lambda  \Theta_{n-1}^{ (i) }
		}
		.
	  \end{equation}
	}
}

\newcommand{\iterationStepDetermAdamW}{
	\State
 	$
		\mathbf{m} 
	\assign
		\alpha_n  \mathbf{m} 
		+ 
		( 1 - \alpha_n ) \grad( \Theta )
	$
	\State
	$
		\mathbb{M}
	\assign
		\beta_n\mathbb{M}
		+ 
		(1-\beta_n)
		\br*{ \grad( \Theta ) }^2
	$
	\State
	$
		\Theta
	\assign
		\Theta 
		- 
		\gamma_n 
		\pr*{
			\br*{
			\varepsilon 
			+ 
			\br*{ 
			\frac{
			\mathbb{M}
			}{ 
				1 - \prod_{ k = 1 }^n \beta_k
			}
			}^{ \nicefrac{ 1 }{ 2 } } 
			}^{ - 1 }
			\br*{
			\frac{
			\mathbf{m}
			}{
				1 - \prod_{ k = 1 }^n \alpha_k
			}
			}
			+
			\lambda  \Theta
		}
	$
}
\newcommand{\algDescrDetermAdamW}{\algorithmicDescription{\AdamW\ \GD\ optimization method}{def:determ_adamW}{ 
	$ ( \gamma_n )_{ n \in \N } \subseteq [0,\infty) $, 
	$ ( \alpha_n )_{ n \in \N } \subseteq [0,1) $, 
	$ ( \beta_n )_{ n \in \N } \subseteq [0,1) $,
	$ \lambda \in \R $,
	$ \varepsilon \in (0,\infty) $, 
	$ \xi \in \R^\defaultParamDim$,
}{
	\AdamW\ \GD\ process 
	for the objective function $ \defaultLossFunction $ 
	with learning rates $ ( \gamma_n )_{ n \in \N } $, 
	momentum decay factors $ ( \alpha_n )_{ n \in \N } $, 
	second moment decay factors $ ( \beta_n )_{ n \in \N } $, 
	weight decay factor $\lambda$,
	regularizing factor $ \varepsilon $, 
	and initial value $ \xi $
}{
	$\Theta \assign \xi$\separator 
	$\mathbf{m} \assign 0 \in \R^\defaultParamDim$\separator
	$\mathbb{M} \assign 0 \in \R^\defaultParamDim$
}{
 	\iterationStepDetermAdamW
}
}

\newcommand{\iterationStepAdamW}{
	\pseudoCodeGradientAssign
	\State
 	$
		\mathbf{m} 
	\assign
		\alpha_n  \mathbf{m} 
		+ 
		( 1 - \alpha_n ) \pseudoCodeGradient
	$
	\State
	$
		\mathbb{M}
	\assign
		\beta_n\mathbb{M}
		+ 
		(1-\beta_n)
		\pseudoCodeGradient^2
	$
	\State
	$
		\Theta
	\assign
		\Theta 
		- 
		\gamma_n 
		\pr*{
			\br*{
			\varepsilon 
			+ 
			\br*{ 
			\frac{
			\mathbb{M}
			}{ 
				1 - \prod_{ k = 1 }^n \beta_k
			}
			}^{ \nicefrac{ 1 }{ 2 } } 
			}^{ - 1 }
			\br*{
			\frac{
			\mathbf{m}
			}{
				1 - \prod_{ k = 1 }^n \alpha_k
			}
			}
			+
			\lambda  \Theta
		}
	$
}
\newcommand{\algDescrAdamW}{\algorithmicDescriptionStochastic{\AdamW\ \SGD\ optimization method}{def:adamW}{ 
	$ ( \gamma_n )_{ n \in \N } \subseteq [0,\infty) $, 
	$ ( \alpha_n )_{ n \in \N } \subseteq [0,1) $, 
	$ ( \beta_n )_{ n \in \N } \subseteq [0,1) $,
	$ \lambda \in \R $,
	$ \varepsilon \in (0,\infty) $,
}{
	\AdamW\ \SGD\ process 
	for the loss function $\defaultStochLoss$ 
	with learning rates $ ( \gamma_n )_{ n \in \N } $, 
	momentum decay factors $ ( \alpha_n )_{ n \in \N } $, 
	second moment decay factors $ ( \beta_n )_{ n \in \N } $, 
	weight decay factor $\lambda$,
	regularizing factor $ \varepsilon $, 
	initial value $ \xi $,
}{
	$\Theta \assign \xi$\separator 
	$\mathbf{m} \assign 0 \in \R^\defaultParamDim$\separator
	$\mathbb{M} \assign 0 \in \R^\defaultParamDim$
}{
 	\iterationStepAdamW
}
}

\newcommand{\defAdamW}{
\SGDdef[true]
	{def:adamW}
	{\AdamW\ \SGD\ optimization method}
	{ 
	$(\gamma_n)_{n \in \N} \subseteq [0,\infty)$, 
	$(\alpha_n)_{n \in \N} \subseteq [0,1)$, 
	$(\beta_n)_{n \in \N} \subseteq [0,1)$, 
	$\lambda \in \R$, 
	$\varepsilon \in (0,\infty)$,}
	{\AdamW\ \SGD\ process for the loss function $\defaultStochLoss$ with 
	learning rates $(\gamma_n)_{n \in \N}$, 
	momentum decay factors $(\alpha_n)_{n \in \N}$, 
	second moment decay factors $(\beta_n)_{n \in \N}$, 
	weight decay factor $\lambda$,
	regularizing factor $\varepsilon$, 
	initial value $\xi$,}
	{there exist 
		$\mathbf{m} = (\mathbf{m}^{(1)},\ldots,\mathbf{m}^{(\defaultParamDim)}) \colon \N_0\times \Omega \to \R^\defaultParamDim$
		and
		$\mathbb{M} = (\mathbb{M}^{(1)},\ldots,\mathbb{M}^{(\defaultParamDim)}) \colon \N_0\times \Omega \to \R^\defaultParamDim$
	such that for all $n \in \N$, $i \in \{1,2,\ldots,\defaultParamDim\}$ it holds that
	\begin{equation}
	\Theta_0 = \xi, 
	\qquad 
	\mathbf{m}_0 = 0, 
	\qquad 
	\mathbb{M}_0 = 0, 
	\end{equation}
	\begin{equation}
	\mathbf{m}_n = \alpha_n \mathbf{m}_{ n - 1 } 
	+ ( 1 - \alpha_n ) \br*{\frac{1}{J_n}\sum_{j = 1}^{J_n}\defaultStochGradient(\Theta_{ n - 1 }, X_{ n, j }) },
	\end{equation}
	\begin{equation}
	\mathbb{M}_n^{(i)}
	= \beta_n\mathbb{M}_{n-1}^{(i)}
	+ (1-\beta_n)
	\br*{ 
		\frac{1}{J_n}\sum_{j = 1}^{J_n}\defaultStochGradient_i(\Theta_{n-1}, X_{n,j})
	}^2
	,
	\qand
	\end{equation}
	\begin{equation}
		\Theta_n^{ (i) } 
		= 
		\Theta_{n-1}^{ (i) } - 
		\gamma_n 
		\pr*{
		  \textstyle 
		  \br*{
			\varepsilon 
			+ 
			\br*{ 
			  \frac{
				\mathbb{M}_n^{ (i)} 
			  }{ 
				( 1 - \prod_{ l = 1 }^n \beta_l ) 
			  }
			}^{ \nicefrac{ 1 }{ 2 } } 
		  }^{ - 1 }
		\br*{
		  \frac{
			\mathbf{m}_n^{ (i) } 
		  }{
			( 1 - \prod_{ l = 1 }^n \alpha_l )
		  }
		}
		+
		\lambda  \Theta_{n-1}^{ (i) }
		}
		.
	\end{equation}
	}
}

\newcommand{\defdetermShampoo}{
\deterministicGDdef[false][][true]
	{def:determ_Shampoo}
	{Shampoo \GD\ optimization method}
	{
	$(\gamma_n)_{n \in \N} \subseteq [0,\infty)$,
	$\varepsilon \in (0,\infty)$,
	}
	{Shampoo \GD\ process for the objective function $\defaultLossFunction$ with learning rates $(\gamma_n)_{n \in \N}$,
	regularizing factor $\varepsilon$,
	and initial value $\xi$}
	{there exist 
		$\mathbf{L} \colon \N_0 \to \R^{\defaultParamDim_1 \times \defaultParamDim_1}$ and
		$\mathbf{R} \colon \N_0 \to \R^{\defaultParamDim_2 \times \defaultParamDim_2}$
	such that for all $n \in \N$ it holds that
	\begin{equation}
	\label{def:determ_Shampoo:eq1}
	\Theta_0 = \xi, \qquad \mathbf{L}_0 = \varepsilon \idMatrix_{\defaultParamDim_1}, \qquad \mathbf{R}_0 = \varepsilon \idMatrix_{\defaultParamDim_2},
	\end{equation}
	\begin{equation}
	\label{def:determ_Shampoo:eq2}
		\mathbf{L}_n = \mathbf{L}_{n-1} + \defaultGradientFunction(\Theta_{n-1}) (\defaultGradientFunction(\Theta_{n-1}))^\ast,		
	\end{equation}
	\begin{equation}
	\label{def:determ_Shampoo:eq3}
	\begin{split} 
		\mathbf{R}_n = \mathbf{R}_{n-1} + (\defaultGradientFunction(\Theta_{n-1}))^\ast \defaultGradientFunction(\Theta_{n-1}),
	\end{split}
	\end{equation}
	\begin{equation}
	\label{def:determ_Shampoo:eq4}
	\andq \Theta_n = \Theta_{n-1} - \gamma_n  (\mathbf{L}_n)^{-1/4} \defaultGradientFunction(\Theta_{n-1}) (\mathbf{R}_n)^{-1/4}
	\end{equation}
	\cfadd{lemma:sums_of_positive_definite_2}\cfload.
	}
}

\newcommand{\iterationStepDetermShampoo}{
	\State 
	$
		\mathbf{L} \assign \mathbf{L} + \grad(\Theta) (\grad(\Theta))^\ast
	$
	\State
	$
		\mathbf{R} \assign \mathbf{R} + (\grad(\Theta))^\ast \grad(\Theta)
	$	
	\State
	$
		\Theta \assign \Theta - \gamma_n  (\mathbf{L})^{-1/4} \grad(\Theta) (\mathbf{R})^{-1/4}
	$
}
\newcommand{\algDescrDetermShampoo}{\algorithmicDescription[false][true]{Shampoo \GD\ optimization method}{def:determ_Shampoo}
	{ 
	$(\gamma_n)_{n \in \N} \subseteq [0,\infty)$,
	$\varepsilon \in (0,\infty)$,
   	}
{
	Shampoo \GD\ process for the objective function $\defaultLossFunction$ with learning rates $(\gamma_n)_{n \in \N}$,
	regularizing factor $\varepsilon$,
	and initial value $\xi$
}{
	$\Theta \assign \xi$\separator $\mathbf{L} \assign \varepsilon \idMatrix_{\defaultParamDim_1}$\separator $\mathbf{R} \assign \varepsilon \idMatrix_{\defaultParamDim_2}$
}{
	\iterationStepDetermShampoo
}
}

\newcommand{\iterationStepShampoo}{
	\pseudoCodeGradientAssign
	\State 
	$
		\mathbf{L} \assign \mathbf{L} + \pseudoCodeGradient \pseudoCodeGradient^\ast
	$
	\State
	$
		\mathbf{R} \assign \mathbf{R} + \pseudoCodeGradient^\ast \pseudoCodeGradient
	$	
	\State
	$
		\Theta \assign \Theta - \gamma_n  (\mathbf{L})^{-1/4} \pseudoCodeGradient (\mathbf{R})^{-1/4}
	$
}

\newcommand{\algDescrShampoo}{\algorithmicDescriptionStochastic[false][true]
	{Shampoo \SGD\ optimization method}
	{def:Shampoo}
	{ 
	$(\gamma_n)_{n \in \N} \subseteq [0,\infty)$,
	$\varepsilon \in (0,\infty)$,}
	{
	Shampoo \SGD\ process for the objective function $\defaultLossFunction$ with learning rates $(\gamma_n)_{n \in \N}$,
	regularizing factor $\varepsilon$,
	initial value $\xi$,
}{
	$\Theta \assign \xi$\separator $\mathbf{L} \assign \varepsilon \idMatrix_{\defaultParamDim_1}$\separator $\mathbf{R} \assign \varepsilon \idMatrix_{\defaultParamDim_2}$
}{
	\iterationStepShampoo
}
}

\newcommand{\defShampoo}{
\SGDdef[false][][true]
	{def:Shampoo}
	{Shampoo \SGD\ optimization method}
	{ 
	$(\gamma_n)_{n \in \N} \subseteq [0,\infty)$, 
	$\varepsilon \in (0,\infty)$,
	}
	{Shampoo \SGD\ process for the objective function $\defaultLossFunction$ with learning rates $(\gamma_n)_{n \in \N}$,
	regularizing factor $\varepsilon$,
	initial value $\xi$,}
	{there exist 
		$\mathbf{L} \colon \N_0 \times \Omega \to \R^{\defaultParamDim_1 \times \defaultParamDim_1}$ and
		$\mathbf{R} \colon \N_0 \times \Omega \to \R^{\defaultParamDim_2 \times \defaultParamDim_2}$
	such that for all $n \in \N$ it holds that
	\begin{equation}
	\label{def:Shampoo:eq1}
	\Theta_0 = \xi, \qquad \mathbf{L}_0 = \varepsilon \idMatrix_{\defaultParamDim_1}, \qquad \mathbf{R}_0 = \varepsilon \idMatrix_{\defaultParamDim_2},
	\end{equation}
	\begin{equation}
	\label{def:Shampoo:eq2}
		\mathbf{L}_n = \mathbf{L}_{n-1} + \br*{\frac{1}{J_n}\sum_{j = 1}^{J_n}\defaultStochGradient(\Theta_{ n - 1 }, X_{ n, j }) } \br*{\frac{1}{J_n}\sum_{j = 1}^{J_n}\defaultStochGradient(\Theta_{ n - 1 }, X_{ n, j }) }^\ast,		
	\end{equation}
	\begin{equation}
	\label{def:Shampoo:eq3}
	\begin{split} 
		\mathbf{R}_n = \mathbf{R}_{n-1} + \br*{\frac{1}{J_n}\sum_{j = 1}^{J_n}\defaultStochGradient(\Theta_{ n - 1 }, X_{ n, j }) }^\ast \br*{\frac{1}{J_n}\sum_{j = 1}^{J_n}\defaultStochGradient(\Theta_{ n - 1 }, X_{ n, j }) },
	\end{split}
	\end{equation}
	\begin{equation}
	\label{def:Shampoo:eq4}
	\andq \Theta_n = \Theta_{n-1} - \gamma_n  (\mathbf{L}_n)^{-1/4} \br*{\frac{1}{J_n}\sum_{j = 1}^{J_n}\defaultStochGradient(\Theta_{ n - 1 }, X_{ n, j }) } (\mathbf{R}_n)^{-1/4}
	\end{equation}
	\cfadd{lemma:sums_of_positive_definite_2}\cfload.
	}
}

\newcommand{\MuonProjection}{\Pi}
\newcommand{\MuonProjImg}{\mathcal{O}}

\newcommand{\defdetermidealMuon}{
\deterministicGDdef[false][][true]
	[$ 
		\MuonProjImg 
	= 
		\{
			O \in \R^{\defaultParamDim_1 \times \defaultParamDim_2} \colon $ $
				\pr*{
					(O O^\ast = \idMatrix_{\defaultParamDim_1}) \lor   (O^\ast O = \idMatrix_{\defaultParamDim_2})
				}
		\}
	$,
	let $\MuonProjection \colon \R^{\defaultParamDim_1 \times \defaultParamDim_2} \to \MuonProjImg$ satisfy for all $A \in \R^{\defaultParamDim_1 \times \defaultParamDim_2}$ that
		\begin{equation}
		\label{def:determ_ideal_Muon:eq1}
			\hilbertSchmidtNorm{\MuonProjection(A) - A} 
		=
			\inf_{O \in \MuonProjImg}
				\hilbertSchmidtNorm{O - A},
		\end{equation}]
	{def:determ_ideal_Muon}
	{\Muon\ \GD\ optimization method}
	{
	$(\gamma_n)_{n \in \N} \subseteq [0,\infty)$,
	$(\alpha_n)_{n \in \N} \subseteq [0,\infty)$,
	}
	{\Muon\ \GD\ process for the objective function $\defaultLossFunction$ with learning rates $(\gamma_n)_{n \in \N}$,
	momentum decay factors $(\alpha_n)_{n \in \N}$,
	orthogonal projection $\MuonProjection$,
	and initial value $\xi$}
	{there exists $\mathbf{m} \colon \N_0 \to \R^{\defaultParamDim_1 \times \defaultParamDim_2}$ such that for all $n \in \N$ it holds that
	\begin{equation}
	\label{eq:def:idealMuon_1}
	\Theta_0 = \xi, \qquad \mathbf{m}_0 = 0,
	\end{equation}
	\begin{equation}
	\label{eq:def:idealMuon_2}
	\mathbf{m}_n = \alpha_n \mathbf{m}_{n-1} + \defaultGradientFunction(\Theta_{n-1}),
	\end{equation}
	\begin{equation}
	\label{eq:def:idealMuon_3}
	\andq \Theta_n = \Theta_{n-1} - \gamma_n  \MuonProjection(\mathbf{m}_n).
	\end{equation}
	}
}

\newcommand{\iterationStepDetermidealMuon}{
	\State 
	$
		\mathbf{m} \assign \alpha_n \mathbf{m} + \grad(\Theta)
	$
	\State
	$
		\Theta \assign \Theta - \gamma_n  \MuonProjection(\mathbf{m})
	$
}
\newcommand{\algDescrDetermidealMuon}{\algorithmicDescription[false][true]
[$\MuonProjection \colon \R^{\defaultParamDim_1 \times \defaultParamDim_2} \to 
	\cu*{
			O \in \R^{\defaultParamDim_1 \times \defaultParamDim_2} \colon
				\pr*{
					(O O^\ast = \idMatrix_{\defaultParamDim_1}) \lor   (O^\ast O = \idMatrix_{\defaultParamDim_2})
				}
		}
	$ with 
	$\forall \, A \in \R^{\defaultParamDim_1 \times \defaultParamDim_2} \colon \hilbertSchmidtNorm{\MuonProjection(A) - A} 
	=
		\inf_{\mathcal{O} \in \cu*{
			O \in \R^{\defaultParamDim_1 \times \defaultParamDim_2} \colon
				\pr*{
					(O O^\ast = \idMatrix_{\defaultParamDim_1}) \lor   (O^\ast O = \idMatrix_{\defaultParamDim_2})
				}
		}}
			\hilbertSchmidtNorm{\mathcal{O} - A}
	$
]{Idealized \Muon\ \GD\ optimization method}{def:determ_ideal_Muon}{
	$(\gamma_n)_{n \in \N} \subseteq [0,\infty)$,
	$(\alpha_n)_{n \in \N} \subseteq [0,\infty)$,
}{
	idealized \Muon\ \GD\ process for the objective function $\defaultLossFunction$ with learning rates $(\gamma_n)_{n \in \N}$,
	momentum decay factors $(\alpha_n)_{n \in \N}$,
	orthogonal projection $\MuonProjection$,
	and initial value $\xi$
}{
	$\Theta \assign \xi$\separator $\mathbf{m} \assign 0 \in \R^\defaultParamDim$
}{
	\iterationStepDetermidealMuon
}
}

\newcommand{\iterationStepidealMuon}{
	\pseudoCodeGradientAssign
	\State 
	$
		\mathbf{m} \assign \alpha_n \mathbf{m} + \pseudoCodeGradient
	$
	\State
	$
		\Theta \assign \Theta - \gamma_n  \MuonProjection(\mathbf{m})
	$
}
\newcommand{\algDescridealMuon}{\algorithmicDescriptionStochastic[false][true]
[$\MuonProjection \colon \R^{\defaultParamDim_1 \times \defaultParamDim_2} \to 
	\cu*{
			O \in \R^{\defaultParamDim_1 \times \defaultParamDim_2} \colon
				\pr*{
					(O O^\ast = \idMatrix_{\defaultParamDim_1}) \lor   (O^\ast O = \idMatrix_{\defaultParamDim_2})
				}
		}
	$ with 
	$\forall \, A \in \R^{\defaultParamDim_1 \times \defaultParamDim_2} \colon \hilbertSchmidtNorm{\MuonProjection(A) - A} 
	=
		\inf_{\mathcal{O} \in \cu*{
			O \in \R^{\defaultParamDim_1 \times \defaultParamDim_2} \colon
				\pr*{
					(O O^\ast = \idMatrix_{\defaultParamDim_1}) \lor   (O^\ast O = \idMatrix_{\defaultParamDim_2})
				}
		}}
			\hilbertSchmidtNorm{\mathcal{O} - A}
	$
]{Idealized \Muon\ \SGD\ optimization method}{def:idealMuon}{ 
	$(\gamma_n)_{n \in \N} \subseteq [0,\infty)$,
	$(\alpha_n)_{n \in \N} \subseteq [0,\infty)$,
}{
	idealized \Muon\ \SGD\ process for the loss function $\defaultStochLoss$ with learning rates $(\gamma_n)_{n \in \N}$, 
	momentum decay factors $(\alpha_n)_{n \in \N}$, 
	orthogonal projection $\MuonProjection$,
	initial value $\xi$,
}{
	$\Theta \assign \xi$\separator $\mathbf{m} \assign 0 \in \R^\defaultParamDim$
}{
	\iterationStepidealMuon
}
}

\newcommand{\defidealMuon}{
\SGDdef[false][][muon]
	[$ 
		\MuonProjImg 
	=  \allowbreak
		\{
			O \in \R^{\defaultParamDim_1 \times \defaultParamDim_2} \colon
				\pr{
					(O O^\ast = \idMatrix_{\defaultParamDim_1}) $ $ \lor   (O^\ast O = \idMatrix_{\defaultParamDim_2})
				}
		\}
	$,
	let $\MuonProjection \colon \R^{\defaultParamDim_1 \times \defaultParamDim_2} \to \MuonProjImg$ satisfy for all $A \in \R^{\defaultParamDim_1 \times \defaultParamDim_2}$ that
		\begin{equation}
		\label{def:ideal_Muon:eq1}
			\hilbertSchmidtNorm{\MuonProjection(A) - A} 
		=
			\inf_{O \in \MuonProjImg}
				\hilbertSchmidtNorm{O - A},
		\end{equation}]
	{def:idealMuon}
	{Idealized \Muon\ \SGD\ optimization method}
	{
	$(\gamma_n)_{n \in \N} \subseteq [0,\infty)$,
	$(\alpha_n)_{n \in \N} \subseteq [0,\infty)$,}
	{idealized \Muon\ \SGD\ process for the loss function $\defaultStochLoss$ with learning rates $(\gamma_n)_{n \in \N}$,
	momentum decay factors $(\alpha_n)_{n \in \N}$,
	orthogonal projection $\MuonProjection$,
	initial value $\xi$,}
	{there exists $\mathbf{m} \colon \N_0 \to \R^\defaultParamDim$ such that for all $n \in \N$ it holds that
		it holds for all $n \in \N$ that 
		\begin{equation}
		\Theta_0 = \xi, \qquad \mathbf{m}_0 = 0,
		\end{equation}
		\begin{equation}
		\label{def:ideal_Muon:eq1}
		  \mathbf{m}_n = \alpha_n \mathbf{m}_{n-1} + \br*{\frac{1}{J_n}\sum_{j = 1}^{J_n}\defaultStochGradient(\Theta_{n-1}, X_{n,j}) }, 
		\end{equation}
		\begin{equation}
		  \andq 
		  \Theta_n = \Theta_{n-1} - \gamma_n  \MuonProjection(\mathbf{m}_n) . 
		\end{equation}
	}
}

\newcommand{\NewtonSchulzAlgorithm}{\cfadd{def:NewtonSchulzAlgorithm}\operatorname{NS}}

\newcommand{\defdetermMuon}{
\deterministicGDdef[false][][muon]
	{def:determ_Muon}
	{\Muon\ \GD\ optimization method}
	{
	$(\gamma_n)_{n \in \N} \subseteq [0,\infty)$,
	$(\alpha_n)_{n \in \N} \subseteq [0,\infty)$,
	$a, b, c \in \R$, $\varepsilon \in (0,\infty)$,
	}
	{\Muon\ \GD\ process for the objective function $\defaultLossFunction$ with learning rates $(\gamma_n)_{n \in \N}$,
	momentum decay factors $(\alpha_n)_{n \in \N}$,
	Newton-Schulz method with polynomial coefficients $a, b, c$, regularization parameter $\varepsilon$, and $N$ iterations,
	and initial value $\xi$}
	{there exists $\mathbf{m} \colon \N_0 \to \R^{\defaultParamDim_1 \times \defaultParamDim_2}$ such that for all $n \in \N$ it holds that
	\begin{equation}
	\label{eq:def:Muon_1}
	\Theta_0 = \xi, \qquad \mathbf{m}_0 = 0,
	\end{equation}
	\begin{equation}
	\label{eq:def:Muon_2}
	\mathbf{m}_n = \alpha_n \mathbf{m}_{n-1} + \defaultGradientFunction(\Theta_{n-1}),
	\end{equation}
	\begin{equation}
	\label{eq:def:Muon_3}
	\andq \Theta_n = \Theta_{n-1} - \gamma_n  \NewtonSchulzAlgorithm_{a, b, c, \varepsilon}(\mathbf{m}_n, K)
	\end{equation}
	\cfload.
	}
}

\newcommand{\iterationStepDetermMuon}{
	\State 
	$
		\mathbf{m} \assign \alpha_n \mathbf{m} + \grad(\Theta)
	$
	\State
	$
		\Theta \assign \Theta - \gamma_n  \NewtonSchulzAlgorithm_{a, b, c, \varepsilon}(\mathbf{m}, K)
	$
}
\newcommand{\algDescrDetermMuon}{\algorithmicDescription[false][muon]{\Muon\ \GD\ optimization method}{def:determ_Muon}{
	$(\gamma_n)_{n \in \N} \subseteq [0,\infty)$,
	$(\alpha_n)_{n \in \N} \subseteq [0,\infty)$,
	$a, b, c \in \R$, $\varepsilon \in (0,\infty)$,
}{
	\Muon\ \GD\ process for the objective function $\defaultLossFunction$ with learning rates $(\gamma_n)_{n \in \N}$,
	momentum decay factors $(\alpha_n)_{n \in \N}$,
	Newton-Schulz method with polynomial coefficients $a, b, c$, regularization parameter $\varepsilon$, and $K$ iterations,
	and initial value $\xi$
}{
	$\Theta \assign \xi$\separator $\mathbf{m} \assign 0 \in \R^\defaultParamDim$
}{
	\iterationStepDetermMuon
}
}

\newcommand{\iterationStepMuon}{
	\pseudoCodeGradientAssign
	\State 
	$
		\mathbf{m} \assign \alpha_n \mathbf{m} + \pseudoCodeGradient
	$
	\State
	$
		\Theta \assign \Theta - \gamma_n  \NewtonSchulzAlgorithm_{a, b, c, \varepsilon}(\mathbf{m}, K)
	$
}
\newcommand{\algDescrMuon}{\algorithmicDescriptionStochastic[false][muon]{\Muon\ \SGD\ optimization method}{def:Muon}{ 
	$(\gamma_n)_{n \in \N} \subseteq [0,\infty)$,
	$(\alpha_n)_{n \in \N} \subseteq [0,\infty)$,
	$a, b, c \in \R$, $\varepsilon \in (0,\infty)$,
}{
	\Muon\ \SGD\ process for the loss function $\defaultStochLoss$ with learning rates $(\gamma_n)_{n \in \N}$, 
	momentum decay factors $(\alpha_n)_{n \in \N}$, 
	Newton-Schulz method with polynomial coefficients $a, b, c$, regularization parameter $\varepsilon$, and $K$ iterations,
	initial value $\xi$,
}{
	$\Theta \assign \xi$\separator $\mathbf{m} \assign 0 \in \R^\defaultParamDim$
}{
	\iterationStepMuon
}
}

\newcommand{\defMuon}{
\SGDdef[false][][muon][]
	{def:Muon}
	{\Muon\ \SGD\ optimization method}
	{
	$(\gamma_n)_{n \in \N} \subseteq [0,\infty)$,
	$(\alpha_n)_{n \in \N} \subseteq [0,\infty)$,
	$a, b, c \in \R$, $\varepsilon \in (0,\infty)$,
	}
	{\Muon\ \SGD\ process for the loss function $\defaultStochLoss$ with learning rates $(\gamma_n)_{n \in \N}$,
	momentum decay factors $(\alpha_n)_{n \in \N}$,
	Newton-Schulz method with polynomial coefficients $a, b, c$, regularization parameter $\varepsilon$, and $K$ iterations,
	initial value $\xi$,}
	{there exists $\mathbf{m} \colon \N_0 \to \R^\defaultParamDim$ such that for all $n \in \N$ it holds that
		it holds for all $n \in \N$ that 
		\begin{equation}
		\Theta_0 = \xi, \qquad \mathbf{m}_0 = 0,
		\end{equation}
		\begin{equation}
		\label{def:Muon:eq1}
		  \mathbf{m}_n = \alpha_n \mathbf{m}_{n-1} + \br*{\frac{1}{J_n}\sum_{j = 1}^{J_n}\defaultStochGradient(\Theta_{n-1}, X_{n,j}) }, 
		\end{equation}
		\begin{equation}
		  \andq 
		  \Theta_n = \Theta_{n-1} - \gamma_n  \NewtonSchulzAlgorithm_{a, b, c, \varepsilon}(\mathbf{m}_n, K) . 
		\end{equation}
	}
}

\newcommand{\defdetermAMSGrad}{
\deterministicGDdef[true]
	{def:determ_AMSGrad}
	{AMSgrad \GD\ optimization method}
	{ 
	$(\gamma_n)_{n \in \N} \subseteq [0,\infty)$, 
	$(\alpha_n)_{n \in \N} \subseteq [0,1]$, 
	$(\beta_n)_{n \in \N} \subseteq [0,1]$, 
	$\varepsilon \in (0,\infty)$, }
	{AMSgrad \GD\ process for the objective function $\defaultLossFunction$ with learning rates $(\gamma_n)_{n \in \N}$, 
	momentum decay factors $(\alpha_n)_{n \in \N}$, 
	second moment decay factors $(\beta_n)_{n \in \N}$, 
	regularizing factor $\varepsilon$, 
	and initial value $\xi$}
	{there exist $\mathbf{m}= ( \mathbf{m}^{(1)},\ldots,\mathbf{m}^{(\defaultParamDim)}) \colon \N_0 \to \R^\defaultParamDim$, $\mathbb{M} = (\mathbb{M}^{(1)}, \ldots, \mathbb{M}^{(\defaultParamDim)}) \colon \N_0 \to \R^\defaultParamDim$, and $\mathfrak{M} = (\mathfrak{M}^{(1)}, \ldots, \mathfrak{M}^{(\defaultParamDim)}) \colon \N_0 \to \R^\defaultParamDim$ such that for all $n \in \N$, $i \in \{1,2,\ldots,\defaultParamDim\}$ it holds that
	\begin{equation}
	\Theta_0 = \xi, 
	\qquad 
	\mathbf{m}_0 = 0, 
	\qquad 
	\mathbb{M}_0 = 0, 
	\qquad
	\mathfrak{M}_0 = 0, 
	\end{equation}
	\begin{equation}
	\mathbf{m}_n = \alpha_n  \mathbf{m}_{ n - 1 } 
	+ ( 1 - \alpha_n )  \defaultGradientFunction( \Theta_{ n - 1 } ) ,
	\end{equation}
	\begin{equation}
	\mathbb{M}_n^{(i)} 
	= \beta_n\mathbb{M}_{n-1}^{(i)} 
	+ (1-\beta_n)
	\abs*{ \defaultGradientFunction_i( \Theta_{n-1} ) }^2
	,
	\end{equation}
	\begin{equation}
	\mathfrak{M}_n^{(i)} 
	= 
	\max \cu[\big]{
		\mathfrak{M}_{n-1}^{(i)}
		,
		\mathbb{M}_n^{(i)}  
	}
	,
	\qand
	\end{equation}
	\begin{equation}
	\Theta_n^{ (i) }
	=
	\Theta_{n-1}^{ (i) }
	-
	\frac{
		\gamma_n \mathbf{m}_n^{ (i) } 
	}{
		( \mathfrak{M}_n^{ (i)})^{\nicefrac{ 1 }{ 2 }  }
		+ 
		\varepsilon 
	}
	.
	\end{equation}
	}
}

\newcommand{\iterationStepDetermAMSGrad}{
	\State
 	$
		\mathbf{m} 
	\assign
		\alpha_n  \mathbf{m} 
		+ 
		( 1 - \alpha_n ) \grad( \Theta )
	$
	\State
	$
		\mathbb{M}
	\assign
		\beta_n\mathbb{M}
		+ 
		(1-\beta_n)
		\br*{ \grad( \Theta ) }^2
	$
	\State
	$
		\mathfrak{M}
	\assign
		\max \cu[\big]{
			\mathfrak{M}
			,
			\mathbb{M}
		}
	$
	\State
	$
		\Theta
	\assign
		\Theta
		- 
		\gamma_n
		\br[\big]{
			\varepsilon
			+
			\mathfrak{M}^{ 1/2  }
		}^{-1}
	 \mathbf{m}
	$
}
\newcommand{\algDescrDetermAMSGrad}{\algorithmicDescription{AMSgrad \GD\ optimization method}{def:determ_AMSGrad}{ 
	$ ( \gamma_n )_{ n \in \N } \subseteq [0,\infty) $, 
	$ ( \alpha_n )_{ n \in \N } \subseteq [0,1] $, 
	$ ( \beta_n )_{ n \in \N } \subseteq [0,1] $,
	$ \varepsilon \in (0,\infty) $, 
}{
	AMSgrad \GD\ process 
	for the objective function $ \defaultLossFunction $ 
	with learning rates $ ( \gamma_n )_{ n \in \N } $, 
	momentum decay factors $ ( \alpha_n )_{ n \in \N } $, 
	second moment decay factors $ ( \beta_n )_{ n \in \N } $, 
	regularizing factor $ \varepsilon $, 
	and initial value $ \xi $
}{
	$\Theta \assign \xi$\separator 
	$\mathbf{m} \assign 0 \in \R^\defaultParamDim$\separator
	$\mathbb{M} \assign 0 \in \R^\defaultParamDim$\separator
	$ \mathfrak{M} \assign 0  \in \R^\defaultParamDim$
}{
 	\iterationStepDetermAMSGrad
}
}

\newcommand{\iterationStepAMSGrad}{
	\pseudoCodeGradientAssign
	\State
 	$
		\mathbf{m} 
	\assign
		\alpha_n  \mathbf{m} 
		+ 
		( 1 - \alpha_n ) \pseudoCodeGradient
	$
	\State
	$
		\mathbb{M}
	\assign
		\beta_n\mathbb{M}
		+ 
		(1-\beta_n)
		\pseudoCodeGradient^2
	$
	\State
	$
		\mathfrak{M}
	\assign
		\max \cu[\big]{
			\mathfrak{M}
			,
			\mathbb{M}
		}
	$
	\State
	$
		\Theta
	\assign
		\Theta
		- 
		\gamma_n
		\br[\big]{
			\varepsilon
			+
			\mathfrak{M}^{ 1/2  }
		}^{-1}
	 \mathbf{m}
	$
}
\newcommand{\algDescrAMSGrad}{\algorithmicDescriptionStochastic{AMSgrad \SGD\ optimization method}{def:AMSGrad}{ 
	$ ( \gamma_n )_{ n \in \N } \subseteq [0,\infty) $, 
	$ ( \alpha_n )_{ n \in \N } \subseteq [0,1] $, 
	$ ( \beta_n )_{ n \in \N } \subseteq [0,1] $,
	$ \varepsilon \in (0,\infty) $, 
}{
	AMSgrad \SGD\ process 
	for the loss function $\defaultStochLoss$ 
	with learning rates $ ( \gamma_n )_{ n \in \N } $, 
	momentum decay factors $ ( \alpha_n )_{ n \in \N } $, 
	second moment decay factors $ ( \beta_n )_{ n \in \N } $, 
	regularizing factor $ \varepsilon $, 
	initial value $ \xi $,
}{
	$\Theta \assign \xi$\separator 
	$\mathbf{m} \assign 0 \in \R^\defaultParamDim$\separator
	$\mathbb{M} \assign 0 \in \R^\defaultParamDim$\separator
	$ \mathfrak{M} \assign 0  \in \R^\defaultParamDim$
}{
 	\iterationStepAMSGrad
}
}

\newcommand{\defAMSGrad}{
\SGDdef[true]
	{def:AMSGrad}
	{AMSGrad \SGD\ optimization method}
	{ 
	$(\gamma_n)_{n \in \N} \subseteq [0,\infty)$, 
	$(\alpha_n)_{n \in \N} \subseteq [0,1]$, 
	$(\beta_n)_{n \in \N} \subseteq [0,1]$, 
	$\varepsilon \in (0,\infty)$,}
	{AMSGrad \SGD\ process for the loss function $\defaultStochLoss$ with 
	learning rates $(\gamma_n)_{n \in \N}$, 
	momentum decay factors $(\alpha_n)_{n \in \N}$, 
	second moment decay factors $(\beta_n)_{n \in \N}$, 
	regularizing factor $\varepsilon$, 
	initial value $\xi$,}
	{there exist 
		$\mathbf{m} = (\mathbf{m}^{(1)},\ldots,\mathbf{m}^{(\defaultParamDim)}) \colon \N_0\times \Omega \to \R^\defaultParamDim$
		and
		$\mathbb{M} = (\mathbb{M}^{(1)},\ldots,\mathbb{M}^{(\defaultParamDim)}) \colon \N_0\times \Omega \to \R^\defaultParamDim$
		and
		$\mathfrak{M} = (\mathfrak{M}^{(1)},\ldots,\mathfrak{M}^{(\defaultParamDim)}) \colon \N_0\times \Omega \to \R^\defaultParamDim$
	such that for all $n \in \N$, $i \in \{1,2,\ldots,\defaultParamDim\}$ it holds that
	\begin{equation}
	\Theta_0 = \xi, 
	\qquad 
	\mathbf{m}_0 = 0, 
	\qquad 
	\mathbb{M}_0 = 0, 
	\qquad
	\mathfrak{M}_0 = 0, 
	\end{equation}
	\begin{equation}
	\mathbf{m}_n = \alpha_n \mathbf{m}_{ n - 1 }
	+ ( 1 - \alpha_n ) \br*{\frac{1}{J_n}\sum_{j = 1}^{J_n}\defaultStochGradient(\Theta_{ n - 1 }, X_{ n, j }) },
	\end{equation}
	\begin{equation}
	\mathbb{M}_n^{(i)}
	= \beta_n\mathbb{M}_{n-1}^{(i)}
	+ (1-\beta_n)
	\br*{ 
		\frac{1}{J_n}\sum_{j = 1}^{J_n}\defaultStochGradient_i(\Theta_{n-1}, X_{n,j})
	}^2
	,
	\end{equation}
	\begin{equation}
	\mathfrak{M}_n^{(i)}
	=
	\max \cu[\big]{
		\mathfrak{M}_{n-1}^{(i)}
		,
		\mathbb{M}_n^{(i)}  
	}
	,
	\qand
	\end{equation}
	\begin{equation}
	\Theta_n^{ (i) }
	=
	\Theta_{n-1}^{ (i) }
	-
	\gamma_n
	\br[\big]{
		\varepsilon
		+
		\mathfrak{M}_n^{ 1/2  }
	}^{-1}
	\mathbf{m}_n^{(i)}
	.
	\end{equation}
	}
}

\todoc{This is not done yet. I think I'm not sure about what the actual definition is. Need to go back to the literature here.}

\newcommand{\defdetermAdan}{
\deterministicGDdef[true]
	{def:determ_adan}
	{Adan \GD\ optimization method}
	{ 
	$(\gamma_n)_{n \in \N} \subseteq [0,\infty)$, 
	$(\alpha_n)_{n \in \N} \subseteq [0,1]$, 
	$(\beta_n)_{n \in \N} \subseteq [0,1]$, 
	$\varepsilon \in (0,\infty)$, }
	{Adan \GD\ process for the objective function $\defaultLossFunction$ with learning rates $(\gamma_n)_{n \in \N}$, 
	momentum decay factors $(\alpha_n)_{n \in \N}$, 
	second moment decay factors $(\beta_n)_{n \in \N}$, 
	regularizing factor $\varepsilon$, 
	and initial value $\xi$}
	{it holds for all $n \in \N$, $i \in \{1,2,\ldots,\defaultParamDim\}$ that 
	\begin{equation}
	\Theta_0 = \xi, 
	\qquad 
	\mathbf{m}_0 = 0, 
	\qquad 
	\mathbb{M}_0 = 0, 
	\end{equation}
	\begin{equation}
	\mathbf{m}_n = \alpha_n  \mathbf{m}_{ n - 1 } 
	+ ( 1 - \alpha_n )  \defaultGradientFunction( \Theta_{ n - 1 } ) ,
	\end{equation}
	\begin{equation}
	\mathbb{M}_n^{(i)} 
	= \beta_n\mathbb{M}_{n-1}^{(i)} 
	+ (1-\beta_n)
	\abs*{ \defaultGradientFunction_i( \Theta_{n-1} ) }^2
	,
	\end{equation}
	\begin{equation}
	\andq
	\Theta_n^{ (i) }
	=
	\Theta_{n-1}^{ (i) } -
	\gamma_n 
	{
	\textstyle
	\br*{
	\varepsilon
	+
	\br*{
	\frac{
	\mathbb{M}_n^{ (i)} 
	}{
	( 1 - \prod_{ l = 1 }^n \beta_l ) 
	}
	}^{ \nicefrac{ 1 }{ 2 } } 
	}^{ - 1 }
	}
	\br*{
	\frac{
	\mathbf{m}_n^{ (i) } 
	}{
	( 1 - \prod_{ k = 1 }^n \alpha_k )
	}
	}
	.
	\end{equation}
	}
}

\newcommand{\defdetermFullHistGD}{
\deterministicGDdef[false]
	{def:full_hist_GD}
	{Full history \GD\ optimization method}
	{ 
	$(\psi_n)_{n \in \N}$}
	{full history \GD\ process for the objective function $\defaultLossFunction$ with full history gradient steps $(\psi_n)_{n \in \N}$ and initial value $\xi$}
	{it holds for all $n \in \N$ that 
	\begin{equation}
	\Theta_0 = \xi 
	\qandq 
	\Theta_n 
	= 
	\Theta_{n-1} 
	- 
	\psi_n(\defaultGradientFunction(\Theta_0), \defaultGradientFunction(\Theta_1), \ldots, \defaultGradientFunction(\Theta_{n-1})).
	\end{equation}
	}
}

%% file: parts/Deterministic_GD_type_optimization_methods.tex
\cchapter{Deterministic gradient descent (GD) optimization methods}{chapter:deterministic}
\chaptermark{Deterministic GD optimization methods}

\todoc{Cite this somewhere: https://arxiv.org/abs/2312.07042
Probably best in some overview of approximation results for ANNs which needs to be written.}

\begin{introductions}
	This chapter reviews and studies deterministic \GD-type optimization methods 
	such as the classical plain-vanilla \GD\ optimization method 
	(see \cref{sec:gradient_descent} below) 
	as well as more sophisticated \GD-type optimization methods 
	including \GD\ optimization methods 
	with momenta (cf.\ \cref{sect:determ_momentum,sect:determ_nesterov,sect:determ_adam} below)
	and \GD\ optimization methods 
	with adaptive modifications of the learning rates 
	(cf.\ \cref{sect:determ_adagrad,sect:determ_RMSprop,sect:determ_adadelta,sect:determ_adam} below).
	
	There are several other outstanding reviews on gradient based optimization methods in the literature; cf., \eg, the books
	\cite[Chapter 5]{bach2023learning},
	\cite[Chapter 9]{BoydVandenberghe04},
	\cite[Chapter 3]{Bubeck15},
	\cite[Sections 4.3 and 5.9 and Chapter 8]{Goodfellow2016},
	\cite{Nesterov13},
	and
	\cite[Chapter 14]{shalev2014understanding} and the references therein
	and, \eg, the survey articles
	\cite{Ruder16,Bottou2018,Sun2019,ChaoWuLei2020,Bercu2011}
	and the references therein.

\end{introductions}

\section{GD optimization}
\label{sec:gradient_descent}

In this section we review and study the classical plain-vanilla \GD\ optimization method 
(cf., for example, 
\cite[Section 1.2.3]{Nesterov13}, 
\cite[Section 9.3]{BoydVandenberghe04}, 
and 
\cite[Chapter 3]{Bubeck15}). 
A simple intuition behind the \GD\ optimization method is the idea to solve 
a minimization problem by performing successive steps in direction of the steepest descents of the objective function, 
that is, 
by performing successive steps in the opposite direction of the gradients of the objective function. 

A slightly different and maybe a bit more accurate perspective for the \GD\ optimization method is 
to view the \GD\ optimization method as a plain-vanilla Euler discretization
of the associated \GF\ \ODE\ (see, for example, Theorem~\ref{flow} in Chapter~\ref{chapter:flow} above)

\defGD
\algDescrDetermGD

\cfclear
\begingroup
\providecommand{\d}{}
\renewcommand{\d}{\defaultParamDim}
\providecommand{\f}{}
\renewcommand{\f}{\defaultLossFunction}
\providecommand{\g}{}
\renewcommand{\g}{\defaultGradientFunction}
\begin{exercise}{quest:momentum}
Let
	$\xi = (\xi_1, \xi_2, \xi_3) \in \R^3$ 
satisfy $\xi = (1, 2, 3)$,
let 
	$\f \colon \R^3 \to \R$ 
satisfy for all 
	$\theta = (\theta_1, \theta_2, \theta_3 ) \in \R^3$
that
\begin{equation}
\begin{split} 
	\f(\theta) = 2 (\theta_1)^2 + (\theta_2 + 1)^2 + (\theta_3-1)^2,
\end{split}
\end{equation}
and let	
	$\Theta$ 
be the \GD\ process\cfadd{def:GD}
for the objective function $\f$ with
learning rates $\N \ni n \mapsto \frac{1}{2^n}$, 
and initial value $\xi$ \cfout.
Specify 
	$\Theta_1$, 
	$\Theta_2$, and
	$\Theta_3$
explicitly and prove that your results are correct!
\end{exercise}

\endgroup

\cfclear
\begingroup
\providecommand{\d}{}
\renewcommand{\d}{\defaultParamDim}
\providecommand{\f}{}
\renewcommand{\f}{\defaultLossFunction}
\providecommand{\g}{}
\renewcommand{\g}{\defaultGradientFunction}
\begin{exercise}{gradient_descent}
Let \(\xi = (\xi_1,\xi_2,\xi_3)\in \R^3 \) satisfy  \(\xi=(\xi_1,\xi_2,\xi_3)=(3,4,5)\),
let \(\f \colon\R^3\to\R\) satisfy for all \(\theta=(\theta_1,\theta_3)\in\R^3\) that
\[
	\f(\theta) =
		( \theta_1 )^2 + (\theta_2 - 1)^2 + 2\, (\theta_3+1)^2 
,\]
and let \(\Theta\) be the \GD\ process\cfadd{def:GD} for the objective function $\f$ with learning rates \(\N\ni n\mapsto \nicefrac{1}{3}\in[0,\infty)\) and initial value \(\xi\) \cfload.
Specify \(\Theta_1, \Theta_2\), and \(\Theta_3\) explicitly and prove that your results are correct.
\end{exercise}

\endgroup

\subsection{GD optimization in the training of ANNs}

In the next example we apply the \GD\ optimization method in the context of the training of fully-connected feedforward \anns\ in the vectorized description (see \cref{subsec:vectorized_description})
with the loss function being the mean squared error loss function in \cref{def:mseloss} (see \cref{sect:MSE}).

\begingroup
\renewcommand{\d}{\mathfrak d}
\renewcommand{\th}[1]{\Theta_{#1}}
\begin{athm}{example}{lem:gd_example}
	Let 
		$d,h,\mathfrak d\in\N$,
		$l_1,l_2,\dots,l_h\in\N$
	satisfy
		$\mathfrak d=l_1(d+1)+\br[\big]{\sum_{k=2}^h l_k(l_{k-1}+1)}+l_h+1$,
	let
		$a\colon \R\to\R$ be differentiable,
	let
		$M\in\N$,
		$x_1,x_2,\dots,x_M\in\R^d$,
		$y_1,y_2,\dots,y_M\in\R$,
	let
		$\mathscr L\colon \R^{\mathfrak d}\to\R$
	satisfy for all
		$\theta\in\R^{\mathfrak d}$
	that
	\begin{equation}
		\llabel{eq:defL}
		\mathscr L(\theta)
		=
		\frac1M\br*{\sum_{m=1}^{M}
		\abs*{ \bpr{ \RealV{ \theta}{0}{\defaultInputDim}{ \multdim_{a, l_1}, \multdim_{a, l_2}, \dots, \multdim_{a, l_h} , \id_{ \R } } } ( x_m ) - y_m }^2
		}
		,
	\end{equation}
	let 	
		$\xi\in\R^d$,
	let 
		$(\gamma_n)_{n\in\N}\subseteq\N$,
	and let
		$\th{} \colon \N_0\to\R^\d$
	satisfy for all
		$n\in\N$
	that
	\begin{equation}
		\llabel{eq:gd}
		\th 0=\xi
		\qquad\text{and}\qquad
		\th n
		=
		\th{n-1} - \gamma_n(\nabla \mathscr L)(\th{n-1})\cfadd{lem:differentiability_loss2}
	\end{equation}
	\cfload.
	Then 
		$\th{}$ is the \GD\ process for the objective function
		$\mathscr L$ with learning rates $(\gamma_n)_{n\in\N}$ and
		initial value $\xi$.
\end{athm}
\begin{aproof}
	\Nobs that
		\cref{def:GD:eq1} and 
		\lref{eq:gd}
	demonstrate that
	$\th{}$ is the \GD\ process for the objective function
	$\mathscr L$ with learning rates $(\gamma_n)_{n\in\N}$ and
	initial value $\xi$.
\end{aproof}
\endgroup

\subsection{Euler discretizations for GF ODEs}

\begin{athm}{theorem}{thm:taylor_formula}[Taylor's formula]
Let $ N \in \N $, 
$ \alpha \in \R $, $ \beta \in ( \alpha, \infty) $, 
$ a, b \in [ \alpha, \beta ] $,
$ f \in C^N( [ \alpha, \beta ] , \R ) $. 
Then 
\begin{equation}
\label{eq:Taylor_formula}
  f(b) 
  =
  \br*{
    \sum_{ n = 0 }^{ N - 1 }
    \frac{ 
      f^{ (n) }( a ) ( b - a )^n
    }{ n! }
  }
  +
  \int_0^1
    \frac{
      f^{ (N) }( a + r ( b - a ) )
      ( b - a )^N
      ( 1 - r )^{N-1}
    }{ (N-1)! }
  \,
  \diff r
  .
\end{equation}
\end{athm}
\begin{aproof}
\Nobs that the fundamental theorem of calculus 
assures that  for all
$ 
  g \in C^1( [0,1], \R ) 
$
it holds that
\begin{equation}
\label{eq:fundamental_theorem_of_calculus}
  g(1)
=
  g(0)
  +
  \int_0^1 g'(r) \, \diff r
=
  g(0)
  +
  \int_0^1 
  \frac{ g'(r) ( 1 - r )^0 }{ 0! } \, \diff r
  .
\end{equation}
\Moreover integration by parts ensures that 
for all 
$ n \in \N $, 
$ g \in C^{ n+1 }( [0,1], \R ) $
it holds that
\begin{equation}
\begin{split}
  \int_0^1 
  \frac{ 
    g^{ (n) }( r ) ( 1 - r )^{ n - 1 }
  }{ 
    (n - 1)!
  }
  \, \diff r
& 
  =
	-
  \biggl[ 
    \frac{ 
      g^{ (n) }( r ) ( 1 - r )^n
    }{
      n!
    }
  \biggr]^{ r = 1 }_{ r = 0 }
  +
  \int_0^1 
  \frac{ 
    g^{ (n+1) }( r ) ( 1 - r )^n 
  }{ 
    n!
  }
  \, \diff r
\\ &
=
  \frac{ 
    g^{ (n) }( 0 ) 
  }{
    n!
  }
  +
  \int_0^1 
  \frac{ 
    g^{ (n+1) }( r ) ( 1 - r )^n 
  }{ 
    n!
  }
  \, \diff r
  .
\end{split}
\end{equation}
Combining this with \cref{eq:fundamental_theorem_of_calculus} 
and induction shows that for all
$
  g \in C^N( [0,1], \R )
$
it holds that
\begin{equation}
	g(1)
	=
	\br*{
		\sum_{n=0}^{N-1}\frac{g^{(n)}(0)}{n!}
	} 
	+ 
	\int_0^1\frac{g^{(N)}(r)(1-r)^{N-1}}{(N-1)!}\,\diff r
	.
\end{equation}
This
establishes \cref{eq:Taylor_formula}. 
\end{aproof}

\cfclear
\begingroup
\providecommand{\d}{}
\renewcommand{\d}{\defaultParamDim}
\providecommand{\f}{}
\renewcommand{\f}{\defaultGradientFunction}
\providecommand{\g}{}
\renewcommand{\g}{g}
\begin{lemma}[Local error of the Euler method]
\label{one_step_euler}
Let 
	$ \d \in \N $, 
	$T, \gamma, c \in [0,\infty)$,
	$\f \in C^1( \R^\d, \R^\d ) $,
	$ \Theta \in C( [0,\infty), \R^\d ) $,  
	$\theta \in\R^\d$
satisfy for all
	$x, y \in \R^\d$, 
	$ t \in [0,\infty) $ 
that
\begin{equation}
\label{one_step_euler:ass1}
\begin{split} 
 	\Theta_t = \Theta_0 + \int_0^t \f( \Theta_s ) \, \diff s,
 \qquad
 	\theta
 =
 	\Theta_T + \gamma \f(\Theta_T),
\end{split}
\end{equation} 
\begin{equation}
\begin{split} 
	\pnorm2{\f(x)}
\leq
	c ,
\qandq
	\pnorm{2}{\f'(x) y}
\leq
	c\pnorm2{y}
\end{split}
\end{equation}
\cfload.
Then 
\begin{equation}
\begin{split} 
	\pnorm2{\Theta_{T + \gamma} - \theta}
\leq
	c^2\gamma^2.
\end{split}
\end{equation}
\end{lemma}

\begin{proof}[Proof of \cref{one_step_euler}]
Note that 
\enum{
	the fundamental theorem of calculus;
	the hypothesis that $ \f \in C^1( \R^\d, \R^\d ) $;
	\eqref{one_step_euler:ass1}
}[assure]
that for all
	$t \in (0,\infty)$
it holds that $\Theta \in C^1( [0,\infty), \R^\d )$ and
\begin{equation}
\label{one_step_euler:eq1}
\begin{split} 
	\dot \Theta_t
=
	\f( \Theta_t ).
\end{split}
\end{equation}
Combining this with
\enum{
	the hypothesis that $ \f \in C^1( \R^\d, \R^\d ) $;
	the chain rule
}
ensures that for all
	$t \in (0,\infty)$
it holds that $\Theta \in C^2( [0,\infty), \R^\d )$ and
\begin{equation}
\begin{split} 
	\ddot \Theta_t
=
	\f'( \Theta_t ) \dot \Theta_t
=
	\f'( \Theta_t ) \f( \Theta_t ).
\end{split}
\end{equation}
\enum{
	\Cref{thm:taylor_formula};
	\eqref{one_step_euler:eq1}
}
\hence
imply that
\begin{equation}
\begin{split} 
	\Theta_{T + \gamma}
&=
	\Theta_{T}
	+
	\gamma
	\dot \Theta_{T}
	+
	\int_{0}^{1}
		(1-r)
		\gamma^2
		\ddot \Theta_{T + r\gamma}
	\,\diff r\\
&=
	\Theta_{T}
	+
	\gamma
	\f( \Theta_T )
	+
	\gamma^2
	\int_{0}^{1}
		(1-r)
		\f'( \Theta_{T + r\gamma} ) \f( \Theta_{T + r\gamma} )
	\,\diff r.
\end{split}
\end{equation}
\enum{
	This;
	\eqref{one_step_euler:ass1};
}[demonstrate]
that
\begin{equation}
\begin{split} 
	&\pnorm2{\Theta_{T + \gamma} - \theta}\\
&=
	\Pnorm*2{
		\Theta_{T}
		+
		\gamma
		\f( \Theta_T )
		+
		\gamma^2
		\int_{0}^{1}
			(1-r)
			\f'( \Theta_{T + r\gamma} ) \f( \Theta_{T + r\gamma} )
		\,\diff r
		- 
		\pr{
			\Theta_T + \gamma \f(\Theta_T)
		}
	}\\
&\leq
	\gamma^2
	\int_{0}^{1}
		(1-r)
		\pnorm2{\f'( \Theta_{T + r\gamma} ) \f( \Theta_{T + r\gamma} )} 
	\,\diff r\\
&\leq
	c^2
	\gamma^2
	\int_{0}^{1}
		r
	\,\diff r
=
	\frac{c^2\gamma^2}{2}
\leq
	c^2\gamma^2.
\end{split}
\end{equation}
The proof of \cref{one_step_euler} is thus complete.
\end{proof}
\endgroup

\cfclear
\begingroup
\providecommand{\d}{}
\renewcommand{\d}{\defaultParamDim}
\providecommand{\f}{}
\renewcommand{\f}{\defaultLossFunction}
\providecommand{\g}{}
\renewcommand{\g}{\defaultGradientFunction}
\begin{cor}[Local error of the Euler method for \GF\ \ODEs]
\label{gradient_one_step_euler}
Let 
	$ \d \in \N $, 
	$T, \gamma, c \in [0,\infty)$,
	$ \f \in C^2( \R^\d, \R ) $,
	$ \Theta \in C( [0,\infty), \R^\d ) $,  
	$\theta \in\R^\d$
satisfy for all
	$x, y \in \R^\d$, 
	$ t \in [0,\infty) $ 
that
\begin{equation}
\begin{split} 
 	\Theta_t = \Theta_0 - \int_0^t (\nabla \f)( \Theta_s ) \, \diff s,
\qquad
	\theta
=
	\Theta_T - \gamma (\nabla \f)(\Theta_T),
\end{split}
\end{equation}
\begin{equation}
\label{gradient_one_step_euler:ass1}
\begin{split} 
	\Pnorm2{(\nabla \f) (x)}
\leq
	c,
\qandq
	\pnorm{2}{
		(\Hess \f)(x) y
	}
\leq
	c
	\pnorm2{y}
\end{split}
\end{equation}
\cfload.
Then 
\begin{equation}
\begin{split} 
	\Pnorm2{\Theta_{T + \gamma} - \theta}
\leq
	c^2\gamma^2.
\end{split}
\end{equation}
\end{cor}

\begin{proof}[Proof of \cref{gradient_one_step_euler}]
Throughout this proof, let $\g \colon \R^\d \to \R^\d$ satisfy for all 
	$\theta \in \R^\d$
that
\begin{equation}
\begin{split} 
	\g(\theta)
=
	- ( \nabla \f )(\theta).
\end{split}
\end{equation}
Note that 
\enum{
	the fact that for all
		$t \in [0,\infty)$
	it holds that
	$
	 	\Theta_t = \Theta_0 + \int_0^t \g( \Theta_s ) \, \diff s
	$;
	the fact that 
	$
		\theta
	=
		\Theta_T + \gamma \g(\Theta_T)
	$;
	the fact that for all
		$x \in \R^\d$
	it holds that
	$
		\pnorm2{\g(x)}
	\leq
		c
	$;
	the fact that for all
		$x, y \in \R^\d$
	it holds that
	$
		\pnorm{2}{\g'(x) y}
	\leq
		c\pnorm2{y}
	$;
	 \cref{one_step_euler}
}[imply]
that
$
	\pnorm2{\Theta_{T + \gamma} - \theta}
\leq
	c^2\gamma^2
$.
The proof of \cref{gradient_one_step_euler} is thus complete.
\end{proof}
\endgroup

\subsection{Lyapunov-type stability for GD optimization}
\label{subsect:objective_function}

\cref{Lyapunov} in \cref{subsec:Lyapunov_flow} and 
\cref{Lyapunov_cor} in \cref{subsec:Lyapunov_coercivity}
in \cref{chapter:flow} above,
in particular, illustrate how Lyapunov-type functions can 
be employed to establish convergence properties for \GFs. 
Roughly speaking, the next two results, 
\cref{prop:lyapunov} and \cref{cor:lyapunov} below, 
are the time-discrete analogons of 
\cref{Lyapunov} and 
\cref{Lyapunov_cor}, respectively.

\begingroup
\providecommand{\d}{}
\renewcommand{\d}{\defaultParamDim}
\providecommand{\f}{}
\renewcommand{\f}{\defaultLossFunction}
\providecommand{\g}{}
\renewcommand{\g}{\defaultGradientFunction}
\begin{prop}[Lyapunov-type stability for discrete-time dynamical systems]
\label{prop:lyapunov}
Let $ \d \in \N $, $ \xi \in \R^\d $, $ c \in (0,\infty) $,
$ ( \gamma_n )_{ n \in \N } \subseteq [0,c] $,
let 
$ V \colon \R^\d \to \R $,
$ \Phi \colon \R^\d \times [0,\infty) \to \R^\d $,
and
$ \varepsilon \colon [0,c] \to [0,\infty) $
satisfy
for all $ \theta \in \R^\d $, $ t \in [0,c] $ 
that
\begin{equation}
\label{eq:assumption_Lyapunov}
  V( \Phi( \theta, t ) ) \leq \varepsilon( t ) V( \theta )
  ,
\end{equation}
and let 
$ \Theta \colon \N_0 \to \R^\d $ satisfy for all $ n \in \N $ that
\begin{equation}
\label{eq:assumption_Lyapunov_dynamic_T}
  \Theta_0 = \xi 
  \qandq
  \Theta_n = \Phi( \Theta_{ n - 1 }, \gamma_n )
  .
\end{equation}
Then it holds for all $ n \in \N_0 $ that
\begin{equation}
\label{eq:Lyapunov}
  V( \Theta_n ) 
  \leq 
  \br*{ 
    \textstyle
    \prod\limits_{ k = 1 }^n
    \displaystyle
    \varepsilon( \gamma_k )
  }
  V( \xi )
  .
\end{equation}
\end{prop}

\begin{proof}[Proof of \cref{prop:lyapunov}]
We prove \eqref{eq:Lyapunov} by induction on $ n \in \N_0 $. 
For the base case $ n = 0 $ note that the assumption that $ \Theta_0 = \xi $ 
ensures that $ V( \Theta_0 ) = V( \xi ) $. 
This establishes \eqref{eq:Lyapunov} in the base case $ n = 0 $. 
For the induction step observe that
\eqref{eq:assumption_Lyapunov_dynamic_T} 
and
\eqref{eq:assumption_Lyapunov} ensure that 
for all $ n \in \N_0 $ with 
$
  V( \Theta_n ) 
  \leq 
  \pr*{ 
    \prod_{ k = 1 }^n
    \varepsilon( \gamma_k )
  }
  V( \xi )
$
it holds that
\begin{equation}
\begin{split} 
  V( \Theta_{ n + 1 } ) 
& =
  V\pr*{
    \Phi( \Theta_n , \gamma_{ n + 1 } )
  }
  \leq 
  \varepsilon( \gamma_{ n + 1 } )
  V( \Theta_n )
\\ &
  \leq 
  \varepsilon( \gamma_{ n + 1 } )
  \pr*{
  \br*{ 
    \textstyle
    \prod\limits_{ k = 1 }^n
    \displaystyle
    \varepsilon( \gamma_k )
  }
  V( \xi )
  }
=
  \br*{ 
    \textstyle
    \prod\limits_{ k = 1 }^{ n + 1 }
    \displaystyle
    \varepsilon( \gamma_k )
  }
  V( \xi )
  .
\end{split}
\end{equation}
Induction thus establishes \eqref{eq:Lyapunov}. 
The proof of
\cref{prop:lyapunov} is thus complete.
\end{proof}
\endgroup

\begingroup
\providecommand{\d}{}
\renewcommand{\d}{\defaultParamDim}
\providecommand{\f}{}
\renewcommand{\f}{\defaultLossFunction}
\providecommand{\g}{}
\renewcommand{\g}{\defaultGradientFunction}
\begin{cor}[On quadratic Lyapunov-type functions for the \GD\ optimization method]
\label{cor:lyapunov}
Let $ \d \in \N $, $ \vartheta, \xi \in \R^\d $, $c\in(0,\infty)$, 
$ ( \gamma_n )_{ n \in \N } \subseteq [0,c] $,
$ \f \in C^1( \R^\d, \R ) $,
let $\normmm{\cdot} \colon \R^\d \to [0,\infty)$ be a norm, 
let 
$ \varepsilon \colon [0,c] \to [0,\infty) $
satisfy
for all $ \theta \in \R^\d $, $ t \in [0,c] $ 
that
\begin{equation}
\label{eq:assumption_LyapunovB}
  \normmm{ \theta - t ( \nabla \f )( \theta ) - \vartheta }^2 \leq \varepsilon( t ) 
  \normmm{ \theta - \vartheta }^2
  ,
\end{equation}
and let 
$ \Theta \colon \N_0 \to \R^\d $ satisfy for all $ n \in \N $ that
\begin{equation}
\label{eq:assumption_Lyapunov_dynamic_TB}
  \Theta_0 = \xi 
  \qandq
  \Theta_n = 
  \Theta_{ n - 1 } 
  -
  \gamma_n ( \nabla \f )( \Theta_{ n - 1 } )
  .
\end{equation}
Then it holds for all $ n \in \N_0 $ that
\begin{equation}
\label{eq:LyapunovB}
  \normmm{ \Theta_n - \vartheta }
  \leq 
  \br*{ 
    \textstyle
    \prod\limits_{ k = 1 }^n
    \displaystyle
    \br*{
      \varepsilon( \gamma_k )
    }^{ \nicefrac{ 1 }{ 2 } }
  }
  \normmm{ \xi - \vartheta }
  .
\end{equation}
\end{cor}

\begin{proof}[Proof of \cref{cor:lyapunov}]
Throughout this proof, let $ V \colon \R^\d \to \R $ and $ \Phi \colon \R^\d \times [0,\infty) \to \R^\d $ satisfy for all 
$ \theta \in \R^\d $, $t \in [0,\infty)$ that
\begin{equation}
\label{cor:lyapunov:eq1}
  V( \theta ) = \normmm{ \theta - \vartheta }^2
\qandq
  \Phi(\theta, t)
=
  \theta - t (\nabla \f)(\theta)
  .
\end{equation}
Observe that 
\enum{
	\cref{prop:lyapunov} (applied with $ V \is V $, $\Phi \is \Phi$ in the notation of \cref{prop:lyapunov}); \eqref{cor:lyapunov:eq1}
}[imply]
that for all $ n \in \N_0 $ 
it holds that
\begin{equation}
  \normmm{ \Theta_n - \vartheta }^2
  =
  V( \Theta_n )
  \leq 
  \br*{ 
    \textstyle
    \prod\limits_{ k = 1 }^n
    \displaystyle
    \varepsilon( \gamma_k )
  }
  V( \xi )
  =
  \br*{ 
    \textstyle
    \prod\limits_{ k = 1 }^n
    \displaystyle
    \varepsilon( \gamma_k )
  }
  \normmm{ \xi - \vartheta }^2
  .
\end{equation}
This establishes \eqref{eq:LyapunovB}.
The proof of \cref{cor:lyapunov} is thus complete.
\end{proof}
\endgroup

\cref{cor:lyapunov}, in particular, illustrates that 
the one-step Lyapunov stability 
assumption in \eqref{eq:assumption_LyapunovB} may provide us suitable estimates 
for the approximation errors associated to the \GD\ optimization method; 
see \eqref{eq:LyapunovB} above. The next result, \cref{fcond} below, now provides 
us sufficient conditions which ensure that 
the one-step Lyapunov stability 
condition in \eqref{eq:assumption_LyapunovB} is satisfied so that we are in the position
to apply \cref{cor:lyapunov} above to obtain estimates for the approximation 
errors associated to the \GD\ optimization method. 
\cref{fcond} employs the growth condition and the coercivity-type condition in 
\eqref{cor_flow:assumption1} in \cref{cor_flow} above.
Results similar to \cref{fcond} can, \eg, be found in 
\cite[Remark 2.1]{DereichGronbach17} 
and 
\cite[Lemma~2.1]{Jentzen18strong}. 
We will employ the statement of \cref{fcond} in our error analysis for the 
\GD\ optimization method in \cref{subsect:GD_convergence} below.

\begingroup
\providecommand{\d}{}
\renewcommand{\d}{\defaultParamDim}
\providecommand{\f}{}
\renewcommand{\f}{\defaultLossFunction}
\providecommand{\g}{}
\renewcommand{\g}{\defaultGradientFunction}
\begin{lemma}[Sufficient conditions for a one-step Lyapunov-type stability condition] %
\label{fcond}
Let $\d \in \N$, 
let $\altscp{ \cdot, \cdot } \colon \R^\d \times \R^\d \to \R$ be a scalar product, 
let $\normmm{\cdot} \colon \R^\d \to \R$ satisfy for all $v \in \R^\d$ that $\normmm{v} = \sqrt{\altscp{ v , v }}$, 
and let $c,L \in (0,\infty)$, $r \in (0,\infty]$, $\vartheta \in \R^\d$, $\mathbb{B} = \{w \in \R^\d \colon \normmm{w-\vartheta} \leq r \}$, $\f \in C^1(\R^\d,\R)$ satisfy for all $\theta \in \mathbb{B}$ that
\begin{equation}
\label{fcond:assumption1}
\altscp{\theta-\vartheta, (\nabla \f)(\theta) } \geq  c \normmm{\theta-\vartheta}^2 \qandq \normmm{(\nabla \f)(\theta)} \leq L \normmm{\theta-\vartheta}.
\end{equation}
Then 
\begin{enumerate}[label=(\roman *)]
\item \label{fcond:item1}
it holds that $c \leq L$,

\item \label{fcond:item2}
it holds for all  $\theta\in\mathbb B$, $\gamma \in [0,\infty)$ that
\begin{equation}
\normmm{\theta - \gamma (\nabla \f)(\theta) - \vartheta}^2  \leq (1 -2\gamma c + \gamma^2 L^2 )\normmm{\theta-\vartheta}^2,
\end{equation}

\item \label{fcond:item3}
it holds for all $\gamma \in (0,\frac{2c}{L^2})$ that 
$0\leq 1 -2\gamma c + \gamma^2 L^2 <1 $, and

\item \label{fcond:item4}
it holds for all $\theta \in \mathbb{B}$, $\gamma \in [0,\frac{c}{L^2}]$ that 
\begin{equation}
\normmm{\theta - \gamma (\nabla \f)(\theta) - \vartheta}^2 \leq (1-c\gamma) \normmm{\theta-\vartheta}^2.
\end{equation}
\end{enumerate}
\end{lemma}

\begin{proof}[Proof of \cref{fcond}]
First of all, note that (\ref{fcond:assumption1}) ensures that for all $\theta \in \mathbb{B}$, $\gamma \in [0,\infty)$ it holds that
\begin{equation}
\label{fcond:eq1}
\begin{split}
0 \leq \normmm{\theta - \gamma (\nabla \f)(\theta) - \vartheta}^2 &=   \normmm{(\theta - \vartheta)- \gamma (\nabla \f)(\theta) }^2\\
&= \normmm{\theta-\vartheta}^2 -2\gamma \,\altscp{ \theta-\vartheta, (\nabla \f)(\theta) } + \gamma^2 \normmm{ (\nabla \f)(\theta)}^2 \\
&\leq \normmm{\theta-\vartheta}^2 -2\gamma c\normmm{\theta-\vartheta}^2 + \gamma^2 L^2 \normmm{\theta-\vartheta}^2 \\
&= (1 -2\gamma c + \gamma^2 L^2 )\normmm{\theta-\vartheta}^2.
\end{split}
\end{equation}
This establishes \cref{fcond:item2}. 
Moreover, note that the fact that $\mathbb{B} \backslash \{ \vartheta \} \neq \emptyset$ and (\ref{fcond:eq1}) assure that for all $\gamma \in [0,\infty)$ it holds that 
\begin{equation}
\label{fcond:eq2}
1 -2\gamma c + \gamma^2 L^2 \geq 0.
\end{equation}
Hence, we obtain that 
\begin{equation}
\begin{split}
1- \tfrac{c^2}{L^2} = 1 -\tfrac{2c^2}{L^2} + \tfrac{c^2}{L^2} 
&= 
1 -2\br*{\tfrac{c}{L^2}}c + \br[\big]{\tfrac{c^2}{L^4}}L^2 \\
&=
1 -2\br*{\tfrac{c}{L^2}}c + \br*{\tfrac{c}{L^2}}^2 L^2 
\geq 0.
\end{split}
\end{equation}
This implies that $\tfrac{c^2}{L^2} \leq 1$.
Therefore, we obtain that $c^2 \leq L^2$.
This establishes \cref{fcond:item1}.
Furthermore, observe that (\ref{fcond:eq2}) ensures that for all $\gamma \in (0,\frac{2c}{L^2})$ it holds that
\begin{equation}
0 \leq 1 -2\gamma c + \gamma^2 L^2 = 1 - \underbrace{\gamma}_{ > 0} \underbrace{(2c - \gamma L^2)}_{ > 0} < 1.
\end{equation}
This proves \cref{fcond:item3}. 
In addition, note that for all $\gamma \in [0,\frac{c}{L^2}]$ it holds that
\begin{equation}
1 -2\gamma c + \gamma^2 L^2 \leq 1 -2\gamma c + \gamma \br*{\tfrac{c}{L^2}} L^2 = 1-c\gamma.
\end{equation}
Combining this with (\ref{fcond:eq1}) establishes \cref{fcond:item4}.
The proof of \cref{fcond} is thus complete.
\end{proof}
\endgroup

\begingroup
\providecommand{\d}{}
\renewcommand{\d}{\defaultParamDim}
\providecommand{\f}{}
\renewcommand{\f}{\defaultLossFunction}
\providecommand{\g}{}
\renewcommand{\g}{\defaultGradientFunction}
\begin{exercise}{ex:lyapunov1}
	Prove or disprove the following statement:
	There exist $\d \in \N$,
	$\gamma\in(0,\infty)$,
	$\eps\in(0,1)$,
	$r\in(0,\infty]$,
	$\vartheta,\theta\in\R^\d$
	and there exists a function
	$\g \colon\R^\d\to\R^\d$
	such that
	$\pnorm2{\theta-\vartheta}\leq r$,
	$\forall\, \xi\in\{w\in\R^\d\colon\pnorm2{w-\vartheta}\leq r\}\colon \pnorm2{\xi-\gamma \g(\xi)-\vartheta}\leq\eps\pnorm2{\xi-\vartheta}$,
	and
	\begin{equation}
		\scp{\theta-\vartheta,\g(\theta)}
		<
		\min\cu[\big]{
			\tfrac{1-\eps^2}{2\gamma},
			\tfrac\gamma2
		}
		\max\cu*{
			\pnorm2{\theta-\vartheta}^2,
			\pnorm2{\g(\theta)}^2
		}.
	\end{equation}
\end{exercise}
\endgroup

\begingroup
\providecommand{\d}{}
\renewcommand{\d}{\defaultParamDim}
\providecommand{\f}{}
\renewcommand{\f}{\defaultLossFunction}
\providecommand{\g}{}
\renewcommand{\g}{\defaultGradientFunction}
\begin{exercise}{ex:lyapunov2}
	Prove or disprove the following statement:
	For all 
		$\d \in \N$,
		$r\in(0,\infty]$,
		$\vartheta\in\R^\d$
	and for every function
		$\g \colon\R^\d\to\R^\d$ which satisfies
		$\forall\, \theta\in\{w\in\R^\d\colon\pnorm2{w-\vartheta}\leq r\}\colon 
		\scp{\theta-\vartheta,\g(\theta)}\geq \tfrac12\max\{\pnorm2{\theta-\vartheta}^2,\pnorm2{\g(\theta)}^2\}$
	it holds that
	\begin{equation}
		\forall \theta\in\{w\in\R^\d\colon\pnorm2{w-\vartheta}\leq r\}\colon
			\pr[\big]{\scp{\theta-\vartheta,\g(\theta)}\geq\tfrac12\pnorm2{\theta-\vartheta}^2
			\;\land\;
			\pnorm2{\g(\theta)}\leq 2\pnorm2{\theta-\vartheta}}.
	\end{equation}
\end{exercise}
\endgroup

\begingroup
\providecommand{\d}{}
\renewcommand{\d}{\defaultParamDim}
\providecommand{\f}{}
\renewcommand{\f}{\defaultLossFunction}
\providecommand{\g}{}
\renewcommand{\g}{\defaultGradientFunction}
\begin{exercise}{ex:lyapunov3}
	Prove or disprove the following statement:
	For all $\d \in \N$,
	$c\in(0,\infty)$,
	$r\in(0,\infty]$,
	$\vartheta,v\in\R^\d$,
	$\f\in C^1(\R^\d,\R)$,
	$s,t\in[0,1]$
	such that
		$\pnorm2 v\leq r$,
		$s\leq t$, and
		$\forall\, \theta\in\{w\in\R^\d\colon \pnorm2{w-\vartheta}\leq r\}\colon
		\scp{\theta-\vartheta,(\nabla \f)(\theta)}\geq c\pnorm2{\theta-\vartheta}^2$
	it holds that
	\begin{equation}
		\f(\vartheta+tv)-\f(\vartheta+sv)\geq \tfrac c2(t^2-s^2)\pnorm2v^2.
	\end{equation}
\end{exercise}
\endgroup

\begingroup
\providecommand{\d}{}
\renewcommand{\d}{\defaultParamDim}
\providecommand{\f}{}
\renewcommand{\f}{\defaultLossFunction}
\providecommand{\g}{}
\renewcommand{\g}{\defaultGradientFunction}
\begin{exercise}{ex:lyapunov4}
	Prove or disprove the following statement:
	For every $\d \in \N$,
	$c\in(0,\infty)$,
	$r\in(0,\infty]$,
	$\vartheta\in\R^\d$
	and for every
	$\f \in C^1(\R^\d,\R)$
	which satisfies for all
		$v\in\R^\d$,
		$s,t\in[0,1]$
		with $\pnorm2 v\leq r$ and $s\leq t$
	that
	  $\f(\vartheta+tv)-\f(\vartheta+sv)\geq c(t^2-s^2)\pnorm2v^2$
	it holds that
	\begin{equation}
		\forall\, \theta\in\{w\in\R^\d\colon \pnorm2{w-\vartheta}\leq r\}\colon
		\scp{\theta-\vartheta,(\nabla \f)(\theta)}\geq 2c\pnorm2{\theta-\vartheta}^2
		.
	\end{equation}
\end{exercise}
\endgroup

\begingroup
\providecommand{\d}{}
\renewcommand{\d}{\defaultParamDim}
\providecommand{\f}{}
\renewcommand{\f}{\defaultLossFunction}
\providecommand{\g}{}
\renewcommand{\g}{\defaultGradientFunction}
\begin{exercise}{ex:lyapunov5}
	Let $\d \in \N$ and for every $v\in\R^\d$, $R\in[0,\infty]$ let
	$\mathbb B_{R}(v)=\{w\in\R^\d\colon \pnorm2{w-v}\leq R\}$.
	Prove or disprove the following statement:
	For all $r\in(0,\infty]$,
	$\vartheta\in\R^\d$,
	$\f \in C^1(\R^\d,\R)$
	the following two statements are equivalent:
	\begin{enumerate}[(i)]
		\item There exists $c\in(0,\infty)$ such that for all $\theta\in\mathbb B_r(\vartheta)$
		it holds that
		\begin{equation}
			\scp{\theta-\vartheta,(\nabla \f)(\theta)}
			\geq
			c\pnorm2{\theta-\vartheta}^2
			.
		\end{equation}
		\item There exists $c\in(0,\infty)$ such that for all
		  $v,w\in\mathbb B_r(\vartheta)$, $s,t\in[0,1]$ with $s\leq t$ it holds that
		\begin{equation}
			\f(\vartheta+t(v-\vartheta))-\f(\vartheta+s(v-\vartheta))\geq c(t^2-s^2)\pnorm2{v-\vartheta}^2.
		\end{equation}
	\end{enumerate}
\end{exercise}
\endgroup

\begingroup
\providecommand{\d}{}
\renewcommand{\d}{\defaultParamDim}
\providecommand{\f}{}
\renewcommand{\f}{\defaultLossFunction}
\providecommand{\g}{}
\renewcommand{\g}{\defaultGradientFunction}
\begin{exercise}{ex:lyapunov6}
	Let $\d \in \N$ and for every $v\in\R^\d$, $R\in[0,\infty]$ let
	$\mathbb B_{R}(v)=\{w\in\R^\d\colon \pnorm2{v-w}\leq R\}$.
	Prove or disprove the following statement:
	For all $r\in(0,\infty]$,
	$\vartheta\in\R^\d$,
	$\f \in C^1(\R^\d,\R)$
	the following three statements are equivalent:
	\begin{enumerate}[(i)]
		\item There exist $c,L\in(0,\infty)$ such that for all 
		$\theta\in\mathbb B_r(\vartheta)$ it holds that
		\begin{equation}
			\scp{\theta-\vartheta,(\nabla \f)(\theta)}
			\geq
			c\pnorm2{\theta-\vartheta}^2
			\qquad\text{and}\qquad
			\pnorm2{(\nabla \f)(\theta)}
			\leq
			L\pnorm2{\theta-\vartheta}
			.
		\end{equation}
		\item There exist $\gamma\in(0,\infty)$, $\eps\in(0,1)$ such that for all
		$\theta\in\mathbb B_r(\vartheta)$ it holds that
		\begin{equation}
			\pnorm2{\theta-\gamma(\nabla \f)(\theta)-\vartheta}
			\leq
			\eps\pnorm2{\theta-\vartheta}
			.
		\end{equation}
		\item
		There exists $c \in (0,\infty)$ such that for all 
		$\theta \in \mathbb B_r(\vartheta)$ it holds that
		\begin{equation}
			\scp{\theta-\vartheta, (\nabla \f)(\theta) }
			\geq 
			c \max\cu[\big]{\pnorm2{\theta-\vartheta}^2, \pnorm2{(\nabla \f)(\theta)}^2}.
		\end{equation}
	\end{enumerate}
\end{exercise}
\endgroup

\subsection{Error analysis for GD optimization}
\label{subsect:GD_convergence}

In this subsection we provide an error analysis for the \GD\ optimization method. 
In particular, we show under suitable hypotheses (cf.\ \cref{GD_convergence} below) 
that the considered \GD\ process converges to a local minimum point 
of the objective function of the considered optimization problem.

\subsubsection{Error estimates for GD optimization}
\label{sec:error_estimates_GD}

\cfclear
\begingroup
\providecommand{\d}{}
\renewcommand{\d}{\defaultParamDim}
\providecommand{\f}{}
\renewcommand{\f}{\defaultLossFunction}
\providecommand{\g}{}
\renewcommand{\g}{\defaultGradientFunction}
\begin{prop}[Error estimates for the \GD\ optimization method]
\label{GD_convergence}
Let $\d \in \N$, 
$c,L \in (0,\infty)$, $r \in (0,\infty]$, $(\gamma_n)_{n \in \N} \subseteq [0,\frac{2c}{L^2}]$, $\vartheta \in \R^\d$, $\mathbb{B} = \{w \in \R^\d \colon \pnorm2{w-\vartheta} \leq r \}$, $\xi \in \mathbb{B}$, $\f \in C^1(\R^\d,\R)$ satisfy for all $\theta \in \mathbb{B}$ that
\begin{equation}
\label{GD_convergence:assumption1}
\scp{\theta-\vartheta, (\nabla \f)(\theta) } \geq  c \pnorm2{\theta-\vartheta}^2 \qandq \pnorm2{(\nabla \f)(\theta)} \leq L \pnorm2{\theta-\vartheta},
\end{equation}
and let $\Theta \colon \N_0 \to \R^\d$ satisfy for all $n \in \N$ that 
\begin{equation}
	\label{eq:GD_convergence:1}
\Theta_0 = \xi \qandq \Theta_n = \Theta_{n-1} - \gamma_n (\nabla \f)(\Theta_{n-1})
\end{equation}
\cfload.
Then
\begin{enumerate}[label=(\roman *)]
\item \label{GD_convergence:item1}
it holds that $\{\theta \in \mathbb{B}  \colon  \f(\theta)= \inf\nolimits_{w \in \mathbb{B}} \f(w)  \} = \{\vartheta \}$,
\item \label{GD_convergence:item2}
it holds for all $n  \in \N$ that 
$0\leq 1-2c\gamma_n + (\gamma_n)^2L^2\leq 1$,
\item \label{GD_convergence:item3}
it holds for all $n \in \N$ that 
$
\pnorm2{\Theta_n-\vartheta} \leq  (1-2c\gamma_n + (\gamma_n)^2L^2)^{\nicefrac{1}{2}} \pnorm2{\Theta_{n-1} - \vartheta} 
\leq r
$,
\item \label{GD_convergence:item4}
it holds for all $n \in \N_0$ that
\begin{equation}
\label{GD_convergence:eq2}
\pnorm2{\Theta_n-\vartheta} 
\leq 
\br*{
  \textstyle
  \prod\limits_{ k = 1 }^n 
  \displaystyle 
  (1-2c\gamma_k + (\gamma_k)^2L^2)^{\nicefrac{1}{2}}
} \pnorm2{ \xi - \vartheta },
\end{equation}
and
\item \label{GD_convergence:item5}
it holds for all $n \in \N_0$ that
\begin{equation}
\label{GD_convergence:eq3}
0 \leq
\f(\Theta_n)-\f(\vartheta) \leq \tfrac{L}{2}\pnorm2{\Theta_n-\vartheta}^2 
\leq \tfrac{L}{2}
\br*{
  \textstyle
  \prod\limits_{ k = 1 }^n 
  \displaystyle 
  (1-2c\gamma_k + (\gamma_k)^2L^2)
}
\pnorm2{ \xi - \vartheta }^2 .
\end{equation}
\end{enumerate}
\end{prop}

\begin{proof}[Proof of \cref{GD_convergence}]
First, \nobs that (\ref{GD_convergence:assumption1}) and \cref{fcond1:item2} in \cref{fcond1} prove \cref{GD_convergence:item1}. 
Moreover, \nobs that \eqref{GD_convergence:assumption1}, 
\cref{fcond:item3} in \cref{fcond}, 
the assumption that for all $n \in \N$ it holds that $\gamma_n \in [0,\frac{2c}{L^2}]$, and the fact that
\begin{equation}
\begin{split}
1-2c \br*{ \tfrac{2c}{L^2}} +\br*{ \tfrac{2c}{L^2}} ^2L^2 
= 
1-\tfrac{4c^2}{L^2} +\br[\big]{\tfrac{4c^2}{L^4} }L^2 
=
1-\tfrac{4c^2}{L^2} +\tfrac{4c^2}{L^2} = 1
\end{split}
\end{equation}
and  establish \cref{GD_convergence:item2}.
Next we claim that for all $n \in \N$ it holds that
\begin{equation}
\label{GD_convergence:eq1}
\pnorm2{\Theta_n-\vartheta} \leq  (1-2c\gamma_n + (\gamma_n)^2L^2)^{\nicefrac{1}{2}} \pnorm2{\Theta_{n-1} - \vartheta} \leq r.
\end{equation}
We now prove (\ref{GD_convergence:eq1}) by induction on $n \in \N$. 
For the base case $n = 1$ \nobs that 
	\eqref{eq:GD_convergence:1},
	the assumption that $\Theta_0 = \xi \in \mathbb{B}$, 
	\cref{fcond:item2} in \cref{fcond}, and 
	\cref{GD_convergence:item2} 
ensure that
\begin{equation}
\begin{split}
\pnorm2{\Theta_1-\vartheta}^2 
&= 
\pnorm2{\Theta_0 - \gamma_1(\nabla \f)(\Theta_0)-\vartheta}^2 \\
&\leq 
(1-2c\gamma_1 + (\gamma_1)^2L^2)\pnorm2{\Theta_0-\vartheta}^2 \\
&\leq
\pnorm2{\Theta_0-\vartheta}^2 \leq r^2.
\end{split}
\end{equation}
This establishes (\ref{GD_convergence:eq1}) in the base case $n = 1$. 
For the induction step \nobs that 
	\eqref{eq:GD_convergence:1},
	\cref{fcond:item2} in \cref{fcond}, and 
	\cref{GD_convergence:item2} 
imply that for all $n \in \N$ with $\Theta_n \in \mathbb{B}$ it holds that
\begin{equation}
\begin{split}
\pnorm2{\Theta_{n+1}-\vartheta}^2 
&= 
\pnorm2{\Theta_n - \gamma_{n+1}(\nabla \f)(\Theta_n)-\vartheta}^2 \\
&\leq 
\underbrace{(1-2c\gamma_{n+1} + (\gamma_{n+1})^2L^2)}_{\in [0,1]}\pnorm2{\Theta_n-\vartheta}^2 \\ &\leq 
\pnorm2{\Theta_n-\vartheta}^2 
\leq 
r^2.
\end{split}
\end{equation}
This demonstrates that for all $n \in \N$ with $\pnorm2{\Theta_{n}-\vartheta} \leq r$ it holds that
\begin{equation}
\pnorm2{\Theta_{n+1}-\vartheta} 
\leq (1-2c\gamma_{n+1} + (\gamma_{n+1})^2L^2)^{\nicefrac{1}{2}}\pnorm2{\Theta_n-\vartheta} \leq r.
\end{equation}
Induction thus proves (\ref{GD_convergence:eq1}). 
Next \nobs that (\ref{GD_convergence:eq1}) establishes \cref{GD_convergence:item3}. 
Moreover, \nobs that 
	induction,
	\cref{GD_convergence:item2}, and 
	\cref{GD_convergence:item3} 
prove 
	\cref{GD_convergence:item4}. 
Furthermore, \nobs that \cref{GD_convergence:item3} 
and the fact that $\Theta_0=\xi\in\mathbb B$
ensure that for all $n \in \N_0$ it holds that $\Theta_n \in \mathbb{B}$. 
Combining this, \eqref{GD_convergence:assumption1}, 
and \cref{fcond2}
with \cref{GD_convergence2:item1,GD_convergence:item4} 
establishes \cref{GD_convergence:item5}. 
The proof of \cref{GD_convergence} is thus complete.
\end{proof}
\endgroup

\subsubsection{Size of the learning rates}
\label{sec:size_of_learningrates}

In the next result, \cref{GD_convergence_constant_optimal} below, 
we, roughly speaking, specialize \cref{GD_convergence} to the case where 
the learning rates $ ( \gamma_n )_{ n \in \N } \subseteq [0, \frac{ 2 c }{ L^2 } ] $
are a constant sequence.

\begingroup
\cfclear
\providecommand{\d}{}
\renewcommand{\d}{\defaultParamDim}
\providecommand{\f}{}
\renewcommand{\f}{\defaultLossFunction}
\providecommand{\g}{}
\renewcommand{\g}{\defaultGradientFunction}
\begin{cor}[Convergence of \GD\ for constant learning rates]
\label{GD_convergence_constant_optimal}
Let $\d \in \N$, 
$c,L \in (0,\infty)$, $r \in (0,\infty]$, $\gamma\in (0,\frac{2c}{L^2})$, $\vartheta \in \R^\d$, $\mathbb{B} = \{w \in \R^\d \colon \pnorm2{w-\vartheta} \leq r \}$, $\xi \in \mathbb{B} $, $\f \in C^1(\R^\d,\R)$ satisfy for all $\theta \in \mathbb{B}$ that
\begin{equation}
\label{GD_convergence_constant_optimal:assumption1}
\scp{\theta-\vartheta, (\nabla \f)(\theta) } \geq  c \pnorm2{\theta-\vartheta}^2 \qandq \pnorm2{(\nabla \f)(\theta)} \leq L \pnorm2{\theta-\vartheta},
\end{equation}
and let $\Theta \colon \N_0 \to \R^\d$ satisfy for all $n \in \N$ that 
\begin{equation}
\Theta_0 = \xi \qandq \Theta_n = \Theta_{n-1} - \gamma (\nabla \f)(\Theta_{n-1})
\end{equation}
\cfload.
Then
\begin{enumerate}[label=(\roman *)]
\item \label{GD_convergence_constant_optimal:item1}
it holds that 
$
\{\theta \in \mathbb{B}  \colon  \f(\theta) =  \inf\nolimits_{w \in \mathbb{B}} \f(w) \} = \{\vartheta \},
$
\item \label{GD_convergence_constant_optimal:item2}
it holds that $0\leq 1-2c\gamma + \gamma^2L^2<1$, 
\item \label{GD_convergence_constant_optimal:item3}
it holds for all $n \in \N_0$ that
\begin{equation}
\label{GD_convergence_constant_optimal:eq1}
\pnorm2{\Theta_n-\vartheta} \leq \br*{1-2c\gamma + \gamma^2L^2}^{\nicefrac{n}{2}} \pnorm2{\xi - \vartheta},
\end{equation} 
and
\item \label{GD_convergence_constant_optimal:item4}
it holds for all $n \in \N_0$ that
\begin{equation}
0 \leq \f(\Theta_n)-\f(\vartheta) 
\leq 
\tfrac{ L }{ 2 } 
\pnorm2{ \Theta_n - \vartheta }^2 
\leq  \tfrac{L}{2}  \br*{1-2c\gamma + \gamma^2L^2}^n   \pnorm2{\xi - \vartheta}^2.
\end{equation}
\end{enumerate}
\end{cor}

\begin{proof}[Proof of Corollary \ref{GD_convergence_constant_optimal}]
Observe that \cref{fcond:item3} in \cref{fcond} proves \cref{GD_convergence_constant_optimal:item2}.
In addition, note that \cref{GD_convergence} establishes 
\cref{GD_convergence_constant_optimal:item1,GD_convergence_constant_optimal:item3,GD_convergence_constant_optimal:item4}.
The proof of Corollary \ref{GD_convergence_constant_optimal} is thus complete.
\end{proof}
\endgroup

\cref{GD_convergence_constant_optimal} above establishes 
under suitable hypotheses convergence of the considered \GD\ process 
in the case where the learning rates are constant and strictly smaller 
than $ \frac{ 2 c }{ L^2 } $. The next result, \cref{GD_convergence_sharp} below, 
demonstrates that the condition that the learning rates are strictly smaller 
than $ \frac{ 2 c }{ L^2 } $ in \cref{GD_convergence_constant_optimal} 
can, in general, not be relaxed.

\cfclear
\begingroup
\providecommand{\d}{}
\renewcommand{\d}{\defaultParamDim}
\providecommand{\f}{}
\renewcommand{\f}{\defaultLossFunction}
\providecommand{\g}{}
\renewcommand{\g}{\defaultGradientFunction}
\begin{theorem}[Sharp bounds on the learning rate for the convergence of \GD]
\label{GD_convergence_sharp}
Let $\d \in \N$, $\alpha \in (0,\infty)$, $\gamma \in \R$, $\vartheta \in \R^\d$, $\xi \in \R^\d \backslash \{ \vartheta \}$,
let $\f \colon \R^\d \to \R$ satisfy for all $\theta \in \R^\d$ that
\begin{equation}
\label{GD_convergence_sharp:ass1}
\f(\theta) = \tfrac{\alpha}{2} \pnorm2{\theta-\vartheta}^2,
\end{equation}
and let $\Theta \colon \N_0 \to \R^\d$ satisfy for all $n \in \N$ that
\begin{equation}
\label{GD_convergence_sharp:ass2}
\Theta_0 = \xi \qandq \Theta_n = \Theta_{n-1} - \gamma (\nabla \f)(\Theta_{n-1})
\end{equation}
\cfload.
Then 
\begin{enumerate}[label=(\roman *)]
\item \label{GD_convergence_sharp:item1}
it holds for all $\theta \in \R^\d$ that
$
\scp{\theta-\vartheta, (\nabla \f)(\theta) } = \alpha  \pnorm2{\theta-\vartheta}^2,
$

\item \label{GD_convergence_sharp:item2}
it holds for all $\theta \in \R^\d$ that
$
\pnorm2{(\nabla \f)(\theta)} = \alpha   \pnorm2{\theta-\vartheta},
$

\item \label{GD_convergence_sharp:item3}
it holds for all $n \in \N_0$ that
$
\pnorm2{\Theta_n-\vartheta}  = \abs{1-\gamma \alpha}^n\pnorm2{\xi- \vartheta}
$,
and

\item \label{GD_convergence_sharp:item4}
it holds that
\begin{equation}
\liminf_{n \to \infty} \pnorm2{\Theta_n-\vartheta} = \limsup_{n \to \infty} \pnorm2{\Theta_n-\vartheta} = 
\begin{cases}
0 &\colon \gamma \in (0,\nicefrac{2}{\alpha}) \\
\pnorm2{\xi-\vartheta} &\colon \gamma \in \{0, \nicefrac{2}{\alpha} \} \\
\infty &\colon \gamma \in \R\backslash[0,\nicefrac2\alpha]
\end{cases}
\end{equation}
\end{enumerate}
\cfout.
\end{theorem}

\begin{proof}[Proof of \cref{GD_convergence_sharp}]
First of all, \nobs that \cref{der_of_norm} ensures that for all $\theta \in \R^\d$ it holds that $\f \in C^{\infty}(\R^\d,\R)$ and 
\begin{equation}
\label{GD_convergence_sharp:eq1}
(\nabla \f)(\theta) = \tfrac{\alpha}{2} (2 (\theta-\vartheta)) = \alpha(\theta - \vartheta).
\end{equation}
This proves \cref{GD_convergence_sharp:item2}.
Moreover, observe that (\ref{GD_convergence_sharp:eq1}) assures that for all $\theta \in \R^\d$ it holds that
\begin{equation}
\scp{\theta-\vartheta, (\nabla \f)(\theta) } = \scp{\theta-\vartheta, \alpha (\theta-\vartheta) } =  \alpha \pnorm2{\theta-\vartheta}^2
\end{equation}
\cfload.
This establishes \cref{GD_convergence_sharp:item1}.
\Nobs that \eqref{GD_convergence_sharp:ass2} and \eqref{GD_convergence_sharp:eq1} demonstrate that for all $n \in \N$ it holds that
\begin{equation}
\begin{split}
\Theta_n -\vartheta
&=
\Theta_{n-1} - \gamma (\nabla \f)(\Theta_{n-1})  - \vartheta\\
&=
\Theta_{n-1} - \gamma \alpha (\Theta_{n-1}- \vartheta)  - \vartheta \\
&= (1-\gamma \alpha)(\Theta_{n-1}- \vartheta).
\end{split}
\end{equation}
The assumption that $\Theta_0=\xi$ and induction 
hence prove that for all $n \in \N_0$ it holds that
\begin{equation}
\Theta_n -\vartheta = (1-\gamma \alpha)^n(\Theta_{0}- \vartheta) = (1-\gamma \alpha)^n(\xi- \vartheta).
\end{equation}
Therefore, we obtain for all $n \in \N_0$ that
\begin{equation}
\pnorm2{\Theta_n -\vartheta} = \abs{1-\gamma \alpha}^n\pnorm2{\xi- \vartheta}.
\end{equation}
This establishes \cref{GD_convergence_sharp:item3}.
Combining \cref{GD_convergence_sharp:item3} with 
the fact that for all $t \in (0,\nicefrac{2}{\alpha})$ it holds that $\abs{1-t\alpha} \in [0,1)$, 
the fact that for all $t \in \{0, \nicefrac{2}{\alpha}\}$ it holds that $\abs{1-t\alpha} = 1$,
the fact that for all $t \in \R \backslash [0,\nicefrac{2}{\alpha}]$ it holds that $\abs{1-t\alpha} \in (1,\infty)$,
and the fact that $\pnorm2{\xi- \vartheta} > 0$ 
establishes \cref{GD_convergence_sharp:item4}.
The proof of \cref{GD_convergence_sharp} is thus complete.
\end{proof}
\endgroup

\begingroup
\providecommand{\d}{}
\renewcommand{\d}{\defaultParamDim}
\providecommand{\f}{}
\renewcommand{\f}{\defaultLossFunction}
\providecommand{\g}{}
\renewcommand{\g}{\defaultGradientFunction}
\begin{exercise}{quest:summable_learningrates}
Let $\f \colon \R \to \R$ satisfy for all $\theta \in \R$ that
\begin{equation}
\f(\theta) = 2 \theta^2
\end{equation}
and let $\Theta \colon \N_0 \to \R$ satisfy for all $n \in \N$ that $\Theta_0 = 1$ and
\begin{equation}
\begin{split}
\Theta_n 
=
\Theta_{n-1} - n^{-2} (\nabla \f)(\Theta_{n-1}).
\end{split}
\end{equation}
Prove or disprove the following statement: 
It holds that
\begin{equation}
\limsup_{n \to \infty} |\Theta_n| = 0.
\end{equation}
\end{exercise}

\endgroup

\begingroup
\providecommand{\d}{}
\renewcommand{\d}{\defaultParamDim}
\providecommand{\f}{}
\renewcommand{\f}{\defaultLossFunction}
\providecommand{\g}{}
\renewcommand{\g}{\defaultGradientFunction}
\begin{exercise}{quest:summable_learningrates_2}
Let $\f \colon \R \to \R$ satisfy for all $\theta \in \R$ that
\begin{equation}
\f(\theta) = 4 \theta^2
\end{equation}
and 
for every $r \in (1,\infty)$
let $\Theta^{(r)} \colon \N_0 \to \R$ satisfy for all $n \in \N$ that $\Theta^{(r)}_0 = 1$ and
\begin{equation}
\begin{split}
\Theta^{(r)}_n 
=
\Theta^{(r)}_{n-1} - n^{-r} (\nabla \f)(\Theta^{(r)}_{n-1}).
\end{split}
\end{equation}
Prove or disprove the following statement: 
It holds for all $r \in (1,\infty)$ that
\begin{equation}
\liminf_{n \to \infty} |\Theta^{(r)}_n| > 0.
\end{equation}

\end{exercise}

\endgroup

\begingroup
\providecommand{\d}{}
\renewcommand{\d}{\defaultParamDim}
\providecommand{\f}{}
\renewcommand{\f}{\defaultLossFunction}
\providecommand{\g}{}
\renewcommand{\g}{\defaultGradientFunction}
\begin{exercise}{quest:summable_learningrates_3}
Let $\f \colon \R \to \R$ satisfy for all $\theta \in \R$ that
\begin{equation}
\f(\theta) = 5 \theta^2
\end{equation}
and 
for every $r \in (1,\infty)$ 
let $\Theta^{(r)} = (\Theta^{(r)}_n)_{n \in \N_0} \colon \N_0 \to \R$ satisfy for all $n \in \N$ that $\Theta^{(r)}_0 = 1$ and
\begin{equation}
\begin{split}
	\Theta^{(r)}_n 
=
	\Theta^{(r)}_{n-1} - n^{-r} (\nabla \f)(\Theta^{(r)}_{n-1}).
\end{split}
\end{equation}
Prove or disprove the following statement: 
It holds for all $r \in (1,\infty)$ that
\begin{equation}
\liminf_{n \to \infty} |\Theta^{(r)}_n| > 0.
\end{equation}

\end{exercise}

\endgroup

\subsubsection{Convergence rates}
\label{sec:convergence_rates}

The next result, \cref{GD_convergence_nonexplicit} below, 
establishes a convergence rate for 
the \GD\ optimization method in the case 
of possibly non-constant learning rates. 
We prove \cref{GD_convergence_nonexplicit} through 
an application of \cref{GD_convergence} above.

\cfclear
\begingroup
\providecommand{\d}{}
\renewcommand{\d}{\defaultParamDim}
\providecommand{\f}{}
\renewcommand{\f}{\defaultLossFunction}
\providecommand{\g}{}
\renewcommand{\g}{\defaultGradientFunction}
\begin{cor}[Qualitative convergence of \GD]
\label{GD_convergence_nonexplicit}
Let 
$ \d \in \N $, 
$ \f \in C^1(\R^\d,\R) $,  
$ (\gamma_n )_{ n \in \N } \subseteq \R $, 
$ c, L \in (0,\infty) $,  
$ \xi, \vartheta \in \R^\d $ 
satisfy for all $\theta \in \R^\d$ that
\begin{equation}
\label{GD_convergence_nonexplicit:assumption1}
\scp{\theta-\vartheta, (\nabla \f)(\theta) } \geq  c \pnorm2{\theta-\vartheta}^2 ,\qquad \pnorm2{(\nabla \f)(\theta)} \leq L \pnorm2{\theta-\vartheta},
\end{equation}
\begin{equation}
\label{GD_convergence_nonexplicit:assumption2}
\andq 0 < \liminf_{n \to \infty} \gamma_n \leq \limsup_{n \to \infty} \gamma_n < \tfrac{2c}{L^2},
\end{equation}
and let $\Theta \colon \N_0 \to \R^\d$ satisfy for all $n \in \N$ that 
\begin{equation}
\label{GD_convergence_nonexplicit:assumption3}
\Theta_0 = \xi \qandq \Theta_n = \Theta_{n-1} - \gamma_n (\nabla \f)(\Theta_{n-1})
\end{equation}
\cfload.
Then
\begin{enumerate}[label=(\roman *)]
\item \label{GD_convergence_nonexplicit:item1}
it holds that $\{\theta \in \R^\d  \colon  \f(\theta)= \inf\nolimits_{w \in \R^\d} \f(w)  \} = \{\vartheta \}$,
\item \label{GD_convergence_nonexplicit:item2}
there exist $ \epsilon \in (0,1)$, $C \in \R$ such that for all $n \in \N_0$ it holds that
\begin{equation}
\pnorm2{\Theta_n - \vartheta} \leq \epsilon^n C ,
\end{equation}
and
\item \label{GD_convergence_nonexplicit:item3}
there exist $ \epsilon \in (0,1)$, $C \in \R$ such that for all $n \in \N_0$ it holds that
\begin{equation}
0 \leq \f(\Theta_n)-\f(\vartheta) \leq \epsilon^n C .
\end{equation}
\end{enumerate}
\end{cor}

\begin{proof}[Proof of Corollary \ref{GD_convergence_nonexplicit}]
Throughout this proof, let $\alpha, \beta \in \R$ satisfy 
\begin{equation}
0 < \alpha < \liminf_{n \to \infty} \gamma_n \leq \limsup_{n \to \infty} \gamma_n  < \beta < \tfrac{2c}{L^2}
\end{equation}
(cf. \eqref{GD_convergence_nonexplicit:assumption2}), let $m \in \N$ satisfy for all $n \in \N$ that $\gamma_{m+n} \in [\alpha,\beta]$, and
let $h \colon \R \to \R$ satisfy for all $t \in \R$ that 
\begin{equation}
h(t) = 1-2ct + t^2 L^2.
\end{equation}
Observe that (\ref{GD_convergence_nonexplicit:assumption1}) and item~\ref{fcond1:item2} in \cref{fcond1} 
prove \cref{GD_convergence_nonexplicit:item1}.
In addition, observe that the fact that for all $t \in \R$ it holds that $h'(t) = -2c + 2t L^2$ implies that for all $ t \in (-\infty,\frac{c}{L^2}]$ 
it holds that 
\begin{equation}
h'(t) \leq -2c + 2 \br*{\tfrac{c}{L^2} } L^2 = 0.
\end{equation}
The fundamental theorem of calculus hence assures that for all $t \in [\alpha,\beta] \cap (-\infty,\frac{c}{L^2}]$ it holds that
\begin{equation}
\label{GD_convergence_nonexplicit:eq1}
h(t) = h(\alpha) + \int_{\alpha}^t h'(s) \, \diff s \leq h(\alpha) + \int_{\alpha}^t 0 \, \diff s = h(\alpha)
\leq \max\{h(\alpha),h(\beta)\}.
\end{equation}
Furthermore, observe that the fact that for all $t \in \R$ it holds that $h'(t) = -2c + 2t L^2$ implies that for all $ t \in [\frac{c}{L^2},\infty)$ 
it holds that 
\begin{equation}
  h'(t) \geq  h'( \tfrac{ c }{ L^2 } ) = - 2 c + 2 \br*{\tfrac{c}{L^2} } L^2 = 0.
\end{equation}
The fundamental theorem of calculus hence ensures that for all $t \in [\alpha,\beta] \cap [\frac{c}{L^2},\infty)$ it holds that
\begin{equation}
\label{GD_convergence_nonexplicit:eq2}
\max\{h(\alpha),h(\beta)\}\geq
h(\beta) = h(t) + \int_{t}^\beta h'(s) \, \diff s \geq h(t) + \int_{t}^\beta 0 \, \diff s = h(t).
\end{equation}
Combining this and (\ref{GD_convergence_nonexplicit:eq1}) establishes that for all $t \in [\alpha,\beta]$ it holds that
\begin{equation}
\label{GD_convergence_nonexplicit:eq3}
h(t) \leq \max \{h(\alpha), h(\beta) \}.
\end{equation}
Moreover, observe that the fact that $\alpha,\beta \in (0,\tfrac{2c}{L^2})$ and \cref{fcond:item3} in \cref{fcond} ensure that
\begin{equation}
\{h(\alpha), h(\beta)\}\subseteq [0,1).
\end{equation}
Hence, we obtain that
\begin{equation}
\max \{h(\alpha), h(\beta) \} \in [0,1).
\end{equation}
This implies that there exists $\varepsilon \in \R$ such that 
\begin{equation}
0\leq \max \{h(\alpha), h(\beta) \} < \varepsilon < 1.
\end{equation}
Next note that the fact 
that for all $n \in  \N$ it holds that $\gamma_{m+n} \in [\alpha,\beta] \subseteq [0,\frac{2c}{L^2}]$, 
\cref{GD_convergence:item2,GD_convergence:item4} in \cref{GD_convergence} 
(applied with $ \d \is \d $, $ c \is c $, 
$ L \is L $, $ r \is \infty $, 
$ ( \gamma_n )_{ n \in \N } \is ( \gamma_{ m + n } )_{ n \in \N } $, 
$ \vartheta \is \vartheta $, 
$ \xi \is \Theta_m $, 
$ \f \is \f $ in the notation of 
\cref{GD_convergence}), 
\eqref{GD_convergence_nonexplicit:assumption1}, \eqref{GD_convergence_nonexplicit:assumption3},  
and \eqref{GD_convergence_nonexplicit:eq3} demonstrate that for all $n \in \N$ it holds that
\begin{equation}
\begin{split}
\pnorm2{\Theta_{m+n}-\vartheta} 
&\leq 
\br*{\prod_{k = 1}^{n} (1-2c\gamma_{m+k} + (\gamma_{m+k})^2 L^2)^{\nicefrac{1}{2}}} \pnorm2{\Theta_m - \vartheta} \\
&= 
\br*{\prod_{k = 1}^{n} (h(\gamma_{m+k}))^{\nicefrac{1}{2}}} \pnorm2{\Theta_m - \vartheta} \\
&\leq
(\max \{h(\alpha), h(\beta) \})^{\nicefrac{n}{2}} \pnorm2{\Theta_m - \vartheta} \\
&\leq 
\varepsilon^{\nicefrac{n}{2}} \pnorm2{\Theta_m - \vartheta}.
\end{split}
\end{equation}
This shows that for all $n \in \N$ with $n>m$ it holds that
\begin{equation}
\pnorm2{\Theta_{n}-\vartheta}  \leq \varepsilon^{\nicefrac{(n-m)}{2}}\pnorm2{\Theta_m - \vartheta}.
\end{equation}
The fact that for all $n \in \N_0$ with $n \leq m$ it holds that
\begin{equation}
\pnorm2{\Theta_{n}-\vartheta} 
= 
\br*{\frac{\pnorm2{\Theta_{n}-\vartheta} }{\varepsilon^{\nicefrac{n}{2}}} } \varepsilon^{\nicefrac{n}{2}}
\leq
\br*{\max\cu*{ \frac{\pnorm2{\Theta_{k}-\vartheta}}{\varepsilon^{\nicefrac{k}{2}}} \colon k \in \{0,1,\ldots, m\} } }\varepsilon^{\nicefrac{n}{2}}
\end{equation}
hence assures that for all $n \in \N_0$ it holds that
\begin{equation}
\label{GD_convergence_nonexplicit:eq4}
\begin{split}
\pnorm2{\Theta_{n}-\vartheta} 
&\leq 
\max \cu*{ \br*{ \max\cu*{ \frac{\pnorm2{\Theta_{k}-\vartheta}}{\varepsilon^{\nicefrac{k}{2}}} \colon k \in \{0,1,\ldots, m\} } } \varepsilon^{\nicefrac{n}{2}}, \varepsilon^{\nicefrac{(n - m)}{2}} \pnorm2{\Theta_m - \vartheta}} \\
&=
(\varepsilon^{\nicefrac{1}{2}})^n \br*{ \max \cu*{  \max\cu*{ \frac{\pnorm2{\Theta_{k}-\vartheta}}{\varepsilon^{\nicefrac{k}{2}}} \colon k \in \{0,1,\ldots, m\} } , \varepsilon^{-\nicefrac{m}{2}} \pnorm2{\Theta_m - \vartheta}} }
\\ & 
=
(\varepsilon^{\nicefrac{1}{2}})^n 
\br*{ 
  \max\cu*{ 
    \frac{ 
      \pnorm2{ \Theta_k - \vartheta } 
    }{ 
      \varepsilon^{ \nicefrac{ k }{ 2 } } 
    } 
    \colon k \in \{0,1,\ldots, m\} 
  } 
}
.
\end{split}
\end{equation}
This proves \cref{GD_convergence_nonexplicit:item2}.
In addition, note that 
\cref{fcond2}, \cref{GD_convergence_nonexplicit:item1}, and \eqref{GD_convergence_nonexplicit:eq4} 
assure that for all $n \in \N_0$ it holds that
\begin{equation}
0 
\leq 
\f(\Theta_n)-\f(\vartheta) 
\leq 
\tfrac{L}{2}\pnorm2{\Theta_n-\vartheta}^2 
\leq 
 \frac{\varepsilon^{n}L}{2} 
\br*{ 
  \max\cu*{ 
    \frac{ 
      \pnorm2{ \Theta_k - \vartheta }^2 
    }{ 
      \varepsilon^{ k } 
    } 
    \colon k \in \{0,1,\ldots, m\} 
  } 
}
   .
\end{equation}
This establishes \cref{GD_convergence_nonexplicit:item3}. The proof of Corollary \ref{GD_convergence_nonexplicit} is thus complete.
\end{proof}
\endgroup

\subsubsection{Error estimates in the case of small learning rates}
\label{sec:small_learning_rates}

The inequality in \eqref{GD_convergence:eq2} in \cref{GD_convergence:item4}
in \cref{GD_convergence} above provides us an error estimate 
for the \GD\ optimization method in the case where the learning rates 
$ ( \gamma_n )_{ n \in \N } $ 
in \cref{GD_convergence}
satisfy 
that for all $ n \in \N $ it holds that
$
  \gamma_n \leq \frac{ 2 c }{ L^2 }
$.
The error estimate in \eqref{GD_convergence:eq2} can be simplified 
in the special case where the learning rates 
$ ( \gamma_n )_{ n \in \N } $ 
satisfy the more restrictive condition that
for all $ n \in \N $ it holds that
$
  \gamma_n \leq \frac{ c }{ L^2 }
$. 
This is the subject of the next result, \cref{GD_convergence2} below.
We prove \cref{GD_convergence2} through 
an application of \cref{GD_convergence} above.

\cfclear
\begingroup
\providecommand{\d}{}
\renewcommand{\d}{\defaultParamDim}
\providecommand{\f}{}
\renewcommand{\f}{\defaultLossFunction}
\providecommand{\g}{}
\renewcommand{\g}{\defaultGradientFunction}
\begin{cor}[Error estimates in the case of small learning rates]
\label{GD_convergence2}
Let $ \d \in \N $, 
$ c, L \in (0,\infty) $, 
$ r \in (0,\infty] $, 
$ ( \gamma_n )_{ n \in \N } \subseteq [0,\frac{c}{L^2}] $, 
$\vartheta \in \R^\d$, 
$ \mathbb{B} = \{ w \in \R^\d \colon \pnorm2{ w - \vartheta } \leq r \} $, 
$\xi \in \mathbb{B}$, $\f \in C^1(\R^\d,\R)$ satisfy for all $\theta \in \mathbb{B}$ that
\begin{equation}
\scp{\theta-\vartheta, (\nabla \f)(\theta) } \geq  c \pnorm2{\theta-\vartheta}^2 \qandq \pnorm2{(\nabla \f)(\theta)} \leq L \pnorm2{\theta-\vartheta},
\end{equation}
and let $\Theta \colon \N_0 \to \R^\d$ satisfy for all $n \in \N$ that 
\begin{equation}
\Theta_0 = \xi \qandq \Theta_n = \Theta_{n-1} - \gamma_n (\nabla \f)(\Theta_{n-1})
\end{equation}
\cfload.
Then
\begin{enumerate}[label=(\roman *)]
\item \label{GD_convergence2:item1}
it holds that $\{\theta \in \mathbb{B}  \colon  \f(\theta)= \inf\nolimits_{w \in \mathbb{B}} \f(w)  \} = \{\vartheta \}$,
\item \label{GD_convergence2:item2}
it holds for all $n \in \N$ that $0\leq 1 - c \gamma_n\leq 1$, 
\item \label{GD_convergence2:item3}
it holds for all $n \in \N_0$ that
\begin{equation}
  \pnorm2{\Theta_n-\vartheta} 
\leq 
  \br*{ {\textstyle \prod\limits_{k = 1}^n (1-c\gamma_k)^{\nicefrac{1}{2}} }} \! \pnorm2{\xi - \vartheta}
\leq
  \exp\bigl( 
  \textstyle 
    - \frac{ c }{ 2 } \bigl[ \sum_{ k = 1 }^n \gamma_k \bigr] 
  \displaystyle 
  \bigr)
  \pnorm2{ \xi - \vartheta }
  ,
\end{equation}
and
\item \label{GD_convergence2:item4}
it holds for all $n \in \N_0$ that
\begin{equation}
  0 
\leq 
  \f(\Theta_n)-\f(\vartheta) 
\leq 
  \tfrac{L}{2}
  \br*{ {\textstyle \prod\limits_{k = 1}^n (1-c\gamma_k) }} 
  \! \pnorm2{\xi - \vartheta}^2
\leq 
  \tfrac{L}{2}
  \exp\bigl( 
  {\textstyle 
    - c \bigl[ \sum_{ k = 1 }^n \gamma_k \bigr] 
  }
  \bigr)
  \pnorm2{ \xi - \vartheta }^2
.
\end{equation}
\end{enumerate}
\end{cor}

\begin{proof}[Proof of Corollary \ref{GD_convergence2}]
Note that \cref{GD_convergence:item2} in \cref{GD_convergence} and the assumption that for all $n \in \N$ it holds that $\gamma_n \in [0,\frac{c}{L^2}]$ ensure that for all $n \in \N$ it holds that
\begin{equation}
\label{GD_convergence2:eq1}
0 \leq 1-2c\gamma_n + (\gamma_n)^2L^2 \leq 1-2c\gamma_n + \gamma_n\br*{\frac{c}{L^2} }L^2 = 1-2c\gamma_n + \gamma_n c  =  1 - c\gamma_n \leq 1. 
\end{equation}
This proves \cref{GD_convergence2:item2}. 
Moreover, note that (\ref{GD_convergence2:eq1}) and 
\cref{GD_convergence} establish 
\cref{GD_convergence2:item1,GD_convergence2:item3,GD_convergence2:item4}.
The proof of Corollary \ref{GD_convergence2} is thus complete.
\end{proof}
\endgroup

In the next result, \cref{GD_convergence_constant} below, 
we, roughly speaking, specialize \cref{GD_convergence2} above 
to the case where the learning rates 
$ ( \gamma_n )_{ n \in \N } \subseteq [0, \frac{ c }{ L^2 } ] $
are a constant sequence.

\cfclear
\begingroup
\providecommand{\d}{}
\renewcommand{\d}{\defaultParamDim}
\providecommand{\f}{}
\renewcommand{\f}{\defaultLossFunction}
\providecommand{\g}{}
\renewcommand{\g}{\defaultGradientFunction}
\begin{cor}[Error estimates in the case of small and constant learning rates]
\label{GD_convergence_constant}
Let $\d \in \N$, 
$c,L \in (0,\infty)$, $r \in (0,\infty]$, $\gamma\in (0,\frac{c}{L^2}]$, $\vartheta \in \R^\d$, $\mathbb{B} = \{w \in \R^\d \colon \pnorm2{w-\vartheta} \leq r \}$, $\xi \in \mathbb{B} $, $\f \in C^1(\R^\d,\R)$ satisfy for all $\theta \in \mathbb{B}$ that
\begin{equation}
\label{GD_convergence_constant:assumption1}
\scp{\theta-\vartheta, (\nabla \f)(\theta) } \geq  c \pnorm2{\theta-\vartheta}^2 \qandq \pnorm2{(\nabla \f)(\theta)} \leq L \pnorm2{\theta-\vartheta},
\end{equation}
and let $\Theta \colon \N_0 \to \R^\d$ satisfy for all $n \in \N$ that 
\begin{equation}
\Theta_0 = \xi \qandq \Theta_n = \Theta_{n-1} - \gamma (\nabla \f)(\Theta_{n-1})
\end{equation}
\cfload.
Then
\begin{enumerate}[label=(\roman *)]
\item \label{GD_convergence_constant:item1}
it holds that
$
\{\theta \in \mathbb{B}  \colon  \f(\theta) =  \inf\nolimits_{w \in \mathbb{B}} \f(w) \} = \{\vartheta \}
$,
\item \label{GD_convergence_constant:item2}
it holds that $0\leq 1 - c \gamma< 1$, 
\item \label{GD_convergence_constant:item3}
it holds for all $n \in \N_0$ that
$
\pnorm2{\Theta_n-\vartheta} \leq (1-c\gamma)^{\nicefrac{n}{2}} \pnorm2{\xi - \vartheta}
$,
and
\item \label{GD_convergence_constant:item4}
it holds for all $n \in \N_0$ that
$
0 \leq \f(\Theta_n)-\f(\vartheta) \leq \tfrac{L}{2} \, (1-c\gamma)^n  \, \pnorm2{\xi - \vartheta}^2.
$
\end{enumerate}
\end{cor}

\begin{proof}[Proof of \cref{GD_convergence_constant}]
\cref{GD_convergence_constant} is an immediate consequence of
\cref{GD_convergence2}.
The proof of \cref{GD_convergence_constant} is thus complete.
\end{proof}
\endgroup

\subsubsection{On the spectrum of the Hessian of the objective function at a local minimum point}

A crucial ingredient in our error analysis for the \GD\ optimization method in 
\cref{sec:error_estimates_GD,sec:size_of_learningrates,sec:convergence_rates,sec:small_learning_rates} above 
is to employ the growth and the coercivity-type hypotheses, \eg, 
in \eqref{GD_convergence:assumption1} in \cref{GD_convergence} above. 
In this subsection we disclose in \cref{conditions_nesterov_vs_coercivity} below
suitable 
conditions on the Hessians 
of the objective function of the considered optimization problem which are sufficient 
to ensure that \eqref{GD_convergence:assumption1} is satisfied so that we are in the 
position to apply the error analysis in 
\cref{sec:error_estimates_GD,sec:size_of_learningrates,sec:convergence_rates,sec:small_learning_rates} above
(cf.\ \cref{GD_convergence_Nesterov} below). 
Our proof of \cref{conditions_nesterov_vs_coercivity} employs the following classical result (see \cref{norm_sym_matrices} below) for 
symmetric matrices with real entries.

\cfclear
\begingroup
\providecommand{\d}{}
\renewcommand{\d}{\defaultParamDim}
\providecommand{\f}{}
\renewcommand{\f}{\defaultLossFunction}
\providecommand{\g}{}
\renewcommand{\g}{\defaultGradientFunction}
\begin{lemma}[Properties of the spectrum of real symmetric matrices]
\label{norm_sym_matrices}
Let $\d \in \N$, 
let $A \in \R^{\d \times \d}$ be a symmetric matrix, and let  
\begin{equation}
  \mathscr{S} = \{\lambda \in \C \colon (\exists \, v \in \C^\d\backslash \{0\} \colon Av = \lambda v) \} 
  .
\end{equation}
Then
\begin{enumerate}[label=(\roman *)]
\item \label{norm_sym_matrices:item1}
it holds that $\mathscr{S} = \{\lambda \in \R \colon (\exists \, v \in \R^\d\backslash \{0\} \colon Av = \lambda v) \}  \subseteq \R$,
\item \label{norm_sym_matrices:item2}
it holds that
\begin{equation}
\sup_{v\in \R^\d\backslash \{0\}} \br*{\frac{\pnorm2{Av}}{\pnorm2{v}}} 
= 
\max_{\lambda \in \mathscr{S}}\abs{\lambda},
\end{equation}
and
\item \label{norm_sym_matrices:item3}
it holds for all $v \in \R^\d$ that
\begin{equation}
\min (\mathscr{S}) \pnorm2{v}^2 \leq \scp{ v , Av } \leq \max (\mathscr{S}) \pnorm2{v}^2
\end{equation}
\end{enumerate}
\cfout.
\end{lemma}

\begin{proof}[Proof of \cref{norm_sym_matrices}]
Throughout this proof, let $e_1,e_2,\ldots,e_\d \in \R^\d$ be the vectors given by 
\begin{equation}
e_1 = (1,0,\ldots,0), \qquad e_2 = (0,1,0,\ldots,0), \qquad \ldots, \qquad e_\d = (0,\ldots,0,1).
\end{equation}
Observe that the spectral theorem for symmetric matrices (see, \eg, Petersen~\cite[Theorem 4.3.4]{Petersen12}) 
proves that there exist $ ( \d \times \d ) $-matrices $\Lambda = (\Lambda_{i,j})_{(i,j) \in \{1, 2, \ldots, \d \}^2}$, $O = (O_{i,j})_{(i,j) \in \{1, 2, \ldots, \d \}^2} \in \R^{\d \times \d}$ such that 
$\mathscr{S} = \{\Lambda_{1,1},\Lambda_{2,2},\ldots,\Lambda_{\d,\d} \}$, 
$O^*O = OO^* = \idMatrix_{\d}$, 
$A = O \Lambda O^*$,
and
\begin{equation}
\label{norm_sym_matrices:eq1}
\Lambda 
=  
\begin{pmatrix}
\Lambda_{1,1} & &0 \\
& \ddots & \\
0 & & \Lambda_{\d,\d} 
\end{pmatrix} \in \R^{\d \times \d}\ifnocf.
\end{equation}
\cfload[.]%
Hence, we obtain that $\mathscr{S} \subseteq \R$.
Next note that the assumption that $ \mathscr{S} = \{\lambda \in \C \colon (\exists \, v \in \C^\d\backslash \{0\} \colon Av = \lambda v) \}$ ensures that for every $\lambda \in \mathscr{S}$ there exists $v \in \C^\d\backslash \{0\}$ such that 
\begin{equation}
A\Real{v} + {\bf i} A\Ima{v} = Av  = \lambda v = \lambda \Real{v}+ {\bf i}\lambda \Ima{v}.
\end{equation} 
The fact that $\mathscr{S} \subseteq \R$ therefore demonstrates that for every $\lambda \in \mathscr{S}$ there exists $v \in \R^\d\backslash \{0\}$ such that $Av = \lambda v$. 
This and the fact that $\mathscr{S} \subseteq \R$ ensure that $\mathscr{S} \subseteq \{\lambda \in \R \colon (\exists \, v \in \R^\d\backslash \{0\} \colon Av = \lambda v) \}$.
Combining this and the fact that $\{\lambda \in \R \colon (\exists \, v \in \R^\d\backslash \{0\} \colon Av = \lambda v) \} \subseteq \mathscr{S}$ 
proves \cref{norm_sym_matrices:item1}. 
Furthermore, note that (\ref{norm_sym_matrices:eq1}) assures that for all $v = (v_1,v_2,\ldots,v_\d) \in \R^\d$ it holds that
\begin{equation}
\begin{split}
\pnorm2{\Lambda v} &= \br*{\sum_{i = 1}^\d \abs{\Lambda_{i,i}v_i}^2}^{\nicefrac{1}{2}} 
\leq 
\br*{ \sum_{i = 1}^\d\max\bcu{\abs{\Lambda_{1,1}}^2,\ldots,\abs{\Lambda_{\d,\d}}^2 } \abs{v_i}^2}^{\nicefrac{1}{2}} \\
&= 
\bbbr{\max\bcu{\abs{\Lambda_{1,1}},\ldots,\abs{\Lambda_{\d,\d}} }^2\pnorm2{v}^2}^{\nicefrac{1}{2}}\\
&= 
\max\bcu{\abs{\Lambda_{1,1}},\ldots,\abs{\Lambda_{\d,\d}} }\pnorm2{v} \\
&= 
\bpr{ \max\nolimits_{\lambda \in \mathscr{S}}\abs{\lambda}} \pnorm2{v}
\end{split}
\end{equation}
\cfload.
The fact that $O$ is an orthogonal matrix and the fact that $A = O \Lambda O^*$ therefore imply that for all  $v  \in \R^\d$ it holds that
\begin{equation}
\begin{split}
\pnorm2{A v}& =\pnorm2{O \Lambda O^* v} = \pnorm2{\Lambda O^* v} \\
&\leq \bpr{ \max\nolimits_{\lambda \in \mathscr{S}}\abs{\lambda}}\pnorm2{O^*v} \\
&= \bpr{\max\nolimits_{\lambda \in \mathscr{S}}\abs{\lambda}}\pnorm2{v}.
\end{split}
\end{equation}
This implies that
\begin{equation}
\label{norm_sym_matrices:eq2}
\begin{split}
\sup_{v \in \R^\d\backslash \{0\}} \br*{\frac{\pnorm2{Av}}{\pnorm2{v}} }
\leq  
\sup_{v \in \R^\d\backslash \{0\}} \br*{\frac{\bpr{\max\nolimits_{\lambda \in \mathscr{S}}\abs{\lambda}}\pnorm2{v}}{\pnorm2{v}} }
=  
\max\nolimits_{\lambda \in \mathscr{S}}\abs{\lambda}.
\end{split}
\end{equation}
In addition, note that the fact that $\mathscr{S} = \{\Lambda_{1,1},\Lambda_{2,2}\ldots,\Lambda_{\d,\d} \}$ ensures that there exists $j \in \{1, 2, \ldots, \d\}$ such that 
\begin{equation}
\label{norm_sym_matrices:eq2.1}
\abs{\Lambda_{j,j}} =  \max\nolimits_{\lambda \in \mathscr{S}}\abs{\lambda}. 
\end{equation}
Next observe that the fact that $A = O\Lambda O^*$, the fact that $O$ is an orthogonal matrix, and \eqref{norm_sym_matrices:eq2.1} imply that 
\begin{equation}
\begin{split}
&\sup_{v \in \R^\d\backslash \{0\}} \br*{\frac{\pnorm2{Av}}{\pnorm2{v}} }
\geq
\frac{\pnorm2{AOe_j}}{\pnorm2{Oe_j}} = \pnorm2{O \Lambda O^*Oe_j}
= 
\pnorm2{O\Lambda e_j}\\
& = \pnorm2{\Lambda e_j} = \pnorm2{\Lambda_{j,j} e_j} 
= \abs{\Lambda_{j,j}} =  \max\nolimits_{\lambda \in \mathscr{S}}\abs{\lambda}.
\end{split}
\end{equation}
Combining this and \eqref{norm_sym_matrices:eq2} establishes \cref{norm_sym_matrices:item2}. 
It thus remains to prove \cref{norm_sym_matrices:item3}. 
For this note that \eqref{norm_sym_matrices:eq1} ensures that for all $v  = (v_1,v_2,\ldots,v_\d) \in \R^\d$ it holds that
\begin{equation}
\begin{split}
\scp{ v, \Lambda v } &= \sum_{i = 1}^\d \Lambda_{i,i}\abs{v_i}^2 \leq \sum_{i = 1}^\d  \max \{\Lambda_{1,1}, \ldots , \Lambda_{\d,\d} \}\abs{v_i}^2 \\
&= \max \{\Lambda_{1,1}, \ldots , \Lambda_{\d,\d} \}\pnorm2{v}^2 = \max( \mathscr{S})\pnorm2{v}^2
\end{split}
\end{equation}
\cfload.
The fact that $O$ is an orthogonal matrix and the fact that $A = O\Lambda O^*$ therefore demonstrate that for all $v \in \R^\d$ it holds that
\begin{equation}
\label{norm_sym_matrices:eq3}
\begin{split}
\scp{ v, A v } &= \scp{ v, O \Lambda O^* v } = \scp{ O^*v,  \Lambda O^* v } \\
&\leq \max( \mathscr{S})\pnorm2{O^* v}^2 = \max( \mathscr{S})\pnorm2{v}^2
.
\end{split}
\end{equation}
Moreover, observe that (\ref{norm_sym_matrices:eq1}) implies that for all $v  = (v_1,v_2,\ldots,v_\d) \in \R^\d$ it holds that
\begin{equation}
\begin{split}
\scp{ v, \Lambda v } &= \sum_{i = 1}^\d \Lambda_{i,i}\abs{v_i}^2 \geq \sum_{i = 1}^\d  \min \{\Lambda_{1,1}, \ldots , \Lambda_{\d,\d} \}\abs{v_i}^2 \\
&= \min \{\Lambda_{1,1}, \ldots , \Lambda_{\d,\d} \}\pnorm2{v}^2 = \min( \mathscr{S})\pnorm2{v}^2.
\end{split}
\end{equation}
The fact that $O$ is an orthogonal matrix and the fact that $A = O\Lambda O^*$  hence demonstrate that for all $v \in \R^\d$ it holds that
\begin{equation}
\begin{split}
\scp{ v, A v } &= \scp{ v, O \Lambda O^* v } = \scp{ O^*v,  \Lambda O^* v } \\
&\geq \min( \mathscr{S})\pnorm2{O^* v}^2 = \min( \mathscr{S})\pnorm2{v}^2.
\end{split}
\end{equation}
Combining this with (\ref{norm_sym_matrices:eq3}) establishes \cref{norm_sym_matrices:item3}. The proof of \cref{norm_sym_matrices} is thus complete.
\end{proof}
\endgroup

We now present the promised \cref{conditions_nesterov_vs_coercivity} which 
discloses suitable 
conditions 
(cf.\ \eqref{conditions_nesterov_vs_coercivity:assumption1} and \eqref{conditions_nesterov_vs_coercivity:assumption2} below)
on the Hessians 
of the objective function of the considered optimization problem which are sufficient 
to ensure that \eqref{GD_convergence:assumption1} is satisfied so that we are in the 
position to apply the error analysis in 
\cref{sec:error_estimates_GD,sec:size_of_learningrates,sec:convergence_rates,sec:small_learning_rates} above.

\cfclear
\begingroup
\newcommand{\Rdnorm}[2][x]{\normmm{#2}}
\newcommand{\rdnorm}[2][x]{\normmm{#2}}
\newcommand{\norm}[1]{\pnorm2{#1}}
\providecommand{\d}{}
\renewcommand{\d}{\defaultParamDim}
\providecommand{\f}{}
\renewcommand{\f}{\defaultLossFunction}
\providecommand{\g}{}
\renewcommand{\g}{\defaultGradientFunction}
\begin{prop}[Conditions on the spectrum of the Hessian of the objective function at a local minimum point]
\label{conditions_nesterov_vs_coercivity}
Let $\d \in \N$, 
let $\Rdnorm[\d \times \d]{\cdot} \colon \R^{\d \times \d} \to [0,\infty)$ 
satisfy for all $A \in \R^{\d \times \d}$ that $\rdnorm[\d \times \d]{A} = \sup_{v\in \R^\d\backslash \{0\}} \frac{\norm{Av}}{\norm{v}}$, and 
let $ \lambda, \alpha \in (0,\infty)$, $\beta \in [\alpha, \infty)$, $\vartheta\in \R^\d$, $\f \in C^2(\R^\d,\R)$ satisfy for all $v,w \in \R^\d$ that 
\begin{equation}
\label{conditions_nesterov_vs_coercivity:assumption1}
(\nabla \f)(\vartheta) = 0, \qquad \rdnorm[\d \times \d]{ (\Hess \f)(v) - (\Hess \f)(w)} \leq \lambda \norm{v- w},
\end{equation}
\begin{equation}
\label{conditions_nesterov_vs_coercivity:assumption2}
\andq  \{\mu \in \R \colon (\exists \, u \in \R^\d\backslash \{0\} \colon [(\Hess \f)(\vartheta)]  u = \mu u) \} \subseteq [\alpha,\beta]
\end{equation}
\cfload.
Then it holds for all $\theta \in \{w \in \R^\d \colon \norm{w-\vartheta} \leq \frac{\alpha}{\lambda} \}$ that
\begin{equation}
\label{conditions_nesterov_vs_coercivity:conclusion}
\scp{\theta-\vartheta, (\nabla \f)(\theta) } \geq \tfrac{\alpha}{2}\norm{\theta-\vartheta}^2 \qandq \norm{(\nabla \f)(\theta)} \leq \tfrac{3\beta}{2}\norm{\theta-\vartheta}
\end{equation}
\cfout.
\end{prop}

\begin{proof}[Proof of \cref{conditions_nesterov_vs_coercivity}]
Throughout this proof, let $ \mathbb{B} \subseteq \R^\d $ be the set given by 
\begin{equation}
\mathbb{B} = \cu*{w \in \R^\d \colon \norm{w-\vartheta} \leq \tfrac{\alpha}{\lambda} }
\end{equation}
and let $\mathscr{S} \subseteq \C$ be the set given by 
\begin{equation}
\mathscr{S} = \{\mu \in \C \colon (\exists \, u \in \C^\d\backslash \{0\} \colon [(\Hess \f)(\vartheta)]u = \mu u) \}.
\end{equation}
Note that 
the fact that $(\Hess \f)(\vartheta) \in \R^{\d \times \d}$ is a symmetric matrix, 
\cref{norm_sym_matrices:item1} in \cref{norm_sym_matrices}, 
and \eqref{conditions_nesterov_vs_coercivity:assumption2} 
imply that
\begin{equation}
\label{conditions_nesterov_vs_coercivity:eq0}
\mathscr{S} = \{\mu \in \R \colon (\exists \, u \in \R^\d\backslash \{0\} \colon [(\Hess \f)(\vartheta)]u = \mu u) \} \subseteq [\alpha,\beta].
\end{equation}
Next observe that the assumption that $(\nabla \f)(\vartheta) = 0$ and 
the fundamental theorem of calculus ensure that for all $\theta,w \in \R^\d$ it holds that
\begin{equation}
\label{conditions_nesterov_vs_coercivity:eq1}
\begin{split}
&\scp{ w,(\nabla \f)(\theta) } = \scp{ w,(\nabla \f)(\theta) - (\nabla \f)(\vartheta) }
\\
&= \scp[\Big]{ w, [(\nabla \f)(\vartheta + t(\theta-\vartheta))]_{t=0}^{t=1} }
\\
&= \scp*{ w, \smallint_0^1\br*{(\Hess \f)(\vartheta + t(\theta-\vartheta))} (\theta-\vartheta) \,\diff t }
\\
&= \int_0^1\scp[\big]{ w, \br*{(\Hess \f)(\vartheta + t(\theta-\vartheta))} (\theta-\vartheta)} \,\diff t\\
&= \scp[\big]{ w, \br*{(\Hess \f)(\vartheta)}(\theta-\vartheta)}  \\
&\quad + \int_0^1\scp[\big]{ w, \bbr{(\Hess \f)(\vartheta + t(\theta-\vartheta))-(\Hess \f)(\vartheta)} (\theta-\vartheta)} \,\diff t
\end{split}
\end{equation}
\cfload.
The fact that $(\Hess \f)(\vartheta) \in \R^{\d \times \d}$ is a symmetric matrix, 
\cref{norm_sym_matrices:item3} in \cref{norm_sym_matrices}, 
and the Cauchy-Schwarz inequality
therefore imply that for all $\theta \in \mathbb{B}$ it holds that
\begin{equation}
	\begin{split}
	&\scp{ \theta-\vartheta,(\nabla \f)(\theta) }
	\\
	&\geq 
	\scp[\big]{ \theta-\vartheta, \br*{(\Hess \f)(\vartheta)} (\theta-\vartheta)}  
	\\
	&\quad 
		- \abs*{ \int_0^1\scp[\big]{ \theta-\vartheta, \bbr{(\Hess \f)(\vartheta + t(\theta-\vartheta))-(\Hess \f)(\vartheta)}  (\theta-\vartheta)} \, \diff t }
	\\
	&\geq 
	\min(\mathscr{S})\norm{\theta-\vartheta}^2   
	\\
	&
		\quad - \int_0^1 \norm{ \theta - \vartheta } \apnorm2{ \bbr{(\Hess \f)(\vartheta + t(\theta-\vartheta))-(\Hess \f)(\vartheta)}(\theta-\vartheta) } 
		\, \diff t 
		.
	\end{split}
\end{equation}
Combining
	this
with	
\eqref{conditions_nesterov_vs_coercivity:eq0}
and 
\eqref{conditions_nesterov_vs_coercivity:assumption1}
shows that for all $\theta \in \mathbb{B}$ it holds that
	\begin{equation}
\label{conditions_nesterov_vs_coercivity:eq2}
\begin{split}
&\scp{ \theta-\vartheta,(\nabla \f)(\theta) }
\\
&\geq 
\alpha \norm{\theta-\vartheta}^2  \\
&\quad - \int_0^1\norm{\theta-\vartheta} \rdnorm[\d \times \d]{(\Hess \f)(\vartheta + t(\theta-\vartheta))-(\Hess \f)(\vartheta)} \norm{\theta-\vartheta}\,\diff t\\
&\geq 
\alpha \norm{\theta-\vartheta}^2 - \br*{\int_0^1 \lambda \norm{\vartheta + t(\theta-\vartheta) - \vartheta}\,\diff t } \norm{\theta-\vartheta}^2 \\
&=  
\pr*{\alpha  - \br*{ \int_0^1t\,\diff t } \lambda\norm{\theta-\vartheta} }\norm{\theta-\vartheta}^2 
=
\pr*{\alpha - \tfrac{\lambda}{2}\pnorm2{\theta-\vartheta} }\norm{\theta-\vartheta}^2 \\
&\geq  
\pr*{\alpha  - \tfrac{\lambda\alpha}{2\lambda} }  \norm{\theta-\vartheta}^2= \tfrac{\alpha}{2}\norm{\theta-\vartheta}^2.
\end{split}
\end{equation}
Moreover, observe that (\ref{conditions_nesterov_vs_coercivity:assumption1}), (\ref{conditions_nesterov_vs_coercivity:eq0}), (\ref{conditions_nesterov_vs_coercivity:eq1}), the fact that $(\Hess \f)(\vartheta) \in \R^{\d \times \d}$ is a symmetric matrix, \cref{norm_sym_matrices:item2} in \cref{norm_sym_matrices}, the Cauchy-Schwarz inequality, and the assumption that $\alpha \leq \beta$ ensure that for all $\theta \in \mathbb{B}$, $w \in \R^\d$ with $\norm{w} = 1$ it holds that
\begin{equation}
\begin{split}
&\scp{ w,(\nabla \f)(\theta) } 
\\
&\leq \babs{\scp[\big]{ w, \br*{(\Hess \f)(\vartheta)}(\theta-\vartheta)}}  
\\
&\quad + \abs*{ \int_0^1\scp[\big]{ w, \bbr{(\Hess \f)(\vartheta + t(\theta-\vartheta))-(\Hess \f)(\vartheta)} (\theta-\vartheta)} \,\diff t } 
\\
&\leq  \norm{w} \apnorm2{\br*{(\Hess \f)(\vartheta)} (\theta-\vartheta) } 
\\
&\quad 
+ 
\int_0^1
  \norm{ w }
  \apnorm2{
    \br*{
      (\Hess \f)(\vartheta + t(\theta-\vartheta))-(\Hess \f)(\vartheta)
    } 
    (\theta-\vartheta)
  }
  \,
  \diff t 
\\
&\leq \br*{ \sup_{v\in \R^\d\backslash \{0\}} \frac{\norm{[(\Hess \f)(\vartheta)]v}}{\norm{v}} } \!\norm{\theta-\vartheta} \\
&\quad + \int_0^1  \rdnorm[\d \times \d]{(\Hess \f)(\vartheta + t(\theta-\vartheta))-(\Hess \f)(\vartheta)}\norm{\theta-\vartheta} \,\diff t \\
&\leq \max\bpr{\mathscr{S}}\norm{\theta-\vartheta} + \br*{\int_0^1 \lambda\norm{\vartheta + t(\theta-\vartheta) - \vartheta} \,\diff t } \! \norm{\theta-\vartheta}\\
&\leq \pr*{\beta + \lambda \br*{\int_0^1t\,\diff t } \! \norm{\theta-\vartheta}} \! \norm{\theta-\vartheta} 
= \pr*{ \beta + \tfrac{\lambda}{2}  \norm{\theta-\vartheta} } \! \norm{\theta-\vartheta} \\
&\leq \pr*{\beta + \tfrac{\lambda\alpha}{2\lambda}} \norm{\theta-\vartheta}= \br*{\tfrac{2\beta + \alpha}{2} } \norm{\theta-\vartheta} \leq \tfrac{3\beta}{2}\norm{\theta-\vartheta}.
\end{split}
\end{equation}
Therefore, we obtain for all $\theta \in \mathbb{B}$ that
\begin{equation}
\norm{(\nabla \f)(\theta)} = \sup_{w \in \R^\d ,\,\norm{w}=1} \br*{\scp{ w,(\nabla \f)(\theta) } } \leq \tfrac{3\beta}{2}\norm{\theta-\vartheta}.
\end{equation}
Combining this and (\ref{conditions_nesterov_vs_coercivity:eq2}) establishes (\ref{conditions_nesterov_vs_coercivity:conclusion}).
The proof of \cref{conditions_nesterov_vs_coercivity} is thus complete.
\end{proof}
\endgroup

The next result, 
\cref{GD_convergence_Nesterov} below, 
combines \cref{conditions_nesterov_vs_coercivity} 
with \cref{GD_convergence} to obtain an error analysis 
which assumes the conditions in \eqref{conditions_nesterov_vs_coercivity:assumption1} and \eqref{conditions_nesterov_vs_coercivity:assumption2} 
in \cref{conditions_nesterov_vs_coercivity} above. 
A result similar to \cref{GD_convergence_Nesterov} 
can, \eg, be found in Nesterov~\cite[Theorem 1.2.4]{Nesterov13}.

\cfclear
\begingroup
\newcommand{\Rdnorm}[2][x]{\normmm{#2}}
\newcommand{\rdnorm}[2][x]{\normmm{#2}}
\newcommand{\norm}[1]{\pnorm2{#1}}
\newcommand{\lrnorm}[1]{\apnorm2{#1}}
\providecommand{\d}{}
\renewcommand{\d}{\defaultParamDim}
\providecommand{\f}{}
\renewcommand{\f}{\defaultLossFunction}
\providecommand{\g}{}
\renewcommand{\g}{\defaultGradientFunction}
\begin{cor}[Error analysis for the \GD\ optimization method under conditions on the Hessian of the objective function]
\label{GD_convergence_Nesterov}
Let $ \d \in \N $, 
let $\Rdnorm[\d \times \d]{\cdot} \colon \R^{\d \times \d} \to \R$ 
satisfy for all $A \in \R^{\d \times \d}$ that $\rdnorm[\d \times \d]{A} = \sup_{v\in \R^\d\backslash \{0\}} \frac{\norm{Av}}{\norm{v}}$, 
and let $\lambda, \alpha \in (0,\infty)$, $\beta \in [\alpha, \infty)$, $(\gamma_n)_{n \in \N} \subseteq  [0,\frac{4\alpha}{9\beta^2}]$, $\vartheta, \xi \in \R^\d$, $\f \in C^2(\R^\d,\R)$ satisfy for all $v,w \in \R^\d$ that 
\begin{equation}
\label{GD_convergence_Nesterov:assumption1}
  (\nabla \f)(\vartheta) = 0,  \qquad   \rdnorm[\d \times \d]{ (\Hess \f)(v) - (\Hess \f)(w)} \leq \lambda \norm{v- w}, 
\end{equation}
\begin{equation}
\label{GD_convergence_Nesterov:assumption2}
 \{\mu \in \R \colon (\exists \, u \in \R^\d\backslash \{0\} \colon [(\Hess \f)(\vartheta)]u = \mu u) \} \subseteq [\alpha,\beta],
\end{equation}
and $\norm{\xi-\vartheta} \leq \frac{\alpha}{\lambda}$, and 
let $\Theta \colon \N_0 \to \R^\d$ satisfy for all $n \in \N$ that 
\begin{equation} 
\label{GD_convergence_Nesterov:assumption3}
\Theta_0 = \xi \qandq \Theta_n = \Theta_{n-1} - \gamma_n(\nabla \f)(\Theta_{n-1})
\end{equation}
\cfload.
Then 
\begin{enumerate}[label=(\roman*)]
\item 
\label{GD_convergence_Nesterov:item1}
it holds that $\{\theta \in \mathbb{B}  \colon  \f(\theta)= \inf\nolimits_{w \in \mathbb{B}} \f(w)  \} = \{\vartheta \}$,
\item 
\label{GD_convergence_Nesterov:item2}
it holds for all $k \in \N$ that 
$
  0
  \leq 1 - \alpha \gamma_k + \frac{ 9 \beta^2 ( \gamma_k )^2 }{ 4 } 
  \leq 1
$, 
\item 
\label{GD_convergence_Nesterov:item3}
it holds for all $n \in \N_0$ that
\begin{equation}
\label{GD_convergence_Nesterov:conclusion1}
\norm{\Theta_n-\vartheta} \leq 
\br*{{ \textstyle \prod\limits_{k = 1}^n} \br*{ 1-\alpha\gamma_k + \tfrac{9\beta^2(\gamma_k)^2}{4} }^{\nicefrac{1}{2}}} 
  \norm{\xi - \vartheta} 
  ,
\end{equation}
and 
\item 
\label{GD_convergence_Nesterov:item4}
it holds for all $n \in \N_0$ that
\begin{equation}
\label{GD_convergence_Nesterov:conclusion2}
0 \leq \f(\Theta_n)-\f(\vartheta) 
\leq 
\tfrac{3\beta}{4}\br*{{ \textstyle \prod\limits_{k = 1}^n} \br*{ 1-\alpha\gamma_k + \tfrac{9\beta^2(\gamma_k)^2}{4} } } 
  \norm{\xi - \vartheta}^2.
\end{equation}
\end{enumerate}
\end{cor}

\begin{proof}[Proof of Corollary \ref{GD_convergence_Nesterov}]
Note that (\ref{GD_convergence_Nesterov:assumption1}), (\ref{GD_convergence_Nesterov:assumption2}), and \cref{conditions_nesterov_vs_coercivity} prove that for all $\theta \in \{w \in \R^\d \colon \norm{w-\vartheta} \leq \frac{\alpha}{\lambda} \}$ it holds that
\begin{equation}
\scp{\theta-\vartheta, (\nabla \f)(\theta) } \geq \tfrac{\alpha}{2}\norm{\theta-\vartheta}^2 \qandq \norm{(\nabla \f)(\theta)} \leq \tfrac{3\beta}{2}\norm{\theta-\vartheta}
\end{equation}
\cfload.
Combining this, the assumption that 
\begin{equation}
  \norm{\xi-\vartheta} \leq \frac{\alpha}{\lambda} ,
\end{equation}
\cref{GD_convergence_Nesterov:assumption3}, 
and \cref{GD_convergence:item4,GD_convergence:item5} in \cref{GD_convergence} 
(applied with 
$ c \is \frac{\alpha}{2} $, 
$ L \is \frac{3\beta}{2}$, $r \is \frac{\alpha}{\lambda}$
in the notation of \cref{GD_convergence}) 
establishes 
\cref{GD_convergence_Nesterov:item1,GD_convergence_Nesterov:item2,GD_convergence_Nesterov:item3,GD_convergence_Nesterov:item4}. 
The proof of Corollary \ref{GD_convergence_Nesterov} is thus complete.
\end{proof}
\endgroup

\cfclear
\begingroup
\newcommand{\Rdnorm}[2][x]{\normmm{#2}}
\newcommand{\rdnorm}[2][x]{\normmm{#2}}
\newcommand{\norm}[1]{\pnorm2{#1}}
\newcommand{\lrnorm}[1]{\apnorm2{#1}}
\providecommand{\d}{}
\renewcommand{\d}{\defaultParamDim}
\providecommand{\f}{}
\renewcommand{\f}{\defaultLossFunction}
\providecommand{\g}{}
\renewcommand{\g}{\defaultGradientFunction}
\begin{remark}
In \cref{GD_convergence_Nesterov} we establish convergence of the considered \GD\ process 
under, amongst other things, the assumption that all eigenvalues 
of the Hessian of 
$ \f \colon \R^{ \d } \to \R $ 
at the local minimum point $ \vartheta $
are strictly positive (see \eqref{GD_convergence_Nesterov:assumption2}). 
In the situation where $ \cL $ is the cost function (integrated loss function)
associated to a supervised learning problem in the training of \anns, 
this assumption is basically not satisfied. 
Nonetheless, the convergence analysis in \cref{GD_convergence_Nesterov} can, roughly speaking, 
also be performed under the essentially (up to the smoothness conditions) more general assumption that 
there exists $ k \in \N_0 $ such that 
the set of local minimum points 
is locally a smooth $ k $-dimensional 
submanifold of $ \R^{ \d } $ 
and that the rank of the Hessian of $ \f $ is on 
this set of local minimum points locally (at least) $ \defaultNetDim - k $ (cf.\ Fehrman et al.~\cite{Fehrman20} for details). 
In certain situations this essentially generalized assumption has also been shown 
to be satisfied in the training of \anns\ in suitable supervised learning problems 
(see Jentzen \& Riekert~\cite{JentzenRiekert22}).
\end{remark}
\endgroup

\subsubsection{Equivalent conditions on the objective function}

\begingroup
\providecommand{\d}{}
\renewcommand{\d}{\defaultParamDim}
\providecommand{\f}{}
\renewcommand{\f}{\defaultLossFunction}
\providecommand{\g}{}
\renewcommand{\g}{\defaultGradientFunction}
\begin{lemma}
\label{equivalent_conditions1}
Let $\d \in \N$, 
let $\altscp{ \cdot, \cdot } \colon \R^\d \times \R^\d \to \R$ be a scalar product, 
let $\normmm{\cdot} \colon \R^\d \to \R$ satisfy for all $v \in \R^\d$ 
that $\normmm{v} = \sqrt{\altscp{  v, v } }$, 
let $\gamma \in (0,\infty)$, $\varepsilon \in (0,1)$, 
$r \in (0,\infty]$, $\vartheta \in \R^\d$, 
$\mathbb{B} = \{w \in \R^\d \colon \normmm{w-\vartheta} \leq r \}$, 
and let
$ \g \colon \R^\d \to \R^\d $ 
satisfy
for all $ \theta \in \mathbb{B} $ that
\begin{equation}
\label{equivalent_conditions1:ass1}
\normmm{\theta - \gamma \g(\theta) - \vartheta} \leq \varepsilon \normmm{\theta-\vartheta}.
\end{equation}
Then it holds for all $\theta \in \mathbb{B}$ that
\begin{equation}
\begin{split}
\altscp{\theta-\vartheta, \g(\theta) }
&\geq  
\max\cu*{\br*{\tfrac{1-\varepsilon^2}{2\gamma} } \normmm{\theta-\vartheta}^2, \tfrac{\gamma}{2}\normmm{ \g(\theta)}^2 } \\
&\geq  
\min\cu*{ \tfrac{1-\varepsilon^2}{2\gamma}, \tfrac{\gamma}{2}} \max\bcu{\normmm{\theta-\vartheta}^2, \normmm{ \g(\theta)}^2 } .
\end{split}
\end{equation}
\end{lemma}

\begin{proof}[Proof of \cref{equivalent_conditions1}]
First, note that \eqref{equivalent_conditions1:ass1} ensures that for all $\theta \in \mathbb{B}$ it holds that
\begin{equation}
\begin{split}
\varepsilon^2 \normmm{\theta-\vartheta}^2 & \geq 
\normmm{\theta - \gamma \g(\theta) - \vartheta}^2 = \normmm{(\theta  - \vartheta)- \gamma \g(\theta)}^2 \\
& = \normmm{\theta - \vartheta}^2 - 2\gamma\, \altscp{\theta-\vartheta, \g(\theta) } + \gamma^2\normmm{ \g(\theta)}^2.
\end{split}
\end{equation}
Hence, we obtain for all $\theta \in \mathbb{B}$ that
\begin{equation}
\begin{split}
2\gamma\altscp{\theta-\vartheta, \g(\theta) } 
&\geq 
(1-\varepsilon^2)\normmm{\theta-\vartheta}^2+ \gamma^2\normmm{ \g(\theta)}^2 \\
&\geq 
\max\bcu{(1-\varepsilon^2)\normmm{\theta-\vartheta}^2,\gamma^2 \normmm{ \g(\theta)}^2 } 
\geq 0.
\end{split}
\end{equation}
This demonstrates that for all $ \theta \in \mathbb{B} $ it holds that
\begin{equation}
\begin{split}
\altscp{\theta-\vartheta, \g(\theta) } & \geq \tfrac{1}{2\gamma} \max\bcu{(1-\varepsilon^2)\normmm{\theta-\vartheta}^2,\gamma^2 \normmm{ \g(\theta)}^2 }\\
&= 
\max\cu*{ \br*{\tfrac{1-\varepsilon^2}{2\gamma} }  \normmm{\theta-\vartheta}^2, \tfrac{\gamma}{2}\normmm{\g(\theta)}^2 }  \\
&\geq 
\min\cu*{ \tfrac{1-\varepsilon^2}{2\gamma}, \tfrac{\gamma}{2}} \max\bcu{\normmm{\theta-\vartheta}^2, \normmm{\g(\theta)}^2 }.
\end{split}
\end{equation}
The proof of \cref{equivalent_conditions1} is thus complete.
\end{proof}
\endgroup

\begingroup
\newcommand{\lrnorm}[1]{\normmm{#1}}
\newcommand{\norm}[1]{\normmm{#1}}
\providecommand{\d}{}
\renewcommand{\d}{\defaultParamDim}
\providecommand{\f}{}
\renewcommand{\f}{\defaultLossFunction}
\providecommand{\g}{}
\renewcommand{\g}{\defaultGradientFunction}
\begin{lemma}
\label{equivalent_conditions2}
Let $\d \in \N$, 
let $\altscp{ \cdot, \cdot } \colon \R^\d \times \R^\d \to \R$ be a scalar product, 
let $\lrnorm{\cdot} \colon \R^\d \to \R$ satisfy for all $v \in \R^\d$ that $\norm{v} = \sqrt{\altscp{  v, v } }$,  
let $c \in (0,\infty)$, $r \in (0,\infty]$, $\vartheta \in \R^\d$, $\mathbb{B} = \{w \in \R^\d \colon \norm{w-\vartheta} \leq r \}$, 
and let
$ \g \colon \R^\d \to \R^\d $ satisfy for all $\theta \in \mathbb{B}$ that
\begin{equation}
\label{equivalent_conditions2:ass1}
\altscp{\theta-\vartheta, \g(\theta) } \geq 
c 
\max\bcu{ \norm{\theta-\vartheta}^2, \norm{\g(\theta)}^2 }.
\end{equation}
Then it holds for all $\theta \in \mathbb{B}$ that
\begin{equation}
\altscp{\theta-\vartheta, \g(\theta) } \geq c \norm{\theta-\vartheta}^2 \qandq \norm{\g(\theta)} \leq \tfrac{1}{c} \norm{\theta-\vartheta}.
\end{equation}
\end{lemma}

\begin{proof}[Proof of \cref{equivalent_conditions2}]
Observe that \eqref{equivalent_conditions2:ass1} and the Cauchy-Schwarz inequality assure that for all $\theta \in \mathbb{B}$ it holds that
\begin{equation}
\norm{\g(\theta)}^2 
\leq
\max\bcu{ \norm{\theta-\vartheta}^2, \norm{\g(\theta)}^2 }
\leq
\tfrac{1}{c} \altscp{\theta-\vartheta, \g(\theta) } \leq \tfrac{1}{c} \norm{\theta-\vartheta} \norm{\g(\theta)}.
\end{equation}
Therefore, we obtain for all $\theta \in \mathbb{B}$ that
\begin{equation}
\norm{\g(\theta)} \leq \tfrac{1}{c} \norm{\theta-\vartheta}.
\end{equation}
Combining this with \eqref{equivalent_conditions2:ass1} completes the proof of \cref{equivalent_conditions2}.
\end{proof}
\endgroup

\begingroup
\newcommand{\norm}[1]{\pnorm2{#1}}
\providecommand{\d}{}
\renewcommand{\d}{\defaultParamDim}
\providecommand{\f}{}
\renewcommand{\f}{\defaultLossFunction}
\providecommand{\g}{}
\renewcommand{\g}{\defaultGradientFunction}
\begin{lemma}
\label{equivalent_conditions3_1}
Let $\d \in \N$, 
$c \in (0,\infty)$, $r \in (0,\infty]$, $\vartheta \in \R^\d$, 
$\mathbb{B} = \{w \in \R^\d \colon \norm{w-\vartheta} \leq r \}$, $\f \in C^1(\R^\d,\R)$ satisfy for all $\theta \in \mathbb{B}$ that 
\begin{equation}
\label{equivalent_conditions3_1:ass1}
\scp{\theta-\vartheta, (\nabla \f)(\theta) } \geq c \norm{\theta-\vartheta}^2.
\end{equation}
Then it holds for all $v \in \R^\d$, $s,t \in [0,1]$ with $\norm{v}\leq r$ and $s \leq t$ that
\begin{equation}
\f(\vartheta + tv) - \f(\vartheta + sv) \geq \tfrac{c}{2}(t^2-s^2)\norm{v}^2.
\end{equation}
\end{lemma}

\begin{proof}[Proof of \cref{equivalent_conditions3_1}]
First of all, observe that \eqref{equivalent_conditions3_1:ass1} implies that for all $v \in \R^\d$ with $\norm{v} \leq r$ it holds that
\begin{equation}
\scp{ (\nabla \f)(\vartheta + v), v } \geq  c\norm{v}^2.
\end{equation}
The fundamental theorem of calculus hence ensures that for all $v \in \R^\d$, $s,t \in [0,1]$ with $\norm{v} \leq r$ and $s \leq t$ it holds that
\begin{equation}
\label{equivalent_conditions3_1:eq1}
\begin{split}
\f(\vartheta + tv) - \f(\vartheta + sv) 
&= \bbr{\f(\vartheta +hv) }_{h = s}^{h =t} \\
&= \int_s^t \f'(\vartheta + hv)v \,\diff h \\
&= \int_s^t \tfrac{1}{h} \scp{(\nabla \f)(\vartheta + hv),hv }  \,\diff h \\
& \geq \int_s^t \tfrac{c}{h}\norm{hv}^2 \,\diff h \\
&= c \br*{\int_s^t h \,\diff h}  \norm{v}^2= \tfrac{c}{2}(t^2-s^2)\norm{v}^2.
\end{split}
\end{equation}
The proof of \cref{equivalent_conditions3_1} is thus complete.
\end{proof}
\endgroup

\cfclear
\begingroup
\newcommand{\norm}[1]{\pnorm2{#1}}
\newcommand{\lrnorm}[1]{\apnorm2{#1}}
\providecommand{\d}{}
\renewcommand{\d}{\defaultParamDim}
\providecommand{\f}{}
\renewcommand{\f}{\defaultLossFunction}
\providecommand{\g}{}
\renewcommand{\g}{\defaultGradientFunction}
\begin{lemma}
\label{equivalent_conditions4_1}
Let $\d \in \N$, 
$c \in (0,\infty)$, $r \in (0,\infty]$, $\vartheta \in \R^\d$, $\mathbb{B} = \{w \in \R^\d \colon \norm{w-\vartheta} \leq r \}$, $\f \in C^1(\R^\d,\R)$ satisfy for all $v \in \R^\d$, $s,t \in [0,1]$ with $\norm{v} \leq r$ and $s \leq t$ that
\begin{equation}
\label{equivalent_conditions4_1:ass1}
\f(\vartheta + tv) - \f(\vartheta + sv) \geq c(t^2-s^2)\norm{v}^2
\end{equation}
\cfload.
Then it holds for all $\theta \in \mathbb{B}$ that 
\begin{equation}
\label{equivalent_conditions4_1:conclusion1}
\scp{\theta-\vartheta, (\nabla \f)(\theta) } \geq 2c \norm{\theta-\vartheta}^2
\end{equation}
\cfout.
\end{lemma}

\begin{proof}[Proof of \cref{equivalent_conditions4_1}]
Observe that \cref{equivalent_conditions4_1:ass1} ensures that 
for all 
$
  s \in (0,r] \cap \R 
$, 
$ \theta \in \R^\d \backslash \{ \vartheta \} $ 
with $ \norm{ \theta - \vartheta } < s $ 
it holds that
\begin{equation}
\begin{split}
&\scp{\theta-\vartheta, (\nabla \f)(\theta) } 
=
\f'(\theta)(\theta-\vartheta) 
= 
\lim_{h \searrow 0} \pr*{ \tfrac{1}{h}\bbr{\f(\theta + h(\theta-\vartheta)) -\f(\theta)} } \\
&=
\lim_{h \searrow 0} \bbbpr{ \frac{1}{h}\bbbbr{\f\bbpr{\vartheta + \tfrac{(1+h)\norm{\theta-\vartheta}}{s} \bbpr{\tfrac{s}{\norm{\theta-\vartheta}}(\theta-\vartheta) }} \\
&\quad 
	-\f\bbpr{\vartheta + \tfrac{\norm{\theta-\vartheta}}{s}\bbpr{\tfrac{s}{\norm{\theta-\vartheta}}(\theta-\vartheta) }}} } \\
&\geq 
\limsup_{h \searrow 0} \bbbpr{ 
\frac{c}{h}  
 \bbpr{ 
 \br*{\tfrac{(1+h)\norm{\theta-\vartheta}}{s}}^{ 2} -  \br*{\tfrac{\norm{\theta-\vartheta}}{s}}^{ 2}
 } 
\lrnorm{\tfrac{s}{\norm{\theta-\vartheta}}(\theta-\vartheta) }^2 
}\\
&= 
c  \br*{\limsup_{h \searrow 0} \pr*{\tfrac{(1+h)^2-1}{h} }} 
\br*{\tfrac{\norm{\theta-\vartheta}}{s}}^{ 2} \lrnorm{\tfrac{s}{\norm{\theta-\vartheta}}(\theta-\vartheta) }^2 \\
&= 
c\br*{\limsup_{h \searrow 0} \pr*{ \tfrac{2h + h^2}{h} }} \norm{\theta-\vartheta}^2 \\
&=
c\br*{\limsup_{h \searrow 0} \pr*{ 2 + h }} \norm{\theta-\vartheta}^2
= 2c\norm{\theta-\vartheta}^2
\end{split}
\end{equation}
\cfload.
Hence, we obtain that for all 
$
  \theta \in \R^\d \backslash \{ \vartheta \}
$ 
with $\norm{\theta - \vartheta} < r$ it holds that
\begin{equation}
  \scp{\theta-\vartheta, (\nabla \f)(\theta) } \geq 2c\norm{\theta-\vartheta}^2.
\end{equation}
Combining this with the fact that the function 
\begin{equation}
  \R^\d \ni v \mapsto (\nabla \f)(v) \in \R^\d
\end{equation}
is continuous establishes \eqref{equivalent_conditions4_1:conclusion1}. The proof of \cref{equivalent_conditions4_1} is thus complete.
\end{proof}
\endgroup

\cfclear
\begingroup
\newcommand{\norm}[1]{\pnorm2{#1}}
\newcommand{\lrnorm}[1]{\apnorm2{#1}}
\providecommand{\d}{}
\renewcommand{\d}{\defaultParamDim}
\providecommand{\f}{}
\renewcommand{\f}{\defaultLossFunction}
\providecommand{\g}{}
\renewcommand{\g}{\defaultGradientFunction}
\begin{lemma}
\label{equivalent_conditions3_2}
Let 
$\d \in \N$, 
$L \in (0,\infty)$, $r \in (0,\infty]$, $\vartheta \in \R^\d$, 
$\mathbb{B} = \{w \in \R^\d \colon \norm{w-\vartheta} \leq r \}$, 
$\f \in C^1(\R^\d,\R)$ satisfy for all $\theta \in \mathbb{B}$ that 
\begin{equation}
\label{equivalent_conditions3_2:ass1}
\norm{(\nabla \f)(\theta)} \leq L \norm{\theta-\vartheta}
\end{equation}
\cfload.
Then it holds for all $v,w \in \mathbb{B}$ that
\begin{equation}
\abs{\f(v)-\f(w)} \leq L \max\bcu{\norm{v-\vartheta}, \norm{w-\vartheta}} \norm{v-w}.
\end{equation}
\end{lemma}

\begin{proof}[Proof of \cref{equivalent_conditions3_2}]
Observe that (\ref{equivalent_conditions3_2:ass1}), the fundamental theorem of calculus, and the Cauchy-Schwarz inequality assure that for all $v, w \in \mathbb{B}$ it holds that
\begin{equation}
\begin{split}
\abs{\f(v) - \f(w)} &=   \abs*{ \bbr{\f(w + h (v - w))}_{h = 0}^{h=1}} \\
&=\abs*{ \int_0^1 \f'(w + h(v-w))(v-w) \,\diff h }\\
&= \abs*{ \int_0^1 \scp[\big]{(\nabla \f)\bpr{w + h(v-w)},v-w } \,\diff h  } \\
&\leq  \int_0^1 \norm{(\nabla \f)\bpr{h v  + (1- h )w}}\norm{v-w} \,\diff h \\
&\leq \int_0^1 L\norm{hv + (1-h)w - \vartheta}\norm{v-w } \,\diff h \\
& \leq 
\int_0^1 L\bpr{ h\norm{v - \vartheta}+(1-h)\norm{w - \vartheta}}\norm{v-w } \, \diff h 
\\
& =
L \,
\norm{ v - w }
\br*{
  \int_0^1 \bpr{ h \norm{v - \vartheta} + h \norm{w - \vartheta}} \, \diff h 
}
\\
& =
L \,
  \bpr{ \norm{v - \vartheta} + \norm{w - \vartheta}} 
  \,
\norm{ v - w }
\br*{
  \int_0^1 
  h \, \diff h 
}
\\
&\leq  L \max \{ \norm{v - \vartheta}, \norm{w - \vartheta}\}\norm{v-w}
\end{split}
\end{equation}
\cfload.
The proof of \cref{equivalent_conditions3_2} is thus complete.
\end{proof}
\endgroup

\cfclear
\begingroup
\newcommand{\norm}[1]{\pnorm2{#1}}
\newcommand{\lrnorm}[1]{\apnorm2{#1}}
\providecommand{\d}{}
\renewcommand{\d}{\defaultParamDim}
\providecommand{\f}{}
\renewcommand{\f}{\defaultLossFunction}
\providecommand{\g}{}
\renewcommand{\g}{\defaultGradientFunction}
\begin{lemma}
\label{equivalent_conditions4_2}
Let $\d \in \N$, 
$L \in (0,\infty)$, $r \in (0,\infty]$, 
$\vartheta \in \R^\d$, 
$\mathbb{B} = \{w \in \R^\d \colon \norm{w-\vartheta} \leq r \}$, 
$\f \in C^1(\R^\d,\R)$ satisfy for all $v,w \in \mathbb{B}$ that
\begin{equation}
\label{equivalent_conditions4_2:ass1}
\abs{\f(v)-\f(w)} \leq L \max\bcu{\norm{v-\vartheta}, \norm{w-\vartheta}} \norm{v-w}
\end{equation}
\cfload.
Then it holds for all $\theta \in \mathbb{B}$ that 
\begin{equation}
\label{equivalent_conditions4_2:conclusion1}
\norm{(\nabla \f)(\theta)} \leq L \norm{\theta-\vartheta}.
\end{equation}
\end{lemma}

\begin{proof}[Proof of \cref{equivalent_conditions4_2}]
Note that (\ref{equivalent_conditions4_2:ass1}) implies that for all $\theta \in \R^\d$ with $\norm{\theta-\vartheta} < r$ it holds that
\begin{equation}
\begin{split}
\norm{(\nabla \f)(\theta)}
&= \sup_{w \in \R^\d , \norm{w}=1} \bbbr{ \f'(\theta)(w) } \\
&=\sup_{w \in \R^\d , \norm{w}=1} \bbbr{ \lim_{h \searrow 0}\bbr{\tfrac{1}{h} ( \f(\theta + hw) -\f(\theta)) } } \\
&\leq \sup_{w \in \R^\d , \norm{w}=1} 
\bbbbr{ \liminf_{h \searrow 0}\bbbr{ \tfrac{L }{h} \max\bcu{\norm{\theta + hw-\vartheta}, \norm{\theta-\vartheta}} \norm{\theta + hw - \theta} } } \\
&= \sup_{w \in \R^\d , \norm{w}=1} \bbbbr{ \liminf_{h \searrow 0} \bbbr{ L \max\bcu{\norm{\theta + hw-\vartheta}, \norm{\theta-\vartheta}} \tfrac{1}{h}\norm{ hw } } }\\
&= \sup_{w \in \R^\d , \norm{w}=1} \bbbbr{ \liminf_{h \searrow 0} \bbbr{ L \max\bcu{\norm{\theta + hw-\vartheta}, \norm{\theta-\vartheta}} } } \\
& = \sup_{w \in \R^\d , \norm{w}=1}\bbbr{  L\norm{\theta-\vartheta}} = L\norm{\theta-\vartheta} .
\end{split}
\end{equation}
The fact that the function $\R^\d \ni v \mapsto (\nabla \f)(v) \in \R^\d$ is continuous therefore establishes \eqref{equivalent_conditions4_2:conclusion1}.
The proof of \cref{equivalent_conditions4_2} is thus complete.
\end{proof}
\endgroup

\cfclear
\begingroup
\newcommand{\norm}[1]{\pnorm2{#1}}
\newcommand{\lrnorm}[1]{\apnorm2{#1}}
\providecommand{\d}{}
\renewcommand{\d}{\defaultParamDim}
\providecommand{\f}{}
\renewcommand{\f}{\defaultLossFunction}
\providecommand{\g}{}
\renewcommand{\g}{\defaultGradientFunction}
\begin{cor}
\label{equivalent_conditions}
Let $\d \in \N$, 
$r \in (0,\infty]$, $\vartheta \in \R^\d$, $\mathbb{B} = \{w \in \R^\d \colon \norm{w-\vartheta} \leq r \}$, $\f \in C^1(\R^\d,\R)$
\cfload.
Then the following four statements are equivalent:
\begin{enumerate}[label=(\roman *)]
\item \label{equivalent_conditions:item1}
There exist $c,L \in (0,\infty)$ such that for all $\theta \in \mathbb{B}$ it holds that
\begin{equation}
\scp{\theta-\vartheta, (\nabla \f)(\theta) } \geq c \norm{\theta-\vartheta}^2 \qandq \norm{(\nabla \f)(\theta)} \leq L \norm{\theta-\vartheta}.
\end{equation}
\item \label{equivalent_conditions:item2}
There exist $\gamma \in (0,\infty)$, $\varepsilon \in (0,1)$ such that for all $\theta \in \mathbb{B}$ it holds that
\begin{equation}
\lrnorm{\theta - \gamma (\nabla \f)(\theta) - \vartheta} \leq \varepsilon \norm{\theta-\vartheta}.
\end{equation}
\item \label{equivalent_conditions:item3}
There exists $c \in (0,\infty)$ such that for all $\theta \in \mathbb{B}$ it holds that
\begin{equation}
\scp{\theta-\vartheta, (\nabla \f)(\theta) } \geq c \max\bcu{\norm{\theta-\vartheta}^2, \norm{(\nabla \f)(\theta)}^2}.
\end{equation}
\item \label{equivalent_conditions:item4}
There exist $c,L \in (0,\infty)$ such that for all $v,w \in \mathbb{B}$, $s,t \in [0,1]$ with $s \leq t$ it holds that
\begin{equation}
\f\bpr{\vartheta + t(v-\vartheta)} - \f\bpr{\vartheta + s(v-\vartheta)} \geq c(t^2-s^2)\norm{v - \vartheta}^2
\end{equation}
\begin{equation}
\andq \abs{\f(v)-\f(w)} \leq L \max\bcu{\norm{v-\vartheta}, \norm{w-\vartheta}} \norm{v-w}
\end{equation}
\end{enumerate}
\cfout.
\end{cor}

\begin{proof}[Proof of Corollary \ref{equivalent_conditions}]
\Nobs that 
\cref{fcond:item2,fcond:item3} in \cref{fcond} 
prove that 
(\ref{equivalent_conditions:item1} $\rightarrow$ \ref{equivalent_conditions:item2}). 
\Nobs that \cref{equivalent_conditions1} demonstrates that 
(\ref{equivalent_conditions:item2} $\rightarrow$ \ref{equivalent_conditions:item3}). 
\Nobs that \cref{equivalent_conditions2} establishes that 
(\ref{equivalent_conditions:item3} $\rightarrow$ \ref{equivalent_conditions:item1}). 
\Nobs that \cref{equivalent_conditions3_1} and \cref{equivalent_conditions3_2} 
show that 
(\ref{equivalent_conditions:item1} $\rightarrow$ \ref{equivalent_conditions:item4}). 
\Nobs that \cref{equivalent_conditions4_1} and \cref{equivalent_conditions4_2} 
establish that 
(\ref{equivalent_conditions:item4} $\rightarrow$ \ref{equivalent_conditions:item1}). 
The proof of Corollary \ref{equivalent_conditions} is thus complete.
\end{proof}
\endgroup

\section{Explicit midpoint optimization}
\label{sect:determ_expl_midpoint}

\begin{introductions}

As discussed in \cref{sec:gradient_descent} above, the \GD\ optimization method can be viewed as an Euler discretization of the associated \GF\ \ODE\ in Theorem~\ref{flow} in Chapter~\ref{chapter:flow}.
In the literature also more sophisticated methods than the Euler method have been employed to approximate the \GF\ \ODE. 
In particular, higher order Runge-Kutta methods have been used to approximate local minimum points of optimization problems (cf., \eg, Zhang et al.~\cite{Zhang2018}).
In this section we illustrate this in the case of the explicit midpoint method.

\end{introductions}

\defmidpointGD
\algDescrDetermmidpointGD

\subsection{Explicit midpoint discretizations for GF ODEs}

\cfclear
\begingroup
\providecommand{\d}{}
\renewcommand{\d}{\defaultParamDim}
\providecommand{\f}{}
\renewcommand{\f}{\defaultLossFunction}
\providecommand{\g}{}
\renewcommand{\g}{\defaultGradientFunction}
\begin{lemma}[Local error of the explicit midpoint method]
\label{one_step_midpoint}
Let 
	$ \d \in \N $, 
	$T, \gamma, c \in [0,\infty)$,
	$ \g \in C^2( \R^\d, \R^\d ) $,
	$ \Theta \in C( [0,\infty), \R^\d ) $,  
	$\theta \in\R^\d$
satisfy for all
	$x, y, z \in \R^\d$, 
	$ t \in [0,\infty) $ 
that
\begin{equation}
\label{one_step_midpoint:ass0}
\begin{split} 
 	\Theta_t = \Theta_0 + \int_0^t \g( \Theta_s ) \, \diff s,
\qquad
	\theta
=
	\Theta_T + \gamma \g \big(\Theta_T + \tfrac{\gamma}{2}\g(\Theta_T)\big),
\end{split}
\end{equation} 
\begin{equation}
\label{one_step_midpoint:ass1}
\begin{split} 
	\pnorm2{\g(x)}
\leq
	c,
\qquad
	\pnorm{2}{\g'(x) y}
\leq
	c
	\pnorm2{y},
\qandq
	\pnorm{2}{\g''(x)(y, z)}
\leq
	c
	\pnorm2{y} \pnorm2{z}
\end{split}
\end{equation}
\cfload.
Then 
\begin{equation}
\begin{split} 
	\Pnorm2{\Theta_{T+\gamma} - \theta}
\leq
	c^3 \gamma^3.
\end{split}
\end{equation}
\end{lemma}

\begin{proof}[Proof of \cref{one_step_midpoint}]
Note that 
\enum{
	the fundamental theorem of calculus;
	the assumption that $ \g \in C^2( \R^\d, \R^\d ) $;
	\eqref{one_step_midpoint:ass0}
}[assure]
that for all
	$t \in [0,\infty)$
it holds that $\Theta \in C^1( [0,\infty), \R^\d )$ and
\begin{equation}
\label{one_step_midpoint:eq1}
\begin{split} 
	\dot \Theta_t
=
	\g( \Theta_t ).
\end{split}
\end{equation}
Combining this with
\enum{
	the assumption that $ \g \in C^2( \R^\d, \R^\d ) $;
	the chain rule
}
ensures that for all
	$t \in [0,\infty)$
it holds that $\Theta \in C^2( [0,\infty), \R^\d )$ and
\begin{equation}
\label{one_step_midpoint:eq2}
\begin{split} 
	\ddot \Theta_t
=
	\g'( \Theta_t ) \dot \Theta_t
=
	\g'( \Theta_t ) \g( \Theta_t )
	.
\end{split}
\end{equation}
\enum{
	\Cref{thm:taylor_formula};
	\eqref{one_step_midpoint:eq1}
}
\hence 
\prove that 
\begin{equation}
\label{one_step_midpoint:eq3}
\begin{split} 
	 \Theta_{T+ \frac{\gamma}{2}}
&=
	\Theta_{T}
	+
	\br*{
        \frac{\gamma}{2}
    }
	\dot \Theta_{T}
	+
	\int_{0}^{1}
		(1-r)
		\br*{\frac{\gamma}{2}}^2
		\ddot \Theta_{T + r\gamma/2}
	\, \diff r\\
&=
	\Theta_{T}
	+
	\br*{
        \frac{\gamma}{2}
    }
	\g( \Theta_T )
	+
	\frac{\gamma^2}{4}
	\int_{0}^{1}
		(1-r)
		\g'( \Theta_{T + r\gamma/2} ) \g( \Theta_{T + r\gamma/2} )
	\, \diff r.
\end{split}
\end{equation}
\Hence that
\begin{equation}
\label{one_step_midpoint:eq31}
\begin{split} 
	\Theta_{T+ \frac{\gamma}{2}} - \Theta_{T} - 
	\br*{
        \frac{\gamma}{2}
    }
	\g(\Theta_T)
=
	\frac{\gamma^2}{4}
	\int_{0}^{1}
		(1-r)
		\g'( \Theta_{T + r\gamma/2} ) \g( \Theta_{T + r\gamma/2} )
	\, \diff r.
\end{split}
\end{equation}
Combining
\enum{
	this;
	the fact that for all
		$x, y \in \R^\d$
	it holds that
	$
		\Pnorm2{\g(x) - \g(y)} \leq c \Pnorm2{x-y}
	$;
	\eqref{one_step_midpoint:ass1};
}
ensures
that
\begin{equation}
\label{one_step_midpoint:eq32}
\begin{split} 
	\Pnorm[\big]2{
		\g( \Theta_{T+ \frac{\gamma}{2}} )
		-
		\g \pr[\big]{\Theta_T + \tfrac{\gamma}{2}\g(\Theta_T)}
	}
&\leq
	c
	\Pnorm[\big]2{ 
		\Theta_{T+ \frac{\gamma}{2}} - \Theta_{T} - \tfrac{\gamma}{2}\g(\Theta_T) 
	}\\
&\leq
	\frac{c\gamma^2}{4}
	\int_{0}^{1}
		(1-r)
		\Pnorm[\big]2{ \g'( \Theta_{T + r\gamma/2} ) \g( \Theta_{T + r\gamma/2} )}
	\, \diff r\\
&\leq
	\frac{c^3\gamma^2}{4}
	\int_{0}^{1}
		r
	\, \diff r
=
	\frac{c^3\gamma^2}{8}.
\end{split}
\end{equation}
\Moreover
\enum{
	\eqref{one_step_midpoint:eq1};
	\eqref{one_step_midpoint:eq2};
	the hypothesis that $ \g\in C^2( \R^\d, \R^\d ) $;
	the product rule;
	the chain rule;
}[assure]
that for all
$ t \in [0,\infty) $
it holds that $\Theta \in C^3( [0,\infty), \R^\d )$ and
\begin{equation}
\label{one_step_midpoint:eq4}
\begin{split} 
	\dddot \Theta_t
&=
	\g''( \Theta_t ) (\dot \Theta_t, \g( \Theta_t ))
	+
	\g'( \Theta_t ) \g'( \Theta_t ) \dot \Theta_t\\
&=
	\g''( \Theta_t ) (\g( \Theta_t ), \g( \Theta_t ))
	+
	\g'( \Theta_t ) \g'( \Theta_t ) \g( \Theta_t )
	.
\end{split}
\end{equation}
\enum{
	\Cref{thm:taylor_formula};
	\eqref{one_step_midpoint:eq1};
	\eqref{one_step_midpoint:eq2}
}
\hence
imply that for all 
	$s, t \in [0,\infty)$ 
it holds that
\begin{equation}
\begin{split} 
	\Theta_{s}
&=
	\Theta_{t}
	+
	(s-t)
	\dot \Theta_{t}
	+
	\br*{
        \frac{(s-t)^2}{2}
    }
	\ddot \Theta_{t}
	+
	\int_{0}^{1}
        \br*{
            \frac{(1-r)^2(s-t)^3}{2}
        }
		\dddot \Theta_{t + r(s-t)}
	\, \diff r
	\\
&=
	\Theta_{t}
	+
	(s-t)
	\g( \Theta_t )
	+
	\br*{
        \frac{(s-t)^2}{2}
    }
	\g'( \Theta_t ) \g( \Theta_t )\\
&\quad
	+
	\frac{(s-t)^3}{2}
	\int_{0}^{1}
		(1-r)^2
		\pr[\big]{
			\g''( \Theta_{t + r(s-t)} ) (\g( \Theta_{t + r(s-t)} ), \g( \Theta_{t + r(s-t)} ))\\
&\quad 
			+
			\g'( \Theta_{t + r(s-t)} ) \g'( \Theta_{t + r(s-t)} ) \g( \Theta_{t + r(s-t)} )
		}
	\, \diff r.
\end{split}
\end{equation}
\enum{
	This
}[assure]
that
\begin{equation}
\begin{split} 
	&\Theta_{T+\gamma}
	-
	\Theta_T\\
&=
	\Theta_{T+ \frac{\gamma}{2}}
	+
	\br*{
        \frac{\gamma}{2}
    }
	\g( \Theta_{T+ \frac{\gamma}{2}} )
	+
	\br*{
        \frac{\gamma^2}{8}
    }
	\g'( \Theta_{T+ \frac{\gamma}{2}} ) \g( \Theta_{T+ \frac{\gamma}{2}} )\\
&\quad
	+
	\frac{\gamma^3}{16}
	\int_{0}^{1}
		(1-r)^2
		\pr[\big]{
			\g''( \Theta_{T+ (1+r)\gamma/2} ) (\g( \Theta_{T+ (1+r)\gamma/2} ), \g( \Theta_{T+ (1+r)\gamma/2} ))\\
&\quad
			+
			\g'( \Theta_{T+ (1+r)\gamma/2} ) \g'( \Theta_{T+ (1+r)\gamma/2} ) \g( \Theta_{T+ (1+r)\gamma/2} )
		}
	\, \diff r\\
&\quad
	-
	\br[\Bigg]{
		\Theta_{T+ \frac{\gamma}{2}}
		-
		\br*{ \frac{\gamma}{2} }
		\g( \Theta_{T+ \frac{\gamma}{2}} )
		+
		\br*{ \frac{\gamma^2}{8} }
		\g'( \Theta_{T+ \frac{\gamma}{2}} ) \g( \Theta_{T+ \frac{\gamma}{2}} )\\
&\quad
		-
		\frac{\gamma^3}{16}
		\int_{0}^{1}
			(1-r)^2
			\pr[\big]{
				\g''( \Theta_{T+ (1- r)\gamma/2} ) (\g( \Theta_{T+ (1- r)\gamma/2} ), \g( \Theta_{T+ (1- r)\gamma/2} ))\\
&\quad
				+
				\g'( \Theta_{T+ (1- r)\gamma/2} ) \g'( \Theta_{T+ (1- r)\gamma/2} ) \g( \Theta_{T+ (1- r)\gamma/2} )
			}
		\, \diff r
	}\\
&=
	\gamma \g( \Theta_{T+ \frac{\gamma}{2}} )
	+
	\frac{\gamma^3}{16}
		\int_{0}^{1}
			(1-r)^2
			\pr[\Big]{
				\g''( \Theta_{T+ (1+r)\gamma/2} ) (\g( \Theta_{T+ (1+r)\gamma/2} ), \g( \Theta_{T+ (1+r)\gamma/2} ))\\
&\quad
				+
				\g'( \Theta_{T+ (1+r)\gamma/2} ) \g'( \Theta_{T+ (1+r)\gamma/2} ) \g( \Theta_{T+ (1+r)\gamma/2} )\\
&\quad
				+
				\g''( \Theta_{T+ (1- r)\gamma/2} ) (\g( \Theta_{T+ (1- r)\gamma/2} ), \g( \Theta_{T+ (1- r)\gamma/2} ))\\
&\quad
				+
				\g'( \Theta_{T+ (1- r)\gamma/2} ) \g'( \Theta_{T+ (1- r)\gamma/2} ) \g( \Theta_{T+ (1- r)\gamma/2} )
			}
		\, \diff r.
\end{split}
\end{equation}
\enum{
	This;
	\eqref{one_step_midpoint:ass1};
	\eqref{one_step_midpoint:eq32}
}[assure]
that 
\begin{equation}
\begin{split} 
	\Pnorm2{\Theta_{T+\gamma} - \theta} 
&=
	\Pnorm[\big]2{
		\Theta_{T+\gamma}
		-
		\Theta_T 
		- 
		\gamma \g(\Theta_T + \tfrac{\gamma}{2}\g(\Theta_T))
	}
\\
&\leq 
	\Pnorm[\big]2{
		\Theta_{T+\gamma}
		-
		\br{
            \Theta_T 
            + 
            \gamma \g( \Theta_{ T + \frac{ \gamma }{ 2 } } )
        }
    }
    +
    \gamma 
    \Pnorm[\big]2{
		\gamma \g( \Theta_{ T + \frac{ \gamma }{ 2 } } )
		-
		\g( 
		  \Theta_T + \tfrac{ \gamma }{ 2 } \g( \Theta_T )
		)
	}
\\
&\leq
	\gamma 
	\Pnorm[\big]2{
		\g( \Theta_{T+ \frac{\gamma}{2}} )
		-
		\g (\Theta_T + \tfrac{\gamma}{2}\g(\Theta_T))
	}\\
&\quad
	+
	\frac{\gamma^3}{16}
		\int_{0}^{1}
			(1-r)^2
			\pr[\Big]{
				\Pnorm[\big]2{
					\g''( \Theta_{T+ (1+r)\gamma/2} ) (\g( \Theta_{T+ (1+r)\gamma/2} ), \g( \Theta_{T+ (1+r)\gamma/2} ))
				}\\
&\quad
				+
				\Pnorm[\big]2{
					\g'( \Theta_{T+ (1+r)\gamma/2} ) \g'( \Theta_{T+ (1+r)\gamma/2} ) \g( \Theta_{T+ (1+r)\gamma/2} )
				}\\
&\quad
				+
				\Pnorm[\big]2{
					\g''( \Theta_{T+ (1- r)\gamma/2} ) (\g( \Theta_{T+ (1- r)\gamma/2} ), \g( \Theta_{T+ (1- r)\gamma/2} ))
				}\\
&\quad
				+
				\Pnorm[\big]2{
					\g'( \Theta_{T+ (1- r)\gamma/2} ) \g'( \Theta_{T+ (1- r)\gamma/2} ) \g( \Theta_{T+ (1- r)\gamma/2} )
				}
			}
		\, \diff r\\
&\leq
	\frac{c^3\gamma^3}{8}
	+
	\frac{c^3\gamma^3}{4}
	\int_{0}^{1}
		r^2
	\, \diff r
=
	\frac{5c^3\gamma^3}{24}
\leq
	c^3\gamma^3.	
\end{split}
\end{equation}
The proof of \cref{one_step_midpoint} is thus complete.
\end{proof}
\endgroup

\cfclear
\begingroup
\providecommand{\d}{}
\renewcommand{\d}{\defaultParamDim}
\providecommand{\f}{}
\renewcommand{\f}{\defaultLossFunction}
\providecommand{\g}{}
\renewcommand{\g}{\defaultGradientFunction}
\begin{cor}[Local error of the explicit midpoint method for \GF\ \ODEs]
\label{gradient_one_step_midpoint}
Let 
	$ \d \in \N $, 
	$T, \gamma, c \in [0,\infty)$,
	$\f \in C^3( \R^{ \d }, \R ) $,
	$ \Theta \in C( [0,\infty), \R^{ \d } ) $,  
	$\theta \in \R^{ \d } $
satisfy for all
	$ x, y, z \in \R^{ \d } $, 
	$ t \in [0,\infty) $ 
that
\begin{equation}
\label{gradient_one_step_midpoint:ass0}
\begin{split} 
  \Theta_t = \Theta_0 - \int_0^t (\nabla \f)( \Theta_s ) \, \diff s,
\qquad
  \theta
=
  \Theta_T 
  - \gamma 
  ( \nabla \f )\bigl(
    \Theta_T - \tfrac{ \gamma }{ 2 } ( \nabla \f )( \Theta_T )
  \bigr) ,
\end{split}
\end{equation}
\begin{equation}
\label{gradient_one_step_midpoint:ass1}
\begin{split} 
	\Pnorm2{(\nabla \f) (x)}
\leq
	c,
\quad
	\pnorm{2}{
		(\Hess \f)(x) y
	}
\leq
	c
	\pnorm2{y},
\quad\text{and}\quad
	\pnorm{2}{(\nabla \f)''(x)(y, z)}
\leq
	c
	\pnorm2{y} \pnorm2{z}
\end{split}
\end{equation}
\cfload.
Then 
\begin{equation}
\begin{split} 
	\Pnorm2{\Theta_{T+\gamma} - \theta}
\leq
	c^3 \gamma^3.
\end{split}
\end{equation}
\end{cor}

\begin{proof}[Proof of \cref{gradient_one_step_midpoint}]
Throughout this proof, let $\g \colon \R^\d \to \R^\d$ satisfy for all 
	$\theta \in \R^\d$
that
\begin{equation}
\begin{split} 
	\g(\theta)
=
	- ( \nabla \f )(\theta).
\end{split}
\end{equation}
\Nobs that
the fact that for all
$ t \in [0,\infty) $
it holds that
\begin{equation}
  \Theta_t = \Theta_0 + \int_0^t \g( \Theta_s ) \, \diff s
  ,
\end{equation}
the fact that
\begin{equation}
  \theta
=
  \Theta_T 
  + \gamma 
  \g\bigl(
    \Theta_T + \tfrac{ \gamma }{ 2 } \g( \Theta_T )
  \bigr) 
  ,
\end{equation}
the fact that for all
		$x \in \R^\d$
	it holds that
$
		\pnorm2{\g(x)}
	\leq
		c
$,
the fact that for all
		$x, y \in \R^\d$
	it holds that
$
  \pnorm{2}{\g'(x) y}
	\leq
		c\pnorm2{y}
$,
the fact that for all
		$x, y, z \in \R^\d$
	it holds that
\begin{equation}
		\pnorm{2}{\g''(x) (y, z)}
	\leq
		c\pnorm2{y}\pnorm2{z}
		,
\end{equation}
and \cref{one_step_midpoint}
\prove that
\begin{equation}
  \pnorm2{\Theta_{T + \gamma} - \theta}
\leq
  c^3 \gamma^3
  .
\end{equation}
The proof of \cref{gradient_one_step_midpoint} is thus complete.
\end{proof}
\endgroup

\section{Momentum optimization}
\label{sect:determ_momentum}

\begin{introductions}
In \cref{sec:gradient_descent} above we have 
introduced and analyzed the classical plain-vanilla 
\GD\ optimization method. 
In the literature 
there are a number of somehow more sophisticated 
\GD-type optimization methods 
which aim to improve the convergence speed 
of the classical plain-vanilla 
\GD\ optimization method
(see, for example, Ruder~\cite{Ruder16} and \cref{sect:determ_nesterov,sect:determ_RMSprop,sect:determ_adadelta,sect:determ_adagrad,sect:determ_adam,sect:determ_AMSgrad,sect:determ_nadam,sect:determ_adamW} below).
In this section we introduce 
one of such more sophisticated 
\GD-type optimization methods, that is, 
we introduce 
the so-called momentum \GD\ optimization method 
(see \cref{def:determ_momentum} below). 
The idea to improve \GD\ optimization methods with a momentum term was first introduced in Polyak~\cite{Polyak64}. 
To illustrate the advantage of the momentum \GD\ optimization method
over the plain-vanilla \GD\ optimization method
we now review a result proving that
the momentum \GD\ optimization method does indeed 
outperform the classical plain-vanilla \GD\ optimization method 
in the case of a simple class of optimization problems 
(see \cref{sec:error_analysis_momentum_GD_all} below).

In the scientific literature there are several very similar, but not exactly equivalent optimization techniques which are referred to as optimization with momentum.
Our definition of the momentum \GD\ optimization method in \cref{def:determ_momentum} below is based on \cite{NgCoursera,KingmaBa14} and (7) in \cite{Dozat16}.
We discuss two alternative definitions from the literature in \cref{sec:momentum_GD_alternatives} below and present relationships between these definitions in \cref{sec:momentum_GD_relationships} below.

\end{introductions}

\defdetermMomentum
\algDescrDetermMomentum

\cfclear
\begingroup
\providecommand{\d}{}
\renewcommand{\d}{\defaultParamDim}
\providecommand{\f}{}
\renewcommand{\f}{\defaultLossFunction}
\providecommand{\g}{}
\renewcommand{\g}{\defaultGradientFunction}
\begin{exercise}{quest:momentum2}
Let 
	$\f \colon \R \to \R$ 
satisfy for all 
	$\theta \in \R$
that
	$\f(\theta) = 2 \theta^2$
and let	
	$\Theta$ 
be the momentum \GD\ process\cfadd{def:determ_momentum}
for the objective function $\f$ with
with learning rates \(\N\ni n\mapsto \nicefrac{1}{2^n} \in[0,\infty)\),
momentum decay factors \(\N\ni n\mapsto \nicefrac{1}{2}\in[0,1]\),
and initial value $1$
\cfload.
Specify 
	$\Theta_1$, 
	$\Theta_2$, and
	$\Theta_3$
explicitly and prove that your results are correct!
\end{exercise}

\cfclear
\begingroup
\providecommand{\d}{}
\renewcommand{\d}{\defaultParamDim}
\providecommand{\f}{}
\renewcommand{\f}{\defaultLossFunction}
\providecommand{\g}{}
\renewcommand{\g}{\defaultGradientFunction}
\begin{exercise}{momentum_gradient_descent}
Let \(\xi = (\xi_1,\xi_2)\in \R^2 \) satisfy \((\xi_1,\xi_2)=(2,3)\),
let \(\f \colon\R^2\to\R\) satisfy for all \(\theta=(\theta_1,\theta_2)\in\R^2\) that
\[
\f(\theta) = (\theta_1 - 3)^2 + \tfrac{1}{2}(\theta_2 - 2)^2 + \theta_1 + \theta_2
,\]
and let \(\Theta\) be the momentum \GD\ process\cfadd{def:determ_momentum} for the objective function $\f$
with learning rates \(\N\ni n\mapsto \nicefrac{2}{n}\in[0,\infty)\),
momentum decay factors \(\N\ni n\mapsto \nicefrac{1}{2}\in[0,1]\),
and initial value \(\xi\) \cfload.
Specify \(\Theta_1 \) and \(\Theta_2\) explicitly and prove that your results are correct!
\end{exercise}

\subsection{Alternative definitions}
\label{sec:momentum_GD_alternatives}

In this section we discuss two definitions similar to the momentum \GD\ optimization method in \cref{def:determ_momentum} which are sometimes also referred to as momentum \GD\ optimization methods in the scientific literature.
The differences between the methods lie in two aspects:
\begin{itemize}
\item Whether the momentum terms are accumulated over the gradients of the objective function or over the increments of the optimization process and
\item whether the momentum terms are given as weighted averages or as general linear combinations.
\end{itemize}
The method in \cref{def:determ_momentum_two} below can, \eg, be found in \cite[Algorithm 2]{Dozat15}.
The method in \cref{def:determ_momentum_three} below can, \eg, be found in 
	(9) in \cite{Polyak64}, 
	(2) in \cite{Qian99}, and
	(4) in \cite{Ruder16}.
Some relationships between these definitions are discussed in \cref{sec:momentum_GD_relationships} below.

\defdetermMomentumTwo
\algDescrDetermMomentumTwo

\defdetermMomentumThree
\algDescrDetermMomentumThree

\defdetermMomentumFour
\algDescrDetermMomentumFour

\subsection{Relationships between different definitions}
\label{sec:momentum_GD_relationships}

In this section we discuss relationships between the different versions of the momentum \GD\ optimization method introduced in \cref{def:determ_momentum,def:determ_momentum_two,def:determ_momentum_three,def:determ_momentum_four} above.

\begingroup
\providecommandordefault{\a}{\mathfrak{a}}
\providecommandordefault{\b}{\mathfrak{b}}
\providecommandordefault{\c}{\mathfrak{c}}

\cfclear
\begin{athm}{prop}{MomentumGeneral}[Comparison of general momentum-type \GD\ optimization methods]
Let 
	$\defaultParamDim \in \N$, 
	$(\a^{(1)}_n)_{n \in \N} \subseteq (0,\infty)$,
	$(\a^{(2)}_n)_{n \in \N} \subseteq (0,\infty)$,
	$(\b^{(1)}_n)_{n \in \N} \subseteq (0,\infty)$,
	$(\b^{(2)}_n)_{n \in \N} \subseteq (0,\infty)$,
	$(\c^{(1)}_n)_{n \in \N} \subseteq (0,\infty)$,
	$(\c^{(2)}_n)_{n \in \N} \subseteq (0,\infty)$,
	$\xi \in \R^\defaultParamDim$,
   	$\defaultLossFunction \in C^1(\R^\defaultParamDim, \R)$
satisfy for all 
	$n \in \N$
that
\begin{equation}
\label{MomentumGeneral:ass1}
\begin{split}
	\b^{(1)}_n \c^{(1)}_n 
=
	\b^{(2)}_n \c^{(2)}_n
\qandq
	\frac{
		\a^{(1)}_{n+1} \b^{(1)}_{n}
	}{
		\b^{(1)}_{n+1}
	}
=
	\frac{
		\a^{(2)}_{n+1} \b^{(2)}_{n}
	}{
		\b^{(2)}_{n+1}
	},
\end{split}
\end{equation}
and for every $i \in \{1,2\}$
let 
	$\Theta^{(i)} \colon \N_0 \to \R^\defaultParamDim$	
	and
	$\mathbf{m}^{(i)} \colon \N_0 \to \R^\defaultParamDim$
satisfy for all 
	$n \in \N$
that
\begin{equation}
\label{MomentumGeneral:ass2}
\begin{split}
	\Theta^{(i)}_0 = \xi, \qquad \mathbf{m}^{(i)}_0 = 0,
\end{split}
\end{equation}
\begin{equation}
\label{MomentumGeneral:ass3}
\begin{split}
	\mathbf{m}^{(i)}_n = \a^{(i)}_n \mathbf{m}^{(i)}_{n-1} + \b^{(i)}_n \grad(\Theta^{(i)}_{n-1}),
\end{split}
\end{equation}
\begin{equation}
\label{MomentumGeneral:ass4}
\begin{split}
	\andq \Theta^{(i)}_n = \Theta^{(i)}_{n-1} - \c^{(i)}_n \mathbf{m}^{(i)}_n.
\end{split}
\end{equation}
Then
\begin{equation}
\label{MomentumGeneral:concl1}
\Theta^{(1)} = \Theta^{(2)}.
\end{equation}
\end{athm}

\begin{aproof}
Throughout this proof,
let 
	$\defaultGradientFunction \colon \R^\defaultParamDim \to \R^\defaultParamDim$ 
satisfy for all 
	$\theta \in \R^\defaultParamDim$
that 
\begin{equation}
\label{T_B_D}
\begin{split}
	\defaultGradientFunction(\theta) = (\nabla \defaultLossFunction)(\theta).
\end{split}
\end{equation}
\Nobs that
\enum{
	the fact that for all $n \in \N$ it holds that
	\begin{equation}
	\begin{split}
		\c^{(1)}_{n+1} 
	=
		\frac{\c^{(2)}_{n+1} \b^{(2)}_{n+1}}{\b^{(1)}_{n+1}},
	\qquad
		\frac{\c^{(2)}_{n}}{\c^{(1)}_{n}}
	=
		\frac{\b^{(1)}_{n+1}}{\b^{(2)}_{n+1}},
	\qandq
		\frac{\b^{(2)}_{n+1} \a^{(1)}_{n+1} \b^{(1)}_{n}}{\b^{(1)}_{n+1} \b^{(2)}_{n}}
	=
		\a^{(2)}_{n+1}
	\end{split}
	\end{equation}
}
\proves
that for all
	$n \in \N$
it holds that
\begin{equation}
\label{MomentumGeneral:eq0}
\begin{split}
	\frac{\c^{(1)}_{n+1} \a^{(1)}_{n+1} \c^{(2)}_{n}}{\c^{(1)}_{n}}
=
	\frac{\c^{(2)}_{n+1} \b^{(2)}_{n+1} \a^{(1)}_{n} \b^{(1)}_{n}}{\b^{(1)}_{n+1} \b^{(2)}_{n}}
=
	\c^{(2)}_{n+1} \a^{(2)}_{n+1}.
\end{split}
\end{equation}
\Moreover
\enum{
	\cref{MomentumGeneral:ass2};
}
\proves
that
\begin{equation}
\label{MomentumGeneral:eq01}
\begin{split}
	\mathbf{m}^{(1)}_0 
=
	0
=
	\mathbf{m}^{(2)}_0
\qandq
	\Theta^{(1)}_0
=
	\xi
=
	\Theta^{(2)}_0.
\end{split}
\end{equation}
Next we claim that for all
	$n \in \N$
it holds that
\begin{equation}
\label{MomentumGeneral:eq1}
\begin{split}
	\c^{(1)}_n \mathbf{m}^{(1)}_n 
= 
	\c^{(2)}_n \mathbf{m}^{(2)}_n
\qandq
	\Theta^{(1)}_n
=
	\Theta^{(2)}_n.
\end{split}
\end{equation}
We now prove \cref{MomentumGeneral:eq1} by induction on $n \in \N$.
For the base case $n = 1$ \nobs that
\enum{
	\cref{MomentumGeneral:ass1};
	\cref{MomentumGeneral:ass2};
	\cref{MomentumGeneral:eq01};
}
\prove
that
\begin{equation}
\label{MomentumGeneral:eq2}
\begin{split}
	\c^{(1)}_1 \mathbf{m}^{(1)}_1
&=
	\c^{(1)}_1 (\a^{(1)}_1 \mathbf{m}^{(1)}_0 + \b^{(1)}_1 \defaultGradientFunction(\Theta^{(1)}_0))\\
&=
	\c^{(1)}_1 \b^{(1)}_1 \defaultGradientFunction(\Theta^{(1)}_0)\\
&=
	\c^{(2)}_1 \b^{(2)}_1 \defaultGradientFunction(\Theta^{(2)}_0)\\
&=
	\c^{(2)}_1 (\a^{(2)}_1 \mathbf{m}^{(2)}_0 + \b^{(2)}_1 \defaultGradientFunction(\Theta^{(2)}_0))\\
&=
	\c^{(2)}_1 \mathbf{m}^{(2)}_1.
\end{split}
\end{equation}
\enum{
	This;
	\cref{MomentumGeneral:ass4};
	\cref{MomentumGeneral:eq01}
}
\proves
\begin{equation}
	\Theta^{(1)}_1
=
	\Theta^{(1)}_0 - \c^{(1)}_1 \mathbf{m}^{(1)}_1
=
	\Theta^{(2)}_0 - \c^{(2)}_1 \mathbf{m}^{(2)}_1
=
	\Theta^{(2)}_1.
\end{equation}
Combining this and \cref{MomentumGeneral:eq2} establishes \cref{MomentumGeneral:eq1} in the base case $n = 1$.
For the induction step
$ \N \ni n \to n+1 \in \{2, 3, \ldots\}$ 
let $n \in \N$ and assume that
\begin{equation}
\label{MomentumGeneral:eq3}
\begin{split}
	\c^{(1)}_{n} \mathbf{m}^{(1)}_{n}
=
	\c^{(2)}_{n} \mathbf{m}^{(2)}_{n}
\qandq
	\Theta^{(1)}_{n}
=
	\Theta^{(2)}_{n}.
\end{split}
\end{equation}
\Nobs that
\enum{
	\cref{MomentumGeneral:ass1};
	\cref{MomentumGeneral:ass3};
	\cref{MomentumGeneral:eq0};
	\cref{MomentumGeneral:eq3};
}
\prove that
\begin{equation}
\label{MomentumGeneral:eq4}
\begin{split}
	\c^{(1)}_{n+1} \mathbf{m}^{(1)}_{n+1}
&=
	\c^{(1)}_{n+1} (\a^{(1)}_{n+1} \mathbf{m}^{(1)}_{n} + \b^{(1)}_{n+1} \defaultGradientFunction(\Theta^{(1)}_{n}))\\
&=
	\frac{\c^{(1)}_{n+1} \a^{(1)}_{n+1} \c^{(2)}_{n}}{\c^{(1)}_{n}} \mathbf{m}^{(2)}_{n} + \c^{(1)}_{n+1} \b^{(1)}_{n+1} \defaultGradientFunction(\Theta^{(2)}_{n})\\
&=
	\c^{(2)}_{n+1} \a^{(2)}_{n+1} \mathbf{m}^{(2)}_{n} + \c^{(2)}_{n+1} \b^{(2)}_{n+1} \defaultGradientFunction(\Theta^{(2)}_{n})\\
&=
	\c^{(2)}_{n+1} (\a^{(2)}_{n+1} \mathbf{m}^{(2)}_{n} + \b^{(2)}_{n+1} \defaultGradientFunction(\Theta^{(2)}_{n}))\\
&=
	\c^{(2)}_{n+1} \mathbf{m}^{(2)}_{n+1}.
\end{split}
\end{equation}
\enum{
	This;
	\cref{MomentumGeneral:ass4};
	\cref{MomentumGeneral:eq3};
}
\prove that
\begin{equation}
\label{T_B_D}
\begin{split}
	\Theta^{(1)}_{n+1}
=
	\Theta^{(1)}_{n} - \c^{(1)}_{n+1} \mathbf{m}^{(1)}_{n+1}
=
	\Theta^{(2)}_{n} - \c^{(2)}_{n+1} \mathbf{m}^{(2)}_{n+1}
=
	\Theta^{(2)}_{n+1}.
\end{split}
\end{equation}
Induction thus proves \cref{MomentumGeneral:eq1}.
Combining \cref{MomentumGeneral:eq01} and \cref{MomentumGeneral:eq1} establishes \cref{MomentumGeneral:concl1}.
\end{aproof}

\cfclear
\begin{athm}{cor}{Momentum1vs2}[Comparison of the \first and \second version of the momentum \GD\ optimization method]
Let 
	$\defaultParamDim \in \N$, 
	$(\gamma^{(1)}_n)_{n \in \N} \subseteq (0,\infty)$,
	$(\gamma^{(2)}_n)_{n \in \N} \subseteq (0,\infty)$,
	$(\alpha^{(1)}_n)_{n \in \N} \subseteq (0,1)$,
	$(\alpha^{(2)}_n)_{n \in \N} \subseteq (0,\infty)$,
	$\xi \in \R^\defaultParamDim$,
   	$\defaultLossFunction \in C^1(\R^\defaultParamDim, \R)$
satisfy for all 
	$n \in \N$
that
\begin{equation}
\label{Momentum1vs2:ass1}
\begin{split}
	\gamma^{(1)}_n (1-\alpha^{(1)}_n)
=
	\gamma^{(2)}_n
\qandq
	\frac{
		\alpha^{(1)}_{n+1}(1-\alpha^{(1)}_{n})
	}{
		1-\alpha^{(1)}_{n+1}
	}
=
	\alpha^{(2)}_{n+1},
\end{split}
\end{equation}
for every $i \in \{1,2\}$\cfadd{def:determ_momentum}\cfadd{def:determ_momentum_two}
let 
	$\Theta^{(i)} \colon \N_0 \to \R^\defaultParamDim$
be the momentum \GD\ process (\ith version) for the objective function $\defaultLossFunction$ with learning rates $(\gamma^{(i)}_n)_{n \in \N}$,
momentum decay factors $(\alpha^{(i)}_n)_{n \in \N}$,
and initial value $\xi$ \cfload.
Then
\begin{equation}
\label{Momentum1vs2:concl1}
\Theta^{(1)} = \Theta^{(2)}.
\end{equation}
\end{athm}

\begin{aproof}
Throughout this proof let 
	$(\a^{(1)}_n)_{n \in \N} \subseteq (0,\infty)$,
	$(\a^{(2)}_n)_{n \in \N} \subseteq (0,\infty)$,
	$(\b^{(1)}_n)_{n \in \N} \subseteq (0,\infty)$,
	$(\b^{(2)}_n)_{n \in \N} \subseteq (0,\infty)$,
	$(\c^{(1)}_n)_{n \in \N} \subseteq (0,\infty)$,
	$(\c^{(2)}_n)_{n \in \N} \subseteq (0,\infty)$
satisfy for all 
	$n \in \N$
that
\begin{equation}
\label{Momentum1vs2:setting1}
\begin{split}
	\a^{(1)}_n = \alpha^{(1)}_n, \qquad \b^{(1)}_n = 1 - \alpha^{(1)}_n, \qquad \c^{(1)}_n = \gamma^{(1)}_n,
\end{split}
\end{equation}
\begin{equation}
\label{Momentum1vs2:setting2}
\begin{split} 
	\a^{(2)}_n = \alpha^{(2)}_n, \qquad \b^{(2)}_n = 1, \qandq \c^{(2)}_n = \gamma^{(2)}_n.
\end{split}
\end{equation}
\Nobs that  
\enum{
	\cref{eq:def:momentum_1_1};
	\cref{eq:def:momentum_1_2};
	\cref{eq:def:momentum_1_3};
	\cref{eq:def:momentum_2_1};
	\cref{eq:def:momentum_2_2};
	\cref{eq:def:momentum_2_3};
}
\prove 
that for all 
	$i \in \{1,2\}$,
	$n \in \N$
it holds that 
\begin{equation}
\label{Momentum1vs2:eq1}
\begin{split}
	\Theta^{(i)}_0 = \xi, \qquad \mathbf{m}^{(i)}_0 = 0,
\end{split}
\end{equation}
\begin{equation}
\label{Momentum1vs2:eq2}
\begin{split}
	\mathbf{m}^{(i)}_n = \a^{(i)}_n \mathbf{m}^{(i)}_{n-1} + \b^{(i)}_n \grad(\Theta^{(i)}_{n-1}),
\end{split}
\end{equation}
\begin{equation}
\label{Momentum1vs2:eq3}
\begin{split}
	\andq \Theta^{(i)}_n = \Theta^{(i)}_{n-1} - \c^{(i)}_n \mathbf{m}^{(i)}_n.
\end{split}
\end{equation}
\Moreover 
\enum{
	\cref{Momentum1vs2:ass1};
	\cref{Momentum1vs2:setting1};
	\cref{Momentum1vs2:setting2};
}
\proves
that for all 
	$n \in \N$
it holds that 
\begin{equation}
\label{Momentum1vs2:eq4}
\begin{split}
	\b^{(1)}_n \c^{(1)}_n 
=
	(1-\alpha^{(1)}_n) \gamma^{(1)}_n
=
	\gamma^{(2)}_n
=
	\b^{(2)}_n \c^{(2)}_n.
\end{split}
\end{equation}
\Moreover 
\enum{
	\cref{Momentum1vs2:ass1};
	\cref{Momentum1vs2:setting1};
	\cref{Momentum1vs2:setting2};
}
\proves
that for all 
	$n \in \N$
it holds that
\begin{equation}
\label{Momentum1vs2:eq5}
\begin{split} 
	\frac{
		\a^{(1)}_{n+1} \b^{(1)}_{n}
	}{
		\b^{(1)}_{n+1}
	}
=
	\frac{
		\alpha^{(1)}_{n+1} (1-\alpha^{(1)}_{n})
	}{
		1-\alpha^{(1)}_{n+1}
	}
=
	\alpha^{(2)}_{n+1}
=
	\frac{
		\a^{(2)}_{n+1} \b^{(2)}_{n}
	}{
		\b^{(2)}_{n+1}
	}.
\end{split}
\end{equation} 
Combining 
\enum{
	this;
	\cref{Momentum1vs2:eq1};
	\cref{Momentum1vs2:eq2};
	\cref{Momentum1vs2:eq3};
	\cref{Momentum1vs2:eq4};
}
with 
\cref{MomentumGeneral}
\proves
\eqref{Momentum1vs2:concl1}.
\end{aproof}

\cfclear
\begin{athm}{lemma}{Momentum1vs3}[Comparison of the \first and \third version of the momentum \GD\ optimization method]
Let 
	$\defaultParamDim \in \N$, 
	$(\gamma^{(1)}_n)_{n \in \N} \subseteq (0,\infty)$,
	$(\gamma^{(3)}_n)_{n \in \N} \subseteq (0,\infty)$,
	$(\alpha^{(1)}_n)_{n \in \N} \subseteq (0,1)$,
	$(\alpha^{(3)}_n)_{n \in \N} \subseteq (0,1)$,
	$\xi \in \R^\defaultParamDim$,
   	$\defaultLossFunction \in C^1(\R^\defaultParamDim, \R)$
satisfy for all 
	$n \in \N$
that
\begin{equation}
\label{Momentum1vs3:ass1}
\begin{split}
	\gamma^{(1)}_n (1-\alpha^{(1)}_n)
=
	\gamma^{(3)}_n (1-\alpha^{(3)}_n)
\qandq
	\frac{
		\gamma^{(1)}_{n+1} \alpha^{(1)}_{n+1}
	}{
		\gamma^{(1)}_{n}
	}
=
	\alpha^{(3)}_{n+1},
\end{split}
\end{equation}
for every $i \in \{1,3\}$\cfadd{def:determ_momentum}\cfadd{def:determ_momentum_three}
let 
	$\Theta^{(i)} \colon \N_0 \to \R^\defaultParamDim$
be the momentum \GD\ process (\ith version) for the objective function $\defaultLossFunction$ with learning rates $(\gamma^{(i)}_n)_{n \in \N}$,
momentum decay factors $(\alpha^{(i)}_n)_{n \in \N}$,
and initial value $\xi$ \cfload.
Then
\begin{equation}
\label{Momentum1vs3:concl1}
\Theta^{(1)} = \Theta^{(3)}.
\end{equation}
\end{athm}

\begin{aproof}
Throughout this proof let 
	$(\a^{(1)}_n)_{n \in \N} \subseteq (0,\infty)$,
	$(\a^{(3)}_n)_{n \in \N} \subseteq (0,\infty)$,
	$(\b^{(1)}_n)_{n \in \N} \subseteq (0,\infty)$,
	$(\b^{(3)}_n)_{n \in \N} \subseteq (0,\infty)$,
	$(\c^{(1)}_n)_{n \in \N} \subseteq (0,\infty)$,
	$(\c^{(3)}_n)_{n \in \N} \subseteq (0,\infty)$
satisfy for all 
	$n \in \N$
that
\begin{equation}
\label{Momentum1vs3:setting1}
\begin{split}
	\a^{(1)}_n = \alpha^{(1)}_n, \qquad \b^{(1)}_n = 1 - \alpha^{(1)}_n, \qquad \c^{(1)}_n = \gamma^{(1)}_n,
\end{split}
\end{equation}
\begin{equation}
\label{Momentum1vs3:setting2}
\begin{split} 
	\a^{(3)}_n = \alpha^{(3)}_n, \qquad \b^{(3)}_n = (1-\alpha^{(3)}_n) \gamma^{(3)}_n, \qandq \c^{(3)}_n = 1.
\end{split}
\end{equation}
\Nobs that  
\enum{
	\cref{eq:def:momentum_1_1};
	\cref{eq:def:momentum_1_2};
	\cref{eq:def:momentum_1_3};
	\cref{eq:def:momentum_3_1};
	\cref{eq:def:momentum_3_2};
	\cref{eq:def:momentum_3_3};
}
\prove 
that for all 
	$i \in \{1,3\}$,
	$n \in \N$
it holds that 
\begin{equation}
\label{Momentum1vs3:eq1}
\begin{split}
	\Theta^{(i)}_0 = \xi, \qquad \mathbf{m}^{(i)}_0 = 0,
\end{split}
\end{equation}
\begin{equation}
\label{Momentum1vs3:eq2}
\begin{split}
	\mathbf{m}^{(i)}_n = \a^{(i)}_n \mathbf{m}^{(i)}_{n-1} + \b^{(i)}_n \grad(\Theta^{(i)}_{n-1}),
\end{split}
\end{equation}
\begin{equation}
\label{Momentum1vs3:eq3}
\begin{split}
	\andq \Theta^{(i)}_n = \Theta^{(i)}_{n-1} - \c^{(i)}_n \mathbf{m}^{(i)}_n.
\end{split}
\end{equation}
\Moreover 
\enum{
	\cref{Momentum1vs3:ass1};
	\cref{Momentum1vs3:setting1};
	\cref{Momentum1vs3:setting2};
}
\proves
that for all 
	$n \in \N$
it holds that 
\begin{equation}
\label{Momentum1vs3:eq4}
\begin{split}
	\b^{(1)}_n \c^{(1)}_n 
=
	(1-\alpha^{(1)}_n) \gamma^{(1)}_n
=
	(1-\alpha^{(3)}_n) \gamma^{(3)}_n 
=
	\b^{(3)}_n \c^{(3)}_n.
\end{split}
\end{equation}
\Moreover 
\enum{
	\cref{Momentum1vs3:ass1};
	\cref{Momentum1vs3:setting1};
	\cref{Momentum1vs3:setting2};
}
\proves
that for all 
	$n \in \N$
it holds that
\begin{equation}
\label{Momentum1vs3:eq5}
\begin{split} 
	\frac{
		\a^{(1)}_{n+1} \b^{(1)}_{n}
	}{
		\b^{(1)}_{n+1}
	}
&= 
	\frac{
		\alpha^{(1)}_{n+1} (1-\alpha^{(1)}_{n})
	}{
		1-\alpha^{(1)}_{n+1}
	}
=
	\frac{
		\alpha^{(1)}_{n+1} \gamma^{(3)}_{n} (1-\alpha^{(3)}_{n}) \gamma^{(1)}_{n+1}
	}{
		\gamma^{(1)}_{n} \gamma^{(3)}_{n+1} (1-\alpha^{(3)}_{n+1})
	}\\
&=
	\frac{
		\alpha^{(3)}_{n+1} \gamma^{(3)}_{n} (1-\alpha^{(3)}_{n}) 
	}{
		\gamma^{(3)}_{n+1} (1-\alpha^{(3)}_{n+1})
	}
=
	\frac{
		\a^{(3)}_{n+1} \b^{(3)}_{n}
	}{
		\b^{(3)}_{n+1}
	}.
\end{split}
\end{equation} 
Combining 
\enum{
	this;
	\cref{Momentum1vs3:eq1};
	\cref{Momentum1vs3:eq2};
	\cref{Momentum1vs3:eq3};
	\cref{Momentum1vs3:eq4};
}
with 
\cref{MomentumGeneral}
\proves
\eqref{Momentum1vs3:concl1}.
\end{aproof}

\cfclear
\begin{athm}{lemma}{Momentum1vs4}[Comparison of the \first and \fourth version of the momentum \GD\ optimization method]
Let 
	$\defaultParamDim \in \N$, 
	$(\gamma^{(1)}_n)_{n \in \N} \subseteq (0,\infty)$,
	$(\gamma^{(4)}_n)_{n \in \N} \subseteq (0,\infty)$,
	$(\alpha^{(1)}_n)_{n \in \N} \subseteq (0,1)$,
	$(\alpha^{(4)}_n)_{n \in \N} \subseteq (0,\infty)$,
	$\xi \in \R^\defaultParamDim$,
   	$\defaultLossFunction \in C^1(\R^\defaultParamDim, \R)$
satisfy for all 
	$n \in \N$
that
\begin{equation}
\label{Momentum1vs4:ass1}
\begin{split}
	\gamma^{(1)}_n (1-\alpha^{(1)}_n)
=
	\gamma^{(4)}_n
\qandq
	\frac{
		\gamma^{(1)}_{n+1} \alpha^{(1)}_{n+1}
	}{
		\gamma^{(1)}_{n}
	}
=
	\alpha^{(4)}_{n+1},
\end{split}
\end{equation}
for every $i \in \{1,4\}$\cfadd{def:determ_momentum}\cfadd{def:determ_momentum_three}
let 
	$\Theta^{(i)} \colon \N_0 \to \R^\defaultParamDim$
be the momentum \GD\ process (\ith version) for the objective function $\defaultLossFunction$ with learning rates $(\gamma^{(i)}_n)_{n \in \N}$,
momentum decay factors $(\alpha^{(i)}_n)_{n \in \N}$,
and initial value $\xi$ \cfload.
Then
\begin{equation}
\label{Momentum1vs4:concl1}
\Theta^{(1)} = \Theta^{(4)}.
\end{equation}
\end{athm}

\begin{aproof}
Throughout this proof let 
	$(\a^{(1)}_n)_{n \in \N} \subseteq (0,\infty)$,
	$(\a^{(4)}_n)_{n \in \N} \subseteq (0,\infty)$,
	$(\b^{(1)}_n)_{n \in \N} \subseteq (0,\infty)$,
	$(\b^{(4)}_n)_{n \in \N} \subseteq (0,\infty)$,
	$(\c^{(1)}_n)_{n \in \N} \subseteq (0,\infty)$,
	$(\c^{(4)}_n)_{n \in \N} \subseteq (0,\infty)$
satisfy for all 
	$n \in \N$
that
\begin{equation}
\label{Momentum1vs4:setting1}
\begin{split}
	\a^{(1)}_n = \alpha^{(1)}_n, \qquad \b^{(1)}_n = 1 - \alpha^{(1)}_n, \qquad \c^{(1)}_n = \gamma^{(1)}_n,
\end{split}
\end{equation}
\begin{equation}
\label{Momentum1vs4:setting2}
\begin{split} 
	\a^{(4)}_n = \alpha^{(4)}_n, \qquad \b^{(4)}_n = \gamma^{(4)}_n, \qandq \c^{(4)}_n = 1.
\end{split}
\end{equation}
\Nobs that  
\enum{
	\cref{eq:def:momentum_1_1};
	\cref{eq:def:momentum_1_2};
	\cref{eq:def:momentum_1_3};
	\cref{eq:def:momentum_4_1};
	\cref{eq:def:momentum_4_2};
	\cref{eq:def:momentum_4_3};
}
\prove 
that for all 
	$i \in \{1,4\}$,
	$n \in \N$
it holds that 
\begin{equation}
\label{Momentum1vs4:eq1}
\begin{split}
	\Theta^{(i)}_0 = \xi, \qquad \mathbf{m}^{(i)}_0 = 0,
\end{split}
\end{equation}
\begin{equation}
\label{Momentum1vs4:eq2}
\begin{split}
	\mathbf{m}^{(i)}_n = \a^{(i)}_n \mathbf{m}^{(i)}_{n-1} + \b^{(i)}_n \grad(\Theta^{(i)}_{n-1}),
\end{split}
\end{equation}
\begin{equation}
\label{Momentum1vs4:eq3}
\begin{split}
	\andq \Theta^{(i)}_n = \Theta^{(i)}_{n-1} - \c^{(i)}_n \mathbf{m}^{(i)}_n.
\end{split}
\end{equation}
\Moreover 
\enum{
	\cref{Momentum1vs4:ass1};
	\cref{Momentum1vs4:setting1};
	\cref{Momentum1vs4:setting2};
}
\proves
that for all 
	$n \in \N$
it holds that 
\begin{equation}
\label{Momentum1vs4:eq4}
\begin{split}
	\b^{(1)}_n \c^{(1)}_n 
=
	(1-\alpha^{(1)}_n) \gamma^{(1)}_n
= 
	\gamma^{(4)}_n
=
	\b^{(4)}_n \c^{(4)}_n.
\end{split}
\end{equation}
\Moreover 
\enum{
	\cref{Momentum1vs4:ass1};
	\cref{Momentum1vs4:setting1};
	\cref{Momentum1vs4:setting2};
}
\proves
that for all 
	$n \in \N$
it holds that
\begin{equation}
\label{Momentum1vs4:eq5}
\begin{split} 
	\frac{
		\a^{(1)}_{n+1} \b^{(1)}_{n}
	}{
		\b^{(1)}_{n+1}
	}
= 
	\frac{
		\alpha^{(1)}_{n+1} (1-\alpha^{(1)}_{n})
	}{
		1-\alpha^{(1)}_{n+1}
	}
=
	\frac{
		\alpha^{(1)}_{n+1} 
		\gamma^{(4)}_{n}
		\gamma^{(1)}_{n+1}
	}{
		\gamma^{(1)}_{n}
		\gamma^{(4)}_{n+1}
	}
=
	\frac{
		\alpha^{(4)}_{n+1} 
		\gamma^{(4)}_{n}
	}{
		\gamma^{(4)}_{n+1}
	}
=
	\frac{
		\a^{(4)}_{n+1} \b^{(4)}_{n}
	}{
		\b^{(4)}_{n+1}
	}.
\end{split}
\end{equation} 
Combining 
\enum{
	this;
	\cref{Momentum1vs4:eq1};
	\cref{Momentum1vs4:eq2};
	\cref{Momentum1vs4:eq3};
	\cref{Momentum1vs4:eq4};
}
with 
\cref{MomentumGeneral}
\proves
\eqref{Momentum1vs4:concl1}.
\end{aproof}

\cfclear
\begin{athm}{cor}{Momentum2vs3}[Comparison of the \second and \third version of the momentum \SGD\ optimization method]
Let 
	$\defaultParamDim \in \N$, 
	$(\gamma^{(2)}_n)_{n \in \N} \subseteq (0,\infty)$,
	$(\gamma^{(3)}_n)_{n \in \N} \subseteq (0,\infty)$,
	$(\alpha^{(2)}_n)_{n \in \N} \subseteq (0,\infty)$,
	$(\alpha^{(3)}_n)_{n \in \N} \subseteq (0,1)$,
	$\xi \in \R^\defaultParamDim$,
   	$\defaultLossFunction \in C^1(\R^\defaultParamDim, \R)$
satisfy for all 
	$n \in \N$
that
\begin{equation}
\label{Momentum2vs3:ass1}
\begin{split}
	\gamma^{(2)}_n
=
	\gamma^{(3)}_n (1-\alpha^{(3)}_n)
\qandq
	\frac{
		\gamma^{(2)}_{n+1} \alpha^{(2)}_{n+1}
	}{
		\gamma^{(2)}_{n}
	}
=
	\alpha^{(3)}_{n+1},
\end{split}
\end{equation}
for every $i \in \{2,3\}$\cfadd{def:determ_momentum_two}\cfadd{def:determ_momentum_four}
let 
	$\Theta^{(i)} \colon \N_0 \to \R^\defaultParamDim$
be the momentum \GD\ process (\ith version) for the objective function $\defaultLossFunction$ with learning rates $(\gamma^{(i)}_n)_{n \in \N}$,
momentum decay factors $(\alpha^{(i)}_n)_{n \in \N}$,
and initial value $\xi$ \cfload.
Then
\begin{equation}
\label{Momentum2vs3:concl1}
\Theta^{(2)} = \Theta^{(3)}.
\end{equation}
\end{athm}

\begin{aproof}
Throughout this proof let 
	$(\a^{(2)}_n)_{n \in \N} \subseteq (0,\infty)$,
	$(\a^{(3)}_n)_{n \in \N} \subseteq (0,\infty)$,
	$(\b^{(2)}_n)_{n \in \N} \subseteq (0,\infty)$,
	$(\b^{(3)}_n)_{n \in \N} \subseteq (0,\infty)$,
	$(\c^{(2)}_n)_{n \in \N} \subseteq (0,\infty)$,
	$(\c^{(3)}_n)_{n \in \N} \subseteq (0,\infty)$
satisfy for all 
	$n \in \N$
that
\begin{equation}
\label{Momentum2vs3:setting1}
\begin{split}
	\a^{(2)}_n = \alpha^{(2)}_n, \qquad \b^{(2)}_n = 1, \qquad \c^{(2)}_n = \gamma^{(2)}_n,
\end{split}
\end{equation}
\begin{equation}
\label{Momentum2vs3:setting2}
\begin{split} 
	\a^{(3)}_n = \alpha^{(3)}_n, \qquad \b^{(3)}_n = (1-\alpha^{(3)}_n) \gamma^{(3)}_n, \qandq \c^{(3)}_n = 1.
\end{split}
\end{equation}
\Nobs that  
\enum{
	\cref{eq:def:momentum_2_1};
	\cref{eq:def:momentum_2_2};
	\cref{eq:def:momentum_2_3};
	\cref{eq:def:momentum_3_1};
	\cref{eq:def:momentum_3_2};
	\cref{eq:def:momentum_3_3};
}
\prove 
that for all 
	$i \in \{2,3\}$,
	$n \in \N$
it holds that 
\begin{equation}
\label{Momentum2vs3:eq1}
\begin{split}
	\Theta^{(i)}_0 = \xi, \qquad \mathbf{m}^{(i)}_0 = 0,
\end{split}
\end{equation}
\begin{equation}
\label{Momentum2vs3:eq2}
\begin{split}
	\mathbf{m}^{(i)}_n = \a^{(i)}_n \mathbf{m}^{(i)}_{n-1} + \b^{(i)}_n \grad(\Theta^{(i)}_{n-1}),
\end{split}
\end{equation}
\begin{equation}
\label{Momentum2vs3:eq3}
\begin{split}
	\andq \Theta^{(i)}_n = \Theta^{(i)}_{n-1} - \c^{(i)}_n \mathbf{m}^{(i)}_n.
\end{split}
\end{equation}
\Moreover 
\enum{
	\cref{Momentum2vs3:ass1};
	\cref{Momentum2vs3:setting1};
	\cref{Momentum2vs3:setting2};
}
\proves
that for all 
	$n \in \N$
it holds that 
\begin{equation}
\label{Momentum2vs3:eq4}
\begin{split}
	\b^{(2)}_n \c^{(2)}_n 
=
	\gamma^{(2)}_n
=
	\gamma^{(3)}_n (1-\alpha^{(3)}_n)
=
	\b^{(3)}_n \c^{(3)}_n.
\end{split}
\end{equation}
\Moreover 
\enum{
	\cref{Momentum2vs3:ass1};
	\cref{Momentum2vs3:setting1};
	\cref{Momentum2vs3:setting2};
}
\proves
that for all 
	$n \in \N$
it holds that
\begin{equation}
\label{Momentum2vs3:eq5}
\begin{split} 
	\frac{
		\a^{(2)}_{n+1} \b^{(2)}_{n}
	}{
		\b^{(2)}_{n+1}
	}
= 
	\alpha^{(2)}_{n+1}
=
	\frac{
		\alpha^{(3)}_{n+1} \gamma^{(3)}_{n} (1-\alpha^{(3)}_{n})
	}{
		\gamma^{(3)}_{n+1} (1-\alpha^{(3)}_{n+1})
	}
=
	\frac{
		\a^{(3)}_{n+1} \b^{(3)}_{n}
	}{
		\b^{(3)}_{n+1}
	}.
\end{split}
\end{equation} 
Combining 
\enum{
	this;
	\cref{Momentum2vs3:eq1};
	\cref{Momentum2vs3:eq2};
	\cref{Momentum2vs3:eq3};
	\cref{Momentum2vs3:eq4};
}
with 
\cref{MomentumGeneral}
\proves
\eqref{Momentum2vs3:concl1}.
\end{aproof}

\cfclear
\begin{athm}{lemma}{Momentum2vs4}[Comparison of the \second and \fourth version of the momentum \GD\ optimization method] 
Let 
	$\defaultParamDim \in \N$, 
	$(\gamma^{(2)}_n)_{n \in \N} \subseteq (0,\infty)$,
	$(\gamma^{(4)}_n)_{n \in \N} \subseteq (0,\infty)$,
	$(\alpha^{(2)}_n)_{n \in \N} \subseteq (0,1)$,
	$(\alpha^{(4)}_n)_{n \in \N} \subseteq (0,1)$,
	$\xi \in \R^\defaultParamDim$,
   	$\defaultLossFunction \in C^1(\R^\defaultParamDim, \R)$
satisfy for all  
	$n \in \N$
that
\begin{equation}
\label{Momentum2vs4:ass1}
\begin{split}
	\gamma^{(2)}_n 
=
	\gamma^{(4)}_n
\qandq
	\frac{
		\gamma^{(2)}_{n+1} \alpha^{(2)}_{n+1}
	}{
		\gamma^{(2)}_{n}
	}
=
	\alpha^{(4)}_{n+1},
\end{split}
\end{equation}
for every $i \in \{2,4\}$\cfadd{def:determ_momentum_three}\cfadd{def:determ_momentum_two}
let 
	$\Theta^{(i)} \colon \N_0 \to \R^\defaultParamDim$
be the momentum \GD\ process (\ith version) for the objective function $\defaultLossFunction$ with learning rates $(\gamma^{(i)}_n)_{n \in \N}$,
momentum decay factors $(\alpha^{(i)}_n)_{n \in \N}$,
and initial value $\xi$ \cfload.
Then
\begin{equation}
\label{Momentum2vs4:concl1}
\Theta^{(2)} = \Theta^{(4)}.
\end{equation}
\end{athm}

\begin{aproof}
Throughout this proof let 
	$(\a^{(2)}_n)_{n \in \N} \subseteq (0,\infty)$,
	$(\a^{(4)}_n)_{n \in \N} \subseteq (0,\infty)$,
	$(\b^{(2)}_n)_{n \in \N} \subseteq (0,\infty)$,
	$(\b^{(4)}_n)_{n \in \N} \subseteq (0,\infty)$,
	$(\c^{(2)}_n)_{n \in \N} \subseteq (0,\infty)$,
	$(\c^{(4)}_n)_{n \in \N} \subseteq (0,\infty)$
satisfy for all 
	$n \in \N$
that
\begin{equation}
\label{Momentum2vs4:setting1}
\begin{split}
	\a^{(2)}_n = \alpha^{(2)}_n, \qquad \b^{(2)}_n = 1, \qquad \c^{(2)}_n = \gamma^{(2)}_n,
\end{split}
\end{equation}
\begin{equation}
\label{Momentum2vs4:setting2}
\begin{split} 
	\a^{(4)}_n = \alpha^{(4)}_n, \qquad \b^{(4)}_n = \gamma^{(4)}_n, \qandq \c^{(4)}_n = 1.
\end{split}
\end{equation}
\Nobs that  
\enum{
	\cref{eq:def:momentum_2_1};
	\cref{eq:def:momentum_2_2};
	\cref{eq:def:momentum_2_3};
	\cref{eq:def:momentum_4_1};
	\cref{eq:def:momentum_4_2};
	\cref{eq:def:momentum_4_3};
}
\prove 
that for all 
	$i \in \{2,4\}$,
	$n \in \N$
it holds that 
\begin{equation}
\label{Momentum2vs4:eq1}
\begin{split}
	\Theta^{(i)}_0 = \xi, \qquad \mathbf{m}^{(i)}_0 = 0,
\end{split}
\end{equation}
\begin{equation}
\label{Momentum2vs4:eq2}
\begin{split}
	\mathbf{m}^{(i)}_n = \a^{(i)}_n \mathbf{m}^{(i)}_{n-1} + \b^{(i)}_n \grad(\Theta^{(i)}_{n-1}),
\end{split}
\end{equation}
\begin{equation}
\label{Momentum2vs4:eq3}
\begin{split}
	\andq \Theta^{(i)}_n = \Theta^{(i)}_{n-1} - \c^{(i)}_n \mathbf{m}^{(i)}_n.
\end{split}
\end{equation}
\Moreover 
\enum{
	\cref{Momentum2vs4:ass1};
	\cref{Momentum2vs4:setting1};
	\cref{Momentum2vs4:setting2};
}
\proves
that for all 
	$n \in \N$
it holds that 
\begin{equation}
\label{Momentum2vs4:eq4}
\begin{split}
	\b^{(2)}_n \c^{(2)}_n 
=
	\gamma^{(2)}_n
= 
	\gamma^{(4)}_n
=
	\b^{(4)}_n \c^{(4)}_n.
\end{split}
\end{equation}
\Moreover 
\enum{
	\cref{Momentum2vs4:ass1};
	\cref{Momentum2vs4:setting1};
	\cref{Momentum2vs4:setting2};
}
\proves
that for all 
	$n \in \N$
it holds that
\begin{equation}
\label{Momentum2vs4:eq5}
\begin{split} 
	\frac{
		\a^{(2)}_{n+1} \b^{(2)}_{n}
	}{
		\b^{(2)}_{n+1}
	}
= 
	\alpha^{(2)}_{n+1}
=
	\frac{
		\alpha^{(4)}_{n+1} \gamma^{(4)}_{n}
	}{
		\gamma^{(4)}_{n+1}
	}
=
	\frac{
		\a^{(4)}_{n+1} \b^{(4)}_{n}
	}{
		\b^{(4)}_{n+1}
	}.
\end{split}
\end{equation} 
Combining 
\enum{
	this;
	\cref{Momentum2vs4:eq1};
	\cref{Momentum2vs4:eq2};
	\cref{Momentum2vs4:eq3};
	\cref{Momentum2vs4:eq4};
}
with 
\cref{MomentumGeneral}
\proves
\eqref{Momentum2vs4:concl1}.
\end{aproof}

\cfclear
\begin{athm}{cor}{Momentum3vs4}[Comparison of the \third and \fourth version of the momentum \GD\ optimization method]
Let 
	$\defaultParamDim \in \N$,
	$(\gamma^{(3)}_n)_{n \in \N} \subseteq (0,\infty)$, 
	$(\gamma^{(4)}_n)_{n \in \N} \subseteq (0,\infty)$,
	$(\alpha^{(3)}_n)_{n \in \N} \subseteq (0,1)$,
	$(\alpha^{(4)}_n)_{n \in \N} \subseteq (0,\infty)$,
	$\xi \in \R^\defaultParamDim$,
   	$\defaultLossFunction \in C^1(\R^\defaultParamDim, \R)$
satisfy for all 
	$n \in \N$
that
\begin{equation}
\label{Momentum3vs4:ass1}
\begin{split}
	\gamma^{(3)}_n (1-\alpha^{(3)}_n)
=
	\gamma^{(4)}_n
\qandq
	\alpha^{(3)}_{n+1}
=
	\alpha^{(4)}_{n+1},
\end{split}
\end{equation}
for every $i \in \{3,4\}$\cfadd{def:determ_momentum_three}\cfadd{def:determ_momentum_four}
let 
	$\Theta^{(i)} \colon \N_0 \to \R^\defaultParamDim$
be the momentum \GD\ process (\ith version) for the objective function $\defaultLossFunction$ with learning rates $(\gamma^{(i)}_n)_{n \in \N}$,
momentum decay factors $(\alpha^{(i)}_n)_{n \in \N}$,
and initial value $\xi$ \cfload.
Then
\begin{equation}
\label{Momentum3vs4:concl1}
	\Theta^{(3)} = \Theta^{(4)}.
\end{equation}
\end{athm}

\begin{aproof}
Throughout this proof let 
	$(\a^{(3)}_n)_{n \in \N} \subseteq (0,\infty)$,
	$(\a^{(4)}_n)_{n \in \N} \subseteq (0,\infty)$,
	$(\b^{(3)}_n)_{n \in \N} \subseteq (0,\infty)$,
	$(\b^{(4)}_n)_{n \in \N} \subseteq (0,\infty)$,
	$(\c^{(3)}_n)_{n \in \N} \subseteq (0,\infty)$,
	$(\c^{(4)}_n)_{n \in \N} \subseteq (0,\infty)$
satisfy for all 
	$n \in \N$
that
\begin{equation}
\label{Momentum3vs4:setting1}
\begin{split}
	\a^{(3)}_n = \alpha^{(3)}_n, \qquad \b^{(3)}_n = (1-\alpha^{(3)}_n) \gamma^{(3)}_n, \qquad \c^{(3)}_n = 1
\end{split}
\end{equation}
\begin{equation}
\label{Momentum3vs4:setting2}
\begin{split} 
	\a^{(4)}_n = \alpha^{(4)}_n, \qquad \b^{(4)}_n = \gamma^{(4)}_n, \qandq \c^{(4)}_n = 1,
\end{split}
\end{equation}
\Nobs that  
\enum{
	\cref{eq:def:momentum_3_1};
	\cref{eq:def:momentum_3_2};
	\cref{eq:def:momentum_3_3};
	\cref{eq:def:momentum_4_1};
	\cref{eq:def:momentum_4_2};
	\cref{eq:def:momentum_4_3};
}
\prove 
that for all 
	$i \in \{3,4\}$,
	$n \in \N$
it holds that 
\begin{equation}
\label{Momentum3vs4:eq1}
\begin{split}
	\Theta^{(i)}_0 = \xi, \qquad \mathbf{m}^{(i)}_0 = 0,
\end{split}
\end{equation}
\begin{equation}
\label{Momentum3vs4:eq2}
\begin{split}
	\mathbf{m}^{(i)}_n = \a^{(i)}_n \mathbf{m}^{(i)}_{n-1} + \b^{(i)}_n \grad(\Theta^{(i)}_{n-1}),
\end{split}
\end{equation}
\begin{equation}
\label{Momentum3vs4:eq3}
\begin{split}
	\andq \Theta^{(i)}_n = \Theta^{(i)}_{n-1} - \c^{(i)}_n \mathbf{m}^{(i)}_n.
\end{split}
\end{equation}
\Moreover 
\enum{
	\cref{Momentum3vs4:ass1};
	\cref{Momentum3vs4:setting1};
	\cref{Momentum3vs4:setting2};
}
\proves
that for all 
	$n \in \N$
it holds that 
\begin{equation}
\label{Momentum3vs4:eq4}
\begin{split}
	\b^{(3)}_n \c^{(3)}_n
=
	\gamma^{(3)}_n (1-\alpha^{(3)}_n)
=
	\gamma^{(4)}_n
=
	\b^{(4)}_n \c^{(4)}_n.
\end{split}
\end{equation}
\Moreover 
\enum{
	\cref{Momentum3vs4:ass1};
	\cref{Momentum3vs4:setting1};
	\cref{Momentum3vs4:setting2};
}
\proves
that for all 
	$n \in \N$
it holds that
\begin{equation}
\label{Momentum3vs4:eq5}
\begin{split} 
	\frac{
		\a^{(3)}_{n+1} \b^{(3)}_{n}
	}{
		\b^{(3)}_{n+1}
	}
=
	\frac{
		\alpha^{(3)}_{n+1} (1-\alpha^{(3)}_{n})\gamma^{(3)}_{n} 
	}{
		(1-\alpha^{(3)}_{n+1}) \gamma^{(3)}_{n+1} 
	}
=
	\frac{
		\alpha^{(4)}_{n+1} \gamma^{(4)}_{n} 
	}{
		\gamma^{(4)}_{n+1}
	}
=
	\frac{
		\a^{(4)}_{n+1} \b^{(4)}_{n}
	}{
		\b^{(4)}_{n+1}
	}.
\end{split}
\end{equation} 
Combining 
\enum{
	this;
	\cref{Momentum3vs4:eq1};
	\cref{Momentum3vs4:eq2};
	\cref{Momentum3vs4:eq3};
	\cref{Momentum3vs4:eq4};
}
with 
\cref{MomentumGeneral}
\proves
\eqref{Momentum3vs4:concl1}.
\end{aproof}

\endgroup

\subsection{Representations for momentum optimization}

In \cref{eq:def:momentum_1_1,eq:def:momentum_1_2,eq:def:momentum_1_3} above
the momentum \GD\ optimization method is formulated 
by means of a one-step recursion. 
This one-step recursion can efficiently be exploited 
in an implementation. 
In 
\cref{explicit_momentum_GD} below
we
provide a suitable full-history recursive 
representation for the momentum \GD\ optimization method, 
which enables us to develop a better intuition for the 
momentum \GD\ optimization method.
Our proof of \cref{explicit_momentum_GD} employs the explicit representation of momentum terms in \cref{explicit_momentum} below.
Our proof of \cref{explicit_momentum}, in turn, uses an application of the following result.

\cfclear
\begin{lemma}
\label{bias_adjustment}
Let $(\alpha_n)_{n \in \N} \subseteq \R$ and 
let $(m_n)_{n \in \N_0} \subseteq \R$ satisfy for all
	$n \in \N$
that
$m_0 = 0$ and
\begin{equation}
\label{bias_adjustment:ass1}
\begin{split} 
	m_n = \alpha_n m_{n-1} + 1 - \alpha_n.
\end{split}
\end{equation}
Then it holds for all
	$n \in \N_0$
that
\begin{equation}
\label{bias_adjustment:concl1}
\begin{split} 
	m_n
=
	1 - \prod_{k = 1}^n \alpha_k.
\end{split}
\end{equation}
\end{lemma}

\begin{proof}[Proof of \cref{bias_adjustment}]
We prove \eqref{bias_adjustment:concl1} by induction on $n \in \N_0$. 
For the base case $ n = 0 $ \nobs that 
the assumption that $m_0=0$
\proves that
\begin{equation}
\begin{split} 
	m_0
= 0
=
  1 - \prod_{ k = 1 }^0 \alpha_k
  .
\end{split}
\end{equation} 
This establishes \eqref{bias_adjustment:concl1} in the base case $n = 0$.
For the induction step note that 
\enum{
	\eqref{bias_adjustment:ass1}
}[assure]
that
for all $n \in \N_0$ with 
$
	m_n
=
	1 - \prod_{k = 1}^n \alpha_k
$
it holds that
\begin{equation}
\begin{split} 
	m_{n+1}
&=
	\alpha_{n+1} m_{n} + 1 - \alpha_{n+1}
=
	\alpha_{n+1} 
	\br*{ 
		1 - \prod_{k = 1}^n \alpha_k
	}
	+ 1 - \alpha_{n+1}\\
&=
	\alpha_{n+1} 
	- 
		\prod_{k = 1}^{n+1} \alpha_k
	+ 
	1 
	- 
	\alpha_{n+1}
=
	1
	-
	\prod_{k = 1}^{n+1} \alpha_k.
\end{split}
\end{equation}
Induction hence establishes \eqref{bias_adjustment:concl1}.
The proof of \cref{bias_adjustment} is thus complete.
\end{proof}

\begingroup
\providecommand{\d}{}
\renewcommand{\d}{\defaultParamDim}
\providecommand{\f}{}
\renewcommand{\f}{\defaultLossFunction}
\providecommand{\g}{}
\renewcommand{\g}{\defaultGradientFunction}
\begin{lemma}[An explicit representation of momentum terms]
\label{explicit_momentum}
Let 
	$\d \in \N$, 
	$(\alpha_n)_{n \in \N} \subseteq \R$,
	$(a_{n, k})_{(n, k) \in (\N_0)^2} \subseteq \R$, 
	$(\g_n)_{n \in \N_0} \subseteq \R^\d$,
	$(\bfm_n)_{n \in \N_0} \subseteq \R^\d$
satisfy for all
	$n \in \N$,
	$k \in \{0, 1, \ldots, n-1\}$
that
\begin{equation}
\label{explicit_momentum:ass1}
\begin{split} 
	\mathbf{m}_0 
= 
	0, 
\quad
	\mathbf{m}_{n} 
= 
	\alpha_{n} \mathbf{m}_{n-1} + 
	(1-\alpha_{n})\g_{n-1},
\quad\text{and}\quad
	a_{n, k}
=
	(1-\alpha_{k+1})
	\br*{
		\prod_{l = k + 2}^n
			\alpha_l
	}
\end{split}
\end{equation}
Then 
\begin{enumerate}[label=(\roman *)]
\item \label{explicit_momentum:item1}
it holds for all 
	$n \in \N_0$ 
that
\begin{equation}
	\mathbf{m}_n 
=
	\sum_{k = 0}^{n-1} a_{n, k} \g_{k}
\end{equation}
and

\item \label{explicit_momentum:item2}
it holds for all 
	$n \in \N_0$ 
that
\begin{equation}
	\sum_{k = 0}^{n-1} a_{n, k}
=
	1 - \prod_{k = 1}^n \alpha_k.
\end{equation}
\end{enumerate}
\end{lemma}

\begin{proof}[Proof of \cref{explicit_momentum}]
Throughout this proof, let 
	$(m_n)_{n \in \N_0} \subseteq \R$
satisfy for all
	$n \in \N_0$ 
that
\begin{equation}
\label{explicit_momentum:setting1}
	{m}_n 
= 
	\sum_{k = 0}^{n-1} a_{n, k}.
\end{equation}
We now prove \cref{explicit_momentum:item1} by induction on $n \in \N_0$.
For the base case $n = 0$ note that
\enum{
	\eqref{explicit_momentum:ass1}
}[ensure]
that
\begin{equation}
\label{explicit_momentum:eq1}
\begin{split} 
	\mathbf{m}_0
=
    0
=
  \sum_{ k = 0 }^{ - 1 }
	a_{0, k}\g_{k}.
\end{split}
\end{equation}
This establishes \cref{explicit_momentum:item1} in the base case $n = 0$.
For the induction step \nobs that
\enum{
	\eqref{explicit_momentum:ass1}
}[assure]
that
for all
	$n \in \N_0$
with 
$
	\mathbf{m}_n 
=
	\sum_{k = 0}^{n-1} a_{n, k} \g_{k}
$
it holds that
\begin{equation}
\label{explicit_momentum:eq2}
\begin{split} 
	\mathbf{m}_{n+1}
&=
	\alpha_{n+1}\mathbf{m}_{n} 
	+ 
	(1-\alpha_{n+1})\g_{n}\\
&=
	\br*{\sum_{k = 0}^{n-1} \alpha_{n+1} a_{n, k} \g_{k} }
	+ 
	(1-\alpha_{n+1})\g_{n}\\
&=
	\br*{\sum_{k = 0}^{n-1} 
		\alpha_{n+1}  
		(1-\alpha_{k+1})	
		\br*{\prod_{l = k + 2}^{n}
			\alpha_l 
		}
		\g_{k} 
	}
	+ 
	(1-\alpha_{n+1})\g_{n}
\\
&=
  \br*{
		\sum_{k = 0}^{n-1} 
			(1-\alpha_{k+1})	
			\br*{\prod_{l = k + 2}^{n+1}
				\alpha_l 
			}
			\g_{k}
    }
	+ 
	(1-\alpha_{n+1})\g_{n}
\\
&=
		\sum_{k = 0}^{n} 
			(1-\alpha_{k+1})	
			\br*{\prod_{l = k + 2}^{n+1}
				\alpha_l 
			}
			\g_{k}
=
	\sum_{k = 0}^{n} a_{n+1, k} \g_{k}.
\end{split}
\end{equation}
Induction thus proves \cref{explicit_momentum:item1}.
\Moreover
\enum{
	\cref{explicit_momentum:ass1};
	\cref{explicit_momentum:setting1}
}[demonstrate]
that for all
	$n \in \N$
it holds that $m_0 = 0$ and
\begin{equation}
\label{explicit_momentum:eq3}
\begin{split} 
	m_n
&=
	\sum_{k = 0}^{n-1} a_{n, k}
=
	\sum_{k = 0}^{n-1} 
		(1-\alpha_{k+1})
		\br*{
			\prod_{l = k + 2}^n
				\alpha_l
		}
=
	1-\alpha_n
	+
	\sum_{k = 0}^{n-2} 
		(1-\alpha_{k+1})
		\br*{
			\prod_{l = k + 2}^n
				\alpha_l
		}
\\ &
=
	1-\alpha_n
	+
	\sum_{k = 0}^{n-2}
		(1-\alpha_{k+1})
		\alpha_n
		\br*{
			\prod_{l = k + 2}^{n-1}
				\alpha_l
		}
=
	1-\alpha_n + \alpha_n \sum_{k = 0}^{n-2}  a_{n-1, k}
=
	1 - \alpha_n  + \alpha_n m_{n-1}.
\end{split}
\end{equation}
Combining this with \cref{bias_adjustment} implies that for all
	$n \in \N_0$
it holds that 
\begin{equation}
\label{explicit_momentum:eq4}
\begin{split} 
	m_n 
=
	1 - \prod_{k = 1}^n \alpha_k.
\end{split}
\end{equation}
This establishes \cref{explicit_momentum:item2}.
The proof of \cref{explicit_momentum} is thus complete.
\end{proof}
\endgroup

\begingroup
\providecommand{\d}{}
\renewcommand{\d}{\defaultParamDim}
\providecommand{\f}{}
\renewcommand{\f}{\defaultLossFunction}
\providecommand{\g}{}
\renewcommand{\g}{(\nabla \f)}
\begin{cor}[On a representation of the momentum \GD\ optimization method]
\label{explicit_momentum_GD}
Let 
	$\d \in \N$, 
	$(\gamma_n)_{n \in \N} \subseteq [0,\infty)$, 
	$(\alpha_n)_{n \in \N} \subseteq [0,1]$,
	$(a_{n, k})_{(n, k) \in ( \N_0 )^2} \subseteq \R$, 
	$\xi \in \R^\d$
satisfy for all
	$n \in \N$,
	$k \in \{0, 1, \ldots, n-1\}$
that
\begin{equation}
\label{explicit_momentum_GD:ass1}
\begin{split} 
	a_{n, k}
=
	(1-\alpha_{k+1})
	\br*{
		\prod_{l = k + 2}^n
			\alpha_l
	},
\end{split}
\end{equation}
let $\f \in C^1(\R^\d,\R)$,
and let 
$\Theta$ be the momentum \GD\cfadd{def:determ_momentum}
process for the objective function $\f$ with learning rates $(\gamma_n)_{n \in \N}$, 
momentum decay factors $(\alpha_n)_{n \in \N}$, 
and initial value $\xi$
\cfload.
Then 
\begin{enumerate}[label=(\roman *)]
\item \label{explicit_momentum_GD:item1}
it holds for all 
$ n \in \N $, $ k \in \{ 0, 1, \ldots, n-1\}$
that
$
	0 \leq a_{n, k} \leq 1
$,

\item \label{explicit_momentum_GD:item2}
it holds for all 
$
  n \in \N_0 
$ 
that
\begin{equation}
	\sum_{k = 0}^{n-1} a_{n, k}
=
	1 - \prod_{k = 1}^n \alpha_k,
\end{equation}
and

\item \label{explicit_momentum_GD:item3}
it holds for all 
$ n \in \N $ 
that
\begin{equation}
	\Theta_n 
= 
	\Theta_{ n - 1 }
	- 
	\gamma_n 
	\br*{ \sum_{k = 0}^{n-1} a_{n, k} \g(\Theta_{k}) }.
\end{equation}
\end{enumerate}
\end{cor}

\begin{proof}[Proof of \cref{explicit_momentum_GD}]
Throughout this proof, let 
$
  \mathbf{m} \colon \N_0 \to \R^\d
$
satisfy for all
$ n \in \N $ 
that
\begin{equation}
\label{explicit_momentum_GD:setting1}
  \mathbf{m}_0 = 0
\qandq
  \mathbf{m}_n = \alpha_n \mathbf{m}_{n - 1} + (1 - \alpha_n) \g( \Theta_{n-1} ) .
\end{equation}
\Nobs that
\enum{\eqref{explicit_momentum_GD:ass1}} establishes
\cref{explicit_momentum_GD:item1}. 
\Nobs that 
\enum{
	\eqref{explicit_momentum_GD:ass1};
	\eqref{explicit_momentum_GD:setting1};
	\cref{explicit_momentum}
}[assure]
that for all 
$ n \in \N_0 $ 
it holds that 
\begin{equation}
\label{explicit_momentum_GD:eq1}
  \mathbf{m}_n 
=
  \sum_{k = 0}^{n-1} a_{n, k} \g(\Theta_{k})
\qandq
  \sum_{k = 0}^{n-1} a_{n, k}
=
  1 - \prod_{k = 1}^n \alpha_k 
  .
\end{equation}
This proves \cref{explicit_momentum_GD:item2}. 
\Nobs that 
\enum{
	\eqref{eq:def:momentum_1_1};
	\eqref{eq:def:momentum_1_2};
	\eqref{eq:def:momentum_1_3};
	\eqref{explicit_momentum_GD:setting1};
	\eqref{explicit_momentum_GD:eq1}
}[demonstrate]
that for all
	$n \in \N$
it holds that
\begin{equation}
\label{explicit_momentum_GD:eq2}
\begin{split} 
	\Theta_n 
= 
	\Theta_{n-1} - \gamma_n  \mathbf{m}_n
= 
	\Theta_{ n - 1 }
	- 
	\gamma_n 
	\br*{ \sum_{k = 0}^{n-1} a_{n, k} \g(\Theta_{k}) }.
\end{split}
\end{equation}
This establishes \cref{explicit_momentum_GD:item3}. 
The proof of \cref{explicit_momentum_GD} is thus complete.
\end{proof}
\endgroup

\subsection{Bias-adjusted momentum optimization}
\label{sect:determ_momentum_bias}

\defdetermMomentumBias
\algDescrDetermMomentumBias

\cfclear 
\begingroup
\providecommand{\d}{}
\renewcommand{\d}{\defaultParamDim}
\providecommand{\f}{}
\renewcommand{\f}{\defaultLossFunction}
\providecommand{\g}{}
\renewcommand{\g}{(\nabla \f)}
\begin{cor}[On a representation of the bias-adjusted momentum \GD\ optimization method]
\label{explicit_bias_momentum_GD}
Let 
$ \d \in \N $, 
$
  (\gamma_n)_{n \in \N} \subseteq [0,\infty)
$, 
$ (\alpha_n)_{n \in \N} \subseteq [0,1) $, 
$ \xi \in \R^\d $, 
$
  ( a_{ n, k } )_{ ( n, k ) \in ( \N_0 )^2 } \subseteq \R
$
satisfy for all
	$n \in \N$,
	$k \in \{0, 1, \ldots, n-1\}$
that
\begin{equation}
\label{explicit_bias_momentum_GD:ass1}
\begin{split} 
	a_{n, k}
=
	\frac{
		(1-\alpha_{k+1})
		\br*{
			\prod_{l = k + 2}^n
				\alpha_l
		}
	}{
		1 - \prod_{l = 1}^n \alpha_l
	},
\end{split}
\end{equation}
let $\f \in C^1(\R^\d,\R)$,
and let 
$\Theta$ be the bias-adjusted momentum \GD\cfadd{def:determ_momentum_bias}
process for the objective function $\f$ with 
learning rates $(\gamma_n)_{n \in \N}$, 
momentum decay factors $(\alpha_n)_{n \in \N}$, 
and initial value $\xi$
\cfload.
Then 
\begin{enumerate}[label=(\roman *)]
\item \label{explicit_bias_momentum_GD:item1}
it holds for all 
	$n \in \N$,
	$k \in \{0, 1, \ldots, n-1\}$
that
$
	0 \leq a_{n, k} \leq 1
$,

\item \label{explicit_bias_momentum_GD:item2}
it holds for all 
	$n\in\N$ 
that
\begin{equation}
	\sum_{k = 0}^{n-1} a_{n, k}
=
	1,
\end{equation}
and

\item \label{explicit_bias_momentum_GD:item3}
it holds for all 
	$n\in\N$ 
that
\begin{equation}
	\Theta_n 
= 
	\Theta_{ n - 1 }
	- 
	\gamma_n 
	\br*{ \sum_{k = 0}^{n-1} a_{n, k} \g(\Theta_{k}) }.
\end{equation}
\end{enumerate}
\end{cor}

\begin{proof}[Proof of \cref{explicit_bias_momentum_GD}]
Throughout this proof, let 
$
  \mathbf{m} \colon \N_0 \to \R^\d
$	
satisfy for all
$
  n \in \N 
$ 
that
\begin{equation}
\label{explicit_bias_momentum_GD:setting1}
  \mathbf{m}_0 = 0
\qandq
  \mathbf{m}_n = 
  \alpha_n \mathbf{m}_{n - 1} 
  + 
  (1 - \alpha_n) \g( \Theta_{n-1} )
\end{equation}
and let 
$
  ( b_{ n, k } )_{ (n, k) \in (\N_0)^2 } \subseteq \R
$
satisfy for all
$ n \in \N $, $ k \in \{ 0, 1, \dots, n - 1 \} $
that
\begin{equation}
\label{explicit_bias_momentum_GD:setting2}
\begin{split} 
	b_{n, k}
=
	(1-\alpha_{k+1})
	\br*{
		\prod_{l = k + 2}^n
			\alpha_l
	}.
\end{split}
\end{equation}
\Nobs that
\enum{\eqref{explicit_bias_momentum_GD:ass1}}[imply]
\cref{explicit_bias_momentum_GD:item1}. 
\Nobs that 
\enum{
	\eqref{explicit_bias_momentum_GD:ass1};
	\eqref{explicit_bias_momentum_GD:setting1};
	\eqref{explicit_bias_momentum_GD:setting2};
	\cref{explicit_momentum}
}[assure]
that for all 
	$n \in \N$ 
it holds that 
\begin{equation}
\label{explicit_bias_momentum_GD:eq1}
	\mathbf{m}_n 
=
	\sum_{k = 0}^{n-1} b_{n, k} \g(\Theta_{k})
\qandq
	\sum_{k = 0}^{n-1} a_{n, k}
=
	\frac{
		\sum_{k = 0}^{n-1} b_{n, k}
	}{
		1 - \prod_{k = 1}^n \alpha_k
	}
=
	\frac{
		1 - \prod_{k = 1}^n \alpha_k
	}{
		1 - \prod_{k = 1}^n \alpha_k
	}
=
	1.
\end{equation}
This proves \cref{explicit_bias_momentum_GD:item2}.
\Nobs that 
\enum{
	\eqref{determ_momentum_bias:eq1};
	\eqref{determ_momentum_bias:eq2};
	\eqref{determ_momentum_bias:eq3};
	\eqref{explicit_bias_momentum_GD:ass1};
	\eqref{explicit_bias_momentum_GD:setting1};
	\eqref{explicit_bias_momentum_GD:setting2};
	\eqref{explicit_bias_momentum_GD:eq1}
}[demonstrate]
that for all
	$n \in \N$
it holds that
\begin{equation}
\label{explicit_bias_momentum_GD:eq2}
\begin{split} 
	\Theta_n 
&= 
	\Theta_{n-1} 
	- 
	\frac{\gamma_n  \mathbf{m}_n}{1- \prod_{l=1}^n\alpha_l}
= 
	\Theta_{ n - 1 }
	- 
	\gamma_n 
	\br*{ \sum_{k = 0}^{n-1} 
		\br*{
			\frac{b_{n,k}}{1- \prod_{l=1}^n\alpha_l}
		}
	 	\g(\Theta_{k}) 
	 }\\
&= 
	\Theta_{ n - 1 }
	- 
	\gamma_n 
	\br*{ \sum_{k = 0}^{n-1} a_{n, k} \g(\Theta_{k}) }.
\end{split}
\end{equation}
This establishes \cref{explicit_bias_momentum_GD:item3}. 
The proof of \cref{explicit_bias_momentum_GD} is thus complete.
\end{proof}
\endgroup

\subsection{Error analysis for momentum optimization}
\label{sec:error_analysis_momentum_GD_all}

In this subsection we provide in \cref{subsec:error_analysis_momentum_GD} below 
an error analysis for the momentum \GD\ optimization method 
in the case of a class of quadratic objective functions (cf.\ \cref{convergencerate_momentum} in 
\cref{subsec:error_analysis_momentum_GD} for the precise statement). 
In this specific case we also provide in \cref{subsec:comparison_momentum} below 
a comparison of the convergence speeds of the plain-vanilla \GD\ optimization method 
and the momentum \GD\ optimization method. 
In particular, we prove, roughly speeking, that the momentum \GD\ optimization method outperfoms 
the plain-vanilla \GD\ optimization method in the case of the considered class of quadratic objective functions; 
see \cref{comparison_plain_vs_momentum} in \cref{subsec:comparison_momentum} for the precise statement. 
For this comparison between the plain-vanilla \GD\ optimization method and the 
momentum \GD\ optimization method we employ a refined error analysis of the 
plain-vanilla \GD\ optimization method for the considered class of quadratic objective functions. 
This refined error analysis is the subject of the next section 
(\cref{subsec:error_analysis_momentum_GD_without} below).

In the literature similar error analyses for the momentum \GD\ optimization method can, \eg, be found in
 \cite[Section 7.1]{Bottou2018} and \cite{Polyak64}.

\subsubsection
{Error analysis for GD optimization in the case of 
quadratic objective functions}
\label{subsec:error_analysis_momentum_GD_without}

\cfclear
\begingroup
\newcommand{\lrnorm}[1]{\apnorm2{#1}}
\newcommand{\norm}[1]{\pnorm2{#1}}
\providecommand{\d}{}
\renewcommand{\d}{\defaultParamDim}
\providecommand{\f}{}
\renewcommand{\f}{\defaultLossFunction}
\providecommand{\g}{}
\renewcommand{\g}{\defaultGradientFunction}
\begin{lemma}[Error analysis for the \GD\ optimization method in the case of 
quadratic objective functions]
\label{GD_convergence_quadratic}
Let $\d \in \N$, 
$\xi\in\R^\d$, $\vartheta = (\vartheta_1,\dots,\vartheta_\d) \in \R^\d$, 
 $\kappa, \cK, \lambda_1,\lambda_2,\ldots, \lambda_\d \in (0,\infty)$ satisfy
$\kappa = \min\{\lambda_1,\lambda_2,\ldots, \lambda_\d \}$ and
$\mathcal{K} = \max\{\lambda_1,\lambda_2,\ldots, \lambda_\d \}$,
let $\f \colon \R^\d \to \R$ satisfy for all $\theta = (\theta_1,\dots,\theta_\d) \in \R^\d$ that 
\begin{equation}
\label{GD_convergence_quadratic:assumption0}
\f(\theta) = \tfrac{1}{2} \br*{\sum_{i = 1}^\d\lambda_i \abs{\theta_i - \vartheta_i}^2 },
\end{equation}
and let $\Theta \colon \N_0 \to \R^\d$ satisfy for all $n \in \N$ that
\begin{equation}
\label{GD_convergence_quadratic:assumption1}
\Theta_0  = \xi \qandq \Theta_n = \Theta_{n-1} -  \tfrac{2}{ \mathcal{K} + \kappa }  (\nabla \f)(\Theta_{n-1}).
\end{equation}
Then it holds for all $n \in \N_0$ that
\begin{equation}
\label{eq:GD_convergence_quadratic_statement}
\norm{\Theta_n-\vartheta} \leq \br*{ \tfrac{\mathcal{K} - \kappa}{\mathcal{K} + \kappa} }^n \norm{\xi-\vartheta}
\end{equation}
\cfout.
\end{lemma}

\begin{proof}[Proof of \cref{GD_convergence_quadratic}]
Throughout this proof, 
let $\Theta^{(i)} \colon \N_0 \to \R$, $i \in \{1, 2, \ldots, \d \}$,
satisfy for all $n \in \N_0$ that
$\Theta_n = (\Theta_n^{(1)},\Theta^{(2)}_n, \ldots, \Theta_n^{(\d)})$. 
Note that (\ref{GD_convergence_quadratic:assumption0}) implies that 
for all $\theta = (\theta_1,\ldots,\theta_\d) \in \R^\d$, $i \in \{1, 2, \ldots, \d \}$ it holds that
\begin{equation}
\bpr{\tfrac{\partial \f}{\partial \theta_i}}(\theta) 
=
\lambda_i(\theta_i - \vartheta_i) 
.
\end{equation}
Combining this and \eqref{GD_convergence_quadratic:assumption1} 
ensures that for all $n \in \N$, $i \in \{1, 2, \ldots, \d \}$ it holds that
\begin{equation}
\begin{split}
\Theta_n^{(i)} - \vartheta_i 
&= 
\Theta_{n-1}^{(i)} -  \tfrac{2}{ \mathcal{K} + \kappa } \bpr{\tfrac{\partial \f}{\partial \theta_i}}(\Theta_{n-1}) - \vartheta_i 
\\&= 
\Theta_{n-1}^{(i)} - \vartheta_i -  \tfrac{2}{ \mathcal{K} + \kappa } \bbr{\lambda_i(\Theta_{n-1}^{(i)} - \vartheta_i)} 
\\&= 
\bpr{1- \tfrac{2\lambda_i}{ \mathcal{K} + \kappa }}(\Theta_{n-1}^{(i)} - \vartheta_i)
.
\end{split}
\end{equation}
Hence, we obtain that for all $n \in \N$ it holds that
\begin{equation}
\label{GD_convergence_quadratic:eq1}
\begin{split}
\norm{\Theta_n-\vartheta}^2 &= \sum_{i=1}^\d \abs{\Theta_n^{(i)} - \vartheta_i}^2 
\\
&= \sum_{i=1}^\d
\br*{ 
  \babs{1- \tfrac{2\lambda_i}{ \mathcal{K} + \kappa }}^2  \abs{\Theta_{n-1}^{(i)} - \vartheta_i}^2 
} 
\\
&\leq 
\br*{ 
  \max\bcu{  
    \babs{1- \tfrac{2\lambda_1}{ \mathcal{K} + \kappa }}^2, \ldots, \babs{ 1 - \tfrac{ 2 \lambda_\d }{ \mathcal{K} + \kappa } }^2 
  }
}
  \br*{
    \sum_{i=1}^\d \abs{\Theta_{n-1}^{(i)} - \vartheta_i}^2 
  }
  \\
& = 
  \bbbr{ 
    \max\bcu{  
      \babs{ 1 - \tfrac{ 2 \lambda_1 }{ \mathcal{K} + \kappa } }, \ldots, \babs{ 1 - \tfrac{ 2 \lambda_\d }{ \mathcal{K} + \kappa } } 
    }
  }^2  
  \norm{\Theta_{n-1}-\vartheta}^2
\end{split}
\end{equation}
\cfload.
Moreover, note that the fact that for all $i \in \{1, 2, \ldots, \d \}$ it holds that $\lambda_i \geq \kappa$ 
implies that for all $i \in \{1, 2, \ldots, \d \}$ it holds that
\begin{equation}
\label{GD_convergence_quadratic:eq2}
\begin{split}
1-\tfrac{2\lambda_i }{ \mathcal{K} + \kappa } \leq  1- \tfrac{2\kappa }{ \mathcal{K} + \kappa } 
= \tfrac{\mathcal{K} + \kappa - 2\kappa }{\mathcal{K} + \kappa} = \tfrac{\mathcal{K} - \kappa}{\mathcal{K} + \kappa} \geq 0.
\end{split}
\end{equation}
In addition, observe that the fact that for all $i \in \{1, 2, \ldots, \d \}$ it holds that $\lambda_i \leq \mathcal{K}$ implies that for all $i \in \{1, 2, \ldots, \d \}$ it holds that
\begin{equation}
\begin{split}
1-\tfrac{2\lambda_i }{\mathcal{K} + \kappa} 
\geq  
1- \tfrac{2\mathcal{K} }{\mathcal{K} + \kappa} = \tfrac{\mathcal{K} + \kappa - 2\mathcal{K} }{\mathcal{K} + \kappa} 
= - \br*{ \tfrac{\mathcal{K} - \kappa}{\mathcal{K} + \kappa} } 
\leq 0.
\end{split}
\end{equation}
This and \eqref{GD_convergence_quadratic:eq2} ensure that for all $i \in \{1, 2, \ldots, \d \}$ it holds that
\begin{equation}
\babs{1- \tfrac{2\lambda_i}{\mathcal{K} + \kappa}} 
\leq \tfrac{\mathcal{K} - \kappa}{\mathcal{K} + \kappa} .
\end{equation}
Combining this with (\ref{GD_convergence_quadratic:eq1}) demonstrates that for all $n \in \N$ it holds that
\begin{equation}
\begin{split}
\norm{\Theta_n-\vartheta} &\leq 
\br*{ 
  \max  \bbcu{  \babs{1- \tfrac{2\lambda_1}{\mathcal{K} + \kappa}}, \ldots, \babs{1- \tfrac{2\lambda_\d}{\mathcal{K} + \kappa}}} 
} 
\norm{\Theta_{n-1}-\vartheta} 
\\
&\leq \br*{ \tfrac{\mathcal{K} - \kappa}{\mathcal{K} + \kappa} } 
\norm{\Theta_{n-1}-\vartheta}.
\end{split}
\end{equation}
Induction therefore establishes that for all $n \in \N_0$ it holds that
\begin{equation}
\begin{split}
\norm{\Theta_n-\vartheta} \leq 
\br*{ 
  \tfrac{\mathcal{K} - \kappa}{\mathcal{K} + \kappa}
}^n 
\norm{\Theta_{0}-\vartheta} 
=  
\br*{ \tfrac{\mathcal{K} - \kappa}{\mathcal{K} + \kappa} }^n 
\norm{\xi-\vartheta}.
\end{split}
\end{equation}
The proof of \cref{GD_convergence_quadratic} is thus complete.
\end{proof}
\endgroup

\cref{GD_convergence_quadratic} above establishes, roughly speaking, 
the convergence rate 
$
  \frac{ \mathcal{K} - \kappa }{ \mathcal{K} + \kappa }
$
(see \eqref{eq:GD_convergence_quadratic_statement} above for the precise statement)
for the \GD\ optimization method in the case of 
the objective function in \eqref{GD_convergence_quadratic:assumption0}. 
The next result, \cref{lowerbound_convergencerate} below, 
essentially proves in the situation of \cref{GD_convergence_quadratic} that this convergence rate cannot 
be improved by means of 
a difference choice of the learning rate.

\cfclear
\begingroup
\newcommand{\lrnorm}[1]{\apnorm2{#1}}
\newcommand{\norm}[1]{\pnorm2{#1}}
\providecommand{\d}{}
\renewcommand{\d}{\defaultParamDim}
\providecommand{\f}{}
\renewcommand{\f}{\defaultLossFunction}
\providecommand{\g}{}
\renewcommand{\g}{\defaultGradientFunction}
\begin{lemma}[Lower bound for the convergence rate of \GD\ for quadratic objective functions]
\label{lowerbound_convergencerate}
Let $\d \in \N$, 
$\xi =(\xi_1,\ldots,\xi_\d)$, $\vartheta = (\vartheta_1,\ldots,\vartheta_\d) \in \R^\d$, 
$\gamma,\kappa,\cK, \lambda_1,\lambda_2\ldots, \lambda_\d \in (0,\infty)$
satisfy
$\kappa = \min\{\lambda_1,\lambda_2,\ldots, \lambda_\d \}$
and
$\mathcal{K} = \max\{\lambda_1,\lambda_2,\ldots, \lambda_\d \}$,
let $\f \colon \R^\d \to \R$ satisfy for all 
$\theta = (\theta_1,\theta_2\ldots,\theta_\d) \in \R^\d$ that 
\begin{equation}
\label{lowerbound_convergencerate:assumption0}
\f(\theta) = \tfrac{1}{2} \br*{\sum_{i = 1}^\d\lambda_i \abs{\theta_i - \vartheta_i}^2 },
\end{equation}
and let $\Theta \colon \N_0 \to \R^\d$ satisfy for all $n \in \N$ that
\begin{equation}
\label{lowerbound_convergencerate:assumption1}
\Theta_0  = \xi \qandq \Theta_n = \Theta_{n-1} -  \gamma (\nabla \f)(\Theta_{n-1}).
\end{equation}
Then it holds for all $n \in \N_0$ that
\begin{equation}
\begin{split}
\norm{\Theta_n-\vartheta} 
& \geq 
\bbr{ 
	\max\{\gamma\mathcal{K} - 1, 1- \gamma\kappa \} 
}^n
\bbr{
	\min \bcu{   \abs{\xi_1 - \vartheta_1}, \ldots,  \abs{\xi_\d - \vartheta_\d} }
} \\
&\geq 
\br*{
	\tfrac{\mathcal{K} - \kappa}{\mathcal{K} + \kappa}
}^n
\bbr{
	\min  \bcu{   \abs{\xi_1 - \vartheta_1}, \ldots,  \abs{\xi_\d - \vartheta_\d} }
}
\end{split}
\end{equation}
\cfout.
\end{lemma}

\begin{proof}[Proof of \cref{lowerbound_convergencerate}]
Throughout this proof, 
let $\Theta^{(i)} \colon \N_0 \to \R$, $i \in \{1, 2, \ldots, \d\}$,
satisfy for all $n \in \N_0$ that
$\Theta_n = (\Theta_n^{(1)}, \ldots, \Theta_n^{(\d)})$ and let $\iota, \mathcal{I} \in \{1, 2, \ldots, \d\}$ satisfy  $
\lambda_\iota = \kappa$ and $\lambda_{\mathcal{I}} = \mathcal{K}$. Observe that (\ref{lowerbound_convergencerate:assumption0}) implies that for all $\theta = (\theta_1,\ldots,\theta_\d) \in \R^\d$, $i \in \{1, 2, \ldots, \d \}$ it holds that
\begin{equation}
\bpr{\tfrac{\partial \f}{\partial \theta_i}}(\theta) =\lambda_i(\theta_i - \vartheta_i) .
\end{equation}
Combining this with (\ref{lowerbound_convergencerate:assumption1}) implies that for all $n \in \N$, $i \in \{1, 2, \ldots, \d \}$ it holds that
\begin{equation}
\begin{split}
\Theta_n^{(i)} - \vartheta_i &= \Theta_{n-1}^{(i)} - \gamma \bpr{\tfrac{\partial \f}{\partial \theta_i}}(\Theta_{n-1}) - \vartheta_i \\
&= \Theta_{n-1}^{(i)} - \vartheta_i -  \gamma \lambda_i(\Theta_{n-1}^{(i)} - \vartheta_i) \\
&= (1- \gamma \lambda_i)(\Theta_{n-1}^{(i)} - \vartheta_i).
\end{split}
\end{equation}
Induction 
and 
\cref{lowerbound_convergencerate:assumption1}
hence prove that for all $n \in \N_0$, $i \in \{1, 2, \ldots, \d \}$  it holds that
\begin{equation}
\begin{split}
\Theta_n^{(i)} - \vartheta_i 
= 
(1- \gamma \lambda_i)^n(\Theta_{0}^{(i)} - \vartheta_i) 
= 
(1- \gamma \lambda_i)^n(\xi_i - \vartheta_i).
\end{split}
\end{equation}
This shows that for all $n \in \N_0$ it holds that
\begin{equation}
\label{lowerbound_convergencerate:eq1}
\begin{split}
&\norm{\Theta_n-\vartheta}^2 = \sum_{i=1}^\d \abs{\Theta_n^{(i)} - \vartheta_i}^2 
= 
\sum_{i=1}^\d \bbbr{ \abs{1- \gamma\lambda_i}^{2n} \abs{\xi_i - \vartheta_i}^2} \\
&\geq 
\bbr{
	\min\bcu{   \abs{\xi_1 - \vartheta_1}^2, \ldots,  \abs{\xi_\d - \vartheta_\d}^2 } 
}
\br*{
	\sum_{i=1}^\d \abs{1- \gamma\lambda_i}^{2n}
} \\
&\geq 
\bbr{
	\min\bcu{   \abs{\xi_1 - \vartheta_1}^2, \ldots,  \abs{\xi_\d - \vartheta_\d}^2 }  
}
\bbr{
	\max\{\abs{1- \gamma\lambda_1}^{2n}, \ldots, \abs{1- \gamma\lambda_\d}^{2n} \}
} \\
&= 
\bbr{ 
	\min\bcu{   \abs{\xi_1 - \vartheta_1}, \ldots,  \abs{\xi_\d - \vartheta_\d} } 
}^2 
\bbr{
	\max\{\abs{1- \gamma\lambda_1}, \ldots, \abs{1- \gamma\lambda_\d} \}
}^{2n}
\end{split}
\end{equation}
\cfload.
Furthermore, note that
\begin{equation}
\label{lowerbound_convergencerate:eq2}
\begin{split}
&\max\{\abs{1- \gamma\lambda_1}, \ldots, \abs{1- \gamma\lambda_\d} \}  
\geq 
\max\{\abs{1- \gamma\lambda_\mathcal{I}},  \abs{1- \gamma\lambda_\iota}\}  \\
&= 
\max\{\abs{1- \gamma\mathcal{K}}, \abs{1- \gamma\kappa} \} 
=
\max\{1-\gamma\mathcal K,\gamma\mathcal K-1, 1-\gamma\kappa, \gamma\kappa-1\}
\\&=
\max\{\gamma\mathcal{K} - 1, 1- \gamma\kappa \}
.
\end{split}
\end{equation}
In addition, observe that for all  $\alpha \in (-\infty, \tfrac{2}{\mathcal{K} + \kappa}] $ it holds that
\begin{equation}
\label{lowerbound_convergencerate:eq3}
\max\{\alpha\mathcal{K} - 1, 1- \alpha\kappa \} 
\geq 
1 - \alpha\kappa \geq 1- \br*{\tfrac{2}{\mathcal{K} + \kappa} } \kappa = \tfrac{\mathcal{K} + \kappa-2\kappa}{\mathcal{K} + \kappa} = \tfrac{\mathcal{K} - \kappa}{\mathcal{K} + \kappa}.
\end{equation}
Moreover, note that for all $\alpha \in  [\tfrac{2}{\mathcal{K} + \kappa},\infty)$ it holds that 
\begin{equation}
\max\{\alpha\mathcal{K} - 1, 1- \alpha\kappa \} \geq \alpha\mathcal{K} - 1\geq \br*{\tfrac{2}{\mathcal{K} + \kappa}} \mathcal{K} - 1 = \tfrac{2\mathcal{K} - (\mathcal{K} + \kappa) }{\mathcal{K} + \kappa}  = \tfrac{\mathcal{K} - \kappa}{\mathcal{K} + \kappa}.
\end{equation}
Combining this, (\ref{lowerbound_convergencerate:eq2}), and  (\ref{lowerbound_convergencerate:eq3}) proves that
\begin{equation}
\label{lowerbound_convergencerate:eq4}
\max\{\abs{1- \gamma\lambda_1}, \ldots, \abs{1- \gamma\lambda_\d} \} 
\geq
\max\{\gamma\mathcal{K} - 1, 1- \gamma\kappa \} 
\geq 
\tfrac{\mathcal{K} - \kappa}{\mathcal{K} + \kappa}
\geq 0.
\end{equation}
This and \cref{lowerbound_convergencerate:eq1} demonstrate that for all $n \in \N_0$ it holds that
\begin{equation}
\begin{split}
\norm{\Theta_n-\vartheta}
&\geq  
\bbr{ 
	\max\{\abs{1- \gamma\lambda_1}, \ldots, \abs{1- \gamma\lambda_\d} \}
}^{n}
\bbr{
	\min\bcu{   \abs{\xi_1 - \vartheta_1}, \ldots,  \abs{\xi_\d - \vartheta_\d} }
}\\
& \geq 
\bbr{ 
	\max\{\gamma\mathcal{K} - 1, 1- \gamma\kappa \} 
}^{n}
\bbr{
	\min\bcu{   \abs{\xi_1 - \vartheta_1}, \ldots,  \abs{\xi_\d - \vartheta_\d} }
}\\
&\geq
\br*{
	\tfrac{\mathcal{K} - \kappa}{\mathcal{K} + \kappa}
}^n 
\bbr{
	\min\bcu{   \abs{\xi_1 - \vartheta_1}, \ldots,  \abs{\xi_\d - \vartheta_\d} }
}.
\end{split}
\end{equation}
The proof of \cref{lowerbound_convergencerate} is thus complete.
\end{proof}
\endgroup

\subsubsection
{Error analysis for momentum GD optimization in the case 
of quadratic objective functions}
\label{subsec:error_analysis_momentum_GD}

In this subsection we provide 
in \cref{convergencerate_momentum} below
an error analysis for the momentum \GD\ optimization method 
in the case of a class of quadratic objective functions. 
Our proof of \cref{convergencerate_momentum} 
employs the two auxiliary results on quadratic 
matrices in \cref{spectral_radius} 
and \cref{determinant_block} below.
\cref{spectral_radius} is a special case 
of the so-called Gelfand spectral radius formula 
in the literature. 
\cref{determinant_block} establishes a formula 
for the determinants of quadratic block matrices 
(see \eqref{eq:determinant_block} below for 
the precise statement). 
\cref{determinant_block} and its proof can, \eg, 
be found in Silvester~\cite[Theorem~3]{Silvester2000}.

\begingroup
\newcommand{\lrnorm}[1]{\normmm{#1}}
\newcommand{\norm}[1]{\normmm{#1}}
\providecommand{\d}{}
\renewcommand{\d}{\defaultParamDim}
\providecommand{\f}{}
\renewcommand{\f}{\defaultLossFunction}
\providecommand{\g}{}
\renewcommand{\g}{\defaultGradientFunction}
\begin{lemma}[A special case of Gelfand's spectral radius formula for real matrices]
\label{spectral_radius}
Let $\d \in \N$, $A \in \R^{\d \times \d}$, 
$\mathscr{S} = \{ \lambda \in \C \colon (\exists \, v \in \C^\d\backslash \{0\} \colon Av = \lambda v ) \}$ 
and 
let $\lrnorm{\cdot} \colon \R^\d \to [0,\infty)$ be a norm. Then 
\begin{equation}
	\label{spectral_radius.1}
\liminf_{n \to \infty} 
\pr*{
	\br*{
		\sup_{v\in \R^\d\backslash \{0\}} \frac{\norm{A^n v}}{\norm{v}} 
	}^{\nicefrac{1}{n}} 
}
= 
\limsup_{n \to \infty} 
\pr*{
	\br*{
		\sup_{v\in \R^\d\backslash \{0\}} \frac{\norm{A^n v}}{\norm{v}} 
	}^{ \nicefrac{1}{n}}
}
=
\max_{\lambda \in \mathscr{S} \cup \{0\} } \abs{\lambda}.
\end{equation}
\end{lemma}
\begin{proof}[Proof of \cref{spectral_radius}]
	Note that, \eg, Einsiedler \& Ward~\cite[Theorem 11.6]{EinsiedlerWard15}
	establishes
	\cref{spectral_radius.1} (cf., \eg, Tropp~\cite{Tropp01}).
	The proof of \cref{spectral_radius} is thus complete.
\end{proof}
\endgroup

\cfclear
\begingroup
\newcommand{\D}[1]{\mathcal{D}_{#1}}

\begin{lemma}[Determinants for block matrices]
\label{determinant_block}
Let $\d \in \N$, $A,B,C,D \in \C^{\d \times \d}$ satisfy  $CD = DC$. Then
\begin{equation}
\label{eq:determinant_block}
\det \!
\underbrace{\begin{pmatrix}
A & B \\
C & D
\end{pmatrix}}_{\in \, \R^{(2\d)\times(2\d)}}
= 
\det (AD-BC)
\end{equation}
\end{lemma}

\begin{proof}[Proof of \cref{determinant_block}]
Throughout this proof, let $\D{x} \in \C^{\d \times \d}$, $x \in \C$, satisfy for all $x \in \C$ that
\begin{equation}
\D{x} = D-x \idMatrix_{\d}\ifnocf.
\end{equation}
\cfload[.]%
Observe that the fact that for all $x\in \C$ it holds that $C\D{x} = \D{x}C$ and the fact that for all $X,Y,Z \in \C^{\d \times \d}$ it holds that 
\begin{equation}
	\label{eq:determinant_block.1}
\det
\begin{pmatrix}
X & Y \\
0 & Z
\end{pmatrix}  = \det(X)\det(Z) = 
\det
\begin{pmatrix}
X & 0 \\
Y & Z
\end{pmatrix}
\end{equation}
(cf., \eg, Petersen~\cite[Proposition 5.5.3 and Proposition 5.5.4]{Petersen12}) imply that for all $x \in \C$ it holds that
\begin{equation}
\label{determinant_block:eq1}
\begin{split}
\det \pr*{\!
\begin{pmatrix}
A & B \\
C & \D{x}
\end{pmatrix}
\begin{pmatrix}
\D{x} & 0 \\
-C & \idMatrix_\d
\end{pmatrix}\!
}
 &= 
 \det
 \begin{pmatrix}
(A\D{x}-BC) & B \\
(C\D{x} - \D{x}C) & \D{x}
\end{pmatrix} \\
&= 
 \det
 \begin{pmatrix}
(A\D{x}-BC) & B \\
0 & \D{x}
\end{pmatrix} \\
&=  \det(A\D{x}-BC)\det(\D{x}).
\end{split}
\end{equation}
Moreover, note that 
	\eqref{eq:determinant_block.1} and
	the multiplicative property of the determinant (see, \eg, Petersen~\cite[(1) in Proposition 5.5.2]{Petersen12}) 
imply that for all $x \in \C$ it holds that
\begin{equation}
\begin{split}
\det \pr*{ \!
\begin{pmatrix}
A & B \\
C & \D{x}
\end{pmatrix}
\begin{pmatrix}
\D{x} & 0 \\
-C & \idMatrix_\d
\end{pmatrix}
\! } 
&= 
\det 
\begin{pmatrix}
A & B \\
C & \D{x}
\end{pmatrix}
\det 
\begin{pmatrix}
\D{x} & 0 \\
-C & \idMatrix_\d
\end{pmatrix}\\
&=
\det 
\begin{pmatrix}
A & B \\
C & \D{x}
\end{pmatrix}
\det(\D{x})\det(\idMatrix_\d) \\
&= 
\det 
\begin{pmatrix}
A & B \\
C & \D{x}
\end{pmatrix}
\det(\D{x}).
\end{split} 
\end{equation}
Combining this and (\ref{determinant_block:eq1}) demonstrates that for all $x \in \C$ it holds that
\begin{equation}
\det
\begin{pmatrix}
A & B \\
C & \D{x}
\end{pmatrix}
\det(\D{x}) 
=
\det(A\D{x}-BC)\det(\D{x}).
\end{equation}
Hence, we obtain for all $x \in \C$ that
\begin{equation}
\pr*{ \det
\begin{pmatrix}
A & B \\
C & \D{x}
\end{pmatrix}
- \det(A\D{x}-BC)
}
\det(\D{x}) = 0.
\end{equation}
This implies that for all $x \in \C$ with $\det(\D{x}) \neq 0$ it holds that
\begin{equation}
\label{determinant_block:eq2}
\det\begin{pmatrix}
A & B \\
C & \D{x}
\end{pmatrix}
- \det(A\D{x}-BC) = 0.
\end{equation}
Moreover, note that the fact that
    $\C\ni x\mapsto \det(D-x\idMatrix_\d)\in\C$
is a polynomial function of degree $\d$
ensures that
$
 \{ x \in \C \colon \det(\D{x}) = 0\} = \{ x \in \C \colon \det(D-x\idMatrix_\d) = 0\}
$
is a finite set.
Combining
	this
	and \cref{determinant_block:eq2}
with
the fact that the function 
\begin{equation}
	\C \ni x \mapsto \det\begin{pmatrix}
A & B \\
C & \D{x}
\end{pmatrix}
- \det(A\D{x}-BC) \in \C
\end{equation}
is continuous
shows that for all $x \in \C $ it holds that
\begin{equation}
\det\begin{pmatrix}
A & B \\
C & \D{x}
\end{pmatrix}
- \det(A\D{x}-BC) = 0.
\end{equation}
Hence, we obtain for all $x \in \C$ that
\begin{equation}
\det\begin{pmatrix}
A & B \\
C & \D{x}
\end{pmatrix}
= \det(A\D{x}-BC).
\end{equation}
This establishes that
\begin{equation}
\det\begin{pmatrix}
A & B \\
C & D
\end{pmatrix} 
=
\det\begin{pmatrix}
A & B \\
C & \D0
\end{pmatrix} 
=
\det(A\D0-BC) = \det(A\D0-BC).
\end{equation}
The proof of \cref{determinant_block} is thus completed.
\end{proof}

We are now in the position to formulate and prove 
the promised error analysis 
for the momentum \GD\ optimization method 
in the case of the considered class of quadratic objective functions; 
see \cref{convergencerate_momentum} below.

\begingroup
\newcommand{\lrnorm}[1]{\apnorm2{#1}}
\newcommand{\norm}[1]{\apnorm2{#1}}
\newcommand{\Norm}[1]{\bpnorm2{#1}}
\renewcommand{\normmm}[1]{\apnorm2{#1}}
\newcommand{\Rdnorm}[2][x]{\opnorm{#2}}
\newcommand{\rdnorm}[2][x]{\opnorm{#2}}
\providecommand{\d}{}
\renewcommand{\d}{\defaultParamDim}
\providecommand{\f}{}
\renewcommand{\f}{\defaultLossFunction}
\providecommand{\g}{}
\renewcommand{\g}{\defaultGradientFunction}
\cfclear
\begin{prop}[Error analysis for the momentum \GD\ optimization method in the case of quadratic objective functions]
\label{convergencerate_momentum}
Let $\d \in \N$, 
$\xi\in\R^\d$, $\vartheta = (\vartheta_1,\dots,\vartheta_\d) \in \R^\d$, 
 $\kappa,\cK,\lambda_1,\lambda_2,\dots, \lambda_\d \in (0,\infty)$ satisfy
$\kappa = \min\{\lambda_1,\lambda_2,\dots, \lambda_\d \}$ and $\mathcal{K} = \max\{\lambda_1,\lambda_2,\dots,\allowbreak \lambda_\d \}$, 
let $\f \colon \R^\d \to \R$ satisfy for all $\theta = (\theta_1,\dots,\theta_\d) \in \R^\d$ that 
\begin{equation}
\label{convergencerate_momentum:assumption0}
\f(\theta) 
=
\tfrac{1}{2} \br*{\sum_{i = 1}^\d\lambda_i \abs{\theta_i - \vartheta_i}^2 },
\end{equation}
and let $\Theta \colon \N_0 \cup \{-1 \} \to \R^\d$ satisfy for all $n \in \N$ that $\Theta_{-1} = \Theta_0  = \xi$ and
\begin{equation}
\label{convergencerate_momentum:assumption1}
	\Theta_n 
= 
	{\Theta}_{n-1} -  \tfrac{4}{(\sqrt{\mathcal{K}} +\sqrt{\kappa} )^2}  (\nabla \f)({\Theta}_{n-1}) 
	+ \br*{\tfrac{\sqrt{\mathcal{K}} -\sqrt{\kappa} }{\sqrt{\mathcal{K}}+\sqrt{\kappa}}}^2 ({\Theta}_{n-1} - {\Theta}_{n-2}).
\end{equation}
Then 
\begin{enumerate}[label=(\roman *)]
\item \label{convergencerate_momentum:item1}
it holds that
$
\Theta|_{\N_0} \colon \N_0 \to \R^\d
$ 
is the momentum \GD\ process for the objective function $\f$ with learning rates 
$
\N \ni n \mapsto \frac{1}{\sqrt{\mathcal{K}\kappa}} \in [0,\infty)
$,
momentum decay factors 
$
\N \ni n \mapsto \bbr{ \tfrac{\mathcal{K}^{1/2} -\kappa^{1/2} }{\mathcal{K}^{1/2}  + \kappa^{1/2}} }^2 \in [0,1]
$, 
and initial value $\xi$  and

\item \label{convergencerate_momentum:item2}
for every $\varepsilon \in (0,\infty)$ there exists $\mathfrak{c} \in \R$ 
such that for all $n \in \N_0$ it holds that
\begin{equation}
	\norm{\Theta_n - \vartheta} 
\leq 
	\mathfrak{c}
	\br*{
		\tfrac{\sqrt{\mathcal{K}} -\sqrt{\kappa} }{\sqrt{\mathcal{K}}+\sqrt{\kappa}} + \varepsilon 
	}^n
\end{equation}
\end{enumerate}
\cfadd{def:determ_momentum}
\cfout.
\end{prop}

\cfclear
\begin{proof}[Proof of \cref{convergencerate_momentum}]
Throughout this proof, 
let $\varepsilon \in (0,\infty)$,
let $\Rdnorm[(2\d)\times(2\d)]{\cdot} \colon \R^{(2\d)\times(2\d)} \to [0,\infty)$ satisfy for all $B \in \R^{(2\d)\times(2\d)}$ that 
\begin{equation}
	\rdnorm[2\d\times2\d]{B}
= 
	\sup_{v\in \R^{2\d}\backslash \{0\}} \br*{\frac{\normmm{Bv}}{\normmm{v}} },
\end{equation}
let $\Theta^{(i)} \colon \N_0 \to \R$, $i \in \{1, 2, \ldots, \d\}$,
satisfy for all $n \in \N_0$ that
$\Theta_n = (\Theta_n^{(1)}, \ldots, \Theta_n^{(\d)})$,
let $\mathbf{m} \colon \N_0 \to \R^\d$ satisfy for all $n \in \N_0$ that
\begin{equation}
\label{convergencerate_momentum:eq-1}
\mathbf{m}_n = -\sqrt{\mathcal{K}\kappa}(\Theta_n - \Theta_{n-1}),
\end{equation} 
let $\varrho \in  (0,\infty)$, $\alpha \in [0,1)$ be given by 
\begin{equation}
\label{convergencerate_momentum:eq0}
	\varrho = \tfrac{4}{(\sqrt{\mathcal{K}} +\sqrt{\kappa} )^2} 
\qandq 
	\alpha 
= 
	\br*{\tfrac{\sqrt{\mathcal{K}} -\sqrt{\kappa} }{\sqrt{\mathcal{K}}+\sqrt{\kappa}}}^2,
\end{equation} 
let $M \in \R^{\d \times \d}$ be the diagonal $(\d \times \d)$-matrix given by 
\begin{equation}
\label{convergencerate_momentum:eq1}
M = 
\begin{pmatrix}
(1-\varrho \lambda_1 + \alpha) & &0 \\
& \ddots & \\
0& & (1-\varrho \lambda_\d + \alpha)
\end{pmatrix},
\end{equation}
\cfclear
let $A \in \R^{2\d \times 2\d}$ be the $((2\d)\times (2\d))$-matrix given by 
\begin{equation}
\label{convergencerate_momentum:eq1.0}
	A 
= 
	\begin{pmatrix}
		M& (-\alpha \idMatrix_{\d}) \\
		\idMatrix_\d& 0
	\end{pmatrix},
\end{equation}
and let $\mathscr{S} \subseteq \C$ be the set given by
\begin{equation}
\label{convergencerate_momentum:eq1.3}
\mathscr{S} = \{\mu \in \C \colon (\exists \, v \in \C^{2\d}\backslash \{0\} \colon Av = \mu v) \}=\{\mu \in \C \colon \det(A-\mu \idMatrix_{2\d}) = 0 \}
\end{equation} 
\cfload. 
Observe that (\ref{convergencerate_momentum:assumption1}), (\ref{convergencerate_momentum:eq-1}), and the fact that 
\begin{equation}
\begin{split}
	\tfrac{(\sqrt{\mathcal{K}} + \sqrt{\kappa})^2 - (\sqrt{\mathcal{K}} - \sqrt{\kappa})^2}{4} 
&=
	\tfrac{1}{4} 
	\br*{
		(\sqrt{\mathcal{K}} + \sqrt{\kappa} + \sqrt{\mathcal{K}} - \sqrt{\kappa}) (\sqrt{\mathcal{K}} + \sqrt{\kappa} - [\sqrt{\mathcal{K}} - \sqrt{\kappa}] )
	} \\
&=
	\tfrac{1}{4} 
	\br*{
		(2\sqrt{\mathcal{K}}) (2\sqrt{\kappa})
	} 
=
	\sqrt{\mathcal{K}\kappa}
\end{split}
\end{equation}
assure that for all $n \in \N$ it holds that
\begin{equation}
\label{convergencerate_momentum:eq1.1}
\begin{split}
	\mathbf{m}_n 
&= 
	-\sqrt{\mathcal{K}\kappa}(\Theta_n - \Theta_{n-1}) \\
&= 
	-\sqrt{\mathcal{K}\kappa} 
	\pr*{ 
		\Theta_{n-1} 
		-  \br*{ \tfrac{4}{(\sqrt{\mathcal{K}} +\sqrt{\kappa} )^2}} (\nabla \f)({\Theta}_{n-1}) 
		+ \br*{\tfrac{\sqrt{\mathcal{K}} -\sqrt{\kappa} }{\sqrt{\mathcal{K}}+\sqrt{\kappa}}}^2 ({\Theta}_{n-1} - {\Theta}_{n-2}) - \Theta_{n-1} 
	} \\
&= 
	\sqrt{\mathcal{K}\kappa} 
	\pr*{ 
		\br*{ \tfrac{4}{(\sqrt{\mathcal{K}} +\sqrt{\kappa} )^2}} (\nabla \f)({\Theta}_{n-1}) 
		- \br*{\tfrac{\sqrt{\mathcal{K}} -\sqrt{\kappa} }{\sqrt{\mathcal{K}}+\sqrt{\kappa}}}^2 ({\Theta}_{n-1} - {\Theta}_{n-2})
	} \\	
&=    
	\tfrac{(\sqrt{\mathcal{K}} + \sqrt{\kappa})^2 - (\sqrt{\mathcal{K}} - \sqrt{\kappa})^2}{4} 
	\br*{\tfrac{4}{(\sqrt{\mathcal{K}} +\sqrt{\kappa} )^2}} (\nabla \f)({\Theta}_{n-1}) \\
	&\quad 
	-\sqrt{\mathcal{K}\kappa}
	\br*{\tfrac{\sqrt{\mathcal{K}} -\sqrt{\kappa} }{\sqrt{\mathcal{K}}+\sqrt{\kappa}}}^2 ({\Theta}_{n-1} - {\Theta}_{n-2}) \\
&=   
	\br*{ 1 - \tfrac{(\sqrt{\mathcal{K}} - \sqrt{\kappa})^2}{(\sqrt{\mathcal{K}} +\sqrt{\kappa} )^2}} (\nabla \f)({\Theta}_{n-1})  
	+ \br*{\tfrac{\sqrt{\mathcal{K}} -\sqrt{\kappa} }{\sqrt{\mathcal{K}}+\sqrt{\kappa}}}^2
	\br*{-\sqrt{\mathcal{K}\kappa} ({\Theta}_{n-1} - {\Theta}_{n-2})} \\
&= 
	\br*{ 1 -  \br*{\tfrac{\sqrt{\mathcal{K}} -\sqrt{\kappa} }{\sqrt{\mathcal{K}}+\sqrt{\kappa}}}^2} (\nabla \f)({\Theta}_{n-1}) 
	+ \br*{\tfrac{\sqrt{\mathcal{K}} -\sqrt{\kappa} }{\sqrt{\mathcal{K}}+\sqrt{\kappa}}}^2\mathbf{m}_{n-1}.
\end{split}
\end{equation}
Moreover, note that (\ref{convergencerate_momentum:eq-1}) implies that for all $n \in \N_0$ it holds that
\begin{equation}
\label{convergencerate_momentum:eq1.2}
\begin{split}
	\Theta_n 
&=
	\Theta_{n-1} + (\Theta_n - \Theta_{n-1}) \\
&=  
	\Theta_{n-1} 
	-\tfrac{1}{\sqrt{\mathcal{K}\kappa}}
	\pr*{ \br*{-\sqrt{\mathcal{K}\kappa}} (\Theta_n - \Theta_{n-1})} 
= 
	\Theta_{n-1} - \tfrac{1}{\sqrt{\mathcal{K}\kappa}}\mathbf{m}_{n}.
\end{split}
\end{equation}
In addition, observe that the assumption that $\Theta_{-1} = \Theta_0 = \xi$ and (\ref{convergencerate_momentum:eq-1}) ensure that 
\begin{equation}
  \mathbf{m}_{0} = -\sqrt{\mathcal{K}\kappa} \, \bpr{ \Theta_0 - \Theta_{-1} }= 0 .
\end{equation}
Combining this and the assumption that $\Theta_0 = \xi$ with (\ref{convergencerate_momentum:eq1.1}) and (\ref{convergencerate_momentum:eq1.2}) proves \cref{convergencerate_momentum:item1}.
 It thus remains to prove \cref{convergencerate_momentum:item2}. 
 For this observe that (\ref{convergencerate_momentum:assumption0}) implies that for all $\theta = (\theta_1,\ldots,\theta_\d) \in \R^\d$, $i \in \{1, 2, \ldots, \d \}$ it holds that
\begin{equation}
	\bpr{\tfrac{\partial \f}{\partial \theta_i}}(\theta) 
=
	\lambda_i(\theta_i - \vartheta_i) .
\end{equation}
This, \eqref{convergencerate_momentum:assumption1}, 
and \eqref{convergencerate_momentum:eq0} imply that for all $n \in \N$, $i \in \{1, 2, \ldots, \d \}$ it holds that
\begin{equation}
\begin{split}
\Theta_n^{(i)} - \vartheta_i 
&=
	\Theta_{n-1}^{(i)} - \varrho \bpr{\tfrac{\partial \f}{\partial \theta_i}}(\Theta_{n-1}) + \alpha (\Theta_{n-1}^{(i)} - \Theta_{n-2}^{(i)}) - \vartheta_i \\
&= 
	(\Theta_{n-1}^{(i)} - \vartheta_i) - \varrho \lambda_i(\Theta_{n-1}^{(i)} - \vartheta_i) + \alpha \bpr{(\Theta_{n-1}^{(i)} - \vartheta_i) - (\Theta_{n-2}^{(i)}-\vartheta_i) } \\
&= 
	(1 - \varrho \lambda_i + \alpha)(\Theta_{n-1}^{(i)} - \vartheta_i) -\alpha (\Theta_{n-2}^{(i)}-\vartheta_i).
\end{split}
\end{equation}
Combining this with \eqref{convergencerate_momentum:eq1} demonstrates that for all $n \in \N$ it holds that
\begin{equation}
\begin{split}
  \R^\d \ni 
  \pr*{ 
    \Theta_n - \vartheta 
  }
& =
  M
  \pr*{
    \Theta_{ n - 1 } - \vartheta
  }
  -
  \alpha 
  \pr*{
    \Theta_{ n - 2 } - \vartheta
  }
\\ & = 
	\underbrace{
		\begin{pmatrix}
			M & (-\alpha \idMatrix_{\d})
		\end{pmatrix}
		}
	_{\in\, \R^{\d \times 2\d}}
	\underbrace{
		\begin{pmatrix}
			\Theta_{n-1} - \vartheta  \\
			\Theta_{n-2} - \vartheta 
		\end{pmatrix}
		}
	_{\in \,\R^{2\d}}.
\end{split}
\end{equation}
This and \eqref{convergencerate_momentum:eq1.0} assure that for all $n \in \N$ it holds that
\begin{equation}
	\R^{2\d} 
\ni 
	\begin{pmatrix}
		\Theta_n - \vartheta  \\
		\Theta_{n-1} - \vartheta 
	\end{pmatrix}
= 
	\begin{pmatrix}
		M& (-\alpha \idMatrix_{\d}) \\
		\idMatrix_{\d}& 0
	\end{pmatrix}
	\begin{pmatrix}
		\Theta_{n-1} - \vartheta  \\
		\Theta_{n-2} - \vartheta 
	\end{pmatrix} 
=
	A 
	\begin{pmatrix}
		\Theta_{n-1} - \vartheta  \\
		\Theta_{n-2} - \vartheta 
	\end{pmatrix}.
\end{equation}
Induction hence proves that for all $n \in \N_0$ it holds that
\begin{equation}
\	\R^{2\d} 
\ni 
	\begin{pmatrix}
		\Theta_n - \vartheta  \\
		\Theta_{n-1} - \vartheta 
	\end{pmatrix}
= 
	A^n
	\begin{pmatrix}
		\Theta_0 - \vartheta  \\
		\Theta_{-1} - \vartheta 
	\end{pmatrix}
=
	A^n
	\begin{pmatrix}
		\xi - \vartheta  \\
		\xi - \vartheta 
	\end{pmatrix}.
\end{equation}
This implies that for all $n \in \N_0$ it holds that
\begin{equation}
\label{convergencerate_momentum:eq2}
\begin{split}
	\norm{\Theta_n-\vartheta} &\leq \sqrt{\norm{\Theta_n-\vartheta}^2 + \norm{\Theta_{n-1}-\vartheta}^2} \\
&= 
	\normmm{
		\begin{pmatrix}
			\Theta_n - \vartheta  \\
			\Theta_{n-1} - \vartheta 
		\end{pmatrix}
	} \\
&= 
	\normmm{
		A^n
		\begin{pmatrix}
			\xi - \vartheta  \\
			\xi - \vartheta 
		\end{pmatrix}
	} \\
&\leq 
	\rdnorm[(2\d)\times(2\d)]{A^n}
	\normmm{
		\begin{pmatrix}
			\xi - \vartheta  \\
			\xi - \vartheta 
		\end{pmatrix}
	} \\
&= 
	\rdnorm[(2\d)\times(2\d)]{A^n} \sqrt{\norm{\xi-\vartheta}^2 + \norm{\xi-\vartheta}^2} \\
&= 
	\rdnorm[(2\d)\times(2\d)]{A^n} \sqrt{2}\norm{\xi-\vartheta}.
\end{split}
\end{equation}
Next note that \cref{convergencerate_momentum:eq1.3} and \cref{spectral_radius} demonstrate that
\begin{equation}
	\limsup_{n \to \infty} \pr*{ \bbr{\rdnorm[(2\d)\times(2\d)]{A^n} }^{\nicefrac{1}{n}} } 
= 
	\liminf_{n \to \infty} \pr*{ \bbr{\rdnorm[(2\d)\times(2\d)]{A^n} }^{\nicefrac{1}{n}} } 
=
	\max_{\mu \in \mathscr{S} \cup \{0\}} \abs{\mu}.
\end{equation}
This implies that there exists $m \in \N$ which satisfies for all $n \in \N \cap [m,\infty)$ that
\begin{equation}
\label{convergencerate_momentum:eq2.1}
	\bbr{\rdnorm[(2\d)\times(2\d)]{A^n} }^{\nicefrac{1}{n}} 
\leq 
	\varepsilon + \max_{\mu \in \mathscr{S}\cup \{0\}} \abs{\mu}.
\end{equation}
Note that 
\cref{convergencerate_momentum:eq2.1}
implies that 
for all $n \in \N \cap [m,\infty)$ it holds that
\begin{equation}
\label{convergencerate_momentum:eq2.2}
	\rdnorm[(2\d)\times(2\d)]{A^n} 
\leq 
	\bbbr{\varepsilon + \max_{\mu \in \mathscr{S}\cup \{0\}} \abs{\mu}   }^{n}.
\end{equation}
Furthermore, note that for all $n \in \N \cap [0,m)$ it holds that
\begin{equation}
\begin{split}
	\rdnorm[(2\d)\times(2\d)]{A^n} 
&= 
	\bbbr{\varepsilon + \max_{\mu \in \mathscr{S}\cup \{0\}} \abs{\mu}   }^{n} 
	\br*{\tfrac{\rdnorm[(2\d)\times(2\d)]{A^n}}{(\varepsilon + \max_{\mu \in \mathscr{S}\cup \{0\}} \abs{\mu})^{n}}} \\
&\leq 
	\bbbr{ 
		\varepsilon +\max_{\mu \in \mathscr{S}\cup \{0\}} \abs{\mu} 
	}^{n} 
	\br*{\max  
		\pr*{
	 		\cu*{
	 			\tfrac{\rdnorm[(2\d)\times(2\d)]{A^k}}{(\varepsilon + \max_{\mu \in \mathscr{S}\cup \{0\}} \abs{\mu})^{k}} \colon k \in \N_0 \cap [0,m)
	 		} 
	 		\cup \{ 1\}
		}
	}.
\end{split}
\end{equation}
Combining this and \eqref{convergencerate_momentum:eq2.2} proves that for all $n \in \N_0$ it holds that
\begin{equation}
\label{convergencerate_momentum:eq3}
\begin{split}
	\rdnorm[(2\d)\times(2\d)]{A^n} 
&\leq 
	\bbbr{ 
		\varepsilon +\max_{\mu \in \mathscr{S}\cup \{0\}} \abs{\mu} 
	}^{n} 
	\br*{
		\max 
		\pr*{
	 		\cu*{
	 			\tfrac{\rdnorm[(2\d)\times(2\d)]{A^k}}{(\varepsilon + \max_{\mu \in \mathscr{S}\cup \{0\}} \abs{\mu})^{k}} \colon k \in \N_0 \cap [0,m)
	 		} 
	 		\cup \{ 1\}
		}
	}.
\end{split}
\end{equation}
Next observe that \cref{determinant_block}, \eqref{convergencerate_momentum:eq1.0}, and the fact that 
for all $\mu \in \C$ it holds that 
$
	\idMatrix_{\d}(-\mu \idMatrix_{\d})
=
	-\mu \idMatrix_{\d}
=
	(-\mu \idMatrix_{\d})\idMatrix_{\d}
$
ensure that for all $\mu \in \C$ it holds that
\begin{equation}
\begin{split}
	\det(A-\mu \idMatrix_{2\d}) 
&= 
	\det  
	\begin{pmatrix}
		(M-\mu \idMatrix_{\d})& (-\alpha \idMatrix_{\d})& \\
		\idMatrix_{\d}& -\mu \idMatrix_{\d}&
	\end{pmatrix} \\
&= 
	\det 
	\bpr{
		(M-\mu \idMatrix_{\d})( -\mu \idMatrix_{\d})- (-\alpha \idMatrix_{\d})\idMatrix_{\d}
	} \\
&=
	\det 
	\bpr{
		(M-\mu \idMatrix_{\d})( -\mu \idMatrix_{\d}) + \alpha \idMatrix_{\d}
	}.
\end{split}
\end{equation}
This and \eqref{convergencerate_momentum:eq1} demonstrate that for all
	$\mu \in \C$
it holds that
\begin{equation}
\label{convergencerate_momentum:eq3.1}
\begin{split}
\det(A-\mu \idMatrix_{2\d}) 
&= 
	\det 
	\begin{pmatrix}
		\bpr{(1-\varrho \lambda_1 + \alpha - \mu)(-\mu) + \alpha} & &0 \\
		& \ddots & \\
		0& & \bpr{(1-\varrho \lambda_\d + \alpha - \mu)(-\mu) + \alpha}
	\end{pmatrix} \\
&= 
	\prod _{i = 1}^\d \bpr{ (1-\varrho \lambda_i + \alpha - \mu)(-\mu) + \alpha} \\
&= 
	\prod _{i = 1}^\d \bpr{ \mu^2 - (1-\varrho \lambda_i + \alpha) \mu + \alpha}.
\end{split}
\end{equation}
Moreover, note that for all $\mu \in \C$, $ i \in \{1, 2, \ldots, \d \}$ it holds that 
\begin{equation}
\begin{split}
	\mu^2 - (1-\varrho \lambda_i + \alpha) \mu + \alpha 
&=
	\mu^2 
	- 2\mu \br*{\tfrac{(1-\varrho \lambda_i + \alpha)}{2} }  
	+ \br*{\tfrac{(1-\varrho \lambda_i + \alpha)}{2} }^2 
	+ \alpha
	- \br*{\tfrac{(1-\varrho \lambda_i + \alpha)}{2} }^2 \\
&=
	\br*{ \mu - \tfrac{(1-\varrho \lambda_i + \alpha)}{2} }^2
	+ \alpha
	- \tfrac{ 1 }{ 4 } \br*{ 1 - \varrho \lambda_i + \alpha }^2 \\
&=
	\br*{ \mu - \tfrac{(1-\varrho \lambda_i + \alpha)}{2} }^2
	-
	\tfrac{1}{4} 
	\bbbr{
	  \bbr{ 1 - \varrho \lambda_i + \alpha }^2 
	  - 4 \alpha 
	} 
	.
\end{split}
\end{equation}
Hence, we obtain that for all $ i \in \{1, 2, \ldots, \d \}$ it holds that
\begin{equation}
\begin{split}
	&\cu*{
		\mu \in \C \colon
		\mu^2 - (1-\varrho \lambda_i + \alpha) \mu + \alpha = 0
	} 
\\&=
	\cu*{
		\mu \in \C \colon 
		\br*{ \mu - \tfrac{(1-\varrho \lambda_i + \alpha)}{2} }^2
		=
	\tfrac{1}{4} 
	\bbbr{
	  \bbr{ 1 - \varrho \lambda_i + \alpha }^2 
	  - 4 \alpha 
	} 
	} \\
&=
	\cu*{
		\tfrac{ \pr*{ 1 - \varrho \lambda_i + \alpha } + \sqrt{\br*{1-\varrho \lambda_i + \alpha}^2 - 4\alpha}}{2},
		\tfrac{ \pr*{ 1 - \varrho \lambda_i + \alpha } - \sqrt{\br*{1-\varrho \lambda_i + \alpha}^2 - 4\alpha}}{2},
	} \\
&=
	\bigcup_{s \in \{-1,1\}}
	\cu*{
          \tfrac{ 1 }{ 2 }
          \br*{
            1 - \varrho \lambda_i + \alpha 
            + s \sqrt{ \pr*{ 1-\varrho \lambda_i + \alpha }^2 - 4\alpha}
          }
	}.
\end{split}
\end{equation}
Combining this, \eqref{convergencerate_momentum:eq1.3}, and \eqref{convergencerate_momentum:eq3.1} demonstrates that
\begin{equation}
\label{convergencerate_momentum:eq4}
\begin{split}
	\mathscr{S} 
&=
	\cu*{
		\mu \in \C \colon
		\det(A-\mu \idMatrix_{2\d})  = 0
	} \\
&=
	\cu*{
		\mu \in \C \colon
		\br*{ \prod _{i = 1}^\d \bpr{ \mu^2 - (1-\varrho \lambda_i + \alpha) \mu + \alpha} = 0 }
	} \\
&=
	\bigcup_{i = 1}^\d 
	\cu*{
		\mu \in \C \colon
		\mu^2 - (1-\varrho \lambda_i + \alpha) \mu + \alpha = 0 
	} \\
&=
	\adjustlimits {\bigcup^\d}_{i = 1} \bigcup_{s \in \{-1,1\}}
	\cu*{
          \tfrac{ 1 }{ 2 }
          \br*{
            1 - \varrho \lambda_i + \alpha 
            + s \sqrt{ \pr*{ 1 - \varrho \lambda_i + \alpha }^2 - 4\alpha}
          }
	}
	.
\end{split}
\end{equation}
Moreover, observe that the fact that for all $i \in \{1, 2, \ldots, \d \}$ it holds that $\lambda_i \geq \kappa$ and (\ref{convergencerate_momentum:eq0}) ensure that for all $i \in \{1, 2, \ldots, \d \}$ it holds that
\begin{equation}
\label{convergencerate_momentum:eq5}
\begin{split}
	1-\varrho \lambda_i + \alpha 
&\leq 
	1-\varrho \kappa+ \alpha  
= 
	1- \br*{\tfrac{4}{(\sqrt{\mathcal{K}} +\sqrt{\kappa} )^2}}\kappa+ \tfrac{(\sqrt{\mathcal{K}} -\sqrt{\kappa})^2 }{(\sqrt{\mathcal{K}}+\sqrt{\kappa})^2} \\
&= 
	\tfrac{(\sqrt{\mathcal{K}} +\sqrt{\kappa})^2 - 4\kappa + (\sqrt{\mathcal{K}} -\sqrt{\kappa})^2 }{(\sqrt{\mathcal{K}} +\sqrt{\kappa})^2 } 
= 
	\tfrac{ \mathcal{K} +2\sqrt{\mathcal{K}}\sqrt{\kappa}+ \kappa  - 4\kappa +  \mathcal{K} -2\sqrt{\mathcal{K}}\sqrt{\kappa}+  \kappa }
	{(\sqrt{\mathcal{K}} +\sqrt{\kappa})^2 } 	\\
&= 
	\tfrac{ 2\mathcal{K} - 2\kappa }{(\sqrt{\mathcal{K}} +\sqrt{\kappa})^2 } 
= 
	\tfrac{ 2(\sqrt{\mathcal{K}} -\sqrt{\kappa})(\sqrt{\mathcal{K}} +\sqrt{\kappa})}{(\sqrt{\mathcal{K}} + \sqrt{\kappa})^2 } 
= 
	2 \br*{\tfrac{ \sqrt{\mathcal{K}} -\sqrt{\kappa}}{\sqrt{\mathcal{K}} + \sqrt{\kappa} } }
\geq 
	0.
\end{split}
\end{equation}
In addition, note that 
	the fact that for all $i \in \{1, 2, \ldots, \d \}$ it holds that $\lambda_i \leq \mathcal{K}$ 
	and \cref{convergencerate_momentum:eq0} 
assure that for all 
	$i \in \{1, 2, \ldots, \d \}$ 
it holds that
\begin{equation}
\begin{split}
	1-\varrho \lambda_i + \alpha  
&\geq	
	1-\varrho \mathcal{K}+ \alpha 
= 
	1- \br*{\tfrac{4}{(\sqrt{\mathcal{K}} +\sqrt{\kappa} )^2}}\mathcal{K}+ \tfrac{(\sqrt{\mathcal{K}} -\sqrt{\kappa})^2 }{(\sqrt{\mathcal{K}}+\sqrt{\kappa})^2} \\
&= 
	\tfrac{(\sqrt{\mathcal{K}} +\sqrt{\kappa})^2 - 4\mathcal{K} + (\sqrt{\mathcal{K}} -\sqrt{\kappa})^2 }{(\sqrt{\mathcal{K}} +\sqrt{\kappa})^2 } 
= 
	\tfrac{ \mathcal{K} +2\sqrt{\mathcal{K}}\sqrt{\kappa}+ \kappa  - 4\mathcal{K} +  \mathcal{K} -2\sqrt{\mathcal{K}}\sqrt{\kappa}+  \kappa }
	{(\sqrt{\mathcal{K}} +\sqrt{\kappa})^2 } \\
&= 
	\tfrac{ -2\mathcal{K} + 2\kappa }{(\sqrt{\mathcal{K}} +\sqrt{\kappa})^2 } 
= 
 	-2 \br*{\tfrac{ \mathcal{K}- \kappa }{(\sqrt{\mathcal{K}} +\sqrt{\kappa})^2 }} 
=  
	-2 \br*{\tfrac{(\sqrt{\mathcal{K}} -\sqrt{\kappa})(\sqrt{\mathcal{K}} +\sqrt{\kappa})}{(\sqrt{\mathcal{K}} + \sqrt{\kappa})^2 }} \\
&= 
	-2 \br*{\tfrac{ \sqrt{\mathcal{K}} -\sqrt{\kappa}}{\sqrt{\mathcal{K}} + \sqrt{\kappa} }} 
\leq 
	0.
\end{split}
\end{equation}
Combining 
	this,
	\cref{convergencerate_momentum:eq5},
	and \cref{convergencerate_momentum:eq0} 
implies that for all 
	$i \in \{1, 2, \ldots, \d \}$ 
it holds that
\begin{equation}
\label{convergencerate_momentum:eq6}
	(1-\varrho \lambda_i + \alpha)^2 
\leq 
	\br*{2\pr*{ \tfrac{ \sqrt{\mathcal{K}} -\sqrt{\kappa}}{\sqrt{\mathcal{K}} + \sqrt{\kappa} }}}^{2}  
= 
	4 \br*{\tfrac{ \sqrt{\mathcal{K}} -\sqrt{\kappa}}{\sqrt{\mathcal{K}} + \sqrt{\kappa} }}^{2} 
= 
	4 \alpha.
\end{equation}
This and \eqref{convergencerate_momentum:eq4} demonstrate that
\begin{equation}
\begin{split}
	\max_{\mu \in \mathscr{S} \cup \{0\}} \abs{\mu}
&=
	\max_{\mu \in \mathscr{S}} \abs{\mu} \\
&= 
	\max_{i \in \{1,2,\ldots,d\}} \max_{ s \in \{ -1, 1 \}}
	\abs*{
	  \frac{1}{2}  \br*{ 1 - \varrho \lambda_i + \alpha + s \sqrt{ \pr*{ 1 - \varrho \lambda_i + \alpha }^2 - 4 \alpha}}
	}   
	\\
&= 
	\frac{1}{2}
	\br*{
		\max_{i \in \{1,2,\ldots,d\}} \max_{ s \in \{ -1, 1 \}}		
        \abs*{ 
          \br*{1-\varrho \lambda_i + \alpha + s \sqrt{(-1)(4 \alpha - [1-\varrho \lambda_i + \alpha]^2)}}
        }
	}  
	\\
&= 
	\frac{1}{2}
	\br*{
		\max_{i \in \{1,2,\ldots,d\}} \max_{ s \in \{ -1, 1 \} }
		\abs*{ 
		  \br*{
		    1-\varrho \lambda_i + \alpha + 
		    s \mathbf{i}\sqrt{4 \alpha - ( 1 - \varrho \lambda_i + \alpha )^2 } 
		  }
		}^2 
	}^{\nicefrac{1}{2}}.
\end{split}
\end{equation}
Combining this with \eqref{convergencerate_momentum:eq6} proves that
\begin{equation}
\begin{split}
	\max_{\mu \in \mathscr{S} \cup \{0\}} \abs{\mu} 
&= 
	\tfrac{1}{2}
	\br*{
		\max_{i \in \{1,2,\ldots,d\}} \max_{ s \in \{ -1, 1 \} }
		\pr*{ 
			\babs{1-\varrho \lambda_i + \alpha }^2 
			+ \babs{s\sqrt{4 \alpha - ( 1-\varrho \lambda_i + \alpha )^2}}^2 
		}
	}^{\nicefrac{1}{2}} \\
&= 
	\tfrac{1}{2}
	\br*{
		\max_{i \in \{1,2,\ldots,d\}} \max_{ s \in \{ -1, 1 \} }
		\pr*{ 
			\pr*{ 1 - \varrho \lambda_i + \alpha }^2 
			+ 4 \alpha - ( 1-\varrho \lambda_i + \alpha )^2
		}
	}^{\nicefrac{1}{2}} \\
&= 
	\tfrac{1}{2} \br*{ 4 \alpha}^{\nicefrac{1}{2}}
=
	\sqrt{\alpha}.
\end{split}
\end{equation}
Combining \eqref{convergencerate_momentum:eq2} 
and \eqref{convergencerate_momentum:eq3} hence ensures that for all $n \in \N_0$ it holds that
\begin{equation}
\begin{split}
	\Norm{\Theta_n-\vartheta}
&\leq 
	\sqrt{2} \, \norm{\xi-\vartheta} \rdnorm[(2\d)\times(2\d)]{A^n} \\
&\leq 
	\sqrt{2} \, \norm{\xi-\vartheta} 
	\br*{\varepsilon  + \max_{\mu \in \mathscr{S} \cup \{0\}} \abs{\mu}}^{n} \\
	&\quad \cdot
	\br*{ \max  
		\pr*{ 
			\cu*{ \tfrac{\rdnorm[(2\d)\times(2\d)]{A^k}}{(\varepsilon + \max_{\mu \in \mathscr{S}\cup \{0\}} \abs{\mu} )^{k}} \in \R \colon k \in \N_0 \cap [0,m) }\cup \{1\} 
		}
	} \\
&= 
	\sqrt{2} \, \norm{\xi-\vartheta}
	\br*{\varepsilon  + \alpha^{\nicefrac{1}{2}}}^{n}
	\br*{ \max 
		 \pr*{ 
		 	\cu*{ \tfrac{\rdnorm[(2\d)\times(2\d)]{A^k}}{(\varepsilon + \alpha^{1/2} )^{k}} \in \R \colon k \in \N_0 \cap [0,m) }\cup \{1\}
		 }
	} \\
&= 
	\sqrt{2} \, \norm{\xi-\vartheta}
	\br*{\varepsilon + \tfrac{\sqrt{\mathcal{K}} -\sqrt{\kappa}}{\sqrt{\mathcal{K}} + \sqrt{\kappa}} }^{n} 
	\br*{ \max 
		 \pr*{ 
		 	\cu*{ \tfrac{\rdnorm[(2\d)\times(2\d)]{A^k}}{(\varepsilon + \alpha^{1/2} )^{k}} \in \R \colon k \in \N_0 \cap [0,m) }\cup \{1\}
		 }
	}.
\end{split}
\end{equation}
This establishes \cref{convergencerate_momentum:item2}.
The proof of \cref{convergencerate_momentum} it thus completed.
\end{proof}
\endgroup

\subsubsection
{Comparison 
of the convergence speeds of GD and momentum optimization}
\label{subsec:comparison_momentum}

In this subsection we provide in 
\cref{comparison_plain_vs_momentum} below 
a comparison between the convergence speeds 
of the plain-vanilla \GD\ optimization method 
and the momentum \GD\ optimization method. 
Our proof of \cref{comparison_plain_vs_momentum} 
employs the auxiliary and elementary estimate 
in \cref{comparison_rates} below, 
the refined error analysis for the plain-vanilla \GD\ optimization method
in \cref{subsec:error_analysis_momentum_GD_without} above 
(see \cref{GD_convergence_quadratic} 
and  
\cref{lowerbound_convergencerate} 
in \cref{subsec:error_analysis_momentum_GD_without}),
as well as the error analysis for the momentum \GD\ optimization 
method in \cref{subsec:error_analysis_momentum_GD} above 
(see \cref{convergencerate_momentum} 
in \cref{subsec:error_analysis_momentum_GD}).

\begingroup
\providecommand{\d}{}
\renewcommand{\d}{\defaultParamDim}
\providecommand{\f}{}
\renewcommand{\f}{\defaultLossFunction}
\providecommand{\g}{}
\renewcommand{\g}{\defaultGradientFunction}
\begin{lemma}[Comparison of the convergence rates 
of the \GD\ optimization method and 
the momentum \GD\ optimization method]
\label{comparison_rates}
Let $\mathcal{K}, \kappa \in (0,\infty)$ satisfy  $\kappa < \mathcal{K}$. Then
\begin{equation}
\frac{\sqrt{\mathcal{K}} -\sqrt{\kappa}}{\sqrt{\mathcal{K}} + \sqrt{\kappa}}  < \frac{\mathcal{K} -{\kappa}}{{\mathcal{K}} + {\kappa}}.
\end{equation}
\end{lemma}

\begin{proof}[Proof of \cref{comparison_rates}]
Note that the fact that ${\mathcal{K}} -{\kappa} > 0 < 2\sqrt{\mathcal{K}}\sqrt{\kappa}$ ensures that
\begin{equation}
\frac{\sqrt{\mathcal{K}} -\sqrt{\kappa}}{\sqrt{\mathcal{K}} + \sqrt{\kappa}} 
= \frac{(\sqrt{\mathcal{K}} -\sqrt{\kappa})(\sqrt{\mathcal{K}} + \sqrt{\kappa})}{(\sqrt{\mathcal{K}} + \sqrt{\kappa})^2}
= \frac{{\mathcal{K}} -{\kappa}}{{\mathcal{K}} +2\sqrt{\mathcal{K}}\sqrt{\kappa} + {\kappa}} 
< \frac{{\mathcal{K}} -{\kappa}}{{\mathcal{K}} +{\kappa}}.
\end{equation}
The proof of \cref{comparison_rates} it thus completed.
\end{proof}
\endgroup

\cfclear
\begingroup
\newcommand{\lrnorm}[1]{\apnorm2{#1}}
\newcommand{\norm}[1]{\pnorm2{#1}}
\providecommand{\d}{}
\renewcommand{\d}{\defaultParamDim}
\providecommand{\f}{}
\renewcommand{\f}{\defaultLossFunction}
\providecommand{\g}{}
\renewcommand{\g}{\defaultGradientFunction}
\begin{cor}[Convergence speed comparisons between the \GD\ optimization method and the momentum \GD\ optimization method]
\label{comparison_plain_vs_momentum}
Let $\d \in \N$, 
$\kappa,\cK,\lambda_1,\lambda_2,\ldots, \lambda_\d\in (0,\infty)$, 
$
  \xi = (\xi_1,\dots,\xi_\d)
$, 
$
  \vartheta = (\vartheta_1,\dots,\vartheta_\d) \in \R^\d
$ satisfy 
$\kappa= \min\{\lambda_1,\lambda_2,\ldots, \lambda_\d \} < \max\{\lambda_1,\lambda_2,\ldots, \lambda_\d \} = \mathcal{K}$,
let $\f \colon \R^\d \to \R$ satisfy 
for all 
$
  \theta = ( \theta_1, \dots, \theta_\d ) \in \R^\d 
$ 
that 
\begin{equation}
\label{comparison_plain_vs_momentum:assumption0}
\f(\theta) 
=
\tfrac{1}{2} \br*{\sum_{i = 1}^\d\lambda_i \abs{\theta_i - \vartheta_i}^2 },
\end{equation}
for every $ \gamma \in (0,\infty) $
let 
$
  \Theta^\gamma \colon \N_0 \to \R^\d 
$ 
satisfy for all $n \in \N$ that
\begin{equation}
\label{eq:def_classicalGD_comparison}
\Theta^\gamma_0  = \xi \qandq \Theta^\gamma_n = \Theta^\gamma_{n-1} -  \gamma (\nabla \f)(\Theta^\gamma_{n-1}),
\end{equation}
and let $\mathcal{M} \colon \N_0 \cup \{-1\} \to \R^\d$ satisfy for all $n \in \N$ that $\mathcal{M}_{-1} = \mathcal{M}_0  = \xi$ and
\begin{equation}
\label{eq:def_momentum_comparison}
	\mathcal{M}_n 
= 
	{\mathcal{M}}_{n-1} 
	- \tfrac{4}{(\sqrt{\mathcal{K}} +\sqrt{\kappa} )^2}  (\nabla \f)({\mathcal{M}}_{n-1}) 
	+ \br*{\tfrac{\sqrt{\mathcal{K}} -\sqrt{\kappa} }{\sqrt{\mathcal{K}}+\sqrt{\kappa}}}^2 ({\mathcal{M}}_{n-1} - {\mathcal{M}}_{n-2}).
\end{equation}
Then 
\begin{enumerate}[label=(\roman *)]
\item \label{comparison_plain_vs_momentum:item1}
there exist $\gamma, \mathfrak{c} \in (0,\infty)$ such that for all $n \in \N_0$ it holds that
\begin{equation}
	\norm{\Theta^\gamma_n-\vartheta} 
\leq 
	\mathfrak{c} \br*{\tfrac{\mathcal{K} - \kappa}{\mathcal{K} + \kappa}}^n ,
\end{equation}

\item \label{comparison_plain_vs_momentum:item2}
it holds for all $\gamma \in (0,\infty), n \in \N_0$ that
\begin{equation}
	\norm{\Theta^\gamma_n-\vartheta} 
\geq 
	\bbr{ \min \{\abs{\xi_1-\vartheta_1}, \ldots, \abs{\xi_\d-\vartheta_\d} \}} 
	\br*{\tfrac{\mathcal{K} - \kappa}{\mathcal{K} + \kappa}}^n ,
\end{equation}

\item \label{comparison_plain_vs_momentum:item3}
for every $\varepsilon \in (0,\infty)$ there exists $\mathfrak{c} \in (0,\infty)$ such that for all $n \in \N_0$ it holds that
\begin{equation}
	\norm{\mathcal{M}_n - \vartheta} 
\leq 
	\mathfrak{c} \br*{\tfrac{\sqrt{\mathcal{K}} - \sqrt{\kappa} }{\sqrt{\mathcal{K}}+\sqrt{\kappa}} + \varepsilon }^n ,
\end{equation}
and 

\item \label{comparison_plain_vs_momentum:item4}
it holds that 
$
\tfrac{\sqrt{\mathcal{K}} -\sqrt{\kappa}}{\sqrt{\mathcal{K}} + \sqrt{\kappa}}  < \tfrac{\mathcal{K} -{\kappa}}{{\mathcal{K}} + {\kappa}}
$
\end{enumerate}
\cfout.
\end{cor}

\begin{proof}[Proof of \cref{comparison_plain_vs_momentum}]
First, note that \cref{GD_convergence_quadratic} proves \cref{comparison_plain_vs_momentum:item1}.
Next observe that \cref{lowerbound_convergencerate} establishes \cref{comparison_plain_vs_momentum:item2}.
In addition, note that \cref{convergencerate_momentum} proves \cref{comparison_plain_vs_momentum:item3}.
Finally, observe that \cref{comparison_rates} establishes \cref{comparison_plain_vs_momentum:item4}.
The proof of \cref{comparison_plain_vs_momentum} is thus complete.
\end{proof}
\endgroup

\cref{comparison_plain_vs_momentum} above, roughly speaking, 
shows in the case of the considered class of quadratic objective functions 
that the momentum \GD\ optimization method 
in 
\eqref{eq:def_momentum_comparison} 
outperforms 
the classical plain-vanilla \GD\ optimization method 
(and, in particular, 
the classical plain-vanilla \GD\ optimization method 
in 
\eqref{GD_convergence_quadratic:assumption1}
in 
\cref{GD_convergence_quadratic}
above)
provided that the parameters 
$ \lambda_1, \lambda_2, \dots, \lambda_\defaultParamDim \in (0,\infty) $
in the objective function 
in \eqref{comparison_plain_vs_momentum:assumption0}
satisfy the assumption that 
\begin{equation}
\begin{split} 
  \min\{ \lambda_1, \dots, \lambda_\defaultParamDim \} 
  < 
  \max\{ \lambda_1, \dots, \lambda_\defaultParamDim \}.
\end{split}
\end{equation}
The next elementary result, 
\cref{example_GD_equal_momentum} below, 
demonstrates that the momentum \GD\ optimization method 
in \eqref{eq:def_momentum_comparison} 
and the plain-vanilla \GD\ optimization method 
in 
\eqref{GD_convergence_quadratic:assumption1}
in 
\cref{GD_convergence_quadratic}
above
coincide in the case where 
$
  \min\{ \lambda_1, \dots, \lambda_\defaultParamDim \} 
  =
  \max\{ \lambda_1, \dots, \lambda_\defaultParamDim \}
$.

\begingroup
\newcommand{\lrnorm}[1]{\apnorm2{#1}}
\newcommand{\norm}[1]{\pnorm2{#1}}
\providecommand{\d}{}
\renewcommand{\d}{\defaultParamDim}
\providecommand{\f}{}
\renewcommand{\f}{\defaultLossFunction}
\providecommand{\g}{}
\renewcommand{\g}{\defaultGradientFunction}
\cfclear
\begin{lemma}[Concurrence of the \GD\ optimization method and the momentum \GD\ optimization method]
\label{example_GD_equal_momentum}
Let $\d \in \N$, $\xi, \vartheta \in \R^\d$, $\alpha \in (0,\infty)$,
let $\f \colon \R^\d \to \R$ satisfy for all $\theta \in \R^\d$ that 
\begin{equation}
\label{example_GD_equal_momentum:ass1}
\f(\theta) = \tfrac{\alpha}{2} \norm{\theta-\vartheta}^2,
\end{equation}
let $\Theta \colon \N_0 \to \R^\d$ satisfy for all $n \in \N$ that
\begin{equation}
\label{example_GD_equal_momentum:ass2}
	\Theta_0  
= 
	\xi 
\qandq 
	\Theta_n 
= 
	\Theta_{n-1} -  \tfrac{ 2 }{ ( \alpha + \alpha ) }  (\nabla \f)(\Theta_{n-1}),
\end{equation}
and let $\mathcal{M} \colon \N_0 \cup \{-1 \}\to \R^\d$ satisfy for all $n \in \N$ that 
$ \mathcal{M}_{ - 1 } = \mathcal{M}_0 = \xi $ and
\begin{equation}
\label{example_GD_equal_momentum:ass3}
\mathcal{M}_n 
= 
	{\mathcal{M}}_{n-1} 
	- \tfrac{4}{(\sqrt{\alpha} +\sqrt{\alpha} )^2}  (\nabla \f)({\mathcal{M}}_{n-1}) 
	+ \br*{\tfrac{\sqrt{\alpha} -\sqrt{\alpha} }{\sqrt{\alpha}+\sqrt{\alpha}}}^2 ({\mathcal{M}}_{n-1} - {\mathcal{M}}_{n-2})
\end{equation}
\cfload.
Then
\begin{enumerate}[label=(\roman *)]
\item \label{example_GD_equal_momentum:item1}
it holds that 
$
\mathcal{M}|_{\N_0} \colon \N_0 \to \R^\d 
$ is the momentum \GD\ process for the objective function $\f$ with learning rates 
$\N \ni n \mapsto \nicefrac{1}{\alpha} \in [0,\infty)$, 
momentum decay factors 
$
  \N \ni n \mapsto 0 \in [0,1] 
$,
and initial value $\xi$,

\item \label{example_GD_equal_momentum:item2}
it holds for all $ n \in \N_0 $ that $\mathcal{M}_n = \Theta_n $, and

\item \label{example_GD_equal_momentum:item3}
it holds for all $n \in \N$ that
$
\Theta_n = \vartheta = \mathcal{M}_n
$
\end{enumerate}
\cfadd{def:determ_momentum}
\cfout.
\end{lemma}

\begin{proof}[Proof of \cref{example_GD_equal_momentum}]
First, note that \eqref{example_GD_equal_momentum:ass3} implies that for all $n \in \N$ it holds that
\begin{equation}
\label{example_GD_equal_momentum:eq1}
	\mathcal{M}_n 
= 
	{\mathcal{M}}_{n-1} 
	- \tfrac{4}{(2\sqrt{\alpha})^2} (\nabla \f)({\mathcal{M}}_{n-1}) 
=
	{\mathcal{M}}_{n-1} 
	- \tfrac{1}{\alpha} (\nabla \f)({\mathcal{M}}_{n-1}) .
\end{equation}
Combining this with the assumption that $ \mathcal{M}_0 = \xi$  establishes \cref{example_GD_equal_momentum:item1}.
Next note that \eqref{example_GD_equal_momentum:ass2} ensures that for all $n \in \N$ it holds that
\begin{equation}
\label{example_GD_equal_momentum:eq2}
	\Theta_n 
= 
	\Theta_{n-1} -  \tfrac{1}{\alpha} (\nabla \f)(\Theta_{n-1}).
\end{equation}
Combining this with \eqref{example_GD_equal_momentum:eq1} and the assumption that $\Theta_0 =  \xi = \mathcal{M}_0$ proves \cref{example_GD_equal_momentum:item2}.
Furthermore, observe that \cref{der_of_norm} assures that for all $\theta \in \R^\d$ it holds that
\begin{equation}
\label{example_GD_equal_momentum:eq3}
	(\nabla \f)(\theta) 
= 
	\tfrac{\alpha}{2}(2(\theta-\vartheta))
=
	\alpha(\theta-\vartheta).
\end{equation}
Next we claim that for all $n \in \N$ it holds that
\begin{equation}
\label{example_GD_equal_momentum:eq4}
\Theta_n = \vartheta .
\end{equation}
We now prove \eqref{example_GD_equal_momentum:eq4} by induction on $n \in \N$.
For the base case $n = 1$ note that \eqref{example_GD_equal_momentum:eq2} and  \eqref{example_GD_equal_momentum:eq3} imply that 
\begin{equation}
	\Theta_1 
= 
	\Theta_{0} -  \tfrac{1}{\alpha} (\nabla \f)(\Theta_{0})
=
	\xi -  \tfrac{1}{\alpha} (\alpha(\xi-\vartheta)) 
=
	\xi - (\xi - \vartheta)
=
	\vartheta.
\end{equation}
This establishes \eqref{example_GD_equal_momentum:eq4} in the base case $n = 1$. 
For the induction step observe that \eqref{example_GD_equal_momentum:eq2} and \eqref{example_GD_equal_momentum:eq3} assure that for all $n \in \N$ with $\Theta_n = \vartheta$ it holds that
\begin{equation}
	\Theta_{n+1} 
= 
	\Theta_{n} -  \tfrac{1}{\alpha} (\nabla \f)(\Theta_{n})
=
	\vartheta -  \tfrac{1}{\alpha} (\alpha(\vartheta-\vartheta)) 
=
	\vartheta.
\end{equation}
Induction thus proves \eqref{example_GD_equal_momentum:eq4}.
Combining \eqref{example_GD_equal_momentum:eq4} and \cref{example_GD_equal_momentum:item2} establishes \cref{example_GD_equal_momentum:item3}.
The proof of \cref{example_GD_equal_momentum} is thus complete.
\end{proof}
\endgroup

\subsection{Numerical comparisons for GD and momentum optimization}

In this subsection we provide in 
\cref{illu:momentum}, 
\cref{code:comparison_GD_momentum}, 
and \cref{fig:comparison_GD_momentum}
 a numerical comparison 
of the plain-vanilla \GD\ optimization method 
and the momentum \GD\ optimization method 
in the case of the specific quadratic 
optimization problem in 
\cref{eq:illu1}--\cref{eq:illu2} below.

\begingroup
\providecommand{\d}{}
\renewcommand{\d}{\defaultParamDim}
\providecommand{\f}{}
\renewcommand{\f}{\defaultLossFunction}
\providecommand{\g}{}
\renewcommand{\g}{\defaultGradientFunction}
\begin{athm}{example}{illu:momentum}
Let $\mathcal{K} = 10$, 
$\kappa = 1$, $\vartheta = (\vartheta_1, \vartheta_2) \in \R^2$,  $\xi = (\xi_1, \xi_2) \in \R^2$ satisfy 
\begin{equation}
\label{eq:illu1}
\vartheta = 
\begin{pmatrix}
	\vartheta_1 \\
	\vartheta_2
\end{pmatrix}
=
\begin{pmatrix}
	1 \\
	1
\end{pmatrix}
\qandq
\xi =
\begin{pmatrix}
	\xi_1 \\
	\xi_2
\end{pmatrix}
=
\begin{pmatrix}
	5 \\
	3
\end{pmatrix},
\end{equation}
let $\f \colon \R^2 \to \R$ satisfy for all $\theta = (\theta_1,\theta_2) \in \R^2$ that
\begin{equation}
\label{eq:illu2}
	\f(\theta) 
= 
	\pr*{\tfrac{\kappa}{2}} \abs{\theta_1-\vartheta_1}^2 + \pr*{\tfrac{\mathcal{K}}{2}} \abs{\theta_2-\vartheta_2}^2,
\end{equation}
let $\Theta \colon \N_0 \to \R^\d$ satisfy for all $n \in \N$ that $	\Theta_0  =\xi$ and 
\begin{equation}
\begin{split}
	\Theta_n 
&= 
	\Theta_{n-1} -  \tfrac{2}{\mathcal{K} + \kappa} (\nabla \f)(\Theta_{n-1})
=
	\Theta_{n-1} -  \tfrac{2}{11} (\nabla \f)(\Theta_{n-1}) \\
&=
	\Theta_{n-1} -  0.\overline{18}  (\nabla \f)(\Theta_{n-1}) 
\approx
	\Theta_{n-1} -  0.18   (\nabla \f)(\Theta_{n-1}),
\end{split}
\end{equation}
and let $\mathcal{M} \colon \N_0 \to \R^\d$ and $ \mathbf{m} \colon \N_0 \to \R^\d$ satisfy for all $n \in \N$ that
$
\mathcal{M}_0 = \xi
$,
$\mathbf{m}_0 = 0$,
$\mathcal{M}_n = \mathcal{M}_{n-1} - 0.3 \, \mathbf{m}_n$, and 
\begin{equation}
\begin{split}
	\mathbf{m}_n 
&= 
	0.5 \, \mathbf{m}_{n-1} + (1-0.5) (\nabla \f)(\mathcal{M}_{n-1}) \\
&=
	0.5 \, (\mathbf{m}_{n-1} + (\nabla \f)(\mathcal{M}_{n-1})).
\end{split}
\end{equation}
Then 
\begin{enumerate}[label=(\roman *)]
\item
it holds for all $\theta = (\theta_1, \theta_2) \in \R^2$ that
\begin{equation}
	(\nabla \f)(\theta) 
=
	\begin{pmatrix}
		\kappa (\theta_1-\vartheta_1) \\
		\mathcal{K} (\theta_2-\vartheta_2)
	\end{pmatrix}
=
	\begin{pmatrix}
		 \theta_1-1 \\
		10 \, (\theta_2-1)
	\end{pmatrix},
\end{equation}

\item
it holds that 
	
	\begin{equation}
	\Theta_0 
	=  
		\begin{pmatrix}
			5 \\
			3
		\end{pmatrix},
	\end{equation}

	\begin{equation}
	\begin{split}
		\Theta_1 
	&=
		\Theta_{0} -  \tfrac{2}{11} (\nabla \f)(\Theta_{0})
	\approx
		\Theta_{0} -  0.18 (\nabla \f)(\Theta_{0})  \\
	&=
		\begin{pmatrix}
			5 \\
			3
		\end{pmatrix}
		-
		 0.18 \,
		\begin{pmatrix}
			5-1 \\
			10(3-1)
		\end{pmatrix} 
	=
		\begin{pmatrix}
			5 -0.18 \cdot 4\\
			3 - 0.18 \cdot 10 \cdot 2
		\end{pmatrix} \\
	&=
		\begin{pmatrix}
			5 -0.72\\
			3 - 3.6
		\end{pmatrix}
	=
		\begin{pmatrix}
			4.28 \\
			-0.6
		\end{pmatrix},
	\end{split}
	\end{equation}

	\begin{equation}
	\begin{split}
		\Theta_2 
	&\approx
		\Theta_{1} - 0.18 (\nabla \f)(\Theta_{1}) 
	=
		\begin{pmatrix}
			4.28 \\
			-0.6
		\end{pmatrix}
		-
		 0.18 \,
		\begin{pmatrix}
			4.28-1 \\
			10(-0.6-1)
		\end{pmatrix} \\
	&=
		\begin{pmatrix}
			4.28 - 0.18 \cdot 3.28\\
			-0.6 - 0.18 \cdot 10 \cdot (-1.6)
		\end{pmatrix}
	=
		\begin{pmatrix}
			4.10 -  0.18\cdot2  - 0.18 \cdot 0.28\\
			-0.6 + 1.8  \cdot 1.6
		\end{pmatrix} \\
	&=
		\begin{pmatrix}
			4.10 -  0.36  - 2 \cdot 9 \cdot 4 \cdot  7 \cdot 10^{-4}\\
			-0.6 + 1.6  \cdot 1.6 + 0.2  \cdot 1.6
		\end{pmatrix}
	=
		\begin{pmatrix}
			3.74 -  9 \cdot 56 \cdot 10^{-4}\\
			-0.6 + 2.56 + 0.32
		\end{pmatrix} \\
	&=
		\begin{pmatrix}
			3.74 -  504 \cdot 10^{-4}\\
			2.88 - 0.6 
		\end{pmatrix} 
	=
		\begin{pmatrix}
			3.6896\\
			2.28
		\end{pmatrix}
	\approx
		\begin{pmatrix}
			3.69 \\
			2.28
		\end{pmatrix},
	\end{split}
	\end{equation}

	\begin{equation}
	\begin{split}
		\Theta_3
	&\approx
		\Theta_{2} - 0.18 (\nabla \f)(\Theta_{2}) 
	\approx
		\begin{pmatrix}
			3.69 \\
			2.28
		\end{pmatrix}
		-
		0.18 \,
		\begin{pmatrix}
			3.69-1 \\
			10(2.28-1)
		\end{pmatrix} \\
	&=
		\begin{pmatrix}
			3.69 - 0.18 \cdot 2.69\\
			2.28 -  0.18 \cdot 10 \cdot 1.28
		\end{pmatrix}
	=
		\begin{pmatrix}
			3.69 - 0.2 \cdot 2.69 + 0.02 \cdot 2.69\\
			2.28 -  1.8  \cdot 1.28
		\end{pmatrix} \\
	&=
		\begin{pmatrix}
			3.69 - 0.538 + 0.0538\\
			2.28 -   1.28 - 0.8  \cdot 1.28
		\end{pmatrix}
	=
		\begin{pmatrix}
			3.7438 - 0.538 \\
			1-1.28 + 0.2 \cdot 1.28
		\end{pmatrix} \\
	&=
		\begin{pmatrix}
			3.2058 \\
			0.256  - 0.280
		\end{pmatrix} 
	=
		\begin{pmatrix}
			3.2058 \\
			-0.024
		\end{pmatrix}
	\approx
		\begin{pmatrix}
			3.21 \\
			-0.02
		\end{pmatrix},
	\end{split}
	\end{equation}
$$ \vdots  $$

and
\item
it holds that 
	
	\begin{equation}
		\mathcal{M}_0 
	=  
		\begin{pmatrix}
			5 \\
			3
		\end{pmatrix},
	\end{equation}

	\begin{equation}
	\begin{split}
		\mathbf{m}_1 
	&= 
		0.5 \, (\mathbf{m}_{0} + (\nabla \f)(\mathcal{M}_{0}))
	=
		0.5 \, 
		\pr*{
		\begin{pmatrix}
			0 \\
			0
		\end{pmatrix}
		+
		\begin{pmatrix}
			5-1 \\
			10(3-1)
		\end{pmatrix}
		} \\
	&=
		\begin{pmatrix}
			0.5 \, (0 + 4)\\
			0.5 \, (0 + 10 \cdot 2)
		\end{pmatrix}
	=
		\begin{pmatrix}
			2 \\
			10
		\end{pmatrix},
	\end{split}
	\end{equation}

	\begin{equation}
		\mathcal{M}_1 
	=
		\mathcal{M}_{0} - 0.3  \, \mathbf{m}_1
	 =
		\begin{pmatrix}
			5 \\
			3
		\end{pmatrix}
		- 0.3 \, 
		\begin{pmatrix}
			2 \\
			10
		\end{pmatrix}
	=
		\begin{pmatrix}
			4.4 \\
			0
		\end{pmatrix},
	\end{equation}

	\begin{equation}
	\begin{split}
		\mathbf{m}_2 
	&= 
		0.5 \, (\mathbf{m}_{1} + (\nabla \f)(\mathcal{M}_{1}))
	=
		0.5 \, 
		\pr*{
		\begin{pmatrix}
			2 \\
			10
		\end{pmatrix}
		+
		\begin{pmatrix}
			4.4-1 \\
			10(0-1)
		\end{pmatrix}
		}\\
	&=
		\begin{pmatrix}
			0.5 \, (2+3.4) \\
			0.5 \, (10 - 10)
		\end{pmatrix}
	=
		\begin{pmatrix}
			2.7 \\
			0
		\end{pmatrix},
	\end{split}
	\end{equation}

	\begin{equation}
		\mathcal{M}_2 
	=
		\mathcal{M}_{1} - 0.3  \, \mathbf{m}_2
	 =
		\begin{pmatrix}
			4.4 \\
			0
		\end{pmatrix}
		- 0.3 \, 
		\begin{pmatrix}
			2.7 \\
			0
		\end{pmatrix}
	=
		\begin{pmatrix}
			4.4 - 0.81 \\
			0
		\end{pmatrix}
	=
		\begin{pmatrix}
			3.59 \\
			0
		\end{pmatrix},
	\end{equation}

	\begin{equation}
	\begin{split}
		\mathbf{m}_3 
	&= 
		0.5 \, (\mathbf{m}_{2} + (\nabla \f)(\mathcal{M}_{2}))
	=
		0.5 \, 
		\pr*{
		\begin{pmatrix}
			2.7 \\
			0
		\end{pmatrix}
		+
		\begin{pmatrix}
			3.59-1 \\
			10(0-1)
		\end{pmatrix}
		} \\
	&=
		\begin{pmatrix}
			0.5 \, (2.7+2.59) \\
			0.5 \, (0 - 10)
		\end{pmatrix}
	=
		\begin{pmatrix}
			0.5 \cdot 5.29 \\
			0.5 (-10)
		\end{pmatrix} \\
	&=
		\begin{pmatrix}
			2.5 + 0.145 \\
			-5
		\end{pmatrix}
	=
		\begin{pmatrix}
			2.645 \\
			-5
		\end{pmatrix}
	\approx
		\begin{pmatrix}
			2.65 \\
			-5
		\end{pmatrix},
	\end{split}
	\end{equation}

	\begin{equation}
	\begin{split}
		\mathcal{M}_3
	&=
		\mathcal{M}_{2} - 0.3  \, \mathbf{m}_3
	 \approx
		\begin{pmatrix}
			3.59 \\
			0
		\end{pmatrix}
		- 0.3 \, 
		\begin{pmatrix}
			2.65 \\
			-5
		\end{pmatrix}\\
	&=
		\begin{pmatrix}
			3.59 - 0.795 \\
			1.5
		\end{pmatrix} 
	=
		\begin{pmatrix}
			3 - 0.205 \\
			1.5
		\end{pmatrix}
	=
		\begin{pmatrix}
			 2.795 \\
			1.5
		\end{pmatrix}
	\approx
		\begin{pmatrix}
			 2.8 \\
			1.5
		\end{pmatrix},
	\end{split}
	\end{equation}
$$ \vdots $$
	\vspace{-40pt}	
\flushright .
\end{enumerate}
\end{athm}
\endgroup

\filelisting{code:comparison_GD_momentum}{code/example_GD_momentum_plots.py}{{\sc Python} code for \cref{fig:comparison_GD_momentum}}

	\begin{center}
		\IfFileExists{plots/GD_momentum_plots.pdf}{\includegraphics[width=0.8\linewidth]{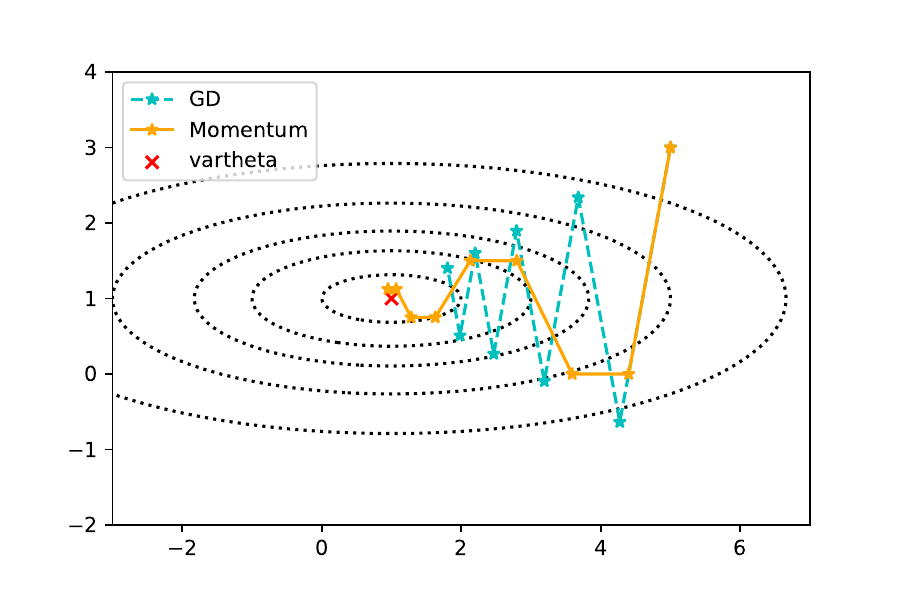}}{}
  \captionof{figure}{\label{fig:comparison_GD_momentum}Result of a call of {\sc Python} code \ref{code:comparison_GD_momentum} }
	\end{center}

\cfclear
\begingroup
\providecommand{\d}{}
\renewcommand{\d}{\defaultParamDim}
\providecommand{\f}{}
\renewcommand{\f}{\defaultLossFunction}
\providecommand{\g}{}
\renewcommand{\g}{\defaultGradientFunction}
\begin{exercise}{ex:momentum_example}
Let
	$(\gamma_n)_{n \in \N} \subseteq [0,\infty)$, 
	$(\alpha_n)_{n \in \N} \subseteq [0,1]$
satisfy for all $n \in \N$ that
$\gamma_n = \frac{1}{n}$ and $\alpha_n = \frac{1}{2}$,
let 
	$\f \colon \R \to \R$ 
satisfy for all 
	$\theta \in \R$
that
	$\f(\theta) = \theta^2$,
and let	
	$\Theta$ 
be the momentum \GD\ process 
for the objective function $\f$ with
learning rates $(\gamma_n)_{n \in \N}$, 
momentum decay factors $(\alpha_n)_{n \in \N}$, 
and initial value $1$\cfadd{def:determ_momentum}
\cfload.
Specify 
	$\Theta_1$, 
	$\Theta_2$, 
	$\Theta_3$, and
	$\Theta_4$
explicitly and prove that your results are correct!
\end{exercise}
\endgroup

\section{Nesterov accelerated momentum optimization}
\label{sect:determ_nesterov}

\todoc{Add statement and result on stability of SGD type methods. See notes: DDA6203 6th Version 11-29. Page 9}

\todoc{Add explanation for why we chose the Nesterov momentum term that way.}

In this section we review the Nesterov accelerated \GD\ optimization method, which was first introduced in Nesterov~\cite{Nesterov83} (cf., \eg, Sutskever et al.~\cite{sutskever2013importance}).
The Nesterov accelerated \GD\ optimization method can be viewed as building on the momentum \GD\ optimization method (see \cref{def:determ_momentum}) by attempting to provide some kind of foresight to the scheme. 
A similar perspective is to see the Nesterov accelerated \GD\ optimization method as a combination of the momentum \GD\ optimization method (see \cref{def:determ_momentum}) and the explicit midpoint \GD\ optimization method (see \cref{def:midpointGD}).

\defdetermNesterov
\algDescrDetermNesterov

\subsection{Alternative definitions}
\label{sect:nesterov_alternatives}

In analogy to the alternative definitions of the momentum \GD\ optimization method in \cref{sec:momentum_GD_alternatives} above, we now provide alternative definitions of the Nesterov accelerated \GD\ optimization method.
Relationships between the considered versions of the Nesterov accelerated \GD\ optimization method are discussed in \cref{sec:relationships_Nesterov} below.

\defdetermNesterovTwo
\algDescrDetermNesterovTwo

\defdetermNesterovThree
\algDescrDetermNesterovThree

\defdetermNesterovFour
\algDescrDetermNesterovFour

\subsection{Relationships between different definitions}
\label{sec:relationships_Nesterov}

In this section we discuss relationships between the different versions of the Nesterov accelerated \GD\ optimization method
introduced in \cref{def:determ_nesterov,def:determ_nesterov_two,def:determ_nesterov_three,def:determ_nesterov_four} above.

\begingroup
\providecommandordefault{\a}{\mathfrak{a}}
\providecommandordefault{\b}{\mathfrak{b}}
\providecommandordefault{\c}{\mathfrak{c}}

\cfclear
\begin{athm}{prop}{NesterovGeneral}[Comparison of general Nesterov-type \GD\ optimization methods]
Let 
	$\defaultParamDim \in \N$, 
	$(\a^{(1)}_n)_{n \in \N} \subseteq (0,\infty)$,
	$(\a^{(2)}_n)_{n \in \N} \subseteq (0,\infty)$,
	$(\b^{(1)}_n)_{n \in \N} \subseteq (0,\infty)$,
	$(\b^{(2)}_n)_{n \in \N} \subseteq (0,\infty)$,
	$(\c^{(1)}_n)_{n \in \N} \subseteq (0,\infty)$,
	$(\c^{(2)}_n)_{n \in \N} \subseteq (0,\infty)$,
	$\xi \in \R^\defaultParamDim$,
   	$\defaultLossFunction \in C^1(\R^\defaultParamDim, \R)$
satisfy for all 
	$n \in \N$
that
\begin{equation}
\label{NesterovGeneral:ass1}
\begin{split}
	\b^{(1)}_n \c^{(1)}_n 
=
	\b^{(2)}_n \c^{(2)}_n
\qandq
	\frac{
		\a^{(1)}_{n+1} \b^{(1)}_{n}
	}{
		\b^{(1)}_{n+1}
	}
=
	\frac{
		\a^{(2)}_{n+1} \b^{(2)}_{n}
	}{
		\b^{(2)}_{n+1}
	},
\end{split}
\end{equation}
and for every $i \in \{1,2\}$
let 
	$\Theta^{(i)} \colon \N_0 \to \R^\defaultParamDim$	
	and
	$\mathbf{m}^{(i)} \colon \N_0 \to \R^\defaultParamDim$
satisfy for all 
	$n \in \N$
that
\begin{equation}
\label{NesterovGeneral:ass2}
\begin{split}
	\Theta^{(i)}_0 = \xi, \qquad \mathbf{m}^{(i)}_0 = 0,
\end{split}
\end{equation}
\begin{equation}
\label{NesterovGeneral:ass3}
\begin{split}
	\mathbf{m}^{(i)}_n = \a^{(i)}_n \mathbf{m}^{(i)}_{n-1} + \b^{(i)}_n \grad(\Theta^{(i)}_{n-1} - \c^{(i)}_n \a^{(i)}_n \mathbf{m}^{(i)}_{n-1}),
\end{split}
\end{equation}
\begin{equation}
\label{NesterovGeneral:ass4}
\begin{split}
	\andq \Theta^{(i)}_n = \Theta^{(i)}_{n-1} - \c^{(i)}_n \mathbf{m}^{(i)}_n.
\end{split}
\end{equation}
Then
\begin{equation}
\label{NesterovGeneral:concl1}
\Theta^{(1)} = \Theta^{(2)}.
\end{equation}
\end{athm}

\begin{aproof}
Throughout this proof,
let 
	$\defaultGradientFunction \colon \R^\defaultParamDim \to \R^\defaultParamDim$ 
satisfy for all 
	$\theta \in \R^\defaultParamDim$
that 
\begin{equation}
\label{T_B_D}
\begin{split}
	\defaultGradientFunction(\theta) = (\nabla \defaultLossFunction)(\theta).
\end{split}
\end{equation}
\Nobs that
\enum{
	the fact that for all $n \in \N$ it holds that
	\begin{equation}
	\begin{split}
		\c^{(1)}_{n+1} 
	=
		\frac{\c^{(2)}_{n+1} \b^{(2)}_{n+1}}{\b^{(1)}_{n+1}},
	\qquad
		\frac{\c^{(2)}_{n}}{\c^{(1)}_{n}}
	=
		\frac{\b^{(1)}_{n+1}}{\b^{(2)}_{n+1}},
	\qandq
		\frac{\b^{(2)}_{n+1} \a^{(1)}_{n+1} \b^{(1)}_{n}}{\b^{(1)}_{n+1} \b^{(2)}_{n}}
	=
		\a^{(2)}_{n+1}
	\end{split}
	\end{equation}
}
\proves
that for all
	$n \in \N$
it holds that
\begin{equation}
\label{NesterovGeneral:eq0}
\begin{split}
	\frac{\c^{(1)}_{n+1} \a^{(1)}_{n+1} \c^{(2)}_{n}}{\c^{(1)}_{n}}
=
	\frac{\c^{(2)}_{n+1} \b^{(2)}_{n+1} \a^{(1)}_{n} \b^{(1)}_{n}}{\b^{(1)}_{n+1} \b^{(2)}_{n}}
=
	\c^{(2)}_{n+1} \a^{(2)}_{n+1}.
\end{split}
\end{equation}
\Moreover
\enum{
	\cref{NesterovGeneral:ass2};
}
\proves
that
\begin{equation}
\label{NesterovGeneral:eq01}
\begin{split}
	\mathbf{m}^{(1)}_0 
=
	0
=
	\mathbf{m}^{(2)}_0
\qandq
	\Theta^{(1)}_0
=
	\xi
=
	\Theta^{(2)}_0.
\end{split}
\end{equation}
Next we claim that for all
	$n \in \N$
it holds that
\begin{equation}
\label{NesterovGeneral:eq1}
\begin{split}
	\c^{(1)}_n \mathbf{m}^{(1)}_n 
= 
	\c^{(2)}_n \mathbf{m}^{(2)}_n
\qandq
	\Theta^{(1)}_n
=
	\Theta^{(2)}_n.
\end{split}
\end{equation}
We now prove \cref{NesterovGeneral:eq1} by induction on $n \in \N$.
For the base case $n = 1$ \nobs that
\enum{
	\cref{NesterovGeneral:ass1};
	\cref{NesterovGeneral:ass2};
	\cref{NesterovGeneral:eq01};
}
\prove
that
\begin{equation}
\label{NesterovGeneral:eq2}
\begin{split}
	\c^{(1)}_1 \mathbf{m}^{(1)}_1
&=
	\c^{(1)}_1 (\a^{(1)}_1 \mathbf{m}^{(1)}_0 + \b^{(1)}_1 \defaultGradientFunction(\Theta^{(1)}_0 - \c^{(1)}_1 \a^{(1)}_1 \mathbf{m}^{(1)}_0))\\
&=
	\c^{(1)}_1 \b^{(1)}_1 \defaultGradientFunction(\Theta^{(1)}_0)\\
&=
	\c^{(2)}_1 \b^{(2)}_1 \defaultGradientFunction(\Theta^{(2)}_0)\\
&=
	\c^{(2)}_1 (\a^{(2)}_1 \mathbf{m}^{(2)}_0 + \b^{(2)}_1 \defaultGradientFunction(\Theta^{(2)}_0 - \c^{(2)}_1 \a^{(2)}_1 \mathbf{m}^{(2)}_0))\\
&=
	\c^{(2)}_1 \mathbf{m}^{(2)}_1.
\end{split}
\end{equation}
\enum{
	This;
	\cref{NesterovGeneral:ass4};
	\cref{NesterovGeneral:eq01}
}
\prove 
that
\begin{equation}
	\Theta^{(1)}_1
=
	\Theta^{(1)}_0 - \c^{(1)}_1 \mathbf{m}^{(1)}_1
=
	\Theta^{(2)}_0 - \c^{(2)}_1 \mathbf{m}^{(2)}_1
=
	\Theta^{(2)}_1.
\end{equation}
Combining this and \cref{NesterovGeneral:eq2} establishes \cref{NesterovGeneral:eq1} in the base case $n = 1$.
For the induction step
$ \N \ni n \to n+1 \in \{2, 3, \ldots\}$ 
let $n \in \N$ and assume that
\begin{equation}
\label{NesterovGeneral:eq3}
\begin{split}
	\c^{(1)}_{n} \mathbf{m}^{(1)}_{n}
=
	\c^{(2)}_{n} \mathbf{m}^{(2)}_{n}
\qandq
	\Theta^{(1)}_{n}
=
	\Theta^{(2)}_{n}.
\end{split}
\end{equation}
\Nobs that
\enum{
	\cref{NesterovGeneral:ass1};
	\cref{NesterovGeneral:ass3};
	\cref{NesterovGeneral:eq0};
	\cref{NesterovGeneral:eq3};
}
\prove that
\begin{equation}
\label{NesterovGeneral:eq4}
\begin{split}
	\c^{(1)}_{n+1} \mathbf{m}^{(1)}_{n+1}
&=
	\c^{(1)}_{n+1} (\a^{(1)}_{n+1} \mathbf{m}^{(1)}_{n} + \b^{(1)}_{n+1} \defaultGradientFunction\pr{\Theta^{(1)}_{n} - \c^{(1)}_{n+1} \a^{(1)}_{n+1} \mathbf{m}^{(1)}_{n}}\\
&=
	\frac{\c^{(1)}_{n+1} \a^{(1)}_{n+1} \c^{(2)}_{n}}{\c^{(1)}_{n}} \mathbf{m}^{(2)}_{n} + \c^{(1)}_{n+1} \b^{(1)}_{n+1} \defaultGradientFunction\pr*{\Theta^{(2)}_{n} - \frac{\c^{(1)}_{n+1} \a^{(1)}_{n+1} \c^{(2)}_{n}}{\c^{(1)}_{n}} \mathbf{m}^{(2)}_{n}}\\
&=
	\c^{(2)}_{n+1} \a^{(2)}_{n+1} \mathbf{m}^{(2)}_{n} + \c^{(2)}_{n+1} \b^{(2)}_{n+1} \defaultGradientFunction\pr{\Theta^{(2)}_{n} - \c^{(2)}_{n+1} \a^{(2)}_{n+1} \mathbf{m}^{(2)}_{n}}\\
&=
	\c^{(2)}_{n+1} (\a^{(2)}_{n+1} \mathbf{m}^{(2)}_{n} + \b^{(2)}_{n+1} \defaultGradientFunction\pr{\Theta^{(2)}_{n} - \c^{(2)}_{n+1} \a^{(2)}_{n+1} \mathbf{m}^{(2)}_{n}}\\
&=
	\c^{(2)}_{n+1} \mathbf{m}^{(2)}_{n+1}.
\end{split}
\end{equation}
\enum{
	This;
	\cref{NesterovGeneral:ass4};
	\cref{NesterovGeneral:eq3};
}
\prove that
\begin{equation}
\label{T_B_D}
\begin{split}
	\Theta^{(1)}_{n+1}
=
	\Theta^{(1)}_{n} - \c^{(1)}_{n+1} \mathbf{m}^{(1)}_{n+1}
=
	\Theta^{(2)}_{n} - \c^{(2)}_{n+1} \mathbf{m}^{(2)}_{n+1}
=
	\Theta^{(2)}_{n+1}.
\end{split}
\end{equation}
Induction thus proves \cref{NesterovGeneral:eq1}.
Combining \cref{NesterovGeneral:eq01} and \cref{NesterovGeneral:eq1} establishes \cref{NesterovGeneral:concl1}.
\end{aproof}

\cfclear
\begin{athm}{cor}{Nesterov1vs2}[Comparison of the \first and \second version of the momentum \GD\ optimization method]
Let 
	$\defaultParamDim \in \N$, 
	$(\gamma^{(1)}_n)_{n \in \N} \subseteq (0,\infty)$,
	$(\gamma^{(2)}_n)_{n \in \N} \subseteq (0,\infty)$,
	$(\alpha^{(1)}_n)_{n \in \N} \subseteq (0,1)$,
	$(\alpha^{(2)}_n)_{n \in \N} \subseteq (0,\infty)$,
	$\xi \in \R^\defaultParamDim$,
   	$\defaultLossFunction \in C^1(\R^\defaultParamDim, \R)$
satisfy for all 
	$n \in \N$
that
\begin{equation}
\label{Nesterov1vs2:ass1}
\begin{split}
	\gamma^{(1)}_n (1-\alpha^{(1)}_n)
=
	\gamma^{(2)}_n
\qandq
	\frac{
		\alpha^{(1)}_{n+1}(1-\alpha^{(1)}_{n})
	}{
		1-\alpha^{(1)}_{n+1}
	}
=
	\alpha^{(2)}_{n+1},
\end{split}
\end{equation}
for every $i \in \{1,2\}$\cfadd{def:determ_nesterov}\cfadd{def:determ_nesterov_two}
let 
	$\Theta^{(i)} \colon \N_0 \to \R^\defaultParamDim$
be the Nesterov accelerated \GD\ process (\ith version) for the objective function $\defaultLossFunction$ with learning rates $(\gamma^{(i)}_n)_{n \in \N}$,
momentum decay factors $(\alpha^{(i)}_n)_{n \in \N}$,
and initial value $\xi$ \cfload.
Then
\begin{equation}
\label{Nesterov1vs2:concl1}
\Theta^{(1)} = \Theta^{(2)}.
\end{equation}
\end{athm}

\begin{aproof}
Throughout this proof let 
	$(\a^{(1)}_n)_{n \in \N} \subseteq (0,\infty)$,
	$(\a^{(2)}_n)_{n \in \N} \subseteq (0,\infty)$,
	$(\b^{(1)}_n)_{n \in \N} \subseteq (0,\infty)$,
	$(\b^{(2)}_n)_{n \in \N} \subseteq (0,\infty)$,
	$(\c^{(1)}_n)_{n \in \N} \subseteq (0,\infty)$,
	$(\c^{(2)}_n)_{n \in \N} \subseteq (0,\infty)$
satisfy for all 
	$n \in \N$
that
\begin{equation}
\label{Nesterov1vs2:setting1}
\begin{split}
	\a^{(1)}_n = \alpha^{(1)}_n, \qquad \b^{(1)}_n = 1 - \alpha^{(1)}_n, \qquad \c^{(1)}_n = \gamma^{(1)}_n,
\end{split}
\end{equation}
\begin{equation}
\label{Nesterov1vs2:setting2}
\begin{split} 
	\a^{(2)}_n = \alpha^{(2)}_n, \qquad \b^{(2)}_n = 1, \qandq \c^{(2)}_n = \gamma^{(2)}_n.
\end{split}
\end{equation}
\Nobs that  
\enum{
	\cref{eq:def:nesterov_1_1};
	\cref{eq:def:nesterov_1_2};
	\cref{eq:def:nesterov_1_3};
	\cref{eq:def:nesterov_2_1};
	\cref{eq:def:nesterov_2_2};
	\cref{eq:def:nesterov_2_3};
}
\prove 
that for all 
	$i \in \{1,2\}$,
	$n \in \N$
it holds that 
\begin{equation}
\label{Nesterov1vs2:eq1}
\begin{split}
	\Theta^{(i)}_0 = \xi, \qquad \mathbf{m}^{(i)}_0 = 0,
\end{split}
\end{equation}
\begin{equation}
\label{Nesterov1vs2:eq2}
\begin{split}
	\mathbf{m}^{(i)}_n = \a^{(i)}_n \mathbf{m}^{(i)}_{n-1} + \b^{(i)}_n \grad(\Theta^{(i)}_{n-1} - \c^{(i)}_n \a^{(i)}_n \mathbf{m}^{(i)}_{n-1}),
\end{split}
\end{equation}
\begin{equation}
\label{Nesterov1vs2:eq3}
\begin{split}
	\andq \Theta^{(i)}_n = \Theta^{(i)}_{n-1} - \c^{(i)}_n \mathbf{m}^{(i)}_n.
\end{split}
\end{equation}
\Moreover 
\enum{
	\cref{Nesterov1vs2:ass1};
	\cref{Nesterov1vs2:setting1};
	\cref{Nesterov1vs2:setting2};
}
\proves
that for all 
	$n \in \N$
it holds that 
\begin{equation}
\label{Nesterov1vs2:eq4}
\begin{split}
	\b^{(1)}_n \c^{(1)}_n 
=
	(1-\alpha^{(1)}_n) \gamma^{(1)}_n
=
	\gamma^{(2)}_n
=
	\b^{(2)}_n \c^{(2)}_n.
\end{split}
\end{equation}
\Moreover 
\enum{
	\cref{Nesterov1vs2:ass1};
	\cref{Nesterov1vs2:setting1};
	\cref{Nesterov1vs2:setting2};
}
\proves
that for all 
	$n \in \N$
it holds that
\begin{equation}
\label{Nesterov1vs2:eq5}
\begin{split} 
	\frac{
		\a^{(1)}_{n+1} \b^{(1)}_{n}
	}{
		\b^{(1)}_{n+1}
	}
=
	\frac{
		\alpha^{(1)}_{n+1} (1-\alpha^{(1)}_{n})
	}{
		1-\alpha^{(1)}_{n+1}
	}
=
	\alpha^{(2)}_{n+1}
=
	\frac{
		\a^{(2)}_{n+1} \b^{(2)}_{n}
	}{
		\b^{(2)}_{n+1}
	}.
\end{split}
\end{equation} 
Combining 
\enum{
	this;
	\cref{Nesterov1vs2:eq1};
	\cref{Nesterov1vs2:eq2};
	\cref{Nesterov1vs2:eq3};
	\cref{Nesterov1vs2:eq4};
}
with 
\cref{NesterovGeneral}
\proves
\eqref{Nesterov1vs2:concl1}.
\end{aproof}

\cfclear
\begin{athm}{lemma}{Nesterov1vs3}[Comparison of the \first and \third version of the momentum \GD\ optimization method]
Let 
	$\defaultParamDim \in \N$, 
	$(\gamma^{(1)}_n)_{n \in \N} \subseteq (0,\infty)$,
	$(\gamma^{(3)}_n)_{n \in \N} \subseteq (0,\infty)$,
	$(\alpha^{(1)}_n)_{n \in \N} \subseteq (0,1)$,
	$(\alpha^{(3)}_n)_{n \in \N} \subseteq (0,1)$,
	$\xi \in \R^\defaultParamDim$,
   	$\defaultLossFunction \in C^1(\R^\defaultParamDim, \R)$
satisfy for all 
	$n \in \N$
that
\begin{equation}
\label{Nesterov1vs3:ass1}
\begin{split}
	\gamma^{(1)}_n (1-\alpha^{(1)}_n)
=
	\gamma^{(3)}_n (1-\alpha^{(3)}_n)
\qandq
	\frac{
		\gamma^{(1)}_{n+1} \alpha^{(1)}_{n+1}
	}{
		\gamma^{(1)}_{n}
	}
=
	\alpha^{(3)}_{n+1},
\end{split}
\end{equation}
for every $i \in \{1,3\}$\cfadd{def:determ_nesterov}\cfadd{def:determ_nesterov_four}
let 
	$\Theta^{(i)} \colon \N_0 \to \R^\defaultParamDim$
be the Nesterov accelerated \GD\ process (\ith version) for the objective function $\defaultLossFunction$ with learning rates $(\gamma^{(i)}_n)_{n \in \N}$,
momentum decay factors $(\alpha^{(i)}_n)_{n \in \N}$,
and initial value $\xi$ \cfload.
Then
\begin{equation}
\label{Nesterov1vs3:concl1}
\Theta^{(1)} = \Theta^{(3)}.
\end{equation}
\end{athm}

\begin{aproof}
Throughout this proof let 
	$(\a^{(1)}_n)_{n \in \N} \subseteq (0,\infty)$,
	$(\a^{(3)}_n)_{n \in \N} \subseteq (0,\infty)$,
	$(\b^{(1)}_n)_{n \in \N} \subseteq (0,\infty)$,
	$(\b^{(3)}_n)_{n \in \N} \subseteq (0,\infty)$,
	$(\c^{(1)}_n)_{n \in \N} \subseteq (0,\infty)$,
	$(\c^{(3)}_n)_{n \in \N} \subseteq (0,\infty)$
satisfy for all 
	$n \in \N$
that
\begin{equation}
\label{Nesterov1vs3:setting1}
\begin{split}
	\a^{(1)}_n = \alpha^{(1)}_n, \qquad \b^{(1)}_n = 1 - \alpha^{(1)}_n, \qquad \c^{(1)}_n = \gamma^{(1)}_n,
\end{split}
\end{equation}
\begin{equation}
\label{Nesterov1vs3:setting2}
\begin{split} 
	\a^{(3)}_n = \alpha^{(3)}_n, \qquad \b^{(3)}_n = (1-\alpha^{(3)}_n) \gamma^{(3)}_n, \qandq \c^{(3)}_n = 1.
\end{split}
\end{equation}
\Nobs that  
\enum{
	\cref{eq:def:nesterov_1_1};
	\cref{eq:def:nesterov_1_2};
	\cref{eq:def:nesterov_1_3};
	\cref{eq:def:nesterov_3_1};
	\cref{eq:def:nesterov_3_2};
	\cref{eq:def:nesterov_3_3};
}
\prove 
that for all 
	$i \in \{1,3\}$,
	$n \in \N$
it holds that 
\begin{equation}
\label{Nesterov1vs3:eq1}
\begin{split}
	\Theta^{(i)}_0 = \xi, \qquad \mathbf{m}^{(i)}_0 = 0,
\end{split}
\end{equation}
\begin{equation}
\label{Nesterov1vs3:eq2}
\begin{split}
	\mathbf{m}^{(i)}_n = \a^{(i)}_n \mathbf{m}^{(i)}_{n-1} + \b^{(i)}_n \grad(\Theta^{(i)}_{n-1} - \c^{(i)}_n \a^{(i)}_n \mathbf{m}^{(i)}_{n-1}),
\end{split}
\end{equation}
\begin{equation}
\label{Nesterov1vs3:eq3}
\begin{split}
	\andq \Theta^{(i)}_n = \Theta^{(i)}_{n-1} - \c^{(i)}_n \mathbf{m}^{(i)}_n.
\end{split}
\end{equation}
\Moreover 
\enum{
	\cref{Nesterov1vs3:ass1};
	\cref{Nesterov1vs3:setting1};
	\cref{Nesterov1vs3:setting2};
}
\proves
that for all 
	$n \in \N$
it holds that 
\begin{equation}
\label{Nesterov1vs3:eq4}
\begin{split}
	\b^{(1)}_n \c^{(1)}_n 
=
	(1-\alpha^{(1)}_n) \gamma^{(1)}_n
=
	(1-\alpha^{(3)}_n) \gamma^{(3)}_n 
=
	\b^{(3)}_n \c^{(3)}_n.
\end{split}
\end{equation}
\Moreover 
\enum{
	\cref{Nesterov1vs3:ass1};
	\cref{Nesterov1vs3:setting1};
	\cref{Nesterov1vs3:setting2};
}
\proves
that for all 
	$n \in \N$
it holds that
\begin{equation}
\label{Nesterov1vs3:eq5}
\begin{split} 
	\frac{
		\a^{(1)}_{n+1} \b^{(1)}_{n}
	}{
		\b^{(1)}_{n+1}
	}
&= 
	\frac{
		\alpha^{(1)}_{n+1} (1-\alpha^{(1)}_{n})
	}{
		1-\alpha^{(1)}_{n+1}
	}
=
	\frac{
		\alpha^{(1)}_{n+1} \gamma^{(3)}_{n} (1-\alpha^{(3)}_{n}) \gamma^{(1)}_{n+1}
	}{
		\gamma^{(1)}_{n} \gamma^{(3)}_{n+1} (1-\alpha^{(3)}_{n+1})
	}\\
&=
	\frac{
		\alpha^{(3)}_{n+1} \gamma^{(3)}_{n} (1-\alpha^{(3)}_{n}) 
	}{
		\gamma^{(3)}_{n+1} (1-\alpha^{(3)}_{n+1})
	}
=
	\frac{
		\a^{(3)}_{n+1} \b^{(3)}_{n}
	}{
		\b^{(3)}_{n+1}
	}.
\end{split}
\end{equation} 
Combining 
\enum{
	this;
	\cref{Nesterov1vs3:eq1};
	\cref{Nesterov1vs3:eq2};
	\cref{Nesterov1vs3:eq3};
	\cref{Nesterov1vs3:eq4};
}
with 
\cref{NesterovGeneral}
\proves
\eqref{Nesterov1vs3:concl1}.
\end{aproof}

\cfclear
\begin{athm}{lemma}{Nesterov1vs4}[Comparison of the \first and \fourth version of the momentum \GD\ optimization method]
Let 
	$\defaultParamDim \in \N$, 
	$(\gamma^{(1)}_n)_{n \in \N} \subseteq (0,\infty)$,
	$(\gamma^{(4)}_n)_{n \in \N} \subseteq (0,\infty)$,
	$(\alpha^{(1)}_n)_{n \in \N} \subseteq (0,1)$,
	$(\alpha^{(4)}_n)_{n \in \N} \subseteq (0,\infty)$,
	$\xi \in \R^\defaultParamDim$,
   	$\defaultLossFunction \in C^1(\R^\defaultParamDim, \R)$
satisfy for all 
	$n \in \N$
that
\begin{equation}
\label{Nesterov1vs4:ass1}
\begin{split}
	\gamma^{(1)}_n (1-\alpha^{(1)}_n)
=
	\gamma^{(4)}_n
\qandq
	\frac{
		\gamma^{(1)}_{n+1} \alpha^{(1)}_{n+1}
	}{
		\gamma^{(1)}_{n}
	}
=
	\alpha^{(4)}_{n+1},
\end{split}
\end{equation}
for every $i \in \{1,4\}$\cfadd{def:determ_nesterov}\cfadd{def:determ_nesterov_three}
let 
	$\Theta^{(i)} \colon \N_0 \to \R^\defaultParamDim$
be the Nesterov accelerated \GD\ process (\ith version) for the objective function $\defaultLossFunction$ with learning rates $(\gamma^{(i)}_n)_{n \in \N}$,
momentum decay factors $(\alpha^{(i)}_n)_{n \in \N}$,
and initial value $\xi$ \cfload.
Then
\begin{equation}
\label{Nesterov1vs4:concl1}
\Theta^{(1)} = \Theta^{(4)}.
\end{equation}
\end{athm}

\begin{aproof}
Throughout this proof let 
	$(\a^{(1)}_n)_{n \in \N} \subseteq (0,\infty)$,
	$(\a^{(4)}_n)_{n \in \N} \subseteq (0,\infty)$,
	$(\b^{(1)}_n)_{n \in \N} \subseteq (0,\infty)$,
	$(\b^{(4)}_n)_{n \in \N} \subseteq (0,\infty)$,
	$(\c^{(1)}_n)_{n \in \N} \subseteq (0,\infty)$,
	$(\c^{(4)}_n)_{n \in \N} \subseteq (0,\infty)$
satisfy for all 
	$n \in \N$
that
\begin{equation}
\label{Nesterov1vs4:setting1}
\begin{split}
	\a^{(1)}_n = \alpha^{(1)}_n, \qquad \b^{(1)}_n = 1 - \alpha^{(1)}_n, \qquad \c^{(1)}_n = \gamma^{(1)}_n,
\end{split}
\end{equation}
\begin{equation}
\label{Nesterov1vs4:setting2}
\begin{split} 
	\a^{(4)}_n = \alpha^{(4)}_n, \qquad \b^{(4)}_n = \gamma^{(4)}_n, \qandq \c^{(4)}_n = 1.
\end{split}
\end{equation}
\Nobs that  
\enum{
	\cref{eq:def:nesterov_1_1};
	\cref{eq:def:nesterov_1_2};
	\cref{eq:def:nesterov_1_3};
	\cref{eq:def:nesterov_4_1};
	\cref{eq:def:nesterov_4_2};
	\cref{eq:def:nesterov_4_3};
}
\prove 
that for all 
	$i \in \{1,4\}$,
	$n \in \N$
it holds that 
\begin{equation}
\label{Nesterov1vs4:eq1}
\begin{split}
	\Theta^{(i)}_0 = \xi, \qquad \mathbf{m}^{(i)}_0 = 0,
\end{split}
\end{equation}
\begin{equation}
\label{Nesterov1vs4:eq2}
\begin{split}
	\mathbf{m}^{(i)}_n = \a^{(i)}_n \mathbf{m}^{(i)}_{n-1} + \b^{(i)}_n \grad(\Theta^{(i)}_{n-1} - \c^{(i)}_n \a^{(i)}_n \mathbf{m}^{(i)}_{n-1}),
\end{split}
\end{equation}
\begin{equation}
\label{Nesterov1vs4:eq3}
\begin{split}
	\andq \Theta^{(i)}_n = \Theta^{(i)}_{n-1} - \c^{(i)}_n \mathbf{m}^{(i)}_n.
\end{split}
\end{equation}
\Moreover 
\enum{
	\cref{Nesterov1vs4:ass1};
	\cref{Nesterov1vs4:setting1};
	\cref{Nesterov1vs4:setting2};
}
\proves
that for all 
	$n \in \N$
it holds that 
\begin{equation}
\label{Nesterov1vs4:eq4}
\begin{split}
	\b^{(1)}_n \c^{(1)}_n 
=
	(1-\alpha^{(1)}_n) \gamma^{(1)}_n
= 
	\gamma^{(4)}_n
=
	\b^{(4)}_n \c^{(4)}_n.
\end{split}
\end{equation}
\Moreover 
\enum{
	\cref{Nesterov1vs4:ass1};
	\cref{Nesterov1vs4:setting1};
	\cref{Nesterov1vs4:setting2};
}
\proves
that for all 
	$n \in \N$
it holds that
\begin{equation}
\label{Nesterov1vs4:eq5}
\begin{split} 
	\frac{
		\a^{(1)}_{n+1} \b^{(1)}_{n}
	}{
		\b^{(1)}_{n+1}
	}
= 
	\frac{
		\alpha^{(1)}_{n+1} (1-\alpha^{(1)}_{n})
	}{
		1-\alpha^{(1)}_{n+1}
	}
=
	\frac{
		\alpha^{(1)}_{n+1} 
		\gamma^{(4)}_{n}
		\gamma^{(1)}_{n+1}
	}{
		\gamma^{(1)}_{n}
		\gamma^{(4)}_{n+1}
	}
=
	\frac{
		\alpha^{(4)}_{n+1} 
		\gamma^{(4)}_{n}
	}{
		\gamma^{(4)}_{n+1}
	}
=
	\frac{
		\a^{(4)}_{n+1} \b^{(4)}_{n}
	}{
		\b^{(4)}_{n+1}
	}.
\end{split}
\end{equation} 
Combining 
\enum{
	this;
	\cref{Nesterov1vs4:eq1};
	\cref{Nesterov1vs4:eq2};
	\cref{Nesterov1vs4:eq3};
	\cref{Nesterov1vs4:eq4};
}
with 
\cref{NesterovGeneral}
\proves
\eqref{Nesterov1vs4:concl1}.
\end{aproof}

\cfclear
\begin{athm}{cor}{Nesterov2vs3}[Comparison of the \second and \third version of the momentum \SGD\ optimization method]
Let 
	$\defaultParamDim \in \N$, 
	$(\gamma^{(2)}_n)_{n \in \N} \subseteq (0,\infty)$,
	$(\gamma^{(3)}_n)_{n \in \N} \subseteq (0,\infty)$,
	$(\alpha^{(2)}_n)_{n \in \N} \subseteq (0,\infty)$,
	$(\alpha^{(3)}_n)_{n \in \N} \subseteq (0,1)$,
	$\xi \in \R^\defaultParamDim$,
   	$\defaultLossFunction \in C^1(\R^\defaultParamDim, \R)$
satisfy for all 
	$n \in \N$
that
\begin{equation}
\label{Nesterov2vs3:ass1}
\begin{split}
	\gamma^{(2)}_n
=
	\gamma^{(3)}_n (1-\alpha^{(3)}_n)
\qandq
	\frac{
		\gamma^{(2)}_{n+1} \alpha^{(2)}_{n+1}
	}{
		\gamma^{(2)}_{n}
	}
=
	\alpha^{(3)}_{n+1},
\end{split}
\end{equation}
for every $i \in \{2,3\}$\cfadd{def:determ_nesterov_two}\cfadd{def:determ_nesterov_four}
let 
	$\Theta^{(i)} \colon \N_0 \to \R^\defaultParamDim$
be the Nesterov accelerated \GD\ process (\ith version) for the objective function $\defaultLossFunction$ with learning rates $(\gamma^{(i)}_n)_{n \in \N}$,
momentum decay factors $(\alpha^{(i)}_n)_{n \in \N}$,
and initial value $\xi$ \cfload.
Then
\begin{equation}
\label{Nesterov2vs3:concl1}
\Theta^{(2)} = \Theta^{(3)}.
\end{equation}
\end{athm}

\begin{aproof}
Throughout this proof let 
	$(\a^{(2)}_n)_{n \in \N} \subseteq (0,\infty)$,
	$(\a^{(3)}_n)_{n \in \N} \subseteq (0,\infty)$,
	$(\b^{(2)}_n)_{n \in \N} \subseteq (0,\infty)$,
	$(\b^{(3)}_n)_{n \in \N} \subseteq (0,\infty)$,
	$(\c^{(2)}_n)_{n \in \N} \subseteq (0,\infty)$,
	$(\c^{(3)}_n)_{n \in \N} \subseteq (0,\infty)$
satisfy for all 
	$n \in \N$
that
\begin{equation}
\label{Nesterov2vs3:setting1}
\begin{split}
	\a^{(2)}_n = \alpha^{(2)}_n, \qquad \b^{(2)}_n = 1, \qquad \c^{(2)}_n = \gamma^{(2)}_n,
\end{split}
\end{equation}
\begin{equation}
\label{Nesterov2vs3:setting2}
\begin{split} 
	\a^{(3)}_n = \alpha^{(3)}_n, \qquad \b^{(3)}_n = (1-\alpha^{(3)}_n) \gamma^{(3)}_n, \qandq \c^{(3)}_n = 1.
\end{split}
\end{equation}
\Nobs that  
\enum{
	\cref{eq:def:nesterov_2_1};
	\cref{eq:def:nesterov_2_2};
	\cref{eq:def:nesterov_2_3};
	\cref{eq:def:nesterov_3_1};
	\cref{eq:def:nesterov_3_2};
	\cref{eq:def:nesterov_3_3};
}
\prove 
that for all 
	$i \in \{2,3\}$,
	$n \in \N$
it holds that 
\begin{equation}
\label{Nesterov2vs3:eq1}
\begin{split}
	\Theta^{(i)}_0 = \xi, \qquad \mathbf{m}^{(i)}_0 = 0,
\end{split}
\end{equation}
\begin{equation}
\label{Nesterov2vs3:eq2}
\begin{split}
	\mathbf{m}^{(i)}_n = \a^{(i)}_n \mathbf{m}^{(i)}_{n-1} + \b^{(i)}_n \grad(\Theta^{(i)}_{n-1} - \c^{(i)}_n \a^{(i)}_n \mathbf{m}^{(i)}_{n-1}),
\end{split}
\end{equation}
\begin{equation}
\label{Nesterov2vs3:eq3}
\begin{split}
	\andq \Theta^{(i)}_n = \Theta^{(i)}_{n-1} - \c^{(i)}_n \mathbf{m}^{(i)}_n.
\end{split}
\end{equation}
\Moreover 
\enum{
	\cref{Nesterov2vs3:ass1};
	\cref{Nesterov2vs3:setting1};
	\cref{Nesterov2vs3:setting2};
}
\proves
that for all 
	$n \in \N$
it holds that 
\begin{equation}
\label{Nesterov2vs3:eq4}
\begin{split}
	\b^{(2)}_n \c^{(2)}_n 
=
	\gamma^{(2)}_n
=
	\gamma^{(3)}_n (1-\alpha^{(3)}_n)
=
	\b^{(3)}_n \c^{(3)}_n.
\end{split}
\end{equation}
\Moreover 
\enum{
	\cref{Nesterov2vs3:ass1};
	\cref{Nesterov2vs3:setting1};
	\cref{Nesterov2vs3:setting2};
}
\proves
that for all 
	$n \in \N$
it holds that
\begin{equation}
\label{Nesterov2vs3:eq5}
\begin{split} 
	\frac{
		\a^{(2)}_{n+1} \b^{(2)}_{n}
	}{
		\b^{(2)}_{n+1}
	}
= 
	\alpha^{(2)}_{n+1}
=
	\frac{
		\alpha^{(3)}_{n+1} \gamma^{(3)}_{n} (1-\alpha^{(3)}_{n})
	}{
		\gamma^{(3)}_{n+1} (1-\alpha^{(3)}_{n+1})
	}
=
	\frac{
		\a^{(3)}_{n+1} \b^{(3)}_{n}
	}{
		\b^{(3)}_{n+1}
	}.
\end{split}
\end{equation} 
Combining 
\enum{
	this;
	\cref{Nesterov2vs3:eq1};
	\cref{Nesterov2vs3:eq2};
	\cref{Nesterov2vs3:eq3};
	\cref{Nesterov2vs3:eq4};
}
with 
\cref{NesterovGeneral}
\proves
\eqref{Nesterov2vs3:concl1}.
\end{aproof}

\cfclear
\begin{athm}{lemma}{Nesterov2vs4}[Comparison of the \second and \fourth version of the momentum \GD\ optimization method] 
Let 
	$\defaultParamDim \in \N$, 
	$(\gamma^{(2)}_n)_{n \in \N} \subseteq (0,\infty)$,
	$(\gamma^{(4)}_n)_{n \in \N} \subseteq (0,\infty)$,
	$(\alpha^{(2)}_n)_{n \in \N} \subseteq (0,1)$,
	$(\alpha^{(4)}_n)_{n \in \N} \subseteq (0,1)$,
	$\xi \in \R^\defaultParamDim$,
   	$\defaultLossFunction \in C^1(\R^\defaultParamDim, \R)$
satisfy for all  
	$n \in \N$
that
\begin{equation}
\label{Nesterov2vs4:ass1}
\begin{split}
	\gamma^{(2)}_n 
=
	\gamma^{(4)}_n
\qandq
	\frac{
		\gamma^{(2)}_{n+1} \alpha^{(2)}_{n+1}
	}{
		\gamma^{(2)}_{n}
	}
=
	\alpha^{(4)}_{n+1},
\end{split}
\end{equation}
for every $i \in \{2,4\}$\cfadd{def:determ_nesterov_three}\cfadd{def:determ_nesterov_two}
let 
	$\Theta^{(i)} \colon \N_0 \to \R^\defaultParamDim$
be the Nesterov accelerated \GD\ process (\ith version) for the objective function $\defaultLossFunction$ with learning rates $(\gamma^{(i)}_n)_{n \in \N}$,
momentum decay factors $(\alpha^{(i)}_n)_{n \in \N}$,
and initial value $\xi$ \cfload.
Then
\begin{equation}
\label{Nesterov2vs4:concl1}
\Theta^{(2)} = \Theta^{(4)}.
\end{equation}
\end{athm}

\begin{aproof}
Throughout this proof let 
	$(\a^{(2)}_n)_{n \in \N} \subseteq (0,\infty)$,
	$(\a^{(4)}_n)_{n \in \N} \subseteq (0,\infty)$,
	$(\b^{(2)}_n)_{n \in \N} \subseteq (0,\infty)$,
	$(\b^{(4)}_n)_{n \in \N} \subseteq (0,\infty)$,
	$(\c^{(2)}_n)_{n \in \N} \subseteq (0,\infty)$,
	$(\c^{(4)}_n)_{n \in \N} \subseteq (0,\infty)$
satisfy for all 
	$n \in \N$
that
\begin{equation}
\label{Nesterov2vs4:setting1}
\begin{split}
	\a^{(2)}_n = \alpha^{(2)}_n, \qquad \b^{(2)}_n = 1, \qquad \c^{(2)}_n = \gamma^{(2)}_n,
\end{split}
\end{equation}
\begin{equation}
\label{Nesterov2vs4:setting2}
\begin{split} 
	\a^{(4)}_n = \alpha^{(4)}_n, \qquad \b^{(4)}_n = \gamma^{(4)}_n, \qandq \c^{(4)}_n = 1.
\end{split}
\end{equation}
\Nobs that  
\enum{
	\cref{eq:def:nesterov_2_1};
	\cref{eq:def:nesterov_2_2};
	\cref{eq:def:nesterov_2_3};
	\cref{eq:def:nesterov_4_1};
	\cref{eq:def:nesterov_4_2};
	\cref{eq:def:nesterov_4_3};
}
\prove 
that for all 
	$i \in \{2,4\}$,
	$n \in \N$
it holds that 
\begin{equation}
\label{Nesterov2vs4:eq1}
\begin{split}
	\Theta^{(i)}_0 = \xi, \qquad \mathbf{m}^{(i)}_0 = 0,
\end{split}
\end{equation}
\begin{equation}
\label{Nesterov2vs4:eq2}
\begin{split}
	\mathbf{m}^{(i)}_n = \a^{(i)}_n \mathbf{m}^{(i)}_{n-1} + \b^{(i)}_n \grad(\Theta^{(i)}_{n-1} - \c^{(i)}_n \a^{(i)}_n \mathbf{m}^{(i)}_{n-1}),
\end{split}
\end{equation}
\begin{equation}
\label{Nesterov2vs4:eq3}
\begin{split}
	\andq \Theta^{(i)}_n = \Theta^{(i)}_{n-1} - \c^{(i)}_n \mathbf{m}^{(i)}_n.
\end{split}
\end{equation}
\Moreover 
\enum{
	\cref{Nesterov2vs4:ass1};
	\cref{Nesterov2vs4:setting1};
	\cref{Nesterov2vs4:setting2};
}
\proves
that for all 
	$n \in \N$
it holds that 
\begin{equation}
\label{Nesterov2vs4:eq4}
\begin{split}
	\b^{(2)}_n \c^{(2)}_n 
=
	\gamma^{(2)}_n
= 
	\gamma^{(4)}_n
=
	\b^{(4)}_n \c^{(4)}_n.
\end{split}
\end{equation}
\Moreover 
\enum{
	\cref{Nesterov2vs4:ass1};
	\cref{Nesterov2vs4:setting1};
	\cref{Nesterov2vs4:setting2};
}
\proves
that for all 
	$n \in \N$
it holds that
\begin{equation}
\label{Nesterov2vs4:eq5}
\begin{split} 
	\frac{
		\a^{(2)}_{n+1} \b^{(2)}_{n}
	}{
		\b^{(2)}_{n+1}
	}
= 
	\alpha^{(2)}_{n+1}
=
	\frac{
		\alpha^{(4)}_{n+1} \gamma^{(4)}_{n}
	}{
		\gamma^{(4)}_{n+1}
	}
=
	\frac{
		\a^{(4)}_{n+1} \b^{(4)}_{n}
	}{
		\b^{(4)}_{n+1}
	}.
\end{split}
\end{equation} 
Combining 
\enum{
	this;
	\cref{Nesterov2vs4:eq1};
	\cref{Nesterov2vs4:eq2};
	\cref{Nesterov2vs4:eq3};
	\cref{Nesterov2vs4:eq4};
}
with 
\cref{NesterovGeneral}
\proves
\eqref{Nesterov2vs4:concl1}.
\end{aproof}

\cfclear
\begin{athm}{cor}{Nesterov3vs4}[Comparison of the \third and \fourth version of the momentum \GD\ optimization method]
Let 
	$\defaultParamDim \in \N$,
	$(\gamma^{(3)}_n)_{n \in \N} \subseteq (0,\infty)$, 
	$(\gamma^{(4)}_n)_{n \in \N} \subseteq (0,\infty)$,
	$(\alpha^{(3)}_n)_{n \in \N} \subseteq (0,1)$,
	$(\alpha^{(4)}_n)_{n \in \N} \subseteq (0,\infty)$,
	$\xi \in \R^\defaultParamDim$,
   	$\defaultLossFunction \in C^1(\R^\defaultParamDim, \R)$
satisfy for all 
	$n \in \N$
that
\begin{equation}
\label{Nesterov3vs4:ass1}
\begin{split}
	\gamma^{(3)}_n (1-\alpha^{(3)}_n)
=
	\gamma^{(4)}_n
\qandq
	\alpha^{(3)}_{n+1}
=
	\alpha^{(4)}_{n+1},
\end{split}
\end{equation}
for every $i \in \{3,4\}$\cfadd{def:determ_nesterov_three}\cfadd{def:determ_nesterov_four}
let 
	$\Theta^{(i)} \colon \N_0 \to \R^\defaultParamDim$
be the Nesterov accelerated \GD\ process (\ith version) for the objective function $\defaultLossFunction$ with learning rates $(\gamma^{(i)}_n)_{n \in \N}$,
momentum decay factors $(\alpha^{(i)}_n)_{n \in \N}$,
and initial value $\xi$ \cfload.
Then
\begin{equation}
\label{Nesterov3vs4:concl1}
	\Theta^{(3)} = \Theta^{(4)}.
\end{equation}
\end{athm}

\begin{aproof}
Throughout this proof let 
	$(\a^{(3)}_n)_{n \in \N} \subseteq (0,\infty)$,
	$(\a^{(4)}_n)_{n \in \N} \subseteq (0,\infty)$,
	$(\b^{(3)}_n)_{n \in \N} \subseteq (0,\infty)$,
	$(\b^{(4)}_n)_{n \in \N} \subseteq (0,\infty)$,
	$(\c^{(3)}_n)_{n \in \N} \subseteq (0,\infty)$,
	$(\c^{(4)}_n)_{n \in \N} \subseteq (0,\infty)$
satisfy for all 
	$n \in \N$
that
\begin{equation}
\label{Nesterov3vs4:setting1}
\begin{split}
	\a^{(3)}_n = \alpha^{(3)}_n, \qquad \b^{(3)}_n = (1-\alpha^{(3)}_n) \gamma^{(3)}_n, \qquad \c^{(3)}_n = 1
\end{split}
\end{equation}
\begin{equation}
\label{Nesterov3vs4:setting2}
\begin{split} 
	\a^{(4)}_n = \alpha^{(4)}_n, \qquad \b^{(4)}_n = \gamma^{(4)}_n, \qandq \c^{(4)}_n = 1,
\end{split}
\end{equation}
\Nobs that  
\enum{
	\cref{eq:def:nesterov_3_1};
	\cref{eq:def:nesterov_3_2};
	\cref{eq:def:nesterov_3_3};
	\cref{eq:def:nesterov_4_1};
	\cref{eq:def:nesterov_4_2};
	\cref{eq:def:nesterov_4_3};
}
\prove 
that for all 
	$i \in \{3,4\}$,
	$n \in \N$
it holds that 
\begin{equation}
\label{Nesterov3vs4:eq1}
\begin{split}
	\Theta^{(i)}_0 = \xi, \qquad \mathbf{m}^{(i)}_0 = 0,
\end{split}
\end{equation}
\begin{equation}
\label{Nesterov3vs4:eq2}
\begin{split}
	\mathbf{m}^{(i)}_n = \a^{(i)}_n \mathbf{m}^{(i)}_{n-1} + \b^{(i)}_n \grad(\Theta^{(i)}_{n-1} - \c^{(i)}_n \a^{(i)}_n \mathbf{m}^{(i)}_{n-1}),
\end{split}
\end{equation}
\begin{equation}
\label{Nesterov3vs4:eq3}
\begin{split}
	\andq \Theta^{(i)}_n = \Theta^{(i)}_{n-1} - \c^{(i)}_n \mathbf{m}^{(i)}_n.
\end{split}
\end{equation}
\Moreover 
\enum{
	\cref{Nesterov3vs4:ass1};
	\cref{Nesterov3vs4:setting1};
	\cref{Nesterov3vs4:setting2};
}
\proves
that for all 
	$n \in \N$
it holds that 
\begin{equation}
\label{Nesterov3vs4:eq4}
\begin{split}
	\b^{(3)}_n \c^{(3)}_n
=
	\gamma^{(3)}_n (1-\alpha^{(3)}_n)
=
	\gamma^{(4)}_n
=
	\b^{(4)}_n \c^{(4)}_n.
\end{split}
\end{equation}
\Moreover 
\enum{
	\cref{Nesterov3vs4:ass1};
	\cref{Nesterov3vs4:setting1};
	\cref{Nesterov3vs4:setting2};
}
\proves
that for all 
	$n \in \N$
it holds that
\begin{equation}
\label{Nesterov3vs4:eq5}
\begin{split} 
	\frac{
		\a^{(3)}_{n+1} \b^{(3)}_{n}
	}{
		\b^{(3)}_{n+1}
	}
=
	\frac{
		\alpha^{(3)}_{n+1} (1-\alpha^{(3)}_{n})\gamma^{(3)}_{n} 
	}{
		(1-\alpha^{(3)}_{n+1}) \gamma^{(3)}_{n+1} 
	}
=
	\frac{
		\alpha^{(4)}_{n+1} \gamma^{(4)}_{n} 
	}{
		\gamma^{(4)}_{n+1}
	}
=
	\frac{
		\a^{(4)}_{n+1} \b^{(4)}_{n}
	}{
		\b^{(4)}_{n+1}
	}.
\end{split}
\end{equation} 
Combining 
\enum{
	this;
	\cref{Nesterov3vs4:eq1};
	\cref{Nesterov3vs4:eq2};
	\cref{Nesterov3vs4:eq3};
	\cref{Nesterov3vs4:eq4};
}
with 
\cref{NesterovGeneral}
\proves
\eqref{Nesterov3vs4:concl1}.
\end{aproof}

\endgroup

\subsection{Bias-adjusted Nesterov accelerated momentum optimization}
\label{sect:determ_nesterov_bias}

In this section, we introduce a bias-adjusted version of the Nesterov accelerated \GD\ optimization method.
Roughly speaking, the bias-adjusted Nesterov accelerated \GD\ optimization method is obtained by adding the same kind of foresight to the bias-adjusted momentum \GD\ optimization method in \cref{def:determ_momentum_bias}
as the foresight that is added to the momentum \GD\ optimization method in \cref{def:determ_momentum} to obtain the Nesterov accelerated \GD\ optimization method in \cref{def:determ_nesterov}.

\defdetermNesterovBias
\algDescrDetermNesterovBias

\subsection{Shifted representations}
\label{sect:determ_nesterov_shifted}

In this section, we introduce shifted representations of all the Nesterov accelerated \GD\ optimization methods introduced in this section (cf.\ \cref{def:determ_nesterov,def:determ_nesterov_two,def:determ_nesterov_three,def:determ_nesterov_four,def:determ_nesterov_bias}).
Roughly speaking, the shifted representations are obtained by taking the point at which the gradient of the objective function is evaluated as the current state of the optimization process.

\subsubsection{Shifted representation for the first version of Nesterov accelerated momentum optimization}

\newcommand{\ShiftedProcess}{\Psi}

\cfclear
\begin{athm}{lemma}{Shifted_Nesterov_1}[Shifting the Nesterov accelerated \GD\ process]
Let 
	$\defaultParamDim \in \N$, 
	$(\gamma_n)_{n \in \N} \subseteq [0,\infty)$,
	$(\alpha_n)_{n \in \N} \subseteq [0,1]$,
	$\xi \in \R^\defaultParamDim$,
   	$\defaultLossFunction \in C^1(\R^\defaultParamDim, \R)$,
let 
	$\Theta \colon \N_0 \to \R^\defaultParamDim$
be the Nesterov accelerated \GD\ process\cfadd{def:determ_nesterov} for the objective function $\defaultLossFunction$ with learning rates $(\gamma_n)_{n \in \N}$,
momentum decay factors $(\alpha_n)_{n \in \N}$,
and initial value $\xi$,
let 
	$\mathbf{m} \colon \N_0 \to \R^\defaultParamDim$
satisfy for all
	$n \in \N$
that
	$\mathbf{m}_0 = 0$ 
and
\begin{equation}
\label{Shifted_Nesterov_1:ass1}
\begin{split}
	\mathbf{m}_n
=
	\alpha_n \mathbf{m}_{n-1} + (1 - \alpha_n) \grad(\Theta_{n-1} - \gamma_n \alpha_n \mathbf{m}_{n-1}),
\end{split}
\end{equation}
and let 
	$\ShiftedProcess \colon \N_0 \to \R^\defaultParamDim$
satisfy for all
	$n \in \N_0$
that
\begin{equation}
\label{Shifted_Nesterov_1:ass2}
\begin{split}
	\ShiftedProcess_n
=
	\Theta_n - \gamma_{n+1} \alpha_{n+1} \mathbf{m}_{n}
\end{split}
\end{equation}
\cfload.
Then it holds for all
	$n \in \N$
that
\begin{equation}
\label{Shifted_Nesterov_1:concl1}
\ShiftedProcess_0 = \xi, \qquad \mathbf{m}_0 = 0,
\end{equation}
\begin{equation}
\label{Shifted_Nesterov_1:concl2}
\mathbf{m}_n = \alpha_n \mathbf{m}_{n-1} + (1-\alpha_n)  \grad(\ShiftedProcess_{n-1}),
\end{equation}
\begin{equation}
\label{Shifted_Nesterov_1:concl3}
\andq \ShiftedProcess_n = \ShiftedProcess_{n-1} - \gamma_{n+1} \alpha_{n+1} \mathbf{m}_n - \gamma_n (1-\alpha_n) \grad(\ShiftedProcess_{n-1}). 
\end{equation}
\end{athm}

\begin{aproof}
\Nobs that
\enum{
	\cref{eq:def:nesterov_1_1};
	\cref{eq:def:nesterov_1_2};
	\cref{eq:def:nesterov_1_3};
	\cref{Shifted_Nesterov_1:ass1}
}
\prove
that for all
	$n \in \N$
it holds that
\begin{equation}
\label{Shifted_Nesterov_1:eq3}
\begin{split}
	\Theta_n 
=
	\Theta_{n-1} - \gamma_n \mathbf{m}_n.
\end{split}
\end{equation}
\Moreover
\enum{
	\cref{Shifted_Nesterov_1:ass1};
	the fact that $\Theta_0 = \xi$;
	the assumption that $\mathbf{m}_0 = 0$;
}
\prove
that 
\begin{equation}
\label{Shifted_Nesterov_1:eq4.1}
\begin{split}
	\ShiftedProcess_0
=
	\Theta_0 - \gamma_1 \alpha_1 \mathbf{m}_0
=
	\xi.
\end{split}
\end{equation}
\Moreover 
\enum{
	\cref{Shifted_Nesterov_1:ass1};
	\cref{Shifted_Nesterov_1:ass2};
}
\prove
that for all
	$n \in \N$
it holds that
\begin{equation}
\label{Shifted_Nesterov_1:eq4.2}
\begin{split}
	\mathbf{m}_n 
&=
	\alpha_n \mathbf{m}_{n-1} + (1-\alpha_n)  \grad(\Theta_{n-1} - \gamma_n \alpha_n \mathbf{m}_{n-1})\\
&=
	\alpha_n \mathbf{m}_{n-1} + (1-\alpha_n)  \grad(\ShiftedProcess_{n-1}).
\end{split}
\end{equation}
\enum{
	This;
	\cref{Shifted_Nesterov_1:ass2};
	\cref{Shifted_Nesterov_1:eq3};
}
\prove
that for all
	$n \in \N$
it holds that
\begin{equation}
\label{Shifted_Nesterov_1:eq5}
\begin{split}
	\ShiftedProcess_n
&=
	\Theta_n - \gamma_{n+1} \alpha_{n+1} \mathbf{m}_n\\
&=
	\Theta_{n-1} - \gamma_n \mathbf{m}_n - \gamma_{n+1} \alpha_{n+1} \mathbf{m}_n\\
&=
	\Theta_{n-1} - \gamma_n (\alpha_n \mathbf{m}_{n-1} + (1-\alpha_n)  \grad(\ShiftedProcess_{n-1})) - \gamma_{n+1} \alpha_{n+1} \mathbf{m}_n\\
&=
	\ShiftedProcess_{n-1} - \gamma_n (1-\alpha_n)  \grad(\ShiftedProcess_{n-1}) - \gamma_{n+1} \alpha_{n+1} \mathbf{m}_n.
\end{split}
\end{equation}
Combining 
\enum{
	this;
	\cref{Shifted_Nesterov_1:eq4.1};
	\cref{Shifted_Nesterov_1:eq4.2}
}
establishes
\cref{Shifted_Nesterov_1:concl1,Shifted_Nesterov_1:concl2,Shifted_Nesterov_1:concl3}.
\end{aproof}

\defdetermNesterovAlt
\algDescrDetermNesterovAlt

\subsubsection{Shifted representation for the second version of Nesterov accelerated momentum optimization}

\cfclear
\begin{athm}{lemma}{Shifted_Nesterov_2}[Shifting the Nesterov accelerated \GD\ process (\second version)]
Let 
	$\defaultParamDim \in \N$, 
	$(\gamma_n)_{n \in \N} \subseteq [0,\infty)$,
	$(\alpha_n)_{n \in \N} \subseteq [0,\infty)$,
	$\xi \in \R^\defaultParamDim$,
   	$\defaultLossFunction \in C^1(\R^\defaultParamDim, \R)$,
let 
	$\Theta \colon \N_0 \to \R^\defaultParamDim$
be the Nesterov accelerated \GD\ process\cfadd{def:determ_nesterov_two} (\second version) for the objective function $\defaultLossFunction$ with learning rates $(\gamma_n)_{n \in \N}$,
momentum decay factors $(\alpha_n)_{n \in \N}$,
and initial value $\xi$,
let 
	$\mathbf{m} \colon \N_0 \to \R^\defaultParamDim$
satisfy for all
	$n \in \N$
that
	$\mathbf{m}_0 = 0$ 
and
\begin{equation}
\label{Shifted_Nesterov_2:ass1}
\begin{split}
	\mathbf{m}_n
=
	\alpha_n \mathbf{m}_{n-1} + \grad(\Theta_{n-1} - \gamma_n \alpha_n \mathbf{m}_{n-1}),
\end{split}
\end{equation}
and let 
	$\ShiftedProcess \colon \N_0 \to \R^\defaultParamDim$
satisfy for all
	$n \in \N_0$
that
\begin{equation}
\label{Shifted_Nesterov_2:ass2}
\begin{split}
	\ShiftedProcess_n
=
	\Theta_n - \gamma_{n+1} \alpha_{n+1} \mathbf{m}_{n}
\end{split}
\end{equation}
\cfload.
Then it holds for all
	$n \in \N$
that
\begin{equation}
\label{Shifted_Nesterov_2:concl1}
\ShiftedProcess_0 = \xi, \qquad \mathbf{m}_0 = 0,
\end{equation}
\begin{equation}
\label{Shifted_Nesterov_2:concl2}
\mathbf{m}_n = \alpha_n \mathbf{m}_{n-1} + \grad(\ShiftedProcess_{n-1}),
\end{equation}
\begin{equation}
\label{Shifted_Nesterov_2:concl3}
\andq \ShiftedProcess_n = \ShiftedProcess_{n-1} - \gamma_{n+1} \alpha_{n+1} \mathbf{m}_n - \gamma_n \grad(\ShiftedProcess_{n-1}). 
\end{equation}
\end{athm}

\begin{aproof}
\Nobs that
\enum{
	\cref{eq:def:nesterov_2_1};
	\cref{eq:def:nesterov_2_2};
	\cref{eq:def:nesterov_2_3};
	\cref{Shifted_Nesterov_2:ass1}
}
\prove
that for all
	$n \in \N$
it holds that
\begin{equation}
\label{Shifted_Nesterov_2:eq3}
\begin{split}
	\Theta_n 
=
	\Theta_{n-1} - \gamma_n \mathbf{m}_n.
\end{split}
\end{equation}
\Moreover
\enum{
	\cref{Shifted_Nesterov_2:ass1};
	the fact that $\Theta_0 = \xi$;
	the assumption that $\mathbf{m}_0 = 0$;
}
\prove
that 
\begin{equation}
\label{Shifted_Nesterov_2:eq4.1}
\begin{split}
	\ShiftedProcess_0
=
	\Theta_0 - \gamma_1 \alpha_1 \mathbf{m}_0
=
	\xi.
\end{split}
\end{equation}
\Moreover 
\enum{
	\cref{Shifted_Nesterov_2:ass1};
	\cref{Shifted_Nesterov_2:ass2};
}
\prove
that for all
	$n \in \N$
it holds that
\begin{equation}
\label{Shifted_Nesterov_2:eq4.2}
\begin{split}
	\mathbf{m}_n 
&=
	\alpha_n \mathbf{m}_{n-1} + \grad(\Theta_{n-1} - \gamma_n \alpha_n \mathbf{m}_{n-1})\\
&=
	\alpha_n \mathbf{m}_{n-1} + \grad(\ShiftedProcess_{n-1}).
\end{split}
\end{equation}
\Moreover 
\enum{
	this;
	\cref{Shifted_Nesterov_2:ass1};
	\cref{Shifted_Nesterov_2:ass2};
	\cref{Shifted_Nesterov_2:eq3};
}
\prove
that for all
	$n \in \N$
it holds that
\begin{equation}
\label{Shifted_Nesterov_2:eq5}
\begin{split}
	\ShiftedProcess_n
&=
	\Theta_n - \gamma_{n+1} \alpha_{n+1} \mathbf{m}_n\\
&=
	\Theta_{n-1} - \gamma_n \mathbf{m}_n - \gamma_{n+1} \alpha_{n+1} \mathbf{m}_n\\
&=
	\Theta_{n-1} - \gamma_n (\alpha_n \mathbf{m}_{n-1} + \grad(\ShiftedProcess_{n-1})) - \gamma_{n+1} \alpha_{n+1} \mathbf{m}_n\\
&=
	\ShiftedProcess_{n-1} - \gamma_n \grad(\ShiftedProcess_{n-1}) - \gamma_{n+1} \alpha_{n+1} \mathbf{m}_n.
\end{split}
\end{equation}
Combining 
\enum{
	this;
	\cref{Shifted_Nesterov_2:eq4.1};
	\cref{Shifted_Nesterov_2:eq4.2}
}
establishes
\cref{Shifted_Nesterov_2:concl1,Shifted_Nesterov_2:concl2,Shifted_Nesterov_2:concl3}.
\end{aproof}

\defdetermNesterovAltTwo
\algDescrDetermNesterovAltTwo

\subsubsection{Shifted representation for the third version of Nesterov accelerated momentum optimization}

\cfclear
\begin{athm}{lemma}{Shifted_Nesterov_3}[Shifting the Nesterov accelerated \GD\ process (\third version)]
Let 
	$\defaultParamDim \in \N$, 
	$(\gamma_n)_{n \in \N} \subseteq [0,\infty)$,
	$(\alpha_n)_{n \in \N} \subseteq [0,\infty)$,
	$\xi \in \R^\defaultParamDim$,
   	$\defaultLossFunction \in C^1(\R^\defaultParamDim, \R)$,
let 
	$\Theta \colon \N_0 \to \R^\defaultParamDim$
be the Nesterov accelerated \GD\ process\cfadd{def:determ_nesterov_three} (\third version) for the objective function $\defaultLossFunction$ with learning rates $(\gamma_n)_{n \in \N}$,
momentum decay factors $(\alpha_n)_{n \in \N}$,
and initial value $\xi$,
let 
	$\mathbf{m} \colon \N_0 \to \R^\defaultParamDim$
satisfy for all
	$n \in \N$
that
	$\mathbf{m}_0 = 0$ 
and
\begin{equation}
\label{Shifted_Nesterov_3:ass1}
\begin{split}
	\mathbf{m}_n
=
	\alpha_n \mathbf{m}_{n-1} + (1 - \alpha_n) \gamma_n \grad(\Theta_{n-1} - \alpha_n \mathbf{m}_{n-1}),
\end{split}
\end{equation}
and let 
	$\ShiftedProcess \colon \N_0 \to \R^\defaultParamDim$
satisfy for all
	$n \in \N_0$
that
\begin{equation}
\label{Shifted_Nesterov_3:ass2}
\begin{split}
	\ShiftedProcess_n
=
	\Theta_n - \alpha_{n+1} \mathbf{m}_{n}
\end{split}
\end{equation}
\cfload.
Then it holds for all
	$n \in \N$
that
\begin{equation}
\label{Shifted_Nesterov_3:concl1}
\ShiftedProcess_0 = \xi, \qquad \mathbf{m}_0 = 0,
\end{equation}
\begin{equation}
\label{Shifted_Nesterov_3:concl2}
\mathbf{m}_n = \alpha_n \mathbf{m}_{n-1} + (1 - \alpha_n) \gamma_n \grad(\ShiftedProcess_{n-1}),
\end{equation}
\begin{equation}
\label{Shifted_Nesterov_3:concl3}
\andq \ShiftedProcess_n = \ShiftedProcess_{n-1} - \alpha_{n+1} \mathbf{m}_n - (1 - \alpha_{n}) \gamma_n \grad(\ShiftedProcess_{n-1}). 
\end{equation}
\end{athm}

\begin{aproof}
\Nobs that
\enum{
	\cref{eq:def:nesterov_3_1};
	\cref{eq:def:nesterov_3_2};
	\cref{eq:def:nesterov_3_3};
	\cref{Shifted_Nesterov_3:ass1}
}
\prove
that for all
	$n \in \N$
it holds that
\begin{equation}
\label{Shifted_Nesterov_3:eq3}
\begin{split}
	\Theta_n 
=
	\Theta_{n-1} - \mathbf{m}_n.
\end{split}
\end{equation}
\Moreover
\enum{
	\cref{Shifted_Nesterov_3:ass1};
	the fact that $\Theta_0 = \xi$;
	the assumption that $\mathbf{m}_0 = 0$;
}
\prove
that 
\begin{equation}
\label{Shifted_Nesterov_3:eq4.1}
\begin{split}
	\ShiftedProcess_0
=
	\Theta_0 - \alpha_1 \mathbf{m}_0
=
	\xi.
\end{split}
\end{equation}
\Moreover 
\enum{
	\cref{Shifted_Nesterov_3:ass1};
	\cref{Shifted_Nesterov_3:ass2};
}
\prove
that for all
	$n \in \N$
it holds that
\begin{equation}
\label{Shifted_Nesterov_3:eq4.2}
\begin{split}
	\mathbf{m}_n 
&=
	\alpha_n \mathbf{m}_{n-1} + (1 - \alpha_n) \gamma_n  \grad(\Theta_{n-1} - \alpha_n \mathbf{m}_{n-1})\\
&=
	\alpha_n \mathbf{m}_{n-1} + (1 - \alpha_n) \gamma_n  \grad(\ShiftedProcess_{n-1}).
\end{split}
\end{equation}
\Moreover 
\enum{
	this;
	\cref{Shifted_Nesterov_3:ass1};
	\cref{Shifted_Nesterov_3:ass2};
	\cref{Shifted_Nesterov_3:eq3};
}
\prove
that for all
	$n \in \N$
it holds that
\begin{equation}
\label{Shifted_Nesterov_3:eq5}
\begin{split}
	\ShiftedProcess_n
&=
	\Theta_n - \alpha_{n+1} \mathbf{m}_n\\
&=
	\Theta_{n-1} - \mathbf{m}_n - \alpha_{n+1} \mathbf{m}_n\\
&=
	\Theta_{n-1} - (\alpha_n \mathbf{m}_{n-1} + (1 - \alpha_n) \gamma_n \grad(\ShiftedProcess_{n-1})) - \alpha_{n+1} \mathbf{m}_n\\
&=
	\ShiftedProcess_{n-1} - (1 - \alpha_n) \gamma_n \grad(\ShiftedProcess_{n-1}) - \alpha_{n+1} \mathbf{m}_n.
\end{split}
\end{equation}
Combining 
\enum{
	this;
	\cref{Shifted_Nesterov_3:eq4.1};
	\cref{Shifted_Nesterov_3:eq4.2}
}
establishes
\cref{Shifted_Nesterov_3:concl1,Shifted_Nesterov_3:concl2,Shifted_Nesterov_3:concl3}.
\end{aproof}

\defdetermNesterovAltThree
\algDescrDetermNesterovAltThree

\subsubsection{Shifted representation for the fourth version of Nesterov accelerated GD optimization}

\cfclear
\begin{athm}{lemma}{Shifted_Nesterov_4}[Shifting the Nesterov accelerated \GD\ process (\fourth version)]
Let 
	$\defaultParamDim \in \N$, 
	$(\gamma_n)_{n \in \N} \subseteq [0,\infty)$,
	$(\alpha_n)_{n \in \N} \subseteq [0,\infty)$,
	$\xi \in \R^\defaultParamDim$,
   	$\defaultLossFunction \in C^1(\R^\defaultParamDim, \R)$,
let 
	$\Theta \colon \N_0 \to \R^\defaultParamDim$
be the Nesterov accelerated \GD\ process\cfadd{def:determ_nesterov_four} (\fourth version) for the objective function $\defaultLossFunction$ with learning rates $(\gamma_n)_{n \in \N}$,
momentum decay factors $(\alpha_n)_{n \in \N}$,
and initial value $\xi$,
let 
	$\mathbf{m} \colon \N_0 \to \R^\defaultParamDim$
satisfy for all
	$n \in \N$
that
	$\mathbf{m}_0 = 0$ 
and
\begin{equation}
\label{Shifted_Nesterov_4:ass1}
\begin{split}
	\mathbf{m}_n
=
	\alpha_n \mathbf{m}_{n-1} + \gamma_n \grad(\Theta_{n-1} - \alpha_n \mathbf{m}_{n-1}),
\end{split}
\end{equation}
and let 
	$\ShiftedProcess \colon \N_0 \to \R^\defaultParamDim$
satisfy for all
	$n \in \N_0$
that
\begin{equation}
\label{Shifted_Nesterov_4:ass2}
\begin{split}
	\ShiftedProcess_n
=
	\Theta_n - \alpha_{n+1} \mathbf{m}_{n}
\end{split}
\end{equation}
\cfload.
Then it holds for all
	$n \in \N$
that
\begin{equation}
\label{Shifted_Nesterov_4:concl1}
\ShiftedProcess_0 = \xi, \qquad \mathbf{m}_0 = 0,
\end{equation}
\begin{equation}
\label{Shifted_Nesterov_4:concl2}
\mathbf{m}_n = \alpha_n \mathbf{m}_{n-1} + \gamma_n  \grad(\ShiftedProcess_{n-1}),
\end{equation}
\begin{equation}
\label{Shifted_Nesterov_4:concl3}
\andq \ShiftedProcess_n = \ShiftedProcess_{n-1} - \alpha_{n+1} \mathbf{m}_n - \gamma_n \grad(\ShiftedProcess_{n-1}). 
\end{equation}
\end{athm}

\begin{aproof}
\Nobs that
\enum{
	\cref{eq:def:nesterov_4_1};
	\cref{eq:def:nesterov_4_2};
	\cref{eq:def:nesterov_4_3};
	\cref{Shifted_Nesterov_4:ass1}
}
\prove
that for all
	$n \in \N$
it holds that
\begin{equation}
\label{Shifted_Nesterov_4:eq3}
\begin{split}
	\Theta_n 
=
	\Theta_{n-1} - \mathbf{m}_n.
\end{split}
\end{equation}
\Moreover
\enum{
	\cref{Shifted_Nesterov_4:ass1};
	the fact that $\Theta_0 = \xi$;
	the assumption that $\mathbf{m}_0 = 0$;
}
\prove
that 
\begin{equation}
\label{Shifted_Nesterov_4:eq4.1}
\begin{split}
	\ShiftedProcess_0
=
	\Theta_0 - \alpha_1 \mathbf{m}_0
=
	\xi.
\end{split}
\end{equation}
\Moreover 
\enum{
	\cref{Shifted_Nesterov_4:ass1};
	\cref{Shifted_Nesterov_4:ass2};
}
\prove
that for all
	$n \in \N$
it holds that
\begin{equation}
\label{Shifted_Nesterov_4:eq4.2}
\begin{split}
	\mathbf{m}_n 
&=
	\alpha_n \mathbf{m}_{n-1} + \gamma_n  \grad(\Theta_{n-1} - \alpha_n \mathbf{m}_{n-1})\\
&=
	\alpha_n \mathbf{m}_{n-1} + \gamma_n  \grad(\ShiftedProcess_{n-1}).
\end{split}
\end{equation}
\Moreover 
\enum{
	this;
	\cref{Shifted_Nesterov_4:ass1};
	\cref{Shifted_Nesterov_4:ass2};
	\cref{Shifted_Nesterov_4:eq3};
}
\prove
that for all
	$n \in \N$
it holds that
\begin{equation}
\label{Shifted_Nesterov_4:eq5}
\begin{split}
	\ShiftedProcess_n
&=
	\Theta_n - \alpha_{n+1} \mathbf{m}_n\\
&=
	\Theta_{n-1} - \mathbf{m}_n -  \alpha_{n+1} \mathbf{m}_n\\
&=
	\Theta_{n-1} - (\alpha_n \mathbf{m}_{n-1} + \gamma_n \grad(\ShiftedProcess_{n-1})) - \alpha_{n+1} \mathbf{m}_n\\
&=
	\ShiftedProcess_{n-1} - \gamma_n  \grad(\ShiftedProcess_{n-1}) - \alpha_{n+1} \mathbf{m}_n.
\end{split}
\end{equation}
Combining 
\enum{
	this;
	\cref{Shifted_Nesterov_4:eq4.1};
	\cref{Shifted_Nesterov_4:eq4.2}
}
establishes
\cref{Shifted_Nesterov_4:concl1,Shifted_Nesterov_4:concl2,Shifted_Nesterov_4:concl3}.
\end{aproof}

\defdetermNesterovAltFour
\algDescrDetermNesterovAltFour

\subsubsection{Shifted representation for the bias-adjusted Nesterov accelerated GD optimization}

\cfclear
\begin{athm}{lemma}{Shifted_bias_Nesterov}[Shifting the bias-adjusted Nesterov accelerated \GD\ process]
Let 
	$\defaultParamDim \in \N$, 
	$(\gamma_n)_{n \in \N} \subseteq [0,\infty)$,
	$(\alpha_n)_{n \in \N} \subseteq [0,1)$,
	$\xi \in \R^\defaultParamDim$,
   	$\defaultLossFunction \in C^1(\R^\defaultParamDim, \R)$,
let 
	$\Theta \colon \N_0 \to \R^\defaultParamDim$
be the bias-adjusted Nesterov accelerated \GD\ process\cfadd{def:determ_nesterov_bias} for the objective function $\defaultLossFunction$ with learning rates $(\gamma_n)_{n \in \N}$,
momentum decay factors $(\alpha_n)_{n \in \N}$,
and initial value $\xi$,
let 
	$\mathbf{m} \colon \N_0 \to \R^\defaultParamDim$
satisfy for all
	$n \in \N$
that
	$\mathbf{m}_0 = 0$ 
and
\begin{equation}
\label{Shifted_bias_Nesterov_3:ass1}
\begin{split}
	\mathbf{m}_n
=
	\alpha_n \mathbf{m}_{n-1} + (1-\alpha_n)  \grad\pr*{\Theta_{n-1} - \frac{\gamma_n\alpha_n \mathbf{m}_{n-1}}{1- \prod_{l=1}^n\alpha_l}},
\end{split}
\end{equation}
and let 
	$\ShiftedProcess \colon \N_0 \to \R^\defaultParamDim$
satisfy for all
	$n \in \N_0$
that
\begin{equation}
\label{Shifted_bias_Nesterov_3:ass2}
\begin{split}
	\ShiftedProcess_n
=
	\Theta_n - \frac{\gamma_{n+1} \alpha_{n+1} \mathbf{m}_{n}}{1- \prod_{l=1}^{n+1}\alpha_l}
\end{split}
\end{equation}
\cfload.
Then it holds for all
	$n \in \N$
that
\begin{equation}
\label{Shifted_bias_Nesterov_3:concl1}
\ShiftedProcess_0 = \xi, \qquad \mathbf{m}_0 = 0,
\end{equation}
\begin{equation}
\label{Shifted_bias_Nesterov_3:concl2}
\mathbf{m}_n = \alpha_n \mathbf{m}_{n-1} + (1-\alpha_n)  \grad(\ShiftedProcess_{n-1}),
\qand
\end{equation}
\begin{equation}
\label{Shifted_bias_Nesterov_3:concl3}
\ShiftedProcess_n = \ShiftedProcess_{n-1} - \frac{\gamma_{n+1} \alpha_{n+1} \mathbf{m}_n}{1- \prod_{l=1}^{n+1}\alpha_l} - \frac{\gamma_n (1-\alpha_n) \grad(\ShiftedProcess_{n-1})}{1- \prod_{l=1}^{n}\alpha_l}.
\end{equation}
\end{athm}

\begin{aproof}
\Nobs that
\enum{
	\cref{eq:def:nesterov_bias_1};
	\cref{eq:def:nesterov_bias_2};
	\cref{eq:def:nesterov_bias_3};
	\cref{Shifted_bias_Nesterov_3:ass1}
}
\prove
that for all
	$n \in \N$
it holds that
\begin{equation}
\label{Shifted_bias_Nesterov_3:eq3}
\begin{split}
	\Theta_n 
=
	\Theta_{n-1} - \frac{\gamma_n \mathbf{m}_n}{1- \prod_{l=1}^{n}\alpha_l}.
\end{split}
\end{equation}
\Moreover
\enum{
	\cref{Shifted_bias_Nesterov_3:ass1};
	the fact that $\Theta_0 = \xi$;
	the assumption that $\mathbf{m}_0 = 0$;
}
\prove
that 
\begin{equation}
\label{Shifted_bias_Nesterov_3:eq4.1}
\begin{split}
	\ShiftedProcess_0
=
	\Theta_0 - \alpha_1 \mathbf{m}_0
=
	\xi.
\end{split}
\end{equation}
\Moreover 
\enum{
	\cref{Shifted_bias_Nesterov_3:ass1};
	\cref{Shifted_bias_Nesterov_3:ass2};
}
\prove
that for all
	$n \in \N$
it holds that
\begin{equation}
\label{Shifted_bias_Nesterov_3:eq4.2}
\begin{split}
	\mathbf{m}_n 
&=
	\alpha_n \mathbf{m}_{n-1} + (1 - \alpha_n) \grad\pr[\Big]{\Theta_{n-1} - \frac{\gamma_n \alpha_n \mathbf{m}_{n-1}}{1- \prod_{l=1}^n\alpha_l}}\\
&=
	\alpha_n \mathbf{m}_{n-1} + (1 - \alpha_n) \grad(\ShiftedProcess_{n-1}).
\end{split}
\end{equation}
\Moreover 
\enum{
	this;
	\cref{Shifted_bias_Nesterov_3:ass1};
	\cref{Shifted_bias_Nesterov_3:ass2};
	\cref{Shifted_bias_Nesterov_3:eq3};
}
\prove
that for all
	$n \in \N$
it holds that
\begin{equation}
\label{Shifted_bias_Nesterov_3:eq5}
\begin{split}
	\ShiftedProcess_n
&=
	\Theta_n - \frac{\gamma_{n+1} \alpha_{n+1} \mathbf{m}_n}{1- \prod_{l=1}^{n+1}\alpha_l}\\
&=
	\Theta_{n-1} - \frac{\gamma_n \mathbf{m}_n}{1- \prod_{l=1}^{n}\alpha_l} - \frac{\gamma_{n+1} \alpha_{n+1} \mathbf{m}_n}{1- \prod_{l=1}^{n+1}\alpha_l}\\
&=
	\Theta_{n-1} - \frac{\gamma_n (\alpha_n\mathbf{m}_{n-1} + (1-\alpha_n) \grad(\ShiftedProcess_{n-1}))}{1- \prod_{l=1}^n\alpha_l} - \frac{\gamma_{n+1} \alpha_{n+1} \mathbf{m}_n}{1- \prod_{l=1}^{n+1}\alpha_l}\\
&=
	\ShiftedProcess_{n-1} - \frac{\gamma_n (1-\alpha_n) \grad(\ShiftedProcess_{n-1})}{1- \prod_{l=1}^{n}\alpha_l} - \frac{\gamma_{n+1} \alpha_{n+1} \mathbf{m}_n}{1- \prod_{l=1}^{n+1}\alpha_l}.
\end{split}
\end{equation}
Combining 
\enum{
	this;
	\cref{Shifted_bias_Nesterov_3:eq4.1};
	\cref{Shifted_bias_Nesterov_3:eq4.2}
}
establishes
\cref{Shifted_bias_Nesterov_3:concl1,Shifted_bias_Nesterov_3:concl2,Shifted_bias_Nesterov_3:concl3}.
\end{aproof}

\defdetermNesterovAltBias
\algDescrDetermNesterovAltBias

\subsection{Simplified Nesterov accelerated momentum optimization}
\label{sect:determ_nesterov_simple}

For reasons of algorithmic simplicity, in several deep learning libraries including {\sc PyTorch} (see \cite{PytorchNAG} and cf., \eg, \cite[Section 3.5]{bengio2013advances}) optimization with Nesterov momentum is not implemented such that it precisely corresponds to any of the definitions presented above.
Rather, an alternative definition for Nesterov accelerated \GD\ optimization is used, which we present in \cref{def:determ_nesterov_simple} below.
Roughly speaking, the simplified version of Nesterov accelerated \GD\ optimization in \cref{def:determ_nesterov_simple} employed by {\sc PyTorch} 
is obtained by reducing some indices in the update rule of the shifted Nesterov accelerated \GD\ optimization method (\second version) in \cref{def:determ_nesterov_shifted_two} so that for each update step only one learning rate and one momentum decay factor are used.

\todoc{Maybe introduce a variable $g$ in pseudocode when gradient is used more than once.}

\defdetermNesterovSimple
\algDescrDetermNesterovSimple

\section{Adaptive gradient (Adagrad) optimization}
\label{sect:determ_adagrad}

In this section we review the \Adagrad\ \GD\ optimization method.
Roughly speaking, the idea of the \Adagrad\ \GD\ optimization method is to modify the plain-vanilla \GD\ optimization method by adapting the learning rates separately for every component of the optimization process. 
\Adagrad\ was first presented in Duchi et al.~\cite{DuchiHazanSinger11} in the context of stochastic optimization.
For pedagogical purposes we present in this section a deterministic version of \Adagrad\ optimization %
and we refer to \cref{sect:adagrad} below for the original stochastic version of \Adagrad\ optimization.

\defdetermAdagrad

\begin{adef}{def:componentwise_operations}[Componentwise operations]
For every 
	$d \in \N$,
	$c \in \R$,
	$v = (v_1, \ldots, v_d) \in \R^d$
we denote by 
	$c + v$
and
	$v + c$
the vectors which satisfy
\begin{equation}
\begin{split} 
	c + v
=
	v + c
=
	\pr*{
		c + v_1, c + v_2, \ldots, c + v_d
	},
\end{split}
\end{equation}
for every 
	$d \in \N$,
	$p \in \R$,
	$v = (v_1, \ldots, v_d) \in [0,\infty)^d$
	with 
	$(p > 0) \lor (v \in (0,\infty)^d)$
we denote by
	$v^p$
the vector which satisfies
\begin{equation}
\begin{split} 
	v^p
=
	\pr*{
		\pr{v_1}^p, \pr{v_2}^p, \ldots, \pr{v_d}^p
	},
\end{split}
\end{equation}
for every 
	$d \in \N$,
	$v = (v_1, \ldots, v_d) \in \R^d$
we denote by
	$\abs{v}$
the vector which satisfies
\begin{equation}
\begin{split} 
	\abs{v}
=
	\pr*{
		\abs{v_1}, \abs{v_2}, \ldots, \abs{v_d}
	},
\end{split}
\end{equation}
for every 
	$d \in \N$,
	$v = (v_1, \ldots, v_d)$, 
	$w = (w_1, \ldots, w_d) \in \R^d$
we denote by
	$v \compMulti w \in \R^d$
the vector which satisfies
\begin{equation}
\begin{split} 
	v \compMulti w
=
	(v_1 w_1, v_2 w_2, \ldots, v_d w_d),
\end{split}
\end{equation}
for every 
	$d \in \N$,
	$v = (v_1, \ldots, v_d) \in \R^d$, 
	$w = (w_1, \ldots, w_d) \in (\R\backslash\{0\})^d$
we denote by
	$\frac{v}{w} \in \R^d$
the vector which satisfies
\begin{equation}
\begin{split} 
	\frac{v}{w}
=
	\pr*{
		\tfrac{v_1}{w_1}, \tfrac{v_2}{w_2}, \ldots, \tfrac{v_d}{w_d}
	},
\end{split}
\end{equation}
and
for every 
	$d \in \N$,
	$v = (v_1, \ldots, v_d)$, 
	$w = (w_1, \ldots, w_d) \in \R^d$
we denote by
	$\max \cu{v, w} \in \R^d$
the vector which satisfies
\begin{equation}
\begin{split} 
	\max\{v, w\}
=
	\pr*{
		\max\cu{{v_1},{w_1}}, \max\cu{{v_2},{w_2}}, \ldots, \max\cu{{v_d},{w_d}}
	}.
\end{split}
\end{equation}
\end{adef}

\algDescrDetermAdagrad

\section{Root mean square propagation (RMSprop) optimization}
\label{sect:determ_RMSprop}

In this section we review the \RMSprop\ \GD\ optimization method.
Roughly speaking, the \RMSprop\ \GD\ optimization method is a modification of the \Adagrad\ \GD\ optimization method where the sum over the squares of previous partial derivatives of the objective function (cf.\ \eqref{def:determ_adagrad:eq1} in \cref{def:determ_adagrad}) 
is replaced by an exponentially decaying average over the squares of previous partial derivatives of the objective function 
(cf.\ \cref{def:determ_RMSprop:eq1,def:determ_RMSprop:eq2}  in \cref{def:determ_RMSprop}). 
\RMSprop\ optimization was introduced by Geoffrey Hinton in his coursera class on \emph{Neural Networks for Machine Learning}
(see 
Hinton et al.~\cite{HintonSlides})
in the context of stochastic optimization.
As in the case of \Adagrad\ optimization, we present for pedagogical purposes first a deterministic version of \RMSprop\ optimization in this section and we refer to \cref{sect:RMSprop} below for the original stochastic version of \RMSprop\ optimization.

\defdetermRMSprop
\algDescrDetermRMSprop

\todoc{We should stop referencing tensorflow.}

\begin{athm}{remark}{RMSprop_rem}
In Hinton et al.~\cite{HintonSlides}
it is proposed to choose
$
0.9 = \beta_1 = \beta_2 = \ldots
$
as default values for the second moment decay factors $ (\beta_n )_{ n \in \N } \subseteq [0,1] $ in \cref{def:rmsprop}.

Moreover, we note that in Hinton et al.~\cite{HintonSlides} the regularizing factor  $\varepsilon$ is omitted in the definition of \RMSprop. We chose the include the regularizing factor $\varepsilon$ in the definition of \RMSprop\ in \cref{def:determ_RMSprop} as it is usually implemented in machine learning libraries (cf., e.g., \cite{pytorchRMSprop}).

\end{athm}

\subsection
{Representations of the mean square terms in RMSprop}

\cfclear
\begingroup
\providecommand{\d}{}
\renewcommand{\d}{\defaultParamDim}
\providecommand{\f}{}
\renewcommand{\f}{\defaultLossFunction}
\providecommand{\g}{}
\renewcommand{\g}{\defaultGradientFunction}
\begin{lemma}[On a representation of the second order terms in \RMSprop]
\label{explicit_RMSprop}
Let 
	$\d \in \N$, 
	$(\gamma_n)_{n \in \N} \subseteq [0,\infty)$, 
	$(\beta_n)_{n \in \N} \subseteq [0,1]$,
	$(b_{n, k})_{(n, k) \in ( \N_0 )^2} \subseteq \R$, 
	$\varepsilon \in (0,\infty)$,
	$\xi \in \R^\d$
satisfy for all
	$n \in \N$,
	$k \in \{0, 1, \ldots, n-1\}$
that
\begin{equation}
\label{explicit_RMSprop:ass1}
\begin{split} 
	b_{n, k}
=
	(1-\beta_{k+1})
	\br*{
		\prod_{l = k + 2}^n
			\beta_l
	},
\end{split}
\end{equation}
let $\f \in C^1(\R^\d,\R)$, $\g = (\g_1, \ldots, \g_{\d}) \in C(\R^\d,\R^\d)$
satisfy for all 
	$\theta\in\R^\d$
that
\begin{equation}
\g(\theta) = (\nabla \f)(\theta), 
\end{equation}
and let 
$
  \Theta = (\Theta^{(1)}, \ldots, \Theta^{(\d)})
  \colon \N_0 \to \R^{ \fd }
$ be the \RMSprop\ \GD\ process\cfadd{def:determ_RMSprop} for the 
objective function $\f$ with 
learning rates $(\gamma_n)_{n \in \N}$, second moment decay factors $(\beta_n)_{n \in \N}$, regularizing factor $\varepsilon$, and initial value $\xi$
\cfload.
Then 
\begin{enumerate}[label=(\roman *)]
\item \label{explicit_RMSprop:item1}
it holds for all 
	$n \in \N$,
	$k \in \{0, 1, \ldots, n-1\}$
that
$
	0 \leq b_{n, k} \leq 1
$,

\item \label{explicit_RMSprop:item2}
it holds for all 
	$n\in\N$ 
that
\begin{equation}
	\sum_{k = 0}^{n-1} b_{n, k}
=
	1 - \prod_{k = 1}^n \beta_k,
\end{equation}
and

\item \label{explicit_RMSprop:item3}
it holds for all 
	$n\in\N$,
	$i \in \{1, 2, \ldots, \d\}$
that
\begin{equation}
	\Theta_n^{(i)}
= 
	\Theta_{n-1}^{(i)} 
	- 
	\gamma_n 
	\br*{
		\varepsilon 
		+ 
		\br*{
		\sum_{k = 0}^{n-1}
			b_{n, k}
			\abs{\g_i(\Theta_k)}^2
		}^{ 1 / 2 }
	}^{ - 1 } 
	\g_i( \Theta_{ n - 1 } ) 
	. 
\end{equation}
\end{enumerate}
\end{lemma}

\begin{proof}[Proof of \cref{explicit_RMSprop}]
Throughout this proof, let 
	$\mathbb{M} = (\mathbb{M}^{(1)},\ldots,\mathbb{M}^{(\d)}) \colon \N_0 \to \R^\d$ 
satisfy for all 
	$n \in \N$, 
	$i \in \{1,2,\ldots,\d\}$ 
that 
$
	\mathbb{M}_0^{ (i) }
=
	0
$ 
and
\begin{equation}
\label{explicit_RMSprop:setting1}
	\mathbb{M}_n^{ (i) } 
= 
	\beta_n  \mathbb{M}_{ n - 1 }^{ (i) } 
	+ 
	( 1 - \beta_n ) 
	\abs*{ \g_i( \Theta_{ n - 1 } ) }^2.
\end{equation}
\Nobs that
\enum{
	\eqref{explicit_RMSprop:ass1}
}[imply]
\cref{explicit_RMSprop:item1}.
\Moreover
\enum{
	\eqref{explicit_RMSprop:ass1};
	\eqref{explicit_RMSprop:setting1};
	\cref{explicit_momentum}
}[assure]
that for all 
	$n \in \N$,
	$i \in \{1, 2, \ldots, \d\}$
it holds that 
\begin{equation}
\label{explicit_RMSprop:eq1}
	\mathbb{M}_n^{ (i) } 
=
	\sum_{k = 0}^{n-1} b_{n, k} \abs{\g_i(\Theta_{k})}^2
\qandq
	\sum_{k = 0}^{n-1} b_{n, k}
=
	1 - \prod_{k = 1}^n \beta_k.
\end{equation}
This proves \cref{explicit_RMSprop:item2}.
\Moreover
\enum{
	\eqref{def:determ_RMSprop:eq1};
	\eqref{def:determ_RMSprop:eq2};
	\eqref{explicit_RMSprop:setting1};
	\eqref{explicit_RMSprop:eq1}
}[demonstrate]
that for all
	$n \in \N$,
	$i \in \{1, 2, \ldots, \d\}$
it holds that
\begin{equation}
\label{explicit_RMSprop:eq2}
\begin{split} 
	\Theta_n^{(i)}
&= 
	\Theta_{n-1}^{(i)} 
	- 
	\gamma_n 
	\br*{
		\varepsilon 
		+ 
		\br*{
		\mathbb{M}_n^{ (i) } 
		}^{ 1 / 2 }
	}^{ - 1 } 
	\g_i( \Theta_{ n - 1 } ) \\
&=
	\Theta_{n-1}^{(i)} 
	- 
	\gamma_n 
	\br*{
		\varepsilon 
		+ 
		\br*{
			\sum_{k = 0}^{n-1}
				b_{n, k}
				\abs{\g_i(\Theta_k)}^2
		}^{ 1 / 2 }
	}^{ - 1 } 
	\g_i( \Theta_{ n - 1 } ) 
	. 
\end{split}
\end{equation}
This establishes \cref{explicit_RMSprop:item3}. 
The proof of \cref{explicit_RMSprop} is thus complete.
\end{proof}
\endgroup

\subsection{Bias-adjusted RMSprop optimization}
\label{sect:determ_bias_adj_RMSprop}

\defdetermRMSpropBias
\algDescrDetermRMSpropBias

\cfclear
\begingroup
\providecommand{\d}{}
\renewcommand{\d}{\defaultParamDim}
\providecommand{\f}{}
\renewcommand{\f}{\defaultLossFunction}
\providecommand{\g}{}
\renewcommand{\g}{\defaultGradientFunction}
\begin{lemma}[On a representation of the second order terms in bias-adjusted \RMSprop]
\label{explicit_bias_RMSprop}
Let 
	$\d \in \N$, 
	$(\gamma_n)_{n \in \N} \subseteq [0,\infty)$, 
	$(\beta_n)_{n \in \N} \subseteq [0,1)$,
	$(b_{n, k})_{(n, k) \in ( \N_0 )^2} \subseteq \R$, 
	$\varepsilon \in (0,\infty)$,
	$\xi \in \R^\d$
satisfy for all
	$n \in \N$,
	$k \in \{0, 1, \ldots, n-1\}$
that
\begin{equation}
\label{explicit_bias_RMSprop:ass1}
\begin{split} 
	b_{n, k}
=
	\frac{
		(1-\beta_{k+1})
		\br*{
			\prod_{l = k + 2}^n
				\beta_l
		}
	}{
		1 - \prod_{l = 1}^n \beta_l
	},
\end{split}
\end{equation}
let 
	$\f \in C^1(\R^\d,\R)$, $\g = (\g_1, \ldots, \g_{\d}) \in C(\R^\d,\R^\d)$
satisfy for all 
	$\theta\in\R^\d$
that
\begin{equation}
\g(\theta) = (\nabla \f)(\theta),
\end{equation}
and let 
$
  \Theta = (\Theta^{(1)}, \ldots, \Theta^{(\d)})
  \colon \N_0 \to \R^{ \fd }
$ be the bias-adjusted \RMSprop\ \GD\ process\cfadd{def:determ_RMSprop_bias} for the objective function $\f$ with 
learning rates $(\gamma_n)_{n \in \N}$, second moment decay factors $(\beta_n)_{n \in \N}$, regularizing factor $\varepsilon$, and initial value $\xi$
\cfload.
Then 
\begin{enumerate}[label=(\roman *)]
\item \label{explicit_bias_RMSprop:item1}
it holds for all 
	$n \in \N$,
	$k \in \{0, 1, \ldots, n-1\}$
that
$
	0 \leq b_{n, k} \leq 1
$,

\item \label{explicit_bias_RMSprop:item2}
it holds for all 
	$n\in\N$ 
that
\begin{equation}
	\sum_{k = 0}^{n-1} b_{n, k}
=
	1,
\end{equation}
and

\item \label{explicit_bias_RMSprop:item3}
it holds for all 
	$n\in\N$,
	$i \in \{1, 2, \ldots, \d\}$
that
\begin{equation}
	\Theta_n^{(i)}
= 
	\Theta_{n-1}^{(i)} 
	- 
	\gamma_n 
	\br*{
		\varepsilon 
		+ 
		\br*{
		\sum_{k = 0}^{n-1}
			b_{n, k}
			\abs{\g_i(\Theta_k)}^2
        }^{ 1 / 2 }
	}^{ - 1 } 
	\g_i( \Theta_{ n - 1 } ) 
	. 
\end{equation}
\end{enumerate}
\end{lemma}

\begin{proof}[Proof of \cref{explicit_bias_RMSprop}]
Throughout this proof, let 
	$\mathbb{M} = (\mathbb{M}^{(1)},\ldots,\mathbb{M}^{(\d)}) \colon \N_0 \to \R^\d$ 
satisfy for all 
	$n \in \N$, 
	$i \in \{1,2,\ldots,\d\}$ 
that 
$
	\mathbb{M}_0^{ (i) }
=
	0
$ 
and
\begin{equation}
\label{explicit_bias_RMSprop:setting1}
	\mathbb{M}_n^{ (i) } 
= 
	\beta_n \mathbb{M}_{ n - 1 }^{ (i) } 
	+ 
	( 1 - \beta_n ) 
	\abs*{ \g_i( \Theta_{ n - 1 } ) }^2
\end{equation}
and let 
	$ ( B_{n, k} )_{ (n, k) \in (\N_0)^2 } \subseteq \R$
satisfy for all
	$n \in \N$,
	$k \in \{0, 1, \ldots, n-1\}$
that
\begin{equation}
\label{explicit_bias_RMSprop:setting2}
\begin{split} 
	B_{n, k}
=
	(1-\beta_{k+1})
	\br*{
		\prod_{l = k + 2}^n
			\beta_l
	}
	.
\end{split}
\end{equation}
\Nobs that
\enum{
	\eqref{explicit_bias_RMSprop:ass1}	
}[imply]
\cref{explicit_bias_RMSprop:item1}.
\Nobs that 
\enum{
	\eqref{explicit_bias_RMSprop:ass1};
	\eqref{explicit_bias_RMSprop:setting1};
	\eqref{explicit_bias_RMSprop:setting2};
	\cref{explicit_momentum}
}[assure]
that for all 
	$n \in \N$,
	$i \in \{1,2,\ldots,\d\}$ 
it holds that 
\begin{equation}
\label{explicit_bias_RMSprop:eq1}
	\mathbb{M}_n^{ (i) } 
=
	\sum_{k = 0}^{n-1} B_{n, k} \abs{\g_i(\Theta_{k})}^2
\qandq
	\sum_{k = 0}^{n-1} b_{n, k}
=
	\frac{
		\sum_{k = 0}^{n-1} B_{n, k}
	}{
		1 - \prod_{l = 1}^n \beta_l
	}
=
	\frac{
		1 - \prod_{k = 1}^n \beta_k
	}{
		1 - \prod_{k = 1}^n \beta_k
	}
=
	1 .
\end{equation}
This proves \cref{explicit_bias_RMSprop:item2}. 
\Nobs that
\enum{
	\eqref{determ_RMSprop_bias:eq1};
	\eqref{determ_RMSprop_bias:eq2};
	\eqref{explicit_bias_RMSprop:setting1};
	\eqref{explicit_bias_RMSprop:eq1}
}[demonstrate]
that for all
$ n \in \N $, $ i \in \{1, 2, \dots, \d\} $
it holds that
\begin{equation}
\label{explicit_bias_RMSprop:eq2}
\begin{split} 
  \Theta_n^{(i)}
&= 
  \Theta_{n-1}^{(i)} 
  - 
  \gamma_n 
  \br*{\textstyle
    \varepsilon 
    + 
    \br*{
      \frac{ 
        \mathbb{M}_n^{ (i) } 
      }{ 
        1 - \prod_{k = 1}^n \beta_k
      }
    }^{ 1 / 2 }
  }^{ - 1 } 
  \g_i( \Theta_{ n - 1 } ) 
\\
&
=
	\Theta_{n-1}^{(i)} 
	- 
	\gamma_n 
	\br*{\textstyle
		\varepsilon 
		+ 
		\br*{
		\frac{ 
			\sum_{k = 0}^{n-1} B_{n, k} \abs{\g_i(\Theta_k)}^2
		}{ 
			1 - \prod_{k = 1}^n \beta_k
		}
		}^{ 1 / 2 }
	}^{ - 1 } 
	\g_i( \Theta_{ n - 1 } ) 
\\
&
=
	\Theta_{n-1}^{(i)} 
	-
	\gamma_n 
	\br*{
		\varepsilon 
		+ 
		\br*{
		\sum_{k = 0}^{n-1} \textstyle
		\pr*{
			\frac{ 
				(1-\beta_{k+1})
				\br*{
					\prod_{l = k + 2}^n
						\beta_l
				}
			}{ 
				1 - \prod_{l = 1}^n \beta_l
			}
		}
		\abs{\g_i(\Theta_k)}^2
		}^{ 1 / 2 }
	}^{ - 1 } 
	\g_i( \Theta_{ n - 1 } ) 
\\
&
=
  \Theta_{n-1}^{(i)} 
  - 
  \gamma_n 
  \br*{
    \varepsilon 
    + 
    \br*{
      \sum_{k = 0}^{n-1}
      b_{n, k}
      \abs{\g_i(\Theta_k)}^2
    }^{ 1 / 2 }
  }^{ \! - 1 } 
  \g_i( \Theta_{ n - 1 } ) 
  . 
\end{split}
\end{equation}
This establishes \cref{explicit_bias_RMSprop:item3}. 
The proof of \cref{explicit_bias_RMSprop} is thus complete.
\end{proof}
\endgroup

\section{Adadelta optimization}
\label{sect:determ_adadelta}

The Adadelta \GD\ optimization method reviewed in this section is an extension of the \RMSprop\ \GD\ optimization method.
Like the \RMSprop\ \GD\ optimization method, 
the Adadelta \GD\ optimization method adapts the learning rates
for every component of the optimization process separately. 
To do this, the Adadelta \GD\ optimization method 
uses two exponentially decaying averages: 
one over the squares of the past partial derivatives of the objective function as does the \RMSprop\ \GD\ optimization method (cf.\ \cref{def:determ_adadelta:eq2} below)
and 
another one over the squares of the past increments (cf.\ \cref{def:determ_adadelta:eq4} below).
As in the case of \Adagrad\ and \RMSprop\ optimization, 
Adadelta optimization was introduced in a stochastic setting (see Zeiler~\cite{Zeiler12}), 
but for pedagogical purposes we present in this section a deterministic version of Adadelta optimization.
We refer to \cref{sect:adadelta} below for the original stochastic version of Adadelta optimization.

\defdetermAdadelta
\algDescrDetermAdadelta

\section{Adaptive moment estimation (Adam) optimization}
\label{sect:determ_adam}

In this section we introduce the \Adam\ \GD\ optimization method 
(see Kingma \& Ba~\cite{KingmaBa14}). 
Roughly speaking, the \Adam\ \GD\ optimization method can be viewed as a combination of 
the bias-adjusted momentum \GD\ optimization method (see \cref{sect:determ_momentum_bias}) and
the bias-adjusted \RMSprop\ \GD\ optimization method (see \cref{sect:determ_bias_adj_RMSprop}).
As in the case of previously considered optimization methods,
\Adam\ optimization was introduced in a stochastic setting in Kingma \& Ba~\cite{KingmaBa14}, 
but for pedagogical purposes we present in this section a deterministic version of \Adam\ optimization.
We refer to \cref{sect:adam} below for the original stochastic version of \Adam\ optimization. 

\defdetermAdam
\algDescrDetermAdam

\subsection{Adamax optimization}
\label{sect:determ_adamax}

In this section we consider the deterministic Adamax \GD\ optimization method which was introduced together with the \Adam\ \GD\ optimization method in Kingma \& Ba~\cite{KingmaBa14}.
We refer to \cref{sec:adamax} below for the original stochastic version of Adamax optimization.

\defdetermAdamax
\algDescrDetermAdamax

\todoc{Somewhat unclear where $\varepsilon$ should appear. Probably does not matter much in practice.
In original article it does not appear and in pytorch it is added to the gradient. In tensorflow they do what we do here: ARNULF}

\section{Nesterov accelerated adaptive moment estimation (Nadam) optimization}
\label{sect:determ_nadam}

In this section we review the \Nadam\ \GD\ optimization method (cf.\ Dozat~\cite{Dozat16,Dozat15}).
Roughly speaking, the \Nadam\ \GD\ optimization method can be viewed as a combination of the shifted bias-adjusted Nesterov \GD\ optimization method in \cref{def:determ_nesterov_shifted_bias} and the bias-adjusted \RMSprop\ \GD\ optimization method in \cref{def:determ_RMSprop_bias}.
Alternatively, it can be seen as adding Nesterov acceleration to the \Adam\ \GD\ optimization method in \cref{def:determ_adam}.
As in the case of previously considered optimization methods,
the \Nadam\ \GD\ optimization method was introduced in a stochastic setting in Dozat~\cite{Dozat16,Dozat15},
but for pedagogical purposes we present in this section a deterministic version of \Nadam\ optimization.
We refer to \cref{sect:nadam} below for the original stochastic version of \Nadam\ optimization.

\defdetermNadam
\algDescrDetermNadam

\subsection{Simplified Nadam optimization}
\label{sect:determ_simplified_nadam}

For reasons of algorithmic simplicity, in Dozat~\cite{Dozat16,Dozat15} a simplified version of the \Nadam\ \GD\ optimization method has been proposed.
Roughly speaking, the simplified version presented in \cref{def:determ_simplified_nadam} below is obtained by reducing the index of the second learning rate in the update step of the \Nadam\ \GD\ optimization method in \cref{def:determ_nadam} above to ensure that for each update step only one learning rate is used.

\defdetermSimplifiedNadam
\algDescrDetermSimplifiedNadam

\subsection{Nadamax optimization}

In this section we consider the deterministic Nadamax \GD\ optimization method which was introduced together with the \Nadam\ \GD\ optimization method in Dozat~\cite{Dozat16,Dozat15}.
We refer to \cref{sec:nadamax} below for the original stochastic version of Nadamax optimization.

\defdetermNadamax
\algDescrDetermNadamax

\section{Adam with decoupled weight decay (AdamW) optimization }
\label{sect:determ_adamW}

\todoc{Cite this somewhere: https://arxiv.org/abs/2010.05627}

In this section we introduce the \AdamW\ \GD\ optimization method
(see Loshchilov \& Hutter~\cite{Loshchilov2019}).
Roughly speaking, the \AdamW\ \GD\ optimization method can be viewed as modification of the \Adam\ \GD\ optimization method in \cref{sect:determ_adam} with a weight decay term added to the update step.
As in the case of previously considered optimization methods,
the \AdamW\ \GD\ optimization method was introduced in a stochastic setting in Loshchilov \& Hutter~\cite{Loshchilov2019},
but for pedagogical purposes we present in this section a deterministic version of \AdamW\ optimization.
We refer to \cref{sect:adamW} below for the original stochastic version of \AdamW\ optimization.

\defdetermAdamW
\algDescrDetermAdamW

\subsection{Adam with $L^2$-regularization optimization}
\label{sect:determ_adamLtwo}

As an alternative way to regularize the parameters in the \Adam\ \GD\ optimization method, 
in Loshchilov \& Hutter~\cite{Loshchilov2019} it was also suggested to add a $L^2$-regularization term to the objective function.
This results in the \Adam\ \GD\ optimization method with $L^2$-regularization in \cref{def:determ_adamLtwo} below.
We refer to \cref{sec:adamLtwo} below for the original stochastic version of \Adam\ \GD\ optimization with $L^2$-regularization.

\defdetermAdamLtwo
\algDescrDetermAdamLtwo

\section{Shampoo optimization}
\label{sect:determ_shampoo}

In this section we introduce the Shampoo \GD\ optimization method (cf.\ Gupta et al.~\cite{Gupta2018}).
The Shampoo \GD\ optimization method was introduced in Gupta et al.~\cite{Gupta2018} for loss functions which are defined on spaces of multidimensional matrices (sometimes called tensors).
However, for simplicity we present in this section the Shampoo \GD\ optimization method only in the case of loss functions which are defined on spaces of (two-dimensional) matrices (cf.\ \cref{def:determ_Shampoo} below).
Roughly speaking, the Shampoo \GD\ optimization method aims to apply suitable preconditioners to transform the gradient of the loss function before doing a step in the direction of the negative gradient.
The use of preconditioners motivated the authors in Gupta et al.~\cite{Gupta2018} to name their method \emph{Shampoo}.
We refer to \cref{sect:shampoo} below for the stochastic version of Shampoo optimization.

To ensure that the preconditioners used in the definition of the Shampoo \GD\ optimization method in \cref{def:determ_Shampoo} are well-defined, we first show in \cref{lemma:sums_of_positive_definite,lemma:sums_of_positive_definite_2} below that sums of symmetric positive definite matrices remain symmetric positive definite and thus their roots are well-defined.

\newcommand{\symmetricPositiveDefiniteMatrix}{\cfadd{def:symmetric_positive_definite_matrix}symmetric positive definite matrix}
\newcommand{\symmetricPositiveDefiniteMatrices}{\cfadd{def:symmetric_positive_definite_matrix}symmetric positive definite matrices}

\begin{adef}{def:symmetric_positive_definite_matrix}[Symmetric positive definite matrix]
Let 
$d \in \N$,
$A \in \R^{d \times d}$.
Then we say that $A$ is a \symmetricPositiveDefiniteMatrix\ if and only if it holds for all 
	$v \in \R^d \backslash \{0\}$ 
that\cfclear
\begin{equation}
A^\ast = A 
\qandq
\scp{A v, v} > 0
\end{equation}
\cfload.
\end{adef}

\cfclear
\begingroup
\begin{athm}{lemma}{lemma:sums_of_positive_definite}%
	Let 
		$n, m \in \N$,
	let 
		$A \in \R^{n \times n}$ 
	be a \symmetricPositiveDefiniteMatrix,
	and let 
		$B \in \R^{n \times m}$
	\cfload.
	Then
		$A + B^\ast B$
	is a \symmetricPositiveDefiniteMatrix.
\end{athm}
\begin{aproof}
\argument{
	the assumption that $A$ is a \symmetricPositiveDefiniteMatrix;
}
{
	that for all $v \in \R^n \backslash \{0\}$ it holds that
	\begin{equation}
		\llabel{eq1}
		\scp{
			(A + B^\ast B) v, v
		}
		=
		\scp{A v, v} + \scp{B^\ast B v, v}
		=
		\scp{A v, v} + \scp{B v, B v}
		=
		\scp{A v, v} + \pnorm2{B v}
		>
		0
	\end{equation}
}
\argument{
	the fact that $A$ is a \symmetricPositiveDefiniteMatrix;
}
{
	that 
	\begin{equation}
	\llabel{eq2}
	\begin{split} 
		(A + B^\ast B)^\ast
		=
		A^\ast + (B^\ast B)^\ast
		=
		A + B^\ast B.
	\end{split}
	\end{equation}
}
\argument{
	\lref{eq1};
	\lref{eq2};
}
{
	that $A + B^\ast B$ is a \symmetricPositiveDefiniteMatrix.
}
\end{aproof}
\endgroup

\cfclear
\begingroup
\providecommandordefault{\n}{n}
\providecommandordefault{\m}{m}
\providecommandordefault{\k}{k}
\begin{athm}{cor}{lemma:sums_of_positive_definite_2}%
	Let 
		$\n, \m \in \N$,
		$\varepsilon \in (0, \infty)$,
	let $(B_{\k})_{\k \in \N_0} \subseteq \R^{\n \times \m}$,
	and let $(A_{\k})_{\k \in \N_0} \subseteq \R^{\n \times \n}$ satisfy for all
		$\k \in \N$ 
	that
	$A_0 = \varepsilon \idMatrix_{\n}$
	and
	$A_{\k} = A_{\k-1} + (B_{\k-1})^\ast B_{\k-1}$.
	Then it holds for all $\k \in \N_0$ that $A_{\k}$ is a \symmetricPositiveDefiniteMatrix\ \cfout.
\end{athm}
\begin{aproof}
\argument{
	induction;
	\cref{lemma:sums_of_positive_definite};
}
{
	that for all $\k \in \N_0$ it holds that $A_{\k}$ is a \symmetricPositiveDefiniteMatrix\ \cfload.
}
\end{aproof}
\endgroup

\defdetermShampoo
\algDescrDetermShampoo

\section{Momentum orthogonalized by Newton-Schulz (Muon) optimization}
\label{sect:determ_muon}

In this section we review the \Muon\ \GD\ optimization method (cf.\ Jordan et al.~\cite{Jordan2024}).
The \Muon\ \GD\ optimization method in Jordan et al.~\cite{Jordan2024} (cf.\ \cref{def:determ_Muon} below) employs the Newton-Schulz method (cf.\ \cref{def:NewtonSchulzAlgorithm} below) to approximately orthogonalize momentum terms.
We note that the version of the momentum \GD\ optimization method considered in Jordan et al.~\cite{Jordan2024} (and \cref{def:determ_Muon} below respectively) corresponds to the \second version of the momentum \GD\ optimization method introduced in this book (cf.\ \cref{def:determ_momentum_two} above). 
For work related to the \Muon\ \GD\ optimization method we refer, \eg, to \cite{Liu2025,Li2025,Bernstein2024}.
We refer to \cref{sect:muon} below for the stochastic version of \Muon\ optimization.

To better understand the aim of the \Muon\ \GD\ optimization method we first define an idealized version of the \Muon\ \GD\ optimization method in \cref{def:determ_ideal_Muon} which employs an exact orthogonalization of momentum terms instead of an approximation by the Newton-Schulz method.

\begin{adef}{def:hilbertSchmidtNorm}[Hilbert-Schmidt norm]
Let
	$n, m \in \N$,
	$A = (A_{i,j})_{(i,j) \in \{1, \dots, n\} \times \{1, \dots, m\}} \in \R^{n \times m}$.
Then we denote by $\hilbertSchmidtNorm{A} \in \R$ the real number given by
\begin{equation}
\textstyle 
  	\hilbertSchmidtNorm A 
  =
	\br[\big]{
		\smallsum_{ i = 1 }^{n}
		\smallsum_{ j = 1 }^{m}
		\lvert A_{i,j} \rvert^2
	}^{ \nicefrac{ 1 }{ 2 } }
  .
\end{equation}
\end{adef}

\defdetermidealMuon
\algDescrDetermidealMuon

\todoc{Add remark what newton-schulz aims to do}

\cfclear
\providecommandordefault{\done}{n}
\providecommandordefault{\dtwo}{m}
\providecommandordefault{\nn}{k}
\cfconsiderloaded{def:NewtonSchulzAlgorithm}
\begin{adef}{def:NewtonSchulzAlgorithm}[Newton-Schulz method]
Let 
	$a, b, c \in \R$, $\varepsilon \in (0, \infty)$,
	$\done, \dtwo \in \N$,
	$A \in \R^{\done \times \dtwo}$.
Then we denote by
$
	(\NewtonSchulzAlgorithm_{a, b, c, \varepsilon}(A, \nn))_{\nn \in \N_0} \subseteq \R^{\done \times \dtwo}
$
the matrices which satisfy for all
	$\nn \in \N$ 
that
$\NewtonSchulzAlgorithm_{a, b, c, \varepsilon}(A, 0) = (\hilbertSchmidtNorm{A} + \varepsilon)^{-1} A$
and
\begin{equation}
\label{T_B_D}
\begin{split} 
	\NewtonSchulzAlgorithm_{a, b, c, \varepsilon}(A, \nn)
&=
	a \NewtonSchulzAlgorithm_{a, b, c, \varepsilon}(A, \nn-1) \\
&\quad
	+ 
	b \NewtonSchulzAlgorithm_{a, b, c, \varepsilon}(A, \nn-1) (\NewtonSchulzAlgorithm_{a, b, c, \varepsilon}(A, \nn-1))^\ast \NewtonSchulzAlgorithm_{a, b, c, \varepsilon}(A, \nn-1) \\
&\quad
	+ 
	c \NewtonSchulzAlgorithm_{a, b, c, \varepsilon}(A, \nn-1) (\NewtonSchulzAlgorithm_{a, b, c, \varepsilon}(A, \nn-1))^\ast \\
&\quad
	\NewtonSchulzAlgorithm_{a, b, c, \varepsilon}(A, \nn-1) (\NewtonSchulzAlgorithm_{a, b, c, \varepsilon}(A, \nn-1))^\ast \NewtonSchulzAlgorithm_{a, b, c, \varepsilon}(A, \nn-1)
\end{split}
\end{equation}
and for every $\nn \in \N$ we call $\NewtonSchulzAlgorithm_{a, b, c, \varepsilon}(A, \nn)$ the $\nn$-th iteration of the Newton-Schulz method applied to $A$ with polynomial coefficients $a, b, c$ and regularization parameter $\varepsilon$
\cfload.
\end{adef}

\defdetermMuon
\algDescrDetermMuon

\begin{athm}{remark}{Muon_rem_params}
In Jordan et al.~\cite{Jordan2024}
it is proposed to choose
$
	a = 3.4445
$, 
$
	b = -4.7750
$, 
$
	c = 2.0315
$, 
$
	\varepsilon = 10^{-7}
$, and
$
	N = 5
$
as values for the polynomial coefficients $a, b, c$, regularization parameter $\varepsilon$, and number of iterations $N$ for the Newton-Schulz method in \cref{def:determ_Muon}.
\end{athm}

\begin{athm}{remark}{Muon_rem_matrices}
In \cref{def:determ_Muon} we have defined the \Muon\ \GD\ optimization method for loss functions that take matrices as inputs.
For loss functions that are defined on more complicated domains (such as, \eg, \anns) the \Muon\ \GD\ optimization method can be applied independently on components of the domain which are spaces of matrices and another \GD-type optimization method can be applied to the remaining components.
For example, in Jordan et al.~\cite{Jordan2024} it is suggested to apply the \Muon\ \GD\ optimization method to all \ann\ parameter matrices except for the matrices in the input and output layers and to train all remaining parameters with the \AdamW\ \GD\ optimization method.
\end{athm}

\section{AMSGrad optimization}
\label{sect:determ_AMSgrad}

In this section we consider the AMSGrad \GD\ optimization method (see Reddi et al.~\cite{Reddi2019}).
As in the case of previously considered optimization methods,
the AMSGrad \GD\ optimization method was introduced in a stochastic setting in Reddi et al.~\cite{Reddi2019},
but for pedagogical purposes we present in this section a deterministic version of AMSGrad optimization.
We refer to \cref{sect:amsgrad} below for the original stochastic version of AMSGrad optimization.

\defdetermAMSGrad
\algDescrDetermAMSGrad

\section{Compact summary of deterministic GD optimization methods}

In this section we provide an overview over all deterministic \GD-type optimization methods considered in \cref{chapter:deterministic}.
Roughly speaking, in this summary we provide for each considered method  the iteration step of the respective pseudo-code. 
The formulas in this summary make use of the componentwise operations in \cref{def:componentwise_operations}.

\algorithmSummaryDescription{\GD\ optimization method}{def:GD}{\iterationStepDetermGD}
\algorithmSummaryDescription{Explicit midpoint \GD\ optimization method}{def:midpointGD}{\iterationStepDetermmidpointGD}
\algorithmSummaryDescription{Momentum \GD\ optimization method}{def:determ_momentum}{\iterationStepDetermMomentum}
\algorithmSummaryDescription{Momentum \GD\ optimization method (\second version)}{def:determ_momentum_two}{\iterationStepDetermMomentumTwo}
\algorithmSummaryDescription{Momentum \GD\ optimization method (\third version)}{def:determ_momentum_three}{\iterationStepDetermMomentumThree}
\algorithmSummaryDescription{Momentum \GD\ optimization method (\fourth version)}{def:determ_momentum_four}{\iterationStepDetermMomentumFour}
\algorithmSummaryDescription{Bias-adjusted momentum \GD\ optimization method}{def:determ_momentum_bias}{\iterationStepDetermMomentumBias}
\algorithmSummaryDescription{Nesterov accelerated \GD\ optimization method}{def:determ_nesterov}{\iterationStepDetermNesterov}
\algorithmSummaryDescription{Nesterov accelerated \GD\ optimization method (\second version)}{def:determ_nesterov_two}{\iterationStepDetermNesterovTwo}
\algorithmSummaryDescription{Nesterov accelerated \GD\ optimization method (\third version)}{def:determ_nesterov_three}{\iterationStepDetermNesterovThree}
\algorithmSummaryDescription{Nesterov accelerated \GD\ optimization method (\fourth version)}{def:determ_nesterov_four}{\iterationStepDetermNesterovFour}
\algorithmSummaryDescription{Bias-adjusted Nesterov accelerated \GD\ optimization method}{def:determ_nesterov_bias}{\iterationStepDetermNesterovBias}
\algorithmSummaryDescription{Shifted Nesterov accelerated \GD\ optimization method}{def:determ_nesterov_shifted}{\iterationStepDetermNesterovAlt}
\algorithmSummaryDescription{Shifted Nesterov accelerated \GD\ optimization method (\second version)}{def:determ_nesterov_shifted_two}{\iterationStepDetermNesterovAltTwo}
\algorithmSummaryDescription{Shifted Nesterov accelerated \GD\ optimization method (\third version)}{def:determ_nesterov_shifted_three}{\iterationStepDetermNesterovAltThree}
\algorithmSummaryDescription{Shifted Nesterov accelerated \GD\ optimization method (\fourth version)}{def:determ_nesterov_shifted_four}{\iterationStepDetermNesterovAltFour}
\algorithmSummaryDescription{Shifted bias-adjusted Nesterov accelerated \GD\ optimization method}{def:determ_nesterov_shifted_bias}{\iterationStepDetermNesterovAltBias}
\algorithmSummaryDescription{Simplified Nesterov accelerated \GD\ optimization method}{def:determ_nesterov_simple}{\iterationStepDetermNesterovSimple}
\algorithmSummaryDescription{\Adagrad\ \GD\ optimization method}{def:determ_adagrad}{\iterationStepDetermAdagrad}
\algorithmSummaryDescription{\RMSprop\ \GD\ optimization method}{def:determ_RMSprop}{\iterationStepDetermRMSprop}
\algorithmSummaryDescription{Bias-adjusted \RMSprop\ \GD\ optimization method}{def:determ_RMSprop_bias}{\iterationStepDetermRMSpropBias}
\algorithmSummaryDescription{Adadelta \GD\ optimization method}{def:determ_adadelta}{\iterationStepDetermAdadelta}
\algorithmSummaryDescription{\Adam\ \GD\ optimization method}{def:determ_adam}{\iterationStepDetermAdam}
\algorithmSummaryDescription{Adamax \GD\ optimization method}{def:determ_adamax}{\iterationStepDetermAdamax}
\algorithmSummaryDescription{\Nadam\ \GD\ optimization method}{def:determ_nadam}{\iterationStepDetermNadam}
\algorithmSummaryDescription{Simplified \Nadam\ \GD\ optimization method}{def:determ_simplified_nadam}{\iterationStepDetermSimplifiedNadam}kil
\algorithmSummaryDescription{Nadamax \GD\ optimization method}{def:determ_nadamax}{\iterationStepDetermNadamax}
\algorithmSummaryDescription{\Adam\ \GD\ optimization method with $L^2$-regularization}{def:determ_adamLtwo}{\iterationStepDetermAdamLtwo}
\algorithmSummaryDescription{\AdamW\ \GD\ optimization method}{def:determ_adamW}{\iterationStepDetermAdamW}
\algorithmSummaryDescription{Shampoo \GD\ optimization method}{def:determ_Shampoo}{\iterationStepDetermShampoo}
\algorithmSummaryDescription{Idealized \Muon\ \GD\ optimization method}{def:determ_ideal_Muon}{\iterationStepDetermidealMuon}
\algorithmSummaryDescription{\Muon\ \GD\ optimization method}{def:determ_Muon}{\iterationStepDetermMuon}
\algorithmSummaryDescription{AMSgrad \GD\ optimization method}{def:determ_AMSGrad}{\iterationStepDetermAMSGrad}

%% file: parts/Stochastic_GD_type_optimization_methods.tex
\setlength{\headheight}{27.11469pt}
\addtolength{\topmargin}{-12.61469pt}

\cchapter{Stochastic gradient descent (SGD) optimization methods}{chapter:stochastic}

This chapter reviews and studies 
\SGD-type optimization methods 
such as the classical plain-vanilla \SGD\ optimization method
(see \cref{sect:SGD})
as well as more sophisticated \SGD-type optimization methods including 
\SGD-type optimization methods with momenta 
(cf.\ \cref{sect:momentum,sect:nesterov,sect:adam} below) 
and \SGD-type optimization methods 
with adaptive modifications of the learning rates
(cf.\ \cref{sect:adagrad,sect:RMSprop,sect:adadelta,sect:adam} below). 

For a brief list of resources in the scientific literature providing reviews on gradient based optimization methods we refer to the beginning of \cref{chapter:deterministic}.

\section{Introductory comments for the training of ANNs with SGD}
\label{sec:intro_SGD}

\cfclear
\begingroup
\providecommandordefault{\inputDim}{\defaultInputDim}
\providecommandordefault{\netDim}{{\defaultNetDim}}
\providecommandordefault{\LossFunction}{\defaultLossFunction}
\providecommandordefault{\targetFunction}{\cE}
\providecommandordefault{\empRiskInifite}{\mathfrak{L}}
\providecommandordefault{\x}{\defaultx}
\providecommandordefault{\y}{\defaulty}

\renewcommand{\d}{\mathfrak d}
\newcommand{\m}[2]{{\mathsf m}_{#1,#2}}
\renewcommand{\th}[1]{\theta_{#1}}

In \cref{chapter:deterministic} we have introduced and studied deterministic \GD-type optimization methods.
In deep learning algorithms usually not deterministic \GD-type optimization methods but stochastic variants of  \GD-type optimization methods are employed.
Such \SGD-type optimization methods can be viewed as suitable Monte Carlo approximations of deterministic \GD-type methods
and in this section we now roughly sketch some of the main ideas of such \SGD-type optimization methods.
To do this, we now briefly recall the deep supervised learning framework developed in the \hyperref[sec:intro]{introduction} and \cref{sec:intro_training_anns} above.

Specifically, 
let  
$ \defaultInputDim, M \in \N $, 
$ \targetFunction \in C( \R^\defaultInputDim, \R ) $,
$ \x_1, \x_2, \dots, \x_{ M + 1 } \in \R^\defaultInputDim $, 
$ \y_1, \y_2, \dots, \y_M \in \R $ 
satisfy for all $ m \in \{ 1, 2, \dots, M \} $ that
\begin{equation}
	\label{intro_SGD:eq1}
  \y_m = \targetFunction( \x_m ).
\end{equation}
As in the  \hyperref[sec:intro]{introduction} and in \cref{sec:intro_training_anns} 
we think of $ M \in \N $ as the number of available known input-output data pairs, 
we think of $ \defaultInputDim \in \N $ as the dimension of the input data, 
we think of $ \targetFunction \colon \R^\defaultInputDim \to \R $ as an unknown function 
which we want to approximate,
we think of $ \x_1, \x_2, \dots, \x_{ M + 1 } \in \R^\defaultInputDim $ as the available known 
input data, 
we think of $ \y_1, \y_2, \dots, \y_M \in \R $ as the available known output data,
and
we are trying to use the available known input-output data pairs to approximate the unknown function $\targetFunction$ by means of \anns.

Specifically,
let
$\activation \colon \R \to \R$ be differentiable,
let
$h\in\N$,
$l_1,l_2,\dots,l_h,\defaultParamDim\in\N$ satisfy
$\defaultParamDim=l_1(d+1)+\br[\big]{\sum_{k=2}^h l_k(l_{k-1}+1)}+l_h+1$,
and let
 $\defaultLossFunction \colon\R^\defaultParamDim\to[0,\infty)$ 
 satisfy
for all $\theta\in\R^\defaultParamDim$ that
\begin{equation}
\label{intro_SGD:eq2}
  \defaultLossFunction( \theta )
  = 
	\frac1M
  \br*{
    \sum_{ m = 1 }^M 
    \abs[\big]{ \bpr{ \RealV{ \theta}{0}{\defaultInputDim}{ \multdim_{\activation, l_1}, \multdim_{\activation, l_2}, \dots, \multdim_{\activation, l_h} , \id_{ \R } } } ( \x_m ) - \y_m }^2
  }
\end{equation}
\cfload.
Note that $h$ is the number of hidden layers of the \anns\ in \cref{intro_SGD:eq2},
note for every $i\in\{1,2,\dots,h\}$ that $l_i\in\N$
is the number of neurons in the $i$-th hidden layer of
the \anns\ in \cref{intro_SGD:eq2},
and
note that $\defaultParamDim$ is the number of real parameters used to describe
the \anns\ in \cref{intro_SGD:eq2}.
We recall that we are trying 
to approximate the function $\targetFunction$ by, first, 
computing an approximate minimizer $\vartheta \in \R^\netDim$
of the function
 $\defaultLossFunction \colon\R^\defaultParamDim\to[0,\infty)$
and, thereafter, 
employing the 
realization
\begin{equation}
\begin{split} 
	\R^{\defaultInputDim} \ni x 
\mapsto
	\RealV{ \vartheta}{0}{\defaultInputDim}{ \multdim_{\activation, l_1}, \multdim_{\activation, l_2}, \dots, \multdim_{\activation, l_h} , \id_{ \R } }
	\in \R
\end{split}
\end{equation}
of the \ann\ 
associated to the approximate minimizer $\vartheta \in \R^\netDim$
as an approximation of $\targetFunction$.

Deep learning algorithms typically solve optimization problems of the type \cref{intro_SGD:eq2} by means of gradient based optimization methods, which aim to minimize the considered objective function by performing successive steps based on the direction of the negative gradient of the objective function.
We recall that one of the simplest gradient based optimization method is the plain-vanilla \GD\ optimization method which performs successive steps in the direction of the negative gradient. 
In the context of the optimization problem in \cref{intro_SGD:eq2} this \GD\ optimization method reads as follows.
Let $\xi\in\R^\defaultParamDim$,
let $(\gamma_n)_{n\in\N}\subseteq[0,\infty)$, 
and let $\theta = (\theta_n)_{n \in \N_0} \colon  \N_0 \to \R^\defaultParamDim$ satisfy for all 
	$n \in \N$
that
\begin{equation}
	\label{intro_SGD:eq3}
   \theta_0 = \xi \qandq \theta_n = \theta_{n-1} - \gamma_n (\nabla \defaultLossFunction)(\theta_{n-1}).
\end{equation}
Note that the process $(\theta_n)_{n \in \N_0}$ is the \GD\ process for the objective function $\defaultLossFunction$ with learning rates $(\gamma_n)_{n\in\N}$  and initial value $\xi$ (cf.\ \cref{def:GD}).
Moreover, observe that the assumption that $\activation$ is differentiable ensures that $\defaultLossFunction$ in \eqref{intro_SGD:eq3}
is also differentiable (see \cref{sec:diff_anns} above for details).

In typical practical deep learning applications the number $M$ of available known input-output data pairs is very large, say, \eg, $M \geq 10^6$.
As a consequence it is typically computationally prohibitively expensive to determine the exact gradient of the objective function to perform steps of deterministic \GD-type optimization methods.
As a remedy for this, deep learning algorithms usually employ stochastic variants of \GD-type optimization methods, where in each step of the optimization method the precise gradient of the objective function is replaced by a Monte Carlo approximation of the gradient of the objective function.
We now sketch this approach for the \GD\ optimization method in \cref{intro_SGD:eq3} resulting in the popular \SGD\ optimization method applied to \cref{intro_SGD:eq2}.

Specifically, 
let
$J\in\N$,
let
$(\Omega, \mathcal{F}, \P)$ be a probability space,
for every
$n,j\in\N$,
let 
$\m nj\colon \Omega\to \R$
be a $\{1, 2, \ldots, M\}$-uniformly distributed random variable,
let
$\defaultStochLoss\colon\R^\d\times \R \to\R$
satisfy for all
$\theta\in\R^\d$,
$m \in \{1, 2, \ldots, M\}$
that
\begin{equation}
\label{intro_SGD:eq31}
\defaultStochLoss(\theta,m)
=
\babs{ \bpr{ \RealV{ \theta}{0}{\defaultInputDim}{ \multdim_{\activation, l_1}, \multdim_{\activation, l_2}, \dots, \multdim_{\activation, l_h} , \id_{ \R } } } ( \x_m ) - \y_m }^2
,
\end{equation}
and
let
$\Theta = (\Theta_n)_{n \in \N_0} \colon\N_0\times\Omega\to\R^\d$ 
satisfy for all 
$n\in\N$ that
\begin{equation}
\label{intro_SGD:eq4}
\Theta_0=\xi
\qquad\text{and}\qquad
\Theta_n
=
\Theta_{n-1}-\gamma_n\br*{\frac1J\sum_{j=1}^{J}(\nabla_\theta\defaultStochLoss)(\Theta_{n-1}, \m nj)}.
\end{equation}
The stochastic process $(\Theta_n)_{n \in \N_0} $ is an \SGD\ process for the minimization problem associated to \cref{intro_SGD:eq2} with
learning rates $(\gamma_n)_{n\in\N}$,
constant number of Monte Carlo samples (batch sizes) $J$,
initial value $\xi$,
and
data $(\m nj)_{ (n,j) \in \N^2}$ 
(see \cref{def:SGD} below for the precise definition).
Note that in \eqref{intro_SGD:eq4} in each step $n \in \N$ we only employ a Monte Carlo approximation 
\begin{equation}
\label{intro_SGD:eq5}
\begin{split} 
	\frac1J\sum_{j=1}^{J}(\nabla_\theta\defaultStochLoss)(\Theta_{n-1}, \m nj)
\approx
	\frac1M\sum_{m=1}^{M}(\nabla_\theta\defaultStochLoss)(\Theta_{n-1}, m)
=
	(\nabla \defaultLossFunction)(\Theta_{n-1})
\end{split}
\end{equation}
of the exact gradient of the objective function.
Nonetheless, in deep learning applications the \SGD\ optimization method (or other \SGD-type optimization methods) typically  result in good approximate minimizers of the objective function.
Note that employing approximate gradients in the \SGD\ optimization method in \eqref{intro_SGD:eq4} means that performing any step of the  \SGD\ process involves the computation of a sum with only $J$ summands,
while employing the exact gradient in the \GD\ optimization method in \eqref{intro_SGD:eq3} means that performing any step of the process involves the computation of a sum with $M$ summands.
In deep learning applications when $M$ is very large (\eg, $M \geq 10^6$) and $J$ is chosen to be reasonably small (\eg, $J = 128$),
this means that performing steps of the \SGD\ process is much more computationally affordable than performing steps of the \GD\ process.
Combining this with the fact that \SGD-type optimization methods do in the training of \anns\ often find good approximate minimizers (cf., \eg, \cref{suboptimal_points} and \cite{Dereich2021,Tadic2015}) is the key reason making the \SGD\ optimization method and other \SGD-type optimization methods the optimization methods chosen in almost all deep learning applications.
It is the topic of this chapter to introduce and study \SGD-type optimization methods such as the plain-vanilla \SGD\ optimization method in \cref{intro_SGD:eq4} above.

\endgroup

\section{SGD optimization}
\label{sect:SGD}

In the next notion we present the promised stochastic version of the plain-vanilla \GD\ optimization method from \cref{sec:gradient_descent}, that is, in the next notion we present the plain-vanilla \SGD\ optimization method.

\defSGD
\algDescrSGD

\subsection{SGD optimization in the training of ANNs}

\todoc{Add another example where data are RV}

In the next example we apply the \SGD\ optimization method in the context of the training of fully-connected feedforward \anns\ in the vectorized description (see \cref{subsec:vectorized_description})
with the loss function being the mean squared error loss function in \cref{def:mseloss} (see \cref{sect:MSE}).
Note that this is a very similar framework as the one developed in \cref{sec:intro_SGD}.

\todoc{Maybe cite another result for the differentiability.}

\begingroup
\renewcommand{\d}{\mathfrak d}
\newcommand{\m}[2]{{\mathsf m}_{#1,#2}}
\renewcommand{\th}[1]{\vartheta_{#1}}
\providecommandordefault{\x}{\defaultx}
\providecommandordefault{\y}{\defaulty}
\begin{athm}{example}{lem:sgd_example_1}
	Let 
		$d,h,\mathfrak d\in\N$,
		$l_1,l_2,\dots,l_h\in\N$
	satisfy
		$\mathfrak d=l_1(d+1)+\br[\big]{\sum_{k=2}^h l_k(l_{k-1}+1)}+l_h+1$,
	let
		$a\colon \R\to\R$ be differentiable,
	let
		$M\in\N$,
		$\x_1,\x_2,\dots,\x_M\in\R^d$,
		$\y_1,\y_2,\dots,\y_M\in\R$,
	let
		$\mathscr L\colon \R^{\mathfrak d}\to[0,\infty)$
	satisfy for all
		$\theta\in\R^{\mathfrak d}$
	that
	\begin{equation}
		\llabel{eq:defL}
		\mathscr L(\theta)
		=
		\frac1M\br*{\sum_{m=1}^{M}
		\abs*{ \bpr{ \RealV{ \theta}{0}{\defaultInputDim}{ \multdim_{a, l_1}, \multdim_{a, l_2}, \dots, \multdim_{a, l_h} , \id_{ \R } } } ( \x_m ) - \y_m }^2
		}
		,
	\end{equation}
	let
		$\ell \colon \R^\d\times \R^{d+1} \to \R$
	satisfy for all
		$\theta\in\R^\d$,
		$x \in \R^{d}$, 
		$y \in \R$
	that
	\begin{equation}
		\llabel{eq:defell}
		\ell(\theta,(x, y))
		=
		\babs{ \bpr{ \RealV{ \theta}{0}{\defaultInputDim}{ \multdim_{a, l_1}, \multdim_{a, l_2}, \dots, \multdim_{a, l_h} , \id_{ \R } } } ( x ) - y }^2
		,
	\end{equation}
	let
		$\xi\in\R^\d$,
	let 
		$(\gamma_n)_{n\in\N}\subseteq [0,\infty)$,
	let
		$\th{} \colon \N_0\to\R^\d$
	satisfy for all
		$n\in\N$
	that
	\begin{equation}
		\llabel{eq:gd}
		\th 0=\xi
		\qquad\text{and}\qquad
		\th n
		=
		\th{n-1} - \gamma_n(\nabla \mathscr L)(\th{n-1})
		,
	\end{equation}
	let
		$(\Omega,\mathcal F,\P)$ be a probability space,
	let
		$(J_n)_{n\in\N}\subseteq\N$,
	for every 
		$n, j \in\N$,
	let 
		$\m nj\colon \Omega\to \R$
		be a $\{1,2,\dots,M\}$-uniformly distributed random variable,
	for every 
		$n, j \in\N$
	let
		$X_{n,j}\colon \Omega\to \R^{d}$ and 
		$Y_{n,j}\colon \Omega\to \R$
	satisfy
	\begin{equation}
		\llabel{eq:X_Y_def}
		X_{n,j} = \x_{\m nj}
		\qandq
		Y_{n,j} = \y_{\m nj},
	\end{equation}
	and let
		$\Theta\colon\N_0\times\Omega\to\R^\d$ 
	satisfy for all 
		$n\in\N$ 
	that
	\begin{equation}
		\llabel{eq:sgd}
		\Theta_0=\xi
		\qquad\text{and}\qquad
		\Theta_n
		=
		\Theta_{n-1}-\gamma_n\br*{\frac1{J_n}\sum_{j=1}^{J_n}(\nabla_\theta\ell)(\Theta_{n-1},(X_{n,j},Y_{n,j}))}
		\cfadd{lem:differentiability_loss2}
	\end{equation}
	\cfload.
	Then
	\begin{enumerate}[(i)]
		\item \llabel{it:gd}
		it holds that $\th{}$ is the\cfadd{def:GD} \GD\ process for the objective function
		$\mathscr L$ with learning rates $(\gamma_n)_{n\in\N}$ and
		initial value $\xi$,
		\item \llabel{it:sgd}
		it holds that $\Theta$ is 
		the\cfadd{def:SGD} \SGD\ process
		for the loss function $\ell$ with learning rates
		$(\gamma_n)_{n\in\N}$, batch sizes $(J_n)_{n\in\N}$,
		initial value $\xi$, and data $((X_{n,j},Y_{n,j}))_{(n,j)\in \N^2}$,
		and
		\item \llabel{it:exp1}
		it holds for all
			$n\in\N$,
			$\theta\in\R^\d$
		that
		\begin{equation}
			\E\!\br*{\rule{0pt}{0.9cm}\theta - \gamma_n\bbbbbr{\frac1{J_n}\sum_{j=1}^{J_n}(\nabla_\theta\ell)(\theta,(X_{n,j},Y_{n,j}))}}
			=
			\theta - \gamma_n(\nabla\mathscr L)(\theta)
		\end{equation}
	\end{enumerate}
	\cfout.
\end{athm}
\begin{aproof}
	\Nobs that
		\lref{eq:gd}
	\proves
		\lref{it:gd}.
	\Nobs that 
		\lref{eq:sgd}
	proves 
		\lref{it:sgd}.
	\Nobs that
		\lref{eq:defL},
 		\lref{eq:defell},
		\lref{eq:X_Y_def},
		and the assumption that
			for all
				$n,j \in\N$
			it holds that
				$\m nj$ is uniformly distributed on $\{1,2,\dots,M\}$
	\prove that for all
		$\theta \in \R^\d$,
		$n, j \in\N$
	it holds that
	\begin{equation}
		\begin{split}
		\E[(\nabla_\theta\ell)(\theta, (X_{n,j},Y_{n,j}))]
		&=
			\E[(\nabla_\theta\ell)(\theta, (\x_{\m nj}, \y_{\m nj}))]\\
		&=
			\Exp{
				\sum_{m=1}^M
					(\nabla_\theta\ell)(\theta, (\x_k,\y_k))
					\mathbbm{1}_{\{\m nj = k\}}
			}\\
		&=
			\sum_{m=1}^M
				(\nabla_\theta\ell)(\theta, (\x_m,\y_m))
				\Exp{\mathbbm{1}_{\{\m nj = m\}}}\\
		&=
			\sum_{m=1}^M
				(\nabla_\theta\ell)(\theta, (\x_m,\y_m))
				\P(\{\m nj = m\})
		\\&=
			\frac1M\br*{\sum_{m=1}^{M}(\nabla_\theta\ell)(\theta,(\x_m,\y_m))}
		\\&=
		\frac1M\br*{\sum_{m=1}^{M}
		\babs{ \bpr{ \nabla_\theta \RealV{ \theta}{0}{\defaultInputDim}{ \multdim_{\activation, l_1}, \multdim_{\activation, l_2}, \dots, \multdim_{\activation, l_h} , \id_{ \R } } } ( \x_m ) - \y_m }^2
		}\\
		&=
		(\nabla\mathscr L)(\theta)
		.
		\end{split}
	\end{equation}
		\Hence
	for all 
		$n\in\N$,
		$\theta\in\R^\d$
	that
	\begin{equation}
		\begin{split}
		&\E\!\br*{\rule{0pt}{0.9cm}\theta - \gamma_n\bbbbbr{\frac1{J_n}\sum_{j=1}^{J_n}(\nabla_\theta\ell)(\theta,(X_{n,j},Y_{n,j}))}}\\
		&=
		\theta - \gamma_n\bbbbbr{\frac1{J_n}\sum_{j=1}^{J_n}\bExp{(\nabla_\theta\ell)(\theta,(X_{n,j},Y_{n,j}))}}
		\\&=
		\theta - \gamma_n\bbbbbr{\frac1{J_n}\sum_{j=1}^{J_n}(\nabla\mathscr L)(\theta)}
		\\&=
		\theta - \gamma_n(\nabla\mathscr L)(\theta).
		\end{split}
	\end{equation}
\end{aproof}

\begin{athm}{example}{lem:sgd_example}
	Let 
		$d,h,\mathfrak d\in\N$,
		$l_1,l_2,\dots,l_h\in\N$
	satisfy
		$\mathfrak d=l_1(d+1)+\br[\big]{\sum_{k=2}^h l_k(l_{k-1}+1)}+l_h+1$,
	let
		$a\colon \R\to\R$ be differentiable,
	let
		$M\in\N$,
		$\x_1,\x_2,\dots,\x_M\in\R^d$,
		$\y_1,\y_2,\dots,\y_M\in\R$,
	let
		$\mathscr L\colon \R^{\mathfrak d}\to[0,\infty)$
	satisfy for all
		$\theta\in\R^{\mathfrak d}$
	that
	\begin{equation}
		\llabel{eq:defL}
		\mathscr L(\theta)
		=
		\frac1M\br*{\sum_{m=1}^{M}
		\abs*{ \bpr{ \RealV{ \theta}{0}{\defaultInputDim}{ \multdim_{a, l_1}, \multdim_{a, l_2}, \dots, \multdim_{a, l_h} , \id_{ \R } } } ( \x_m ) - \y_m }^2
		}
		,
	\end{equation}
	let
		$S=\{1,2,\dots,M\}$,
	let
		$\ell\colon \R^\d\times S \to \R$
	satisfy for all
		$\theta\in\R^\d$,
		$m\in S$
	that
	\begin{equation}
		\llabel{eq:defell}
		\ell(\theta,m)
		=
		\babs{ \bpr{ \RealV{ \theta}{0}{\defaultInputDim}{ \multdim_{a, l_1}, \multdim_{a, l_2}, \dots, \multdim_{a, l_h} , \id_{ \R } } } ( \x_m ) - \y_m }^2
		,
	\end{equation}
	let
		$\xi\in\R^\d$,
	let 
		$(\gamma_n)_{n\in\N}\subseteq\N$,
	let
		$\th{} \colon \N_0\to\R^\d$
	satisfy for all
		$n\in\N$
	that
	\begin{equation}
		\llabel{eq:gd}
		\th 0=\xi
		\qquad\text{and}\qquad
		\th n
		=
		\th{n-1} - \gamma_n(\nabla \mathscr L)(\th{n-1})
		,
	\end{equation}
	let
		$(\Omega,\mathcal F,\P)$ be a probability space,
	let
		$(J_n)_{n\in\N}\subseteq\N$,
	for every 
		$n,j \in\N$
	let 
		$\m nj\colon \Omega\to S$
		be a uniformly distributed random variable,
	and let
		$\Theta\colon\N_0\times\Omega\to\R^\d$ 
	satisfy for all 
		$n\in\N$ 
	that
	\begin{equation}
		\llabel{eq:sgd}
		\Theta_0=\xi
		\qquad\text{and}\qquad
		\Theta_n
		=
		\Theta_{n-1}-\gamma_n\br*{\frac1{J_n}\sum_{j=1}^{J_n}(\nabla_\theta\ell)(\Theta_{n-1},\m nj)}
		\cfadd{lem:differentiability_loss2}
	\end{equation}
	\cfload.
	Then
	\begin{enumerate}[(i)]
		\item \llabel{it:gd}
		it holds that $\th{}$ is the\cfadd{def:GD} \GD\ process for the objective function
		$\mathscr L$ with learning rates $(\gamma_n)_{n\in\N}$ and
		initial value $\xi$,
		\item \llabel{it:sgd}
		it holds that $\Theta$ is 
		the\cfadd{def:SGD} \SGD\ process
		for the loss function $\ell$ with learning rates
		$(\gamma_n)_{n\in\N}$, batch sizes $(J_n)_{n\in\N}$,
		initial value $\xi$, and data $(\m nj)_{(n,j)\in \N^2}$,
		and
		\item \llabel{it:exp1}
		it holds for all
			$n\in\N$,
			$\theta\in\R^\d$
		that
		\begin{equation}
			\E\!\br*{\rule{0pt}{0.9cm}\theta - \gamma_n\bbbbbr{\frac1{J_n}\sum_{j=1}^{J_n}(\nabla_\theta\ell)(\theta,\m nj)}}
			=
			\theta - \gamma_n(\nabla\mathscr L)(\theta)
		\end{equation}
	\end{enumerate}
	\cfout.
\end{athm}
\begin{aproof}
	\Nobs that
		\lref{eq:gd}
	\proves
		\lref{it:gd}.
	\Nobs that 
		\lref{eq:sgd}
	proves 
		\lref{it:sgd}.
	\Nobs that
 		\lref{eq:defell},
		\lref{eq:defL},
		and the assumption that
			for all
				$n,j \in\N$
			it holds that
				$\m nj$ is uniformly distributed on $\{1,2,\dots,M\}$
	\prove that for all
		$\theta \in \R^\d$,
		$n\in\N$
	it holds that
	\begin{equation}
		\begin{split}
		&\E[(\nabla_\theta\ell)(\theta, (\x_{\m nj}, \y_{\m nj}))]
		=
		\frac1M\br*{\sum_{m=1}^{M}(\nabla_\theta\ell)(\theta, (\x_m,\y_m))}
		\\&=
		\frac1M\br*{\sum_{m=1}^{M}
		\babs{ \bpr{ \RealV{ \theta}{0}{\defaultInputDim}{ \multdim_{\activation, l_1}, \multdim_{\activation, l_2}, \dots, \multdim_{\activation, l_h} , \id_{ \R } } } ( \x_m ) - \y_m }^2
		}
		=
		(\nabla\mathscr L)(\theta)
		.
		\end{split}
	\end{equation}
		\Hence
	for all 
		$n\in\N$,
		$\theta\in\R^\d$
	that
	\begin{equation}
		\begin{split}
		\E\!\br*{\rule{0pt}{0.9cm}\theta - \gamma_n\bbbbbr{\frac1{J_n}\sum_{j=1}^{J_n}(\nabla_\theta\ell)(\theta,\m nj)}}
		&=
		\theta - \gamma_n\bbbbbr{\frac1{J_n}\sum_{j=1}^{J_n}\bExp{(\nabla_\theta\ell)(\theta,\m nj)}}
		\\&=
		\theta - \gamma_n\bbbbbr{\frac1{J_n}\sum_{j=1}^{J_n}(\nabla\mathscr L)(\theta)}
		\\&=
		\theta - \gamma_n(\nabla\mathscr L)(\theta).
		\end{split}
	\end{equation}
\end{aproof}
\Cref{code:sgd,code:sgd2} give two concrete implementations in {\sc PyTorch} 
of the framework described in \cref{lem:sgd_example} with different data and
network architectures. The plots generated by these codes can be found in 
in \cref{fig:sgd,fig:sgd2}, respectively. They show the approximations of the
respective target functions by the realization functions of the \anns\ at various
points during the training.

\filelisting{code:sgd}{code/optimization_methods/sgd.py}{{\sc Python} code implementing the \SGD\ optimization
	method in the training of an \ann\ as described in
	\cref{lem:sgd_example} in {\sc PyTorch}. 
	In this code a fully-connected feedforward \ann\ with a single hidden layer
	with 200 neurons using the hyperbolic tangent activation function
	is trained so that the realization function approximates
	the target function $\sin\colon\R\to\R$.
	\Cref{lem:sgd_example} is implemented with
	$d = 1$, $h = 1$, $\mathfrak d = 301$, $l_1 = 200$,
	$a=\tanh$, $M = 10000$,
	$\x_1,\x_2,\dots,\x_M\in\R$, $\y_i = \sin(x_i)$ for all $i\in\{1,2,\dots,M\}$,
	$\gamma_n = 0.003$ for all $n\in\N$, and $J_n=32$ for all $n\in\N$
	in the notation of \cref{lem:sgd_example}. The plot
	generated by this code is shown in \cref{fig:sgd}.}

\endgroup

\begin{figure}[!ht]
	\centering
	\includegraphics[width=1\linewidth]{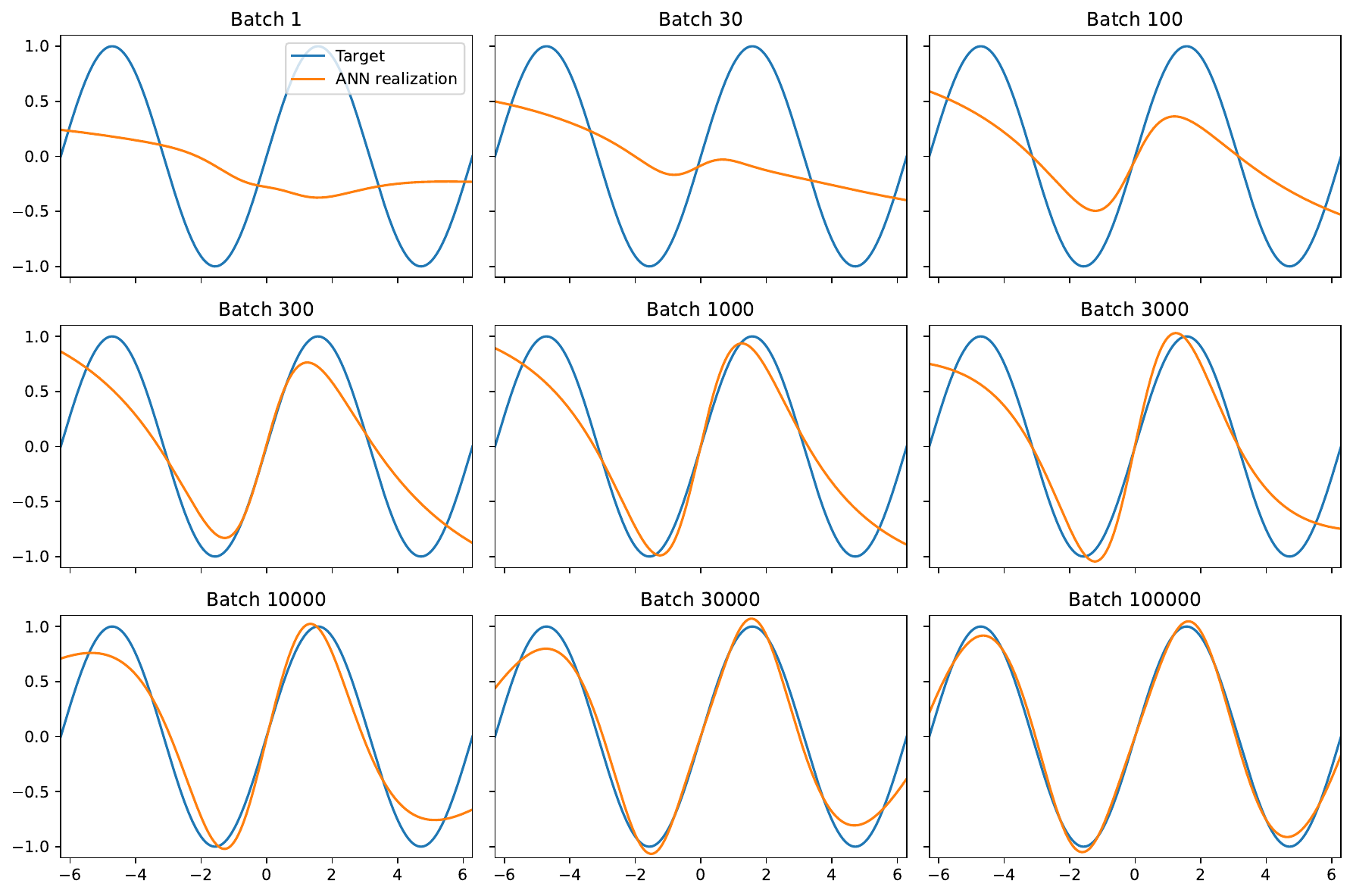}
	\caption{\label{fig:sgd}
		A plot showing the realization function of an \ann\
		at several points during training with the \SGD\ optimization method.
		This plot is generated by the code in \cref{code:sgd}.
	}
\end{figure}

\filelisting{code:sgd2}{code/optimization_methods/sgd2.py}{{\sc Python} code implementing the \SGD\ optimization
	method in the training of an \ann\ as described in
	\cref{lem:sgd_example} in {\sc PyTorch}. 
	In this code a fully-connected feedforward \ann\ with two hidden layers
	with 50 neurons each using the softplus activation funcction 
	is trained so that the realization function approximates
	the target function $f\colon\R^2\to\R$ which 
	satisfies for all $x,y\in\R$ that $f(x,y) = \sin(x)\sin(y)$.
	\Cref{lem:sgd_example} is implemented with
	$d = 1$, $h = 2$, $\mathfrak d = 2701$, $l_1 = l_2 = 50$,
	$a$ being the softplus activation function, $M = 10000$,
	$x_1,x_2,\dots,x_M\in\R^2$, $y_i = f(x_i)$ for all $i\in\{1,2,\dots,M\}$,
	$\gamma_n = 0.003$ for all $n\in\N$, and $J_n=32$ for all $n\in\N$
	in the notation of \cref{lem:sgd_example}. The plot generated by this
	code is shown in \cref{fig:sgd2}.}

\begin{figure}[!ht]
	\centering
	\includegraphics[width=1\linewidth]{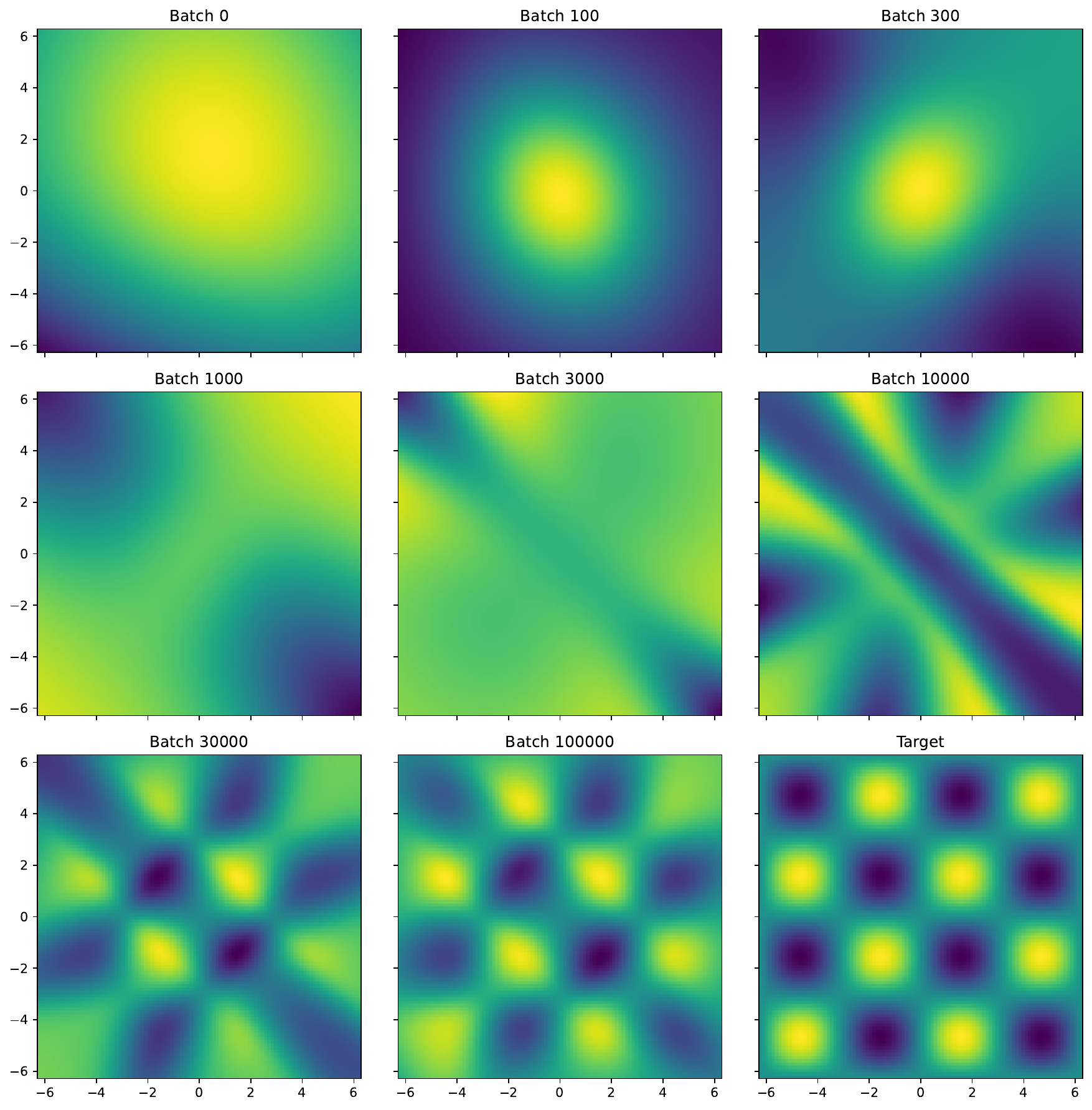}
	\caption{\label{fig:sgd2}
		A plot showing the realization function of an \ann\
		at several points during training with the \SGD\ optimization method.
		This plot is generated by the code in \cref{code:sgd2}.
	}
\end{figure}

\subsection{Non-convergence of SGD for not appropriately decaying learning rates}

In this section we present two results that, roughly speaking, motivate that the sequence of 
learning rates of the \SGD\ optimization method should be chosen such that they 
converge to zero (see \cref{comment_MB_LR1} below) but not too fast (see \cref{comment_learningrates2} below).

\subsubsection{Bias-variance decomposition of the mean square error}
\label{subsubsect:L2_distance}

\begin{athm}{lemma}{L2_distance}[Bias-variance decomposition of the mean square error]
Let $d \in \N$, $\vartheta \in \R^d$,
let $\altscp{ \cdot, \cdot } \colon \R^d \times \R^d \to \R$ be a scalar product,
let $\opnorm{\cdot} \colon \R^d \to [0,\infty)$ satisfy for all $v \in \R^d$ that 
\begin{equation}
	\opnorm{v} = \sqrt{\altscp{ v, v }},
\end{equation}
let $(\Omega, \mathcal{F}, \P)$ be a probability space, and
let $Z \colon \Omega \to \R^d$ be a random variable with $\EXp{\opnorm{Z}} < \infty$.
Then 
\begin{equation}
  \Exp{\opnorm{Z-\vartheta}^2} 
  = \Exp{\opnorm{Z-\EXp{Z}}^2} + \opnorm{\EXp{Z} - \vartheta}^2.
\end{equation}
\end{athm}

\begin{aproof}
\Nobs that the assumption that $\EXp{\opnorm{Z}} < \infty$ 
and the Cauchy-Schwarz inequality \prove that 
\begin{equation}
\begin{split}
\EXP{ \abs{\altscp{ Z - \EXp{Z}, \EXp{Z}-\vartheta}} } 
&\leq 
\EXP{\opnorm{ Z - \EXp{Z}}\opnorm{\EXp{Z}-\vartheta } }\\ 
&\leq
 (\Exp{\opnorm{ Z}} + \opnorm{\EXp{Z}})\opnorm{\EXp{Z}-\vartheta } < \infty.
\end{split}
\end{equation}
The linearity of the expectation \hence \proves that
\begin{equation}
\begin{split}
&\EXP{\opnorm{Z-\vartheta}^2} 
=
\EXP{\opnorm{(Z - \EXp{Z}) + (\EXp{Z}-\vartheta)}^2} \\
&=
\EXP{\opnorm{Z - \EXp{Z}}^2 + 2\altscp{ Z - \EXp{Z}, \EXp{Z}-\vartheta} + \opnorm{\EXp{Z}-\vartheta}^2} \\
&=
\EXP{\opnorm{Z - \EXp{Z}}^2} + 2\altscp{ \EXp{Z} - \EXp{Z}, \EXp{Z}-\vartheta} + \opnorm{\EXp{Z}-\vartheta}^2 \\
&=
\EXP{\opnorm{Z - \EXp{Z}}^2} + \opnorm{\EXp{Z}-\vartheta}^2.
\end{split}
\end{equation}
\end{aproof}

\subsubsection{Non-convergence of SGD for constant learning rates}

In this section we present
\cref{comment_minibatches_and_learningrates},
\cref{comment_MB_LR1}, and
\cref{concrete_example_comment_learningrates1}.
Our proof of \cref{comment_minibatches_and_learningrates} employs 
the auxiliary results in 
\cref{fubini_for_RV,lem:expnorm,sum_of_indep,factorization_lemma,meas_grad} below.
\cref{fubini_for_RV} recalls an elementary and well known property for the expectation of the product of 
independent random variables (see, \eg, Klenke~\cite[Theorem 5.4]{Klenke14}). 
In the elementary \cref{meas_grad} we prove under suitable hypotheses 
the measurability of certain derivatives of a function. 
A result similar to \cref{meas_grad} can, \eg, be found in Jentzen et al.~\cite[Lemma~4.4]{Jentzen18strong_arxiv}.

\begin{athm}{lemma}{fubini_for_RV}
Let $ ( \Omega , \mathcal{F}, \P ) $ be a probability space and
let $X,Y \colon \Omega \to \R$ be independent random variables 
with $\Exp{ \abs{X} + \abs{Y} } <\infty$.
Then
\begin{enumerate}[label=(\roman *)]
\item \label{fubini_for_RV:item1}
it holds that 
$
\EXP{ \abs{XY} } =  \EXP{ \abs{X} }\EXP{ \abs{Y} }<\infty
$ and
\item \label{fubini_for_RV:item2}
it holds that
$
\Exp{ XY } =  \Exp{ X }\Exp{ Y }
$.
\end{enumerate}
\end{athm}

\begin{aproof}
\Nobs that 
the fact that $(X,Y)(\P) = (X(\P)) \otimes (Y(\P))$, 
the integral transformation theorem, 
Fubini's theorem, 
and the assumption that  $\Exp{ \abs{X} + \abs{Y} } <\infty$ 
\prove that
\begin{equation}
\begin{split}
	\EXP{ \abs{XY} } 
&=
	\int_{ \Omega }
		\abs{X(\omega)Y(\omega)}
	\, \P( \diff \omega ) \\
&=
	\int_{ \R \times \R}
		\abs{xy}
	\, \bpr{(X,Y)(\P)}( \diff x, \diff y ) \\
&=
	\int_{ \R}
	\br*{
		\int_{ \R} \abs{xy} \, (X(\P))(\diff x)
	} 
	(Y(\P))(\diff y)\\
&=
	\int_{ \R}
	\abs{y}
	\br*{
		\int_{ \R} \abs{x} \, (X(\P))(\diff x)
	} 
	(Y(\P))(\diff y)\\
&=
	\br*{
		\int_{ \R} \abs{x} \, (X(\P))(\diff x)
	} 
	\br*{
		\int_{ \R} \abs{y} \, (Y(\P))(\diff y)
	}\\
&=
	\EXP{ \abs{X} } \EXP{ \abs{Y} } 
<
	\infty.
\end{split}
\end{equation} 
This \proves[ep] \cref{fubini_for_RV:item1}.
\Nobs that  
\cref{fubini_for_RV:item1}, 
the fact that $(X,Y)(\P) = (X(\P)) \otimes (Y(\P))$, 
the integral transformation theorem, 
and Fubini's theorem 
\prove that
\begin{equation}
\begin{split}
	\EXP{ XY } 
&=
	\int_{ \Omega }
		X(\omega)Y(\omega)
	\, \P( \diff \omega ) \\
&=
	\int_{ \R \times \R}
		xy
	\, \bpr{(X,Y)(\P)}( \diff x, \diff y ) \\
&=
	\int_{ \R}
	\br*{
		\int_{ \R} xy \, (X(\P))(\diff x)
	} 
	(Y(\P))(\diff y)\\
&=
	\int_{ \R}
	y
	\br*{
		\int_{ \R} x \, (X(\P))(\diff x)
	} 
	(Y(\P))(\diff y)\\
&=
	\br*{
		\int_{ \R} x \, (X(\P))(\diff x)
	} 
	\br*{
		\int_{ \R} y \, (Y(\P))(\diff y)
	}\\
&=
	\Exp{ X } \Exp{ Y }.
\end{split}
\end{equation}
This \proves[ep] \cref{fubini_for_RV:item2}.
\end{aproof}

\begingroup
\newcommand{\Cov}{\operatorname{Cov}}
\begin{athm}{lemma}{lem:expnorm}
	Let $(\Omega, \mc F, \P)$ be a probability space,
	let $d\in\N$,
	let $\altscp{ \cdot, \cdot } \colon \R^d \times \R^d \to \R$ be a scalar product,
	let $\opnorm{\cdot} \colon \R^d \to [0,\infty)$ 
	satisfy 
		for all 
			$v \in \R^d$ 
		that 
			\begin{equation}
				\opnorm{v} = \sqrt{\altscp{ v, v }},
			\end{equation}
	let $X\colon \Omega\to\R^d$ be a random variable,
	assume $\bEbr{\opnorm{X}^2}<\infty$,
	let $e_1,e_2,\dots,e_d\in\R^d$ 
	satisfy 
		for all
			$i,j\in\{1,2,\dots,d\}$
		that
		  $\altscp{e_i,e_j}=\indicator{\{i\}}(j)$,
	and for every 
		random variable $Y\colon\Omega\to\R^d$ 
			with $\bEbr{\opnorm{Y}^2}<\infty$
		let $\Cov(Y)\in\R^{d \times d}$ 
		satisfy 
		\begin{equation}
			\Cov(Y) = \bpr{\Ebr{\altscp{e_i,Y-\Ebr{Y}}\altscp{e_j,Y-\Ebr{Y}}}}{}_{(i,j)\in\{1,2,\dots,d\}^2}
			.
		\end{equation}
	Then
	\begin{equation}
		\llabel{claim}
		\Trace(\Cov(X))
		=
		\bEbr{\opnorm{X-\Ebr{X}}^2}
		.
	\end{equation}
\end{athm}
\begin{aproof}
	First, \nobs that 
		the fact that 
			$\forall\,i,j\in\{1,2,\dots,d\}\colon \altscp{e_i,e_j}=\indicator{\{i\}}(j)$
	\proves that for all
		$v\in\R^d$
	it holds that
		$\sum_{i=1}^d\altscp{e_i,v}e_i=v$.
	Combining 
		this 
	with
		the fact that
			$\forall\,i,j\in\{1,2,\dots,d\}\colon \altscp{e_i,e_j}=\indicator{\{i\}}(j)$
	\proves that
	\begin{eqsplit}
		\llabel{2}
		\Trace(\Cov(X))
		&=
		\sum_{i=1}^d\bEbr{\altscp{e_i,X-\Ebr{X}}\altscp{e_i,X-\Ebr{X}}}
		\\&=
		\sum_{i=1}^d\sum_{j=1}^d\Ebr{\altscp{e_i,X-\Ebr{X}}\altscp{e_j,X-\Ebr{X}}\altscp{e_i, e_j}}
		\\&=
		\bEbr{\baltscp{\ssumnl_{i=1}^d \altscp{e_i,X-\Ebr{X}} e_i,\ssumnl_{j=1}^d \altscp{e_j,X-\Ebr{X}} e_j}}
		\\&=
		\Ebr{\altscp{X-\Ebr{X},X-\Ebr{X}}}
		=
		\bEbr{\opnorm{X-\Ebr{X}}^2}
		.
	\end{eqsplit}
\end{aproof}
\endgroup

\begin{athm}{lemma}{sum_of_indep}
Let $d,n \in \N$, 
let $\altscp{ \cdot, \cdot } \colon \R^d \times \R^d \to \R$ be a scalar product,
let $\opnorm{\cdot} \colon \R^d \to [0,\infty)$ satisfy for all $v \in \R^d$ that 
\begin{equation}
	\opnorm{v} = \sqrt{\altscp{ v, v }}
	,
\end{equation} 
let $ ( \Omega , \mathcal{F}, \P ) $ be a probability space,
let $ X_k \colon \Omega \to \R^d$, $k\in\{1,2,\dots,n\}$, be independent random variables,
and assume
$
\sum_{k=1}^n\EXP{\opnorm{X_k}} < \infty
$.
Then
\begin{equation}
	\Exp{
    \opnorm*{
      	 {\textstyle\sum_{k = 1}^{n}} 
      	 (X_k -\Exp{X_k})
    }^2
    }
=
	\sum_{k = 1}^{n}
	\Exp{
    \opnorm*{
      	 X_k -\Exp{X_k}
    }^2
    }.
\end{equation}
\end{athm}

\begin{aproof}
First, \nobs that \cref{fubini_for_RV} and the assumption that 
 $\E\br{ \opnorm{X_1} +\allowbreak \opnorm{X_2} + \allowbreak \ldots + \opnorm{X_n}} < \infty$ 
 \prove that for all $k_1,k_2 \in \{1,2,\ldots,n \}$ with $k_1 \neq k_2$ it holds that
\begin{equation}
	\EXP{ \abs*{ \altscp{ X_{k_1}- \Exp{X_{k_1}}, X_{k_2}- \Exp{X_{k_2}} } } }
\leq	
	\EXP{ \opnorm{ X_{k_1}- \EXp{X_{k_1}}} \opnorm{ X_{k_2}- \EXp{X_{k_2}}}  }
< 
	\infty
\end{equation}
and
\begin{equation}
\begin{split}
	&\EXP{\altscp{ X_{k_1}- \Exp{X_{k_1}}, X_{k_2}- \Exp{X_{k_2}} }} \\
&=
	\aaltscp{ \Exp{X_{k_1}- \Exp{X_{k_1}}}, \Exp{X_{k_2}- \Exp{X_{k_2}}}  } \\
&= 
	\aaltscp{ \Exp{X_{k_1}}- \Exp{X_{k_1}}, \Exp{X_{k_2}}- \Exp{X_{k_2}}  } 
=
	0.
\end{split}
\end{equation}
\Hence that
\begin{equation}
\begin{split}
	&\Exp{
    \opnorm*{
      	 {\textstyle\sum_{k = 1}^{n}} 
      	 (X_k -\Exp{X_k})
    }^2
    } \\
&=
	\Exp{
    \aaltscp{
      	 {\textstyle\sum_{k_1 = 1}^{n}} 
      	 (X_{k_1} -\Exp{X_{k_1}})
      	 , 
      	 {\textstyle\sum_{k_2 = 1}^{n}} 
      	 (X_{k_2} -\Exp{X_{k_2}})
    }
    }\\
&=
	\Exp{{\textstyle\sum_{k_1,k_2 = 1}^{n}} 
    \aaltscp{
      	 X_{k_1} -\Exp{X_{k_1}}
      	 , 
      	 X_{k_2} -\Exp{X_{k_2}}
    }
    }\\
&=
	 \Exp{
	 	\pr*{
	 		\sum_{k = 1}^{n}
	 		\opnorm*{ 
      	 		X_k -\Exp{X_k}
    		}^2
    	}
    	+
    	\pr*{
     		\sum_{ \substack{k_1,k_2 \in \{1,2,\dots,n\}, \\k_1 \neq k_2}}
    		\altscp{
      			X_{k_1} -\Exp{X_{k_1}}
      	 		, 
      	 		X_{k_2} -\Exp{X_{k_2}}
    		}
    	}
    } \\
&=
	 \pr*{
	 	\sum_{k = 1}^{n}
	 	\Exp{
	 		\opnorm*{ 
      	 		X_k -\Exp{X_k}
    		}^2
   	 	}
   	 }
    +
    \pr*{
     \sum_{ \substack{k_1,k_2 \in \{1,2,\dots,n\},\\ k_1 \neq k_2}}
    \bExp{
    \altscp{
      	 X_{k_1} -\Exp{X_{k_1}}
      	 , 
      	 X_{k_2} -\Exp{X_{k_2}}
    }
    }
    } \\
&=
	\sum_{k = 1}^{n}
	\Exp{
		\opnorm*{ 
      	 	X_k -\Exp{X_k}
    	}^2
    }.
\end{split}
\end{equation}
\end{aproof}

\begin{athm}{lemma}{factorization_lemma}[Factorization lemma for independent random variables]
Let $ ( \Omega , \mathcal{F}, \P ) $ be a probability space,
let $(\mathbb{X},\mathcal{X})$ and $(\mathbb{Y},\mathcal{Y})$ be measurable spaces,
let $ X  \colon \Omega \to \mathbb{X}$ and $ Y \colon \Omega \to \mathbb{Y}$ be independent random variables,
let $\Phi \colon \mathbb{X}\times \mathbb{Y} \to [0,\infty]$ be $(\mathcal{X}\otimes \mathcal{Y})$/$\mathcal{B}([0,\infty])$-measurable, 
and let $\phi \colon \mathbb{Y} \to [0,\infty]$ satisfy for all $y\in \mathbb{Y}$ that 
\begin{equation}
	\phi(y)= \Exp{\Phi(X,y)}
	.
\end{equation}
Then
\begin{enumerate}[label=(\roman *)]
\item \label{factorization_lemma:item1}
it holds that the function $\phi$ is $\mathcal{Y}$/$\mathcal{B}([0,\infty])$-measurable and

\item \label{factorization_lemma:item2}
it holds that
\begin{equation}
	\Exp{\Phi(X,Y)} 
= 
	\Exp{\phi(Y)}.
\end{equation}
\end{enumerate}
\end{athm}

\begin{aproof}
First, \nobs that 
Fubini's theorem (cf., \eg,   Klenke~\cite[(14.6) in Theorem 14.16]{Klenke14}), 
the assumption that the function $ X  \colon \Omega \to \mathbb{X}$ is $\mathcal{F}$/$\mathcal{X}$-measurable, 
and the assumption that the function $\Phi \colon \mathbb{X}\times \mathbb{Y} \to [0,\infty]$ is $(\mathcal{X}\otimes \mathcal{Y}) $/$ \mathcal{B}([0,\infty])$-measurable 
\prove that the function
\begin{equation}
\mathbb{Y} \ni y \mapsto \phi(y)= \Exp{\Phi(X,y)} = \int_\Omega \Phi(X(\omega),y) \,\P(\diff \omega) \in [0,\infty]
\end{equation}
is $\mathcal{Y}/\mathcal{B}([0,\infty])$-measurable. %
This \proves[ep] \cref{factorization_lemma:item1}.
\Nobs that 
the integral transformation theorem, 
the fact that  $(X,Y)(\P) = (X(\P)) \otimes (Y(\P))$, 
and Fubini's theorem 
\prove that
\begin{equation}
\begin{split}
	\EXP{\Phi(X,Y)} 
&=
	\int_{ \Omega }
		\Phi(X(\omega),Y(\omega))
	\, \P( \diff \omega ) \\
&=
	\int_{ \mathbb{X} \times \mathbb{Y}}
		\Phi(x,y)
	\, \bpr{(X,Y)(\P)}( \diff x, \diff y ) \\
&=
	\int_{ \mathbb{Y}}
	\br*{
		\int_{ \mathbb{X}} \Phi(x,y) \, (X(\P))(\diff x)
	} 
	(Y(\P))(\diff y)\\
&=
	\int_{ \mathbb{Y}}
	\EXP{ \Phi(X,y)}
	\, (Y(\P))(\diff y) \\
&=
	\int_{ \mathbb{Y}}
	\phi(y)
	\, (Y(\P))(\diff y)
=
	\EXP{ \phi(Y)}.
\end{split}
\end{equation}
This \proves[ep] \cref{factorization_lemma:item2}.
\end{aproof}

\cfclear
\begingroup
\providecommand{\d}{}
\renewcommand{\d}{\defaultParamDim}
\providecommand{\F}{}
\renewcommand{\F}{\defaultStochLoss}
\providecommand{\G}{}
\renewcommand{\G}{\defaultStochGradient}
\begin{athm}{lemma}{meas_grad}
Let $\d \in \N$,  let $(S, \mathcal{S})$ be a measurable space, 
let $\F = ( \F(\theta,x) )_{(\theta,x) \in \R^\d\times S}\colon$ $\R^\d \times S \to \R$ 
be $(\mathcal{B}(\R^\d) \otimes \mathcal{S})$/$\mathcal{B}(\R)$-measurable, 
and assume for every $x \in S$ that the function 
$
 \R^\d \ni \theta \mapsto \F(\theta,x) \in \R
$
is differentiable. 
Then the function 
\begin{equation}
\R^\d \times S \ni (\theta, x) \mapsto(\nabla_\theta \F)(\theta,x) \in \R^\d
\end{equation}
 is $(\mathcal{B}(\R^\d) \otimes \mathcal{S}) $/$\mathcal{B}(\R^\d)$-measurable.
\end{athm}
\begin{aproof}
Throughout this proof, 
let $\G = (\G_1, \dots, \G_\d) \colon \R^\d \times S \to \R^\d$ 
satisfy for all $\theta \in \R^\d$, $x\in S$ that
\begin{equation}
\label{meas_grad:eq1}
\G(\theta,x)= (\nabla_\theta \F)(\theta,x).
\end{equation}
The assumption that the function $\F \colon \R^\d \times S \to \R$ is 
$(\mathcal{B}(\R^\d) \otimes \mathcal{S})$/$\mathcal{B}(\R)$-measurable 
\proves that for all $i \in \{1,2,\dots,\d\}$, $h\in \R\backslash\{0\}$ 
it holds that the function
\begin{equation}
\begin{split}
&\R^\d \times S \ni(\theta,x)=((\theta_1,\dots,\theta_\d),x) \mapsto \pr*{\tfrac{\F((\theta_1,\dots,\theta_{i-1},\theta_{i}+h,\theta_{i+1},\dots,\theta_\d),x)-\F(\theta,x)}{h}
} \in \R
\end{split}
\end{equation}
 is $(\mathcal{B}(\R^\d) \otimes \mathcal{S}) $/$ \mathcal{B}(\R)$-measurable. 
 The fact that for all $i \in \{1,2,\dots,\d\}$, 
 $\theta=(\theta_1,\dots,\theta_\d) \in \R^\d$, $x\in S$ it holds that
\begin{equation}
\G_i(\theta,x)= \lim\limits_{n \to \infty} \pr*{\tfrac{\F((\theta_1,\dots,\theta_{i-1},\theta_{i}+2^{-n},\theta_{i+1},\dots,\theta_\d),x)-\F(\theta,x)}{2^{-n}}
}
\end{equation}
\hence \proves that  for all $i \in \{1,2,\dots,\d\}$ 
it holds that the function $\G_i \colon \R^\d \times S \to \R$ is
$(\mathcal{B}(\R^\d) \otimes \mathcal{S}) $/$ \mathcal{B}(\R)$-measurable. 
This \proves that $\G$ is $(\mathcal{B}(\R^\d) \otimes \mathcal{S}) $/$ \mathcal{B}(\R^\d)$-measurable.
\end{aproof}
\endgroup

\cfclear
\begingroup
\providecommand{\d}{}
\renewcommand{\d}{\defaultParamDim}
\providecommand{\F}{}
\renewcommand{\F}{\defaultStochLoss}
\providecommand{\G}{}
\renewcommand{\G}{\defaultStochGradient}
\begin{athm}{lemma}{comment_minibatches_and_learningrates}
Let $\d \in \N$, 
$(\gamma_n)_{n \in \N} \subseteq (0,\infty)$, 
$(J_n)_{n \in \N} \subseteq \N$,
let $\altscp{\cdot, \cdot } \colon \R^\d \times \R^\d \to \R$ be a scalar product,
let $\opnorm{\cdot} \colon \R^\d \to [0,\infty)$ satisfy for all 
$v \in \R^\d$ that 
\begin{equation}
	\opnorm{v} = \sqrt{\altscp{v, v}}
	,
\end{equation}
let $ ( \Omega , \mathcal{F}, \P) $ be a probability space, 
let $\xi \colon \Omega \to \R^\d$ be a random variable, 
let $(S, \mathcal{S})$ be a measurable space, 
let $X_{n,j}\colon \Omega \to S$, $j \in \{1,2,\ldots,J_n\}$, $n \in \N$, be i.i.d.\ random variables,
assume that $\xi$ and $(X_{n,j})_{(n,j) \in  \{(k,l)\in\N^2\colon l\leq J_k\}}$ are independent,
let $\F = ( \F(\theta,x) )_{(\theta,x) \in \R^\d \times S} \colon \R^\d \times S \to \R$ 
  be $(\mathcal{B}(\R^\d) \otimes \mathcal{S})$/$\mathcal{B}(\R)$-measurable, 
assume for all $x \in S$ that $ ( \R^\d \ni \theta \mapsto \F(\theta,x) \in \R ) \in C^1(\R^\d, \R) $, 
assume for all $\theta \in \R^\d$ that
$
  \EXP{ \opnorm{(\nabla_\theta \F)(\theta, X_{1,1})}} < \infty  
$
(cf.\ \cref{meas_grad}), 
let $\mathcal{V} \colon \R^\d \to [0,\infty]$ satisfy for all $\theta \in \R^\d$ that
\begin{equation}
\label{comment_minibatches_and_learningrates:ass1}
	\mathcal{V}(\theta) 
= 
	\Exp{\opnorm{(\nabla_\theta \F)(\theta , X_{1,1} ) - \EXP{(\nabla_\theta \F)(\theta , X_{1,1})  } }^2 }, 
\end{equation}
and let $\Theta \colon \N_0 \times \Omega \to \R^\d$ be the stochastic process which satisfies for all $n \in \N$ that 
\begin{equation}
\label{comment_minibatches_and_learningrates:ass2}
\Theta_0 = \xi   \qandq
\Theta_n = \Theta_{n-1} - \gamma_n \br*{\frac{1}{J_n}\sum_{j = 1}^{J_n} ( \nabla_\theta \F )(\Theta_{n-1}, X_{n,j})}.
\end{equation}
Then 
it holds for all $ n \in \N $, $ \vartheta \in \R^\d $ that
\begin{equation}
\label{comment_minibatches_and_learningrates:concl1}
  \bpr{ \EXP{\opnorm{\Theta_n-\vartheta}^2} }^{ \nicefrac{ 1 }{ 2 } } 
  \geq 
  \bbbbr{\frac{\gamma_n}{(J_n)^{\nicefrac{1}{2}}}}
  \,
  \bpr{ 
    \EXP{ \mathcal{V}(\Theta_{ n - 1 } ) } 
  }^{ \nicefrac{ 1 }{ 2 } }
  .
\end{equation}
\end{athm}

\begin{aproof}
Throughout this proof, 
for every $n\in\N$
let $\phi_n \colon \R^\d \to [0,\infty]$
  satisfy for all $\theta \in \R^\d$ that
\begin{equation}
\label{comment_minibatches_and_learningrates:eq1}
	\phi_n(\theta) 
=
	\Exp{
    \opnorm*{
      	\theta -  \tfrac{\gamma_n}{J_n} \br*{{\textstyle\sum_{j = 1}^{J_n}} ( \nabla_\theta \F )(\theta, X_{n,j})} - \vartheta
    }^2
  	}.
\end{equation}
\Nobs that \cref{L2_distance} \proves that
for all $\vartheta \in \R^\d$ 
and all random variables $Z \colon \Omega \to \R^\d$ 
with $\Exp{ \opnorm{Z} } < \infty$ 
it holds that
\begin{equation}
\begin{split}
  \EXP{\opnorm{Z-\vartheta}^2} 
& 
  = \Exp{\opnorm{Z-\EXp{Z}}^2} + \opnorm{\EXp{Z} - \vartheta}^2
 \geq
  \Exp{\opnorm{Z-\EXp{Z}}^2}
  .
\end{split}
\end{equation}
\Hence for all $n \in \N$, $\theta \in \R^\d$ that
\begin{equation}
\label{comment_minibatches_and_learningrates:eq2}
\begin{split}
	\phi_n(\theta) 
&=
	\Exp{
    \opnorm*{
      	\tfrac{\gamma_n}{J_n}\br*{{\textstyle\sum_{j = 1}^{J_n}} ( \nabla_\theta \F )(\theta, X_{n,j})} - (\theta - \vartheta)
    }^2
  	} \\
&\geq	
	\Exp{
    \opnorm*{
      	\tfrac{\gamma_n}{J_n} \br*{{\textstyle\sum_{j = 1}^{J_n}} ( \nabla_\theta \F )(\theta, X_{n,j})} - 
      	\Exp{\tfrac{\gamma_n}{J_n} \br*{{\textstyle\sum_{j = 1}^{J_n}} ( \nabla_\theta \F )(\theta, X_{n,j})} }
    }^2
  	} \\
&=
	\tfrac{(\gamma_n)^2}{(J_n)^2} \,
	\Exp{
    \opnorm*{
      	 {\textstyle\sum_{j = 1}^{J_n}} 
      	 \bpr{ ( \nabla_\theta \F )(\theta, X_{n,j}) - 
      	\EXP{( \nabla_\theta \F )(\theta, X_{n,j})}}
    }^2
  	}.
\end{split}
\end{equation}
\Cref{sum_of_indep}, the fact that 
$
X_{n,j} \colon \Omega \to S
$,
$j \in \{1,2,\ldots,J_n\}, n \in \N,$ are i.i.d.\ random variables, and the fact that for all $n \in \N, j \in \{1,2,\ldots,J_n\}$, $\theta \in \R^\d$ it holds that 
\begin{equation}
\EXP{ \opnorm{(\nabla_\theta \F)(\theta, X_{n,j})}} = \EXP{ \opnorm{(\nabla_\theta \F)(\theta, X_{1,1})}} < \infty
\end{equation}
\hence \prove that for all $n \in \N$, $\theta \in \R^\d$ it holds that
\begin{equation}
\label{comment_minibatches_and_learningrates:eq3}
\begin{split}
	\phi_n(\theta) 
&\geq	
	\tfrac{(\gamma_n)^2}{(J_n)^2} 
	\br*{ \sum_{j = 1}^{J_n} 
		\Exp{
   	 	\opnorm*{
			( \nabla_\theta \F )(\theta, X_{n,j}) - 
      		\EXP{( \nabla_\theta \F )(\theta, X_{n,j})}
   	 	}^2
  		}
  	} \\
&=
	\tfrac{(\gamma_n)^2}{(J_n)^2} 
	\br*{ \sum_{j = 1}^{J_n} 
		\Exp{
   	 	\opnorm*{
			( \nabla_\theta \F )(\theta, X_{1,1}) - 
      		\EXP{( \nabla_\theta \F )(\theta, X_{1,1})}
   	 	}^2
  		}
  	} \\
&=
	\tfrac{(\gamma_n)^2}{(J_n)^2} 
	\br*{ \sum_{j = 1}^{J_n} 
		\mathcal{V}(\theta)
  	} 
=
	\tfrac{(\gamma_n)^2}{(J_n)^2} \,
	\bbr{
		J_n \mathcal{V}(\theta)
	}
=
	\pr*{\tfrac{(\gamma_n)^2}{J_n} }
	\mathcal{V}(\theta).
\end{split}
\end{equation}
\Moreover 
\eqref{comment_minibatches_and_learningrates:ass2}, 
\eqref{comment_minibatches_and_learningrates:eq1},
the fact that for all $ n \in \N $ it holds that 
$ \Theta_{ n - 1 } $ and $ (X_{n,j})_{j\in\{1,2,\dots,J_n\}} $ are independent random variables, 
and \cref{factorization_lemma} 
\prove that for all $n \in \N$, $ \vartheta \in \R^\d $ it holds that
\begin{equation}
\begin{split}
  \EXP{\opnorm{\Theta_n-\vartheta}^2} 
&=
  \Exp{ 
    \opnorm*{
      \Theta_{n-1} - \tfrac{\gamma_n}{J_n} \br*{{\textstyle\sum_{j = 1}^{J_n}} ( \nabla_\theta \F )(\Theta_{n-1}, X_{n,j})} - \vartheta
    }^2
  } \\
&=
  \E \bbr{ 
      \phi_n(\Theta_{n-1})
  }.
\end{split}
\end{equation}
Combining this with \eqref{comment_minibatches_and_learningrates:eq3} \proves that 
for all $n \in \N$, $ \vartheta \in \R^\d $ it holds that
\begin{equation}
\begin{split}
  \EXP{\opnorm{\Theta_n-\vartheta}^2} 
\geq
  \Exp{
      \pr*{\tfrac{(\gamma_n)^2}{J_n} }
	\mathcal{V}(\Theta_{n-1})
  }
=
	\pr*{\tfrac{(\gamma_n)^2}{J_n} }
	\EXP{\mathcal{V}(\Theta_{n-1})}.
\end{split}
\end{equation}
This \proves[ep] \eqref{comment_minibatches_and_learningrates:concl1}.
\end{aproof}
\endgroup

\cfclear
\begingroup
\providecommand{\d}{}
\renewcommand{\d}{\defaultParamDim}
\providecommand{\F}{}
\renewcommand{\F}{\defaultStochLoss}
\providecommand{\G}{}
\renewcommand{\G}{\defaultStochGradient}
\begin{athm}{cor}{comment_MB_LR1}
Let $\d \in \N$, $\varepsilon \in (0,\infty)$, $(\gamma_n)_{n \in \N} \subseteq (0,\infty)$, $(J_n)_{n \in \N} \subseteq \N$,
let $\altscp{ \cdot, \cdot } \colon \R^\d \times \R^\d \to \R$ be a scalar product,
let $\opnorm{\cdot} \colon \R^\d \to [0,\infty)$ satisfy for all $v \in \R^\d$ that 
\begin{equation}
	\opnorm{v} = \sqrt{\altscp{v, v}}
	,
\end{equation}
let $ ( \Omega , \mathcal{F}, \P) $ be a probability space, 
let $\xi \colon \Omega \to \R^\d$ be a random variable, 
let $(S, \mathcal{S})$ be a measurable space, 
let $X_{n,j}\colon \Omega \to S$, $j \in \{1,2,\ldots,J_n\}$, $n \in \N$, be i.i.d.\ random variables,
assume that $\xi$ and $(X_{n,j})_{(n,j)\in\{(k,l)\in\N^2\colon l\leq J_k\}}$ are independent,
let $\F = ( \F(\theta,x) )_{(\theta,x) \in \R^\d \times S} \colon \R^\d  \times S \to \R$ be $(\mathcal{B}(\R^\d) \otimes \mathcal{S})$/$\mathcal{B}(\R)$-measurable, 
assume for all $x \in S$ that $ ( \R^\d \ni \theta \mapsto \F(\theta,x) \in \R ) \in C^1(\R^\d, \R) $, 
assume for all $\theta \in \R^\d$ that
$
\EXP{ \opnorm{(\nabla_\theta \F)(\theta, X_{1,1})}}<\infty
$
(cf.\ \cref{meas_grad})
and
\begin{equation}
\label{comment_MB_LR1:ass1}
	\pr*{
		\Exp{\opnorm{(\nabla_\theta \F)(\theta , X_{1,1} ) - \EXP{(\nabla_\theta \F)(\theta , X_{1,1})  } }^2 }
	}^{\nicefrac{1}{2}}
\geq 
	\varepsilon,
\end{equation}
and let $\Theta \colon \N_0 \times \Omega \to \R^\d$ be the stochastic process which satisfies for all $n \in \N$ that 
\begin{equation}
\label{comment_MB_LR1:ass2}
\Theta_0 = \xi   \qandq
\Theta_n = \Theta_{n-1} - \gamma_n \br*{\frac{1}{J_n}\sum_{j = 1}^{J_n} ( \nabla_\theta \F )(\Theta_{n-1}, X_{n,j})}.
\end{equation}
Then 
\begin{enumerate}[label=(\roman *)]
\item \label{comment_MB_LR1:item1}
it holds for all $ n \in \N $, $ \vartheta \in \R^\d $ that
\begin{equation}
\label{comment_MB_LR1:concl1}
  \bpr{ \EXP{\opnorm{\Theta_n-\vartheta}^2} }^{ \nicefrac{ 1 }{ 2 } } 
  \geq 
  \varepsilon \pr*{ \frac{\gamma_n}{(J_n)^{\nicefrac{1}{2}}}}
\end{equation}
and

\item \label{comment_MB_LR1:item2}
it holds for all $\vartheta \in \R^\d$ that
\begin{equation}
	\liminf_{n \to \infty}\bpr{ \EXP{\opnorm{\Theta_n-\vartheta}^2} }^{ \nicefrac{ 1 }{ 2 } }
\geq 
	\varepsilon \pr*{\liminf_{n\to \infty}\,\br*{\frac{\gamma_n}{(J_n)^{\nicefrac{1}{2}}}}}.
\end{equation}
\end{enumerate}
\end{athm}

\begin{aproof}
Throughout this proof, let $\mathcal{V} \colon \R^\d \to [0,\infty]$ 
satisfy for all $\theta \in \R^\d$ that
\begin{equation}
	\mathcal{V}(\theta) 
= 
	\Exp{\opnorm{(\nabla_\theta \F)(\theta , X_{1,1} ) - \EXP{(\nabla_\theta \F)(\theta , X_{1,1})  } }^2 }.
\end{equation}
\Nobs that \eqref{comment_MB_LR1:ass1} \proves that for all $\theta \in \R^\d$ it holds that
\begin{equation}
\mathcal{V}(\theta) \geq \varepsilon^2.
\end{equation}
\Cref{comment_minibatches_and_learningrates} \hence \proves that 
for all $ n \in \N $, $ \vartheta \in \R^\d $ it holds that
\begin{equation}
  \bpr{ \EXP{\opnorm{\Theta_n-\vartheta}^2} }^{ \nicefrac{ 1 }{ 2 } } 
\geq 
  \frac{\gamma_n}{(J_n)^{\nicefrac{1}{2}}}
  \,
  \bpr{ 
    \EXP{ \mathcal{V}(\Theta_{ n - 1 } ) } 
  }^{ \nicefrac{ 1 }{ 2 } }
\geq
  \br*{ 
  \frac{\gamma_n}{(J_n)^{\nicefrac{1}{2}}}
  }
  ( 
    \varepsilon^2
  )^{ \nicefrac{ 1 }{ 2 } }
=
  \frac{\gamma_n \varepsilon }{(J_n)^{\nicefrac{1}{2}}}.
\end{equation}
This \proves \cref{comment_MB_LR1:item1}. 
\Nobs that \cref{comment_MB_LR1:item1} implies \cref{comment_MB_LR1:item2}.
\end{aproof}
\endgroup

\cfclear
\begingroup
\providecommand{\d}{}
\renewcommand{\d}{\defaultParamDim}
\providecommand{\F}{}
\renewcommand{\F}{\defaultStochLoss}
\providecommand{\G}{}
\renewcommand{\G}{\defaultStochGradient}
\begin{athm}{lemma}{concrete_example_comment_learningrates1}[Lower bound for the \SGD\ optimization method]
Let $\d \in \N$, $(\gamma_n)_{n \in \N} \subseteq (0,\infty)$, 
$(J_n)_{n \in \N} \subseteq \N$,
let $ ( \Omega , \mathcal{F}, \P) $ be a probability space, 
let $\xi \colon \Omega \to \R^\d$ be a random variable, 
let $X_{n,j}\colon \Omega \to \R^\d$, $j \in \{1,2,\ldots,J_n\}$, $n \in \N$, be i.i.d.\ random variables with $\Exp{\pnorm2{X_{1,1}}} < \infty$,
assume that $\xi$ and $(X_{n,j})_{(n,j)\in\{(k,l)\in\N^2\colon l\leq J_k\}}$ are independent,
let $\F = ( \F(\theta,x) )_{(\theta,x) \in \R^\d \times \R^\d} \colon \R^\d \times \R^\d \to \R$ satisfy for all $\theta,x \in \R^\d$ that
\begin{equation}
\label{concrete_example_comment_learningrates1:ass1}
\F(\theta,x) = \tfrac{1}{2}\pnorm2{\theta-x}^2,
\end{equation}
and let $\Theta \colon \N_0 \times \Omega \to \R^\d$ be the stochastic process which satisfies for all $n \in \N$ that 
\begin{equation}
\label{concrete_example_comment_learningrates1:ass2}
\Theta_0 = \xi   \qandq
\Theta_n = \Theta_{n-1} - \gamma_n \br*{\frac{1}{J_n}\sum_{j = 1}^{J_n} ( \nabla_\theta \F )(\Theta_{n-1}, X_{n,j})}.
\end{equation}
Then 
\begin{enumerate}[label=(\roman *)]
\item \label{concrete_example_comment_learningrates1:item1}
it holds for all $\theta \in \R^\d$ that
\begin{equation}
	\EXP{ \pnorm2{(\nabla_\theta \F)(\theta, X_{1,1})}}
<
	\infty,
\end{equation}

\item \label{concrete_example_comment_learningrates1:item2}
it holds for all $\theta \in \R^\d$ that
\begin{equation}
\Exp{\bpnorm2{(\nabla_\theta \F)(\theta , X_{1,1} ) - \EXP{(\nabla_\theta \F)(\theta , X_{1,1})  } }^2 } = \Exp{\pnorm2{X_{1,1}-\Exp{X_{1,1}}}^2},
\end{equation}
and

\item \label{concrete_example_comment_learningrates1:item3}
it holds for all $n \in \N$, $\vartheta \in \R^\d$ that
\begin{equation}
	\bpr{ \EXP{\pnorm2{\Theta_n-\vartheta}^2} }^{ \nicefrac{ 1 }{ 2 } } 
\geq 
	 \bpr{ \bExp{ \pnorm2{ X_{1,1} - \E[ X_{1,1} ] }^2 } }^{ \nicefrac{ 1 }{ 2 } }
	 \br*{ \frac{\gamma_n}{(J_n)^{\nicefrac{1}{2}}}}.
\end{equation}
\end{enumerate}
\end{athm}
 
\begin{aproof}
First, \nobs that 
\eqref{concrete_example_comment_learningrates1:ass1} 
and \cref{der_of_norm} \prove that for all $\theta, x \in \R^\d$ it holds that 
\begin{equation}
\label{concrete_example_comment_learningrates1:eq1}
(\nabla_\theta \F)(\theta,x) 
= 
\tfrac{1}{2}(2( \theta-x))
=
\theta-x.
\end{equation}
The assumption that $\Exp{\pnorm2{X_{1,1}}} < \infty$ \hence \proves that 
for all $\theta \in \R^\d$ it holds that
\begin{equation}
	\EXP{ \pnorm2{(\nabla_\theta \F)(\theta, X_{1,1})}}
=
	\EXP{ \pnorm2{ \theta - X_{1,1}}}
\leq
	\pnorm2{ \theta} + \EXP{ \pnorm2{X_{1,1}}}
<
	\infty.
\end{equation}
This \proves[ep] \cref{concrete_example_comment_learningrates1:item1}.
\Moreover \eqref{concrete_example_comment_learningrates1:eq1} 
and \cref{concrete_example_comment_learningrates1:item1} 
\prove that for all $\theta \in \R^\d$ it holds that
\begin{equation}
\begin{split}
&\EXP{\pnorm2{(\nabla_\theta \F)(\theta, X_{1,1}) - \Exp{(\nabla_\theta \F)(\theta, X_{1,1})}}^2 }\\
&=
\EXP{\pnorm2{(\theta -  X_{1,1}) - \Exp{\,\theta - X_{1,1}}}^2 } 
=
\EXP{\pnorm2{X_{1,1} - \Exp{X_{1,1}}}^2 }.
\end{split}
\end{equation}
This \proves[ep] \cref{concrete_example_comment_learningrates1:item2}. 
\Nobs that \cref{comment_MB_LR1:item1} in \cref{comment_MB_LR1} 
and \cref{concrete_example_comment_learningrates1:item1,concrete_example_comment_learningrates1:item2} 
\prove[ep] \cref{concrete_example_comment_learningrates1:item3}. 
\end{aproof}
\endgroup

\subsubsection{Non-convergence of GD for summable learning rates}

In the next auxiliary result, \cref{LB_log} below, 
we recall a well known lower bound for the natural logarithm.

\begin{athm}{lemma}{LB_log}[A lower bound for the natural logarithm]
It holds for all $x \in (0,\infty)$ that
\begin{equation}
\label{LB_log:concl1}
\log(x) \geq \frac{(x-1)}{x}.
\end{equation}
\end{athm}

\begin{aproof}
First, \nobs that 
the fundamental theorem of calculus 
\proves that for all $x \in [1,\infty)$ it holds that
\begin{equation}
\label{LB_log:eq1}
	\log(x) 
= 
	\log(x) - \log(1)
=
	\int_1^x \frac{1}{t} \, \diff t
\geq
 	\int_1^x \frac{1}{x} \, \diff t
=
	\frac{(x-1)}{x}.
\end{equation}
\Moreover the fundamental theorem of calculus 
\proves that for all $x \in (0,1]$ it holds that
\begin{equation}
\begin{split}
	\log(x) 
&= 
	\log(x) - \log(1)
= 
	-(\log(1) - \log(x)) 
=
	-\br*{\int_x^1 \frac{1}{t} \, \diff t} \\
&=
	\int_x^1\pr*{- \frac{1}{t}}  \diff t
\geq
 	\int_x^1\pr*{- \frac{1}{x}}  \diff t
=
	(1-x)\pr*{- \frac{1}{x} }
=
	\frac{(x-1)}{x}.
\end{split}
\end{equation}
This and \eqref{LB_log:eq1} \prove[ep] \eqref{LB_log:concl1}.
\end{aproof}

\cfclear
\begingroup
\providecommand{\d}{}
\renewcommand{\d}{\defaultParamDim}
\providecommand{\f}{}
\renewcommand{\f}{\defaultLossFunction}
\begin{athm}{lemma}{comment_learningrates2}[\GD\ fails to converge for a summable sequence of learning rates]
Let $\d \in \N$, $\vartheta \in \R^\d$, $\xi \in \R^\d\backslash \{\vartheta\}$,  
$\alpha \in (0,\infty)$, $(\gamma_n)_{n \in \N} \subseteq [0,\infty)\backslash \{\nicefrac{1}{\alpha} \}$ satisfy 
$
\sum_{n = 1}^\infty \gamma_n < \infty
$,
let $\f \colon \R^\d \to \R$ satisfy for all $\theta \in \R^\d$ that
\begin{equation}
\f(\theta) = \tfrac{\alpha}{2} \pnorm2{\theta-\vartheta}^2, 
\end{equation}
and let $\Theta \colon \N_0 \to \R^\d$ satisfy for all $n \in \N$ that $\Theta_0 = \xi$ and
\begin{equation}
\begin{split}
\Theta_n 
=
\Theta_{n-1} - \gamma_n (\nabla \f)(\Theta_{n-1}).
\end{split}
\end{equation}
Then
\begin{enumerate}[label=(\roman *)]
\item \label{comment_learningrates2:item1}
it holds for all $n \in \N_0$ that
\begin{equation}
\Theta_n - \vartheta = \br*{ \prod_{k = 1}^n (1-\gamma_k\alpha) }(\xi-\vartheta),
\end{equation}
\item \label{comment_learningrates2:item2}
it holds that
\begin{equation}
\liminf_{n \to \infty} \br*{ \prod_{k = 1}^n \babs{ 1-\gamma_k\alpha } }
>
0,
\end{equation}
and
\item \label{comment_learningrates2:item3}
it holds that
\begin{equation}
\liminf_{n \to \infty} \pnorm2{\Theta_n - \vartheta} > 0.
\end{equation}
\end{enumerate}
\end{athm}

\begin{aproof}
Throughout this proof, let $m \in \N$ satisfy for all $k \in \N \cap [m,\infty)$ that $\gamma_k < \nicefrac{1}{(2\alpha)}$.
\Nobs that \cref{der_of_norm} \proves that for all $\theta \in \R^\d$ it holds that
\begin{equation}
(\nabla \f)(\theta) 
=
\tfrac{\alpha}{2}(2(\theta -\vartheta))
=
\alpha (\theta -\vartheta).
\end{equation}
\Hence for all $n \in \N$ that
\begin{equation}
\begin{split}
\Theta_n -\vartheta
&=
\Theta_{n-1} - \gamma_n (\nabla \f)(\Theta_{n-1}) -\vartheta \\
&=
\Theta_{n-1} - \gamma_n \alpha(\Theta_{n-1}-\vartheta) -  \vartheta \\
&=
(1-\gamma_n \alpha)(\Theta_{n-1} - \vartheta).
\end{split}
\end{equation}
Induction \hence \proves that for all $n \in \N$ it holds that
\begin{equation}
\Theta_n - \vartheta = \br*{ \prod_{k = 1}^n (1-\gamma_k\alpha) }(\Theta_0-\vartheta),
\end{equation}
This and the assumption that $\Theta_0 = \xi$ \prove[ep] \cref{comment_learningrates2:item1}. 
\Nobs that the fact that for all $k \in \N$ it holds that $\gamma_k \alpha \neq 1$ \proves that 
\begin{equation}
\label{comment_learningrates2:eq1}
\prod_{k = 1}^{m-1} \babs{ 1-\gamma_k\alpha } > 0.
\end{equation}
Moreover, note that the fact that for all $k \in \N \cap [m,\infty)$ it holds that $\gamma_k \alpha \in [0,\nicefrac{1}{2})$ assures that for all $k \in \N \cap [m,\infty)$ it holds that
\begin{equation}
(1-\gamma_k \alpha) \in (\nicefrac{1}{2},1].
\end{equation}
This, \cref{LB_log}, and the assumption that
$
\sum_{n = 1}^\infty \gamma_n < \infty
$
\prove that for all $n \in \N \cap [m,\infty)$ it holds that
\begin{equation}
\begin{split}
	&\log \pr*{\prod_{k = m}^n \babs{ 1-\gamma_k\alpha } } 
=
	\sum_{k = m}^n \log (1-\gamma_k\alpha) \\
&\geq
	\sum_{k = m}^n \frac{(1-\gamma_k\alpha)-1}{(1-\gamma_k\alpha)} 
=
	\sum_{k = m}^n \br*{ - \frac{\gamma_k\alpha}{(1-\gamma_k\alpha)} } \\
&\geq
	\sum_{k = m}^n \br*{ - \frac{\gamma_k\alpha}{(\frac{1}{2})} } 
=
	 - 2\alpha \br*{\sum_{k = m}^n \gamma_k} 
\geq
	- 2\alpha \br*{\sum_{k = 1}^\infty \gamma_k} 
> 
	-\infty.
\end{split}
\end{equation}
Combining this with \eqref{comment_learningrates2:eq1} \proves that for all 
$n \in \N \cap [m,\infty)$ it holds that
\begin{equation}
\begin{split}
	 \prod_{k = 1}^n \babs{ 1-\gamma_k\alpha }
&=
	\br*{ \prod_{k = 1}^{m-1} \babs{ 1-\gamma_k\alpha } }
	\exp  
	\pr*{  
		\log 
		\pr*{ 
			\prod_{k = m}^{n} \babs{ 1-\gamma_k\alpha } 
		}
	} \\
&\geq
	\br*{ \prod_{k = 1}^{m-1} \babs{ 1-\gamma_k\alpha } }
	\exp 
	\pr*{ 
		- 2\alpha \br*{\sum_{k = 1}^\infty \gamma_k} 
	}
> 
	0.
\end{split}
\end{equation}
\Hence that
\begin{equation}
	\liminf_{n \to \infty} \br*{ \prod_{k = 1}^n \babs{ 1-\gamma_k\alpha } }
\geq
	\br*{ \prod_{k = 1}^{m-1} \babs{ 1-\gamma_k\alpha } }
	\exp 
	\pr*{ 
		- 2\alpha \br*{\sum_{k = 1}^\infty \gamma_k} 
	}
> 
	0.
\end{equation}
This \proves[ep] \cref{comment_learningrates2:item2}. 
\Nobs that \cref{comment_learningrates2:item1,comment_learningrates2:item2} 
and the assumption that $\xi \neq \vartheta$ \prove that
\begin{equation}
\begin{split}
	\liminf_{n \to \infty} \pnorm2{\Theta_n - \vartheta} 
&= 
	\liminf_{n \to \infty} \,\apnorm2{\br*{ \prod_{k = 1}^n (1-\gamma_k\alpha) }(\xi-\vartheta)} \\
&=
	\liminf_{n \to \infty} \pr*{ \abs*{ \prod_{k = 1}^n (1-\gamma_k\alpha) } \pnorm2{\xi-\vartheta}} \\
&=
	\apnorm2{\xi-\vartheta} \pr*{\liminf_{n \to \infty} \br*{ \prod_{k = 1}^n \babs{ 1-\gamma_k\alpha }} }
> 
	0.
\end{split}
\end{equation}
This \proves[ep] \cref{comment_learningrates2:item3}.
\end{aproof}
\endgroup

\subsection{Convergence rates for SGD for quadratic objective functions}
\label{ssec:lower_bounds}

\Cref{example_SGD} below, in particular, provides 
an error analysis for the \SGD\ optimization method 
in the case of one specific stochastic optimization problem 
(see \eqref{eq:example_stochastic_optimization} below). 
More general error analyses for the \SGD\ optimization method 
can, \eg, be found in \cite{Jentzen18strong,JentzenvonWurstemberger2020} 
and the references therein 
(cf.\ \cref{ssec:lower_bounds} below).

\cfclear
\begingroup
\providecommand{\d}{}
\renewcommand{\d}{\defaultParamDim}
\providecommand{\F}{}
\renewcommand{\F}{\defaultStochLoss}
\providecommand{\f}{}
\renewcommand{\f}{\defaultLossFunction}
\begin{athm}{example}{example_SGD}[Example of an \SGD\ process]
Let $\d \in \N$, 
let $(\Omega, \mathcal{F}, \P)$ be a probability space, 
let $X_n \colon \Omega \to \R^\d$, $n \in \N$, be i.i.d.\ random variables with $\Exp{\pnorm2{X_1}^2} < \infty$, 
let $\F = ( \F(\theta,x) )_{(\theta,x) \in \R^\d \times \R^\d} \colon \R^\d \times \R^\d \to \R$ 
and $\f \colon \R^\d \to \R$ satisfy for all $\theta, x  \in \R^\d$ that
\begin{equation}
\label{eq:example_stochastic_optimization}
\F(\theta,x) = \tfrac{1}{2} \pnorm2{\theta-x}^2 \qandq \f(\theta) = \EXP{\F(\theta,X_1)},
\end{equation}
and let $\Theta \colon \N_0 \times \Omega \to \R^\d$ be the stochastic process which satisfies for all $n \in \N$ that $\Theta_0= 0$ and
\begin{equation}
\label{example_SGD:assumption2}
\Theta_n = \Theta_{n-1} - \tfrac{1}{n} (\nabla_\theta \F) (\Theta_{n-1},X_n)
\end{equation}
\cfload.
Then
\begin{enumerate}[label=(\roman *)]
\item \label{example_SGD:item1}
it holds that
$
\{\theta \in \R^\d  \colon  \f(\theta)= \inf\nolimits_{w \in \R^\d} \f(w)  \} = \{ \EXp{X_1} \},
$

\item \label{example_SGD:item2}
it holds for all $n \in \N$ that $\Theta_n = \tfrac{1}{n} (X_1 + X_2+  \ldots + X_n)$,

\item \label{example_SGD:item3}
it holds for all $n \in \N$ that
\begin{equation}
\label{example_SGD:conclusion2}
\bpr{\Exp{ \apnorm2{\Theta_n-\EXp{X_1}}^2} }^{\nicefrac{1}{2}} = \bpr{\Exp{ \apnorm2{X_1-\EXp{X_1}}^2} }^{\nicefrac{1}{2}} \, n^{-\nicefrac{1}{2}},
\end{equation}
and 

\item \label{example_SGD:item4}
it holds for all $n \in \N$ that
\begin{equation}
\Exp{\f(\Theta_n)} - \f(\EXp{X_1}) = \tfrac{1}{2}\,\Exp{ \pnorm2{X_1-\EXp{X_1}}^2} n^{-1}.
\end{equation}
\end{enumerate} 
\end{athm}

\begin{aproof}
\Nobs that 
the assumption that $\Exp{\pnorm2{X_1}^2} < \infty$ 
and \cref{L2_distance} \prove that for all $\theta \in \R^\d$ it holds that
\begin{equation}
\label{example_SGD:eq0}
\begin{split}
\f(\theta)
&= 
\EXP{\F(\theta,X_1)}
=
\tfrac{1}{2} \,\EXP{ \pnorm2{ X_1 - \theta}^2} \\
&=  
\tfrac{1}{2}\pr*{ \EXP{\pnorm2{X_1 - \EXp{X_1}}^2} + \pnorm2{\theta- \EXp{X_1}}^2 }.
\end{split}
\end{equation}
This \proves[ep] \cref{example_SGD:item1}.
\Nobs that \cref{der_of_norm} \proves that for all $\theta,x \in \R^\d$ it holds that
\begin{equation}
(\nabla_\theta \F) (\theta,x) = \tfrac{1}{2}(2(\theta-x)) = \theta - x.
\end{equation}
This and \eqref{example_SGD:assumption2} assure that for all $n \in \N$ it holds that
\begin{equation}
\label{example_SGD:eq1}
	\Theta_n 
=
	\Theta_{n-1} - \tfrac{1}{n} (\Theta_{n-1} - X_n)
=
	(1-\tfrac{1}{n})\, \Theta_{n-1} + \tfrac{1}{n} X_n
=
	\tfrac{(n-1)}{n}\,\Theta_{n-1} + \tfrac{1}{n} X_n.
\end{equation}
Next we claim that for all $n \in \N$ it holds that
\begin{equation}
\label{example_SGD:eq2}
\Theta_n = \tfrac{1}{n} (X_1 + X_2+ \ldots + X_n).
\end{equation}
We now prove \eqref{example_SGD:eq2} by induction on $n \in \N$.
For the base case $n = 1$ note that \eqref{example_SGD:eq1} implies that 
\begin{equation}
	\Theta_1
=
	\bpr{\tfrac{0}{1}}\Theta_{0} +  X_1
=
	\bpr{\tfrac{1}{1}}(X_1).
\end{equation}
This establishes \eqref{example_SGD:eq2} in the base case $n = 1$.
For the induction step note that \eqref{example_SGD:eq1} \proves that 
for all $n \in \{2,3,4,\ldots \}$ with 
$
\Theta_{n-1} = \tfrac{1}{(n-1)} (X_1 + X_2+ \ldots + X_{n-1})
$
it holds that
\begin{equation}
\begin{split}
	\Theta_n 
&= 
	\tfrac{(n-1)}{n}\,\Theta_{n-1} + \tfrac{1}{n} X_n 
=
	\br*{\tfrac{(n-1)}{n}} \br*{\tfrac{1}{(n-1)}} (X_1 + X_2+ \ldots + X_{n-1}) + \tfrac{1}{n} X_n  \\
&=
	\tfrac{1}{n} (X_1 + X_2+ \ldots + X_{n-1}) + \tfrac{1}{n} X_n  
=
	\tfrac{1}{n} (X_1 + X_2+ \ldots + X_n).
\end{split}
\end{equation}
Induction \hence \proves \eqref{example_SGD:eq2}. 
\Moreover \eqref{example_SGD:eq2} \proves[ep] \cref{example_SGD:item2}.
\Nobs that \cref{sum_of_indep}, \cref{example_SGD:item2}, 
and the fact that $(X_n)_{n \in \N}$ are i.i.d.\ random variables 
with $\Exp{\pnorm2{X_1}} < \infty$ \prove that for all $n \in \N$ it holds that
\begin{equation}
\label{example_SGD:eq3}
\begin{split}
	\Exp{ \pnorm2{\Theta_n-\EXp{X_1}}^2} 
&=
	\Exp{ \pnorm2{\tfrac{1}{n} (X_1 + X_2+ \ldots + X_n)-\EXp{X_1}}^2} \\
&= 
	\Exp{ \apnorm2{  \frac{1}{n} \br*{ {\textstyle\sum\limits_{k = 1}^n}( X_k- \EXp{X_1})}}^2  } \\
&=  
	\frac{1}{n^2}  \pr*{ \Exp{ \apnorm2{  {\textstyle\sum\limits_{k = 1}^n}( X_k- \EXp{X_k})}^2  } }\\
&= 
	\frac{1}{n^2} \textstyle  \br*{ \sum\limits_{k = 1}^n \EXP{\pnorm2{X_k- \EXp{X_k}}^2} } \\
&= 
	\frac{1}{n^2} \textstyle  \bbbr{ n \,  \EXP{\pnorm2{X_1- \EXp{X_1}}^2} } \\
&=
	\frac{\Exp{\pnorm2{X_1- \EXp{X_1}}^2}}{n}.
\end{split}
\end{equation}
This \proves[epi] \cref{example_SGD:item3}.
It thus remains to prove \cref{example_SGD:item4}.
For this \nobs that \eqref{example_SGD:eq0} and \eqref{example_SGD:eq3} \prove that 
for all $n \in \N$ it holds that
\begin{equation}
\begin{split}
\Exp{\f(\Theta_n)} - \f(\EXp{X_1}) 
&= 
\Exp{\tfrac{1}{2} \bpr{ \EXP{\pnorm2{\EXp{X_1} - X_1}^2} + \pnorm2{\Theta_n- \EXp{X_1}}^2 }} \\
&\qquad 
- \tfrac{1}{2}\bpr{ \EXP{\pnorm2{\EXp{X_1} - X_1}^2} + \pnorm2{\EXp{X_1}- \EXp{X_1}}^2 } \\
&=
\tfrac{1}{2}\,\Exp{\pnorm2{\Theta_n- \EXp{X_1}}^2} \\
&=
\tfrac{1}{2} \, \EXP{\pnorm2{X_1- \EXp{X_1}}^2} \, n^{-1}.
\end{split}
\end{equation} 
This \proves[ep] \cref{example_SGD:item4}.
\end{aproof}
\endgroup

The next result, \cref{intro_thm} below, 
specifies strong and weak convergence rates for 
the \SGD\ optimization method in dependence 
on the asymptotic behavior of the sequence of learning rates. 
The statement and the proof of \cref{intro_thm} can 
be found in Jentzen et al.~\cite[Theorem~1.1]{JentzenvonWurstemberger2020}.

\cfclear
\begingroup
\providecommand{\d}{}
\renewcommand{\d}{\defaultParamDim}
\providecommand{\F}{}
\renewcommand{\F}{\defaultStochLoss}
\providecommand{\f}{}
\renewcommand{\f}{\defaultLossFunction}
\begin{athm}{theorem}{intro_thm}[Convergence rates in dependence of learning rates]
Let $\d \in \N$, $\alpha, \gamma,\nu  \in (0,\infty)$, $\xi \in \R^\d$, 
let $(\Omega, \mathcal{F}, \P)$ be a probability space, 
let $X_n \colon \Omega \to \R^\d$, $n \in \N$, be i.i.d.\ random variables with
$\Exp{\pnorm2{X_1}^2} < \infty$
and
$\P(X_1= \EXp{X_1}) < 1$,
let $(r_{\varepsilon, i})_{(\varepsilon,i) \in (0,\infty)\times\{0,1\}} \subseteq \R$ satisfy for all $\varepsilon \in (0,\infty)$, $i \in \{0,1\}$ that
\begin{equation}
r_{\varepsilon, i} =
\begin{cases}
	\nicefrac{\nu}{2} 																			&\colon \nu < 1 \\
	\min \{\nicefrac{1}{2}, \gamma \alpha + (-1)^{i} \varepsilon\}	&\colon \nu = 1  \\
	0																										&\colon \nu > 1,
\end{cases}
\end{equation}
let $\F = ( \F(\theta,x) )_{(\theta,x) \in \R^\d \times \R^\d} \colon \R^\d \times \R^\d \to \R$ and $\f \colon \R^\d \to \R$ be the functions which satisfy for all $\theta, x  \in \R^\d$ that
\begin{equation}
\label{intro_thm:ass1}
\F(\theta,x) = \tfrac{\alpha}{2} \pnorm2{\theta-x}^2 \qandq \f(\theta) = \EXP{\F(\theta,X_1)},
\end{equation}
and let $\Theta \colon \N_0 \times \Omega \to \R^\d$ be the stochastic process which satisfies for all $n \in \N$ that 
\begin{equation}
\label{intro_thm:ass2}
\begin{split}
\Theta_0 = \xi \qandq \Theta_n = \Theta_{n-1} - \tfrac{\gamma}{n^\nu} (\nabla_\theta \F) (\Theta_{n-1},X_n).
\end{split}
\end{equation}
Then 
\begin{enumerate}[label=(\roman *)]
\item \label{intro_thm:item1}
there exists a unique $\vartheta \in \R^\d$ which satisfies that 
$
\{\theta \in \R^\d  \colon  \f(\theta) = \inf\nolimits_{w \in \R^\d} \f(w)  \} = \{ \vartheta \},
$

\item \label{intro_thm:item2}
for every $\varepsilon \in (0,\infty)$ there exist $c_0,c_1 \in (0,\infty)$ such that for all $n \in \N$ it holds that
\begin{equation}
c_0n^{-r_{\varepsilon,0}}
\leq
\bpr{\EXP{\pnorm2{\Theta_n-\vartheta}^2}}^{\nicefrac{1}{2}} 
\leq
c_1  n^{-r_{\varepsilon,1}},
\end{equation}

and
\item \label{intro_thm:item3}
for every $\varepsilon \in (0,\infty)$ there exist $\mathscr c_0,\mathscr c_1 \in (0,\infty)$ such that for all $n \in \N$ it holds that
\begin{equation}
\mathscr c_0  n^{-2r_{\varepsilon,0}} 
\leq
\Exp{\f(\Theta_n)} - \f(\vartheta) 
\leq
\mathscr c_1 n^{-2r_{\varepsilon,1}}.
\end{equation}
\end{enumerate}
\end{athm}
\begin{aproof}
	\Nobs[note] that
	Jentzen et al.~\cite[Theorem~1.1]{JentzenvonWurstemberger2020}
	\proves[ep]
	\cref{intro_thm:item1,intro_thm:item2,intro_thm:item3}.
\end{aproof}
\endgroup

\subsection{Convergence rates for SGD for coercive objective functions}

The statement and the proof of the next result, \cref{SGD_intro} below, 
can be found in Jentzen et al.~\cite[Theorem~1.1]{Jentzen18strong}.

\cfclear
\begingroup
\providecommand{\d}{}
\renewcommand{\d}{\defaultParamDim}
\providecommand{\F}{}
\renewcommand{\F}{\defaultStochLoss}
\providecommand{\G}{}
\renewcommand{\G}{\defaultStochGradient}
\providecommand{\f}{}
\renewcommand{\f}{\defaultLossFunction}
\begin{athm}{theorem}{SGD_intro}
Let $\d \in \N$, $p, \alpha,\kappa,c \in (0,\infty)$, $ \nu \in (0,1)$, $q=\min(\{2,4,6,\dots\} \cap [p,\infty))$, $\xi, \vartheta \in \R^\d$,
let $ ( \Omega , \mathcal{F}, \P) $ be a probability space, 
let $(S, \mathcal{S})$ be a measurable space, 
let $X_{n} \colon \Omega \to S$, $n\in\N$, be i.i.d.\ random variables, 
let $\F = ( \F(\theta,x) )_{\theta \in \R^\d, x \in S} \colon \R^\d \times S \to \R$ 
be $(\mathcal{B}(\R^\d) \otimes \mathcal{S})$/$\mathcal{B}(\R)$-measurable, 
assume for all $x \in S$ that $(\R^\d \ni \theta \mapsto \F(\theta,x) \in \R) \in C^1(\R^\d, \R)$, 
assume for all $\theta \in \R^\d$ that
\begin{equation}
\label{SGD_intro:assumption2}
\bExp{\abs{\F(\theta,X_1)} + \pnorm2{(\nabla_\theta \F)(\theta, X_1)}}<\infty,
\end{equation}
\vspace{-.6cm}
\begin{equation}
\label{SGD_intro:assumption3}
\scp[\big]{\theta - \vartheta, \Exp{(\nabla_\theta \F)(\theta, X_1)} }   \geq c\max\bcu{\pnorm2{\theta - \vartheta}^2, \pnorm2{\Exp{(\nabla_\theta \F)(\theta, X_1)}}^2 },  
\end{equation}
\begin{equation}
\label{SGD_intro:assumption4}
\andq
\bExp{\pnorm2{(\nabla_\theta \F)(\theta, X_1) - \Exp{(\nabla_\theta \F)(\theta, X_1)}}^q } \leq \kappa \bpr{1+ \pnorm2{\theta}^q},
\end{equation}
let $\f \colon \R^\d\to\R$ satisfy for all $\theta\in\R^\d$ that
$\f(\theta)=\Exp{\F(\theta,X_1)}$,
and let $\Theta \colon \N_0 \times \Omega \to \R^\d$ be the stochastic process which satisfies for all $n \in \N$ that 
\begin{equation}
\label{SGD_intro:assumption5}
\Theta_0 = \xi   \qquad \text{ and } 
\qquad \Theta_n = \Theta_{n-1} - \tfrac{\alpha}{n^\nu}(\nabla_\theta \F)(\Theta_{n-1},X_n)
\end{equation}
\cfload.
Then 
\begin{enumerate}[(i)]
\item \label{SGD_intro:item1}
it holds that $\bcu{\theta \in \R^\d \colon \f(\theta) = \inf\nolimits_{w \in \R^\d}\f(w)} 
= \{\vartheta\}$ 
and 
\item \label{SGD_intro:item2}
there exists $\mathscr c \in \R$ such that for all $n \in \N$ it holds that
\begin{equation}
\bigl(\bExp{ \pnorm2{\Theta_n-\vartheta}^p }\bigr)^{\nicefrac{1}{p}} \leq \mathscr c n^{-\nicefrac{\nu}{2}}.
\end{equation}
\end{enumerate} 
\end{athm}
\begin{aproof}
	\Nobs that
	Jentzen et al.~\cite[Theorem~1.1]{Jentzen18strong}
	\proves[ep]
	\cref{SGD_intro:item1,SGD_intro:item2}.
\end{aproof}
\endgroup

\subsection{Measurability of SGD processes}
\label{sect:meas_sgd}

\begin{athm}{lemma}{measurability_gradient}
Let 
	$\defaultParamDim, \mathbf{d} \in \N$, 
let 
	$\defaultStochLoss = \allowbreak (\defaultStochLoss(\theta,x))_{(\theta,x) \in \R^\defaultParamDim \times \R^{\mathbf{d}}} \colon \allowbreak \R^\defaultParamDim \times \R^{\mathbf{d}} \to \R$
be differentiable,
and
let 
	$\defaultStochGradient = (\defaultStochGradient_1,\dots,\defaultStochGradient_{\defaultParamDim}) \colon \R^\defaultParamDim \times \R^{\mathbf{d}} \to \R^\defaultParamDim$
satisfy for all 
	$\theta\in \R^{\defaultParamDim}$,
	$x \in \R^{\mathbf{d}}$
that
\begin{equation}
	\defaultStochGradient(\theta,x) = (\nabla_\theta \defaultStochLoss)(\theta, x).
\end{equation}
Then $\defaultStochGradient$ is measurable.
\end{athm}

\begin{aproof}
Throughout this proof, 
let $e_1, e_2, \ldots, e_{\defaultParamDim} \in \R^{\defaultParamDim}$ satisfy 
\begin{equation}
	e_1 = (1, 0, \ldots, 0), \qquad
	e_2 = (0, 1, \ldots, 0), \qquad
	\ldots, \qquad
	e_{\defaultParamDim} = (0, 0, \ldots, 1)
\end{equation}
and
for every 
	$i \in \{1, 2, \dots, \defaultParamDim\}$, 
	$n \in \N$
let 
	$g_{i, n} \colon \R^\defaultParamDim \times \R^{\mathbf{d}} \to \R$
satisfy for all $\theta \in \R^\defaultParamDim$, $x \in \R^{\mathbf{d}}$ that
\begin{equation}
	g_{i,n}(\theta, x) = n [\defaultStochLoss(\theta + \tfrac{1}{n}e_i, x) - \defaultStochLoss(\theta, x)].
\end{equation}
\Nobs that 
\enum{
	the fact that $\defaultStochLoss$ is measurable;
}
\proves that
for all 
	$i \in \{1, 2, \dots, \defaultParamDim\}$, 
	$n \in \N$
it holds that 
$g_{i,n}$ is measurable.
\Moreover 
\enum{
	the fact that $\defaultStochLoss$ is differentiable;
}
\proves that
for all 
	$\theta \in \R^\defaultParamDim$,
	$x \in \R^{\mathbf{d}}$,
	$i \in \{1, 2, \dots, \defaultParamDim\}$,
it holds that
\begin{equation}
	\defaultStochGradient_i(\theta,x) = \limsup_{n \to \infty} g_{i,n}(\theta,x).
\end{equation}
Combining this with 
	the fact that 
	for all 
		$i \in \{1, 2, \dots, \defaultParamDim\}$, 
		$n \in \N$
	it holds that 
	$g_{i,n}$ is measurable
	and, \eg,
	\cite[Theorem 1.92]{Klenke14}
\proves that
for all 
	$i \in \{1, 2, \dots, \defaultParamDim\}$
it holds that 
$\defaultStochGradient_i$ is measurable.
This and, \eg, \cite[Theorem 1.90]{Klenke14}
\prove that
$\defaultStochGradient$ is measurable.
\end{aproof}

\cfclear
\begin{athm}{cor}{cor:SGD_stochastic_process}
Let $\defaultParamDim, \mathbf{d} \in \N$, 
let $\defaultStochLoss \colon \R^\defaultParamDim \times \R^{\mathbf{d}} \to \R$ be differentiable,
let $(\gamma_n)_{n \in \N} \subseteq [0,\infty)$, 
let $(J_n)_{n \in \N} \subseteq \N$, 
let $(\Omega, \mathcal{F}, \P)$ be a probability space, 
let $\xi \colon \Omega \to \R^\defaultParamDim$ be a random variable, 
for every $n,j \in \N$ let $X_{n,j} \colon \Omega \to \R^{\mathbf{d}}$ be a random variable,
and let $\Theta \colon \N_0 \times \Omega \to \R^\defaultParamDim$ be the \SGD\ process\cfadd{def:SGD} 
for the loss function $\defaultStochLoss$ with learning rates $(\gamma_n)_{n \in \N}$, 
initial value $\xi$, batch sizes $(J_n)_{n \in \N}$, and data $(X_{n,j})_{(n,j)\in \N^2}$
\cfload.
Then $\Theta$ is a stochastic process.
\end{athm}

\begin{aproof}
\Nobs that
\enum{
	\eqref{def:SGD:eq1};
	\cref{meas_grad};
	the fact that $\xi$ is a random variable;
	induction
}
\prove that for all $n \in \N_0$ it holds that $\Theta_n$ is a random variable.
\end{aproof}

\section{Explicit midpoint optimization}
\label{sect:stoch_expl_midpoint}

In this section we introduce the stochastic version of the explicit midpoint \GD\ optimization method from \cref{sect:determ_expl_midpoint}.

\defmidpointSGD
\algDescrSGD

An implementation of the explicit midpoint \SGD\ optimization method 
in {\sc PyTorch} is given in \cref{code:midpoint_sgd}.

\filelisting{code:midpoint_sgd}{code/optimization_methods/midpoint_sgd.py}{{\sc Python} code implementing the explicit midpoint \SGD\ optimization
	method in {\sc PyTorch}}

\section{Momentum optimization}
\label{sect:momentum}

In this section we introduce the stochastic version of the  momentum \GD\ optimization method from \cref{sect:determ_momentum} (cf.\ Polyak~\cite{Polyak64} and, \eg, \cite{KingmaBa14,Dozat16}).

\defmomentum 
\algDescrMomentum

An implementation in {\sc PyTorch} of the momentum \SGD\ optimization method
as described in \cref{def:momentum} above
is given in \cref{code:momentum_sgd}.
This code produces a plot which illustrates how different choices of the 
momentum decay rate and of the learning rate influence the progression of the
the loss during the training of a simple \ann\ with a single hidden layer, learning
an approximation of the sine function.
We note that while \cref{code:momentum_sgd} serves to illustrate a concrete
implementation of the momentum \SGD\ optimization method, for applications
it is generally much preferable to use {\sc PyTorch}'s built-in implementation
of the momentum \SGD\ optimization method in the
{\tt torch.optim.SGD} optimizer, rather than implementing it from scratch.

\filelisting{code:momentum_sgd}{code/optimization_methods/momentum_sgd.py}{{\sc Python} code implementing the \SGD\ optimization method with classical momentum in {\sc PyTorch}}

\begin{figure}[!ht]
	\centering
	\includegraphics[width=1\linewidth]{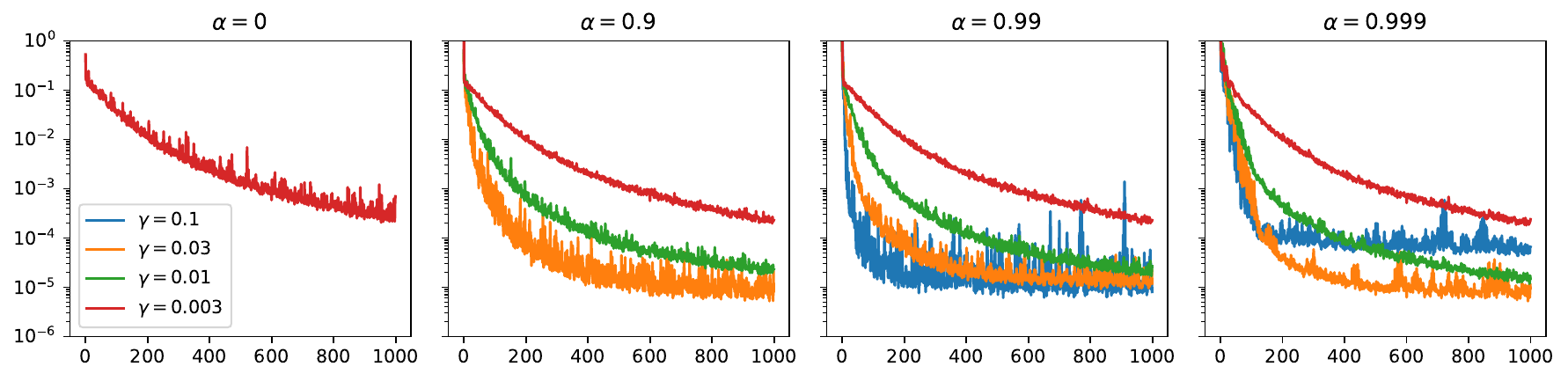}
	\caption{\label{fig:sgd_momentum}
		A plot showing the influence of the momentum decay rate and learning rate
		on the loss during the training of an \ann\ using the
		\SGD\ optimization method with classical momentum
	}
\end{figure}

\subsection{Alternative definitions}

In this section we introduce the stochastic versions of the alternative momentum \GD\ optimization methods from \cref{sec:momentum_GD_alternatives}.

\defmomentumTwo
\algDescrMomentumTwo

\defmomentumThree
\algDescrMomentumThree

\defmomentumFour
\algDescrMomentumFour

\subsection{Bias-adjusted momentum optimization}

In this section we introduce the stochastic version of the bias-adjusted momentum \GD\ optimization method from \cref{sect:determ_momentum_bias}.

\defStochMomentumBias
\algDescrMomentumBias

An implementation of the bias-adjusted momentum \SGD\ optimization method
in {\sc PyTorch} is given in \cref{code:momentum_sgd_bias_adj}.

\filelisting{code:momentum_sgd_bias_adj}{code/optimization_methods/momentum_sgd_bias_adj.py}{{{\sc Python} code implementing the bias-adjusted momentum \SGD\ optimization	method in {\sc PyTorch}}}

\section{Nesterov accelerated momentum optimization}
\label{sect:nesterov}

In this section we introduce the stochastic version of the Nesterov accelerated \GD\ optmization method from \cref{sect:determ_nesterov} (cf.\ \cite{Nesterov83,sutskever2013importance}).

\defStochNesterov
\algDescrNesterov

An implementation of the Nesterov accelerated \SGD\ optimization method
in {\sc PyTorch} is given in \cref{code:nesterov_sgd}.

\filelisting{code:nesterov_sgd}{code/optimization_methods/nesterov_sgd.py}{{{\sc Python} code implementing the Nesterov accelerated \SGD\ optimization	method in {\sc PyTorch}}}

\subsection{Alternative definitions}

In this section we introduce the stochastic versions of the alternative Nesterov accelerated \GD\ optimization methods from \cref{sect:nesterov_alternatives}.

\defStochNesterovTwo
\algDescrNesterovTwo

\defStochNesterovThree
\algDescrNesterovThree

\defStochNesterovFour
\algDescrNesterovFour

\subsection{Bias-adjusted Nesterov accelerated momentum optimization}

In this section we introduce the stochastic version of the bias-adjusted Nesterov accelerated \GD\ optimization method from \cref{sect:determ_nesterov_bias}.

\defNesterovBias
\algDescrNesterovBias

\subsection{Shifted representations}

In this section we introduce the stochastic versions of the shifted representations of the Nesterov accelerated \GD\ optimization methods from \cref{sect:determ_nesterov_shifted}.

\subsubsection{Shifted representation for the first version of Nesterov accelerated momentum optimization}
\defNesterovAlt
\algDescrNesterovAlt

\subsubsection{Shifted representation for the second version of Nesterov accelerated momentum optimization}

\defNesterovAltTwo
\algDescrNesterovAltTwo

\subsubsection{Shifted representation for the third version of Nesterov accelerated momentum optimization}
\defNesterovAltThree
\algDescrNesterovAltThree

\subsubsection{Shifted representation for the fourth version of Nesterov accelerated momentum optimization}
\defNesterovAltFour
\algDescrNesterovAltFour

\subsubsection{Shifted representation for the bias-adjusted Nesterov accelerated momentum optimization}
\defNesterovAltBias
\algDescrNesterovAltBias

\subsection{Simplified Nesterov accelerated momentum optimization}

In this section we introduce the stochastic version of the simplified Nesterov accelerated \GD\ optimization method from \cref{sect:determ_nesterov_simple}.

\defNesterovSimple
\algDescrNesterovSimple
	
The simplified Nesterov accelerated \SGD\ optimization method as described
in \cref{def:nesterov_simple} is implemented in {\sc PyTorch} in the form of the
{\tt torch.optim.SGD} optimizer with the {\tt nesterov=True} option.

\section{Adagrad optimization}
\label{sect:adagrad}

In this section we introduce the stochastic version of the \Adagrad\ \GD\ optimization method from \cref{sect:determ_adagrad} (cf.\ Duchi et al.~\cite{DuchiHazanSinger11}).

\defAdagrad
\algDescrAdagrad

An implementation in {\sc PyTorch} of the \Adagrad\ \SGD\ optimization method as 
described in \cref{def:adagrad} above is given in \cref{code:adagrad}.
The \Adagrad\ \SGD\ optimization method as described in \cref{def:adagrad} above
is also available in {\sc PyTorch} in the form of the built-in {\tt torch.optim.Adagrad} 
optimizer (which, for applications, is generally much preferable to
implementing it from scratch).

\filelisting{code:adagrad}{code/optimization_methods/adagrad.py}{{{\sc Python} code implementing the \Adagrad\ \SGD\ optimization method in {\sc PyTorch}}}

\section{RMSprop optimization}
\label{sect:RMSprop}

In this section we introduce the stochastic version of the  \RMSprop\ \GD\ optimization method from \cref{sect:determ_RMSprop}
(cf.\
Hinton et al.~\cite{HintonSlides}).

\defRMSprop
\algDescrRMSprop

An implementation in {\sc PyTorch} of the \RMSprop\ \SGD\ optimization method as 
described in \cref{def:rmsprop} above is given in \cref{code:rmsprop}.
The \RMSprop\ \SGD\ optimization method as described in \cref{def:rmsprop} above
is also available in {\sc PyTorch} in the form of the built-in {\tt torch.optim.RMSprop} 
optimizer (which, for applications, is generally much preferable to
implementing it from scratch).

\filelisting{code:rmsprop}{code/optimization_methods/rmsprop.py}{{{\sc Python} code implementing the \RMSprop\ \SGD\ optimization method in {\sc PyTorch}}}

\subsection{Bias-adjusted RMSprop optimization}
\label{sect:RMSprop_bias}

\defRMSpropBias

An implementation in {\sc PyTorch} of the bias-adjusted 
\RMSprop\ \SGD\ optimization method as 
described in \cref{def:RMSprop_bias} above is given in \cref{code:rmsprop_bias}.

\filelisting{code:rmsprop_bias}{code/optimization_methods/rmsprop_bias_adj.py}{{{\sc Python} code implementing the bias-adjusted \RMSprop\ \SGD\ optimization method in {\sc PyTorch}}}

\section{Adadelta optimization}
\label{sect:adadelta}

In this section we introduce the stochastic version of the Adadelta \GD\ optimization method from \cref{sect:determ_adadelta} (cf.\ Zeiler~\cite{Zeiler12}).

\defAdadelta
\algDescrAdadelta

An implementation in {\sc PyTorch} of the Adadelta \SGD\ optimization method as
described in \cref{def:adadelta} above is given in \cref{code:adadelta}.
The Adadelta \SGD\ optimization method as described in \cref{def:adadelta} above
is also available in {\sc PyTorch} in the form of the built-in {\tt torch.optim.Adadelta} 
optimizer (which, for applications, is generally much preferable to
implementing it from scratch).

\filelisting{code:adadelta}{code/optimization_methods/adadelta.py}{{{\sc Python} code implementing the Adadelta \SGD\ optimization	method in {\sc PyTorch}}}

\section{Adam optimization}
\label{sect:adam}

In this section we introduce the stochastic version of the  \Adam\ \GD\ optimization method from \cref{sect:determ_adam}
(cf.\ Kingma \& Ba~\cite{KingmaBa14}).

\defAdam
\algDescrAdam

\begin{athm}{remark}{Adam_rem}
In Kingma \& Ba~\cite{KingmaBa14}
it is proposed to choose
\begin{equation}
\begin{split} 
	0.001 = \gamma_1 = \gamma_2 = \ldots, \qquad
	0.9 = \alpha_1 = \alpha_2 = \ldots, \qquad
	0.999 = \beta_1 = \beta_2 = \ldots, \qquad
\end{split}
\end{equation}
and
$10^{-8} = \varepsilon$
as default values for 
$
  ( \gamma_n )_{ n \in \N } 
  \allowbreak 
  \subseteq [0,\infty) 
$, 
$ 
  ( \alpha_n )_{ n \in \N } \subseteq [0,1] 
$, 
$
  ( \beta_n )_{ n \in \N } \subseteq [0,1] 
$,
$
  \varepsilon \in (0,\infty) 
$
in \cref{def:adam}.
\end{athm}

An implementation in {\sc PyTorch} of the \Adam\ \SGD\ optimization method as
described in \cref{def:adam} above is given in \cref{code:adam}.
The \Adam\ \SGD\ optimization method as described in \cref{def:adam} above
is also available in {\sc PyTorch} in the form of the built-in {\tt torch.optim.Adam} 
optimizer (which, for applications, is generally much preferable to
implementing it from scratch).

\filelisting{code:adam}{code/optimization_methods/adam.py}{{{\sc Python} code implementing the \Adam\ \SGD\ optimization method in {\sc PyTorch}}}

\begin{figure}[!ht]
	\centering
	\includegraphics[width=0.9\linewidth]{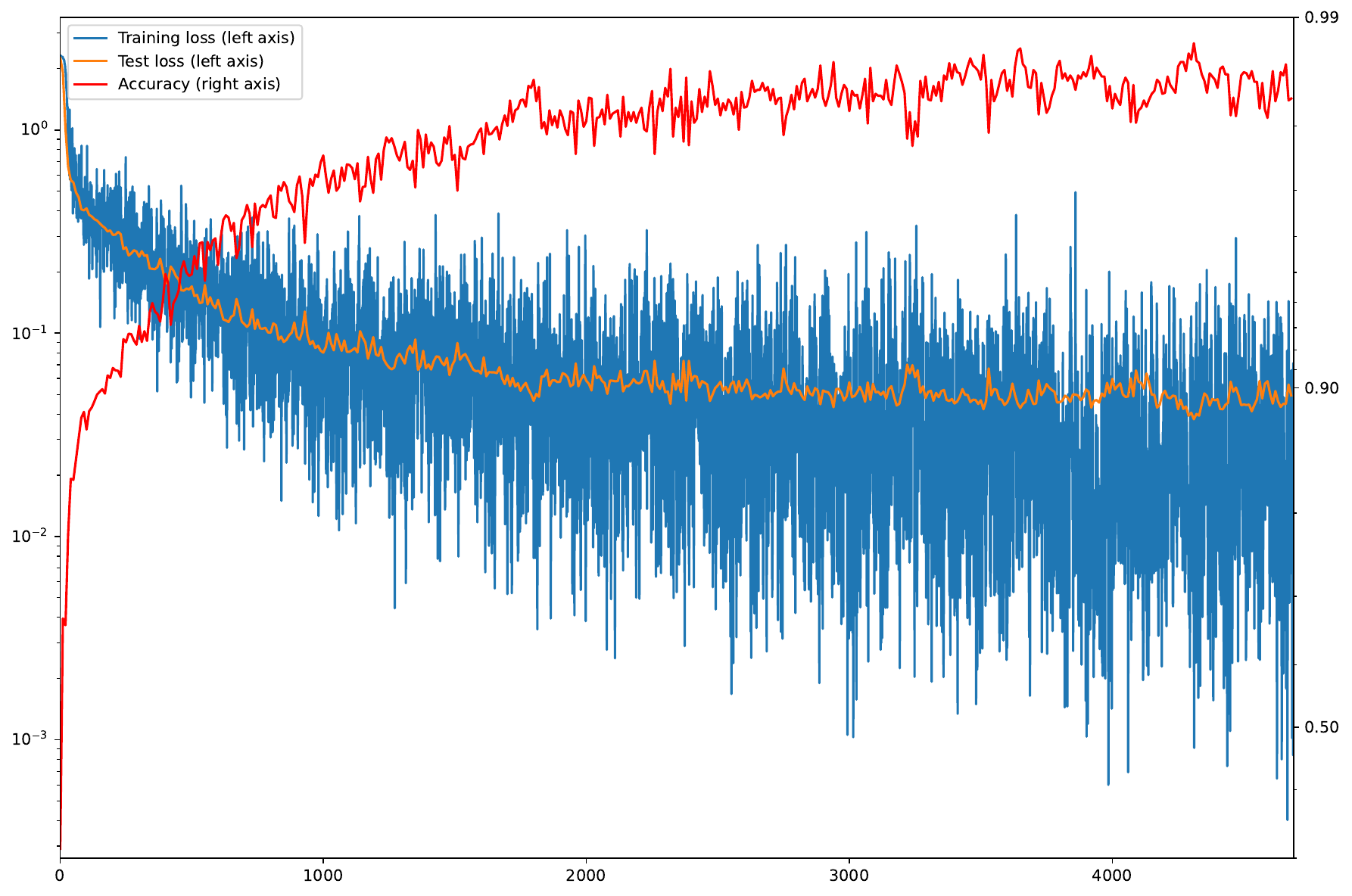}
	\caption{\label{fig:mnist}
	The plot produced by \cref{code:mnist},
	showing the average loss over each minibatch
	used during training
	(training loss) as well as the average loss over the test set
	and the accuracy of the model's predictions over the course of
	the training.}
\end{figure}

Whereas \cref{code:adam} and the other source codes presented
in this chapter so far served mostly
to elucidate the definitions of the various optimization methods introduced
in this chapter by giving example implementations, in \cref{code:mnist} 
we demonstrate how an actual machine learning problem might be solved using the
built-in functionality of {\sc PyTorch}. This code trains a neural network
with 3 convolutional layers and 2 fully connected layers (with each
hidden layer followed by a \ReLU\ activation function) on the MNIST dataset (introduced in Bottou et al.~\cite{bottou1994mnist}), which
consists of $28\times 28$ pixel grayscale images of handwritten digits from 0 to 9
and the corresponding labels and is one of the most commonly used benchmarks
for training machine learning systems in the literature. 
\Cref{code:mnist}  uses the cross-entropy loss function
and the \Adam\ \SGD\ optimization method and outputs a graph showing the progression
of the average loss on the training set and on a test set that is not used for training
as well as the accuracy of the model's predictions over the course of the training,
see \cref{fig:mnist}.

\filelisting{code:mnist}{code/mnist.py}{
	{{\sc Python} code training an \ann\ on the MNIST dataset in {\sc PyTorch}.
	This code produces a plot showing the progression of the average loss
	on the test set and the accuracy of the model's predictions, see
	\cref{fig:mnist}.}
}

\begin{figure}[!ht]
	\centering
	\includegraphics[width=0.85\linewidth]{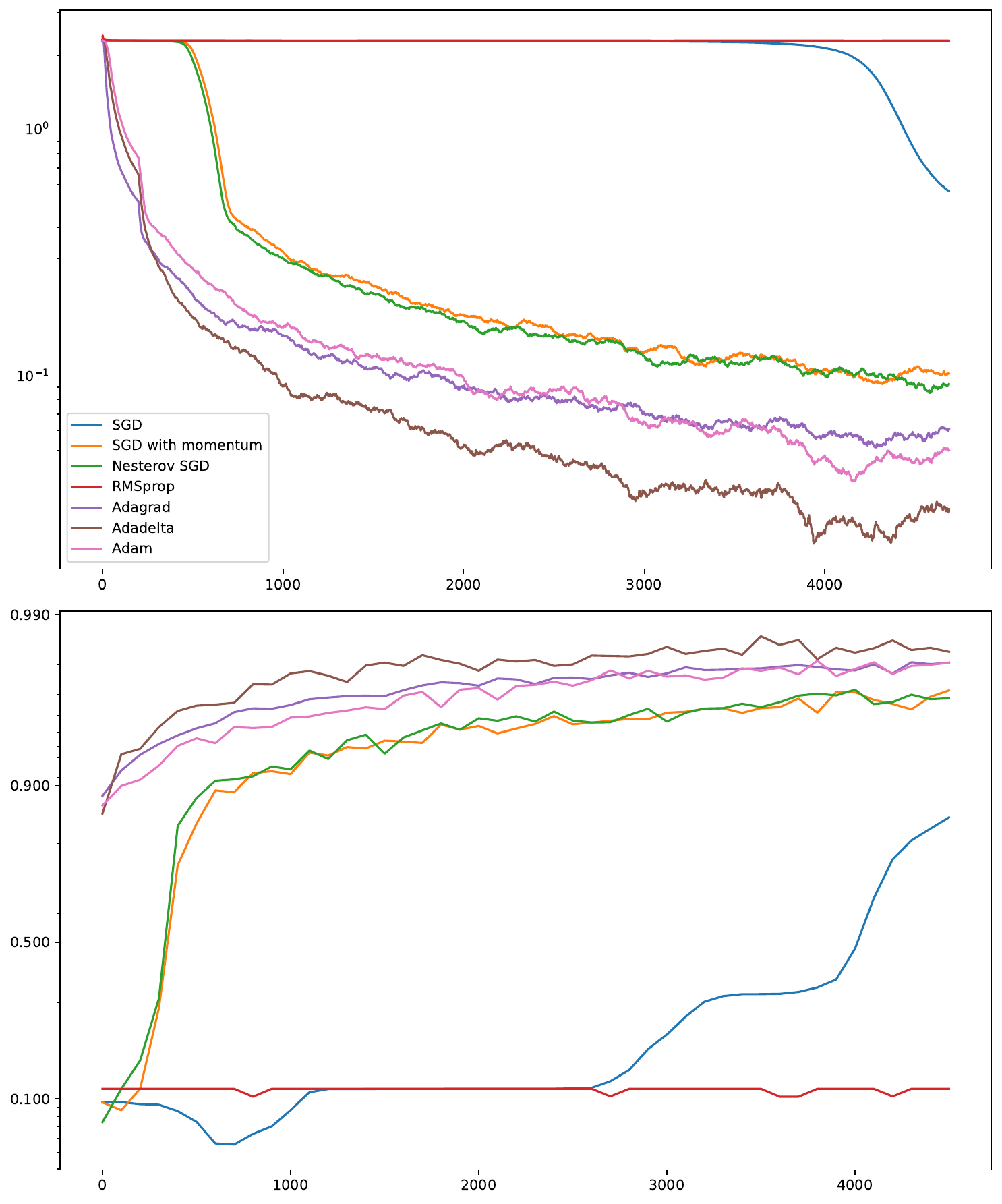}
	\caption{\label{fig:mnist_optim}
	The plots produced by \cref{code:mnist_optim}.
	The upper plot shows the progression of the training loss
	during the training of the \anns. More precisely, each line
	shows a moving average of the training loss over 200 minibatches
	during the training of an \ann\ with the corresponding optimization method.
	The lower plot shows the accuracy of the \ann's predictions
	on the test set over the course of the training with each optimization method.}
\end{figure}

\Cref{code:mnist_optim} compares the performance of several of the optimization
methods introduced in this chapter, namely the plain vanilla
\SGD\ optimization method introduced in \cref{def:SGD},
the momentum \SGD\ optimization method introduced in \cref{def:momentum},
the simplified Nesterov accelerated \SGD\ optimization method introduced 
in \cref{def:nesterov_simple}, the \Adagrad\ \SGD\ optimization method introduced
in \cref{def:adagrad}, the \RMSprop\ \SGD\ optimization method introduced
in \cref{def:rmsprop}, the Adadelta \SGD\ optimization method introduced
in \cref{def:adadelta}, and the \Adam\ \SGD\ optimization method introduced
in \cref{def:adam}, during training of an \ann\ on the MNIST dataset.
The code produces two plots showing the progression of the 
training loss as well as the accuracy of the model's predictions
on the test set, see \cref{fig:mnist_optim}.
Note that this compares the performance of the optimization methods
only on one particular problem and without any efforts towards
choosing good hyperparameters for the considered optimization methods.
Thus, the results are not necessarily representative of the performance
of these optimization methods in general.

\filelisting{code:mnist_optim}{code/mnist_optim.py}{
	{{\sc Python} code comparing the performance of several optimization
	methods during training of an \ann\ on the MNIST dataset. See
	\cref{fig:mnist_optim} for the plots produced by this code.}
}

\subsection{Adamax optimization}
\label{sec:adamax}

In this section we introduce the stochastic version of the Adamax \GD\ optimization method from \cref{sect:determ_adamax}
(cf.\ Kingma \& Ba~\cite{KingmaBa14}).

\defAdamax
\algDescrAdamax

\section{Nadam optimization}
\label{sect:nadam}

In this section we introduce the stochastic version of the \Nadam\ \GD\ optimization method from \cref{sect:determ_nadam}
(cf.\ Dozat~\cite{Dozat15,Dozat16}).

\defNadam
\algDescrNadam

\subsection{Simplified Nadam optimization}
\label{sect:nadam_simple}

In this section we introduce the stochastic version of the simplified \Nadam\ \GD\ optimization method from \cref{sect:determ_simplified_nadam}.

\defSimplifiedNadam
\algDescrSimplifiedNadam

\subsection{Nadamax optimization}
\label{sec:nadamax}

In this section we introduce the stochastic version of the Nadamax \GD\ optimization method from \cref{sect:determ_adamax}
(cf.\ Dozat~\cite{Dozat15,Dozat16}).

\defNadamax
\algDescrNadamax

\section{AdamW optimization}
\label{sect:adamW}

In this section we introduce the stochastic version of the \AdamW\ \GD\ optimization method from \cref{sect:determ_adamW}
(cf.\ Loshchilov \& Hutter~\cite{Loshchilov2019}).

\defAdamW
\algDescrAdamW

\subsection{Adam with $L^2$-regularization optimization}
\label{sec:adamLtwo}

In this section we introduce the stochastic version of the \Adam\ \GD\ optimization method with $L^2$-regularization from \cref{sect:determ_adamLtwo}
(cf.\ Loshchilov \& Hutter~\cite{Loshchilov2019}).

\defAdamLtwo
\algDescrAdamLtwo

\begin{athm}{remark}{AdamLtwo_rem}
In the {\sc PyTorch} implementation of the \Adam\ \SGD\ optimization method with $L^2$-regularization in \cref{def:adamLtwo}
the default value for $\lambda \in \R$ is $\lambda = 0.01$.
\end{athm}

\todoc{Add default value for $\lambda$: 0.01 in Pytorch: DONE}

\section{Shampoo optimization}
\label{sect:shampoo}

In this section we introduce the stochastic version of the Shampoo \GD\ optimization method from \cref{sect:determ_shampoo}
(cf.\ Gupta et al.~\cite{Gupta2018}).

\defShampoo
\algDescrShampoo

\section{Muon optimization}
\label{sect:muon}

In this section we introduce the stochastic version of the Muon \GD\ optimization method from \cref{sect:determ_muon}
(cf.\ Jordan et al.~\cite{Jordan2024}).

\defidealMuon
\algDescridealMuon

\defMuon
\algDescrMuon

\section{AMSGrad optimization}
\label{sect:amsgrad}

In this section we introduce the stochastic version of the AMSGrad \GD\ optimization method from \cref{sect:determ_AMSgrad}
(cf.\ Reddi et al.~\cite{Reddi2019}).

\defAMSGrad
\algDescrAMSGrad

\begin{athm}{remark}{analysis_higher_order_methods}[Analysis of accelerated \SGD-type optimization methods]
In the literature there are numerous research articles which study the accelerated \SGD-type optimization methods reviewed in this chapter.
In particular, 
we refer, \eg, to
\cite{Gess2023,Loizou2020,Liu2018,sutskever2013importance,Qian99} and the references therein for articles on \SGD-type optimization methods with momentum
and
we refer, \eg, to
\cite{GodichonBaggioni2023,Defossez2020,Zhang2022,Reddi2019,Ma2020}
and the references therein for articles on adaptive \SGD-type optimization methods.
\end{athm}

\todoc{Reintroduce section above and add more methods here (ARNULF)}

\section{Compact summary of SGD optimization methods}

In this section we provide an overview over all \SGD-type optimization methods considered in this chapter.
Roughly speaking, in this summary we provide for each considered method only the iteration step of the respective pseudo-code.
The formulas in this summary make use of the componentwise operations in \cref{def:componentwise_operations}.

\algorithmSummaryDescription{\SGD\ optimization method}{def:SGD}{\iterationStepSGD}
\algorithmSummaryDescription{Explicit midpoint \SGD\ optimization method}{def:midpointSGD}{\iterationStepmidpointSGD}
\algorithmSummaryDescription{Momentum \SGD\ optimization method}{def:momentum}{\iterationStepMomentum}
\algorithmSummaryDescription{Momentum \SGD\ optimization method (\second version)}{def:momentum_two}{\iterationStepMomentumTwo}
\algorithmSummaryDescription{Momentum \SGD\ optimization method (\third version)}{def:momentum_three}{\iterationStepMomentumThree}
\algorithmSummaryDescription{Momentum \SGD\ optimization method (\fourth version)}{def:momentum_four}{\iterationStepMomentumFour}
\algorithmSummaryDescription{Bias-adjusted momentum \SGD\ optimization method}{def:momentum_bias}{\iterationStepMomentumBias}
\algorithmSummaryDescription{Nesterov accelerated \SGD\ optimization method}{def:nesterov}{\iterationStepNesterov}
\algorithmSummaryDescription{Nesterov accelerated \SGD\ optimization method (\second version)}{def:nesterov_two}{\iterationStepNesterovTwo}
\algorithmSummaryDescription{Nesterov accelerated \SGD\ optimization method (\third version)}{def:nesterov_three}{\iterationStepNesterovThree}
\algorithmSummaryDescription{Nesterov accelerated \SGD\ optimization method (\fourth version)}{def:nesterov_four}{\iterationStepNesterovFour}
\algorithmSummaryDescription{Shifted Nesterov accelerated \SGD\ optimization method}{def:nesterov_shifted}{\iterationStepNesterovAlt}
\algorithmSummaryDescription{Shifted Nesterov accelerated \SGD\ optimization method (\second version)}{def:nesterov_shifted_two}{\iterationStepNesterovAltTwo}
\algorithmSummaryDescription{Shifted Nesterov accelerated \SGD\ optimization method (\third version)}{def:nesterov_shifted_three}{\iterationStepNesterovAltThree}
\algorithmSummaryDescription{Shifted Nesterov accelerated \SGD\ optimization method (\fourth version)}{def:nesterov_shifted_four}{\iterationStepNesterovAltFour}
\algorithmSummaryDescription{Bias-adjusted Nesterov accelerated \SGD\ optimization method}{def:nesterov_bias}{\iterationStepNesterovBias}
\algorithmSummaryDescription{Shifted bias-adjusted Nesterov accelerated \SGD\ optimization method}{def:nesterov_shifted_bias}{\iterationStepNesterovAltBias}
\algorithmSummaryDescription{Simplified Nesterov accelerated \SGD\ optimization method}{def:nesterov_simple}{\iterationStepNesterovSimple}
\algorithmSummaryDescription{\Adagrad\ \SGD\ optimization method}{def:adagrad}{\iterationStepAdagrad}
\algorithmSummaryDescription{\RMSprop\ \SGD\ optimization method}{def:rmsprop}{\iterationStepRMSprop}
\algorithmSummaryDescription{Bias-adjusted \RMSprop\ \SGD\ optimization method}{def:RMSprop_bias}{\iterationStepRMSpropBias}
\algorithmSummaryDescription{Adadelta \SGD\ optimization method}{def:adadelta}{\iterationStepAdadelta}
\algorithmSummaryDescription{\Adam\ \SGD\ optimization method}{def:adam}{\iterationStepAdam}
\algorithmSummaryDescription{Adamax \SGD\ optimization method}{def:adamax}{\iterationStepAdamax}
\algorithmSummaryDescription{\Nadam\ \SGD\ optimization method}{def:nadam}{\iterationStepNadam}
\algorithmSummaryDescription{Simplified \Nadam\ \SGD\ optimization method}{def:simplified_nadam}{\iterationStepSimplifiedNadam}
\algorithmSummaryDescription{Nadamax \SGD\ optimization method}{def:nadamax}{\iterationStepNadamax}
\algorithmSummaryDescription{\Adam\ \SGD\ optimization method with $L^2$-regularization}{def:adamLtwo}{\iterationStepAdamLtwo}
\algorithmSummaryDescription{Shampoo \SGD\ optimization method}{def:Shampoo}{\iterationStepShampoo}
\algorithmSummaryDescription{Idealized \Muon\ \SGD\ optimization method}{def:idealMuon}{\iterationStepidealMuon}
\algorithmSummaryDescription{\Muon\ \SGD\ optimization method}{def:Muon}{\iterationStepMuon}
\algorithmSummaryDescription{\AdamW\ \SGD\ optimization method}{def:adamW}{\iterationStepAdamW}
\algorithmSummaryDescription{AMSgrad \SGD\ optimization method}{def:AMSGrad}{\iterationStepAMSGrad}

%% file: parts/Backpropagation.tex
\setlength{\headheight}{27.11469pt}
\addtolength{\topmargin}{-12.61469pt}

\cchapter{Backpropagation}{chapter:backprop}

In \cref{chapter:deterministic,chapter:stochastic} we reviewed common deterministic and stochastic \GD-type optimization methods used for the training of \anns.
The specific implementation of such methods requires efficient explicit computations of gradients.
The most popular and somehow most natural method to explicitly compute such gradients in the case of the training of \anns\ is the \emph{backpropagation} method.
In this chapter we derive and present this method in detail.

Further material on the backpropagation method can, \eg, be found in the books and overview articles 
\cite{Griewank2008},
\cite[Section 11.7]{alpaydin2020introduction},
\cite[Section 6.2.3]{Calin2020},
\cite[Section 3.2.3]{Caterini2018},
\cite[Section 5.6]{Deisenroth2020}, and 
\cite[Section 20.6]{shalev2014understanding}.

\section{Backpropagation for parametric functions}

\cfclear
\begingroup
\providecommand{\d}{}
\renewcommand{\d}{\defaultParamDim}
\providecommand{\Targetfunction}{}
\renewcommand{\Targetfunction}{f}
\providecommandordefault{\Function}{F}

\begin{prop}[Backpropagation for parametric functions]
\label{backprop_1}
Let
	$L \in \N$,
	$l_0, l_1, \ldots, l_L,\allowbreak  \d_1,\allowbreak \d_2, \allowbreak \ldots, \d_L\in \N$,
for every 
	$k \in \{1, 2, \ldots, L\}$
let
$
	\Function_k 
= 
	(\Function_k(\thetavar{k}, \xvar{k-1}))_{(\thetavar{k}, \xvar{k-1}) \in \R^{\d_k} \times \R^{l_{k-1}}}
\colon \allowbreak
	\R^{\d_k} \times \R^{l_{k-1}} \to \R^{l_k}
$
be  differentiable,
for every
	$k \in \{1, 2, \ldots, L\}$
let
$
	\Targetfunction_k 
= 
	(\Targetfunction_k(\theta_{k}, \theta_{k+1}, \ldots, \allowbreak \theta_{L}, \linebreak \xvar{k-1}))_{
		(\theta_{k}, \theta_{k+1}, \ldots, \theta_{L}, \xvar{k-1}) 
		\in 
		\R^{\d_k} \times \R^{\d_{k+1}} \times \ldots \times \R^{\d_L} \times \R^{l_{k-1}}
	}
\colon 
	\R^{\d_k} \times \R^{\d_{k+1}} \times \ldots \times \R^{\d_L}  \times \R^{l_{k-1}} 
\to 
	\R^{l_L}
$
satisfy for all
	$\theta = (\theta_k, \theta_{k+1}, \ldots, \theta_L) \in \R^{\d_k} \times \R^{\d_{k+1}} \times \ldots \times \R^{\d_L}$,
	$\xvar{k-1} \in \R^{l_{k-1}}$
that
\begin{equation}
\label{backprop_1:ass1}
\begin{split} 
	\Targetfunction_k(\theta, \xvar{k-1})
=
	\bpr{
		\Function_L(\theta_L, \cdot) \circ 
		\Function_{L-1}(\theta_{L-1}, \cdot) \circ
		\ldots 
		\circ
		\Function_{k}(\theta_{k}, \cdot)
	} 
	(\xvar{k-1}),
\end{split}
\end{equation}
let
	$\vartheta = (\vartheta_1, \vartheta_{2}, \ldots, \vartheta_L) \in \R^{\d_1} \times \R^{\d_{2}} \times \ldots \times \R^{\d_L}$,
	$\fx_0 \in \R^{l_0}$, 
	$\fx_1 \in \R^{l_1}$, 
	$\dots$,
	$\fx_L \in \R^{l_L}$
satisfy for all
	$k \in \{1, 2, \ldots, L\}$
that
\begin{equation}
\label{backprop_1:ass2}
\begin{split} 
	\fx_k 
= 
	\Function_{k}(\vartheta_k, \fx_{k-1}),
\end{split}
\end{equation}
and let 
	$D_k \in \R^{l_L \times l_{k-1}}$, $k \in \{1, 2, \ldots, L+1\}$,
satisfy for all 
	$k \in \{1, 2, \ldots, L\}$
that
$
	D_{L+1}
=
	\idMatrix_{l_L}
$
and
\begin{equation}
\label{backprop_1:ass3}
\begin{split} 
	D_k
= 
	D_{k+1}
	\br*{
		\pr*{
			\frac{\partial \Function_k}{\partial \xvar{k-1}}
		}
		(\vartheta_k, \fx_{k-1})
	}
\end{split}
\end{equation}
\cfload.
Then 
\begin{enumerate}[label=(\roman *)]
\item 
\label{backprop_1:item1}
it holds for all 
	$k \in \{1, 2, \ldots, L\}$
that
$
	\Targetfunction_k 
\colon 
	\R^{\d_k} \times \R^{\d_{k+1}} \times \ldots \times \R^{\d_L}  \times \R^{l_{k-1}} 
\to 
	\R^{l_L}
$ 
is differentiable,

\item 
\label{backprop_1:item2}
it holds for all 
	$k \in \{1, 2, \ldots, L\}$
that
\begin{equation}
\label{backprop_1:concl1}
\begin{split} 
	D_k
=
	\pr*{
		\frac{\partial \Targetfunction_k}{\partial \xvar{k-1}}
	}
	((\vartheta_k, \vartheta_{k+1}, \ldots, \vartheta_L), \fx_{k-1}),
\end{split}
\end{equation}
and

\item 
\label{backprop_1:item3}
it holds for all
	$k \in \{1, 2, \ldots, L\}$
that
\begin{equation}
\label{backprop_1:concl3}
\begin{split} 
	\pr*{
		\frac{\partial \Targetfunction_1}{\partial \theta_k}
	}
	(\vartheta, \fx_0)
=
	D_{k+1}
	\br*{
		\pr*{
			\frac{\partial \Function_k}{\partial \thetavar{k}}
		}
		(\vartheta_k, \fx_{k-1})
	}
	.
\end{split}
\end{equation}
\end{enumerate}
\end{prop}

\begin{proof}[Proof of \cref{backprop_1}]
Note that 
	\cref{backprop_1:ass1},
	the fact that for all
		$k \in \N \cap (0,L)$, \allowbreak
		$(\theta_{k}, \theta_{k+1}, \ldots,\allowbreak \theta_L) \in \R^{\d_k} \times \R^{\d_{k+1}} \times \ldots \times \R^{\d_L} $,
		$\xvar{k-1} \in \R^{l_{k-1}}$
	it holds that
	\begin{equation}
	\label{backprop_1:eq00}
	\begin{split} 
		\Targetfunction_k((\theta_{k}, \theta_{k+1}, \ldots, \theta_L), \xvar{k-1})
	=
		(
			\Targetfunction_{k+1}((\theta_{k+1}, \theta_{k+2}, \ldots, \theta_L), \cdot) 
			\circ
			\Function_k(\theta_k, \cdot)
		)(\xvar{k-1}),
	\end{split}
	\end{equation}
	the assumption that for all 
		$k \in \{1, 2, \ldots, L\}$
	it holds that 
	$
		\Function_k 
	\colon 
		\R^{\d_k} \times \R^{l_{k-1}} \to \R^{l_k}
	$
	is differentiable,
	\cref{param_comp_diff}, and
	induction
imply
that
for all 
	$k \in \{1, 2, \ldots, L\}$
it holds that
\begin{equation}
\label{backprop_1:eq0}
\begin{split} 
	\Targetfunction_k 
\colon 
	\R^{\d_k} \times \R^{\d_{k+1}} \times \ldots \times \R^{\d_L}  \times \R^{l_{k-1}} 
\to 
	\R^{l_L}
\end{split}
\end{equation}
is differentiable.
This proves \cref{backprop_1:item1}.
Next we prove \eqref{backprop_1:concl1} by induction on $k \in \{L, L-1, \ldots, 1\}$.
Note that
\enum{
	\eqref{backprop_1:ass3};
	the assumption that
	$
		D_{L+1}
	=
		\idMatrix_{l_L}
	$;
	the fact that 
	$\Targetfunction_L = \Function_L$
}[assure]
that
\begin{equation}
\label{backprop_1:eq1}
\begin{split} 
	D_L
=
	D_{L+1}
	\br*{
		\pr*{
			\frac{\partial \Function_L}{\partial \xvar{L-1}}
		}
		(\vartheta_L, \fx_{L-1})
	}
=
	\pr*{
		\frac{\partial \Targetfunction_L}{\partial \xvar{L-1}}
	}
	(\vartheta_L, \fx_{L-1}).
\end{split}
\end{equation}
This establishes \eqref{backprop_1:concl1} in the base case $k = L$.
For the induction step note that 
\enum{
	\eqref{backprop_1:ass3};
	the chain rule;
	the fact that for all
		$k \in \N \cap (0,L)$,
		$\xvar{k-1} \in \R^{l_{k-1}}$
	it holds that
	\begin{equation}
	\label{backprop_1:eq1.1}
	\begin{split} 
		\Targetfunction_k((\vartheta_{k}, \vartheta_{k+1}, \ldots, \vartheta_L), \xvar{k-1})
	=
		\Targetfunction_{k+1}((\vartheta_{k+1}, \vartheta_{k+2}, \ldots, \vartheta_L), \Function_k(\vartheta_k, \xvar{k-1}))
	\end{split}
	\end{equation}
	;
}[imply]
that for all 
	$k \in \N \cap (0,L)$
with
$
	D_{k+1}
=
	\pr[\big]{
		\frac{\partial \Targetfunction_{k+1}}{\partial \xvar{k}}
	}
	((\vartheta_{k+1}, \vartheta_{k+2}, \ldots, \vartheta_L), \fx_{k})
$
it holds that
\begin{equation}
\label{backprop_1:eq2}
\begin{split} 
	&\pr*{
		\frac{\partial \Targetfunction_k}{\partial \xvar{k-1}}
	}
	((\vartheta_k, \vartheta_{k+1}, \ldots, \vartheta_L), \fx_{k-1}) \\
&=
	\pr*{
				\R^{l_{k-1}} \ni \xvar{k-1} 
				\mapsto
				\Targetfunction_k((\vartheta_{k}, \vartheta_{k+1}, \ldots, \vartheta_L), \xvar{k-1})
				\in \R^{l_L}
	}'
	( \fx_{k-1}) \\
&=
	\pr*{
				\R^{l_{k-1}} \ni \xvar{k-1} 
				\mapsto
				\Targetfunction_{k+1}((\vartheta_{k+1}, \vartheta_{k+2}, \ldots, \vartheta_L), \Function_k(\vartheta_k, \xvar{k-1}))
				\in \R^{l_L}
	}'
	( \fx_{k-1}) \\
&=
	\br*{
		\pr*{
					\R^{l_{k-1}} \ni \xvar{k} 
					\mapsto
					\Targetfunction_{k+1}((\vartheta_{k+1}, \vartheta_{k+2}, \ldots, \vartheta_L),  \xvar{k}))
					\in \R^{l_L}
		}'
		( \Function_k(\vartheta_k, \fx_{k-1}))
	} \\
&\quad\,\,
	\br*{
		\pr*{
					\R^{l_{k-1}} \ni \xvar{k-1} 
					\mapsto
					\Function_k(\vartheta_k, \xvar{k-1}))
					\in \R^{l_k}
		}'
		(\fx_{k-1}) 
	}\\
&=
	\br*{
		\pr*{
			\frac{
				\partial \Targetfunction_{k+1}
			}{
				\partial \xvar{k}
			}
		}
		((\vartheta_{k+1}, \vartheta_{k+2}, \ldots, \vartheta_L),  \fx_{k})
	} 
	\br*{
		\pr*{
			\frac{
				\partial \Function_k
			}{
				\partial \xvar{k-1}
			}
		}
		(\vartheta_k, \fx_{k-1}) 
	}\\
&=
	D_{k+1}
	\br*{
		\pr*{
			\frac{
				\partial \Function_k
			}{
				\partial \xvar{k-1}
			}
		}
		(\vartheta_k, \fx_{k-1}) 
	}
=
	D_k.
\end{split}
\end{equation}
Induction thus proves \cref{backprop_1:concl1}. 
This establishes \cref{backprop_1:item2}.
Moreover, observe that
\enum{
	\cref{backprop_1:ass1};
	\cref{backprop_1:ass2}
}[assure]
that for all
	$k \in \N \cap (0,L)$,
	$\theta_k \in \R^{l_k}$
it holds that
\begin{equation}
\label{backprop_1:eq3}
\begin{split} 
	&\Targetfunction_1((\vartheta_1, \ldots, \vartheta_{k-1}, \theta_k, \vartheta_{k+1}, \ldots, \vartheta_L), \fx_0)\\
&=
	\bpr{
		\Function_L(\vartheta_L, \cdot) \circ 
		\ldots \circ
		\Function_{k+1}(\vartheta_{k+1}, \cdot) \circ 
		\Function_{k}(\theta_{k}, \cdot) \circ 
		\Function_{k-1}(\vartheta_{k-1}, \cdot) \circ 
		\ldots \circ
		\Function_{1}(\vartheta_{1}, \cdot)
	} 
	(\fx_0)\\
&=
	\bpr{
		\Targetfunction_{k+1}((\vartheta_{k+1}, \vartheta_{k+2}, \ldots, \vartheta_L), \Function_{k}(\theta_{k}, \cdot))
	} 
	\bpr{ 
		(
			\Function_{k-1}(\vartheta_{k-1}, \cdot) \circ 
			\ldots \circ
			\Function_{1}(\vartheta_{1}, \cdot)
		)
		(\fx_0)
	} \\
&=
		\Targetfunction_{k+1}((\vartheta_{k+1}, \vartheta_{k+2}, \ldots, \vartheta_L), \Function_{k}(\theta_{k}, \fx_{k-1})).
\end{split}
\end{equation}
Combining this with 
	the chain rule, 
	\cref{backprop_1:ass2}, and
	\cref{backprop_1:concl1}
demonstrates that for all
	$k \in \N \cap (0,L)$
it holds that
\begin{equation}
\label{backprop_1:eq4}
\begin{split} 
	\pr*{
		\frac{\partial \Targetfunction_1}{\partial \theta_k}
	}
	(\vartheta, \fx_0)
&=
	\pr*{
				\R^{n_k} \ni \theta_k
				\mapsto
				\Targetfunction_{k+1}((\vartheta_{k+1}, \vartheta_{k+2}, \ldots, \vartheta_L), \Function_{k}(\theta_{k}, \fx_{k-1}))
				\in \R^{l_L}
	}'
	(\vartheta_k)\\
&=
	\br*{
		\pr*{
					\R^{l_k} \ni \xvar{k}
					\mapsto
					\Targetfunction_{k+1}((\vartheta_{k+1}, \vartheta_{k+2}, \ldots, \vartheta_L), \xvar{k})
					\in \R^{l_L}
		}'
		(\Function_{k}(\vartheta_{k}, \fx_{k-1}))
	} \\
&\quad\,\,
	\br*{
		\pr*{
					\R^{n_k} \ni \theta_k
					\mapsto
					\Function_{k}(\theta_{k}, \fx_{k-1})
					\in \R^{l_k}
		}'
		(\vartheta_k)
	}\\
&=
	\br*{
		\pr*{
			\frac{
				\partial \Targetfunction_{k+1}
			}{
				\partial \xvar{k}
			}
		}
		((\vartheta_{k+1}, \vartheta_{k+2}, \ldots, \vartheta_L), \fx_{k})
	} 
	\br*{
		\pr*{
			\frac{\partial \Function_k}{\partial \thetavar{k}}
		}
		(\vartheta_k, \fx_{k-1})
	} \\
&=
	D_{k+1}
	\br*{
		\pr*{
			\frac{\partial \Function_k}{\partial \thetavar{k}}
		}
		(\vartheta_k, \fx_{k-1})
	}
	.
\end{split}
\end{equation}
\Moreover 
\enum{
	\cref{backprop_1:ass1};
	the fact that $D_{L+1} = 	\idMatrix_{l_L}$
}[ensure]
that 
\begin{equation}
\label{backprop_1:eq5}
\begin{split} 
	\pr*{
		\frac{\partial \Targetfunction_1}{\partial \theta_L}
	}
	(\vartheta, \fx_0)
&=
	\pr*{
				\R^{n_L} \ni \theta_L
				\mapsto
				\Function_{L}(\theta_{L}, \fx_{L-1}))
				\in \R^{l_L}
	}'
	(\vartheta_L)\\
&=
	\br*{
		\pr*{
			\frac{\partial \Function_L}{\partial \thetavar{L}}
		}
		(\vartheta_L, \fx_{L-1})
	} \\
&=
	D_{L+1}
	\br*{
		\pr*{
			\frac{\partial \Function_L}{\partial \thetavar{L}}
		}
		(\vartheta_L, \fx_{L-1})
	}
	.
\end{split}
\end{equation}
Combining this and \eqref{backprop_1:eq4} establishes \cref{backprop_1:item3}.
The proof of \cref{backprop_1} is thus complete.
\end{proof}
\endgroup

\cfclear
\begingroup
\providecommand{\d}{}
\renewcommand{\d}{\defaultParamDim}
\providecommand{\J}{}
\renewcommand{\J}{\defaultLossFunction}
\providecommand{\F}{}
\renewcommand{\F}{F}
\providecommand{\auxTarget}{}
\renewcommand{\auxTarget}{f}
\providecommandordefault{\Function}{F}
\begin{cor}[Backpropagation for parametric functions with loss]
\label{backprop_with_loss_gradients}
Let
	$L \in \N$,
	$l_0, l_1, \ldots, l_{L}, \allowbreak \d_1, \d_2, \ldots, \d_L  \in \N$,
	$\vartheta = (\vartheta_1, \vartheta_{2}, \ldots, \vartheta_L) \in \R^{\d_1} \times \R^{\d_{2}} \times \ldots \times \R^{\d_L}$,
	$\fx_0 \in \R^{l_0}$, 
	$\fx_1 \in \R^{l_1}$, 
	$\dots$,
	$\fx_L \in \R^{l_L}$,
	$\fy \in \R^{l_L}$,
let
$
	\fC = (\fC(x, y))_{(x,y) \in \R^{l_L} \times \R^{l_L}}
\colon
	\R^{l_L} \times \R^{l_L}
	\to
	\R
$
be differentiable,
for every 
	$k \in \{1, 2, \ldots, L\}$
let
$
	\Function_k 
= 
	(\Function_k(\thetavar{k}, \xvar{k-1}))_{(\thetavar{k}, \xvar{k-1}) \in \R^{\d_k} \times \R^{l_{k-1}}}
\colon 
	\R^{\d_k} \times \R^{l_{k-1}} \to \R^{l_k}
$
be differentiable,
let
$
	\Targetfunction
= 
	(\Targetfunction(\theta_{1}, \theta_{2}, \ldots, \theta_{L}))_{
		(\theta_{1}, \theta_{2}, \ldots, \theta_{L}) 
		\in 
		\R^{\d_1} \times \R^{\d_{2}} \times \ldots \times \R^{\d_L}
	} \allowbreak
\colon 
	\R^{\d_1} \times \R^{\d_{2}} \times \ldots \times \R^{\d_L}
\to 
	\R
$
satisfy for all
	$\theta = (\theta_1, \theta_{2}, \ldots, \theta_L) \in \R^{\d_1} \times \R^{\d_{2}} \times \ldots \times \R^{\d_L}$
that
\begin{equation}
\label{backprop_with_loss_gradients:ass1}
\begin{split} 
	\Targetfunction(\theta)
=
	\bpr{
		\fC(\cdot, \fy)
		\circ
		\Function_L(\theta_L, \cdot) \circ 
		\Function_{L-1}(\theta_{L-1}, \cdot) \circ
		\ldots 
		\circ
		\Function_{1}(\theta_{1}, \cdot)
	}
	(\fx_0),
\end{split}
\end{equation}
assume for all
	$k \in \{1, 2, \ldots, L\}$
that
\begin{equation}
\label{backprop_with_loss_gradients:ass2}
\begin{split} 
	\fx_k 
= 
	\Function_{k}(\vartheta_k, \fx_{k-1}),
\end{split}
\end{equation}
and let 
	$D_k \in \R^{l_{k-1}}$, $k \in \{1, 2, \ldots, L+1\}$,
satisfy for all 
	$k \in \{1, 2, \ldots, L\}$
that
\begin{equation}
\label{backprop_with_loss_gradients:ass3}
\begin{split} 
	D_{L+1}
=
	\pr*{
		\nabla_x \fC
	}
	(\fx_{L}, \fy)
\qandq
	D_k
=
	\biggl[
		\pr*{
			\frac{\partial \Function_k}{\partial \xvar{k-1}}
		}
		(\vartheta_k, \fx_{k-1})
	\biggr]^*
	D_{k+1}.
\end{split}
\end{equation}
Then 
\begin{enumerate}[label=(\roman *)]
\item 
\label{backprop_with_loss_gradients:item1}
it holds that 
$
	\Targetfunction
\colon 
	\R^{\d_1} \times \R^{\d_{2}} \times \ldots \times \R^{\d_L}
\to 
	\R
$ 
is differentiable
and

\item 
\label{backprop_with_loss_gradients:item2}
it holds for all
	$k \in \{1, 2, \ldots, L\}$
that 
\begin{equation}
\label{backprop_with_loss_gradients:concl3}
\begin{split} 
	\pr*{
		\nabla_{\theta_k} \Targetfunction
	}
	(\vartheta)
=
	\bbbbr{
		\pr*{
			\frac{\partial \Function_k}{\partial \thetavar{k}}
		}
		(\vartheta_k, \fx_{k-1})
	}^*
	D_{k+1}.
\end{split}
\end{equation}
\end{enumerate}
\end{cor}

\begin{proof}[Proof of \cref{backprop_with_loss_gradients}]
Throughout this proof,
let
	$\bfD_k \in \R^{l_L \times l_{k-1}}$, $k \in \{1, 2, \ldots, L+1\}$,
satisfy for all 
	$k \in \{1, 2, \ldots, L\}$
that
$
	\bfD_{L+1}
=
	\idMatrix_{l_L}
$
and
\begin{equation}
\label{backprop_with_loss_gradients:setting2}
\begin{split} 
	\bfD_k
= 
	\bfD_{k+1}
	\br*{
		\pr*{
			\frac{\partial \Function_k}{\partial \xvar{k-1}}
		}
		(\vartheta_k, \fx_{k-1})
	}
\end{split}
\end{equation}
and 
let
$
	\auxTarget
= 
	(\auxTarget(\theta_{1}, \theta_{2}, \ldots, \theta_{L}))_{
		(\theta_{1}, \theta_{2}, \ldots, \theta_{L}) 
		\in 
		\R^{\d_1} \times \R^{\d_{2}} \times \ldots \times \R^{\d_L}
	} \allowbreak
\colon 
	\R^{\d_1} \times \R^{\d_{2}} \times \ldots \times \R^{\d_L}
\to 
	\R^{l_L}
$
satisfy for all
	$\theta = (\theta_1, \theta_{2}, \ldots, \theta_L) \in \R^{\d_1} \times \R^{\d_{2}} \times \ldots \times \R^{\d_L}$
that
\begin{equation}
\label{backprop_with_loss_gradients:setting1}
\begin{split} 
	\auxTarget(\theta)
=
	\bpr{
		\Function_L(\theta_L, \cdot) \circ 
		\Function_{L-1}(\theta_{L-1}, \cdot) \circ
		\ldots 
		\circ
		\Function_{1}(\theta_{1}, \cdot)
	}
	(\fx_0)
\end{split}
\end{equation}
\cfload.
Note that \cref{backprop_1:item1} in \cref{backprop_1} ensures that 
$\auxTarget 
\colon 
\R^{\d_1} \times \R^{\d_{2}} \times \ldots \times \R^{\d_L}
\to 
	\R^{l_L}$ is differentiable.
\enum{
	This;
	the assumption that
	$
		\fC
	\colon
		\R^{l_L} \times \R^{l_L}
		\to
		\R
	$
	is differentiable;
	the fact that $\Targetfunction = \fC(\cdot, \fy) \circ \auxTarget$
}[ensure]
that
$
	\Targetfunction
\colon 
	\R^{\d_1} \times \R^{\d_{2}} \times \ldots \times \R^{\d_L}
\to 
	\R
$ 
is differentiable.
This establishes \cref{backprop_with_loss_gradients:item1}.
Next we claim that for all 
	$k \in \{1, 2, \ldots, L+1\}$
it holds that
\begin{equation}
\label{backprop_with_loss_gradients:eq1}
\begin{split} 
	[D_k]^*
=
	\br*{
		\pr*{
			\frac{\partial \fC }{\partial x}
		}
		(\fx_L, \fy)
	}
	\bfD_k.
\end{split}
\end{equation}
We now prove \cref{backprop_with_loss_gradients:eq1} by induction on $k \in \{L+1, L, \ldots, 1\}$.
For the base case $k = L+1$ note that 
\enum{
	\cref{backprop_with_loss_gradients:ass3};
	\cref{backprop_with_loss_gradients:setting2}
}[assure]
that
\begin{equation}
\label{backprop_with_loss_gradients:eq2}
\begin{split} 
	[D_{L+1}]^*
&=
	\br*{
		\pr*{
				\nabla_x \fC
		}
		(\fx_{L}, \fy)	
	}^*
=
	\pr*{
		\frac{\partial \fC }{\partial x}
	}
	(\fx_L, \fy)\\
&=
	\br*{
		\pr*{
			\frac{\partial \fC }{\partial x}
		}
		(\fx_L, \fy)
	}
	\idMatrix_{l_L}
=
	\br*{
		\pr*{
			\frac{\partial \fC }{\partial x}
		}
		(\fx_L, \fy)
	}
	\bfD_{L+1}.
\end{split}
\end{equation}
This establishes \cref{backprop_with_loss_gradients:eq1} in the base case $k = L+1$.
For the induction step observe 
\enum{
	\cref{backprop_with_loss_gradients:ass3};
	\cref{backprop_with_loss_gradients:setting2}
}[demonstrate]
that
for all 
	$k \in \{L, L-1, \ldots, 1\}$
with
$
	[D_{k+1}]^*
=
	\br*{
		\pr*{
			\frac{\partial \fC }{\partial x}
		}
		(\fx_L, \fy)
	}
	\bfD_{k+1}
$
it holds that
\begin{equation}
\label{backprop_with_loss_gradients:eq3}
\begin{split} 
	[D_{k}]^*
&=
	[D_{k+1}]^*
	\biggl[
		\pr*{
			\frac{\partial \Function_k}{\partial \xvar{k-1}}
		}
		(\vartheta_k, \fx_{k-1})
	\biggr] \\
&=
	\br*{
		\pr*{
			\frac{\partial \fC }{\partial x}
		}
		(\fx_L, \fy)
	}
	\bfD_{k+1}
	\biggl[
		\pr*{
			\frac{\partial \Function_k}{\partial \xvar{k-1}}
		}
		(\vartheta_k, \fx_{k-1})
	\biggr] 
=
	\br*{
		\pr*{
			\frac{\partial \fC }{\partial x}
		}
		(\fx_L, \fy)
	}
	\bfD_{k}.
\end{split}
\end{equation}
Induction thus establishes \cref{backprop_with_loss_gradients:eq1}.
Furthermore, note that 
\enum{
	\cref{backprop_1:item3} in \cref{backprop_1}
}[assure]
that
for all
	$k \in \{1, 2, \ldots, L\}$
it holds that
\begin{equation}
\label{backprop_with_loss_gradients:eq4}
\begin{split} 
	\pr*{
		\frac{\partial \auxTarget}{\partial \theta_k}
	}
	(\vartheta)
=
	\bfD_{k+1}
	\br*{
		\pr*{
			\frac{\partial \Function_k}{\partial \thetavar{k}}
		}
		(\vartheta_k, \fx_{k-1})
	}
	.
\end{split}
\end{equation}
Combining this with
\enum{
	chain rule;
	the fact that
	$\Targetfunction = \fC(\cdot, \fy) \circ \auxTarget$;
	\cref{backprop_with_loss_gradients:eq1}
}
ensures
that
for all
	$k \in \{1, 2, \ldots, L\}$
it holds that
\begin{equation}
\label{backprop_with_loss_gradients:eq5}
\begin{split} 
	\pr*{
		\frac{\partial \Targetfunction}{\partial \theta_k}
	}
	(\vartheta)
&=
	\br*{
		\pr*{
			\frac{\partial \fC }{\partial x}
		}
		(\auxTarget(\vartheta), \fy)
	}
	\br*{
		\pr*{
			\frac{\partial \auxTarget }{\partial \theta_k}
		}
		(\vartheta)
	}\\
&=
	\br*{
		\pr*{
			\frac{\partial \fC }{\partial x}
		}
		(\fx_L, \fy)
	}
	\bfD_{k+1}
	\br*{
		\pr*{
			\frac{\partial \Function_k}{\partial \thetavar{k}}
		}
		(\vartheta_k, \fx_{k-1})
	}\\
&=
	[D_{k+1}]^*
	\br*{
		\pr*{
			\frac{\partial \Function_k}{\partial \thetavar{k}}
		}
		(\vartheta_k, \fx_{k-1})
	}.
\end{split}
\end{equation}
Hence, we obtain that for all
	$k \in \{1, 2, \ldots, L\}$
it holds that 
\begin{equation}
\label{backprop_with_loss_gradients:eq6}
\begin{split} 
	\pr*{
		\nabla_{\theta_k} \Targetfunction
	}
	(\vartheta)
=
	\br*{
		\pr*{
			\frac{\partial \Targetfunction}{\partial \theta_k}
		}
		(\vartheta)
	}^*
=
	\bbbbr{
		\pr*{
			\frac{\partial \Function_k}{\partial \thetavar{k}}
		}
		(\vartheta_k, \fx_{k-1})
	}^*
	D_{k+1}.
\end{split}
\end{equation}
This establishes \cref{backprop_with_loss_gradients:item2}.
The proof of \cref{backprop_with_loss_gradients} is thus complete.
\end{proof}
\endgroup

\section{Backpropagation for ANNs}

\cfclear
\begin{adef}{def:diag_matrix}[Diagonal matrices]
We denote by 
	$\diag \colon (\bigcup_{d \in \N} \R^d) \to (\bigcup_{d \in \N} \R^{d \times d})$ 
the function which satisfies for all
	$d \in \N$,
	$x = (x_1, \ldots, x_d) \in \R^d$
that
\begin{equation}
\label{diag_matrix:eq1}
\begin{split} 
	\diag(x)
=
	\begin{pmatrix}
		x_1 & 0 & \cdots & 0 \\
		0 & x_2 & \cdots & 0 \\
		\vdots & \vdots & \ddots & \vdots \\
		0 & 0 & \cdots & x_d \\
	\end{pmatrix}
\in 
	\R^{d \times d}.
\end{split}
\end{equation}
\end{adef}

\cfclear
\begingroup
\providecommand{\J}{}
\renewcommand{\J}{\defaultLossFunction}
\providecommandordefault{\Function}{F}
\begin{cor}[Backpropagation for \anns]
\label{backprop_for_ANNs}
Let
	$L \in \N$,
	$l_0, l_1, \ldots, l_{L} \in \N$,
	$
		\Phi 
	=
		(
			(
				\bfW_1
				, 
				\bfB_1
			)
			,
			\ldots
			, \allowbreak
			(
				\bfW_L
				, 
				\bfB_L
			)
		)
	\in
		\bigtimes_{k = 1}^L (\R^{l_k \times l_{k-1}} \times \R^{l_k})
	$,
let 
$
	\fC = \allowbreak (\fC(x, y))_{(x,y) \in \R^{l_L} \times \R^{l_L}}
\colon
	\R^{l_L} \times \R^{l_L}
	\to
	\R
$
and
$a \colon \R \to \R$ 
be differentiable,
let
	$\fx_0  \in \R^{l_0}$, 
	$\fx_1 \in \R^{l_1}$, 
	$\dots$,
	$\fx_L \in \R^{l_L}$,
	$\fy \in \R^{l_L}$
satisfy for all
	$k \in \{1, 2, \ldots, L \}$
that
\begin{equation}
\label{backprop_for_ANNs:ass2}
\begin{split} 
	\fx_k 
= 
	\multdim_{a\ind{[0,L)}(k) + \id_{\R} \ind{\{L\}}(k), l_k}(\bfW_k \fx_{k-1} +  \bfB_k),
\end{split}
\end{equation}
let
$
	\Targetfunction
= 
	\bpr{
		\Targetfunction(
			(W_{1}, B_{1}), \ldots, (W_{L}, B_{L})
		)
	}_{
		((W_{1}, B_{1}), \ldots, (W_{L}, B_{L}))
		\in 
			\bigtimes_{k = 1}^L (\R^{l_k \times l_{k-1}} \times \R^{l_k})
	} \allowbreak
\colon 
	\bigtimes_{k = 1}^L (\R^{l_k \times l_{k-1}}\allowbreak \times \R^{l_k})
\to 
	\R
$
satisfy for all
	$
		\Psi 
	\in 
		\bigtimes_{k = 1}^L (\R^{l_k \times l_{k-1}} \allowbreak \times \R^{l_k})
	$
that
\begin{equation}
\label{backprop_for_ANNs:ass1}
\begin{split} 
	\Targetfunction(\Psi)
=
	\fC( (\functionANN{a}(\Psi))(\fx_0), \fy),
\end{split}
\end{equation}
and let 
	$
		D_k 
		\in \R^{l_{k-1}}
	$, $k \in \{1, 2, \ldots, L+1\}$,
satisfy for all 
	$k \in \{1, 2, \ldots, L-1\}$
that
\begin{equation}
\label{backprop_for_ANNs:ass3}
\begin{split} 
	D_{L+1}
=
	\pr*{
		\nabla_x \fC
	}
	(\fx_{L}, \fy)
,
\qquad
	D_L
=
	[
		\bfW_L
	]^*
	D_{L+1},
\qand
\end{split}
\end{equation}
\begin{equation}
\label{backprop_for_ANNs:ass4}
\begin{split} 
	D_k
=
	[\bfW_{k}]^*
	[
		\diag(
			\multdim_{a', l_k}(
				\bfW_{k} \fx_{k-1} + \bfB_{k}
			)
		)
	]
	D_{k+1}
\end{split}
\end{equation}
\cfload.
Then 
\begin{enumerate}[label=(\roman *)]
\item 
\label{backprop_for_ANNs:item1}
it holds that 
$
	\Targetfunction
\colon 
	\bigtimes_{k = 1}^L (\R^{l_k \times l_{k-1}} \times \R^{l_k})
\to 
	\R
$ 
is differentiable,

\item 
\label{backprop_for_ANNs:item2}
it holds
that
$
	\pr*{
		\nabla_{B_L}  \Targetfunction
	}
	(\Phi)
=
	D_{L+1}
$,

\item 
\label{backprop_for_ANNs:item3}
it holds for all
	$k \in \{1, 2, \ldots, L-1\}$
that
\begin{equation}
\label{backprop_for_ANNs:concl1}
\begin{split} 
	\pr*{
		\nabla_{B_k}  \Targetfunction
	}
	(\Phi)
=
	[
		\diag(
			\multdim_{a', l_k}(
				\bfW_{k} \fx_{k-1} + \bfB_{k}
			)
		)
	]
	D_{k+1},
\end{split}
\end{equation}

\item 
\label{backprop_for_ANNs:item4}
it holds that
$
	\pr*{
		\nabla_{W_L} \Targetfunction
	}
	(\Phi)
=
	D_{L+1}
	[\fx_{L-1}]^*
$, and

\item 
\label{backprop_for_ANNs:item5}
it holds for all
	$k \in \{1, 2, \ldots, L-1\}$
that
\begin{equation}
\label{backprop_for_ANNs:concl2}
\begin{split} 
	\pr*{
		\nabla_{W_k} \Targetfunction
	}
	(\Phi)
=
	[
		\diag(
			\multdim_{a', l_k}(
				\bfW_{k} \fx_{k-1} + \bfB_{k}
			)
		)
	]
	D_{k+1}
	[\fx_{k-1}]^*
	.
\end{split}
\end{equation}

\end{enumerate}

\end{cor}

\begin{proof}[Proof of \cref{backprop_for_ANNs}]
Throughout this proof, 
for every 
	$k \in \{1, 2, \ldots, L\}$
let
\begin{equation}
\begin{split} 
	&\Function_k 
= 
	(\Function_k^{(m)})_{m \in \{1, 2, \ldots, l_k\}} \\
&=
	\bpr{\Function_k\bpr{
		(
			(W_{k,i,j})_{
				(i,j) \in \{1, 2, \ldots, l_k\} \times \{1, 2, \ldots, l_{k-1}\}}, 
			B_k
		),\\
&\quad 
		\xvar{k-1}
	}}_{
		(
			(
				(W_{k,i,j})_{(i,j) \in \{1, 2, \ldots, l_k\} \times \{1, 2, \ldots, l_{k-1}\}}, 
				B_k
			), 
			\xvar{k-1}
		)
		\in 
		(\R^{l_k \times l_{k-1}} \times \R^{l_{k-1}}) \times \R^{l_{k-1}}
	} \\
&\colon 
	(\R^{l_k \times l_{k-1}} \times \R^{l_{k-1}}) \times \R^{l_{k-1}} \to \R^{l_k}
\end{split}
\end{equation}
satisfy for all
	$(W_k, B_k) \in \R^{l_k \times l_{k-1}} \times \R^{l_{k-1}}$,
	$\xvar{k-1} \in \R^{l_{k-1}}$
that
\begin{equation}
\label{backprop_for_ANNs:setting1}
\begin{split} 
	\Function_k((W_k, B_k), \xvar{k-1}) 
= 
	\multdim_{a\ind{[0,L)}(k) + \id_{\R} \ind{\{L\}}(k), l_k}(W_k \xvar{k-1} +  B_k)
\end{split}
\end{equation}
and
for every $d \in \N$ let
	$\mathbf{e}^{(d)}_{1}, \mathbf{e}^{(d)}_{2}, \ldots, \mathbf{e}^{(d)}_{ d} \in \R^d$
satisfy
	$\mathbf{e}^{(d)}_{1} = (1, 0, \ldots, 0)$,
	$\mathbf{e}^{(d)}_{2} = (0, 1, 0, \ldots,\allowbreak 0)$,
	$\dots$,
	$\mathbf{e}^{(d)}_{d} = (0, \ldots, 0, 1)$.
Observe that
the assumption that $a$ is differentiable and \cref{backprop_for_ANNs:ass2} imply that
$
	\Targetfunction
\colon 
	\bigtimes_{k = 1}^L (\R^{l_k \times l_{k-1}} \times \R^{l_k})
\to 
	\R
$ 
is differentiable.
This establishes \cref{backprop_for_ANNs:item1}.
Next note that 
\enum{
	\cref{setting_NN:ass2};
	\cref{backprop_for_ANNs:ass1}; 
	\cref{backprop_for_ANNs:setting1}
}[ensure]
that for all
	$
		\Psi = ((W_{1}, B_{1}), \ldots, (W_{L}, B_{L}))
	\in 
			\bigtimes_{k = 1}^L (\R^{l_k \times l_{k-1}} \times \R^{l_k})
	$
it holds that
\begin{equation}
\label{backprop_for_ANNs:eq1}
\begin{split} 
	\Targetfunction(\Psi)
=
	\bpr{
		\fC(\cdot, \fy)
		\circ
		\Function_L((W_{L}, B_{L}), \cdot) \circ 
		\Function_{L-1}((W_{L-1}, B_{L-1}), \cdot) \circ
		\ldots 
		\circ
		\Function_{1}((W_{1}, B_{1}), \cdot)
	}
	(\fx_0).
\end{split}
\end{equation}
Moreover, observe that
\enum{
	\cref{backprop_for_ANNs:ass2};
	\cref{backprop_for_ANNs:setting1}
}[imply]
that for all
	$k \in \{1, 2, \ldots, L\}$
it holds that
\begin{equation}
\label{backprop_for_ANNs:eq2}
\begin{split} 
	\fx_k 
= 
	\Function_{k}((\bfW_k, \bfB_k), \fx_{k-1}).
\end{split}
\end{equation}
In addition, observe that 
\enum{
	\cref{backprop_for_ANNs:setting1}
}[assure]
that
\begin{equation}
\label{backprop_for_ANNs:eq4}
\begin{split} 
	\pr*{
		\frac{\partial \Function_L}{\partial \xvar{L-1}}
	}
	((\bfW_L, \bfB_L), \fx_{L-1})
=
	\bfW_L.
\end{split}
\end{equation}
Moreover, note that
\enum{
	\cref{backprop_for_ANNs:setting1}
}[imply]
that
for all
	$k \in \{1, 2, \ldots, L-1\}$
it holds that
\begin{equation}
\label{backprop_for_ANNs:eq5}
\begin{split} 
	\pr*{
		\frac{\partial \Function_k}{\partial \xvar{k-1}}
	}
	((\bfW_k, \bfB_k), \fx_{k-1})
=
	[
		\diag(
			\multdim_{a', l_k}(
				\bfW_{k} \fx_{k-1} + \bfB_{k}
			)
		)
	]
	\bfW_{k}.
\end{split}
\end{equation}
Combining 
this
and \eqref{backprop_for_ANNs:eq4}
with
\cref{backprop_for_ANNs:ass3} and \cref{backprop_for_ANNs:ass4}
demonstrates that
for all 
	$k \in \{1, 2, \ldots, L\}$
it holds that
\begin{equation}
\label{backprop_for_ANNs:eq6}
\begin{split} 
	D_{L+1}
=
	\pr*{
		\nabla_x \fC
	}
	(\fx_{L}, \fy)
\qandq
	D_k
=
	\biggl[
		\pr*{
			\frac{\partial \Function_k}{\partial \xvar{k-1}}
		}
		(\vartheta_k, \fx_{k-1})
	\biggr]^*
	D_{k+1}.
\end{split}
\end{equation}
Next note that
\enum{
	this;
	\cref{backprop_for_ANNs:eq1};
	\cref{backprop_for_ANNs:eq2};
	\cref{backprop_with_loss_gradients}	
}[prove]
that
for all
	$k \in \{1, 2, \ldots, L\}$
it holds that 
\begin{equation}
\label{backprop_for_ANNs:eq7}
\begin{split} 
	\pr*{
		\nabla_{B_k} \Targetfunction
	}
	(\Phi)
=
	\bbbbr{
		\pr*{
			\frac{\partial \Function_k}{\partial B_{k}}
		}
		((\bfW_k, \bfB_k), \fx_{k-1})
	}^*
	D_{k+1}
\qand
\end{split} 
\end{equation}
\begin{equation}
\label{backprop_for_ANNs:eq8}
\begin{split} 
	\pr*{
		\nabla_{W_k} \Targetfunction
	}
	(\Phi)
=
		\bbbbr{
			\pr*{
				\frac{\partial \Function_k}{\partial W_{k}}
			}
			((\bfW_k, \bfB_k), \fx_{k-1})
		}^*
	D_{k+1}
	.
\end{split}
\end{equation}
Moreover, observe that
\enum{
	\cref{backprop_for_ANNs:setting1}
}[imply]
that
\begin{equation}
\label{backprop_for_ANNs:eq9}
\begin{split} 
	\pr*{
		\frac{\partial \Function_L}{\partial B_L}
	}
	((\bfW_L, \bfB_L), \fx_{L-1})
=
	\idMatrix_{l_L}
\end{split}
\end{equation}
\cfload.
Combining this with \cref{backprop_for_ANNs:eq7} demonstrates that 
\begin{equation}
\label{backprop_for_ANNs:eq9.2}
\begin{split} 
	\pr*{
		\nabla_{B_L}  \Targetfunction
	}
	(\Phi)
=
	[\idMatrix_{l_L}]^* D_{L+1}
=
	D_{L+1}
.
\end{split}
\end{equation}
This establishes \cref{backprop_for_ANNs:item2}.
Furthermore, note that 
\enum{
	\cref{backprop_for_ANNs:setting1}
}[assure]
that
for all
	$k \in \{1, 2, \ldots, L-1\}$
it holds that
\begin{equation}
\label{backprop_for_ANNs:eq10}
\begin{split} 
	\pr*{
		\frac{\partial \Function_k}{\partial B_k}
	}
	((\bfW_k, \bfB_k), \fx_{k-1})
=
	\diag(
		\multdim_{a', l_k}(
			\bfW_{k} \fx_{k-1} + \bfB_{k}
		)
	).
\end{split}
\end{equation}
Combining this with \cref{backprop_for_ANNs:eq7} implies that 
for all
	$k \in \{1, 2, \ldots, L-1\}$
it holds that
\begin{equation}
\label{backprop_for_ANNs:eq11}
\begin{split} 
	\pr*{
		\nabla_{B_k}  \Targetfunction
	}
	(\Phi)
&=
	[
		\diag(
			\multdim_{a', l_k}(
				\bfW_{k} \fx_{k-1} + \bfB_{k}
			)
		)
	]^*
	D_{k+1}\\
&=
	[
		\diag(
			\multdim_{a', l_k}(
				\bfW_{k} \fx_{k-1} + \bfB_{k}
			)
		)
	]
	D_{k+1}.
\end{split}
\end{equation}
This establishes \cref{backprop_for_ANNs:item3}.
In addition, observe that
\enum{
	\cref{backprop_for_ANNs:setting1}
}[ensure]
that for all
	$m, i \in \{1, 2, \ldots, l_{L}\}$,
	$j \in \{1, 2, \ldots, l_{L-1}\}$
it holds that
\begin{equation}
\label{backprop_for_ANNs:eq12}
\begin{split} 
	\pr*{
		\frac{\partial \Function_L^{(m)}}{\partial W_{L, i, j}}
	}
	((\bfW_L, \bfB_L), \fx_{L-1})
=
	\mathbbm{1}_{\{m\}}(i)
	\scp{\fx_{L-1}, \mathbf{e}^{(l_{L-1})}_{ j}}
\end{split}
\end{equation}
\cfload.
Combining this with \cref{backprop_for_ANNs:eq8} demonstrates that
\begin{equation}
\label{backprop_for_ANNs:eq13}
\begin{split} 
	&\pr*{
		\nabla_{W_L} \Targetfunction
	}
	(\Phi) \\
&=
	\pr*{
		{\sum_{m = 1}^{l_L}}
			\br*{
				\pr*{
					\frac{\partial \Function_L^{(m)}}{\partial W_{L, i, j}}
				}
				((\bfW_L, \bfB_L), \fx_{L-1})
			}
			\scp{D_{L+1}, \mathbf{e}^{(l_L)}_{ m}}
	}_{(i, j) \in \{1, 2, \ldots, l_L\} \times \{1, 2, \ldots, l_{L-1}\}}\\
&=
\textstyle
	\pr*{
		\sum_{m = 1}^{l_L}
			\mathbbm{1}_{\{m\}}(i)
			\scp{\mathbf{e}^{(l_{L-1})}_{ j}, \fx_{L-1}}
			\scp{\mathbf{e}^{(l_L)}_{ m}, D_{L+1}}
	}_{(i, j) \in \{1, 2, \ldots, l_L\} \times \{1, 2, \ldots, l_{L-1}\}}\\
&=
	\pr*{
			\scp{\mathbf{e}^{(l_{L-1})}_{ j}, \fx_{L-1}}
			\scp{\mathbf{e}^{(l_L)}_{ i}, D_{L+1}}
	}_{(i, j) \in \{1, 2, \ldots, l_L\} \times \{1, 2, \ldots, l_{L-1}\}}\\
&=
	D_{L+1}
	[\fx_{L-1}]^*.
\end{split}
\end{equation}
This establishes \cref{backprop_for_ANNs:item4}.
Moreover, note that 
\enum{
	\cref{backprop_for_ANNs:setting1}
}[imply]
that
for all
	$k \in \{1, 2, \ldots, L-1\}$,
	$m, i \in \{1, 2, \ldots, l_{k}\}$,
	$j \in \{1, 2, \ldots, l_{k-1}\}$
it holds that
\begin{equation}
\label{backprop_for_ANNs:eq14}
\begin{split} 
	\pr*{
		\frac{\partial \Function_k^{(m)}}{\partial W_{k, i, j}}
	}
	((\bfW_k, \bfB_k), \fx_{k-1})
=
	\mathbbm{1}_{\{m\}}(i)
	a'(\scp{
		\mathbf{e}^{(l_k)}_{ i},
		\bfW_{k} \fx_{k-1} + \bfB_{k}
	})
	\scp{\mathbf{e}^{(l_{k-1})}_{ j}, \fx_{k-1}}.
\end{split}
\end{equation}
Combining this with \cref{backprop_for_ANNs:eq8} demonstrates that for all
	$k \in \{1, 2, \ldots, L-1\}$
it holds that
\begin{equation}
\label{backprop_for_ANNs:eq15}
\begin{split} 
	&\pr*{
		\nabla_{W_k} \Targetfunction
	}
	(\Phi)\\
&=
	\pr*{
		\sum_{m = 1}^{l_k}
			\br*{
				\pr*{
					\frac{\partial \Function_k^{(m)}}{\partial W_{k, i, j}}
				}
				((\bfW_k, \bfB_k), \fx_{k-1})
			}
			\scp{\mathbf{e}^{(l_k)}_{ m}, D_{k+1}}
	}_{\!(i, j) \in \{1, 2, \ldots, l_k\} \times \{1, 2, \ldots, l_{k-1}\}}\\	
&=
\textstyle
	\pr*{
		\sum_{m = 1}^{l_k}
			\mathbbm{1}_{\{m\}}(i)
			a'(\scp{
				\mathbf{e}^{(l_k)}_{ i},
				\bfW_{k} \fx_{k-1} + \bfB_{k}
			})
			\scp{\mathbf{e}^{(l_{k-1})}_{ j}, \fx_{k-1}}
			\scp{\mathbf{e}^{(l_k)}_{ m}, D_{k+1}}
	}_{\!(i, j) \in \{1, 2, \ldots, l_k\} \times \{1, 2, \ldots, l_{k-1}\}}\\	
&=
\textstyle
	\pr*{
		a'(\scp{
			\mathbf{e}^{(l_k)}_{ i},
			\bfW_{k} \fx_{k-1} + \bfB_{k}
		})
		\scp{\mathbf{e}^{(l_{k-1})}_{ j}, \fx_{k-1}}
		\scp{\mathbf{e}^{(l_k)}_{ i}, D_{k+1}}
	}_{\!(i, j) \in \{1, 2, \ldots, l_k\} \times \{1, 2, \ldots, l_{k-1}\}}\\	
&=
	[
		\diag(
			\multdim_{a', l_k}(
				\bfW_{k} \fx_{k-1} + \bfB_{k}
			)
		)
	]
	D_{k+1}
	[\fx_{k-1}]^*
	.
\end{split}
\end{equation}
This establishes \cref{backprop_for_ANNs:item5}.
The proof of \cref{backprop_for_ANNs} is thus complete.
\end{proof}
\endgroup

\cfclear
\begingroup
\providecommand{\auxTarget}{}
\renewcommand{\auxTarget}{\mathbf{L}}
\begin{athm}{cor}{backprop_for_ANNs_minibatch}[Backpropagation for \anns\ with minibatches]
Let
	$L, M \in \N$,
	$l_0, l_1, \ldots,\allowbreak l_{L} \in \N$,
	$
		\Phi 
	=
		(
			(
				\bfW_1
				, 
				\bfB_1
			)
			,
			\ldots
			, \allowbreak
			(
				\bfW_L
				, 
				\bfB_L
			)
		)
	\in
		\bigtimes_{k = 1}^L (\R^{l_k \times l_{k-1}} \times \R^{l_k})
	$,
let 
$a \colon \R \to \R$ 
and 
$
	\fC = \allowbreak (\fC(x, y))_{(x,y) \in \R^{l_L} \times \R^{l_L}}
\colon
	\R^{l_L} \times \R^{l_L}
	\to
	\R
$
be differentiable,
for every 
	$m \in \{1, 2, \ldots, M\}$ 
let	
	$\fx^{(m)}_0  \in \R^{l_0}$, 
	$\fx^{(m)}_1  \in \R^{l_1}$, 
	$\dots$,
	$\fx^{(m)}_L \in \R^{l_L}$,
	$\fy^{(m)} \in \R^{l_L}$
satisfy for all
	$k \in \{1, 2, \ldots, L \}$
that
\begin{equation}
\label{backprop_for_ANNs_minibatch:ass2}
\begin{split} 
	\fx^{(m)}_k 
= 
	\multdim_{a\ind{[0,L)}(k) + \id_{\R} \ind{\{L\}}(k), l_k}(\bfW_k \fx^{(m)}_{k-1} +  \bfB_k),
\end{split}
\end{equation}
let
$
	\Targetfunction
= 
	\bpr{
		\Targetfunction(
			(W_{1}, B_{1}), \ldots, (W_{L}, B_{L})
		)
	}_{
		((W_{1}, B_{1}), \ldots, (W_{L}, B_{L}))
		\in 
			\bigtimes_{k = 1}^L (\R^{l_k \times l_{k-1}} \times \R^{l_k})
	} \allowbreak
\colon 
	\bigtimes_{k = 1}^L (\R^{l_k \times l_{k-1}} \allowbreak \times \R^{l_k})
\to 
	\R
$
satisfy for all
	$
		\Psi 
	\in 
			\bigtimes_{k = 1}^L (\R^{l_k \times l_{k-1}} \times \R^{l_k})
	$
that
\begin{equation}
\label{backprop_for_ANNs_minibatch:ass1}
\begin{split} 
	\Targetfunction(\Psi)
=
	\frac{1}{M}
	\bbbbr{
		\smallsum_{m = 1}^M
			\fC ((\functionANN{a}(\Psi))(\fx^{(m)}_0), \fy^{(m)})
	},
\end{split}
\end{equation}
and for every
	$m \in \{1, 2, \ldots, M\}$ 
let 
	$
		D^{(m)}_k 
	\in 
		\R^{l_{k-1}}
	$, $k \in \{1, 2, \ldots, L+1\}$,
satisfy for all 
	$k \in \{1, 2, \ldots, L-1\}$
that
\begin{equation}
\label{backprop_for_ANNs_minibatch:ass3}
\begin{split} 
	D^{(m)}_{L+1}
=
	\pr*{
		\nabla_x \fC
	}
	(\fx^{(m)}_{L}, \fy^{(m)})
,
\qquad
	D^{(m)}_L
=
	[
		\bfW_L
	]^*
	D^{(m)}_{L+1},
\qand
\end{split}
\end{equation}
\begin{equation}
\label{backprop_for_ANNs_minibatch:ass4}
\begin{split} 
	D^{(m)}_k
=
	[\bfW_{k}]^*
	[
		\diag(
			\multdim_{a', l_k}(
				\bfW_{k} \fx^{(m)}_{k-1} + \bfB_{k}
			)
		)
	]
	D^{(m)}_{k+1}
\end{split}
\end{equation}
\cfload.
Then 
\begin{enumerate}[label=(\roman *)]
\item 
\label{backprop_for_ANNs_minibatch:item1}
it holds that 
$
	\Targetfunction
\colon 
	\bigtimes_{k = 1}^L (\R^{l_k \times l_{k-1}} \times \R^{l_k})
\to 
	\R
$ 
is differentiable,

\item 
\label{backprop_for_ANNs_minibatch:item2}
it holds 
that
$
	\pr*{
		\nabla_{B_L}  \Targetfunction
	}
	(\Phi)
=
	\frac{1}{M}
	\bbr{
		\sum_{m = 1}^M
			D^{(m)}_{L+1}
	}
$,

\item 
\label{backprop_for_ANNs_minibatch:item3}
it holds for all
	$k \in \{1, 2, \ldots, L-1\}$
that
\begin{equation}
\label{backprop_for_ANNs_minibatch:concl1}
\begin{split} 
	\pr*{
		\nabla_{B_k}  \Targetfunction
	}
	(\Phi)
=
	\frac{1}{M}
	\bbbbr{
		\smallsum_{m = 1}^M
			[
				\diag(
					\multdim_{a', l_k}(
						\bfW_{k} \fx^{(m)}_{k-1} + \bfB_{k}
					)
				)
			]
			D^{(m)}_{k+1}
	},
\end{split}
\end{equation}

\item 
\label{backprop_for_ANNs_minibatch:item4}
it holds 
that
$
	\pr*{
		\nabla_{W_L} \Targetfunction
	}
	(\Phi)
=
	\frac{1}{M}
	\bbr{
		\sum_{m = 1}^M
			D^{(m)}_{L+1}
			[\fx^{(m)}_{L-1}]^*
	}
$,
and

\item 
\label{backprop_for_ANNs_minibatch:item5}
it holds for all
	$k \in \{1, 2, \ldots, L-1\}$
that
\begin{equation}
\label{backprop_for_ANNs_minibatch:concl2}
\begin{split} 
	\pr*{
		\nabla_{W_k} \Targetfunction
	}
	(\Phi)
=
	\frac{1}{M}
	\bbbbr{
		\smallsum_{m = 1}^M
			[
				\diag(
					\multdim_{a', l_k}(
						\bfW_{k} \fx^{(m)}_{k-1} + \bfB_{k}
					)
				)
			]
			D^{(m)}_{k+1}
			[\fx^{(m)}_{k-1}]^*
	}
	.
\end{split}
\end{equation}

\end{enumerate}

\end{athm}

\begin{aproof}
	Throughout this proof,
	let $\auxTarget^{(m)}\colon \bigtimes_{k = 1}^L (\R^{l_k \times l_{k-1}} \times \R^{l_k})\to \R$, $m\in\{1,2,\dots,M\}$,
	satisfy 
		for all
			$m\in\{1,2,\dots,M\}$,
			$\Psi \in \bigtimes_{k = 1}^L (\R^{l_k \times l_{k-1}} \times \R^{l_k})$
		that
		\begin{equation}
			\llabel{eq:1}
			\auxTarget^{(m)}(\Psi)
			=
			\fC ((\functionANN{a}(\Psi))(\fx^{(m)}_0), \fy^{(m)})
			.
		\end{equation}
	\Nobs that
		\lref{eq:1}
		and \cref{backprop_for_ANNs_minibatch:ass1}
	ensure that for all
		$\Psi \in \bigtimes_{k = 1}^L (\R^{l_k \times l_{k-1}} \times \R^{l_k})$
	it holds that
	\begin{equation}
		\Targetfunction(\Psi)
		=
		\frac{1}{M}
		\bbbbr{
			\smallsum_{m = 1}^M
			\auxTarget^{(m)}(\Psi)
		}
		.	
	\end{equation}
		\Cref{backprop_for_ANNs} 
		hence
	establishes 
		\cref{backprop_for_ANNs_minibatch:item1,backprop_for_ANNs_minibatch:item2,backprop_for_ANNs_minibatch:item3,backprop_for_ANNs_minibatch:item4,backprop_for_ANNs_minibatch:item5}.
\end{aproof}
\endgroup

\cfclear
\begingroup
\providecommand{\J}{}
\renewcommand{\J}{\defaultLossFunction} 
\begin{cor}[Backpropagation for \anns\ with quadratic loss and minibatches]
\label{backprop_for_ANNs_minibatch_quadr_loss}
Let
	$L, M \in \N$,
	$l_0, l_1, \ldots, l_{L} \in \N$,
	$
		\Phi 
	=
		(
			(
				\bfW_1
				, 
				\bfB_1
			)
			,
			\ldots
			, \allowbreak
			(
				\bfW_L
				, 
				\bfB_L
			)
		)
	\in
		\bigtimes_{k = 1}^L (\R^{l_k \times l_{k-1}} \times \R^{l_k})
	$,
let
	$a \colon \R \to \R$
be differentiable,
for every 
	$m \in \{1, 2, \ldots, M\}$ 
let	
	$\fx^{(m)}_0  \in \R^{l_0}$, 
	$\fx^{(m)}_1  \in \R^{l_1}$, 
	$\dots$,
	$\fx^{(m)}_L \in \R^{l_L}$,
	$\fy^{(m)} \in \R^{l_L}$
satisfy for all
	$k \in \{1, 2, \ldots, L \}$
that
\begin{equation}
\label{backprop_for_ANNs_minibatch_quadr_loss:ass1}
\begin{split} 
	\fx^{(m)}_k 
= 
	\multdim_{a\ind{[0,L)}(k) + \id_{\R} \ind{\{L\}}(k), l_k}(\bfW_k \fx^{(m)}_{k-1} +  \bfB_k),
\end{split}
\end{equation}
let
$
	\Targetfunction
= 
	\bpr{
		\Targetfunction(
			(W_{1}, B_{1}), \ldots, (W_{L}, B_{L})
		)
	}_{
		((W_{1}, B_{1}), \ldots, (W_{L}, B_{L}))
		\in 
			\bigtimes_{k = 1}^L (\R^{l_k \times l_{k-1}} \times \R^{l_k})
	} \allowbreak
\colon 
	\bigtimes_{k = 1}^L (\R^{l_k \times l_{k-1}} \allowbreak \times \R^{l_k})
\to 
	\R
$
satisfy for all
	$
		\Psi 
	\in 
		\bpr{
			\bigtimes_{k = 1}^L (\R^{l_k \times l_{k-1}} \times \R^{l_k})
		}
	$
that
\begin{equation}
\label{backprop_for_ANNs_minibatch_quadr_loss:ass2}
\begin{split} 
	\Targetfunction(\Psi)
=
	\frac{1}{M}
	\bbbbr{
		\smallsum_{m = 1}^M
			\pnorm2{ (\functionANN{a}(\Psi))(\fx^{(m)}_0) - \fy^{(m)}}^2
	}
	,
\end{split}
\end{equation}
and for every
	$m \in \{1, 2, \ldots, M\}$ 
let 
	$D^{(m)}_k  \in \R^{l_{k-1}}$, $k \in \{1, 2, \ldots, L+1\}$,
satisfy for all 
	$k \in \{1, 2, \ldots, L-1\}$
that
\begin{equation}
\label{backprop_for_ANNs_minibatch_quadr_loss:ass3}
\begin{split} 
	D^{(m)}_{L+1}
=
	2 (\fx^{(m)}_{L} - \fy^{(m)})
,
\qquad
	D^{(m)}_L
=
	[
		\bfW_L
	]^*
	D^{(m)}_{L+1},
\qand
\end{split}
\end{equation}
\begin{equation}
\label{backprop_for_ANNs_minibatch_quadr_loss:ass4}
\begin{split} 
	D^{(m)}_k
=
	[\bfW_{k}]^*
	[
		\diag(
			\multdim_{a', l_k}(
				\bfW_{k} \fx^{(m)}_{k-1} + \bfB_{k}
			)
		)
	]
	D^{(m)}_{k+1}
\end{split}
\end{equation}
\cfload.
Then 
\begin{enumerate}[label=(\roman *)]
\item 
\label{backprop_for_ANNs_minibatch_quadr_loss:item1}
it holds that 
$
	\Targetfunction
\colon 
	\bigtimes_{k = 1}^L (\R^{l_k \times l_{k-1}} \times \R^{l_k})
\to 
	\R
$ 
is differentiable,

\item 
\label{backprop_for_ANNs_minibatch_quadr_loss:item2}
it holds 
that
$
	\pr*{
		\nabla_{B_L}  \Targetfunction
	}
	(\Phi)
=
	\textstyle
	\frac{1}{M}
	\bbr{
		\sum_{m = 1}^M
			D^{(m)}_{L+1}
	}
$,

\item 
\label{backprop_for_ANNs_minibatch_quadr_loss:item3}
it holds for all
	$k \in \{1, 2, \ldots, L-1\}$
that
\begin{equation}
\label{backprop_for_ANNs_minibatch_quadr_loss:concl1}
\begin{split} 
	\pr*{
		\nabla_{B_k}  \Targetfunction
	}
	(\Phi)
=
	\frac{1}{M}
	\bbbbr{
		\smallsum_{m = 1}^M
			[
				\diag(
					\multdim_{a', l_k}(
						\bfW_{k} \fx^{(m)}_{k-1} + \bfB_{k}
					)
				)
			]
			D^{(m)}_{k+1}
	}
	,
\end{split}
\end{equation}

\item 
\label{backprop_for_ANNs_minibatch_quadr_loss:item4}
it holds 
that
$
	\pr*{
		\nabla_{W_L} \Targetfunction
	}
	(\Phi)
=
	\frac{1}{M}
	\bbr{
		\sum_{m = 1}^M
			D^{(m)}_{L+1}
			[\fx^{(m)}_{L-1}]^*
	}
$,
and

\item 
\label{backprop_for_ANNs_minibatch_quadr_loss:item5}
it holds for all
	$k \in \{1, 2, \ldots, L-1\}$
that
\begin{equation}
\label{backprop_for_ANNs_minibatch_quadr_loss:concl2}
\begin{split} 
	\pr*{
		\nabla_{W_k} \Targetfunction
	}
	(\Phi)
=
	\frac{1}{M}
	\bbbbr{
		\smallsum_{m = 1}^M
			[
				\diag(
					\multdim_{a', l_k}(
						\bfW_{k} \fx^{(m)}_{k-1} + \bfB_{k}
					)
				)
			]
			D^{(m)}_{k+1}
			[\fx^{(m)}_{k-1}]^*
	}
	.
\end{split}
\end{equation}
\end{enumerate}
\end{cor}

\begin{proof}[Proof of \cref{backprop_for_ANNs_minibatch_quadr_loss}]
Throughout this proof,
let
	$
		\fC = (\fC(x, y))_{(x,y) \in \R^{l_L} \times \R^{l_L}}
	\colon
		\R^{l_L} \times \R^{l_L}
		\to
		\R
	$
satisfy for all 
	$x,y \in \R^{l_L}$
that
\begin{equation}
\label{backprop_for_ANNs_minibatch_quadr_loss:setting0}
\begin{split} 
	\fC(x, y) = \pnorm{2}{x-y}^2,
\end{split}
\end{equation}
Observe that
\enum{
	\cref{backprop_for_ANNs_minibatch_quadr_loss:setting0}
}[ensure]
that for all
	$m\in\{1,2,\dots,M\}$
it holds that
\begin{equation}
\label{backprop_for_ANNs:eq3}
\begin{split} 
	\pr*{
		\nabla_x \fC
	}
	(\fx_{L}^{(m)}, \fy^{(m)})
=
	2 (\fx_{L}^{(m)} - \fy^{(m)})
=
	D_{L+1}^{(m)}.
\end{split}
\end{equation}
Combining 
\enum{
	this;
	\cref{backprop_for_ANNs_minibatch_quadr_loss:ass1};
	\cref{backprop_for_ANNs_minibatch_quadr_loss:ass2};
	\cref{backprop_for_ANNs_minibatch_quadr_loss:ass3};
	\cref{backprop_for_ANNs_minibatch_quadr_loss:ass4};
}
with
\cref{backprop_for_ANNs_minibatch} establishes \cref{backprop_for_ANNs_minibatch_quadr_loss:item1,backprop_for_ANNs_minibatch_quadr_loss:item2,backprop_for_ANNs_minibatch_quadr_loss:item3,backprop_for_ANNs_minibatch_quadr_loss:item4,backprop_for_ANNs_minibatch_quadr_loss:item5}.
The proof of \cref{backprop_for_ANNs_minibatch_quadr_loss} is thus complete.
\end{proof}
\endgroup

%% file: parts/KL-inequalities.tex
\cchapter{Kurdyka--\L ojasiewicz (KL) inequalities}{chapter:KL}

In \cref{chapter:flow} (\GF\ trajectories), \cref{chapter:deterministic} (deterministic \GD-type processes), and \cref{chapter:stochastic} (\SGD-type processes) we reviewed and studied gradient based processes for the approximate solution of certain optimization problems. In particular, we sketched the approach of general Lyapunov-type functions as well as the special case where the Lyapunov-type function is the squared standard norm around a minimizer resulting in the coercivity-type conditions used in several convergence results in \cref{chapter:flow,chapter:deterministic,chapter:stochastic}.
However, the coercivity-type conditions in \cref{chapter:flow,chapter:deterministic,chapter:stochastic} are usually too restrictive to cover the situation of the training of \anns\ (cf., \eg, \cref{fcond1:item2} in \cref{fcond1}, \cite[item (vi) in Corollary 29]{JentzenRiekert22}, and \cite[Corollary 2.19]{Hutzenthaler2021}).

In this chapter we introduce another general class of Lyapunov-type functions which does indeed cover the mathematical analysis of many of the \ann\ training situations.
Specifically, in this chapter we study Lyapunov-type functions that are given by suitable fractional powers of differences of the risk function (cf., \eg\  \eqref{prop:gf.conv.simple.eq:defVU} in the proof of \cref{prop:gf.conv.simple} below).
In that case the resulting Lyapunov-type conditions (cf., \eg, \cref{def:KLinequality_standard:eq1,gf.conv.simple:ass1,prop:gf.conv.simple:eq1} below) are referred to as \KL\ inequalities in the literature.

Further investigations related to \KL\ inequalities in the scientific literature can, \eg, be found in \cite{bierstone1988semianalytic,colding2015lojasiewicz,Dereich2021,Bolte2006,Eberle2023,JentzenRiekert2022Existence}.
The specific presentation of this chapter is in parts closely based on \cite[Sections~3, 7, and 8]{jentzen2022existence_arxiv}.

\section{Standard KL functions}

\cfclear
\begingroup
\providecommand{\d}{}
\renewcommand{\d}{\defaultParamDim}
\providecommand{\f}{}
\renewcommand{\f}{\defaultLossFunction}
\begin{adef}{def:KLinequality_standard}[Standard \KL\ inequalities]
Let $ \d \in \N $, $ c \in \R $, 
$ \alpha \in (0,\infty) $, 
let $ U \subseteq \R^{ \d } $ be open,
let $ \cpoint \in U $, and 
let $ \f \colon U \to \R $ be a function. 
Then we say that $ \f $ satisfies the standard \KL\ inequality 
at $ \cpoint $
with exponent $ \alpha $
and constant $ c $ 
(we say that $ \f $ satisfies the standard \KL\ inequality 
at $ \cpoint $)
if and only if 
\begin{enumerate}[(i)]
\item
it holds that $ \f $ is differentiable and 
\item
it holds for all 
$
  \altpoint \in U 
$
that 
\begin{equation}
\label{def:KLinequality_standard:eq1}
  \abs{ \f( \cpoint ) - \f( \altpoint ) }^{ \alpha }
  \leq 
  c 
  \Pnorm2{ ( \nabla \f )( \altpoint ) }
\end{equation}
\end{enumerate}
\cfload.
\end{adef}
\endgroup

\cfclear
\begingroup
\providecommand{\d}{}
\renewcommand{\d}{\defaultParamDim}
\providecommand{\f}{}
\renewcommand{\f}{\defaultLossFunction}
\begin{adef}{def:KLfunction_standard}[Standard \KL\ functions]
Let $ \d \in \N $ and 
let $ \f \colon \R^{ \d } \to \R $ be a function. 
Then we say that $ \f $ is a standard \KL\ function if and only if 
for all $ \cpoint \in \R^{ \d } $ there exist 
$ \varepsilon, c \in (0,\infty) $, 
$ \alpha \in (0,1) $
such that\cfadd{def:KLinequality_standard} $ \f\!\mid_{\{ \altpoint \in \R^{ \d } \colon \Pnorm2{ \altpoint - \cpoint } < \varepsilon \} } $ satisfies the standard \KL\ inequality at $ \cpoint $ with exponent $ \alpha $ and constant $ c $
\cfload.
\end{adef}
\endgroup

\section{Convergence analysis using standard KL functions (regular regime)}

\cfclear
\begingroup
\providecommand{\d}{}
\renewcommand{\d}{\defaultParamDim}
\providecommand{\c}{}
\renewcommand{\c}{\mathfrak{c}}
\providecommand{\C}{}
\renewcommand{\C}{\mathfrak{C}}
\providecommand{\g}{}
\renewcommand{\g}{\defaultGradientFunction}
\providecommand{\f}{}
\renewcommand{\f}{\defaultLossFunction}
\begin{athm}{prop}{prop:gf.conv.simple}
	Let
		$\d\in\N$,
		$\cpoint\in\R^\d$,
		$\c,\C,\eps\in(0,\infty)$,
		$\alpha\in(0,1)$,
		$\f\in C^1(\R^\d,\R)$, 
		let 
		$O\subseteq\R^\d$
satisfy
\begin{equation}
  O=\{\altpoint\in\R^\d\colon \Pnorm2{\altpoint-\cpoint}<\eps\}\backslash\{\cpoint\}
\qandq 
  \c=\C^{2}\bbr{\sup\nolimits_{\altpoint\in O}\abs{\f(\altpoint)-\f(\cpoint)}}^{2-2\alpha}
  ,
\end{equation}
assume 
			for all
				$\altpoint\in O$
			that
				$\f(\altpoint)>\f(\cpoint)$
				and
				\begin{equation}
				\label{gf.conv.simple:ass1}
				\begin{split} 
					\abs{\f(\altpoint)-\f(\cpoint)}^\alpha\leq \C\Pnorm2{(\nabla\f)(\altpoint)},
				\end{split}
				\end{equation}
	and
	let
		$ \Theta\in C([0,\infty),O) $
	satisfy
		for all
			$t\in[0,\infty)$
		that
        \begin{equation}
			\Theta_t=\Theta_0-\int_0^t(\nabla\f)(\Theta_s)\,\diff s
        \end{equation}
	\cfload.
	Then there exists $\psi\in\R^\d$ such that
	\begin{enumerate}[(i)]
		\item \llabel{it:1}
		it holds that
			$\f(\psi)=\f(\cpoint)$,
		\item \llabel{it:2}
		it holds for all
			$t\in[0,\infty)$
		that
			\begin{equation}
			  0
              \leq 
			    \f( \Theta_t ) - \f( \psi ) 
              \leq 
                [ ( \f( \Theta_0 ) - \f( \psi ) )^{ - 1 } + \c^{ - 1 } t ]^{ - 1 } 
			  ,
			\end{equation}
		and
		\item \llabel{it:3}
		it holds for all
			$t\in[0,\infty)$
		that
		\begin{eqsplit}
			\Pnorm2{\Theta_t-\psi}
			&\leq
			\int_t^\infty\Pnorm2{(\nabla\f)(\Theta_s)}\,\diff s
			\\&
			\leq
			\C(1-\alpha)^{-1}\br{\f(\Theta_t)-\f(\psi)}^{1-\alpha}
			\\&
			\leq
			\C(1-\alpha)^{-1}
                [ ( \f( \Theta_0 ) - \f( \psi ) )^{ - 1 } + \c^{ - 1 } t ]^{ \alpha - 1 } 
			.
		\end{eqsplit}
	\end{enumerate}
\end{athm}
\providecommand{\ff}{}
\renewcommand{\ff}{\mathbf{L}}
\begin{aproof}
	\providecommand{\CC}{}
	\renewcommand{\CC}{\c}
	Throughout this proof,
		let	$ V \colon O \to \R $ 
		and $ U \colon O \to \R $
		satisfy for all
			$\altpoint\in O$
		that
		\begin{equation}
			\llabel{eq:defVU}
			V(\altpoint)
			=
			-\abs{\f(\altpoint)-\f(\cpoint)}^{-1}
			\qquad\text{and}\qquad
			U(\altpoint)
			=
			\abs{\f(\altpoint)-\f(\cpoint)}^{1-\alpha}
			.
		\end{equation}
	\Nobs that
		the assumption that
			for all
				$\altpoint\in O$
			it holds that
				$\abs{\f(\altpoint)-\f(\cpoint)}^\alpha\leq\C\Pnorm2{(\nabla\f)(\altpoint)}$
	\proves that for all
		$\altpoint\in O$
	it holds that
	\begin{equation}
		\llabel{eq:kl}
		\Pnorm2{(\nabla\f)(\altpoint)}^2
		\geq 
		\C^{-2}\abs{\f(\altpoint)-\f(\cpoint)}^{2\alpha}
		.
	\end{equation}
	\Moreover
		\lref{eq:defVU}
	ensures that for all
		$\altpoint\in O$
	it holds that
		$V\in C^1(O,\R)$
	and
	\begin{equation}
		(\nabla V)(\altpoint)
		=
		\abs{\f(\altpoint)-\f(\cpoint)}^{-2}(\nabla \f)(\altpoint)
		.
	\end{equation}
	Combining
		this
	with
		\lref{eq:kl}
	\proves that for all
		$\altpoint\in O$
	it holds that
	\begin{eqsplit}
	\label{prop:gf.conv.simple:eq1}
		\scp{(\nabla V)(\altpoint),-(\nabla\f)(\altpoint)}
		&=
		-\abs{\f(\altpoint)-\f(\cpoint)}^{-2}\Pnorm2{(\nabla\f)(\altpoint)}^2
		\\&\leq
		-\C^{-2}\abs{\f(\altpoint)-\f(\cpoint)}^{2\alpha-2}
		\leq
		-\CC^{-1}
		.
	\end{eqsplit}
		The assumption that
			for all
				$t\in[0,\infty)$
			it holds that
				$\Theta_t\in O$,
		the assumption that 
			for all
				$t\in[0,\infty)$
			it holds that
				$\Theta_t=\Theta_0-\int_0^t(\nabla\f)(\Theta_s)\,\diff s$,
		and \Cref{lem:lyapunov_general}
		\hence
	establish that for all
		$t\in[0,\infty)$
	it holds that
	\begin{equation}
	\begin{split}
		\mathop{-}
		\abs{
		  \f( \Theta_t ) - \f( \cpoint ) 
        }^{ - 1 }
    & =
		V(\Theta_t)
		\leq
		V(\Theta_0)
		+
		\int_0^t-\CC^{-1}\,\diff s
		=
		V(\Theta_0)
		- \CC^{-1} t
    \\ &
    =
		\mathop{-}
		\abs{
		  \f( \Theta_0 ) - \f( \cpoint ) 
        }^{ - 1 }
		- \CC^{-1} t
		.
    \end{split}
	\end{equation}
	\Hence for all $ t \in [0,\infty) $ that 
	\begin{equation}
	\label{eq:weak_error_with_rate_f_estimate}
	\begin{split}
	0
	<
		  \f( \Theta_t ) - \f( \cpoint ) 
    & \leq 
    [
		\abs{
		  \f( \Theta_0 ) - \f( \cpoint ) 
        }^{ - 1 }
		+ \CC^{-1} t
    ]^{ - 1 }
		.
    \end{split}
	\end{equation}
	\Moreover
		\lref{eq:defVU}
	ensures that for all
		$\altpoint\in O$
	it holds that
		$U\in C^1(O,\R)$
	and
	\begin{equation}
		(\nabla U)(\altpoint)
		=
		(1-\alpha)\abs{\f(\altpoint)-\f(\cpoint)}^{-\alpha}(\nabla\f)(\altpoint)
		.
	\end{equation}
		The assumption that
		  for all
				$\altpoint\in O$
			it holds that
			  $\abs{\f(\altpoint)-\f(\cpoint)}^{\alpha}\leq\C\Pnorm2{(\nabla\f)(\altpoint)}$
		\hence
	\proves that for all
		$\altpoint\in O$
	it holds that
	\begin{eqsplit}
		\scp{(\nabla U)(\altpoint),-(\nabla\f)(\altpoint)}
		&=
		-(1-\alpha)\abs{\f(\altpoint)-\f(\cpoint)}^{-\alpha}\Pnorm2{(\nabla\f)(\altpoint)}^2
		\\&\leq
		-\C^{-1}(1-\alpha)\Pnorm2{(\nabla\f)(\altpoint)}
		.
	\end{eqsplit}
	Combining
		this,
		the assumption that
			for all
				$t\in[0,\infty)$
			it holds that
				$\Theta_t\in O$,
		the fact that 
			for all
				$ s, t \in [0,\infty) $
			it holds that
\begin{equation}
  \Theta_{s+t}=\Theta_s-\int_0^t (\nabla\f)(\Theta_{s+u})\,\diff u 
  ,
\end{equation}
and 
		\cref{lem:lyapunov_general}
		(applied for every $ s \in [0,\infty) $, $ t \in (s,\infty) $ 
		with
			$\d\is\d$,
			$T\is t-s$,
			$\alpha\is 0$,
			$O\is O$,
			$\beta\is (O\ni\altpoint\mapsto -\C^{-1}(1-\alpha)\Pnorm2{(\nabla\f)(\altpoint)}\in\R)$,
			$\g\is(\nabla \f)|_O$,
			$\Theta\is([0,t-s]\ni u\mapsto \Theta_{s+u}\in O)$
		in the notation of \cref{lem:lyapunov_general})
	\proves that for all
		$s,t\in[0,\infty)$
		with $s<t$
	it holds that
	\begin{eqsplit}
		&0< \abs{\f(\Theta_t)-\f(\cpoint)}^{1-\alpha}
		=
		U(\Theta_t)
		\\&\leq
		U(\Theta_s)+\int_s^t -\C^{-1}(1-\alpha)\Pnorm2{(\nabla\f)(\Theta_u)}\,\diff u
		\\&=
		\abs{\f(\Theta_s)-\f(\cpoint)}^{1-\alpha}-\C^{-1}(1-\alpha)\br*{\int_s^t \Pnorm2{(\nabla\f)(\Theta_u)}\,\diff u}
		.
	\end{eqsplit}
	\enum{
		This
	}[imply] that for all
		$s,t\in[0,\infty)$
		with $s< t$
	it holds that
	\begin{equation}
		\llabel{eq:1}
		\int_s^t \Pnorm2{(\nabla\f)(\Theta_u)}\,\diff u
		\leq
		\C(1-\alpha)^{-1}\abs{\f(\Theta_s)-\f(\cpoint)}^{1-\alpha}
		.
	\end{equation}
	\Hence that
	\begin{equation}
		\int_0^\infty \Pnorm2{(\nabla\f)(\Theta_s)}\,\diff s
		\leq
		\C(1-\alpha)^{-1}\abs{\f(\Theta_0)-\f(\cpoint)}^{1-\alpha}
		<
		\infty
	\end{equation}
		This
	demonstrates that
	\begin{equation}
		\llabel{eq:conv}
		\limsup_{r\to\infty}\int_r^\infty \Pnorm2{(\nabla\f)(\Theta_s)}\,\diff s
		=
		0
		.
	\end{equation}
	\Moreover
		the fundamental theorem of calculus
		and the assumption that 
			for all
				$t\in[0,\infty)$
			it holds that
				$\Theta_t=\Theta_0-\int_0^t(\nabla\f)(\Theta_s)\,\diff s$
	\prove that for all
		$r,s,t\in[0,\infty)$
		with $r\leq s\leq t$
	it holds that
	\begin{equation}
		\llabel{eq:5}
		\Pnorm2{\Theta_t-\Theta_s}
		=
		\Pnorm*2{\int_s^t (\nabla\f)(\Theta_u)\,\diff u}
		\leq
		\int_s^t\Pnorm2{(\nabla\f)(\Theta_u)}\,\diff u
		\leq
		\int_r^\infty \Pnorm2{(\nabla\f)(\Theta_u)}\,\diff u
		.
	\end{equation}
		This
		and \lref{eq:conv}
	\prove that there exists
		$\psi\in\R^\d$
	which satisfies
	\begin{equation}
		\llabel{eq:conv2}
		\limsup_{t\to\infty}\Pnorm2{\Theta_t-\psi}
		=
		0
		.
	\end{equation}
	Combining
		this
		and the assumption that
			$\f$ is continuous
	with
	\cref{eq:weak_error_with_rate_f_estimate}
	demonstrates that
	\begin{equation}
	\llabel{eq:risk_of_psi_equal_to_risk_of_vartheta}
		\f(\psi)
		=
		\f\bpr{\lim\nolimits_{t\to\infty}\Theta_t}
		=
		\lim\nolimits_{t\to\infty}\f(\Theta_t)
		=
		\f(\cpoint)
		.
	\end{equation}
	\Moreover 
		\lref{eq:conv2},
		\lref{eq:1},
		and \lref{eq:5}
	\prove[sindep] that for all
		$t\in[0,\infty)$
	it holds that
	\begin{eqsplit}
		\Pnorm2{\Theta_t-\psi}
		&=
		\Pnorm[\big]2{\Theta_t-\bbr{\lim\nolimits_{s\to\infty}\Theta_s}}
		\\&=
		\lim_{s\to\infty}\Pnorm2{\Theta_t-\Theta_s}
		\\&\leq
		\int_t^\infty \Pnorm2{(\nabla\f)(\Theta_u)}\,\diff u
		\\&\leq
		\C(1-\alpha)^{-1}\abs{\f(\Theta_t)-\f(\cpoint)}^{1-\alpha}
		.
	\end{eqsplit}
	Combining
		this
	with
	\cref{eq:weak_error_with_rate_f_estimate}
	and 
	\lref{eq:risk_of_psi_equal_to_risk_of_vartheta}
	establishes
    \cref{prop:gf.conv.simple.it:1,prop:gf.conv.simple.it:2,prop:gf.conv.simple.it:3}.
\end{aproof}
\endgroup

\section{Standard KL inequalities for monomials}

\cfclear
\begingroup
\providecommand{\d}{}
\renewcommand{\d}{\defaultParamDim}
\providecommand{\f}{}
\renewcommand{\f}{\defaultLossFunction}
\begin{athm}{lemma}{lem:kl-monom}[Standard \KL\ inequalities for monomials]
	Let 
		$\d \in \N$,
		$p\in(1,\infty)$,
		$\eps,c,\alpha\in(0,\infty)$
	satisfy
		$c\geq p^{-1}\eps^{p(\alpha-1)+1}$
		and $\alpha\geq 1-\frac1p$
	and let 
		$\f \colon \R^\d\to\R$ 
	satisfy 
		for all 
			$\altpoint\in\R^\d$
		that
		\begin{equation}
			\f(\altpoint)=\Pnorm2\altpoint^p.
		\end{equation}
	Then
	\begin{enumerate}[(i)]
		\item \llabel{it:1}
		it holds that $\f\in C^1(\R^\d,\R)$ and
		\item \llabel{it:2}
		it holds for all
			$\altpoint\in\{\altpointTwo\in\R^\d\colon \Pnorm2 \altpointTwo\leq\eps\}$
		that 
		\begin{equation}
			\abs{\f(0)-\f(\altpoint)}^\alpha
			\leq 
			c\Pnorm2{(\nabla \f)(\altpoint)}
			.
		\end{equation}	
	\end{enumerate}
\end{athm}

\begin{aproof}
	First, \nobs that 
		the fact that
			for all
				$\altpoint\in\R^\d$
			it holds that
\begin{equation}
				\f( \altpoint ) = ( \Pnorm2\altpoint^2 )^{ \nicefrac{ p }{ 2 } } 
\end{equation}
	\proves that for all
		$\altpoint\in\R^\d$
	it holds that $\f \in C^1(\R^\d,\R)$ and
	\begin{equation}
		\llabel{eq:1}
		\Pnorm2{(\nabla \f)(\altpoint)}
		=
		p\Pnorm2\altpoint^{p-1}
		.
	\end{equation}
	\Moreover
		the assumption that
			$\alpha\geq 1-\tfrac1p$
		ensures that
			$p(\alpha-1)+1\geq 0$.
	The assumption that
		$c\geq p^{-1}\eps^{p(\alpha-1)+1}$
		\hence
	\proves that for all
		$\altpoint\in\{\altpointTwo\in\R^\d\colon \Pnorm2 \altpointTwo\leq\eps\}$
	it holds that
	\begin{equation}
		\llabel{eq:2}
		\Pnorm2{\altpoint}^{p\alpha}\Pnorm2{\altpoint}^{-(p-1)}
		=
		\Pnorm2{\altpoint}^{p(\alpha-1)+1}
		\leq
		\eps^{p(\alpha-1)+1}
		\leq 
		cp.
	\end{equation}
	Combining
		\lref{eq:1}
		and \lref{eq:2}
	\proves that for all
		$\altpoint\in\{\altpointTwo\in\R^\d\colon \Pnorm2 \altpointTwo\leq\eps\}$
	it holds that
	\begin{equation}
		\abs{\f(0)-\f(\altpoint)}^\alpha
		=
		\Pnorm2{\altpoint}^{p\alpha}
		\leq
		cp\Pnorm2{\altpoint}^{p-1}
		=
		c\Pnorm2{(\nabla \f)(\altpoint)}.
	\end{equation}
	\finishproofthis
\end{aproof}
\endgroup

\section{Standard KL inequalities around non-critical points}

\cfclear
\begingroup
\providecommand{\d}{}
\renewcommand{\d}{\defaultParamDim}
\providecommand{\f}{}
\renewcommand{\f}{\defaultLossFunction}
\begin{athm}{lemma}{lem:non_critical}[Standard \KL\ inequality around non-critical points]
Let $ \d \in \N $, 
let $ U \subseteq \R^{ \d } $ be open, 
and 
let 
$ \f \in C^1( U, \R ) $, 
$ \cpoint \in U $, 
$ c \in [0,\infty) $, $ \alpha \in (0,\infty) $
satisfy for all 
$
  \altpoint \in U
$
that 
\begin{equation}
\label{eq:epsilon_property_nabla_f}
\textstyle 
    \max\{
      \abs{ \f( \cpoint ) - \f( \altpoint ) }^{ \alpha }
      ,
      c
      \Pnorm2{ ( \nabla \f )( \cpoint ) - ( \nabla \f )( \altpoint ) }
    \}
  \leq 
  \frac{ c \Pnorm2{ ( \nabla \f )( \cpoint ) } }{ 2 }
\end{equation}
\cfload.
Then it holds for all 
$
  \altpoint \in U
$
that 
\begin{equation}
  \abs{ \f( \cpoint ) - \f( \altpoint ) }^{ \alpha }
  \leq 
  c 
  \pnorm2{ ( \nabla \f )( \altpoint ) }
  .
\end{equation}
\end{athm}
\begin{aproof}
\Nobs that 
\cref{eq:epsilon_property_nabla_f}  
and the triangle inequality 
ensure that 
for all $ \altpoint \in U $ it holds that
\begin{equation}
\begin{split}
&
  c \pnorm2{ ( \nabla \f )( \cpoint ) }
\\ 
& =
  c \pnorm2{ ( \nabla \f )( \altpoint ) + [ ( \nabla \f )( \cpoint ) - ( \nabla \f )( \altpoint ) ] }
\\ & 
\leq 
  c \pnorm2{ ( \nabla \f )( \altpoint ) } 
  + 
  c \pnorm2{ ( \nabla \f )( \cpoint ) - ( \nabla \f )( \altpoint ) }
\leq 
  c \pnorm2{ ( \nabla \f )( \altpoint ) } 
  + 
  \tfrac{
    c \pnorm2{ ( \nabla \f )( \cpoint ) } 
  }{
    2
  }
  .
\end{split}
\end{equation}
\Hence 
for all 
$ \altpoint \in U $
that 
\begin{equation}
\textstyle 
  \frac{ 
    c \pnorm2{ ( \nabla \f )( \cpoint ) }
  }{ 2 }
\leq 
  c \pnorm2{ ( \nabla \f )( \altpoint ) } 
  .
\end{equation}
Combining this with \cref{eq:epsilon_property_nabla_f} \proves 
that for all $ \altpoint \in U $ it holds that 
\begin{equation}
\begin{split}
\textstyle 
  \abs{ \f( \cpoint ) - \f( \altpoint ) }^{ \alpha }
&
\textstyle 
  \leq 
  \frac{ c \pnorm2{ ( \nabla \f )( \cpoint ) } }{ 2 }
  \leq 
  c \pnorm2{ ( \nabla \f )( \altpoint ) }
  .
\end{split}
\end{equation}
\end{aproof}
\endgroup

\cfclear
\begingroup
\providecommand{\d}{}
\renewcommand{\d}{\defaultParamDim}
\providecommand{\f}{}
\renewcommand{\f}{\defaultLossFunction}
\begin{athm}{cor}{cor:non_critical2}[Standard \KL\ inequality around non-critical points]
Let $ \d \in \N $, 
let $U \subseteq \R^{ \d }$ be open, 
let $ \f \in C^1( U, \R ) $, 
$ \cpoint \in U $, 
$ c, \alpha \in (0,\infty) $
satisfy 
$
  ( \nabla \f )( \cpoint ) \neq 0
$. 
Then there exists $ \varepsilon \in (0,1) $ such that 
for all 
$
  \altpoint \in \{ \altpointTwo \in U \colon \pnorm2{ \altpointTwo - \cpoint } < \varepsilon \}
$
it holds that 
\begin{equation}
\label{eq:KL_non_critical_statement}
  \abs{ \f( \cpoint ) - \f( \altpoint ) }^{ \alpha }
  \leq 
  c 
  \Pnorm2{ ( \nabla \f )( \altpoint ) }
\end{equation}
\cfout.
\end{athm}
\begin{aproof}
\Nobs that the assumption 
that $ \f \in C^1( U, \R ) $ ensures that 
\begin{equation}
\textstyle 
  \limsup_{ \varepsilon \searrow 0 }
  \bigl(
    \sup_{
      \altpoint \in \{ \altpointTwo \in U \colon \pnorm2{ \altpointTwo - \cpoint } < \varepsilon \}
    }
    \Pnorm2{ ( \nabla \f )( \cpoint ) - ( \nabla \f )( \altpoint ) }
  \bigr)
  = 0
\end{equation}
\cfload.
Combining this and the fact that $ \alpha > 0 $ 
with the fact that $ \f $ is continuous demonstrates that 
\begin{equation}
\textstyle 
  \limsup_{ \varepsilon \searrow 0 }
  \bigl(
    \sup_{
      \altpoint \in \{ \altpointTwo \in U \colon \Pnorm2{ \altpointTwo - \cpoint } < \varepsilon \}
    }
    \max\bigl\{
      \abs{ \f( \cpoint ) - \f( \altpoint ) }^{ \alpha }
      ,
      c
      \pnorm2{ ( \nabla \f )( \cpoint ) - ( \nabla \f )( \altpoint ) }
    \bigr\}
  \bigr)
  = 0
  .
\end{equation}
The fact that $ c > 0 $
and the fact that 
$
  \pnorm2{ ( \nabla \f )( \cpoint ) } > 0
$
\hence \prove that 
there exists $ \varepsilon \in (0,1) $ which satisfies 
\begin{equation}
\label{eq:in_proof_construction_of_positive_eps}
\textstyle 
    \sup_{
      \altpoint \in \{ \altpointTwo \in U \colon \pnorm2{ \altpointTwo - \cpoint } < \varepsilon \}
    }
    \max\{
      \abs{ \f( \cpoint ) - \f( \altpoint ) }^{ \alpha }
      ,
      c
      \pnorm2{ ( \nabla \f )( \cpoint ) - ( \nabla \f )( \altpoint ) }
    \}
  <
  \frac{ c \pnorm2{ ( \nabla \f )( \cpoint ) } }{ 2 }
  .
\end{equation}
\Nobs that \cref{eq:in_proof_construction_of_positive_eps} ensures that 
for all 
$
  \altpoint \in \{ \altpointTwo \in U \colon \Pnorm2{ \altpointTwo - \cpoint } < \varepsilon \}
$
it holds that 
\begin{equation}
\textstyle 
  \max\{
    \abs{ \f( \cpoint ) - \f( \altpoint ) }^{ \alpha }
    ,
    c
    \pnorm2{ ( \nabla \f )( \cpoint ) - ( \nabla \f )( \altpoint ) }
  \}
  \leq 
  \frac{ c \pnorm2{ ( \nabla \f )( \cpoint ) } }{ 2 }
  .
\end{equation}
This and \cref{lem:non_critical} establish \cref{eq:KL_non_critical_statement}. 
\end{aproof}
\endgroup

\section{Standard KL inequalities with increased exponents}

\begingroup
\providecommand{\d}{}
\renewcommand{\d}{\defaultParamDim}
\providecommand{\f}{}
\renewcommand{\f}{\defaultLossFunction}
\providecommand{\g}{}
\renewcommand{\g}{\defaultGradientFunction}
\begin{athm}{lemma}{lem:KL_increasing_alpha}[Standard \KL\ inequalities with increased exponents]
Let $ \d \in \N $, 
let $ U \subseteq \R^{ \d } $ be a set, 
let $ \cpoint \in U $, 
$ \fc, \alpha \in (0,\infty) $, 
let 
$ \f \colon U \to \R $ and 
$ \g \colon U \to \R $
satisfy for all $ \altpoint \in U $ that 
\begin{equation}
\label{eq:KL_increasing_alpha_assumption}
  \abs{ \f( \cpoint ) - \f( \altpoint ) }^{ \alpha }
  \leq 
  \fc \abs{ \g( \altpoint ) }
  ,
\end{equation}
and let 
$ \beta \in ( \alpha, \infty) $, 
$ \fC \in \R $
satisfy 
$
  \fC 
  =   
  \fc 
  ( 
    \sup_{ \altpoint \in U }
    \abs{
      \f( \cpoint ) - \f( \altpoint )
    }^{
      \beta - \alpha
    }
  )
$. 
Then it holds for all 
$ 
  \altpoint \in U 
$
that 
\begin{equation}
\label{eq:KL_increasing_alpha_statement}
\textstyle 
  \abs{ \f( \cpoint ) - \f( \altpoint ) }^{ \beta }
  \leq 
  \fC
  \abs{ \g( \altpoint ) }
  .
\end{equation}
\end{athm}
\begin{aproof}
\Nobs that 
\cref{eq:KL_increasing_alpha_assumption}
\proves that 
for all $ \altpoint \in U $ it holds that 
\begin{equation}
\begin{split}
\textstyle 
  \abs{ \f( \cpoint ) - \f( \altpoint ) }^{ \beta }
& =
  \abs{ \f( \cpoint ) - \f( \altpoint ) }^{ \alpha }
  \abs{ \f( \cpoint ) - \f( \altpoint ) }^{ \beta - \alpha }
\leq 
  \bigl(
    \fc 
    \abs{ \g( \altpoint ) }
  \bigr)
  \abs{ \f( \cpoint ) - \f( \altpoint ) }^{ \beta - \alpha }
\\ &
=
  \bigl(
    \fc 
    \abs{ \f( \cpoint ) - \f( \altpoint ) }^{ \beta - \alpha }
  \bigr)
  \abs{ \g( \altpoint ) }
\leq 
  \fC 
  \abs{ \g( \altpoint ) }
  .
\end{split}
\end{equation}
This establishes \cref{eq:KL_increasing_alpha_statement}. 
\end{aproof}

\cfclear
\begingroup
\providecommand{\d}{}
\renewcommand{\d}{\defaultParamDim}
\providecommand{\f}{}
\renewcommand{\f}{\defaultLossFunction}
\begin{athm}{cor}{cor:KL_increasing_alpha}[Standard \KL\ inequalities with increased exponents]
Let $ \d \in \N $, 
$ \f \in C^1( \R^{ \d }, \R ) $,  
$ \cpoint \in \R^{ \d } $, 
$ \varepsilon, \fc, \alpha \in (0,\infty) $, 
$ \beta \in [ \alpha, \infty) $
satisfy for all 
$ 
  \altpoint \in \{ \altpointTwo \in \R^{ \d } \colon \pnorm2{ \altpointTwo - \cpoint } < \varepsilon \}
$ 
that 
\begin{equation}
  \abs{ \f( \cpoint ) - \f( \altpoint ) }^{ \alpha }
  \leq 
  \fc 
  \Pnorm2{ ( \nabla \f )( \altpoint ) }
\end{equation}
\cfload.
Then there exists $ \fC \in (0,\infty) $ 
such that for all 
$ 
  \altpoint \in \{ \altpointTwo \in \R^{ \d } \colon \pnorm2{ \altpointTwo - \cpoint } < \varepsilon \}
$
it holds that 
\begin{equation}
\label{eq:KL_estimate_statement}
\textstyle 
  \abs{ \f( \cpoint ) - \f( \altpoint ) }^{ \beta }
  \leq 
  \fC
  \pnorm2{ ( \nabla \f )( \altpoint ) }
  .
\end{equation}
\end{athm}
\begin{aproof}
\Nobs that \cref{lem:KL_increasing_alpha}
establishes \cref{eq:KL_estimate_statement}. 
\end{aproof}
\endgroup

\section{Standard KL inequalities for coercive-type functions}

\cfclear
\begingroup
\providecommand{\d}{}
\renewcommand{\d}{\defaultParamDim}
\providecommand{\f}{}
\renewcommand{\f}{\defaultLossFunction}
\providecommand{\g}{}
\renewcommand{\g}{\defaultGradientFunction}
\begin{athm}{lemma}{lem:growth_for_KL_coercive}[On a growth bound on the gradient]
Let 
	$\d \in \N$, $\cpoint \in \R^{\d}$, $L, \rho \in (0,\infty)$, $r \in (0,\infty]$, 
	$\mathbb{B} = \{\altpointTwo \in \R^\d \colon \pnorm2{\altpointTwo-\cpoint} < r \}$, 
	$\f \in C^1(\R^{\d}, \R)$ satisfy for all $\altpoint \in \mathbb{B}$ that
\begin{equation}
\label{lem:growth_for_KL_coercive:ass1}
	\pnorm2{(\nabla \f)(\altpoint)} \leq L \pnorm2{\altpoint-\cpoint}^\rho
\end{equation}
\cfload.
Then it holds for all $\altpoint \in \mathbb{B}$ that
\begin{equation}
\label{lem:growth_for_KL_coercive:concl1}
	\abs{\f(\altpoint) - \f(\cpoint)} \leq L \pnorm2{\altpoint-\cpoint}^{\rho+1}.
\end{equation}
\end{athm}

\begin{aproof}
Observe that \cref{lem:growth_for_KL_coercive:ass1}, the fundamental theorem of calculus, and the Cauchy-Schwarz inequality assure that for all $\altpoint \in \mathbb{B}$ it holds that
\begin{equation}
\begin{split}
\abs{\f(\altpoint) - \f(\cpoint)} &=   \abs*{ \bbr{\f(\cpoint + h (\altpoint - \cpoint))}_{h = 0}^{h=1}} \\
&=\abs*{ \int_0^1 \f'(\cpoint + h(\altpoint-\cpoint))(\altpoint-\cpoint) \,\diff h }\\
&= \abs*{ \int_0^1 \scp[\big]{(\nabla \f)\bpr{\cpoint + h(\altpoint-\cpoint)},\altpoint-\cpoint } \,\diff h  } \\
&\leq  \int_0^1 \pnorm2{(\nabla \f)\bpr{\cpoint + h(\altpoint-\cpoint)}}\pnorm2{\altpoint-\cpoint} \,\diff h \\
&\leq \int_0^1 L\pnorm2{\cpoint + h(\altpoint-\cpoint) - \cpoint}^\rho\pnorm2{\altpoint-\cpoint } \,\diff h \\
& = L\pnorm2{\altpoint-\cpoint}^{\rho+1} \int_0^1 h^\rho \,\diff h 
= \frac{L \pnorm2{\altpoint-\cpoint}^{\rho+1}}{\rho+1} 
\leq L \pnorm2{\altpoint-\cpoint}^{\rho+1}.
\end{split}
\end{equation}
\end{aproof}
\endgroup

\cfclear
\begingroup
\providecommand{\d}{}
\renewcommand{\d}{\defaultParamDim}
\providecommand{\f}{}
\renewcommand{\f}{\defaultLossFunction}
\providecommand{\g}{}
\renewcommand{\g}{\defaultGradientFunction}
\begin{athm}{lemma}{lem:coercive_are_KL_explicit}[Explicit \KL\ inequalities for coercive-type functions]
Let 
	$\d \in \N$, $\cpoint \in \R^{\d}$, $c, \rho \in (0,\infty)$, $\f \in C^1(\R^{\d}, \R)$
satisfy for all $\altpoint \in \R^{\d}$ that
\begin{equation}
\label{lem:coercive_are_KL_explicit:ass1}
\begin{split}
	\scp{\altpoint-\cpoint, (\nabla \f)(\altpoint) }\geq  c\pnorm2{\altpoint-\cpoint}^2
\end{split}
\end{equation}
and assume that $\nabla \f$ is locally $\rho$-H\"older continuous \cfload.
Then
\begin{enumerate}[(i)]
\item \label{lem:coercive_are_KL_explicit:item1}
for every 
	$\altpoint \in \R^\d \backslash \{\cpoint\}$,
	$\alpha, \mathfrak{C} \in (0, \infty)$
	there exists $\varepsilon \in (0,1)$ such that for all
    $\altpointThree \in \{ \altpointTwo \in \R^\d \colon \pnorm2{\altpointTwo-\altpoint} < \varepsilon \}$
	it holds that
	\begin{equation}
	\abs{\f(\altpoint) - \f(\altpointThree)}^\alpha \leq \mathfrak{C} \pnorm2{(\nabla \f)(\altpointThree)}
	\end{equation}
and
\item \label{lem:coercive_are_KL_explicit:item2}
for every $\altpoint \in \R^\d$, $\alpha \in [1/(1+\rho), \infty)$
there exist  $\mathfrak{C} \in (0,\infty)$, $\varepsilon \in (0,1)$ such that for all $\altpointThree \in \{ \altpointTwo \in \R^\d \colon \pnorm2{\altpointTwo-\altpoint} < \varepsilon \}$ it holds that 
\begin{equation}
	\abs{\f(\altpoint) - \f(\altpointThree)}^{\alpha} \leq \mathfrak{C} \pnorm2{(\nabla \f)(\altpointThree)}
\end{equation}
\end{enumerate}
\cfload.
\end{athm}

\begin{aproof}
Throughout this proof, let 
$L \in (0,\infty)$, $\epsilon \in (0,1)$
satisfy for all
$\altpoint, \altpointThree \in \{ \altpointTwo \in \R^\d \colon \pnorm2{\altpointTwo-\cpoint} < \epsilon \}$
that
\begin{equation}
\label{lem:coercive_are_KL_explicit:setting1}
	\pnorm2{(\nabla \f)(\altpoint) - (\nabla \f)(\altpointThree) } \leq L \pnorm2{\altpoint-\altpointThree}^\rho.
\end{equation}
\Nobs thatx
\enum{
	the Cauchy-Schwarz inequality;
	\cref{lem:coercive_are_KL_explicit:ass1}
}
\prove that
for all $\altpoint \in \R^\d \backslash \{\cpoint\}$ it holds that
\begin{equation}
	\pnorm2{\altpoint-\cpoint} \pnorm2{(\nabla \f)(\altpoint)}
\geq
	\scp{\altpoint-\cpoint, (\nabla \f)(\altpoint) }
\geq  
	c\pnorm2{\altpoint-\cpoint}^2.
\end{equation}
This \proves that for all $\altpoint \in \R^\d \backslash \{\cpoint\}$ it holds that
\begin{equation}
\label{lem:coercive_are_KL_explicit:eq1}
	\pnorm2{(\nabla \f)(\altpoint)}
\geq
	c \pnorm2{\altpoint-\cpoint}
>
	0.
\end{equation}
\Hence that for all $\altpoint \in \R^\d \backslash \{\cpoint\}$ it holds that
$(\nabla \f)(\altpoint) \neq 0$.
\cref{cor:non_critical2} \hence
\proves that 
for all $\altpoint \in \R^\d \backslash \{\cpoint\}$,
$ \alpha, \mathfrak{C} \in (0,\infty) $
there exists $ \varepsilon \in (0,1) $ such that 
for all 
$
  \altpoint \in \{ \altpointTwo \in \R^{ \d } \colon \pnorm2{ \altpointTwo - \altpoint } < \varepsilon \}
$
it holds that 
\begin{equation}
\label{lem:coercive_are_KL_explicit:eq2}
\abs{\f(\altpoint) - \f(\altpoint)}^\alpha \leq \mathfrak{C} \pnorm2{(\nabla \f)(\altpoint)}.
\end{equation}
This establishes \cref{lem:coercive_are_KL_explicit:item1}.
\Moreover 
\enum{
	\cref{lem:coercive_are_KL_explicit:ass1};
	\cref{fcond1:item3} in \cref{fcond1};
}
\prove that $(\nabla \f)(\cpoint) = 0$.
\enum{
	This;
	\cref{lem:coercive_are_KL_explicit:setting1}
} \prove that for all $\altpoint \in \{ \altpointTwo \in \R^\d \colon \pnorm2{\altpointTwo-\cpoint} < \epsilon \}$ it holds that
\begin{equation}
	\pnorm2{(\nabla \f)(\altpoint)} = \pnorm2{(\nabla \f)(\altpoint) - (\nabla \f)(\cpoint)} \leq L \pnorm2{\altpoint-\cpoint}^\rho.
\end{equation}
Combining this with
\enum{
	\cref{lem:coercive_are_KL_explicit:eq1};
	\cref{lem:growth_for_KL_coercive};
}
\proves
that for all $\altpoint \in \{ \altpointTwo \in \R^\d \colon \pnorm2{\altpointTwo-\cpoint} < \epsilon \}$ it holds that
\begin{equation}
	\abs{\f(\altpoint) - \f(\cpoint)}
\leq
	L \pnorm2{\altpoint-\cpoint}^{\rho+1}
\leq
	L \pr[\big]{
		\tfrac{1}{c}
		\pnorm2{(\nabla \f)(\altpoint)}
	}^{\rho+1}.
\end{equation}
\Hence that for all $\altpoint \in \{ \altpointTwo \in \R^\d \colon \pnorm2{\altpointTwo-\cpoint} < \epsilon \}$ it holds that
\begin{equation}
	\abs{\f(\altpoint) - \f(\cpoint)}^{\nicefrac{1}{(1+\rho)}}
\leq
	L^{\nicefrac{1}{(1+\rho)}}
	c \pnorm2{(\nabla \f)(\altpoint)}.
\end{equation}
This and \cref{cor:KL_increasing_alpha} 
\prove that for all
$\alpha \in [1/(1+\rho), \infty)$
there exists $\mathfrak{C} \in (0,\infty)$ such that for all $\altpoint \in \{ \altpointTwo \in \R^\d \colon \pnorm2{\altpointTwo-\cpoint} < \epsilon \}$ it holds that
\begin{equation}
	\abs{\f(\altpoint) - \f(\cpoint)}^{\alpha} \leq \mathfrak{C} \pnorm2{(\nabla \f)(\altpoint)}.
\end{equation}
This and \cref{lem:coercive_are_KL_explicit:eq2} establish \cref{lem:coercive_are_KL_explicit:item2}.
\end{aproof}
\endgroup

\cfclear
\begingroup
\providecommand{\d}{}
\renewcommand{\d}{\defaultParamDim}
\providecommand{\f}{}
\renewcommand{\f}{\defaultLossFunction}
\providecommand{\g}{}
\renewcommand{\g}{\defaultGradientFunction}
\begin{athm}{prop}{prop:template}[Coercive-type functions are standard \KL\ functions]
Let 
	$\d \in \N$
and let\cfadd{def:coercivityI} $\f \in C^2(\R^{\d}, \R)$ be a coercive-type function \cfload.
Then\cfadd{def:KLfunction_standard} $\f$ is a standard \KL\ function \cfload.
\end{athm}

\begin{aproof}
\Nobs that the fact that $\nabla \f$ is continuously differentiable 
\proves that $\nabla \f$ is locally 1-H\"older continuous.
Combining
\enum{
	this;
	the assumption that $\f$ is a coercive-type function;
	\cref{lem:coercive_are_KL_explicit:item2} in \cref{lem:coercive_are_KL_explicit};
}
\proves that for every $\altpoint \in \R^\d$, $\alpha \in [1/2, \infty)$
there exist $c \in (0,\infty)$, $\varepsilon \in (0,1)$
such that for all $\altpointThree \in \{ \altpointTwo \in \R^\d \colon \pnorm2{\altpointTwo-\altpoint} < \varepsilon \}$ it holds that
\begin{equation}
	\abs{\f(\altpoint) - \f(\altpointThree)}^{\alpha} \leq c \pnorm2{(\nabla \f)(\altpointThree)}.
\end{equation}
\enum{
	This;
	\cref{def:KLinequality_standard:eq1}
} \prove that $\f$ is a standard \KL\ function.
\end{aproof}
\endgroup

\section{Standard KL inequalities for one-dimensional polynomials}
\label{sect:KL_onedim_poly}

\begingroup
\providecommand{\f}{}
\renewcommand{\f}{p}
\begin{adef}{def:polynomial}[Polynomial]
Let $d, \delta \in \N$ and let $ \f \colon \R^d \to \R^\delta$ be a function.
Then we say that $\f$ is a polynomial if and only if there exist 
	$N \in \N$
	and
	$c = (c_\alpha)_{\alpha \in \{0, 1, \dots, N\}^d} \colon \{0, 1,\allowbreak \ldots,\allowbreak N\}^d \to \R^\delta$
such that for all $\altpoint = (\altpoint_1, \dots, \altpoint_d) \in \R^d$ it holds that
\begin{equation}
	\label{def:polynomial:eq1}
	\f(\altpoint) = \sum_{\alpha = (\alpha_1, \dots, \alpha_d) \in \{0, 1, \dots, N\}^d} c_\alpha (\altpoint_1)^{\alpha_1} (\altpoint_2)^{\alpha_2} \cdot \ldots \cdot (\altpoint_d)^{\alpha_d}.
\end{equation}
\end{adef}
\endgroup

\begingroup
\providecommand{\p}{}
\renewcommand{\p}{p}
\begin{athm}{cor}{cor:reparametrization}[Reparametrization]
Let 
$ \cpoint \in \R $, $ N \in \N $, 
$ \p \in C^{ \infty }( \R , \R ) $ satisfy for all $ \altpoint \in \R $ that 
$
  \p^{ (N+1) }( \altpoint ) = 0
$
and let $ \beta_0, \beta_1, \dots, \beta_N \in \R $
satisfy for all 
$ n \in \{ 0, 1, \dots, N \} $ that
$
  \beta_n = 
  \frac{ \p^{ (n) }( \cpoint ) }{
    n! 
  }
$. 
Then it holds for all $ \altpoint \in \R $ that 
\begin{equation}
	\llabel{claim}
\textstyle 
  \p( \altpoint ) = \sum_{ n = 0 }^N \beta_n ( \altpoint - \cpoint )^n
  .
\end{equation}
\end{athm}
\begin{aproof}
\Nobs that \cref{thm:taylor_formula}
establishes \lref{claim}.
\end{aproof}
\endgroup

\cfclear
\begingroup
\providecommand{\p}{}
\renewcommand{\p}{p}
\begin{athm}{cor}{cor:equivalent_cond_oned_poly}[Equivalent conditions for one-dimensional polynomials]
Let $\p \colon \R \to \R$ be a function.
Then the following four statements are equivalent.
\begin{enumerate}[label=(\roman*)]
	\item \label{cor:equivalent_cond_oned_poly:item1} It holds that $\p$ is\cfadd{def:polynomial} a polynomial \cfload.
	\item \label{cor:equivalent_cond_oned_poly:item2} There exists $N \in \N$ such that for all $\altpoint \in \R$ it holds that $\p \in C^{N}(\R,\R)$ and $\p^{(N)}(\altpoint) = 0$.
	\item \label{cor:equivalent_cond_oned_poly:item3} 
	For every $\cpoint \in \R$ there exist $N \in \N$, $\beta_0, \beta_1, \dots, \beta_N \in \R$
	such that for all $\altpoint \in \R$ it holds that
	\begin{equation}
		\label{cor:equivalent_cond_oned_poly:eq0}
		\p(\altpoint) = \sum_{n=0}^N \beta_n (\altpoint-\cpoint)^n.
	\end{equation}
	\item \label{cor:equivalent_cond_oned_poly:item4} There exists $N \in \N$, $\beta_0, \beta_1, \dots, \beta_N \in \R$
	such that for all $\altpoint \in \R$ it holds that
	\begin{equation}
		\label{cor:equivalent_cond_oned_poly:eq1}
		\p(\altpoint) = \sum_{n=0}^N \beta_n \altpoint^n.
	\end{equation}
\end{enumerate}
\end{athm}
\begin{aproof}
\Nobs that 
\enum{
	\cref{def:polynomial:eq1};
	\cref{cor:equivalent_cond_oned_poly:eq1}
}establish that 
(\ref{cor:equivalent_cond_oned_poly:item1} $\leftrightarrow$ \ref{cor:equivalent_cond_oned_poly:item4}).
\Nobs that 
\enum{
	the fact that for all $N \in \N$, $g \in C^N(\R,\R)$ with $\forall \, \altpoint \in \R \colon g^{(N)}(\altpoint) = 0$ it holds that $g \in C^\infty(\R,\R)$;
	\cref{cor:reparametrization};
}establish that
(\ref{cor:equivalent_cond_oned_poly:item2} $\rightarrow$ \ref{cor:equivalent_cond_oned_poly:item3}).
\Nobs that 
\enum{
	\cref{cor:equivalent_cond_oned_poly:eq0};
	\cref{cor:equivalent_cond_oned_poly:eq1};
}establish that 
(\ref{cor:equivalent_cond_oned_poly:item3} $\rightarrow$ \ref{cor:equivalent_cond_oned_poly:item4}).
\Nobs that 
\enum{
	\cref{cor:equivalent_cond_oned_poly:eq1};
}establishes that 
(\ref{cor:equivalent_cond_oned_poly:item4} $\rightarrow$ \ref{cor:equivalent_cond_oned_poly:item2}).
\end{aproof}
\endgroup

\begingroup
\providecommand{\p}{}
\renewcommand{\p}{p}
\begin{athm}{cor}{cor:Kurdyka}[Quantitative standard \KL\ inequalities for non-constant one-dimensional polynomials]
Let 
$ \cpoint \in \R $, $ N \in \N $, 
$ \p \in C^{ \infty }( \R , \R ) $ satisfy for all $ \altpoint \in \R $ that 
$
  \p^{ (N+1) }( \altpoint ) = 0
$, 
let 
$ \beta_0, \beta_1, \dots, \beta_N \in \R $
satisfy for all 
$ n \in \{ 0, 1, \dots, N \} $ that
$
  \beta_n = \frac{ \p^{ (n) }( \cpoint ) }{ n! }
$,
and let 
$ \mathscr{m} \in \{ 1, 2, \dots, N \} $, 
$ \alpha \in [0,1] $, 
$ c, \varepsilon \in \R $
satisfy 
\begin{equation}
\textstyle 
   \beta_{ \mathscr{m} } \neq 0 
  = \sum_{ n = 1 }^{ \mathscr{m} - 1 } \abs{ \beta_n } 
  ,
  \qquad 
  \alpha \geq 1 - \mathscr{m}^{ - 1 }
  ,
  \qquad 
  c = 
  2
  \bigl[ 
    \sum_{ n = 1 }^N
    \frac{ 
      \abs{ \beta_n }^{ \alpha } 
    }{
      \abs{ \beta_{ \mathscr{m} } \mathscr{m} }
    }
  \bigr]   
  ,
\end{equation}
and 
$
  \varepsilon = 
  \frac{ 1 }{ 2 }
  [
      \sum_{ n = 1 }^N
      \frac{ 
        \abs{ \beta_n n } 
      }{
        \abs{ \beta_{ \mathscr{m} } \mathscr{m} }
      }
  ]^{ - 1 }
$.
Then it holds for all 
$ \altpoint \in [ \cpoint - \varepsilon, \cpoint + \varepsilon ] $ 
that 
\begin{equation} 
\label{eq:KL_inequality_1d_polynomial}
  \abs{ \p(\altpoint) - \p( \cpoint ) }^{ \alpha }
\leq 
  c \abs{ \p'(\altpoint) }
  .
\end{equation}
\end{athm}
\begin{aproof}
\Nobs that 
\cref{cor:reparametrization} 
ensures that for all $ \altpoint \in \R $ it holds that
\begin{equation}
\textstyle
  \p(\altpoint) - \p( \cpoint )
  =
  \sum_{ n = 1 }^N
  \beta_n ( \altpoint - \cpoint )^n.
\end{equation}
\Hence for all $ \altpoint \in \R $ that
\begin{equation}
\textstyle
  \p'(\altpoint)
  =
  \sum_{ n = 1 }^N
  \beta_n n ( \altpoint - \cpoint )^{ n - 1 }
\end{equation}
\Hence for all $ \altpoint \in \R $ that
\begin{equation}\llabel{1}
\textstyle
  \p(\altpoint) - \p(\cpoint)
  =
  \sum_{ n = \mathscr{m} }^N
  \beta_n ( \altpoint - \cpoint )^n
\qandq
  \p'(\altpoint)
  =
  \sum_{ n = \mathscr{m} }^N
  \beta_n n ( \altpoint - \cpoint )^{ n - 1 }
  .
\end{equation}
\Hence for all $ \altpoint \in \R $ that
\begin{equation}
\textstyle
  \abs{ \p(\altpoint) - \p(\cpoint) }^{ \alpha }
\leq 
  \sum_{ n = \mathscr{m} }^N
  \bigl(
    \abs{ \beta_n }^{ \alpha } 
    \abs{ \altpoint - \cpoint }^{ n \alpha }
  \bigr)
  .
\end{equation}
The fact that 
for all 
$ n \in \{ \mathscr{m}, \mathscr{m} + 1, \dots, N \} $, 
$ \altpoint \in \R $ with 
$ \abs{ \altpoint - \cpoint } \leq 1 $
it holds that
$ 
  \abs{ \altpoint - \cpoint }^{ n \alpha }
  \leq 
  \abs{ \altpoint - \cpoint }^{ n (1-\mathscr{m}^{-1}) }
  \leq 
  \abs{ \altpoint - \cpoint }^{ \mathscr{m}(1-\mathscr{m}^{-1}) }
  =
  \abs{ \altpoint - \cpoint }^{ \mathscr{m} - 1 } 
$
\hence 
\proves that 
for all $ \altpoint \in \R $ 
with $ \abs{ \altpoint - \cpoint } \leq 1 $ 
it holds that
\begin{equation}
\begin{split}
\textstyle
  \abs{ \p(\altpoint) - \p(\cpoint) }^{ \alpha }
& 
\textstyle 
  \leq 
  \sum_{ n = \mathscr{m} }^N
  \abs{ \beta_n }^{ \alpha } 
  \abs{ \altpoint - \cpoint }^{ n \alpha }
\leq
  \sum_{ n = \mathscr{m} }^N
  \abs{ \beta_n }^{ \alpha } 
  \abs{ \altpoint - \cpoint }^{ \mathscr{m} - 1 }
\\ & 
\textstyle 
=
  \abs{ \altpoint - \cpoint }^{ \mathscr{m} - 1 }
  \bigl[ 
    \sum_{ n = \mathscr{m} }^N
    \abs{ \beta_n }^{ \alpha } 
  \bigr] 
=
  \abs{ \altpoint - \cpoint }^{ \mathscr{m} - 1 }
  \bigl[ 
    \sum_{ n = 1 }^N
    \abs{ \beta_n }^{ \alpha } 
  \bigr]
  .
\end{split}
\end{equation}
\Hence 
for all $ \altpoint \in \R $ 
with $ \abs{ \altpoint - \cpoint } \leq 1 $ 
that
\begin{equation}
\label{eq:polnomial_difference_estimate_in_proof_KL}
\begin{split}
\textstyle
  \abs{ \p(\altpoint) - \p(\cpoint) }^{ \alpha }
& 
\textstyle 
  \leq 
  \abs{ \altpoint - \cpoint }^{ \mathscr{m} - 1 }
  \bigl[ 
    \sum_{ n = 1 }^N
    \abs{ \beta_n }^{ \alpha } 
  \bigr]
= 
  \frac{ c }{ 2 }
  \abs{ \altpoint - \cpoint }^{ \mathscr{m} - 1 }
  \abs{ \beta_{ \mathscr{m} } \mathscr{m} }
  .
\end{split}
\end{equation}
\Moreover 
\lref{1}
ensures that
for all $ \altpoint \in \R $ 
with $\abs{\altpoint-\cpoint}\leq 1$
it holds that
\begin{equation}
\begin{split}
\textstyle 
  \abs{ \p'(\altpoint) }
& 
\textstyle 
  =
  \babs{
    \sum_{ n = \mathscr{m} }^N
    \beta_n n ( \altpoint - \cpoint )^{ n - 1 }
  }
\geq 
  \abs{ \beta_{ \mathscr{m} } \mathscr{m} }
  \abs{ \altpoint - \cpoint }^{ \mathscr{m} - 1 }
  -
  \babs{
    \sum_{ n = \mathscr{m} + 1 }^N
    \beta_n n ( \altpoint - \cpoint )^{ n - 1 }
  }
\\ &
\textstyle 
\geq 
  \abs{ \altpoint - \cpoint }^{ \mathscr{m} - 1 }
  \abs{ \beta_{ \mathscr{m} } \mathscr{m} }
  -
  \bigl[
    \sum_{ n = \mathscr{m} + 1 }^N
    \abs{ \altpoint - \cpoint }^{ n - 1 }
    \abs{ \beta_n n } 
  \bigr]
\\ & 
\textstyle 
\geq 
  \abs{ \altpoint - \cpoint }^{ \mathscr{m} - 1 }
  \abs{ \beta_{ \mathscr{m} } \mathscr{m} }
  -
  \bigl[
    \sum_{ n = \mathscr{m} + 1 }^N
    \abs{ \altpoint - \cpoint }^{ \mathscr{m} }
    \abs{ \beta_n n } 
  \bigr]
\\ & 
\textstyle 
=
  \abs{ \altpoint - \cpoint }^{ \mathscr{m} - 1 }
  \abs{ \beta_{ \mathscr{m} } \mathscr{m} }
  -
  \abs{ \altpoint - \cpoint }^{ \mathscr{m} }
  \bigl[
    \sum_{ n = \mathscr{m} + 1 }^N
    \abs{ \beta_n n } 
  \bigr].
\end{split}
\end{equation}
\Hence 
for all 
$ \altpoint \in \R $
with 
$
  \abs{ \altpoint - \cpoint } 
  \leq 
  \frac{ 1 }{ 2 }
  \bigl[
      \sum_{ n = \mathscr{m} }^N
      \frac{ 
        \abs{ \beta_n n } 
      }{
        \abs{ \beta_{ \mathscr{m} } \mathscr{m} }
      }
  \bigr]^{ - 1 }
$
that 
\begin{equation}
\begin{split}
\textstyle 
  \abs{ \p'(\altpoint) }
& 
\textstyle 
\geq 
  \abs{ \altpoint - \cpoint }^{ \mathscr{m} - 1 }
  \bigl(
    \abs{ \beta_{ \mathscr{m} } \mathscr{m} }
    -
    \abs{ \altpoint - \cpoint }
    \bigl[
      \sum_{ n = \mathscr{m} + 1 }^N
      \abs{ \beta_n n } 
    \bigr]
  \bigr)
\\ & 
\textstyle
  \geq
  \abs{ \altpoint - \cpoint }^{ \mathscr{m} - 1 }
  \Bigl(
    \abs{ \beta_{ \mathscr{m} } \mathscr{m} }
    -
    \frac{ \abs{ \beta_{ \mathscr{m} } \mathscr{m} } }{ 2 }
    \Bigl(
    \abs{ \altpoint - \cpoint }
    \bigl[
      \sum_{ n = \mathscr{m} }^N
      \frac{ 
        2 \abs{ \beta_n n } 
      }{
        \abs{ \beta_{ \mathscr{m} } \mathscr{m} }
      }
    \bigr]
    \Bigr)
  \Bigr)
\\ & 
\textstyle 
\geq 
  \abs{ \altpoint - \cpoint }^{ \mathscr{m} - 1 }
  \Bigl(
    \abs{ \beta_{ \mathscr{m} } \mathscr{m} }
    -
    \frac{ \abs{ \beta_{ \mathscr{m} } \mathscr{m} } }{ 2 }
  \Bigr)
=
  \frac{ 1 }{ 2 }
  \abs{ \altpoint - \cpoint }^{ \mathscr{m} - 1 }
  \abs{ \beta_{ \mathscr{m} } \mathscr{m} }
  .
\end{split}
\end{equation}
Combining this with 
\cref{eq:polnomial_difference_estimate_in_proof_KL} 
\proves that 
for all 
$ \altpoint \in \R $
with 
$
  \abs{ \altpoint - \cpoint } 
  \leq 
  \frac{ 1 }{ 2 }
  \bigl[
      \sum_{ n = \mathscr{m} }^N
      \frac{ 
        \abs{ \beta_n n } 
      }{
        \abs{ \beta_{ \mathscr{m} } \mathscr{m} }
      }
  \bigr]^{ - 1 }
$
it holds that 
\begin{equation}
\textstyle 
  \abs{ \p(\altpoint) - \p( \cpoint ) }^{ \alpha }
\leq 
  \frac{ c }{ 2 } 
  \abs{ \altpoint - \cpoint }^{ \mathscr{m} - 1 }
  \abs{ \beta_{ \mathscr{m} } \mathscr{m} }
\leq 
  c \abs{ \p'(\altpoint) }
  .
\end{equation}
This establishes \cref{eq:KL_inequality_1d_polynomial}.
\end{aproof}
\endgroup

\begingroup
\providecommand{\p}{}
\renewcommand{\p}{p}
\begin{athm}{cor}{cor:Kurdyka_1B}[Quantitative standard \KL\ inequalities for general one-dimensional polynomials]
Let 
$ \cpoint \in \R $, $ N \in \N $, 
$ \p \in C^{ \infty }( \R , \R ) $ satisfy for all $ \altpoint \in \R $ that 
$
  \p^{ (N+1) }( \altpoint ) = 0
$, 
let 
$ \beta_0, \beta_1, \dots, \beta_N \in \R $
satisfy for all 
$ n \in \{ 0, 1, \dots, N \} $ that
$
  \beta_n = \frac{ \p^{ (n) }( \cpoint ) }{ n! }
$, 
let 
$ \rho \in \R $
satisfy 
$
  \rho = 
  \mathbbm{1}_{
    \{ 0 \}
  }\bpr{
    \sum_{ n = 1 }^N \abs{ \beta_n }
	}
  +
  \min\bigl(
    \bpr{
      \bpr{
        \bigcup_{n = 1}^N
          \{ \abs{ \beta_n n } \}
			}
      \backslash
      \{0\}
		}
    \cup
    \bcu{
      \sum_{ n = 1 }^N \abs{ \beta_n n }
    }
  \bigr) 
$, 
and let 
$ \alpha \in (0,1] $,
$ c, \varepsilon \in [0,\infty) $
satisfy 
\begin{equation}
\textstyle 
  \alpha \geq 1 - N^{ - 1 }
  ,
\quad
  c \geq 
      2 
      \rho^{ - 1 }
      [
        \sum_{ n = 1 }^N
        \abs{ \beta_n }^{ \alpha } 
      ]
  ,
\quad\text{and}\quad
  \varepsilon \leq 
    \rho
    [
      \mathbbm{1}_{
        \{ 0 \}
      }(
        \sum_{ n = 1 }^N \abs{ \beta_n }
      )
      +
      2
      (
        \sum_{ n = 1 }^N
        \abs{ \beta_n n } 
      )
    ]^{ - 1 }
  .
\end{equation}
Then it holds for all 
$ \altpoint \in [ \cpoint - \varepsilon, \cpoint + \varepsilon ] $ 
that 
\begin{equation} 
\label{eq:KL_inequality_1d_polynomial_1B}
  \abs{ \p(\altpoint) - \p( \cpoint ) }^{ \alpha }
\leq 
  c \abs{ \p'(\altpoint) }
  .
\end{equation}
\end{athm}
\begin{aproof}
Throughout this proof, 
assume without loss of generality that 
\begin{equation}
\label{eq:strictly_bigger_than_0_1B}
\textstyle 
  \sup_{ \altpoint \in \R } 
  \abs{ \p( \altpoint ) - \p( \cpoint ) } > 0.
\end{equation}
\Nobs that 
\cref{cor:reparametrization}
and 
\cref{eq:strictly_bigger_than_0_1B} ensure that 
$
  \sum_{ n = 1 }^N \abs{ \beta_n } > 0
$. 
\Hence that there exists 
$ \mathscr{m} \in \{ 1, 2, \dots, N \} $
which satisfies 
\begin{equation}
\label{eq:existence_of_script_m_in_proof}
  \abs{ \beta_{ \mathscr{m} } }
  >
  0
  =
  \sum_{ n = 1 }^{ \mathscr{m} - 1 }
  \abs{ \beta_n }
  .
\end{equation}
\Nobs that \cref{eq:existence_of_script_m_in_proof}, 
the fact that 
$
  \alpha \geq 1 - N^{ - 1 }  
$,
and \cref{cor:Kurdyka} \prove that 
for all $ \altpoint \in \R $ 
with 
$
  \abs{ \altpoint - \cpoint } 
  \leq 
  \frac{ 1 }{ 2 }
  [ 
    \sum_{ n = 1 }^N
    \frac{ 
      \abs{ \beta_n n }
    }{
      \abs{ \beta_{ \mathscr{m} } \mathscr{m} }
    }
  ]^{ - 1 }
$
it holds that 
\begin{equation}
  \abs{ \p(\altpoint) - \p(\cpoint) }^{ \alpha }
  \leq 
  \br*{
      \sum_{ n = 1 }^N 
      \frac{ 2 \abs{ \beta_n }^{ \alpha } }{
        \abs{ \beta_{ \mathscr{m} } \mathscr{m} }
      }
  }
  \abs{ \p'(\altpoint) }
\leq 
  \br*{
    \frac{ 2 }{ \rho }
    \br*{
      \sum_{ n = 1 }^N 
      \abs{ \beta_n }^{ \alpha } 
    }
  }
  \abs{ \p'(\altpoint) }
\leq
  c 
  \abs{ \p'(\altpoint) }
  .
\end{equation}
This establishes \cref{eq:KL_inequality_1d_polynomial_1B}.
\end{aproof}
\endgroup

\begingroup
\providecommand{\p}{}
\renewcommand{\p}{p}
\begin{athm}{cor}{cor:Kurdyka2_prep}
Let 
$ \cpoint \in \R$,
$N \in \N$,
$ \alpha \in [1-\frac{1}{N},\infty) \cap (0,\infty)$,
$ \p \in C^\infty(\R,\R) $
satisfy for all $ \altpoint \in \R $ that
$ \p^{ (N+1) }( \altpoint ) = 0 $.
Then there exist 
$ \varepsilon, c \in (0,\infty) $
such that for all 
$ \theta \in [ \cpoint - \varepsilon, \cpoint + \varepsilon ] $ 
it holds that 
\begin{equation}
	\label{cor:Kurdyka2_prep:eq1}
\abs{ \p(\theta) - \p( \cpoint ) }^{ \alpha }
\leq 
	c \abs{ \p'(\theta) }.
\end{equation}
\end{athm}
\begin{aproof}
\Nobs that 
\enum{
	\cref{cor:KL_increasing_alpha};
	\cref{cor:Kurdyka_1B}
}
\prove \cref{cor:Kurdyka2_prep:eq1}.
\end{aproof}
\endgroup

\begingroup
\providecommand{\p}{}
\renewcommand{\p}{p}
\begin{athm}{cor}{cor:Kurdyka2}[Qualitative standard \KL\ inequalities for general one-dimensional polynomials]
Let 
$ \cpoint \in \R $, $ N \in \N $, 
$ \p \in C^{ \infty }( \R , \R ) $ satisfy for all $ \altpoint \in \R $ that 
$
  \p^{ (N) }( \altpoint ) = 0
$. 
Then there exist 
$ \varepsilon, c \in (0,\infty) $, 
$ \alpha \in (0,1) $
such that for all 
$ \altpoint \in [ \cpoint - \varepsilon, \cpoint + \varepsilon ] $ 
it holds that 
\begin{equation} 
\label{eq:KL_inequality_1d_polynomial2}
  \abs{ \p(\altpoint) - \p( \cpoint ) }^{ \alpha }
\leq 
  c \abs{ \p'(\altpoint) }
  .
\end{equation}  
\end{athm}
\begin{aproof}
\Nobs that \cref{cor:Kurdyka_1B} \proves[ep] \cref{eq:KL_inequality_1d_polynomial2}.
\end{aproof}
\endgroup

\cfclear
\begingroup
\providecommand{\d}{}
\renewcommand{\d}{\defaultParamDim}
\providecommand{\f}{}
\renewcommand{\f}{\defaultLossFunction}
\begin{athm}{cor}{cor:Kurdyka3}
Let 
$ \f \colon \R \to \R $ be a polynomial.
Then $ \f $ 
is a standard \KL\ function\cfadd{def:KLfunction_standard}
\cfout.
\end{athm}
\begin{aproof}
\Nobs that 
\cref{def:KLinequality_standard:eq1}
and 
\cref{cor:Kurdyka2} 
establish 
that $ \f $ 
\cfadd{def:KLfunction_standard}
is a 
standard \KL\ function
\cfload.
\end{aproof}
\endgroup

\section{Power series and analytic functions}

\cfclear
\begingroup
\providecommand{\f}{}
\renewcommand{\f}{f}
\begin{adef}{def:analytic_function}[Analytic functions]
Let $ m, n \in \N $, 
let 
$ U \subseteq \R^m $ 
be open, 
and let $ \f \colon U \to \R^n $ be a function. 
Then we say that $ \f $ is analytic 
if and only if 
\begin{enumerate}[label=(\roman*)]
\item it holds that  
$
  \f \in C^{ \infty }( U, \R^n ) 
$
and 
\item
for all $ \cpoint \in U $ 
there exists  
$ \varepsilon \in (0,\infty) $
such that 
for all $ \altpoint \in \{ \altpointTwo \in U \colon \Pnorm2{ \cpoint - \altpointTwo } < \varepsilon \} $
it holds that 
\begin{eqsplit}
\label{def:analytic_function:eq1}
  \limsup_{ K \to \infty }
  \Pnorm[\Big]2{
    \f( \altpoint ) 
    - 
    \smallsum_{ k = 0 }^K
    \tfrac{ 1 }{ k! } 
    f^{ (k) }( \cpoint )( \altpoint - \cpoint, \altpoint - \cpoint, \dots, \altpoint - \cpoint ) 
  }
  = 0
\end{eqsplit}
\end{enumerate}
\cfload.
\end{adef}
\endgroup

\begingroup
\providecommand{\f}{}
\renewcommand{\f}{f}
\begin{athm}{lemma}{lemma:multivariate_derivatives}[Higher derivatives of multidimensional functions]
Let $k, n, m \in \N$, let $U \subseteq \R^m$ be open, 
and
let $f \in C^k(U,\R^n)$, $\altpoint = (\altpoint_1, \dots, \altpoint_m) \in U$, $v^1 = (v^1_1, \dots, v^1_m)$, $v^2 = (v^2_1, \dots, v^2_m)$, $\dots$, $v^k = (v^k_1, \dots, v^k_m) \in \R^m$.
Then 
\begin{equation}
	f^{ (k) }( \altpoint )( v^1, v^2, \dots, v^k )
=
	\sum_{i_1, i_2, \dots, i_k \in \{ 1, \dots, m \} }
	v^{ 1 }_{ i_1 } v^{ 2 }_{ i_2 } \cdots v^{ k }_{ i_k }
	\pr[\big]{
		\tfrac{ \partial^k \f }{ \partial \altpoint_{ i_1 } \partial \altpoint_{ i_2 } \cdots \partial \altpoint_{ i_k } }
	}( \altpoint )
	.
\end{equation}

\end{athm}
\begin{aproof}
Throughout this proof, let $e_1, e_2, \dots, e_m \in \R^m$ satisfy
\begin{equation}
	e_1 = (1, 0, \ldots, 0), \quad
	e_2 = (0, 1, 0, \ldots, 0), \quad
	\dots, \quad
	e_m = (0, \ldots, 0, 1).
\end{equation}
\Nobs that
\enum{
	the fact that $f^{ (k) }( \altpoint )$ is $k$-linear;
	the fact that for all $i_1, i_2, \dots, i_k \in \{ 1, \dots, m \}$ it holds that
	$f^{ (k) }( \altpoint )(e_{i_1}, e_{i_2}, \dots, e_{i_k}) = \pr*{ \tfrac{ \partial^k \f }{ \partial \altpoint_{ i_1 } \partial \altpoint_{ i_2 } \cdots \partial \altpoint_{ i_k } } }( \altpoint )$
}
\prove 
\begin{equation}
\begin{split}
	f^{ (k) }( \altpoint )( v^1, v^2, \dots, v^k )
&=
	f^{ (k) }( \altpoint )
	\pr*{
		\sum_{i_1 = 1 }^m v^{ 1 }_{ i_1 } e_{ i_1 },
		\sum_{i_2 = 1 }^m v^{ 2 }_{ i_2 } e_{ i_2 },
		\dots,
		\sum_{i_k = 1 }^m v^{ k }_{ i_k } e_{ i_k }
	} \\
&=
	\sum_{i_1, i_2, \dots, i_k \in \{ 1, \dots, m \} }
	f^{ (k) }( \altpoint )(v^{ 1 }_{ i_1 } e_{ i_1 }, v^{ 2 }_{ i_2 } e_{ i_2 }, \dots, v^{ k }_{ i_k } e_{ i_k })\\
&=
	\sum_{i_1, i_2, \dots, i_k \in \{ 1, \dots, m \} }
	v^{ 1 }_{ i_1 } v^{ 2 }_{ i_2 } \cdots v^{ k }_{ i_k }
	f^{ (k) }( \altpoint )( e_{ i_1 }, e_{ i_2 }, \dots, e_{ i_k } )\\
&=
	\sum_{i_1, i_2, \dots, i_k \in \{ 1, \dots, m \} }
	v^{ 1 }_{ i_1 } v^{ 2 }_{ i_2 } \cdots v^{ k }_{ i_k }
	\pr[\big]{ \tfrac{ \partial^k \f }{ \partial \altpoint_{ i_1 } \partial \altpoint_{ i_2 } \cdots \partial \altpoint_{ i_k } } }( \altpoint )
	.
\end{split}
\end{equation}
\end{aproof}
\endgroup

\begingroup
\providecommand{\f}{}
\renewcommand{\f}{f}
\begin{athm}{lemma}{lemma:analytic_functions_onedim}[One-dimensional analytic functions]
Let $ U \subseteq \R $ be open, 
let $ \f \colon U \to \R $ be a function.
Then the following two statements are equivalent:
\begin{enumerate}[label=(\roman*)]
	\item \label{lemma:analytic_functions_onedim:item1}
	It holds that $ \f $ is\cfadd{def:analytic_function} analytic \cfload.
	\item \label{lemma:analytic_functions_onedim:item2}
	For every $ \cpoint \in U $ there exists $ \varepsilon \in (0,\infty) $
	such that for all $ \altpoint \in U \cap (\cpoint-\varepsilon,\cpoint+\varepsilon) $ it holds that
	$\f \in C^\infty(U,\R)$ and
	\begin{equation}
		\limsup_{ K \to \infty }
		\abs*{ \f( \altpoint ) - 
		\smallsum_{ k = 0 }^K
		\tfrac{ \f^{ (k) }( \cpoint )( \altpoint - \cpoint) ^k }{ k! } 
		}
		= 0
		.
	\end{equation}
\end{enumerate}
\end{athm}
\begin{aproof}
\Nobs that
\enum{
	\cref{def:analytic_function:eq1};
	\cref{lemma:multivariate_derivatives}
}
establish that (\ref{lemma:analytic_functions_onedim:item1} $\leftrightarrow$ \ref{lemma:analytic_functions_onedim:item2}).
\end{aproof}

\cfclear
\begingroup
\providecommand{\f}{}
\renewcommand{\f}{f}
\begin{athm}{prop}{prop:power_series}[Power series]
Let $ m, n \in \N $, 
$ \varepsilon \in (0,\infty) $, 
let 
$ U \subseteq \R^m $
satisfy 
$ 
  U = \{ \altpointThree \in \R^m \colon \Pnorm2{ \altpointThree } \leq \varepsilon \}
$, 
for every $ k \in \N $ let 
$ A_k \colon ( \R^m )^k \to \R^n $
be $ k $-linear and symmetric, 
and let 
$ \f \colon U \to \R^n $ satisfy
for all $ \altpoint \in U $ that
\begin{eqsplit}
\label{eq:convergence_assumption}
  \limsup_{ K \to \infty }
  \Pnorm[\Big]2{
    \f( \altpoint ) 
    - 
    \f( 0 )
    -
    \smallsum_{ k = 1 }^K
    A_k( \altpoint, \altpoint, \dots, \altpoint ) 
  }
  = 0
\end{eqsplit}
\cfload.
Then 
\begin{enumerate}[label=(\roman*)]
\item \llabel{it:1}
it holds for all 
$ \altpoint \in \{ \altpointThree \in U \colon \Pnorm2{ \altpointThree } < \varepsilon \} $ 
that 
$
  \sum_{ k = 1 }^{ \infty }
  \Pnorm2{ 
    A_k( \altpoint, \altpoint, \dots, \altpoint )
  }
  < \infty 
$
and 
\begin{eqsplit}
  \f(\altpoint) = \f(0) + \smallsum_{ k = 1 }^{ \infty } A_k( \altpoint, \altpoint, \dots, \altpoint ), 
\end{eqsplit}
\item \llabel{it:2}
it holds that 
$
  \f|_{
    \{ \altpointThree \in U \colon \Pnorm2{ \altpointThree } < \varepsilon \}
  } 
$
is infinitely often differentiable, 
\item \llabel{it:3}
it holds for all 
$
  \altpoint \in 
  \{ \altpointThree \in U \colon \Pnorm2{ \altpointThree } < \varepsilon \}
$, 
$ l \in \N $, 
$ \altpointTwo_1, \altpointTwo_2, \dots, \altpointTwo_l \in \R^m $
that 
\begin{equation}
\textstyle 
  \sum\limits_{ k = l }^{ \infty }
  \bigl(
    \bigl[
      \frac{ k! }{ ( k - l)! }
    \bigr]
    \allowbreak 
    \Pnorm2{
      A_k( 
        \altpointTwo_1, \altpointTwo_2, \dots, \altpointTwo_l, 
        \altpoint, \altpoint, \dots, \altpoint 
      )
    }
  \bigr)
  < \infty 
\end{equation}
and 
\begin{equation}
	\llabel{eq:deriv}
\textstyle 
  \f^{ (l) }( \altpoint )( \altpointTwo_1, \dots, \altpointTwo_l )
  =
  \sum\limits_{ k = l }^{ \infty }
  \bigl(
  \bigl[ 
    \frac{ k! }{ ( k - l )! }
  \bigr]
  A_k( 
    \altpointTwo_1, \altpointTwo_2, \dots, \altpointTwo_l, \altpoint, \altpoint, \dots, \altpoint 
  )
  \bigr)
  ,
\end{equation}
and 
\item \llabel{it:4}
it holds for all $ k \in \N $ that 
$
  \f^{ (k) }( 0 ) 
  =
  k!
  A_k
$. 
\end{enumerate}
\end{athm}

\begin{aproof}
Throughout this proof, for every $ K \in \N_0 $ 
let $ F_K \colon \R^m \to \R^n $ 
satisfy for all $ \altpoint \in \R^m $ that 
\begin{eqsplit}
\llabel{eq:F_K}
  F_K(\altpoint) = \f(0) + \sum_{ k = 1 }^K A_k( \altpoint, \altpoint, \dots, \altpoint )
  .
\end{eqsplit}
\Nobs that 
\enum{
	\cref{eq:convergence_assumption};
	\lref{eq:F_K}
} ensure that
for all $ \altpoint \in U $ it holds that 
\begin{equation}
\textstyle 
  \limsup_{ K \to \infty }
  \Pnorm2{
    \f(\altpoint) - F_K(\altpoint)
  }  
  = 0 
  .
\end{equation}
\Hence for all $ \altpoint \in U $ that
\begin{equation}
\textstyle 
  \limsup_{ K \to \infty }
  \Pnorm2{
    F_{ K + 1 }( \altpoint ) - F_K(\altpoint)
  }  
  = 0 
  .
\end{equation}
This \proves for all $ \altpoint \in U $ that 
\begin{equation}
\textstyle 
  \sup_{ k \in \N }
  \Pnorm2{
    A_k( \altpoint, \altpoint, \dots, \altpoint )
  }
  =
  \sup_{ K \in \N_0 }
  \Pnorm2{
    F_{ K + 1 }( \altpoint ) - F_K(\altpoint)
  }  
  < \infty 
  .
\end{equation}
\Hence 
for all
$
  \altpoint \in \{ \altpointThree \in U \colon \Pnorm2{ \altpointThree } < \varepsilon \} \backslash \{ 0 \} 
$
that 
\begin{eqsplit}
  \sum_{ k = 1 }^{ \infty }
  \Pnorm2{  
    A_k( \altpoint, \altpoint, \dots, \altpoint )
  }
&
=
  \sum_{ k = 1 }^{ \infty }
  \pr*{
  \br*{
    \frac{
      \Pnorm2{ \altpoint }
    }{
      \varepsilon
    }
  }^k
  \Pnorm[\big]2{  
    A_k\bigl( 
      \tfrac{ \varepsilon \altpoint }{ \Pnorm2{ \altpoint } }, 
      \tfrac{ \varepsilon \altpoint }{ \Pnorm2{ \altpoint } }, 
      \dots, 
      \tfrac{ \varepsilon \altpoint }{ \Pnorm2{ \altpoint } } 
    \bigr)
  }
  }
\\ &
\leq 
  \br*{
    \sum_{ k = 1 }^{ \infty }
    \br*{
      \frac{
        \Pnorm2{ \altpoint }
      }{
        \varepsilon
      }
    }^k
  }
  \br*{
    \sup_{ k \in \N }
    \Pnorm[\big]2{  
      A_k\bigl( 
        \tfrac{ \varepsilon \altpoint }{ \Pnorm2{ \altpoint } }, 
        \tfrac{ \varepsilon \altpoint }{ \Pnorm2{ \altpoint } }, 
        \dots, 
        \tfrac{ \varepsilon \altpoint }{ \Pnorm2{ \altpoint } } 
      \bigr)
    }
  }
  < 
  \infty 
  .
\end{eqsplit}
	This
\proves that for all
	$\altpoint\in \{\altpointThree\in U\colon \Pnorm2 \altpointThree<\eps\}$
it holds that
\begin{equation}
	\sum_{ k = 1 }^{ \infty }
  \Pnorm2{  
    A_k( \altpoint, \altpoint, \dots, \altpoint )
  }
	<
	\infty
	.
\end{equation}
Combining
	this
with
	\cref{eq:convergence_assumption}
establishes
	\lref{it:1}.
\Nobs that, \eg,
	Krantz \& Parks~\cite[Proposition~2.2.3]{krantz2002primer}
\proves
	\cref{prop:power_series.it:2,prop:power_series.it:3}.
\Nobs that 
	\lref{eq:deriv}
implies
	\lref{it:4}.
\end{aproof}
\endgroup

\cfclear
\begingroup
\providecommand{\f}{}
\renewcommand{\f}{f}
\begin{athm}{prop}{prop:charac_analytic}[Characterization for analytic functions]
Let $ m, n \in \N $, 
let 
$ U \subseteq \R^m $ 
be open, 
and let $ \f \in C^{ \infty }( U, \R^n ) $. 
Then the following three statements are equivalent: 
\begin{enumerate}[label=(\roman*)]
\item \cfadd{def:analytic_function}
It holds that $ \f $ is analytic \cfload.
\item 
It holds for all $ \cpoint \in U $ that there exists 
$ \varepsilon \in (0,\infty) $ such that 
for all $ \altpoint \in \{ \altpointThree \in U \colon \Pnorm2{ \cpoint - \altpointThree } < \varepsilon \} $
it holds that 
$
  \sum_{ k = 0 }^{ \infty }
  \frac{ 
    1
  }{ k! } 
  \Pnorm2{
    \f^{ (k) }( \cpoint )( \altpoint - \cpoint, \altpoint - \cpoint, \dots, \altpoint - \cpoint ) 
  }
  < \infty 
$
and 
\begin{eqsplit}
    \f( \altpoint ) 
  =
    \smallsum_{ k = 0 }^{ \infty }
    \tfrac{ 1 }{ k! } 
    \f^{ (k) }( \cpoint )( \altpoint - \cpoint, \altpoint - \cpoint, \dots, \altpoint - \cpoint ) 
  .
\end{eqsplit}
\item 
It holds 
for all compact $ \fC \subseteq U $
that there exists $ c \in \R $ 
such that for all 
$ \altpoint \in \fC $, 
$ k \in \N $, 
$ \altpointTwo \in \R^m $
it holds that 
\begin{eqsplit}
    \Pnorm2{ \f^{ (k) }( \altpoint )( \altpointTwo, \altpointTwo, \dots, \altpointTwo ) }
  \leq 
  k! \, c^k \, \Pnorm2{ \altpointTwo }^k
  .
\end{eqsplit}
\end{enumerate}
\end{athm}
\begin{aproof}
The equivalence is a direct consequence from \cref{prop:power_series}. 
\end{aproof}
\endgroup

\section{Standard KL inequalities for one-dimensional analytic functions}

In \cref{sect:KL_onedim_poly} above we have seen that one-dimensional polynomials are standard \KL\ functions (see \cref{cor:Kurdyka3}).
In this section we verify that one-dimensional analytic functions are also standard \KL\ functions (see \cref{cor:kl-analytic1d_2} below).
The main arguments for this statement are presented in the proof of \cref{lem:kl-analytic} and are inspired by \cite{EoMLoja}.

\cfclear
\providecommand{\f}{}
\renewcommand{\f}{f}
\providecommand{\x}{}
\renewcommand{\x}{\altpoint}
\providecommand{\y}{}
\renewcommand{\y}{y}
\begingroup
\begin{athm}{lemma}{lem:powerseries1d}
	Let
		$\eps\in(0,\infty)$,
	let 
		$(a_k)_{k\in\N}\subseteq \R$,
	and let
		$\f\colon [-\eps,\eps]\to\R$
	satisfy for all
		$\x\in [-\eps,\eps]$
	that
	\begin{equation}
		\limsup_{K\to\infty}\abs*{\f(\x)-\f(0)-\ssuml_{k=1}^K a_k\x^k}
		=
		0
		.
	\end{equation}
	Then
	\begin{enumerate}[(i)]
		\item \llabel{it:1}
		it holds for all
			$\x\in(-\eps,\eps)$
		that
			$\sum_{k=1}^\infty \abs{a_k}\abs{\x}^k<\infty$
		and
		\begin{equation}
			\f(\x)
			=
			\f(0)+\ssuml_{k=1}^\infty a_k\x^k
			,
		\end{equation}
		\item \llabel{it:2}
		it holds that
			$\f|_{(-\eps,\eps)}$
		is infinitely often differentiable,
		\item \llabel{it:3}
		it holds for all
			$\x\in(-\eps,\eps)$,
			$l\in\N$
		that
		$
			\sum_{k=l}^\infty \bbr{\tfrac{k!}{(k-l)!}}\abs{a_k}\abs{\x}^{k-l}<\infty
		$
		and
		\begin{equation}
			\f^{(l)}(\x)
			=
			\ssuml_{k=l}^\infty \bbr{\tfrac{k!}{(k-l)!}}a_k\x^{k-l},
		\end{equation}
		and
		\item \llabel{it:4}
		it holds for all
			$k\in\N$
		that
			$\f^{(k)}(0)=k!a_k$.
	\end{enumerate}
\end{athm}
\begin{aproof}
	\Nobs that
		\cref{prop:power_series}
		(applied with
		  $m\is 1$,
			$n\is 1$,
			$\eps\is\eps$,
			$U\is [-\eps,\eps]$,
			$(A_k)_{k\in\N}\is \bpr{(\R^k\ni (\altpointTwo_1,\altpointTwo_2,\dots,\altpointTwo_k)\mapsto a_k\altpointTwo_1\altpointTwo_2\cdots \altpointTwo_k\in\R)}{}_{k\in\N}$,
			$\f\is \f$
		in the notation of \cref{prop:power_series})
	establishes
		\cref{lem:powerseries1d.it:1,lem:powerseries1d.it:2,lem:powerseries1d.it:3,lem:powerseries1d.it:4}.
\end{aproof}
\endgroup

\cfclear
\begingroup
\providecommand{\f}{}
\renewcommand{\f}{\defaultLossFunction}
\begin{athm}{lemma}{lem:kl-analytic}
	Let 
		$\eps,\delta\in(0,1)$,
		$N\in\N\backslash \{1\}$,
	let
		$(a_k)_{k\in\N}\subseteq\R$
	satisfy
		$N=\min(\{k\in\N\colon a_k\neq0\}\cup\{\infty\})$,
	let
		$\f\colon [-\eps,\eps]\to\R$
	satisfy for all
		$\altpoint\in [-\eps,\eps]$
	that
	\begin{equation}
		\llabel{eq:limsup}
		\limsup_{K\to\infty}\abs*{\f(\altpoint)-\f(0)-\br*{\smallsum_{k=1}^K a_k \altpoint^k}}
		=
		0
		,
	\end{equation}
	and let
		$M\in\N\cap(N,\infty)$
	satisfy for all
		$k\in\N\cap[M,\infty)$
	that
		$k\abs{a_k}\leq (2\eps^{-1})^k$
		and
		\begin{equation}
		\llabel{eq:delta}
			\delta
			=
			\min\bcu{\tfrac\eps4,\abs{a_N}
				\bbr{
					(2\eps^{-1})^{N+1}
					+
					\pr[\big]{\max\nolimits_{k\in\{1, 2, \dots, M\}} \abs{2 k a_k}}
				}^{-1}
			}
			.
		\end{equation}
	Then it holds for all
		$\altpoint\in(-\delta,\delta)$
	that
	\begin{equation}
		\abs{\f(\altpoint)-\f(0)}^{\tfrac{N-1}N}
		\leq
		2
			\abs{a_N}^{-\frac{1}{N}}
		\abs{\f'(\altpoint)}
		.
	\end{equation}
\end{athm}
\begin{aproof}
	\Nobs that
		the fact that
			for all
				$k\in\N\cap[M,\infty)$
			it holds that
				$\abs{a_k}\leq k\abs{a_k}\leq (2\eps^{-1})^k$
	ensures that for all
		$\altpoint \in \R$
	it holds that
	\begin{eqsplit}
		&\ssuml_{k=N+1}^{\infty} \abs{a_k} \abs{\altpoint}^k
		\\&=
		\abs{\altpoint}^{N+1}\br*{\ssuml_{k=N+1}^{\infty} \abs{a_k} \abs{\altpoint}^{k-N-1}}
		\\&=
		\abs{\altpoint}^{N+1}\br*{
			\pr*{\ssuml_{k=N+1}^M \abs{a_k} \abs{\altpoint}^{k-N-1}}
			+
			\pr*{\ssuml_{k=M+1}^{\infty} \abs{a_k} \abs{\altpoint}^{k-N-1}}
		}
		\\&\leq
		\abs{\altpoint}^{N+1}\br*{
			\pr*{\max_{k\in\{1, 2, \dots, M\}} \abs{a_k}} \pr*{\ssuml_{k=N+1}^M \abs{\altpoint}^{k-N-1}}
			+
			\pr*{\ssuml_{k=M+1}^{\infty} (2\eps^{-1})^{k} \abs{\altpoint}^{k-N-1}}
		}
		\\&=
		\abs{\altpoint}^{N+1}\br*{
			\pr*{\max_{k\in\{1, 2, \dots, M\}} \abs{a_k}} \pr*{\ssuml_{k=0}^{M-N-1} \abs{\altpoint}^{k}}
			+
			(2\eps^{-1})^{N+1}\pr*{\ssuml_{k=M-N}^{\infty} (2\eps^{-1}\abs{\altpoint})^{k}}
		}
		\\&\leq
		\abs{\altpoint}^{N+1}\br*{
			\pr*{\max_{k\in\{1, 2, \dots, M\}} \abs{a_k}} \pr*{\ssuml_{k=0}^{\infty} \abs{\altpoint}^{k}}
			+
			(2\eps^{-1})^{N+1}\pr*{\ssuml_{k=M-N}^{\infty} (2\eps^{-1}\abs{\altpoint})^{k}}
		}
		.
	\end{eqsplit}
	\Hence for all
		$\altpoint\in (-\frac\eps4, \frac\eps4)$
	that
	\begin{eqsplit}
		\ssuml_{k=N+1}^\infty \abs{a_k} \abs{\altpoint}^k
		&\leq
		\abs{\altpoint}^{N+1}\br*{
			\pr*{\max_{k\in\{1, 2, \dots, M\}} \abs{a_k}} \pr*{\ssuml_{k=0}^\infty \babs{\tfrac14}^{k}}
			+
			(2\eps^{-1})^{N+1}\pr*{\ssuml_{k=1}^\infty \babs{\tfrac12}^{k}}
		}
		\\&\leq
		\abs{\altpoint}^{N+1}\br*{
			2\pr*{\max_{k\in\{1, 2, \dots, M\}} \abs{a_k}}
			+
			(2\eps^{-1})^{N+1}
		}
		.
	\end{eqsplit}
		This
		and 
		\lref{eq:delta}
	\prove for all
		$\altpoint\in(-\delta,\delta)$
	that
	\begin{equation}
		\llabel{eq:1}
		\ssuml_{k=N+1}^\infty \abs{a_k} \abs{\altpoint}^k
		\leq
		\abs{a_N}\abs{\altpoint}^N
		.
	\end{equation}
	Combining this and \lref{eq:limsup}
	\proves for all
		$\altpoint\in(-\delta,\delta)$
	that
	\begin{equation}
		\llabel{eq:2}
		\abs{\f(\altpoint)-\f(0)}
		=
		\abs*{\ssuml_{k=N}^\infty a_k \altpoint^k}
		\leq
		\abs{a_N}\abs{\altpoint}^N+ \br*{\ssuml_{k=N+1}^\infty \abs{a_k} \abs{\altpoint}^k}
		\leq
		2\abs{a_N}\abs{\altpoint}^N
		.
	\end{equation}
	Next \nobs that
		the assumption that 
			for all
				$k\in\N\cap[M,\infty)$
			it holds that
				$k\abs{a_k}\leq(2\eps^{-1})^k$
	\proves that for all
		$\altpoint\in \R$ 
	it holds that
	\begin{eqsplit}
		&\ssuml_{k=N+1}^{\infty} k\abs{a_k}\abs{\altpoint}^{k-1}
		\\&=
		\abs{\altpoint}^{N}
		\br*{
			\pr*{\ssuml_{k=N+1}^{\infty} k \abs{a_{k}}\abs{\altpoint}^{k-N-1}}
		}
		\\&\leq
		\abs{\altpoint}^{N}
		\br*{
			\pr*{\ssuml_{k=N+1}^{M} k \abs{a_{k}}\abs{\altpoint}^{k-N-1}}
			+
			\pr*{\ssuml_{k=M+1}^{\infty} (2\eps^{-1})^{k} \abs{\altpoint}^{k-N-1}}
		}
		\\&\leq
		\abs{\altpoint}^{N}
		\br*{
			\pr*{\max_{k\in\{1, 2, \dots, M\}} k\abs{a_k}} \pr*{\ssuml_{k=N+1}^{M} \abs{\altpoint}^{k-N-1}}
			+
			\pr*{\ssuml_{k=M+1}^{\infty} (2\eps^{-1})^{k} \abs{\altpoint}^{k-N-1}}
		}
		\\&=
		\abs{\altpoint}^{N}
		\br*{
			\pr*{\max_{k\in\{1, 2, \dots, M\}} k\abs{a_k}} \pr*{\ssuml_{k=0}^{M-N-1} \abs{\altpoint}^{k}}
			+
			\pr*{\ssuml_{k=M-N}^{\infty} (2\eps^{-1})^{k+N+1} \abs{\altpoint}^{k}}
		}
		\\&\leq
		\abs{\altpoint}^{N}
		\br*{
			\pr*{\max_{k\in\{1, 2, \dots, M\}} k\abs{a_k}} \pr*{\ssuml_{k=0}^{\infty} \abs{\altpoint}^{k}}
			+
			(2\eps^{-1})^{N+1}\pr*{\ssuml_{k=M-N}^{\infty} (2\eps^{-1}\abs{\altpoint})^{k}}
		}
		.
	\end{eqsplit}
	\Hence for all
		$\altpoint\in(-\frac\eps4, \frac\eps4)$
	that
	\begin{eqsplit}
		&\ssuml_{k=N+1}^\infty k\abs{a_k}\abs{\altpoint}^{k-1}
		\\&\leq
		\abs{\altpoint}^{N}
		\br*{
			\pr*{\max_{k\in\{1, 2, \dots, M\}} k\abs{a_k}} \pr*{\ssuml_{k=0}^\infty \babs{\tfrac14}^{k}}
			+
			(2\eps^{-1})^{N+1}\pr*{\ssuml_{k=1}^\infty \babs{\tfrac12}^{k}}
		}
		\\&\leq
		\abs{\altpoint}^{N}
		\br*{
			2\pr*{\max_{k\in\{1, 2, \dots, M\}} k\abs{a_k}}
			+
			(2\eps^{-1})^{N+1}
		}
		.
	\end{eqsplit}
	Combining 
		\enum{
			this;
			\lref{eq:delta}
		}
	\proves for all
		$\altpoint\in(-\delta,\delta)$
	that
	\begin{equation}
		\ssuml_{k=N+1}^\infty k\abs{a_k}\abs{\altpoint}^{k-1}
		\leq
		\abs{a_N}\abs{\altpoint}^{N-1}
		.
	\end{equation}
	\Hence for all
		$K\in\N\cap[N,\infty)$,
		$\altpoint\in(-\delta,\delta)$
	that
	\begin{equation}
		\abs*{\ssuml_{k=1}^K ka_k\altpoint^{k-1}}
		=
		\abs*{\ssuml_{k=N}^K ka_k\altpoint^{k-1}}
		\geq
		N\abs{a_N}\abs{\altpoint}^{N-1}-\ssuml_{k=N+1}^\infty k\abs{a_k}\abs{\altpoint}^{k-1}
		\geq
		(N-1)\abs{a_N}\abs{\altpoint}^{N-1}
		.
	\end{equation}
	\enum{
		\Cref{lem:powerseries1d};
		\lref{eq:limsup}
	}
		\hence
	\prove that for all
		$\altpoint\in(-\delta,\delta)$
	it holds that
	$\sum_{k=1}^\infty k\abs{a_k\altpoint^{k-1}}<\infty$
	and
	\begin{equation}
		\abs{\f'(\altpoint)}
		=
		\abs*{\ssuml_{k=1}^\infty ka_k\altpoint^{k-1}}
		\geq
		(N-1)\abs{a_N}\abs{\altpoint}^{N-1}
		.
	\end{equation}
	Combining
		this
	with
		\lref{eq:2}
	\proves that for all
		$\altpoint\in(-\delta,\delta)$
	it holds that
	\begin{equation}
	\begin{split}
		\abs{\f(\altpoint)-\f(0)}^{\frac{N-1}N}
		&\leq
		\br*{2\abs{a_N}\abs{\altpoint}^N}^{\frac{N-1}N}\\
		&\leq
		\abs{2a_N}^{\frac{N-1}N}\abs{\altpoint}^{N-1}\\
		&\leq
		\abs{2a_N}^{\frac{N-1}N}(N-1)^{-1}\abs{a_N}^{-1}\abs{\f'(\altpoint)} \\
		&\leq 
		2 \abs{a_N}^{-\frac1N}(N-1)^{-1}\abs{\f'(\altpoint)}
		\\&\leq
		2\abs{a_N}^{-\frac1N}\abs{\f'(\altpoint)}
		.
	\end{split}
	\end{equation}
\end{aproof}
\endgroup

\cfclear
\providecommand{\f}{}
\renewcommand{\f}{\defaultLossFunction}
\begingroup
\begin{athm}{cor}{cor:kl-powseries1d}
	Let
		$\eps\in(0,\infty)$,
	let $(a_k)_{k \in \N} \subseteq \R$,
	and let
		$\f\colon [-\eps,\eps]\to\R$
	satisfy for all
		$\altpoint\in[-\eps,\eps]$
	that
	\begin{equation}
		\limsup_{K\to\infty}\abs*{\f(\altpoint)-\f(0)-\ssuml_{k=1}^K a_k\altpoint^k}
		=
		0
		.
	\end{equation}
	Then there exist
		$\delta\in(0,\eps)$,
		$c\in(0,\infty)$,
		$\alpha\in(0,1)$
	such that for all
		$\altpoint\in(-\delta,\delta)$
	it holds that
	\begin{equation}
		\abs{\f(\altpoint)-\f(0)}^\alpha
		\leq 
		c\,\abs{\f'(\altpoint)}
		.
	\end{equation}
	
\end{athm}
\begin{aproof}
	Throughout this proof,
	  assume without loss of generality that
			$\eps<1$,
		let
			$N\in\N\cup\{\infty\}$
		satisfy
			$N=\min(\{k\in\N\colon a_k\neq0\}\cup\{\infty\})$,
		and assume without loss of generality that
			$1<N<\infty$
	(cf.\ \cref{lem:powerseries1d.it:3,lem:powerseries1d.it:4} in \cref{lem:powerseries1d} and \cref{cor:non_critical2}).
	\Nobs that
		\cref{lem:powerseries1d.it:3} in \cref{lem:powerseries1d}
	ensures that for all
		$\altpoint\in(-\eps,\eps)$
	it holds that
	\begin{equation}
		\ssuml_{k=1}^\infty k\abs{a_k}\abs{\altpoint}^{k-1}<\infty
		.
	\end{equation}
	\Hence that
	\begin{equation}
		\ssuml_{k=1}^\infty k\abs{a_k}\babs{\tfrac{\eps}{2}}^k<\infty
		.
	\end{equation}
	\Hence that 
	\begin{equation}
		\limsup_{k\to\infty} \pr*{
			k \abs{a_k} \babs{\tfrac{\eps}{2}}^k
		}
		=
		0
		.
	\end{equation}
		This
	\proves that there exists
		$M\in\N\cap(N,\infty)$
	which satisfies for all
		$k\in\N\cap[M,\infty)$
	that
	\begin{equation}
		k\abs{a_k}
		\leq 
		(2\eps^{-1})^{k}
		.
	\end{equation}
	\Cref{lem:kl-analytic}
	\hence[hence]
	shows that for all
		$\altpoint\in\{\altpointThree\in\R\colon\abs{\altpointThree}<\min\{\tfrac\eps4,\abs{a_N}
		\br{
			\pr{\max_{k\in\{1, 2, \dots, M\}} \abs{2 k a_k}}
			+
			(2\eps^{-1})^{N+1}
		}^{-1}
		\}\}$
	it holds that
	\begin{equation}
		\abs{\f(\altpoint)-\f(0)}^{\frac{N-1}N}
		\leq
		2
		\abs{a_N}^{-\frac1N}
		\abs{\f'(\altpoint)}
		.
	\end{equation}
\end{aproof}
\endgroup

\cfclear
\providecommand{\f}{}
\renewcommand{\f}{\defaultLossFunction}
\providecommand{\g}{}
\renewcommand{\g}{\mathcal{M}}
\begingroup
\begin{athm}{cor}{cor:kl-powseries1d.2}
	Let
		$\eps\in(0,\infty)$,
		$\cpoint\in\R$,
	let
		$(a_k)_{k \in \N} \subseteq \R$,
	and let
		$\f\colon [\cpoint-\eps,\cpoint+\eps]\to\R$
	satisfy for all
		$\altpoint\in[\cpoint-\eps,\cpoint+\eps]$
	that
	\begin{equation}
		\llabel{eq:1}
		\limsup_{K\to\infty}\abs*{\f(\altpoint)-\f(\cpoint)-\ssuml_{k=1}^K a_k(\altpoint-\cpoint)^k}
		=
		0
		.
	\end{equation}
	Then there exist
		$\delta\in(0,\eps)$,
		$c\in(0,\infty)$,
		$\alpha\in(0,1)$
	such that for all
		$\altpoint\in(\cpoint-\delta,\cpoint+\delta)$
	it holds that
	\begin{equation}
		\llabel{claim}
		\abs{\f(\altpoint)-\f(\cpoint)}^\alpha
		\leq 
		c\,\abs{\f'(\altpoint)}
		.
	\end{equation}
	
\end{athm}
\begin{aproof}
	Throughout this proof,
	  let $\g\colon [-\eps,\eps]\to\R$ satisfy 
			for all
				$\altpoint\in[-\eps,\eps]$
			that
				$\g(\altpoint)=\f(\altpoint+\cpoint)$.
	\Nobs that
		\lref{eq:1}
		and the fact that 
			for all
				$\altpoint\in[-\eps,\eps]$
			it holds that
				$\altpoint+\cpoint\in[\cpoint-\eps,\cpoint+\eps]$
	\prove that for all
		$\altpoint\in[-\eps,\eps]$
	it holds that
	\begin{eqsplit}
		&\limsup_{K\to\infty}\abs*{\g(\altpoint)-\g(0)-\ssuml_{k=1}^K a_k\altpoint^k}
		\\&=
		\limsup_{K\to\infty}\abs*{\f(\altpoint+\cpoint)-\f(\cpoint)-\ssuml_{k=1}^K a_k((\altpoint+\cpoint)-\cpoint)^k}
		=
		0
		.
	\end{eqsplit}
		\Cref{cor:kl-powseries1d}
		\hence
	establishes that there exist
		$\delta\in(0,\eps)$,
		$c\in(0,\infty)$,
		$\alpha\in(0,1)$
	which satisfy for all
		$\altpoint\in(-\delta,\delta)$
	that
	\begin{equation}
		\abs{\g(\altpoint)-\g(0)}^{\alpha}
		\leq 
		c\,\abs{\g'(\altpoint)}
		.
	\end{equation}
		\Hence 
	for all
		$\altpoint\in(-\delta,\delta)$
	that
	\begin{equation}
		\abs{\f(\altpoint+\cpoint)-\f(\cpoint)}^\alpha
		\leq 
		c\,\abs{\f'(\altpoint+\cpoint)}
		.
	\end{equation}
		This
	implies
		\lref{claim}.
\end{aproof}
\endgroup

\cfclear
\begingroup
\providecommand{\f}{}
\renewcommand{\f}{\defaultLossFunction}
\begin{athm}{cor}{cor:kl-analytic1d}
	Let
		$U\subseteq\R$ be open,
	let
		$\f\colon U\to\R$ be \cfadd{def:analytic_function}analytic,
	and let
		$\cpoint\in U$
	\cfload.
	Then there exist
		$\eps,c\in(0,\infty)$,
		$\alpha\in(0,1)$
	such that for all
		$\altpoint\in(\cpoint-\eps,\cpoint+\eps)$
	it holds that
	\begin{equation}
		\llabel{claim}
		\abs{\f(\altpoint)-\f(\cpoint)}^\alpha
		\leq
		c\,\abs{(\nabla \f)(\altpoint)}
		.
	\end{equation}
\end{athm}
\begin{aproof}
	\Nobs that 
	\enum{
		\cref{lemma:analytic_functions_onedim};
		\cref{cor:kl-powseries1d.2}
	}
	\prove
		\lref{claim}.
\end{aproof}
\endgroup

\cfclear
\begingroup
\providecommand{\d}{}
\renewcommand{\d}{\defaultParamDim}
\providecommand{\f}{}
\renewcommand{\f}{\defaultLossFunction}
\begin{athm}{cor}{cor:kl-analytic1d_2}
\cfadd{def:analytic_function}
Let $ \f \colon \R \to \R $ 
be analytic \cfload.
Then $ \f $\cfadd{def:KLfunction_standard}
is a standard \KL\ function
\cfout.
\end{athm}
\begin{aproof}
\Nobs that \cref{def:KLinequality_standard:eq1}
and 
\cref{cor:kl-analytic1d} 
establish 
that $ \f $\cfadd{def:KLfunction_standard}
is a 
standard \KL\ function
\cfload.
\end{aproof}
\endgroup

\section{Standard KL inequalities for analytic functions}  

\cfclear
\begingroup
\providecommand{\d}{}
\renewcommand{\d}{\defaultParamDim}
\providecommand{\f}{}
\renewcommand{\f}{\defaultLossFunction}
\begin{athm}{theorem}{thm:KL}[Standard \KL\ inequalities for analytic functions]
\cfadd{def:analytic_function}
Let $ \d \in \N $, 
let $ U \subseteq \R^{ \d } $ be open, 
let $ \f \colon U \to \R $ be analytic, 
and let $ \cpoint \in U $
\cfload.
Then there exist 
$ \varepsilon, c \in (0,\infty) $, 
$ \alpha \in (0,1) $
such that 
for all 
$
  \altpoint \in \{ \altpointTwo \in U \colon \Pnorm2{ \cpoint - \altpointTwo} < \varepsilon \} 
$
it holds that 
\begin{equation}
	\llabel{eq:claim}
  \abs{ \f( \cpoint ) - \f( \altpoint ) }^{ \alpha }
  \leq 
  c
  \,
  \Pnorm2{ 
    ( \nabla \f )( \altpoint )
  }
\end{equation}
\cfload.
\end{athm}
\begin{aproof}
	\Nobs that
		\L{}ojasiewicz~\cite[Proposition~1]{lojasiewicz1964ensembles}
	\proves
		\lref{eq:claim} (cf., \eg, also Bierstone \& Milman~\cite[Proposition~6.8]{bierstone1988semianalytic}).
\end{aproof}
\endgroup

\cfclear
\begingroup
\providecommand{\d}{}
\renewcommand{\d}{\defaultParamDim}
\providecommand{\f}{}
\renewcommand{\f}{\defaultLossFunction}
\begin{athm}{cor}{cor:KL}
\cfadd{def:analytic_function}
Let $ \d \in \N $ 
and let $ \f \colon \R^{ \d } \to \R $ 
be analytic \cfload.
Then $ \f $\cfadd{def:KLfunction_standard}
is a standard \KL\ function
\cfout.
\end{athm}
\begin{aproof}
\Nobs that 
\cref{def:KLinequality_standard:eq1}
and 
\cref{thm:KL} 
establish 
that $ \f $\cfadd{def:KLfunction_standard}
is a 
standard \KL\ function
\cfload.
\end{aproof}
\endgroup

\section{Counterexamples}

\todoc{add exercise from exam (DDA6206 with $x^{-2}$): DONE}

\cfclear
\begingroup
\providecommand{\f}{}
\renewcommand{\f}{\defaultLossFunction}
\begin{athm}{example}{prop:counterexample_KL_1}[Example of a smooth function that is not a standard \KL\ function]
Let $ \f \colon \R \to \R $ satisfy 
for all $ \altpoint \in \R $ that 
\begin{equation}
\label{eq:counterexample_KL_1}
  \f(\altpoint)
  =
  \begin{cases}
    \exp( - \altpoint^{ - 1 } )
  &
    \colon 
    \altpoint > 0
  \\
    0
  &
    \colon
    \altpoint \leq 0.
  \end{cases}
\end{equation}
Then 
\begin{enumerate}[label=(\roman*)]
\item \llabel{it:1}
it holds that $ \f \in C^{ \infty }( \R, \R ) $, 
\item \llabel{it:2}
it holds for all $ \altpoint \in (0,\infty) $ that 
$
  \f'(\altpoint) 
  = 
  \altpoint^{ - 2 } \exp( - \altpoint^{ - 1 } )
$, 
\item \llabel{it:3}
it holds 
for all $ \alpha \in (0,1) $, $ \varepsilon \in (0,\infty) $
that 
\begin{eqsplit}
  \sup_{ \altpoint \in (0, \varepsilon) }
  \pr*{
  \frac{ \abs{ \f( \altpoint ) - \f(0) }^{ \alpha } }{ \abs{ \f'( \altpoint ) } }  
  }
  = \infty
  ,
\end{eqsplit}
and 
\item \llabel{it:4}
it holds that $ \f $ is not a 
standard \KL\ function
\cfadd{def:KLfunction_standard}
\end{enumerate}
\cfload.
\end{athm}
\begin{aproof}
	Throughout this proof, 
		let
		\begin{equation}
			P
			=
			\{f\in C((0,\infty),\R) \colon \text{$f$ is a polynomial}\}
		\end{equation}
		and for every
			$f\in C((0,\infty),\R)$
		let
			$G_f\colon (0,\infty)\to\R$
			satisfy for all
				$\altpoint\in (0,\infty)$
			that
			\begin{equation}
				G_f(\altpoint)
				=
				f(\altpoint^{-1})\exp(-\altpoint^{-1})
				.
			\end{equation}
	\Nobs that
		the chain rule
		and the product rule
	ensure that for all
		$f\in C^1((0,\infty),\R)$,
		$\altpoint\in (0,\infty)$
	it holds that
		$G_f\in C^1((0,\infty),\R)$
	and
	\begin{eqsplit}
		\llabel{eq:1}
		(G_f)'(\altpoint)
		&=
		-f'(\altpoint^{-1})\altpoint^{-2}\exp(-\altpoint^{-1}) + f(\altpoint^{-1})\altpoint^{-2}\exp(-\altpoint^{-1})
		\\&=
		(f(\altpoint^{-1})-f'(\altpoint^{-1}))\altpoint^{-2}\exp(-\altpoint^{-1})
		.
	\end{eqsplit}
	\Hence for all
		$p\in P$
	that there exists
		$q\in P$
	such that
	\begin{equation}
		(G_p)' = G_q
		.
	\end{equation}
	Combining
		this
		and \lref{eq:1}
	with
		induction
	\proves that for all 
		$p\in P$,
		$n\in\N$
	it holds that
	\begin{equation}
		G_p\in C^\infty((0,\infty),\R)
		\qquad\text{and}\qquad
		(\exists\, q\in P\colon (G_p)^{(n)} = G_q)
		.
	\end{equation}
		This
		and the fact that 
			for all
				$p\in P$
			it holds that
				$\lim_{\altpoint\searrow 0} G_p(\altpoint)=0$
	\prove that for all
		$p\in P$,
		$n\in\N$
	it holds that
	\begin{equation}
		\lim_{\altpoint\searrow 0}(G_p)^{(n)}(\altpoint) = 0
		.
	\end{equation}
		The fact that
			$\f|_{(0,\infty)} = G_{(0,\infty)\ni \altpoint\mapsto 1\in\R}$
		and \lref{eq:1}
		\hence
	establish
		\cref{prop:counterexample_KL_1.it:1,prop:counterexample_KL_1.it:2}.
\Nobs that \cref{eq:counterexample_KL_1} 
and the fact that for all 
$ \altpoint \in (0,\infty) $ it holds that 
\begin{equation}
    \exp( \altpoint )
  =
    \sum_{ k = 0 }^{ \infty }
    \frac{ \altpoint^k }{ k! }
  \geq 
    \frac{ \altpoint^3 }{ 3! }
  =
    \frac{ \altpoint^3 }{ 6 }  
\end{equation}
ensure that 
for all 
$ \alpha \in (0,1) $, $ \varepsilon \in (0,\infty) $, 
$ \altpoint \in (0, \varepsilon) $
it holds that 
\begin{eqsplit}
  \frac{ \abs{ \f( \altpoint ) - \f(0) }^{ \alpha } }{ \abs{ \f'( \altpoint ) } }  
&
  =
  \frac{ \abs{ \f( \altpoint ) }^{ \alpha } }{ \abs{ \f'( \altpoint ) } }  
  =
  \frac{ \altpoint^2 \abs{ \f( \altpoint ) }^{ \alpha } }{ \f( \altpoint ) }  
  =
    \altpoint^2 \abs{ \f( \altpoint ) }^{ \alpha - 1 } 
\\ &
  = 
    \altpoint^2 
    \exp\pr*{
      \frac{ ( 1 - \alpha ) }{ \altpoint } 
    }
  \geq 
  \frac{ \altpoint^2 ( 1 - \alpha )^3 }{ 6 \altpoint^3 }
  =
  \frac{ ( 1 - \alpha )^3 }{ 6 \altpoint }
  .
\end{eqsplit}
\Hence 
for all 
$ \alpha \in (0,1) $, $ \varepsilon \in (0,\infty) $
that 
\begin{equation}
  \sup_{ \altpoint \in (0, \varepsilon) }
  \pr*{
    \frac{ \abs{ \f( \altpoint ) - \f(0) }^{ \alpha } }{ \abs{ \f'( \altpoint ) } }  
  }
  \geq 
  \sup_{ \altpoint \in (0, \varepsilon) }
  \pr*{
    \frac{ ( 1 - \alpha )^3 }{ 6 \altpoint }
  }
  =
  \infty .
\end{equation}
This establishes \cref{prop:counterexample_KL_1.it:3,prop:counterexample_KL_1.it:4}.
\end{aproof}
\endgroup

\cfclear
\begingroup
\providecommand{\f}{}
\renewcommand{\f}{\defaultLossFunction}
\providecommand{\g}{}
\renewcommand{\g}{\defaultGradientFunction}
\begin{athm}{example}{prop:counterexample_KL_2}[Example of a differentiable function that fails to satisfy the standard \KL\ inequality]
Let $ \f \colon \R \to \R $ satisfy 
for all $ \altpoint \in \R $ that 
\begin{equation}
\label{eq:counterexample_KL_2}
\textstyle 
  \f(\altpoint)
  =
  \int_0^{ \max\{ \altpoint, 0 \} }
  x \abs{ \sin( x^{ - 1 } ) }
  \, 
  \diff x
  .
\end{equation}
Then 
\begin{enumerate}[label=(\roman*)]
\item 
it holds that $ \f \in C^1( \R, \R ) $, 
\item 
it holds for all 
$ c \in \R $, 
$ \alpha, \varepsilon \in (0,\infty) $ that 
there exist $ \altpoint \in (0,\varepsilon) $
such that 
\begin{eqsplit}
  \abs{ \f( \altpoint ) - \f( 0 ) }^{ \alpha }
  >
  c \abs{ \f'(\altpoint) }
  ,
\end{eqsplit}
and 
\item 
it holds 
for all 
$ c \in \R $, 
$ \alpha, \varepsilon \in (0,\infty) $
that 
we do not have that 
$ \f $ satisfies the  
standard \KL\ inequality 
at $ 0 $ on $ [ 0, \varepsilon ) $ 
with exponent $ \alpha $ and constant $ c $
\cfadd{def:KLinequality_standard}
\end{enumerate}
\cfout.
\end{athm}
\begin{aproof}
Throughout this proof, let 
$ \g \colon \R \to \R $ satisfy for all 
$ \altpoint \in \R $ that 
\begin{eqsplit}
\label{eq:definition_of_g_in_proof}
  \g(\altpoint) 
  = 
  \begin{cases}
    \altpoint \abs{ \sin( \altpoint^{ - 1 } ) }
  &
    \colon 
    \altpoint > 0
  \\
    0
  &
    \colon 
    \altpoint \leq 0.
  \end{cases}
\end{eqsplit}
\Nobs that \cref{eq:definition_of_g_in_proof} 
\proves that for all $ k \in \N $ it holds that 
\begin{equation}
\label{eq:zeros_of_g}
  \g( ( k \pi )^{ - 1 } )
  =
  ( k \pi )^{ - 1 } 
  \abs{
    \sin( k \pi )
	}
  =
  0 
  .
\end{equation}
\Moreover \cref{eq:definition_of_g_in_proof} 
\proves for all 
$ \altpoint \in (0,\infty) $ 
that 
\begin{equation}
  \abs{ \g(\altpoint) - \g(0) }
  =
  \abs{ \altpoint \sin( \altpoint^{ - 1 } ) }
  \leq 
  \abs{ \altpoint }
  .
\end{equation}
\Hence that $ \g $ is continuous. 
This, \cref{eq:counterexample_KL_2},  
and the fundamental theorem of calculus ensure that 
$ \f $ is continuously differentiable with  
\begin{equation}
  \f' = \g.
\end{equation}
Combining this with \cref{eq:zeros_of_g} demonstrates that 
for all 
$ c \in \R $, 
$ \alpha \in (0,\infty) $, 
$ k \in \N $ it holds that 
\begin{eqsplit}
  \abs{ \f( ( k \pi )^{ - 1 } ) - \f(0) }^{ \alpha }
  =
  [ \f( ( k \pi )^{ - 1 } ) ]^{ \alpha }
  > 0
  = 
  c \abs{ \g( ( k \pi )^{ - 1 } ) }
  =
  c \abs{ \f'( ( k \pi )^{ - 1 } ) }
  .
\end{eqsplit}
\end{aproof}
\endgroup

\begin{exercise}{exercise:KL_counterexample}
Let $\defaultLossFunction \colon \R \to \R$ satisfy for all $\altpoint \in \R$ that
\begin{equation*}
    \defaultLossFunction(\altpoint)
=
    \begin{cases}
        \exp(-\altpoint^{-2}) & \colon \altpoint > 0 \\
        0 & \colon \altpoint \leq 0.
    \end{cases}
\end{equation*}
Prove or disprove the following statement:
It holds that $\defaultLossFunction$ is a standard \KL\ function. 
\end{exercise}

\section{Convergence analysis for solutions of GF ODEs}
\label{section:gf:loja}

In this section we employ standard \KL\ inequalities to establish convergence of solutions of \GF\ \ODEs\ to critical points (see \cref{sec:critical_points}). The specific presentation of this section is closely based on \cite[Section~7]{jentzen2022existence_arxiv}.

\subsection{Abstract local convergence results for GF processes}
\label{subsection:gf:local:conv}

\cfclear
\begingroup
\providecommand{\d}{}
\renewcommand{\d}{\defaultParamDim}
\providecommand{\f}{}
\renewcommand{\f}{\defaultLossFunction}
\providecommand{\g}{}
\renewcommand{\g}{\defaultGradientFunction}
\begin{athm}{lemma}{prop:gf:chainrule.2}
	Let 
	$ \d \in \N $, 
	$ \Theta \in C( [0,\infty), \R^{ \d } ) $, 
	$ \f \in C^1( \R^{ \d }, \R ) $,
	let 
		$\g\colon \R^\d\to\R^\d$
	satisfy for all
		$\altpoint\in\R^d$
	that
		$\g(\altpoint)=(\nabla\f)(\altpoint)$,
	and assume
	for all $ t \in [0,\infty) $ 
	that
	$
  	\Theta_t = \Theta_0 - \int_0^t \g( \Theta_s ) \, \diff s
	$
	\cfload. 
	Then 
		it holds
		for all $ t \in [0,\infty) $ 
		that
		\begin{equation}
			\llabel{claim}
			\f( \Theta_t ) = \f( \Theta_0 ) - \int_0^t \pnorm2{ \g( \Theta_s ) }^2 \, \diff s
		\end{equation}
	\cfout.
\end{athm}

\begin{aproof}
\Nobs that 
	\cref{prop:gf:chainrule}
implies
	\lref{claim}.
	\finishproofthis
\end{aproof}
\endgroup

\cfclear
\begingroup
\providecommand{\d}{}
\renewcommand{\d}{\defaultParamDim}
\providecommand{\f}{}
\renewcommand{\f}{\defaultLossFunction}
\providecommand{\g}{}
\renewcommand{\g}{\defaultGradientFunction}
\begin{athm}{prop}{prop:gf:conv:local}
	Let 
	$ \d \in \N $, 
	$ \cpoint \in \R^{ \d } $, 
	$ \consttt \in \R $, 
	$ \const, \varepsilon \in (0, \infty) $, 
	$ \alpha \in (0,1) $, 
	$ \Theta \in C( [0,\infty), \R^{ \d } ) $, 
	$ \cL \in C( \R^{ \d }, \R ) $, 
	let $ \g \colon \R^{ \d } \to \R^{ \d } $ 
	be $\Borel(\R^\d)$/$\Borel(\R^\d)$-measurable, 
	assume 
	for all $ t \in [0,\infty) $ 
	that
	\begin{equation}
		\llabel{eq:defTheta}
		\cL( \Theta_t ) = \cL( \Theta_0 ) - \int_0^t \pnorm2{ \g( \Theta_s ) }^2 \, \diff s
		\qquad\text{and}\qquad
  	\Theta_t = \Theta_0 - \int_0^t \g( \Theta_s ) \, \diff s
		,
	\end{equation}
	and assume for all 
	$
	\altpoint \in \R^{ \d } 
	$
	with 
	$
	\Pnorm2{ \altpoint - \cpoint } < \varepsilon
	$
	that
	\begin{equation} 
	\label{prop:gf:conv:local:eq:loj}
	\abs{ 
		\cL( \altpoint ) - \cL( \cpoint ) 
	}^\alpha 
	\leq 
	\const \Pnorm2{ \g( \altpoint ) } ,
	\quad 
	\consttt = 
	\abs{
	\cL( \Theta_0 ) - \cL( \cpoint )
	},
	\quad 
	\const 
	( 1 - \alpha )^{ - 1 }
	\consttt^{ 1 - \alpha }
	+
	\Pnorm2{ \Theta_0 - \cpoint }
	< 
	\varepsilon
	,
	\end{equation}
	and 
	$
	\inf_{ 
		t \in 
		\cu{ 
			s \in  [0, \infty) \colon 
			\forall \, r \in [0,s] \colon 
			\Pnorm2{ \Theta_r - \cpoint } < 
			\varepsilon
		} 
	} 
	\cL( \Theta_t ) \geq \cL( \cpoint )
	$
	\cfload. 
	Then there exists 
	$ \psi \in \R^{\d}$ 
	such that 
	\begin{enumerate}[label = (\roman*)]
		\item 
		\label{prop:quantitative:item_0}
		it holds that $\cL( \psi ) = \cL( \cpoint )$,
		\item 
		\label{prop:quantitative:item_i}
		it holds for all $ t \in [0, \infty) $ that 
		$
		\Pnorm2{ \Theta_t - \cpoint } < \varepsilon
		$,
		\item 
		\label{prop:quantitative:item_ii}
		it holds for all $ t \in [0,\infty) $ that
		$
		0 \leq 
		\cL( \Theta_t ) - \cL( \psi ) 
		\leq 
		\const^2
		\consttt^2
		(
		\indicator{
			\{ 0 \}
		}( \consttt )
		+
		\const^2 \consttt
		+ 
		\consttt^{
			2 \alpha 
		}
		t 
		)^{ - 1 }
		$,
		and 
		\item 
		\label{prop:quantitative:item_iii}
		it holds for all $ t \in [0,\infty) $ that 
		\begin{equation}
		\label{prop:gf:conv:local:eq:statement}
		\begin{split}
		\Pnorm2{
			\Theta_t - \psi 
		}
		& \leq
		\int_t^{ \infty }
		\Pnorm2{ 
			\g( \Theta_s )
		}
		\,
		\diff s
		\leq
		\const 
		\rbr{ 1 - \alpha }^{ - 1 } 
		[ 
		\cL( \Theta_t ) - \cL( \psi )
		]^{ 1 - \alpha } 
		\\ & \leq 
		\const^{ 3 - 2 \alpha } 
		\consttt^{ 2 - 2 \alpha }
		\rbr{ 1 - \alpha }^{ - 1 } 
		(
		\indicator{
			\{ 0 \}
		}( \consttt )
		+
		\const^2 \consttt
		+ 
		\consttt^{
			2 \alpha 
		}
		t 
		)^{ \alpha - 1 }
		.
		\end{split}
		\end{equation}
	\end{enumerate}
\end{athm}

\begin{aproof}
	Throughout this proof, let $ \mathbf{L} \colon [0, \infty) \to \R $ 
	satisfy for all $ t \in [0, \infty) $ that 
	\begin{equation} 
	\label{prop:gf:convergence:eq:defl}
	\mathbf{L}(t) = \cL ( \Theta_t ) - \cL ( \cpoint ) ,
	\end{equation}
	let $ \bB \subseteq \R^{ \d } $ satisfy 
	\begin{equation}
	\bB = \{ \altpoint \in \R^{ \d } \colon \pnorm2{ \altpoint - \cpoint } < \varepsilon \} ,
	\end{equation}
	let $ T \in [0, \infty] $ 
	satisfy
	\begin{equation} 
	\label{prop:gf:convergence:eq:tout}
	T = 
	\inf\rbr*{ 
		\cu*{ t \in [0, \infty ) \colon \Theta_t \notin \bB } 
		\cup \cu{\infty} 
	} ,
	\end{equation}
	let $ \tau \in [0,T] $ satisfy
	\begin{equation}
	\label{prop:gf:convergence:eq:tau}
	\tau = 
	\inf\rbr*{ 
		\cu*{ t \in [0,T) \colon \mathbf{L}( t ) = 0 } \cup \cu{T } 
	} 
	,
	\end{equation}
	let 
	$ 
	\mathscr{g} = ( \mathscr{g}_t )_{ t \in [0,\infty) }
	\colon [0,\infty) \to [0,\infty] 
	$ 
	satisfy for all $ t \in [0,\infty) $ that 
	$
	\mathscr{g}_t = \int_t^{ \infty } \Pnorm2{ \g( \Theta_s ) } \, \diff s
	$, 
	and let 
	$
	\constt \in \R 
	$ satisfy 
	$
	\constt 
	=
	\const^2 
	\consttt^{ ( 2 - 2 \alpha ) } 
	$.
	In the first step of our proof 
	of 
	\cref{prop:quantitative:item_0,prop:quantitative:item_i,prop:quantitative:item_ii,prop:quantitative:item_iii}
	we show that for all $ t \in [0,\infty) $
	it holds that 
	\begin{equation}
	\label{eq:to_prove}
	\Theta_t \in \bB
	.
	\end{equation}
	For this we \nobs that 
	\cref{prop:gf:conv:local:eq:loj}, 
	the triangle inequality, 
	and 
	the assumption that 
	for all $ t \in [0,\infty) $ it holds that 
	$
	\Theta_t = \Theta_0 - \int_0^t \g( \Theta_s ) \, \diff s
	$
	\prove that 
	for all $ t \in [0,\infty) $ it holds that
	\begin{equation} 
	\label{eq:theta_t}
	\begin{split}
	&
	\Pnorm2{ \Theta_t - \cpoint } 
	\leq  
	\Pnorm2{ \Theta_t - \Theta_0 } 
	+ 
	\Pnorm2{ \Theta_0 - \cpoint }  
	\leq
	\Pnorm*2{ \int_{0}^t \g ( \Theta_s )  \, \diff s} 
	+
	\Pnorm2{ \Theta_0 - \cpoint }  
	\\ & \leq 
	\int_0^t \Pnorm2{\g(\Theta_s ) } \, \diff s 
	+
	\Pnorm2{ \Theta_0 - \cpoint }  
	< 
	\int_0^t \Pnorm2{\g(\Theta_s ) } \, \diff s 
	-
	\const 
	( 1 - \alpha )^{ - 1 }
	\abs{
		\cL( \Theta_0 ) - \cL( \cpoint ) 
	}^{ 1 - \alpha } 
	+
	\varepsilon
	.
	\end{split}
	\end{equation} 
	To establish \cref{eq:to_prove}, it is thus sufficient 
	to prove that 
	$
	\int_0^T \Pnorm2{\g(\Theta_s ) } \, \diff s 
	\leq 
	\const 
	( 1 - \alpha )^{ - 1 }
	\abs{
		\cL( \Theta_0 ) - \cL( \cpoint ) 
	}^{ 1 - \alpha } 
	$. 
	We will accomplish this 
	by employing an appropriate differential inequality 
	for a fractional power of the function $ \mathbf{L} $ in 
	\cref{prop:gf:convergence:eq:defl} 
	(see \cref{prop:gf:conv:eq:derivative} below for details). 
	For this we need several technical preparations. 
	More formally, 
	\nobs that 
	\cref{prop:gf:convergence:eq:defl} 
	and the assumption that for all $ t \in [0,\infty) $ it holds that 
	\begin{equation}
	\label{eq:weak_chain_rule}
	\cL( \Theta_t ) = \cL( \Theta_0 ) - \int_0^t \pnorm2{ \g( \Theta_s ) }^2 \, \diff s
	\end{equation} 
	\prove that for almost all $ t \in [0,\infty) $ it holds 
	that $ \mathbf{L} $ is differentiable at $ t $ and 
	satisfies 
	\begin{equation}
	\label{eq:derivative_bL}
	\mathbf{L}'( t ) = \tfrac{ \d }{ \d t }( \cL( \Theta_t ) ) 
	= - \Pnorm2{ \g( \Theta_t ) }^2
	.
	\end{equation}
	\Moreover the assumption that 
	$
	\inf_{ 
		t \in 
		\cu{ 
			s \in  [0, \infty) \colon 
			\forall \, r \in [0,s] \colon 
			\Pnorm2{ \Theta_r - \cpoint } < 
			\varepsilon
		} 
	} 
	\cL( \Theta_t ) \geq \cL( \cpoint )
	$
	\proves that for all $ t \in [0, T) $ it holds that 
	\begin{equation}
	\label{eq:Lgeq0}
	\mathbf{L}(t) \geq 0 
	.
	\end{equation}
	Combining this with 
	\cref{prop:gf:conv:local:eq:loj}, 
	\cref{prop:gf:convergence:eq:defl}, 
	and 
	\cref{prop:gf:convergence:eq:tau}
	\proves that for all $t \in [0, \tau)$ it holds that
	\begin{equation}
	0 < [ \mathbf{L}( t ) ]^{ \alpha } 
	= \abs{ \cL( \Theta_t ) - \cL( \cpoint ) }^{ \alpha } 
	\leq \const \Pnorm2{\g ( \Theta_t ) } 
	.
	\end{equation}
	The chain rule and \cref{eq:derivative_bL} 
	\hence \prove that for almost all $ t \in [0, \tau) $ it holds that
	\begin{equation} 
	\label{prop:gf:conv:eq:derivative}
	\begin{split}
	\tfrac{ \d }{ \d t } 
	( [ \mathbf{L}( t ) ]^{ 1 - \alpha } ) 
	&
	= ( 1 - \alpha ) [ \mathbf{L}( t ) ]^{ - \alpha } \rbr{ - \Pnorm2{\g(\Theta_t)}^2  } 
	\\
	&
	\leq 
	- ( 1 - \alpha ) 
	\const^{ - 1 } 
	\Pnorm2{ \g( \Theta_t ) }^{ - 1 } 
	\Pnorm2{ \g( \Theta_t ) }^2 
	= - \const^{ - 1 } ( 1 - \alpha ) 
	\Pnorm2{ \g( \Theta_t ) } .
	\end{split}
	\end{equation}
	\Moreover \cref{eq:weak_chain_rule} \proves that 
	$ [0, \infty ) \ni t \mapsto \mathbf{L}(t) \in \R $ is absolutely continuous. 
	This and the fact that for all $ r \in (0, \infty) $ it holds that 
	$ [ r , \infty ) \ni y \mapsto y^{ 1 - \alpha } \in \R $ is Lipschitz continuous 
	\prove that for all $ t \in [0, \tau) $ 
	it holds that $ [0, t] \ni s \mapsto [ \mathbf{L}( s ) ]^{ 1 - \alpha } \in \R $ 
	is absolutely continuous. 
	Combining this with \cref{prop:gf:conv:eq:derivative} \proves 
	that for all $ s, t \in [0, \tau ) $ with $ s \le t $ it holds that
	\begin{equation} 
	\label{eq:integral_inequality}
	\int_s^t 
	\Pnorm2{ \g( \Theta_u ) } 
	\, \d u 
	\leq 
	- \const \rbr{ 1 - \alpha }^{ - 1 } 
	\rbr{  
		[ \mathbf{L}( t ) ]^{ 1 - \alpha } - [ \mathbf{L}( s ) ]^{ 1 - \alpha } 
	} 
	\leq 
	\const 
	\rbr{ 1 - \alpha }^{ - 1 } 
	[ \mathbf{L}( s ) ]^{ 1 - \alpha } .
	\end{equation}
	In the next step we \nobs that 
	\cref{eq:weak_chain_rule}
	\proves that 
	$
	[0, \infty) \ni t \mapsto \cL ( \Theta_t ) \in \R 
	$ 
	is non-increasing. 
	This and \cref{prop:gf:convergence:eq:defl} 
	\prove that $ \mathbf{L} $ is non-increasing. 
	Combining \cref{prop:gf:convergence:eq:tau} 
	and \cref{eq:Lgeq0} \hence \proves
	that for all 
	$ t \in [\tau , T) $ 
	it holds that $ \mathbf{L}(t) = 0 $. 
	\Hence that for all 
	$ t \in ( \tau, T ) $ 
	it holds that 
	\begin{equation}
	\label{eq:derivative_zero}
	\mathbf{L}'( t ) = 0
	.
	\end{equation}
	This and \cref{eq:derivative_bL} \prove that 
	for almost all $ t \in (\tau, T) $ it holds that 
	\begin{equation}
	\label{eq:G_vanishes}
	\g( \Theta_t ) = 0 
	.
	\end{equation}
	Combining this with \cref{eq:integral_inequality} \proves that 
	for all $ s, t \in [0, T) $ with $ s \le t $ 
	it holds that 
	\begin{equation} 
	\label{prop:gf_conv:eq:integrated}
	\int_s^t \Pnorm2{ \g( \Theta_u ) } \, \diff u 
	\leq 
	\const \rbr{ 1 - \alpha}^{-1} [ \mathbf{L}( s ) ]^{ 1 - \alpha } 
	.
	\end{equation}
	\Hence that 
	for all $ t \in [0, T) $ 
	it holds that 
	\begin{equation}
	\label{prop:gf_conv:eq:integrated1b}
	\int_0^t \Pnorm2{ \g( \Theta_u ) } \, \diff u 
	\leq 
	\const \rbr{ 1 - \alpha}^{-1} [ \mathbf{L}( 0 ) ]^{ 1 - \alpha } 
	.
	\end{equation}
	\Moreover \cref{prop:gf:conv:local:eq:loj} 
	\proves that $ \Theta_0 \in \bB $. 
	Combining this with \cref{prop:gf:convergence:eq:tout} 
	\proves that $ T > 0 $. 
	This, \cref{prop:gf_conv:eq:integrated1b}, 
	and \cref{prop:gf:conv:local:eq:loj} 
	\prove that 
	\begin{equation} 
	\label{prop:gf_conv:eq:integrated2}
	\int_0^T \Pnorm2{ \g( \Theta_u ) } \, \diff u 
	\leq 
	\const \rbr{ 1 - \alpha}^{-1} [ \mathbf{L}( 0 ) ]^{ 1 - \alpha } 
	< 
	\varepsilon 
	< \infty 
	.
	\end{equation}
	Combining \cref{prop:gf:convergence:eq:tout} and \cref{eq:theta_t} 
	hence \proves that 
	\begin{equation}
	\label{eq:T_infinity} 
	T = \infty 
	.
	\end{equation}
	This \proves[ep] \cref{eq:to_prove}. 
	In the next step of our proof of 
	\cref{prop:quantitative:item_0,prop:quantitative:item_i,prop:quantitative:item_ii,prop:quantitative:item_iii}
	we verify that 
	$ \Theta_t \in \R^{ \d } $, $ t \in [0,\infty) $, 
	is convergent (see \cref{eq:convergent} below). 
	For this \nobs that 
	the assumption that 
	for all $ t \in [0,\infty) $ it holds that 
	$
	\Theta_t = \Theta_0 - \int_0^t \g( \Theta_s ) \, \diff s
	$
	\proves that for all 
	$ r, s, t \in [0,\infty) $ 
	with $ r \leq s \leq t $ 
	it holds that 
	\begin{equation}
	\label{eq:Cauchy_family}
	\Pnorm2{ \Theta_t - \Theta_s } 
	=
	\Pnorm*2{ 
		\int_s^t
		\g( \Theta_u ) \, \diff u
	}
	\leq 
	\int_s^t 
	\Pnorm2{ 
		\g( \Theta_u )
	}
	\, \diff u
	\leq 
	\int_r^{ \infty }
	\Pnorm2{
		\g( \Theta_u )
	}
	\, \diff u
	= \mathscr{g}_r 
	.
	\end{equation}
	\Moreover
	\cref{prop:gf_conv:eq:integrated2}
	and 
	\cref{eq:T_infinity} 
	\prove that 
	$
	\infty >
	\mathscr{g}_0
	\geq 
	\limsup_{ r \to \infty } 
	\mathscr{g}_r = 0
	$. 
	Combining this with 
	\cref{eq:Cauchy_family} \proves that there exist 
	$ \psi \in \R^{ \d } $ 
	which satisfies 
	\begin{equation}
	\label{eq:convergent}
	\limsup\nolimits_{ t \to \infty } \Pnorm2{ \Theta_t - \psi } = 0 .
	\end{equation}
	In the next step of our proof of 
	\cref{prop:quantitative:item_0,prop:quantitative:item_i,prop:quantitative:item_ii,prop:quantitative:item_iii}
	we show that $ \cL( \Theta_t ) $, $ t \in [0,\infty) $, 
	converges to $ \cL( \psi ) $ with convergence order $ 1 $. 
	We accomplish this by bringing a suitable differential inequality 
	for the reciprocal of the function $ \mathbf{L} $ 
	in \cref{prop:gf:convergence:eq:defl} into play 
	(see \cref{eq:d_dt_reciprocal_L} below for details). More specifically, 
	\nobs that 
	\cref{eq:derivative_bL}, 
	\cref{eq:T_infinity}, 
	\cref{prop:gf:convergence:eq:tout}, 
	and \cref{prop:gf:conv:local:eq:loj} 
	\prove that for almost all $ t \in [0, \infty) $ 
	it holds that
	\begin{equation}
	\mathbf{L}'( t ) 
	= - \Pnorm2{ \g( \Theta_t ) }^2 
	\leq - \const^{ - 2 } [ \mathbf{L}( t ) ]^{ 2 \alpha } .
	\end{equation}
	\Hence that $ \mathbf{L} $ is non-increasing. 
	This \proves that for all $ t \in [0,\infty) $ 
	it holds that 
	$ \mathbf{L}(t) \leq \mathbf{L}( 0 ) $. 
	This and the fact that for all $ t \in [0, \tau) $ 
	it holds that $ \mathbf{L}( t ) > 0 $ 
	\prove that for almost all $ t \in [0, \tau) $ 
	it holds that
	\begin{equation}
	\mathbf{L}'( t )
	\leq 
	- \const^{ - 2 } 
	[ \mathbf{L}( t ) ]^{ ( 2 \alpha - 2 ) } 
	[ \mathbf{L}( t ) ]^{ 2 } 
	\leq 
	- \const^{ - 2 } 
	[ \mathbf{L}( 0 ) ]^{ ( 2 \alpha - 2 ) } 
	[ \mathbf{L}( t ) ]^{ 2 } 
	=
	- \constt^{ - 1 }
	[ \mathbf{L}( t ) ]^{ 2 } 
	.
	\end{equation}
	\Hence that for almost all $ t \in [0, \tau) $ 
	it holds that
	\begin{equation}
	\label{eq:d_dt_reciprocal_L}
	\frac{ \d }{ \d t } 
	\rbr*{ 
		\frac{ 
			\constt  
		}{ 
			\mathbf{L}( t ) 
		} 
	} 
	= 
	- 
	\rbr*{
		\frac{
			\constt \, 
			\mathbf{L}'( t )
		}{
			[ \mathbf{L}( t ) ]^2 
		} 
	}
	\geq 1 .
	\end{equation}
	\Moreover the fact that 
	for all $ t \in [0, \tau) $ 
	it holds that 
	$ [0, t] \ni s \mapsto \mathbf{L}( s ) \in (0, \infty) $ 
	is absolutely continuous 
	\proves that for all $ t \in [0,\tau) $ 
	it holds that 
	$ [0, t] \ni s \mapsto \constt [ \mathbf{L}( s ) ]^{ - 1 } \in (0, \infty) $ 
	is absolutely continuous. 
	This and \cref{eq:d_dt_reciprocal_L} \prove that 
	for all $ t \in [0,\tau) $ it holds that 
	\begin{equation}
	\frac{ 
		\constt
	}{
		\mathbf{L}(t)
	}
	-
	\frac{ 
		\constt
	}{
		\mathbf{L}(0)
	}
	\geq t
	.
	\end{equation}
	\Hence that for all $ t \in [0,\tau) $ 
	it holds that 
	\begin{equation}
	\frac{ \constt }{ \mathbf{L}(t) } 
	\geq 
	\frac{ \constt }{ \mathbf{L}(0) } 
	+ t 
	. 
	\end{equation}
	\Hence that for all 
	$ t \in [0,\tau) $ it holds that 
	\begin{equation}
	\constt 
	\,
	\pr*{
	\frac{ \constt }{ \mathbf{L}(0) } 
	+ t 
	}^{ - 1 }
	\geq 
	\mathbf{L}(t)
	. 
	\end{equation}
	This \proves that for all $ t \in [0,\tau) $ 
	it holds that 
	\begin{equation}
	\begin{split}
	\mathbf{L}(t)
	& \leq
	\constt 
	\,
	(
	\constt 
	[ \mathbf{L}(0) ]^{ - 1 }
	+ 
	t 
	)^{ - 1 }
	=
	\const^2
	\consttt^{
		2 - 2 \alpha 
	}
	(
	\const^2
	\consttt^{
		1 - 2 \alpha 
	}
	+ 
	t 
	)^{ - 1 }
	=
	\const^2
	\consttt^2
	(
	\const^2 \consttt
	+ 
	\consttt^{
		2 \alpha 
	}
	t 
	)^{ - 1 }
	.
	\end{split}
	\end{equation}
	The fact that for all 
	$ t \in [\tau, \infty) $ 
	it holds that $ \mathbf{L}(t) = 0 $ 
	and \cref{prop:gf:convergence:eq:tau} 
	\hence \prove 
	that for all $ t \in [0, \infty) $ 
	it holds that 
	\begin{equation}
	\label{prop:gf:convergence:eq:lossest}
	0 
	\leq
	\mathbf{L}( t ) 
	\leq 
	\const^2
	\consttt^2
	(
	\indicator{
		\{ 0 \}
	}( \consttt )
	+
	\const^2 \consttt
	+ 
	\consttt^{
		2 \alpha 
	}
	t 
	)^{ - 1 }
	.
	\end{equation}
	\Moreover[observe] \cref{eq:convergent} and 
	the assumption that $ \cL \in C( \R^{ \d }, \R ) $ 
	\prove that 
	$
	\limsup_{ t \to \infty } 
	\abs{
		\cL( \Theta_t ) - \cL( \psi ) 
	} = 0
	$. 
	Combining this with 
	\cref{prop:gf:convergence:eq:lossest} 
	\proves that 
	$ \cL( \psi ) = \cL( \cpoint ) $. 
	This and \cref{prop:gf:convergence:eq:lossest} 
	\prove that for all $ t \in [0,\infty) $ 
	it holds that
	\begin{equation}
	\label{prop:gf:convergence:eq:claim2}
	0 
	\leq
	\cL( \Theta_t ) - \cL( \psi ) 
	\leq 
	\const^2
	\consttt^2
	(
	\indicator{
		\{ 0 \}
	}( \consttt )
	+
	\const^2 \consttt
	+ 
	\consttt^{
		2 \alpha 
	}
	t 
	)^{ - 1 }
	.
	\end{equation}
	In the final step of our proof 
	of 
	\cref{prop:quantitative:item_0,prop:quantitative:item_i,prop:quantitative:item_ii,prop:quantitative:item_iii}
	we establish convergence rates for the 
	real numbers  
	$ \Pnorm2{ \Theta_t - \psi } $, 
	$ t \in [0, \infty) $. 
	\Nobs that \cref{eq:convergent}, 
	\cref{eq:Cauchy_family}, 
	and \cref{prop:gf_conv:eq:integrated} 
	\prove
	that for all $ t \in [0,\infty) $ 
	it holds that 
	\begin{equation}
	\Pnorm2{
		\Theta_t - \psi 
	}
	=
	\apnorm2{
		\Theta_t 
		- 
		\br*{
		\lim\nolimits_{ s \to \infty } 
		\Theta_s 
		}
	}
	=
	\lim\nolimits_{ s \to \infty }
	\Pnorm2{
		\Theta_t - \Theta_s 
	}
	\leq 
	\mathscr{g}_t
	\leq 
	\const 
	\rbr{ 1 - \alpha }^{ - 1 } [ \mathbf{L}( t ) ]^{ 1 - \alpha } 
	.
	\end{equation}
	This and \cref{prop:gf:convergence:eq:claim2} 
	\prove that for all $ t \in [0,\infty) $ 
	it holds that 
	\begin{equation}
	\begin{split}
	\Pnorm2{
		\Theta_t - \psi 
	}
	& \leq 
	\mathscr{g}_t
	\leq
	\const 
	\rbr{ 1 - \alpha }^{ - 1 } 
	[ 
	\cL( \Theta_t ) - \cL( \psi )
	]^{ 1 - \alpha } 
	\\ & \leq 
	\const 
	\rbr{ 1 - \alpha }^{ - 1 } 
	\br*{
	\const^2
	\consttt^2
	(
	\indicator{
		\{ 0 \}
	}( \consttt )
	+
	\const^2 \consttt
	+ 
	\consttt^{
		2 \alpha 
	}
	t 
	)^{ - 1 }
	}^{ 1 - \alpha } 
	\\ & =
	\const^{ 3 - 2 \alpha } 
	\consttt^{ 2 - 2 \alpha }
	\rbr{ 1 - \alpha }^{ - 1 } 
	(
	\indicator{
		\{ 0 \}
	}( \consttt )
	+
	\const^2 \consttt
	+ 
	\consttt^{
		2 \alpha 
	}
	t 
	)^{ \alpha - 1 }
	.
	\end{split}
	\end{equation}
	Combining this with \cref{eq:to_prove,prop:gf:convergence:eq:claim2} 
	\proves[ep] 
	\cref{prop:quantitative:item_0,prop:quantitative:item_i,prop:quantitative:item_ii,prop:quantitative:item_iii}. 
\end{aproof}
\endgroup

\cfclear 
\begingroup
\providecommand{\d}{}
\renewcommand{\d}{\defaultParamDim}
\providecommand{\f}{}
\renewcommand{\f}{\defaultLossFunction}
\providecommand{\g}{}
\renewcommand{\g}{\defaultGradientFunction}
\begin{athm}{cor}{cor:gf:conv:local}
Let 
$ \d \in \N $, 
$ \cpoint \in \R^{ \d } $, 
$ \consttt \in [0,1] $, 
$ \const, \varepsilon \in (0, \infty) $, 
$ \alpha \in (0,1) $, 
$ \Theta \in C( [0,\infty), \R^{ \d } ) $, 
$ \f \in C( \R^{ \d }, \R ) $, 
let $ \g \colon \R^{ \d } \to \R^{ \d } $ 
be $\Borel(\R^\d)$/$\Borel(\R^\d)$-measurable, 
assume 
for all $ t \in [0,\infty) $ 
that
\begin{equation}
  \f( \Theta_t ) = \f( \Theta_0 ) - \int_0^t \pnorm2{ \g( \Theta_s ) }^2 \, \diff s
	\qquad\text{and}\qquad
  \Theta_t = \Theta_0 - \int_0^t \g( \Theta_s ) \, \diff s
	, 
\end{equation}
and assume for all 
$
  \altpoint \in \R^{ \d } 
$
with 
$
  \Pnorm2{ \altpoint - \cpoint } < \varepsilon
$
that
\begin{equation} 
  \abs{ 
    \f( \altpoint ) - \f( \cpoint ) 
  }^\alpha 
  \leq 
  \const \Pnorm2{ \g( \altpoint ) } ,
\quad 
  \consttt = 
  \abs{
    \f( \Theta_0 ) - \f( \cpoint )
	},
\quad 
  \const 
  ( 1 - \alpha )^{ - 1 }
  \consttt^{ 1 - \alpha }
  +
  \Pnorm2{ \Theta_0 - \cpoint }
  < 
  \varepsilon
  ,
\end{equation}
and 
$
  \inf_{ 
    t \in 
    \cu{ 
      s \in  [0, \infty) \colon 
      \forall \, r \in [0,s] \colon 
      \Pnorm2{ \Theta_r - \cpoint } < 
      \varepsilon
    } 
  } 
  \f( \Theta_t ) \geq \f( \cpoint )
$
\cfload. 
	Then there 
	exists 
	$ \psi \in \R^\d$
	such that for all $ t \in [0,\infty) $ 
	it holds that 
	$
	\f(\psi)=\f(\cpoint)
	$,
	$
	\Pnorm2{ \Theta_t - \cpoint } < \varepsilon
	$,
	$
	0 \leq 
	\f( \Theta_t ) - \f( \psi ) 
	\leq 
	(
	1
	+ 
	\const^{
		- 2
	}
	t 
	)^{ - 1 }
	$,
	and 
	\begin{equation}
	\label{cor:eq:statement}
	\Pnorm2{
		\Theta_t - \psi 
	}
	\leq
	\int_t^{ \infty }
	\Pnorm2{ 
		\g( \Theta_s )
	}
	\,
	\diff s
	\leq 
	\const
	\rbr{ 1 - \alpha }^{ - 1 } 
	(
	1
	+ 
	\const^{
		- 2
	}
	t 
	)^{ \alpha - 1 }
	.
	\end{equation}
\end{athm}

\begin{aproof}
	\Nobs that \cref{prop:gf:conv:local} \proves that 
	there exists 
	$ \psi \in \R^\d $ 
	which satisfies that
	\begin{enumerate}[label = (\roman*)]
		\item
		\llabel{cor:quantitative:proof:item_0}
		it holds that $\f(\psi)=\f(\cpoint)$,
		\item 
		\llabel{cor:quantitative:proof:item_i}
		it holds for all $ t \in [0, \infty) $ that 
		$
		\Pnorm2{ \Theta_t - \cpoint } < \varepsilon
		$,
		\item 
		\llabel{cor:quantitative:proof:item_ii}
		it holds for all $ t \in [0,\infty) $ that
		$
		0 \leq 
		\f( \Theta_t ) - \f( \psi ) 
		\leq 
		\const^2
		\consttt^2
		(
		\indicator{
			\{ 0 \}
		}( \consttt )
		+
		\const^2 \consttt
		+ 
		\consttt^{
			2 \alpha 
		}
		t 
		)^{ - 1 }
		$,
		and 
		\item 
		\llabel{cor:quantitative:proof:item_iii}
		it holds for all $ t \in [0,\infty) $ that 
		\begin{equation}
		\begin{split}
		\Pnorm2{
			\Theta_t - \psi 
		}
		& \leq
		\int_t^{ \infty }
		\Pnorm2{ 
			\g( \Theta_s )
		}
		\,
		\diff s
		\leq
		\const 
		\rbr{ 1 - \alpha }^{ - 1 } 
		[ 
		\f( \Theta_t ) - \f( \psi )
		]^{ 1 - \alpha } 
		\\ & \leq 
		\const^{ 3 - 2 \alpha } 
		\consttt^{ 2 - 2 \alpha }
		\rbr{ 1 - \alpha }^{ - 1 } 
		(
		\indicator{
			\{ 0 \}
		}( \consttt )
		+
		\const^2 \consttt
		+ 
		\consttt^{
			2 \alpha 
		}
		t 
		)^{ \alpha - 1 }
		.
		\end{split}
		\end{equation}
	\end{enumerate}
	\Nobs that \lref{cor:quantitative:proof:item_ii} 
	and the assumption that $ \consttt \leq 1 $ \prove 
	that 
	for all $ t \in [0,\infty) $ it holds that
	\begin{equation}
	\label{eq:weak_estimate:proof:cor}
	0 
	\leq 
	\f( \Theta_t ) - \f( \psi ) 
	\leq 
	\consttt^2
	(
	\const^{ - 2 }
	\indicator{
		\{ 0 \}
	}( \consttt )
	+
	\consttt
	+ 
	\const^{ - 2 }
	\consttt^{
		2 \alpha 
	}
	t 
	)^{ - 1 }
	\leq 
	(
	1
	+ 
	\const^{ - 2 }
	t 
	)^{ - 1 }
	.
	\end{equation} 
	This and \lref{cor:quantitative:proof:item_iii} 
	\prove that 
	for all $ t \in [0,\infty) $ 
	it holds that 
	\begin{equation}
	\begin{split}
	\Pnorm2{
		\Theta_t - \psi 
	}
	& \leq
	\int_t^{ \infty }
	\Pnorm2{ 
		\g( \Theta_s )
	}
	\,
	\diff s
	\leq
	\const 
	\rbr{ 1 - \alpha }^{ - 1 } 
	[ 
	\f( \Theta_t ) - \f( \psi )
	]^{ 1 - \alpha } 
	\\ & \leq 
	\const 
	\rbr{ 1 - \alpha }^{ - 1 } 
	(
	1
	+ 
	\const^{ - 2 }
	t 
	)^{ \alpha - 1 }
	. 
	\end{split}
	\end{equation} 
	Combining this with
	\lref{cor:quantitative:proof:item_0},
	\lref{cor:quantitative:proof:item_i},
	and \cref{eq:weak_estimate:proof:cor}
	\proves[ep] \cref{cor:eq:statement}. 
\end{aproof}
\endgroup

\subsection{Abstract global convergence results for GF processes}
\label{subsection:gf:global:conv}

\cfclear 
\begingroup
\providecommand{\d}{}
\renewcommand{\d}{\defaultParamDim}
\providecommand{\f}{}
\renewcommand{\f}{\defaultLossFunction}
\providecommand{\g}{}
\renewcommand{\g}{\defaultGradientFunction}
\begin{athm}{prop}{prop:gf:global:abstract}
Let 
$ \d \in \N $, $ \Theta \in C( [0,\infty), \R^{ \d } ) $, 
let $ \f \in C( \R^{ \d }, \R ) $ be a standard\cfadd{def:KLfunction_standard} \KL\ function,
let $ \g \colon \R^{ \d } \to \R^{ \d } $ 
be $\Borel(\R^\d)$/$\Borel(\R^\d)$-measurable, 
assume for all $ t \in [0,\infty) $ 
that
\begin{equation} 
\label{prop:gf:global:eq:ass}
  \f( \Theta_t ) = 
  \f( \Theta_0 ) - \int_0^t \Pnorm2{ \g( \Theta_s ) }^2 \, \diff s
	\qquad\text{and}\qquad
  \Theta_t = \Theta_0 - \int_0^t \g( \Theta_s ) \, \diff s
  ,
\end{equation}	
and assume $\liminf_{ t \to \infty } \Pnorm2{ \Theta_t } < \infty$
\cfload.
Then there exist 
$ \cpoint \in \R^{ \d } $, $ \fC, \tau, \beta \in (0, \infty) $ 
such that for all $ t \in [ \tau, \infty) $ 
it holds that
\begin{equation} 
\label{prop:gf:global:eq:claim}
  \Pnorm2{
    \Theta_t - \cpoint
  } 
  \leq 
  \bigl( 
    1 + \fC ( t - \tau ) 
  \bigr)^{ - \beta } 
\qandq
  0
  \leq
  \f( \Theta_t ) - \f( \cpoint ) 
  \leq 
  \bigl( 
    1 + \fC ( t - \tau ) 
  \bigr)^{ - 1 } .
\end{equation}
\end{athm}
\begin{aproof}
\Nobs that \cref{prop:gf:global:eq:ass} \proves that 
$
  [0, \infty) \ni t \mapsto \f( \Theta_t ) \in \R 
$ 
is non-increasing. 
\Hence that there exists $ \bfm \in [-\infty, \infty) $ 
which satisfies
\begin{equation} 
\label{proof:gf:global:eq:defm}
  \bfm = \limsup\nolimits_{ t \to \infty } \f( \Theta_t ) 
  = 
  \liminf\nolimits_{ t \to \infty } 
  \f( \Theta_t ) 
  = 
  \inf\nolimits_{ t \in [0, \infty) } 
  \f( \Theta_t )
  .
\end{equation}
\Moreover the assumption that 
$ 
  \liminf_{ t \to  \infty } \Pnorm2{ \Theta_t } < \infty 
$ 
\proves that there exist 
$
  \cpoint \in \R^{ \d } 
$ 
and 
$
  \delta = ( \delta_n )_{ n \in \N } \colon \N \to [0, \infty) 
$ 
which satisfy 
\begin{equation} 
\label{proof:gf:global:eq:taun}
\textstyle
  \liminf_{ n \to \infty } \delta_n = \infty 
\qandq
  \limsup\nolimits_{ n \to \infty } \Pnorm2{ \Theta_{ \delta_n } - \cpoint } = 0 
  .
\end{equation} 
\Nobs that 
\cref{proof:gf:global:eq:defm}, 
\cref{proof:gf:global:eq:taun}, 
and the fact that $ \f $ is continuous \prove that 
\begin{equation} 
\label{proof:gf:global:eq:bfm}
  \f( \cpoint ) 
  = \bfm \in \R
\qandq 
  \forall \, t \in [0, \infty) \colon 
  \f( \Theta_t ) \geq \f( \cpoint ) 
  .
\end{equation}
Next let
$
  \varepsilon, \const \in (0, \infty) 
$, 
$
  \alpha \in (0, 1)
$
satisfy for all 
$
  \altpoint \in \R^{ \d } 
$ 
with 
$
  \Pnorm2{ \altpoint - \cpoint } < \varepsilon 
$ 
that
\begin{equation}
\label{eq:loja_in_proof}
  \abs{ \f( \altpoint ) - \f( \cpoint ) }^{ \alpha } 
  \leq \const \Pnorm2{ \g( \altpoint ) } 
  .
\end{equation}
\Nobs that \cref{proof:gf:global:eq:taun} and 
the fact that $ \f $ is continuous demonstrate 
that there exist $ n \in \N $, $ \consttt \in [0,1] $ 
which satisfy
\begin{equation} 
\label{proof:gf:global:eq:consttt}
  \consttt = 
  \abs{
    \f( \Theta_{ \delta_n } ) - \f( \cpoint ) 
  } 
\qandq 
  \const 
  ( 1 - \alpha )^{ - 1 }
  \consttt^{ 1 - \alpha }
  +
  \Pnorm2{ \Theta_{ \delta_n } - \cpoint }
  < 
  \varepsilon
  .
\end{equation}
Next let $ \Phi \colon [0, \infty ) \to \R^{ \d } $ 
satisfy for all $ t \in [0, \infty) $ that
\begin{equation} 
\label{proof:gf:global:eq:defphi}
  \Phi_t = \Theta_{ \delta_n + t }
  .
\end{equation}
\Nobs that \cref{prop:gf:global:eq:ass,proof:gf:global:eq:bfm,proof:gf:global:eq:defphi}
\prove that for all $ t \in [0, \infty) $
it holds that
\begin{equation}
  \f( \Phi_t ) = \f( \Phi_0 ) - \int_0^t \Pnorm2{ \g( \Phi_s ) }^2 \, \diff s ,
\;\;\;
  \Phi_t = \Phi_0 - \int_0^t \g ( \Phi_s ) \, \diff s ,
\;\;\;\text{and}\;\;\;
  \f( \Phi_t ) 
  \ge \f( \cpoint )
  .
\end{equation}
Combining this with 
\cref{eq:loja_in_proof}, 
\cref{proof:gf:global:eq:consttt}, 
\cref{proof:gf:global:eq:defphi}, 
and \cref{cor:gf:conv:local} 
(applied with 
$
  \Theta \with \Phi
$
in the notation of \cref{cor:gf:conv:local}) establishes 
that there exists 
$
  \psi \in \R^\d
$
which satisfies 
for all $ t \in [0, \infty) $ that
\begin{equation} 
\label{proof:gf:global:eq:phiest}
  0 \leq 
  \f( \Phi_t ) - \f( \psi ) 
  \leq 
  ( 1 + \const^{- 2 } t  )^{ - 1 }
,
  \Pnorm2{ \Phi_t - \psi }
  \leq
  \const \rbr{ 1 - \alpha }^{ - 1 } ( 1 + \const^{- 2 } t )^{ \alpha - 1 },
\end{equation} 
and $\f(\psi)=\f(\cpoint)$.
\Nobs that \cref{proof:gf:global:eq:defphi,proof:gf:global:eq:phiest} 
\prove for all $ t \in [ 0, \infty) $ that 
$
  0 \leq 
  \f( \Theta_{ \delta_n + t } ) - \f( \psi ) 
  \leq 
  ( 1 + \const^{- 2 } t  )^{ - 1 }
$
and 
$
  \Pnorm2{ \Theta_{ \delta_n + t } - \psi }
  \leq
  \const \rbr{ 1 - \alpha }^{ - 1 } ( 1 + \const^{- 2 } t )^{ \alpha - 1 } 
$.
\Hence for all 
$ \tau \in [ \delta_n, \infty ) $, 
$ t \in [ \tau, \infty ) $ that 
\begin{equation}
	\label{eq:prop:gf:global:abstract.1}
\begin{split}
  0 
& \leq 
  \f( \Theta_t ) - \f( \psi ) 
  \leq 
  ( 1 + \const^{ - 2 } ( t - \delta_n ) )^{ - 1 }
  =
  ( 1 + \const^{ - 2 } ( t - \tau ) + \const^{ - 2 } ( \tau - \delta_n ) )^{ - 1 }
\\ &
\leq 
  ( 1 + \const^{ - 2 } ( t - \tau ) )^{ - 1 }
\end{split}
\end{equation}
and 
\begin{equation}
\label{eq:norm_Theta_t_psi_estimate}
\begin{split}
&
  \Pnorm2{ \Theta_t - \psi }
\leq
  \const \rbr{ 1 - \alpha }^{ - 1 } ( 1 + \const^{- 2 } ( t - \delta_n ) )^{ \alpha - 1 } 
\\&= 
  \Bigl[
    \bigl[
      \const \rbr{ 1 - \alpha }^{ - 1 } 
    \bigr]^{ \frac{ 1 }{ \alpha - 1 } } 
    ( 1 + \const^{- 2 } ( t - \delta_n ) )
  \Bigr]^{ \alpha - 1 } 
\\ & =
  \biggl[
    \bigl[
      \const \rbr{ 1 - \alpha }^{ - 1 } 
    \bigr]^{ \frac{ 1 }{ \alpha - 1 } } 
    \bigl[
      1 
      +
      \const^{ - 2 } ( \tau - \delta_n ) 
    \bigr]
    + 
    \Bigl[
      \bigl[
        \const ( 1 - \alpha )^{ - 1 } 
      \bigr]^{ \frac{ 1 }{ 1 - \alpha } } 
      \const^2
    \Bigr]^{ - 1 }
    ( t - \tau ) 
  \biggr]^{ \alpha - 1 } 
  .
\end{split}
\end{equation}
Next let 
$ \scrC, \tau \in (0, \infty) $ 
satisfy 
\begin{equation}
\label{eq:def_tau_in_proof}
  \scrC 
  = 
  \max\bigl\{ 
    \fC^2 , 
    \bigl[
      \const ( 1 - \alpha )^{ - 1 } 
    \bigr]^{ \frac{ 1 }{ 1 - \alpha } } 
    \const^2
  \bigr\} 
\qandq
  \tau 
  =
  \delta_n 
  + 
  \const^2
  \bigl[
    \const \rbr{ 1 - \alpha }^{ - 1 } 
  \bigr]^{ \frac{ 1 }{ 1 - \alpha } } 
  .
\end{equation}
\Nobs that 
\cref{eq:prop:gf:global:abstract.1,eq:norm_Theta_t_psi_estimate,eq:def_tau_in_proof} 
demonstrate for all $ t \in [ \tau, \infty ) $ that
\begin{equation}
\begin{split}
  0 
  \leq
  \f( \Theta_t ) - \f( \psi ) 
& \leq 
  ( 1 + \const^{ - 2 } ( t - \tau ) )^{ - 1 }
  \leq   
  ( 1 + \scrC^{ - 1 } ( t - \tau ) )^{ - 1 }
\end{split}
\end{equation}
and 
\begin{equation}
\begin{split}
  \Pnorm2{ \Theta_t - \psi }
&\leq
  \Bigl[
    \bigl[
      \const \rbr{ 1 - \alpha }^{ - 1 } 
    \bigr]^{ \frac{ 1 }{ \alpha - 1 } } 
    \bigl[
      1 
      +
      \const^{ - 2 } ( \tau - \delta_n ) 
    \bigr]
    + 
    \scrC^{ - 1 }
    ( t - \tau ) 
  \Bigr]^{ \alpha - 1 } 
\\ & =
  \Bigl[
    \bigl[
      \const \rbr{ 1 - \alpha }^{ - 1 } 
    \bigr]^{ \frac{ 1 }{ \alpha - 1 } } 
    \bigl[
      1 
      +
      \bigl[
        \const \rbr{ 1 - \alpha }^{ - 1 } 
      \bigr]^{ \frac{ 1 }{ 1 - \alpha } } 
    \bigr]
    + 
    \scrC^{ - 1 }
    ( t - \tau ) 
  \Bigr]^{ \alpha - 1 } 
\\ & 
\leq 
  \bigl[
    1
    + 
    \scrC^{ - 1 }
    ( t - \tau ) 
  \bigr]^{ \alpha - 1 } 
  .
\end{split}
\end{equation}
\end{aproof}
\endgroup

\cfclear 
\begingroup
\providecommand{\d}{}
\renewcommand{\d}{\defaultParamDim}
\providecommand{\f}{}
\renewcommand{\f}{\defaultLossFunction}
\providecommand{\g}{}
\renewcommand{\g}{\defaultGradientFunction}
\begin{athm}{cor}{cor:gf:global:abstract}
	Let 
	$ \d \in \N $, 
	$ \Theta \in C( [0,\infty), \R^{ \d } ) $, 
	let $ \f \in C( \R^{ \d }, \R ) $ be a standard\cfadd{def:KLfunction_standard} \KL\ function,
	let $ \g \colon \R^{ \d } \to \R^{ \d } $ 
	be $\Borel(\R^\d)$/$\Borel(\R^\d)$-measurable, 
	assume for all $ t \in [0,\infty) $ 
	that
\begin{equation} 
	\label{cor:gf:global:eq:ass}
	\f( \Theta_t ) = \f( \Theta_0 ) - \int_0^t \Pnorm2{ \g( \Theta_s ) }^2 \, \diff s
	\qandq
	\Theta_t = \Theta_0 - \int_0^t \g( \Theta_s ) \, \diff s,
\end{equation}
and assume 
$
	\liminf_{ t \to \infty } \Pnorm2{ \Theta_t } < \infty 
$
\cfload.
Then there exist 
$ \cpoint \in \R^{ \d } $, $ \scrC, \beta \in (0, \infty) $ 
which satisfy for all $ t \in [0, \infty) $ 
that
\begin{equation} 
\label{cor:gf:global:eq:claim}
  \Pnorm2{ \Theta_t - \cpoint } 
  \leq \scrC ( 1 + t )^{ - \beta } 
\qandq
  0 \leq \f( \Theta_t ) - \f( \cpoint ) \leq \scrC ( 1 + t )^{ - 1 } .
\end{equation}
\end{athm}
\begin{aproof}
	\Nobs that
	\cref{prop:gf:global:abstract} demonstrates that
	there exist $\cpoint \in \R^\d$,
	$ \fC, \tau , \beta \in (0, \infty)$ which satisfy for all $t \in [ \tau , \infty )$ that
	\begin{equation} \label{proof:cor:gf:convergence:eq1}
	\Pnorm2{\Theta_t - \cpoint} \leq \rbr[\big]{ 1 + \fC ( t - \tau ) }^{- \beta} 
\qandq
	0 \leq \f( \Theta_t ) - \f( \cpoint ) 
	\leq \rbr[\big]{ 1 + \fC ( t - \tau ) }^{ - 1 } .
	\end{equation}
	In the following let $\scrC \in (0, \infty)$ satisfy
	\begin{equation} 
	\label{proof:cor:gf:conv:eq3}
	\scrC =
	\max\bigl\{
	  1 + \tau ,
	  ( 1 + \tau )^{ \beta } , 
	  \fC^{-1} , 
	  \fC^{ -\beta } , 
	  ( 1 + \tau )^{ \beta } 
	  \bigl( 
	    \sup\nolimits_{ s \in [0, \tau ] } \Pnorm2{ \Theta_s - \cpoint } 
	  \bigr),
	  ( 1 + \tau ) (\f( \Theta_0 )  - \f(\cpoint))
        \bigr\} .
	\end{equation}
	\Nobs that
	\cref{proof:cor:gf:convergence:eq1}, 
	\cref{proof:cor:gf:conv:eq3},
	and the fact that $ [0, \infty) \ni t \mapsto \f( \Theta_t ) \in \R $ 
	is non-increasing \prove for all $ t \in [0, \tau] $ that
	\begin{equation}
	  \Pnorm2{ \Theta_t - \cpoint } 
	  \le 
	  \sup\nolimits_{ s \in [0, \tau] } 
	  \Pnorm2{ \Theta_s - \cpoint } 
	  \le 
	  \scrC ( 1 + \tau )^{ - \beta } 
	  \le 
	  \scrC ( 1 + t )^{ - \beta }
	\end{equation}
	and
	\begin{equation}
	  0 \leq \f( \Theta_t ) - \f( \cpoint ) 
	  \le \f( \Theta_0 ) - \f( \cpoint ) 
	  \le \scrC ( 1 + \tau )^{ - 1 } 
	  \le \scrC ( 1 + t )^{ - 1 } . 
	\end{equation}
	\Moreover \cref{proof:cor:gf:convergence:eq1,proof:cor:gf:conv:eq3} \prove for all $t \in [ \tau , \infty ) $ that
	\begin{equation}
	\begin{split}
	\Pnorm2{\Theta_t - \cpoint } 
	&\leq 
	\bigl( 1 + \fC ( t - \tau ) \bigr)^{ - \beta } 
	= \scrC 
	\bigl( 
	  \scrC^{ \nicefrac{ 1 }{ \beta } } 
	  + 
          \scrC^{ 
            \nicefrac{ 1 }{ \beta } 
          } 
          \fC 
          ( t - \tau ) 
        \bigr)^{ - \beta } 
\\
& 
\le  
  \scrC 
  \bigl(
    \scrC^{ 
      \nicefrac{ 1 }{ \beta } 
    } 
    + 
    t - \tau 
  \bigr)^{ - \beta }
\le 
  \scrC ( 1 + t )^{ - \beta } .
\end{split}
\end{equation}
\Moreover \cref{proof:cor:gf:convergence:eq1,proof:cor:gf:conv:eq3} demonstrate 
for all $ t \in [ \tau, \infty) $ that 
\begin{equation}
  0 \leq \f( \Theta_t ) - \f( \cpoint )  
  \le \scrC \rbr[\big]{ \scrC + \fC \scrC ( t - \tau ) }^{ - 1 }
  \le \scrC \rbr[\big]{ \scrC - \tau + t }^{-1} \le \scrC ( 1 + t )^{ - 1 } .
\end{equation}
\end{aproof}
\endgroup

\cfclear 
\begingroup
\providecommand{\d}{}
\renewcommand{\d}{\defaultParamDim}
\providecommand{\f}{}
\renewcommand{\f}{\defaultLossFunction}
\providecommand{\g}{}
\renewcommand{\g}{\defaultGradientFunction}
\begin{athm}{cor}{cor:gf:global:abstract.c1}
	Let 
	$ \d \in \N $, 
	$ \Theta \in C( [0,\infty), \R^{ \d } ) $, 
	let $ \f \in C^1( \R^{ \d }, \R ) $ be a standard\cfadd{def:KLfunction_standard} \KL\ function,
	assume for all $ t \in [0,\infty) $ 
	that
\begin{equation} 
	\llabel{ass}
	\Theta_t = \Theta_0 - \int_0^t (\nabla\f)( \Theta_s ) \, \diff s,
\end{equation}
and assume 
$
	\liminf_{ t \to \infty } \Pnorm2{ \Theta_t } < \infty 
$
\cfload.
Then
	there exist 
	$ \cpoint \in \R^{ \d } $, $ \scrC, \beta \in (0, \infty) $ 
	which satisfy for all $ t \in [0, \infty) $ 
	that
	\begin{equation} 
	\llabel{claim}
		\Pnorm2{ \Theta_t - \cpoint } 
		\leq \scrC ( 1 + t )^{ - \beta } 
		,\quad
		0 \leq \f( \Theta_t ) - \f( \cpoint ) \leq \scrC ( 1 + t )^{ - 1 },
		\quad\text{and}\quad
		(\nabla\f)(\cpoint)=0.
	\end{equation} 
\end{athm}
\begin{aproof}
	\Nobs that 
		\cref{prop:gf:chainrule.2}
	\proves that for all 
		$t\in[0,\infty)$
	it holds that
	\begin{equation}
		\llabel{eq:1}
		\f(\Theta_t)
		=
		\f(\Theta_0) - \int_0^t \Pnorm2{(\nabla\f)(\Theta_s)}^2 \,\diff s
		.
	\end{equation}
		\cref{cor:gf:global:abstract}
		\hence
	establishes that there exist 
	$ \cpoint \in \R^{ \d } $, $ \scrC, \beta \in (0, \infty) $ 
	which satisfy for all $ t \in [0, \infty) $ 
	that
	\begin{equation} 
		\llabel{eq:2}
		\Pnorm2{ \Theta_t - \cpoint } 
		\leq \scrC ( 1 + t )^{ - \beta } 
		\qquad\text{and}\qquad
		0 \leq \f( \Theta_t ) - \f( \cpoint ) \leq \scrC ( 1 + t )^{ - 1 }.
	\end{equation}
		This
	\proves that
	\begin{equation}
		\limsup_{t\to\infty}\Pnorm2{\Theta_t-\cpoint}=0.
	\end{equation}
	Combining
		this
	with
		the assumption that
			$\f\in C^1(\R^\d,\R)$
	\proves that
	\begin{equation}
		\limsup_{t\to\infty}\Pnorm2{(\nabla\f)(\Theta_t)-(\nabla\f)(\cpoint)}=0.
	\end{equation}
	\Hence that
	\begin{equation}
		\llabel{eq:3}
		\limsup_{t\to\infty}\babs{\Pnorm2{(\nabla\f)(\Theta_t)}-\Pnorm2{(\nabla\f)(\cpoint)}}=0.
	\end{equation}
	\Moreover
		\lref{eq:1}
		and \lref{eq:2}
	ensure that
	\begin{equation}
		\int_0^\infty \Pnorm2{(\nabla\f)(\Theta_s)}^2 \,\diff s<\infty.
	\end{equation}
		This
		and \lref{eq:3}
	demonstrate that
	\begin{equation}
		(\nabla\f)(\cpoint)=0.
	\end{equation}
	Combining
		this
	with
		\lref{eq:2}
	establishes
	\lref{claim}.
\end{aproof}
\endgroup

\cfclear 
\begingroup
\providecommand{\d}{}
\renewcommand{\d}{\defaultParamDim}
\providecommand{\f}{}
\renewcommand{\f}{\defaultLossFunction}
\providecommand{\g}{}
\renewcommand{\g}{\defaultGradientFunction}
\begin{athm}{cor}{cor:gf:global:abstract.analytic}
	Let 
	$ \d \in \N $, 
	$ \Theta \in C( [0,\infty), \R^{ \d } ) $, 
	let
	$ \f \colon\R^\d\to\R $
	be \cfadd{def:analytic_function}analytic,
	assume for all $ t \in [0,\infty) $ 
	that
\begin{equation} 
	\llabel{ass}
	\Theta_t = \Theta_0 - \int_0^t (\nabla\f)( \Theta_s ) \, \diff s,
\end{equation}
and assume 
$
	\liminf_{ t \to \infty } \Pnorm2{ \Theta_t } < \infty 
$
\cfload.
Then there exist 
$ \cpoint \in \R^{ \d } $, $ \scrC, \beta \in (0, \infty) $ 
which satisfy for all $ t \in [0, \infty) $ 
that
\begin{equation} 
	\llabel{claim}
		\Pnorm2{ \Theta_t - \cpoint } 
		\leq \scrC ( 1 + t )^{ - \beta } 
		,\quad
		0 \leq \f( \Theta_t ) - \f( \cpoint ) \leq \scrC ( 1 + t )^{ - 1 },
		\quad\text{and}\quad
		(\nabla\f)(\cpoint)=0.
	\end{equation}
\end{athm}
\begin{aproof}
	\Nobs that 
		\cref{thm:KL}
	\proves that for all $\cpoint \in \R^\d$ there exist
	$\varepsilon, \const \in (0, \infty)$, $\alpha \in ( 0 , 1 )$ such that 
	for all $\altpoint \in \R^\d$ with $\Pnorm2{\altpoint - \cpoint } < \varepsilon$ 
	it holds that
	\begin{equation}
		\abs{\f ( \altpoint ) - \f ( \cpoint ) } ^\alpha 
		\leq 
		\const \Pnorm2 {(\nabla\f)( \altpoint ) }
		.
	\end{equation}
		\cref{cor:gf:global:abstract.c1}
		\hence
	establishes
		\lref{claim}.
\end{aproof}
\endgroup

\cfclear
\begingroup
\providecommand{\d}{}
\renewcommand{\d}{\defaultParamDim}
\providecommand{\f}{}
\renewcommand{\f}{\defaultLossFunction}
\begin{exercise}{question:descent}
	Prove or disprove the following statement: 
	For all 
		$\fd \in \N$,
		$L \in (0 , \infty )$,
		$\gamma \in [0, L^{-1} ]$,
	all open and convex sets  $U \subseteq \R^\fd$,
	and all
		$\f \in C^1 (U, \R)$,
		$ \altpoint \in U$ 
	with 
		$\altpoint - \gamma ( \nabla \f ) ( \altpoint ) \in U$ 
	and
		$\forall \, \altpointTwo, \altpointThree \in U \colon \Pnorm2{ ( \nabla \f ) ( \altpointTwo ) - ( \nabla \f ) ( \altpointThree )} \le L \Pnorm2{\altpointTwo-\altpointThree}$
	it holds that
	\begin{equation} 
	\label{question:descent:eq:claim}
		\f(\altpoint - \gamma ( \nabla \f ) ( \altpoint )) \le \f(\altpoint) - \tfrac{\gamma}{2} \Pnorm2{ ( \nabla \f ) ( \altpoint ) }^2
	\end{equation}
	\cfload.
\end{exercise}
\endgroup

\section{Convergence analysis for GD processes} 
\label{sec:convergence_GD}

In this section we employ standard \KL\ inequalities to establish convergence of the \GD\ method to critical points (see \cref{sec:critical_points}). The specific presentation of this section is closely based on \cite[Section~8]{jentzen2022existence_arxiv}.

\subsection{One-step descent property for GD processes}

\cfclear
\begingroup
\providecommand{\d}{}
\renewcommand{\d}{\defaultParamDim}
\providecommand{\f}{}
\renewcommand{\f}{\defaultLossFunction}
\begin{athm}{lemma}{lem:descent}
	Let $\d \in \N$,
	$L \in \R$,
	let $U \subseteq \R^\d$ be open and convex,
	let $\f \in C^1 (U, \R)$,
	and assume for all $\altpointTwo, \altpointThree \in U$ that 
	\begin{equation}
		\llabel{eq:Lip}
		\Pnorm2{ ( \nabla \f ) ( \altpointTwo ) - ( \nabla \f ) ( \altpointThree )} \le L \Pnorm2{\altpointTwo-\altpointThree}
	\end{equation}
	\cfload.
	Then it holds for all $\altpointTwo, \altpointThree \in U$ that
	\begin{equation}
	\f(\altpointThree) \le  \f(\altpointTwo) + \scp{ (\nabla \f ) ( \altpointTwo ) , \altpointThree - \altpointTwo } + \tfrac{L}{2} \Pnorm2{\altpointTwo-\altpointThree}^2
	\end{equation}
	\cfout.
\end{athm}
\begin{aproof}
	\Nobs that the fundamental theorem of calculus,
	the Cauchy-Schwarz inequality,
	and \lref{eq:Lip}
	\prove that for all $\altpointTwo,\altpointThree \in U$ we have that
	\begin{equation}
	\begin{split}
	&\f(\altpointThree) - \f(\altpointTwo) \\
	&= \br[\big]{ \f(\altpointTwo+r(\altpointThree-\altpointTwo)) }_{r=0}^{r=1} = \int_0^1 \scp{ (\nabla \f ) ( \altpointTwo + r ( \altpointThree - \altpointTwo ) ) , \altpointThree - \altpointTwo } \, \diff r \\
	&= \scp{ (\nabla \f ) ( \altpointTwo ) , \altpointThree - \altpointTwo } + \int_0^1 \scp{ (\nabla \f ) ( \altpointTwo + r ( \altpointThree - \altpointTwo ) ) - ( \nabla \f ) ( \altpointTwo ), \altpointThree - \altpointTwo } \, \diff r \\
	& \le \scp{ (\nabla \f ) ( \altpointTwo ) , \altpointThree - \altpointTwo } + \int_0^1 \abs*{ \scp{ (\nabla \f ) ( \altpointTwo + r ( \altpointThree - \altpointTwo ) ) - ( \nabla \f ) ( \altpointTwo ), \altpointThree - \altpointTwo } } \, \diff r \\
	& \le  \scp{ (\nabla \f ) ( \altpointTwo ) , \altpointThree - \altpointTwo } + \br*{\int_0^1 \Pnorm2{ (\nabla \f ) ( \altpointTwo + r ( \altpointThree - \altpointTwo ) ) - ( \nabla \f ) ( \altpointTwo )} \, \diff r } \Pnorm2{\altpointThree-\altpointTwo} \\
	& \le \scp{ (\nabla \f ) ( \altpointTwo ) , \altpointThree - \altpointTwo } + L \Pnorm2{\altpointThree-\altpointTwo} \br*{ \int_0^1 \Pnorm2{r(\altpointThree-\altpointTwo)} \, \diff r } \\
	&= \scp{ (\nabla \f ) ( \altpointTwo ) , \altpointThree - \altpointTwo } + \tfrac{L}{2} \Pnorm2{\altpointTwo-\altpointThree}^2
	\end{split}
	\end{equation}
	\cfload.
\end{aproof}
\endgroup

\cfclear
\begingroup
\providecommand{\d}{}
\renewcommand{\d}{\defaultParamDim}
\providecommand{\f}{}
\renewcommand{\f}{\defaultLossFunction}
\begin{athm}{cor}{cor:descent:general}
	Let $\d \in \N$,
	$L , \gamma  \in \R$,
	let $U \subseteq \R^\d$ be open and convex,
	let $\f \in C^1 (U, \R)$,
	and assume for all $\altpointTwo,\altpointThree \in U$ that 
	\begin{equation}
		\Pnorm2{ ( \nabla \f ) ( \altpointTwo ) - ( \nabla \f ) ( \altpointThree )} \le L \Pnorm2{\altpointTwo-\altpointThree}
	\end{equation}
	\cfload.
	Then it holds for all $\altpoint \in U $ with $\altpoint - \gamma ( \nabla \f ) ( \altpoint ) \in U$ that
	\begin{equation} \label{cor:descent:general:eq:claim}
	\f(\altpoint - \gamma ( \nabla \f ) ( \altpoint )) 
	\le \f ( \altpoint ) + \gamma \bpr{ \tfrac{L \gamma}{2} - 1 } \Pnorm2{( \nabla \f ) ( \altpoint ) }^2 .
	\end{equation}
\end{athm}
\begin{aproof}
	\Nobs[observe] that \cref{lem:descent} ensures that for all $\altpoint \in U $ with $\altpoint - \gamma ( \nabla \f ) ( \altpoint ) \in U$ it holds that
	\begin{equation}
	\begin{split}
	\f(\altpoint - \gamma ( \nabla \f ) ( \altpoint )) 
	&\le \f(\altpoint) + \scp{ ( \nabla \f ) ( \altpoint ) , - \gamma ( \nabla \f ) (\altpoint)} + \tfrac{L}{2} \Pnorm2{\gamma (\nabla \f ) (\altpoint )}^2 \\
	&= \f(\altpoint) - \gamma \Pnorm2{ ( \nabla \f ) ( \altpoint ) }^2 + \tfrac{L \gamma^2}{2} \Pnorm2{ ( \nabla \f ) ( \altpoint ) }^2.
	\end{split}
	\end{equation}
	This establishes \cref{cor:descent:general:eq:claim}.
\end{aproof}
\endgroup

\cfclear
\begingroup
\providecommand{\d}{}
\renewcommand{\d}{\defaultParamDim}
\providecommand{\f}{}
\renewcommand{\f}{\defaultLossFunction}
\begin{athm}{cor}{cor:descent}
	Let $\d \in \N$,
	$L \in (0 , \infty )$,
	$\gamma \in [0, L^{-1} ]$,
	let $U \subseteq \R^\d$ be open and convex,
	let $\f \in C^1 (U, \R)$,
	and assume for all $\altpointTwo,\altpointThree \in U$ that 
	\begin{equation}
		\Pnorm2{ ( \nabla \f ) ( \altpointTwo ) - ( \nabla \f ) ( \altpointThree )} \le L \Pnorm2{\altpointTwo-\altpointThree}
	\end{equation}
	\cfload.
	Then it holds for all $\altpoint \in U $ with $\altpoint - \gamma ( \nabla \f ) ( \altpoint ) \in U$ that
	\begin{equation} \label{cor:descent:eq:claim}
	\f(\altpoint - \gamma ( \nabla \f ) ( \altpoint )) \le \f(\altpoint) - \tfrac{\gamma}{2} \Pnorm2{ ( \nabla \f ) ( \altpoint ) }^2 \le \f(\altpoint).
	\end{equation}
\end{athm}
\begin{aproof}
	\Nobs that \cref{cor:descent:general},
	the fact that $\gamma \ge 0$,
	and the fact that $\frac{L \gamma}{2} - 1 \le - \frac{1}{2}$
	establish
	\cref{cor:descent:eq:claim}.
\end{aproof}
\endgroup

\begingroup
\providecommand{\d}{}
\renewcommand{\d}{\defaultParamDim}
\providecommand{\f}{}
\renewcommand{\f}{\defaultLossFunction}
\begin{exercise}{quest:Descent}
Let $(\gamma_n)_{n \in \N} \subseteq (0,\infty)$ satisfy for all 
	$n \in \N$
that
$\gamma_n = \frac{1}{n+1}$
and
let $\f \colon \R \to \R$ satisfy for all 
	$\altpoint \in \R$ 
that
\begin{equation}
\begin{split} 
	\f(\altpoint) 
=
	2\altpoint + \sin(\altpoint).
\end{split}
\end{equation}
Prove or disprove the following statement:
For every $\Theta = (\Theta_k)_{k \in \N_0} \colon \N_0 \to \R$ with
$	
	\forall \, k \in \N
\colon
	\Theta_k = \Theta_{k-1} - \gamma_k (\nabla \f)(\Theta_{k-1})
$
and every
	$n \in \N$
it holds that
\begin{equation}
\begin{split} 
	\f(\Theta_n)
\leq
	\f(\Theta_{n-1}) - \tfrac{1}{n+1}\big(1 - \tfrac{3}{2(n+1)}\big) |2 + \cos(\Theta_{n-1})|^2.
\end{split}
\end{equation}
\end{exercise}
\endgroup

\begingroup
\providecommand{\d}{}
\renewcommand{\d}{\defaultParamDim}
\providecommand{\f}{}
\renewcommand{\f}{\defaultLossFunction}
\begin{exercise}{quest:Descent2}
Let $\f \colon \R \to \R$ satisfy for all 
	$\altpoint \in \R$ 
that
\begin{equation}
\begin{split} 
	\f(\altpoint) 
=
	4 \altpoint + 3 \sin(\altpoint). 
\end{split}
\end{equation}
Prove or disprove the following statement:
For every $\Theta = (\Theta_n)_{n \in \N_0} \colon \N_0 \to \R$ with
$	
	\forall \, n \in \N
\colon
	\Theta_n = \Theta_{n-1} - \tfrac{1}{n+1} (\nabla \f)(\Theta_{n-1})
$
and every
	$k \in \N$
it holds that
\begin{equation}
\begin{split} 
	\f(\Theta_k)
<
	\f(\Theta_{k-1}).
\end{split}
\end{equation}
\end{exercise}

\endgroup

\subsection{Abstract local convergence results for GD processes}
\cfclear
\begingroup
\providecommand{\d}{}
\renewcommand{\d}{\defaultParamDim}
\providecommand{\f}{}
\renewcommand{\f}{\defaultLossFunction}
\providecommand{\g}{}
\renewcommand{\g}{\defaultGradientFunction}
\begin{athm}{prop}{prop:gd:loja}
Let $ \d \in \N $, $ \consttt \in \R $,
$ \varepsilon, L, \const \in (0, \infty) $,
	$\alpha \in (0,1)$,
	$\gamma \in (0, L^{-1}]$,
	$\cpoint \in \R^\d$,
	let $\bB \subseteq \R^\d$ satisfy $\bB = \cu{ \altpoint \in \R^\d \colon \Pnorm2{\altpoint - \cpoint } < \varepsilon}$,
	let $\f \in C ( \R^\d , \R)$ satisfy $\f |_{\bB} \in C^1 ( \bB , \R)$,
	let $\g \colon \R^\d \to \R^\d$ satisfy for all $\altpoint \in \bB$ that $\g (\altpoint ) = ( \nabla \f ) ( \altpoint )$,
	assume $\g ( \cpoint ) = 0$,
	assume for all $\altpointTwo, \altpointThree \in \bB$ that
	\begin{equation}
		\Pnorm2{ \g ( \altpointTwo ) - \g ( \altpointThree )} \le L \Pnorm2{\altpointTwo - \altpointThree}
		,
	\end{equation}
	let $\Theta \colon \N_0 \to \R^\d$ satisfy for all $n \in \N_0$ that $\Theta_{n+1} = \Theta_n - \gamma \g (\Theta_n )$,
	and assume for all $\altpoint \in \bB$ that
	\begin{equation} \label{prop:gd:eq:ass:loja}
	\abs{ 
		\f( \altpoint ) - \f( \cpoint ) 
	}^\alpha 
	\leq 
	\const \Pnorm2{ \g( \altpoint ) } ,
	\;\: 
	\consttt = 
	\abs{\f( \Theta_0 ) - \f( \cpoint )},
	\;\: 
	2 \const 
	( 1 - \alpha )^{ - 1 }
	\consttt^{ 1 - \alpha }
	+
	\Pnorm2{ \Theta_0 - \cpoint }
	< 
	\tfrac{\varepsilon}{\gamma L + 1 },
	\end{equation}
	and $\inf_{n \in \cu{m \in \N_0 \colon \forall \, k \in \N_0 \cap [0, m ] \colon \Theta_k \in \bB}} \f (\Theta_n  ) \ge \f ( \cpoint )$
	\cfload.
	Then there exists $\psi \in \f^{-1} ( \cu{\f ( \cpoint )}) \cap \g^{-1} ( \cu{0}) \cap \bB$ such that 
	\begin{enumerate} [label = (\roman*)]
		\item \label{prop:gd:item1} it holds for all $n \in \N_0$ that $\Theta_n \in \bB$,
		\item \label{prop:gd:item2} it holds for all $n \in \N_0$ that
		$
		0 \le \f ( \Theta_n ) - \f ( \psi )  
		\le 
		2 \const^2 \consttt^2 
		\rbr{ 
			\indicator{ \cu{0}} ( \consttt ) + \consttt^{2 \alpha} n \gamma + 2 \const^2 \consttt 
		}^{ - 1 } ,
		$
		and
		\item \label{prop:gd:item3} it holds for all $n \in \N_0$ that
		\begin{equation}
		\begin{split}
		\Pnorm2{\Theta_n - \psi } 
		&\le \smallsum_{k=n}^\infty \Pnorm2{\Theta_{k+1} - \Theta_k } 
		\le 2 \const \rbr{1 - \alpha}^{-1}
		\abs{\f ( \Theta_n ) - \f ( \psi ) }^{1 - \alpha} \\
		& \le
		2^{2 - \alpha} \const^{3 - 2 \alpha} \consttt^{2 - 2 \alpha }(1 - \alpha)^{-1}
		\rbr{ 
			\indicator{ \cu{0}} ( \consttt ) + \consttt^{2 \alpha} n \gamma + 2 \const^2 \consttt 
		}^{ \alpha - 1 } .
		\end{split}
		\end{equation}
	\end{enumerate}
\end{athm}

\begin{aproof}
	Throughout this proof, 
	let $T \in \N_0 \cup \cu{\infty}$ satisfy
	\begin{equation}
	T = \inf \rbr*{ \cu{n \in \N_0 \colon \Theta_n \notin \bB } \cup \cu{\infty}} 
	,
	\end{equation} 
	let $ \bL \colon \N_0 \to \R $ 
	satisfy for all $ n \in \N_0 $ that 
	$ 
	\bL( n ) = \f( \Theta_n ) - \f( \cpoint ) 
	$, 
	and let $ \tau \in \N_0 \cup \cu{ \infty } $ satisfy
	\begin{equation} \label{prop:gd:eq:deftau}
	\tau = \inf \rbr*{ \cu{n \in \N_0 \cap [0, T ) \colon \bL(n) = 0} \cup \cu{T}}.
	\end{equation}
	\Nobs that the 
	assumption that $\g ( \cpoint ) = 0$
	\proves for all $\altpoint \in \bB$ that
	\begin{equation} \label{prop:gd:eq:stepsize}
	\gamma \Pnorm2{\g ( \altpoint ) } = \gamma  \Pnorm2{\g ( \altpoint ) - \g ( \cpoint ) } \le  \gamma L \Pnorm2{\altpoint - \cpoint } .
	\end{equation}
	This,
	the fact that $\Pnorm2{\Theta_0 - \cpoint }  < \varepsilon$,
	and the fact that 
	\begin{equation}
	\Pnorm2{\Theta_1 - \cpoint } \le \Pnorm2{\Theta_1 - \Theta_0 } + \Pnorm2{\Theta_0 - \cpoint } = \gamma \Pnorm2{\g ( \Theta_0 ) } + \Pnorm2{\Theta_0 - \cpoint} \le (\gamma L + 1) \Pnorm2{\Theta_0 - \cpoint}  < \varepsilon
	\end{equation}
	ensure that $T \ge 2$.
	Next \nobs that the assumption that 
	\begin{equation}
		\inf_{n \in \cu{m \in \N_0 \colon \forall \, k \in \N_0 \cap [0, m ] \colon \Theta_k \in \bB}} \f (\Theta_n  ) 
		\ge 
		\f ( \cpoint )
	\end{equation}
	\proves for all $n \in \N_0 \cap [0, T)$ that 
	\begin{equation}
		\bL ( n ) \ge 0.
	\end{equation}
	\Moreover the fact that $\bB \subseteq \R^\d$ is open and convex, \cref{cor:descent},
	and \cref{prop:gd:eq:ass:loja} demonstrate for all $n \in \N_0 \cap [0, T - 1 )$ that
	\begin{equation} \label{prop:gd:eq:descentest}
	\begin{split}
	\bL ( n+1 ) - \bL (n) 
	&= \f ( \Theta_{n+1} ) - \f ( \Theta_n) \le - \tfrac{\gamma}{2} \Pnorm2{\g ( \Theta_n ) } ^2 
	= - \tfrac{1}{2} \Pnorm2{\g ( \Theta_n ) } \Pnorm2{\gamma \g ( \Theta_n ) } \\
	&= - \tfrac{1}{2} \Pnorm2{\g ( \Theta_n ) } \Pnorm2{\Theta_{n+1} - \Theta_n } 
	\le - (2 \const)^{-1} \abs{\f ( \Theta_n ) - \f ( \cpoint ) }^\alpha  \Pnorm2{\Theta_{n+1} - \Theta_n } \\
	& = - (2 \const)^{-1} [ \bL ( n ) ] ^{\alpha}  \Pnorm2{\Theta_{n+1} - \Theta_n } \le 0 .
	\end{split}
	\end{equation}
	\Hence that 
	\begin{equation}
		\llabel{eq:1}
		\N_0 \cap [0, T ) \ni n \mapsto \bL(n) \in [0, \infty)
	\end{equation}
	is non-increasing.
	Combining 
		this 
	with 
		\cref{prop:gd:eq:deftau} 
	\proves for all 
		$n \in \N_0 \cap [\tau , T )$ 
	that \begin{equation}
		\bL(n) = 0.
	\end{equation}
		This 
		and \cref{prop:gd:eq:descentest}
	demonstrate for all 
		$n \in \N_0 \cap [\tau , T - 1 )$ 
	that 
	\begin{equation}
		0 = \bL ( n+1 ) - \bL ( n ) \le - \tfrac{\gamma}{2} \Pnorm2{\g ( \Theta_n ) } ^2 \le 0.
	\end{equation}
	The fact that $\gamma > 0$ \hence \proves 
	for all $n \in \N_0 \cap [\tau , T - 1 )$  that $\g ( \Theta_n ) = 0$.
	\Hence for all $n \in \N_0 \cap[\tau , T)$ that 
	\begin{equation} \label{prop:gd:eq:theta:stop}
	\Theta_n  = \Theta_{\tau}.
	\end{equation}
	\Moreover \cref{prop:gd:eq:descentest,prop:gd:eq:deftau} ensure for all $n \in \N_0 \cap [0, \tau) \cap [0, T - 1 )$ that
	\begin{equation}
	\begin{split}
	\Pnorm2{\Theta_{n+1} - \Theta_n } 
	&\le \frac{2 \const ( \bL (n) - \bL (n+1))}{[\bL (n)]^\alpha} = 2 \const \int_{\bL (n+1)}^{\bL (n)} [\bL (n)]^{-\alpha} \, \diff u \\
	& \le 2 \const \int_{\bL (n+1)}^{\bL (n)} u^{-\alpha} \, \diff u = \frac{2 \const \rbr*{[\bL (n)]^{1 - \alpha} - [ \bL ( n+1) ]^{1 - \alpha}}}{1 - \alpha}.
	\end{split}
	\end{equation}
	This and \cref{prop:gd:eq:theta:stop} \prove for all $n  \in \N_0 \cap [0, T - 1 )$ that
	\begin{equation}
	\Pnorm2{\Theta_{n+1} - \Theta_n } \le \frac{2 \const \rbr*{[\bL (n)]^{1 - \alpha} - [ \bL ( n+1) ]^{1 - \alpha}}}{1 - \alpha}.
	\end{equation}
	Combining this with the triangle inequality
	\proves for all $m , n \in \N_0 \cap [0, T )$ with $m \le n$ that
	\begin{equation} \label{prop:gd:eq:triangleest}
	\begin{split}
	\Pnorm2{\Theta_n - \Theta_m} 
	& \le \sum_{k=m}^{n-1} \Pnorm2{\Theta_{k+1} - \Theta_k } \le \frac{2 \const}{1 - \alpha} \br*{ \sum_{k=m}^{n-1} \rbr*{[\bL ( k ) ]^{1 - \alpha} - [ \bL ( k + 1) ]^{1 - \alpha}} } \\
	& 
	= \frac{2 \const \rbr*{[ \bL ( m ) ]^{1 - \alpha} - [ \bL ( n ) ]^{1 - \alpha} } }{1 - \alpha}
	\le \frac{2 \const [ \bL ( m ) ]^{1 - \alpha}}{1 - \alpha}.
	\end{split}
	\end{equation}
	This and \cref{prop:gd:eq:ass:loja} demonstrate for all $n \in \N_0 \cap [0, T )$ that
	\begin{equation} \label{prop:gd:eq:dist:theta0}
	\Pnorm2{\Theta_n - \Theta_0 } \le \frac{2 \const [ \bL ( 0 )]^{1 - \alpha}}{1 - \alpha} = \frac{2 \const \abs{\f ( \Theta_0 ) - \f ( \cpoint )}^{1 - \alpha}}{1 - \alpha} = 2 \const 
	( 1 - \alpha )^{ - 1 }
	\consttt^{ 1 - \alpha } .
	\end{equation}
	Combining this with \cref{prop:gd:eq:stepsize}, \cref{prop:gd:eq:ass:loja},
	and the triangle inequality \proves for all $n \in \N_0 \cap [0, T )$ that
	\begin{equation} \label{prop:gd:eq:induct:dist}
	\begin{split}
	\Pnorm2{\Theta_{n+1} - \cpoint }
	& \le \Pnorm2{\Theta_{n+1} - \Theta_n } + \Pnorm2{\Theta_n - \cpoint} 
	= \gamma \Pnorm2{ \g ( \Theta_n ) } + \Pnorm2{\Theta_n - \cpoint } \\
	& \le ( \gamma L + 1 ) \Pnorm2{\Theta_n - \cpoint } \le( \gamma L + 1 ) (\Pnorm2{\Theta_n - \Theta_0} + \Pnorm2{\Theta_0 - \cpoint }) \\
	& \le ( \gamma L + 1 ) ( 2 \const 
	( 1 - \alpha )^{ - 1 }
	\consttt^{ 1 - \alpha } + \Pnorm2{\Theta_0 - \cpoint}) <  \varepsilon.
	\end{split}
	\end{equation}
	\Hence that 
	\begin{equation} \label{prop:gd:eq:tau:infty}
	T = \infty .
	\end{equation}
	Combining 
		this 
	with 
		\cref{prop:gd:eq:ass:loja}, 
		\cref{prop:gd:eq:triangleest},
		and \lref{eq:1}
	\proves that
	\begin{equation} \label{prop:gd:eq:traj:length}
	\sum_{k=0}^\infty \Pnorm2{\Theta_{k+1} - \Theta_k } = \lim_{n \to \infty} \br*{	\sum_{k=0}^n \Pnorm2{\Theta_{k+1} - \Theta_k } }
	\le \frac{2 \const [ \bL ( 0 )]^{1 - \alpha}}{1 - \alpha} = \frac{2 \const \consttt^{1 - \alpha}}{1 - \alpha} < \varepsilon < \infty. 
	\end{equation}
	\Hence that there exists $\psi \in \R^\d$ which satisfies
	\begin{equation} \label{prop:gd:eq:limitpsi}
	\limsup\nolimits_{n \to \infty} \Pnorm2{\Theta_n - \psi } = 0.
	\end{equation}
	\Nobs that 
		\cref{prop:gd:eq:induct:dist,prop:gd:eq:tau:infty,prop:gd:eq:limitpsi} 
	\prove that 
	\begin{equation}
		\Pnorm2{\psi - \cpoint } \le ( \gamma L + 1 ) ( 2 \const 
		( 1 - \alpha )^{ - 1 }
		\consttt^{ 1 - \alpha } + \Pnorm2{\Theta_0 - \cpoint}) <  \varepsilon .
	\end{equation}
	\Hence that 
	\begin{equation}
		\psi \in \bB.
	\end{equation}
	Next \nobs that \cref{prop:gd:eq:descentest},
	\cref{prop:gd:eq:ass:loja},
	and the fact that for all $n \in \N_0$ 
	it holds that $ \bL( n ) \le \bL ( 0 ) = \consttt $ ensure 
	that for all $ n \in \N_0 \cap [0, \tau) $ we have that
	\begin{equation} \label{prop:gd:eq:ln:descent}
	- \bL ( n ) 
	\le 
	\bL (n+1) - \bL (n) 
	\le - \tfrac{\gamma}{2} \Pnorm2{\g ( \Theta_n)}^2 \le - \tfrac{\gamma}{2 \const^2} [\bL ( n ) ]^{2 \alpha} \le -  \tfrac{\gamma}{2 \const^2 \consttt^{2 - 2 \alpha}} [ \bL ( n ) ]^{2 }.
	\end{equation}
	This \proves for all $ n \in \N_0 \cap [0, \tau) $ that 
	\begin{equation}
		0 < \bL (n) \le \frac{ 2 \const^2 \consttt^{ 2 - 2 \alpha } }{ \gamma } 
		.
	\end{equation}
	Combining this and \cref{prop:gd:eq:ln:descent} demonstrates 
	for all $ n \in \N_0 \cap [0, \tau - 1 ) $ that
	\begin{equation}
	\begin{split}
	\frac{ 1 }{ \bL(n) } - \frac{ 1 }{ \bL(n+1) } 
	& 
	\le 
	\frac{ 1 }{ 
		\bL(n) 
	} 
	- 
	\frac{ 1 }{ 
		\bL(n) 
		( 
		1 - 
		\tfrac{ 
			\gamma 
		}{ 2 \const^2 \consttt^{ 2 - 2 \alpha } 
		} 
		\bL( n ) 
		) 
	} 
	=
	\frac{
		\bpr{ 
		1 
		- 
		\frac{ \gamma }{ 2 \const^2 \consttt^{ 2 - 2 \alpha } } 
		\bL( n )
		} 
		- 1
	}{
		\bL(n) 
		\bpr{ 
		1 
		- 
		\frac{ \gamma }{ 2 \const^2 \consttt^{ 2 - 2 \alpha } } 
		\bL( n )
		}
	}
	\\ &   =
	\frac{
		- 
		\frac{ \gamma }{ 2 \const^2 \consttt^{ 2 - 2 \alpha } } 
	}{
		\bpr{ 
		1 
		- 
		\frac{ \gamma }{ 2 \const^2 \consttt^{ 2 - 2 \alpha } } 
		\bL( n )
		}
	}
	=
	- 
	\frac{ 
		1 
	}{ 
		(
		\tfrac{ 2 \const^2 \consttt^{ 2 - 2 \alpha } 
		}{ \gamma } 
		- \bL(n)
		)
	} 
	< 
	- \frac{ \gamma }{ 2 \const^2 \consttt^{ 2 - 2 \alpha } } 
	.
	\end{split}
	\end{equation}
	Therefore,
	we get for all $n \in \N_0 \cap [0, \tau )$ that
	\begin{equation}
	\frac{1}{ \bL ( n ) } 
	= \frac{1}{\bL ( 0 ) } + \sum_{k=0}^{n-1} \br*{ \frac{1}{\bL ( k+1 )} - \frac{1}{\bL ( k ) } }
	> \frac{1}{ \bL (0)} + \frac{n \gamma}{2 \const^2 \consttt^{2 - 2 \alpha} } = \frac{1}{\consttt} +  \frac{n \gamma}{2 \const^2 \consttt^{2 - 2 \alpha} }.
	\end{equation}
	\Hence for all $n \in \N_0 \cap [0, \tau )$ that $ \bL (n) < \frac{2 \const^2 \consttt^{2 - 2 \alpha} }{n \gamma + 2 \const^2 \consttt^{1 - 2 \alpha}}$. Combining this with 
	the fact that for all $n \in \N_0 \cap  [ \tau , \infty )$ it holds that $\bL (n) = 0$ \proves that for all $n \in \N_0$ we have that
	\begin{equation} \label{prop:gd:eq:ln:finalest}
	\bL (n) \le \frac{2 \const^2 \consttt^{2} }{\indicator{ \cu{0 } } ( \consttt) + \consttt ^{2 \alpha} n \gamma + 2 \const^2 \consttt} .
	\end{equation}
	This, \cref{prop:gd:eq:limitpsi}, and the assumption that $\f$ is continuous \prove that 
	\begin{equation} \label{prop:gd:eq:limit:risk}
	\f( \psi ) 
	= 
	\lim\nolimits_{n \to \infty} \f( \Theta_n ) 
	= \f ( \cpoint ) .
	\end{equation} 
	Combining this with \cref{prop:gd:eq:ln:finalest}
	\proves for all $n \in \N_0$ that
	\begin{equation} \label{prop:gd:eq:risk:rate}
	0 \le 
	\f ( \Theta_n ) - \f ( \psi ) 
	\le \frac{2 \const^2 \consttt^{2} }{\indicator{ \cu{0} } ( \consttt) + \consttt ^{2 \alpha} n \gamma + 2 \const^2 \consttt} .
	\end{equation}
	Furthermore, \nobs that 
	the fact that $ \bB \ni \altpoint \mapsto \g ( \altpoint ) \in \R^\d $ is continuous,
	the fact that $ \psi \in \bB$,
	and \cref{prop:gd:eq:limitpsi} \prove that 
	\begin{equation} \label{prop:gd:eq:limit:gradient}
	\g( \psi ) 
	=
	\lim\nolimits_{ n \to \infty } \g( \Theta_n )
	= 
	\lim\nolimits_{ n \to \infty } 
	( \gamma^{ - 1 } ( \Theta_{ n } - \Theta_{n+1} ) ) 
	= 0 .
	\end{equation}
	Next \nobs that \cref{prop:gd:eq:ln:finalest} and \cref{prop:gd:eq:triangleest} 
	ensure for all $n \in \N_0$ that
	\begin{equation}
	\begin{split}
	\Pnorm2{\Theta_n - \psi } 
	&= \lim_{m \to \infty} \Pnorm2{\Theta_n - \Theta_m } \le \sum_{k=n}^\infty \Pnorm2{\Theta_{k+1} - \Theta_k }
	\le \frac{2 \const [ \bL ( n ) ]^{1 - \alpha}}{1 - \alpha} \\
	& \le 
	\frac{ 2^{2 - \alpha} \const^{3 - 2 \alpha} \consttt^{2  - 2 \alpha } }{(1 - \alpha)
		\rbr{ \indicator{ \cu{0}} ( \consttt ) + \consttt^{2 \alpha} n \gamma + 2 \const^2 \consttt } ^{1 - \alpha } }.
	\end{split}
	\end{equation}
	Combining this with \cref{prop:gd:eq:limit:risk}, 
	\cref{prop:gd:eq:tau:infty}, \cref{prop:gd:eq:limit:gradient}, 
	and \cref{prop:gd:eq:risk:rate} establishes \cref{prop:gd:item1,prop:gd:item2,prop:gd:item3}.
\end{aproof}
\endgroup

\begingroup
\providecommand{\d}{}
\renewcommand{\d}{\defaultParamDim}
\providecommand{\f}{}
\renewcommand{\f}{\defaultLossFunction}
\providecommand{\g}{}
\renewcommand{\g}{\defaultGradientFunction}
\begin{athm}{cor}{cor:gd:loja}
	Let $\d \in \N$,
	$\consttt \in [0 , 1 ]$,
	$\varepsilon, L, \const \in (0, \infty)$,
	$\alpha \in (0,1)$,
	$\gamma \in (0, L^{-1}]$,
	$\cpoint \in \R^\d$,
	let $ \bB \subseteq \R^{ \d } $ satisfy 
	$
	  \bB = \cu{ \altpoint \in \R^\d \colon \Pnorm2{\altpoint - \cpoint } < \varepsilon } 
	$,
	let $ \f \in C( \R^{ \d }, \R) $ satisfy $ \f|_{\bB} \in C^1( \bB , \R) $,
	let $\g \colon \R^\d \to \R^\d$ satisfy for all $\altpoint \in \bB$ that 
	$ \g( \altpoint ) = ( \nabla \f ) ( \altpoint )$,
	assume for all $\altpointTwo, \altpointThree \in \bB$ that
	\begin{equation}
		\Pnorm2{ \g ( \altpointTwo ) - \g ( \altpointThree )} \le L \Pnorm2{\altpointTwo - \altpointThree}
		,
	\end{equation}
	let $ \Theta = ( \Theta_n )_{ n \in \N_0 } \colon \N_0 \to \R^{ \d } $ satisfy for all $n \in \N_0$ that 
	\begin{equation}
	\begin{split} 
	\Theta_{n+1} = \Theta_n - \gamma \g (\Theta_n ),
	\end{split}
	\end{equation}
	and assume for all  $\altpoint \in \bB$ that
	\begin{equation} \label{cor:gd:eq:ass:loja}
	\abs{ 
		\f( \altpoint ) - \f( \cpoint ) 
	}^\alpha 
	\leq 
	\const \Pnorm2{ \g( \altpoint ) } ,
	\;\: 
	\consttt = 
	\abs{\f( \Theta_0 ) - \f( \cpoint )},
	\;\: 
	2 \const 
	( 1 - \alpha )^{ - 1 }
	\consttt^{ 1 - \alpha }
	+
	\Pnorm2{ \Theta_0 - \cpoint }
	< 
	\tfrac{\varepsilon}{\gamma L + 1 },
	\end{equation}
	and $ \f( \altpoint ) \geq \f( \cpoint ) $.
	Then there exists $\psi \in \f^{-1} ( \cu{\f ( \cpoint )}) \cap \g^{-1} ( \cu{0})$ such that for all $n \in \N_0$ it holds that $\Theta_n \in \bB$, $0 \le \f ( \Theta_n ) - \f ( \psi ) \le 2 ( 2 + \const^{-2} \gamma n )^{-1}$, and
	\begin{equation}
	\Pnorm2{\Theta_n - \psi } \le \smallsum_{k=n}^\infty \Pnorm2{\Theta_{k+1} - \Theta_k } \le 
	2^{2 - \alpha} \const (1 - \alpha)^{-1} ( 2 + \const^{-2} \gamma n )^{\alpha - 1} .
	\end{equation}
\end{athm}
\begin{aproof}
	\Nobs that
	the fact that $\f (\cpoint ) = \inf_{\altpoint \in \bB} \f ( \altpoint )$
	ensures that $\g ( \cpoint ) = ( \nabla \f ) ( \cpoint ) = 0$
	and
	$\inf_{n \in \cu{m \in \N_0 \colon \forall \, k \in \N_0 \cap [0, m ] \colon \Theta_k \in \bB}} \f (\Theta_n  ) \ge \f ( \cpoint )$.
	Combining this with \cref{prop:gd:loja} \proves that there exists
	$\psi \in \f^{-1} ( \cu{\f ( \cpoint )}) \cap \g^{-1} ( \cu{0})$ such that 
	\begin{enumerate}[label = (\Roman*)]
		\item \label{cor:gd:item1} it holds for all $n \in \N_0$ that $\Theta_n \in \bB$,
		\item \label{cor:gd:item2} it holds for all $n \in \N_0$ that
		$0 \le \f ( \Theta_n ) - \f ( \psi )  \le \frac{ 2 \const^2 \consttt^{2} }{\indicator{ \cu{0}} ( \consttt ) + \consttt^{2 \alpha} n \gamma + 2 \const^2 \consttt } $,
		and
		\item \label{cor:gd:item3} it holds for all $n \in \N_0$ that
		\begin{equation}
		\begin{split}
		\Pnorm2{\Theta_n - \psi } 
		&\le \sum_{k=n}^\infty \Pnorm2{\Theta_{k+1} - \Theta_k } 
		\le \frac{2 \const \abs{\f ( \Theta_n ) - \f ( \psi ) }^{1 - \alpha}}{1 - \alpha} \\
		& \le
		\frac{ 2^{2 - \alpha} \const^{3 - 2 \alpha} \consttt^{2  - 2 \alpha } }{(1 - \alpha)
			\rbr{ \indicator{ \cu{0}} ( \consttt ) + \consttt^{2 \alpha} n \gamma + 2 \const^2 \consttt } ^{1 - \alpha } }.
		\end{split}
		\end{equation}
	\end{enumerate}
	\Nobs that \cref{cor:gd:item2} and the assumption that $\consttt \le 1$ \prove for all $n \in \N_0$ that
	\begin{equation}
	0 \le \f ( \Theta_n ) - \f ( \psi ) \le  2  \consttt^{2} \rbr*{\const^{-2} \indicator{ \cu{0}} ( \consttt ) + \const^{-2} \consttt^{2 \alpha} n \gamma + 2 \consttt }^{-1} \le 2 ( 2 + \const^{-2} \gamma n )^{-1}.
	\end{equation}
	This and \cref{cor:gd:item3} demonstrate for all $n \in \N_0$ that
	\begin{equation}
	\begin{split}
	\Pnorm2{\Theta_n - \psi } 
	\le \sum_{k=n}^\infty \Pnorm2{\Theta_{k+1} - \Theta_k } 
	\le \frac{2 \const \abs{\f ( \Theta_n ) - \f ( \psi ) }^{1 - \alpha}}{1 - \alpha} 
	\le 
	\br*{
	\frac{ 2^{2 - \alpha} \const }{1 - \alpha } 
	}
	( 2 + \const^{-2} \gamma n )^{\alpha - 1} .
	\end{split}
	\end{equation}
\end{aproof}
\endgroup

\begingroup
\providecommand{\f}{}
\renewcommand{\f}{\defaultLossFunction}
\begin{exercise}{quest:Kurdzyka}
Let $\f \in C^1(\R, \R)$ satisfy for all 
	$\altpoint \in \R$
that
\begin{equation}
\begin{split} 
	\f(\altpoint)
=
	\altpoint^4
+
	\int_0^1
		(\sin(x) - \altpoint x)^2
	\, \diff x. 
\end{split}
\end{equation}
Prove or disprove the following statement:
For every continuous
	$\Theta = (\Theta_t)_{t \in [0,\infty)} \colon [0,\infty) \to \R $
with
	$\sup_{t \in [0,\infty)} |\Theta_t| < \infty$
and
	$\forall \, t \in [0,\infty) \colon \Theta_t = \Theta_0 - \int_0^t (\nabla \f)(\Theta_s) \, \diff s$
there exists
	$\cpoint \in \R$
such that
\begin{equation}
\begin{split} 
	\limsup_{t \to \infty} |\Theta_t - \cpoint|
=
	0.
\end{split}
\end{equation}
\end{exercise}

\endgroup

\begingroup
\providecommand{\f}{}
\renewcommand{\f}{\defaultLossFunction}
\begin{exercise}{quest:Kurdzyka2}
Let $\f \in C^\infty(\R, \R)$ satisfy for all 
	$\altpoint \in \R$
that
\begin{equation}
\begin{split} 
	\f(\altpoint)
=
	\int_0^1
		(\sin(x) - \altpoint x + \altpoint^2)^2
	\diff x.
\end{split}
\end{equation}
Prove or disprove the following statement:
For every
	$\Theta \in C([0,\infty), \R)$
with
	$\sup_{t \in [0,\infty)} |\Theta_t| < \infty$
and
	$\forall \, t \in [0,\infty) \colon \Theta_t = \Theta_0 - \int_0^t (\nabla \f)(\Theta_s)\, \diff s$
there exists
	$\cpoint \in \R$,
	$\mathscr{C}, \beta \in (0,\infty)$
such that for all
	$t \in [0,\infty)$
it holds that
\begin{equation}
\begin{split} 
	|\Theta_t - \cpoint|
=
	\mathscr{C}(1+t)^{-\beta}.
\end{split}
\end{equation}
\end{exercise}
\endgroup

\section{On the analyticity of realization functions of ANNs}
\cfclear
\begingroup
\providecommand{\d}{}
\renewcommand{\d}{\defaultParamDim}
\providecommand{\f}{}
\renewcommand{\f}{f}
\providecommand{\g}{}
\renewcommand{\g}{g}
\begin{athm}{prop}{lem:composition_analytic}[Compositions of analytic functions] 
\cfadd{def:analytic_function}
Let $ l, m, n \in \N $, 
let $ U \subseteq \R^l $
and $ V \subseteq \R^m $
be open, 
let 
$
  \f \colon U \to \R^m
$
and 
$
  \g \colon V \to \R^n
$
be analytic, 
and assume 
$
  \f( U ) \subseteq V
$
\cfload.
Then 
\begin{equation} 
  U \ni u \mapsto \g( \f(u) ) \in \R^n 
\end{equation}
is analytic. 
\end{athm}
\begin{aproof}
\Nobs that 
Fa\`{a} di Bruno's formula (cf., \eg, Fraenkel \cite{frankel1978formulae})
establishes that $ \f \circ \g $ is analytic (cf.~also, \eg, 
Krantz \& Parks~\cite[Proposition~2.8]{krantz2002primer}). 
\end{aproof}
\endgroup

\cfclear
\begingroup 
\providecommand{\d}{}
\renewcommand{\d}{\defaultParamDim}
\providecommand{\H}{}
\renewcommand{\H}{f}
\providecommand{\F}{}
\renewcommand{\F}{F}
\providecommand{\A}{}
\renewcommand{\A}{A}
\providecommand{\B}{}
\renewcommand{\B}{B}
\begin{athm}{lemma}{lem:stacking_analytic}
Let
	$\d_1, \d_2, l_1, l_2 \in \N$,
for every 
	$k \in \{1, 2\}$
let
$
	\F_k
\colon 
	\R^{\d_k} \to \R^{l_k}
$
be  \cfadd{def:analytic_function}analytic,
and let $\H \colon \R^{\d_1} \times \R^{\d_2} \to \R^{l_1} \times \R^{l_2}$ satisfy for all
	$x_1 \in \R^{\d_1}$,
	$x_2 \in \R^{\d_2}$
that
\begin{equation}
\llabel{ass1}
\begin{split} 
	\H(x_1, x_2)
=
	(
		\F_1(x_1),
		\F_2(x_2)
	)
\end{split}
\end{equation}
\cfload.
Then
	$\H%
	$ is analytic.
\end{athm}

\begin{aproof}
	Throughout this proof,
	let 
$\A_1 \colon \R^{l_1} \to \R^{l_1} \times \R^{l_2}$ and 
$\A_2 \colon \R^{l_2} \to \R^{l_1} \times \R^{l_2}$
satisfy for all 
	$x_1 \in \R^{l_1}$,
	$x_2 \in \R^{l_2}$
that
\begin{equation}
	\label{stacking_diff:setting1}
	\begin{split} 
		\A_1(x_1) = (x_1, 0)
	\qandq
		\A_2(x_2) = (0, x_2)
	\end{split}
	\end{equation}
and for every 
	$k \in \{1, 2\}$ 
let $\B_k \colon \R^{l_1} \times \R^{l_2} \to \R^{l_k}$ satisfy for all 
	$x_1 \in \R^{l_1}$,
	$x_2 \in \R^{l_2}$
that
\begin{equation}
\label{stacking_diff:setting2}
\begin{split} 
	\B_k(x_1, x_2) = x_k.
\end{split}
\end{equation}
\Nobs that
	\cref{stacking_diff.it:repr} in \cref{stacking_diff}
\proves that
\begin{equation}
	\H 
	=
		\A_1 \circ \F_1 \circ \B_1 + \A_2 \circ \F_2 \circ \B_2
		.
\end{equation}
	This,
	the fact that
		$\A_1$, $\A_2$, $\F_1$, $\F_2$, $\B_1$, and $\B_2$
		are analytic,
	and \cref{lem:composition_analytic}
	establishes that $\H$ is differentiable.
\end{aproof}
\endgroup

\cfclear
\begingroup
\providecommand{\d}{}
\renewcommand{\d}{\defaultParamDim}
\providecommand{\H}{}
\renewcommand{\H}{f}
\providecommand{\F}{}
\renewcommand{\F}{F}
\providecommand{\A}{}
\renewcommand{\A}{A}
\providecommand{\B}{}
\renewcommand{\B}{B}
\begin{athm}{lemma}{param_comp_analytic}
Let
	$\d_1, \d_2, l_0, l_1, l_2 \in \N$,
for every 
	$k \in \{1, 2\}$
let
$
	\F_k 
\colon 
	\R^{\d_k} \times \R^{l_{k-1}} \to \R^{l_k}
$
be  \cfadd{def:analytic_function}analytic,
and
let
$
	\H
\colon 
	\R^{\d_1} \times \R^{\d_{2}} \times \R^{l_{0}} 
\to 
	\R^{l_2}
$
satisfy for all
	$\altpoint_1 \in \R^{\d_1}$,
	$\altpoint_2 \in \R^{\d_2}$,
	$x \in \R^{l_0}$
that
\begin{equation}
\label{param_comp_analytic:ass1}
\begin{split} 
	\H(\altpoint_1, \altpoint_2, x)
=
	\bpr{
		\F_2(\altpoint_2, \cdot) \circ 
		\F_{1}(\altpoint_{1}, \cdot)
	} 
	(x)
\end{split}
\end{equation}
\cfload.
Then 
	$\H%
	$ is analytic.
\end{athm}

\begin{aproof}
Throughout this proof,
	let
	$\A \colon \R^{\d_1} \times \R^{\d_{2}} \times \R^{l_{0}} \to \R^{\d_{2}} \times \R^{\d_1 + l_0} $ and
	$\B \colon \R^{\d_{2}} \times \R^{\d_1 + l_0} \to \R^{\d_2} \times \R^{l_1}$
satisfy for all
	$\altpoint_1 \in \R^{\d_1}$,
	$\altpoint_2 \in \R^{\d_2}$,
	$x \in \R^{l_0}$
that
\begin{equation}
\begin{split} 
	\A(\altpoint_1, \altpoint_2, x)
=
	(\altpoint_2, (\altpoint_1, x))
\qandq
	\B(\altpoint_2, (\altpoint_1, x))
=
	(
		\altpoint_2,
		\F_1(\altpoint_1, x)
	),
\end{split}
\end{equation}
\Nobs that 
	\cref{param_comp_diff.it:repr} in \cref{param_comp_diff}
\proves that
\begin{equation}
	\llabel{eq:1}
	\H=\F_2\circ\B\circ\A
	.
\end{equation}
\Moreover \cref{lem:stacking_analytic} 
(with 
$\d_1 \is \d_2$,
$\d_2 \is \d_1 + l_1$,
$l_1 \is \d_2$,
$l_2 \is l_1$,
$\F_1 \is (\R^{\d_2} \ni \altpoint_2 \mapsto \altpoint_2 \in \R^{\d_2})$,
$\F_2 \is (\R^{\d_1 + l_1} \ni (\altpoint_1, x) \mapsto \F_1(\altpoint_1, x) \in \R^{l_1})$
in the notation of \cref{lem:stacking_analytic})
\proves that $\B$ is analytic.
Combining 
\enum{
	this;
	the fact that $\A$ is analytic;
	the fact that $\F_2$ is analytic;
	\lref{eq:1}
}
with
\cref{lem:composition_analytic}
demonstrates that $\H$ is analytic.
\end{aproof}
\endgroup

\cfclear
\begingroup
\renewcommand{\d}{\mathfrak d}
\begin{athm}{cor}{lem:analyticity_ANNs}[Analyticity of realization functions of \anns]
	Let 
		$L \in \N$, 
		$l_0,l_1,\ldots,\allowbreak l_L \in \N$
	and	for every
		$k\in \{1,2,\dots,L\}$
		let 
			$\Psi_k \colon \R^{l_k} \to \R^{l_k}$
			be analytic\cfadd{def:analytic_function}
	\cfload.
	Then
		\begin{equation}
			\R^{\sum_{k=1}^{L}l_k(l_{k-1}+1)}\times\R^{l_0}\ni(\theta,x)\mapsto 
			\bpr{\RealV \theta0{l_0}{\Psi_1,\Psi_2,\dots,\Psi_L}}(x)\in\R^{l_L}
		\end{equation}
	is analytic
	\cfout.
\end{athm}
\begin{aproof}
	Throughout this proof,
	for every
		$k\in \{1,2,\dots,L\}$
		let
			$\d_k=l_k(l_{k-1}+1)$
	and for every
		$k\in \{1,2,\dots,L\}$
		let 
			$F_k\colon \R^{\d_k}\times\R^{l_{k-1}}\to\R^{l_k}$
		satisfy for all
			$\theta\in\R^{\d_k}$,
			$x\in \R^{l_{k-1}}$
		that
		\begin{equation}
			F_k(\theta,x)
			=
			\Psi_k\bpr{\Aff_{l_k,l_{k-1}}^{\theta,0}(x)}
		\end{equation}
	\cfload.
	\Nobs that
		\cref{lem:differentiability_ANNs.it:repr} in \cref{lem:differentiability_ANNs}
	\proves that for all
		$\theta_1\in\R^{\d_1}$,
		$\theta_2\in\R^{\d_2}$,
		$\ldots$,
		$\theta_L\in\R^{\d_L}$,
		$x\in\R^{l_0}$
	it holds that
	\begin{equation}
		\llabel{eq:repr}
		\bpr{\RealV{(\theta_1,\theta_2,\dots,\theta_L)}{0}{l_0}{\Psi_1,\Psi_2,\dots,\Psi_L}}(x)
		=
		(F_L(\theta_L,\cdot)\circ F_{L-1}(\theta_{L-1},\cdot)\circ\ldots\circ F_1(\theta_1,\cdot))(x)
	\end{equation}
	\cfload.
	\Nobs that
		the assumption that
			for all
				$k\in \{1,2,\dots,L\}$
			it holds that
				$\Psi_k$ is analytic,
		the fact that
			for all
				$m,n\in\N$,
				$\theta\in\R^{m(n+1)}$
			it holds that
				$\R^{m(n+1)}\times\R^n\ni (\theta,x)\mapsto \Aff^{\theta,0}_{m,n}(x)\in\R^m$
				is analytic,
		and \cref{lem:composition_analytic}
	ensure that for all
		$k\in \{1,2,\dots,L\}$
	it holds that
		$F_k$ is analytic.
		\Cref{param_comp_diff}
		and induction
		hence
	\prove that
	\begin{multline}
		\R^{\d_1}\times\R^{\d_2}\times\ldots\times\R^{\d_L}\times\R^{l_0}
		\ni
		(\theta_1,\theta_2,\dots,\theta_L,x)
		\\\mapsto
		(F_L(\theta_L,\cdot)\circ F_{L-1}(\theta_{L-1},\cdot)\circ\ldots\circ F_1(\theta_1,\cdot))(x)
		\in\R^{l_L}
	\end{multline}
	is analytic.
		This
		and \lref{eq:repr}
	\prove that
	\begin{equation}
		\R^{\sum_{k=1}^{L}l_k(l_{k-1}+1)}\times\R^{l_0}\ni(\theta,x)\mapsto \RealV \theta0{l_0}{\Psi_1,\Psi_2,\dots,\Psi_L}(x)\in\R^{l_L}
	\end{equation}
is analytic.
\end{aproof}
\endgroup

\cfclear
\begingroup
\renewcommand{\d}{\mathfrak d}
\providecommand{\loss}{def}
\renewcommand{\loss}{\mathbf{L}}
\newcommand{\Loss}{\mathscr{L}}
\renewcommand{\x}[1]{x_{#1}}
\renewcommand{\y}[1]{y_{#1}}
\begin{athm}{cor}{cor:analyticity_empirical_risk}[Analyticity of the empirical risk function]
	Let
		$L,\d\in\N\backslash\{1\}$,
		$M,l_0,l_1,\allowbreak\dots,\allowbreak l_L\in\N$,
		$\x 1,\x 2,\dots,\x M\in \R^{l_0}$,
		$\y 1,\y 2,\dots,\allowbreak \y M\in \R^{l_L}$
	satisfy
		$\d=\sum_{k=1}^{L}l_k(l_{k-1}+1)$,
	let
		$a\colon\R\to\R$
		and $\loss\colon\R^{l_L}\times\R^{l_L}\to\R$
		be \cfadd{def:analytic_function}analytic,
	let
		$\Loss\colon\R^\d\to\R$
	satisfy for all
		$\theta\in\R^\d$ that
	\begin{equation}
		\llabel{eq:defLoss}
		\Loss(\theta)
		=
		\frac1M\br*{\sum_{m=1}^{M}\loss\bpr{\bpr{\RealV \theta0{l_0}{\multdim_{ a, l_1 } , 
		\multdim_{ a, l_2 } , 
		\dots, 
		\multdim_{ a, l_{ L - 1 } } , 
		\operatorname{id}_{ \R^{ l_L }}}\!}(\x m),\y m}}
	\end{equation}
	\cfload.
	Then $\Loss$ is analytic.
\end{athm}
\begin{aproof}
	\Nobs that
		the assumption that 
			$a$ is analytic,
		\cref{lem:stacking_analytic},
		and induction
	\prove that for all
		$m\in\N$
	it holds that
		$\multdim_{a,m}$
	is analytic.
		This,
		\cref{lem:analyticity_ANNs}
		and \cref{lem:stacking_analytic}
		(applied with
			$\mathfrak d_1\is \d+l_0$,
			$\mathfrak d_2\is l_L$,
			$l_1\is l_L$,
			$l_2\is l_L$,
			$F_1\is(\R^{\d}\times\R^{l_0}\ni (\theta,x)\mapsto \bpr{\RealV \theta0{l_0}{\multdim_{ a, l_1 } , 
			\multdim_{ a, l_2 } , 
			\dots, 
			\multdim_{ a, l_{ L - 1 } } , 
			\operatorname{id}_{ \R^{ l_L } }}\!}(x)\in\R^{l_L})$,
			$F_2\is\id_{\R^{l_L}}$
		in the notation of \cref{lem:stacking_analytic})
	ensure that
	\begin{equation}
		\R^\d\times\R^{l_0}\times\R^{l_L}
		\ni
		(\theta,x,y)
		\mapsto
		\bpr{\bpr{\RealV \theta0{l_0}{\multdim_{ a, l_1 } , 
		\multdim_{ a, l_2 } , 
		\dots, 
		\multdim_{ a, l_{ L - 1 } } , 
		\operatorname{id}_{ \R^{ l_L } }}\!}(x),y}
		\in
		\R^{l_L}\times \R^{l_L}
	\end{equation}
	is analytic.
		The assumption that
			$\loss$ is differentiable
		and the chain rule
		\hence
	establish that for all
		$x\in\R^{l_0}$,
		$y\in\R^{l_L}$
	it holds that
	\begin{equation}
		\R^\d\ni\theta\mapsto
		\loss\bpr{\bpr{\RealV \theta0{l_0}{\multdim_{ a, l_1 } , 
		\multdim_{ a, l_2 } , 
		\dots, 
		\multdim_{ a, l_{ L - 1 } } , 
		\operatorname{id}_{ \R^{ l_L }}}\!}(\x m),\y m}
		\in\R
	\end{equation}
	is analytic.
		This
	\proves
		\lref{eq:defLoss}.
\end{aproof}
\endgroup

\section{Standard KL inequalities for empirical risks in the training of ANNs with analytic activation functions}
\label{sec:KL_for_ANNs}

\cfclear
\begingroup
\providecommand{\d}{}
\renewcommand{\d}{\defaultParamDim}
\providecommand{\f}{}
\renewcommand{\f}{\defaultLossFunction}
\providecommandordefault{\x}{\defaultx}
\providecommandordefault{\y}{\defaulty}
\begin{athm}{theorem}{thm:convergence_gradient_flow}[Empirical risk minimization for \anns\ 
with analytic activation functions]
\cfadd{def:analytic_function}
Let $ L, \d \in \N \backslash \{ 1 \} $, 
$ M, l_0, l_1, \dots, l_L \in \N $, 
$ \x_1, \x_2, \dots, \x_M \in \R^{ l_0 } $, 
$ \y_1, \y_2, \dots, \y_M \in \R^{ l_L } $
satisfy 
$
  \d = \sum_{ k = 1 }^L l_k ( l_{ k - 1 } + 1 )
$, 
let $ a \colon \R \to \R $ 
and 
$
  \bfL \colon \R^{ l_L } \times \R^{ l_L }
  \to \R
$
be analytic, 
let 
$ 
  \f \colon \R^{ \d } \to \R 
$
satisfy for all 
$
  \altpoint \in \R^{ \d } 
$
that 
\begin{eqsplit}
  \f( \altpoint )
  =
  \frac{ 1 }{ M }
  \br*{
  \sum_{ m = 1 }^M
  \bfL\bigl( 
    \RealVshort^{ \altpoint, l_0 }_{
      \multdim_{ a, l_1 } , 
      \multdim_{ a, l_2 } , 
      \dots, 
      \multdim_{ a, l_{ L - 1 } } , 
      \operatorname{id}_{ \R^{ l_L } }
    }\!( 
      \x_m 
    )
    ,
    \y_m
  \bigr)
  }
  ,
\end{eqsplit}
and let $ \Theta \in C( [0,\infty), \R^{ \d } ) $ 
satisfy 
\begin{equation}
\textstyle 
  \liminf_{ t \to \infty } \Pnorm2{ \Theta_t } < \infty 
\qquad 
  \text{and}
\qquad 
  \forall \, t \in [0,\infty) \colon 
  \Theta_t 
  =
  \Theta_0 
  -
  \int_0^t
  ( \nabla \f )( \Theta_s ) \,
  \diff s
\end{equation}
\cfload.
Then there exist $ \cpoint \in \R^{ \d } $, 
$ c, \beta \in (0,\infty) $ 
such that for all $ t \in (0,\infty) $ 
it holds that 
\begin{equation}
\label{eq:cor_convergence_GF}
  \Pnorm2{ \Theta_t - \cpoint }
  \leq 
  c t^{ - \beta }
,\quad 
  0 \leq 
  \f( \Theta_t ) - \f( \cpoint )
  \leq 
  c t^{ - 1 },
	\qquad\text{and}\qquad
	(\nabla\f)(\cpoint)=0
  .
\end{equation}
\end{athm}
\begin{aproof}
\Nobs that
\cref{cor:analyticity_empirical_risk} 
demonstrates that $ \f $ is analytic\cfadd{def:analytic_function} 
\cfload.
Combining this with 
\cref{cor:gf:global:abstract.analytic}
establishes \cref{eq:cor_convergence_GF}. 
\end{aproof}
\endgroup

\cfclear
\begingroup
\begin{athm}{lemma}{lem:softplus_analytic}
	Let $a\colon \R\to\R$ be the softplus\cfadd{def:softplus1} activation function
	\cfload.
	Then $a$ is \cfadd{def:analytic_function}analytic
	\cfout.
\end{athm}
\begin{aproof}
	Throughout this proof, let
		$f\colon \R\to (0,\infty)$
	satisfy for all
		$x\in\R$
	that $f(x) = 1+\exp(x)$.
	\Nobs that
		the fact that
			$\R\ni x\mapsto \exp(x)\in\R$ is analytic
	\proves[in] that
		$f$ is analytic\cfadd{def:analytic_function}
		\cfload.
	Combining
		this
		and the fact that
			$(0,\infty)\ni x\mapsto \ln(x)\in\R$
			is analytic
	with
		\cref{lem:composition_analytic}
		and \cref{eq:softplus1.def}
	\proves that
		$a$ is analytic.
\end{aproof}
\endgroup

\cfclear
\begingroup
\providecommand{\loss}{}
\renewcommand{\loss}{\mathbf{L}}
\begin{athm}{lemma}{lem:mseloss_analytic2}
	Let $d\in\N$ and let $\loss$ be the
	mean squared error\cfadd{def:mseloss} loss function
	based on $\R^d\ni x\mapsto\Pnorm2 x\in[0,\infty)$
	\cfload.
	Then $\loss$ is analytic\cfadd{def:analytic_function} \cfout.
\end{athm}
\begin{aproof}
	\Nobs that
		\cref{lem:mseloss_analytic}
	\proves that
		$\loss$ is analytic\cfadd{def:analytic_function} \cfload.
\end{aproof}
\endgroup

\cfclear
\begingroup
\providecommand{\d}{}
\renewcommand{\d}{\defaultParamDim}
\providecommand{\f}{}
\renewcommand{\f}{\defaultLossFunction}
\providecommandordefault{\x}{\defaultx}
\providecommandordefault{\y}{\defaulty}

\begin{athm}{cor}{ex:softplus}[Empirical risk minimization for \anns\ 
with softplus activation]
Let 
$ L, \d \in \N \backslash \{ 1 \} $, 
$ M, l_0, l_1, \dots, l_L \in \N $, 
$ \x_1, \x_2, \dots, \x_M \in \R^{ l_0 } $, 
$ \y_1, \y_2, \dots, \y_M \in \R^{ l_L } $
satisfy 
$
  \d = \sum_{ k = 1 }^L l_k ( l_{ k - 1 } + 1 )
$, 
let $\activation$ be the \softplusfunc{},
let 
$ 
  \f \colon \R^{ \d } \to \R 
$
satisfy for all 
$
  \altpoint \in \R^{ \d } 
$
that 
\begin{eqsplit}
  \f( \altpoint )
  =
  \frac{ 1 }{ M }
  \br*{
  \sum_{ m = 1 }^M
    \Pnorm[\big]2{
      \y_m
      -
      \RealVshort^{ \altpoint, l_0 }_{
        \multdim_{ a, l_1 } , 
        \multdim_{ a, l_2 } , 
        \dots, 
        \multdim_{ a, l_{ L - 1 } } , 
        \operatorname{id}_{ \R^{ l_L } }
      }\!( 
        \x_m 
      )
    }^2
  }
  ,
\end{eqsplit}
and let $ \Theta \in C( [0,\infty), \R^{ \d } ) $ 
satisfy 
\begin{equation}
\label{convergence_GF_softplus:ass1}
\textstyle 
  \liminf_{ t \to \infty } \Pnorm2{ \Theta_t } < \infty 
\qquad 
  \text{and}
\qquad 
  \forall \, t \in [0,\infty) \colon 
  \Theta_t 
  =
  \Theta_0 
  -
  \int_0^t
  ( \nabla \f )( \Theta_s ) \,
  \diff s
\end{equation}
\cfload.
Then there exist $ \cpoint \in \R^{ \d } $, 
$ c, \beta \in (0,\infty) $ 
such that for all $ t \in (0,\infty) $ 
it holds that 
\begin{equation}
\label{eq:convergence_GF_softplus}
\Pnorm2{ \Theta_t - \cpoint }
\leq 
c t^{ - \beta }
,\quad 
0 \leq 
\f( \Theta_t ) - \f( \cpoint )
\leq 
c t^{ - 1 }
\quad\text{and}\quad
(\nabla\f)(\cpoint)=0
.
\end{equation}
\end{athm}
\begin{aproof}
	\Nobs that
		\cref{lem:softplus_analytic},
		\cref{lem:mseloss_analytic2},
		and \cref{thm:convergence_gradient_flow} 
	\prove
		\cref{eq:convergence_GF_softplus}. 
\end{aproof}

\begin{athm}{remark}{suboptimal_points}[Convergence to a good suboptimal critical point whose risk value is close to the optimal risk value]
\cref{ex:softplus} establishes convergence of a non-divergent \GF\ trajectory in the training of fully-connected feedforward \anns\ to a critical point $\cpoint \in \R^\d$ of the objective function.
In several scenarios in the training of \anns\ such limiting critical points seem to be with high probability not global minimum points but suboptimal critical points at which the value of the objective function is, however, not far away from the minimal value of the objective function (cf.\ Ibragimov et al.~\cite{Ibragimov2022} and also \cite{Gentile2022,Welper2023}).
In view of this, there has been an increased interest in landscape analyses associated to the objective function
to gather more information on critical points of the objective function (cf., \eg,  \cite{Zhang2021,Zhang2021a,Cheridito2022b,Ibragimov2022,Ibragimov2022a,Fukumizu2000,Baldi1989,Kawaguchi2016,SaraoMannelli2020,Safran2016,Choromanska2015,Choromanska2015a,Soudry2017,Dauphin2014,Safran2018,Du2018a,Venturi2019,Soudry2016,Soltanolkotabi2019,Nguyen2017} and the references therein).

In general in most cases it remains an open problem to rigorously prove that the value of the objective function at the limiting critical point is indeed with high probability close to the minimal/infimal value\footnote{
It is of interest to note that it seems to strongly depend on 
	the activation function, 
	the architecture of the \ann, 
	and 
	the underlying probability distribution of the data of the considered learning problem
whether the infimal value of the objective function is also a minimal value of the objective function or whether there exists no minimal value of the objective function (cf., \eg, \cite{Gallon2022,Dereich2023} and \cref{Esistence_of_min} below).
} of the objective function  and thereby establishing a full convergence analysis.
However, in the so-called overparametrized regime where there are much more \ann\ parameters than input-output training data pairs, several convergence analyses for the training of \anns\ have been achieved (cf., \eg, \cite{Du2019,Chizat2018,Chizat2019,Jacot2018} and the references therein).
\end{athm}

\begin{athm}{remark}{GD_non_saddle}[Almost surely excluding strict saddle points]
We also note that 
in several situations 
it has been shown that the limiting critical point of the considered \GF\ trajectory with random initialization or 
of the considered \GD\ process with random initialization 
is almost surely not a saddle points 
but a local minimizers;
cf., \eg, \cite{Lee2016,Lee2019,Panageas2016,Panageas2019,Cheridito2022}.
\end{athm}

\begin{athm}{remark}{Esistence_of_min}[A priori bounds and existence of minimizers]
Under the assumption that the considered \GF\ trajectory is non-divergent in the sense that
\begin{equation}
\label{Esistence_of_min:eq1}
\begin{split} 
	 \liminf_{ t \to \infty } \Pnorm2{ \Theta_t } < \infty 
\end{split}
\end{equation}
(see \cref{convergence_GF_softplus:ass1} above) we have that \cref{ex:softplus} establishes convergence of a \GF\ trajectory in the training of fully-connected feedforward \anns\ to a critical point $\cpoint \in \R^\d$ of the objective function (see \cref{eq:convergence_GF_softplus} above).
Such kind of non-divergence and slightly stronger boundedness assumptions, respectively, are very common hypotheses in convergence results for gradient based optimization methods in the training of \anns\
(cf., \eg, 
	\cite{Dereich2021,Bolte2006,Eberle2023,JentzenRiekert2022Existence,Dereich2022,Absil2005,Attouch2009,Tadic2015}, 
	\cref{subsection:gf:global:conv}, 
	and \cref{thm:convergence_gradient_flow} in the context of the \KL\ approach 
and 
	\cite{Davis2020,Jentzen2023,Dereich2022,Mertikopoulos2020}
	in the context of other approaches).
	
In most scenarios in the training of \anns\ it remains an open problem to prove or disprove such non-divergence and boundedness assumptions.
In Gallon et al.~\cite{Gallon2022} the condition in \cref{Esistence_of_min:eq1} has been disproved and divergence of \GF\ trajectories in the training of shallow fully-connected feedforward \anns\ has been established for specific target functions; see also Petersen et al.~\cite{Petersen2021}.

The question of non-divergence of gradient based optimization methods seems to be closely related to the question whether there exist minimizers in the optimization landscape of the objective function.
We refer to \cite{JentzenRiekert2022Existence,Dereich2023,Dereich2023a,Kainen2003} for results proving the existence of minimizers in optimization landscapes for the training of \anns\ and 
we refer to \cite{Gallon2022,Petersen2021} for results disproving the existence of minimizers in optimization landscapes for the training of \anns.
We also refer to, \eg, \cite{Eberle2022,Ibragimov2022} for strongly simplified \ann\ training scenarios where non-divergence and boundedness conditions of the form \cref{Esistence_of_min:eq1} have been established.
\end{athm}

\endgroup

\section{Generalized KL-inequalities}

In this section we present and study suitable generalized gradients (Fr\'echet subgradients and limiting Fr\'echet subgradients) and we briefly present generalized \KL\ inequalities that are based on such generalized gradients. The specific presentation of this section is based on \cite[Section~3.8]{jentzen2022existence_arxiv}.

\subsection{Fr\'{e}chet subgradients and limiting Fr\'{e}chet subgradients}
\label{sec:subdifferential}

\cfclear
\begingroup
\providecommand{\d}{}
\renewcommand{\d}{\defaultParamDim}
\providecommand{\f}{}
\renewcommand{\f}{\defaultLossFunction}
\begin{adef}{def:limit:subdiff}[Fr\'{e}chet subgradients and limiting Fr\'{e}chet subgradients]
\cfconsiderloaded{def:limit:subdiff}
Let $ \d \in \N $, $ \f \in C( \R^\d, \R) $, $ \altpoint \in \R^\d $.
Then we denote by 
	$(\FrechetSubdiff \f)(\altpoint) \subseteq \R^\d $ 
the set given by
\begin{equation}
\label{def:subdiff:eq}
  ( \FrechetSubdiff \f)( \altpoint )
  = 
  \cu*{ 
    \altpointTwo \in \R^\d \colon 
    \br*{
      \liminf_{\R^\d \backslash \cu{  0 } \ni h \to 0 } 
      \rbr*{ \frac{\f(\altpoint + h ) - \f ( \altpoint ) - \scp{\altpointTwo , h } }{ \Pnorm2{h}} } \geq 0  
    }
  }, 
\end{equation}
we call $( \FrechetSubdiff \f)( \altpoint )$ the set of Fr\'{e}chet subgradients of $f$ at $\altpoint$,
we denote by 
$
  (\limitingFrechetSubdiff \f)(\altpoint) \subseteq \R^\d 
$ 
the set given by
\begin{equation}
\label{def:limit:subdiff:eq}
  (\limitingFrechetSubdiff \f)( \altpoint ) =
  \textstyle\bigcap_{ \varepsilon \in (0, \infty) } 
  \overline{
    \br*{
      \bigcup_{ 
        \altpointTwo \in 
        \cu{ 
          z \in \R^\d \colon \Pnorm2{ \altpoint - z } < \varepsilon 
        }
      } 
      ( \FrechetSubdiff \f )( \altpointTwo ) 
    } 
  }
  ,
\end{equation}
and we call $(\limitingFrechetSubdiff \f)( \altpoint )$ the set of limiting Fr\'{e}chet subgradients of $f$ at $\altpoint$
\cfload.
\end{adef}
\endgroup

\cfclear
\begingroup
\providecommand{\d}{}
\renewcommand{\d}{\defaultParamDim}
\providecommand{\f}{}
\renewcommand{\f}{\defaultLossFunction}
\begin{athm}{lemma}{lem:convex_differentials}[Convex differentials]
Let $ \d \in \N $, 
$ \f \in C( \R^\d, \R ) $, 
$ \altpoint, a \in \R^\d $, 
$ b \in \R $, 
$ \varepsilon \in (0,\infty) $
and let 
$
  A \colon \R^\d \to \R
$
satisfy for all 
$ \altpointTwo \in \{ \altpointThree \in \R^\d \colon \Pnorm2{ \altpointThree - \altpoint } < \varepsilon \} $
that 
\begin{eqsplit}
\label{eq:A_definition_local_affine_linear}
  A(\altpointTwo) = \scp{ a, \altpointTwo } + b 
  \leq \f(\altpointTwo)
\qquad\text{and}\qquad 
  A(\altpoint) = \f(\altpoint)
\end{eqsplit}
\cfload.
Then 
\begin{enumerate}[label = (\roman*)]
\item 
\label{item:convex_differentials_item_i}
it holds for all 
$ \altpointTwo \in \{ \altpointThree \in \R^\d \colon \Pnorm2{ \altpointThree - \altpoint } < \varepsilon \} $
that 
$
  A(\altpointTwo) = \scp{ a, \altpointTwo - \altpoint } + \f( \altpoint )
$
and 
\item 
\label{item:convex_differentials_item_ii}
it holds that
$
  a \in ( \FrechetSubdiff \f )( \altpoint )
$
\end{enumerate}
\cfload.
\end{athm}
\begin{aproof}
\Nobs that 
\cref{eq:A_definition_local_affine_linear} 
\proves 
for all 
$ \altpointTwo \in \{ \altpointThree \in \R^\d \colon \Pnorm2{ \altpointThree - \altpoint } < \varepsilon \} $
that 
\begin{eqsplit}
  A(\altpointTwo) 
& 
  = 
  [ A(\altpointTwo) - A(\altpoint) ] + A(\altpoint)
  =
  [ ( \scp{ a, \altpointTwo } + b ) - ( \scp{ a, \altpoint } + b ) ] + A(\altpoint)
\\ & =
  \scp{ a, \altpointTwo - \altpoint } + A(\altpoint)
  =
  \scp{ a, \altpointTwo - \altpoint } + \f(\altpoint)
  .
\end{eqsplit}
This establishes 
\cref{item:convex_differentials_item_i}. 
\Nobs that 
\cref{eq:A_definition_local_affine_linear}
and 
\cref{item:convex_differentials_item_i} 
ensure
for all 
$ h \in \{ \altpointThree \in \R^\d \colon 0 < \Pnorm2{ \altpointThree } < \varepsilon \} $
that 
\begin{eqsplit}
  \frac{
    \f( \altpoint + h ) - \f(\altpoint) - \scp{ a, h } 
  }{
    \Pnorm2{ h }
  }
=
  \frac{
    \f( \altpoint + h ) - A( \altpoint + h )
  }{
    \Pnorm2{ h }
  }
\geq 
  0
  .
\end{eqsplit}
This and \cref{def:subdiff:eq} 
establish \cref{item:convex_differentials_item_ii}. 
\end{aproof}
\endgroup

\cfclear
\begingroup
\providecommand{\d}{}
\renewcommand{\d}{\defaultParamDim}
\providecommand{\f}{}
\renewcommand{\f}{\defaultLossFunction}
\begin{athm}{lemma}{lem:subdifferential:c1}[Properties of Fr\'{e}chet subgradients]
\cfadd{def:limit:subdiff}
Let $ \d \in \N $, $ \f \in C( \R^\d , \R ) $. 
Then 
\begin{enumerate}[label = (\roman*)]
\item 
\label{item:subdiff_item_i}
it holds for all $ \altpoint \in \R^\d $ that
\begin{multline} 
\label{lem:subdiff:eq} 
\cfadd{def:limit:subdiff}
  ( \limitingFrechetSubdiff \f )( \altpoint ) 
  = 
  \bigl\{ 
    \altpointTwo \in \R^\d \colon 
    \bigl[
      \exists \, z = (z_1, z_2) \colon \N \to \R^\d \times \R^\d \colon 
      \bigl( 
        \bigl[
          \forall \, k \in \N \colon\\ z_2( k ) \in ( \FrechetSubdiff \f )( z_1(k) ) 
        \bigr]
        \wedge
          \br[\big]{
            \limsup\nolimits_{ k \to \infty } 
            ( 
              \Pnorm2{ z_1(k) - \altpoint } + \Pnorm2{ z_2(k) - \altpointTwo } 
            ) 
            = 0
          } 
      \bigr) 
    \bigr]
  \bigr\}
  ,
\end{multline}
\item 
\label{item:subdiff_item_ii}
it holds for all $ \altpoint \in \R^\d $ that
$ 
  ( \FrechetSubdiff \f)( \altpoint )
  \subseteq 
  ( \limitingFrechetSubdiff \f)( \altpoint )
$,
\item 
\label{item:subdiff_item_iii}
it holds for all 
$ \altpoint \in \{ \altpointTwo \in \R^\d \colon \f \text{ is differentiable at } \altpointTwo \} $
that
$
  ( \FrechetSubdiff \f )( \altpoint ) 
  =
  \cu{ ( \nabla \f )( \altpoint ) }  
$,
\item 
\label{item:subdiff_item_iv}
it holds for all 
$
  \altpoint \in \bigcup_{ U \subseteq \R^\d, \, \text{$U$ is open}, \, \f|_U \in C^1( U , \R) } U $
that
$ ( \limitingFrechetSubdiff \f )( \altpoint ) = \cu{ ( \nabla \f ) ( \altpoint ) } $,
and
\item 
\label{item:subdiff_item_v}
it holds for all $ \altpoint \in \R^\d $ that 
$
  ( \limitingFrechetSubdiff \f )( \altpoint ) 
$
is closed. 
\end{enumerate}
\cfout. 
\end{athm}
\begin{aproof}
Throughout this proof, 
for every
	$\altpoint,\altpointTwo\in\R^\d$
let 
$
  Z^{ \altpoint, \altpointTwo } = ( Z^{ \altpoint, \altpointTwo }_1,\allowbreak Z^{ \altpoint, \altpointTwo }_2 ) 
  \colon
  \N \to \R^\d \times \R^\d
$
satisfy for all $ k \in \N $ 
that 
\begin{equation}
\label{eq:def_Z_process}
  Z^{ \altpoint, \altpointTwo }_1( k ) = \altpoint
\qquad\text{and}\qquad
  Z^{ \altpoint, \altpointTwo }_2( k ) = \altpointTwo
  .
\end{equation}
\Nobs that \cref{def:limit:subdiff:eq} \proves that 
for all 
$ \altpoint \in \R^\d $, 
$ \altpointTwo \in ( \limitingFrechetSubdiff \f)( \altpoint ) $, 
$ \varepsilon \in (0,\infty) $
it holds that 
\begin{eqsplit}
	\textstyle
  \altpointTwo \in 
  \overline{
    \bbr{
      \bigcup_{ 
        \altpointThree \in 
        \cu{ 
          \altpointFour \in \R^n \colon \Pnorm2{ \altpoint - \altpointFour } < \varepsilon 
        }
      } 
      ( \FrechetSubdiff \f )( \altpointThree ) 
    } 
  }
  .
\end{eqsplit}
This \proves that  
for all 
$ \altpoint \in \R^\d $, 
$ \altpointTwo \in ( \limitingFrechetSubdiff \f)( \altpoint ) $
and all 
$ \varepsilon, \delta \in (0,\infty) $
there exists  
$ 
  \AltpointTwo \in 
  \bpr{
    \bigcup_{ 
      \altpointThree \in 
      \cu{ 
        \altpointFour \in \R^\d \colon \Pnorm2{ \altpoint - \altpointFour } < \varepsilon 
      }
    } 
    ( \FrechetSubdiff \f )( \altpointThree ) 
	}
$
such that 
\begin{eqsplit}
  \Pnorm2{ \altpointTwo - \AltpointTwo } 
  < \delta 
  .
\end{eqsplit}
\Hence that
for all 
$ \altpoint \in \R^\d $, 
$ \altpointTwo \in ( \limitingFrechetSubdiff \f)( \altpoint ) $, 
$ \varepsilon, \delta \in (0,\infty) $
there exist  
$
  \altpointThree \in 
  \cu{ 
    \altpointFour \in \R^\d \colon \Pnorm2{ \altpoint - \altpointFour } < \varepsilon 
  }
$, 
$ 
  \AltpointTwo \in 
  ( \FrechetSubdiff \f )( \altpointThree ) 
$
such that 
$
  \Pnorm2{ \altpointTwo - \AltpointTwo } 
  < \delta 
$. 
This \proves 
that 
for all 
$ \altpoint \in \R^\d $, 
$ \altpointTwo \in ( \limitingFrechetSubdiff \f)( \altpoint ) $, 
$ \varepsilon, \delta \in (0,\infty) $
there exist  
$
  \Altpoint \in \R^\d 
$, 
$ 
  \AltpointTwo \in 
  ( \FrechetSubdiff \f )( \Altpoint ) 
$
such that 
\begin{eqsplit}
  \Pnorm2{ \altpoint - \Altpoint } < \varepsilon 
\qquad\text{and}\qquad
  \Pnorm2{ \altpointTwo - \AltpointTwo } 
  < \delta 
  .
\end{eqsplit}
\Hence  that
for all 
$ \altpoint \in \R^\d $, 
$ \altpointTwo \in ( \limitingFrechetSubdiff \f)( \altpoint ) $, 
$ k \in \N $
there exist  
$
  z_1, z_2 \in \R^\d 
$
such that 
\begin{eqsplit}
\label{eq:in_proof_item_i_direction_A}
  z_2 \in 
  ( \FrechetSubdiff \f )( z_1 ) 
\qquad\text{and}\qquad
  \Pnorm2{ z_1 - \altpoint } 
  +
  \Pnorm2{ z_2 - \altpointTwo } 
  < \tfrac{ 1 }{ k }
  .
\end{eqsplit}
\Moreover for all 
$ \altpoint, \altpointTwo \in \R^\d $, 
$ \varepsilon \in (0,\infty) $
and all 
$
  z = ( z_1, z_2 ) \colon \N \to \R^\d \times \R^\d
$
with 
$ 
  \limsup_{ k \to \infty }
  (
    \Pnorm2{ z_1(k) - \altpoint }
    +
    \Pnorm2{ z_2(k) - \altpointTwo }
  )
  = 0
$
and 
$
  \forall \, k \in \N \colon z_2(k) \in ( \cD \f )( z_1(k) )
$
there exist 
$
  \Altpoint, \AltpointTwo \in \R^\d
$
such that 
\begin{eqsplit}
  \AltpointTwo \in ( \cD \f )( \Altpoint )
\qquad\text{and}\qquad
  \Pnorm2{ \Altpoint - \altpoint }
  +
  \Pnorm2{ \AltpointTwo - \altpointTwo }
  < 
  \varepsilon
  .
\end{eqsplit}
\Hence  that
for all 
$ \altpoint, \altpointTwo \in \R^\d $, 
$ \varepsilon, \delta \in (0,\infty) $
and all 
$
  z = ( z_1, z_2 ) \colon \N \to \R^\d \times \R^\d
$
with 
$ 
  \limsup_{ k \to \infty }
  (
    \Pnorm2{ z_1(k) - \altpoint }
    +
    \Pnorm2{ z_2(k) - \altpointTwo }
  )
  = 0
$
and 
$
  \forall \, k \in \N \colon z_2(k) \in ( \cD \f )( z_1(k) )
$
there exist 
$
  \Altpoint, \AltpointTwo \in \R^\d
$
such that 
\begin{eqsplit}
  \AltpointTwo \in ( \cD \f )( \Altpoint )
  ,
\qquad
  \Pnorm2{ \altpoint - \Altpoint }
  < 
  \varepsilon ,
\qquad\text{and}\qquad
  \Pnorm2{ \altpointTwo - \AltpointTwo }
  < 
  \delta
  .
\end{eqsplit}
This \proves that 
for all 
$ \altpoint, \altpointTwo \in \R^\d $, 
$ \varepsilon, \delta \in (0,\infty) $
and all 
$
  z = ( z_1, z_2 ) \colon \N \to \R^\d \times \R^\d
$
with 
$ 
  \limsup_{ k \to \infty }
  (
    \Pnorm2{ z_1(k) - \altpoint }
    +
    \Pnorm2{ z_2(k) - \altpointTwo }
  )
  = 0
$
and 
$
  \forall \, k \in \N \colon z_2(k) \in ( \cD \f )( z_1(k) )
$
there exist 
$
  \altpointThree \in \{ 
    \altpointFour \in \R^\d \colon \Pnorm2{ \altpoint - \altpointFour } < \varepsilon
  \} 
$, 
$
  \AltpointTwo \in ( \cD \f )( \altpointThree )
$
such that 
$
  \Pnorm2{ \altpointTwo - \AltpointTwo }
  < 
  \delta
$. 
\Hence that
for all 
$ \altpoint, \altpointTwo \in \R^\d $, 
$ \varepsilon, \delta \in (0,\infty) $
and all 
$
  z = ( z_1, z_2 ) \colon \N \to \R^\d \times \R^\d
$
with 
$ 
  \limsup_{ k \to \infty }
  (
    \Pnorm2{ z_1(k) - \altpoint }
    +
    \Pnorm2{ z_2(k) - \altpointTwo }
  )
  = 0
$
and 
$
  \forall \, k \in \N \colon z_2(k) \in ( \cD \f )( z_1(k) )
$
there exists  
$
  \AltpointTwo \in 
  \bbr{
    \bigcup_{
      \altpointThree 
      \in 
      \{ 
        \altpointFour \in \R^\d \colon \Pnorm2{ \altpoint - \altpointFour } < \varepsilon
      \} 
    }
    ( \cD \f )( \altpointThree )
	}
$
such that 
\begin{eqsplit}
  \Pnorm2{ \altpointTwo - \AltpointTwo }
  < 
  \delta
  .
\end{eqsplit}
This \proves that 
for all 
$ \altpoint, \altpointTwo \in \R^\d $, 
$ \varepsilon \in (0,\infty) $
and all 
$
  z = ( z_1, z_2 ) \colon \N \to \R^\d \times \R^\d
$
with 
$ 
  \limsup_{ k \to \infty }
  (
    \Pnorm2{ z_1(k) - \altpoint }
    +
    \Pnorm2{ z_2(k) - \altpointTwo }
  )
  = 0
$
and 
$
  \forall \, k \in \N \colon z_2(k) \in ( \cD \f )( z_1(k) )
$
it holds that 
\begin{eqsplit}
	\textstyle
  \altpointTwo \in 
  \overline{
    \bbr{
      \bigcup_{
        \altpointThree 
        \in 
        \{ 
          \altpointFour \in \R^n \colon \Pnorm2{ \altpoint - \altpointFour } < \varepsilon
        \} 
      }
      ( \cD \f )( \altpointThree )
		}
  }
  .
\end{eqsplit}
This and 
\cref{def:limit:subdiff:eq} 
\prove that 
for all 
$ \altpoint, \altpointTwo \in \R^\d $ 
and all 
$
  z = ( z_1, z_2 ) \colon \N \to \R^\d \times \R^\d
$
with 
$ 
  \limsup_{ k \to \infty }
  (
    \Pnorm2{ z_1(k) - \altpoint }
    +
    \Pnorm2{ z_2(k) - \altpointTwo }
  )
  = 0
$
and 
$
  \forall \, k \in \N \colon z_2(k) \in ( \cD \f )( z_1(k) )
$
it holds that 
\begin{eqsplit}
\label{eq:in_proof_item_i_direction_B}
  \altpointTwo \in ( \limitingFrechetSubdiff \f )( \altpoint )
  .
\end{eqsplit}
Combining this with 
\cref{eq:in_proof_item_i_direction_A}
\proves[ep] \cref{item:subdiff_item_i}. 
\Nobs that \cref{eq:def_Z_process} \proves that 
for all $ \altpoint \in \R^\d $, $ \altpointTwo \in ( \FrechetSubdiff \f )( \altpoint ) $ 
it holds that
\begin{equation}
  \biggl[
    \forall \, k \in \N \colon 
    \Bigl(
      Z^{ \altpoint, \altpointTwo }_2( k ) 
      \in ( \FrechetSubdiff \f )( Z^{ \altpoint, \altpointTwo }_1( k ) )
    \Bigr)
  \biggr]
  \wedge
  \biggl[
    \limsup_{ k \to \infty }
    \bigl(
      \Pnorm2{ Z^{ \altpoint, \altpointTwo }_1( k ) - \altpoint }
      +
      \Pnorm2{ Z^{ \altpoint, \altpointTwo }_2( k ) - \altpointTwo }
    \bigr)
    = 0
  \biggr]
\end{equation}
\cfload.
Combining this with \cref{item:subdiff_item_i}
\proves[ep] \cref{item:subdiff_item_ii}.
\Nobs that the fact that for all 
$ a \in \R $ it holds that 
$
  - a \leq \abs{ a }
$
demonstrates that
for all 
$ \altpoint \in \{ \altpointTwo \in \R^\d \colon \f \text{ is}\text{ differentiable}\text{ at } \altpointTwo \} $
it holds that 
\begin{align}
\nonumber
&
\textstyle
  \liminf_{
    \R^\d \backslash \cu{ 0 } \ni h \to 0 
  } 
  \rbr*{ 
    \frac{ \f( \altpoint + h ) - \f( \altpoint ) - \scp{ ( \nabla \f )( \altpoint ) , h } }{ \Pnorm2{h} } 
  } 
  \geq
  -
  \abs*{
    \liminf_{
      \R^\d \backslash \cu{ 0 } \ni h \to 0 
    } 
    \rbr*{ 
      \frac{ \f( \altpoint + h ) - \f( \altpoint ) - \scp{ ( \nabla \f )( \altpoint ) , h } }{ \Pnorm2{h} } 
    } 
  }
\\ & \geq
\textstyle
  -
  \br*{
    \limsup_{
      \R^\d \backslash \cu{ 0 } \ni h \to 0 
    } 
    \pr*{
      \frac{ 
        \abs{ \f( \altpoint + h ) - \f( \altpoint ) - \scp{ ( \nabla \f )( \altpoint ), h } } 
      }{ \Pnorm2{h} }
    }
  }
= 0  
\end{align}
\cfload.
This \proves that for all 
$ \altpoint \in \{ \altpointTwo \in \R^\d \colon \f \text{ is differentiable at } \altpointTwo \} $
it holds that 
\begin{equation}
\label{eq:item_iii_part_A}
  ( \nabla \f )( \altpoint )
  \in 
  ( \FrechetSubdiff \f)( \altpoint )
  .
\end{equation}
\Moreover for all $ \altpointThree \in \R^\d \backslash \{ 0 \} $ it holds that 
\begin{equation}
\begin{split}
&
\textstyle
  \liminf_{
    \R^\d \backslash \cu{ 0 } \ni h \to 0 
  } 
  \rbr*{ 
    \frac{ \scp{ \altpointThree, h } }{ \Pnorm2{h} } 
  } 
  =
  \sup_{ \varepsilon \in (0,\infty) }
  \inf_{ 
    h \in \{ \altpointFour \in \R^\d \colon \pnorm2{ \altpointFour } \leq \varepsilon \}
  }
  \rbr*{ 
    \frac{ \scp{ \altpointThree, h } }{ \Pnorm2{h} } 
  } 
\\ &
\textstyle
\leq
  \sup_{ \varepsilon \in (0,\infty) }
  \rbr*{ 
    \frac{ \scp{ \altpointThree, - \varepsilon \pnorm2{ \altpointThree }^{ - 1 } \altpointThree } }{ \Pnorm2{ - \varepsilon \pnorm2{ \altpointThree }^{ - 1 } \altpointThree } } 
  } 
=
  \sup_{ \varepsilon \in (0,\infty) }
  \rbr*{ 
    \scp{ \altpointThree, - \pnorm2{ \altpointThree }^{ - 1 } \altpointThree } 
  } 
=
  - \pnorm2{ \altpointThree } 
  < 0
  .
\end{split}
\end{equation}
\Hence for all 
$ \altpoint \in \{ \altpointTwo \in \R^\d \colon \f \text{ is differentiable at } \altpointTwo \} $, 
$ \altpointFour \in ( \FrechetSubdiff \f)( \altpoint ) $
that 
\begin{equation}
\begin{split}
\textstyle
&
  0
\leq 
  \liminf_{
    \R^\d \backslash \cu{ 0 } \ni h \to 0 
  } 
  \rbr*{ 
    \frac{ \f( \altpoint + h ) - \f( \altpoint ) - \scp{ \altpointFour, h } }{ \Pnorm2{h} } 
  } 
\\ & 
\textstyle
  =
  \liminf_{
    \R^\d \backslash \cu{ 0 } \ni h \to 0 
  } 
  \rbr*{ 
    \frac{ 
      \f( \altpoint + h ) - \f( \altpoint ) 
      - \scp{ ( \nabla \f )( \altpoint ), h }
      - \scp{ \altpointFour - ( \nabla \f )( \altpoint ), h } 
    }{ \Pnorm2{h} } 
  } 
\\ & 
\textstyle
  \leq
  \liminf_{
    \R^\d \backslash \cu{ 0 } \ni h \to 0 
  } 
  \rbr*{ 
    \frac{ 
      \abs{
        \f( \altpoint + h ) - \f( \altpoint ) 
        - \scp{ ( \nabla \f )( \altpoint ), h }
			}
      + \scp{ ( \nabla \f )( \altpoint ) - \altpointFour, h } 
    }{ \Pnorm2{h} } 
  } 
\\ &
\textstyle
  \leq
  \br*{
    \liminf_{
      \R^\d \backslash \cu{ 0 } \ni h \to 0 
    } 
    \rbr*{ 
      \frac{ 
        \scp{ ( \nabla \f )( \altpoint ) - \altpointFour, h } 
      }{ \Pnorm2{h} } 
    } 
  }
  +
  \br*{
    \limsup_{
      \R^\d \backslash \cu{ 0 } \ni h \to 0 
    } 
    \rbr*{ 
      \frac{ 
        \abs{
          \f( \altpoint + h ) - \f( \altpoint ) 
          - \scp{ ( \nabla \f )( \altpoint ), h }
				}
      }{ \Pnorm2{h} } 
    } 
  }
\\ &
\textstyle
  =
    \liminf_{
      \R^\d \backslash \cu{ 0 } \ni h \to 0 
    } 
    \rbr*{ 
      \frac{ 
        \scp{ ( \nabla \f )( \altpoint ) - \altpointFour, h } 
      }{ \Pnorm2{h} } 
    } 
  \leq 
  -
  \Pnorm2{
    ( \nabla \f )( \altpoint ) - \altpointFour 
  }
  .
\end{split}
\end{equation}
Combining this with \cref{eq:item_iii_part_A} 
\proves[ep] \cref{item:subdiff_item_iii}.
\Nobs that 
\cref{item:subdiff_item_ii,item:subdiff_item_iii} 
\prove that for all open
$
  U \subseteq \R^{ n }
$
and all 
$ \altpoint \in U $
with 
$
  \f|_U \in C^1( U, \R )
$
it holds that 
\begin{equation}
\label{eq:in_proof_prop_subdiff_A}
  \{ 
    ( \nabla \f )( \altpoint )
  \}
  =
  ( \cD \f )( \altpoint )
  \subseteq 
  ( \limitingFrechetSubdiff \f )( \altpoint )
  .
\end{equation}
\Moreover for all 
open 
$ U \subseteq \R^\d $,
all
$ \altpoint \in U $,
$
  \altpointTwo \in \R^\d 
$
and all 
$
  z = ( z_1, z_2 ) \colon \N \to \R^\d \times \R^\d
$
with 
$
  \limsup_{ k \to \infty } 
  (
    \pnorm2{ 
      z_1( k ) - \altpoint
		}
    + 
    \pnorm2{ 
      z_2( k ) - \altpointTwo
		}
  )
  = 0
$
and
$
  \forall \, k \in \N \colon 
  z_2( k ) \in ( \cD \f )( z_1( k ) )
$ 
there exists $ K \in \N $ 
such that for all 
$ k \in \N \cap [ K, \infty) $
it holds that 
\begin{equation}
  z_1(k) \in U
  .
\end{equation}
Combining this with 
\cref{item:subdiff_item_iii}
\proves that for all 
open 
$ U \subseteq \R^\d $,
all
$ \altpoint \in U $,
$
  \altpointTwo \in \R^\d 
$
and all 
$
  z = ( z_1, z_2 ) \colon \N \to \R^\d \times \R^\d
$
with 
$
  \f|_U \in C^1( U, \R )
$,
$
  \limsup_{ k \to \infty } 
  (
    \pnorm2{
      z_1( k ) - \altpoint
		}
    + 
    \pnorm2{
      z_2( k ) - \altpointTwo
		}
  )
  = 0
$
and
$
  \forall \, k \in \N \colon 
  z_2( k ) \in ( \cD \f )( z_1( k ) )
$ 
there exists $ K \in \N $ 
such that 
$
  \forall \, k \in \N \cap [K,\infty) \colon 
  z_1( k ) \in U
$
and
\begin{equation}
\begin{split}
&
\textstyle 
  \limsup_{ \N \cap [K,\infty) \ni k \to \infty } 
  (
    \pnorm2{
      z_1( k ) - \altpoint
    }
    + 
    \pnorm2{
      ( \nabla \f )( z_1( k ) ) - \altpointTwo
    }
  )
\\ &
\textstyle 
=
  \limsup_{ k \to \infty } 
  (
    \pnorm2{
      z_1( k ) - \altpoint
    }
    + 
    \pnorm2{
      z_2( k ) - \altpointTwo
    }
  )
  = 0
  .
\end{split}
\end{equation}
This and \cref{item:subdiff_item_i} \prove that 
for all open $ U \subseteq \R^\d $
and all $ \altpoint \in U $, 
$ \altpointTwo \in ( \limitingFrechetSubdiff \f )( \altpoint ) $
with $ \f|_U \in C^1( U, \R ) $
it holds that 
\begin{equation}
  \altpointTwo = ( \nabla \f )( \altpoint )
  .
\end{equation}
Combining this with \cref{eq:in_proof_prop_subdiff_A} 
\proves[ep] \cref{item:subdiff_item_iv}. 
\Nobs that \cref{def:limit:subdiff:eq} 
\proves that for all $ \altpoint \in \R^\d $ it holds that 
\begin{equation}
\textstyle 
  \R^\d \backslash ( ( \limitingFrechetSubdiff \f )( \altpoint ) )
  =
    \bigcup_{ \varepsilon \in (0,\infty) }
    \bigl(
      \R^\d \backslash 
      \bigl( 
      \,
  \overline{
      {\scriptstyle\bigcup}_{ 
        \altpointTwo \in 
        \cu{ 
          z \in \R^\d \colon \Pnorm2{ \altpoint - z } < \varepsilon 
        }
      } 
      ( \FrechetSubdiff \f )( \altpointTwo ) 
  }
  \,
      \bigr) 
    \bigr)
\end{equation}
\Hence for all $ \altpoint \in \R^\d $ that 
$
  \R^\d \backslash ( ( \limitingFrechetSubdiff \f )( \altpoint ) )
$
is open. This \proves[ep] \cref{item:subdiff_item_v}. 
\end{aproof}
\endgroup

\cfclear
\begingroup
\providecommand{\f}{}
\renewcommand{\f}{\defaultLossFunction}
\begin{athm}{lemma}{lem:example1_subdifferentials}[Fr\'{e}chet subgradients for maxima]
Let 
$ c \in \R $
and let 
$ \f \colon \R \to \R $ satisfy 
for all 
$ \altpoint \in \R $ 
that 
$
  \f( \altpoint ) = \max\{ \altpoint, c \} 
$. 
Then 
\begin{enumerate}[label = (\roman*)]
\item 
\label{item:example1_subdifferentials_item_i}
it holds for all $ \altpoint \in (-\infty,c) $ that
$
  ( \FrechetSubdiff \f )( \altpoint )
  =
  \{ 0 \}
$,
\item 
\label{item:example1_subdifferentials_item_ii}
it holds for all $ \altpoint \in (c,\infty) $ that
$
  ( \FrechetSubdiff \f )( \altpoint )
  =
  \{ 1 \}
$, 
and 
\item 
\label{item:example1_subdifferentials_item_iii}
it holds that 
$
  ( \FrechetSubdiff \f )( c )
  =
  [0,1]
$
\end{enumerate}
\cfload.
\end{athm}
\begin{aproof}
\Nobs that 
\cref{item:subdiff_item_iii}
in \cref{lem:subdifferential:c1}
\proves[ep]
\cref{item:example1_subdifferentials_item_i,item:example1_subdifferentials_item_ii}. 
\Nobs that \cref{lem:convex_differentials}
establishes
\begin{eqsplit}
\label{item:example1_subdifferentials_item_ii_partA}
  [0,1]
  \subseteq 
  ( \FrechetSubdiff \f )( c )
  .
\end{eqsplit}
\Moreover 
the assumption that for all 
$ \altpoint \in \R $ it holds that 
$
  \f(\altpoint) = \max\{ \altpoint, c \} 
$
\proves that
for all $ a \in (1,\infty) $, 
$ h \in (0,\infty) $
it holds that 
\begin{eqsplit}
\label{eq:example_subdiff_1}
  \frac{
    \f( c + h )
    -
    \f( c )
    -
    a h
  }{
    \abs{ h }
  }
  =
  \frac{
    ( c + h )
    -
    c
    -
    a h
  }{
    h
  }
  =
  1 - a 
  < 0
  .
\end{eqsplit}
\Moreover 
the assumption that for all 
$ \altpoint \in \R $ it holds that 
$
  \f(\altpoint) = \max\{ \altpoint, c \} 
$
\proves that
for all $ a, h \in (-\infty,0) $, 
it holds that 
\begin{eqsplit}
  \frac{
    \f( c + h )
    -
    \f( c )
    -
    a h
  }{
    \abs{ h }
  }
  =
  \frac{
    c
    -
    c
    -
    a h
  }{
    - h
  }
  =
  a
  < 0
  .
\end{eqsplit}
Combining this with \cref{eq:example_subdiff_1} demonstrates that 
\begin{eqsplit}
  ( \FrechetSubdiff \f )( c )
  \subseteq [0,1]
  .
\end{eqsplit}
This and \cref{item:example1_subdifferentials_item_ii_partA} 
establish 
\cref{item:example1_subdifferentials_item_iii}. 
\end{aproof}
\endgroup

\cfclear
\begingroup
\providecommand{\d}{}
\renewcommand{\d}{\defaultParamDim}
\providecommand{\f}{}
\renewcommand{\f}{\defaultLossFunction}
\providecommand{\n}{}
\renewcommand{\n}{n}
\begin{athm}{lemma}{lem:limiting_derivatives}[Limits of limiting Fr\'{e}chet subgradients]
\cfadd{def:limit:subdiff}
Let $ \d \in \N $, $ \f \in C( \R^\d , \R ) $, 
let 
$ ( \altpoint_k )_{ k \in \N_0 } \subseteq \R^\d $
and 
$ ( \altpointTwo_k )_{ k \in \N_0 } \subseteq \R^\d $
satisfy 
\begin{equation}
\begin{split} 
\textstyle
  \limsup_{ k \to \infty }
  ( 
    \Pnorm2{ \altpoint_k - \altpoint_0 }
    +
    \Pnorm2{ \altpointTwo_k - \altpointTwo_0 }
  )
  = 0,
\end{split}
\end{equation} 
and assume for all $ k \in \N $ that 
$
  \altpointTwo_k \in ( \limitingFrechetSubdiff \f)( \altpoint_k )
$
\cfload. 
Then 
$
  \altpointTwo_0 \in ( \limitingFrechetSubdiff \f)( \altpoint_0 )
$.
\end{athm}

\begin{aproof}
\Nobs that 
\cref{item:subdiff_item_i} in 
\cref{lem:subdifferential:c1}
and the fact that for all 
$ k \in \N $ it holds that 
$
  \altpointTwo_k \in ( \limitingFrechetSubdiff \f )( \altpoint_k )
$
\prove that for every $ k \in \N $
there exists  
$
  z^{ (k) } = ( z^{ (k) }_1, z^{ (k) }_2 ) \colon \N \to \R^\d \times \R^\d 
$
which satisfies 
for all
$\n\in\N$
that
\begin{equation}
\label{eq:limit_graph_z}
\textstyle
    z^{ (k) }_2(\n) \in ( \FrechetSubdiff \f )( z^{ (k) }_1(\n) )
\;\,\text{and}\;\,
  \limsup_{ w \to \infty }
  \bigl(
    \Pnorm2{ z^{ (k) }_1( w ) - \altpoint_k }
    +
    \Pnorm2{ z^{ (k) }_2( w ) - \altpointTwo_k }
  \bigr) = 0
  .
\end{equation}
\Nobs that \cref{eq:limit_graph_z} \proves that 
there exists 
$
  \altpointThree = ( \altpointThree_k )_{ k \in \N } \colon \N \to \N
$
which satisfies for all $ k \in \N $ that
\begin{equation}
\label{eq:limit_graph_z2}
  \pnorm2{ z^{ (k) }_1( \altpointThree_k ) - \altpoint_k }
  +
  \pnorm2{ z^{ (k) }_2( \altpointThree_k ) - \altpointTwo_k }
  \leq 
  2^{ - k }
  .
\end{equation}
Next let $ Z = ( Z_1, Z_2 ) \colon \N \to \R^\d \times \R^\d $ 
satisfy for all 
$ j \in \{ 1, 2 \} $, $ k \in \N $ that 
\begin{equation}
\label{eq:limit_graph_Z}
  Z_j(k) = z^{ (k) }_j( \altpointThree_k ) .
\end{equation}
\Nobs that \cref{eq:limit_graph_z},
\cref{eq:limit_graph_z2},
\cref{eq:limit_graph_Z}, 
and the assumption that 
$
  \limsup_{ k \to \infty }
  ( 
    \Pnorm2{ \altpoint_k - \altpoint_0 }
    +
    \Pnorm2{ \altpointTwo_k - \altpointTwo_0 } 
  )
  = 0
$
\prove that 
\begin{equation}
\label{eq:limit_big_Z}
\begin{split}
&
\textstyle
  \limsup_{ k \to \infty }
  \bigl(
    \Pnorm2{
      Z_1(k) - \altpoint_0
    }
    +
    \Pnorm2{
      Z_2(k) - \altpointTwo_0
    }
  \bigr)
\\ & 
\textstyle
  \leq
  \bigl[
    \limsup_{ k \to \infty }
    \bigl(
      \Pnorm2{
        Z_1(k) - \altpoint_k
      }
      +
      \Pnorm2{
        Z_2(k) - \altpointTwo_k
      }
    \bigr)
  \bigr]
  \\&\qquad +
  \bigl[\textstyle
    \limsup_{ k \to \infty }
    \bigl(
      \Pnorm2{ 
        \altpoint_k - \altpoint_0
      }
      +
      \Pnorm2{
        \altpointTwo_k - \altpointTwo_0
      }
    \bigr)
  \bigr]
\\ & 
\textstyle
  =
    \limsup_{ k \to \infty }
    \bigl(
      \Pnorm2{
        Z_1(k) - \altpoint_k
      }
      +
      \Pnorm2{
        Z_2(k) - \altpointTwo_k
      }
    \bigr)
\\ &
\textstyle
  =
    \limsup_{ k \to \infty }
    \bigl(
      \Pnorm2{
        z_1^{ (k) }( \altpointThree_k ) - \altpoint_k
      }
      +
      \Pnorm2{
        z_2^{ (k) }( \altpointThree_k ) - \altpointTwo_k
      }
    \bigr)
  \\&\leq\textstyle
    \limsup_{ k \to \infty }
    \bigl(
      2^{ - k }
    \bigr)
    = 0
    .
\end{split}
\end{equation}
\Moreover \cref{eq:limit_graph_z,eq:limit_graph_Z} \prove that 
for all $ k \in \N $ it holds that 
$
  Z_2( k ) \in ( \FrechetSubdiff \f )( Z_1( k ) )
$. 
Combining this and \cref{eq:limit_big_Z} 
with 
\cref{item:subdiff_item_i} in 
\cref{lem:subdifferential:c1} 
\proves[ep] that 
$ \altpointTwo_0 \in ( \limitingFrechetSubdiff \f )( \altpoint_0 ) $. 
\end{aproof}
\endgroup

\cfclear
\begingroup
\providecommand{\d}{}
\renewcommand{\d}{\defaultParamDim}
\providecommand{\f}{}
\renewcommand{\f}{\defaultLossFunction}
\begin{exercise}{quest:Frechet1}
Prove or disprove the following statement: 
It holds for all $\d \in \N$, $\f \in C^1( \R^\d, \R)$, $\altpoint \in \R^\d$ that
$ 
  ( \limitingFrechetSubdiff \f)( \altpoint )
  = 
  ( \FrechetSubdiff \f)( \altpoint )
$
\cfload.
\end{exercise}
\endgroup

\cfclear
\begingroup
\providecommand{\d}{}
\renewcommand{\d}{\defaultParamDim}
\providecommand{\f}{}
\renewcommand{\f}{\defaultLossFunction}
\begin{exercise}{quest:Frechet2}
Prove or disprove the following statement: 
There exists $\d \in \N$ such that for all $\f \in C( \R^\d, \R)$, $\altpoint \in \R^\d$ it holds that 
$ 
  ( \limitingFrechetSubdiff \f)( \altpoint )
  \subseteq 
  ( \FrechetSubdiff \f)( \altpoint )
$
\cfload.
\end{exercise}
\endgroup

\cfclear
\begingroup
\providecommand{\d}{}
\renewcommand{\d}{\defaultParamDim}
\providecommand{\f}{}
\renewcommand{\f}{\defaultLossFunction}
\begin{exercise}{quest:Frechet3}
Prove or disprove the following statement: 
It holds for all $\d \in \N$, $\f \in C( \R^\d, \R)$, $\altpoint \in \R^\d$ that
$ 
  ( \FrechetSubdiff \f)( \altpoint )
$
is convex
\cfload.
\end{exercise}
\endgroup

\cfclear
\begingroup
\providecommand{\d}{}
\renewcommand{\d}{\defaultParamDim}
\providecommand{\f}{}
\renewcommand{\f}{\defaultLossFunction}
\begin{exercise}{quest:Frechet4}
Prove or disprove the following statement: 
It holds for all $\d \in \N$, $\f \in C( \R^\d, \R)$, $\altpoint \in \R^n$ that
$ 
  ( \limitingFrechetSubdiff \f)( \altpoint )
$
is convex
\cfload.
\end{exercise}
\endgroup

\cfclear
\begingroup
\providecommand{\d}{}
\renewcommand{\d}{\defaultParamDim}
\providecommand{\f}{}
\renewcommand{\f}{\defaultLossFunction}
\providecommand{\g}{}
\renewcommand{\g}{\defaultGradientFunction}
\begin{exercise}{question:compute_Frechetdiff}
For every 
	$\alpha \in (0,\infty)$,
	$s \in \{-1, 1\}$
let $\f_{\alpha, s} \colon \R \to \R$ satisfy for all 
	$\altpoint \in \R$ 
that
\begin{equation}
\begin{split} 
	\f_{\alpha, s}(\altpoint) 
=
	\begin{cases}
		\altpoint &\colon \altpoint> 0 \\
		s\abs{\altpoint}^\alpha &\colon \altpoint \leq 0.
	\end{cases}
\end{split}
\end{equation}
For every 
	$\alpha \in (0,\infty)$,
	$s \in \{-1, 1\}$,
	$\altpoint \in \R$ 
specify
	$(\FrechetSubdiff \f_{\alpha, s})(\altpoint)$ and 
	$(\limitingFrechetSubdiff \f_{\alpha, s})(\altpoint)$ 
explicitly and prove that your results are correct
\cfload!
\end{exercise}
\endgroup

\subsection{Non-smooth slope}

\cfclear
\begingroup
\providecommand{\d}{}
\renewcommand{\d}{\defaultParamDim}
\providecommand{\f}{}
\renewcommand{\f}{\defaultLossFunction}
\begin{adef}{def:nonsmoothSlope}[Non-smooth slope]
\cfconsiderloaded{def:nonsmoothSlope}
Let $ \d \in \N $, $ \f \in C(\R^{ \d }, \R) $. 
Then we denote by 
$
  \NSslope_f \colon \R^{ \d } \to [0,\infty] 
$
the function which satisfies for all $ \altpoint \in \R^{ \d } $ that 
\begin{equation}
\label{eq:def_non-smooth_slope}
\cfadd{def:limit:subdiff}
  \NSslope_\f( \altpoint ) =
  \inf\bigl(
    \bigl\{
      r \in \R \colon 
      (
        \exists \,
        \altpointTwo \in ( \limitingFrechetSubdiff \f )( \altpoint ) 
        \colon
        r =
        \Pnorm2{\altpointTwo} 
      )
    \bigr\} 
    \cup \{ \infty \}
  \bigr) 
\end{equation}
and we call $\NSslope_f$ the non-smooth slope of $f$
\cfload. 
\end{adef}
\endgroup

\subsection{Generalized KL functions}

\cfclear
\begingroup
\providecommand{\d}{}
\renewcommand{\d}{\defaultParamDim}
\providecommand{\f}{}
\renewcommand{\f}{\defaultLossFunction}
\begin{adef}{def:KLinequality}[Generalized \KL\ inequalities]
Let $ \d \in \N $, $c \in \R $, $ \alpha \in (0,\infty) $, $ \f \in C(\R^{ \d }, \R) $,
let $ U \subseteq \R^{ \d } $ be a set, 
and let $\cpoint \in U$.
Then we say that $ \f $ satisfies the generalized \KL\ inequality at $\cpoint$ on $U$ 
with exponent $\alpha$ 
and constant $c$ 
(we say that $ \f $ satisfies the generalized \KL\ inequality at $\cpoint$)
if and only if 
for all $ \altpoint \in U $ it holds that 
\begin{equation}
  \abs{ \f( \cpoint ) - \f( \altpoint ) }^{ \alpha }
  \leq 
  c 
  \,
  \abs{
    \NSslope_\f( \altpoint )
	}
\end{equation}
\cfload.
\end{adef}
\endgroup

\cfclear
\begingroup
\providecommand{\d}{}
\renewcommand{\d}{\defaultParamDim}
\providecommand{\f}{}
\renewcommand{\f}{\defaultLossFunction}
\begin{adef}{def:KLfunction}[Generalized \KL\ functions]
Let $ \d \in \N $, $ \f \in C(\R^{ \d }, \R) $. 
Then we say that $ \f $ is a generalized \KL\ function if and only if 
for all $ \cpoint \in \R^{ \d } $ there exist 
$ \varepsilon, c \in (0,\infty) $, 
$ \alpha \in (0,1) $
such that for all 
$
  \altpoint \in \{ \altpointTwo \in \R^{ \d } \colon \Pnorm2{ \altpointTwo - \cpoint } < \varepsilon \} 
$
it holds that 
\begin{equation}
  \abs{ \f( \cpoint ) - \f( \altpoint ) }^{ \alpha }
  \leq 
  c 
  \,
  \abs{
    \NSslope_\f( \altpoint )
	}
\end{equation}
\cfload.
\end{adef}
\endgroup

\begin{athm}{remark}{rem:gen_KL}[Examples and convergence results for generalized \KL\ functions]
In \cref{thm:KL,cor:analyticity_empirical_risk} above we have seen that in the case of an analytic activation function we have that the associated empirical risk function is also analytic and therefore a standard \KL\ function.
In deep learning algorithms often deep \anns\ with non-analytic activation functions such as the \ReLU\ activation (cf.\ \cref{subsect:Relu}) and the leaky \ReLU\ activation (cf.\ \cref{subsect:leaky_relu}) are used.
In the case of such non-differentiable activation functions, the associated risk function is typically not a standard \KL\ function. However, under suitable assumptions on the target function and the underlying probability measure of the input data of the considered learning problem, using Bolte et al.~\cite[Theorem 3.1]{Bolte2006} one can verify in the case of such non-differentiable activation functions that the risk function is a generalized \KL\ function in the sense of \cref{def:KLfunction} above;
cf., \eg, \cite{Eberle2023,JentzenRiekert2022Existence}.
Similar as for standard \KL\ functions (cf., \eg, Dereich \& Kassing~\cite{Dereich2021} and \cref{section:gf:loja,sec:convergence_GD}) one can then also develop a convergence theory for gradient based optimization methods for generalized \KL\ function (cf., \eg, Bolte et al.~\cite[Section 4]{Bolte2006} and \cref{cor:gf:global:abstract}).
\end{athm}

\begin{athm}{remark}{rem:other_cv_approaches}[Further convergence analyses]
We refer, \eg, to \cite{An2023,Dereich2021,Tadic2015,Absil2005,Attouch2009,Bolte2006} 
and the references therein for convergence analyses under \KL-type conditions
for gradient based optimization methods in the literature.
Beyond the \KL\ approach reviewed in this chapter there are also several other approaches in the literature with which one can conclude convergence of gradient based optimization methods to suitable generalized critical points; cf., \eg, \cite{Bolte2021,Davis2020,Chatterjee2022} and the references therein.
\end{athm}

\section{Non-convergence for stochastic gradient descent}

In \cref{section:gf:loja,sec:convergence_GD} above we present convergence results for gradient based optimization procedures to critical points.
In general one cannot expect that these limiting critical points are global minimizers.
This is the subject of the next statement, \cref{non_convergence_SGD} below.
\cref{non_convergence_SGD} shows in particular that the probability to converge to a global minimizer converges to zero as the number of parameters of the \ann\ architecture converges to infinity.

\begin{samepage}
\begingroup 
\newcommand{\cLnri}[3]{\mathcal{L}_{#1}^{#2,#3}}
\newcommand{\cki}{c}
\newcommand{\NNelll}{\Theta}
\renewcommand{\cN}{\mathfrak{N}}
\renewcommand{\f}{f}
\cfclear
\begin{athm}{theorem}{non_convergence_SGD}
Let $d\in\N$,
$ a \in \R $, 
$ b \in (a, \infty)  $, let $ ( \Omega, \mathcal{F}, \P) $ be a probability space, for every $ m, n \in \N_0 $ 
let 
$ X^m_n \colon \Omega \to [a,b]^d $
and 
$ Y^m_n \colon \Omega \to \R $
be random variables, assume for all $i\in \N$, $j\in \N\backslash\{i\}$ that $\P( X_0^i=X_0^j)=0$, let $\f\colon\N\to \N$ be a function, for every $k\in \N_0$ let $\fd_k,L_k\in \N$, $\ell_k =(\ell_k^0,\ell_k^1\dots,\ell_k^{L_k}) \in \N^{L_k+1}$ satisfy $\ell_k^0=d$, $\ell_k^{L_k}=1$, $\fd_k=\sum_{ i = 1 }^{L}\ell_k^i ( \ell_k^{i-1} + 1 )$, and $\max\{\ell_k^1,\ell_k^2,\dots,\ell_k^{L_k}\}\leq \f(\ell_k^1)$,  let $\mathbb{A}_r\colon \mathbb{R} \rightarrow \mathbb{R}$, $r \in \N_0$, satisfy for all $x\in \R$
that there exists $m \in \N$ such that 
	$(\cup_{r = m}^{\infty} \{\mathbb{A}_r\}) \subseteq C^1(\R,\R)$
	and 
\begin{equation}\label{equation 1: main theorem}
    \sum_{r=m}^{\infty} 
		\pr{
			\abs{
				\mathbb{A}_0(x)
				-
				\mathbb{A}_r(x)
			}
			+
			\abs{
				\mathbbm{1}_{(0,\infty)}(x)
				-
				( \mathbb{A}_r )'( x )
			}
			+
			\abs{
				\max\{0,x\}
				-
				\mathbb{A}_0(x)
			}
		}
=
	0,
\end{equation}
for every $r, k \in \N_0$, $v \in \{1,2,\dots,L_k\}$ let $\Psi^{r, k}_v \colon \R^{\ell_k^v} \to \R^{\ell_k^v}$ satisfy
\begin{equation}
	\Psi^{r, k}_v
=
	\begin{cases}	
		\multdim_{\mathbb{A}_r, \ell_k^v} & \colon v < L_k \\
		\id_{\R^{\ell_k^v}} & \colon v = L_k,
	\end{cases}
\end{equation}
for every $k,n\in \N_0$ let $ M^{ k }_n \in  \N $, $\gamma_n^k\in \R$, for every $r, k,n\in \N_0$ let
$ 
  \cLnri{n}{r}{k} \colon \R^{ \fd_{ k } } \times \Omega \to \R 
$
satisfy for all 
$ \altpoint \in \R^{ \fd_{ k} }$
that
\begin{equation}\label{equation 3: main theorem}
\displaystyle
  \cLnri{n}{r}{k}( \altpoint) 
  = 
  \frac{ 1 }{ M^{ k }_n } 
  \biggl[ \textstyle
  \sum\limits_{ m = 1 }^{ M^{ k }_n} 
    \abs[\big]{
      \pr[\big]{
	  	\RealV{\altpoint}{.}{d}{
	  		\Psi^{r, k}_1,	
			\Psi^{r, k}_2,
			\ldots,
			\Psi^{r, k}_{L_k}
	  	}
	  }
	  ( X^m_n )
      - 
      Y^m_n 
    }^2
  \biggr]
  ,
\end{equation}
 for every $k,n\in \N_0$ let 
$ 
  \fG_n^{ k }  
  \colon \R^{ \fd_{ k} } \times \Omega \to \R^{ \fd_{ k } } 
$ 
satisfy for all $\omega\in \Omega$, $\altpoint\in \{\altpointTwo\in \R^{\fd_k}\colon (\nabla_{\altpointTwo} \cLnri{n}{r}{k}(\altpointTwo,\allowbreak\omega))_{r\in \N}$ is convergent$\}$
that
\begin{equation}\label{equation 4: main theorem}
  \fG^{k}_n( \altpoint,\omega) 
  = 
  \lim_{r\to\infty}\bigl[\nabla_\altpoint \cLnri{n}{r}{k}(\altpoint,\omega)\bigr]
\end{equation}
and let 
	$
	\Theta_n^k 
	\colon \Omega  \to \R^{\fd_{ k } }
	$
	be a random variable, 
	 assume for all $k,n\in \N$ that 
	\begin{equation}\label{equation 5: main theorem}
		\Theta_{ n  } ^k
		= 
		\Theta_{n-1} ^k - \gamma_{n}^k
		\fG_n^k ( \Theta_{n-1}^k ),
	\end{equation} 
  assume $\liminf_{k\to \infty} \fd_{k}=\infty$ and $\liminf_{k \to \infty}\allowbreak \P\big(\inf_{\altpoint\in \R^{\fd_{k}}}\cL_0^{0,k}(\altpoint) >0\big)=1$, for every $k\in \N$ let $\cki_{k}\in (0,\infty)$, and assume for all $k\in \N$ that $\cki_k\Theta_0^k$ is standard normal \cfload. Then
\begin{equation}\label{equation 6: main theorem}
\textstyle
\liminf\limits_{ k \to \infty } \P\biggl(\inf\limits_{ n\in \N_0 }  \cLnri{0}{0}{k}(\NNelll^{k}_n)>\inf\limits_{\altpoint\in \R^{\fd_k}}\cLnri{0}{0}{k}(\altpoint)\biggr)=1.
\end{equation}
\end{athm}
\endgroup
\end{samepage}

\cref{non_convergence_SGD} is, in a slightly modified form, proved as \cite[Theorem 1.1]{Hannibal2024} (cf.\ also \cite[Theorem 1.1]{Dereich2025}).
We also refer, \eg, to 
\cite{Cheridito2021,Dereich2024,Dereich2024a,Gallon2022,Grohs2021b,Jentzen2024,Lu2020,Reddi2019} for further lower bounds and non-convergence results for \SGD\ optimization methods.

%% file: parts/ANNs_with_batch_normalization.tex
\cchapter{ANNs with batch normalization}{chapter:BN}

In data-driven learning problems popular methods that aim to accelerate \ann\ training procedures are \BN\ methods.
In this chapter we rigorously review such methods in detail.
In the literature \BN\ methods have first been introduced in Ioffe \& Szegedi~\cite{Ioffe2015}.

Further investigation on \BN\ techniques and applications of such methods can, \eg, be found in %
\cite[Section 12.3.3]{alpaydin2020introduction},
\cite[Section 6.2.3]{Fan2021},
\cite[Section 8.7.1]{Goodfellow2016}, and 
\cite{Santurkar2018,Bjorck2018}.

\section{Batch normalization (BN)}

\cfclear
\begin{adef}{def:batch}[Batch]
Let $d, M \in \N$.
Then we say that $x$ is a batch of $d$-dimensional data points of size $M$ 
(we say that $x$ is a batch of $M$ $d$-dimensional data points, we say that $x$ is a batch)  if and only if it holds that $x \in (\R^d)^M$.
\end{adef}

\cfclear
\begin{adef}{def:batch_mean}[Batch mean]
Let 
	$d, M \in \N$,
	$x = (x^{(m)})_{m \in \{1, 2, \ldots, M\}} \in (\R^d)^M$.
Then
we denote by 
$\batchmean(x) = (\batchmean_1(x), \ldots, \batchmean_d(x)) \in \R^d$ the vector given by
\begin{equation}
\label{batch_mean:eq1}
\begin{split} 
	\batchmean(x)
=
	\frac{1}{M}
	\br*{
		\sum_{m = 1}^M
			x^{(m)}
	}
\end{split}
\end{equation}
and we call $\batchmean(x)$ the batch mean of the batch $x$.
\end{adef}

\cfclear
\begin{adef}{def:batch_var}[Batch variance]
Let 
	$d, M \in \N$,
	$x =  ((x^{(m)}_i)_{i \in \{1, 2, \ldots, d\}})_{m \in \{1, 2, \ldots, M\}} \in (\R^d)^M$.
Then
we denote by 
\begin{equation}
  \batchvar(x) = (\batchvar_1(x), \ldots, \batchvar_d(x)) \in \R^d
\end{equation}
the vector which satisfies for all
$ i \in \{ 1, 2, \dots, d \} $
that
\begin{equation}
\label{batch_var:eq1}
\begin{split} 
	\batchvar_i(x) 
=
	\frac{1}{M}
	\br*{
		\sum_{m = 1}^M
			(x^{(m)}_i - \batchmean_i(x) )^2
	}
\end{split}
\end{equation}
and we call $\batchvar(x)$ the batch variance of the batch $x$
(cf.\ \cref{def:batch_mean}).
\end{adef}

\cfclear
\begin{lemma}
\label{batch_random_lemma}
Let 
	$d, M \in \N$,
	$x = (x^{(m)})_{m \in \{1, 2, \ldots, M\}} = ((x^{(m)}_i)_{i \in \{1, 2, \ldots, d\}})_{m \in \{1, 2, \ldots, M\}} \in (\R^d)^M$,
let $(\Omega, \cF, \P)$ be a probability space,
and let
$U \colon \Omega \to \{1, 2, \ldots, M \}$ be a \unif{$\{1, 2, \ldots, \allowbreak M \}$} random variable.
Then
\begin{enumerate}[label=(\roman *)]
\item 
\label{batch_random_lemma:item1}
it holds that $\batchmean(x) = \Exp{x^{(U)}}$
and

\item 
\label{batch_random_lemma:item2}
it holds for all
	$i \in \{1, 2, \ldots, d \}$
that
$\batchvar_i(x) = \var({x^{(U)}_i})$.
\end{enumerate}
\end{lemma}

\begin{proof}[Proof of \cref{batch_random_lemma}]
Note that \cref{batch_mean:eq1} proves \cref{batch_random_lemma:item1}.
\Moreover
\enum{
	\cref{batch_random_lemma:item1};
	\cref{batch_var:eq1}
}[establish]
\cref{batch_random_lemma:item2}.
The proof of \cref{batch_random_lemma} is thus complete.
\end{proof}

\cfclear
\begin{adef}{def:batch_norm_op}[\BN\ operations for given batch mean and batch variance]
Let 
$ d \in \N $,
$ \varepsilon \in (0,\infty) $,
$ \beta = ( \beta_1, \dots, \beta_d ) $,
$ \gamma = ( \gamma_1, \dots, \gamma_d ) $,
$ \mu = (\mu_1, \dots, \mu_d ) \in \R^d $, 
$ \variance = ( \variance_1, \dots, \variance_d ) \in [0,\infty)^d $.
Then we denote by 
\begin{equation}
  \Batchnormop_{\beta, \gamma, \mu, \variance, \varepsilon}
  \colon 
  \R^d \to \R^d
\end{equation}
the function which satisfies for all
	$x = (x_1, \ldots, x_d) \allowbreak \in \R^d$
that
\begin{equation}
\label{def:batch_norm_op:eq1}
\begin{split} 
	\Batchnormop_{\beta, \gamma, \mu, \variance, \varepsilon}(x)
=
	\pr[\Big]{
		\gamma_i
		\br[\Big]{
			\frac{x_i - \mu_i}{\sqrt{{\variance_i + \varepsilon}}}
		}
		+
		\beta_i
	}_{\!i \in \{1, 2, \ldots, d\}}
\end{split}
\end{equation}
and we call 
$
	\Batchnormop_{\beta, \gamma, \mu, \variance, \varepsilon}
$
the \BN\ operation with mean parameter $\beta$, standard deviation parameter $\gamma$, and regularization parameter $\varepsilon$ given the batch mean $\mu$ and batch variance $\variance$.
\end{adef}

\cfclear
\begin{adef}{def:batch_norm}[Batch normalization]
Let 
$ d \in \N $,
$ \varepsilon \in (0,\infty) $,
$ \beta, \gamma \in \R^d $.
Then we denote by 
\begin{equation}
	\textstyle
  \Batchnorm_{\beta, \gamma, \varepsilon}
  \colon 
  \pr*{
    \bigcup_{M \in \N} (\R^d)^M
  } 
  \to
  \pr*{
    \bigcup_{M \in \N} (\R^d)^M
  }
\end{equation}
the function which satisfies for all
	$M \in \N$,
	$x = (x^{(m)})_{m \in \{1, 2, \ldots, M\}} \in (\R^d)^M$
that
\begin{equation}
\label{def:batch_norm:eq1}
\begin{split} 
	\Batchnorm_{\beta, \gamma, \varepsilon}(x)
= 
	\pr*{
		\Batchnormop_{\beta, \gamma, \batchmean(x), \batchvar(x), \varepsilon}(x^{(m)})
	}_{m \in \{1, 2, \ldots, M\}}
\in
	(\R^d)^M
\end{split}
\end{equation}
and we call $\Batchnorm_{\beta, \gamma, \varepsilon}$ the \BN\ 
with mean parameter $\beta$, 
standard deviation parameter $\gamma$,
and regularization parameter $\varepsilon$
\cfclear\cfadd{def:batch_mean}\cfadd{def:batch_var}\cfadd{def:batch_norm_op}\cfload.
\end{adef}

\cfclear
\begin{lemma}
\label{batch_norm_lemma}
Let 
	$d, M \in \N$,
	$\beta = (\beta_1, \ldots, \beta_d)$,
	$\gamma = (\gamma_1, \ldots, \gamma_d) \in \R^d$.
Then 
\begin{enumerate}[label=(\roman *)]
\item 
\label{batch_norm_lemma:item1}
it holds for all
	$\varepsilon \in (0,\infty)$,
	$x = ((x^{(m)}_i)_{i \in \{1, 2, \ldots, d\}})_{m \in \{1, 2, \ldots, M\}} \in (\R^d)^M$
that
\begin{equation}
\begin{split} 
	\Batchnorm_{\beta, \gamma, \varepsilon}(x)
=
	\pr[\Big]{
		\pr[\Big]{
			\gamma_i
			\br[\Big]{
				\frac{x^{(m)}_i - \batchmean_i(x)}{\sqrt{\smash[b]{\batchvar_i(x) + \varepsilon}} }
			}
			+
			\beta_i
		}_{i \in \{1, 2, \ldots, d\}}
	}_{m \in \{1, 2, \ldots, M\}},
\end{split}
\end{equation}

\item 
\label{batch_norm_lemma:item2}
it holds for all
	$\varepsilon \in (0,\infty)$,
	$x \in (\R^d)^M$
that
\begin{equation}
\begin{split} 
	\batchmean(\Batchnorm_{\beta, \gamma, \varepsilon}(x))
=
	\beta,
\end{split}
\end{equation}
and 
\item 
\label{batch_norm_lemma:item3}
it holds for all
	$x = ((x^{(m)}_i)_{i \in \{1, 2, \ldots, d\}})_{m \in \{1, 2, \ldots, M\}} \in (\R^d)^M$,
	$i \in \{1, 2, \ldots, d\}$
with
	$\#(\bigcup_{m=1}^M \{x^{(m)}_i\})\allowbreak >1$	
that
\begin{equation}
\begin{split} 
	\limsup\nolimits_{\varepsilon \searrow 0} \,
	\abs[\big]{
		\batchvar_i(\Batchnorm_{\beta, \gamma, \varepsilon}(x))
		-
		(\gamma_i)^2
	}
=
	0
\end{split}
\end{equation}
\end{enumerate}
\cfout.
\end{lemma}

\begin{proof}[Proof of \cref{batch_norm_lemma}]
Note that 
\enum{
	\cref{batch_mean:eq1};
	\cref{batch_var:eq1};
	\cref{def:batch_norm_op:eq1};
	\cref{def:batch_norm:eq1}
}[establish]
\cref{batch_norm_lemma:item1}.
In addition, note that 
\enum{
	\cref{batch_norm_lemma:item1}
}[ensure]
that
for all
	$\varepsilon \in (0,\infty)$,
	$x = ((x^{(m)}_i)_{i \in \{1, 2, \ldots, d\}})_{m \in \{1, 2, \ldots, M\}}  \in (\R^d)^M$,
	$i \in \{1, 2, \ldots, d\}$
it holds that
\begin{equation}
\label{batch_norm_lemma:eq1}
\begin{split} 
	\batchmean_i(\Batchnorm_{\beta, \gamma, \varepsilon}(x))
&=
	\frac{1}{M}
	\sum_{m = 1}^M
		\pr[\Big]{
			\gamma_i
			\br[\Big]{
				\frac{x^{(m)}_i - \batchmean_i(x)}{\sqrt{\smash[b]{\batchvar_i(x) + \varepsilon}}}
			}
			+
			\beta_i
		}\\
&=
	\gamma_i
	\br[\Big]{
		\frac{
			\frac{1}{M}
			\pr[\big]{
				\sum_{m = 1}^M
					x^{(m)}_i
			}
			- 
			\batchmean_i(x)
		}{
			\sqrt{\smash[b]{\batchvar_i(x) + \varepsilon}}
		}
	}
	+
	\beta_i \\
&=
	\gamma_i
	\br[\Big]{
		\frac{
			\batchmean_i(x)
			- 
			\batchmean_i(x)
		}{
			\sqrt{\smash[b]{\batchvar_i(x) + \varepsilon}}
		}
	}
	+
	\beta_i
=
	\beta_i
\end{split}
\end{equation}
\cfload.
This implies \cref{batch_norm_lemma:item2}.
\Moreover
\enum{
	\cref{batch_norm_lemma:eq1};
	\cref{batch_norm_lemma:item1}
}[ensure]
that for all
	$\varepsilon \in (0,\infty)$,
	$x = ((x^{(m)}_i)_{i \in \{1, 2, \ldots, d\}})_{m \in \{1, 2, \ldots, M\}}  \in (\R^d)^M$,
	$i \in \{1, 2, \ldots, d\}$
it holds that
\begin{equation}
\begin{split} 
	&\batchvar_i(\Batchnorm_{\beta, \gamma, \varepsilon}(x))\\
&=
	\frac{1}{M}
	\sum_{m = 1}^M
		\br[\Big]{
			\gamma_i
			\br[\Big]{
				\frac{x^{(m)}_i - \batchmean_i(x)}{\sqrt{\smash[b]{\batchvar_i(x) + \varepsilon}}}
			}
			+
			\beta_i
			- 
			\batchmean_i(\Batchnorm_{\beta, \gamma, \varepsilon}(x))
		}^2 \\
&=
	\frac{1}{M}
	\sum_{m = 1}^M
			(\gamma_i)^2
			\br[\Big]{
				\frac{x^{(m)}_i - \batchmean_i(x)}{\sqrt{\smash[b]{\batchvar_i(x) + \varepsilon}}}
			}^2\\
&=
	(\gamma_i)^2
	\br[\Big]{
		\frac{
			\frac{1}{M}
			\sum_{m = 1}^M
			\pr{x^{(m)}_i - \batchmean_i(x)}^2
		}{
			\batchvar_i(x) + \varepsilon
		}
	}
=
	(\gamma_i)^2
	\br[\Big]{
		\frac{
			\batchvar_i(x)
		}{
			\batchvar_i(x) + \varepsilon
		}
	}.
\end{split}
\end{equation}
Combining this with the fact that for all
	$x = ((x^{(m)}_i)_{i \in \{1, 2, \ldots, d\}})_{m \in \{1, 2, \ldots, M\}} \in (\R^d)^M$,
	$i \in \{1, 2, \ldots, d\}$
with
	$\#(\bigcup_{m=1}^M \{x^{(m)}_i\})>1$	
it holds that 
\begin{equation}
\begin{split} 
	\batchvar_i(x) > 0
\end{split}
\end{equation}
implies \cref{batch_norm_lemma:item3}.
The proof of \cref{batch_norm_lemma} is thus complete.
\end{proof}

\section[Structured description of ANNs with BN (training)]{Structured description of 
ANNs with BN for training}

\begin{adef}{def:BNANN}[Structured description of fully-connected feedforward \anns\ with \BN]
We denote by $\BNANNs$ the set given by 
\begin{equation}
\begin{split}
\textstyle
   	\BNANNs
=
	\bigcup_{L \in \N}
	\bigcup_{ l_0,l_1,\ldots, l_L \in \N }
	\bigcup_{\Normlayers \subseteq \{0, 1, \ldots, L\}}
		\pr[\Big]{
			\pr[\big]{
			  \bigtimes_{k = 1}^L 
			  (
			     	\R^{l_{k} \times l_{k-1}} \times \R^{l_{k}}
			  )
			}
			\times
			\pr[\big]{
				\bigtimes_{k \in \Normlayers}
				(
					\R^{l_{k}}
				)^2
			}
		}.
\end{split}
\end{equation}
\end{adef}

\cfclear
\begin{adef}{def:bnanns}[Fully-connected feedforward \anns\ with \BN]
  We say that $\Phi$ is a fully-connected feedforward \ann\ with \BN\ 
  (we say that $\Phi$ is an \ann\ with \BN)
  if and only
  if it holds that 
  \begin{equation}
    \Phi \in \BNANNs
  \end{equation}
  \cfload.
\end{adef}

\section[Realizations of fully-connected feedforward ANNs with BN (training)]{Realizations of fully-connected feedforward ANNs with BN for training}

\begin{introductions}
In the next definition we apply the multi-dimensional version of \cref{def:multidim_version} with batches as input. For this we implicitly identify batches with matrices. This identification is exemplified in the following exercise.
\end{introductions} 

\cfclear
\begin{exercise}{ex:identification_batches_matrices}
Let 
	$l_0 = 2$,
	$l_1 = 3$,
	$M = 4$, 
	$W \in \R^{l_1 \times l_0}$,
	$B \in \R^{l_1}$,
	$y \in (\R^{l_0})^M$,
	$x \in (\R^{l_1})^M$
satisfy
\begin{equation}
\begin{split} 
	W 
=
	\begin{pmatrix}
		3 & -1\\ -1 & 3\\ 3 & -1
	\end{pmatrix},
\qquad
	B
=
	\begin{pmatrix}
		1\\ -1\\ 1
	\end{pmatrix},
\qquad
	y
=
	\pr*{
		\begin{pmatrix}	0\\1 \end{pmatrix}
		,
		\begin{pmatrix}	1\\0 \end{pmatrix}
		,
		\begin{pmatrix}	2\\-2 \end{pmatrix}
		,
		\begin{pmatrix}	-1\\1 \end{pmatrix}
	},
\end{split}
\end{equation}
and 
$
	x 
= 
	\multdim_{
		\rect
		,
		l_1
		,
		M
	}
	(
		W y + (B, B, B, B)
	)
$
\cfload.
Prove the following statement:
It holds that
\begin{equation}
\begin{split} 
	x
=
	\pr*{
		\begin{pmatrix}	0\\2\\0 \end{pmatrix}
		,
		\begin{pmatrix}	4\\0\\4 \end{pmatrix}
		,
		\begin{pmatrix}	9\\0\\9 \end{pmatrix}
		,
		\begin{pmatrix}	0\\3\\0 \end{pmatrix}
	}.
\end{split}
\end{equation}
\end{exercise}

\begin{adef}{def:BNANNrealisation}[Realizations associated to fully-connected feedforward \anns\ with \BN]
Let $\varepsilon \in (0,\infty)$, $a\in C(\R,\R)$.
Then we denote by 
\begin{equation}
\begin{split} \textstyle
	\BNANNRealisation{a, \varepsilon} 
\colon 
		\BNANNs
	\to 
		\bigl(\bigcup_{k,l\in\N}\,
			C(
				{\scriptstyle\bigcup}_{M \in \N}(\R^k)^M, \allowbreak
				{\scriptstyle\bigcup}_{M \in \N}(\R^l)^M
			)
		\bigr)
\end{split}
\end{equation}
the function which satisfies
for all 
	$L, M \in\N$, $l_0,l_1,\ldots, l_L \in \N$, 
	$\Normlayers \subseteq \{0, 1, \ldots, L\}$,
	$
		\Phi 
	=
		(
			((W_k, B_k))_{k \in \{1, 2, \ldots, L\}}, 
			((\beta_k, \gamma_k))_{k \in \Normlayers}
		)
		\in \allowbreak
		\pr[\big]{
			\bigtimes_{k = 1}^L 
				\big(
					\R^{l_{k} \times l_{k-1}} \times \R^{l_{k}}
				\big)
		}
		\times
		\pr[\big]{
			\bigtimes_{k \in \Normlayers}
				(
					\R^{l_{k}}
				)^2
		}
	$,
	$x_0, y_0 \in (\R^{l_0})^M$, 
	$x_1, y_1 \in (\R^{l_1})^M$, 
	$\ldots$, 
	$x_L, y_L \in (\R^{l_L})^M$
with
\begin{equation}
\label{def:BNANNrealisation:eq1}
\begin{split}
	\forall \, k \in  \{0, 1, \ldots, L\} 
\colon \qquad
		y_k 
	= 
		\begin{cases}
			\Batchnorm_{\beta_k, \gamma_k, \varepsilon}(x_k) &\colon k \in \Normlayers \\
			x_k &\colon k \notin \Normlayers
		\end{cases}
\qand
\end{split}
\end{equation}
\begin{equation}
\label{def:BNANNrealisation:eq2}
\begin{split}  
	\forall \, k \in  \{1, 2, \ldots, L\} 
\colon \qquad
		x_k
	=
		\multdim_{
			a \indicator{(0,L)}(k)
			+
			\id_\R \indicator{\{L\}}(k)
			,
			l_k
			,
			M
		}
		(
			W_k y_{k-1} + (B_k, B_k, \ldots, B_k)
		)
\end{split}
\end{equation}
that
\begin{equation}
\label{def:BNANNrealisation:eq3}
	\BNANNRealisation{a, \varepsilon}(\Phi) 
\in 
	C(
		{\scriptstyle\bigcup}_{B \in \N}(\R^{l_0})^B, 
		{\scriptstyle\bigcup}_{B \in \N}(\R^{l_L})^B
	)
\qandq
	\pr[\big]{ \BNANNRealisation{a, \varepsilon}(\Phi) } (x_0) 
= 
	y_L
\in 
	(\R^{l_L})^M
\end{equation}
and
for every
	$\Phi \in \BNANNs$ 
we call $\BNANNRealisation{a, \varepsilon}(\Phi)$ the realization function of the fully-connected feedforward \ann\ with \BN\ $\Phi$
with activation function $a$ and \BN\ regularization parameter $\varepsilon$
	(we call $\BNANNRealisation{a, \varepsilon}(\Phi)$ the realization of the fully-connected feedforward \ann\ with \BN\ $\Phi$
with activation $a$ and \BN\ regularization parameter $\varepsilon$)
(cf.\ \cref{def:BNANN,def:batch_norm,def:multidim_version}).
\end{adef}

\section[Structured description of ANNs with BN (inference)]{Structured description of 
ANNs with BN for inference}

\begin{adef}{def:BNANNData}[Structured description of fully-connected feedforward \anns\ with \BN\ for given batch means and batch variances]
We denote by $\BNANNsData$ the set given by
\begin{equation}
\textstyle
   	\BNANNsData
=
	\bigcup_{ L \in \N}
	\bigcup_{ l_0,l_1,\ldots, l_L \in \N }
	\bigcup_{\Normlayers \subseteq \{0, 1, \ldots, L\}}
\textstyle
		\pr[\Big]{
			\pr[\big]{
			  \bigtimes_{k = 1}^L 
			  (
			     	\R^{l_{k} \times l_{k-1}} \times \R^{l_{k}}
			  )
			}
			\times
			\pr[\big]{
				\bigtimes_{k \in \Normlayers}
				(
				  (
			  		\R^{l_{k}}
				)^3
				\times [0,\infty)^{ l_k }
				)
			}
		}.
\end{equation}
\end{adef}

\cfclear
\begin{adef}{def:bnannsdata}[Fully-connected feedforward \anns\ with \BN\ for given batch means and batch variances]
  We say that $\Phi$ is a fully-connected feedforward \ann\ with \BN\ for given batch means and batch variances 
  (we say that $\Phi$ is an \ann\ with \BN\ for given batch means and batch variances)
  if and only
  if it holds that 
  \begin{equation}
    \Phi\in\BNANNsData
  \end{equation}
  \cfload.
\end{adef}

\section[Realizations of ANNs with BN (inference)]{Realizations of ANNs with BN for inference}

\begin{adef}{def:BNANNgivenrealisation}[Realizations associated to fully-connected feedforward \anns\ with \BN\ for given batch means and batch variances]
Let $\varepsilon \in (0,\infty)$, $a\in C(\R,\R)$.
Then we denote by 
\begin{equation}
\begin{split} \textstyle
	\BNANNDataRealisation{a, \varepsilon} 
\colon 
		\BNANNsData
	\to 
		\bigl(\bigcup_{k,l\in\N}\,
			C(
				\R^k, \allowbreak
				\R^l
			)
		\bigr)
\end{split}
\end{equation}
the function which satisfies
for all 
	$ L \in \N $, $l_0,l_1,\ldots, l_L \in \N$, 
	$\Normlayers \subseteq \{0, 1, \ldots, L\}$,
	$
		\Phi 
	=
	\allowbreak 
		(
			((
			  W_k, \allowbreak 
			  B_k))_{k \in \{1, 2, \ldots, L\}}, 
			((\beta_k, \gamma_k, \mu_k, \variance_k))_{k \in \Normlayers}
		)
		\in \allowbreak
		\pr[\big]{
			\bigtimes_{k = 1}^L 
				\big(
					\R^{l_{k} \times l_{k-1}} \times \R^{l_{k}}
				\big)
		}
		\times
		\pr[\big]{
			\bigtimes_{k \in \Normlayers}
			(
				(
					\R^{l_{k}}
				)^3
				\times 
				[0,\infty)^{ l_k }
            )
		}
	$,
	$x_0, y_0 \in \R^{l_0}$, 
	$x_1, y_1 \in \R^{l_1}$, 
	$\ldots$, 
	$x_L, y_L \in \R^{l_L}$
with
\begin{equation}
\label{def:BNANNgivenrealisation:eq1}
\begin{split}
	\forall \, k \in  \{0, 1, \ldots, L\} 
\colon \qquad
		y_k 
	= 
		\begin{cases}
			\Batchnormop_{\beta_k, \gamma_k, \mu_k, \variance_k, \varepsilon}(x_k) &\colon k \in \Normlayers \\
			x_k &\colon k \notin \Normlayers
		\end{cases}
\qand
\end{split}
\end{equation}
\begin{equation}
\label{def:BNANNgivenrealisation:eq2}
\begin{split}  
	\forall \, k \in  \{1, 2, \ldots, L\} 
\colon \qquad
		x_k
	=
		\multdim_{
			a \indicator{(0,L)}(k)
			+
			\id_\R \indicator{\{L\}}(k)
			,
			l_k
		}
		(
			W_k y_{k-1} + B_k
		)
\end{split}
\end{equation}
that
\begin{equation}
\label{def:BNANNgivenrealisation:eq3}
	\BNANNDataRealisation{a, \varepsilon}(\Phi) 
\in 
	C(
		\R^{l_0}, 
		\R^{l_L}
	)
\qandq
	\pr[\big]{ \BNANNDataRealisation{a, \varepsilon}(\Phi) } (x_0) 
= 
	y_L
\end{equation}
and
for every
	$\Phi \in \BNANNsData$ 
we call $\BNANNDataRealisation{a, \varepsilon}(\Phi)$ the realization function of the fully-connected feedforward \ann\ with \BN\ for given batch means and batch variances $\Phi$
with activation function $a$ and \BN\ regularization parameter $\varepsilon$
(cf.\ \cref{def:BNANNData,def:batch_norm_op}).
\end{adef}

\section{On the connection between BN for training and BN for inference}

\cfclear
\begin{adef}{def:BNANNsDataAssociated}[Fully-connected feed-forward \anns\ with \BN\ for given batch means and batch variances associated to fully-connected feedforward \anns\ with \BN\ and given input batches]
Let 
	$\varepsilon \in (0,\infty)$, 
	$a\in C(\R,\R)$,
	$L, M \in\N$, $l_0,l_1,\ldots, l_L \in \N$, 
	$\Normlayers \subseteq \{0, 1, \ldots, L\}$,
	$
		\Phi 
	=
		(
			((W_k, B_k))_{k \in \{1, 2, \ldots, L\}}, 
			((\beta_k, \gamma_k))_{k \in \Normlayers}
		)
		\in \allowbreak
		\pr[\big]{
			\bigtimes_{k = 1}^L 
				\big(
					\R^{l_{k} \times l_{k-1}} \times \R^{l_{k}}
				\big)
		}
		\times
		\pr[\big]{
			\bigtimes_{k \in \Normlayers}
				(
					\R^{l_{k}}
				)^2
		}
	$,
	$\fx \in (\R^{l_0})^M$.
Then we say that $\Psi$ is the fully-connected feedforward \anns\ with \BN\ for given batch means and batch variances associated to $(\Phi, \fx, a, \varepsilon)$
if and only if there exists
	$x_0, y_0 \in (\R^{l_0})^M$, 
	$x_1, y_1 \in (\R^{l_1})^M$, 
	$\ldots$, 
	$x_L, y_L \in (\R^{l_L})^M$
such that
\begin{enumerate}[label=(\roman *)]
\item 
\label{def:BNANNsDataAssociated:item1}
it holds that $x_0 = \fx$,
	
\item 
\label{def:BNANNsDataAssociated:item2}
it holds for all
	$k \in  \{0, 1, \ldots, L\} $
that
\begin{equation}
\label{def:BNANNsDataAssociated:eq1}
\begin{split}
	y_k 
= 
	\begin{cases}
		\Batchnorm_{\beta_k, \gamma_k, \varepsilon}(x_k) &\colon k \in \Normlayers \\
		x_k &\colon k \notin \Normlayers,
	\end{cases}
\end{split}
\end{equation}

\item 
\label{def:BNANNsDataAssociated:item3}
it holds for all
	$k \in  \{1, 2, \ldots, L\}$
that
\begin{equation}
\label{def:BNANNsDataAssociated:eq2}
\begin{split}  
		x_k
	=
		\multdim_{
			a \indicator{(0,L)}(k)
			+
			\id_\R \indicator{\{L\}}(k)
			,
			l_k
			,
			M
		}
		(
			W_k y_{k-1} + (B_k, B_k, \ldots, B_k)
		),
\end{split}
\end{equation}
and
\item 
\label{def:BNANNsDataAssociated:item4}
it holds that
\begin{equation}
\label{def:BNANNsDataAssociated:eq3}
\begin{split} 
	\Psi
&=
	(
		((W_k, B_k))_{k \in \{1, 2, \ldots, L\}}, 
		((\beta_k, \gamma_k, \batchmean(x_k), \batchvar(x_k)))_{k \in \Normlayers}
	)\\
&\in
 \allowbreak
	\pr[\big]{
		\textstyle
		\bigtimes_{k = 1}^L 
			(
				\R^{l_{k} \times l_{k-1}} \times \R^{l_{k}}
			)
	}
	\times
	\pr[\big]{
		\bigtimes_{k \in \Normlayers}
			(
				\R^{l_{k}}
			)^4
	}
\end{split}
\end{equation}
\end{enumerate}
\cfload.
\end{adef}

\cfclear
\begin{lemma}
\label{BNANNsDataAssociated_property}
Let 
	$\varepsilon \in (0,\infty)$, 
	$a\in C(\R,\R)$,
	$L, M \in\N$, $l_0,l_1,\ldots, l_L \in \N$, 
	$\Normlayers \subseteq \{0, \allowbreak 1, \ldots, \allowbreak L\}$,
	$
		\Phi 
	=
		(
			((W_k, B_k))_{k \in \{1, 2, \ldots, L\}}, 
			((\beta_k, \gamma_k))_{k \in \Normlayers}
		)
		\in \allowbreak
		\pr[\big]{
			\bigtimes_{k = 1}^L 
				\big(
					\R^{l_{k} \times l_{k-1}} \times \R^{l_{k}}
				\big)
		}
		\times\allowbreak
		\pr[\big]{
			\bigtimes_{k \in \Normlayers}
				(
					\R^{l_{k}}
				)^2
		}
	$,
	$x = (x^{(m)})_{m \in \{1, 2, \ldots, M\}} \in (\R^{l_0})^M$
and let $\Psi$ be the fully-connected feedforward \ann\ with \BN\ for given batch means and batch variances associated to $(\Phi, x, a, \varepsilon)$ \cfadd{def:BNANNsDataAssociated}\cfload.
Then
\begin{equation}
\label{BNANNsDataAssociated_property:concl1}
\begin{split} 
	(\BNANNRealisation{a, \varepsilon}(\Phi))(x)
=
	((\BNANNDataRealisation{a, \varepsilon}(\Psi))(x^{(m)}))_{m \in \{1, 2, \ldots, M\}}
\end{split}
\end{equation}
\cfout.
\end{lemma}

\begin{proof}[Proof of \cref{BNANNsDataAssociated_property}]
Observe that
	\cref{def:BNANNrealisation:eq1},
	\cref{def:BNANNrealisation:eq2},
	\cref{def:BNANNrealisation:eq3},
	\cref{def:BNANNgivenrealisation:eq1},
	\cref{def:BNANNgivenrealisation:eq2},
	\cref{def:BNANNgivenrealisation:eq3},
	\cref{def:BNANNsDataAssociated:eq1},
	\cref{def:BNANNsDataAssociated:eq2}, and
	\cref{def:BNANNsDataAssociated:eq3}
establish
\cref{BNANNsDataAssociated_property:concl1}.
The proof of \cref{BNANNsDataAssociated_property} is thus complete.
\end{proof}

\cfclear
\begin{exercise}{ex:compute_realization_batchnorm}
Let 
	$l_0 = 2$,
	$l_1 = 3$,
	$l_2 = 1$,
	$N = \{0, 1\}$,
	$\gamma_0 =(2 , 2)$,
	$\beta_0 =(0 , 0)$,
	$\gamma_1=(1 , 1 , 1)$,
	$\beta_1=(0 , 1 , 0)$,
	$
		x 
	= 
		(
			(0, 1),
			(1, 0),
			(-2, 2),
			(2, -2)
		)
	$,
$
	\Phi
\in 
	\BNANNs
$
satisfy
\begin{equation}
\begin{split} 
	\Phi
&=
	\pr*{
		\pr*{
			\pr*{
				\begin{pmatrix}
					1 & 2 \\
					3 & 4 \\
					5 & 6 
				\end{pmatrix},
				\begin{pmatrix}
					-1\\
					-1
				\end{pmatrix}
			}
			,
			\pr*{
				\begin{pmatrix}
					-1 & 1 & -1 \\
					1 & -1 & 1
				\end{pmatrix},
				\begin{pmatrix}
					-2
				\end{pmatrix}
			}
		}
		,
		((\gamma_k, \beta_k))_{k \in \Normlayers}
	}\\
&\in
	\pr[\big]{
	\textstyle
		\bigtimes_{k = 1}^2
			\big(
				\R^{l_{k} \times l_{k-1}} \times \R^{l_{k}}
			\big)
	}
	\times
	\pr[\big]{
		\bigtimes_{k \in \Normlayers}
			(
				\R^{l_{k}}
			)^2
	}
\end{split}
\end{equation}
and let $\Psi \in \BNANNsData$ be the fully-connected feedforward \anns\ with \BN\ for given batch means and batch variances associated to $(\Phi, x, \rect, 0.01)$.\cfadd{def:BNANNsDataAssociated}
Compute $(\BNANNRealisation{\rect, \frac{1}{100}}(\Phi) ) (x)$ and $(\BNANNDataRealisation{\rect, \frac{1}{100}}(\Psi) ) (-1, 1)$ explicitly and prove that your results are correct \cfload!
\end{exercise}

%% file: parts/Optimization_through_random_initializations.tex
\cchapter{Optimization through random initializations}{chapter:optimization_error}

In addition to minimizing an objective function through iterative steps of an \SGD-type optimization method,
another approach to minimize an objective function is 
	to sample different random initializations, 
	to iteratively calculate \SGD\ optimization processes starting at these random initializations,
and, 
	thereafter, 
	to pick a \SGD\ trajectory with the smallest final evaluation of the objective function. 
The approach to consider different random initializations is reviewed and analyzed within this chapter in detail.
The specific presentation of this chapter is strongly based on Jentzen \& Welti~\cite[Section 5]{JentzenWelti2023}.

\section{Analysis of the optimization error}
 
\subsection{The complementary distribution function formula}
\label{sec:CplDistForm}
\begin{athm}{lemma}{lem:moment_formula}[Complementary distribution function formula]
Let $ \mu \colon \mathcal{B}( [0,\infty) ) \to [0,\infty] $ be a sigma-finite measure.
Then
\begin{equation}
  \int_0^{ \infty } x \, \mu( \diff x )
  =
  \int_0^{ \infty }
  \mu( [x,\infty) )
  \, \diff x
  =
  \int_0^{ \infty }
  \mu( (x,\infty) )
  \, \diff x
  .
\end{equation}
\end{athm}
\begin{aproof}
First, \nobs that 
\begin{equation}
\label{eq:mu_calc_1}
\begin{split}
  \int_0^{ \infty } x \, \mu( \diff x )
& =
  \int_0^{ \infty } \br*{ \int_0^x \diff y } \mu( \diff x )  
  =
  \int_0^{ \infty } \br*{ \int_0^{ \infty } \ind{ (-\infty,x] }( y ) \, \diff y } \mu( \diff x )  
\\ & =
  \int_0^{ \infty } \int_0^{ \infty } \ind{ [y,\infty) }( x ) \, \diff y \, \mu( \diff x )  
  .
\end{split}
\end{equation}
\Moreover the fact that
$
  [0,\infty)^2 
  \ni
  (x,y)
  \mapsto 
  \ind{ 
    [ y, \infty )
  }( x )
  \in \R
$
is $ ( \mathcal{B}( [0,\infty) ) \otimes \mathcal{B}( [0,\infty) ) ) $/$ \mathcal{B}( \R ) 
$-measurable,
the assumption that $ \mu $ is a sigma-finite measure,
and Fubini's theorem 
\prove that
\begin{equation}
\begin{split}
  \int_0^{ \infty } \int_0^{ \infty } \ind{ [y,\infty) }( x ) \, \diff y \, \mu( \diff x )  
&
=
  \int_0^{ \infty } \int_0^{ \infty } \ind{ [y,\infty) }( x ) \, \mu( \diff x ) \, \diff y 
=
  \int_0^{ \infty } \mu( [y,\infty) ) \, \diff y  .
\end{split}
\end{equation}
Combining this with \eqref{eq:mu_calc_1} \proves
that for all
	$\varepsilon \in (0,\infty)$
it holds that
\begin{equation}
\begin{split}
  \int_0^{ \infty } x \, \mu( \diff x )
& =
  \int_0^{ \infty } \mu( [y,\infty) ) \, \diff y 
\geq
  \int_0^{ \infty } \mu( (y,\infty) ) \, \diff y 
\\ & 
\geq
  \int_0^{ \infty } \mu( [y+\varepsilon,\infty) ) \, \diff y 
=
  \int_{ \varepsilon }^{ \infty } \mu( [y,\infty) ) \, \diff y 
  .
\end{split}
\end{equation}
Beppo Levi's monotone convergence theorem \hence \proves that
\begin{equation}
\begin{split}
  \int_0^{ \infty } x \, \mu( \diff x )
& =
  \int_0^{ \infty } \mu( [y,\infty) ) \, \diff y 
\geq
  \int_0^{ \infty } \mu( (y,\infty) ) \, \diff y 
\\ & 
\geq
  \sup_{ \varepsilon \in (0,\infty) }
  \br*{
    \int_{ \varepsilon }^{ \infty } \mu( [y,\infty) ) \, \diff y 
  }
\\ & =
  \sup_{ \varepsilon \in (0,\infty) }
  \br*{
    \int_{ 0 }^{ \infty } 
    \mu( [y,\infty) ) 
    \,
    \ind{
      ( \varepsilon, \infty)
    }( y )
    \, \diff y 
  }
=
  \int_{ 0 }^{ \infty } \mu( [y,\infty) ) \, \diff y 
  .
\end{split}
\end{equation}
\end{aproof}

\subsection{Estimates for the optimization error involving complementary distribution functions}

\begin{athm}{lemma}{lem:minimum_MC_Lp}
Let
$ ( E, \delta ) $
be a metric space,
let
$ x\in E$,
$ K \in \N $,
$ p, L \in ( 0, \infty ) $,
let
$ ( \Omega, \cF, \P ) $
be a probability space,
let
$ \emprisk \colon E \times \Omega \to \R $
be
$ ( \cB( E ) \otimes \cF ) $/$ \cB( \R ) $-measurable,
assume for all
$ y \in E $,
$ \omega \in \Omega $
that
$ \lvert \emprisk( x, \omega ) - \emprisk( y, \omega ) \rvert
\leq L \delta( x, y ) $,
and let
$ X_k \colon \Omega \to E $, $ k \in \{ 1, 2, \ldots, K \} $,
be i.i.d.\ random variables.
Then
\begin{equation}
\label{eq:minimum_MC_Lp}
\E\bigl[
    \min\nolimits_{ k \in \{ 1, 2, \ldots, K \} } \lvert \emprisk( X_k ) - \emprisk( x ) \rvert^p
\bigr]
\leq
L^p
\int_{0}^{\infty}
    [ \P( \delta( X_1, x ) > \varepsilon^{ \nicefrac{1}{p} } ) ]^K
\,\diff \varepsilon
.
\end{equation}
\end{athm}
\begin{aproof}
Throughout this proof, 
let
$ Y \colon \Omega \to [ 0, \infty ) $
satisfy for all
$ \omega \in \Omega $
that
$ Y( \omega ) = \min\nolimits_{ k \in \{ 1, 2, \ldots, K \} } [ \delta( X_k( \omega ), x ) ]^p $.
\Nobs that
the fact that
$ Y $ is a random variable,
the assumption that
$ \forall \, y \in E,
\, \omega \in \Omega \colon
\lvert \emprisk( x, \omega ) - \emprisk( y, \omega ) \rvert
\leq L \delta( x, y ) $,
and
\cref{lem:moment_formula}
\prove that
\begin{equation}
\label{eq:expectation_min}
\begin{split}
&
\E\bigl[
    \min\nolimits_{ k \in \{ 1, 2, \ldots, K \} } \lvert \emprisk( X_k ) - \emprisk( x ) \rvert^p
\bigr]
\leq
L^p
\, \E\bigl[
    \min\nolimits_{ k \in \{ 1, 2, \ldots, K \} } [ \delta( X_k, x ) ]^p
\bigr]
\\ &
=
L^p
\, \E[ Y ]
=
L^p
\int_{0}^{\infty}
y
\, \P_Y( \diff y )
=
L^p
\int_{0}^{\infty}
\P_Y( ( \varepsilon, \infty ) )
\,\diff \varepsilon
\\ &
=
L^p
\int_{0}^{\infty}
\P( Y > \varepsilon )
\,\diff \varepsilon
=
L^p
\int_{0}^{\infty}
    \P\bigl( \min\nolimits_{ k \in \{ 1, 2, \ldots, K \} } [ \delta( X_k, x ) ]^p > \varepsilon \bigr)
\,\diff \varepsilon
.
\end{split}
\end{equation}
\Moreover
the assumption that
$ X_k $, $ k \in \{ 1, 2, \ldots, K \} $,
are i.i.d.\ random variables
\proves that for all
$ \varepsilon \in ( 0, \infty ) $
it holds that
\begin{equation}
\label{eq:probability_min}
\begin{split}
&
\P\bigl( \min\nolimits_{ k \in \{ 1, 2, \ldots, K \} } [ \delta( X_k, x ) ]^p > \varepsilon \bigr)
=
\P\bigl(
    \forall \, k \in \{ 1, 2, \ldots, K \} \colon
    [ \delta( X_k, x ) ]^p > \varepsilon
\bigr)
\\ &
=
\smallprod_{ k = 1 }^K
    \P( [ \delta( X_k, x ) ]^p > \varepsilon )
=
[ \P( [ \delta( X_1, x ) ]^p > \varepsilon ) ]^K
=
[ \P( \delta( X_1, x ) > \varepsilon^{ \nicefrac{1}{p} } ) ]^K
.
\end{split}
\end{equation}
Combining this with \cref{eq:expectation_min}
proves \cref{eq:minimum_MC_Lp}.
\end{aproof}

\section{Strong convergences rates for the optimization error}

\subsection{Properties of the gamma and the beta function}

\begin{athm}{lemma}{lem:Gamma_basic}
Let
$ \Gamma \colon ( 0, \infty ) \to ( 0, \infty ) $ and
$ \bbB \colon ( 0, \infty )^2 \to ( 0, \infty ) $
satisfy for all
$ x, y \in ( 0, \infty ) $
that
$ \Gamma( x ) = \int_0^{ \infty } t^{ x - 1 } e^{ - t } \,\diff t $
and
$ \bbB( x, y )
=
\int_{0}^{1}
    t^{ x - 1 } ( 1 - t )^{ y - 1 }
\,\diff t $.
Then
\begin{enumerate}[label=(\roman *)]
\item
\label{item:lem:Gamma_basic:1}
it holds for all
$ x \in ( 0, \infty ) $
that
$ \Gamma( x + 1 ) = x \, \Gamma( x ) $,
\item
\label{item:lem:Gamma_basic:2}
it holds that
$ \Gamma(1) = \Gamma(2) = 1 $,
and
\item
\label{item:lem:Gamma_basic:3}
it holds for all
$ x, y \in ( 0, \infty ) $
that
$ \bbB( x, y )
=
\frac{ \Gamma( x ) \Gamma( y ) }{ \Gamma( x + y ) } $.
\end{enumerate}
\end{athm}

\begin{aproof}
Throughout this proof, 
let $x,y\in(0,\infty)$, 
let 
$ 
  \Phi \colon 
  (0,\infty) \times (0,1)
  \to
  (0,\infty)^2 
$
satisfy
for all $ u \in (0,\infty)$, $ v \in (0,1) $
that
\begin{equation}
  \Phi( u, v )
  =
  ( u ( 1 - v ), uv )
  ,
\end{equation}
and let 
$f\colon (0,\infty)^2 \to (0,\infty)$
satisfy for all 
$s,t \in (0,\infty)$
that 
\begin{equation}
  f(s,t)=s^{(x-1)} \, t^{(y-1)} \, e^{-(s+t)}
  .
\end{equation}
\Nobs that 
  the integration by parts formula 
\proves that for all $ x \in (0,\infty) $
it holds that
\begin{equation}
\begin{split}
&
  \Gamma( x + 1 )
  =
  \int_0^{ \infty }
  t^{ ( ( x + 1 ) - 1 ) }
  \,
  e^{ - t }
  \, \diff t
  =
  -
  \int_0^{ \infty }
  t^{ x }
  \br*{ 
    -
    e^{ - t }
  }
  \diff t
\\ & =
  -
  \pr*{
  \br*{ 
    t^x e^{ - t }
  }^{ t = \infty }_{ t = 0 }
  -
  x
  \int_0^{ \infty }
  t^{ ( x - 1 ) }
  \,	
  e^{ - t }
  \, \diff t
  }
=
  x
  \int_0^{ \infty }
  t^{ ( x - 1 ) }
  \,	
  e^{ - t }
  \, \diff t
=
  x \cdot \Gamma( x )
  .
\end{split}
\end{equation}
This \proves[ep] \cref{item:lem:Gamma_basic:1}.
\Moreover 
\begin{equation}
	\Gamma(1)
=
	\int_0^{ \infty } t^{ 0 } e^{ - t } \,\diff t 
=
	[-e^{ - t } ]_{t = 0}^{t = \infty} 
=
	1.
\end{equation}
This and \cref{item:lem:Gamma_basic:1} \prove[ep] \cref{item:lem:Gamma_basic:2}.
\Moreover the integral transformation 
theorem with the diffeomorphism 
$
  (1,\infty) \ni t \mapsto \frac{1}{t} \in (0,1)
$
\proves that
\begin{equation}
\begin{split}
  B( x, y )
& =
  \int_0^1
  t^{ ( x - 1 ) }
  \,
  \pr*{ 1 - t }^{ ( y - 1 ) }
  \diff t
=
  \int_1^{ \infty }
  \br*{ \tfrac{ 1 }{ t } }^{ ( x - 1 ) }
  \,
  \br*{ 1 - \tfrac{ 1 }{ t } }^{ ( y - 1 ) }
  \tfrac{ 1 }{ t^2 }
  \,
  \diff t
\\ & =
  \int_1^{ \infty }
  t^{
    \pr*{
      - x - 1
    }
  }
  \br*{ \tfrac{ t - 1 }{ t } }^{ ( y - 1 ) }
  \diff t
=
  \int_1^{ \infty }
  t^{
    \pr*{
      - x - y
    }
  }
  \pr*{ t - 1 }^{ ( y - 1 ) }
  \diff t
\\ & =
  \int_0^{ \infty }
  \pr*{ t + 1 }^{
    \pr*{
      - x - y
    }
  }
  t^{ ( y - 1 ) }
  \diff t
=
  \int_0^{ \infty }
  \frac{
    t^{ ( y - 1 ) }
  }{
    \pr*{ t + 1 }^{ ( x + y ) }
  }
  \,
  \diff t
  .
\end{split}
\end{equation}
\Moreover the fact that
for all 
$ (u,v) \in (0,\infty) \times (0,1) $
it holds that
\begin{equation}
  \Phi'(u,v)
  =
  \begin{bmatrix}
  1-v & -u \\
  v & u
  \end{bmatrix}
\end{equation}
\proves that for all 
$ (u,v) \in (0,\infty) \times (0,1) $
it holds that
\begin{equation}
  \operatorname{det}(\Phi'(u,v))
  =
  (1-v)u - v(-u)
  =
  u-vu+vu
  =
  u \in (0,\infty)
  .
\end{equation}
This, 
the fact that
\begin{equation}
  \label{eq:pre.transform}
  \begin{split}
  \Gamma( x ) 
  \cdot 
  \Gamma( y )
  & =
  \br*{
  \int_0^{ \infty }
  t^{ ( x - 1 ) }
  \,
  e^{ - t } \, \diff t
  }
  \br*{
  \int_0^{ \infty }
  t^{ ( y - 1 ) }
  \,
  e^{ - t } \, \diff t
  }
  \\
  & =
  \br*{
  \int_0^{ \infty }
  s^{ ( x - 1 ) }
  \,
  e^{ - s } \, \diff s
  }
  \br*{
  \int_0^{ \infty }
  t^{ ( y - 1 ) }
  \,
  e^{ - t } \, \diff t
  }
  \\ & 
  =
  \int_0^{ \infty }
  \int_0^{ \infty }
  s^{ ( x - 1 ) }
  \,
  t^{ ( y - 1 ) }
  \,
  e^{ - ( s + t ) } 
  \, \diff t \, \diff s
  \\&
  =
  \int_{(0,\infty)^2}
  f(s,t) 
  \, \diff (s,t)
  ,
  \end{split}
  \end{equation}
and the integral transformation 
theorem \prove that
\begin{equation}
\begin{split}
  \Gamma( x ) 
  \cdot 
  \Gamma( y )
 &=
 \int_{(0,\infty)\times(0,1)}
 f(\Phi(u,v)) \,
 \abs{\operatorname{det}(\Phi'(u,v))} 
 \, \diff (u,v)
 \\&=
 \int^\infty_0
 \int^1_0
 (u(1-v))^{(x-1)} \, (uv)^{(y-1)} \,
 e^{-(u(1-v)+uv)} \, u
 \, \diff v \, \diff u
 \\&=
 \int^\infty_0
 \int^1_0
 u^{(x+y-1)} \, e^{-u} \,
 v^{(y-1)} \, (1-v)^{(x-1)}
 \, \diff v \, \diff u
 \\&
 =
 \br*{
 \int^\infty_0
 u^{(x+y-1)} \, e^{-u}
 \, \diff u
 }
 \br*{
 \int^1_0
 v^{(y-1)} \, (1-v)^{(x-1)}
 \, \diff v
 }
 \\&=
 \Gamma(x+y) \, B(y,x).
\end{split}
\end{equation}
This establishes \cref{item:lem:Gamma_basic:3}.
\end{aproof}

\begin{athm}{lemma}{lem:unit_interval_basic}
It holds for all
$ \alpha, x \in [ 0, 1 ] $
that
$ ( 1 - x )^\alpha \leq 1 - \alpha x $.
\end{athm}
\begin{aproof}
\Nobs that
the fact that
for all
$ y \in [ 0, \infty ) $
it holds that
$ [ 0, \infty ) \ni z \mapsto y^z \in [ 0, \infty ) $
is convex
\proves that for all
$ \alpha, x \in [ 0, 1 ] $
it holds that
\begin{equation}
\begin{split}
( 1 - x )^\alpha
&
\leq
\alpha ( 1 - x )^1
+
( 1 - \alpha ) ( 1 - x )^0
\\ &
=
\alpha - \alpha x
+
1 - \alpha
=
1 - \alpha x
.
\end{split}
\end{equation}
\end{aproof}

\begin{athm}{prop}{prop:Gamma_function}
Let
$ \Gamma \colon ( 0, \infty ) \to ( 0, \infty ) $
and
$ \llfloor \cdot \rrfloor \colon ( 0, \infty ) \to \N_0 $
satisfy for all
$ x \in ( 0, \infty ) $
that
$ \Gamma( x ) =
\int_0^{ \infty } t^{ x - 1 } e^{ - t } \,\diff t $
and
$ \llfloor x \rrfloor = \max( [ 0, x ) \cap \N_0 ) $.
Then
\begin{enumerate}[label=(\roman *)]
\item
\label{item:prop:Gamma_function:1}
it holds that
$ \Gamma \colon ( 0, \infty ) \to ( 0, \infty ) $
is convex,
\item
\label{item:prop:Gamma_function:2}
it holds for all
$ x \in ( 0, \infty ) $
that
$
\Gamma( x + 1 )
=
x \, \Gamma( x )
\leq
x^{ \llfloor x \rrfloor }
\leq
\max\{ 1, x^x \}
$,
\item
\label{item:prop:Gamma_function:3}
it holds for all
$ x \in ( 0, \infty ) $,
$ \alpha \in [ 0, 1 ] $
that
\begin{equation}
\label{eq:Gamma_ratio}
( \max\{ x + \alpha - 1, 0 \} )^{ \alpha }
\leq
\frac{x}{( x + \alpha )^{ 1 - \alpha }}
\leq
\frac{ \Gamma( x + \alpha ) }{ \Gamma( x ) }
\leq
x^\alpha
,
\end{equation}
and
\item
\label{item:prop:Gamma_function:4}
it holds for all
$ x \in ( 0, \infty ) $,
$ \alpha \in [ 0, \infty ) $
that
\begin{equation}
( \max\{ x + \min\{ \alpha - 1, 0 \}, 0 \} )^{ \alpha }
\leq
\frac{ \Gamma( x + \alpha ) }{ \Gamma( x ) }
\leq
( x + \max\{ \alpha - 1, 0 \} )^{ \alpha }
.
\end{equation}
\end{enumerate}
\end{athm}
\begin{aproof}
Throughout this proof, let
$ \lfloor \cdot \rfloor \colon [ 0, \infty ) \to \N_0 $
satisfy for all
$ x \in [ 0, \infty ) $
that
$ \lfloor x \rfloor = \max( [ 0, x ] \cap \N_0 ) $.
\Nobs that
the fact that
for all
$ t \in ( 0, \infty ) $
it holds that
$ \R \ni x \mapsto t^x \in ( 0, \infty ) $
is convex 
\proves that for all
$ x, y \in ( 0, \infty ) $,
$ \alpha \in [ 0, 1 ] $
it holds that
\begin{equation}
\begin{split}
\Gamma( \alpha x + ( 1 - \alpha ) y )
& =
\int_0^{ \infty } t^{ \alpha x + ( 1 - \alpha ) y - 1 } e^{ - t } \,\diff t
=
\int_0^{ \infty } t^{ \alpha x + ( 1 - \alpha ) y } t^{ -1 } e^{ - t } \,\diff t
\\ &
\leq
\int_0^{ \infty } ( \alpha t^x + ( 1 - \alpha ) t^y ) t^{ -1 } e^{ - t } \,\diff t
\\ &
=
\alpha
\int_0^{ \infty } t^{ x - 1 } e^{ - t } \,\diff t
+
( 1 - \alpha )
\int_0^{ \infty } t^{ y - 1 } e^{ - t } \,\diff t
\\ &
=
\alpha
\, \Gamma( x )
+
( 1 - \alpha )
\Gamma( y )
.
\end{split}
\end{equation}
This \proves[ep]
\cref{item:prop:Gamma_function:1}.
\Moreover
\cref{item:lem:Gamma_basic:2} in \cref{lem:Gamma_basic}
and \cref{item:prop:Gamma_function:1} \prove that for all
$ \alpha \in [ 0, 1 ] $
it holds that
\begin{equation}
\Gamma( \alpha + 1 )
=
\Gamma( \alpha \cdot 2 + ( 1 - \alpha ) \cdot 1 )
\leq
\alpha
\, \Gamma( 2 )
+
( 1 - \alpha )
\Gamma( 1 )
=
\alpha + ( 1 - \alpha )
= 1
.
\end{equation}
This \proves for all
$ x \in ( 0, 1 ] $
that
\begin{equation}
\label{eq:Gamma_bound_small}
\Gamma( x + 1 )
\leq
1
=
x^{ \llfloor x \rrfloor }
=
\max\{ 1, x^x \}
.
\end{equation}
Induction,
\cref{item:lem:Gamma_basic:1} in \cref{lem:Gamma_basic},
and
the fact that
$ \forall \, x \in ( 0, \infty ) \colon
x - \llfloor x \rrfloor \in ( 0, 1 ] $
\hence
\prove that for all
$ x \in [ 1, \infty ) $
it holds that
\begin{equation}
\Gamma( x + 1 )
=
\biggl[
    \smallprod_{i=1}^{ \llfloor x \rrfloor }
    ( x - i + 1 )
\biggr]
\Gamma( x - \llfloor x \rrfloor + 1 )
\leq
x^{ \llfloor x \rrfloor }
\Gamma( x - \llfloor x \rrfloor + 1 )
\leq
x^{ \llfloor x \rrfloor }
\leq
x^x
=
\max\{ 1, x^x \}
.
\end{equation}
Combining this and \cref{eq:Gamma_bound_small} with
\cref{item:lem:Gamma_basic:1} in \cref{lem:Gamma_basic}
\proves[ep] \cref{item:prop:Gamma_function:2}.
\Moreover
H\"older's inequality
and
\cref{item:lem:Gamma_basic:1} in \cref{lem:Gamma_basic}
\prove that for all
$ x \in ( 0, \infty ) $,
$ \alpha \in [ 0, 1 ] $
it holds that
\begin{equation}
\label{eq:Gamma_ratio_UB}
\begin{split}
\Gamma( x + \alpha )
& =
\int_0^{ \infty } t^{ x + \alpha - 1 } e^{ - t } \,\diff t
=
\int_0^{ \infty }
    t^{ \alpha x } e^{ - \alpha t }
    t^{ ( 1 - \alpha ) x - ( 1 - \alpha ) } e^{ - ( 1 - \alpha ) t }
\,\diff t
\\ &
=
\int_0^{ \infty }
    [ t^{ x } e^{ - t } ]^\alpha
    [ t^{ x - 1 } e^{ - t } ]^{ 1 - \alpha }
\,\diff t
\\ &
\leq
\biggl(
    \int_0^{ \infty } t^{ x } e^{ - t } \,\diff t
\biggr)^{ \!\! \alpha }
\biggl(
    \int_0^{ \infty } t^{ x - 1 } e^{ - t } \,\diff t
\biggr)^{ \!\! 1 - \alpha }
\\ &
=
[ \Gamma( x + 1 ) ]^{ \alpha }
[ \Gamma( x ) ]^{ 1 - \alpha }
=
x^\alpha [ \Gamma( x ) ]^{ \alpha }
[ \Gamma( x ) ]^{ 1 - \alpha }
\\ &
=
x^\alpha
\Gamma( x )
.
\end{split}
\end{equation}
This
and
\cref{item:lem:Gamma_basic:1} in \cref{lem:Gamma_basic}
\prove that for all
$ x \in ( 0, \infty ) $,
$ \alpha \in [ 0, 1 ] $
it holds that
\begin{equation}
\label{eq:Gamma_ratio_LB}
x \, \Gamma( x )
=
\Gamma( x + 1 )
=
\Gamma( x + \alpha + ( 1 - \alpha ) )
\leq
( x + \alpha )^{ 1 - \alpha }
\Gamma( x + \alpha )
.
\end{equation}
Combining~\cref{eq:Gamma_ratio_UB} and~\cref{eq:Gamma_ratio_LB}
\proves that for all
$ x \in ( 0, \infty ) $,
$ \alpha \in [ 0, 1 ] $
it holds that
\begin{equation}
\label{eq:sharper_bound}
\frac{x}{( x + \alpha )^{ 1 - \alpha }}
\leq
\frac{ \Gamma( x + \alpha ) }{ \Gamma( x ) }
\leq
x^\alpha
.
\end{equation}
\Moreover
\cref{item:lem:Gamma_basic:1} in \cref{lem:Gamma_basic}
and \cref{eq:sharper_bound}
\prove that for all
$ x \in ( 0, \infty ) $,
$ \alpha \in [ 0, 1 ] $
it holds that
\begin{equation}
\frac{ \Gamma( x + \alpha ) }{ \Gamma( x + 1 ) }
=
\frac{ \Gamma( x + \alpha ) }{ x \, \Gamma( x ) }
\leq
x^{ \alpha - 1 }
.
\end{equation}
This \proves for all
$ \alpha \in [ 0, 1 ] $,
$ x \in ( \alpha, \infty ) $
that
\begin{equation}
\frac{ \Gamma( x ) }{ \Gamma( x + ( 1 - \alpha ) ) }
=
\frac{ \Gamma( ( x - \alpha ) + \alpha ) }{ \Gamma( ( x - \alpha ) + 1 ) }
\leq
( x - \alpha )^{ \alpha - 1 }
=
\frac{1}{( x - \alpha )^{ 1 - \alpha }}
.
\end{equation}
This, in turn,
\proves for all
$ \alpha \in [ 0, 1 ] $,
$ x \in ( 1 - \alpha, \infty ) $
that
\begin{equation}
( x + \alpha - 1 )^{ \alpha }
=
( x - ( 1 - \alpha ) )^{ \alpha }
\leq
\frac{ \Gamma( x + \alpha ) }{ \Gamma( x ) }
.
\end{equation}
\Moreover
\cref{lem:unit_interval_basic}
\proves that for all
$ x \in ( 0, \infty ) $,
$ \alpha \in [ 0, 1 ] $
it holds that
\begin{equation}
\begin{split}
( \max\{ x + \alpha - 1, 0 \} )^{ \alpha }
& =
( x + \alpha )^\alpha
\biggl(
    \frac{ \max\{ x + \alpha - 1, 0 \} }{ x + \alpha }
\biggr)^{ \!\! \alpha }
\\ &
=
( x + \alpha )^\alpha
\biggl(
    \max\biggl\{ 1 - \frac{1}{ x + \alpha }, 0 \biggr\}
\biggr)^{ \!\! \alpha }
\\ &
\leq
( x + \alpha )^\alpha
\biggl(
    1 - \frac{\alpha}{ x + \alpha }
\biggr)
=
( x + \alpha )^\alpha
\biggl(
    \frac{x}{ x + \alpha }
\biggr)
\\ &
=
\frac{x}{( x + \alpha )^{ 1 - \alpha }}
.
\end{split}
\end{equation}
This and
\cref{eq:sharper_bound}
\prove[ep]
\cref{item:prop:Gamma_function:3}.
\Moreover
induction,
\cref{item:lem:Gamma_basic:1} in \cref{lem:Gamma_basic},
the fact that
$ \forall \, \alpha \in [ 0, \infty ) \colon
\alpha - \lfloor \alpha \rfloor \in [ 0, 1 ) $,
and
\cref{item:prop:Gamma_function:3}
\prove that for all
$ x \in ( 0, \infty ) $,
$ \alpha \in [ 0, \infty ) $
it holds that
\begin{equation}
\label{eq:Gamma_ratio_UB_general}
\begin{split}
\frac{ \Gamma( x + \alpha ) }{ \Gamma( x ) }
& =
\biggl[
    \smallprod_{i=1}^{ \lfloor \alpha \rfloor }
    ( x + \alpha - i )
\biggr]
\frac{ \Gamma( x + \alpha - \lfloor \alpha \rfloor ) }{ \Gamma( x ) }
\leq
\biggl[
    \smallprod_{i=1}^{ \lfloor \alpha \rfloor }
    ( x + \alpha - i )
\biggr]
x^{ \alpha - \lfloor \alpha \rfloor }
\\ &
\leq
( x + \alpha - 1 )^{ \lfloor \alpha \rfloor }
x^{ \alpha - \lfloor \alpha \rfloor }
\\ &
\leq
( x + \max\{ \alpha - 1, 0 \} )^{ \lfloor \alpha \rfloor }
( x + \max\{ \alpha - 1, 0 \} )^{ \alpha - \lfloor \alpha \rfloor }
\\ &
=
( x + \max\{ \alpha - 1, 0 \} )^{ \alpha }
.
\end{split}
\end{equation}
\Moreover
the fact that
$ \forall \, \alpha \in [ 0, \infty ) \colon
\alpha - \lfloor \alpha \rfloor \in [ 0, 1 ) $,
\cref{item:prop:Gamma_function:3},
induction,
and
\cref{item:lem:Gamma_basic:1} in \cref{lem:Gamma_basic}
\prove that for all
$ x \in ( 0, \infty ) $,
$ \alpha \in [ 0, \infty ) $
it holds that
\begin{equation}
\begin{split}
\frac{ \Gamma( x + \alpha ) }{ \Gamma( x ) }
& =
\frac{ \Gamma( x + \lfloor \alpha \rfloor + \alpha - \lfloor \alpha \rfloor ) }{ \Gamma( x ) }
\\ &
\geq
( \max\{ x + \lfloor \alpha \rfloor + \alpha - \lfloor \alpha \rfloor - 1, 0 \} )^{ \alpha - \lfloor \alpha \rfloor }
\biggl[
    \frac{ \Gamma( x + \lfloor \alpha \rfloor ) }{ \Gamma( x ) }
\biggr]
\\ &
=
( \max\{ x + \alpha - 1, 0 \} )^{ \alpha - \lfloor \alpha \rfloor }
\biggl[
    \smallprod_{i=1}^{ \lfloor \alpha \rfloor }
    ( x + \lfloor \alpha \rfloor - i )
\biggr]
\frac{ \Gamma( x ) }{ \Gamma( x ) }
\\ &
\geq
( \max\{ x + \alpha - 1, 0 \} )^{ \alpha - \lfloor \alpha \rfloor }
x^{ \lfloor \alpha \rfloor }
\\ &
=
( \max\{ x + \alpha - 1, 0 \} )^{ \alpha - \lfloor \alpha \rfloor }
( \max\{ x, 0 \} )^{ \lfloor \alpha \rfloor }
\\ &
\geq
( \max\{ x + \min\{ \alpha - 1, 0 \}, 0 \} )^{ \alpha - \lfloor \alpha \rfloor }
( \max\{ x + \min\{ \alpha - 1, 0 \}, 0 \} )^{ \lfloor \alpha \rfloor }
\\ &
=
( \max\{ x + \min\{ \alpha - 1, 0 \}, 0 \} )^{ \alpha }
.
\end{split}
\end{equation}
Combining this
with
\cref{eq:Gamma_ratio_UB_general}
\proves[ep]
\cref{item:prop:Gamma_function:4}.
\end{aproof}

\begin{athm}{cor}{cor:Beta_function}
Let
$ \bbB \colon ( 0, \infty )^2 \to ( 0, \infty ) $
satisfy for all
$ x, y \in ( 0, \infty ) $
that
$ \bbB( x, y )
=
\int_{0}^{1}
    t^{ x - 1 } ( 1 - t )^{ y - 1 }
\,\diff t $
and
let
$ \Gamma \colon ( 0, \infty ) \to ( 0, \infty ) $
satisfy for all
$ x \in ( 0, \infty ) $
that
$ \Gamma( x ) = \int_0^{ \infty } t^{ x - 1 } e^{ - t } \,\diff t $.
Then it holds for all
$ x, y \in ( 0, \infty ) $
with
$ x + y > 1 $
that
\begin{equation}
\label{eq:cor:Beta_function}
\frac{\Gamma( x )}{ ( y + \max\{ x - 1, 0 \} )^{ x } }
\leq
\bbB( x, y )
\leq
\frac{\Gamma( x )}{ ( y + \min\{ x - 1, 0 \} )^{ x } }
\leq
\frac{ \max\{ 1, x^{ x } \} }{ x ( y + \min\{ x - 1, 0 \} )^{ x } }
.
\end{equation}
\end{athm}
\begin{aproof}
\Nobs that
\cref{item:lem:Gamma_basic:3} in \cref{lem:Gamma_basic}
\proves that for all
$ x, y \in ( 0, \infty ) $
it holds that
\begin{equation}
\label{eq:Beta_Gamma_relation}
\bbB( x, y )
=
\frac{ \Gamma( x ) \Gamma( y ) }{ \Gamma( y + x ) }
.
\end{equation}
\Moreover
the fact that for all
$ x, y \in ( 0, \infty ) $
with
$ x + y > 1 $
 it holds that
$ y + \min\{ x - 1, 0 \} > 0 $
and
\cref{item:prop:Gamma_function:4} in \cref{prop:Gamma_function}
\prove that for all
$ x, y \in ( 0, \infty ) $
with
$ x + y > 1 $
it holds that%
\begin{equation}
0 <
( y + \min\{ x - 1, 0 \} )^{ x }
\leq
\frac{ \Gamma( y + x ) }{ \Gamma( y ) }
\leq
( y + \max\{ x - 1, 0 \} )^{ x }
.
\end{equation}
Combining this
with \cref{eq:Beta_Gamma_relation}
and
\cref{item:prop:Gamma_function:2} in \cref{prop:Gamma_function}
\proves that for all
$ x, y \in ( 0, \infty ) $
with
$ x + y > 1 $
it holds that
\begin{equation}
\frac{\Gamma( x )}{ ( y + \max\{ x - 1, 0 \} )^{ x } }
\leq
\bbB( x, y )
\leq
\frac{\Gamma( x )}{ ( y + \min\{ x - 1, 0 \} )^{ x } }
\leq
\frac{ \max\{ 1, x^{ x } \} }{ x ( y + \min\{ x - 1, 0 \} )^{ x } }
.
\end{equation}
\end{aproof}

\subsection{Product measurability of continuous random fields}

\begin{athm}{lemma}{lem:projection_new}[Projections in metric spaces]
	Let $(E,d)$ be a metric space, 
	let $n \in \N$, $e_1, e_2, \ldots, e_n \in E$,
	and let $P \colon E \rightarrow E $
	satisfy for all $ x \in E $ that
	\begin{equation}
	\label{eq:projection_1_new} 
	P(x)
	=
	e_{
		\min\{ 
		k \in \{1,2,\ldots,n\} 
		\colon 
		d(x,e_k) 
		= 
		\min\{ y
		d(x,e_1), 
		d(x,e_2),
		\ldots,
		d(x,e_n)
		\}
		\}
	} .  
	\end{equation}
	Then
	\begin{enumerate}[label=(\roman *)]
		\item\label{it:projection_2_new} 
		it holds for all $ x \in E $ that
		\begin{equation}
		d(x, P(x))
		=
		\min_{k\in\{1,2,\ldots, n\}} d(x,e_k)
		\end{equation}
		and
		\item\label{it:projection_1_new} 
		it holds for all $A\subseteq E$ that 
		$P^{-1}(A)\in \B(E)$.
	\end{enumerate}
\end{athm}
\begin{aproof}
	Throughout this proof, let 
	$ D = (D_1, \ldots, D_n) \colon E \rightarrow \R^n $
	satisfy for all $ x \in E $
	that
	\begin{equation}\label{eq:definition_of_D}
	D(x)
	=
	\pr*{
	D_1(x),
	D_2(x),
	\ldots,
	D_n(x)
	}
	=
	\pr*{
	d(x, e_1),
	d(x, e_2),
	\ldots,
	d(x, e_n)
	} .
	\end{equation}
	\Nobs that \eqref{eq:projection_1_new} 
	\proves that for all $x\in E$ it holds that 
	\begin{eqsplit}
	d(x,P(x)) 
	&= 
	d(x,e_{
		\min\{ 
		k \in \{1,2,\ldots,n\} 
		\colon 
		d(x,e_k) 
		= 
		\min\{ 
		d(x,e_1), 
		d(x,e_2),
		\ldots,
		d(x,e_n)
		\}
		\}
	})
	\\&= 
	\min_{k\in\{1,2,\ldots,n\}} d(x,e_k) . 
	\end{eqsplit}
	This \proves[ep] \cref{it:projection_2_new}. 
	It thus remains to prove \cref{it:projection_1_new}. For this \nobs that the fact that 
	$d\colon E\times E\to [0,\infty)$ is continuous \proves 
	that $D\colon E\to\R^n$ is continuous. 
  \Hence that  $D\colon E\to \R^n$ is 
	$ \mathcal{B}(E)$/$\mathcal{B}(\R^n)$-measurable. 
	\Moreover \cref{it:projection_2_new} 
	\proves that for all $k\in \{1,2,\ldots,n\}$, 
	$x\in P^{-1}(\{e_k\})$ it holds that 
	\begin{equation}
	d(x,e_k) 
	= 
	d(x,P(x)) 
	= 
	\min_{l\in\{1,2,\ldots,n\}} d(x,e_l). 
	\end{equation}
	\Hence that for all $k\in\{1,2,\ldots,n\}$, 
	$x\in P^{-1}(\{e_k\})$ it holds that 
	\begin{equation}\label{eq:first_ineq_for_k_and_min}
	k\geq \min\{l\in\{1,2,\ldots,n\}\colon d(x,e_l) 
	= \min\{d(x,e_1),d(x,e_2),\ldots,d(x,e_n)\}\} . 
	\end{equation}
	\Moreover~\eqref{eq:projection_1_new} 
	\proves that for all 
	$ k \in \{1,2,\ldots,n\}$, $x\in P^{-1}(\{e_k\})$ 
	it holds that 
	\begin{align}
	\begin{split}
	&\min\cu*{l\in\{1,2,\ldots,n\}\colon d(x,e_l) 
	= \min_{u\in\{1,2,\ldots,n\}} d(x,e_u)} \\
	&\in 
	\bcu{ l \in \{1,2,\ldots,n\} \colon e_l = e_k}
	\subseteq\bcu{k, k+1, \ldots, n}. 
	\end{split}
	\end{align}
	\Hence that for all $k\in\{1,2,\ldots,n\}$, $x\in P^{-1}(\{e_k\})$  
	with $e_k\notin \bpr{\bigcup_{l\in \N\cap [0,k)} \{e_l\}}$ it holds that 
	\begin{equation}
	\min\cu*{l\in\{1,2,\ldots,n\}\colon d(x,e_l) 
	= \min_{u\in\{1,2,\ldots,n\}} d(x,e_u)} 
	\geq k. 
	\end{equation}
	Combining this with~\eqref{eq:first_ineq_for_k_and_min} \proves 
	that for all $k\in\{1,2,\ldots,n\}$, $x\in P^{-1}(\{e_k\})$ with 
	$e_k\notin  \bpr{\bigcup_{l\in \N\cap [0,k)} \{e_l\}}$  
	it holds that 
	\begin{equation}
	\min\cu*{l\in\{1,2,\ldots,n\}\colon d(x,e_l) 
	= \min_{u\in\{1,2,\ldots,n\}} d(x,e_u)} 
	=
	k . 
	\end{equation}
	\Hence that for all $k\in\{1,2,\ldots,n\}$ 
	with 
	$e_k\notin  \bpr{\bigcup_{l\in \N\cap [0,k)} \{e_l\}}$ it holds 
	that 
	\begin{equation}
	P^{-1}(\{e_k\}) \subseteq
	\cu*{x\in E\colon 
	\min\cu*{l\in\{1,2,\ldots,n\}\colon d(x,e_l) 
	= \min_{u\in\{1,2,\ldots,n\}} d(x,e_u)} 
	=
	k} . 
	\end{equation}
	This and \eqref{eq:projection_1_new} \prove that 
	for all $k\in\{1,2,\ldots,n\}$ with 
	$e_k\notin  \bpr{\bigcup_{l\in \N\cap [0,k)} \{e_l\}}$ it holds 
	that
	\begin{equation}
	P^{-1}(\{e_k\}) 
	=
	\cu*{x\in E\colon 
	\min\cu*{l\in\{1,2,\ldots,n\}\colon d(x,e_l) 
	= \min_{u\in\{1,2,\ldots,n\}} d(x,e_u)} 
	=
	k} .
	\end{equation}
	Combining \eqref{eq:definition_of_D} with 
	the fact that $D\colon E\to\R^n$ 
	is $\B(E)$/$\B(\R^n)$-measurable \hence \proves that 
	for all $ k \in \{1,2,\ldots,n\}$ with 
	$e_k \notin \bpr{\bigcup_{l \in \N \cap [0,k)} \{e_l\}} $
	it holds that
	\begin{align}
	\begin{split}
	&
	P^{-1}(\{e_k\})
	\\&=
	\cu*{ 
	x \in E 
	\colon
	\min\cu*{ 
	l \in \{1, 2, \ldots, n \}
	\colon
	d(x, e_l)
	=
	\min_{u\in\{1,2,\ldots,n\}} d( x, e_u )
	} 
	= k
	}
	\\&=
	\cu*{ 
	x \in E 
	\colon
	\min\cu*{ 
	l \in \{1, 2, \ldots, n \}
	\colon
	D_l(x)
	=
	\min_{ u \in \{1,2,\ldots,n\} }
	D_u(x)
	}
	= 
	k
	} 
	\\&=
	\cu*{ 
	x \in E 
	\colon
	\pr*{ 
	\begin{array}{c}
	\forall \, l \in\N\cap[0,k) 
	\colon
	D_k(x) < D_l(x) \,\, \text{and} \\
	\forall \, l \in \{1,2,\ldots,n\}
	\colon
	D_k(x) \leq D_l(x)
	\end{array}
	}
	} 
	\\&=
	\br*{ 
	\bigcap_{l=1}^{k-1}
	\underbrace{
		\{ 
		x \in E 
		\colon 
		D_k(x) < D_l(x) 
		\}
	}_{ \in \mathcal{B}(E) }
	}
	\bigcap 
	\br*{ 
	\bigcap_{l=1}^{n}
	\underbrace{
		\{ 
		x \in E 
		\colon 
		D_k(x) \leq D_l(x) 
		\}
	}_{ \in \mathcal{B}(E) }
	} 
	\in 
	\mathcal{B}(E) .
	\end{split}
	\end{align}
	\Hence that for all
	$ f \in \{e_1,e_2,\ldots,e_n\} $
	it holds that
	\begin{equation}
	P^{-1}(\{f\})
	\in
	\mathcal{B}(E) .
	\end{equation}
	\Hence that 
	for all $ A \subseteq E $ it holds that
	\begin{equation}
	P^{-1}(A)
	=
	P^{-1}\pr*{
	A \cap \{e_1,e_2,\ldots,e_n\}
	}
	=\textstyle
	\bigcup_{f \in A \cap \{e_1,e_2,\ldots,e_n\}}
	\underbrace{ 
		P^{-1}(\{f\})
	}_{ \in \mathcal{B}(E) }
	\in \mathcal{B}(E) .
	\end{equation}
	This \proves[ep] \cref{it:projection_1_new}.
\end{aproof}

\begin{athm}{lemma}{lem:productMeasurable_new}
	Let $ (E, d) $ be a separable metric space, 
	let $ (\cE, \delta) $ be a metric space,
	let $ (\Omega, \F) $ be a measurable space,
	let $ X \colon E \times \Omega \rightarrow \cE $,
	assume for all
	$ e \in E $ that 
	$ \Omega \ni \omega \mapsto X(e, \omega) \in \cE $
	is $\F$/$\mathcal{B}(\cE)$-measurable, 
	and assume for all $ \omega \in \Omega $
	that $ E \ni e \mapsto X(e, \omega) \in \cE $
	is continuous. Then
	$ X \colon E\times \Omega \to \cE $ is $(\mathcal{B}(E) \otimes \F)$/$\mathcal{B}(\cE) $-measurable.
\end{athm}

\begin{aproof}
	Throughout this proof, let
	$ e = (e_m)_{m \in \N} \colon \N \to E $ satisfy
	\begin{equation}
    \overline{\{ e_m \colon m \in \N \}} = E,  
  \end{equation}
	let 
	$ P_n \colon E \rightarrow E $, $ n \in \N $,
	satisfy for all $ n \in \N $, $ x \in E $ that
	\begin{align}
	P_n(x)
	=
	e_{
		\min\{ 
		k \in \{1,2,\ldots,n\} 
		\colon 
		d(x,e_k) 
		= 
		\min\{
		d(x,e_1), d(x,e_2), \ldots, d(x,e_n)
		\} 
		\}
	} ,
	\end{align}
	and let $\mathcal{X}_n \colon E \times \Omega \rightarrow \cE $, $ n \in \N $,
	satisfy for all $ n \in \N $, $ x \in E $, $ \omega \in \Omega $ that
	\begin{equation}
	\label{eq:curlyXn}
	\mathcal{X}_n(x, \omega)
	=
	X( P_n(x), \omega) .
	\end{equation}
  \Nobs that 
	\eqref{eq:curlyXn} \proves that for all $ n \in \N $, $ B \in \mathcal{B}(\cE) $
	it holds that
	\begin{align}
	(\mathcal{X}_n)^{-1}(B)
	&=
	\cu*{ 
	(x, \omega)
	\in E \times \Omega
	\colon
	\mathcal{X}_n(x, \omega) \in B
  }\notag
	\\&=
	\bigcup_{ y \in \operatorname{Im}(P_n) }
	\bbpr{
	\br*{ 
	(\mathcal{X}_n)^{-1}(B)
	}
	\cap 
	\br*{ 
	(P_n)^{-1}(\{y\}) \times \Omega
	} 
	}
	\\&=
	\bigcup_{ y \in \operatorname{Im}(P_n) }
	\cu*{ 
	(x, \omega)
	\in E \times \Omega
	\colon 
	\bbbr{
	\mathcal{X}_n(x, \omega) \in B
	\,\, \text{and} \,\,
	x \in (P_n)^{-1}(\{y\})
	}
  }\notag
	\\&=
	\bigcup_{ y \in \operatorname{Im}(P_n) }
	\cu*{ 
	(x, \omega)
	\in E \times \Omega
	\colon
	\bbbr{
	X(P_n(x), \omega) \in B
	\,\, \text{and} \,\,
	x \in (P_n)^{-1}(\{y\})
	}
	} 
	.\notag
	\end{align}
	\Cref{it:projection_1_new} in \cref{lem:projection_new}
	\hence \proves that for all $ n \in \N $, $ B \in \mathcal{B}(\cE) $
	it holds that
	\begin{align}
	(\mathcal{X}_n)^{-1}(B)
	&=
	\bigcup_{ y \in \operatorname{Im}(P_n) }
	\cu*{ 
	(x, \omega)
	\in E \times \Omega
	\colon
	\bbbr{
	X(y, \omega) \in B
	\,\, \text{and} \,\,
	x \in (P_n)^{-1}(\{y\})
	}
  } \notag
	\\&=
	\bigcup_{ y \in \operatorname{Im}(P_n) }
	\bbpr{
	\cu*{ 
	(x, \omega)
	\in E \times \Omega
	\colon
	X(y, \omega) \in B
	}
	\cap 
	\br*{ 
	(P_n)^{-1}(\{y\}) \times \Omega
	}
	}
	\\&=
	\bigcup_{ y \in \operatorname{Im}(P_n) }
	\bbpr{
	\bbr{ 
	\underbrace{
		E
		\times \pr*{
		\pr*{X(y, \cdot)}^{-1}(B)}
	}_{\in (\mathcal{B}(E) \otimes \F)}
	} 
	\cap 
	\bbr{ 
	\underbrace{
		(P_n)^{-1}(\{y\})
		\times 
		\Omega
	}_{\in (\mathcal{B}(E) \otimes \F)}
	}
	}
	\in (\mathcal{B}(E) \otimes \F) .\notag
	\end{align}
	This \proves that for all $ n \in \N $ it holds that
	$
	\mathcal{X}_n
	$
	is
	$
	(\mathcal{B}(E) \otimes \F)
	$/$
	\mathcal{B}(\cE)
	$-measurable.
	\Moreover 
	\cref{it:projection_2_new} in \cref{lem:projection_new}
	and the assumption that for all $\omega\in\Omega$ it holds that 
	 $E\ni x\mapsto X(x,\omega)\in\cE$ is continuous 
	\prove that for all $ x \in E $, $ \omega \in \Omega $
	it holds that
	\begin{equation}
	\label{eq:productMeasurable_5_new}
	\lim_{ n \rightarrow \infty }
	\mathcal{X}_n(x, \omega)
	=
	\lim_{ n \rightarrow \infty }
	X( P_n(x), \omega)
	=
	X(x, \omega) .
	\end{equation}
	Combining this with the fact that for all $n\in\N$ it holds that 
	$X_n\colon E\times\Omega\to\cE$ is 
	$(\B(E)\otimes\F)$/$\B(\cE)$-measurable 
	\proves that $X\colon E\times\Omega\to \cE$ 
	is $(\B(E)\otimes\F)$/$\B(\cE)$-measurable. 
\end{aproof}

\subsection{Strong convergences rates for the optimization error}

\cfclear
\begin{athm}{prop}{prop:minimum_MC_rate}
Let
$ \bfd, K \in \N $,
$ L, \alpha \in \R $,
$ \beta \in ( \alpha, \infty ) $,
let
$ ( \Omega, \cF, \P ) $
be a probability space,
let
$ \emprisk \colon [ \alpha, \beta ]^\bfd \times \Omega \to \R $
be a random field,
assume for all
$ \theta, \vartheta \in [ \alpha, \beta ]^\bfd $,
$ \omega \in \Omega $
that
$ \lvert \emprisk( \theta, \omega ) - \emprisk( \vartheta, \omega ) \rvert
\leq L \pnorm\infty{ \theta - \vartheta } $,
let
$ \Theta_k \colon \Omega \to [ \alpha, \beta ]^\bfd $, $ k \in \{ 1, 2, \ldots, K \} $,
be i.i.d.\ random variables,
and assume that
$ \Theta_1 $ is continuously uniformly distributed on $ [ \alpha, \beta ]^\bfd $
\cfload.
Then
\begin{enumerate}[label=(\roman *)]
\item
\label{item:prop:minimum_MC_rate:1}
it holds that
$ \emprisk $
is
$ ( \cB( [ \alpha, \beta ]^\bfd ) \otimes \cF ) $/$ \cB( \R ) $-measurable
and
\item
\label{item:prop:minimum_MC_rate:2}
it holds for all
$ \theta \in [ \alpha, \beta ]^\bfd $,
$ p \in ( 0, \infty ) $
that
\begin{equation}
\begin{split}
\bigl(
\E\bigl[
    \min\nolimits_{ k \in \{ 1, 2, \ldots, K \} } \lvert \emprisk( \Theta_k ) - \emprisk( \theta ) \rvert^p
\bigr]
\bigr)^{ \nicefrac{1}{p} }
&
\leq
\frac{ L ( \beta - \alpha ) \max\{ 1, ( \nicefrac{p}{\bfd} )^{ \nicefrac{1}{\bfd} } \} }{ K^{ \nicefrac{1}{\bfd} } }
\\&\leq
\frac{ L ( \beta - \alpha ) \max\{ 1, p \} }{ K^{ \nicefrac{1}{\bfd} } }
.
\end{split}
\end{equation}
\end{enumerate}
\end{athm}
\begin{aproof}
Throughout this proof,
assume without loss of generality that
$ L > 0 $,
let
$ \delta \colon
\allowbreak
( [ \alpha, \beta ]^\bfd ) \times ( [ \alpha, \beta ]^\bfd ) \to [ 0, \infty ) $
satisfy for all
$ \theta, \vartheta \in [ \alpha, \beta ]^\bfd $
that
\begin{equation}
  \delta( \theta, \vartheta ) = \infnorm { \theta - \vartheta },
\end{equation}
let
$ \bbB \colon ( 0, \infty )^2 \to ( 0, \infty ) $
satisfy for all
$ x, y \in ( 0, \infty ) $
that
\begin{equation}
  \bbB( x, y )
=
\int_{0}^{1}
    t^{ x - 1 } ( 1 - t )^{ y - 1 }
\,\diff t,
\end{equation}
and
let
$ \Theta_{ 1, 1 }, \Theta_{ 1, 2 }, \ldots, \Theta_{ 1, \bfd } \colon \Omega \to [ \alpha, \beta ] $
satisfy
$ \Theta_1 = ( \Theta_{ 1, 1 }, \Theta_{ 1, 2 }, \ldots, \Theta_{ 1, \bfd } ) $.
First,
\nobs that
the assumption that
for all $ \theta, \vartheta \in [ \alpha, \beta ]^\bfd$,
$\omega \in \Omega$ it holds that
\begin{equation}
  \lvert \emprisk( \theta, \omega ) - \emprisk( \vartheta, \omega ) \rvert
  \leq L \infnorm{ \theta - \vartheta }
\end{equation}
\proves that for all
$ \omega \in \Omega $
it holds that
$ [ \alpha, \beta ]^\bfd \ni \theta
\mapsto \emprisk( \theta, \omega ) \in \R $
is continuous.
Combining this with
the fact that
$ ( [ \alpha, \beta ]^\bfd, \delta ) $
is a separable metric space,
the fact that
for all
$ \theta \in [ \alpha, \beta ]^\bfd $
it holds that
$ \Omega \ni \omega
\mapsto \emprisk( \theta, \omega ) \in \R $
is $ \cF $/$ \cB( \R ) $-measurable,
and
\cref{lem:productMeasurable_new}
\proves[pe] \cref{item:prop:minimum_MC_rate:1}.
\Nobs that
the fact that for all
$ \theta \in [ \alpha, \beta ] $,
$ \varepsilon \in [ 0, \infty ) $
it holds that
\begin{equation}
\begin{split}
& \min\{ \theta + \varepsilon, \beta \} - \max\{ \theta - \varepsilon, \alpha \}
=
\min\{ \theta + \varepsilon, \beta \} + \min\{ \varepsilon - \theta, -\alpha \}
\\ &
=
\min\bigl\{
    \theta + \varepsilon + \min\{ \varepsilon - \theta, -\alpha \},
    \beta + \min\{ \varepsilon - \theta, -\alpha \}
\bigr\}
\\ &
=
\min\bigl\{
    \min\{ 2 \varepsilon, \theta - \alpha + \varepsilon \},
    \min\{ \beta - \theta + \varepsilon, \beta - \alpha \}
\bigr\}
\\ &
\geq
\min\bigl\{
    \min\{ 2 \varepsilon, \alpha - \alpha + \varepsilon \},
    \min\{ \beta - \beta + \varepsilon, \beta - \alpha \}
\bigr\}
\\ &
=
\min\{
    2 \varepsilon, \varepsilon,
    \varepsilon, \beta - \alpha
\}
=
\min\{ \varepsilon, \beta - \alpha \}
\end{split}
\end{equation}
and the assumption that
$ \Theta_1 $ is continuously uniformly distributed on $ [ \alpha, \beta ]^\bfd $
\prove that for all
$ \theta = ( \theta_1, \theta_2, \ldots, \theta_\bfd ) \in [ \alpha, \beta ]^\bfd $,
$ \varepsilon \in [ 0, \infty ) $
it holds that
\begin{equation}
\begin{split}
& \P( \infnorm{ \Theta_1 - \theta } \leq \varepsilon )
=
\P\bigl(
    \max\nolimits_{ i \in \{ 1, 2, \ldots, \bfd \} } \lvert \Theta_{ 1, i } - \theta_i \rvert \leq \varepsilon
\bigr)
\\ &
=
\P\bigl(
    \forall \, i \in \{ 1, 2, \ldots, \bfd \} \colon
    -\varepsilon \leq \Theta_{ 1, i } - \theta_i \leq \varepsilon
\bigr)
\\ &
=
\P\bigl(
    \forall \, i \in \{ 1, 2, \ldots, \bfd \} \colon
    \theta_i - \varepsilon \leq \Theta_{ 1, i } \leq \theta_i + \varepsilon
\bigr)
\\ &
=
\P\bigl(
    \forall \, i \in \{ 1, 2, \ldots, \bfd \} \colon
    \max\{ \theta_i - \varepsilon, \alpha \} \leq \Theta_{ 1, i } \leq \min\{ \theta_i + \varepsilon, \beta \}
\bigr)
\\ &
=
\P\bigl(
    \Theta_1 \in \bigl[ \textstyle\bigtimes_{ i = 1 }^\bfd [ \max\{ \theta_i - \varepsilon, \alpha \}, \min\{ \theta_i + \varepsilon, \beta \} ] \bigr]
\bigr)
\\ &
=
\tfrac{1}{ ( \beta - \alpha )^\bfd }
\smallprod_{ i = 1 }^\bfd
    ( \min\{ \theta_i + \varepsilon, \beta \} - \max\{ \theta_i - \varepsilon, \alpha \} )
\\ &
\geq
\tfrac{1}{ ( \beta - \alpha )^\bfd }
[ \min\{ \varepsilon, \beta - \alpha \} ]^\bfd
=
\min\Bigl\{
    1, \tfrac{ \varepsilon^\bfd }{ ( \beta - \alpha )^\bfd }
\Bigr\}
.
\end{split}
\end{equation}
\Hence
for all
$ \theta \in [ \alpha, \beta ]^\bfd $,
$ p \in ( 0, \infty ) $,
$ \varepsilon \in [ 0, \infty ) $
that
\begin{equation}
\begin{split}
&
\P( \infnorm{ \Theta_1 - \theta } > \varepsilon^{ \nicefrac{1}{p} } )
=
1 - \P( \infnorm{ \Theta_1 - \theta } \leq \varepsilon^{ \nicefrac{1}{p} } )
\\ &
\leq
1
-
\min\Bigl\{
    1, \tfrac{ \varepsilon^{ \nicefrac{\bfd}{p} } }{ ( \beta - \alpha )^\bfd }
\Bigr\}
=
\max\Bigl\{
    0, 1 - \tfrac{ \varepsilon^{ \nicefrac{\bfd}{p} } }{ ( \beta - \alpha )^\bfd }
\Bigr\}
.
\end{split}
\end{equation}
This,
\cref{item:prop:minimum_MC_rate:1},
the assumption that
for all $\theta, \vartheta \in [ \alpha, \beta ]^\bfd$,
$\omega \in \Omega$ it holds that
\begin{equation}
  \lvert \emprisk( \theta, \omega ) - \emprisk( \vartheta, \omega ) \rvert
\leq L \pnorm\infty{ \theta - \vartheta }
,
\end{equation}
the assumption that
$ \Theta_k $, $ k \in \{ 1, 2, \ldots, K \} $,
are i.i.d.\ random variables,
and
\cref{lem:minimum_MC_Lp}
(applied with
$ ( E, \delta ) \is ( [ \alpha, \beta ]^\bfd, \delta ) $,
$ ( X_k )_{ k \in \{ 1, 2, \ldots, K \} }
\is
( \Theta_k )_{ k \in \{ 1, 2, \ldots, K \} } $
in the notation of \cref{lem:minimum_MC_Lp})
\prove that for all
$ \theta \in [ \alpha, \beta ]^\bfd $,
$ p \in ( 0, \infty ) $
it holds that
\begin{equation}
\begin{split}
&
\E\bigl[
    \min\nolimits_{ k \in \{ 1, 2, \ldots, K \} } \lvert \emprisk( \Theta_k ) - \emprisk( \theta ) \rvert^p
\bigr]
\leq
L^p
\int_{0}^{\infty}
    [ \P( \infnorm{ \Theta_1 - \theta } > \varepsilon^{ \nicefrac{1}{p} } ) ]^K
\,\diff \varepsilon
\\ &
\leq
L^p
\int_{0}^{\infty}
    \Bigl[ \max\Bigl\{
        0, 1 - \tfrac{ \varepsilon^{ \nicefrac{\bfd}{p} } }{ ( \beta - \alpha )^\bfd }
    \Bigr\} \Bigr]^K
\,\diff \varepsilon
=
L^p
\int_{0}^{ ( \beta - \alpha )^p }
    \Bigl(
        1 - \tfrac{ \varepsilon^{ \nicefrac{\bfd}{p} } }{ ( \beta - \alpha )^\bfd }
    \Bigr)^{ \! K }
\,\diff \varepsilon
\\ &
=
\tfrac{p}{\bfd}
L^p
( \beta - \alpha )^p
\int_{0}^{ 1 }
    t^{ \nicefrac{p}{\bfd} - 1 }
    ( 1 - t )^K
\,\diff t
=
\tfrac{p}{\bfd}
L^p
( \beta - \alpha )^p
\int_{0}^{ 1 }
    t^{ \nicefrac{p}{\bfd} - 1 }
    ( 1 - t )^{ K + 1 - 1 }
\,\diff t
\\ &
=
\tfrac{p}{\bfd}
L^p
( \beta - \alpha )^p
\, \bbB( \nicefrac{p}{\bfd}, K + 1 )
.
\end{split}
\end{equation}
\Cref{cor:Beta_function}
(applied with
$ x \is \nicefrac{p}{\bfd} $,
$ y \is K + 1 $
for
$ p \in ( 0, \infty ) $
in the notation of \cref{eq:cor:Beta_function} in \cref{cor:Beta_function})
\hence \proves that for all
$ \theta \in [ \alpha, \beta ]^\bfd $,
$ p \in ( 0, \infty ) $
it holds that
\begin{equation}
\begin{split}
\E\bigl[
    \min\nolimits_{ k \in \{ 1, 2, \ldots, K \} } \lvert \emprisk( \Theta_k ) - \emprisk( \theta ) \rvert^p
\bigr]
 &
\leq
\frac{ \tfrac{p}{\bfd} L^p ( \beta - \alpha )^p \max\{ 1, ( \nicefrac{p}{\bfd} )^{ \nicefrac{p}{\bfd} } \}
}{
\tfrac{p}{\bfd} ( K + 1 + \min\{ \nicefrac{p}{\bfd} - 1, 0 \} )^{ \nicefrac{p}{\bfd} } }
\\&\leq
\frac{ L^p ( \beta - \alpha )^p \max\{ 1, ( \nicefrac{p}{\bfd} )^{ \nicefrac{p}{\bfd} } \}
}{
K^{ \nicefrac{p}{\bfd} } }
.
\end{split}
\end{equation}
This \proves for all
$ \theta \in [ \alpha, \beta ]^\bfd $,
$ p \in ( 0, \infty ) $
that
\begin{equation}
\begin{split}
\bigl(
\E\bigl[
    \min\nolimits_{ k \in \{ 1, 2, \ldots, K \} } \lvert \emprisk( \Theta_k ) - \emprisk( \theta ) \rvert^p
\bigr]
\bigr)^{ \nicefrac{1}{p} }
&
\leq
\frac{ L ( \beta - \alpha ) \max\{ 1, ( \nicefrac{p}{\bfd} )^{ \nicefrac{1}{\bfd} } \} }{ K^{ \nicefrac{1}{\bfd} } }
\\&\leq
\frac{ L ( \beta - \alpha ) \max\{ 1, p \} }{ K^{ \nicefrac{1}{\bfd} } }
.
\end{split}
\end{equation}
This \proves[ep] \cref{item:prop:minimum_MC_rate:2}.
\end{aproof}

\section{Strong convergences rates for the optimization error involving ANNs}

\subsection{Local Lipschitz continuity estimates for the parametrization functions of ANNs}
\label{sec:local_lip_cont_ANNs_param}

\begin{athm}{lemma}{lem:max_estimate}
Let $ a, x, y \in \R $. Then 
\begin{equation}
  \llabel{claim}
  \abs*{ \max\{ x, a \} - \max\{ y, a \} }
\leq
  \max\{ x, y \} - \min\{ x, y \}
=
  \abs{ x - y }
  .
\end{equation}
\end{athm}

\begin{aproof}
\Nobs that the fact that
\begin{equation}
\begin{split}
&
  \abs*{ \max\{ x, a \} - \max\{ y, a \} }
  =
  \abs*{
    \max\{ \max\{ x, y \}, a \} - \max\{ \min\{ x, y \}, a \} 
  }
\\ &
=
  \max\bigl\{ \max\{ x, y \}, a \bigr\} 
  - 
  \max\bigl\{ \min\{ x, y \}, a \bigr\} 
\\ &
=
  \max\Bigl\{ 
    \max\{ x, y \}
    - 
    \max\bigl\{ \min\{ x, y \}, a \bigr\} 
    , 
    a 
    - 
    \max\bigl\{ \min\{ x, y \}, a \bigr\} 
  \Bigr\} 
\\ &
\leq
  \max\Bigl\{ 
    \max\{ x, y \}
    - 
    \max\bigl\{ \min\{ x, y \}, a \bigr\} 
    , 
    a 
    - 
    a 
  \Bigr\} 
\\ & 
=
  \max\Bigl\{ 
    \max\{ x, y \}
    - 
    \max\bigl\{ \min\{ x, y \}, a \bigr\} 
    , 
    0
  \Bigr\} 
\leq
  \max\Bigl\{ 
    \max\{ x, y \}
    - 
    \min\{ x, y \}
    , 
    0
  \Bigr\} 
\\ &
=
  \max\{ x, y \}
  - 
  \min\{ x, y \}
=
  \abs*{
    \max\{ x, y \}
    - 
    \min\{ x, y \}
  }
=
  \abs*{ x - y }
  .
\end{split}
\end{equation}
\proves[ep] \lref{claim}.
\end{aproof}

\begin{athm}{cor}{cor:min_estimate}
Let $ a, x, y \in \R $. Then 
\begin{equation}
  \abs*{ \min\{ x, a \} - \min\{ y, a \} }
\leq
  \max\{ x, y \} - \min\{ x, y \}
=
  \abs{ x - y }
  .
\end{equation}
\end{athm}

\begin{aproof}
\Nobs that \cref{lem:max_estimate} \proves that
\begin{equation}
\begin{split}
  \abs*{ \min\{ x, a \} - \min\{ y, a \} }
& =
  \abs*{ - \pr*{ \min\{ x, a \} - \min\{ y, a \} } }
\\&  =
  \abs*{ \max\{ - x, - a \} - \max\{ - y, - a \} }
\\ & 
\leq
  \abs*{ ( - x ) - ( - y ) }
=
  \abs*{ x - y }
  .
\end{split}
\end{equation}
\end{aproof}

\cfclear
\begin{athm}{lemma}{result:ReLU_contraction}
Let $ d \in \N $. Then it holds for all $ x, y \in \R^d $ that
\begin{equation}
\label{eq:inf_norm}
  \infnorm{
    \Rect_d( x ) 
    -
    \Rect_d( y )
  }
  \leq 
  \infnorm{ x - y }
\end{equation}
\cfout.
\end{athm}
\begin{aproof}
  \Nobs[Observe] that \cref{lem:max_estimate} \proves[d] \cref{eq:inf_norm}.
\end{aproof}
  
\cfclear
\begin{athm}{lemma}{result:Clip_contraction}
  Let $d\in\N$, 
    $u\in[-\infty,\infty)$, 
    $v\in(u,\infty]$.
  Then it holds for all $x,y\in\R^d$ that
  \begin{equation}
    \infnorm{
      \Clip uvd(x)
      -
      \Clip uvd(y)
    }
    \leq
    \infnorm{x-y}
  \end{equation}
  \cfout.
\end{athm}
\begin{aproof}
  \Nobs that
    \cref{lem:max_estimate},
    \cref{cor:min_estimate},
    and the fact that for all
      $x\in\R$
    it holds that
      $\max\{-\infty,x\}=x=\min\{x,\infty\}$
  \prove that for all
    $x,y\in\R$
  it holds that
  \begin{eqsplit}
    \abs{\clip uv(x)-\clip uv(y)}
    &=
    \abs{
      \max\{u,\min\{x,v\}\}
      -
      \max\{u,\min\{y,v\}\}
    }
    \\&\leq
    \abs{
      \min\{x,v\}
      -
      \min\{y,v\}
    }
    \leq
    \abs{x-y}
  \end{eqsplit}
  \cfload.
  \Hence that for all
    $x=(x_1,x_2,\dots,x_d),y=(y_1,y_2,\dots,\allowbreak y_d)\in\R^d$
  it holds that
  \begin{eqsplit}
    \infnorm{\Clip uvd(x)-\Clip uvd(y)}
    &=
    \max_{i\in\{1,2,\dots,d\}} \abs{\clip uv(x_i)-\clip uv(y_i)}
    \\&\leq
    \max_{i\in\{1,2,\dots,d\}} \abs{x_i-y_i}
    =
    \infnorm{x-y}
  \end{eqsplit}
  \cfload.
\end{aproof}

\cfclear
\begin{athm}{lemma}{result:row_sum_norm}[Row sum norm, operator norm induced by the maximum norm]
Let $ a, b \in \N $, 
$ M = ( M_{ i, j } )_{ (i,j) \in \{ 1, 2, \dots, a \} \times \{ 1, 2, \dots, b \} } \in \R^{ a \times b } $.
Then 
\begin{equation}
  \sup_{ v \in \R^b \backslash \{ 0 \} }
  \br*{
    \frac{ 
      \infnorm{ M v }
    }{ 
      \infnorm{ v } 
    }
  }
=
  \max_{ i \in \{ 1, 2, \dots, a \} }
  \br*{
    \textstyle
    \sum\limits_{ j = 1 }^b
    \displaystyle
    \abs*{
      M_{ i, j } 
    }
  }
\leq 
  b 
  \br*{
    \max_{ i \in \{ 1, 2, \dots, a \} }
    \max_{ j \in \{ 1, 2, \dots, b \} }
    \abs*{
      M_{ i, j } 
    }
  }
\end{equation}
\cfout.
\end{athm}
\begin{aproof}
\Nobs that 
\begin{equation}
\begin{split}
  \sup_{ v \in \R^b }
  \br*{
    \frac{ 
      \infnorm{ M v }
    }{ 
      \infnorm{ v } 
    }
  }
& =
  \sup_{ 
    v \in \R^b , \, \infnorm{ v } \leq 1
  }
  \infnorm{ M v }
\\&=
  \sup_{ 
    v = ( v_1, v_2, \dots, v_b ) \in [-1,1]^b 
  }
  \infnorm{ M v }
\\ & 
=
  \sup_{ 
    v = ( v_1, v_2, \dots, v_b ) \in [-1,1]^b 
  }
  \pr*{
    \max_{ i \in \{ 1, 2, \dots, a \} }
    \abs*{
    \textstyle
      \sum\limits_{ j = 1 }^b
    \displaystyle
      M_{ i, j } v_j
    }
  }
\\ & 
=
  \max_{ i \in \{ 1, 2, \dots, a \} }
  \pr*{
  \sup_{ 
    v = ( v_1, v_2, \dots, v_b ) \in [-1,1]^b 
  }
  \abs*{
    \textstyle
    \sum\limits_{ j = 1 }^b
    \displaystyle
    M_{ i, j } v_j
  }
  }
\\&=
  \max_{ i \in \{ 1, 2, \dots, a \} }
  \pr*{
    \textstyle
    \sum\limits_{ j = 1 }^b
    \displaystyle
  \abs*{
    M_{ i, j } 
  }
  }
\end{split}
\end{equation}
\cfload.
\end{aproof}

\cfclear
\begin{athm}{theorem}{thm:RealNNLipsch}
Let 
$ a \in \R $,
$ b \in [a,\infty) $,
$ d, L \in \N$,
$ l = (l_0,l_1,\dots,l_L) \in \N^{ L + 1 } $
satisfy
\begin{equation}
  d \geq 
  \sum_{ k =1 }^L 
  l_k ( l_{ k - 1 } + 1 ) 
  .
\end{equation}
Then it holds for all
$ \theta, \vartheta \in \R^d $
that
\begin{equation}
\begin{split}
&
  \sup_{ x \in [a,b]^{ l_0 } }
  \infnorm{ \UnclippedRealV{\theta}{ l }(x)-\UnclippedRealV\vartheta{ l }(x) }
\\ 
& \leq
  \max\{ 1, \abs{a}, \abs{b} \} 
  \infnorm{ \theta - \vartheta }
  \br*{
    \prod_{ m = 0 }^{ L - 1 }
    ( l_m + 1 )
  }
  \br*{ 
    \sum_{ n = 0 }^{ L - 1 }
    \pr*{
      \max\{1,\infnorm{ \theta }^n\}
      \,
      \infnorm{ \vartheta }^{ L - 1 - n }
    }
  }
\\ 
& \leq 
  L
  \max\{ 1, \abs{a}, \abs{b} \} 
  \pr*{ 
    \max\{ 1, \infnorm{ \theta }, \infnorm{ \vartheta } \}
  }^{ L - 1 }
  \br*{
    \prod_{ m = 0 }^{ L - 1 }
    (
      l_m + 1 
    )
  }
  \infnorm{ \theta - \vartheta }
\\
& \leq 
  L 
  \max\{ 1, \abs{a}, \abs{b} \}
  \, 
  ( \infnorm{ l } + 1 )^L 
  \,
  ( \max\{ 1, \infnorm\theta,\infnorm\vartheta\} )^{ L - 1 }
  \,
  \infnorm{ \theta - \vartheta }
\end{split}
\end{equation}
\cfout.
\end{athm}
\begin{aproof}
Throughout this proof, let 
$ \theta_j = ( \theta_{j, 1}, \theta_{j, 2}, \dots, \theta_{j, d} ) \in \R^d $, 
$ j \in \{ 1, 2 \} $,
let $ \mathfrak{d} \in \N $ satisfy 
\begin{equation}
  \mathfrak{d} = 
  \sum_{ k =1 }^L 
  l_k ( l_{ k - 1 } + 1 ) 
  ,
\end{equation} 
let 
$ W_{ j, k } \in \R^{ l_k \times l_{ k - 1 } } $, 
$ k \in \{ 1, 2, \dots, L \} $, 
$ j \in \{ 1, 2 \} $, 
and 
$ B_{ j, k } \in \R^{ l_k } $, $ k \in \{ 1, 2, \dots, L \} $, $ j \in \{ 1, 2 \} $,
satisfy 
for all $ j \in \{ 1, 2 \} $,
$ k \in \{ 1, 2, \dots, L \} $
that
\begin{equation}
  \MappingStructuralToVectorized\bpr{ 
    \bpr{ 
      ( W_{j, 1}, B_{j, 1} ), 
      ( W_{j, 2}, B_{j, 2} ), 
      \dots, 
      ( W_{j, L}, B_{j, L} ) 
    }
  } = 
  ( \theta_{j, 1}, \theta_{j, 2}, \dots, \theta_{ {j, \mathfrak{d}} } )
  ,
\end{equation}
let $ \phi_{ j, k } \in \ANNs $, $ k \in \{ 1, 2, \dots, L \} $, $ j \in \{ 1, 2 \} $, 
satisfy for all $ j \in \{ 1, 2 \} $,
$ k \in \{ 1, 2, \dots, L \} $ that
\begin{equation}
  \phi_{ j, k } 
  =
  \bpr{ 
    ( W_{j, 1}, B_{j, 1} ), 
    ( W_{j, 2}, B_{j, 2} ), 
    \dots, 
    ( W_{j, k}, B_{j, k} ) 
  }
  \in
  \br*{
    \textstyle
      \bigtimes_{ i = 1 }^k
      \pr*{
        \R^{ l_i \times l_{ i - 1 } } \times \R^{ l_i }
      }
    \displaystyle
  }
  ,
\end{equation}
let 
$ 
  D = [a,b]^{ l_0 }
$,
let $ \mathfrak{m}_{ j, k } \in [0,\infty) $, 
$ j \in \{ 1, 2 \} $,
$ k \in \{ 0, 1, \dots, L \} $,
satisfy for all 
$ j \in \{ 1, 2 \} $, 
$ k \in \{ 0, 1, \dots, L \} $ that 
\begin{equation}
  \mathfrak{m}_{ j, k } 
  = 
  \begin{cases}
    \max\{ 1, \abs{ a } , \abs{ b } \}
  &
    \colon
    k = 0
  \\ 
    \max\cu*{ 
      1 
      , 
      \sup\nolimits_{ x \in D } 
      \,
      \infnorm{ 
        ( \functionANN{\rect}( \phi_{ j, k } ) )( x ) 
      }
    } 
  &
    \colon 
    k > 0
    ,
  \end{cases}
\end{equation}
and let $ \mathfrak{e}_k \in [0,\infty) $, $ k \in \{ 0, 1, \dots, L \} $, 
satisfy for all $ k \in \{ 0, 1, \dots, L \} $ that
\begin{equation}
  \mathfrak{e}_k 
  =
  \begin{cases}
    0
  &
    \colon
    k = 0
  \\
    \sup_{ x \in D }
    \,
    \infnorm{
      ( \functionANN{\rect}( \phi_{ 1, k } ) )( x ) 
      -
      ( \functionANN{\rect}( \phi_{ 2, k } ) )( x ) 
    }
  &
    \colon
    k > 0
  \end{cases}
\end{equation}
\cfload.
\Nobs that \cref{result:row_sum_norm} \proves that
\begin{equation}
\label{eq:e0_estimate_DNN}
\begin{split}
  \mathfrak{e}_1
& 
  =
  \sup_{ x \in D }
  \,
  \infnorm{
    ( \functionANN{\rect}( \phi_{ 1, 1 } ) )( x ) 
    -
    ( \functionANN{\rect}( \phi_{ 2, 1 } ) )( x ) 
  }
\\&=
  \sup_{ x \in D }
  \,
  \infnorm{
    \pr*{
      W_{ 1, 1 } x + B_{ 1, 1 }
    }
    -
    \pr*{
      W_{ 2, 1 } x + B_{ 2, 1 }
    }
  }
\\ &
\leq
  \br*{
    \sup_{ x \in D }
    \,
    \infnorm{
      (
        W_{ 1, 1 } - W_{ 2, 1 }
      )
      x
    }
  }
  +
  \infnorm{
    B_{ 1, 1 }
    -
    B_{ 2, 1 }
  }
\\ & 
\leq 
  \br*{
    \sup_{ v \in \R^{ l_0 } \backslash \{ 0 \} }
    \pr*{
      \frac{
        \infnorm{
          (
            W_{ 1, 1 } - W_{ 2, 1 }
          )
          v
        }
      }{
        \infnorm{ v }
      }
    }
  }
  \br*{ 
    \sup_{ x \in D }
    \,
    \infnorm{ x }
  }
  +
  \infnorm{
    B_{ 1, 1 }
    -
    B_{ 2, 1 }
  }
\\ & 
\leq 
  l_0
  \,
  \infnorm{ \theta_1 - \theta_2 }
  \max\{ \abs{ a } , \abs{ b } \}
  +
  \infnorm{
    B_{ 1, 1 }
    -
    B_{ 2, 1 }
  }
\\&
\leq 
  l_0
  \,
  \infnorm{ \theta_1 - \theta_2 }
  \max\{ \abs{ a } , \abs{ b } \}
  +
  \infnorm{ \theta_1 - \theta_2 }
\\ &
=
  \infnorm{ \theta_1 - \theta_2 }
  \pr*{
    l_0
    \max\{ \abs{ a } , \abs{ b } \}
    +
    1
  }
\leq 
  \mathfrak{m}_{ 1, 0 }
  \,
  \infnorm{ \theta_1 - \theta_2 }
  \pr*{
    l_0
    +
    1
  }
  .
\end{split}
\end{equation}
\Moreover the triangle inequality 
\proves that 
for all $ k \in \{ 1, 2, \dots, L \} \cap (1,\infty) $
it holds that
\begin{equation}
\begin{split}
  \mathfrak{e}_k
& 
  =
  \sup_{ x \in D }
  \,
  \infnorm{
    ( \functionANN{\rect}( \phi_{ 1, k } ) )( x ) 
    -
    ( \functionANN{\rect}( \phi_{ 2, k } ) )( x ) 
  }
\\ &
=
  \sup_{ x \in D }
  \,
  \biggl\lVert
    \br*{
      W_{ 1, k }\bbpr{
        \Rect_{ l_{ k - 1 } }\bpr{
          ( \functionANN{\rect}( \phi_{ 1, k - 1 } ) )( x ) 
        }
      }
      +
      B_{ 1, k }
    }
    \\&\qquad\qquad
    -
    \br*{
      W_{ 2, k }\bbpr{
        \Rect_{ l_{ k - 1 } }\bpr{
          ( \functionANN{\rect}( \phi_{ 2, k - 1 } ) )( x ) 
        }
      }
      +
      B_{ 2, k }
    }
  \biggr\rVert_\infty
\\ &
\leq
  \br*{
    \sup_{ x \in D }
    \asinfnorm{
      W_{ 1, k }\bbpr{
        \Rect_{ l_{ k - 1 } }\bpr{
          ( \functionANN{\rect}( \phi_{ 1, k - 1 } ) )( x ) 
        }
      }
      -
      W_{ 2, k }\bbpr{
        \Rect_{ l_{ k - 1 } }\bpr{
          ( \functionANN{\rect}( \phi_{ 2, k - 1 } ) )( x ) 
        }
      }
    }
  }
  \\&\qquad\qquad+
  \infnorm{ \theta_1 - \theta_2 }
  .
\end{split}
\end{equation}
The triangle inequality \hence \proves that 
for all 
$ j \in \{ 1, 2 \} $, $ k \in \{ 1, 2, \dots, L \} \cap (1,\infty) $
it holds that
\begin{equation}
\begin{split}
  \mathfrak{e}_k
& 
\leq
  \br*{
    \sup_{ x \in D }
    \asinfnorm{
      \bpr{
        W_{ 1, k } - W_{ 2, k }
      }\bpr{
        \Rect_{ l_{ k - 1 } }\bpr{
          ( \functionANN{\rect}( \phi_{ j, k - 1 } ) )( x ) 
        }
      }
    }
  }
\\
&\qquad
+
  \br*{
    \sup_{ x \in D }
    \asinfnorm{
      W_{ 3 - j, k }\bbpr{
        \Rect_{ l_{ k - 1 } }\bpr{
          ( \functionANN{\rect}( \phi_{ 1, k - 1 } ) )( x ) 
        }
        -
        \Rect_{ l_{ k - 1 } }\bpr{
          ( \functionANN{\rect}( \phi_{ 2, k - 1 } ) )( x ) 
        }
      }
    }
  }\\&\qquad
  +
  \infnorm{ \theta_1 - \theta_2 }
\\ & 
\leq
  \br*{ 
    \sup_{ v \in \R^{ l_{ k - 1 } } \backslash \{ 0 \} }
    \pr*{
    \frac{
      \infnorm{ ( W_{ 1, k } - W_{ 2, k } ) v }
    }{
      \infnorm{ v }
    }
    }
  }
  \br*{
    \sup_{ x \in D }
    \asinfnorm{
      \Rect_{ l_{ k - 1 } }\bpr{
        ( \functionANN{\rect}( \phi_{ j, k - 1 } ) )( x ) 
      }
    }
  }\\&\qquad
+
  \br*{ 
    \sup_{ v \in \R^{ l_{ k - 1 } } \backslash \{ 0 \} }
    \pr*{
    \frac{
      \infnorm{ W_{ 3 - j, k } v }
    }{
      \infnorm{ v }
    }
    }
  }
  \biggl[
    \sup_{ x \in D }
    \bigl\lVert
        \Rect_{ l_{ k - 1 } }\bpr{
          ( \functionANN{\rect}( \phi_{ 1, k - 1 } ) )( x ) 
        }
        \\&\qquad\qquad
        -
        \Rect_{ l_{ k - 1 } }\bpr{
          ( \functionANN{\rect}( \phi_{ 2, k - 1 } ) )( x ) 
        }
    \bigr\rVert_\infty
  \biggr]
  +
  \infnorm{ \theta_1 - \theta_2 }
  .
\end{split}
\end{equation}
\cref{result:row_sum_norm} and \cref{result:ReLU_contraction} \hence \prove that 
for all $ j \in \{ 1, 2 \} $, $ k \in \{ 1, 2, \dots, L \} \cap (1,\infty) $
it holds that
\begin{equation}
\begin{split}
  \mathfrak{e}_k
& 
\leq
  l_{ k - 1 }
  \, 
  \infnorm{ 
    \theta_1 - \theta_2
  }
  \br*{
    \sup_{ x \in D }
    \asinfnorm{
      \Rect_{ l_{ k - 1 } }\bpr{
        ( \functionANN{\rect}( \phi_{ j, k - 1 } ) )( x ) 
      }
    }
  }
  +
  \infnorm{ \theta_1 - \theta_2 }
\\
&
+
  l_{ k - 1 }
  \,
  \infnorm{ \theta_{ 3 - j } }
  \br*{
    \sup_{ x \in D }
    \asinfnorm{
        \Rect_{ l_{ k - 1 } }\bpr{
          ( \functionANN{\rect}( \phi_{ 1, k - 1 } ) )( x ) 
        }
        -
        \Rect_{ l_{ k - 1 } }\bpr{
          ( \functionANN{\rect}( \phi_{ 2, k - 1 } ) )( x ) 
        }
    }
  }
\\ & \leq
  l_{ k - 1 }
  \, 
  \infnorm{ 
    \theta_1 - \theta_2
  }
  \br*{
    \sup_{ x \in D }
    \asinfnorm{
      ( \functionANN{\rect}( \phi_{ j, k - 1 } ) )( x ) 
    }
  }
  +
  \infnorm{ \theta_1 - \theta_2 }
\\
&
+
  l_{ k - 1 }
  \,
  \infnorm{ \theta_{ 3 - j } }
  \br*{
    \sup_{ x \in D }
    \asinfnorm{
      ( \functionANN{\rect}( \phi_{ 1, k - 1 } ) )( x ) 
      -
      ( \functionANN{\rect}( \phi_{ 2, k - 1 } ) )( x ) 
    }
  }
\\ &
\leq 
  \infnorm{ \theta_1 - \theta_2 }
  \pr*{
    l_{ k - 1 } \, \mathfrak{m}_{ j, k - 1 } + 1 
  }
  + 
  l_{ k - 1 } 
  \, \infnorm{ \theta_{ 3 - j } }
  \, \mathfrak{e}_{ k - 1 }
  .
\end{split}
\end{equation}
\Hence that
for all $ j \in \{ 1, 2 \} $, $ k \in \{ 1, 2, \dots, L \} \cap (1,\infty) $
it holds that
\begin{equation}
\begin{split}
  \mathfrak{e}_k
& 
\leq 
  \mathfrak{m}_{ j, k - 1 } 
  \,
  \infnorm{ \theta_1 - \theta_2 }
  \pr*{
    l_{ k - 1 } + 1 
  }
  + 
  l_{ k - 1 } 
  \, \infnorm{ \theta_{ 3 - j } }
  \, \mathfrak{e}_{ k - 1 }
  .
\end{split}
\end{equation}
Combining this with \cref{eq:e0_estimate_DNN}, 
the fact that $ \mathfrak{e}_0 = 0 $, 
and the fact that $ \mathfrak{m}_{ 1, 0 } = \mathfrak{m}_{ 2, 0 } $
\proves that 
for all $ j \in \{ 1, 2 \} $, 
$ k \in \{ 1, 2, \dots, L \} $
it holds that
\begin{equation}
\begin{split}
  \mathfrak{e}_k
& 
\leq 
  \mathfrak{m}_{ j, k - 1 } 
  \pr*{
    l_{ k - 1 } + 1 
  }
  \infnorm{ \theta_1 - \theta_2 }
  + 
  l_{ k - 1 } 
  \, \infnorm{ \theta_{ 3 - j } }
  \, \mathfrak{e}_{ k - 1 }
  .
\end{split}
\end{equation}
This \proves that for all 
$
  j = ( j_n )_{ n \in \{ 0, 1, \dots, L \} } \colon \{ 0, 1, \dots, L \} \to \{ 1, 2 \} 
$
and all
$ k \in \{ 1, 2, \dots, L \} $ 
it holds that
\begin{equation}
\begin{split}
  \mathfrak{e}_k
& 
\leq 
  \mathfrak{m}_{ j_{ k - 1 }, k - 1 } 
  \pr*{
    l_{ k - 1 } + 1 
  }
  \infnorm{ \theta_1 - \theta_2 }
  + 
  l_{ k - 1 } 
  \, \infnorm{ \theta_{ 3 - j_{ k - 1 } } }
  \, \mathfrak{e}_{ k - 1 }
  .
\end{split}
\end{equation}
\Hence that for all 
$
  j = ( j_n )_{ n \in \{ 0, 1, \dots, L \} } \colon \{ 0, 1, \dots, L \} \to \{ 1, 2 \} 
$
and all
$ k \in \{ 1, 2, \dots, L \} $ 
it holds that
\begin{equation}
\label{eq:error_difference_estimate_DNN}
\begin{split}
  \mathfrak{e}_k
& \leq
  \sum_{ n = 0 }^{ k - 1 }
  \pr*{
    \br*{
      \prod_{ m = n + 1 }^{ k - 1 }
      \bpr{
        l_m
        \, \infnorm{ \theta_{ 3 - j_m } }
      }
    }
    \mathfrak{m}_{ j_n, n }
    \pr*{
      l_n + 1 
    }
    \infnorm{ \theta_1 - \theta_2 }
  }
\\ & 
= 
  \infnorm{ \theta_1 - \theta_2 }
  \br*{ 
    \sum_{ n = 0 }^{ k - 1 }
    \pr*{
      \br*{
        \prod_{ m = n + 1 }^{ k - 1 }
        \bpr{
          l_m
          \, \infnorm{ \theta_{ 3 - j_m } }
        }
      }
      \mathfrak{m}_{ j_n, n }
      \pr*{
        l_n + 1 
      }
    }
  }
  .
\end{split}
\end{equation}
\Moreover \cref{result:row_sum_norm} \proves that
for all $ j \in \{ 1, 2 \} $, $ k \in \{ 1, 2, \dots, L \} \cap (1, \infty) $, $ x \in D $
it holds that
\begin{equation}
\begin{split}
  &\infnorm{ 
    ( \functionANN{\rect}( \phi_{ j, k } ) )( x ) 
  }
\\&
  =
  \asinfnorm{
    W_{ j, k }
    \bbpr{
      \Rect_{ l_{ k - 1 } }\bpr{
        ( \functionANN{\rect}( \phi_{ j, k - 1 } ) )( x ) 
      }
    }
    +
    B_{ j, k }
  }
\\ &
  \leq 
  \br*{
    \sup_{ 
      v \in \R^{ l_{ k - 1 } } \backslash \{ 0 \}
    }
    \frac{ 
      \infnorm{
        W_{ j, k }
        v
      }
    }{
      \infnorm{ v }
    }
  }
  \asinfnorm{
    \Rect_{ l_{ k - 1 } }\bpr{
      ( \functionANN{\rect}( \phi_{ j, k - 1 } ) )( x ) 
    }
  }
  +
  \infnorm{ B_{ j, k } }
\\ & \leq
  l_{ k - 1 }
  \,
  \infnorm{ \theta_j }
  \asinfnorm{
    \Rect_{ l_{ k - 1 } }\bpr{
      ( \functionANN{\rect}( \phi_{ j, k - 1 } ) )( x ) 
    }
  }
  +
  \infnorm{ \theta_j }
\\ & \leq
  l_{ k - 1 }
  \,
  \infnorm{ \theta_j }
  \asinfnorm{
    ( \functionANN{\rect}( \phi_{ j, k - 1 } ) )( x ) 
  }
  +
  \infnorm{ \theta_j }
\\ & =
  \pr*{
    l_{ k - 1 }
    \asinfnorm{
      ( \functionANN{\rect}( \phi_{ j, k - 1 } ) )( x ) 
    }
    +
    1
  }
  \infnorm{ \theta_j }
\\ & 
\leq
  \pr*{
    l_{ k - 1 }
    \mathfrak{m}_{ j, k - 1 }
    +
    1
  }
  \infnorm{ \theta_j }
\leq
  \mathfrak{m}_{ j, k - 1 }
  \pr*{
    l_{ k - 1 }
    +
    1
  }
  \infnorm{ \theta_j }
  .
\end{split}
\end{equation}
\Hence for all 
$ j \in \{ 1, 2 \} $,
$ k \in \{ 1, 2, \dots, L \} \cap (1,\infty) $ that
\begin{equation}
\label{eq:mk_recursion}
  \mathfrak{m}_{ j, k }
\leq
  \max\{ 1,
  \mathfrak{m}_{ j, k - 1 }
  \pr*{
    l_{ k - 1 }
    +
    1
  }
  \infnorm{ \theta_j }
  \}
  .
\end{equation}
\Moreover 
\cref{result:row_sum_norm} \proves that
for all $ j \in \{ 1, 2 \} $, $ x \in D $
it holds that
\begin{equation}
\begin{split}
  \infnorm{ 
    ( \functionANN{\rect}( \phi_{ j, 1 } ) )( x ) 
  }
&
  =
  \asinfnorm{
    W_{ j, 1 } x
    +
    B_{ j, 1 }
  }
\\ &
  \leq 
  \br*{
    \sup_{ 
      v \in \R^{ l_0 } \backslash \{ 0 \}
    }
    \frac{ 
      \infnorm{
        W_{ j, 1 }
        v
      }
    }{
      \infnorm{ v }
    }
  }
  \infnorm{
    x
  }
  +
  \infnorm{ B_{ j, 1 } }
\\ & \leq
  l_0
  \,
  \infnorm{ \theta_j }
  \,
  \infnorm{
    x
  }
  +
  \infnorm{ \theta_j }
\leq
  l_0
  \,
  \infnorm{ \theta_j }
  \max\{ \abs{ a } , \abs{ b } \}
  +
  \infnorm{ \theta_j }
\\ &
=
  \pr*{
    l_0
    \max\{ \abs{ a } , \abs{ b } \}
    +
    1
  }
  \infnorm{ \theta_j }
  \leq 
  \mathfrak{m}_{ 1, 0 }
  \pr*{
    l_0
    +
    1
  }
  \infnorm{ \theta_j }
  .
\end{split}
\end{equation}
\Hence that
for all $ j \in \{ 1, 2 \} $ 
it holds that
\begin{equation}
  \mathfrak{m}_{ j, 1 } 
  \leq
  \max\{1,
  \mathfrak{m}_{ j, 0 }
  \pr*{
    l_0
    +
    1
  }
  \infnorm{ \theta_j }
  \}
  .
\end{equation}
Combining this with \cref{eq:mk_recursion} \proves that
for all $ j \in \{ 1, 2 \} $, 
$ k \in \{ 1, 2, \dots, L \} $ it holds that
\begin{equation}
  \mathfrak{m}_{ j, k }
\leq
  \max\{1,
  \mathfrak{m}_{ j, k - 1 }
  \pr*{
    l_{ k - 1 }
    +
    1
  }
  \infnorm{ \theta_j }
  \}
  .
\end{equation}
\Hence that
for all $ j \in \{ 1, 2 \} $, $ k \in \{ 0, 1, \dots, L \} $ it holds that
\begin{equation}
  \mathfrak{m}_{ j, k }
\leq
  \mathfrak{m}_{ j, 0 }
  \br*{
    \prod_{ n = 0 }^{ k - 1 }
    \pr*{
      l_n
      +
      1
    }
  }
  \bigl[\max\{1, \infnorm{ \theta_j }\}\bigr]^k
  .
\end{equation}
Combining this with \cref{eq:error_difference_estimate_DNN} \proves that for all
$
  j = ( j_n )_{ n \in \{ 0, 1, \dots, L \} } \colon \{ 0, 1, \dots, L \} \to \{ 1, 2 \} 
$
and all
$ k \in \{ 1, 2, \dots, L \} $ 
it holds that
\begin{equation}
\begin{split}
  \mathfrak{e}_k
& 
\leq
  \infnorm{ \theta_1 - \theta_2 }
  \Biggl[
    \sum_{ n = 0 }^{ k - 1 }
    \Biggl(
      \br*{
        \prod_{ m = n + 1 }^{ k - 1 }
        \bpr{
          l_m
          \, \infnorm{ \theta_{ 3 - j_m } }
        }
      }
      \\&\qquad
      \cdot\pr*{
        \mathfrak{m}_{ j_n, 0 }
        \br*{
        \prod_{ v = 0 }^{ n - 1 }
        ( l_v + 1 )
        }
        \max\{1,\infnorm{ \theta_{ j_n } }^n\}
        \pr*{
          l_n + 1 
        }
      }
    \Biggr)
  \Biggr]
\\ &
=
  \mathfrak{m}_{ 1, 0 }
  \,
  \infnorm{ \theta_1 - \theta_2 }
  \br*{ 
    \sum_{ n = 0 }^{ k - 1 }
    \pr*{
      \br*{
        \prod_{ m = n + 1 }^{ k - 1 }
        \bpr{
          l_m
          \, \infnorm{ \theta_{ 3 - j_m } }
        }
      }
      \pr*{
        \br*{
        \prod_{ v = 0 }^n
        ( l_v + 1 )
        }
        \max\{1,\infnorm{ \theta_{ j_n } }^n\}
      }
    }
  }
\\ &
\leq
  \mathfrak{m}_{ 1, 0 }
  \,
  \infnorm{ \theta_1 - \theta_2 }
  \br*{ 
    \sum_{ n = 0 }^{ k - 1 }
    \pr*{
      \br*{
        \prod_{ m = n + 1 }^{ k - 1 }
        \infnorm{ \theta_{ 3 - j_m } }
      }
      \br*{
        \prod_{ v = 0 }^{ k - 1 }
        ( l_v + 1 )
      }
      \max\{1,\infnorm{ \theta_{ j_n } }^n\}
    }
  }
\\ &
=
  \mathfrak{m}_{ 1, 0 }
  \,
  \infnorm{ \theta_1 - \theta_2 }
  \br*{
    \prod_{ n = 0 }^{ k - 1 }
    ( l_n + 1 )
  }
  \br*{ 
    \sum_{ n = 0 }^{ k - 1 }
    \pr*{
      \br*{
        \prod_{ m = n + 1 }^{ k - 1 }
        \infnorm{ \theta_{ 3 - j_m } }
      }
      \max\{1,\infnorm{ \theta_{ j_n } }^n\}
    }
  }
  .
\end{split}
\end{equation}
\Hence that for all 
$
  j \in \{ 1, 2 \}
$, 
$ k \in \{ 1, 2, \dots, L \} $ 
it holds that
\begin{equation}
\begin{split}
  \mathfrak{e}_k
& 
\leq
  \mathfrak{m}_{ 1, 0 }
  \,
  \infnorm{ \theta_1 - \theta_2 }
  \br*{
    \prod_{ n = 0 }^{ k - 1 }
    ( l_n + 1 )
  }
  \br*{ 
    \sum_{ n = 0 }^{ k - 1 }
    \pr*{
      \br*{
        \prod_{ m = n + 1 }^{ k - 1 }
        \infnorm{ \theta_{ 3 - j } }
      }
      \max\{1,\infnorm{ \theta_{ j } }^n\}
    }
  }
\\
& 
=
  \mathfrak{m}_{ 1, 0 }
  \,
  \infnorm{ \theta_1 - \theta_2 }
  \br*{
    \prod_{ n = 0 }^{ k - 1 }
    ( l_n + 1 )
  }
  \br*{ 
    \sum_{ n = 0 }^{ k - 1 }
    \pr*{
		\max\{1,\infnorm{ \theta_{ j } }^n\}
      \,
      \infnorm{ \theta_{ 3 - j } }^{ k - 1 - n }
    }
  }
\\
& 
\leq
  k \,
  \mathfrak{m}_{ 1, 0 }
  \,
  \infnorm{ \theta_1 - \theta_2 }
  \pr*{ 
    \max\{ 1, \infnorm{ \theta_1 }, \infnorm{ \theta_2 } \}
  }^{ k - 1 }
      \br*{
        \prod_{ m = 0 }^{ k - 1 }
        \bpr{
          l_m + 1 
        }
      }
  .
\end{split}
\end{equation}
\end{aproof}

\cfclear
\begin{athm}{cor}{cor:ClippedRealNNLipsch}
  Let 
    $a\in\R$,
    $b\in[a,\infty)$,
  	$u\in[-\infty,\infty)$,
	  $v\in(u,\infty]$,
    $d,L\in\N$,
    $l=(l_0,l_1,\dots,l_L)\in\N^{L+1}$
  satisfy
  \begin{equation}
    d\geq \sum_{k=1}^L l_k(l_{k-1}+1).
  \end{equation}
  Then it holds for all
    $\theta,\vartheta\in\R^d$
  that
  \begin{eqsplit}
    &\sup_{x\in[a,b]^{l_0}}\infnorm{\ClippedRealV{\theta}{l}uv(x)-\ClippedRealV\vartheta{l}uv(x)}
    \\&\leq
    L 
    \max\{ 1, \abs{a}, \abs{b} \}
    \, 
    ( \infnorm{ l } + 1 )^L 
    \,
    ( \max\{ 1, \infnorm\theta,\infnorm\vartheta\} )^{ L - 1 }
    \,
    \infnorm{ \theta - \vartheta }
    \end{eqsplit}
  \cfout.
\end{athm}
\begin{aproof}
  \Nobs that
    \cref{result:Clip_contraction}
    and \cref{thm:RealNNLipsch}
  \prove that for all
    $\theta,\vartheta\in\R^d$
  it holds that
  \begin{equation}
  \begin{split}
		&
		\sup_{x\in[a,b]^{l_0}}\infnorm{\ClippedRealV{\theta}{l} uv(x)-\ClippedRealV\vartheta{ l} uv(x)}
    \\&=
    \sup_{x\in[a,b]^{l_0}}\infnorm{\Clip uv{l_L}(\UnclippedRealV\theta{l}(x))-\Clip uv{l_L}(\UnclippedRealV\vartheta{l}(x))}
    \\&\leq
    \sup_{x\in[a,b]^{l_0}}\infnorm{\UnclippedRealV\theta{l}(x)-\UnclippedRealV\vartheta{l}(x)}
    \\&\leq
    L 
    \max\{ 1, \abs{a}, \abs{b} \}
    \, 
    ( \infnorm{ l } + 1 )^L 
    \,
    ( \max\{ 1,\infnorm\theta,\infnorm\vartheta\} )^{ L - 1 }
    \,
    \infnorm{ \theta - \vartheta }
    \end{split}
  \end{equation}
  \cfload.
\end{aproof}

\subsection{Strong convergences rates for the optimization error involving ANNs}

\cfclear
\begin{athm}{lemma}{lem:Lipschitz_risk}
Let
$ d, \bfd, \bfL, M \in \N $,
$ B, b \in [ 1, \infty ) $,
$ u \in \R $,
$ v \in ( u, \infty ) $,
$ \bfl = ( \bfl_0, \bfl_1, \allowbreak\ldots, \allowbreak\bfl_\bfL ) \in \N^{ \bfL + 1 } $,
$ D \subseteq [ -b, b ]^d $,
assume
$ \bfl_0 = d $,
$ \bfl_\bfL = 1 $,
and
$ \bfd \geq \sum_{i=1}^{\bfL} \bfl_i( \bfl_{ i - 1 } + 1 ) $,
let
$ \Omega $
be a set,
let
$ X_j \colon \Omega \to D $,
$ j \in \{ 1, 2, \ldots, M \} $,
and
$ Y_j \colon \Omega \to [ u, v ] $,
$ j \in \{ 1, 2, \ldots, M \} $,
be functions,
and
let
$ \emprisk \colon [ -B, B ]^\bfd \times \Omega \to [ 0, \infty ) $
satisfy for all
$ \theta \in [ -B, B ]^\bfd $,
$ \omega \in \Omega $
that
\begin{equation}
\emprisk( \theta, \omega )
=
\frac{1}{M}
\biggl[
\smallsum_{j=1}^M
    \lvert \clippedNN{\theta}{\bfl}{u}{v}( X_j( \omega ) ) - Y_j( \omega ) \rvert^2
\biggr]
\end{equation}
\cfload.
Then it holds for all
$ \theta, \vartheta \in [ -B, B ]^\bfd $,
$ \omega \in \Omega $
that
\begin{equation}
\lvert \emprisk( \theta, \omega ) - \emprisk( \vartheta, \omega ) \rvert
\leq
2 ( v - u ) b
\bfL
( \pnorm\infty{ \bfl } + 1 )^\bfL
B^{ \bfL - 1 }
\pnorm\infty{ \theta - \vartheta  }
\end{equation}
\cfout.
\end{athm}
\begin{aproof}
\Nobs that
the fact that
for all
$ x_1, x_2, y \in \R$
it holds that
$( x_1 - y )^2 - ( x_2 - y)^2
= ( x_1 - x_2 )( ( x_1 - y ) + ( x_2 - y ) ) $,
the fact that
for all
$ \theta \in \R^\bfd$, 
$ x \in \R^d$ it holds that
$\clippedNN{\theta}{\bfl}{u}{v}( x ) \in [ u, v ] $,
and the assumption that
for all
$ j \in \{ 1, 2, \ldots, M \}$,
$\omega \in \Omega$
it holds that
$Y_j( \omega ) \in [ u, v ] $
\prove that for all
$ \theta, \vartheta \in [ -B, B ]^\bfd $,
$ \omega \in \Omega $
it holds that
\begin{equation}
\label{eq:difference_cR}
\begin{split}
& \lvert \emprisk( \theta, \omega ) - \emprisk( \vartheta, \omega ) \rvert
\\ &
=
\frac{1}{M}
\biggl\lvert
\biggl[
\smallsum_{j=1}^M
    \lvert \clippedNN{\theta}{\bfl}{u}{v}( X_j( \omega ) ) - Y_j( \omega ) \rvert^2
\biggr]
-
\biggl[
\smallsum_{j=1}^M
    \lvert \clippedNN{\vartheta}{\bfl}{u}{v}( X_j( \omega ) ) - Y_j( \omega ) \rvert^2
\biggr]
\biggr\rvert
\\ &
\leq
\frac{1}{M}
\biggl[
\smallsum_{j=1}^M
    \bigl\lvert
    [ \clippedNN{\theta}{\bfl}{u}{v}( X_j( \omega ) ) - Y_j( \omega ) ]^2
    -
    [ \clippedNN{\vartheta}{\bfl}{u}{v}( X_j( \omega ) ) - Y_j( \omega ) ]^2
    \bigr\rvert
\biggr]
\\ &
=
\frac{1}{M}
\biggl[
\smallsum_{j=1}^M
    \bigl(
    \bigl\lvert
        \clippedNN{\theta}{\bfl}{u}{v}( X_j( \omega ) ) - \clippedNN{\vartheta}{\bfl}{u}{v}( X_j( \omega ) )
    \bigr\rvert
\\ &
\hphantom{
\; =
\frac{1}{M}
\biggl[
\smallsum_{j=1}^M
    \bigl(
}
    \cdot
    \bigl\lvert
    [ \clippedNN{\theta}{\bfl}{u}{v}( X_j( \omega ) ) - Y_j( \omega ) ]
    +
    [ \clippedNN{\vartheta}{\bfl}{u}{v}( X_j( \omega ) ) - Y_j( \omega ) ]
    \bigr\rvert
    \bigr)
\biggr]
\\ &
\leq
\frac{2}{M}
\biggl[
\smallsum_{j=1}^M
    \bigl(
    \bigl[
    \sup\nolimits_{ x \in D }
    \lvert
        \clippedNN{\theta}{\bfl}{u}{v}( x ) - \clippedNN{\vartheta}{\bfl}{u}{v}( x )
    \rvert
    \bigr]
    \bigl[
    \sup\nolimits_{ y_1, y_2 \in [ u, v ] }
        \lvert y_1 - y_2 \rvert
    \bigr]
    \bigr)
\biggr]
\\ &
=
2 ( v - u )
\bigl[
\sup\nolimits_{ x \in D }
    \lvert
        \clippedNN{\theta}{\bfl}{u}{v}( x ) - \clippedNN{\vartheta}{\bfl}{u}{v}( x )
    \rvert
\bigr]
.
\end{split}
\end{equation}
\Moreover
the assumption that
$ D \subseteq [ -b, b ]^d $,
$ \bfd \geq \sum_{i=1}^{\bfL} \bfl_i( \bfl_{ i - 1 } + 1 ) $,
$ \bfl_0 = d $,
$ \bfl_\bfL = 1 $,
$ b \geq 1 $,
and
$ B \geq 1 $
and
\cref{cor:ClippedRealNNLipsch}
(applied with
$ a \is -b $,
$ b \is b $,
$ u \is u $,
$ v \is v $,
$ d \is \bfd $,
$ L \is \bfL $,
$ l \is \bfl $
in the notation of
\cref{cor:ClippedRealNNLipsch})
\prove that for all
$ \theta, \vartheta \in [ -B, B ]^\bfd $
it holds that
\begin{equation}
\begin{split}
&
\sup\nolimits_{ x \in D }
    \lvert
        \clippedNN{\theta}{\bfl}{u}{v}( x ) - \clippedNN{\vartheta}{\bfl}{u}{v}( x )
    \rvert
\leq
\sup\nolimits_{ x \in [ -b, b ]^d }
    \lvert
        \clippedNN{\theta}{\bfl}{u}{v}( x ) - \clippedNN{\vartheta}{\bfl}{u}{v}( x )
    \rvert
\\ &
\leq
\bfL
\max\{ 1, b \}
( \infnorm{ \bfl } + 1 )^\bfL
( \max\{ 1, \infnorm{ \theta }, \infnorm{ \vartheta } \} )^{ \bfL - 1 }
\infnorm{ \theta - \vartheta }
\\ &
\leq
b
\bfL
( \pnorm\infty {\bfl } + 1 )^\bfL
B^{ \bfL - 1 }
\pnorm\infty {\theta - \vartheta }
\ifnocf.
\end{split}
\end{equation}
\cfload[.]%
This and \cref{eq:difference_cR}
\prove that for all
$ \theta, \vartheta \in [ -B, B ]^\bfd $,
$ \omega \in \Omega $
it holds that
\begin{equation}
\lvert \emprisk( \theta, \omega ) - \emprisk( \vartheta, \omega ) \rvert
\leq
2 ( v - u ) b
\bfL
( \infnorm{ \bfl } + 1 )^\bfL
B^{ \bfL - 1 }
\infnorm{ \theta - \vartheta }
.
\end{equation}
\end{aproof}

\cfclear
\begin{athm}{cor}{cor:minimum_MC_rate}
Let
$ d, \bfd, \fd, \bfL, M, K \in \N $,
$ B, b \in [ 1, \infty ) $,
$ u \in \R $,
$ v \in ( u, \infty ) $,
$ \bfl = ( \bfl_0, \bfl_1, \ldots, \bfl_\bfL ) \in \N^{ \bfL + 1 } $,
$ D \subseteq [ -b, b ]^d $,
assume
$ \bfl_0 = d $,
$ \bfl_\bfL = 1 $,
and
$ \bfd \geq \fd = \sum_{i=1}^{\bfL} \bfl_i( \bfl_{ i - 1 } + 1 ) $,
let
$ ( \Omega, \cF, \P ) $
be a probability space,
let
$ \Theta_k \colon \Omega \to [ -B, B ]^\bfd $, $ k \in \{ 1, 2, \ldots, K \} $,
be i.i.d.\ random variables,
assume that
$ \Theta_1 $ is continuously uniformly distributed on $ [ -B, B ]^\bfd $,
let
$ X_j \colon \Omega \to D $,
$ j \in \{ 1, 2, \ldots, M \} $,
and
$ Y_j \colon \Omega \to [ u, v ] $,
$ j \in \{ 1, 2, \ldots, M \} $,
be random variables,
and
let
$ \emprisk \colon [ -B, B ]^\bfd \times \Omega \to [ 0, \infty ) $
satisfy for all
$ \theta \in [ -B, B ]^\bfd $,
$ \omega \in \Omega $
that
\begin{equation}
\emprisk( \theta, \omega )
=
\frac{1}{M}
\biggl[
\smallsum_{j=1}^M
    \lvert \clippedNN{\theta}{\bfl}{u}{v}( X_j( \omega ) ) - Y_j( \omega ) \rvert^2
\biggr]
\end{equation}
\cfload.
Then
\begin{enumerate}[label=(\roman *)]
\item
\label{item:cor:minimum_MC_rate:1}
it holds that
$ \emprisk $
is a
$ ( \cB( [ -B, B ]^\bfd ) \otimes \cF ) $/$ \cB( [ 0, \infty ) ) $-measurable function
and
\item
\label{item:cor:minimum_MC_rate:2}
it holds
for all
$ \theta \in [ -B, B ]^\bfd $,
$ p \in ( 0, \infty ) $
that
\begin{eqsplit}
&
\bigl(
\E\bigl[
    \min\nolimits_{ k \in \{ 1, 2, \ldots, K \} } \lvert \emprisk( \Theta_k ) - \emprisk( \theta ) \rvert^p
\bigr]
\bigr)^{ \nicefrac{1}{p} }
\\ &
\leq
\frac{
4 ( v - u ) b
\bfL
( \infnorm{\bfl} + 1 )^\bfL
B^{ \bfL }
\sqrt{ \max\{ 1, \nicefrac{p}{\fd} \} }
}{
K^{ \nicefrac{1}{\fd} }
}
\\&\leq
\frac{
4 ( v - u ) b
\bfL
( \infnorm{ \bfl } + 1 )^\bfL
B^{ \bfL }
\max\{ 1, p \}
}{
K^{ [ \bfL^{-1} ( \infnorm{ \bfl } + 1 )^{-2} ] }
}
\end{eqsplit}
\end{enumerate}
\cfout.
\end{athm}
\begin{aproof}
Throughout this proof,
let
$ L =
2 ( v - u ) b
\bfL
( \infnorm{ \bfl } \allowbreak + 1 )^\bfL
B^{ \bfL - 1 } $,
let
$ P \colon [ -B, B ]^\bfd \to [ -B, B ]^\fd $
satisfy for all
$ \theta = ( \theta_1, \theta_2, \ldots, \theta_\bfd ) \in [ -B, B ]^\bfd $
that
$ P( \theta ) = ( \theta_1, \theta_2, \ldots, \theta_\fd ) $,
and
let
$ R \colon [ -B, B ]^\fd \times \Omega \to \R $
satisfy for all
$ \theta \in [ -B, B ]^\fd $,
$ \omega \in \Omega $
that
\begin{equation}
R( \theta, \omega )
=
\frac{1}{M}
\biggl[
\smallsum_{j=1}^M
    \lvert \clippedNN{\theta}{\bfl}{u}{v}( X_j( \omega ) ) - Y_j( \omega ) \rvert^2
\biggr]
.
\end{equation}
\Nobs that
the fact that
$ \forall \, \theta \in [ -B, B ]^\bfd \colon
\clippedNN{\theta}{\bfl}{u}{v}
=
\clippedNN{ \smash{ P( \theta ) } }{\bfl}{u}{v} $
\proves that for all
$ \theta \in [ -B, B ]^\bfd $,
$ \omega \in \Omega $
it holds that
\begin{equation}
\label{eq:clipped_parameters}
\begin{split}
\emprisk( \theta, \omega )
& =
\frac{1}{M}
\biggl[
\smallsum_{j=1}^M
    \lvert \clippedNN{\theta}{\bfl}{u}{v}( X_j( \omega ) ) - Y_j( \omega ) \rvert^2
\biggr]
\\ &
=
\frac{1}{M}
\biggl[
\smallsum_{j=1}^M
    \lvert \clippedNN{ P( \theta ) }{\bfl}{u}{v}( X_j( \omega ) ) - Y_j( \omega ) \rvert^2
\biggr]
=
R( P( \theta ), \omega )
.
\end{split}
\end{equation}
\Moreover
\cref{lem:Lipschitz_risk}
(applied with
$ \bfd \is \fd $,
$ \emprisk \is
( [ -B, B ]^\fd \times \Omega \ni ( \theta, \omega )
\mapsto R( \theta, \omega ) \in [ 0, \infty ) ) $
in the notation of \cref{lem:Lipschitz_risk})
\proves that for all
$ \theta, \vartheta \in [ -B, B ]^\fd $,
$ \omega \in \Omega $
it holds that
\begin{equation}
\label{eq:Lipschitz_R}
\lvert R( \theta, \omega ) - R( \vartheta, \omega ) \rvert
\leq
2 ( v - u ) b
\bfL
( \infnorm{ \bfl } + 1 )^\bfL
B^{ \bfL - 1 }
\infnorm{ \theta - \vartheta }
=
L
\infnorm{ \theta - \vartheta }
.
\end{equation}
\Moreover
the assumption that
$ X_j $,
$ j \in \{ 1, 2, \ldots, M \} $,
and
$ Y_j $,
$ j \in \{ 1, 2, \ldots, \allowbreak M \} $,
are random variables
\proves that
$ R \colon [ -B, B ]^\fd \times \Omega \to \R $
is a random field.
This,
\cref{eq:Lipschitz_R},
the fact that
$ P \circ \Theta_k \colon \Omega \to [ -B, B ]^\fd $, $ k \in \{ 1, 2, \ldots, K \} $,
are i.i.d.\ random variables,
the fact that
$ P \circ \Theta_1 $ is continuously uniformly distributed on $ [ -B, B ]^\fd $,
and
\cref{prop:minimum_MC_rate}
(applied with
$ \bfd \is \fd $,
$ \alpha \is -B $,
$ \beta \is B $,
$ \emprisk \is R $,
$ ( \Theta_k )_{ k \in \{ 1, 2, \ldots, K \} }
\is
( P \circ \Theta_k )_{ k \in \{ 1, 2, \ldots, K \} } $
in the notation of \cref{prop:minimum_MC_rate})
\prove that for all
$ \theta \in [ -B, B ]^\bfd $,
$ p \in ( 0, \infty ) $
it holds that
$ R $
is
$ ( \cB( [ -B, B ]^\fd ) \otimes \cF ) $/$ \cB( \R ) $-measurable 
and
\begin{equation}
\label{eq:prop:minimum_MC_rate}
\begin{split}
&
\bigl(
\E\bigl[
    \min\nolimits_{ k \in \{ 1, 2, \ldots, K \} } \lvert R( P( \Theta_k ) ) - R( P( \theta ) ) \rvert^p
\bigr]
\bigr)^{ \nicefrac{1}{p} }
\\ &
\leq
\frac{ L ( 2 B ) \max\{ 1, ( \nicefrac{p}{\fd} )^{ \nicefrac{1}{\fd} } \} }{ K^{ \nicefrac{1}{\fd} } }
=
\frac{
4 ( v - u ) b
\bfL
( \infnorm{ \bfl } + 1 )^\bfL
B^{ \bfL } 
\max\{ 1, ( \nicefrac{p}{\fd} )^{ \nicefrac{1}{\fd} } \}
}{
K^{ \nicefrac{1}{\fd} }
}
.
\end{split}
\end{equation}
The fact that
$ P $ is 
$ \cB( [ -B, B ]^\bfd ) $/$ \cB( [ -B, B ]^\fd ) $-measurable 
and
\cref{eq:clipped_parameters}
\hence
\prove[ep]
\cref{item:cor:minimum_MC_rate:1}.
\Moreover
\cref{eq:clipped_parameters},
\cref{eq:prop:minimum_MC_rate},
and
the fact that
$ 2
\leq \fd
=
\sum_{i=1}^{\bfL} \bfl_i( \bfl_{ i - 1 } + 1 )
\leq
\bfL
( \infnorm{ \bfl } + 1 )^2 $
\prove that for all
$ \theta \in [ -B, B ]^\bfd $,
$ p \in ( 0, \infty ) $
it holds that
\begin{eqsplit}
&
\bigl(
\E\bigl[
    \min\nolimits_{ k \in \{ 1, 2, \ldots, K \} } \lvert \emprisk( \Theta_k ) - \emprisk( \theta ) \rvert^p
\bigr]
\bigr)^{ \nicefrac{1}{p} }
\\ &
=
\bigl(
\E\bigl[
    \min\nolimits_{ k \in \{ 1, 2, \ldots, K \} } \lvert R( P( \Theta_k ) ) - R( P( \theta ) ) \rvert^p
\bigr]
\bigr)^{ \nicefrac{1}{p} }
\\ &
\leq
\frac{
4 ( v - u ) b
\bfL
( \infnorm{ \bfl } + 1 )^\bfL
B^{ \bfL }
\sqrt{ \max\{ 1, \nicefrac{p}{\fd} \} }
}{
K^{ \nicefrac{1}{\fd} }
}
\\&\leq
\frac{
4 ( v - u ) b
\bfL
( \infnorm{ \bfl } + 1 )^\bfL
B^{ \bfL }
\max\{ 1, p \}
}{
K^{ [ \bfL^{-1} ( \infnorm{ \bfl } + 1 )^{-2} ] }
}
.
\end{eqsplit}
This \proves[ep]
\cref{item:cor:minimum_MC_rate:2}.
\end{aproof}

%% file: parts/Generalization_error.tex
\cchapter{Probabilistic generalization error estimates}{sect:probabilistic_generalization}

In \cref{sec:composed_error} below we establish a full error analysis for the training of \anns\ in the specific situation of \GD-type optimization methods with many independent random initializations (see \cref{cor:SGD_simplfied}).
For this combined error analysis we do not only employ estimates for the approximation error (see \cref{part:approx} above) and the optimization error (see \cref{part:opt} above) but we also employ suitable generalization error estimates.
Such generalization error estimates are the subject of this chapter (cf.\ \cref{lem:cov6} below) and the next (cf.\ \cref{cor:generalisation_error} below).
While in this chapter, we treat probabilistic generalization error estimates, 
in Chapter we will present generalization error estimates in the strong $L^p$-sense.

In the literature, related generalization error estimates can, \eg, be found in the survey articles and books \cite{berner_grohs_kutyniok_petersen_2022,Berner2020,CuckerSmale2002,Beck2019published,shalev2014understanding}
and the references therein.
The specific material in \cref{sect:concentration_inequ} is inspired by Duchi~\cite{duchiprobability},
the specific material in \cref{sect:covering_numbers} is inspired by Cucker \& Smale~\cite[Section~6 in Chapter~I]{CuckerSmale2002} and Carl \& Stephani~\cite[Section~1.1]{CarlStephani90},
and the specific presentation of \cref{sect:emp_risk_min} is strongly based on Beck et al.~\cite[Section 3.2]{Beck2019published}.

\section{Concentration inequalities for random variables}
\label{sect:concentration_inequ}

\subsection{Markov's inequality}

\begin{lemma}[Markov inequality]
\label{lem:markov}
Let 
$ \pr*{ \Omega, \mathcal{F}, \mu } $
be a measure space, 
let 
$ X \colon \Omega \to [0,\infty) $ 
be $ \mathcal{F} $/$ \mathcal{B}( [0,\infty) ) $-measurable, 
and 
let $ \varepsilon \in (0,\infty) $. 
Then
\begin{equation}
\label{eq:markov}
  \mu\bpr{
    X \geq \varepsilon
  }
  \leq
  \frac{
    \int_{ \Omega } X \, \diff\mu
  }{
    \varepsilon
  }
  .
\end{equation}
\end{lemma}

\begin{proof}[Proof
of \cref{lem:markov}]
Observe that the fact that $ X \geq 0 $ proves that
\begin{equation}
\label{eq:markov_use}
  \ind{
    \{ X \geq \varepsilon \}
  }
  =
  \frac{
    \varepsilon 
    \ind{
      \{ X \geq \varepsilon \}
    }
  }{
    \varepsilon
  }
  \leq
  \frac{
    X 
    \ind{
      \{ X \geq \varepsilon \}
    }
  }{
    \varepsilon
  }
  \leq
  \frac{
    X
  }{
    \varepsilon
  }
  .
\end{equation}
Hence, we obtain that
\begin{equation}
  \mu( X \geq \varepsilon )
  =
  \int_{ \Omega }
  \ind{
    \{ X \geq \varepsilon \}
  }
  \, \diff\mu
  \leq
  \frac{
    \int_{ \Omega } X \, \diff \mu
  }{
    \varepsilon
  }
  .
\end{equation}
The proof of \cref{lem:markov} is thus complete.
\end{proof}

\subsection{A first concentration inequality}

\subsubsection{On the variance of bounded random variables}

\begin{lemma}%
\label{redistribution_inequality}
Let
	$x \in [0,1]$, 
	$y \in \R$.
Then
\begin{equation}
\label{redistribution_inequality:concl}
	(x-y)^2
\leq
	(1-x) y^2 + x (1-y)^2.
\end{equation}
\end{lemma}

\begin{proof}[Proof of \cref{redistribution_inequality}]
Observe that the assumption that $x \in [0, 1]$ assures that
\begin{equation}
\begin{split}
	(1-x) y^2 + x (1-y)^2
=
	y^2 -xy^2  + x - 2xy +  xy^2 
\geq
	y^2 + x^2 - 2xy
=
	(x-y)^2.
\end{split}
\end{equation}
This establishes \cref{redistribution_inequality:concl}.
The proof of \cref{redistribution_inequality} is thus complete.
\end{proof}

\begin{lemma}
\label{variance_bd_simple}
It holds that 
$
	\sup_{p \in \R} p(1-p)
	=
	\frac{1}{4}
$.
\end{lemma}

\begin{proof}[Proof of \cref{variance_bd_simple}]
Throughout this proof, let $f \colon \R \to \R$ satisfy for all 
	$p \in \R$ 
that
	$f(p) = p(1-p)$.
Observe that the fact that $\forall \, p \in \R \colon f'(p) = 1-2p$ implies that 
$\{p \in \R \colon  f'(p) = 0\} = \{ \nicefrac{1}{2}\}$.
Combining this with the fact that $f$ is strictly concave implies that 
\begin{equation}
\begin{split}
	\sup_{p \in \R} p(1-p)
=
	\sup_{p \in \R} f(p)
=
	f(\nicefrac{1}{2})
=
	\nicefrac{1}{4}.
\end{split}
\end{equation}
The proof of \cref{variance_bd_simple} is thus complete.
\end{proof}

\begin{lemma}
\label{variance_bd_01}
Let $ ( \Omega, \mathcal{F}, \P ) $ be a probability space and
let $ X \colon \Omega \to [ 0, 1 ] $ be a random variable.
Then 
\begin{equation}
\label{variance_bd_01:concl}
	\var(X)\leq \nicefrac{1}{4}.
\end{equation}
\end{lemma}

\begin{proof}[Proof of \cref{variance_bd_01}]
Observe that \cref{redistribution_inequality} implies that
\begin{equation}
\begin{split}
	\var(X)
=
	\Exp{ (X - \Exp{X})^2} 
&\leq
	\Exp{ (1-X)(\Exp{X})^2 + X(1-\Exp{X})^2 } \\
&=
	(1-\Exp{X})(\Exp{X})^2 + \Exp{X}(1-\Exp{X})^2 \\
&=
	(1-\Exp{X})\Exp{X} (\Exp{X} + (1-\Exp{X})) \\
&=
	(1-\Exp{X})\Exp{X}.
\end{split}
\end{equation}
This and \cref{variance_bd_simple} demonstrate that $\var(X) \leq \nicefrac{1}{4}$.
The proof of \cref{variance_bd_01} is thus complete.
\end{proof}

\begin{lemma}
\label{variance_bd_ab}
Let $ ( \Omega, \mathcal{F}, \P ) $ be a probability space, 
let $a \in \R$, $b \in [a, \infty)$, and
let $ X \colon \Omega \to [ a, b ] $ be a random variable.
Then
\begin{equation}
\label{variance_bd_ab:concl}
	\var(X) \leq \frac{(b-a)^2}{4}.
\end{equation}
\end{lemma}

\begin{proof}[Proof of \cref{variance_bd_ab}]
Throughout this proof, assume without loss of generality that $a < b$.
Observe that \cref{variance_bd_01} implies that 
\begin{equation}
\begin{split}
	\var(X)
=
	\Exp{ (X - \Exp{X})^2} 
&=
	(b-a)^2 \, \Exp{ \pr*{ \tfrac{X-a - (\EXp{X}-a)}{b-a}}^2} \\ 
&=
	(b-a)^2 \, \Exp{ \pr*{ \tfrac{X-a}{b-a} - \Exp{\tfrac{X-a}{b-a}}}^2}  \\
&=
	(b-a)^2 \var \bpr{\tfrac{X-a}{b-a}}
\leq
	(b-a)^2 (\tfrac{1}{4})
=
	\frac{(b-a)^2}{4}.
\end{split}
\end{equation}
The proof of \cref{variance_bd_ab} is thus complete.
\end{proof}

\subsubsection{A concentration inequality}

\begin{lemma}
\label{lemma:easy_hoeffding}
Let $ ( \Omega, \mathcal{F}, \P ) $ be a probability space, 
let $ N \in \N $, 
$ \varepsilon \in (0,\infty) $, $ a_1, a_2, \dots,\allowbreak a_N \in \R $, $ b_1 \in [a_1,\infty) $, $ b_2 \in [a_2, \infty) $, $ \dots $, $ b_N \in [ a_N, \infty) $,
and let $ X_n \colon \Omega \to [ a_n, b_n ] $, $ n \in \{ 1, 2, \dots, N \} $, 
be independent random variables. 
Then
\begin{equation}
  \P\pr*{
    \abs*{
        \sum_{ n = 1 }^N
        \bpr{
          X_n
          -
          \E[ X_n ]
        }
    }
    \geq 
    \varepsilon
  }
  \leq
  \frac{
     \sum_{ n = 1 }^N
     ( b_n - a_n )^2
   }{
      4 \varepsilon^2
   }
.
\end{equation}
\end{lemma}

\begin{proof}[Proof of \cref{lemma:easy_hoeffding}]
Note that \cref{lem:markov} assures that
\begin{equation}
\label{lemma:easy_hoeffding:eq1}
\begin{split}
  \P\pr*{
    \abs*{
        \sum_{ n = 1 }^N
        \bpr{
          X_n
          -
          \E[ X_n ]
        }
    }
    \geq 
    \varepsilon
  }
&=
  \P\pr*{
    \abs*{
        \sum_{ n = 1 }^N
        \bpr{
          X_n
          -
          \E[ X_n ]
        }
    }^2
    \geq 
    \varepsilon^2
  } \\
&\leq
  \frac{
    \EXPP{
      \babs{
          \sum_{ n = 1 }^N
          \bpr{
            X_n
            -
            \E[ X_n ]
          }
        }^2
      }
    }{
    \varepsilon^2
  } .\end{split}
\end{equation}
In addition, note that the assumption that $ X_n \colon \Omega \to [ a_n, b_n ] $, $ n \in \{ 1, 2, \dots, N \} $, are independent variables
and
\cref{variance_bd_ab} demonstrate that 
\begin{equation}
\begin{split}
    \EXPP{
      \babs{
          {\textstyle \sum_{ n = 1 }^N}
          \bpr{
            X_n
            -
            \E[ X_n ]
          }
        }^2
      }
&=
   \sum_{ n,m = 1 }^N
   \EXPP{
          \bpr{
            X_n
            -
            \E[ X_n ]
          }
           \bpr{
            X_m
            -
            \E[ X_m ]
          }
      } \\
&=
  \sum_{ n= 1 }^N
   \EXPP{
          \bpr{
            X_n
            -
            \E[ X_n ]
          }^2
      } 
\leq
 \frac{
     \sum_{ n = 1 }^N
     ( b_n - a_n )^2
   }{
      4
   }
.
\end{split}
\end{equation}
Combining this with \cref{lemma:easy_hoeffding:eq1} establishes  
 \begin{equation}
\begin{split}
  \P\pr*{
    \abs*{
        \sum_{ n = 1 }^N
        \bpr{
          X_n
          -
          \E[ X_n ]
        }
    }
    \geq 
    \varepsilon
  }
&\leq
   \frac{
   \sum_{ n = 1 }^N
          (b_n-a_n)^2
    }{
    4 \varepsilon^2
  }
\end{split}
\end{equation}
The proof of \cref{lemma:easy_hoeffding} is thus complete.
\end{proof}

\subsection{Moment-generating functions}

\begin{adef}{def:mgf}[Moment generating functions]
Let $ ( \Omega, \mathcal{F}, \P ) $ be a probability space 
and let $ X \colon \Omega \to \R $ be a random variable. 
Then we denote by 
$
  \mathbb{M}_{ X, \P } \colon \R \to [0,\infty] 
$
(we denote by 
$
  \mathbb{M}_X \colon \R \to [0,\infty] 
$)
the function which satisfies for all $ t \in \R $ that
\begin{equation}
  \mathbb{M}_{ X, \P }( t ) 
  = 
  \Exp{ 
    e^{ t X }
  }
\end{equation}
and we call $ \mathbb{M}_{ X, \P } $ the moment-generating function of $ X $ with respect to $ \P $ 
(we call $ \mathbb{M}_{ X, \P } $ the moment-generating function of $ X $).
\end{adef}

\subsubsection{Moment-generation function for the sum of independent random variables}

\begin{lemma}
\label{lem:sums_moment}
Let $ ( \Omega, \mathcal{F}, \P ) $ be a probability space, 
let 
$ t \in \R $, $ N \in \N $, 
and 
let $ X_n \colon \Omega \to \R $, $ n \in \{ 1, 2, \dots, N \} $, 
be independent random variables. 
Then
\begin{equation}
  \mathbb{M}_{ \sum_{ n = 1 }^N X_n }( t ) 
  =
  \prod\nolimits_{ n = 1 }^N
  \mathbb{M}_{ X_n }( t )   
  .
\end{equation}
\end{lemma}

\begin{proof}[Proof of \cref{lem:sums_moment}] 
Observe that Fubini's theorem ensures that for all $ t \in \R $ it holds that
\begin{equation}
  \mathbb{M}_{ \sum_{ n = 1 }^N X_n }( t ) 
  =
  \Exp{ 
    e^{
      t \pr*{ \sum_{ n = 1 }^N X_n }
    }
  }
  =
  \Exp{ 
    \prod\nolimits_{ n = 1 }^N
    e^{
      t X_n
    }
  }
  =
  \prod\nolimits_{ n = 1 }^N
  \Exp{ 
    e^{
      t X_n
    }
  }
  =
  \prod\nolimits_{ n = 1 }^N
  \mathbb{M}_{ X_n }( t )   
  .
\end{equation}
The proof of \cref{lem:sums_moment} is thus complete.
\end{proof}

\subsection{Chernoff bounds}

\subsubsection{Probability to cross a barrier}

\begin{prop}
\label{prop:Chernoff}
Let $ ( \Omega, \mathcal{F}, \P ) $ be a probability space, 
let $ X \colon \Omega \to \R $ be a random variable, 
and let $ \varepsilon \in \R $. 
Then 
\begin{equation}
  \P\pr*{ 
    X \geq \varepsilon
  }
  \leq 
  \inf_{ \lambda \in [0,\infty) }
  \pr*{
    e^{ - \lambda \varepsilon }
    \,
    \Exp{ 
      e^{ \lambda X }
    }
  }
  =
  \inf_{ \lambda \in [0,\infty) }
  \pr*{
    e^{ - \lambda \varepsilon }
    \,
    \mathbb{M}_X( \lambda )
  }
  .
\end{equation}
\end{prop}

\begin{proof}[Proof of \cref{prop:Chernoff}]
Note that 
\cref{lem:markov} 
ensures that for all $ \lambda \in [0,\infty) $ it holds that
\begin{equation}
  \P\pr*{ 
    X \geq \varepsilon
  }
  \leq
  \P\pr*{ 
    \lambda X \geq \lambda \varepsilon
  }
  =
  \P\pr*{ 
    \exp\pr*{ \lambda X } 
    \geq 
    \exp\pr*{ \lambda \varepsilon }
  }
  \leq
  \frac{
    \Exp{ \exp( \lambda X ) }
  }{
    \exp( \lambda \varepsilon )
  }
  =
  e^{ - \lambda \varepsilon }
  \,
  \Exp{ 
    e^{ \lambda X }
  }
  .
\end{equation}
The proof of \cref{prop:Chernoff} is thus complete.
\end{proof}

\begin{cor}
\label{cor:Chernoff00}
Let $ ( \Omega, \mathcal{F}, \P ) $ be a probability space, 
let $ X \colon \Omega \to \R $ be a random variable, 
and let $ c, \varepsilon \in \R $. 
Then 
\begin{equation}
  \P\pr*{ 
    X \geq c + \varepsilon
  }
  \leq 
  \inf_{ \lambda \in [0,\infty) }
  \pr*{
    e^{ - \lambda \varepsilon }
    \,
    \mathbb{M}_{ X - c }( \lambda )
  }
  .
\end{equation}
\end{cor}

\begin{proof}[Proof of \cref{cor:Chernoff00}]
Throughout this proof, let $ Y \colon \Omega \to \R $ satisfy 
\begin{equation} 
\label{eq:def_Y_chernoff}
  Y = X - c . 
\end{equation}
Observe that \cref{prop:Chernoff} and \cref{eq:def_Y_chernoff} ensure that
\begin{equation}
  \P\pr*{ 
    X - c \geq \varepsilon
  }
  =
  \P\pr*{ 
    Y \geq \varepsilon
  }
  \leq 
  \inf_{ \lambda \in [0,\infty) }
  \pr*{
    e^{ - \lambda \varepsilon }
    \,
    \mathbb{M}_{ Y }( \lambda )
  }
  =
  \inf_{ \lambda \in [0,\infty) }
  \pr*{
    e^{ - \lambda \varepsilon }
    \,
    \mathbb{M}_{ X - c }( \lambda )
  }
  .
\end{equation}
The proof of \cref{cor:Chernoff00} is thus complete. 
\end{proof}

\begin{cor}
\label{cor:Chernoff}
Let $ ( \Omega, \mathcal{F}, \P ) $ be a probability space, 
let $ X \colon \Omega \to \R $ be a random variable with $ \E[ \abs{ X } ] < \infty $, 
and let $ \varepsilon \in \R $. 
Then 
\begin{equation}
\label{eq:Chernoff_cor}
  \P\pr*{ 
    X \geq \E[ X ] + \varepsilon
  }
  \leq 
  \inf_{ \lambda \in [0,\infty) }
  \pr*{
    e^{ - \lambda \varepsilon }
    \,
    \mathbb{M}_{ X - \E[ X ] }( \lambda )
  }
  .
\end{equation}
\end{cor}

\begin{proof}[Proof of \cref{cor:Chernoff}]
Observe that \cref{cor:Chernoff00} (applied with $ c \is \E[ X ] $ in the notation 
of \cref{cor:Chernoff00}) establishes \cref{eq:Chernoff_cor}.
The proof of \cref{cor:Chernoff} is thus complete. 
\end{proof}

\subsubsection{Probability to fall below a barrier}

\begin{cor}
\label{cor:Chernoff11}
Let $ ( \Omega, \mathcal{F}, \P ) $ be a probability space, 
let $ X \colon \Omega \to \R $ be a random variable, 
and let $ c, \varepsilon \in \R $. 
Then 
\begin{equation}
  \P\pr*{ 
    X \leq c - \varepsilon
  }
  \leq 
  \inf_{ \lambda \in [0,\infty) }
  \pr*{
    e^{ - \lambda \varepsilon }
    \,
    \mathbb{M}_{ c - X }( \lambda )
  }
  .
\end{equation}
\end{cor}

\begin{proof}[Proof of \cref{cor:Chernoff11}]
Throughout this proof, let $ \mathfrak{c} \in \R $ satisfy  $ \mathfrak{c} = - c $ 
and let $ \mathfrak{X} \colon \Omega \to \R $ satisfy  
\begin{equation} 
\label{eq:def_Y_chernoff_11}
  \mathfrak{X} = - X . 
\end{equation}
Observe that \cref{cor:Chernoff00} and \cref{eq:def_Y_chernoff_11} ensure that
\begin{equation}
\begin{split}
  \P\pr*{ 
    X \leq c - \varepsilon
  }
& =
  \P\pr*{ 
    - X \geq - c + \varepsilon
  }
  =
  \P\pr*{ 
    \mathfrak{X} \geq \mathfrak{c} + \varepsilon
  }
\leq 
  \inf_{ \lambda \in [0,\infty) }
  \pr*{
    e^{ - \lambda \varepsilon }
    \,
    \mathbb{M}_{ \mathfrak{X} - \mathfrak{c} }( \lambda )
  }
\\ & =
  \inf_{ \lambda \in [0,\infty) }
  \pr*{
    e^{ - \lambda \varepsilon }
    \,
    \mathbb{M}_{ c - X }( \lambda )
  }
  .
\end{split}
\end{equation}
The proof of \cref{cor:Chernoff11} is thus complete. 
\end{proof}

\subsubsection{Sums of independent random variables}

\begin{cor}
\label{cor:Chernoff_sum}
Let $ ( \Omega, \mathcal{F}, \P ) $ be a probability space, 
let $ \varepsilon \in \R $, $ N \in \N $, 
and 
let $ X_n \colon \Omega \to \R $, $ n \in \{ 1, 2, \dots, N \} $, be independent random variables 
with $ \sum_{ n =1 }^N \E[ \abs{ X_n } ]\allowbreak < \infty $.
Then 
\begin{equation}
  \P\pr*{ 
    \br*{ 
      \sum_{ n = 1 }^N \bpr{ X_n - \E[ X_n ] } 
    } 
    \geq \varepsilon
  }
  \leq 
  \inf_{ \lambda \in [0,\infty) }
  \pr*{
    e^{ - \lambda \varepsilon }
    \br*{
      \prod_{ n = 1 }^N
      \mathbb{M}_{ X_n - \E[ X_n ] }( \lambda )
    }
  }
  .
\end{equation}
\end{cor}

\begin{proof}[Proof of \cref{cor:Chernoff_sum}]
Throughout this proof, let $ Y_n \colon \Omega \to \R $, $ n \in \{ 1, 2, \dots, N \} $, 
satisfy for all $ n \in \{ 1, 2, \dots, N \} $ that
\begin{equation}
\label{eq:def_Yn_Chernoff}
  Y_n = X_n - \E[ X_n ]
  .
\end{equation}
Observe that \cref{prop:Chernoff}, 
\cref{lem:sums_moment}, and \cref{eq:def_Yn_Chernoff} 
ensure that
\begin{equation}
\begin{split}
&
  \P\pr*{ 
    \br*{ 
      \sum_{ n = 1 }^N \bpr{ X_n - \E[ X_n ] } 
    } 
    \geq \varepsilon
  }
  =
  \P\pr*{ 
    \br*{ 
      \sum_{ n = 1 }^N Y_n 
    } 
    \geq \varepsilon
  }
  \leq 
  \inf_{ \lambda \in [0,\infty) }
  \pr*{
    e^{ - \lambda \varepsilon }
    \,
    \mathbb{M}_{ \sum_{ n = 1 }^N Y_n }( \lambda )
  }
\\ & =
  \inf_{ \lambda \in [0,\infty) }
  \pr*{
    e^{ - \lambda \varepsilon }
    \br*{
      \prod_{ n = 1 }^N
      \mathbb{M}_{ Y_n }( \lambda )
    }
  }
  =
  \inf_{ \lambda \in [0,\infty) }
  \pr*{
    e^{ - \lambda \varepsilon }
    \br*{
      \prod_{ n = 1 }^N
      \mathbb{M}_{ X_n - \E[ X_n ] }( \lambda )
    }
  }
  .
\end{split}
\end{equation}
The proof of \cref{cor:Chernoff_sum} is thus complete.
\end{proof}

\subsection{Hoeffding's inequality}

\subsubsection{On the moment-generating function for bounded random variables}

\begin{lemma}
\label{lem:Hoeff0}
Let $ ( \Omega, \mathcal{F}, \P ) $ be a probability space, 
let $ \lambda, a \in \R $, $ b \in (a,\infty) $, $ p \in [0,1] $
satisfy
$ p = \frac{ - a }{ ( b - a ) } $, 
let $ X \colon \Omega \to [a,b] $ be a random variable with 
$ \E[ X ] = 0 $, 
and 
let 
$ \phi \colon \R \to \R $ satisfy 
for all $ x \in \R $ that
$
  \phi( x ) 
  =
  \ln( 1 - p + p e^x ) - p x
$.
Then 
\begin{equation}
  \bExp{
    e^{ \lambda X }
  }
  \leq 
  e^{
    \phi\pr*{ \lambda ( b - a ) }
  }
  .
\end{equation}
\end{lemma}

\begin{proof}[Proof of \cref{lem:Hoeff0}]
Observe that for all $ x \in \R $ it holds that
\begin{equation}
\begin{split}
  x \pr*{ b - a } 
& =
  b x 
  -
  a x 
=
  \br*{ a b - a x }
  +
  \br*{ b x - a b }
=
  \br*{ 
    a 
    \pr*{ b - x }
  }
  +
  \br*{
    b
    \pr*{
      x - a 
    }
  }
\\ & 
=
  a 
  \pr*{ b - x }
  +
  b
  \br*{ 
    b - a 
    - b + x
  }
=
  a 
  \pr*{ b - x }
  +
  b
  \br*{ 
    \pr*{ b - a }
    -
    \pr*{ b - x }
  }
  .
\end{split}
\end{equation}
Hence, we obtain that for all $ x \in \R $ it holds that
\begin{equation}
  x 
  =
  a
  \pr*{
    \frac{ b - x }{ b - a }
  }
  +
  b
  \br*{ 
    1
    -
    \pr*{
      \frac{ b - x }{ b - a }
    }
  }
  .
\end{equation}
This implies that for all $ x \in \R $ it holds that
\begin{equation}
  \lambda x 
  =
  \pr*{
    \frac{ b - x }{ b - a }
  }
  \lambda a
  +
  \br*{ 
    1
    -
    \pr*{
      \frac{ b - x }{ b - a }
    }
  }
  \lambda b
  .
\end{equation}
The fact that 
$ \R \ni x \mapsto e^x \in \R $ is convex
hence demonstrates that for all $ x \in [a,b] $ 
it holds that
\begin{equation}
  e^{ \lambda x }
  =
  \exp\pr*{
    \pr*{
      \frac{ b - x }{ b - a }
    }
    \lambda a
    +
    \br*{ 
      1
      -
      \pr*{
        \frac{ b - x }{ b - a }
      }
    }
    \lambda b
  }
  \leq
  \pr*{
    \frac{ b - x }{ b - a }
  }
  e^{ \lambda a }
  +
  \br*{ 
    1
    -
    \pr*{
      \frac{ b - x }{ b - a }
    }
  }
  e^{ \lambda b }
  .
\end{equation}
The assumption that $ \E[ X ] = 0 $ 
therefore assures that 
\begin{equation}
\begin{split}
  \Exp{ 
    e^{ \lambda X }
  }
& \leq
  \pr*{
    \frac{ b }{ b - a }
  }
  e^{ \lambda a }
  +
  \br*{ 
    1
    -
    \pr*{
      \frac{ b }{ b - a }
    }
  }
  e^{ \lambda b } 
  .
\end{split}
\end{equation}
Combining this with the fact that 
\begin{eqsplit}
\label{eq:p_fact}
    \frac{ b }{ ( b - a ) }
  &=
  1 
  - 
  \br*{ 
    1 - 
    \pr*{
      \frac{ b }{ ( b - a ) }
    }
  }
  \\&=
  1 - 
  \br*{ 
    \pr*{
      \frac{ ( b - a ) }{ ( b - a ) } 
    }
    - 
    \pr*{
      \frac{ b }{ ( b - a ) }
    }
  }
\\&= 
  1 
  - 
  \br*{ 
    \frac{ - a }{ ( b - a ) }
  }
= 
  1 - p
\end{eqsplit}
demonstrates that
\begin{equation}
\label{eq:Hoeffding_convex_bound2}
\begin{split}
  \Exp{ 
    e^{ \lambda X }
  }
& \leq
  \pr*{
    \frac{ b }{ b - a }
  }
  e^{ \lambda a }
  +
  \br*{ 
    1
    -
    \pr*{
      \frac{ b }{ b - a }
    }
  }
  e^{ \lambda b } 
\\&=
  \pr*{
    1 - p
  }
  e^{ \lambda a }
  +
  \br*{ 
    1
    -
    \pr*{
      1 - p 
    }
  }
  e^{ \lambda b } 
  \\&=
  \pr*{
    1 - p
  }
  e^{ \lambda a }
  +
  p \,
  e^{ \lambda b } 
\\ & =
  \br*{ 
    \pr*{ 1 - p }
    +
    p 
    \, 
    e^{ \lambda ( b - a ) }
  } 
  e^{ \lambda a }
  .
\end{split}
\end{equation}
Moreover, note that the assumption that $ p = \frac{ - a }{ ( b - a ) } $ shows that 
$ p ( b - a ) = - a $. Hence, we obtain that 
$
  a = - p ( b - a ) 
$.
This and \cref{eq:Hoeffding_convex_bound2} assure that 
\begin{equation}
\begin{split}
  \Exp{ 
    e^{ \lambda X }
  }
& \leq
  \br*{ 
    \pr*{ 1 - p }
    +
    p 
    \, 
    e^{ \lambda ( b - a ) }
  } 
  e^{ - p \lambda ( b - a ) }
  =
  \exp\pr*{
    \ln\pr*{ 
      \br*{ 
        \pr*{ 1 - p }
        +
        p 
        \, 
        e^{ \lambda ( b - a ) }
      } 
      e^{ - p \lambda ( b - a ) }
    }
  }
\\ & =
  \exp\pr*{
    \ln\pr*{ 
      \pr*{ 1 - p }
      +
      p 
      \, 
      e^{ \lambda ( b - a ) }
    }
    -
    p \lambda ( b - a ) 
  }
  =
  \exp\pr*{
    \phi( \lambda ( b - a ) )
  }
  .
\end{split}
\end{equation}
The proof of \cref{lem:Hoeff0} is thus complete.
\end{proof}

\subsubsection{Hoeffding's lemma}

\begin{lemma}
\label{lem:Hoeffding}
Let $ p \in [0,1] $ and let 
$ \phi \colon \R \to \R $ satisfy 
for all $ x \in \R $ that
$
  \phi( x ) 
  =
  \ln( 1 - p + p e^x ) - p x
$.
Then it holds for all $ x \in \R $ that
$ 
  \phi( x ) 
  \leq 
  \frac{ x^2 }{ 8 }
$.
\end{lemma}

\begin{proof}[Proof of \cref{lem:Hoeffding}]
Observe that the fundamental theorem of calculus ensures that
for all $ x \in \R $ it holds that
\begin{equation}
\label{eq:taylor}
\begin{split}
  \phi( x ) 
& 
  = \phi( 0 ) + \int_0^x \phi'(y) \, \diff y
  \\&
  =
  \phi(0)
  +
  \phi'(0) x
  +
  \int_0^x
  \int_0^y
  \phi''(z) 
  \, \diff z \, \diff y
\\&\leq 
  \phi(0)
  +
  \phi'(0) x
  +
  \frac{ x^2 }{ 2 }
  \br*{ 
    \sup_{ z \in \R }
    \phi''(z)
  }
  .
\end{split}
\end{equation}
Moreover, note that for all $ x \in \R $ it holds that
\begin{equation}
\label{eq:phi_derivatives}
  \phi'(x) = 
  \br*{ \frac{ p e^x }{ 1 - p + p e^x } } - p
\qquad 
  \text{and}
\qquad
  \phi''(x) =
  \br*{ \frac{ p e^x }{ 1 - p + p e^x } } 
  -
  \br*{ \frac{ p^2 e^{ 2 x } }{ \pr*{ 1 - p + p e^x }^2 } } 
  .
\end{equation}
Hence, we obtain that
\begin{equation}
\label{eq:derivative_first}
  \phi'(0) = 
  \br*{ 
    \frac{ p }{ 1 - p + p } 
  }
  - p 
  = 
  0
  .
\end{equation}
In the next step we combine \cref{eq:phi_derivatives} and the fact that 
for all $ a \in \R $ it holds that
\begin{equation}
  a \pr*{ 1 - a }
  =
  a
  -
  a^2
  =
  - 
  \bbbr{
    a^2 - 2 a \br*{ \tfrac{ 1 }{ 2 } } 
    +
    \br*{ 
      \tfrac{ 1 }{ 2 }
    }^2
  }
  +
  \br*{ 
    \tfrac{ 1 }{ 2 }
  }^2
=
  \tfrac{ 1 }{ 4 }
  - 
  \br*{ 
    a - \tfrac{ 1 }{ 2 } 
  }^2
\leq 
  \tfrac{ 1 }{ 4 }
\end{equation}
to obtain that
for all $ x \in \R $ 
it holds that
$
  \phi''(x) \leq \frac{ 1 }{ 4 } 
$. 
This, \cref{eq:taylor}, and \cref{eq:derivative_first} ensure that 
for all $ x \in \R $ it holds that
\begin{equation}
\begin{split}
  \phi( x ) 
& 
\leq 
  \phi(0)
  +
  \phi'(0) x
  +
  \frac{ x^2 }{ 2 }
  \br*{ 
    \sup_{ z \in \R }
    \phi''(z)
  }
  =
  \phi(0)
  +
  \frac{ x^2 }{ 2 }
  \br*{ 
    \sup_{ z \in \R }
    \phi''(z)
  }
\leq 
  \phi(0)
  +
  \frac{ x^2 }{ 8 }
  =
  \frac{ x^2 }{ 8 }
  .
\end{split}
\end{equation}
The proof of \cref{lem:Hoeffding} is thus complete. 
\end{proof}

\begin{lemma}
\label{lem:Hoeff_lemma}
Let $ ( \Omega, \mathcal{F}, \P ) $ be a probability space, 
let $ a \in \R $, $ b \in [a,\infty) $, $ \lambda \in \R $, 
and let $ X \colon \Omega \to [a,b] $ be a random variable with 
$ \E[ X ] = 0 $. Then 
\begin{equation}
  \bExp{
    \exp( \lambda X )
  }
  \leq 
  \exp\pr*{
    \tfrac{ \lambda^2 ( b - a )^2 }{ 8 }
  }
  .
\end{equation}
\end{lemma}

\begin{proof}[Proof of \cref{lem:Hoeff_lemma}]
Throughout this proof, assume without loss of generality that $ a < b $, let 
$ p \in \R $ satisfy $ p = \frac{ - a }{ ( b - a ) } $, 
and let $ \phi_r \colon \R \to \R $, $ r \in [0,1] $, 
satisfy for all $ r \in [0,1] $, $ x \in \R $ that
\begin{equation}
  \phi_r( x ) = \ln( 1 - r + r e^x ) - r x .
\end{equation}
Observe that the assumption that $ \E[ X ] = 0 $
and the fact that $ a \leq \E[ X ] \leq b $ ensures that 
$ a \leq 0 \leq b $. Combining this with the assumption that $ a < b $ implies that 
\begin{equation}
  0 \leq p = \frac{ - a }{ ( b - a ) } \leq \frac{ ( b - a ) }{ ( b - a ) } = 1
  .
\end{equation}
\cref{lem:Hoeff0} and \cref{lem:Hoeffding} hence demonstrate that 
\begin{equation}
  \bExp{ 
    e^{ \lambda X }
  }
  \leq
  e^{ 
    \phi_p( \lambda ( b - a ) )
  }
  =
  \exp\pr*{
    \phi_p( \lambda ( b - a ) )
  }
  \leq 
  \exp\pr*{
    \tfrac{ ( \lambda ( b - a ) )^2 }{ 8 }
  }
  =
  \exp\pr*{
    \tfrac{ \lambda^2 ( b - a )^2 }{ 8 }
  }
  .
\end{equation}
The proof of \cref{lem:Hoeff_lemma} is thus complete.
\end{proof}

\subsubsection{Probability to cross a barrier}

\begin{lemma}
\label{lem:minimize}
Let $ \beta \in (0,\infty) $, $ \varepsilon \in [0,\infty) $
and let 
$ f \colon [0,\infty) \to [0,\infty) $ satisfy for all 
$ \lambda \in [0,\infty) $ that 
$
  f( \lambda ) = \beta \lambda^2 - \varepsilon \lambda
$.
Then 
\begin{equation}
\label{eq:minimize}
  \inf_{ 
    \lambda \in [0,\infty)
  }
  f( \lambda )
  =
  f( 
    \tfrac{ 
      \varepsilon
    }{
      2 \beta
    }
  )  
  =
  -
    \tfrac{ 
      \varepsilon^2
    }{
      4 \beta
    }
  .
\end{equation}
\end{lemma}

\begin{proof}[Proof of \cref{lem:minimize}]
Observe that for all $ \lambda \in \R $ it holds that
\begin{equation}
\label{eq:minimize_derivative}
  f'( \lambda ) = 2 \beta \lambda - \varepsilon
  .
\end{equation}
Moreover, note that 
\begin{equation}
  f( 
    \tfrac{ 
      \varepsilon
    }{
      2 \beta
    }
  )  
  =
  \beta 
  \br*{
    \tfrac{ 
      \varepsilon
    }{
      2 \beta
    }
  }^2
  -
  \varepsilon
  \br*{
    \tfrac{ 
      \varepsilon
    }{
      2 \beta
    }
  }
  =
    \tfrac{ 
      \varepsilon^2
    }{
      4 \beta
    }
  -
    \tfrac{ 
      \varepsilon^2
    }{
      2 \beta
    }
  =
  -
    \tfrac{ 
      \varepsilon^2
    }{
      4 \beta
    }
  .
\end{equation}
Combining this and \cref{eq:minimize_derivative} establishes \cref{eq:minimize}. 
The proof of \cref{lem:minimize} is thus complete.
\end{proof}

\begin{cor}
\label{cor:Hoeffding}
Let $ ( \Omega, \mathcal{F}, \P ) $ be a probability space, 
let 
$ N \in \N $, 
$ \varepsilon \in [0,\infty) $, $ a_1, a_2, \dots, \allowbreak a_N \in \R $, $ b_1 \in [a_1,\infty) $, $ b_2 \in [a_2, \infty) $, $ \dots $, $ b_N \in [ a_N, \infty) $
satisfy 
$
  \sum_{ n = 1 }^N
  ( b_n - a_n )^2 
  \neq 0
$,
and let $ X_n \colon \Omega \to [ a_n, b_n ] $, $ n \in \{ 1, 2, \dots, N \} $, 
be independent random variables. 
Then
\begin{equation}
  \P\pr*{
      \br*{
        \sum_{ n = 1 }^N
        \bpr{
          X_n
          -
          \E[ X_n ]
        }
      }
    \geq 
    \varepsilon
  }
  \leq
  \exp\pr*{
    \frac{
      - 2 \varepsilon^2 
    }{
      \sum_{ n = 1 }^N
      ( b_n - a_n )^2
    }
  }
  .
\end{equation}
\end{cor}

\begin{proof}[Proof of \cref{cor:Hoeffding}]
Throughout this proof, let $ \beta \in (0,\infty) $ 
satisfy 
\begin{equation}
  \beta =
  \frac{ 1 }{ 8 }
      \br*{ 
          \sum_{ n = 1 }^N
          ( b_n - a_n )^2 
      } 
      .
\end{equation}
Observe that \cref{cor:Chernoff_sum} ensures that
\begin{equation}
\label{eq:cor_Hoeffding_0}
  \P\pr*{ 
    \br*{ 
      \sum_{ n = 1 }^N \bpr{ X_n - \E[ X_n ] } 
    } 
    \geq \varepsilon
  }
  \leq 
  \inf_{ \lambda \in [0,\infty) }
  \pr*{
    e^{ - \lambda \varepsilon }
    \br*{
      \prod_{ n = 1 }^N
      \mathbb{M}_{ X_n - \E[ X_n ] }( \lambda )
    }
  }
  .
\end{equation}
Moreover, note that \cref{lem:Hoeff_lemma} proves that for all $ n \in \{ 1, 2, \dots, N \} $
it holds that
\begin{equation}
  \mathbb{M}_{ X_n - \E[ X_n ] }( \lambda )
  \leq 
  \exp\pr*{ 
    \tfrac{ \lambda^2 [ ( b_n - \E[ X_n ] ) - ( a_n - \E[ X_n ] ) ]^2 }{ 8 }
  }
  =
  \exp\pr*{ 
    \tfrac{ \lambda^2 ( b_n - a_n )^2 }{ 8 }
  }
  .
\end{equation}
Combining this with \cref{eq:cor_Hoeffding_0} 
and \cref{lem:minimize}
ensures that 
\begin{equation}
\begin{split}
&
  \P\pr*{ 
    \br*{ 
      \sum_{ n = 1 }^N \bpr{ X_n - \E[ X_n ] } 
    } 
    \geq \varepsilon
  }
  \leq 
  \inf_{ \lambda \in [0,\infty) }
  \pr*{
    \exp\pr*{ 
      \br*{ 
        \sum_{ n = 1 }^N
        \pr*{
          \tfrac{ \lambda^2 ( b_n - a_n )^2 }{ 8 }
        }
      } 
      - \lambda \varepsilon
    }
  }
\\ & =
  \inf_{ \lambda \in [0,\infty) }
  \br*{
    \exp\pr*{ 
      \lambda^2
      \br*{ 
        \frac{ 
          \sum_{ n = 1 }^N
          ( b_n - a_n )^2 
        }{ 8 }
      } 
      - \lambda \varepsilon
    }
  }
  =
  \exp\pr*{ 
    \inf_{ \lambda \in [0,\infty) }
    \br*{
      \beta
      \lambda^2
      - \varepsilon \lambda 
    }
  }
\\ & =
  \exp\pr*{ 
    \frac{ 
      - \varepsilon^2
    }{ 
      4 \beta
    }
  }
  =
  \exp\pr*{ 
    \frac{ 
      - 2 \varepsilon^2
    }{ 
          \sum_{ n = 1 }^N
          ( b_n - a_n )^2 
    }
  }
  .
\end{split}
\end{equation}
The proof of \cref{cor:Hoeffding} is thus complete.
\end{proof}

\subsubsection{Probability to fall below a barrier}

\begin{cor}
\label{cor:Hoeffding_B}
Let $ ( \Omega, \mathcal{F}, \P ) $ be a probability space, 
let 
$ N \in \N $, 
$ \varepsilon \in [0,\infty) $, $ a_1, a_2, \dots,\allowbreak a_N \in \R $, $ b_1 \in [a_1,\infty) $, $ b_2 \in [a_2, \infty) $, $ \dots $, $ b_N \in [ a_N, \infty) $
satisfy 
$
  \sum_{ n = 1 }^N
  ( b_n - a_n )^2 
  \neq 0
$,
and let $ X_n \colon \Omega \to [ a_n, b_n ] $, $ n \in \{ 1, 2, \dots, N \} $, 
be independent random variables. 
Then
\begin{equation}
  \P\pr*{
      \br*{
        \sum_{ n = 1 }^N
        \bpr{
          X_n
          -
          \E[ X_n ]
        }
      }
    \leq 
    - \varepsilon
  }
  \leq
  \exp\pr*{
    \frac{
      - 2 \varepsilon^2 
    }{
      \sum_{ n = 1 }^N
      ( b_n - a_n )^2
    }
  }
  .
\end{equation}
\end{cor}

\begin{proof}[Proof of \cref{cor:Hoeffding_B}]
Throughout this proof, let $ \mathfrak{X}_n \colon \Omega \to [ - b_n, - a_n ] $, $ n \in \{ 1, 2, \dots,\allowbreak N \} $, 
satisfy for all $ n \in \{ 1, 2, \dots, N \} $ that
\begin{equation}
\label{eq:def_frakX}
  \mathfrak{X}_n = - X_n
  .
\end{equation}
Observe that \cref{cor:Hoeffding} and \cref{eq:def_frakX} ensure that
\begin{equation}
\begin{split}
  &\P\pr*{
      \br*{
        \sum_{ n = 1 }^N
        \bpr{
          X_n
          -
          \E[ X_n ]
        }
      }
    \leq 
    - \varepsilon
  }
\\& =
  \P\pr*{
      \br*{
        \sum_{ n = 1 }^N
        \bpr{
          - X_n
          -
          \E[ - X_n ]
        }
      }
    \geq 
    \varepsilon
  }
\\ &  
=
  \P\pr*{
      \br*{
        \sum_{ n = 1 }^N
        \bpr{
          \mathfrak{X}_n
          -
          \E[ \mathfrak{X}_n ]
        }
      }
    \geq
    \varepsilon
  }
  \leq
  \exp\pr*{
    \frac{
      - 2 \varepsilon^2 
    }{
      \sum_{ n = 1 }^N
      ( b_n - a_n )^2
    }
  }
  .
\end{split}
\end{equation}
The proof of \cref{cor:Hoeffding_B} is thus complete.
\end{proof}

\subsubsection{Hoeffding's inequality}

\begin{cor}
\label{cor:Hoeffding2}
Let $ ( \Omega, \mathcal{F}, \P ) $ be a probability space, 
let $ N \in \N $, 
$ \varepsilon \in [0,\infty) $, $ a_1, a_2, \dots, \allowbreak a_N \in \R $, $ b_1 \in [a_1,\infty) $, $ b_2 \in [a_2, \infty) $, $ \dots $, $ b_N \in [ a_N, \infty) $
satisfy 
$
  \sum_{ n = 1 }^N
  ( b_n - a_n )^2 
  \neq 0
$, 
and let $ X_n \colon \Omega \to [ a_n, b_n ] $, $ n \in \{ 1, 2, \dots, N \} $, 
be independent random variables. 
Then
\begin{equation}
\label{eq:Hoeffding_inequality}
  \P\pr*{
    \abs*{
        \sum_{ n = 1 }^N
        \bpr{
          X_n
          -
          \E[ X_n ]
        }
    }
    \geq 
    \varepsilon
  }
  \leq
  2 
  \exp\pr*{
    \frac{
      - 2 \varepsilon^2 
    }{
      \sum_{ n = 1 }^N
      ( b_n - a_n )^2
    }
  }
  .
\end{equation}
\end{cor}

\begin{proof}[Proof of \cref{cor:Hoeffding2}]
Observe that 
\begin{equation}
\begin{split}
&
  \P\pr*{
    \abs*{
        \sum_{ n = 1 }^N
        \bpr{
          X_n
          -
          \E[ X_n ]
        }
    }
    \geq 
    \varepsilon
  }
\\ & =
  \P\pr*{
  \cu*{
      \br*{
        \sum_{ n = 1 }^N
        \bpr{
          X_n
          -
          \E[ X_n ]
        }
      }
    \geq 
    \varepsilon
  }
  \cup
  \cu*{ 
      \br*{
        \sum_{ n = 1 }^N
        \bpr{
          X_n
          -
          \E[ X_n ]
        }
      }
    \leq
    - \varepsilon
  }
  }
\\ & \leq 
  \P\pr*{
      \br*{
        \sum_{ n = 1 }^N
        \bpr{
          X_n
          -
          \E[ X_n ]
        }
      }
    \geq 
    \varepsilon
  }
  +
  \P\pr*{
      \br*{
        \sum_{ n = 1 }^N
        \bpr{
          X_n
          -
          \E[ X_n ]
        }
      }
    \leq
    - \varepsilon
  }
  .
\end{split}
\end{equation}
Combining this with \cref{cor:Hoeffding} and \cref{cor:Hoeffding_B} establishes \cref{eq:Hoeffding_inequality}. 
The proof of \cref{cor:Hoeffding2} is thus complete.
\end{proof}

\begin{cor}
\label{cor:Hoeffding3}
Let $ ( \Omega, \mathcal{F}, \P ) $ be a probability space, 
let $ N \in \N $, 
$ \varepsilon \in [0,\infty) $, $ a_1, a_2, \dots,\allowbreak a_N \in \R $, $ b_1 \in [a_1,\infty) $, $ b_2 \in [a_2, \infty) $, $ \dots $, $ b_N \in [ a_N, \infty) $
satisfy 
$
  \sum_{ n = 1 }^N
  ( b_n - a_n )^2 
  \neq 0
$,
and let $ X_n \colon \Omega \to [ a_n, b_n ] $, $ n \in \{ 1, 2, \dots, N \} $, 
be independent random variables. 
Then
\begin{equation}
\label{eq:Hoeffding_inequality3}
  \P\pr*{
    \frac{ 1 }{ N }
    \abs*{
        \sum_{ n = 1 }^N
        \bpr{
          X_n
          -
          \E[ X_n ]
        }
    }
    \geq 
    \varepsilon
  }
  \leq
  2 
  \exp\pr*{
    \frac{
      - 2 \varepsilon^2 
      N^2
    }{
      \sum_{ n = 1 }^N
      ( b_n - a_n )^2
    }
  }
  .
\end{equation}
\end{cor}

\begin{proof}[Proof of \cref{cor:Hoeffding3}]
Observe that \cref{cor:Hoeffding2} ensures that
\begin{eqsplit}
  \P\pr*{
    \frac{ 1 }{ N }
    \abs*{
        \sum_{ n = 1 }^N
        \bpr{
          X_n
          -
          \E[ X_n ]
        }
    }
    \geq 
    \varepsilon
  }
  &=
  \P\pr*{
    \abs*{
        \sum_{ n = 1 }^N
        \bpr{
          X_n
          -
          \E[ X_n ]
        }
    }
    \geq 
    \varepsilon N
  }
  \\&\leq
  2 
  \exp\pr*{
    \frac{
      - 2 ( \varepsilon N )^2 
    }{
      \sum_{ n = 1 }^N
      ( b_n - a_n )^2
    }
  }
  .
\end{eqsplit}
The proof of \cref{cor:Hoeffding3} is thus complete.
\end{proof}

\begin{exercise}{ex:hoeffding1}
Prove or disprove the following statement: 
For every probability space
  $(\Omega,\mc F,\P)$,
every
  $N\in\N$,
  $\eps\in[0,\infty)$,
  and every
  random variable
    $X=(X_1,X_2,\dots,X_N)\colon \Omega\to[-1,1]^N$
  with
    $\Forall a=(a_1,a_2,\dots,a_N)\in[-1,1]^N\colon \P(\bigcap_{i=1}^N\{X_i\leq a_i\})=\prod_{i=1}^N \frac{a_i+1}2$
  it holds that
\begin{equation}
  \P\Biggl(
    \frac1N\Biggl\lvert
      \sum_{i=1}^N(X_n-\E[X_n])
    \Biggr\rvert
    \geq
    \eps
  \Biggr)
  \leq
  2\exp\biggl(
    \frac{
      -\eps^2N
    }{
      2
    }
  \biggr).
\end{equation}
\end{exercise}

\begin{exercise}{ex:hoeffding2}
  Prove or disprove the following statement:
  For every probability space $(\Omega,\mc F,\P)$,
  every $N\in\N$,
  and every
  random variable
    $X=(X_1,X_2,\dots,X_N)\colon \Omega\to[-1,1]^N$
  with
    $\Forall a=(a_1,a_2,\dots,a_N)\in[-1,1]^N\colon \P(\bigcap_{i=1}^N\{X_i\leq a_i\})=\prod_{i=1}^N \frac{a_i+1}2$
  it holds that
  \begin{equation}
    \P\Biggl(\frac 1N
      \Biggl\lvert
        \sum_{n=1}^N (X_n-\E[X_n])
      \Biggr\rvert
      \geq
      \frac12
    \Biggr)
    \leq
    2\Bigl[
      \frac{
        e
      }{
        4
      }
    \Bigr]^N
    .
  \end{equation}
\end{exercise}

\begin{exercise}{ex:hoeffding3}
	Prove or disprove the following statement:
	For every probability space $(\Omega,\mc F,\P)$,
	every $N\in\N$,
  and every
  random variable
    $X=(X_1,X_2,\dots,X_N)\colon \Omega\to[-1,1]^N$
  with
    $\Forall a=(a_1,a_2,\dots,a_N)\in[-1,1]^N\colon \P(\bigcap_{i=1}^N\{X_i\leq a_i\})=\prod_{i=1}^N \frac{a_i+1}2$
  it holds that
	\begin{equation}
	  \P\Biggl(\frac 1N
      \Biggl\lvert
        \sum_{n=1}^N (X_n-\E[X_n])
      \Biggr\rvert
      \geq
      \frac12
	  \Biggr)
	  \leq
	  2\biggl[
		\frac{
		  e-e^{-3}
		}{
		  4
		}
	  \biggr]^N
	  .
	\end{equation}
\end{exercise}

\begin{exercise}{ex:hoeffding4}
	Prove or disprove the following statement:
	For every probability space $(\Omega,\mc F,\P)$,
	every $N\in\N$, $\eps\in[0,\infty)$,
	and every standard normal random variable
	  $X=(X_1,X_2,\dots,X_N)\colon \Omega\to\R^N$
	it holds that
	\begin{equation}
	  \P\Biggl(\frac 1N
      \Biggl\lvert
        \sum_{n=1}^N (X_n-\E[X_n])
      \Biggr\rvert
      \geq
      \eps
	  \Biggr)
	  \leq
	  2\exp\biggl(
      \frac{-\eps^2N}2
    \biggr)
	  .
	\end{equation}
\end{exercise}

\subsection{A strengthened Hoeffding's inequality}

\begin{lemma}
\label{hoeffding_bound_comparison1}
Let $f, g \colon (0,\infty) \to \R$ satisfy for all $x \in (0,\infty)$ that 
$f(x) = 2 \exp(-2x)$ and $g(x) = \frac{1}{4x}$.
Then
\begin{enumerate}[label=(\roman*)]
\item
\label{hoeffding_bound_comparison1:item1}
	it holds that $\lim_{x \to \infty} \frac{f(x)}{g(x)} = \lim_{x \searrow 0} \frac{f(x)}{g(x)} = 0$
and
\item
\label{hoeffding_bound_comparison1:item2}
it holds that 
	$g(\frac{1}{2}) = \frac{1}{2} < \frac{2}{3} < \frac{2}{e} = f(\frac{1}{2})$.
\end{enumerate}
\end{lemma}

\begin{proof}[Proof of \cref{hoeffding_bound_comparison1}]
Note that the fact that 
$\lim_{x \to \infty} \frac{\exp(-x)}{x^{-1}} = \lim_{x \searrow 0} \frac{\exp(-x)}{x^{-1}} = 0$ establishes \cref{hoeffding_bound_comparison1:item1}.
Moreover, observe that the fact that $e < 3$ implies \cref{hoeffding_bound_comparison1:item2}.
The proof of \cref{hoeffding_bound_comparison1} is thus complete.
\end{proof}

\begin{cor}
\label{cor:Hoeffding_strengthened}
Let $ ( \Omega, \mathcal{F}, \P ) $ be a probability space, 
let $ N \in \N $, 
$ \varepsilon \in (0,\infty) $, $ a_1, a_2, \dots, \allowbreak a_N \in \R $, $ b_1 \in [a_1,\infty) $, $ b_2 \in [a_2, \infty) $, $ \dots $, $ b_N \in [ a_N, \infty) $
satisfy 
$
  \sum_{ n = 1 }^N
  ( b_n - a_n )^2 
  \neq 0
$, 
and let $ X_n \colon \Omega \to [ a_n, b_n ] $, $ n \in \{ 1, 2, \dots, N \} $, 
be independent random variables. 
Then
\begin{equation}
\label{cor:Hoeffding_strengthened:concl}
  \P\pr*{
    \abs*{
        \sum_{ n = 1 }^N
        \bpr{
          X_n
          -
          \E[ X_n ]
        }
    }
    \geq 
    \varepsilon
  }
  \leq
  \min \cu*{
  1,
  2 
  \exp\pr*{
    \frac{
      - 2 \varepsilon^2 
    }{
      \sum_{ n = 1 }^N
      ( b_n - a_n )^2
    }
  }
  , 
  \frac{
     \sum_{ n = 1 }^N
     ( b_n - a_n )^2
   }{
      4 \varepsilon^2
   }
  }
  .
\end{equation}
\end{cor}

\begin{proof}[Proof of \cref{cor:Hoeffding_strengthened}]
Observe that \cref{lemma:easy_hoeffding}, \cref{cor:Hoeffding2}, and the fact that for all $B \in \mathcal{F}$ it holds that $\P(B) \leq 1$ establish \cref{cor:Hoeffding_strengthened:concl}. 
The proof of \cref{cor:Hoeffding_strengthened} is thus complete.
\end{proof}

\section{Covering number estimates}
\label{sect:covering_numbers}

\subsection{Entropy quantities}

\subsubsection{Covering radii (Outer entropy numbers)}

\begin{adef}{def:covering_radius}[Covering radii]
Let $ (X, d) $ be a metric space
and let $ n \in \N $.
Then we denote by $ \CovRad{ (X,d), n } \in [0,\infty] $ 
(we denote by $ \CovRad{ X, n } \in [0,\infty] $)
the extended real number given by
\begin{equation}
\label{eq:covering_radius}
  \CovRad{(X,d),n}
  =
  \inf\Bigl(
    \Bigl\{ 
      r \in [0,\infty] \colon
      \bigl(
        \Exists A\subseteq X \colon \bigl[
        (\card A\leq n)
        \land
        (
          \Forall x\in X\colon
          \Exists a\in A\colon
          d(a,x)\leq r
        )
        \bigr]
      \bigr)
    \Bigr\}
  \Bigr)
\end{equation}
and we call $ \CovRad{ (X,d), n }$ the 
$ n $-covering radius of $ ( X, d ) $ 
(we call $ \CovRad{ X, r } $ the $ n $-covering radius of $ X $). 
\end{adef}

\cfclear
\begin{lemma}
  \label{lem:covering_radius_char}
  Let $(X,d)$ be a metric space 
  and let $n\in\N$.
  Then
  \begin{equation}
    \label{eq:crc.claim}
    \CovRad{(X,d),n}
    =
    \begin{cases}
      0 & \colon X=\emptyset \\
    \begin{aligned}
    \inf\biggl(
      \biggl\{ 
        r \in [0,\infty) \colon
        \biggl(
          &\exists \, x_1, x_2, \dots, x_n \in X \colon
            \\
            &\hspace{-1cm}X \subseteq 
            \br*{ 
              \textstyle
              \bigcup\limits_{ m = 1 }^n 
              \displaystyle
              \cu*{ 
                v \in X \colon
                d( x_m , v ) \leq r
              }
            }
        \biggr)
      \biggr\}
      \cup \{ \infty \}
    \biggr) 
    \end{aligned}
    & \colon X\neq\emptyset
    \end{cases}
  \end{equation}
  \cfout.
\end{lemma}
\begin{proof}[Proof of \cref{lem:covering_radius_char}]
  Throughout this proof,
    assume without loss of generality that $X\neq\emptyset$
    and let $a\in X$.
  Note that the assumption that
    $d$ is a metric
  implies that for all 
    $x\in X$
  it holds that $d(a,x)\leq\infty$.
  Combining
    this
  with
    \cref{lem:covering_radius_aux}
  proves
    \cref{eq:crc.claim}.
  This completes the proof of \cref{lem:covering_radius_char}.
\end{proof}

\cfclear
\begin{exercise}{ex:covering_radius1}
Prove or disprove the following statement: 
For every metric space $ ( X, d ) $ 
and every $ n, m \in \N $
it holds that 
$ \CovRad{ ( X, d ), n } < \infty $
if and only if
$ \CovRad{ ( X, d ), m } < \infty $
\cfload
\end{exercise}

\cfclear
\begin{exercise}{ex:covering_radius2}
Prove or disprove the following statement: 
For every metric space $ ( X, d ) $ 
and every $ n \in \N $
it holds that 
$ ( X, d ) $ is bounded if and only if
$ \CovRad{ ( X, d ), n } < \infty $
\cfload.
\end{exercise}

\cfclear
\begin{exercise}{result:covering_rad_formula}
Prove or disprove the following statement: 
For every $ n \in \N $
and every metric space $ (X,d) $ with $ X \neq \emptyset $ 
it holds that
\begin{eqsplit}
  \CovRad{ (X,d), n } 
  &= 
  \inf\nolimits_{ x_1, x_2, \dots, x_n \in X }
  \sup\nolimits_{ v \in X }
  \min\nolimits_{ i \in \{ 1, 2, \dots, n \} }
  d( x_i, v )
  \\&= 
  \inf\nolimits_{ x_1, x_2, \dots, x_n \in X }
  \sup\nolimits_{ x_{ n + 1 } \in X }
  \min\nolimits_{ i \in \{ 1, 2, \dots, n \} }
  d( x_i, x_{ n + 1 } )
\end{eqsplit}
\cfload.
\end{exercise}

\subsubsection{Packing radii (Inner entropy numbers)}

\begin{adef}{def:packing_radius}[Packing radii]
Let $ (X, d) $ be a metric space
and let $ n \in \N $.
Then we denote by $ \PackRad{ (X,d), n } \in [0,\infty] $ 
(we denote by $ \PackRad{ X, n } \in [0,\infty] $)
the extended real number given by
\begin{multline}
\label{eq:packing_radius}
  \PackRad{(X,d),n}
  =
  \sup\bigl(
  \bigl\{ 
    r \in [0,\infty)
    \colon
    \bigl(
      \exists \, x_1, x_2, \dots, x_{ n + 1 } \in X 
      \colon\\
      \br*{
        \min\nolimits_{ 
          \substack{ 
            i, j \in \{ 1, 2, \dots, n + 1 \}, \,
            i \neq j 
          }
        }
        d( x_i, x_j )
      }
      > 2 r
    \bigr)
  \bigr\}
  \cup \{ 0 \}
  \bigr)
\end{multline}
and we call $ \PackRad{ (X,d), n }$ the 
$ n $-packing radius of $ ( X, d ) $ 
(we call $ \PackNum{ X, r } $ the $ n $-packing radius of $ X $). 
\end{adef}

\cfclear
\begin{exercise}{ex:packing_radius}
Prove or disprove the following statement: 
For every $ n \in \N $
and every metric space $ (X,d) $ with $ X \neq \emptyset $ 
it holds that
\begin{equation}
  \PackRad{ (X,d), n }
  =
  \tfrac{
    1
  }{ 2 }
    \br*{
        \sup\nolimits_{ x_1, x_2, \dots, x_{ n + 1 } \in X }
        \min\nolimits_{ 
          \substack{ 
            i, j \in \{ 1, 2, \dots, n + 1 \}, \,
            i \neq j 
          }
        }
        d( x_i, x_j )
    }
\end{equation}
\cfload.
\end{exercise}

\subsubsection{Packing numbers}

\begin{adef}{def:packing_number}[Packing numbers]
Let $ (X, d) $ be a metric space
and let $ r \in [0,\infty] $.
Then we denote by $ \PackNum{ (X,d), r } \in [0,\infty] $ 
(we denote by $ \PackNum{ X, r } \in [0,\infty] $)
the extended real number given by
\begin{multline}
\label{eq:packing_number}
  \PackNum{(X,d),r}
  =
  \sup\bigl(
  \bigl\{
    n \in \N
    \colon
    \bigl(
      \exists \, x_1, x_2, \dots, x_{ n + 1 } \in X \colon\\
      \br*{
        \min\nolimits_{ 
          \substack{ 
            i, j \in \{ 1, 2, \dots, n + 1 \}, \,
            i \neq j 
          }
        }
        d( x_i, x_j )
      }
      > 2 r
    \bigr)
  \bigr\}
  \cup \{ 0 \}
  \bigr)
\end{multline}
and we call $ \PackNum{ (X,d), r } $ the 
$ r $-packing number of $ ( X, d ) $ 
(we call $ \PackNum{ X, r } $ the $ r $-packing number of $ X $). 
\end{adef}

\subsection{Inequalities for packing entropy quantities in metric spaces}

\subsubsection{Lower bounds for packing radii based on lower bounds for 
packing numbers}

\cfclear
\begin{lemma}[Lower bounds for packing radii]
\label{lem:packing_number_bound0}
Let $ ( X, d ) $ be a metric space and
let $ n \in \N $, $ r \in [0,\infty] $ 
satisfy $ n \leq \PackNum{ (X,d), r } $
\cfload.
Then $ r \leq \PackRad{(X,d),n} $
\cfout.
\end{lemma}

\begin{proof}[Proof of \cref{lem:packing_number_bound0}] 
Note that \cref{eq:packing_number} ensures 
that there exist 
$ x_1, x_2, \dots, x_{ n + 1 } \in X $ such that 
\begin{equation}
      \br*{
        \min\nolimits_{ 
          \substack{ 
            i, j \in \{ 1, 2, \dots, n + 1 \}, \,
            i \neq j 
          }
        }
        d( x_i, x_j )
      }
      > 2 r
      .
\end{equation}
This implies 
that 
$ \PackRad{ (X,d), n } \geq r $
\cfload.
The proof of \cref{lem:packing_number_bound0} is thus complete.
\end{proof}

\subsubsection{Upper bounds for packing numbers based on upper bounds for packing radii}

\cfclear
\begin{lemma}
\label{lem:packing_number_bound}
Let $ ( X, d ) $ be a metric space and
let 
$ n \in \N $, $ r \in [0,\infty] $ 
satisfy $ \PackRad{(X,d),n} < r $
\cfload.
Then 
$
  \PackNum{ (X,d), r} < n
$
\cfout.
\end{lemma}

\begin{proof}[Proof of \cref{lem:packing_number_bound}] 
Observe that \cref{lem:packing_number_bound0} establishes that 
$
  \PackNum{ (X,d), r} < n
$ \cfload.
The proof of \cref{lem:packing_number_bound} is thus complete.
\end{proof}

\subsubsection{Upper bounds for packing radii based on upper bounds for covering radii}

\cfclear
\begin{lemma}
\label{lem:covering_0}
Let $ ( X, d ) $ be a metric space and let $ n \in \N $. 
Then $ \PackRad{ (X,d), n } \leq \CovRad{ (X,d), n } $
\cfout.
\end{lemma}

\begin{proof}[Proof of \cref{lem:covering_0}]
Throughout this proof, assume without loss of generality that 
$ \CovRad{ (X,d), n } < \infty $ and $ \PackRad{ (X,d), n } > 0 $, 
let 
$ r \in [ 0, \infty) $, 
$ x_1, x_2, \dots, x_n \in X $
satisfy 
\begin{equation}
\label{eq:covering_in_proof}
  X \subseteq 
  \br*{ 
    \bigcup_{ m = 1 }^n 
            \cu*{ 
              v \in X \colon
              d( x_m , v ) \leq r
            }
  }
  ,
\end{equation}
let 
$ {\bf r} \in [0,\infty) $,
$ {\bf x}_1, {\bf x}_2, \dots, {\bf x}_{ n + 1 } \in X $ 
satisfy 
\begin{equation}
\label{eq:proof_property_mathfrakd}
  \br*{
    \min\nolimits_{ 
      \substack{ 
        i, j \in \{ 1, 2, \dots, n + 1 \}, \,
        i \neq j 
      }
    }
    d( {\bf x}_i, {\bf x}_j )
  }
  > 2 {\bf r}
  ,
\end{equation}
and let 
$
  \varphi \colon X \to \{ 1, 2, \dots, n \} 
$
satisfy for all $ v \in X $ that 
\begin{equation}
\label{eq:def_varphi}
  \varphi(v) = \min\cu*{ m \in \{ 1, 2, \dots, n \} \colon v \in \{ w \in X \colon d( x_m, w ) \leq r \} }
  \cfadd{lem:covering_radius_char}
\end{equation}
\cfload.
Observe that \cref{eq:def_varphi} shows that for all $ v \in X $ it holds that
\begin{equation}
  v \in \cu*{ w \in X \colon d( x_{ \varphi(v) }, w ) \leq r } .
\end{equation}
Hence, we obtain that for all $ v \in X $ it holds that
\begin{equation}
\label{eq:proof_property_varphi}
  d( v, x_{ \varphi(v) } ) \leq r
\end{equation}
Moreover, note that the fact that 
$
  \varphi( {\bf x}_1 ), \varphi( {\bf x}_2 ), \dots, \varphi( {\bf x}_{ n + 1 } ) \in \{ 1, 2, \dots, n \} 
$
ensures that there exist 
$ i, j \in \{ 1, 2, \dots, n + 1 \} $ 
which satisfy 
\begin{equation}
  i \neq j 
\qquad 
  \text{and}
\qquad
  \varphi( {\bf x}_i ) = \varphi( {\bf x}_j )
  .
\end{equation}
The triangle inequality, \cref{eq:proof_property_mathfrakd}, and \cref{eq:proof_property_varphi} hence show that
\begin{equation}
  2 {\bf r} 
< 
  d( {\bf x}_i, {\bf x}_j ) 
\leq 
  d( {\bf x}_i, x_{ \varphi({\bf x}_i) } )
  +
  d( x_{ \varphi({\bf x}_i) } , {\bf x}_j )
=
  d( {\bf x}_i, x_{ \varphi({\bf x}_i) } )
  +
  d( {\bf x}_j, x_{ \varphi({\bf x}_j) } )
\leq 
  2 r
  .
\end{equation}
This implies that $ {\bf r} < r $.
The proof of \cref{lem:covering_0} is thus complete.
\end{proof}

\subsubsection{Upper bounds for packing radii in balls of metric spaces}

\cfclear
\begin{lemma}
\label{result:packing_radius_1}
Let $ ( X, d ) $ be a metric space, 
let $ n \in \N $, $ x \in X $, $ r \in (0,\infty] $, 
and let 
$ S = \cu*{ v \in X \colon d( x, v ) \leq r } $.
Then 
$ \PackRad{ ( S, d|_{ S \times S } ), n } \leq r $
\cfout.
\end{lemma}

\begin{proof}[Proof of \cref{result:packing_radius_1}]
Throughout this proof, assume without loss of generality that 
$
  \PackRad{ ( S, d|_{ S \times S } ), n } > 0
$ \cfload.
Observe that for all $ {\bf x}_1, {\bf x}_2, \dots, {\bf x}_{ n + 1 } \in S $, $ i, j \in \{ 1, 2, \dots, n + 1 \} $ 
it holds that 
\begin{equation}
  d( {\bf x}_i, {\bf x}_j ) 
  \leq 
  d( {\bf x}_i, x ) 
  +
  d( x, {\bf x}_j )
  \leq 
  2 r .
\end{equation}
Hence, we obtain that for all $ {\bf x}_1, {\bf x}_2, \dots, {\bf x}_{ n + 1 } \in S $ 
it holds that
\begin{equation}
\label{eq:packing_radius_proof_1}
  \min\nolimits_{ i, j \in \{ 1, 2, \dots, n + 1 \}, i \neq j } d( {\bf x}_i, {\bf x}_j )
  \leq 2 r .
\end{equation}
Moreover, note that \cref{eq:packing_radius} ensures that  for all 
$ \rho \in [ 0, \PackRad{ (S,d|_{S \times S}), n } ) $ 
there exist $ {\bf x}_1, {\bf x}_2, \dots,\allowbreak {\bf x}_{ n + 1 } \in S $ such that
\begin{equation}
  \min\nolimits_{ i, j \in \{ 1, 2, \dots, n + 1 \}, i \neq j } d( {\bf x}_i, {\bf x}_j ) > 2 \rho .
\end{equation}
This and \cref{eq:packing_radius_proof_1} demonstrate that for all 
$ \rho \in [ 0, \PackRad{ (S,d|_{S \times S}), n } ) $ 
it holds that $ 2 \rho < 2 r $.
The proof of \cref{result:packing_radius_1} is thus complete.
\end{proof}

\subsection{Inequalities for covering entropy quantities in metric spaces}

\subsubsection{Upper bounds for covering numbers based on upper bounds for covering radii}

\cfclear
\begin{lemma}
\label{result:covering_number_bound}
Let $ ( X, d ) $ be a metric space and let 
$ r \in [0,\infty] $, $ n \in \N $
satisfy 
$
  \CovRad{ (X,d), n } < r 
$ \cfload.
Then 
$
  \CovNum{ (X,d), r } \leq n
$ \cfout.
\end{lemma}
\begin{proof}[Proof of \cref{result:covering_number_bound}]
Observe that the assumption that 
$
  \CovRad{ (X,d), n } < r 
$
ensures that there exists $ A\subseteq X $ such that  
$\card A\leq n$ and 
\begin{equation}
  X \subseteq 
  \br*{ 
    \bigcup_{ a\in A }
    \cu*{ 
      v \in X \colon d( a, v ) \leq r
    }
  }
  .
\end{equation}
This establishes that 
$
  \CovNum{ (X,d), r } \leq n
$ \cfload.
The proof of \cref{result:covering_number_bound} is thus complete.
\end{proof}

\cfclear
\begin{lemma}
\label{result:covering_number_bound2}
Let $ ( X, d ) $ be a compact metric space
and let 
$ r \in [0,\infty] $, $ n \in \N $,
satisfy
$
  \CovRad{ (X,d), n } \leq r 
$ \cfload.
Then 
$
  \CovNum{ (X,d), r } \leq n
$ \cfout.
\end{lemma}
\begin{proof}[Proof of \cref{result:covering_number_bound2}]
Throughout this proof,
assume without loss of generality that $X\neq\emptyset$
and let $ x_{ k, m } \in X $, 
$ m \in \{ 1, 2, \dots, n \} $,
$ k \in \N $,
satisfy for all $ k \in \N $ that
\begin{equation}
\label{eq:covering_number_bound2_B}
  X \subseteq 
  \br*{ 
    \bigcup_{ m = 1 }^n 
    \cu*{ 
      v \in X \colon d( x_{ k, m }, v ) \leq r + \tfrac{ 1 }{ k }
    }
  }
\end{equation}
(cf.\ \cref{lem:covering_radius_aux}).
Note that the assumption that 
$ ( X, d ) $ is a compact metric space 
demonstrates that there exist 
$ 
  \mathfrak{x} = 
  ( \mathfrak{x}_m )_{ m \in \{ 1, 2, \dots, n \} } \colon 
  \{ 1, 2, \dots, n \} \to X
$
and 
$
  k = ( k_l )_{ l \in \N } \colon \N \to \N
$
which satisfy that
\begin{equation}
  \limsup\nolimits_{ l \to \infty }
  \max\nolimits_{ m \in \{ 1, 2, \dots, n \} }
  d( \mathfrak{x}_m , x_{ k_l, m } )
  = 0
\qquad 
  \text{and}
\qquad
  \limsup\nolimits_{ l \to \infty }
  k_l = \infty
  .
\end{equation}
Next observe that the assumption that $ d $ is a metric ensures that 
for all $ v \in X $, $ m \in \{ 1, 2, \dots, n \} $, $ l \in \N $ 
it holds that
\begin{equation}
  d( v, \mathfrak{x}_m )
  \leq 
  d( v, x_{ k_l, m } )
  +
  d( x_{ k_l, m }, \mathfrak{x}_m )
  .
\end{equation}
This and \cref{eq:covering_number_bound2_B} prove that 
for all $ v \in X $, $ l \in \N $ 
it holds that
\begin{equation}
\begin{split}
  \min\nolimits_{ m \in \{ 1, 2, \dots, n \} }
  d( v, \mathfrak{x}_m )
&
\leq 
  \min\nolimits_{ m \in \{ 1, 2, \dots, n \} }
  \br*{
    d( v, x_{ k_l, m } )
    +
    d( x_{ k_l, m }, \mathfrak{x}_m )
  }
\\ & 
\leq
  \br*{
    \min\nolimits_{ m \in \{ 1, 2, \dots, n \} }
    d( v, x_{ k_l, m } )
  }
  +
  \br*{
    \max\nolimits_{ m \in \{ 1, 2, \dots, n \} }
    d( x_{ k_l, m }, \mathfrak{x}_m )
  }
\\ &
\leq
  \bbr{
    r
    +
    \tfrac{ 1 }{ k_l }
  }
  +
  \br*{
    \max\nolimits_{ m \in \{ 1, 2, \dots, n \} }
    d( x_{ k_l, m }, \mathfrak{x}_m )
  }
  .
\end{split}
\end{equation}
Hence, we obtain for all $ v \in X $ that
\begin{equation}
\begin{split}
  \min\nolimits_{ m \in \{ 1, 2, \dots, n \} }
  d( v, \mathfrak{x}_m )
&
\leq 
  \limsup\nolimits_{ l \to \infty }
  \bpr{
    \bbr{
      r
      +
      \tfrac{ 1 }{ k_l }
    }
    +
    \bbr{
      \max\nolimits_{ m \in \{ 1, 2, \dots, n \} }
      d( x_{ k_l, m }, \mathfrak{x}_m )
    }
  }
  = r
  .
\end{split}
\end{equation}
This establishes that 
$
  \CovNum{ (X,d), r } \leq n
$ \cfload.
The proof of \cref{result:covering_number_bound2} is thus complete.
\end{proof}

\subsubsection{Upper bounds for covering radii based on upper bounds 
for covering numbers}

\cfclear
\begin{lemma}
\label{result:covering_radii_bound}
Let $ ( X, d ) $ be a metric space and let 
$ r \in [0,\infty] $, $ n \in \N $
satisfy 
$
  \CovNum{ (X,d), r } \leq n
$ \cfload.
Then 
$
  \CovRad{ (X,d), n } \leq r 
$ \cfout.
\end{lemma}
\begin{proof}[Proof of \cref{result:covering_radii_bound}]
Observe that the assumption that 
$
  \CovNum{ (X,d), r } \leq n
$
ensures that there exists $A\subseteq X$ %
such that $\card A\leq n$ and
\begin{equation}
  X \subseteq 
  \br*{ 
    \bigcup_{ a\in A }
    \cu*{ 
      v \in X \colon d( a, v ) \leq r
    }
  }
  .
\end{equation}
This establishes that 
$
  \CovRad{ (X,d), n } \leq r 
$ \cfload.
The proof of \cref{result:covering_radii_bound} is thus complete.
\end{proof}

\subsubsection{Upper bounds for covering radii based on upper bounds for packing radii}

\cfclear
\begin{lemma}
\label{lem:covering_1}
Let $ ( X, d ) $ be a metric space and let $ n \in \N $. 
Then $ \CovRad{ (X,d), n } \leq 2 \PackRad{ (X,d), n } $ \cfout.
\end{lemma}

\begin{proof}[Proof of \cref{lem:covering_1}]
Throughout this proof, assume w.l.o.g. that $X\neq\emptyset$, 
assume without loss of generality that $ \PackRad{ (X,d), n } < \infty $, 
let $ r \in [0,\infty] $ satisfy $ r > \PackRad{ (X,d), n } $, and let 
$ N \in \N_0 \cup\{\infty\}$ satisfy 
$
  N = \PackNum{ (X,d), r } 
$ \cfload.
Observe that \cref{lem:packing_number_bound} ensures that 
\begin{equation}
  N = \PackNum{ (X,d), r } < n .
\end{equation}
Moreover, note that the fact that 
$ N = \PackNum{ (X,d), r } $ and \cref{eq:packing_number} 
demonstrate that for all 
$ x_1, x_2, \dots, x_{ N + 1 }, x_{ N + 2 } \in X $ 
it holds that
\begin{equation}
\label{eq:N_plus_2_packing}
  \min\nolimits_{ 
    i, j \in \{ 1, 2, \dots, N + 2 \},\, i \neq j
  }
  d( x_i, x_j ) 
  \leq 2 r
  .
\end{equation}
In addition, observe that the fact that $ N = \PackNum{ (X,d), r } $ and \cref{eq:packing_number} 
imply that there exist 
$ x_1, x_2, \dots, x_{ N + 1 } \in X $ which satisfy that
\begin{equation}
  \min\bigl(\{
    d(x_i,x_j)
    \colon
    i, j \in \{ 1, 2, \dots, N + 1 \},\, i \neq j
  \}\cup\{\infty\}\bigr)
  > 2 r
  .
\end{equation}
Combining this with \cref{eq:N_plus_2_packing} establishes that 
for all $ v \in X $ it holds that
\begin{equation}
  \min\nolimits_{ i \in \{ 1, 2, \dots, N \} } d( x_i, v ) \leq 2 r 
  .
\end{equation}
Hence, we obtain that for all $ w \in X $ it holds that
\begin{equation}
  w \in 
  \br*{ 
    \bigcup_{ m = 1 }^n
    \cu*{ v \in X \colon d( x_i, v ) \leq 2 r }
  }
  .
\end{equation}
Therefore, we obtain that
\begin{equation}
  X \subseteq
  \br*{ 
    \bigcup_{ m = 1 }^n
    \cu*{ v \in X \colon d( x_i, v ) \leq 2 r }
  }
  .
\end{equation}
Combining this and %
  \cref{lem:covering_radius_char}
shows that 
$
  \CovRad{ (X,d), n } \leq 2 r
$ \cfload.
The proof of \cref{lem:covering_1} is thus complete.
\end{proof}

\subsubsection{Equivalence of covering and packing radii}

\cfclear
\begin{cor}
\label{cor:covering_2}
Let $ ( X, d ) $ be a metric space and let $ n \in \N $. 
Then 
$ 
  \PackRad{ (X,d), n } \leq \CovRad{ (X,d), n } \leq 2 \PackRad{ (X,d), n } 
$ \cfout.
\end{cor}

\begin{proof}[Proof of \cref{cor:covering_2}]
Observe that \cref{lem:covering_0} and \cref{lem:covering_1} establish that 
$ 
  \PackRad{ (X,d), n } \leq \CovRad{ (X,d), n } \leq 2 \PackRad{ (X,d), n } 
$ \cfload.
The proof of \cref{cor:covering_2} is thus complete.
\end{proof}

\subsection{Inequalities for entropy quantities in finite-dimensional vector spaces}

\subsubsection{Measures induced by Lebesgue--Borel measures}

\begin{lemma}
\label{result:packing_radius_measure}
Let $ ( V, \opnorm{\cdot} ) $ be a normed vector space, 
let $ N \in \N $,
let $ b_1, b_2, \dots, b_N \in V $ be a Hamel-basis of $ V $, 
let $ \lambda \colon \mathcal{B}( \R^N ) \to [0,\infty] $ be the Lebesgue--Borel measure on $ \R^N $, 
let $ \Phi \colon \R^N \to V $ satisfy for all $ r = ( r_1, r_2, \dots, r_N ) \in \R^N $ that
$  
  \Phi( r ) = r_1 b_1 + r_2 b_2 + \ldots + r_N b_N 
$, 
and let $ \nu \colon \mathcal{B}( V ) \to [0,\infty] $ satisfy for all 
$ A \in \mathcal{B}( V ) $ that
\begin{equation}
\label{eq:measure_nu_packing_def}
  \nu( A )
  =
  \lambda( \Phi^{ - 1 }( A ) )
  . 
\end{equation}
Then
\begin{enumerate}[label=(\roman{*})]
\item 
\label{item:linear}
it holds that $ \Phi $ is linear, 
\item 
\label{item:continuity_bound}
it holds for all $ r = ( r_1, r_2, \dots, r_N ) \in \R^N $ that
$
  \opnorm{\Phi( r )}
\leq
  \bigl[ 
    \sum_{ n = 1 }^N 
    \opnorm{b_n}^2
  \bigr]^{ \nicefrac{ 1 }{ 2 } }
  \allowbreak
  \bigl[ 
    \sum_{ n = 1 }^N 
    \abs{ r_n }^2
  \bigr]^{ \nicefrac{ 1 }{ 2 } }
$,
\item 
\label{item:Phi_continuous}
it holds that $ \Phi \in C( \R^N, V ) $,
\item 
\label{item:Phi_bijective}
it holds that $ \Phi $ is bijective, 
\item 
\label{item:measure_space}
it holds that 
$ ( V, \mathcal{B}( V ), \nu ) $ is a measure space, 
\item
\label{item:affine_transform_measure}
it holds for all $ r \in (0,\infty) $, $ v \in V $, $ A \in \mathcal{B}( V ) $ that
$
  \nu\pr*{ \cu*{ ( r a + v ) \in V \colon a \in A } }
  =
  r^N \nu( A ) 
$,
\item 
\label{item:scalar_mult}
it holds for all $ r \in (0,\infty) $ that
$
  \nu\pr*{ 
    \cu*{ v \in V \colon \opnorm{v} \leq r }
  }
  =
  r^N
  \nu\pr*{ 
    \cu*{ v \in V \colon \opnorm{v} \leq 1 }
  }
$,
and 
\item 
\label{item:nu_not_zero}
it holds that 
$ \nu\pr*{ \cu*{ v \in V \colon \opnorm{v} \leq 1 } } > 0 $.
\end{enumerate}
\end{lemma}

\begin{proof}[Proof of \cref{result:packing_radius_measure}]
Note that for all $ r = ( r_1, r_2, \dots, r_N ) $, $ s = (s_1, s_2, \dots, s_N ) \in \R^N $, 
$ \rho \in \R $ it holds that 
\begin{equation}
  \Phi( \rho r + s )
  =
  ( \rho r_1 + s_1 ) b_1 
  +
  ( \rho r_2 + s_2 ) b_2 
  +
  \dots 
  +
  ( \rho r_N + s_N ) b_N
  =
  \rho \Phi( r ) + \Phi( s )
  .
\end{equation}
This establishes \cref{item:linear}.
Next observe that H\"{o}lder's inequality 
shows that 
for all $ r = ( r_1, r_2, \dots, r_N ) \in \R^N $ it holds that
\begin{equation}
  \opnorm{ \Phi( r ) }
  =
  \opnorm{ r_1 b_1 + r_2 b_2 + \dots + r_N b_N }
\leq 
  \sum_{ n = 1 }^N 
  \abs*{ r_n }
  \opnorm{ b_n }
\leq
  \br*{ 
    \sum_{ n = 1 }^N 
    \opnorm{ b_n }^2
  }^{ \nicefrac{ 1 }{ 2 } }
  \br*{ 
    \sum_{ n = 1 }^N 
    \abs*{ r_n }^2
  }^{ \nicefrac{ 1 }{ 2 } }
  .
\end{equation}
This establishes \cref{item:continuity_bound}. 
Moreover, note that \cref{item:continuity_bound} 
proves \cref{item:Phi_continuous}. 
Furthermore, observe that the assumption that $ b_1, b_2, \dots, b_N \in V $
is a Hamel-basis of $ V $ establishes \cref{item:Phi_bijective}. 
Next note that \cref{eq:measure_nu_packing_def} and \cref{item:Phi_continuous} 
prove \cref{item:measure_space}. 
In addition, observe that the integral transformation theorem shows that
for all $ r \in (0,\infty) $, $ v \in \R^N $, $ A \in \mathcal{B}( \R^N ) $ it holds that
\begin{equation}
\begin{split}
  \lambda\pr*{ \cu*{ ( r a + v ) \in \R^N \colon a \in A } }
& =
  \lambda\pr*{ \cu*{ r a \in \R^N \colon a \in A } }
  =
  \int_{ \R^N } \ind{ \{ r a \in \R^N \colon a \in A \} }( x ) \, \diff x 
\\ & =
  \int_{ \R^N } \ind{A}( \tfrac{ x }{ r } ) \, \diff x 
  = 
  r^N
  \int_{ \R^N } \ind{A}( x ) \, \diff x 
  = 
  r^N \lambda( A )
  .
\end{split}
\end{equation}
Combining \cref{item:linear} and \cref{item:Phi_bijective} hence demonstrates that
for all $ r \in (0,\infty) $, $ v \in V $, $ A \in \mathcal{B}( V ) $ it holds that
\begin{equation}
\label{eq:affine_transformation_measure_proof_packing}
\begin{split}
  \nu\pr*{ \cu*{ ( r a + v ) \in V \colon a \in A } }
&
  =
  \lambda\pr*{ \Phi^{ - 1 }\pr*{ \cu*{ ( r a + v ) \in V \colon a \in A } } }
  \\&=
  \lambda\pr*{ \cu*{ \Phi^{ - 1 }( r a + v ) \in \R^N \colon a \in A } }
\\ & =
  \lambda\pr*{ \cu*{ \br*{ r \Phi^{ - 1 }( a ) + \Phi^{ - 1 }( v ) } \in \R^N \colon a \in A } }
\\ &
  =
  \lambda\pr*{ \cu*{ \br*{ r a + \Phi^{ - 1 }( v ) } \in \R^N \colon a \in \Phi^{ - 1 }( A ) } }
  \\&= 
  r^N \lambda( \Phi^{ - 1 }( A ) )
  = 
  r^N \nu( A )
  .
\end{split}
\end{equation}
This establishes \cref{item:affine_transform_measure}. 
Hence, we obtain that for all $ r \in (0,\infty) $ it holds that
\begin{eqsplit}
\label{eq:multiplication_transformation_measure_proof_packing}
  \nu\pr*{ 
    \cu*{ v \in V \colon \opnorm v \leq r }
  }
  &=
  \nu\pr*{ 
    \cu*{ r v \in V \colon \opnorm v \leq 1 }
  }
  \\&=
  r^N
  \nu\pr*{ 
    \cu*{ v \in V \colon \opnorm v \leq 1 }
  }
  \\&=
  r^N
  \nu( X )
  .
\end{eqsplit}
This establishes \cref{item:scalar_mult}.
Furthermore, observe that \cref{eq:multiplication_transformation_measure_proof_packing} demonstrates that 
\begin{eqsplit}
  \infty = \lambda( \R^N ) = \nu( V ) 
  &= 
  \limsup_{ r \to \infty }
  \bbbr{
    \nu\pr*{
      \cu*{ v \in V \colon \opnorm{v} \leq r }
    }
  }
  \\&=
  \limsup_{ r \to \infty }
  \bbbr{
    r^N
    \nu(\{v\in V\colon \opnorm v\leq 1\})
  }
  .
\end{eqsplit}
Hence, we obtain that $ \nu( \{v\in V\colon \opnorm v\leq 1\} ) \neq 0 $. 
This establishes \cref{item:nu_not_zero}. 
The proof of \cref{result:packing_radius_measure} is thus complete.
\end{proof}

\subsubsection{Upper bounds for packing radii}

\cfclear
\begin{lemma}
\label{result:upper_packing_radius}
Let $ ( V, \opnorm{ \cdot } ) $ be a normed vector space, 
let 
$ X = \cu*{ v \in V \colon \opnorm{ v } \leq 1 } $, 
let $ d \colon X \times X \to [0,\infty) $ satisfy for all 
$ v, w \in X $ that 
$ d(v,w) = \opnorm{ v - w } $,
and 
let $ n, N \in \N $ satisfy $ N = \dim(V) $. Then 
\begin{equation}
  \PackRad{ (X,d) , n } 
  \leq 
  2 \, ( n + 1 )^{ - \nicefrac{ 1 }{ N } } 
\end{equation}
\cfout.
\end{lemma}

\begin{proof}[Proof of \cref{result:upper_packing_radius}]
Throughout this proof, assume without loss of generality that 
$
  \PackRad{ (X,d) , n } > 0
$, 
let $ \rho \in [0, \PackRad{ (X,d) , n }) $, 
let $ \lambda \colon \mathcal{B}( \R^N ) \to [0,\infty] $ be the Lebesgue-Borel measure on $ \R^N $, 
let $ b_1, b_2, \dots, b_N \in V $ be a Hamel-basis of $ V $, 
let $ \Phi \colon \R^N \to V $ satisfy 
for all $ r = ( r_1, r_2, \dots, r_N ) \in \R^N $ that
\begin{equation}
  \Phi( r ) = r_1 b_1 + r_2 b_2 + \ldots + r_N b_N ,
\end{equation}
and let $ \nu \colon \mathcal{B}( V ) \to [0,\infty] $ satisfy for all 
$ A \in \mathcal{B}( V ) $ that
\begin{equation}
  \nu( A )
  =
  \lambda( \Phi^{ - 1 }( A ) )
\end{equation}
\cfload.
Observe that \cref{result:packing_radius_1} ensures that 
$
  \rho < \PackRad{ (X,d), n } \leq 1
$.
Moreover, note that \cref{eq:packing_radius} shows that there exist 
$ x_1, x_2, \dots, x_{ n + 1 } \in X $ which satisfy 
\begin{equation}
\label{eq:min_condition_proof}
  \min\nolimits_{ i, j \in \{ 1, 2, \dots, n + 1 \}, i \neq j } \opnorm{ x_i - x_j }
  =
  \min\nolimits_{ i, j \in \{ 1, 2, \dots, n + 1 \}, i \neq j } d( x_i, x_j ) 
  > 2 \rho .
\end{equation}
Observe that \cref{eq:min_condition_proof} ensures that
for all $ i, j \in \{ 1, 2, \dots, n + 1 \} $
with $ i \neq j $ it holds that
\begin{equation}
\label{eq:balls_are_disjoint_proof_packing}
  \cu*{ 
    v \in V \colon 
    \opnorm{ x_i - v } \leq \rho 
  }
  \cap 
  \cu*{ 
    v \in V \colon 
    \opnorm{ x_j - v } \leq \rho 
  }
  = \emptyset .
\end{equation}
Moreover, note that \cref{eq:min_condition_proof} and the fact that 
$
  \rho < 1
$
show that
for all $ j \in \{ 1, 2, \dots, n + 1 \} $, 
$ w \in \{ v \in X \colon d( x_j, v ) \leq \rho \} $ 
it holds that
\begin{equation}
  \opnorm{ w }
  \leq 
  \opnorm{ w - x_j }
  +
  \opnorm{ x_j }
  \leq 
  \rho + 1
  \leq 2 .
\end{equation}
Therefore, we obtain that for all $ j \in \{ 1, 2, \dots, n + 1 \} $
it holds that
\begin{equation}
\label{eq:balls_are_in_B2_proof_packing}
  \cu*{ v \in V \colon \opnorm{ v - x_j } \leq \rho }
  \subseteq
  \cu*{ v \in V \colon \opnorm{ v } \leq 2 }
  .
\end{equation} 
Next observe that \cref{result:packing_radius_measure} 
ensures that $ ( V, \mathcal{B}( V ), \nu ) $ is a measure space. 
Combining this and \cref{eq:balls_are_disjoint_proof_packing} with \cref{eq:balls_are_in_B2_proof_packing} proves that
\begin{eqsplit}
  \sum_{ j = 1 }^{ n + 1 }
  \nu\pr*{
    \cu*{ v \in V \colon \opnorm{ v - x_j } \leq \rho }
  }
  &=
  \nu\pr*{
    \bigcup_{ j = 1 }^{ n + 1 } 
    \cu*{ v \in V \colon \opnorm{ v - x_j } \leq \rho }
  }
  \\&\leq 
  \nu\pr*{
    \cu*{ v \in V \colon \opnorm{ v } \leq 2 }
  }
  .
\end{eqsplit}
\cref{result:packing_radius_measure} 
hence shows that 
\begin{equation}
\label{eq:sigma_additivity_proof_packing}
\begin{split}
    \pr*{ n + 1 }
    \rho^N
    \nu(X)
  &=
  \sum_{ j = 1 }^{ n + 1 }
  \br*{
    \rho^N
    \nu\pr*{
      \cu*{ v \in V \colon \opnorm{ v } \leq 1 }
    }
  }
  \\&=
  \sum_{ j = 1 }^{ n + 1 }
  \nu\pr*{
    \cu*{ v \in V \colon \opnorm{ v } \leq \rho }
  }
\\ &  
  =
  \sum_{ j = 1 }^{ n + 1 }
  \nu\pr*{
    \cu*{ v \in V \colon \opnorm{ v - x_j } \leq \rho }
  }
  \leq 
  \nu\pr*{
    \cu*{ v \in V \colon \opnorm{ v } \leq 2 }
  }
  \\&=
  2^N 
  \nu\pr*{ 
    \cu*{ v \in V \colon \opnorm{ v } \leq 1 }
  }
  =
  2^N 
  \nu( X )
  .
\end{split}
\end{equation}
Next observe that \cref{result:packing_radius_measure} demonstrates that $ \nu( X ) > 0 $. 
Combining this 
with \cref{eq:sigma_additivity_proof_packing} assures that
$
  \pr*{ n + 1 } \rho^N \leq 2^N
$.
Therefore, we obtain that 
$
  \rho^N \leq \pr*{ n + 1 }^{ - 1 } 2^N
$. 
Hence, we obtain that 
$
  \rho \leq 2 \pr*{ n + 1 }^{ - \nicefrac{ 1 }{ N } }
$.
The proof of \cref{result:upper_packing_radius} is thus complete.
\end{proof}

\subsubsection{Upper bounds for covering radii}

\cfclear
\begin{cor}
\label{result:upper_covering_radius}
Let $ ( V, \opnorm{ \cdot } ) $ be a normed vector space, 
let 
$ X = \cu*{ v \in V \colon \opnorm{ v } \leq 1 } $, 
let $ d \colon X \times X \to [0,\infty) $ satisfy for all 
$ v, w \in X $ that 
$ d(v,w) = \opnorm{ v - w } $,
and 
let $ n, N \in \N $ satisfy $ N = \dim(V) $. Then 
\begin{equation}
\label{eq:upper_covering_radius}
  \CovRad{ (X,d) , n } 
  \leq 
  4 \, ( n + 1 )^{ - \nicefrac{ 1 }{ N } } 
\end{equation}
\cfout.
\end{cor}

\begin{proof}[Proof of \cref{result:upper_covering_radius}]
Observe that \cref{cor:covering_2} and \cref{result:upper_packing_radius} establish \cref{eq:upper_covering_radius}. 
The proof of \cref{result:upper_covering_radius} is thus complete.
\end{proof}

\subsubsection{Lower bounds for covering radii}

\cfclear
\begin{lemma}
\label{result:lower_covering_radius}
Let $ ( V, \opnorm{ \cdot } ) $ be a normed vector space, 
let 
$ X = \cu*{ v \in V \colon \opnorm{ v } \leq 1 } $, 
let $ d \colon X \times X \to [0,\infty) $ satisfy for all 
$ v, w \in X $ that 
$ d(v,w) = \opnorm{ v - w } $,
and 
let $ n, N \in \N $ satisfy $ N = \dim(V) $. Then 
\begin{equation}
\label{eq:lower_covering_radius}
  n^{ - \nicefrac{ 1 }{ N } } 
  \leq 
  \CovRad{ (X,d) , n } 
\end{equation}
\cfout.
\end{lemma}

\begin{proof}[Proof of \cref{result:lower_covering_radius}]
Throughout this proof, assume without loss of generality that 
$
  \CovRad{ (X,d) , n } < \infty
$, 
let $ \rho \in ( \CovRad{ (X,d) , n } , \infty ) $, 
let $ \lambda \colon \mathcal{B}( \R^N ) \to [0,\infty] $ be the Lebesgue-Borel measure on $ \R^N $, 
let $ b_1, b_2, \dots, b_N \in V $ be a Hamel-basis of $ V $, 
let $ \Phi \colon \R^N \to V $ satisfy 
for all $ r = ( r_1, r_2, \dots, r_N ) \in \R^N $ that
\begin{equation}
  \Phi( r ) = r_1 b_1 + r_2 b_2 + \ldots + r_N b_N ,
\end{equation}
and let $ \nu \colon \mathcal{B}( V ) \to [0,\infty] $ satisfy for all 
$ A \in \mathcal{B}( V ) $ that
\begin{equation}
  \nu( A )
  =
  \lambda( \Phi^{ - 1 }( A ) )
\end{equation}
\cfload.
The fact that $ \rho > \CovRad{ (X,d), n } $ demonstrates that there exist 
$ x_1, x_2, \dots, \allowbreak x_n \in X $ which satisfy 
\begin{equation}
  X \subseteq 
  \br*{ 
    \bigcup_{ m = 1 }^n 
    \cu*{ 
      v \in X \colon d(x_m,v) \leq \rho 
    }
  }
  .
\end{equation}
\cref{result:packing_radius_measure} hence shows that
\begin{equation}
\begin{split}
  \nu( X )
& \leq 
  \nu\pr*{
    \bigcup_{ m = 1 }^n 
    \cu*{ 
      v \in X \colon d(x_m,v) \leq \rho 
    }
  }
  \leq
  \sum_{ m = 1 }^n
  \nu\pr*{
    \cu*{ 
      v \in X \colon d(x_m,v) \leq \rho 
    }
  }
\\ & =
  \sum_{ m = 1 }^n
  \br*{
    \rho^N
    \nu\pr*{
      \cu*{ 
        v \in X \colon d(x_m,v) \leq 1
      }
    }
  }
  \leq
  n
  \rho^N 
  \nu( X )
  .
\end{split}
\end{equation}
This and \cref{result:packing_radius_measure} demonstrate that $ 1 \leq n \rho^N $. 
Hence, we obtain that 
$
  \rho^N \geq n^{ - 1 }
$. 
This ensures that 
$
  \rho \geq n^{ - \nicefrac{ 1 }{ N } }
$.
The proof of \cref{result:lower_covering_radius} is thus complete.
\end{proof}

\subsubsection{Lower and upper bounds for covering radii}

\cfclear
\begin{cor}
\label{result:lower_and_upper_covering_radius}
Let $ ( V, \opnorm{ \cdot } ) $ be a normed vector space, 
let 
$ X = \cu*{ v \in V \colon \opnorm{ v } \leq 1 } $, 
let $ d \colon X \times X \to [0,\infty) $ satisfy for all 
$ v, w \in X $ that 
$ d(v,w) = \opnorm{ v - w } $,
and 
let $ n, N \in \N $ satisfy $ N = \dim(V) $. Then 
\begin{equation}
\label{eq:lower_and_upper_covering_radius}
  n^{ - \nicefrac{ 1 }{ N } } 
  \leq 
  \CovRad{ (X,d) , n } 
  \leq 
  4 \,
  ( n + 1 )^{ - \nicefrac{ 1 }{ N } } 
\end{equation}
\cfout.
\end{cor}

\begin{proof}[Proof of \cref{result:lower_and_upper_covering_radius}]
Observe that \cref{result:upper_covering_radius} and \cref{result:lower_covering_radius} establish 
\cref{eq:lower_and_upper_covering_radius}.  
The proof of \cref{result:lower_and_upper_covering_radius} is thus complete.
\end{proof}

\subsubsection{Scaling property for covering radii}

\cfclear
\begin{lemma}
\label{result:scaling_covering}
Let $ ( V, \opnorm{ \cdot } ) $ be a normed vector space, 
let $ d \colon V \times V \to [0,\infty) $ satisfy for all $ v, w \in V $ that
$ d( v, w ) = \opnorm{ v - w } $, 
let $ n \in \N $, $ r \in (0,\infty) $,
and 
let $ X \subseteq V $ and $ \mathfrak{X} \subseteq V $ satisfy 
$ \mathfrak{X} = \{ r v \in V \colon v \in X \} $.
Then 
\begin{equation} 
\label{eq:scaling_covering}
  \CovRad{ ( \mathfrak{X}, d|_{ \mathfrak{X} \times \mathfrak{X} } ), n }
  =
  r \,
  \CovRad{ ( X, d|_{ X \times X } ), n }
\end{equation}
\cfout.
\end{lemma}

\begin{proof}[Proof of \cref{result:scaling_covering}]
Throughout this proof, let $ \Phi \colon V \to V $ satisfy 
for all $ v \in V $ that $ \Phi( v ) = r v $. 
Observe that \cref{result:covering_rad_formula} shows that
\begin{equation}
\begin{split}
  r \,
  \CovRad{ (X,d), n } 
& = 
  r 
  \br*{
    \inf\nolimits_{ x_1, x_2, \dots, x_n \in X }
    \sup\nolimits_{ v \in X }
    \min\nolimits_{ i \in \{ 1, 2, \dots, n \} }
    d( x_i, v )
  }
\\ & =
    \inf\nolimits_{ x_1, x_2, \dots, x_n \in X }
    \sup\nolimits_{ v \in X }
    \min\nolimits_{ i \in \{ 1, 2, \dots, n \} }
    \opnorm{ r x_i - r v }
\\ & =
    \inf\nolimits_{ x_1, x_2, \dots, x_n \in X }
    \sup\nolimits_{ v \in X }
    \min\nolimits_{ i \in \{ 1, 2, \dots, n \} }
    \opnorm{ \Phi( x_i ) - \Phi( v ) }
\\ & =
    \inf\nolimits_{ x_1, x_2, \dots, x_n \in X }
    \sup\nolimits_{ v \in X }
    \min\nolimits_{ i \in \{ 1, 2, \dots, n \} }
    d( \Phi( x_i ) , \Phi( v ) )
\\ & =
    \inf\nolimits_{ x_1, x_2, \dots, x_n \in X }
    \sup\nolimits_{ v \in \mathfrak{X} }
    \min\nolimits_{ i \in \{ 1, 2, \dots, n \} }
    d( \Phi( x_i ) , v )
\\ & =
    \inf\nolimits_{ x_1, x_2, \dots, x_n \in \mathfrak{X} }
    \sup\nolimits_{ v \in \mathfrak{X} }
    \min\nolimits_{ i \in \{ 1, 2, \dots, n \} }
    d( x_i, v )
  =
  \CovRad{ ( \mathfrak{X}, d|_{ \mathfrak{X} \times \mathfrak{X} } ), n }
\end{split}
\end{equation}
\cfload.
This establishes \cref{eq:scaling_covering}. 
The proof of \cref{result:scaling_covering} is thus complete.
\end{proof}

\subsubsection{Upper bounds for covering numbers}

\cfclear
\begin{prop}
\label{prop:covering}
Let
$ ( V, \opnorm\cdot ) $ 
be a normed vector space with $ \dim( V ) < \infty $,
let
$ r, R \in (0,\infty) $, 
$ 
  X = \{ v \in V \colon \opnorm{v} \leq R \} 
$, 
and let
$ d \colon X \times X \to [0,\infty) $ 
satisfy for all 
$ v, w \in X $
that
$ d( v, w ) = \opnorm{ v - w } $.
Then 
\begin{equation}
    \CovNum{(X,d),r}
    \leq
    \begin{cases}
      1 & \colon r\geq R \\
    \br*{ 
      \frac{ 4 R }{ r }
    }^{ \dim(V) } &\colon r<R
    \end{cases}
\end{equation}
\cfout.
\end{prop}

\begin{proof}[Proof of \cref{prop:covering}]
Throughout this proof, assume without loss of generality that $ \dim( V ) > 0 $, 
assume without loss of generality that $ r < R $, 
let $ N \in \N $ satisfy 
$ N = \dim( V ) $, 
let $ n \in \N $ satisfy 
\begin{equation}
  n = 
  \ceil*{ 
    \br*{  
      \frac{ 4 R }{ r } 
    }^N
    - 1
  }
  ,
\end{equation}
let $ \mathfrak{X} = \cu*{ v \in V \colon \opnorm{ v } \leq 1 } $,
and let $ \mathfrak{d} \colon \mathfrak{X} \times \mathfrak{X} \to [0,\infty) $
satisfy for all $ v, w \in \mathfrak{X} $ that
\begin{equation}
  \mathfrak{d}( v, w ) = \opnorm{ v - w }
\end{equation}
\cfload.
Observe that \cref{result:upper_covering_radius} proves that 
\begin{equation}
  \CovRad{ 
    ( \mathfrak{X}, \mathfrak{d} ), n
  } 
\leq
  4 \, ( n + 1 )^{ - \nicefrac{ 1 }{ N } }
\end{equation}
\cfload.
The fact that 
\begin{equation}
  n + 1
= 
  \ceil*{ 
    \br*{  
      \frac{ 4 R }{ r } 
    }^N
    - 1
  }
  + 1
\geq 
  \br*{
    \br*{  
      \frac{ 4 R }{ r } 
    }^N
    - 1
  }
  + 1
  =
  \br*{  
    \frac{ 4 R }{ r } 
  }^N
\end{equation}
therefore ensures that 
\begin{equation}
  \CovRad{ 
    ( \mathfrak{X}, \mathfrak{d} ), n
  } 
\leq
  4 \, ( n + 1 )^{ - \nicefrac{ 1 }{ N } }
\leq 
  4
  \br*{ 
    \br*{  
      \frac{ 4 R }{ r } 
    }^N
  }^{ - \nicefrac{ 1 }{ N } }
=
  4
  \br*{  
    \frac{ 4 R }{ r } 
  }^{ - 1 }
=
  \frac{ r }{ R }
  .
\end{equation}
This and \cref{result:scaling_covering} demonstrate that 
\begin{equation}
  \CovRad{ (X,d), n } 
  =
  R 
  \,
  \CovRad{ (\mathfrak{X},\mathfrak{d}), n } 
  \leq 
  R
  \br*{ 
    \frac{ r }{ R }
  }
  =
  r 
  .
\end{equation}
\cref{result:covering_number_bound2} 
hence ensures that 
\begin{equation}
  \CovNum{(X,d),r}
\leq
  n
\leq 
  \br*{ 
    \frac{ 4 R }{ r } 
  }^{ 
    N
  }
=
  \br*{ 
    \frac{ 4 R }{ r } 
  }^{ 
    \dim( V )
  }
\end{equation}
\cfload.
The proof of \cref{prop:covering} is thus complete.
\end{proof}

\cfclear
\begin{athm}{prop}{lem:covering_number_cube_infty}
Let
$ d \in \N $,
$ a \in \R $,
$ b \in ( a, \infty ) $,
$ r \in ( 0, \infty ) $
and
let
$ \delta \colon ( [ a, b ]^d ) \times ( [ a, b ]^d ) \to [ 0, \infty ) $
satisfy for all
$ x, y \in [ a, b ]^d $
that
$ \delta( x, y ) = \pnorm{\infty}{ x - y } $
\cfload.
Then
\begin{equation}
\CovNum{ ( [ a, b ]^d, \delta ), r }
\leq
\pr[\big]{
\ceil[\big]{
    \tfrac{ b - a }{2r}
}
}^d
\leq
\begin{cases}
1
& \colon
r \geq \nicefrac{ ( b - a ) }{2}
\\
\bigl(
    \tfrac{ b - a }{r}
\bigr)^d
& \colon
r < \nicefrac{ ( b - a ) }{2}
\end{cases}
\end{equation}
\cfout.
\end{athm}

\begin{aproof}
Throughout this proof,
let
$ \fN \subseteq \N $
satisfy 
\begin{equation}
\label{eq:def_fN_infty}
\fN
=
\ceil[\big]{
    \tfrac{ b - a }{2r}
}
,
\end{equation}
for every
$ N \in \N $,
$ i \in \{ 1, 2, \ldots, N \} $
let
$ g_{ N, i } \in [ a, b ] $
be given by
\begin{equation}
\label{covering_number_cube:eq1_infty}
\begin{split} 
g_{ N, i } = a + \nicefrac{ ( i - \nicefrac{1}{2} ) ( b - a ) }{N} 
\end{split}
\end{equation}
and let
$ A \subseteq [ a, b ]^d $
be given by
\begin{equation}
\label{covering_number_cube:eq2_infty}
\begin{split} 
A =
\{ g_{ \fN, 1 }, g_{ \fN, 2 }, \ldots, g_{ \fN, \fN } \}^d 
\end{split}
\end{equation}
\cfload.
Observe that it holds for all
$ N \in \N $,
$ i \in \{ 1, 2, \ldots, N \} $,
$ x \in [ a + \nicefrac{ ( i - 1 )( b - a ) }{N}, g_{ N, i } ] $
that
\begin{equation}
\label{eq:grid_estimate1_infty}
\lvert x - g_{ N, i } \rvert
=
a + \tfrac{ ( i - \nicefrac{1}{2} ) ( b - a ) }{N}
-
x
\leq
a + \tfrac{ ( i - \nicefrac{1}{2} ) ( b - a ) }{N}
-
\bigl( a + \tfrac{ ( i - 1 )( b - a ) }{N} \bigr)
=
\tfrac{ b - a }{2N}
.
\end{equation}
In addition,
note that it holds for all
$ N \in \N $,
$ i \in \{ 1, 2, \ldots, N \} $,
$ x \in [ g_{ N, i }, a + \nicefrac{ i ( b - a ) }{N} ] $
that%
\begin{equation}
\label{eq:grid_estimate2_infty}
\lvert x - g_{ N, i } \rvert
=
x
-
\bigl( a + \tfrac{ ( i - \nicefrac{1}{2} ) ( b - a ) }{N} \bigr)
\leq
a + \tfrac{ i ( b - a ) }{N}
-
\bigl( a + \tfrac{ ( i - \nicefrac{1}{2} ) ( b - a ) }{N} \bigr)
=
\tfrac{ b - a }{2N}
.
\end{equation}
Combining this with~\cref{eq:grid_estimate1_infty}
implies for all
$ N \in \N $,
$ i \in \{ 1, 2, \ldots, N \} $,
$ x \in [ a + \nicefrac{ ( i - 1 )( b - a ) }{N}, \allowbreak a + \nicefrac{ i ( b - a ) }{N} ] $
that
$ \lvert x - g_{ N, i } \rvert
\leq
\nicefrac{ ( b - a ) }{(2N)} $.
This proves that
for every
$ N \in \N $,
$ x \in [ a, b ] $
there exists
$ y \in \{ g_{ N, 1 }, g_{ N, 2 }, \ldots, g_{ N, N } \} $
such that
\begin{equation}
\label{eq:grid_estimate_final_infty}
\lvert x - y \rvert
\leq
\tfrac{ b - a }{2N}
.
\end{equation}
This
shows that
for every
$ x = ( x_1, x_2, \ldots, x_d )
\in [ a, b ]^d $
there exists
$ y = ( y_1, y_2, \ldots, y_d )
\in A $
such that
\begin{equation}
\label{eq:covering_L_infty}
\delta( x, y )
=
\infnorm{ x - y }
=
\max_{ i \in \{ 1, 2, \ldots, d \} }
    \lvert x_i - y_i \rvert
\leq
\tfrac{ b - a }{2\fN}
\leq
\tfrac{ ( b - a ) 2r }{ 2 ( b - a ) }
=
r
.
\end{equation}
Combining
  this
with
\cref{def:covering_number:eq1},
\cref{covering_number_cube:eq2_infty},
\cref{eq:def_fN_infty},
and
the fact that
$ \forall \, x \in [ 0, \infty ) \colon
\lceil x \rceil
\leq
\ind{ ( 0, r ] }( r x )
+
2 x \ind{ ( r, \infty ) }( r x ) $
demonstrates that
\begin{equation}
\CovNum{ ( [ a, b ]^d, \delta ), r }
\leq
\lvert A \rvert = ( \fN )^d
=
\bigl(
\bigl\lceil
    \tfrac{ b - a }{2r}
\bigr\rceil
\bigr)^d
\leq
\ind{ ( 0, r ] }\bigl( \tfrac{ b - a }{2} \bigr)
+
\bigl(
\tfrac{ b - a }{r}
\bigr)^d
\ind{ ( r, \infty ) }\bigl( \tfrac{ b - a }{2} \bigr)
\end{equation}
\cfload.
\end{aproof}

\section{Empirical risk minimization}
\label{sect:emp_risk_min}

\subsection{Concentration inequalities for random fields}

\cfclear
\begin{lemma}
\label{lem:separable}
Let $ (E, d) $ be a separable
metric space and let
$ F \subseteq E $ be a set.
Then
\begin{equation}
  ( F, d|_{ F \times F } ) 
\end{equation}
is a separable metric space.
\end{lemma}

\begin{proof}[Proof of \cref{lem:separable}]
Throughout this proof, assume without loss of generality that
$ F \neq \emptyset $,
let $ e = ( e_n )_{ n \in \N } \colon \N\to E $
be a sequence of elements in $ E $
such that
$
  \{ e_n \in E \colon n \in \N \}
$
is dense in $ E $,
and let
$ f = ( f_n )_{ n \in \N } \colon \N\to F $
be a sequence of elements in $ F $
such that for all $ n \in \N $
it holds that
\begin{equation}
  d( f_n, e_n )
\leq 
\begin{cases}
  0
&
  \colon
  e_n \in F
\\
  \bigl[\inf\nolimits_{x\in F} d(x,e_n)\bigr]
  +
  \frac{ 1 }{ 2^n }
&
  \colon
  e_n \notin F.
\end{cases}
\end{equation}
Observe that for all
$ v \in F \backslash \{ e_m \in E \colon m \in \N \} $,
$ n \in \N $
it holds that
\begin{equation}
\label{eq:dist_in_e_2}
\begin{split}
	\inf_{m\in\N} d(v,f_m)
& \leq
  \inf_{m\in\N\cap[n,\infty)} d(v,f_m)
\\ & \leq
  \inf_{ m\in\N\cap[n,\infty) }
  \br*{
    d( v, e_m )
    +
    d( e_m , f_m )
  }
\\ & \leq
  \inf_{ m\in\N\cap[n,\infty) }
  \br*{
    d( v, e_m )
    +
    \bigl[\inf\nolimits_{x\in F} d(x,e_m)\bigr]
    +
    \frac{ 1 }{ 2^m }
  }
\\ & \leq
  \inf_{ m\in\N\cap[n,\infty) }
  \br*{
    2 \,
    d( v, e_m )
    +
    \frac{ 1 }{ 2^m }
  }
\\ & \leq
  2\br*{
    \inf_{ m\in\N\cap[n,\infty) }
    d( v, e_m )
  }
    +
  \frac{ 1 }{ 2^n }
= 
  \frac{ 1 }{ 2^n }
  .
\end{split}
\end{equation}
Combining this with the fact that 
for all $v\in F\cap\{e_m\in E\colon m\in\N\}$
it holds that $\inf_{m\in\N} d(v,f_m)=0$
ensures that 
the set 
$ \{ f_n \in F \colon n \in \N \} $
is dense in $ F $.
The proof of \cref{lem:separable}
is thus complete.
\end{proof}

\cfclear
\begingroup
\newcommand{\vX}{E}
\newcommand{\vY}{\mathbf E}
\newcommand{\vmcF}{\mc F}
\begin{athm}{lemma}{lem:measurability_sup}
  Let $(\vX,\mathscr E)$ be a topological space,
  assume $\vX\neq\emptyset$,
  let $\vY\subseteq \vX$ be an at most countable set,
  assume that $\vY$ is dense in $\vX$,
  let $(\Omega,\vmcF)$ be a measurable space,
  for every 
  $x\in \vX$
  let $f_x\colon\Omega\to\R$ be $\vmcF$/$\mc B(\R)$-measurable,
  assume 
    for all 
      $\omega\in\Omega$ 
    that 
      $\vX\ni x\mapsto f_x(\omega)\in\R$ is continuous,
  and let $F\colon\Omega\to\R\cup\{\infty\}$ satisfy
    for all
      $\omega\in\Omega$
    that
    \begin{equation}
      F(\omega)=\sup_{x\in \vX} f_x(\omega)
      .
    \end{equation}
  Then
  \begin{enumerate}[label=(\roman{*})]
    \item \label{it:meas1}
      it holds for all 
        $\omega\in\Omega$ 
      that 
        $F(\omega)=\sup_{x\in \vY} f_x(\omega)$ 
      and
    \item \label{it:meas2}
      it holds that 
        $F$ is $\vmcF$/$\mc B(\R\cup\{\infty\})$-measurable.
  \end{enumerate}
\end{athm}
\begin{aproof}
  \Nobs that
    the assumption that $\vY$ is dense in $\vX$ 
  \proves that for all
    $g\in C(\vX,\R)$
  it holds that
  \begin{equation}
    \sup_{x\in \vX} g(x)
    =
    \sup_{x\in \vY} g(x)
    .
  \end{equation}
    This
    and the assumption that
      for all
        $\omega\in\Omega$
      it holds that
        $\vX\ni x\mapsto f_x(\omega)\in\R$ is continuous
  \prove that for all
    $\omega\in\Omega$
  it holds that
  \begin{equation}
    \label{eq:meas.1}
    F(\omega)
    =
    \sup_{x\in \vX}f_x(\omega)
    =
    \sup_{x\in \vY}f_x(\omega)
    .
  \end{equation}
  This \proves[ep] \cref{it:meas1}.
  \Moreover
    \cref{it:meas1}
    and the assumption that for all
      $x\in \vX$
    it holds that
      $f_x\colon\Omega\to\R$ is $\vmcF$/$\mc B(\R)$-measurable
  \prove[ep] \cref{it:meas2}.
\end{aproof}
\endgroup

\cfclear
\begingroup
\newcommand{\ZZ}[1]{Z_{#1}}
\newcommand{\vX}{E}
\newcommand{\vd}{\delta}
\newcommand{\vmcF}{\mc F}
\begin{lemma}
  \label{lem:cov0}
  Let $(\vX,\vd)$ be a separable metric space,
  let 
    $\eps,L\in\R$,
    $N\in\N$,
    $z_1,z_2,\dots,\allowbreak z_N\in \vX$
    satisfy
    $\vX\subseteq\bigcup_{i=1}^N\{x\in \vX\colon 2L\vd(x,z_i)\leq \eps\}$,
  let $(\Omega,\vmcF,\P)$ be a probability space,
  and let $\ZZ x\colon\Omega\to\R$, $x\in \vX$, be random variables 
    which satisfy for all
      $x,y\in \vX$
    that
      $\abs{\ZZ x-\ZZ y}\leq L\vd(x,y)$.
  Then
  \begin{equation}
    \P\pr*{\textstyle \sup_{x\in \vX}\abs{\ZZ x}\geq\eps}
    \leq
    \sum_{i=1}^N\P\pr*{\abs{\ZZ{z_i}}\geq\tfrac\eps2}
  \end{equation}
  (cf.\ \cref{lem:measurability_sup}).
\end{lemma}
\begin{proof}[Proof of \cref{lem:cov0}]
  Throughout this proof,
  let $B_1,B_2,\dots,B_N\subseteq \vX$ satisfy
    for all $i\in\{1,2,\dots,N\}$ that
      $B_i=\{x\in \vX\colon 2L\vd(x,z_i)\leq \eps\}$.
  Observe that
    the triangle inequality
    and the assumption that for all
      $x,y\in \vX$
    it holds that
      $\abs{\ZZ x-\ZZ y}\leq L\vd(x,y)$
  show
  that for all
    $i\in\{1,2,\dots,N\}$,
    $x\in B_i$
  it holds that
  \begin{equation}
  \begin{split}
    \abs{\ZZ x}
    =
    \abs{\ZZ x-\ZZ{z_i}+\ZZ{z_i}}
    \leq
    \abs{\ZZ x-\ZZ{z_i}} + \abs{\ZZ{z_i}}
    \leq
    L\vd(x,z_i) + \abs{\ZZ{z_i}}
    \leq \tfrac\eps2+\abs{\ZZ{z_i}}
    .
  \end{split}
  \end{equation}
  Combining
    this
  with 
    \cref{lem:measurability_sup}
	and \cref{lem:separable}
  proves that for all
    $i\in\{1,2,\dots,N\}$
  it holds that
  \begin{equation}
  \label{eq:cov0.1}
    \P\pr*{\sup\nolimits_{x\in B_i}\abs{\ZZ x}\geq\eps}
    \leq
    \P\pr*{\tfrac\eps2+\abs{\ZZ{z_i}}\geq\eps}
    =
    \P\bigl(\abs{\ZZ{z_i}}\geq\tfrac\eps2\bigr)
    .
  \end{equation}
    This,
	  \cref{lem:measurability_sup},
	  and \cref{lem:separable}
  establish that
  \begin{equation}
  \begin{split}
    \textstyle \P\pr*{\sup_{x\in \vX}\abs{\ZZ x}\geq\eps}
    &=
    \textstyle \P\pr*{\sup_{x\in\pr*{\bigcup_{i=1}^N B_i}}\abs{\ZZ x}\geq\eps}
    =
    \P\pr*{\textstyle\bigcup_{i=1}^N\cu*{\sup_{x\in B_i}\abs{\ZZ x}\geq\eps}}
    \\&\leq
    \sum_{i=1}^N\P\pr*{\sup\nolimits_{x\in B_i}\abs{\ZZ x}\geq\eps}
    \leq
    \sum_{i=1}^N\P\bigl(\abs{\ZZ{z_i}}\geq\tfrac\eps2\bigr)
    .
  \end{split}  
  \end{equation}
  This completes the proof of \cref{lem:cov0}.
\end{proof}
\endgroup

\cfclear
\begingroup
\newcommand{\ZZ}[1]{Z_{#1}}
\newcommand{\vX}{E}
\newcommand{\vd}{\delta}
\newcommand{\vmcF}{\mc F}
\begin{lemma}
  \label{lem:cov02}
  Let $(\vX,\vd)$ be a separable metric space,
  assume $\vX\neq\emptyset$,
  let 
    $\eps,L\in(0,\infty)$,
  let $(\Omega,\vmcF,\P)$ be a probability space,
  and let $\ZZ x\colon\Omega\to\R$, $x\in \vX$, be random variables 
    which satisfy for all
      $x,y\in \vX$
    that
      $\abs{\ZZ x-\ZZ y}\leq L\vd(x,y)$.
  Then
  \begin{equation}
  \label{eq:cov02con}
    \bigl[\CovNum{(\vX,\vd),\frac\eps{2L}}\bigr]^{-1}\textstyle \P\pr*{\sup_{x\in \vX}\abs{\ZZ x}\geq\eps}
    \leq
    \sup_{x\in \vX}\P\bigl(\abs{\ZZ x}\geq\tfrac\eps2\bigr)
    .
    \cfadd{lem:measurability_sup}
  \end{equation}
  \cfout.
\end{lemma}
\begin{proof}[Proof of \cref{lem:cov02}]
  Throughout this proof,
    let $N\in\N\cup\{\infty\}$ satisfy $N=\CovNum{(\vX,\vd),\frac\eps{2L}}$,
    assume without loss of generality that $N<\infty$,
    and let $z_1,z_2,\dots,z_N\in \vX$ satisfy
      $\vX\subseteq \bigcup_{i=1}^N\{x\in \vX\colon \vd(x,z_i)\leq \frac\eps{2L}\}$
  \cfload.
  Observe that
    \cref{lem:measurability_sup}
    and \cref{lem:cov0}
  establish that
  \begin{equation}
  \begin{split}
    \P\pr*{\textstyle \sup_{x\in \vX}\abs{\ZZ x}\geq\eps}
    \leq
    \sum_{i=1}^N\P\pr*{\abs{\ZZ{z_i}}\geq\tfrac\eps2}
    \leq
    N\bigl[\sup\nolimits_{x\in \vX}\P\bpr{\abs{\ZZ x}\geq\tfrac\eps2}\bigr]
    .
  \end{split}
  \end{equation}
  This completes the proof of \cref{lem:cov02}.
\end{proof}
\endgroup

\cfclear

\begingroup
\newcommand{\ZZ}[1]{Z_{#1}}
\newcommand{\vX}{E}
\newcommand{\vd}{\delta}
\newcommand{\vmcF}{\mc F}
\begin{athm}{lemma}{lem:measurability_sup_centred}
  Let $(\vX,\vd)$ be a separable metric space,
  assume $\vX\neq\emptyset$,
  let $(\Omega,\vmcF,\P)$ be a probability space,
  let $L\in\R$,
  for every 
    $x\in \vX$
  let $\ZZ x\colon\Omega\to\R$ be a random variable with
    $\E[\abs{\ZZ x}]<\infty$,
  and assume
    for all $x,y\in \vX$ that
      $\abs{\ZZ x-\ZZ y}\leq L\vd(x,y)$.
  Then
  \begin{enumerate}[label=(\roman{*})]
    \item \label{it:meas2.1}
      it holds for all 
        $x,y\in \vX$,
        $\eta\in\Omega$
      that 
      \begin{equation}
        \abs{(\ZZ x(\eta)-\E[\ZZ x])-(\ZZ y(\eta)-\E[\ZZ y])}\leq 2L\vd(x,y)
      \end{equation}
      and
    \item \label{it:meas2.2}
      it holds that 
        $\Omega\ni\eta\mapsto\sup_{x\in \vX}\abs{\ZZ x(\eta)-\E[\ZZ x]}\in[0,\infty]$
      is $\vmcF$/$\Borel([0,\infty])$-measurable.
  \end{enumerate}
\end{athm}
\begin{aproof}
  \Nobs that
    the assumption that for all
      $x,y\in \vX$
    it holds that
      $\abs{\ZZ x -\ZZ y}\leq L\vd(x,y)$
  \proves that for all 
    $x,y\in \vX$,
    $\eta\in\Omega$
  it holds that
  \begin{equation}
  \begin{split}
      \abs{(\ZZ x(\eta)-\E[\ZZ x])-(\ZZ y(\eta)-\E[\ZZ y])}
      &=
      \abs{(\ZZ x(\eta)-\ZZ y(\eta))+(\E[\ZZ y]-\E[\ZZ x])}
      \\&\leq
      \abs{\ZZ x(\eta)-\ZZ y(\eta)}+\abs{\E[\ZZ x]-\E[\ZZ y]}
      \\&\leq
      L\vd(x,y)+\abs{\E[\ZZ x]-\E[\ZZ y]}
      \\&=
      L\vd(x,y)+\abs{\E[\ZZ x-\ZZ y]}
      \\&\leq
      L\vd(x,y)+\E[\abs{\ZZ x-\ZZ y}]
      \\&\leq
      L\vd(x,y)+L\vd(x,y)
      =
      2L\vd(x,y)
      .
  \end{split}  
  \end{equation}
  This \proves \cref{it:meas2.1}.
  \Nobs that
    \cref{it:meas2.1} 
  \proves that
    for all $\eta\in\Omega$ it holds that
      $\vX\ni x\mapsto \abs{\ZZ x(\eta)-\E[\ZZ x]}\in\R$ is continuous.
  Combining 
    this
    and the assumption that $\vX$ is separable
  with
    \cref{lem:measurability_sup}
  \proves[ep]
    \cref{it:meas2.2}.
\end{aproof}
\endgroup

\cfclear
\begingroup
\newcommand{\ZZ}[1]{Z_{#1}}
\newcommand{\vX}{E}
\newcommand{\vd}{\delta}
\newcommand{\vmcF}{\mc F}
\begin{lemma}
  \label{lem:cov2}
  Let $(\vX,\vd)$ be a separable metric space,
  assume $\vX\neq\emptyset$,
  let 
    $\eps,L\in(0,\infty)$,
  let $(\Omega,\vmcF,\P)$ be a probability space,
  and let $\ZZ x\colon\Omega\to\R$, $x\in \vX$, be random variables 
    which satisfy for all
      $x,y\in \vX$
    that
      $\E[\abs{\ZZ x}]<\infty$
      and $\abs{\ZZ x-\ZZ y}\leq L\vd(x,y)$.
  Then
  \begin{equation}
  \label{eq:cov2con}
    \bigl[\CovNum{(\vX,\vd),\frac\eps{4L}}\bigr]^{-1}\textstyle \P\pr*{\sup_{x\in \vX}\abs{\ZZ x-\E[\ZZ x]}\geq\eps}
    \leq
    \sup_{x\in \vX}\P\bigl(\abs{\ZZ x-\E[\ZZ x]}\geq\tfrac\eps2\bigr)
    .
    \cfadd{lem:measurability_sup_centred}
  \end{equation}
  \cfout.
\end{lemma}
\begin{proof}[Proof of \cref{lem:cov2}]
  \newcommand{\YY}[1]{Y_{#1}}
  Throughout this proof,
    let $\YY x\colon\Omega\to\R$, $x\in \vX$, satisfy for all
      $x\in \vX$,
      $\eta\in\Omega$
    that
      $\YY x(\eta)=\ZZ x(\eta)-\E[\ZZ x]$.
  Observe that
    \cref{lem:measurability_sup_centred}
  ensures that for all
    $x,y\in \vX$
  it holds that
  \begin{equation}
    \abs{\YY x-\YY y}\leq 2L\vd(x,y).
  \end{equation}
    This
    and \cref{lem:cov02} (applied with
      $(\vX,\vd)\is(\vX,\vd)$,
      $\eps\is\eps$,
      $L\is 2L$,
      $(\Omega,\vmcF,\P)\is(\Omega,\vmcF,\P)$,
      $(\ZZ x)_{x\in \vX}\is(\YY x)_{x\in \vX}$
    in the notation of \cref{lem:cov02})
  establish \cref{eq:cov2con}.
  The proof of \cref{lem:cov2} is thus complete.
\end{proof}
\endgroup

\newcommand{\lemIcovthreeIvE}[2]{Y_{#1,#2}}
\cfclear
\begingroup
\newcommand{\vE}[2]{\lemIcovthreeIvE{#1}{#2}}
\newcommand{\ZZ}[1]{Z_{#1}}
\newcommand{\vX}{E}
\newcommand{\vd}{\delta}
\newcommand{\vmcF}{\mc F}
\begin{lemma}
  \label{lem:cov3}
  Let $(\vX,\vd)$ be a separable metric space,
  assume $\vX\neq\emptyset$,
  let 
    $M\in\N$,
    $\eps,L,D\in(0,\infty)$,
  let $(\Omega,\vmcF,\P)$ be a probability space,
  for every 
    $x\in \vX$ 
  let 
    $\vE{x}{1},\vE{x}{2},\dots,\allowbreak \vE{x}{M}\colon\Omega\to[0,D]$
  be independent random variables,
  assume
    for all
      $x,y\in \vX$,
      $m\in\{1,2,\dots,\allowbreak M\}$
    that
      $\abs{\vE{x}{m}-\vE{y}{m}}\leq L\vd(x,y)$,
  and let $\ZZ{x}\colon \Omega\to[0,\infty)$, $x\in \vX$, satisfy for all 
    $x\in\vX$ 
  that
  \begin{equation}
    \label{eq:cov3.defZZ}
    \ZZ{x}=\frac1M\Biggl[\sum_{m=1}^M \vE{x}{m}\Biggr].
  \end{equation}
  Then 
  \begin{enumerate}[label=(\roman *)]
    \item \label{it:cov3.1}
      it holds for all 
        $x\in \vX$
      that 
        $\E[\abs{\ZZ{x}}]\leq D<\infty$,
    \item \label{it:cov3.2}
      it holds that 
        $\Omega\ni\eta\mapsto\sup_{x\in \vX} \abs{\ZZ{x}(\eta)-\E[\ZZ{x}]}\in[0,\infty]$
        is $\vmcF$/$\Borel([0,\infty])$-measurable, and
    \item \label{it:cov3.3}
      it holds that
      \begin{equation}
        \label{eq:cov3con}
        \P\pr*{\sup\nolimits_{x\in \vX}\abs{\ZZ{x}-\E[\ZZ{x}]}\geq\eps}
        \leq
        2\CovNum{(\vX,\vd),\frac\eps{4L}}\exp\pr*{\frac{-\eps^2M}{2D^2}}
      \end{equation}
  \end{enumerate}
  \cfout.
\end{lemma}
\begin{proof}[Proof of \cref{lem:cov3}]
  First, observe that
    the triangle inequality
    and the assumption that for all
      $x,y\in \vX$,
      $m\in\{1,2,\dots,M\}$
    it holds that 
      $\abs{\vE{x}{m}-\vE{y}{m}}\leq L\vd(x,y)$
  imply that for all
    $x,y\in \vX$
  it holds that
  \begin{equation}
    \label{eq:cov3lipsch}
  \begin{split}
    \abs{\ZZ{x}-\ZZ{y}}
    &=
    \abs*{ \frac1M\br*{\sum_{m=1}^M \vE{x}{m}} - \frac1M\br*{\sum_{m=1}^M \vE{y}{m}} }
    =
    \frac1M\abs*{\sum_{m=1}^M \bigl(\vE{x}{m} - \vE{y}{m}\bigr)}
    \\&\leq
    \frac1M\br*{\sum_{m=1}^M \babs{\vE{x}{m} - \vE{y}{m}}}
    \leq
    L\vd(x,y)
    .
  \end{split}
  \end{equation}
  Next note that
    the assumption that 
      for all
        $x\in \vX$,
        $m\in\{1,2,\dots,M\}$,
        $\omega\in\Omega$
      it holds that
        $\abs{\vE{x}{m}(\omega)}\in [0,D]$
  ensures that for all
    $x\in \vX$
  it holds that
  \begin{equation}
    \label{eq:Zxint}
    \E\bigl[\abs{\ZZ{x}}\bigr]
    =
    \Exp{\frac1M\br*{\sum_{m=1}^M \vE{x}{m}}}
    =
    \frac1M\br*{\sum_{m=1}^M \E\bigl[\vE{x}{m}\bigr]}
    \leq
    D
    <
    \infty
    .
  \end{equation}
  This proves \cref{it:cov3.1}.
  Furthermore, note that 
    \cref{it:cov3.1}, 
    \cref{eq:cov3lipsch},
    and \cref{lem:measurability_sup_centred}
  establish
  \cref{it:cov3.2}.
  Next observe that
    \cref{eq:cov3.defZZ}
  shows that for all
    $x\in \vX$
  it holds that
  \begin{equation}
    \abs{\ZZ{x}-\E[\ZZ{x}]}
    =
    \abs*{ \frac1M\br*{\sum_{m=1}^M \vE{x}{m}} - \Exp{\frac1M\br*{\sum_{m=1}^M \vE{x}{m}}} }
    =
    \frac1M\abs*{ \sum_{m=1}^M \bigl(\vE{x}{m} - \E\bigl[\vE{x}{m}\bigr]\bigr) }
    .
  \end{equation}
  Combining
    this
  with
    \cref{cor:Hoeffding3}
      (applied with 
        $(\Omega,\vmcF,\P)\is (\Omega,\vmcF,\P)$,
        $N\is M$,
        $\eps\is\frac\eps2$,
        $(a_1,a_2,\dots,a_N)\is (0,0,\dots,0)$,
        $(b_1,b_2,\dots,b_N)\is (D,D,\dots,D)$,
        $(X_n)_{n\in\{1,2,\dots,N\}}\is (\vE{x}{m})_{m\in\{1,2,\dots,M\}}$
        for $x\in \vX$
        in the notation of \cref{cor:Hoeffding3})
  ensures that for all 
    $x\in \vX$
  it holds that
  \begin{equation}
    \P\bigl(\abs{\ZZ{x}-\E[\ZZ{x}]}\geq\tfrac\eps2\bigr)
    \leq
    2\exp\pr*{\frac{-2\bigl[\frac{\eps}2\bigr]^2M^2}{MD^2}}
    =
    2\exp\pr*{\frac{-\eps^2M}{2D^2}}
    .
  \end{equation}
  Combining
    this,
    \cref{eq:cov3lipsch},
    and~\cref{eq:Zxint}
  with
    \cref{lem:cov2}
  establishes~\cref{it:cov3.3}.
  The proof of \cref{lem:cov3} is thus complete.
\end{proof}
\endgroup

\subsection{Uniform estimates for the statistical learning error}

\cfclear
\begingroup
\newcommand{\vA}{E}
\newcommand{\vd}{\delta}
\newcommand{\vX}[2]{X_{#1,#2}}
\newcommand{\vmfE}[1]{\mf E_{#1}}
\newcommand{\vmcE}[1]{\mc E_{#1}}
\newcommand{\vmcF}{\mc F}
\begin{lemma}
  \label{lem:cov4}
  Let $(\vA,\vd)$ be a separable metric space,
  assume $\vA\neq\emptyset$,
  let 
    $M\in\N$,
    $\eps,L,D\in(0,\infty)$,
  let $(\Omega,\vmcF,\P)$ be a probability space,
  let $\vX xm\colon \Omega\to\R$, $x\in \vA$, $m\in\{1,2,\dots,M\}$,
    and $Y_m\colon \Omega\to\R$, $m\in\{1,2,\dots,M\}$,
    be functions,
  assume for all $x\in \vA$ that
    $(\vX{x}{m},Y_m)$, $m\in\{1,2,\dots,M\}$, are i.i.d.\ random variables,
  assume for all
    $x,y\in \vA$,
    $m\in\{1,2,\dots,M\}$
  that
    $\abs{\vX{x}{m}-\vX{y}{m}}\leq L\vd(x,y)$
    and $\lvert \vX{x}{m}-Y_m\rvert\leq D$,
  let $\vmfE{x}\colon \Omega\to[0,\infty)$, $x\in \vA$,
    satisfy for all
      $x\in \vA$
    that
    \begin{equation}
      \label{eq:cov4.defmfE}
      \vmfE{x}=\frac1M\br*{\sum_{m=1}^M \abs{\vX{x}{m}-Y_m}^2},
    \end{equation}
  and let $\vmcE{x}\in[0,\infty)$, $x\in \vA$, 
    satisfy for all
      $x\in \vA$
    that
      $\vmcE{x}=\E[\abs{\vX{x}{1}-Y_1}^2]$.
  Then
    $\Omega\ni\omega\mapsto \sup_{x\in \vA}\abs{\vmfE{x}(\omega)-\vmcE{x}}\in [0,\infty]$
    is $\vmcF$/$\Borel([0,\infty])$-measurable
  and
  \begin{equation}
    \label{eq:cov4con}
    \P\pr*{\sup\nolimits_{x\in \vA}\abs{\vmfE{x}-\vmcE{x}}\geq\eps}
    \leq
    2\CovNum{(\vA,\vd),\frac\eps{8LD}}\exp\pr*{\frac{-\eps^2M}{2D^4}}
  \end{equation}
  \cfout.
\end{lemma}
\begin{proof}[Proof of \cref{lem:cov4}]
\begingroup
\newcommand{\vE}[2]{\mathscr{E}_{#1,#2}}
  Throughout this proof, let 
    $\vE{x}{m}\colon\Omega\to[0,D^2]$, $x\in \vA$, $m\in\{1,2,\dots,M\}$,
  satisfy for all
    $x\in \vA$,
    $m\in\{1,2,\dots,M\}$
  that
  \begin{equation}
    \vE{x}{m}
    =
    \lvert \vX{x}{m}-Y_m\rvert^2
    .
  \end{equation}
  Observe that
    the fact that for all 
      $x_1,x_2,y\in\R$ 
    it holds that 
      $(x_1-y)^2-(x_2-y)^2=(x_1-x_2)((x_1-y)+(x_2-y))$,
    the assumption that
      for all
        $x\in \vA$,
        $m\in\{1,2,\dots,M\}$
      it holds that
        $\abs{\vX{x}{m}-Y_m}\leq D$,
    and the assumption that
      for all
        $x,y\in \vA$,
        $m\in\{1,2,\dots,M\}$
      it holds that
        $\abs{\vX{x}{m}-\vX{y}{m}}\leq L\vd(x,y)$
  imply that for all
    $x,y\in \vA$,
    $m\in\{1,2,\dots,M\}$
  it holds that
  \begin{equation}
    \label{eq:cov4lipsch}
  \begin{split}
    \abs{\vE{x}{m}-\vE ym}
    &=
    \babs{(\vX{x}{m}-Y_m)^2-(\vX{y}{m}-Y_m)^2}
    \\&=
    \abs{\vX{x}{m}-\vX{y}{m}}\babs{(\vX{x}{m}-Y_m)+(\vX{y}{m}-Y_m)}
    \\&\leq
    \abs{\vX{x}{m}-\vX{y}{m}}\bigl(\abs{\vX{x}{m}-Y_m}+\abs{\vX{y}{m}-Y_m}\bigr)
    \\&\leq
    2D\abs{\vX{x}{m}-\vX{y}{m}}
    \leq
    2LD\vd(x,y)
    .
  \end{split}
  \end{equation}
  In addition, note that 
    \cref{eq:cov4.defmfE}
    and the assumption that
      for all 
        $x\in\vA$ 
      it holds that
        $(X_{x,m},Y_m)$, $m\in\{1,2,\dots,M\}$, are i.i.d.\ random variables
  show that for all
    $x\in \vA$
  it holds that
  \begin{equation}
  \label{eq:cov4.1}
    \E\bigl[\vmfE{x}\bigr]
    =
    \frac1M\br*{\sum_{m=1}^M\E\bigl[\abs{\vX{x}{m}-Y_m}^2\bigr]}
    =
    \frac1M\br*{\sum_{m=1}^M\E\bigl[\abs{\vX{x}{1}-Y_1}^2\bigr]}
    =
    \frac1M\br*{\sum_{m=1}^M\vmcE{x}}
    =
    \vmcE{x}
    .
  \end{equation}
  Furthermore, observe that
    the assumption that for all
      $x\in \vA$
    it holds that
      $(\vX{x}{m},Y_m)$, $m\in\{1,2,\dots,M\}$,
    are i.i.d.\ random variables
  ensures that for all
    $x\in \vA$
  it holds that
    $\vE xm$, $m\in\{1,2,\dots,M\}$,
  are i.i.d.\ random variables.
  Combining
    this, 
    \cref{eq:cov4lipsch}, 
    and \cref{eq:cov4.1}
  with
    \cref{lem:cov3}
    (applied with
      $(E,\delta)\is(\vA,\vd)$,
      $M\is M$,
      $\eps\is\eps$,
      $L\is 2LD$,
      $D\is D^2$,
      $(\Omega,\vmcF,\P)\is(\Omega,\vmcF,\P)$,
      $(\lemIcovthreeIvE xm)_{x\in E,\,m\in\{1,2,\dots,M\}}\is (\vE xm)_{x\in \vA,\,m\in\{1,2,\dots,M\}}$,
      $(Z_x)_{x\in E}=(\vmfE{x})_{x\in \vA}$
    in the notation of \cref{lem:cov3})
  establishes~\cref{eq:cov4con}.
  The proof of \cref{lem:cov4} is thus complete.
\endgroup
\end{proof}
\endgroup

\cfclear
\begingroup
\newcommand{\vmcF}{\mc F}
\begin{prop}
  \label{lem:cov5}
  Let 
    $d,\mf d,M\in\N$,
    $R,L,\mc R,\eps\in(0,\infty)$,
  let $D\subseteq\R^d$ be a compact set,
  let $(\Omega,\vmcF,\P)$ be a probability space,
  let $X_m\colon\Omega\to D$, $m\in\{1,2,\dots,M\}$,
    and $Y_m\colon \Omega\to\R$, $m\in\{1,2,\dots,M\}$,
    be functions,
  assume that $(X_m,Y_m)%
      $, $m\in\{1,2,\dots,M\}$,
    are i.i.d.\ random variables,
  let $H=(H_\theta)_{\theta\in [-R,R]^{\mf d}}\colon [-R,R]^{\mf d}\to C(D,\R)$ satisfy
    for all
      $\theta,\vartheta\in [-R,R]^{\mf d}$,
      $x\in D$
    that
      $\abs{H_\theta(x)-H_\vartheta(x)}\leq L\infnorm{\theta-\vartheta}$,
  assume for all 
      $\theta\in [-R,R]^{\mf d}$,
      $m\in\{1,2,\dots,M\}$
    that
      $\abs{H_\theta(X_m)-Y_m}\leq\mc R$
      and $\E[\abs{Y_1}^2]<\infty$,
  let $\mc E\colon C(D,\R)\to[0,\infty)$ satisfy 
    for all
      $f\in C(D,\R)$
    that
      $\mc E(f)=\E[\abs{f(X_1)-Y_1}^2]$,
  and let $\mf E\colon [-R,R]^{\mf d}\times\Omega\to[0,\infty)$ satisfy
    for all
      $\theta\in [-R,R]^{\mf d}$,
      $\omega\in\Omega$
    that
    \begin{equation}
      \mf E(\theta,\omega)
      =
      \frac1M\br*{\sum_{m=1}^M\abs{H_\theta(X_m(\omega))-Y_m(\omega)}^2}
    \end{equation}
  \cfload.
  Then
    $\Omega\ni\omega\mapsto \sup_{\theta\in[-R,R]^{\mf d}}\abs{\mf E(\theta,\omega)-\mc E(H_\theta)}\in [0,\infty]$ is $\vmcF$/$\Borel([0,\infty])$-measurable and
  \begin{equation}
    \label{eq:cov5con}
    \begin{split}
    \P\bigl(\sup\nolimits_{\theta\in [-R,R]^{\mf d}}\abs{\mf E(\theta)-\mc E(H_\theta)}\geq\eps\bigr)
    &\leq
    2\max\biggl\{1,\biggl[
    \frac{16LR\mc R}{\eps}
    \biggr]^{\mf d}\biggr\}
    \exp\pr*{\frac{-\eps^2M}{2\mc R^4}}.
    \end{split}
  \end{equation}
\end{prop}
\begin{proof}[Proof of \cref{lem:cov5}]
  Throughout this proof, 
  let $B\subseteq\R^{\mf d}$ satisfy $B=[-R,R]^{\mf d}=\{\theta\in\R^{\mf d}\colon \infnorm\theta\leq R\}$
  and let
    $\delta\colon B\times B\to[0,\infty)$
      satisfy for all
        $\theta,\vartheta\in B$
      that
      \begin{equation}
        \delta(\theta,\vartheta)
        =
        \infnorm{\theta-\vartheta}
        .
      \end{equation}
  Observe that
    the assumption that $(X_m,Y_m)$, $m\in\{1,2,\dots,M\}$, are i.i.d.\ random variables
    and
    the assumption that for all $\theta\in [-R,R]^{\mf d}$ it holds that $H_\theta$ is continuous
  imply that for all 
    $\theta\in B$
  it holds that
    $(H_\theta(X_m),Y_m)$, $m\in\{1,2,\dots,M\}$, are i.i.d.\ random variables.
  Combining
    this,
    the assumption that
    for all
      $\theta,\vartheta\in B$,
      $x\in D$
    it holds that
      $\abs{H_\theta(x)-H_\vartheta(x)}\leq L\infnorm{\theta-\vartheta}$,
    and the assumption that for all 
      $\theta\in B$,
      $m\in\{1,2,\dots,M\}$
    it holds that
      $\abs{H_\theta(X_m)-Y_m}\leq\mc R$
  with \cref{lem:cov4} 
    (applied with
      $(E,\delta)\is (B,\delta)$,
      $M\is M$,
      $\eps\is\eps$,
      $L\is L$,
      $D\is \mc R$,
      $(\Omega,\vmcF,\P)\is(\Omega,\vmcF,\P)$,
      $(X_{x,m})_{x\in E,\,m\in\{1,2,\dots,M\}}\is (H_\theta(X_m))_{\theta\in B,\,m\in\{1,2,\dots,M\}}$,
      $(Y_m)_{m\in\{1,2,\dots,M\}}\is(Y_m)_{m\in\{1,2,\dots,M\}}$,
      $(\mf E_x)_{x\in E}\is\bigl((\Omega\ni\omega\mapsto\mf E(\theta,\omega)\in[0,\infty))\bigr)_{\theta\in B}$,
      $(\mc E_x)_{x\in E}\is(\mc E(H_\theta))_{\theta\in B}$
   in the notation of \cref{lem:cov4}) establishes that
     $\Omega\ni\omega\mapsto \sup_{\theta\in B}\abs{\mf E(\theta,\omega)-\mc E(H_\theta)}\in [0,\infty]$ is $\vmcF$/$\Borel([0,\infty])$-measurable and
   \begin{equation}
    \label{eq:cov5bnd1}
    \P\bigl(\sup\nolimits_{\theta\in B}\abs{\mf E(\theta)-\mc E(H_\theta)}\geq\eps\bigr)
    \leq
    2\CovNum{(B,\delta),\frac\eps{8L\mc R}}\exp\pr*{\frac{-\eps^2M}{2\mc R^4}}
  \end{equation}
  \cfload.
  Moreover, note that
    \cref{lem:covering_number_cube_infty}
      (applied with
        $d\is\mf d$,
        $a\is -R$,
        $b\is R$,
        $r\is\tfrac{\eps}{8L\mc R}$,
        $\delta\is\delta$
      in the notation of \cref{prop:covering})
  demonstrates that
  \begin{equation}
    \CovNum{(B,\delta),\frac{\eps}{8L\mc R}}
    \leq
    \max\biggl\{1,
    \pr*{\frac{16LR\mc R}{\eps}}^{\mf d}\biggr\}
    .
  \end{equation}
    This
    and \cref{eq:cov5bnd1}
  prove
    \cref{eq:cov5con}.
   The proof of \cref{lem:cov5} is thus complete.
\end{proof}
\endgroup

\cfclear
\begingroup
\newcommand{\vmcF}{\mc F}
\begin{cor}
  \label{lem:cov6}
  Let 
    $\mf d,M,L\in\N$,
    $u\in\R$,
    $v\in(u,\infty)$,
    $R\in[1,\infty)$,
    $\eps,b\in(0,\infty)$,
    $l=(l_0,l_1,\dots,l_L)\in\N^{L+1}$
    satisfy 
      $l_L=1$ 
      and $\sum_{k=1}^Ll_k(l_{k-1}+1) \leq \mf d$,
  let $D\subseteq[-b,b]^{l_0}$ be a compact set,
  let $(\Omega,\vmcF,\P)$ be a probability space,
  let $X_m\colon\Omega\to D$, $m\in\{1,2,\dots,M\}$,
    and $Y_m\colon\Omega\to [u,v]$, $m\in\{1,2,\dots,M\}$,
    be functions,
  assume that
  $(X_m,Y_m)%
     $, $m\in\{1,2,\dots,M\}$,
    are i.i.d.\ random variables,
  let $\mc E\colon C(D,\R)\to[0,\infty)$ satisfy 
    for all
      $f\in C(D,\R)$
    that
      $\mc E(f)=\E[\abs{f(X_1)-Y_1}^2]$,
  and let $\mf E\colon [-R,R]^{\mf d}\times\Omega\to[0,\infty)$ satisfy
    for all
      $\theta\in [-R,R]^{\mf d}$,
      $\omega\in\Omega$
    that
    \begin{equation}
      \mf E(\theta,\omega)
      =
      \frac1M\br*{\sum_{m=1}^M\abs{\ClippedRealV{\theta}{l} uv(X_m(\omega))-Y_m(\omega)}^2}
    \end{equation}
  \cfload.
  Then
  \begin{enumerate}[label=(\roman *)]
  \item 
  \label{lem:cov6:item1}
    it holds that 
    $\Omega\ni\omega\mapsto \sup\nolimits_{\theta\in [-R,R]^{\mf d}}\babs{\mf E(\theta,\omega)-\mc E\bigl(\ClippedRealV\theta{l}uv|_{D}\bigr)}\in[0,\infty]$ is $\vmcF$/$\Borel([0,\infty])$-measurable and
  \item 
  \label{lem:cov6:item2}
  it holds that 
	\begin{equation}
	  \label{eq:cov6con}
	  \begin{split}
	  &\P\pr*{\sup\nolimits_{\theta\in [-R,R]^{\mf d}}\babs{\mf E(\theta)-\mc E\bigl(\ClippedRealV\theta{l}uv|_{D}\bigr)}\geq\eps}
	  \\&\leq
	  2\max\biggl\{1,\biggl[\frac{16L\max\{1,b\}(\infnorm l+1)^LR^{L}(v-u)}{\eps}\biggr]^{\mf d}\biggr\}
	  \exp\pr*{\frac{-\eps^2M}{2(v-u)^4}}
	  .
	  \end{split}
	\end{equation}
  \end{enumerate}
\end{cor}
\begin{proof}[Proof of \cref{lem:cov6}]
  Throughout this proof,
    let $\mf L\in(0,\infty)$ satisfy
    \begin{equation}
      \mf L
      =
      L 
      \max\{ 1, b \}
      \, 
      ( \infnorm{ l } + 1 )^L 
      R^{ L - 1 }
      .
    \end{equation}
  Observe that
    \cref{cor:ClippedRealNNLipsch}
      (applied with
        $a\is -b$,
        $b\is b$,
        $u\is u$,
        $v\is v$,
        $d\is \mf d$,
        $L\is L$,
        $l\is l$
      in the notation of \cref{cor:ClippedRealNNLipsch})
    and the assumption that $D\subseteq[-b,b]^{l_0}$
  show that for all
    $\theta,\vartheta\in [-R,R]^{\mf d}$
  it holds that
  \begin{equation}
  \begin{split}
    \label{eq:cov6lipsch}
    &\sup_{x\in D}\,\abs{\ClippedRealV{\theta}{l}uv(x)-\ClippedRealV\vartheta{l}uv(x)}
    \\&\leq
    \sup_{x\in[-b,b]^{l_0}}\abs{\ClippedRealV{\theta}{l}uv(x)-\ClippedRealV\vartheta{l}uv(x)}
    \\&\leq
    L 
    \max\{ 1, b \}
    \, 
    ( \infnorm{ l } + 1 )^L 
    \,
    ( \max\{ 1, \infnorm\theta,\infnorm\vartheta\} )^{ L - 1 }
    \infnorm{ \theta - \vartheta }
    \\&\leq
    L 
    \max\{ 1, b \}
    \, 
    ( \infnorm{ l } + 1 )^L 
    R^{ L - 1 }
    \infnorm{ \theta - \vartheta }
    =
    \mf L\infnorm{ \theta - \vartheta }
    .
  \end{split}
  \end{equation}
  Furthermore, observe that
    the fact that for all 
      $\theta\in\R^{\mf d}$,
      $x\in\R^{l_0}$
    it holds that
      $\ClippedRealV{\theta}{l}uv(x)\in[u,v]$
    and the assumption that for all 
      $m\in\{1,2,\dots,M\}$,
      $\omega\in\Omega$
    it holds that
      $Y_m(\omega)\in[u,v]$
  demonstrate that for all
    $\theta\in [-R,R]^{\mf d}$,
    $m\in\{1,2,\dots,M\}$
  it holds that
  \begin{equation}
    \abs{\ClippedRealV\theta{l}uv(X_m)-Y_m}
    \leq 
    v-u
    .
  \end{equation}
  Combining
    this
    and~\cref{eq:cov6lipsch}
  with
    \cref{lem:cov5}
      (applied with
        $d\is l_0$,
        $\mf d\is\mf d$,
        $M\is M$,
        $R\is R$,
        $L\is \mf L$,
        $\mc R\is v-u$,
        $\eps\is\eps$,
        $D\is D$,
        $(\Omega,\vmcF,\P)\is(\Omega,\vmcF,\P)$,
        $(X_m)_{m\in\{1,2,\dots,M\}}\is(X_m)_{m\in\{1,2,\dots,M\}}$,
        $(Y_m)_{m\in\{1,2,\dots,M\}}\is((\Omega\ni\omega\mapsto Y_m(\omega)\in\R))_{m\in\{1,2,\dots,M\}}$,
        $H\is([-R,R]^{\mf d}\ni\theta\mapsto \ClippedRealV\theta{l}uv|_{D}\in C(D,\R))$,
        $\mc E\is\mc E$,
        $\mf E\is\mf E$
      in the notation of \cref{lem:cov5})
  establishes that
    $\Omega\ni\omega\mapsto \sup\nolimits_{\theta\in [-R,R]^{\mf d}}\babs{\mf E(\theta,\omega)-\mc E\bigl(\ClippedRealV\theta{l}uv|_{D}\bigr)}\in[0,\infty]$ is $\vmcF$/$\Borel([0,\infty])$-measurable and
  \begin{equation}
    \P\pr*{\sup\nolimits_{\theta\in [-R,R]^{\mf d}}\babs{\mf E(\theta)-\mc E\bigl(\ClippedRealV\theta{l}uv|_{D}\bigr)}\geq\eps}
    \leq
    2\max\biggl\{1,\biggl[\frac{16\mf LR(v-u)}{\eps}\biggr]^{\mf d}\biggr\}
    \exp\pr*{\frac{-\eps^2M}{2(v-u)^4}}
    .
  \end{equation}
  The proof of \cref{lem:cov6} is thus complete.
\end{proof}
\endgroup

%% file: parts/Analysis_of_the_generalization_error.tex
\cchapter{Strong generalization error estimates}{sec:generalisation_error}

In \cref{sect:probabilistic_generalization} above we reviewed generalization error estimates in the  probabilistic sense. 
Besides such probabilistic generalization error estimates, generalization error estimates in the
strong $L^p$-sense are also considered in the literature and in our overall error analysis in \cref{sec:composed_error} below we employ such strong generalization error estimates. 
These estimates are precisely the subject of this chapter 
(cf.\ \cref{cor:generalisation_error} below).

We refer to the beginning of \cref{sect:probabilistic_generalization} for a short list of
references in the literature dealing with similar generalization error estimates.
The specific material in this chapter mostly consists of slightly modified extracts from Jentzen \& Welti~\cite[Section 4]{JentzenWelti2023}.

\section{Monte Carlo estimates}
\label{sec:MC_estimates}

\cfclear
\begin{athm}{prop}{lem:MC_L2}
	Let
$ d, M \in \N $,
let
$ ( \Omega, \cF, \P ) $
be a probability space,
let
$ X_j \colon \Omega \to \R^d $,
$ j \in \{ 1, 2, \ldots, M \} $,
be independent random variables,
and assume
$ \max_{ j \in \{ 1, 2, \ldots, M \} } \E[ \pnorm2{ X_j } ] \allowbreak < \infty $
\cfload.
Then
\begin{eqsplit}
&\biggl(
\E\biggl[
    \bbbpnorm2{
    \frac{1}{M}
    \biggl[
    \smallsum_{j=1}^M
        X_j
    \biggr]
    -
    \E\biggl[
    \frac{1}{M}
    \bbbbr{
    \smallsum_{j=1}^M
        X_j
    }
    \biggr]
    }^2
\biggr]
\biggr)^{ \!\! \nicefrac{1}{2} }
\\&\leq
\frac{ 1 }{ \sqrt{M} }
\biggl[
\max_{ j \in \{ 1, 2, \ldots, M \} }
    \bigl(
    \E\bigl[
        \pnorm2{ X_j - \E[ X_j ] }^2
    \bigr]
    \bigr)^{ \nicefrac{1}{2} }
\biggr]
.
\end{eqsplit}
\end{athm}
\begingroup
\begin{aproof}
	\Nobs that 
		the fact that 
		  for all
				$x\in\R^d$
			it holds that
				$\scp{x,x}=\pnorm2x^2$
	\proves that
	\begin{equation}
		\begin{split}
			&\bbbpnorm2{\frac1M\biggl[\smallsum_{j=1}^MX_j\biggr]-\E\biggl[\frac1M\smallsum_{j=1}^MX_j\biggr]}^2
			\\&=
			\frac1{M^2}\bbbpnorm2{\biggl[\smallsum_{j=1}^MX_j\biggr]-\E\biggl[\smallsum_{j=1}^MX_j\biggr]}^2
			\\&=
			\frac1{M^2}\bbbpnorm2{\smallsum_{j=1}^M\bigl(X_j-\E[X_j]\bigr)}^2
			\\&=
			\frac1{M^2}\biggl[\smallsum_{i,j=1}^M\bscp{ X_i-\E[X_i],X_j-\E[X_j]}\biggr]
			\\&=
			\frac1{M^2}\biggl[\smallsum_{j=1}^M\pnorm2{X_j-\E[X_j]}^2\biggr]
			+
			\frac1{M^2}\biggl[\smallsum_{(i,j)\in\{1,2,\dots,M\}^2,\,i\neq j}\bscp{ X_i-\E[X_i],X_j-\E[X_j]}\biggr]
		\end{split}
	\end{equation}
    \cfload.
		This,
		the fact that 
			for all 
				independent random variables $Y\colon\Omega\to\R^d$ and $Z\colon\Omega\to\R^d$
				with $\E[\pnorm2{Y}+\pnorm2Z]<\infty$
			it holds that
				$\E[\abs{\scp{Y, Z}}]<\infty$
				and $\E[\scp{Y,Z}]=\scp{\E[Y],\E[Z]}$,
		and the assumption that
		  $X_j\colon\Omega\to\R^d$, $j\in\{1,2,\dots,M\}$, are independent random variables
	\prove that
	\begin{align*}
        &\E\biggl[\bbbpnorm2{\frac1M\biggl[\smallsum_{j=1}^MX_j\biggr]-\E\biggl[\frac1M\smallsum_{j=1}^MX_j\biggr]}^2\biggr]
        \\&=
        \frac1{M^2}\biggl[\smallsum_{j=1}^M\E\bigl[\pnorm2{X_j-\E[X_j]}^2\bigr]\biggr]
        +
        \frac1{M^2}\biggl[\smallsum_{(i,j)\in\{1,2,\dots,M\}^2,\,i\neq j}\bscp{ \E\bigl[X_i-\E[X_i]\bigr],\E\bigl[X_j-\E[X_j]\bigr]}\biggr]
        \\&=\yesnumber
        \frac1{M^2}\biggl[\smallsum_{j=1}^M\E\bigl[\pnorm2{X_j-\E[X_j]}^2\bigr]\biggr]
        \\&\leq
        \frac1{M}\biggl[\max_{j\in\{1,2,\dots,M\}}\E\bigl[\pnorm2{X_j-\E[X_j]}^2\bigr]\biggr]
        .
	\end{align*}
\end{aproof}
\endgroup

\begin{adef}{def:Rademacher_family}[Rademacher family]
Let
$ ( \Omega, \cF, \P ) $
be a probability space
and
let $ J $ be a set.
Then we say that 
$ ( r_j )_{ j \in J } $
is a $ \P $-Rademacher family
if and only if
it holds that
$ r_j \colon \Omega \to \{ -1, 1 \} $,
$ j \in J $,
are independent random variables 
with
\begin{equation}
  \forall \, j \in J \colon
\P( r_j = 1 )
= \P( r_j = - 1 )
.
\end{equation}
\end{adef}

\cfclear
\begingroup
\DeclarePairedDelimiterX{\localnorm}[1]
{\vvvert}
{\vvvert}
{\ifblank{#1}{\:\cdot\:}{#1}}
\begin{adef}{def:Kahane_Khintchine}[$ p $-Kahane--Khintchine constant]
\cfconsiderloaded{def:Kahane_Khintchine}
Let $ p \in ( 0, \infty ) $.
Then
we denote by
$ \KKc p \in ( 0, \infty ] $
the extended real number given by
\begin{equation}
\label{def:Kahane_Khintchine:eq1}
\KKc p =
\sup \cu*{
    c \in [ 0, \infty ) \colon
     \br*{
    \arraycolsep=0pt \begin{array}{c}
        \exists \, \R\text{-Banach space}\ ( E, \localnorm{\cdot} ) \colon \\
        \exists \, \text{probability space}\ ( \Omega, \cF, \P ) \colon \\
        \exists \, \P\text{-Rademacher family\cfadd{def:Rademacher_family}}\ ( r_j )_{ j \in \N } \colon \\
        \exists \, k \in \N \colon \exists \, x_1, x_2, \ldots, x_k \in E \backslash \{ 0 \} \colon \\
        \Bigl(
            \E\Bigl[ \localnorm[\big]{
                \sum_{ j = 1 }^k r_j x_j
            }^p \Bigr]
        \Bigr)^{ \! \nicefrac{1}{p} }
        =
        c \Bigl(
            \E\Bigl[ \localnorm[\big]{
                \sum_{ j = 1 }^k r_j x_j
            }^2 \Bigr]
        \Bigr)^{ \! \nicefrac{1}{2} }
    \end{array}
    }
}
\end{equation}
\cfload.
\end{adef}
\endgroup

\cfclear
\begin{athm}{lemma}{lem:Kahane_Khintchine}
It holds for all
$ p \in [ 2, \infty ) $
that
\begin{equation}
\label{Kahane_Khintchine:ass1}
  \KKc p \leq \sqrt{ p - 1 } < \infty
\end{equation}
\cfout.
\end{athm}
\begin{aproof}
\Nobs that
	\cref{def:Kahane_Khintchine:eq1} and
	Grohs et al.\ \cite[Corollary 2.5]{Grohs2023Aproof}
\prove
\cref{Kahane_Khintchine:ass1}.
\end{aproof}

\cfclear
\begin{athm}{prop}{prop:MC_Lp}
Let
$ d, M \in \N $,
$ p \in [ 2, \infty ) $,
let
$ ( \Omega, \cF, \P ) $
be a probability space,
let
$ X_j \colon \Omega \to \R^d $,
$ j \in \{ 1, 2, \ldots, M \} $,
be independent random variables,
and assume
\begin{equation}
    \max_{ j \in \{ 1, 2, \ldots, M \} } \E[ \pnorm2 {X_j} ] < \infty  
\end{equation}
\cfload.
Then
\begin{equation}
\biggl(
\E\biggl[
    \bbbpnorm2{
    \biggl[
    \smallsum_{j=1}^M
        X_j
    \biggr]
    -
    \E\biggl[
    \smallsum_{j=1}^M
        X_j
    \biggr]
    }^p
\biggr]
\biggr)^{ \!\! \nicefrac{1}{p} }
\leq
2 \KKc p
\biggl[
\smallsum_{j=1}^M
    \bigl(
    \E\bigl[
        \pnorm 2{ X_j - \E[ X_j ] }^p
    \bigr]
    \bigr)^{ \nicefrac{2}{p} }
\biggr]^{ \nicefrac{1}{2} }
\cfadd{lem:Kahane_Khintchine}
\end{equation}
\cfout.
\end{athm}
\begin{aproof}
\Nobs that
	\cref{def:Kahane_Khintchine:eq1} and
	Cox et al.~\cite[Corollary 5.11]{Cox2020}
\prove
\cref{Kahane_Khintchine:ass1}.
\end{aproof}

\cfclear
\begin{athm}{cor}{cor:MC_Lp}
Let
$ d, M \in \N $,
$ p \in [ 2, \infty ) $,
let
$ ( \Omega, \cF, \P ) $
be a probability space,
let
$ X_j \colon \Omega \to \R^d $,
$ j \in \{ 1, 2, \ldots, M \} $,
be independent random variables,
and assume
\begin{equation}
    \max_{ j \in \{ 1, 2, \ldots, M \} } \E[ \pnorm2 {X_j} ] < \infty 
\end{equation}
\cfload.
Then
\begin{equation}
\biggl(
\E\biggl[
    \bbbpnorm2{
    \frac{1}{M}
    \biggl[
    \smallsum_{j=1}^M
        X_j
    \biggr]
    -
    \E\biggl[
    \frac{1}{M}
    \smallsum_{j=1}^M
        X_j
    \biggr]
    }^p
\biggr]
\biggr)^{ \!\! \nicefrac{1}{p} }
\leq
\frac{ 2 \sqrt{ p - 1 } }{ \sqrt{M} }
\biggl[
\max_{ j \in \{ 1, 2, \ldots, M \} }
    \bigl(
    \E\bigl[
        \pnorm 2{ X_j - \E[ X_j ] }^p
    \bigr]
    \bigr)^{ \nicefrac{1}{p} }
\biggr]
.
\end{equation}
\end{athm}
\begin{aproof}
\Nobs that
\cref{prop:MC_Lp}
and
\cref{lem:Kahane_Khintchine}
\prove that
\begin{equation}
\begin{split}
&
\biggl(
\E\biggl[
    \bbbpnorm2{
    \frac{1}{M}
    \biggl[
    \smallsum_{j=1}^M
        X_j
    \biggr]
    -
    \E\biggl[
    \frac{1}{M}
    \smallsum_{j=1}^M
        X_j
    \biggr]
    }^p
\biggr]
\biggr)^{ \!\! \nicefrac{1}{p} }
\\ &
=
\frac{1}{M}
\biggl(
\E\biggl[
    \bbbpnorm2{
    \biggl[
    \smallsum_{j=1}^M
        X_j
    \biggr]
    -
    \E\biggl[
    \smallsum_{j=1}^M
        X_j
    \biggr]
    }^p
\biggr]
\biggr)^{ \!\! \nicefrac{1}{p} }
\\ &
\leq
\frac{ 2 \KKc p }{M}
\biggl[
\smallsum_{j=1}^M
    \bigl(
    \E\bigl[
        \pnorm 2{ X_j - \E[ X_j ] }^p
    \bigr]
    \bigr)^{ \nicefrac{2}{p} }
\biggr]^{ \nicefrac{1}{2} }
\\ &
\leq
\frac{ 2 \KKc p }{M}
\biggl[
M
\biggl(
\max_{ j \in \{ 1, 2, \ldots, M \} }
    \bigl(
    \E\bigl[
        \pnorm 2{ X_j - \E[ X_j ] }^p
    \bigr]
    \bigr)^{ \nicefrac{2}{p} }
\biggr)
\biggr]^{ \nicefrac{1}{2} }
\\ &
=
\frac{ 2 \fK_p }{ \sqrt{M} }
\biggl[
\max_{ j \in \{ 1, 2, \ldots, M \} }
    \bigl(
    \E\bigl[
        \pnorm 2{ X_j - \E[ X_j ] }^p
    \bigr]
    \bigr)^{ \nicefrac{1}{p} }
\biggr]
\\ &
\leq
\frac{ 2 \sqrt{ p - 1 } }{ \sqrt{M} }
\biggl[
\max_{ j \in \{ 1, 2, \ldots, M \} }
    \bigl(
    \E\bigl[
        \pnorm 2{ X_j - \E[ X_j ] }^p
    \bigr]
    \bigr)^{ \nicefrac{1}{p} }
\biggr]
\end{split}
\end{equation}
\cfload.
\end{aproof}

\section{Uniform strong error estimates for random fields}
\label{sec:uniform_strong_error}

\cfclear
\begin{athm}{lemma}{lem:Lp_sup_random_field}
Let
$ ( E, \delta ) $
be a separable metric space,
let
$ N \in \N $,
$ r_1, r_2, \ldots, r_N \in [ 0, \infty ) $,
$ z_1, z_2, \ldots, z_N \in E $
satisfy
\begin{equation}
  \llabel{eq:2}
    \textstyle
    E \subseteq 
    \bigcup_{ n = 1 }^N
    \{ x \in E \colon \delta( x, z_n ) \leq r_n \} 
    ,  
\end{equation}
let
$ ( \Omega, \cF, \P ) $
be a probability space,
for every $x\in E$ let
$ Z_x \colon \Omega \to \R $
be a random variable,
let
$L\in[0,\infty)$
satisfy for all
$ x, y \in E $
that
$ \lvert Z_x - Z_y \rvert \leq L \delta( x, y ) $,
and let
$p\in[0,\infty)$.
Then
\begin{equation}
\bExp{
    \sup\nolimits_{ x \in E }
    \, \lvert Z_x \rvert^p
}
\leq
\sum_{ n = 1 }^N
\bExp{
    ( L r_n + \lvert Z_{ z_n } \rvert )^p
}
\cfadd{lem:measurability_sup}
\end{equation}
\cfout.
\end{athm}
\begin{aproof}
Throughout this proof,
for every
  $n\in \{1,2,\dots,N\}$
let
\begin{equation}
  \llabel{eq:1}
    B_n =
    \{ x \in E \colon \delta( x, z_n ) \leq r_n \}
    .  
\end{equation}
\Nobs that
  \lref{eq:2}
  and \lref{eq:1}
\prove that
\begin{equation}
  \textstyle
  E \subseteq \bigcup_{ n = 1 }^N B_n
  \qandq
  E \supseteq \bigcup_{ n = 1 }^N B_n
  .
\end{equation}
\Hence that
\begin{equation}
\sup\nolimits_{ x \in E }
\lvert Z_x \rvert
=
\sup\nolimits_{ x \in \pr*{ \bigcup_{ n = 1 }^N B_n } }
\lvert Z_x \rvert
=
\max\nolimits_{ n \in \{ 1, 2, \ldots, N \} }
\sup\nolimits_{ x \in B_n }
\lvert Z_x \rvert
.
\end{equation}
\Hence that
\begin{equation}
\label{eq:sup_power}
\begin{split}
& \bExp{
    \sup\nolimits_{ x \in E }
    \, \lvert Z_x \rvert^p
}
=
\bExp{
    \max\nolimits_{ n \in \{ 1, 2, \ldots, N \} }
    \sup\nolimits_{ x \in B_n }
    \lvert Z_x \rvert^p
}
\\ &
\leq
\bbbExp{
    \smallsum_{ n = 1 }^N
    \sup\nolimits_{ x \in B_n }
    \lvert Z_x \rvert^p
}
=
\smallsum_{ n = 1 }^N
\bExp{
    \sup\nolimits_{ x \in B_n }
    \lvert Z_x \rvert^p
}
.
\cfadd{lem:measurability_sup}
\end{split}
\end{equation}
\cfload.
\Moreover
the assumption that
for all
$ x, y \in E$
it holds that $\lvert Z_x - Z_y \rvert \leq L \delta( x, y ) $
\proves that for all
$ n \in \{ 1, 2, \ldots, N \} $,
$ x \in B_n $
it holds that
\begin{equation}
\lvert Z_x \rvert
=
\lvert Z_x - Z_{ z_n } + Z_{ z_n } \rvert
\leq
\lvert Z_x - Z_{ z_n } \rvert + \lvert Z_{ z_n } \rvert
\leq
L \delta( x, z_n ) + \lvert Z_{ z_n } \rvert
\leq
L r_n + \lvert Z_{ z_n } \rvert
.
\end{equation}
This and \cref{eq:sup_power}
\prove that
\begin{equation}
\bExp{
    \sup\nolimits_{ x \in E }
    \, \lvert Z_x \rvert^p
}
\leq
\smallsum_{ n = 1 }^N
\bExp{
    ( L r_n + \lvert Z_{ z_n } \rvert )^p
}
.
\end{equation}
\end{aproof}

\cfclear
\begin{athm}{lemma}{lem:Lp_sup_covering_number}
Let
$ ( E, \delta ) $
be a non-empty separable metric space,
let
$ ( \Omega, \cF, \P ) $
be a probability space,
for every $x\in E$ let
$ Z_x \colon \Omega \to \R $
be a random variable,
let 
$L\in(0,\infty)$
satisfy for all
$ x, y \in E $
that
$ \lvert Z_x - Z_y \rvert \leq L \delta( x, y ) $,
and let 
$ p, r \in ( 0, \infty ) $.
Then
\begin{equation}
\bExp{
    \sup\nolimits_{ x \in E }
    \, \lvert Z_x \rvert^p
}\leq
\CovNum{ ( E, \delta ), r }
\biggl[
\sup_{ x \in E }
\bExp{
    ( L r + \lvert Z_{ x } \rvert )^p
}\biggr]
\cfadd{lem:measurability_sup}
\end{equation}
\cfout.
\end{athm}
\begin{aproof}
Throughout this proof,
assume without loss of generality that
$ \CovNum{ ( E, \delta ), r } < \infty $,
let
$ N = \CovNum{ ( E, \delta ), r } $,
and
let
$ z_1, z_2, \ldots, z_N \in E $
satisfy
\begin{equation}
    \textstyle
    E \subseteq
    \bigcup_{ n = 1 }^N
    \{ x \in E \colon \delta( x, z_n ) \leq r \}  
\end{equation}
\cfload.
\Nobs that
\cref{lem:Lp_sup_random_field}
(applied with
$ r_1 \is r $,
$ r_2 \is r $,
\ldots,
$ r_N \is r $
in the notation of \cref{lem:Lp_sup_random_field})
\proves that
\begin{equation}
\begin{split}
& \bExp{
    \sup\nolimits_{ x \in E }
    \, \lvert Z_x \rvert^p
}\leq
\smallsum_{ i = 1 }^N
\bExp{
    ( L r + \lvert Z_{ z_i } \rvert )^p
}\\ &
\leq
\smallsum_{ i = 1 }^N
\biggl[
\sup_{ x \in E }
\bExp{
    ( L r + \lvert Z_{ x } \rvert )^p
}\biggr]
=
N
\biggl[
\sup_{ x \in E }
\bExp{
    ( L r + \lvert Z_{ x } \rvert )^p
}\biggr]
.
\cfadd{lem:measurability_sup}
\end{split}
\end{equation}
\cfload.
\end{aproof}

\cfclear
\begin{athm}{lemma}{lem:Lp_sup_centred}
Let
$ ( E, \delta ) $
be a non-empty separable metric space,
let
$ ( \Omega, \cF, \P ) $
be a probability space,
for every $x\in E$
let
$ Z_x \colon \Omega \to \R $
be a random variable with
$ \E[ \lvert Z_x \rvert ] < \infty $,
let
$L\in(0,\infty)$
satisfy for all
$ x, y \in E $
that
$ \lvert Z_x - Z_y \rvert \leq L \delta( x, y ) $,
and let
$ p \in [ 1, \infty ) $,
$ r \in ( 0, \infty ) $
\cfload.
Then
\begin{equation}
\bigl(
\bExp{
    \sup\nolimits_{ x \in E }
    \lvert Z_x - \E[ Z_x ] \rvert^p
}
\bigr)^{ \nicefrac{1}{p} }
\leq
( \CovNum{ ( E, \delta ), r } )^{ \nicefrac{1}{p} }
\Bigl[
2 L r
+
\sup\nolimits_{ x \in E }
\bigl(
\bExp{
    \lvert Z_x - \E[ Z_x ] \rvert^p
}
\bigr)^{ \nicefrac{1}{p} }
\Bigr]
\cfadd{lem:measurability_sup_centred}
\end{equation}
\cfout.
\end{athm}
\begin{aproof}
Throughout this proof,
for every
$x\in E$
let
$ Y_x \colon \Omega \to \R $
satisfy for all
$ \omega \in \Omega $
that
\begin{equation}
    \llabel{eq:1}
    Y_x( \omega ) = Z_x( \omega ) - \E[ Z_x ]
    .
\end{equation}
\Nobs that
    \lref{eq:1}
    and the triangle inequality
\prove that for all
$ x, y \in E $
it holds that
\begin{equation}
\begin{split}
\lvert Y_x - Y_y \rvert
&=
\lvert
    ( Z_x - \E[ Z_x ] )
    -
    ( Z_y - \E[ Z_y ] )
\rvert
\\&=
\abs{
  (Z_x - Z_y)
  -
  (\E[ Z_x ] - \E[ Z_y ])
}
\\&\leq
\lvert
    Z_x - Z_y
\rvert
+
\lvert
    \E[ Z_x ] - \E[ Z_y ]
\rvert
\\&\leq
L \delta( x, y )
+
\E[ \lvert Z_x - Z_y \rvert ]
\leq
2 L \delta( x, y )
.
\end{split}
\end{equation}
\cref{lem:Lp_sup_covering_number}
(applied with
$ L \is 2 L $,
$ ( \Omega, \cF, \P ) \is ( \Omega, \cF, \P ) $,
$ ( Z_x )_{ x \in E }
\is
( Y_x )_{ x \in E } $
in the notation of \cref{lem:Lp_sup_covering_number})
\hence
\proves that
\begin{equation}
\begin{split}
\bigl(
\E\bigl[
    \sup\nolimits_{ x \in E }
    \lvert Z_x - \E[ Z_x ] \rvert^p
\bigr]
\bigr)^{ \nicefrac{1}{p} }
&=
\bigl(
\E\bigl[
    \sup\nolimits_{ x \in E }
    \lvert Y_x \rvert^p
\bigr]
\bigr)^{ \nicefrac{1}{p} }
\\ &
\leq
( \CovNum{ ( E, \delta ), r } )^{ \nicefrac{1}{p} }
\Bigl[
\sup\nolimits_{ x \in E }
\bigl(
\E\bigl[
    ( 2 L r + \lvert Y_{ x } \rvert )^p
\bigr]
\bigr)^{ \nicefrac{1}{p} }
\Bigr]
\\ &
\leq
( \CovNum{ ( E, \delta ), r } )^{ \nicefrac{1}{p} }
\Bigl[
2 L r
+
\sup\nolimits_{ x \in E }
\bigl(
\E\bigl[
    \lvert Y_{ x } \rvert^p
\bigr]
\bigr)^{ \nicefrac{1}{p} }
\Bigr]
\\ &
=
( \CovNum{ ( E, \delta ), r } )^{ \nicefrac{1}{p} }
\Bigl[
2 L r
+
\sup\nolimits_{ x \in E }
\bigl(
\E\bigl[
    \lvert Z_x - \E[ Z_x ] \rvert^p
\bigr]
\bigr)^{ \nicefrac{1}{p} }
\Bigr]
.
\end{split}
\end{equation}
\end{aproof}

\cfclear
\begin{athm}{lemma}{lem:Lp_sup_MC}
Let
$ ( E, \delta ) $
be a non-empty separable metric space,
let
$ ( \Omega, \cF, \P ) $
be a probability space,
let
$M\in\N$,
for every
$ x \in E $
let
$ Y_{ x, m } \colon \Omega \to \R $,
$ m \in \{ 1, 2, \ldots, M \} $,
be independent random variables
with
$\bExp{ \abs{ Y_{ x, 1 } } + \abs{ Y_{ x, 2 } } + \ldots + \abs{ Y_{ x, m } } } < \infty$,
let
$ L \in (0,\infty)$
satisfy for all
$ x, y \in E $,
$ m \in \{ 1, 2, \ldots, M \} $
that
\begin{equation}
  \llabel{eq:Lipschitz}
  \lvert Y_{ x, m } - Y_{ y, m } \rvert \leq L \delta( x, y )
  ,
\end{equation}
and
for every
$ x \in E $
let
$ Z_x \colon \Omega \to \R $
satisfy
\begin{equation}
Z_x
=
\frac{1}{M}
\biggl[
\smallsum_{m=1}^M
    Y_{ x, m }
\biggr]
.
\end{equation}
Then
\begin{enumerate}[label=(\roman *)]
\item
\label{item:lem:Lp_sup_MC:1}
it holds for all
$ x \in E $
that
$ \E[ \lvert Z_x \rvert ] < \infty $,
\item
\label{item:lem:Lp_sup_MC:2}
it holds that
$ \Omega \ni \omega \mapsto
\sup\nolimits_{ x \in E } \lvert Z_x( \omega ) - \E[ Z_x ] \rvert
\in [ 0, \infty ] $
is $ \cF $/$ \cB( [ 0, \infty ] ) $-measurable,
and
\item
\label{item:lem:Lp_sup_MC:3}
it holds for all
$ p \in [ 2, \infty ) $,
$ r \in ( 0, \infty ) $
that
\begin{equation}
\begin{split}
&
\bigl(
\E\bigl[
    \sup\nolimits_{ x \in E }
    \lvert Z_x - \E[ Z_x ] \rvert^p
\bigr]
\bigr)^{ \nicefrac{1}{p} }
\\ &
\leq
2
( \CovNum{ ( E, \delta ), r } )^{ \nicefrac{1}{p} }
\Bigl[
L r
+
\tfrac{ \sqrt{ p - 1 } }{ \sqrt{M} }
\Bigl(
\sup\nolimits_{ x \in E }
\max\nolimits_{ m \in \{ 1, 2, \ldots, M \} }
    \bigl(
    \E\bigl[
        \lvert Y_{ x, m } - \E[ Y_{ x, m } ] \rvert^p
    \bigr]
    \bigr)^{ \nicefrac{1}{p} }
\Bigr)
\Bigr]
\end{split}
\end{equation}
\end{enumerate}
\cfout.
\end{athm}
\begin{aproof}
\Nobs that
the assumption that
for all $ x \in E$, $m \in \{ 1, 2, \ldots, M \}$
it holds that
$\E[ \lvert Y_{ x, m } \rvert ] < \infty $
\proves that for all
$ x \in E $
it holds that
\begin{equation}
\label{eq:max_expectation_finite}
\E[ \lvert Z_x \rvert ]
=
\E\biggl[
\frac{1}{M}
\biggl\lvert
\smallsum_{m=1}^M
    Y_{ x, m }
\biggr\rvert
\biggr]
\leq
\frac{1}{M}
\biggl[
\smallsum_{m=1}^M
    \E[ \lvert Y_{ x, m } \rvert ]
\biggr]
\leq
\max_{ m \in \{ 1, 2, \ldots, M \} }
    \E[ \lvert Y_{ x, m } \rvert ]
< \infty
.
\end{equation}
This \proves[ep] \cref{item:lem:Lp_sup_MC:1}.
\Nobs that
\lref{eq:Lipschitz}
\proves that for all
$ x, y \in E $
it holds that
\begin{equation}
\label{eq:Lipschitz_Z}
\lvert Z_x - Z_y \rvert
=
\frac{1}{M}
\biggl\lvert
\biggl[
\smallsum_{m=1}^M
    Y_{ x, m }
\biggr]
-
\biggl[
\smallsum_{m=1}^M
    Y_{ y, m }
\biggr]
\biggr\rvert
\leq
\frac{1}{M}
\biggl[
\smallsum_{m=1}^M
    \lvert Y_{ x, m } - Y_{ y, m } \rvert
\biggr]
\leq
L \delta( x, y )
.
\end{equation}
\Cref{item:lem:Lp_sup_MC:1}
and
\cref{lem:measurability_sup_centred}
\hence \prove[ep] \cref{item:lem:Lp_sup_MC:2}.
It thus remains to show
\cref{item:lem:Lp_sup_MC:3}.
For this \nobs that
\cref{item:lem:Lp_sup_MC:1},
\cref{eq:Lipschitz_Z},
and
\cref{lem:Lp_sup_centred}
\prove that for all
$ p \in [ 1, \infty ) $,
$ r \in ( 0, \infty ) $
it holds that
\begin{equation}
\label{eq:Lp_sup_centred}
\bigl(
\E\bigl[
    \sup\nolimits_{ x \in E }
    \lvert Z_x - \E[ Z_x ] \rvert^p
\bigr]
\bigr)^{ \nicefrac{1}{p} }
\leq
( \CovNum{ ( E, \delta ), r } )^{ \nicefrac{1}{p} }
\Bigl[
2 L r
+
\sup\nolimits_{ x \in E }
\bigl(
\E\bigl[
    \lvert Z_x - \E[ Z_x ] \rvert^p
\bigr]
\bigr)^{ \nicefrac{1}{p} }
\Bigr]
\ifnocf.
\end{equation}
\cfload[.]%
\Moreover
\cref{eq:max_expectation_finite}
and
\cref{cor:MC_Lp}
(applied with
$ d \is 1 $,
$ ( X_m )_{ m \in \{ 1, 2, \ldots, M \} } \is ( Y_{ x, m } )_{ m \in \{ 1, 2, \ldots, M \} } $
for $ x \in E $
in the notation of \cref{cor:MC_Lp})
\prove that for all
$ x \in E $,
$ p \in [ 2, \infty ) $,
$ r \in ( 0, \infty ) $
it holds that
\begin{equation}
\begin{split}
\bigl(
\E\bigl[
    \lvert Z_x - \E[ Z_x ] \rvert^p
\bigr]
\bigr)^{ \nicefrac{1}{p} }
& =
\biggl(
\E\biggl[
    \biggl\lvert
    \frac{1}{M}
    \biggl[
    \smallsum_{m=1}^M
        Y_{ x, m }
    \biggr]
    -
    \E\biggl[
    \frac{1}{M}
    \smallsum_{m=1}^M
        Y_{ x, m }
    \biggr]
    \biggr\rvert^p
\biggr]
\biggr)^{ \!\! \nicefrac{1}{p} }
\\ &
\leq
\frac{ 2 \sqrt{ p - 1 } }{ \sqrt{M} }
\biggl[
\max_{ m \in \{ 1, 2, \ldots, M \} }
    \bigl(
    \E\bigl[
        \lvert Y_{ x, m } - \E[ Y_{ x, m } ] \rvert^p
    \bigr]
    \bigr)^{ \nicefrac{1}{p} }
\biggr]
.
\end{split}
\end{equation}
Combining
this
with
\cref{eq:Lp_sup_centred}
\proves that for all
$ p \in [ 2, \infty ) $,
$ r \in ( 0, \infty ) $
it holds that
\begin{equation}
\begin{split}
&
\bigl(
\E\bigl[
    \sup\nolimits_{ x \in E }
    \lvert Z_x - \E[ Z_x ] \rvert^p
\bigr]
\bigr)^{ \nicefrac{1}{p} }
\\ &
\leq
( \CovNum{ ( E, \delta ), r } )^{ \nicefrac{1}{p} }
\Bigl[
2 L r
+
\tfrac{ 2 \sqrt{ p - 1 } }{ \sqrt{M} }
\Bigl(
\sup\nolimits_{ x \in E }
\max\nolimits_{ m \in \{ 1, 2, \ldots, M \} }
    \bigl(
    \E\bigl[
        \lvert Y_{ x, m } - \E[ Y_{ x, m } ] \rvert^p
    \bigr]
    \bigr)^{ \nicefrac{1}{p} }
\Bigr)
\Bigr]
\\ &
=
2
( \CovNum{ ( E, \delta ), r } )^{ \nicefrac{1}{p} }
\Bigl[
L r
+
\tfrac{ \sqrt{ p - 1 } }{ \sqrt{M} }
\Bigl(
\sup\nolimits_{ x \in E }
\max\nolimits_{ m \in \{ 1, 2, \ldots, M \} }
    \bigl(
    \E\bigl[
        \lvert Y_{ x, m } - \E[ Y_{ x, m } ] \rvert^p
    \bigr]
    \bigr)^{ \nicefrac{1}{p} }
\Bigr)
\Bigr]
.
\end{split}
\end{equation}
\end{aproof}

\cfclear
\begin{athm}{cor}{cor:Lp_sup_MC}
Let
$ ( E, \delta ) $
be a non-empty separable metric space,
let
$ ( \Omega, \cF, \P ) $
be a probability space,
let
$M\in\N$,
for every
$ x \in E $
let
$ Y_{ x, m } \colon \Omega \to \R $,
$ m \in \{ 1, 2, \ldots, M \} $,
be independent random variables
with
$\bExp{ \abs{ Y_{ x, 1 } } + \abs{ Y_{ x, 2 } } + \ldots + \abs{ Y_{ x, m } } } < \infty$,
let
$ L \in ( 0, \infty ) $
satisfy for all
$ x, y \in E $,
$ m \in \{ 1, 2, \ldots, M \} $
that
$ \lvert Y_{ x, m } - Y_{ y, m } \rvert \leq L \delta( x, y ) $,
and for every
$ x \in E $
let
$ Z_x \colon \Omega \to \R $
satisfy
\begin{equation}
Z_x
=
\frac{1}{M}
\biggl[
\smallsum_{m=1}^M
    Y_{ x, m }
\biggr]
.
\end{equation}
Then
\begin{enumerate}[label=(\roman *)]
\item
\label{item:cor:Lp_sup_MC:1}
it holds for all
$ x \in E $
that
$ \E[ \lvert Z_x \rvert ] < \infty $,
\item
\label{item:cor:Lp_sup_MC:2}
it holds that 
$ \Omega \ni \omega \mapsto
\sup\nolimits_{ x \in E } \lvert Z_x( \omega ) - \E[ Z_x ] \rvert
\in [ 0, \infty ] $
is $ \cF $/$ \cB( [ 0, \infty ] ) $-measurable,
and
\item
\label{item:cor:Lp_sup_MC:3}
it holds
for all 
$ p \in [ 2, \infty ) $,
$ c \in ( 0, \infty ) $
that
\begin{align*}
&
\bigl(
\E\bigl[
    \sup\nolimits_{ x \in E }
    \lvert Z_x - \E[ Z_x ] \rvert^p
\bigr]
\bigr)^{ \nicefrac{1}{p} }\yesnumber
\\ &
\leq
\tfrac{ 2 \sqrt{ p - 1 } }{ \sqrt{M} }
\Bigl( \CovNum{ ( E, \delta ), \frac{ c \sqrt{ p - 1 } }{ L \sqrt{M} } } \Bigr)^{ \! \nicefrac{1}{p} }
\Bigl[
c
+
\sup\nolimits_{ x \in E }
\max\nolimits_{ m \in \{ 1, 2, \ldots, M \} }
    \bigl(
    \E\bigl[
        \lvert Y_{ x, m } - \E[ Y_{ x, m } ] \rvert^p
    \bigr]
    \bigr)^{ \nicefrac{1}{p} }
\Bigr]
\end{align*}
\end{enumerate}
\cfout.
\end{athm}
\begin{aproof}
\Nobs that
\cref{lem:Lp_sup_MC}
\proves[ep]
\cref{item:cor:Lp_sup_MC:1,item:cor:Lp_sup_MC:2}.
\Nobs that
\cref{lem:Lp_sup_MC}
(applied with
$ r \is \nicefrac{ c \sqrt{ p - 1 } }{ ( L \sqrt{M} ) } $
for $c\in(0,\infty)$
in the notation of
\cref{lem:Lp_sup_MC})
\proves that for all
$ p \in [ 2, \infty ) $,
$ c \in ( 0, \infty ) $
it holds that
\begin{eqsplit}
&
\bigl(
\E\bigl[
    \sup\nolimits_{ x \in E }
    \lvert Z_x - \E[ Z_x ] \rvert^p
\bigr]
\bigr)^{ \nicefrac{1}{p} }
\\ &
\leq
2
\Bigl( \CovNum{ ( E, \delta ), \frac{ c \sqrt{ p - 1 } }{ L \sqrt{M} } } \Bigr)^{ \! \nicefrac{1}{p} }
\Bigl[
L \tfrac{ c \sqrt{ p - 1 } }{ L \sqrt{M} }
\\&\quad+
\tfrac{ \sqrt{ p - 1 } }{ \sqrt{M} }
\Bigl(
\sup\nolimits_{ x \in E }
\max\nolimits_{ m \in \{ 1, 2, \ldots, M \} }
    \bigl(
    \E\bigl[
        \lvert Y_{ x, m } - \E[ Y_{ x, m } ] \rvert^p
    \bigr]
    \bigr)^{ \nicefrac{1}{p} }
\Bigr)
\Bigr]
\\ &
=
\tfrac{ 2 \sqrt{ p - 1 } }{ \sqrt{M} }
\Bigl( \CovNum{ ( E, \delta ), \frac{ c \sqrt{ p - 1 } }{ L \sqrt{M} } } \Bigr)^{ \! \nicefrac{1}{p} }
\Bigl[
c
+
\sup\nolimits_{ x \in E }
\max\nolimits_{ m \in \{ 1, 2, \ldots, M \} }
    \bigl(
    \E\bigl[
        \lvert Y_{ x, m } - \E[ Y_{ x, m } ] \rvert^p
    \bigr]
    \bigr)^{ \nicefrac{1}{p} }
\Bigr]
\end{eqsplit}
\cfload.
This \proves[ep] \cref{item:cor:Lp_sup_MC:3}.
\end{aproof}

\section{Strong convergence rates for the generalisation error}
\label{sec:strong_rates_generalisation_error}

\cfclear
\begin{athm}{lemma}{lem:abstract_generalisation_error}
Let
$ ( E, \delta ) $
be a separable metric space,
assume
$ E \neq \emptyset $,
let
$ ( \Omega, \cF, \P ) $
be a probability space,
let
$ M \in \N $,
let
$ X_{ x, m } \colon \Omega \to \R $,
$ m \in \{ 1, 2, \ldots, M \} $,
$ x \in E $,
and
$ Y_m \colon \Omega \to \R $,
$ m \in \{ 1, 2, \ldots, M \} $,
be functions,
assume for all
$ x \in E $
that
$ ( X_{ x, m }, Y_m ) $,
$ m \in \{ 1, 2, \ldots, M \} $,
are i.i.d.\ random variables,
let
$ L, b \in ( 0, \infty ) $
satisfy for all
$ x, y \in E $,
$ m \in \{ 1, 2, \ldots, M \} $
that
\begin{equation}
  \llabel{eq:Lip}
 \lvert X_{ x, m } - Y_m \rvert \leq b
  \qquad\text{and}\qquad
 \lvert X_{ x, m } - X_{ y, m } \rvert \leq L \delta( x, y )
 ,    
\end{equation}
and let
$ \bfR \colon E \to [ 0, \infty ) $
and
$ \emprisk \colon E \times \Omega \to [ 0, \infty ) $
satisfy for all
$ x \in E $,
$ \omega \in \Omega $
that
\begin{equation}
  \bfR( x )
  = \E\bbr{ \lvert X_{ x, 1 } - Y_1 \rvert^2 }
  \qquad\text{and}\qquad
  \emprisk( x, \omega )
=
\frac{1}{M}
\biggl[
\smallsum_{m=1}^M
    \lvert X_{ x, m }( \omega ) - Y_m( \omega ) \rvert^2
\biggr]
.
\end{equation}
Then
\begin{enumerate}[label=(\roman *)]
\item
\label{item:lem:abstract_generalisation_error:1}
it holds that
$ \Omega \ni \omega \mapsto
\sup\nolimits_{ x \in E } \lvert \emprisk( x, \omega ) - \bfR( x ) \rvert
\in [ 0, \infty ] $
is $ \cF $/$ \cB( [ 0, \infty ] ) $-measurable
and
\item
\label{item:lem:abstract_generalisation_error:2}
it holds for all
$ p \in [ 2, \infty ) $,
$ c \in ( 0, \infty ) $
that
\begin{equation}
\bigl(
\E\bigl[
    \sup\nolimits_{ x \in E }
    \lvert \emprisk( x ) - \bfR( x ) \rvert^p
\bigr]
\bigr)^{ \nicefrac{1}{p} }
\leq
\Bigl( \CovNum{ ( E, \delta ), \frac{ c b \sqrt{ p - 1 } }{ 2 L \sqrt{M} } } \Bigr)^{ \! \nicefrac{1}{p} }
\biggl[ \frac{ 2 ( c + 1 ) b^2 \sqrt{ p - 1 } }{ \sqrt{M} } \biggr]
\end{equation}
\end{enumerate}
\cfout.
\end{athm}
\begin{aproof}
Throughout this proof,
for every
$ x \in E $,
$ m \in \{ 1, 2, \ldots, M \} $
let
$ \cY_{ x, m } \colon \Omega \to \R $
satisfy
$ \cY_{ x, m } = \lvert X_{ x, m } - Y_m \rvert^2 $.
\Nobs that
the assumption that
for all
$ x \in E $
it holds that
$ ( X_{ x, m }, Y_m ) $,
$ m \in \{ 1, 2, \ldots, M \} $,
are i.i.d.\ random variables
\proves that for all
$ x \in E $
it holds that
\begin{equation}
\label{eq:expected_risk}
\E[ \emprisk( x ) ]
=
\frac{1}{M}
\biggl[
\smallsum_{m=1}^M
    \E\bigl[ \lvert X_{ x, m } - Y_m \rvert^2 \bigr]
\biggr]
=
\frac{ M \, \E\bigl[ \lvert X_{ x, 1 } - Y_1 \rvert^2 \bigr] }{M}
=
\bfR( x )
.
\end{equation}
\Moreover
the assumption that
for all
$ x \in E$,
$ m \in \{ 1, 2, \ldots, M \}$
it holds that
$\lvert X_{ x, m } - Y_m \rvert \leq b $
shows that for all
$ x \in E $,
$ m \in \{ 1, 2, \ldots, M \} $
it holds that
\begin{flalign}
\label{eq:finite_expectation}
&& \E[ \lvert \cY_{ x, m } \rvert ]
& =
\E\bbr{ \lvert X_{ x, m } - Y_m \rvert^2 }
\leq
b^2
< \infty, &&
\\ \llabel{eq:22}
&& \cY_{ x, m } - \E[ \cY_{ x, m } ]
& =
\lvert X_{ x, m } - Y_m \rvert^2
-
\E\bigl[
    \lvert X_{ x, m } - Y_m \rvert^2
\bigr]
\leq
\lvert X_{ x, m } - Y_m \rvert^2
\leq
b^2,
\\ \nonumber
& \text{and}\hidewidth
\\ \label{eq:bound_centred2}
&& \E[ \cY_{ x, m } ] - \cY_{ x, m }
& =
\E\bigl[
    \lvert X_{ x, m } - Y_m \rvert^2
\bigr]
-
\lvert X_{ x, m } - Y_m \rvert^2
\leq
\E\bigl[
    \lvert X_{ x, m } - Y_m \rvert^2
\bigr]
\leq
b^2
.
\end{flalign}
\Nobs that
\cref{eq:finite_expectation},
\lref{eq:22},
and \cref{eq:bound_centred2}
\prove
for all
$ x \in E $,
$ m \in \{ 1, 2, \ldots, M \} $,
$ p\in (0,\infty)$
that
\begin{equation}
\label{eq:bound_Lp_centred}
\bigl(
\E\bigl[
    \lvert \cY_{ x, m } - \E[ \cY_{ x, m } ] \rvert^p
\bigr]
\bigr)^{ \nicefrac{1}{p} }
\leq
\bigl(
\E\bigl[
    b^{ 2p }
\bigr]
\bigr)^{ \nicefrac{1}{p} }
=
b^2
.
\end{equation}
\Moreover
\lref{eq:Lip}
and
the fact that
for all
$ x_1, x_2, y \in \R$
it holds that
$( x_1 - y )^2 - ( x_2 - y)^2
= ( x_1 - x_2 )( ( x_1 - y ) + ( x_2 - y ) ) $
\prove that for all
$ x, y \in E $,
$ m \in \{ 1, 2, \ldots, M \} $
it holds that
\begin{equation}
\begin{split}
\lvert \cY_{ x, m } - \cY_{ y, m } \rvert
& =
\lvert
    ( X_{ x, m } - Y_m )^2
    -
    ( X_{ y, m } - Y_m )^2
\rvert
\\ &
\leq
\lvert
    X_{ x, m } - X_{ y, m }
\rvert
( \lvert X_{ x, m } - Y_m \rvert
    +
\lvert X_{ y, m } - Y_m \rvert )
\\ &
\leq
2 b
\lvert
    X_{ x, m } - X_{ y, m }
\rvert
\leq
2 b
L \delta( x, y )
.
\end{split}
\end{equation}
The fact that
for all
$ x \in E $
it holds that
$ \cY_{ x, m } $, %
$ m \in \{ 1, 2, \ldots, M \} $,
are independent random variables,
\cref{eq:finite_expectation},
and \cref{cor:Lp_sup_MC}
(applied with
$ ( Y_{ x, m } )_{ x \in E, \, m \in \{ 1, 2, \ldots, M \} }
\is
( \cY_{ x, m } )_{ x \in E, \, m \in \{ 1, 2, \ldots, M \} } $,
$ L \is 2 b L $,
$ ( Z_x )_{ x \in E }
\is
( \Omega \ni \omega \mapsto \emprisk( x, \omega ) \in \R )_{ x \in E } $
in the notation of \cref{cor:Lp_sup_MC})
\hence
\prove[ep] that
\begin{enumerate}[(I)]
    \item \llabel{it:1}
    it holds that
$ \Omega \ni \omega \mapsto
\sup\nolimits_{ x \in E } \lvert \emprisk( x, \omega ) - \bfR( x ) \rvert
\in [ 0, \infty ] $
is $ \cF $/$ \cB( [ 0, \infty ] ) $-measurable
and
\item \llabel{it:2}
it holds for all
$ p \in [ 2, \infty ) $,
$ c \in ( 0, \infty ) $
that
\begin{multline}
\bigl(
\E\bigl[
    \sup\nolimits_{ x \in E }
    \lvert \emprisk( x ) - \E[ \emprisk( x ) ] \rvert^p
\bigr]
\bigr)^{ \nicefrac{1}{p} }
\leq
\tfrac{ 2 \sqrt{ p - 1 } }{ \sqrt{M} }
\Bigl( \CovNum{ ( E, \delta ), \frac{ c b^2 \sqrt{ p - 1 } }{ 2 b L \sqrt{M} } } \Bigr)^{ \! \nicefrac{1}{p} }
\Bigl[
c b^2
\\+
\sup\nolimits_{ x \in E }
\max\nolimits_{ m \in \{ 1, 2, \ldots, M \} }
    \bigl(
    \E\bigl[
        \lvert \cY_{ x, m } - \E[ \cY_{ x, m } ] \rvert^p
    \bigr]
    \bigr)^{ \nicefrac{1}{p} }
\Bigr]
.
\end{multline}
\end{enumerate}
\Nobs that
\cref{lem:abstract_generalisation_error.it:2},
\cref{eq:expected_risk},
\cref{eq:finite_expectation},
and \cref{eq:bound_Lp_centred}
\prove that for all
$ p \in [ 2, \infty ) $,
$ c \in ( 0, \infty ) $
it holds that
\begin{equation}
\begin{split}
\bigl(
\E\bigl[
    \sup\nolimits_{ x \in E }
    \lvert \emprisk( x ) - \bfR( x ) \rvert^p
\bigr]
\bigr)^{ \nicefrac{1}{p} }
&
\leq
\tfrac{ 2 \sqrt{ p - 1 } }{ \sqrt{M} }
\Bigl( \CovNum{ ( E, \delta ), \frac{ c b \sqrt{ p - 1 } }{ 2 L \sqrt{M} } } \Bigr)^{ \! \nicefrac{1}{p} }
[ c b^2 + b^2 ]
\\&=
\Bigl( \CovNum{ ( E, \delta ), \frac{ c b \sqrt{ p - 1 } }{ 2 L \sqrt{M} } } \Bigr)^{ \! \nicefrac{1}{p} }
\biggl[ \frac{ 2 ( c + 1 ) b^2 \sqrt{ p - 1 } }{ \sqrt{M} } \biggr]
.
\end{split}
\end{equation}
  This
and \cref{lem:abstract_generalisation_error.it:1}
\prove[ep] 
\cref{item:lem:abstract_generalisation_error:1,item:lem:abstract_generalisation_error:2}.
\end{aproof}

\cfclear
\begin{athm}{prop}{prop:generalisation_error}
Let
$ d \in\N $,
$ D \subseteq \R^d $,
let
$ ( \Omega, \cF, \P ) $
be a probability space,
let
$M \in \N$,
let
$ \mathbb X_m=(X_m,Y_m) \colon \Omega \to (D \times\R) $,
$ m \in \{ 1, 2, \ldots, M \} $,
be i.i.d.\ random variables,
let
$ \alpha \in \R $,
$ \beta \in ( \alpha, \infty ) $,
$\bfd\in\N$,
let
$ f = ( f_\theta )_{ \theta \in [ \alpha, \beta ]^\bfd } \colon
[ \alpha, \beta ]^\bfd \to C( D, \R ) $,
let
$ L, b \in ( 0, \infty ) $
satisfy for all
$ \theta, \vartheta \in [ \alpha, \beta ]^\bfd $,
$ m \in \{ 1, 2, \ldots, M \} $,
$ x \in D $
that
\begin{equation}
  \llabel{eq:Lip}
   \lvert f_\theta( X_m ) - Y_m \rvert \leq b 
  \qquad\text{and}\qquad
   \lvert f_\theta( x ) - f_\vartheta( x ) \rvert
   \leq
   L \infnorm{ \theta - \vartheta } ,
  \end{equation}
and let
$ \bfR \colon [ \alpha, \beta ]^\bfd \to [ 0, \infty ) $
and
$ \emprisk \colon [ \alpha, \beta ]^\bfd \times \Omega \to [ 0, \infty ) $
satisfy for all
$ \theta \in [ \alpha, \beta ]^\bfd $,
$ \omega \in \Omega $
that
\begin{equation}
  \bfR( \theta )
  = \E\bbr{ \lvert f_\theta( X_1 ) - Y_1 \rvert^2 }
  \qquad\text{and}\qquad
  \emprisk( \theta, \omega )
=
\frac{1}{M}
\biggl[
\smallsum_{m=1}^M
    \lvert f_\theta( X_m( \omega ) ) - Y_m( \omega ) \rvert^2
\biggr]
\end{equation}
\cfload.
Then
\begin{enumerate}[label=(\roman *)]
\item
\label{item:prop:generalisation_error:1}
it holds that
$ \Omega \ni \omega \mapsto
\sup\nolimits_{ \theta \in [ \alpha, \beta ]^\bfd } \lvert \emprisk( \theta, \omega ) - \bfR( \theta ) \rvert
\in [ 0, \infty ] $
is $ \cF $/$ \cB( [ 0, \infty ] ) $-measurable
and
\item
\label{item:prop:generalisation_error:2}
it holds for all
$ p \in ( 0, \infty ) $
that
\begin{equation}
\begin{split}
&
\bigl(
\E\bigl[
    \sup\nolimits_{ \theta \in [ \alpha, \beta ]^\bfd }
    \lvert \emprisk( \theta ) - \bfR( \theta ) \rvert^p
\bigr]
\bigr)^{ \nicefrac{1}{p} }
\\ &
\leq
\inf_{ c, \varepsilon \in ( 0, \infty ) }
\Biggl[
\frac{ 2 ( c + 1 ) b^2
    \max\{
    1,
    [ 2 \sqrt{M} L ( \beta - \alpha ) ( c b )^{ -1 } ]^{ \varepsilon }
    \}
\sqrt{ \max\{ 1, p, \nicefrac{\bfd}{ \varepsilon } \} }
}{ \sqrt{M} }
\Biggr]
\\ &
\leq
\inf_{ c \in ( 0, \infty ) }
\Biggl[
\frac{ 2 ( c + 1 ) b^2
\sqrt{ e \max\{ 1, p, \bfd \ln( 4 M L^2 ( \beta - \alpha )^2 ( c b )^{ -2 } ) \} }
}{ \sqrt{M} }
\Biggr]
.
\end{split}
\end{equation}
\end{enumerate}
\end{athm}
\begin{aproof}
Throughout this proof,
let
$ ( \kappa_c )_{ c \in ( 0, \infty ) } \subseteq ( 0, \infty ) $
satisfy for all
$ c \in ( 0, \infty ) $
that
\begin{equation}
  \kappa_c=\frac{ 2 \sqrt{M} L ( \beta - \alpha ) }{ c b },
\end{equation}
let
$ \cX_{ \theta, m } \colon \Omega \to \R $,
$ m \in \{ 1, 2, \ldots, M \} $,
$ \theta \in [ \alpha, \beta ]^\bfd $,
satisfy for all
$ \theta \in [ \alpha, \beta ]^\bfd $,
$ m \in \{ 1, 2, \ldots, M \} $
that
\begin{equation}
    \cX_{ \theta, m } = f_\theta( X_m )  
    ,
\end{equation}
and
let
$ \delta \colon [ \alpha, \beta ]^\bfd \times [ \alpha, \beta ]^\bfd \to [ 0, \infty ) $
satisfy for all
$ \theta, \vartheta \in [ \alpha, \beta ]^\bfd $
that
\begin{equation}
    \delta( \theta, \vartheta ) = \infnorm{ \theta - \vartheta }
    .  
\end{equation}
First,
\nobs that
the assumption that
for all
$ \theta \in [ \alpha, \beta ]^\bfd$,
$ m \in \{ 1, 2, \ldots, M \} $
it holds that 
$\lvert f_\theta( X_m ) - Y_m \rvert \leq b $
\proves
for all
$ \theta \in [ \alpha, \beta ]^\bfd $,
$ m \in \{ 1, 2, \ldots, M \} $
that
\begin{equation}
\label{eq:difference_bounded}
\lvert \cX_{ \theta, m } - Y_m \rvert
=
\lvert f_\theta( X_m ) - Y_m \rvert
\leq b
.
\end{equation}
\Moreover
the assumption that
for all
$ \theta, \vartheta \in [ \alpha, \beta ]^\bfd$,
$x \in D$ it holds that 
$\lvert f_\theta( x ) - f_\vartheta( x ) \rvert
\leq
L \infnorm{ \theta - \vartheta } $
ensures
for all
$ \theta, \vartheta \in [ \alpha, \beta ]^\bfd $,
$ m \in \{ 1, 2, \ldots, M \} $
that
\begin{equation}
\lvert \cX_{ \theta, m } - \cX_{ \vartheta, m } \rvert
=
\lvert f_\theta( X_m ) - f_\vartheta( X_m ) \rvert
\leq
\sup\nolimits_{ x \in D }
    \lvert f_\theta( x ) - f_\vartheta( x ) \rvert
\leq
L \infnorm{ \theta - \vartheta }
=
L \delta( \theta, \vartheta )
.
\end{equation}
The fact that
for all
$ \theta \in [ \alpha, \beta ]^\bfd $
it holds that
$ ( \cX_{ \theta, m }, Y_m ) $,
$ m \in \{ 1, 2, \ldots, M \} $,
are i.i.d.\ random variables,
\cref{eq:difference_bounded},
and
\cref{lem:abstract_generalisation_error}
(applied with
$ p \is q $,
$ C \is C $,
$ ( E, \delta ) \is ( [ \alpha, \beta ]^\bfd, \delta ) $,
$ ( X_{ x, m } )_{ x \in E, \, m \in \{ 1, 2, \ldots, M \} }
\is
( \cX_{ \theta, m } )_{ \theta \in [ \alpha, \beta ]^\bfd, \, m \in \{ 1, 2, \ldots, M \} } $
for
$ p \in [ 2, \infty ) $,
$ C \in ( 0, \infty ) $
in the notation of \cref{lem:abstract_generalisation_error})
\hence \prove that for all
$ p \in [ 2, \infty ) $,
$ c \in ( 0, \infty ) $
it holds that
$ \Omega \ni \omega \mapsto
\sup\nolimits_{ \theta \in [ \alpha, \beta ]^\bfd } \lvert \emprisk( \theta, \omega ) - \bfR( \theta ) \rvert
\in [ 0, \infty ] $
is $ \cF $/$ \cB( [ 0, \infty ] ) $-measurable
and%
\begin{equation}
\label{eq:abstract_generalisation_error}
\bigl(
\E\bigl[
    \sup\nolimits_{ \theta \in [ \alpha, \beta ]^\bfd }
    \lvert \emprisk( \theta ) - \bfR( \theta ) \rvert^p
\bigr]
\bigr)^{ \nicefrac{1}{p} }
\leq
\Bigl(
    \CovNum{ ( [ \alpha, \beta ]^\bfd, \delta ), \frac{ c b \sqrt{ p - 1 } }{ 2 L \sqrt{M} } }
\Bigr)^{ \! \nicefrac{1}{p} }
\biggl[ \frac{ 2 ( c + 1 ) b^2 \sqrt{ p - 1 } }{ \sqrt{M} } \biggr]
\end{equation}%
\cfload.
This \proves[ep] \cref{item:prop:generalisation_error:1}.
\Nobs that
\cref{lem:covering_number_cube_infty}
(applied with
$ d \is \bfd $,
$ a \is \alpha $,
$ b \is \beta $,
$ r \is r $
for
$ r \in ( 0, \infty ) $
in the notation of \cref{lem:covering_number_cube_infty})
\proves that for all
$ r \in ( 0, \infty ) $
it holds that
\begin{equation}
\begin{split}
\CovNum{ ( [ \alpha, \beta ]^\bfd, \delta ), r }
& \leq
\ind{ [ 0, r ] }\bigl( \tfrac{ \beta - \alpha }{2} \bigr)
+
\bigl( \tfrac{ \beta - \alpha }{r} \bigr)^\bfd
\ind{ ( r, \infty ) }\bigl( \tfrac{ \beta - \alpha }{2} \bigr)
\\ &
\leq
\max\Bigl\{
    1,
    \bigl( \tfrac{ \beta - \alpha }{r} \bigr)^\bfd
\Bigr\}
\bigl(
\ind{ [ 0, r ] }\bigl( \tfrac{ \beta - \alpha }{2} \bigr)
+
\ind{ ( r, \infty ) }\bigl( \tfrac{ \beta - \alpha }{2} \bigr)
\bigr)
\\ &
=
\max\Bigl\{
    1,
    \bigl( \tfrac{ \beta - \alpha }{r} \bigr)^\bfd
\Bigr\}
.
\end{split}
\end{equation}
\Hence for all
$ c \in ( 0, \infty ) $,
$ p \in [ 2, \infty ) $
that
\begin{equation}
\begin{split}
\Bigl(
\CovNum{ ( [ \alpha, \beta ]^\bfd, \delta ), \frac{ c b \sqrt{ p - 1 } }{ 2 L \sqrt{M} } }
\Bigr)^{ \! \nicefrac{1}{ p } }
&
\leq
\max\biggl\{
    1,
    \Bigl( \tfrac{ 2 ( \beta - \alpha ) L \sqrt{M} }{ c b \sqrt{ p - 1 } } \Bigr)^{ \! \frac{\bfd}{p} }
\biggr\}
\\ &
\leq
\max\biggl\{
    1,
    \Bigl( \tfrac{ 2 ( \beta - \alpha ) L \sqrt{M} }{ c b } \Bigr)^{ \! \frac{\bfd}{p} }
\biggr\}
=
\max\Bigl\{
    1,
    ( \kappa_c )^{ \frac{\bfd}{p} }
\Bigr\}
.
\end{split}
\end{equation}
This,
\cref{eq:abstract_generalisation_error},
and
Jensen's inequality
\prove that for all
$ c, \varepsilon, p \in ( 0, \infty ) $
it holds that
\begin{equation}
\label{eq:generalisation_error_estimate}
\begin{split}
&
\bigl(
\E\bigl[
    \sup\nolimits_{ \theta \in [ \alpha, \beta ]^\bfd }
    \lvert \emprisk( \theta ) - \bfR( \theta ) \rvert^p
\bigr]
\bigr)^{ \nicefrac{1}{p} }
\\ &
\leq
\bigl(
\E\bigl[
    \sup\nolimits_{ \theta \in [ \alpha, \beta ]^\bfd }
    \lvert \emprisk( \theta ) - \bfR( \theta ) \rvert^{ \max\{ 2, p, \nicefrac{\bfd}{ \varepsilon } \} }
\bigr]
\bigr)^{ \frac{1}{ \max\{ 2, p, \nicefrac{\bfd}{ \varepsilon } \} } }
\\ &
\leq
\max\Bigl\{
    1,
    ( \kappa_c )^{ \frac{\bfd}{ \max\{ 2, p, \nicefrac{\bfd}{ \varepsilon } \} } }
\Bigr\}
\frac{ 2 ( c + 1 ) b^2 \sqrt{ \max\{ 2, p, \nicefrac{\bfd}{ \varepsilon } \} - 1 } }{ \sqrt{M} }
\\ &
=
\max\bigl\{
    1,
    ( \kappa_c )^{ \min\{ \nicefrac{\bfd}{2}, \nicefrac{\bfd}{p}, \varepsilon \} }
\bigr\}
\frac{ 2 ( c + 1 ) b^2 \sqrt{ \max\{ 1, p - 1, \nicefrac{\bfd}{ \varepsilon } - 1 \} } }{ \sqrt{M} }
\\ &
\leq
\frac{
2 ( c + 1 ) b^2
\max\{ 1, ( \kappa_c )^{ \varepsilon } \}
\sqrt{ \max\{ 1, p, \nicefrac{\bfd}{ \varepsilon } \} }
}{ \sqrt{M} }
.
\end{split}
\end{equation}
\Moreover
the fact that for all
$ a \in ( 1, \infty )$
it holds that
\begin{equation}
    a^{ \nicefrac{1}{ ( 2 \ln( a ) ) }}
= e^{ \nicefrac{ \ln( a ) }{ ( 2 \ln( a ) ) } }
= e^{ \nicefrac{1}{2} }
= \sqrt{e}
\geq 1 
\end{equation}
\proves that for all
$ c,p \in ( 0, \infty ) $
with $ \kappa_c > 1 $
it holds that
\begin{equation}
\label{eq:generalisation_error_epsilon_choice}
\begin{split}
&
\inf_{ \varepsilon \in ( 0, \infty ) }
\Biggl[
\frac{ 2 ( c + 1 ) b^2
    \max\{
    1,
    ( \kappa_c )^{ \varepsilon }
    \}
\sqrt{ \max\{ 1, p, \nicefrac{\bfd}{ \varepsilon } \} }
}{ \sqrt{M} }
\Biggr]
\\ &
\leq
\frac{ 2 ( c + 1 ) b^2
    \max\{
    1,
    ( \kappa_c )^{ \nicefrac{1}{ ( 2 \ln( \kappa_c ) ) } }
    \}
\sqrt{ \max\{ 1, p, 2 \bfd \ln( \kappa_c ) \} }
}{ \sqrt{M} }
\\ &
=
\frac{ 2 ( c + 1 ) b^2
\sqrt{ e \max\{ 1, p, \bfd \ln( [ \kappa_c ]^2 ) \} }
}{ \sqrt{M} }
.
\end{split}
\end{equation}
The fact that
for all
$ c, p \in ( 0, \infty ) $
with $ \kappa_c \leq 1 $
it holds that
\begin{equation}
\label{eq:generalisation_error_epsilon_auxiliary}
\begin{split}
&
\inf_{ \varepsilon \in ( 0, \infty ) }
\Biggl[
\frac{ 2 ( c + 1 ) b^2
    \max\{
    1,
    ( \kappa_c )^{ \varepsilon }
    \}
\sqrt{ \max\{ 1, p, \nicefrac{\bfd}{ \varepsilon } \} }
}{ \sqrt{M} }
\Biggr]
\\ &
=
\inf_{ \varepsilon \in ( 0, \infty ) }
\Biggl[
\frac{ 2 ( c + 1 ) b^2
\sqrt{ \max\{ 1, p, \nicefrac{\bfd}{ \varepsilon } \} }
}{ \sqrt{M} }
\Biggr]
\leq
\frac{ 2 ( c + 1 ) b^2
\sqrt{ \max\{ 1, p \} }
}{ \sqrt{M} }
\\ &
\leq
\frac{ 2 ( c + 1 ) b^2
\sqrt{ e \max\{ 1, p, \bfd \ln( [ \kappa_c ]^2 ) \} }
}{ \sqrt{M} }
.
\end{split}
\end{equation}
and
\cref{eq:generalisation_error_estimate}
\hence \prove that for all
  $p\in(0,\infty)$
it holds that
\begin{equation}
\begin{split}
&
\bigl(
\E\bigl[
    \sup\nolimits_{ \theta \in [ \alpha, \beta ]^\bfd }
    \lvert \emprisk( \theta ) - \bfR( \theta ) \rvert^p
\bigr]
\bigr)^{ \nicefrac{1}{p} }
\\ &
\leq
\inf_{ c, \varepsilon \in ( 0, \infty ) }
\Biggl[
\frac{ 2 ( c + 1 ) b^2
    \max\{
    1,
    ( \kappa_c )^{ \varepsilon }
    \}
\sqrt{ \max\{ 1, p, \nicefrac{\bfd}{ \varepsilon } \} }
}{ \sqrt{M} }
\Biggr]
\\ &
=
\inf_{ c, \varepsilon \in ( 0, \infty ) }
\Biggl[
\frac{ 2 ( c + 1 ) b^2
    \max\{
    1,
    [ 2 \sqrt{M} L ( \beta - \alpha ) ( c b )^{ -1 } ]^{ \varepsilon }
    \}
\sqrt{ \max\{ 1, p, \nicefrac{\bfd}{ \varepsilon } \} }
}{ \sqrt{M} }
\Biggr]
\\ &
\leq
\inf_{ c \in ( 0, \infty ) }
\Biggl[
\frac{ 2 ( c + 1 ) b^2
\sqrt{ e \max\{ 1, p, \bfd \ln( [ \kappa_c ]^2 ) \} }
}{ \sqrt{M} }
\Biggr]
\\ &
=
\inf_{ c \in ( 0, \infty ) }
\Biggl[
\frac{ 2 ( c + 1 ) b^2
\sqrt{ e \max\{ 1, p, \bfd \ln( 4 M L^2 ( \beta - \alpha )^2 ( c b )^{ -2 } ) \} }
}{ \sqrt{M} }
\Biggr]
.
\end{split}
\end{equation}
This \proves[ep] \cref{item:prop:generalisation_error:2}.
\end{aproof}

\cfclear
\begin{athm}{cor}{cor:generalisation_error}
Let
$ d, M \in \N $,
$ b \in [ 1, \infty ) $,
$ u \in \R $,
$ v \in [ u + 1, \infty ) $,
$ D \subseteq [ -b, b ]^d $,
let
$ ( \Omega, \cF, \P ) $
be a probability space,
let
$ \mathbb X_m = (X_m,Y_m) \colon \Omega \to (D \times [u,v])$,
$ m \in \{ 1, 2, \ldots, M \} $,
be i.i.d.\ random variables,
let
$ B \in [ 1, \infty ) $,
$ \bfL, \bfd \in\N$,
$ \bfl = ( \bfl_0, \bfl_1, \ldots, \bfl_\bfL ) \in \N^{ \bfL + 1 } $
satisfy
$ \bfl_0 = d $,
$ \bfl_\bfL = 1 $,
and
$ \bfd \geq \sum_{i=1}^{\bfL} \bfl_i( \bfl_{ i - 1 } + 1 ) $,
let
$ \bfR \colon [ -B, B ]^\bfd \to [ 0, \infty ) $
and
$ \emprisk \colon [ -B, B ]^\bfd \times \Omega \to [ 0, \infty ) $
satisfy for all
$ \theta \in [ -B, B ]^\bfd $,
$ \omega \in \Omega $
that
\begin{equation}
 \bfR( \theta )
  = \E\bbr{ \lvert \clippedNN{\theta}{\bfl}{u}{v}( X_1 ) - Y_1 \rvert^2 }
  \quad\text{and}\quad
  \emprisk( \theta, \omega )
=
\frac{1}{M}
\biggl[
\smallsum_{m=1}^M
    \lvert \clippedNN{\theta}{\bfl}{u}{v}( X_m( \omega ) ) - Y_m( \omega ) \rvert^2
\biggr]
\end{equation}
\cfload.
Then
\begin{enumerate}[label=(\roman *)]
\item
\label{item:cor:generalisation_error:1}
it holds that
$ \Omega \ni \omega \mapsto
\sup\nolimits_{ \theta \in [ -B, B ]^\bfd } \lvert \emprisk( \theta, \omega ) - \bfR( \theta ) \rvert
\in [ 0, \infty ] $
is $ \cF $/\allowbreak$ \cB( [ 0, \infty ] ) $-mea\-sur\-able
and
\item
\label{item:cor:generalisation_error:2}
it holds for all
$ p \in ( 0, \infty ) $
that
\begin{equation}
\begin{split}
& \bigl(
\E\bigl[
    \sup\nolimits_{ \theta \in [ -B, B ]^\bfd }
    \lvert \emprisk( \theta ) - \bfR( \theta ) \rvert^p
\bigr]
\bigr)^{ \nicefrac{1}{p} }
\\ &
\leq
\frac{
    9 ( v - u )^2
    \bfL ( \infnorm{ \bfl } + 1 )
    \sqrt{
        \max\{
            p,
            \ln( 4 ( M b )^{ \nicefrac{1}{ \bfL } } ( \infnorm{ \bfl } + 1 ) B )
        \}
    }
}{ \sqrt{M} }
\\ &
\leq
\frac{
    9 ( v - u )^2
    \bfL ( \infnorm{ \bfl } + 1 )^2
    \max\{
        p,
        \ln( 3 M B b )
    \}
}{ \sqrt{M} }
\end{split}
\end{equation}
\end{enumerate}
\cfout.
\end{athm}
\begin{aproof}
Throughout this proof,
let
$ \fd = \sum_{i=1}^{\bfL} \bfl_i( \bfl_{ i - 1 } + 1 ) \in\N$,
let
$ L =
b \bfL ( \infnorm{ \bfl  } + 1 )^\bfL B^{ \bfL - 1 } \in ( 0, \infty )$,
for every
$\theta \in [ -B, B ]^\fd$
let
$ f_\theta\colon D\to\R $
satisfy for all
$ x \in D $
that
\begin{equation}
    f_\theta( x ) = \clippedNN{\theta}{\bfl}{u}{v}( x )
    ,  
\end{equation}
let
$ \scrR \colon [ -B, B ]^\fd \to [ 0, \infty ) $
satisfy for all
$ \theta \in [ -B, B ]^\fd $
that
\begin{equation}
  \scrR( \theta )
= \E\bbr{ \lvert f_\theta( X_1 ) - Y_1 \rvert^2 }
= \E\bbr{ \lvert \clippedNN{\theta}{\bfl}{u}{v}( X_1 ) - Y_1 \rvert^2 }
,
\end{equation}
and
let
$ R \colon [ -B, B ]^\fd \times \Omega \to [ 0, \infty ) $
satisfy for all
$ \theta \in [ -B, B ]^\fd $,
$ \omega \in \Omega $
that
\begin{equation}
R( \theta, \omega )
=
\frac{1}{M}
\biggl[
\smallsum_{m=1}^M
    \lvert f_\theta( X_m( \omega ) ) - Y_m( \omega ) \rvert^2
\biggr]
=
\frac{1}{M}
\biggl[
\smallsum_{m=1}^M
    \lvert \clippedNN{\theta}{\bfl}{u}{v}( X_m( \omega ) ) - Y_m( \omega ) \rvert^2
\biggr]
\end{equation}
\cfload.
\Nobs that
the fact that
for all
$ \theta \in \R^\fd$,
$x \in \R^d$ it holds that
$\clippedNN{\theta}{\bfl}{u}{v}( x ) \in [ u, v ] $
and the assumption that
for all
$  m \in \{ 1, 2, \ldots, M \}$ it holds that
$Y_m( \Omega ) \subseteq [ u, v ] $
\prove
for all
$ \theta \in [ -B, B ]^\fd $,
$ m \in \{ 1, 2, \ldots, M \} $
that
\begin{equation}
\label{eq:difference_bounded_NN}
\lvert f_\theta( X_m ) - Y_m \rvert
=
\lvert \clippedNN{\theta}{\bfl}{u}{v}( X_m ) - Y_m \rvert
\leq
\sup\nolimits_{ y_1, y_2 \in [ u, v ] }
    \lvert y_1 - y_2 \rvert
=
v - u
.
\end{equation}
\Moreover
the assumption that
$ D \subseteq [ -b, b ]^d $,
$ \bfl_0 = d $,
and
$ \bfl_\bfL = 1 $,
\cref{cor:ClippedRealNNLipsch}
(applied with
$ a \is -b $,
$ b \is b $,
$ u \is u $,
$ v \is v $,
$ d \is \fd $,
$ L \is \bfL $,
$ l \is \bfl $
in the notation of~%
\cref{cor:ClippedRealNNLipsch}),
and
the assumption that
$ b \geq 1 $ and $ B \geq 1 $
\prove that for all
$ \theta, \vartheta \in [ -B, B ]^\fd $,
$ x \in D $
it holds that
\begin{equation}
\label{eq:Lipschitz_NN}
\begin{split}
\lvert f_\theta( x ) - f_\vartheta( x ) \rvert
& \leq
\sup\nolimits_{ y \in [ -b, b ]^d }
    \lvert \clippedNN{\theta}{\bfl}{u}{v}( y ) - \clippedNN{\vartheta}{\bfl}{u}{v}( y ) \rvert
\\ &
\leq
\bfL
\max\{ 1, b \}
( \infnorm{ \bfl } + 1 )^\bfL
( \max\{ 1, \infnorm{ \theta }, \infnorm{ \vartheta } \} )^{ \bfL - 1 }
\infnorm{ \theta - \vartheta }
\\ &
\leq
b
\bfL
( \infnorm{ \bfl } + 1 )^\bfL
B^{ \bfL - 1 }
\infnorm{ \theta - \vartheta }
=
L \infnorm{ \theta - \vartheta }
.
\end{split}
\end{equation}
\Moreover
the fact that
$ \bfd \geq \fd $
and the fact that for all
$ \theta = ( \theta_1, \theta_2, \ldots, \theta_\bfd ) \in \R^\bfd $
it holds that
$\clippedNN{\theta}{\bfl}{u}{v}
=
\clippedNN{ \smash{ (\theta_1, \theta_2, \ldots, \theta_\fd ) } }{\bfl}{u}{v} $
\proves that for all
$ \omega \in \Omega $
it holds that
\begin{equation}
\label{eq:independent_parameters}
\sup\nolimits_{ \theta \in [ -B, B ]^\bfd }
    \lvert \emprisk( \theta, \omega ) - \bfR( \theta ) \rvert
=
\sup\nolimits_{ \theta \in [ -B, B ]^\fd }
    \lvert R( \theta, \omega ) - \scrR( \theta ) \rvert
.
\end{equation}
\Moreover
\cref{eq:difference_bounded_NN},
\cref{eq:Lipschitz_NN},
\cref{prop:generalisation_error}
(applied with
$ \alpha \is -B $,
$ \beta \is B $,
$ \bfd \is \fd $,
$ b \is v - u $,
$ \bfR \is \scrR $,
$ \emprisk \is R $
in the notation of \cref{prop:generalisation_error}),
the fact that
\begin{equation}
  v - u \geq ( u + 1 ) - u = 1
\end{equation}
and the fact that
\begin{equation}
  \fd \leq
\bfL \infnorm{ \bfl }( \infnorm{ \bfl } + 1 )
\leq
\bfL ( \infnorm{ \bfl } + 1 )^2
\end{equation}
\prove
that for all
$ p \in ( 0, \infty ) $
it holds that
$ \Omega \ni \omega \mapsto
\sup\nolimits_{ \theta \in [ -B, B ]^\fd } \lvert R( \theta, \omega ) - \scrR( \theta ) \rvert
\in [ 0, \infty ] $
is $ \cF $/$ \cB( [ 0, \infty ] ) $-measurable
and
\begin{equation}
\label{eq:prop:generalisation_error}
\begin{split}
&
\bigl(
\E\bigl[
    \sup\nolimits_{ \theta \in [ -B, B ]^\fd }
    \lvert R( \theta ) - \scrR( \theta ) \rvert^p
\bigr]
\bigr)^{ \nicefrac{1}{p} }
\\ &
\leq
\inf_{ C \in ( 0, \infty ) }
\Biggl[
\frac{ 2 ( C + 1 ) ( v - u )^2
\sqrt{ e \max\{ 1, p, \fd \ln( 4 M L^2 ( 2 B )^2 ( C [ v - u ] )^{ -2 } ) \} }
}{ \sqrt{M} }
\Biggr]
\\ &
\leq
\inf_{ C \in ( 0, \infty ) }
\Biggl[
\frac{ 2 ( C + 1 ) ( v - u )^2
\sqrt{ e \max\{ 1, p, \bfL ( \infnorm{ \bfl } + 1 )^2 \ln( 2^4 M L^2 B^2 C^{ -2 } ) \} }
}{ \sqrt{M} }
\Biggr]
.
\end{split}
\end{equation}
Combining this
with
 \cref{eq:independent_parameters}
\proves[ep] \cref{item:cor:generalisation_error:1}.
\Nobs that
\cref{eq:independent_parameters},
\cref{eq:prop:generalisation_error},
the fact that
$ 2^6 \bfL^2
\leq 2^6 \cdot 2^{ 2 ( \bfL - 1 ) }
= 2^{ 4 + 2 \bfL }
\leq 2^{ 4 \bfL + 2 \bfL }
= 2^{ 6 \bfL } $,
the fact that
$ 3 \geq e $,
and the assumption that
$ B \geq 1 $,
$ \bfL \geq 1 $,
$ M \geq 1 $,
and
$ b \geq 1 $
\prove that for all
$ p \in ( 0, \infty ) $
it holds that
\begin{equation}
\label{eq:generalisation_error_NN1}
\begin{split}
& \bigl(
\E\bigl[
    \sup\nolimits_{ \theta \in [ -B, B ]^\bfd }
    \lvert \emprisk( \theta ) - \bfR( \theta ) \rvert^p
\bigr]
\bigr)^{ \nicefrac{1}{p} }
=
\bigl(
\E\bigl[
    \sup\nolimits_{ \theta \in [ -B, B ]^\fd }
    \lvert R( \theta ) - \scrR( \theta ) \rvert^p
\bigr]
\bigr)^{ \nicefrac{1}{p} }
\\ &
\leq
\frac{ 2 ( \nicefrac{1}{2} + 1 ) ( v - u )^2
\sqrt{ e \max\{ 1, p, \bfL ( \infnorm{ \bfl } + 1 )^2 \ln( 2^4 M L^2 B^2 2^{ 2 } ) \} }
}{ \sqrt{M} }
\\ &
=
\frac{ 3 ( v - u )^2
\sqrt{
    e
    \max\{
        p,
        \bfL ( \infnorm{ \bfl } + 1 )^2
        \ln( 2^6 M b^2 \bfL^2 ( \infnorm{ \bfl } + 1 )^{ 2 \bfL } B^{ 2 \bfL } )
    \}
    }
}{ \sqrt{M} }
\\ &
\leq
\frac{ 3 ( v - u )^2
\sqrt{
    e
    \max\{
        p,
        3 \bfL^2 ( \infnorm{ \bfl } + 1 )^2
        \ln( [ 2^{ 6 \bfL } M b^2 ( \infnorm{ \bfl } + 1 )^{ 2 \bfL } B^{ 2 \bfL } ]^{ \nicefrac{1}{ ( 3 \bfL ) } } )
    \}
    }
}{ \sqrt{M} }
\\ &
\leq
\frac{ 3 ( v - u )^2
\sqrt{
    3
    \max\{
        p,
        3 \bfL^2 ( \infnorm{ \bfl } + 1 )^2
        \ln( 2^2 ( M b^2 )^{ \nicefrac{1}{ ( 3 \bfL ) } } ( \infnorm{ \bfl } + 1 ) B )
    \}
    }
}{ \sqrt{M} }
\\ &
\leq
\frac{ 9 ( v - u )^2
\bfL ( \infnorm{ \bfl } + 1 )
\sqrt{
    \max\{
        p,
        \ln( 4 ( M b )^{ \nicefrac{1}{ \bfL } } ( \infnorm{ \bfl } + 1 ) B )
    \}
    }
}{ \sqrt{M} }
.
\end{split}
\end{equation}
\Moreover
the fact that
for all
$  n \in \N $
it holds that 
$n \leq 2^{ n - 1 } $
and
the fact that
$ \infnorm{ \bfl } \geq 1 $
\prove that
\begin{equation}
4 ( \infnorm{ \bfl } + 1 )
\leq 2^2 \cdot 2^{ ( \infnorm{ \bfl } + 1 ) - 1 }
= 2^3 \cdot 2^{ ( \infnorm{ \bfl } + 1 ) - 2 }
\leq 3^2 \cdot 3^{ ( \infnorm{ \bfl } + 1 ) - 2 }
= 3^{ ( \infnorm{ \bfl } + 1 ) }
.
\end{equation}
\Hence that for all
$ p \in ( 0, \infty ) $
it holds that
\begin{equation}
\begin{split}
&
\frac{ 9 ( v - u )^2
\bfL ( \infnorm{ \bfl } + 1 )
\sqrt{
    \max\{
        p,
        \ln( 4 ( M b )^{ \nicefrac{1}{ \bfL } } ( \infnorm{ \bfl } + 1 ) B )
    \}
    }
}{ \sqrt{M} }
\\ &
\leq
\frac{ 9 ( v - u )^2
\bfL ( \infnorm{ \bfl } + 1 )
\sqrt{
    \max\{
        p,
        ( \infnorm{ \bfl } + 1 )
        \ln( [ 3^{ ( \infnorm{ \bfl } + 1 ) } ( M b )^{ \nicefrac{1}{ \bfL } } B ]^{ \nicefrac{1}{ ( \infnorm{ \bfl } + 1 ) } } )
    \}
    }
}{ \sqrt{M} }
\\ &
\leq
\frac{
    9 ( v - u )^2
    \bfL ( \infnorm{ \bfl } + 1 )^2
    \max\{
        p,
        \ln( 3 M B b )
    \}
}{ \sqrt{M} }
.
\end{split}
\end{equation}
This and \cref{eq:generalisation_error_NN1}
\prove[ep] \cref{item:cor:generalisation_error:2}.
\end{aproof}

%% file: parts/Overall_error_analysis.tex
\cchapter{Overall error decomposition}{sect:overall_error_decomp}

In \cref{sec:composed_error} below we combine
parts of the approximation error estimates from \cref{part:approx},
parts of the optimization error estimates from \cref{part:opt}, and 
parts of the generalization error estimates from \cref{part:generalization}
to establish estimates for the overall error in the training of \anns\ in the specific situation of \GD-type optimization methods with many independent random initializations.
For such a combined error analysis we employ a suitable overall error decomposition for supervised learning problems.
It is the subject of this chapter to review and derive this overall error decomposition (see \cref{prop:error_decomposition2} below).

In the literature such kind of error decompositions can, \eg, be found in \cite{CuckerSmale2002,berner_grohs_kutyniok_petersen_2022,JentzenWelti2023,Berner2020,Beck2019published}.
The specific presentation of this chapter is strongly based on \cite[Section 4.1]{Beck2019published} and \cite[Section 6.1]{JentzenWelti2023}.

\section{Bias-variance decomposition}

\newcommand{\functionRisk}{\mathbf{r}}

\begin{athm}{lemma}{lem:decomposition}[Bias-variance decomposition]
Let $ ( \Omega, \mathcal{F}, \P ) $ be a probability space, 
let $ ( S, \mathcal{S} ) $ be a measurable space, 
let $ X \colon \Omega \to S $
and $ Y \colon \Omega \to \R $
be random variables with $ \sExp{ \abs{ Y }^2 } < \infty $, 
and 
let $ \functionRisk \colon \mathcal{L}^2( \P_X ; \R ) \to [0,\infty) $ 
satisfy 
for all $ f \in \mathcal{L}^2( \P_X ; \R ) $ that
\begin{equation}
  \llabel{eq:defrisk}
  \functionRisk( f )
  =
  \Exp{ 
    \abs{ f( X ) - Y }^2
    }
    . 
  \end{equation}
Then 
\begin{enumerate}[label=(\roman{*})]
\item 
\label{item:dec_i}
it holds for all $ f \in \mathcal{L}^2( \P_X ; \R ) $ that
\begin{equation}
  \functionRisk( f ) 
  =
  \Exp{
    \abs*{
      f( X )
      -
      \Exp{ Y \cond X }
    }^2
  }
  +
  \E\bigl[ 
    \abs*{
      Y
      -
      \Exp{ Y \cond X }
    }^2
  \bigr]
  ,
\end{equation}
\item 
\label{item:dec_ii}
it holds for all $ f, g \in \mathcal{L}^2( \P_X ; \R ) $ that
\begin{equation}
  \functionRisk( f ) - \functionRisk( g ) 
  =
  \Exp{ 
    \abs*{ f( X ) - \sExp{ Y \cond X } }^2
  }
  -
  \E\bigl[ 
    \abs*{ g( X ) - \sExp{ Y \cond X } }^2
  \bigr]
  ,
\end{equation}
and 
\item 
\label{item:dec_iii}
it holds for all $ f, g \in \mathcal{L}^2( \P_X ; \R ) $ that
\begin{equation}
  \Exp{ 
    \abs*{ f( X ) - \sExp{ Y \cond X } }^2
  }
=
  \E\bigl[ 
    \abs*{ g( X ) - \sExp{ Y \cond X } }^2
  \bigr]
  +
  \bigl(
    \functionRisk( f ) - \functionRisk( g ) 
  \bigr)
  .
\end{equation}
\end{enumerate}
\end{athm}
\begin{aproof}
First, \nobs that
\lref{eq:defrisk}
\proves that for all $ f \in \mathcal{L}^2( \P_X ; \R ) $ 
it holds that
\begin{equation}
\label{eq:lem_dec1}
\begin{split}
  \functionRisk( f ) 
& =
  \Exp{ 
    \abs*{ f( X ) - Y }^2
  }
  =
  \Exp{ 
    \abs*{ ( f( X ) - \sExp{ Y \cond X } ) + ( \sExp{ Y \cond X } - Y ) }^2
  }
\\ &  
=
  \Exp{ 
    \abs*{ f( X ) - \sExp{ Y \cond X } }^2
  }
  +
  2 
  \,
  \bExp{ 
    \bpr{ f( X ) - \sExp{ Y \cond X } } 
    \bpr{ \sExp{ Y \cond X } - Y } 
  }
  \\&\qquad+
  \Exp{
    \abs*{ \sExp{ Y \cond X } - Y }^2
  }
\end{split}
\end{equation}
\Moreover
the tower rule 
\proves that
for all $ f \in \mathcal{L}^2( \P_X ; \R ) $ 
it holds that
\begin{eqsplit}
  &\bExp{ 
    \bpr{ f( X ) - \sExp{ Y \cond X } } 
    \bpr{ \sExp{ Y \cond X } - Y } 
  }
  \\&=
  \E\bbbr{ 
    \bExp{
      \bpr{ f( X ) - \sExp{ Y \cond X } } 
      \bpr{ \sExp{ Y \cond X } - Y } 
      \cond
      X
    }
  }
  \\&=
  \E\bbbr{ 
    \bpr{ f( X ) - \sExp{ Y \cond X } } 
    \bExp{
      \bpr{ \sExp{ Y \cond X } - Y } 
      \cond
      X
    }
  }
  \\&=
  \bExp{ 
    \bpr{ f( X ) - \sExp{ Y \cond X } } 
    \bpr{ \sExp{ Y \cond X } - \sExp{ Y \cond X } } 
  }
  =
  0
  .
\end{eqsplit}
Combining
  this
with
  \cref{eq:lem_dec1}
\proves that
for all $ f \in \mathcal{L}^2( \P_X ; \R ) $ 
it holds that
\begin{equation}
  \llabel{eq:it1}
    \functionRisk( f ) 
=
  \Exp{ 
    \abs*{ f( X ) - \sExp{ Y \cond X } }^2
  }
  +
  \E\bigl[
    \abs*{ \sExp{ Y \cond X } - Y }^2
  \bigr]
  .
\end{equation}
This \proves that 
for all $ f, g \in \mathcal{L}^2( \P_X ; \R ) $ 
it holds that
\begin{equation}
\label{eq:lem_dec2}
  \functionRisk( f ) - \functionRisk( g ) 
=
  \Exp{ 
    \abs*{ f( X ) - \sExp{ Y \cond X } }^2
  }
  -
  \E\bigl[ 
    \abs*{ g( X ) - \sExp{ Y \cond X } }^2
  \bigr]
  .
\end{equation}
\Hence that
for all $ f, g \in \mathcal{L}^2( \P_X ; \R ) $ 
it holds that
\begin{equation}
\label{eq:lem_dec3}
  \Exp{ 
    \abs*{ f( X ) - \sExp{ Y \cond X } }^2
  }
=
  \Exp{ 
    \abs*{ g( X ) - \sExp{ Y \cond X } }^2
  }
  +
  \functionRisk( f ) - \functionRisk( g ) 
  .
\end{equation}
Combining this
with 
\lref{eq:it1} and \cref{eq:lem_dec2} \proves[ep] 
\cref{item:dec_i,item:dec_ii,item:dec_iii}.
\end{aproof}

\subsection{Risk minimization for measurable functions}

\begin{athm}{prop}{prop:minrisk}
  Let $(\Omega,\mc F,\P)$ be a probability space,
  let $(S,\mc S)$ be a measurable space,
  let $X\colon \Omega\to S$ and $Y\colon \Omega\to\R$
    be random variables,
  assume $\E[\abs{Y}^2]<\infty$,
  let $\mc E\colon \mc L^2(\P_X;\R )\to[0,\infty)$
	satisfy for all 
	  $f\in\mc L^2(\P_X;\R)$
	that
  \begin{equation}
	  \mc E(f)=\E\bigl[\abs{f(X)-Y}^2\bigr].
  \end{equation}
  Then 
  \begin{equation}
    \label{eq:minrisk.claim}
    \begin{split}
    &\bigl\{f\in\mc L^2(\P_X;\R)\colon \mc E(f)=\inf\nolimits_{g\in \mc L^2(\P_X;\R)}\mc E(g)\bigr\}
    \\&=
    \bigl\{f\in\mc L^2(\P_X;\R)\colon \mc E(f)=\E\bigl[
    \abs*{ \sExp{ Y \cond X } - Y }^2
    \bigr]\bigr\}
    \\&=
    \{f\in\mc L^2(\P_X;\R)\colon \text{$f(X)=\sExp{Y\cond X}$ $\P$-a.s.}\}
    .
    \end{split}
  \end{equation}
\end{athm}
\begin{aproof}
  \Nobs that 
    \cref{lem:decomposition}
  \proves that for all
    $g\in\mc L^2(\P_X;\R)$
  it holds that
  \begin{equation}
    \label{eq:minrisk.dec}
    \mc E(g)
    =
    \Exp{ 
      \abs*{ g( X ) - \sExp{ Y \cond X } }^2
    }
    +
    \E\bigl[
      \abs*{ \sExp{ Y \cond X } - Y }^2
    \bigr]
    .
  \end{equation}
  \Hence that for all
    $g\in\mc L^2(\P_X;\R)$
  it holds that
  \begin{equation}
    \label{eq:minrisk.inf}
    \mc E(g)
    \geq
    \E\bigl[
      \abs*{ \sExp{ Y \cond X } - Y }^2
    \bigr]
    .
  \end{equation}
  \Moreover
    \cref{eq:minrisk.dec}
  \proves that
  \begin{equation}
  \begin{split}
    &
    \bigl\{f\in\mc L^2(\P_X;\R)\colon \mc E(f)=\E\bigl[
      \abs*{ \sExp{ Y \cond X } - Y }^2
    \bigr]\bigr\}
    \\&=
    \bigl\{f\in\mc L^2(\P_X;\R)\colon \Exp{ 
      \abs*{ f( X ) - \sExp{ Y \cond X } }^2
    }=0\bigr\}
    \\&=
    \{f\in\mc L^2(\P_X;\R)\colon \text{$f(X)=\sExp{Y\cond X}$ $\P$-a.s.}\}
    .
  \end{split}
  \end{equation}
  Combining
    this
  with
    \cref{eq:minrisk.inf}
  \proves[ep]
    \cref{eq:minrisk.claim}.
\end{aproof}

\begin{athm}{cor}{cor:riskmin2}
  Let $(\Omega,\mc F,\P)$ be a probability space,
  let $(S,\mc S)$ be a measurable space,
  let $X\colon \Omega\to S$ be a random variable,
  let $\mc M=\{(f\colon S\to \R)\colon \text{$f$ is $\mc S$/$\mc B(\R)$-measurable}\}$,
  let $\varphi\in\mc M$, and
  let $\mc E\colon \mc M\to[0,\infty)$
	satisfy for all 
	  $f\in\mc M$
	that
  \begin{equation}
    \llabel{eq:defE}
	  \mc E(f)=\E\bigl[\abs{f(X)-\varphi(X)}^2\bigr].
  \end{equation}
  Then 
  \begin{equation}
    \begin{split}
    \{f\in\mc M\colon \mc E(f)=\inf\nolimits_{g\in \mc M}\mc E(g)\}
    &=
    \{f\in\mc M\colon \mc E(f)=0\}
    \\&=
    \{f\in\mc M\colon \P(f(X)=\varphi(X))=1\}
    .
    \end{split}
  \end{equation}
\end{athm}
\begin{aproof}
  \Nobs that
    \lref{eq:defE}
  \proves that
    $\mc E(\varphi)=0$.
  \Hence that
  \begin{equation}
    \inf_{g\in\mc M}\mc E(g)=0.
  \end{equation}
  Furthermore, observe that
  \begin{equation}
    \begin{split}
    \{f\in\mc M\colon \mc E(f)=0\}
    &=
    \bigl\{f\in\mc M\colon \E\bigl[\abs{f(X)-\varphi(X)}^2\bigr]=0\bigr\}
    \\&=
    \bigl\{f\in\mc M\colon \P\bigl(\{\omega\in\Omega\colon f(X(\omega))\neq \varphi(X(\omega))\}\bigr)=0\bigr\}
    \\&=
    \bigl\{f\in\mc M\colon \P\bigl(X^{-1}(\{x\in S\colon f(x)\neq \varphi(x)\})\bigr)=0\bigr\}
    \\&=
    \{f\in\mc M\colon \P_X(\{x\in S\colon f(x)\neq \varphi(x)\})=0\}
    .
    \end{split}
  \end{equation}
\end{aproof}

\section{Overall error decomposition}
\label{sec:overall_error}

\newcommand{\Timestepsubset}{\bfT}

\cfclear
\begin{athm}{prop}{prop:error_decomposition2}
  Let
  $ ( \Omega, \cF, \P ) $
  be a probability space,
  let
    $M,d\in\N$,
    $ D \subseteq \R^d $,
    $ u \in \R $,
    $ v \in ( u, \infty ) $,    
  for every
  $ j \in \{ 1, 2, \ldots, M \} $
  let
  $ X_j \colon \Omega \to D $
  and
  $ Y_j \colon \Omega \to [ u, v ] $
  be random variables,
  let
  $ \bfR \colon \R^\bfd \to \R $
  satisfy for all
  $ \theta \in \R^\bfd $
  that
  \begin{equation}
    \llabel{eq:defR}
    \bfR( \theta )
  = \E[ \lvert \clippedNN{\theta}{\bfl}{u}{v}( X_1 ) - Y_1 \rvert^2 ]
  ,
  \end{equation}
  let
  $ \bfd, \bfL\in \N $,
  $ \bfl = ( \bfl_0, \bfl_1, \ldots, \bfl_\bfL ) \in \N^{ \bfL + 1 } $
  satisfy
  \begin{equation}
    \textstyle
     \bfl_0 = d ,\qquad
 \bfl_\bfL = 1 ,\qquad
\text{and}\qquad
 \bfd \geq \sum_{i=1}^{\bfL} \bfl_i( \bfl_{ i - 1 } + 1 ) ,
  \end{equation}
    let
  $ \emprisk \colon \R^{\bfd} \times \Omega \to \R $
  satisfy for all
  $ \theta \in \R^{\bfd} $
  that
  \begin{equation}
  \emprisk( \theta )
  =
  \frac{1}{M}
  \biggl[
  \smallsum_{j=1}^M
      \lvert \clippedNN{\theta}{\bfl}{u}{v}( X_j ) - Y_j \rvert^2
  \biggr]
  ,
  \end{equation}        
  let
  $ \cE \colon D \to [ u, v ] $
  be
  $ \cB( D ) $/$ \cB( [ u, v ] ) $-measurable,
  assume
  $ \P $-a.s.\
  that
  \begin{equation}
  \begin{split} 
  \cE( X_1 )
  = \E[ Y_1 \vert X_1 ],
  \end{split}
  \end{equation}
  let
  $ B \in [ 0, \infty ) $,
  for every
  $ k, n \in \N_0 $
  let
  $ \Theta_{ k, n } \colon \Omega \to \R^{\bfd} $
  be a function,
  let
$ K, N \in \N $,
$ \Timestepsubset \subseteq \{ 0, 1, \ldots, N \} $,
let
$ \bfk \colon \Omega \to (\N_0)^2 $
satisfy for all
$ \omega \in \Omega $
that
\begin{align}
  \llabel{eq:condk1}
  \bfk(\omega)
  &\in
  \{(k,n)\in \{1,2,\dots,K\}\times \Timestepsubset\colon \infnorm{\Theta_{k,n}(\omega)}\leq B\}
  \\\text{and}\qquad
  \llabel{eq:condk2}
  \emprisk(\Theta_{\bfk(\omega)}(\omega))
  &=
  \textstyle
  \min_{(k,n)\in \{1,2,\dots,K\}\times \Timestepsubset,\,\infnorm{\Theta_{k,n}(\omega)}\leq B}
  \emprisk(\Theta_{k,n}(\omega))
\end{align}
\cfload.
Then
it holds for all
$ \vartheta \in [ -B, B ]^\bfd $
that
\begin{equation}
\begin{split}
&
\int_D
    \lvert \clippedNN{\Theta_\bfk}{\bfl}{u}{v}( x ) - \cE( x ) \rvert^2
\, \P_{ X_1 }( \diff x )
\\ &
\leq
\bigl[
\sup\nolimits_{ x \in D }
    \lvert \clippedNN{\vartheta}{\bfl}{u}{v}( x ) - \cE( x ) \rvert^2
\bigr]
+
2\bigl[
\sup\nolimits_{ \theta \in [ -B, B ]^\bfd }
\lvert
    \emprisk( \theta )
    -
    \bfR( \theta )
\rvert
\bigr]
\\ & \quad
+
\min\nolimits_{ ( k, n ) \in \{ 1, 2, \ldots, K \} \times \Timestepsubset, \, \infnorm{ \Theta_{ k, n } } \leq B }
\br{ \emprisk( \Theta_{ k, n } )
-
\emprisk( \vartheta ) }
.
\end{split}
\end{equation}
\end{athm}
\begin{aproof}
Throughout this proof,
let
$ \mathbf r \colon \cL^2( \P_{ \smash{ X_1 } }; \R ) \to [ 0, \infty ) $
satisfy for all
$ f \in \cL^2( \P_{ \smash{ X_1 } }; \R ) $
that
\begin{equation}
  \llabel{eq:defr}
  \mathbf r( f ) = \E[ \lvert f( X_1 ) - Y_1 \rvert^2 ]
  .
\end{equation}
\Nobs that
the assumption that
for all
$ \omega \in \Omega $
it holds that
$Y_1( \omega ) \in [ u, v ] $
and
the fact that
for all
$ \theta \in \R^\bfd$,
$ x \in \R^d$
 it holds that
$\clippedNN{\theta}{\bfl}{u}{v}( x ) \in [ u, v ] $
\prove that for all
$ \theta \in \R^\bfd $
it holds that
$ \E[ \lvert Y_1 \rvert^2 ]
\leq \max\{ u^2, v^2 \} < \infty $
and
\begin{equation}
\int_D
        \lvert \clippedNN{\theta}{\bfl}{u}{v}( x ) \rvert^2
\, \P_{ X_1 }( \diff x )
=
\E\bigl[
    \lvert \clippedNN{\theta}{\bfl}{u}{v}( X_1 ) \rvert^2
\bigr]
\leq
\max\{ u^2, v^2 \} < \infty
.
\end{equation}
\Cref{item:dec_iii} in \cref{lem:decomposition}
(applied with
$ ( \Omega, \cF, \P ) \is ( \Omega, \cF, \P ) $,
$ ( S, \cS ) \is ( D, \cB( D ) ) $,
$ X \is X_1 $,
$ Y \is
( \Omega \ni \omega \mapsto Y_1( \omega ) \in \R ) $,
$ \functionRisk \is \mathbf r $,
$ f \is \clippedNN{\theta}{\bfl}{u}{v} \vert_D $,
$ g \is \clippedNN{\vartheta}{\bfl}{u}{v} \vert_D $
for
$ \theta, \vartheta \in \R^\bfd $
in the notation of~%
\cref{item:dec_iii} in \cref{lem:decomposition})
\hence
\proves that for all
$ \theta, \vartheta \in \R^\bfd $
it holds that
\begin{eqsplit}
&
\int_D
    \lvert \clippedNN{\theta}{\bfl}{u}{v}( x ) - \cE( x ) \rvert^2
\, \P_{ X_1 }( \diff x )
\\ &
=
\E\bigl[
    \lvert \clippedNN{\theta}{\bfl}{u}{v}( X_1 )
    -
    \cE( X_1 ) \rvert^2
\bigr]
=
\E\bigl[
    \lvert \clippedNN{\theta}{\bfl}{u}{v}( X_1 )
    -
    \E[ Y_1 \vert X_1 ] \rvert^2
\bigr]
\\ &
=
\E\bigl[
    \lvert \clippedNN{\vartheta}{\bfl}{u}{v}( X_1 )
    -
    \E[ Y_1 \vert X_1 ] \rvert^2
\bigr]
+
\mathbf r( \clippedNN{\theta}{\bfl}{u}{v} \vert_D )
-
\mathbf r( \clippedNN{\vartheta}{\bfl}{u}{v} \vert_D )
\end{eqsplit}
Combining
  this
with
  \lref{eq:defr}
  and \lref{eq:defR}
\proves that for all
$ \theta, \vartheta \in \R^\bfd $
it holds that
\begin{eqsplit}
  &
  \int_D
      \lvert \clippedNN{\theta}{\bfl}{u}{v}( x ) - \cE( x ) \rvert^2
  \, \P_{ X_1 }( \diff x )
  \\ &
=
\E\bigl[
    \lvert \clippedNN{\vartheta}{\bfl}{u}{v}( X_1 )
    -
    \cE( X_1 ) \rvert^2
\bigr]
+
\E\bigl[
    \lvert \clippedNN{\theta}{\bfl}{u}{v}( X_1 )
    -
    Y_1 \rvert^2
\bigr]
-
\E\bigl[
    \lvert \clippedNN{\vartheta}{\bfl}{u}{v}( X_1 )
    -
    Y_1 \rvert^2
\bigr]
\\ &
=
\int_D
    \lvert \clippedNN{\vartheta}{\bfl}{u}{v}( x ) - \cE( x ) \rvert^2
\, \P_{ X_1 }( \diff x )
+
\bfR( \theta )
-
\bfR( \vartheta )
.
\end{eqsplit}
This
\proves that for all
$ \theta, \vartheta \in \R^\bfd $
it holds that
\begin{equation}
\label{eq:first_error_estimate}
\begin{split} 
&
\int_D
    \lvert \clippedNN{\theta}{\bfl}{u}{v}( x ) - \cE( x ) \rvert^2
\, \P_{ X_1 }( \diff x )
\\ &
=
\int_D
    \lvert \clippedNN{\vartheta}{\bfl}{u}{v}( x ) - \cE( x ) \rvert^2
\, \P_{ X_1 }( \diff x )
-
[ \emprisk( \theta )
- \bfR( \theta ) ]
+
\emprisk( \vartheta ) - \bfR( \vartheta )
\\&\quad
+
\emprisk( \theta )
-
\emprisk( \vartheta )
\\ & 
\leq
\int_D
    \lvert \clippedNN{\vartheta}{\bfl}{u}{v}( x ) - \cE( x ) \rvert^2
\, \P_{ X_1 }( \diff x )
+
2\bigl[
\max\nolimits_{ \eta \in \{ \theta, \vartheta \} }
\lvert
    \emprisk( \eta )
    -
    \bfR( \eta )
\rvert
\bigr]
\\&\quad
+
\emprisk( \theta )
-
\emprisk( \vartheta )
.
\end{split}
\end{equation}
\Moreover
\lref{eq:condk1}
\proves[ie] that for all
$ \omega \in \Omega $
it holds that
$ \Theta_{ \bfk( \omega ) }( \omega ) \in [ -B, B ]^\bfd $.
Combining
\cref{eq:first_error_estimate}
with
\lref{eq:condk2}
\hence
\proves that for all
$ \vartheta \in [ -B, B ]^\bfd $
it holds that
\begin{eqsplit}
&
\int_D
    \lvert \clippedNN{\Theta_\bfk}{\bfl}{u}{v}( x ) - \cE( x ) \rvert^2
\, \P_{ X_1 }( \diff x )
\\ &
\leq
\int_D
    \lvert \clippedNN{\vartheta}{\bfl}{u}{v}( x ) - \cE( x ) \rvert^2
\, \P_{ X_1 }( \diff x )
+
2\bigl[
\sup\nolimits_{ \theta \in [ -B, B ]^\bfd }
\lvert
    \emprisk( \theta )
    -
    \bfR( \theta )
\rvert
\bigr]
\\&\quad
+
\emprisk( \Theta_\bfk )
-
\emprisk( \vartheta )
\\ &
=
\int_D
    \lvert \clippedNN{\vartheta}{\bfl}{u}{v}( x ) - \cE( x ) \rvert^2
\, \P_{ X_1 }( \diff x )
+
2\bigl[
\sup\nolimits_{ \theta \in [ -B, B ]^\bfd }
\lvert
    \emprisk( \theta )
    -
    \bfR( \theta )
\rvert
\bigr]
\\ & \quad
+
\min\nolimits_{ ( k, n ) \in \{ 1, 2, \ldots, K \} \times \Timestepsubset, \, \infnorm{ \Theta_{ k, n } } \leq B }
[ \emprisk( \Theta_{ k, n } )
-
\emprisk( \vartheta ) ]
\\ &
\leq
\bigl[
\sup\nolimits_{ x \in D }
    \lvert \clippedNN{\vartheta}{\bfl}{u}{v}( x ) - \cE( x ) \rvert^2
\bigr]
+
2\bigl[
\sup\nolimits_{ \theta \in [ -B, B ]^\bfd }
\lvert
    \emprisk( \theta )
    -
    \bfR( \theta )
\rvert
\bigr]
\\ & \quad
+
\min\nolimits_{ ( k, n ) \in \{ 1, 2, \ldots, K \} \times \Timestepsubset, \, \infnorm{ \Theta_{ k, n } } \leq B }
\br{ \emprisk( \Theta_{ k, n } )
-
\emprisk( \vartheta ) }
.
\end{eqsplit}
\end{aproof}

%% file: parts/Analysis_of_the_overall_error.tex
\cchapter{Composed error estimates}{sec:composed_error}

In \cref{part:approx} we have established several estimates for the approximation error,
in \cref{part:opt} we have established several estimates for the optimization error, and
in \cref{part:generalization} we have established several estimates for the generalization error.
In this chapter we employ the error decomposition from \cref{sect:overall_error_decomp} as well as parts of \cref{part:approx,part:opt,part:generalization} 
(see \cref{prop:approximation_error,cor:minimum_MC_rate,cor:generalisation_error})
to establish estimates for the overall error in the training of \anns\ in the specific situation of \GD-type optimization methods with many independent random initializations.

In the literature such overall error analyses can, \eg, be found in \cite{Beck2019published,JentzenRiekert2023,JentzenWelti2023}.
The material in this chapter consist of slightly modified extracts from Jentzen \& Welti~\cite[Sections 6.2 and 6.3]{JentzenWelti2023}.

\section{Full strong error analysis for the training of ANNs}
\label{sec:overall_error_strong_rate}

\cfclear
\begin{athm}{lemma}{lem:measurability}
Let
$ d, \bfd, \bfL \in \N $,
$ \bfl = ( \bfl_0, \bfl_1, \ldots, \bfl_\bfL ) \in \N^{ \bfL + 1 } $,
$ u \in [ -\infty, \infty ) $,
$ v \in ( u, \infty ] $,
let $ D \subseteq \R^d $,
assume
\begin{equation}
    \textstyle
\bfl_0 = d,\qquad
\bfl_\bfL = 1,
\qquad\text{and}\qquad
\bfd \geq \sum_{i=1}^{\bfL} \bfl_i( \bfl_{ i - 1 } + 1 ),
\end{equation}
let
$ \cE \colon D \to \R $
be
$ \cB( D ) $/$ \cB( \R ) $-measurable,
let
$ ( \Omega, \cF, \P ) $
be a probability space,
and
let
$ X \colon \Omega \to D $,
$ \bfk \colon \Omega \to ( \N_0 )^2 $,
and
$ \Theta_{ k, n } \colon \Omega \to \R^{\bfd} $,
$ k, n \in \N_0 $,
be random variables \cfload.
Then
\begin{enumerate}[label=(\roman *)]
    \item
\label{item:lem:measurability:1}
it holds
that
$ \R^\bfd \times \R^d \ni ( \theta, x )
\mapsto
\clippedNN{\theta}{\bfl}{u}{v}( x )
\in \R $
is $ ( \cB( \R^\bfd ) \otimes \cB( \R^d ) ) $/$ \cB( \R ) $-measurable,
\item \llabel{it:3}
it holds for all
$\omega\in\Omega$
that
$
    \R^d \ni x
\mapsto
\clippedNN{ \smash{ \Theta_{ \smash{ \bfk( \omega ) } }( \omega ) } }{ \smash{ \bfl } }{u}{v}( x )
\in \R
$
is $\Borel(\R^d)$/$\Borel(\R)$-mesaurable,
and
\item
\label{item:lem:measurability:3}
it holds for all
$p\in[0,\infty)$
that
\begin{equation}
\Omega \ni \omega
\mapsto
\int_D
    \lvert \clippedNN{ \smash{ \Theta_{ \smash{ \bfk( \omega ) } }( \omega ) } }{ \bfl }{u}{v}( x ) - \cE( x ) \rvert^p
\, \P_{ X }( \diff x )
\in [ 0, \infty ]
\end{equation}
is
$ \cF $/$ \cB( [ 0, \infty ] ) $-measurable
\end{enumerate}
\cfout.
\end{athm}
\begin{aproof}
    Throughout this proof let
    $ \Xi \colon \Omega \to \R^\bfd $
    satisfy for all
    $ \omega \in \Omega $
    that
    \begin{equation}
        \Xi( \omega ) = \Theta_{ \bfk( \omega ) }( \omega ) 
        .
    \end{equation}
\Nobs that
the assumption that
$ \Theta_{ k, n } \colon \Omega \to \R^{\bfd} $,
$ k, n \in \N_0 $,
and
$ \bfk \colon \Omega \to ( \N_0 )^2 $
are random variables
\proves that for all
$ U \in \cB( \R^\bfd ) $
it holds that
\begin{equation}
\begin{split}
\Xi^{ -1 }( U )
& =
\{ \omega \in \Omega \colon
\Xi( \omega ) \in U \}
=
\{ \omega \in \Omega \colon
\Theta_{ \bfk( \omega ) }( \omega ) \in U \}
\\ &
=
\bigl\{ \omega \in \Omega \colon
\bigl[ \exists \, k, n \in \N_0 \colon
( [ \Theta_{ k, n }( \omega ) \in U ]
\land
[ \bfk( \omega ) = ( k, n ) ]
) \bigr] \bigr\}
\\ &
=
\smallbigcup_{ k = 0 }^\infty
\smallbigcup_{ n = 0 }^\infty
    \bigl(
        \{ \omega \in \Omega \colon \Theta_{ k, n }( \omega ) \in U \}
        \cap
        \{ \omega \in \Omega \colon \bfk( \omega ) = ( k, n ) \}    
    \bigr)
\\ &
=
\smallbigcup_{ k = 0 }^\infty
\smallbigcup_{ n = 0 }^\infty
    \bigl(
         [ ( \Theta_{ k, n } )^{ -1 }( U ) ]
        \cap
        [ \bfk^{ -1 }( \{ ( k, n ) \} ) ]
    \bigr)
\in
\cF
.
\end{split}
\end{equation}
This \proves[ep] that
\begin{equation}
    \llabel{eq:Theta_measurable}
    \Omega \ni \omega
    \mapsto
    \Theta_{ \bfk( \omega ) }( \omega )
    \in \R^\bfd
\end{equation}
is $ \cF $/$ \cB( \R^\bfd ) $-measurable.
\Moreover 
that
\cref{cor:ClippedRealNNLipsch}
(applied with
$ a \is -\infnorm{ x } $,
$ b \is \infnorm{ x } $,
$ u \is u $,
$ v \is v $,
$ d \is \bfd $,
$ L \is \bfL $,
$ l \is \bfl $
for
$ x \in \R^d $
in the notation of~%
\cref{cor:ClippedRealNNLipsch})
\proves that for all
$ \theta, \vartheta \in \R^\bfd $,
$ x \in \R^d $
it holds that
\begin{equation}
\begin{split}
&
\lvert
    \clippedNN{\theta}{\bfl}{u}{v}( x ) - \clippedNN{\vartheta}{\bfl}{u}{v}( x )
\rvert
\leq
\sup\nolimits_{ y \in [ -\infnorm{ x }, \infnorm{ x } ]^{ \bfl_0 } }
    \lvert
        \clippedNN{\theta}{\bfl}{u}{v}( y ) - \clippedNN{\vartheta}{\bfl}{u}{v}( y )
    \rvert
\\ &
\leq
\bfL
\max\{ 1, \infnorm x \}
( \infnorm \bfl + 1 )^\bfL
( \max\{ 1, \infnorm \theta , \infnorm \vartheta \} )^{ \bfL - 1 }
\infnorm{ \theta - \vartheta }
\end{split}
\end{equation}
\cfload.
This
\proves
for all
$ x \in \R^d $
that
\begin{equation}
\label{eq:NN_continuous}
\R^\bfd \ni \theta
\mapsto \clippedNN{\theta}{\bfl}{u}{v}( x ) \in \R
\end{equation}
is continuous.
\Moreover
the fact that for all
$ \theta \in \R^\bfd$ it holds that
$\clippedNN{\theta}{\bfl}{u}{v} \in C( \R^d, \R ) $
\proves that for all
$ \theta \in \R^\bfd $
it holds that
$ 
\clippedNN{\theta}{\bfl}{u}{v}( x ) 
$
is $ \cB( \R^d ) $/$ \cB( \R ) $-measurable.
This,
\cref{eq:NN_continuous},
the fact that
$ ( \R^\bfd, \infnorm{ \cdot } \vert_{ \R^\bfd } ) $
is a separable normed $ \R $-vector space,
and
\cref{lem:productMeasurable_new}
\prove[ep] \cref{item:lem:measurability:1}.
\Nobs that
\cref{item:lem:measurability:1}
and
\lref{eq:Theta_measurable}
\prove that
\begin{equation}
    \Omega \times \R^d \ni ( \omega, x )
\mapsto
\clippedNN{ \smash{ \Theta_{ \smash{ \bfk( \omega ) } }( \omega ) } }{ \smash{ \bfl } }{u}{v}( x )
\in \R
\end{equation}
is $ ( \cF \otimes \cB( \R^d ) ) $/$ \cB( \R ) $-measurable.
This
\proves[epi]
\lref{it:3}.
\Nobs that
\lref{it:3}
and
the assumption that
$ \cE \colon D \to \R $
is
$ \cB( D ) $/$ \cB( \R ) $-measurable
\prove that for all
$p\in[0,\infty)$
it holds that
\begin{equation}
    \Omega \times D \ni ( \omega, x )
\mapsto
\lvert \clippedNN{ \smash{ \Theta_{ \smash{ \bfk( \omega ) } }( \omega ) } }{ \smash{ \bfl } }{u}{v}( x ) - \cE( x ) \rvert^p
\in [ 0, \infty )
\end{equation}
is $ ( \cF \otimes \cB( D ) ) $/$ \cB( [ 0, \infty ) ) $-measurable.
Tonelli's theorem
\hence \proves[ep] \cref{item:lem:measurability:3}.
\end{aproof}

\cfclear
\begin{athm}{prop}{prop:main}
Let
  $ ( \Omega, \cF, \P ) $
  be a probability space,
  let
    $ M,d\in\N$,
    $ b \in [ 1, \infty ) $,
    $ D \subseteq [ -b, b ]^d $,
    $ u \in \R $,
    $ v \in ( u, \infty ) $,
  for every $j\in\N$
  let
  $ X_j \colon \Omega \to D $
  and
  $ Y_j \colon \Omega \to [ u, v ] $
  be random variables,
  assume that
  $ ( X_j, Y_j ) $,
  $ j \in \{ 1, 2, \ldots, M \} $,
  are i.i.d.,
  let
    $ \bfd, \bfL \in \N $,
    $ \bfl = ( \bfl_0, \bfl_1, \ldots, \bfl_\bfL ) \in \N^{ \bfL + 1 } $
  satisfy
  \begin{equation}
    \textstyle 
    \bfl_0 = d ,\qquad
    \bfl_\bfL = 1 ,\qquad
     \text{and}\qquad
    \bfd \geq \sum_{i=1}^{\bfL} \bfl_i( \bfl_{ i - 1 } + 1 ) ,
  \end{equation}   
  let
  $ \emprisk \colon \R^{\bfd} \times \Omega \to [ 0, \infty ) $
  satisfy for all
  $ \theta \in \R^{\bfd} $
  that
  \begin{equation}
  \emprisk( \theta )
  =
  \frac{1}{M}
  \biggl[
  \smallsum_{j=1}^M
      \lvert \clippedNN{\theta}{\bfl}{u}{v}( X_j) - Y_j \rvert^2
  \biggr],
  \end{equation}
  let
  $ \cE \colon D \to [ u, v ] $
  be
  $ \cB( D ) $/$ \cB( [ u, v ] ) $-measurable,
  assume
  $ \P $-a.s.\
  that
  \begin{equation}
    \cE( X_1 )
  = \E[ Y_1 \vert X_1 ]
  ,
  \end{equation}
    let
$ K\in \N $,
$ c \in [ 1, \infty ) $,
$ B \in [ c, \infty ) $,
for every $k,n\in\N_0$
let
$ \Theta_{ k, n } \colon \Omega \to \R^{\bfd} $
be random variables,
assume
$ \bigcup_{ k = 1 }^{ \infty }
\Theta_{ k, 0 }( \Omega ) 
\subseteq [ -B, B ]^\bfd $,
assume that
$ \Theta_{ k, 0 } $,
$ k \in \{ 1, 2, \ldots, K \} $,
are i.i.d.,
assume that
$ \Theta_{ 1, 0 } $ is continuously uniformly distributed on $ [ -c, c ]^\bfd $,
let
  $N\in\N$,
  $ \Timestepsubset \subseteq \{ 0, 1, \ldots, N \} $
satisfy $0 \in \Timestepsubset$,
let
$ \bfk \colon \Omega \to ( \N_0 )^2 $
be a random variable,
and assume for all
  $\omega\in\Omega$
that
\begin{align}
  \llabel{eq:condk1}
  \bfk(\omega)
  &\in
  \{(k,n)\in \{1,2,\dots,K\}\times \Timestepsubset\colon \infnorm{\Theta_{k,n}(\omega)}\leq B\}
  \\\text{and}\qquad
  \llabel{eq:condk2}
  \emprisk(\Theta_{\bfk(\omega)}(\omega))
  &=
  \textstyle
  \min_{(k,n)\in \{1,2,\dots,K\}\times \Timestepsubset,\,\infnorm{\Theta_{k,n}(\omega)}\leq B}
  \emprisk(\Theta_{k,n}(\omega))
\end{align}
\cfload.
Then
it holds for all
$ p \in ( 0, \infty ) $
that
\begin{equation}
  \llabel{eq.claim}
\begin{split}
&
\Bigl(
\E\Bigl[
\Bigl( \medint{D}
        \lvert \clippedNN{\Theta_{\bfk}}{\bfl}{u}{v}( x ) - \cE( x ) \rvert^2
    \, \P_{ X_1 }( \diff x )
\Bigr)^{ \! p \, }
\Bigr]
\Bigr)^{ \! \nicefrac{1}{p} }
\\ &
\leq
\bigl[
\inf\nolimits_{ \theta \in [ -c, c ]^\bfd }
\sup\nolimits_{ x \in D }
    \lvert \clippedNN{\theta}{\bfl}{u}{v}( x ) - \cE( x ) \rvert^{2}
\bigr]
+
\frac{
4 ( v - u ) b
\bfL
( \infnorm{ \bfl } + 1 )^\bfL
c^{ \bfL }
\max\{ 1, p \}
}{
K^{ [ \bfL^{-1} ( \infnorm{ \bfl } + 1 )^{-2} ] }
}
\\ & \quad
+
\frac{
    18 \max\{ 1, ( v - u )^2 \}
    \bfL ( \infnorm{ \bfl } + 1 )^2
    \max\{
        p,
        \ln( 3 M B b )
    \}
}{ \sqrt{M} }
\end{split}
\end{equation}
(cf.~\cref{lem:measurability}).
\end{athm}
\begin{aproof}
Throughout this proof,
let
$ \bfR \colon \R^\bfd \to [ 0, \infty ) $
satisfy for all
$ \theta \in \R^\bfd $
that
\begin{equation}
  \bfR( \theta )
= \E[ \lvert \clippedNN{\theta}{\bfl}{u}{v}( X_1 ) - Y_1 \rvert^2 ]
.
\end{equation}
\Nobs that
\cref{prop:error_decomposition2}
\proves that for all
$ \vartheta \in [ -B, B ]^\bfd $
it holds that%
\begin{equation}
\begin{split}
&
\int_D
    \lvert \clippedNN{\Theta_\bfk}{\bfl}{u}{v}( x ) - \cE( x ) \rvert^2
\, \P_{ X_1 }( \diff x )
\\ &
\leq
\bigl[
\sup\nolimits_{ x \in D }
    \lvert \clippedNN{\vartheta}{\bfl}{u}{v}( x ) - \cE( x ) \rvert^2
\bigr]
+
2\bigl[
\sup\nolimits_{ \theta \in [ -B, B ]^\bfd }
\lvert
    \emprisk( \theta )
    -
    \bfR( \theta )
\rvert
\bigr]
\\ & \qquad
+
\min\nolimits_{ ( k, n ) \in \{ 1, 2, \ldots, K \} \times \Timestepsubset, \, \infnorm{ \Theta_{ k, n } } \leq B }
\lvert \emprisk( \Theta_{ k, n } )
-
\emprisk( \vartheta ) \rvert
.
\end{split}
\end{equation}
The assumption that
$  \bigcup_{ k = 1 }^{ \infty }
\Theta_{ k, 0 }( \Omega ) 
\subseteq [ -B, B ]^\bfd $
and
the assumption that
$ 0 \in \Timestepsubset $
\hence \prove that
\begin{eqsplit}
\label{eq:reduce_to_minimum_MC}
  &
\int_D
    \lvert \clippedNN{\Theta_\bfk}{\bfl}{u}{v}( x ) - \cE( x ) \rvert^2
\, \P_{ X_1 }( \diff x )
\\ &
\leq
\bigl[
\sup\nolimits_{ x \in D }
    \lvert \clippedNN{\vartheta}{\bfl}{u}{v}( x ) - \cE( x ) \rvert^2
\bigr]
+
2\bigl[
\sup\nolimits_{ \theta \in [ -B, B ]^\bfd }
\lvert
    \emprisk( \theta )
    -
    \bfR( \theta )
\rvert
\bigr]
\\ & \qquad
+
\min\nolimits_{ k \in \{ 1, 2, \ldots, K \}, \, \infnorm{ \Theta_{ k, 0 } } \leq B }
\lvert \emprisk( \Theta_{ k, 0 } )
-
\emprisk( \vartheta ) \rvert
\\ &
=
\bigl[
\sup\nolimits_{ x \in D }
    \lvert \clippedNN{\vartheta}{\bfl}{u}{v}( x ) - \cE( x ) \rvert^2
\bigr]
+
2\bigl[
\sup\nolimits_{ \theta \in [ -B, B ]^\bfd }
\lvert
    \emprisk( \theta )
    -
    \bfR( \theta )
\rvert
\bigr]
\\ & \qquad
+
\min\nolimits_{ k \in \{ 1, 2, \ldots, K \} }
\lvert \emprisk( \Theta_{ k, 0 } )
-
\emprisk( \vartheta ) \rvert
.
\end{eqsplit}
Minkowski's inequality
\hence \proves that for all
$ p \in [ 1, \infty ) $,
$ \vartheta \in [ -c, c ]^\bfd
\subseteq [ -B, B ]^\bfd $
it holds that%
\begin{equation}
\label{eq:strong_error_decomposition}
\begin{split}
&
\Bigl(
\E\Bigl[
\Bigl( \medint{D}
        \lvert \clippedNN{\Theta_{\bfk}}{\bfl}{u}{v}( x ) - \cE( x ) \rvert^2
    \, \P_{ X_1 }( \diff x )
\Bigr)^{ \! p \, }
\Bigr]
\Bigr)^{ \! \nicefrac{1}{p} }
\\ &
\leq
\bigl(
\E\bigl[
\sup\nolimits_{ x \in D }
    \lvert \clippedNN{\vartheta}{\bfl}{u}{v}( x ) - \cE( x ) \rvert^{2p}
\bigr]
\bigr)^{ \nicefrac{1}{p} }
+
2 \bigl(
\E\bigl[
\sup\nolimits_{ \theta \in [ -B, B ]^\bfd }
\lvert
    \emprisk( \theta )
    -
    \bfR( \theta )
\rvert^p
\bigr]
\bigr)^{ \nicefrac{1}{p} }
\\ & \qquad
+
\bigl(
\E\bigl[
\min\nolimits_{ k \in \{ 1, 2, \ldots, K \} }
\lvert \emprisk( \Theta_{ k, 0 } )
-
\emprisk( \vartheta ) \rvert^p
\bigr]
\bigr)^{ \nicefrac{1}{p} }
\\ &
\leq
\bigl[
\sup\nolimits_{ x \in D }
    \lvert \clippedNN{\vartheta}{\bfl}{u}{v}( x ) - \cE( x ) \rvert^{2}
\bigr]
+
2 \bigl(
\E\bigl[
\sup\nolimits_{ \theta \in [ -B, B ]^\bfd }
\lvert
    \emprisk( \theta )
    -
    \bfR( \theta )
\rvert^p
\bigr]
\bigr)^{ \nicefrac{1}{p} }
\\ & \qquad
+
\sup\nolimits_{ \theta \in [ -c, c ]^\bfd }
\bigl(
\E\bigl[
\min\nolimits_{ k \in \{ 1, 2, \ldots, K \} }
\lvert \emprisk( \Theta_{ k, 0 } )
-
\emprisk( \theta ) \rvert^p
\bigr]
\bigr)^{ \nicefrac{1}{p} }
\end{split}
\end{equation}
(cf.~\cref{item:cor:generalisation_error:1} in \cref{cor:generalisation_error}
and
\cref{item:cor:minimum_MC_rate:1} in \cref{cor:minimum_MC_rate}).
\Moreover
\cref{cor:generalisation_error}
(applied with
$ v \is \max\{ u + 1, v \} $,
$ \bfR \is \bfR \vert_{ [ -B, B ]^\bfd } $,
$ \emprisk \is \emprisk \vert_{ [ -B, B ]^\bfd \times \Omega } $
in the notation of \cref{cor:generalisation_error})
\proves that for all
$ p \in ( 0, \infty ) $
it holds that
\begin{equation}
\label{eq:strong_generalisation_error}
\begin{split}
& \bigl(
\E\bigl[
    \sup\nolimits_{ \theta \in [ -B, B ]^\bfd }
    \lvert \emprisk( \theta ) - \bfR( \theta ) \rvert^p
\bigr]
\bigr)^{ \nicefrac{1}{p} }
\\ &
\leq
\frac{
    9 ( \max\{ u + 1, v \} - u )^2
    \bfL ( \infnorm{ \bfl } + 1 )^2
    \max\{
        p,
        \ln( 3 M B b )
    \}
}{ \sqrt{M} }
\\ &
=
\frac{
    9 \max\{ 1, ( v - u )^2 \}
    \bfL ( \infnorm{ \bfl } + 1 )^2
    \max\{
        p,
        \ln( 3 M B b )
    \}
}{ \sqrt{M} }
.
\end{split}
\end{equation}
\Moreover
\cref{cor:minimum_MC_rate}
(applied with
$ \fd \is \sum_{i=1}^{\bfL} \bfl_i( \bfl_{ i - 1 } + 1 ) $,
$ B \is c $,
$ ( \Theta_k )_{ k \in \{ 1, 2, \ldots, K \} }
\allowbreak
\is
( \Omega \ni \omega \mapsto
\ind{ \{ \Theta_{ k, 0 } \in [ -c, c ]^\bfd \} }( \omega ) \Theta_{ k, 0 }( \omega ) \in [ -c, c ]^\bfd )_{ k \in \{ 1, 2, \ldots, K \} } $,
$ \emprisk \is \emprisk \vert_{ [ -c, c ]^\bfd \times \Omega } $
in the notation of \cref{cor:minimum_MC_rate})
\proves that  for all
$ p \in ( 0, \infty ) $
it holds that
\begin{equation}
\begin{split}
&
\sup\nolimits_{ \theta \in [ -c, c ]^\bfd }
\bigl(
\E\bigl[
\min\nolimits_{ k \in \{ 1, 2, \ldots, K \} }
    \lvert \emprisk( \Theta_{ k, 0 } )
    -
    \emprisk( \theta ) \rvert^p
\bigr]
\bigr)^{ \nicefrac{1}{p} }
\\ &
=
\sup\nolimits_{ \theta \in [ -c, c ]^\bfd }
\bigl(
\E\bigl[
\min\nolimits_{ k \in \{ 1, 2, \ldots, K \} }
    \lvert \emprisk( \ind{ \{ \Theta_{ k, 0 } \in [ -c, c ]^\bfd \} } \Theta_{ k, 0 } )
    -
    \emprisk( \theta ) \rvert^p
\bigr]
\bigr)^{ \nicefrac{1}{p} }
\\ &
\leq
\frac{
4 ( v - u ) b
\bfL
( \infnorm{ \bfl } + 1 )^\bfL
c^{ \bfL }
\max\{ 1, p \}
}{
K^{ [ \bfL^{-1} ( \infnorm{ \bfl } + 1 )^{-2} ] }
}
.
\end{split}
\end{equation}
Combining
  this
  and \cref{eq:strong_generalisation_error}
with
  \cref{eq:strong_error_decomposition}
\proves that for all
  $p\in[1,\infty)$
it holds that
\begin{equation}
  \llabel{eq:1}
  \begin{split}
  &
  \Bigl(
  \E\Bigl[
  \Bigl( \medint{D}
          \lvert \clippedNN{\Theta_{\bfk}}{\bfl}{u}{v}( x ) - \cE( x ) \rvert^2
      \, \P_{ X_1 }( \diff x )
  \Bigr)^{ \! p \, }
  \Bigr]
  \Bigr)^{ \! \nicefrac{1}{p} }
  \\ &
  \leq
  \bigl[
    \inf\nolimits_{\theta\in[-c,c]^\bfd}
  \sup\nolimits_{ x \in D }
      \lvert \clippedNN{\theta}{\bfl}{u}{v}( x ) - \cE( x ) \rvert^{2}
  \bigr]
  +
  \frac{
    4 ( v - u ) b
    \bfL
    ( \infnorm{ \bfl } + 1 )^\bfL
    c^{ \bfL }
    \max\{ 1, p \}
    }{
      K^{ [ \bfL^{-1} ( \infnorm{ \bfl } + 1 )^{-2} ] }
      }
      \\ & \qquad
      +
  \frac{
    18 \max\{ 1, ( v - u )^2 \}
    \bfL ( \infnorm{ \bfl } + 1 )^2
    \max\{
        p,
        \ln( 3 M B b )
    \}
  }{ \sqrt{M} }
  .
  \end{split}
\end{equation}
\Moreover that
Jensen's inequality
\proves that
for all
  $p\in(0,\infty)$
it holds that
\begin{eqsplit}
  &
  \Bigl(
  \E\Bigl[
  \Bigl( \medint{D}
          \lvert \clippedNN{\Theta_{\bfk}}{\bfl}{u}{v}( x ) - \cE( x ) \rvert^2
      \, \P_{ X_1 }( \diff x )
  \Bigr)^{ \! p \, }
  \Bigr]
  \Bigr)^{ \! \nicefrac{1}{p} }
  \\ &
  \leq
  \biggl(
  \E\biggl[
  \Bigl( \medint{D}
          \lvert \clippedNN{\Theta_{\bfk}}{\bfl}{u}{v}( x ) - \cE( x ) \rvert^2
      \, \P_{ X_1 }( \diff x )
  \Bigr)^{ \! \max\{ 1, p \} }
  \biggr]
  \biggr)^{ \!\! \frac{1}{\max\{ 1, p \}} }    
\end{eqsplit}
This,
\lref{eq:1},
and the fact that
$ \ln( 3 M B b ) \geq 1 $
\prove that for all
$ p \in ( 0, \infty ) $
it holds that
\begin{equation}
\label{eq:prop:main_derivation}
\begin{split}
&
\Bigl(
\E\Bigl[
\Bigl( \medint{D}
        \lvert \clippedNN{\Theta_{\bfk}}{\bfl}{u}{v}( x ) - \cE( x ) \rvert^2
    \, \P_{ X_1 }( \diff x )
\Bigr)^{ \! p \, }
\Bigr]
\Bigr)^{ \! \nicefrac{1}{p} }
\\ &
\leq
\bigl[
\inf\nolimits_{ \theta \in [ -c, c ]^\bfd }
\sup\nolimits_{ x \in D }
    \lvert \clippedNN{\theta}{\bfl}{u}{v}( x ) - \cE( x ) \rvert^{2}
\bigr]
+
\frac{
4 ( v - u ) b
\bfL
( \infnorm{ \bfl } + 1 )^\bfL
c^{ \bfL }
\max\{ 1, p \}
}{
K^{ [ \bfL^{-1} ( \infnorm{ \bfl } + 1 )^{-2} ] }
}
\\ & \qquad
+
\frac{
    18 \max\{ 1, ( v - u )^2 \}
    \bfL ( \infnorm{ \bfl } + 1 )^2
    \max\{
        p,
        \ln( 3 M B b )
    \}
}{ \sqrt{M} }
.
\end{split}
\end{equation}
\end{aproof}

\cfclear
\begingroup
\begin{athm}{lemma}{lem:ln_estimate1}
  Let
  $ a, x, p \in ( 0, \infty ) $.
  Then
  $ a x^p
  \leq
  \exp\bigl(  \tfrac{a^{ \nicefrac{1}{p} } p x }{ e } \bigr) $.  
\end{athm}
\begin{aproof}
  \Nobs that
  the fact that for all
  $ y \in \R$
  it holds that
  $y + 1 \leq e^{ y } $
  \proves that%
  \begin{equation}
  a x^p
  =
  ( a^{ \nicefrac{1}{p} } x )^p
  =
  \bigl[ e \bigl(  \tfrac{ a^{ \nicefrac{1}{p} }x }{ e } - 1 + 1 \bigr) \bigr]^p
  \leq
  \bigl[ e \exp\bigl(  \tfrac{ a^{ \nicefrac{1}{p} }x }{ e }  - 1 \bigr) \bigr]^p
  =
  \exp\bigl(  \tfrac{ a^{ \nicefrac{1}{p} }p x }{ e } \bigr)
  .
  \end{equation}
\end{aproof}
\endgroup

\cfclear
\begin{athm}{lemma}{lem:ln_estimate2}
Let
$ M, c \in [ 1, \infty ) $,
$ B \in [ c, \infty ) $.
Then
$ \ln( 3 M B c )
\leq
\tfrac{23 B}{18}
\ln( e M ) $.
\end{athm}
\begin{aproof}
\Nobs that
\cref{lem:ln_estimate1}
and
the fact that
$ \nicefrac{ 2 \sqrt{3} }{e}
\leq 
\nicefrac{23}{18} $
\prove that
\begin{equation}
3 B^2
\leq
\exp\bigl( \tfrac{ 2 \sqrt{3} B }{e} \bigr)
\leq
\exp\bigl( \tfrac{ 23 B }{18} \bigr)
.
\end{equation}
The fact that
$ B \geq c \geq 1 $
and
$ M \geq 1 $
\hence
\proves that
\begin{equation}
\ln( 3 M B c )
\leq
\ln( 3 B^2 M )
\leq
\ln( [ e M ]^{ \nicefrac{23 B}{18} } )
=
\tfrac{23 B}{18}
\ln( e M )
.
\end{equation}
\end{aproof}

\cfclear
\begin{athm}{theorem}{thm:main}
Let
$ ( \Omega, \cF, \P ) $
be a probability space,
let
  $M,d\in\N$,
  $a,u\in\R$,
  $ b \in ( a, \infty ) $,
  $ v \in ( u, \infty ) $,
for every $j\in\N$
let
$ X_j \colon \Omega \to [ a, b ]^d $
and
$ Y_j \colon \Omega \to [ u, v ] $
be random variables,
assume that
$ ( X_j, Y_j ) $,
$ j \in \{ 1, 2, \ldots, M \} $,
are i.i.d.,
let
$ A \in ( 0, \infty ) $,
$\bfL\in\N$
satisfy
$\bfL \geq \nicefrac{ A \ind{ \smash{ ( 6^d, \infty ) } }( A ) }{ ( 2d ) } + 1$,
let
$ \bfl = ( \bfl_0, \bfl_1, \ldots, \bfl_\bfL ) \in \N^{ \bfL + 1 } $
satisfy for all
$ i \in \{2,3,4,\ldots\} \cap [ 0, \bfL ) $
that
\begin{equation}
  \textstyle
\bfl_0 = d ,\quad
\bfl_1 \geq A \ind{ \smash{ ( 6^d, \infty ) } }( A ) ,\quad
\bfl_i \geq \ind{ \smash{ ( 6^d, \infty ) } }( A ) \max\{ \nicefrac{A}{d} - 2i + 3, 2 \},\quad
\text{and}\quad
\bfl_\bfL = 1,
\end{equation}
let
$\bfd\in\N$ satisfy
$\bfd \geq \sum_{i=1}^{\bfL} \bfl_i( \bfl_{ i - 1 } + 1 ) $,
let
$ \emprisk \colon \R^{\bfd} \times \Omega \to [ 0, \infty ) $
satisfy for all
$ \theta \in \R^{\bfd} $
that
\begin{equation}
\emprisk( \theta )
=
\frac{1}{M}
\biggl[
\smallsum_{j=1}^M
    \lvert \clippedNN{\theta}{\bfl}{u}{v}( X_j ) - Y_j \rvert^2
\biggr],
\end{equation}
let
$ \cE \colon [ a, b ]^d \to [ u, v ] $
satisfy
$ \P $-a.s.\ that
\begin{equation}
  \cE( X_1 )
= \E[ Y_1 \vert X_1 ]
,
\end{equation}
let
$L\in\R$
satisfy for all
$ x, y \in [ a, b ]^d $
that
$ \lvert \cE( x ) - \cE( y ) \rvert \leq L \pnorm1 { x - y } $,
let
$ K \in \N $,
$ c \in
[ \max\{ 1, L, \lvert a \rvert, \lvert b \rvert, 2 \lvert u \rvert, 2 \lvert v \rvert \}, \infty ) $,
$ B \in [ c, \infty ) $,
for every
$ k, n \in \N_0 $
let
$ \Theta_{ k, n } \colon \Omega \to \R^{\bfd} $
be a random variable,
assume
$ \bigcup_{ k = 1 }^{ \infty }
\Theta_{ k, 0 }( \Omega ) 
\subseteq [ -B, B ]^\bfd $,
assume that
$ \Theta_{ k, 0 } $,
$ k \in \{ 1, 2, \ldots, K \} $,
are i.i.d.,
assume that
$ \Theta_{ 1, 0 } $ is continuously uniformly distributed on $ [ -c, c ]^\bfd $,
let
$N\in\N$,
$ \Timestepsubset \subseteq \{ 0, 1, \ldots, N \} $
satisfy
$ 0 \in \Timestepsubset $,
let
$ \bfk \colon \Omega \to ( \N_0 )^2 $
be a random variable,
and  assume for all
  $\omega\in\Omega$
that
\begin{align}
  \bfk(\omega)
  &\in
  \{(k,n)\in \{1,2,\dots,K\}\times \Timestepsubset\colon \infnorm{\Theta_{k,n}(\omega)}\leq B\}
  \\\text{and}\qquad
  \emprisk(\Theta_{\bfk(\omega)}(\omega))
  &=
  \textstyle
  \min_{(k,n)\in \{1,2,\dots,K\}\times \Timestepsubset,\,\infnorm{\Theta_{k,n}(\omega)}\leq B}
  \emprisk(\Theta_{k,n}(\omega))
\end{align}\cfload.
Then
it holds for all
$ p \in ( 0, \infty ) $
that
\begin{eqsplit}
  \label{eq:thm:main}
&
\Bigl(
\E\Bigl[
\Bigl( \medint{[ a, b ]^d}
        \lvert \clippedNN{\Theta_{\bfk}}{\bfl}{u}{v}( x ) - \cE( x ) \rvert^2
    \, \P_{ X_1 }( \diff x )
\Bigr)^{ \! p \, }
\Bigr]
\Bigr)^{ \! \nicefrac{1}{p} }
\\ &
\leq
\frac{ 36 d^2 c^4 }{ A^{ \nicefrac{2}{d} } }
+
\frac{
4
\bfL
( \infnorm{ \bfl } + 1 )^\bfL
c^{ \bfL + 2 }
\max\{ 1, p \}
}{
K^{ [ \bfL^{-1} ( \infnorm{ \bfl } + 1 )^{-2} ] }
}
\\&\quad
+
\frac{
    23 B^3
    \bfL ( \infnorm{ \bfl } + 1 )^2
    \max\{ p, \ln( e M ) \}
}{ \sqrt{M} }
\end{eqsplit}
(cf.~\cref{lem:measurability}).
\end{athm}
\begin{aproof}
\Nobs that
the assumption that
for all
$ x, y \in [ a, b ]^d $
it holds that
$\lvert \cE( x ) - \cE( y ) \rvert \leq L \pnorm1 { x - y } $
\proves that
$ \cE \colon [ a, b ]^d \to [ u, v ] $
is
$ \cB( [ a, b ]^d ) $/$ \cB( [ u, v ] ) $-measurable.
\Cref{prop:main}
(applied with
$ b \is \max\{ 1, \lvert a \rvert, \lvert b \rvert \} $,
$ D \is [ a, b ]^d $
in the notation of \cref{prop:main})
\hence
\proves that for all
$ p \in ( 0, \infty ) $
it holds that
\begin{eqsplit}
&
\Bigl(
\E\Bigl[
\Bigl( \medint{[ a, b ]^d}
        \lvert \clippedNN{\Theta_{\bfk}}{\bfl}{u}{v}( x ) - \cE( x ) \rvert^2
    \, \P_{ X_1 }( \diff x )
\Bigr)^{ \! p \, }
\Bigr]
\Bigr)^{ \! \nicefrac{1}{p} }
\\ &
\leq
\bigl[
\inf\nolimits_{ \theta \in [ -c, c ]^\bfd }
\sup\nolimits_{ x \in [ a, b ]^d }
    \lvert \clippedNN{\theta}{\bfl}{u}{v}( x ) - \cE( x ) \rvert^{2}
\bigr]
\\ & \qquad
+
\frac{
4 ( v - u ) \max\{ 1, \lvert a \rvert, \lvert b \rvert \}
\bfL
( \infnorm{ \bfl } + 1 )^\bfL
c^{ \bfL }
\max\{ 1, p \}
}{
K^{ [ \bfL^{-1} ( \infnorm{ \bfl } + 1 )^{-2} ] }
}
\\ & \qquad
+
\frac{
    18 \max\{ 1, ( v - u )^2 \}
    \bfL ( \infnorm{ \bfl } + 1 )^2
    \max\{
        p,
        \ln( 3 M B \max\{ 1, \lvert a \rvert, \lvert b \rvert \} )
    \}
}{ \sqrt{M} }
.
\end{eqsplit}
The fact that
$ \max\{ 1, \lvert a \rvert, \lvert b \rvert \}
\leq c $
\hence 
\proves that for all
    $p\in(0,\infty)$
it holds that
\begin{eqsplit}
  \label{eq:prop:main_estimate}
  &
\Bigl(
\E\Bigl[
\Bigl( \medint{[ a, b ]^d}
        \lvert \clippedNN{\Theta_{\bfk}}{\bfl}{u}{v}( x ) - \cE( x ) \rvert^2
    \, \P_{ X_1 }( \diff x )
\Bigr)^{ \! p \, }
\Bigr]
\Bigr)^{ \! \nicefrac{1}{p} }
  \\ &
\leq
\bigl[
\inf\nolimits_{ \theta \in [ -c, c ]^\bfd }
\sup\nolimits_{ x \in [ a, b ]^d }
    \lvert \clippedNN{\theta}{\bfl}{u}{v}( x ) - \cE( x ) \rvert^{2}
\bigr]
\\&\quad
+
\frac{
4 ( v - u )
\bfL
( \infnorm{ \bfl } + 1 )^\bfL
c^{ \bfL + 1 }
\max\{ 1, p \}
}{
K^{ [ \bfL^{-1} ( \infnorm{ \bfl } + 1 )^{-2} ] }
}
\\ & \quad
+
\frac{
    18 \max\{ 1, ( v - u )^2 \}
    \bfL ( \infnorm{ \bfl } + 1 )^2
    \max\{
        p,
        \ln( 3 M B c )
    \}
}{ \sqrt{M} }
.
\end{eqsplit}
\Moreover[Furthermore]%
\cref{prop:approximation_error}
(applied with
$ f \is \cE $
in the notation of \cref{prop:approximation_error})
\proves[de] that
there exists $ \vartheta \in \R^\bfd $
such that
$ \infnorm{ \vartheta }
\leq \max\{ 1, L, \abs a , \allowbreak \abs b , \allowbreak 2[ \sup_{ x \in [ a, b ]^d } \lvert \cE( x ) \rvert ] \} $
and
\begin{equation}
\label{eq:prop:approximation_error}
\sup\nolimits_{ x \in [ a, b ]^d }
    \lvert \clippedNN{\vartheta}{\bfl}{u}{v}( x ) - \cE( x ) \rvert
\leq
\frac{ 3 d L ( b - a ) }{ A^{ \nicefrac{1}{d} } }
.
\end{equation}
The fact that
for all
$ x \in [ a, b ]^d$
it holds that 
$\cE( x ) \in [ u, v ] $
\hence
\proves that
\begin{equation}
\infnorm{ \vartheta }
\leq
\max\{ 1, L, \lvert a \rvert, \lvert b \rvert, 2 \lvert u \rvert, 2 \lvert v \rvert \}
\leq c
.
\end{equation}
This and
\cref{eq:prop:approximation_error}
\prove that
\begin{equation}
\begin{split}
\inf\nolimits_{ \theta \in [ -c, c ]^\bfd }
\sup\nolimits_{ x \in [ a, b ]^d }
    \lvert \clippedNN{\theta}{\bfl}{u}{v}( x ) - \cE( x ) \rvert^2
&
\leq
\sup\nolimits_{ x \in [ a, b ]^d }
    \lvert \clippedNN{\vartheta}{\bfl}{u}{v}( x ) - \cE( x ) \rvert^2
\\ &
\leq
\biggl[
\frac{ 3 d L ( b - a ) }{ A^{ \nicefrac{1}{d} } }
\biggr]^2
=
\frac{ 9 d^2 L^2 ( b - a )^2 }{ A^{ \nicefrac{2}{d} } }
.
\end{split}
\end{equation}
Combining
this
with
\cref{eq:prop:main_estimate}
\proves that for all
$ p \in ( 0, \infty ) $
it holds that
\begin{equation}
\label{eq:overall_estimate}
\begin{split}
&
\Bigl(
\E\Bigl[
\Bigl( \medint{[ a, b ]^d}
        \lvert \clippedNN{\Theta_{\bfk}}{\bfl}{u}{v}( x ) - \cE( x ) \rvert^2
    \, \P_{ X_1 }( \diff x )
\Bigr)^{ \! p \, }
\Bigr]
\Bigr)^{ \! \nicefrac{1}{p} }
\\ &
\leq
\frac{ 9 d^2 L^2 ( b - a )^2 }{ A^{ \nicefrac{2}{d} } }
+
\frac{
4 ( v - u )
\bfL
( \infnorm{ \bfl } + 1 )^\bfL
c^{ \bfL + 1 }
\max\{ 1, p \}
}{
K^{ [ \bfL^{-1} ( \infnorm{ \bfl } + 1 )^{-2} ] }
}
\\ & \qquad
+
\frac{
    18 \max\{ 1, ( v - u )^2 \}
    \bfL ( \infnorm{ \bfl } + 1 )^2
    \max\{
        p,
        \ln( 3 M B c )
    \}
}{ \sqrt{M} }
.
\end{split}
\end{equation}
\Moreover
the fact that
$ \max\{ 1, L, \lvert a \rvert, \lvert b \rvert \} \leq c $
and
$ ( b - a )^2 \leq ( \lvert a \rvert + \lvert b \rvert )^2 \leq 2 ( a^2 + b^2 ) $
\proves that
\begin{equation}
\label{eq:coarse_estimate}
9 L^2 ( b - a )^2
\leq
18 c^2 ( a^2 + b^2 )
\leq
18 c^2 ( c^2 + c^2 )
=
36 c^4
.
\end{equation}
\Moreover
the fact that
$ B \geq c \geq 1 $,
the fact that
$ M \geq 1 $,
and 
\cref{lem:ln_estimate2}
\prove that
$ \ln( 3 M B c )
\leq
\tfrac{ 23 B }{18}
\ln( e M ) $.
This,
\cref{eq:coarse_estimate},
the fact that
$ ( v - u )
\leq 2 \max\{ \lvert u \rvert, \lvert v \rvert \}
= \max\{ 2 \lvert u \rvert, 2 \lvert v \rvert \}
\leq c
\leq B $,
and
the fact that
$ B \geq 1 $
\prove that for all
$ p \in ( 0, \infty ) $
it holds that
\begin{eqsplit}
&
\frac{ 9 d^2 L^2 ( b - a )^2 }{ A^{ \nicefrac{2}{d} } }
+
\frac{
4 ( v - u )
\bfL
( \infnorm{ \bfl } + 1 )^\bfL
c^{ \bfL + 1 }
\max\{ 1, p \}
}{
K^{ [ \bfL^{-1} ( \infnorm{ \bfl } + 1 )^{-2} ] }
}
\\ &\quad
+
\frac{
    18 \max\{ 1, ( v - u )^2 \}
    \bfL ( \infnorm{ \bfl } + 1 )^2
    \max\{
        p,
        \ln( 3 M B c )
    \}
}{ \sqrt{M} }
\\ &
\leq
\frac{ 36 d^2 c^4 }{ A^{ \nicefrac{2}{d} } }
+
\frac{
4
\bfL
( \infnorm{ \bfl } + 1 )^\bfL
c^{ \bfL + 2 }
\max\{ 1, p \}
}{
K^{ [ \bfL^{-1} ( \infnorm{ \bfl } + 1 )^{-2} ] }
}
\\&\quad
+
\frac{
    23 B^3
    \bfL ( \infnorm{ \bfl } + 1 )^2
    \max\{ p, \ln( e M ) \}
}{ \sqrt{M} }
.
\end{eqsplit}
Combining this with \cref{eq:overall_estimate}
\proves \cref{eq:thm:main}.
\end{aproof}

\cfclear
\begin{athm}{cor}{cor:min_NN_architecture}
Let
  $ ( \Omega, \cF, \P ) $
  be a probability space,
  let
  $M,d\in\N$,
  $ a, u \in \R $,
  $ b \in ( a, \infty ) $,
  $ v \in ( u, \infty ) $,
  for every $j\in\N$
  let
  $ X_j \colon \Omega \to [ a, b ]^d $
  and
  $ Y_j \colon \Omega \to [ u, v ] $
  be random variables,
  assume that
  $ ( X_j, Y_j ) $,
  $ j \in \{ 1, 2, \ldots, M \} $,
  are i.i.d.,
let
  $\bfd,\bfL\in\N$,
  $ \bfl = ( \bfl_0, \bfl_1, \ldots, \bfl_\bfL ) \in \N^{ \bfL + 1 } $,
assume
\begin{equation}
  \textstyle
  \bfl_0 = d,\qquad
  \bfl_\bfL = 1,\qquad
 \text{and}\qquad
  \bfd \geq \sum_{i=1}^{\bfL} \bfl_i( \bfl_{ i - 1 } + 1 ), 
\end{equation}
  let
$ \emprisk \colon \R^{\bfd} \times \Omega \to [ 0, \infty ) $
satisfy for all
$ \theta \in \R^{\bfd} $
that
\begin{equation}
\emprisk( \theta )
=
\frac{1}{M}
\biggl[
\smallsum_{j=1}^M
    \lvert \clippedNN{\theta}{\bfl}{u}{v}( X_j ) - Y_j \rvert^2
\biggr],
\end{equation}
let
$ \cE \colon [ a, b ]^d \to [ u, v ] $
satisfy
$ \P $-a.s.\
that
\begin{equation}
  \cE( X_1 )
= \E[ Y_1 \vert X_1 ]
,
\end{equation}
let
$L\in\R$
satisfy for all
$ x, y \in [ a, b ]^d $
that
$ \lvert \cE( x ) - \cE( y ) \rvert \leq L \pnorm1 { x - y } $,
let
$K\in\N$,
$ c \in
[ \max\{ 1, L, \lvert a \rvert, \lvert b \rvert, 2 \lvert u \rvert, 2 \lvert v \rvert \}, \infty ) $,
$ B \in [ c, \infty ) $,
for every
$ k, n \in \N_0 $
let
$ \Theta_{ k, n } \colon \Omega \to \R^{\bfd} $
be a random variable,
assume
$ \bigcup_{ k = 1 }^{ \infty }
\Theta_{ k, 0 }( \Omega ) 
\subseteq [ -B, B ]^\bfd $,
assume that
$ \Theta_{ k, 0 } $,
$ k \in \{ 1, 2, \ldots, K \} $,
are i.i.d.,
assume that
$ \Theta_{ 1, 0 } $ is continuously uniformly distributed on $ [ -c, c ]^\bfd $,
let
$ N \in \N $,
$ \Timestepsubset \subseteq \{ 0, 1, \ldots,
\allowbreak
N \} $
satisfy
$
 0 \in \Timestepsubset$,
let
$ \bfk \colon \Omega \to ( \N_0 )^2 $
be a random variable,
and assume for all
$\omega\in\Omega$
that
\begin{align}
  \bfk(\omega)
  &\in
  \{(k,n)\in \{1,2,\dots,K\}\times \Timestepsubset\colon \infnorm{\Theta_{k,n}(\omega)}\leq B\}
  \\\text{and}\qquad
  \emprisk(\Theta_{\bfk(\omega)}(\omega))
  &=
  \textstyle
  \min_{(k,n)\in \{1,2,\dots,K\}\times \Timestepsubset,\,\infnorm{\Theta_{k,n}(\omega)}\leq B}
  \emprisk(\Theta_{k,n}(\omega))
\end{align}
\cfload.
Then
it holds for all
$ p \in ( 0, \infty ) $
that
\begin{eqsplit}
&
\Bigl(
\E\Bigl[
\Bigl( \medint{[ a, b ]^d}
        \lvert \clippedNN{\Theta_{\bfk}}{\bfl}{u}{v}( x ) - \cE( x ) \rvert^2
    \, \P_{ X_1 }( \diff x )
\Bigr)^{ \! \nicefrac{p}{2} \, }
\Bigr]
\Bigr)^{ \! \nicefrac{1}{p} }
\\ &
\leq
\frac{
6 d c^2
}{
[ \min(
    \{ \bfL \}
    \cup
    \{ \bfl_i \colon i \in \N \cap [ 0, \bfL ) \}
) ]^{ \nicefrac{1}{d} }
}
+
\frac{
2
\bfL
( \infnorm{ \bfl } + 1 )^\bfL
c^{ \bfL + 1 }
\max\{ 1, p \}
}{
K^{ [ ( 2 \bfL )^{-1} ( \infnorm{ \bfl } + 1 )^{-2} ] }
}
\\ & \qquad
+
\frac{
    5 B^2
    \bfL ( \infnorm{ \bfl } + 1 )
    \max\{ p, \ln( e M ) \}
}{ M^{ \nicefrac{1}{4} } }
\end{eqsplit}
(cf.~\cref{lem:measurability}).
\end{athm}
\begin{aproof}
Throughout this proof,
let
\begin{equation}
\label{eq:choice_A}
A
=
\min(
    \{ \bfL \}
    \cup
    \{ \bfl_i \colon i \in \N \cap [ 0, \bfL ) \}
)\in(0,\infty)
.
\end{equation}
\Nobs that
\cref{eq:choice_A}
\proves that
\begin{equation}
\label{eq:assumption_bfL}
\begin{split}
\bfL
& \geq
A
=
A - 1 + 1
\geq
( A - 1 ) \ind{ [ 2, \infty ) }( A ) + 1
\\ &
\geq
\bigl( A - \tfrac{A}{2} \bigr) \ind{ [ 2, \infty ) }( A ) + 1
=
\tfrac{ A \ind{ [ 2, \infty ) }( A ) }{ 2 } + 1
\geq
\tfrac{ A \ind{ \smash{ ( 6^d, \infty ) } }( A ) }{ 2d } + 1
.
\end{split}
\end{equation}
\Moreover
the assumption that
$ \bfl_\bfL = 1 $
and
\cref{eq:choice_A}
\prove that
\begin{equation}
\label{eq:assumption_bfl_1}
\bfl_1
=
\bfl_1
\ind{ \{ 1 \} }( \bfL )
+
\bfl_1
\ind{ [ 2, \infty ) }( \bfL )
\geq
\ind{ \{ 1 \} }( \bfL )
+
A
\ind{ [ 2, \infty ) }( \bfL )
=
A
\geq
A \ind{ \smash{ ( 6^d, \infty ) } }( A )
.
\end{equation}
\Moreover
\cref{eq:choice_A}
\proves that for all
$ i \in \{2,3,4,\ldots\} \cap [ 0, \bfL ) $
it holds that
\begin{equation}
\label{eq:assumption_bfl_i}
\begin{split}
\bfl_i
& \geq
A
\geq
A
\ind{ \smash{ [ 2, \infty ) } }( A )
\geq
\ind{ \smash{ [ 2, \infty ) } }( A )
\max\{ A - 1, 2 \}
=
\ind{ \smash{ [ 2, \infty ) } }( A )
\max\{ A - 4 + 3, 2 \}
\\ &
\geq
\ind{ \smash{ [ 2, \infty ) } }( A )
\max\{ A - 2 i + 3, 2 \}
\geq
\ind{ \smash{ ( 6^d, \infty ) } }( A )
\max\{ \nicefrac{A}{d} - 2i + 3, 2 \}
.
\end{split}
\end{equation}
Combining
this,
\cref{eq:assumption_bfL},
and \cref{eq:assumption_bfl_1}
with
\cref{thm:main}
(applied with
$ p \is \nicefrac{p}{2} $
for
$ p \in ( 0, \infty ) $
in the notation of \cref{thm:main})
\proves that for all
$ p \in ( 0, \infty ) $
it holds that
\begin{eqsplit}
&
\Bigl(
\E\Bigl[
\Bigl( \medint{[ a, b ]^d}
        \lvert \clippedNN{\Theta_{\bfk}}{\bfl}{u}{v}( x ) - \cE( x ) \rvert^2
    \, \P_{ X_1 }( \diff x )
\Bigr)^{ \! \nicefrac{p}{2} \, }
\Bigr]
\Bigr)^{ \! \nicefrac{2}{p} }
\\ &
\leq
\frac{ 36 d^2 c^4 }{ A^{ \nicefrac{2}{d} } }
+
\frac{
4
\bfL
( \infnorm{ \bfl } + 1 )^\bfL
c^{ \bfL + 2 }
\max\{ 1, \nicefrac{p}{2} \}
}{
K^{ [ \bfL^{-1} ( \infnorm{ \bfl } + 1 )^{-2} ] }
}
\\&\quad
+
\frac{
    23 B^3
    \bfL ( \infnorm{ \bfl } + 1 )^2
    \max\{ \nicefrac{p}{2}, \ln( e M ) \}
}{ \sqrt{M} }
.
\end{eqsplit}
This,
\cref{eq:choice_A},
and
the fact that
$ \bfL \geq 1 $,
$ c \geq 1 $,
$ B \geq 1 $,
and
$ \ln( e M ) \geq 1 $
\prove that for all
$ p \in ( 0, \infty ) $
it holds that
\begin{eqsplit}
&
\Bigl(
\E\Bigl[
\Bigl( \medint{[ a, b ]^d}
        \lvert \clippedNN{\Theta_{\bfk}}{\bfl}{u}{v}( x ) - \cE( x ) \rvert^2
    \, \P_{ X_1 }( \diff x )
\Bigr)^{ \! \nicefrac{p}{2} \, }
\Bigr]
\Bigr)^{ \! \nicefrac{1}{p} }
\\ &
\leq
\frac{
6 d c^2
}{
[ \min(
    \{ \bfL \}
    \cup
    \{ \bfl_i \colon i \in \N \cap [ 0, \bfL ) \}
) ]^{ \nicefrac{1}{d} }
}
+
\frac{
2
[
\bfL
( \infnorm{ \bfl } + 1 )^\bfL
c^{ \bfL + 2 }
\max\{ 1, \nicefrac{p}{2} \}
]^{ \nicefrac{1}{2} }
}{
K^{ [ ( 2 \bfL )^{-1} ( \infnorm{ \bfl } + 1 )^{-2} ] }
}
\\ & \qquad
+
\frac{
    5 B^3
    [ \bfL ( \infnorm{ \bfl } + 1 )^2
    \max\{ \nicefrac{p}{2}, \ln( e M ) \}
    ]^{ \nicefrac{1}{2} }
}{ M^{ \nicefrac{1}{4} } }
\\ &
\leq
\frac{
6 d c^2
}{
[ \min(
    \{ \bfL \}
    \cup
    \{ \bfl_i \colon i \in \N \cap [ 0, \bfL ) \}
) ]^{ \nicefrac{1}{d} }
}
+
\frac{
2
\bfL
( \infnorm{ \bfl } + 1 )^\bfL
c^{ \bfL + 1 }
\max\{ 1, p \}
}{
K^{ [ ( 2 \bfL )^{-1} ( \infnorm{ \bfl } + 1 )^{-2} ] }
}
\\ & \qquad
+
\frac{
    5 B^2
    \bfL ( \infnorm{ \bfl } + 1 )
    \max\{ p, \ln( e M ) \}
}{ M^{ \nicefrac{1}{4} } }
.
\end{eqsplit}
\end{aproof}

\section{Full strong error analysis with optimization via SGD with random initializations}
\label{sec:SGD}

\cfclear
\begin{athm}{cor}{cor:SGD_full_error}
  let
  $ ( \Omega, \cF, \P ) $
  be a probability space,
  let
  $M,d\in\N$,
  $ a, u \in \R $,
  $ b \in ( a, \infty ) $,
  $ v \in ( u, \infty ) $,
  for every
  $ k, n, j \in \N_0 $
  let
  $ X^{ k, n }_{ \smash{j} }
  \colon \Omega \to [ a, b ]^d $
  and
  $ Y^{ k, n }_{ \smash{j} }
  \colon \Omega \to [ u, v ] $
  be random variables,
  assume that
  $ ( X^{ 0, 0 }_{ \smash{j} }, Y^{ 0, 0 }_{ \smash{j} } ) $,
  $ j \in \{ 1, 2, \ldots, M \} $,
  are i.i.d.,
let
  $\bfd,\bfL\in\N$,
  $ \bfl = ( \bfl_0, \bfl_1, \ldots, \bfl_\bfL ) \in \N^{ \bfL + 1 } $
satisfy
\begin{equation}
  \textstyle
  \bfl_0 = d ,\qquad
  \bfl_\bfL = 1 ,
 \qquad\text{and}\qquad
  \bfd \geq \sum_{i=1}^{\bfL} \bfl_i( \bfl_{ i - 1 } + 1 ) ,
\end{equation}
for every
$ k, n \in \N_0 $, $J\in\N$
let
$ \emprisk^{ k, n }_{ \smash{J} } \colon \R^{\bfd} \times \Omega \to [ 0, \infty ) $
satisfy for all
$ \theta \in \R^{\bfd} $,
$ \omega \in \Omega $
that
\begin{equation}
\emprisk^{ k, n }_{ \smash{J} }( \theta, \omega )
=
\frac{1}{J}
\biggl[
\smallsum_{j=1}^{J}
    \lvert \clippedNN{\theta}{\bfl}{u}{v}( X^{ k, n }_{ \smash{j} }( \omega ) ) - Y^{ k, n }_{ \smash{j} }( \omega ) \rvert^2
\biggr]
,
\end{equation}
let
$ \cE \colon [ a, b ]^d \to [ u, v ] $
satisfy
$ \P $-a.s.\
that
\begin{equation}
  \cE( X_{ \smash{1} }^{ 0, 0 } )
= \E[ Y_{ \smash{1} }^{ 0, 0 } \vert X_{ \smash{1} }^{ 0, 0 } ]
,
\end{equation}
let
$ L \in \R $
satisfy for all
$ x, y \in [ a, b ]^d $
that
$ \lvert \cE( x ) - \cE( y ) \rvert \leq L \pnorm1 { x - y } $,
let
$ ( \bfJ_n )_{ n \in \N } \subseteq \N $,
for every
$ k, n \in \N $
let
$ \cG^{ k, n } \colon \R^{\bfd} \times \Omega \to \R^{\bfd} $
satisfy for all
$ \omega \in \Omega $,
$ \theta \in
\{ \vartheta \in \R^{\bfd} \colon
( \emprisk^{ k, n }_{ \smash{ \bfJ_n } } ( \cdot, \omega ) \colon\allowbreak \R^{\bfd} \to [ 0, \infty )
\text{ is differentiable at } \vartheta ) \} $
that
\begin{equation}
  \cG^{ k, n }( \theta, \omega )
=
( \nabla_\theta \emprisk^{ k, n }_{ \smash{ \bfJ_n } } )( \theta, \omega )
,
\end{equation}
let
  $K\in\N$,
$ c \in
[ \max\{ 1, L, \lvert a \rvert, \lvert b \rvert,\allowbreak 2 \lvert u \rvert, 2 \lvert v \rvert \}, \infty ) $,
$ B \in [ c, \infty ) $,
for every 
$ k, n \in \N_0 $
let
$ \Theta_{ k, n } \colon \Omega \to \R^{\bfd} $
be a random variable,
assume
$ \bigcup_{ k = 1 }^{ \infty }
\Theta_{ k, 0 }( \Omega ) 
\subseteq [ -B, B ]^\bfd $,
assume that
$ \Theta_{ k, 0 } $,
$ k \in \{ 1, 2, \ldots, \allowbreak K \} $,
are i.i.d.,
assume that
$ \Theta_{ 1, 0 } $ is continuously uniformly distributed on $ [ -c, c ]^\bfd $,
let
$ ( \gamma_n )_{ n \in \N } \subseteq \R $
satisfy for all
$ k, n \in \N $
that
\begin{equation}
  \Theta_{ k, n } = \Theta_{ k, n -1 } - \gamma_n \cG^{ k, n }( \Theta_{ k, n -1 } )
  ,
\end{equation}
let
$ N \in \N $,
$ \Timestepsubset \subseteq \{ 0, 1, \ldots,
\allowbreak
N \} $
satisfy
$0 \in \Timestepsubset$,
let
$ \bfk \colon \Omega \to ( \N_0 )^2 $
be a random variable,
and assume for all $\omega\in\Omega$ that
\begin{align}
  \bfk(\omega)
  &\in
  \{(k,n)\in \{1,2,\dots,K\}\times \Timestepsubset\colon \infnorm{\Theta_{k,n}(\omega)}\leq B\}
  \\\text{and}\qquad
  \emprisk(\Theta_{\bfk(\omega)}(\omega))
  &=
  \textstyle
  \min_{(k,n)\in \{1,2,\dots,K\}\times \Timestepsubset,\,\infnorm{\Theta_{k,n}(\omega)}\leq B}
  \emprisk(\Theta_{k,n}(\omega))
\end{align}
\cfload.
Then it holds for all
  $p\in(0,\infty)$
that
\begin{eqsplit}
\label{eq:cor:SGD_full_error}
&
\Bigl(
\E\Bigl[
\Bigl( \medint{[ a, b ]^d}
        \lvert \clippedNN{\Theta_{\bfk}}{\bfl}{u}{v}( x ) - \cE( x ) \rvert^2
    \, \P_{ X^{ 0, 0 }_{ \smash{1}\vphantom{x} } }( \diff x )
\Bigr)^{ \! \nicefrac{p}{2} \, }
\Bigr]
\Bigr)^{ \! \nicefrac{1}{p} }
\\ &
\leq
\frac{
6 d c^2
}{
[ \min(
    \{ \bfL \}
    \cup
    \{ \bfl_i \colon i \in \N \cap [ 0, \bfL ) \}
) ]^{ \nicefrac{1}{d} }
}
+
\frac{
2
\bfL
( \infnorm{ \bfl } + 1 )^\bfL
c^{ \bfL + 1 }
\max\{ 1, p \}
}{
K^{ [ ( 2 \bfL )^{-1} ( \infnorm{ \bfl } + 1 )^{-2} ] }
}
\\ & \qquad
+
\frac{
    5 B^2
    \bfL ( \infnorm{ \bfl } + 1 )
    \max\{ p, \ln( e M ) \}
}{ M^{ \nicefrac{1}{4} } }
\end{eqsplit}
(cf.~\cref{lem:measurability}).
\end{athm}
\begin{aproof}
\Nobs[Note] that
\cref{cor:min_NN_architecture}
(applied with
$ ( X_j )_{ j \in \N } \is ( X^{ 0, 0 }_{ \smash{j} } )_{ j \in \N } $,
$ ( Y_j )_{ j \in \N } \is ( Y^{ 0, 0 }_{ \smash{j} } )_{ j \in \N } $,
$ \emprisk \is \emprisk^{ 0, 0 }_{ \smash{ M } } $
in the notation of \cref{cor:min_NN_architecture})
\proves[epsi] \cref{eq:cor:SGD_full_error}.
\end{aproof}

\cfclear
\begin{athm}{cor}{cor:SGD_L1}
  Let
  $ ( \Omega, \cF, \P ) $
  be a probability space,
  let
  $M,d\in\N$,
  $ a, u \in \R $,
  $ b \in ( a, \infty ) $,
  $ v \in ( u, \infty ) $,
  for every
  $ k, n, j \in \N_0 $
  let
  $ X^{ k, n }_{ \smash{j} }
  \colon \Omega \to [ a, b ]^d $
  and
  $ Y^{ k, n }_{ \smash{j} }
  \colon \Omega \to [ u, v ] $
  be random variables,
  assume that
  $ ( X^{ 0, 0 }_{ \smash{j} }, Y^{ 0, 0 }_{ \smash{j} } ) $,
  $ j \in \{ 1, 2, \ldots, M \} $,
  are i.i.d.,
  let
$ \bfd, \bfL \in \N $,
$ \bfl = ( \bfl_0, \bfl_1, \ldots, \bfl_\bfL ) \in \N^{ \bfL + 1 } $
satisfy
\begin{equation}
  \textstyle
  \bfl_0 = d ,\qquad
  \bfl_\bfL = 1 ,\qquad
 \text{and}\qquad
  \bfd \geq \sum_{i=1}^{\bfL} \bfl_i( \bfl_{ i - 1 } + 1 ) ,
\end{equation}
for every
$ k, n\in\N_0$, $J \in \N $
let
$ \emprisk^{ k, n }_{ \smash{J} } \colon \R^{\bfd} \times \Omega \to [ 0, \infty ) $
satisfy for all
$ \theta \in \R^{\bfd} $
that
\begin{equation}
\emprisk^{ k, n }_{ \smash{J} }( \theta)
=
\frac{1}{J}
\biggl[
\smallsum_{j=1}^{J}
    \lvert \clippedNN{\theta}{\bfl}{u}{v}( X^{ k, n }_{ \smash{j} } ) - Y^{ k, n }_{ \smash{j} } \rvert^2
\biggr]
,
\end{equation}
let
$ \cE \colon [ a, b ]^d \to [ u, v ] $
satisfy
$ \P $-a.s.\ that
\begin{equation}
  \cE( X_{ \smash{1} }^{ 0, 0 } )
= \E[ Y_{ \smash{1} }^{ 0, 0 } \vert X_{ \smash{1} }^{ 0, 0 } ] 
,
\end{equation}
let
  $L\in\R$
satisfy for all
$ x, y \in [ a, b ]^d $
that
$ \lvert \cE( x ) - \cE( y ) \rvert \leq L \pnorm1 { x - y } $,
let
$ ( \bfJ_n )_{ n \in \N } \subseteq \N $,
for every 
$ k, n \in \N $
let
$ \cG^{ k, n } \colon \R^{\bfd} \times \Omega \to \R^{\bfd} $
satisfy for all
$ \omega \in \Omega $,
$ \theta \in
\{ \vartheta \in \R^{\bfd} \colon
( \emprisk^{ k, n }_{ \smash{ \bfJ_n } } ( \cdot, \omega ) \colon\allowbreak \R^{\bfd} \to [ 0, \infty )
\text{ is differentiable at } \vartheta ) \} $
that
\begin{equation}
  \cG^{ k, n }( \theta, \omega )
=
( \nabla_\theta \emprisk^{ k, n }_{ \smash{ \bfJ_n } } )( \theta, \omega ) 
,
\end{equation}
let
$K\in\N$,
$ c \in
[ \max\{ 1, L, \lvert a \rvert, \lvert b \rvert, 2 \lvert u \rvert, 2 \lvert v \rvert \}, \infty ) $,
$ B \in [ c, \infty ) $,
for every
$ k, n \in \N_0 $
let
$ \Theta_{ k, n } \colon \Omega \to \R^{\bfd} $
be a random variable,
assume
$ \bigcup_{ k = 1 }^{ \infty }
\Theta_{ k, 0 }( \Omega ) 
\subseteq [ -B, B ]^\bfd $,
assume that
$ \Theta_{ k, 0 } $,
$ k \in \{ 1, 2, \ldots, K \} $,
are i.i.d.,
assume that
$ \Theta_{ 1, 0 } $ is continuously uniformly distributed on $ [ -c, c ]^\bfd $,
let
$ ( \gamma_n )_{ n \in \N } \subseteq \R $
satisfy for all
$ k, n \in \N $
that
\begin{equation}
  \Theta_{ k, n } = \Theta_{ k, n -1 } - \gamma_n \cG^{ k, n }( \Theta_{ k, n -1 } )
  ,
\end{equation}
let
$ N \in \N $,
$ \Timestepsubset \subseteq \{ 0, 1, \ldots,
\allowbreak
N \} $
satisfy
$0 \in \Timestepsubset$,
let
$ \bfk \colon \Omega \to ( \N_0 )^2 $
be a random variable,
and assume for all
  $\omega\in\Omega$
  that
  \begin{align}
    \bfk(\omega)
    &\in
    \{(k,n)\in \{1,2,\dots,K\}\times \Timestepsubset\colon \infnorm{\Theta_{k,n}(\omega)}\leq B\}
    \\\text{and}\qquad
    \emprisk(\Theta_{\bfk(\omega)}(\omega))
    &=
    \textstyle
    \min_{(k,n)\in \{1,2,\dots,K\}\times \Timestepsubset,\,\infnorm{\Theta_{k,n}(\omega)}\leq B}
    \emprisk(\Theta_{k,n}(\omega))
  \end{align}
  \cfload.
Then
\begin{equation}
  \llabel{eq.claim}
\begin{split}
&
\E\Bigl[
\medint{[ a, b ]^d}
        \lvert \clippedNN{\Theta_{\bfk}}{\bfl}{u}{v}( x ) - \cE( x ) \rvert
    \, \P_{ X^{ 0, 0 }_{ \smash{1}\vphantom{x} } }( \diff x )
\Bigr]
\\ &
\leq
\frac{
6 d c^2
}{
[ \min\{ \bfL, \bfl_1, \bfl_2, \ldots, \bfl_{ \bfL - 1 } \} ]^{ \nicefrac{1}{d} }
}
+
\frac{
    5 B^2
    \bfL ( \infnorm{ \bfl } + 1 )
    \ln( e M )
}{ M^{ \nicefrac{1}{4} } }
+
\frac{
2
\bfL
( \infnorm{ \bfl } + 1 )^\bfL
c^{ \bfL + 1 }
}{
K^{ [ ( 2 \bfL )^{-1} ( \infnorm{ \bfl } + 1 )^{-2} ] }
}
\end{split}
\end{equation}
(cf.~\cref{lem:measurability}).
\end{athm}
\begin{aproof}
\Nobs that
Jensen's inequality
\proves that
\begin{equation}
\E\Bigl[
\medint{[ a, b ]^d}
        \lvert \clippedNN{\Theta_{\bfk}}{\bfl}{u}{v}( x ) - \cE( x ) \rvert
    \, \P_{ X^{ 0, 0 }_{ \smash{1}\vphantom{x} } }( \diff x )
\Bigr]
\leq
\E\Bigl[
\Bigl( \medint{[ a, b ]^d}
        \lvert \clippedNN{\Theta_{\bfk}}{\bfl}{u}{v}( x ) - \cE( x ) \rvert^2
    \, \P_{ X^{ 0, 0 }_{ \smash{1}\vphantom{x} } }( \diff x )
\Bigr)^{ \! \nicefrac{1}{2} \, }
\Bigr]
.
\end{equation}
This and
\cref{cor:SGD_full_error}
(applied with
$ p \is 1 $
in the notation of \cref{cor:SGD_full_error})
\prove[ep] \lref{eq.claim}.
\end{aproof}

\cfclear
\begin{athm}{cor}{cor:SGD_simplfied}
Let
  $ ( \Omega, \cF, \P ) $
  be a probability space,
  $ M,d \in \N $,
  for every
  $ k, n, j \in \N_0 $
  let
  $ X^{ k, n }_{ \smash{j} }
  \colon \Omega \to [ 0, 1 ]^d $
  and
  $ Y^{ k, n }_{ \smash{j} }
  \colon \Omega \to [ 0, 1 ] $
  be random variables,
  assume that
  $ ( X^{ 0, 0 }_{ \smash{j} }, Y^{ 0, 0 }_{ \smash{j} } ) $,
  $ j \in \{ 1, 2, \ldots, M \} $,
  are i.i.d.,
  for every
$ k, n \in \N_0 $,
$J\in\N$
let
$ \emprisk^{ k, n }_{ \smash{J} } \colon \R^{\bfd} \times \Omega \to [ 0, \infty ) $
satisfy for all
$ \theta \in \R^{\bfd} $
that
\begin{equation}
\emprisk^{ k, n }_{ \smash{J} }( \theta, \omega )
=
\frac{1}{J}
\biggl[
\smallsum_{j=1}^{J}
    \lvert \clippedNN{\theta}{\bfl}01( X^{ k, n }_{ \smash{j} }( \omega ) ) - Y^{ k, n }_{ \smash{j} }( \omega ) \rvert^2
\biggr]
,
\end{equation}
let
$ \bfd, \bfL \in \N $,
$ \bfl = ( \bfl_0, \bfl_1, \ldots, \bfl_\bfL ) \in \N^{ \bfL + 1 } $
satisfy
\begin{equation}
  \textstyle
  \bfl_0 = d ,\qquad
  \bfl_\bfL = 1 ,\qquad
 \text{and}\qquad
  \bfd \geq \sum_{i=1}^{\bfL} \bfl_i( \bfl_{ i - 1 } + 1 ) ,
\end{equation}
let
$ \cE \colon [ 0, 1 ]^d \to [ 0, 1 ] $
satisfy
$ \P $-a.s.\
that
\begin{equation}
  \cE( X_{ \smash{1} }^{ 0, 0 } )
= \E[ Y_{ \smash{1} }^{ 0, 0 } \vert X_{ \smash{1} }^{ 0, 0 } ]
,
\end{equation}
let
$ c \in
[ 2, \infty ) $,
satisfy for all
$ x, y \in [ 0, 1 ]^d $
that
$ \lvert \cE( x ) - \cE( y ) \rvert \leq c \pnorm1 { x - y } $,
let
$ ( \bfJ_n )_{ n \in \N } \subseteq \N $,
for every
$ k, n \in \N $
let
$ \cG^{ k, n } \colon \R^{\bfd} \times \Omega \to \R^{\bfd} $
satisfy for all
$ \omega \in \Omega $,
$ \theta \in
\{ \vartheta \in \R^{\bfd} \colon
( \emprisk^{ k, n }_{ \smash{ \bfJ_n } } ( \cdot, \omega ) \colon\allowbreak \R^{\bfd} \to [ 0, \infty )
\text{ is differentiable at } \vartheta ) \} $
that
\begin{equation}
  \cG^{ k, n }( \theta, \omega )
=
( \nabla_\theta \emprisk^{ k, n }_{ \smash{ \bfJ_n } } )( \theta, \omega ) 
,
\end{equation}
let
  $K\in\N$,
for every
$ k, n \in \N_0 $
let
$ \Theta_{ k, n } \colon \Omega \to \R^{\bfd} $
be a random variable,
assume
$ \bigcup_{ k = 1 }^{ \infty }
\Theta_{ k, 0 }( \Omega )
\allowbreak
\subseteq [ -c, c ]^\bfd $,
assume that
$ \Theta_{ k, 0 } $,
$ k \in \{ 1, 2, \ldots, K \} $,
are i.i.d.,
assume that
$ \Theta_{ 1, 0 } $ is continuously uniformly distributed on $ [ -c, c ]^\bfd $,
let
$ ( \gamma_n )_{ n \in \N } \subseteq \R $
satisfy for all
$ k, n \in \N $
that
\begin{equation}
  \Theta_{ k, n } = \Theta_{ k, n -1 } - \gamma_n \cG^{ k, n }( \Theta_{ k, n -1 } )
  ,
\end{equation}
let
$ N \in \N $,
$ \Timestepsubset \subseteq \{ 0, 1, \ldots, N \} $
satisfy
     $0 \in \Timestepsubset$,
let
$ \bfk \colon \Omega \to ( \N_0 )^2 $
be a random variable,
and assume for all
$ \omega \in \Omega $
that
\begin{align}
  \bfk(\omega)
  &\in
  \{(k,n)\in \{1,2,\dots,K\}\times \Timestepsubset\colon \infnorm{\Theta_{k,n}(\omega)}\leq B\}
  \\\text{and}\qquad
  \emprisk(\Theta_{\bfk(\omega)}(\omega))
  &=
  \textstyle
  \min_{(k,n)\in \{1,2,\dots,K\}\times \Timestepsubset,\,\infnorm{\Theta_{k,n}(\omega)}\leq B}
  \emprisk(\Theta_{k,n}(\omega))
\end{align}
\cfload.
Then
\begin{equation}
    \llabel{eq:claim}
\begin{split}
&
\E\Bigl[
\medint{[ 0, 1 ]^d}
        \lvert \clippedNN{\Theta_{\bfk}}{\bfl}01( x ) - \cE( x ) \rvert
    \, \P_{ X^{ 0, 0 }_{ \smash{1}\vphantom{x} } }( \diff x )
\Bigr]
\\ & %
\leq
\frac{
6d c^2
}{
[ \min\{ \bfL, \bfl_1, \bfl_2, \ldots, \bfl_{ \bfL - 1 } \} ]^{ \nicefrac{1}{d} }
}
+
\frac{
    5c^2
    \bfL ( \infnorm{ \bfl } + 1 )
    \ln( e M )
}{ M^{ \nicefrac{1}{4} } }
+
\frac{
\bfL
( \infnorm{ \bfl } + 1 )^\bfL
c^{ \bfL + 1 }
}{
K^{ [ ( 2 \bfL )^{-1} ( \infnorm{ \bfl } + 1 )^{-2} ] }
}
\end{split}
\end{equation}
(cf.~\cref{lem:measurability}).
\end{athm}
\begin{aproof}
\Nobs that
\cref{cor:SGD_L1}
(applied with
$ a \is 0 $,
$ u \is 0 $,
$ b \is 1 $,
$ v \is 1 $,
$ L \is c $,
$ c \is c $,
$ B \is c $
in the notation of \cref{cor:SGD_L1}),
the fact that
$ c \geq 2 $
and
$ M \geq 1 $,
and \cref{lem:ln_estimate2}
\prove 
\lref{eq:claim}.
\end{aproof}

%% file: parts/PINNs.tex
\cchapter{Physics-informed neural networks (PINNs)}{subsec:dgm}

Deep learning methods have not only become very popular for data-driven learning problems, but are nowadays also heavily used for solving mathematical equations such as ordinary and partial differential equations (cf., \eg, \cite{EHanJentzen2017Science,EHanJentzen17,raissi2019physics,Sirignano2018dgm}).
In particular, we refer to the overview articles \cite{beck2020overview,Germain2021,Ruf2019,Cuomo2022,karniadakis2021physics,Brunton2023}
and the references therein for numerical simulations and theoretical investigations for deep learning methods for \PDEs.

Often deep learning methods for \PDEs\ are obtained, 
first,
by reformulating the \PDE\ problem under consideration as an infinite-dimensional stochastic optimization problem, 
then, 
by approximating the infinite-dimensional stochastic optimization problem through finite-dimensional stochastic optimization problems involving deep \anns\ as approximations for the \PDE\ solution and/or its derivatives,
and
thereafter, 
by approximately solving the resulting finite-dimensional stochastic optimization problems through \SGD-type optimization methods.

Among the most basic schemes of such deep learning methods for \PDEs\ are
 \PINNs\ and \DGMs; see \cite{raissi2019physics,Sirignano2018dgm}.
 In this chapter we present in \cref{thm:dgm} in \cref{subsec:dgm_theory} a reformulation of \PDE\ problems as stochastic optimization problems,
 we use the theoretical considerations from \cref{subsec:dgm_theory}  to briefly sketch in \cref{sec:PINNS_derivation} a possible derivation of \PINNs\ and \DGMs,
 and 
 we present in \cref{sec:PINNs,sec:DGM} numerical simulations for \PINNs\ and \DGMs.
For simplicity and concreteness we restrict ourselves in this chapter to the case of semilinear heat \PDEs.
The specific presentation of this chapter is based on Beck et al.~\cite{beck2020overview}.

\todoc{Restructure this section. The method deriation should become the intro text and in which we use the results further down. The theorem should be proved as a consequence of the abstract result. For the intro text we will either keep the PDE or change to the Laplace PDE as it is simpler. (As of the second version on arXiv this is commented out. We will need to do this for the third version)}

\section{Reformulation of PDE problems as stochastic optimization problems}
\label{subsec:dgm_theory}

Both \PINNs\ and \DGMs\ are based on reformulations of the considered \PDEs\ as suitable infinite-dimensional stochastic optimization problems.
In \cref{thm:dgm} below we present the theoretical result behind this reformulation in the special case of semilinear heat \PDEs.

\begingroup 
\providecommand{\Tv}{}
\renewcommand{\Tv}{\mathcal{T}}
\providecommand{\Xv}{}
\renewcommand{\Xv}{\mathcal{X}}
\providecommand{\Td}{} 
\renewcommand{\Td}{\mathscr{t}}
\providecommand{\Xd}{}
\renewcommand{\Xd}{\mathscr{x}} 

\providecommandordefault{\PinnsLossInfinite}{\mathfrak{L}}
\providecommandordefault{\LossFunction}{\defaultLossFunction}

\begin{athm}{theorem}{thm:dgm}
  Let
    $T\in(0,\infty)$,
    $d\in\N$,
    $g\in C^2(\R^d,\R)$,
    $u\in C^{1,2}([0,T]\times\R^d,\R)$,
    $\Td\in C([0,T],(0,\infty))$,
    $\Xd\in C(\R^d,(0,\infty))$,
  assume that $g$ has at most polynomially growing partial derivatives,
  let $(\Omega,\mathcal F,\P)$ be a probability space,
  let
    $\Tv\colon \Omega\to[0,T]$
    and $\Xv\colon \Omega\to\R^d$
    be independent random variables,
  assume
    for all
      $A\in\Borel([0,T])$,
      $B\in\Borel(\R^d)$
    that
    \begin{equation}
      \label{eq:densities}
      \P(\Tv\in A)=\int_A\Td(t)\,\diff t
      \qquad\text{and}\qquad
      \P(\Xv\in B)=\int_B \Xd(x)\,\diff x,
    \end{equation}
  let $f\colon\R\to\R$ be Lipschitz continuous,
  and let $\PinnsLossInfinite\colon C^{1,2}([0,T]\times\R^d,\R)\to[0,\infty]$
  satisfy
    for all
      $v=(v(t,x))_{(t,x)\in[0,T]\times\R^d}\in C^{1,2}([0,T]\times\R^d,\R)$
    that
    \begin{equation}
      \label{eq:dgmloss}
      \PinnsLossInfinite(v)
      =
      \Exp{
        \abs{v(0,\Xv)-g(\Xv)}^2
        +
        \abs[\big]{\bpr{\tfrac{\partial v}{\partial t}}(\Tv,\Xv)-(\Delta_x v)(\Tv,\Xv)-f(v(\Tv,\Xv))}{}^2
      }
      .
    \end{equation}
  Then the following two statements are equivalent:
  \begin{enumerate}[(i)]
    \item \llabel{it:1}
    It holds that
      $\PinnsLossInfinite(u)=\inf_{v\in C^{1,2}([0,T]\times\R^d,\R)}\PinnsLossInfinite(v)$.
    \item \llabel{it:2}
    It holds for all
      $t\in[0,T]$,
      $x\in\R^d$
    that
      $u(0,x)=g(x)$ and
      \begin{equation}
        \label{eq:dgm.pde}
      \bpr{\tfrac{\partial u}{\partial t}}(t,x) = (\Delta_x u)(t,x)+f(u(t,x))
      .
      \end{equation}
  \end{enumerate}
\end{athm}
\begin{aproof}
  \Nobs that
    \cref{eq:dgmloss}
  \proves that for all
    $v\in C^{1,2}([0,T]\times\R^d,\R)$
    with
      $\forall\,x\in\R^d\colon u(0,x)=g(x)$ and
      $\forall\,t\in[0,T],\,x\in\R^d\colon \bpr{\tfrac{\partial u}{\partial t}}(t,x) = (\Delta_x u)(t,x)+f(u(t,x))$
  it holds that
  \begin{equation}
    \llabel{eqphi}
    \PinnsLossInfinite(v)=0.
  \end{equation}
    This
    and the fact that
      for all
        $v\in C^{1,2}([0,T]\times\R^d,\R)$
      it holds that
        $\PinnsLossInfinite(v)\geq 0$
  \prove[ep] that (\ref{thm:dgm.it:2} $\rightarrow$ \ref{thm:dgm.it:1}).
  \Nobs that
    the assumption that
      $f$ is Lipschitz continuous,
    the assumption that $g$ is twice continuously differentiable,
    and the assumption that $g$ has at most polynomially growing partial derivatives
  \prove that there exists
    $v\in C^{1,2}([0,T]\times\R^d,\R)$
  which satisfies for all
    $t\in[0,T]$,
    $x\in\R^d$
  that $v(0,x)=g(x)$ and
  \begin{equation}
    \bpr{\tfrac{\partial v}{\partial t}}(t,x) = (\Delta_x v)(t,x)+f(v(t,x))
  \end{equation}
  (cf., \eg, Beck et al.~\cite[Corollary~3.4]{beck2021nonlinear}).
    This
    and \lref{eqphi}
  show that
  \begin{equation}
    \llabel{2}
    \inf_{v\in C^{1,2}([0,T]\times\R^d,\R)}\!\!\!\PinnsLossInfinite(v) = 0.
  \end{equation}
  \Moreover \enum{
    \cref{eq:dgmloss}
    ;
    \cref{eq:densities}
    ;
    the assumption that $\Tv$ and $\Xv$ are independent
  } \prove that for all
    $v\in C^{1,2}([0,T]\times\R^d,\R)$
  it holds that
  \begin{equation}
    \PinnsLossInfinite(v)
    =
    \int_{[0,T]\times\R^d}
    \Bigl(\abs{v(0,x)-g(x)}^2
    +
    \abs[\big]{\bpr{\tfrac{\partial v}{\partial t}}(t,x)-(\Delta_x v)(t,x)-f(v(t,x))}^2\Bigr)
    \Td(t)\Xd(x)
    \,\diff (t,x)
    .
  \end{equation}
    The assumption that
      $\Td$ and $\Xd$ are continuous
    and the fact that for all
      $t\in[0,T]$,
      $x\in\R^d$
    it holds that
      $\Td(t)\geq 0$
      and $\Xd(x)\geq 0$
  \hence \prove that for all
    $v\in C^{1,2}([0,T]\times\R^d,\R)$,
    $t\in[0,T]$,
    $x\in\R^d$
    with $\PinnsLossInfinite(v)=0$
  it holds that
  \begin{equation}
    \Bigl(\abs{v(0,x)-g(x)}^2
    +
    \abs[\big]{\bpr{\tfrac{\partial v}{\partial t}}(t,x)-(\Delta_x v)(t,x)-f(v(t,x))}^2\Bigr)
    \Td(t)\Xd(x)
    =
    0
    .
  \end{equation}
    This
    and the assumption that for all
      $t\in[0,T]$,
      $x\in\R^d$
    it holds that
      $\Td(t)>0$
      and $\Xd(x)>0$
  \prove that for all
    $v\in C^{1,2}([0,T]\times\R^d,\R)$,
    $t\in[0,T]$,
    $x\in\R^d$
    with $\PinnsLossInfinite(v)=0$
  it holds that
  \begin{equation}
    \abs{v(0,x)-g(x)}^2
    +
    \abs[\big]{\bpr{\tfrac{\partial v}{\partial t}}(t,x)-(\Delta_x v)(t,x)-f(v(t,x))}^2
    =
    0
    .
  \end{equation}
  Combining
    this
  with
    \lref{2}
  \proves[ep] that (\ref{thm:dgm.it:1} $\rightarrow$ \ref{thm:dgm.it:2}).
\end{aproof}
\endgroup

\section{Derivation of PINNs and deep Galerkin methods (DGMs)}
\label{sec:PINNS_derivation}

\begingroup
\providecommand{\Tv}{}
\renewcommand{\Tv}{\mathcal{T}}
\providecommand{\Xv}{}
\renewcommand{\Xv}{\mathcal{X}}
\providecommand{\Td}{}
\renewcommand{\Td}{\mathscr{t}}
\providecommand{\Xd}{}
\renewcommand{\Xd}{\mathscr{x}}

\providecommandordefault{\d}{\defaultParamDim}

\providecommandordefault{\PinnsLossInfinite}{\mathfrak{L}}
\providecommandordefault{\LossFunction}{\defaultLossFunction}
\providecommandordefault{\XRV}{\mathfrak{X}}
\providecommandordefault{\TRV}{\mathfrak{T}}

\newcommand{\localAnn}[1]{\RealV{#1}{0}{\defaultInputDim+1}{ \multdim_{\activation, l_1}, \multdim_{\activation, l_2}, \dots, \multdim_{\activation, l_h} , \id_{ \R } }}

In this section we employ the reformulation of semilinear \PDEs\ as optimization problems from 
\cref{thm:dgm} to sketch an informal derivation of deep learning schemes to approximate solutions of semilinear heat \PDEs.
For this 
let
	$T\in(0,\infty)$,
	$d\in\N$,
	$u\in C^{1,2}([0,T]\times\R^d,\R)$,
	$g\in C^2(\R^d,\R)$
satisfy that $g$ has at most polynomially growing partial derivatives,
let $f\colon\R\to\R$ be Lipschitz continuous,
and assume for all
	$t\in[0,T]$,
	$x\in\R^d$
that
	$u(0,x)=g(x)$ 
and
\begin{equation}
\label{PINNS_derivation:eq1}
	\bpr{\tfrac{\partial u}{\partial t}}(t,x) 
= 
	(\Delta_x u)(t,x)+f(u(t,x))
.
\end{equation}
In the framework described in the previous sentence, we think of $u$ as the unknown \PDE\ solution.
The objective of this derivation is to develop deep learning methods which aim to approximate the unknown function $u$.

In the first step we employ \cref{thm:dgm} to reformulate the \PDE\ problem associated to \eqref{PINNS_derivation:eq1} as an infinite-dimensional stochastic optimization problem over a function space.
For this 
let
	$\Td\in C([0,T],(0,\infty))$,
	$\Xd\in C(\R^d,(0,\infty))$,
let $(\Omega,\mathcal F,\P)$ be a probability space,
let
$\Tv\colon \Omega\to[0,T]$
and $\Xv\colon \Omega\to\R^d$
be independent random variables,
assume
for all
$A\in\Borel([0,T])$,
$B\in\Borel(\R^d)$
that
\begin{equation}
\label{PINNS_derivation:eq2}
\P(\Tv\in A)=\int_A\Td(t)\,\diff t
\qquad\text{and}\qquad
\P(\Xv\in B)=\int_B \Xd(x)\,\diff x,
\end{equation}
and let $\PinnsLossInfinite\colon C^{1,2}([0,T]\times\R^d,\R)\to[0,\infty]$
satisfy
for all
$v=(v(t,x))_{(t,x)\in[0,T]\times\R^d}\in C^{1,2}([0,T]\times\R^d,\R)$
that
\begin{equation}
\label{PINNS_derivation:eq3}
\PinnsLossInfinite(v)
=
\Exp{
\abs{v(0,\Xv)-g(\Xv)}^2
+
\abs[\big]{\bpr{\tfrac{\partial v}{\partial t}}(\Tv,\Xv)-(\Delta_x v)(\Tv,\Xv)-f(v(\Tv,\Xv))}{}^2
}
.
\end{equation}
Observe that \cref{thm:dgm} assures that the unknown function $u$ satisfies 
\begin{equation}
\label{PINNS_derivation:eq4}
\begin{split} 
	\PinnsLossInfinite(u)
=
	0
\end{split}
\end{equation}
and is thus a minimizer of the optimization problem associated to \eqref{PINNS_derivation:eq3}.
Motivated by this, we consider aim to find approximations of $u$ by computing approximate minimizers of the function 
$\PinnsLossInfinite\colon C^{1,2}([0,T]\times\R^d,\R)\to[0,\infty]$.
Due to its infinite-dimensionality this optimization problem is however not yet amenable to numerical computations.

For this reason, in the second step, we reduce this infinite-dimensional stochastic optimization problem to a finite-dimensional stochastic optimization problem involving \anns.
Specifically, 
let
$\activation \colon \R \to \R$ be differentiable,
let
$h\in\N$,
$l_1,l_2,\dots,l_h,\defaultParamDim\in\N$ satisfy
$\defaultParamDim=l_1(d+2)+\br[\big]{\sum_{k=2}^h l_k(l_{k-1}+1)}+l_h+1$,
and let
 $\defaultLossFunction \colon\R^\defaultParamDim\to[0,\infty)$ 
 satisfy
for all $\theta\in\R^\defaultParamDim$ that
\begin{equation}
\label{PINNS_derivation:eq5}
\begin{split}
	\defaultLossFunction( \theta )
&= 
	\PinnsLossInfinite\pr[\big]{\localAnn{\theta} }\\
&=
	\EXPPP{
		\abs[\big]{\localAnn{\theta}(0,\Xv)-g(\Xv)}^2 \\
&\quad
		+
		\abs[\Big]{
			\bbpr{\tfrac{\partial \localAnn{\theta}}{\partial t}}(\Tv,\Xv)
			-
			\bpr{\Delta_x \localAnn{\theta}}(\Tv,\Xv)\\
&\quad
			-
			f\pr[\big]{ \localAnn{\theta}(\Tv,\Xv))} 
		}^2
	}
\end{split}
\end{equation}
\cfload.
We can now compute an approximate minimizer of the function $\PinnsLossInfinite$ 
by computing an approximate minimizer $\vartheta \in \R^{\defaultParamDim}$ of the function $\defaultLossFunction$
and employing the realization 
$\localAnn{\vartheta}$
of the \ann\ associated to this approximate minimizer as an approximate minimizer of $\PinnsLossInfinite$.

The third and last step of this derivation is to approximately compute such an approximate minimizer of $\defaultLossFunction$ by means of \SGD-type optimization methods. 
We now sketch this in the case of the plain-vanilla \SGD\ optimization method (cf.\ \cref{def:SGD}).
Let 
	$\xi\in\R^\defaultParamDim$,
	$J \in \N$,
	$(\gamma_n)_{n\in\N}\subseteq[0,\infty)$, 
for every 
	$n \in \N$,
	$j \in \{1, 2, \ldots, J\}$
let
	$\TRV_{n, j} \colon \Omega \to [0,T]$
and
	$\XRV_{n, j} \colon \Omega \to \R^d$
be random variables,
assume for all
	$n \in \N$,
	$j \in \{1, 2, \ldots, J\}$,
	$A\in\Borel([0,T])$,
	$B\in\Borel(\R^d)$
that
\begin{equation}
	\P(\Tv \in A) = \P(\TRV_{n, j} \in A)
\qandq
	\P(\Xv\in B) = \P(\XRV_{n, j} \in B),
\end{equation}
let
$\defaultStochLoss\colon\R^\defaultParamDim\times [0,T] \times \R^d \to\R$
satisfy for all
$\theta\in\R^\d$,
$t \in [0,T]$,
$x \in \R^d$
that
\begin{equation}
\label{PINNS_derivation:eq6}
\begin{split}
	\defaultStochLoss(\theta,t, x)
&=
	\abs[\big]{
		\localAnn{\theta}(0,x)-g(x)}^2 \\
&\quad
	+
	\abs[\Big]{
		\bbpr{\tfrac{\partial \localAnn{\theta}}{\partial t}}(t,x)
		-
		\bpr{\Delta_x \localAnn{\theta}}(t,x)\\
&\quad
		-
		f\pr[\big]{ \localAnn{\theta}(t,x))} 
	}^2,
\end{split}
\end{equation}
and
let
$\Theta = (\Theta_n)_{n \in \N_0} \colon\N_0\times\Omega\to\R^\d$ 
satisfy for all 
$n\in\N$ that
\begin{equation}
\label{PINNS_derivation:eq7}
\Theta_0=\xi
\qquad\text{and}\qquad
\Theta_n
=
\Theta_{n-1}-\gamma_n\br*{\frac1{J}\sum_{j=1}^{J}(\nabla_\theta\defaultStochLoss)(\Theta_{n-1}, \TRV_{n, j}, \XRV_{n, j} )}.
\end{equation}
Finally, the idea of \PINNs\ and \DGMs\ is then to choose
for large enough $n \in \N$ 
the realization
$\localAnn{\Theta_n}$ as an approximation
\begin{equation}
\begin{split}
	\localAnn{\Theta_n} \approx u 
\end{split}
\end{equation}
of the unknown solution $u$ of the \PDE\ in \eqref{PINNS_derivation:eq1}.

The ideas and the resulting schemes in the above derivation were first introduced as \PINNs\ in Raissi et al.~\cite{raissi2019physics} and as \DGMs\ in Sirignano \& Spiliopoulos \cite{Sirignano2018dgm}.
Very roughly speaking, \PINNs\ and \DGMs\ in their original form differ in the way the joint distribution of the random variables 
$(\TRV_{n, j}, \XRV_{n, j})_{(n,j) \in \N \times \{1, 2, \ldots, J\}}$
would be chosen.
Loosely speaking, in the case of \PINNs\ the originally proposed distribution for 
$(\TRV_{n, j}, \XRV_{n, j})_{(n,j) \in \N \times \{1, 2, \ldots, J\}}$
would be based on drawing a finite number of samples of the random variable $(\Tv, \Xv)$ and then having the random variable $(\TRV_{n, j}, \XRV_{n, j})_{(n,j) \in \N \times \{1, 2, \ldots, J\}}$ be randomly chosen among those samples.
In the case of \DGMs\ the original proposition would be to choose $(\TRV_{n, j}, \XRV_{n, j})_{(n,j) \in \N \times \{1, 2, \ldots, J\}}$ independent and identically distributed.
Implementations of \PINNs\ and \DGMs\ that employ more sophisticated optimization methods, such as the \Adam\ \SGD\ optimization method, can be found in the next section.

\endgroup

\section{Implementation of PINNs}
\label{sec:PINNs}

\begin{figure}[!ht]
	\centering
	\includegraphics[width=\linewidth]{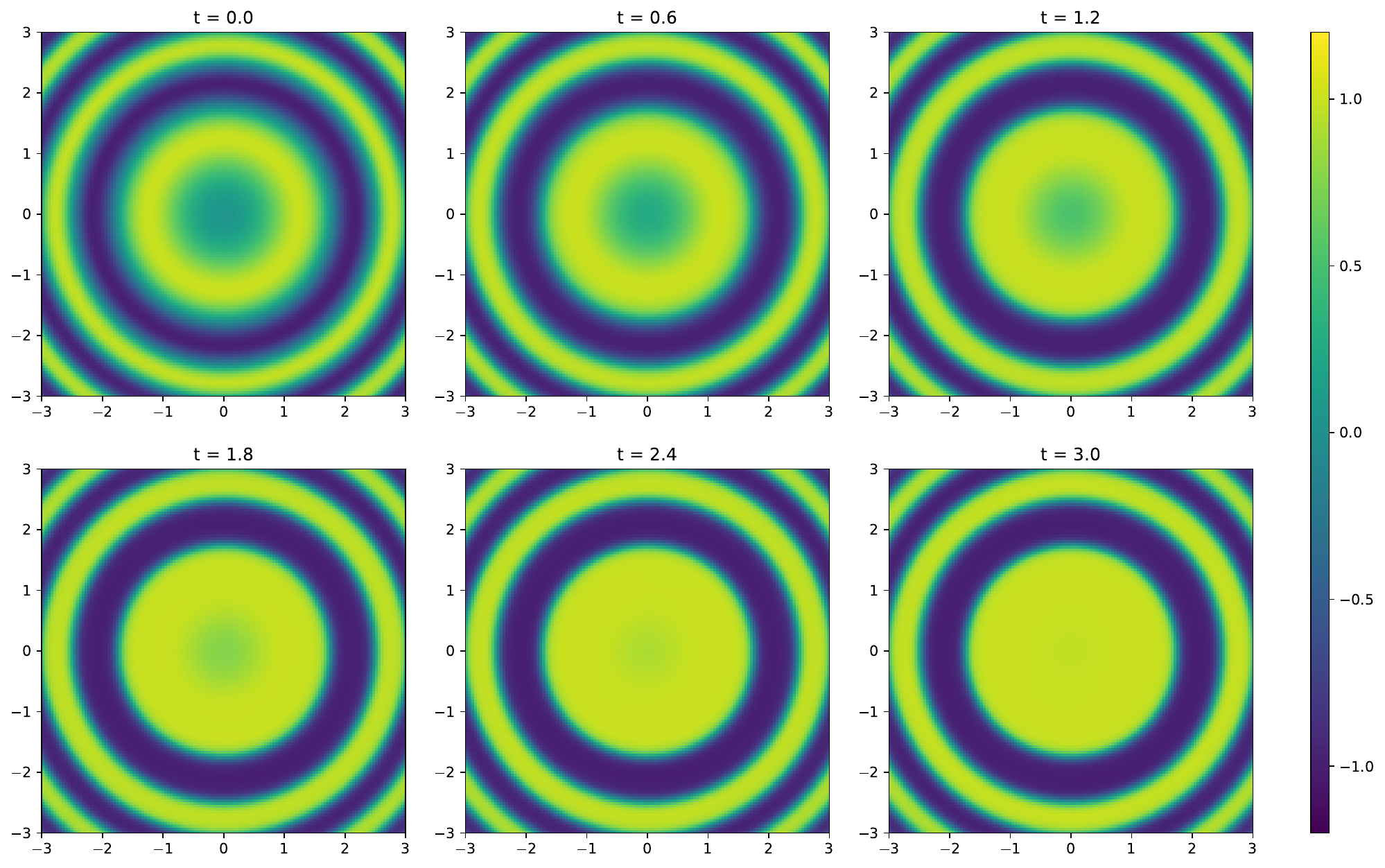}
	\caption{\label{fig:pinn}
  \cfclear
  Plots for the functions
  $[-3,3]^2\ni x\mapsto U(t,x)\in\R$, where
  $t\in\{0,0.6,1.2,1.8, 2.4, 3\}$
  and where
  $U\in C([0,3]\times\R^{2},\R)$
  is an approximation of the function 
  $u\in C^{1,2}([0,3]\times\R^{2},\R)$
  which satisfies for all $t\in[0,3]$, $x\in\R^{2}$ that
  $\bpr{\tfrac{\partial u}{\partial t}}(t,x) = \frac1{200}(\Delta_x u)(t,x) + u(t,x) - [u(t,x)]^3$
  and
  $u(0,x)=\sin(\pnorm2{x}^2)$
  computed by means of the \PINN\ method 
  as implemented in \cref{lst.pinn}
  \cfload.
  }
\end{figure}

In \cref{lst.pinn} below we present a simple implementation of the \PINN\ method,
as explained in \cref{sec:PINNS_derivation} above, for finding an approximation
of a solution $u\in C^{1,2}([0,3]\times\R^2)$ of the two-dimensional
Allen--Cahn-type semilinear heat equation
\begin{align}
  \bpr{\tfrac{\partial u}{\partial t}}(t,x) = \tfrac1{200}(\Delta_x u)(t,x)+u(t,x) - [u(t,x)]^3
\end{align}
with $u(0,x)=\sin(\pnorm2{x}^2)$ for $t\in[0,3]$, $x\in\R^2$.
This implementation follows the original proposal in 
Raissi et al.~\cite{raissi2019physics} in that it
first chooses 20000 realizations 
of the random variable $(\mathcal T,\mathcal X)$, where $\mathcal T$ is continuous
uniformly distributed on $[0,3]$ and where $\mathcal X$ is normally distributed
on $\R^2$
with mean $0\in\R^2$ and covariance $4\idMatrix_2\in\R^{2\times 2}$
(cf.\ \cref{def:identityMatrix}).
It then trains a fully connected feed-forward \ann\ with $4$ hidden layers
(with 50 neurons on each hidden layer)
and using the swish activation function with parameter $1$ (cf.\ \cref{sec:swish}).
The training uses batches of size 256 with each batch chosen from the 20000
realizations of the random variable $(\mathcal T,\mathcal X)$ which were picked
beforehand.
The training is performed using the \Adam\ \SGD\ optimization method
(cf.\ \cref{sect:adam}).
A plot of the resulting approximation of the solution $u$ after
20000 training steps is shown in \cref{fig:pinn}.

\begingroup
\filelisting{lst.pinn}{code/pinn.py}{
  \cfclear
  A simple implementation
  in \textsc{PyTorch} of the \PINN\ method, computing
  an approximation of the function $u\in C^{1,2}(\br{0,3}\times\R^{2},\R)$
  which satisfies for all $t\in\br{0,2}$, $x\in\R^{2}$ that
  $\bpr{\tfrac{\partial u}{\partial t}}(t,x)= \tfrac1{200}(\Delta_x u)(t,x) + u(t,x) - \br{u(t,x)}^3$
  and
  $u(0,x)=\sin(\pnorm2{x}^2)$ 
  \cfload. The plot created by this code is shown in \cref{fig:pinn}.
}
\endgroup

\section{Implementation of DGMs}
\label{sec:DGM}

\begin{figure}[!ht]
	\centering
	\includegraphics[width=\linewidth]{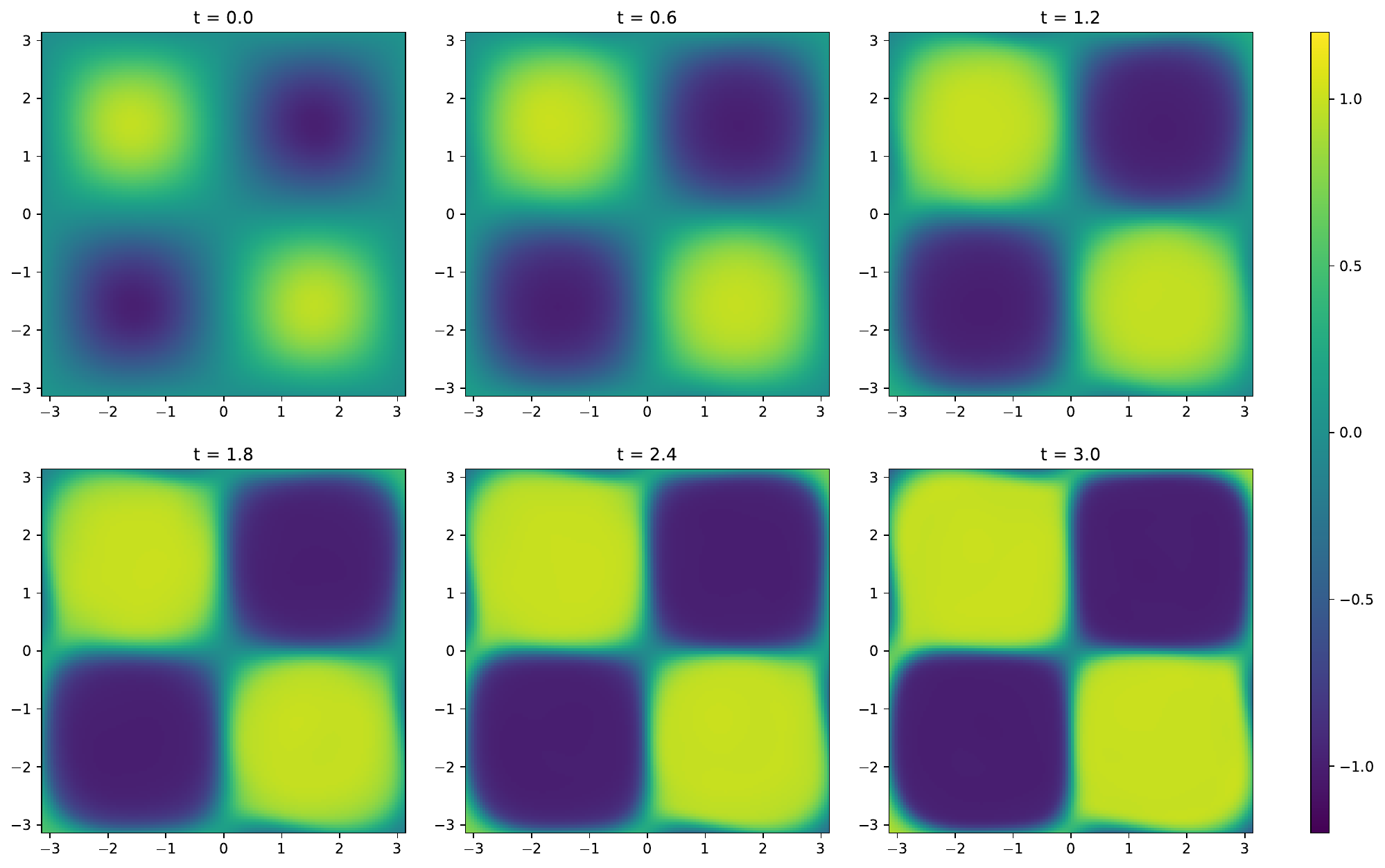}
	\caption{\label{fig:dgm}
  \cfclear
  Plots for the functions  
  $[-\pi,\pi]^2\ni x\mapsto U(t,x)\in\R$, where
  $t\in\{0,0.6,1.2,1.8,2.4,3\}$  
  and where
  $U\in C([0,3]\times\R^{2},\R)$
  is an approximation of the function 
  $u\in C^{1,2}([0,3]\times\R^{2},\R)$ which
  satisfies for all $t\in[0,3]$, $x=(x_1,x_2)\in\R^{2}$ that
  $u(0,x)=\sin(x_1)\sin(x_2)$ and
  $\bpr{\tfrac{\partial u}{\partial t}}(t,x) = \tfrac1{200}(\Delta_x u)(t,x) + u(t,x) - [u(t,x)]^3$
  computed by means of \cref{lst.dgm}
  \cfload.
  }
\end{figure}

In \cref{lst.dgm} below we present a simple implementation of the \DGM,
as explained in \cref{sec:PINNS_derivation} above, for finding an approximation
for a solution $u\in C^{1,2}([0,3]\times\R^2)$ of the two-dimensional
Allen--Cahn-type semilinear heat equation
\begin{align}
  \bpr{\tfrac{\partial u}{\partial t}}(t,x) = \tfrac1{200}(\Delta_x u)(t,x)+u(t,x) - [u(t,x)]^3
\end{align}
with $u(0,x)=\sin(x_1)\sin(x_2)$ for $t\in[0,3]$, $x=(x_1,x_2)\in\R^2$.
As originally proposed in Sirignano \& Spiliopoulos \cite{Sirignano2018dgm},
this implementation 
chooses for each training step a batch of 256 realizations 
of the random variable $(\mathcal T,\mathcal X)$, where $\mathcal T$ is continuously
uniformly distributed on $[0,3]$ and where $\mathcal X$ is normally distributed
on $\R^2$
with mean $0\in\R^2$ and covariance $4\idMatrix_2\in\R^{2\times 2}$
(cf.\ \cref{def:identityMatrix}).
Like the \PINN\ implementation in \cref{lst.pinn},
it trains a fully connected feed-forward \ann\ with $4$ hidden layers
(with 50 neurons on each hidden layer) and using the swish activation function with parameter $1$
(cf.\ \cref{sec:swish}).
The training is performed using the \Adam\ \SGD\ optimization method (cf.\ \cref{sect:adam}).
A plot of the resulting approximation of the solution $u$ after
30000 training steps is shown in \cref{fig:dgm}.

\filelisting{lst.dgm}{code/dgm.py}{
  \cfclear
  A simple implementation
  in \textsc{PyTorch} of the deep Galerkin method, computing
  an approximation of the function $u\in C^{1,2}(\br{0,3}\times\R^{2},\R)$
  which satisfies for all $t\in\br{0,3}$, $x=(x_1,x_2)\in\R^{2}$ that
  $\bpr{\tfrac{\partial u}{\partial t}}(t,x) = \tfrac1{200}( \Delta_x u)(t,x) + u(t,x) - \br{u(t,x)}^3$
  and
  $u(0,x)=\sin(x_1)\sin(x_2)$ 
  \cfload. The plot created by this code is shown in \cref{fig:dgm}.
}

%% file: parts/Deep_Komogorov_method.tex
\cchapter{Deep Kolmogorov methods (DKMs)}{sect:deepKolmogorov}

The \PINNs\ and the \DGMs\ presented in \cref{subsec:dgm} do, on the one hand, not exploit a lot of structure of the underlying \PDE\  in the process of setting up the associated stochastic optimization problems and have as such the key advantage to be very widely applicable deep learning methods for \PDEs.
On the other hand, deep learning methods for \PDEs\ that in some way exploit the specific structure of the considered \PDE\ problem often result in more accurate approximations (cf., \eg, Beck et al.~\cite{beck2020overview} and the references therein).
In particular, there are several deep learning approximation methods in the literature which exploit in the process of setting up stochastic optimization problems that the \PDE\ itself admits a stochastic representation.
In the literature there are a lot of deep learning methods which are based on such stochastic formulations of \PDEs\ and therefore have a strong link to stochastic analysis and formulas of the Feynman--Kac-type (cf., \eg, \cite{Germain2021,Pham2021,Hure2020,EHanJentzen2017Science,EHanJentzen17,BeckWeinanJentzen2019} and the references therein).

The schemes in Beck et al.~\cite{BeckJafaari21}, which we refer to as \DKMs, belong to the simplest of such deep learning methods for \PDEs.
In this chapter we present in  \cref{sec:dkm_RV,sec:dkm_RF,sec:FC,sec:dkm_PDE} theoretical considerations leading to
a reformulation of heat \PDE\ problems as stochastic optimization problems (see \cref{prop:heat_min} below),
we use these theoretical considerations to derive \DKMs\ in the specific case of heat equations in \cref{sec:dkm_derivation},
and we present an implementation of \DKMs\ in the case of a simple two-dimensional heat equation in \cref{sec:impl_dKM}.

\cref{sec:dkm_RV,sec:dkm_RF} are slightly modified extracts from Beck et al.~\cite{BeckBeckerGrohsJaafariJentzen2018arXiv},
\cref{sec:FC} is inspired by Beck et al.~\cite[Section 2]{beck2021nonlinear}, and
\cref{sec:dkm_PDE,sec:dkm_derivation} are inspired by Beck et al.~\cite{BeckBeckerGrohsJaafariJentzen2018arXiv}.

\section
{Stochastic optimization problems for expectations of random variables}
\label{sec:dkm_RV}

\begin{athm}{lemma}{lemma:minimizingProperty}
 Let $(\Omega,\cF,\P)$ be a probability space 
 and let $X\colon \Omega\to\R$ be a random 
 variable with $\E[\abs{X}^2] < \infty$. Then 
 \begin{enumerate}[label=(\roman{*})]
  \item\label{it:minimizingProperty_1} 
  it holds for all $y\in\R$ that 
 \begin{equation}\label{eq:lemma_orthogonality_property}
  \bExp{ \abs{ X - y }^2}
  = 
  \bExp{ \abs{ X - \EXp X }^2 } + \abs{\Exp X-y}^2
  ,
 \end{equation}
  \item\label{it:minimizingProperty_2}
  there exists a unique $z\in\R$ 
  such that 
  \begin{equation}\label{eq:lemma_minimizer_existence}
   \bExp{ \abs{ X - z }^2 } 
   =
   \inf_{y\in\R} \bExp{ \abs{ X - y }^2 }
   , 
  \end{equation}
  and 
  \item\label{it:minimizingProperty_3}
  it holds that 
  \begin{equation}\label{eq:lemma_minimizer_uniqueness}
   \bExp{ \abs{ X - \Exp{X} }^2 }
   =
   \inf_{y\in\R} \bExp{ \abs{ X - y }^2 }
   .  
  \end{equation}
 \end{enumerate}
\end{athm}

\begin{aproof}
  \Nobs that
  \cref{L2_distance}
\proves[ep] \cref{it:minimizingProperty_1}. 
 \Nobs that \cref{it:minimizingProperty_1} \proves[ep] 
 \cref{it:minimizingProperty_2,it:minimizingProperty_3}. 
\end{aproof}

\section
{Stochastic optimization problems for expectations of random fields}
\label{sec:dkm_RF}

\begin{athm}{prop}{proposition:minimizingProperty}
Let 
  $d\in\N$,
  $ a \in \R $, 
  $b\in (a,\infty)$, 
 let 
  $(\Omega,\cF,\P)$ be a probability space,  
 let 
  $ X = ( X_x )_{ x \in [a,b]^d } \colon [a,b]^d \times \Omega \to \R $ 
  be
  $ ( \mathcal{B}( [a,b]^d ) \otimes \mathcal{F} ) $/$ \mathcal{B}( \R ) $-measurable, 
 assume 
  for every $x\in [a,b]^d$ that $\E[\abs{X_x}^2]<\infty$, 
 and 
 assume 
  that $[a,b]^d\ni x\mapsto \E[X_x]\in\R$ is continuous. 
 Then 
 \begin{enumerate}[label=(\roman{*})]
  \item\label{it:minimizingProperty_prop_1} 
  there exists a unique
  $u\in C([a,b]^d,\R)$ such that 
  \begin{equation}\label{eq:existenceAndUniquenessOfMinimizer}
   \int_{[a,b]^d} \bExp{ \abs{ X_x - u(x) }^2 } \,\diff x
   = 
   \inf_{v\in C([a,b]^d,\R)} \bbbpr{\int_{[a,b]^d} \bExp{ \abs{ X_x - v(x) }^2 } \,\diff x}
  \end{equation} 
  and
  \item\label{it:minimizingProperty_prop_2} 
  it holds for all $x\in [a,b]^d$ that $u(x)=\E[X_x]$.
 \end{enumerate}
\end{athm}

\begin{aproof}
\Nobs that \cref{it:minimizingProperty_1} 
in \cref{lemma:minimizingProperty} 
and the assumption that
for all
$x \in [a,b]^d$ it holds that $\E[ \abs{X_x}^2 ] < \infty $
\prove that for every function $ u \colon [a,b]^d \to \R $ 
and every $x\in [a,b]^d$ it holds that
\begin{equation}
\label{eq:minimizingPropertyQuantified}
  \bExp{ \abs{ X_x - u(x) }^2}
  = 
  \bExp{ \abs{ X_x - \E[X_x] }^2 } + \abs{ \E[X_x]- u(x) }^2
  .
\end{equation}
Fubini's theorem (see, \eg, Klenke~\cite[Theorem 14.16]{Klenke14}) 
\hence \proves that for all
$ u \in C([a,b]^d, \R) $ it holds that
\begin{equation}
\label{eq:minimizingPropertyQuantified_2}
  \int_{ [a,b]^d }
  \bExp{ \abs{ X_x - u(x) }^2} \, \diff x
  = 
  \int_{ [a,b]^d }
  \bExp{ \abs{ X_x - \E[X_x] }^2 } \, \diff x 
  + 
  \int_{ [a,b]^d }
  \abs{ \E[X_x]- u(x) }^2 \, \diff x 
  .
\end{equation}
This \proves that
\begin{align}
  \begin{split}
   &\int_{[a,b]^d} 
    \bExp{ \abs{ X_x - \E[X_x] }^2 } \, \diff x
  \\&\geq
    \inf_{v\in C([a,b]^d,\R)} 
    \pr*{
      \int_{[a,b]^d}
      \bExp{ \abs{ X_x - v(x) }^2 } \, \diff x
    }
  \\&=
    \inf_{v\in C([a,b]^d,\R)} 
    \pr*{
      \int_{ [a,b]^d }
      \bExp{ \abs{ X_x - \E[X_x] }^2 } \, \diff x 
      + 
      \int_{ [a,b]^d }
      \abs{ \E[X_x]- v(x) }^2 \, \diff x
    }
  \end{split}
\end{align}  
The assumption that 
$ [a,b]^d \ni x \mapsto \E[ X_x ] \in \R $ 
is continuous \hence \proves that
\begin{align}
\begin{split}
 \int_{[a,b]^d} 
  \bExp{ \abs{ X_x - \E[X_x] }^2 } \, \diff x
&\geq 
  \inf_{v\in C([a,b]^d,\R)} 
  \pr*{
    \int_{ [a,b]^d }
    \bExp{ \abs{ X_x - \E[X_x] }^2 } \, \diff x 
  }
\\&=
  \int_{ [a,b]^d }
  \bExp{ \abs{ X_x - \E[X_x] }^2 } \, \diff x 
  .
\end{split} 
\end{align}
\Hence that
\begin{equation}
\label{eq:minimizingPropertyQuantified_3}
  \int_{[a,b]^d} 
  \bExp{ \abs{ X_x - \E[X_x] }^2 } \, \diff x
  =
  \inf_{v\in C([a,b]^d,\R)} 
  \pr*{
    \int_{[a,b]^d}
    \bExp{ \abs{ X_x - v(x) }^2 } \, \diff x
  } 
  .
\end{equation}
The fact that the function 
$[a,b]^d\ni x\mapsto \E[X_x]\in\R$ is continuous \hence
\proves that there exists $u\in C([a,b]^d,\R)$ 
such that
\begin{equation}
\label{eq:minimizingPropertyQuantified_4}
  \int_{[a,b]^d} \bExp{ \abs{ X_x - u(x) }^2 } \,\diff x
  = 
  \inf_{v\in C([a,b]^d,\R)} 
  \pr*{
    \int_{[a,b]^d} \bExp{ \abs{ X_x - v(x) }^2 } \, \diff x
  } .
\end{equation}
\Moreover
  \cref{eq:minimizingPropertyQuantified_2}
  and~\cref{eq:minimizingPropertyQuantified_3}
\prove that for all $ u \in C([a,b]^d , \R) $ 
with
\begin{equation}
  \int_{[a,b]^d} 
  \bExp{ \abs{ X_x - u(x) }^2 } \, \diff x
  = 
  \inf_{v\in C([a,b]^d,\R)} 
  \pr*{
    \int_{[a,b]^d}\bExp{ \abs{ X_x - v(x) }^2 } \, \diff x
  } 
\end{equation}
it holds that 
\begin{align}
\begin{split}
 &\int_{[a,b]^d} 
  \bExp{ \abs{ X_x - \E[X_x] }^2 } \, \diff x
\\&=
  \inf_{v\in C([a,b]^d,\R)} 
  \pr*{
    \int_{[a,b]^d}
    \bExp{ \abs{ X_x - v(x) }^2 } \, \diff x
  }
  =
  \int_{ [a,b]^d }
  \bExp{ \abs{ X_x - u(x) }^2} \, \diff x
\\&= 
  \int_{ [a,b]^d }
  \bExp{ \abs{ X_x - \E[X_x] }^2 } \, \diff x 
  + 
  \int_{ [a,b]^d }
  \abs{ \E[X_x]- u(x) }^2 \, \diff x 
  .
\end{split}
\end{align}
\Hence
that for all $ u \in C([a,b]^d, \R) $ 
with
\begin{equation}
  \int_{[a,b]^d} 
  \bExp{ \abs{ X_x - u(x) }^2 } \, \diff x
  = 
  \inf_{v\in C([a,b]^d,\R)} 
  \pr*{
    \int_{[a,b]^d}\bExp{ \abs{ X_x - v(x) }^2 } \, \diff x
  } 
\end{equation}
it holds that 
\begin{equation}
  \int_{[a,b]^d} \abs{ \E[X_x] - u(x) }^2 \, \diff x = 0 
  .
\end{equation}
This and the assumption that
$ [a,b]^d \ni x \mapsto \E[ X_x ] \in \R $ 
is continuous \prove that for 
all  $ y \in [a,b]^d $,
$ u \in C([a,b]^d, \R) $ 
with
\begin{equation}
  \int_{[a,b]^d} 
  \bExp{ \abs{ X_x - u(x) }^2 } \, \diff x
  = 
  \inf_{v\in C([a,b]^d,\R)} 
  \pr*{
    \int_{[a,b]^d}\bExp{ \abs{ X_x - v(x) }^2 } \, \diff x
  } 
\end{equation}
it holds that 
$ u(y) = \E[ X_y ] $. 
Combining this with \cref{eq:minimizingPropertyQuantified_4}
\proves[ep] \cref{it:minimizingProperty_prop_1,it:minimizingProperty_prop_2}.
\end{aproof}

\section{Feynman--Kac formulas}
\label{sec:FC}

\subsection{Feynman--Kac formulas providing existence of solutions}

\cfclear
\begin{athm}{lemma}{lem:lebesgue}[A variant of Lebesgue's theorem on dominated convergence]
  Let $(\Omega,\mathcal F,\P)$ be a probability space,
  for every 
    $n\in\N_0$
  let $X_n\colon\Omega\to\R$ be a random variable,
  assume 
    for all
      $\eps\in(0,\infty)$
    that
    \begin{equation}
      \limsup_{n\to\infty}\P(\abs{X_n-X_0}>\eps)=0
      ,
    \end{equation}
  let $Y\colon\Omega\to\R$ be a random variable
  with $\bExp{\abs{Y}}<\infty$,
  and assume
    for all
      $n\in\N$
    that
      $\P(\abs{X_n}\leq Y)=1$.
  Then
  \begin{enumerate}[(i)]
    \item \llabel{it2}
    it holds that
      $\limsup_{n\to\infty}\bExp{\abs{X_n-X_0}}=0$,
    \item \llabel{it1} 
    it holds that
      $\bExp{\abs{X_0}}<\infty$,
    and
    \item \llabel{it3}
    it holds that
      $\limsup_{n\to\infty}\babs{\Exp{X_n}-\Exp{X_0}}=0$.
  \end{enumerate}
\end{athm}
\begin{aproof}
  Note that, \eg,
    the variant of Lebesgue's theorem on dominated convergence in Klenke~\cite[Corollary~6.26]{Klenke14}
  \proves[ep]
    \cref{lem:lebesgue.it1,lem:lebesgue.it2,lem:lebesgue.it3}.
\end{aproof}

\cfclear
\begin{athm}{prop}{lem:classical_solution_smooth_case}
	Let $ T \in (0,\infty) $, 
		$ d,m \in \N $, 
		$ B \in \R^{d\times m} $, 
		$ \varphi \in C^{2}(\R^d,\R) $ 
	satisfy  
  \begin{equation}
  \label{lem:classical_solution_smooth_case:ass1}
    \textstyle
		\sup_{x\in\R^d} 
		\bbr{
			\sum_{i,j=1}^d 
			\bpr{ \abs{ \varphi(x) } + \babs{ \bpr{\tfrac{\partial }{\partial x_i}\varphi}(x) } + \babs{\bpr{\tfrac{\partial^2}{\partial x_i\partial x_j} \varphi}(x)} } 
		} < \infty , 
  \end{equation}
	let $ (\Omega,\mc F,\P) $ be a probability space, 
	let $ Z \colon \Omega \to \R^m $ be a standard normal random variable, 
	and let $ u \colon [0,T] \times \R^d \to \R $ satisfy for all 
		$ t \in [0,T] $, 
		$ x \in \R^d $ 
	that 
		\begin{equation} \label{classical_solution_smooth_case:solution_formula}
		u(t,x) = \bExp{\varphi(x+\sqrt{t}B Z)}. 
		\end{equation} 
	Then  
	\begin{enumerate}[label=(\roman{*})]
		\item\label{classical_solution_smooth_case:item1}
		it holds that 
			$ u \in C^{1,2}([0,T] \times \R^d, \R) $ 
		and 
		\item\label{classical_solution_smooth_case:item2}
		it holds for all 
			$ t \in [0,T] $, 
			$ x \in \R^d $ 
		that 
			\begin{equation} \label{classical_solution_smooth_case:claim}
			\bpr{\tfrac{\partial u}{\partial t}}(t,x) = \tfrac12 \operatorname{Trace}\pr[\big]{B\transpose B(\operatorname{Hess}_x u)(t,x)}
			\end{equation} 
	\end{enumerate}
	\cfout.
\end{athm}

\begin{aproof}
	Throughout this proof, 
	let 
  \begin{equation}
    e_1 = ( 1, 0, \ldots, 0), e_2 = ( 0, 1, \ldots, 0), \ldots, e_m = (0, \ldots, 0, 1) \in \R^m
  \end{equation}
	and for every
  $ t\in [0,T] $, 
  $ x \in \R^d $
  let 
		$ \psi_{t,x} \colon \R^m \to \R $, 
	satisfy for all 
		$ y \in \R^m $
	that 
		$ \psi_{t,x}(y) = \varphi(x+\sqrt{t}By) $. 
	Note that the assumption that $ \varphi \in C^2(\R^d,\R) $, 
	the chain rule, 
  \cref{lem:lebesgue},
  and
  \cref{lem:classical_solution_smooth_case:ass1}
  \prove that 
	\begin{enumerate}[label=(\Roman{*})] 
		\item \label{classical_solution_smooth_case:proof_item1} 
		for all 
			$ x \in \R^d $ 
		it holds that 
			$ (0,T] \ni t \mapsto u(t,x) \in \R $ 
		is differentiable, 
		\item \label{classical_solution_smooth_case:proof_item2} 
		for all  
			$ t \in [0,T] $
		it holds that 
			$ \R^d \ni x \mapsto u(t,x) \in \R $ is twice differentiable, 		
		\item \label{classical_solution_smooth_case:proof_item3} 
		for all 
			$ t \in (0,T] $, 
			$ x \in \R^d $ 
		it holds that 
			\begin{equation} 
			\bpr{\tfrac{\partial u}{\partial t}}(t,x) 
			= 
			\bExp{\scp[\big]{(\nabla\varphi)(x+\sqrt{t}BZ),\tfrac{1}{2\sqrt{t}}BZ}}, 
			\end{equation} 
		and 
		\item \label{classical_solution_smooth_case:proof_item4} 
		for all 
			$ t \in [0,T] $, 
			$ x \in \R^d $
		it holds that 
			\begin{equation} 
			(\operatorname{Hess}_x u)(t,x) 
			= 
			\bExp{(\operatorname{Hess} \varphi)(x+\sqrt{t}BZ) }
			\end{equation} 
		\end{enumerate}
  \cfload.
  \Nobs that 
    \cref{classical_solution_smooth_case:proof_item3,classical_solution_smooth_case:proof_item4}, 
    the assumption that $ \varphi \in C^2(\R^d,\R) $, 
    the assumption that 
    \begin{equation}
    \begin{split} 
      \textstyle
    \sup_{x\in\R^d} \br[\big]{\sum_{i,j=1}^d \bpr{\babs{\varphi(x)} + \abs{\bpr{\frac{\partial}{\partial x_i}\varphi}(x)} + \babs{ \bpr{\frac{\partial^2}{\partial x_i\partial x_j}\varphi}(x) }}} < \infty,
    \end{split}
    \end{equation}
    the fact that $ \EXP{\pnorm2{Z}} < \infty $, 
    and \cref{lem:lebesgue}
  \prove that 
		\begin{equation}
		\begin{split} 
		(0,T] \times \R^d \ni (t,x) \mapsto \bpr{\tfrac{\partial u}{\partial t}}(t,x) \in \R
		\end{split}
		\end{equation}
		and
		\begin{equation}
		\begin{split} 
		[0,T] \times \R^d \ni (t,x) \mapsto (\operatorname{Hess}_x u)(t,x) \in \R^{d\times d}
		\end{split}
		\end{equation}
		are continuous
		\cfload. 
	\Moreover 
    \cref{classical_solution_smooth_case:proof_item4} 
    and the fact that 
      for all 
        $ X \in \R^{m \times d} $, 
        $ Y \in \R^{d \times m} $ 
      it holds that 
        $ \operatorname{Trace}(X Y) = \operatorname{Trace}(Y X) $	
	\prove that for all 
		$ t \in (0,T] $, 
		$ x \in \R^d $ 
	it holds that 
		\begin{equation} 
		\begin{split}
		& 
		\tfrac12 \operatorname{Trace}\bpr{B \transpose B (\operatorname{Hess}_x u)(t,x)} 
		= 
		\Exp{\tfrac12\operatorname{Trace}\bpr{ B \transpose B (\operatorname{Hess} \varphi)(x+\sqrt{t}BZ) }}
		\\
		& = 
		\tfrac{1}{2} \, \Exp{\operatorname{Trace}\bpr{ \transpose B (\operatorname{Hess} \varphi)(x+\sqrt{t}BZ) B } } 
		= 
		\tfrac{1}{2} \, \Exp{\smallsum\limits_{k=1}^m \displaystyle\scp{e_k, \transpose B  (\operatorname{Hess} \varphi)(x+\sqrt{t}BZ) B e_k} }
		\\		
		& = 
		\tfrac{1}{2} \, \Exp{\smallsum\limits_{k=1}^m \displaystyle\scp{B e_k ,  (\operatorname{Hess} \varphi)(x+\sqrt{t}BZ) B e_k} }
		= 
		\tfrac{1}{2} \, \Exp{\smallsum\limits_{k=1}^m \displaystyle\varphi^{\prime\prime}( x + \sqrt{t}BZ )(B e_k, B e_k) }
		\\
		& = 
		 \tfrac{1}{2t} \, \Exp{\smallsum\limits_{k=1}^m \displaystyle(\psi_{t,x})^{\prime\prime}(Z)(e_k,e_k) } 
		= 
		\tfrac{1}{2t} \, \Exp{\smallsum\limits_{k=1}^m \displaystyle\bpr{\tfrac{\partial^2 }{\partial y_k^2}\psi_{t,x}}(Z) }
		= 
		\tfrac{1}{2t} \,
		\Exp{ (\Delta \psi_{t,x})(Z)}
		\end{split}
		\end{equation} 
		\cfload.
	The assumption that $ Z \colon \Omega \to \R^m $ is a standard normal random variable 
  and integration by parts \hence \prove that for all 
		$ t \in (0,T] $, 
		$ x \in \R^d $ 
	it holds that  
		\begin{equation} 
		\begin{split} 
		& \tfrac12 \operatorname{Trace}\bpr{B \transpose B (\operatorname{Hess}_x u)(t,x)} 
		\\
		& = 
		\frac{1}{2t} 
		\int_{\R^m} (\Delta\psi_{t,x})(y)  \br*{ \frac{\exp\bpr{\tfrac{\scp {y,y}}{2}}}{(2\pi)^{\nicefrac{m}{2}}}}\!\,\diff y 
		= 
		\frac{1}{2t} 
		\int_{\R^m} \scp{ (\nabla\psi_{t,x})(y), y } \br*{ \frac{\exp\bpr{-\frac{\scp{y,y}}{2}}}{(2\pi)^{\nicefrac{m}{2}}} }\!\,\diff y 
		\\
		& = 
		\frac{1}{2\sqrt{t}} 
		\int_{\R^m}\scp*{ \transpose B(\nabla \varphi)(x+\sqrt{t}By), y } \br*{\frac{\exp\bpr{-\frac{\scp{ y, y}}{2}}}{(2\pi)^{\nicefrac{m}{2}}}}\!\,\diff y 
		\\
		& =
		\frac{1}{2\sqrt{t}} \, 
		\bExp{ \scp{ \transpose B(\nabla\varphi)(x+\sqrt{t}BZ),Z } }
		= 
		\bExp{ \bscp{ (\nabla\varphi)(x+\sqrt{t}BZ), \tfrac{1}{2\sqrt{t}} B Z} }.
		\end{split}
		\end{equation} 
	\Cref{classical_solution_smooth_case:proof_item3} \hence \proves that for all 
		$ t \in (0,T] $, 
		$ x \in \R^d $ 
	it holds that 
		\begin{equation} \label{classical_solution_smooth_case:t_positive}
		\bpr{\tfrac{\partial u}{\partial t}}(t,x) = \tfrac12 \operatorname{Trace}\bpr{B B^{*} (\operatorname{Hess}_x u)(t,x)}. 
		\end{equation} 
	The fundamental theorem of calculus \hence \proves that for all 
		$ t,s \in (0,T] $, 
		$ x \in \R^d $ 
	it holds that 
		\begin{equation} 
		\begin{split}
		u(t,x) - u(s,x) 
		& = 
		\int_{s}^t \bpr{\tfrac{\partial u}{\partial t}}(r,x) \,\diff r 
		= \int_{s}^t \tfrac12 \operatorname{Trace}\bpr{BB^{*}(\operatorname{Hess}_x u)(r,x)} \,\diff r . 
		\end{split}
		\end{equation} 
	The fact that $ [0,T] \times \R^d \ni (t,x) \mapsto (\operatorname{Hess}_x u)(t,x) \in \R^{d\times d} $ 
	is continuous \hence \proves for all 
		$ t \in (0,T] $, 
		$ x \in \R^d $ 
	that 	
		\begin{equation} 
		\frac{u(t,x) - u(0,x)}{t} 
		= 
		\lim_{ s \searrow 0 } 
		\br*{ \frac{u(t,x)-u(s,x)}{t} } 
		= 
		\frac{1}{t} 
		\int_0^t \tfrac12 \operatorname{Trace}\bpr{ B B^{*} (\operatorname{Hess}_x u)(r,x) } \,\diff r. 
		\end{equation} 
	This and the fact that 
		$ [0,T]\times\R^d \ni (t,x) \mapsto (\operatorname{Hess}_x u)(t,x) \in \R^{d\times d} $ 
	is continuous \prove that for all 
		$ x \in \R^d $ 
	it holds that 	
		\begin{equation} \label{classical_solution_smooth_case:t_0}
		\begin{split}
		& \limsup_{ t \searrow 0 } \abs*{ \frac{u(t,x) - u(0,x)}{t} - \tfrac12 \operatorname{Trace}\bpr{ B B^{*} (\operatorname{Hess}_x u)(0,x) } } 
		\\
		& 
		\leq  
		\limsup_{ t \searrow 0 } \br*{ \frac{1}{t} \int_0^t \abs*{ \tfrac12 \operatorname{Trace}\bpr{ B B^{*} (\operatorname{Hess}_x u)(s,x) } - \tfrac12 \operatorname{Trace}\bpr{ B B^{*} (\operatorname{Hess}_x u)(0,x) } } \,\diff s }
		\\
		& 
		\leq 
		\limsup_{ t \searrow 0 } \br*{ \sup_{s\in [0,t]} \abs*{ \tfrac12 \operatorname{Trace}\bbpr{ B B^{*} \bpr{(\operatorname{Hess}_x u)(s,x) - (\operatorname{Hess}_x u)(0,x) } } } } 
		= 0. 
		\end{split}
		\end{equation} 
	\Cref{classical_solution_smooth_case:proof_item1} \hence \proves that for all $ x \in \R^d $ it holds that $ [0,T] \ni t \mapsto u(t,x) \in \R $ is differentiable. Combining this with \cref{classical_solution_smooth_case:t_0} and \cref{classical_solution_smooth_case:t_positive} ensures that for all 
		$ t \in [0,T] $, 
		$ x \in \R^d $ 
	it holds that 
		\begin{equation} \label{classical_solution_smooth_case:end_of_proof}
		\bpr{\tfrac{\partial u}{\partial t}}(t,x) = \tfrac12 \operatorname{Trace}\bpr{ B B^{*} (\operatorname{Hess}_x u)(t,x) }. 
		\end{equation}
    This 
    and the fact that 
      $ [0,T] \times \R^d \ni (t,x) \mapsto (\operatorname{Hess}_x u)(t,x) \in \R^{d\times d} $ is continuous 
    \prove[ep] \cref{classical_solution_smooth_case:item1}. 
    \Nobs that  
      \cref{classical_solution_smooth_case:end_of_proof} 
    \proves[ep] 
      \cref{classical_solution_smooth_case:item2}.
\end{aproof}

\begin{adef}{def:brownian}[Standard Brownian motions]
	Let $(\Omega,\mc F,\P)$ be a probability space.
	Then we say that $W$ is an $m$-dimensional $\P$-standard Brownian motion
	(we say that $W$ is a $\P$-standard Brownian motion, we say that
	$W$ is a standard Brownian motion)
	if and only if there exists $T\in(0,\infty)$ such that
	\begin{enumerate}[(i)]
		\item it holds that $m\in\N$,
		\item it holds that $W\colon[0,T]\times\Omega\times\R^m$ is a function,
		\item it holds for all $\omega\in\Omega$ that $[0,T]\ni s\mapsto W_s(\omega)\in\R^m$ is continuous,
		\item it holds for all $\omega\in\Omega$ that $W_0(\omega)=0\in\R^m$,
		\item it holds for all $t_1\in[0,T]$, $t_2\in[0,T]$ with $t_1< t_2$ that
			$\Omega\ni \omega\mapsto (t_2-t_1)^{-\nicefrac 12}(W_{t_2}(\omega)-W_{t_1}(\omega))\in\R^m$
			is a standard normal random variable, and
		\item it holds for all $n\in\{3,4,5,\dots\}$, $t_1,t_2,\dots,t_n\in[0,T]$ with
			$t_1\leq t_2\leq \dots\leq t_n$ that
			$W_{t_2}-W_{t_1},\,W_{t_3}-W_{t_2},\,\ldots,\, W_{t_n}-W_{t_{n-1}}$
			are independent.
	\end{enumerate}
\end{adef}

\filelisting{Brownian_Motion}{code/brownian_motion.py}{
\textsc{Python} code producing four trajectories of a one-dimensional standard Brownian motion.
}

\begin{figure}[t]
	\centering
	\includegraphics[scale=0.95]{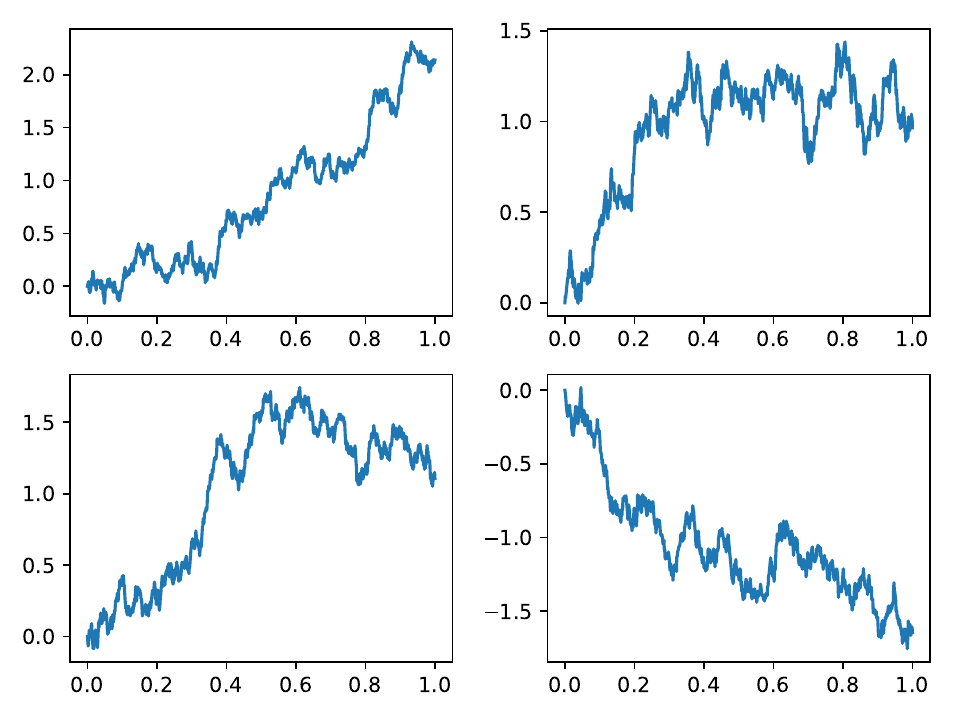}
	\caption{Four trajectories of a one-dimensional standard Brownian motion}
\end{figure}

\cfclear
\begin{athm}{cor}{cor:classical_solution_brownian_motion}
	Let 
    $ T \in (0,\infty) $, 
		$ d,m \in \N $, 
		$ B \in \R^{d\times m} $, 
		$ \varphi \in C^{2}(\R^d,\R) $ 
	satisfy  
  \begin{equation}
    \textstyle
  \sup_{x\in\R^d}
		\br[\big]{
			\sum_{i,j=1}^d
			\pr[\big]{ 
				\abs{ \varphi(x) } 
				+ 
				\babs{ \bpr{\tfrac{ \partial }{\partial x_i}\varphi}(x) } 
				+ 
				\babs{\bpr{\tfrac{\partial^2}{\partial x_i\partial x_j} \varphi}(x)} 
			}
		 } < \infty
     , 
  \end{equation}
	let $ (\Omega,\mc F,\P) $ be a probability space, 
	let $ W \colon [0,T] \times \Omega \to \R^m $ be a standard Brownian motion\cfadd{def:brownian}, 
	and let $ u \colon [0,T] \times \R^d \to \R $ satisfy for all 
		$ t \in [0,T] $, 
		$ x \in \R^d $ 
	that 
		\begin{equation} 
		u(t,x) = \bExp{\varphi(x+BW_t)}\ifnocf.
		\end{equation} 
	\cfout[.]
	Then  
	\begin{enumerate}[label=(\roman{*})]
		\item \label{classical_solution_brownian_motion:item1}
		it holds that 
			$ u \in C^{1,2}([0,T] \times \R^d, \R) $ 
		and 
		\item \label{classical_solution_brownian_motion:item2}
		it holds for all 
			$ t \in [0,T] $, 
			$ x \in \R^d $ 
		that 
			\begin{equation} 
			\bpr{\tfrac{\partial u}{\partial t}}(t,x)
			=
			\tfrac12 \operatorname{Trace}\pr[\big]{B\transpose B(\operatorname{Hess}_x u)(t,x)}
			\end{equation} 
	\end{enumerate}
	\cfout.
\end{athm} 

\begin{aproof}
	First, \nobs that 
		the assumption that $ W\colon [0,T]\times\Omega\to\R^m $ is a standard Brownian motion 
	\proves  that
	for all 
		$ t \in [0,T] $, 
		$ x \in \R^d $
	it holds that 
		\begin{equation} 
		u(t,x) = \Exp{ \varphi(x+BW_t) } = \Exp{ \varphi\pr*{x+\sqrt{t}B \frac{W_T}{\sqrt{T}}} }.
		\end{equation} 
		The fact that 
		  $ \frac{W_T}{\sqrt{T}} \colon \Omega \to \R^m $ is a standard normal random variable 
		and \cref{lem:classical_solution_smooth_case} 
		\hence 
	\prove[ep] 
		\cref{classical_solution_brownian_motion:item1,classical_solution_brownian_motion:item2}. 
\end{aproof}

\subsection{Feynman--Kac formulas providing uniqueness of solutions}

\cfclear
\begin{athm}{lemma}{lem:vitali}[A special case of Vitali's convergence theorem]
	Let $(\Omega,\cF,\P)$ be a probability space,
	let $X_n\colon\Omega\to\R$, $n\in\N_0$, be random variables
		with 
    \begin{equation}
      \llabel{eq:conv}
      \P\bpr{\textstyle\limsup_{n\to\infty} \abs{X_n-X_0}=0}=1,
    \end{equation}
	and let $p\in(1,\infty)$ satisfy $\sup_{n\in\N} \EXp{\abs{X_n}^p}<\infty$.
	Then
	\begin{enumerate}[(i)]
		\item \llabel{it1}
		it holds that $\limsup_{n\to\infty} \EXP{\abs{X_n-X_0}}=0$,
		\item \llabel{it2}
		it holds that
		$\bExp{\abs{X_0}}<\infty$,
		and
		\item \llabel{it3}
    it holds that $\limsup_{n\to\infty} \babs{\EXp{X_n}-\EXp{X_0}}=0$.
	\end{enumerate}
\end{athm}
\begin{aproof}
  First, \nobs that
    the assumption that
      \begin{equation}
      \begin{split} 
      \sup_{n\in\N}\bExp{\abs{X_n}^p}<\infty
      \end{split}
      \end{equation}
    and, \eg, the consequence of de la Vall\'ee-Poussin's theorem in Klenke~\cite[Corollary~6.21]{Klenke14}
  \prove that
    $\{X_n\colon n\in\N\}$ is uniformly integrable.
    This,
    \lref{eq:conv},
    and Vitali's convergence theorem in, \eg, Klenke~\cite[Theorem~6.25]{Klenke14}
  \prove[ep] \cref{lem:vitali.it1,lem:vitali.it2}.
  \Nobs that
    \cref{lem:vitali.it1,lem:vitali.it2}
  \prove[epi]
    \lref{it3}.
\end{aproof}

\cfclear
\begin{athm}{prop}{lem:feynman_kac_representation_inhomogeneous_equation_smooth_solution}
Let $d\in\N$, 
	$T,\rho\in(0,\infty)$,
	$f\in C([0,T]\times\R^d,\R)$,
let $u\in C^{1,2}([0,T]\times\R^d,\R)$
	have at most polynomially growing partial derivatives,
assume for all
	$t\in[0,T]$,
	$x\in\R^d$ 
that
\begin{equation}
	\label{eq:feynman_kac.pde}
	\bpr{\tfrac{\partial u}{\partial t}}(t,x) = \rho\,(\Delta_x u)(t,x) + f(t,x),
\end{equation}
let $(\Omega,\mc F,\P)$ be a probability space, 
and let $W \colon [0,T]\times\Omega \to \R^d$ be a standard Brownian\cfadd{def:brownian} motion
\cfload.
Then it holds for all
	$t\in[0,T]$, 
	$x\in\R^d$ 
that 
\begin{equation} \label{eq:feynman_kac_representation_inhomogeneous_equation_smooth_solution:claim}
	u(t,x) 
	=
	\Exp{u(0,x+\sqrt{2\rho} W_t) + \int_0^t f(t-s, x+\sqrt{2\rho} W_s)\,\diff s}
	. 
\end{equation} 
\end{athm}

\begin{aproof}
Throughout this proof,
let $D_1\colon [0,T]\times\R^d\to\R$ satisfy for all
	$t\in[0,T]$,
	$x\in\R^d$
that
	\begin{equation}
    D_1(t,x)=\bpr{\tfrac{\partial u}{\partial t}}(t,x),
  \end{equation}
let $D_2=(D_{2,1},D_{2,2},\dots,D_{2,d})\colon [0,T]\times\R^d\to \R^d$ satisfy for all
	$t\in[0,T]$,
	$x\in\R^d$
that
	$D_2(t,x)=(\nabla_x u)(t,x)$,
let $H=(H_{i,j})_{i,j\in\{1,2,\dots,d\}}\colon [0,T]\times\R^d\to\R^{d\times d}$ satisfy for all
	$t\in[0,T]$,
	$x\in\R^d$
that
  \begin{equation}
    H(t,x)=(\operatorname{Hess}_x u)(t,x),
  \end{equation}
let 
	$\gamma \colon \R^d \to \R$ 
satisfy for all
	$z \in \R^d$
that 
	\begin{equation} 
	\gamma(z) 
	= 
	(2\pi)^{-\nicefrac{d}{2}} \exp\pr[\big]{-\tfrac{\pnorm2{z}^2}{2}}
	,
	\end{equation} 
and let $v_{t,x} \colon [0,t] \to \R$, $t\in [0,T]$, $x\in\R^d$, satisfy for all
	$t\in[0,T]$, 
	$x\in\R^d$,
	$s\in[0,t]$ 
that 
	\begin{equation} 
	v_{t,x}(s) 
	= 
	\EXP{u(s,x+\sqrt{2\rho} W_{t-s})}
	\end{equation} 
\cfload.
\Nobs that 
	the assumption that $W$ is a standard Brownian motion 
\proves that for all 
	$t\in(0,T]$, 
	$s\in[0,t)$
it holds that $(t-s)^{-\nicefrac12}W_{t-s}\colon\Omega\to\R^d$ is a standard normal random variable.
This \proves that for all
	$t\in(0,T]$, 
	$x\in\R^d$, 
	$s\in[0,t)$ 
it holds that	
	\begin{eqsplit}  
	v_{t,x}(s) 
	&= \EXP{u(s,x+\sqrt{2\rho(t-s)}(t-s)^{-\nicefrac12} W_{t-s})}
	\\&= \int_{\R^d} u(s,x+\sqrt{2\rho(t-s)}z ) \gamma(z)\,\diff z. 
	\end{eqsplit} 
  The assumption that $u$ has at most polynomially growing partial derivatives, 
  the fact that $(0,\infty)\ni s \mapsto \sqrt{s}\in (0,\infty)$ is differentiable, 
	the chain rule, 
	and Vitali's convergence theorem 
	\hence 
\prove that for all 
	$t\in(0,T]$, 
	$x\in\R^d$, 
	$s\in[0,t)$ 
it holds that 
	$v_{t,x}|_{[0,t)} \in C^1([0,t),\R)$ 
and  
\begin{equation} \label{eq:feynman_kac_representation_inhomogeneous_equation_smooth_solution:differentiating_v_tx}
\begin{split}
	(v_{t,x})'(s) 
	= 
	\int_{\R^d} \br*{
		D_1(s,x+\sqrt{2\rho(t-s)}z) 
		+ 
		\scp*{D_2(s,x+\sqrt{2\rho(t-s)}z),\tfrac{-\rho z}{\sqrt{2\rho(t-s)}}} 
	} \gamma(z)	\,\diff z
\end{split}
\end{equation}
\cfload.
\Moreover the fact that for all
	$z\in\R^d$ 
it holds that 
	$(\nabla \gamma)(z) = -\gamma(z)z$ 
\proves that for all
	$t\in(0,T]$, 
	$x\in\R^d$, 
	$s\in[0,t)$ 
it holds that
	\begin{equation} 
		\label{eq:feynman_kac.1}
	\begin{split}
	&\int_{\R^d}
		\scp*{D_2(s,x+\sqrt{2\rho(t-s)}z),\tfrac{-\rho z}{\sqrt{2\rho(t-s)}}} 
	\gamma(z)	\,\diff z
	\\&= 
	\int_{\R^d}
		\scp*{D_2(s,x+\sqrt{2\rho(t-s)}z),\tfrac{\rho (\nabla \gamma)(z)}{\sqrt{2\rho(t-s)}}} 
	\,\diff z
	\\&= 
	\tfrac{\rho}{\sqrt{2\rho(t-s)}}\sum\nolimits_{i=1}^d\br*{
	\int_{\R^d}D_{2,i}(s,x+\sqrt{2\rho(t-s)}z)(\tfrac{\partial\gamma}{\partial z_i})(z_1,z_2,\dots,z_d)
	\,\diff z}
  .
	\end{split} 
	\end{equation}
\Moreover
	integration by parts 
\proves that for all
	$t\in(0,T]$,
	$x\in\R^d$,
	$s\in[0,t)$,
	$i\in\{1,2,\dots,d\}$,
	$a\in\R$,
	$b\in(a,\infty)$
it holds that
\begin{equation}
\begin{split}
	&\int_a^b D_{2,i}(s,x+\sqrt{2\rho(t-s)}(z_1,z_2,\dots,z_d))(\tfrac{\partial\gamma}{\partial z_i})(z_1,z_2,\dots,z_d)\,\diff z_i
	\\&=
	\br*{D_{2,i}(s,x+\sqrt{2\rho(t-s)}(z_1,z_2,\dots,z_d))\gamma(z_1,z_2,\dots,z_d)}_{z_i=a}^{z_i=b}
	\\&\quad-\int_a^b \sqrt{2\rho(t-s)}H_{i,i}(s,x+\sqrt{2\rho(t-s)}(z_1,z_2,\dots,z_d))\gamma(z_1,z_2,\dots,z_d)\,\diff z_i
	.
\end{split}
\end{equation}
  The assumption that
		$u$ has at most polynomially growing derivatives
	\hence
\proves that for all
	$t\in(0,T]$,
	$x\in\R^d$,
	$s\in[0,t)$,
	$i\in\{1,2,\dots,d\}$
it holds that
\begin{equation}
\begin{split}
	&\int_\R D_{2,i}(s,x+\sqrt{2\rho(t-s)}(z_1,z_2,\dots,z_d))\bpr{\tfrac{\partial\gamma}{\partial z_i}}(z_1,z_2,\dots,z_d)\,\diff z_i
	\\&=
	-\sqrt{2\rho(t-s)}\int_\R H_{i,i}(s,x+\sqrt{2\rho(t-s)}(z_1,z_2,\dots,z_d))\gamma(z_1,z_2,\dots,z_d)\,\diff z_i
	.
\end{split}
\end{equation}
Combining
	this
with
	\eqref{eq:feynman_kac.1}
	and Fubini's theorem
\proves that for all
	$t\in(0,T]$,
	$x\in\R^d$,
	$s\in[0,t)$
it holds that
\begin{equation}
\begin{split}
	&\int_{\R^d}
		\scp*{D_2(s,x+\sqrt{2\rho(t-s)}z),\tfrac{-\rho z}{\sqrt{2\rho(t-s)}}} 
	\gamma(z)	\,\diff z
	\\&= 
	-\rho\sum\nolimits_{i=1}^d \int_{\R^d} H_{i,i}(s,x+\sqrt{2\rho(t-s)}(z))\gamma(z)\,\diff z
	\\&= 
	-\int_{\R^d} \rho\operatorname{Trace}\pr[\big]{H(s,x+\sqrt{2\rho(t-s)}(z))}\gamma(z)\,\diff z
	.	
\end{split}
\end{equation}
  This,
	\eqref{eq:feynman_kac_representation_inhomogeneous_equation_smooth_solution:differentiating_v_tx},
	\eqref{eq:feynman_kac.pde},
	and the fact that 
		for all 
			$t\in(0,T]$, 
			$s\in[0,t)$
		it holds that
	  	$(t-s)^{-\nicefrac12}W_{t-s}\colon\Omega\to\R^d$ is a standard normal random variable
\prove that for all
	$t\in(0,T]$, 
	$x\in\R^d$, 
	$s\in[0,t)$ 
it holds that 
	\begin{align} 
	\notag(v_{t,x})^{\prime}(s) 
	&
	= 
	\int_{\R^d} \br[\big]{D_1(s, x+\sqrt{2\rho(t-s)}z) -\rho \Trace \pr[\big]{H(s, x+\sqrt{2\rho(t-s)} z)}} \gamma(z)\,\diff z
	\\
	&= 
	\int_{\R^d} f(s, x+\sqrt{2\rho(t-s)} z) \gamma(z)\,\diff z
	=
	\Exp{f(s, x + \sqrt{2\rho} W_{t-s})}. 
	\end{align} 
  The fact that $W_0=0$,
	the fact that 
		for all
			$t\in[0,T]$, 
			$x\in\R^d$ 
		it holds that 
			$v_{t,x}\colon[0,t]\to\R$ 
		is continuous,
	and the fundamental theorem of calculus 
	\hence 
\prove that for all
	$t\in[0,T]$, 
	$x\in\R^d$ 
it holds that 
	\begin{equation} 
	\begin{split}
	u(t,x) 
	& = \Exp{u(t, x + \sqrt{2\rho} W_{t-t})} 
	= v_{t,x}(t) 
	= v_{t,x}(0) + \int_0^t (v_{t,x})'(s)\,\diff s 
	\\
	&= \Exp{u(0, x + \sqrt{2\rho} W_t)} + \int_0^t \Exp{f(s, x + \sqrt{2\rho} W_{t-s})}\,\diff s. 
	\end{split}
	\end{equation} 
  Fubini's theorem 
  and the fact that $u$ and $f$ are at most polynomially growing 
  \hence 
\prove[epi]
  \eqref{eq:feynman_kac_representation_inhomogeneous_equation_smooth_solution:claim}. 
\end{aproof}

\begingroup
\providecommandordefault{\Wv}{\mathcal{W}}

\begin{athm}{cor}{prop:feynman}
Let 
$ d \in \N $, 
$ T, \rho \in (0,\infty) $, 
$ \varrho = \sqrt{ 2 \rho T } $, 
$ a \in \R $, 
$ b \in (a,\infty) $, 
let 
$ \varphi \colon \R^d \to \R $ be a function,
let 
$
  u 
  \in C^{1,2}([0,T]\times\R^d,\R)
$ 
have at most polynomially growing partial derivatives,
assume for all
$t\in [0,T]$, $x\in\R^d$ that 
$ u(0,x) = \varphi(x) $ and 
\begin{equation}
\label{eq:differentialu0}
  \bpr{\tfrac{ \partial u}{\partial t}}(t,x) 
  = 
  \rho \, (\Delta_x u)(t,x),
\end{equation}
let 
$ (\Omega, \mathcal{F}, \P ) $ be a probability space, 
and let
$ \Wv \colon \Omega \to \R^d $ 
be a standard normal random variable.
Then 
\begin{enumerate}[label=(\roman{*})]
\item \label{it:feynman.1}
it holds that $ \varphi \colon \R^d \rightarrow \R $ is twice continuously differentiable
with at most polynomially growing partial derivatives and
\item \label{it:feynman.2}
it holds for all $ x \in \R^d $ that 
$ 
  u(T,x) = \EXP{\varphi( \varrho \Wv + x )}
$.
\end{enumerate}
\end{athm}
\begin{aproof}
	\Nobs that 
		the assumption that 
			$u\in C^{1,2}([0,T]\times\R^d,\R)$ has at most polynomially growing partial derivatives 
		and the fact that 
			for all
				$x\in\R^d$ 
			it holds that 
				$\varphi(x)=u(0,x)$ 
	\prove[epi] \cref{it:feynman.1}.
	\Moreover \cref{lem:feynman_kac_representation_inhomogeneous_equation_smooth_solution}
	\proves[epi] \cref{it:feynman.2}.
\end{aproof}
\endgroup

\begin{adef}{def:cont_convolution}[Continuous convolutions]
  Let $d\in\N$
  and let $f\colon\R^d\to\R$ and $g\colon\R^d\to\R$ be
  $\mc B(\R^d)$/$\mc B(\R)$-measurable.
  Then we denote by 
  \begin{multline}
    \textstyle
    f \convfunc g\colon \Bigl\{x\in\R^d\colon \min\bigl\{
    \int_{\R^d} \max\{0,f(x-y)g(y)\}\,\diff y,\\\textstyle
    {-{\int_{\R^d} \min\{0,f(x-y)g(y)\}\,\diff y}}
    \bigr\}<\infty\Bigr\}\to[-\infty,\infty]
  \end{multline}
  the function which satisfies for all
    $x\in\R^d$
  with
  \begin{equation}
    \textstyle
    \min\bigl\{
      \int_{\R^d} \max\{0,f(x-y)g(y)\}\,\diff y,\allowbreak
      -{\int_{\R^d} \min\{0,f(x-y)g(y)\}\,\diff y}
    \bigr\}<\infty
  \end{equation}
  that
  \begin{equation}
    (f \convfunc g)(x)
    =
    \int_{\R^d} f(x-y)g(y)\,\diff y
    .
  \end{equation}
\end{adef}

\cfclear
\begin{exercise}{ex:heat_equation_via_convolution}
  Let $d\in\N$,
    $T\in(0,\infty)$,
  for every $\sigma\in(0,\infty)$
  let $\gamma_\sigma\colon \R^d\to\R$ satisfy for all
    $x\in\R^d$
  that
  \begin{equation}
    \gamma_\sigma(x)
    =
		(2\pi\sigma^2)^{-\frac d2}\exp\biggl(\frac{-\pnorm2x^2}{2\sigma^2}\biggr)
		,
  \end{equation}
	and for every
	  $\rho\in(0,\infty)$, 
	  $\varphi\in C^2(\R^d,\R)$ 
   with
    $\sup_{x\in\R^d} \bigl[
      \sum_{i,j=1}^d \bigl(
        \abs{\varphi(x)}
        +\abs{(\frac\partial{\partial x_i}\varphi)(x)}
        +\abs{(\frac{\partial^2}{\partial x_i\partial x_j}\varphi)(x)}
      \bigr)
    \bigr]<\infty$
  let $u_{\rho,\varphi}\colon [0,T]\times\R^d\to\R$ satisfy for all 
    $t\in(0,T]$,
    $x\in\R^d$
  that
  \begin{equation}
    u_{\rho,\varphi}(0,x)=\varphi(x)
    \qquad\text{and}\qquad
    u_{\rho,\varphi}(t,x)=(\varphi \convfunc \gamma_{\sqrt{2t\rho}})(x)
  \end{equation}
  \cfload.
  Prove or disprove the following statement:
  For all 
  	$\rho\in(0,\infty)$,
   $\varphi\in C^2(\R^d,\R)$ 
  with
  $\sup_{x\in\R^d} \bigl[
    \sum_{i,j=1}^d \bigl(
      \abs{\varphi(x)}
      +\abs{(\frac\partial{\partial x_i}\varphi)(x)}
      +\abs{(\frac{\partial^2}{\partial x_i\partial x_j}\varphi)(x)}
    \bigr)
  \bigr]<\infty$
      it holds for all
      $t\in(0,T)$,
      $x\in\R^d$
    that $u_{\rho,\varphi}\in C^{1,2}([0,T]\times\R^d,\R)$ 
      and
    \begin{equation}
      \bpr{\tfrac{\partial u_{\rho,\varphi}}{\partial t}}(t,x)
      =
      \rho\,(\Delta_x u_{\rho,\varphi})(t,x)
      .
    \end{equation}
\end{exercise}

\begin{exercise}{ex:gaussian_fourier}
  Prove or disprove the following statement: 
  For every
    $x\in\R$
  it holds that
  \begin{equation}
    e^{-x^2/2}
    =
    \frac{1}{\sqrt{2\pi}}\biggl[
      \int_\R e^{-t^2/2}e^{-\mathrm{i}xt}\, \diff t
    \biggr]
    .
  \end{equation}  
\end{exercise}

\cfclear
\begin{exercise}{ex:heat_equation_via_fourier}
  Let 
    $d\in\N$,
    $T\in(0,\infty)$,
  for every $\sigma\in(0,\infty)$
  let $\gamma_\sigma\colon \R^d\to\R$ satisfy for all
    $x\in\R^d$
  that
  \begin{equation}
    \gamma_\sigma(x)
    =
    (2\pi\sigma^2)^{-\frac d2}\exp\biggl(\frac{-\pnorm2x^2}{2\sigma^2}\biggr),
  \end{equation}
	for every
    $\varphi\in C^2(\R^d,\R)$ with
    $\sup_{x\in\R^d} \bigl[
      \sum_{i,j=1}^d \bigl(
        \abs{\varphi(x)}
        +\abs{(\frac\partial{\partial x_i}\varphi)(x)}
        +\abs{(\frac{\partial^2}{\partial x_i\partial x_j}\varphi)(x)}
      \bigr)
    \bigr]<\infty$
  let $u_\varphi\colon [0,T]\times\R^d\to\R$ satisfy for all 
    $t\in(0,T]$,
    $x\in\R^d$
  that
  \begin{equation}
    u_{\varphi}(0,x)=\varphi(x)
    \qquad\text{and}\qquad
		u_{\varphi}(t,x)=(\varphi \convfunc \gamma_{\sqrt{2t}})(x)
    \cfadd{def:cont_convolution}
		,
  \end{equation}
  and
  for every 
      $i=(i_1,\dots,i_d)\in\N^d$
  let $\psi_i\colon\R^d\to\R$ satisfy for all
    $x=(x_1,\dots,x_d)\in\R^d$
  that
  \begin{equation}
    \psi_i(x)
    =
    2^{\frac d2}\Biggl[\prod_{k=1}^d \sin(i_k\pi x_k)\Biggr]
  \end{equation}
  \cfload.
  Prove or disprove the following statement:
  For all 
    $i=(i_1,\dots,i_d)\in\N^d$,
    $t\in[0,T]$,
    $x\in\R^d$
  it holds that
  \begin{equation}
    u_{\psi_i}(t,x)
    =
    \exp\bigl(-\pi^2\bigl[{\textstyle\sum}_{k=1}^d \abs{i_k}^2\bigr]t\bigr)\psi_i(x)
    .
  \end{equation}
\end{exercise}

\cfclear
\begin{exercise}{ex:heat_equation_unique_solution}
  Let 
    $d\in\N$,
    $T\in(0,\infty)$,
  for every 
    $\sigma\in(0,\infty)$
  let $\gamma_\sigma\colon \R^d\to\R$ satisfy for all
    $x\in\R^d$
  that
  \begin{equation}
    \gamma_\sigma(x)
    =
		(2\pi\sigma^2)^{-\frac d2}\exp\biggl(\frac{-\pnorm2x^2}{2\sigma^2}\biggr)
		,
  \end{equation}
  and for every
        $i=(i_1,\dots,i_d)\in\N^d$
  let $\psi_i\colon\R^d\to \R$ satisfy for all
    $x=(x_1,\dots,x_d)\in\R^d$
  that
  \begin{equation}
    \psi_i(x)
    =
		2^{\frac d2}\Biggl[\prod_{k=1}^d \sin(i_k\pi x_k)\Biggr]
  \end{equation}
  \cfload.
  Prove or disprove the following statement:
  For every 
    $i=(i_1,\dots,i_d)\in\N^d$,
    $s\in[0,T]$,
    $y\in\R^d$
  and every function
    $u\in C^{1,2}([0,T]\times\R^d,\R)$ 
  with at most polynomially growing partial derivatives
  which satisfies for all
    $t\in(0,T)$,
    $x\in\R^d$
 that 
    $u(0,x)=\psi_i(x)$
  and
  \begin{equation}
    \bpr{\tfrac{\partial u}{\partial t}}(t,x)
    =
    (\Delta_x u)(t,x)
  \end{equation}
  it holds that
  \begin{equation}
    u(s,y)
    =
    \exp\bigl(-\pi^2\bigl[{\textstyle\sum}_{k=1}^d \abs{i_k}^2\bigr]s\bigr)\psi_i(y)
    .
  \end{equation}
\end{exercise}

\section
{Reformulation of PDE problems as stochastic optimization problems}
\label{sec:dkm_PDE}

The proof of the next result, \cref{prop:heat_min} below, is based 
on an application of \cref{proposition:minimizingProperty} and \cref{lem:feynman_kac_representation_inhomogeneous_equation_smooth_solution}.
A more general result than \cref{prop:heat_min} with a detailed proof can, \eg, be found in Beck et al.~\cite[Proposition~2.7]{BeckBeckerGrohsJaafariJentzen2018arXiv}.

\begingroup

\providecommandordefault{\Xv}{\mathcal{X}}
\providecommandordefault{\Wv}{\mathcal{W}}

\begin{athm}{prop}{prop:heat_min}
Let 
$ d \in \N $, 
$ T, \rho \in (0,\infty) $, 
$ \varrho = \sqrt{ 2 \rho T } $, 
$ a \in \R $, 
$ b \in (a,\infty) $, 
let 
$ \varphi \colon \R^d \to \R $ be a function,
let 
$
  u 
  \in C^{1,2}([0,T]\times\R^d,\R)
$ 
have at most polynomially growing partial derivatives,
assume for all
$t\in [0,T]$, $x\in\R^d$ that 
$ u(0,x) = \varphi(x) $ and 
\begin{equation}
\label{eq:differentialu}
  \bpr{\tfrac{ \partial u}{\partial t}}(t,x) 
  = 
  \rho \, (\Delta_x u)(t,x), 
\end{equation}
let 
$ (\Omega, \cF, \P ) $ be a probability space, 
let
$ \Wv \colon \Omega \to \R^d $ 
be a standard normal random variable, 
let 
$ \Xv \colon \Omega \to [a,b]^d $ be 
a continuously uniformly distributed random variable, 
and
assume that $ \Wv $ and $ \Xv $ are independent.
Then 
\begin{enumerate}[label=(\roman{*})]
\item\label{it:varphiContinuous_new} 
it holds that $ \varphi \colon \R^d \rightarrow \R $ is twice continuously differentiable
with at most polynomially growing partial derivatives,
\item \label{it:exuniqueSolutions} 
there exists a unique continuous function 
$ U \colon [a,b]^d \to \R $ such that 
\begin{equation}
  \EXP{ \abs{ \varphi( \varrho \Wv + \Xv ) - U(\Xv) }^2 } 
  = 
  \inf_{v\in C([a,b]^d,\R)} \EXP{ \abs{ \varphi( \varrho \Wv  + \Xv ) - v(\Xv) }^2 } ,
\end{equation} 
and
\item 
\label{it:UboundaryCond}
it holds for every $ x \in [a,b]^d $ that $ U(x) = u(T,x) $.
\end{enumerate}
\end{athm}

\begin{aproof}
First, \nobs that 
	\eqref{eq:differentialu}, 
	the assumption that $\Wv$ is a standard normal random variable, 
	and \cref{prop:feynman} 
\prove that for all
$x \in \R^d$ 
it holds that 
$\varphi\colon\R^d\to\R$ is twice continuously differentiable with at most polynomially growing partial derivatives and
\begin{equation} \label{eq:heat_min:fk_rep}
u(T,x) 
=\EXP{ u(0, \varrho\Wv + x) }
=\EXP{ \varphi(\varrho\Wv + x) }
. 
\end{equation}
\Moreover 
	the assumption that $\Wv$ is a standard normal random variable, 
	the fact that $\varphi$ is continuous, 
	and the fact that $\varphi$ has at most polynomially growing partial derivatives and is continuous
\prove that 
\begin{enumerate}[(I)]
\item\label{it:heat_min:proof_item:measurability} it holds that 
$[a,b]^d \times \Omega \ni (x,\omega) \mapsto \varphi(\varrho \Wv(\omega) + x) \in \R$ 
is $(\Borel([a,b]^d)\otimes\mathcal{F})$/$\Borel(\R)$-measurable and 
\item it holds for all
$x \in [a,b]^d$ 
that $\E[\abs{\varphi(\varrho\Wv+x)}^2] < \infty$. 
\end{enumerate}
\cref{proposition:minimizingProperty} and \eqref{eq:heat_min:fk_rep} hence ensure that 
\begin{enumerate}[(A)] 
\item \label{it:heat_min:proof_item:ex_min} there exists a unique continuous function $U \colon [a,b]^d \to \R$ which satisfies that 
\begin{equation} 
\int_{[a,b]^d} \EXP{ \abs{\varphi( \varrho \Wv + x ) - U(x)}^2 } \,\diff x
= 
\inf_{v\in C([a,b]^d,\R)} \bbbpr{\int_{[a,b]^d} \EXP{ \abs{ \varphi( \varrho \Wv + x ) - v(x) }^2 } \,\diff x}
\end{equation} 
and 	
\item\label{it:heat_min:proof_item:min_is_exp} it holds for all
$x \in [a,b]^d$ 
that 
$U(x) = u(T,x)$. 
\end{enumerate}
\Moreover the assumption that $\Wv$ and $\Xv$ are independent, 
\cref{it:heat_min:proof_item:measurability}, 
and the assumption that $\Xv$ is continuously uniformly distributed on $[a,b]^d$ 
\prove that for all
$v \in C([a,b]^d,\R)$ 
it holds that 
\begin{equation} \label{eq:heat_min:expectations}
\EXP{ \abs{\varphi(\varrho \Wv + \Xv) - v(\Xv) }^2 }
= 
\frac{1}{(b-a)^d}
\int_{[a,b]^d} \EXP{\abs{\varphi(\varrho \Wv + x) - v(x)}^2}\,\diff x. 
\end{equation} 
Combining this with \cref{it:heat_min:proof_item:ex_min} \proves[epi] \cref{it:exuniqueSolutions}. 
\Nobs that 
  \cref{it:heat_min:proof_item:ex_min,it:heat_min:proof_item:min_is_exp} 
  and \eqref{eq:heat_min:expectations} 
\prove[epi]
  \cref{it:UboundaryCond}. 
\end{aproof}

While \cref{prop:heat_min} above recasts the solutions of the \PDE\ in
\eqref{eq:differentialu} at a particular point in time as the solutions
of a stochastic optimization problem, we can also derive from this
a corollary which shows that
the solutions of the \PDE\ over an entire timespan
are similarly the solutions of a stochastic optimization problem.

\begin{athm}{cor}{cor:kolmogorovtime}
  Let
  $ d \in \N $,
  $ T, \rho \in (0,\infty) $,
  $ \varrho = \sqrt{ 2 \rho } $,
  $ a \in \R $,
  $ b \in (a,\infty) $,
  let
  $ \varphi \colon \R^d \to \R $ be a function,
  let
  $
    u
    \in C^{1,2}([0,T]\times\R^d,\R)
  $
  be a function with at most polynomially growing partial derivatives which satisfies
  for all $t\in [0,T]$, $x\in\R^d$ that
  $ u(0,x) = \varphi(x) $ and
  \begin{equation}
    \bpr{\tfrac{ \partial u}{\partial t}}(t,x)
    =
    \rho \, (\Delta_x u)(t,x)
    ,
  \end{equation}
  let
  $ (\Omega, \mathcal{F}, \P ) $ be a probability space,
  let
  $ \Wv \colon \Omega \to \R^d $
  be a standard normal random variable,
  let $\tau\colon \Omega\to[0,T]$ be a continuously uniformly
  distributed random variable,
  let
  $ \Xv \colon \Omega \to [a,b]^d $ be
  a continuously uniformly distributed random variable,
  and
  assume that $ \Wv $, $ \tau $, and $\Xv$ are independent.
  Then
  \begin{enumerate}[(i)]
  \item \label{it:corheat.1}
  there exists a unique
  $ U \in C( [0,T]\times [a,b]^d , \R ) $ which satisfies that
  \begin{equation}
  \begin{split}
    \E\br[\big]{ \abs{ \varphi( \varrho\sqrt{\tau} \Wv + \Xv ) - U(\tau,\Xv) }^2 }
    =\!\!\!\!
    \inf_{v\in C([0,T]\times[a,b]^d,\R)}\!\!\!\! \E\br[\big]{ \abs{ \varphi( \varrho\sqrt{\tau}\Wv  + \Xv ) - v(\tau,\Xv) }^2 }
  \end{split}
  \end{equation}
  and
  \item \label{it:corheat.2}
  it holds for all $t\in[0,T]$, $x\in[a,b]^d$ that $ U(t,x) = u(t,x) $.
  \end{enumerate}
\end{athm}
\renewcommand{\d}[2]{\delta^{#1}_{#2}}
\newcommand{\dee}{\delta}
\newcommand{\e}[1]{\eps_{#1}}
\newcommand{\ee}{\eps}
\begin{aproof}
  Throughout this proof,
  let
    $F\colon C([0,T]\times[a,b]^d,\R)\to[0,\infty]$
  satisfy for all
    $v\in C([0,T]\times[a,b]^d,\R)$
  that
  \begin{equation}
    F(v)
    =
    \E\br[\big]{ \abs{ \varphi( \varrho\sqrt{\tau}\Wv  + \Xv ) - v(\tau,\Xv) }^2 }
    .
  \end{equation}
  \Nobs that
    \cref{prop:heat_min}
  \proves that for all
    $v\in C([0,T]\times[a,b]^d,\R)$,
    $s\in[0,T]$
  it holds that
  \begin{equation}
    \label{eq:corheat.2}
    \E\br[\big]{ \abs{ \varphi( \varrho\sqrt{s}\Wv  + \Xv ) - v(s,\Xv) }^2 }
    \geq
    \E\br[\big]{ \abs{ \varphi( \varrho\sqrt{s}\Wv  + \Xv ) - u(s,\Xv) }^2 }
    .
  \end{equation}
  \Moreover
    the assumption that
      $\Wv$, $\tau$, and $\Xv$ are independent,
    the assumption that
      $\tau\colon \Omega\to [0,T]$ is continuously uniformly distributed,
    and Fubini's theorem
  \prove that for all
    $v\in C([0,T]\times[a,b]^d,\R)$
  it holds that
  \begin{equation}
    \label{eq:corheat.1}
    F(v)
    =
    \E\br[\big]{ \abs{ \varphi( \varrho\sqrt{\tau}\Wv  + \Xv ) - v(\tau,\Xv) }^2 }
    =
    \int_{[0,T]}\E\br[\big]{ \abs{ \varphi( \varrho\sqrt{s}\Wv  + \Xv ) - v(s,\Xv) }^2 } \,\diff s
    .
  \end{equation}
    This
    and \eqref{eq:corheat.2}
  \prove that for all
    $v\in C([0,T]\times[a,b]^d,\R)$
  it holds that
  \begin{equation}
    F(v)
      \geq
      \int_{[0,T]}\E\bbr{\abs{\varphi(\varrho\sqrt{ s}\Wv+\Xv)-u(s,\Xv)}}\,\diff s
      .
  \end{equation}
  Combining
    this
  with
    \eqref{eq:corheat.1}
  \proves that for all
    $v\in C([0,T]\times[a,b]^d,\R)$
  it holds that
  $
    F(v)
    \geq
    F(u)
    $.
  \Hence that
  \begin{equation}
    \label{eq:corheat.3}
    F(u)
    =\!\!\!
    \inf_{v\in C([0,T]\times[a,b]^d,\R)}\!\!\! F(v)
    .
  \end{equation}
    This
    and \eqref{eq:corheat.1}
  \prove that for all
    $U\in C([0,T]\times [a,b]^d,\R)$
    with
    \begin{equation}
      F(U)  = \!\!\! \inf_{v\in C([0,T]\times[a,b]^d,\R)} \!\!\! F(v)
    \end{equation}
  it holds that
  \begin{equation}
    \int_{[0,T]}\E\bbr{\abs{\varphi(\varrho\sqrt{ s}\Wv+\Xv)-U(s,\Xv)}}\,\diff s
    =
    \int_{[0,T]}\E\bbr{\abs{\varphi(\varrho\sqrt{ s}\Wv+\Xv)-u(s,\Xv)}}\,\diff s
    .
  \end{equation}
    Combining
      this
    with
     \eqref{eq:corheat.2}
  \proves that for all
    $U\in C([0,T]\times [a,b]^d,\R)$
    with $F(U) = \inf_{v\in C([0,T]\times[a,b]^d,\R)} F(v)$
  there exists
    $A\subseteq[0,T]$
    with $\int_A 1\,\diff x=T$
  such that for all
    $s\in A$
  it holds that
  \begin{equation}
    \E\br[\big]{ \abs{ \varphi( \varrho\sqrt{s}\Wv  + \Xv ) - U(s,\Xv) }^2 }
    =
    \E\br[\big]{ \abs{ \varphi( \varrho\sqrt{s}\Wv  + \Xv ) - u(s,\Xv) }^2 }
    .
  \end{equation}
    \cref{prop:heat_min}
    \hence[therefore]
  \proves[de] that for all
    $U\in C([0,T]\times [a,b]^d,\R)$
    with $F(U) = \inf_{v\in C([0,T]\times[a,b]^d,\R)} F(v)$
  there exists
    $A\subseteq[0,T]$
    with $\int_A 1\,\diff x=T$
  such that for all
    $s\in A$
  it holds that
    $U(s)=u(s)$.
  The fact that
    $u\in C([0,T]\times[a,b]^d,\R)$
    \hence
  \proves that for all
    $U\in C([0,T]\times [a,b]^d,\R)$
    with $F(U) = \inf_{v\in C([0,T]\times[a,b]^d,\R)} F(v)$
  it holds that
    $U=u$.
  Combining
    this
  with
    \eqref{eq:corheat.3}
  \proves[ep]
    \cref{it:corheat.1,it:corheat.2}.
\end{aproof}
\endgroup

\section{Derivation of DKMs}
\label{sec:dkm_derivation}

\begingroup

\providecommandordefault{\DKMLossInfinite}{\mathfrak{L}}
\providecommandordefault{\LossFunction}{\defaultLossFunction}

\providecommandordefault{\d}{\defaultParamDim}

\newcommand{\localAnn}[1]{\RealV{#1}{0}{\defaultInputDim}{ \multdim_{\activation, l_1}, \multdim_{\activation, l_2}, \dots, \multdim_{\activation, l_h} , \id_{ \R } }}

\providecommandordefault{\Xv}{\mathcal{X}}
\providecommandordefault{\Wv}{\mathcal{W}}
\providecommandordefault{\XRV}{\mathfrak{X}}
\providecommandordefault{\WRV}{\mathfrak{W}}

In this section we present in the special case of the heat equation a rough derivation of the \DKMs\ introduced in Beck et al.~\cite{BeckJafaari21}.
This derivation will proceed along the analogous steps as the derivation of \PINNs\ and \DGMs\ in \cref{sec:PINNS_derivation}.
Firstly, 
we will employ \cref{prop:heat_min} to reformulate the \PDE\ problem under consideration as an infinite-dimensional stochastic optimization problem,
secondly, 
we will employ \anns\ to reduce the infinite-dimensional stochastic optimization problem to a finite-dimensional stochastic optimization problem, 
and thirdly, 
we will aim to approximately solve this finite-dimensional stochastic optimization problem by means of \SGD-type optimization methods.
We start by introducing the setting of the problem.
Let 
$ d \in \N $, 
$ T, \rho \in (0,\infty) $, 
$ a \in \R $, 
$ b \in (a,\infty) $, 
let 
$ \varphi \colon \R^d \to \R $ be a function,
let 
$
  u 
  \in C^{1,2}( [0,T] \times \R^d, \R )
$ 
have at most polynomially growing partial derivatives,
and
assume for all $ t \in [0,T] $, $ x \in \R^d $ that 
$ u(0,x) = \varphi(x) $ and 
\begin{equation}
\label{dkm_derivation:eq1}
  \bpr{\tfrac{ \partial u}{\partial t}}(t,x) 
  = 
  \rho \, (\Delta_x u)(t,x).
\end{equation}
In the framework described in the previous sentence, we think of $u$ as the unknown \PDE\ solution. 
The objective of this derivation is to develop deep learning methods which aim to approximate the unknown \PDE\ solution $u(T, \cdot)|_{[a,b]^d} \colon [a,b]^d \to \R$ at time $T$ restricted on $[a,b]^d$.

In the first step, we employ \cref{prop:heat_min} to recast the unknown target function $u(T, \cdot)|_{[a,b]^d} \colon [a,b]^d \to \R$ as the solution of an optimization problem.
For this 
let $ \varrho = \sqrt{ 2 \rho T } $, 
let
$ (\Omega, \cF, \P ) $ be a probability space, 
let
$ \Wv \colon \Omega \to \R^d $ 
be a standard normally distributed random variable, 
let 
$ \Xv \colon \Omega \to [a,b]^d $ be 
a continuously uniformly distributed random variable,  
assume that $ \Wv $ and $ \Xv $ are independent,
and
let $\DKMLossInfinite \colon C([a,b]^d,\R)\to[0,\infty]$
satisfy
for all
$v \in C([a,b]^d,\R)$
that
\begin{equation}
\label{dkm_derivation:eq2}
	\DKMLossInfinite(v)
=
	\EXP{ \abs{ \varphi( \varrho \Wv  + \Xv ) - v(\Xv) }^2 } 
.
\end{equation}
\cref{prop:heat_min} then ensures 
that the unknown target function $u(T, \cdot)|_{[a,b]^d} \colon [a,b]^d \to \R$ is the unique global 
minimizer of the function $\DKMLossInfinite \colon C([a,b]^d,\R)\to[0,\infty]$.
Minimizing $\DKMLossInfinite$ is, however, not yet amenable to numerical computations.

In the second step, we therefore reduce this infinite-dimensional stochastic optimization problem to a finite-dimensional stochastic optimization problem involving \anns.
Specifically, 
let
$\activation \colon \R \to \R$ be differentiable,
let
$h\in\N$,
$l_1,l_2,\dots,l_h,\defaultParamDim\in\N$ satisfy
$\defaultParamDim=l_1(d+1)+\br[\big]{\sum_{k=2}^h l_k(l_{k-1}+1)}+l_h+1$,
and let
 $\defaultLossFunction \colon\R^\defaultParamDim\to[0,\infty)$ 
 satisfy
for all $\theta\in\R^\defaultParamDim$ that
\begin{equation}
\label{dkm_derivation:eq3}
\begin{split}
	\defaultLossFunction( \theta )
&= 
	\DKMLossInfinite\pr[\Big]{ \pr[\big]{\localAnn{\theta}}|_{[a,b]^d} } \\
&=
	\EXPP{ \abs[\big]{ \varphi( \varrho \Wv  + \Xv ) - \localAnn{\theta}(\Xv) }^2 } 
\end{split}
\end{equation}
\cfload.
We can now compute an approximate minimizer of the function $\DKMLossInfinite$ 
by computing an approximate minimizer $\vartheta \in \R^{\defaultParamDim}$ of the function $\defaultLossFunction$
and employing the realization 
$ \pr[\big]{\localAnn{\theta}}|_{[a,b]^d} \in C([a,b]^d, \R)$
of the \ann\ associated to this approximate minimizer restricted on $[a,b]^d$ as an approximate minimizer of $\DKMLossInfinite$.

In the third step, we use \SGD-type methods to compute such an approximate minimizer of $\defaultLossFunction$.
We now sketch this in the case of the plain-vanilla \SGD\ optimization method (cf.\ \cref{def:SGD}).
Let 
	$\xi \in\R^\defaultParamDim$,
	$J \in \N$,
	$(\gamma_n)_{n\in\N}\subseteq[0,\infty)$, 
for every 
	$n \in \N$,
	$j \in \{1, 2, \ldots, J\}$
let
	$ \WRV_{n, j} \colon \Omega \to \R^d $ 
be a standard normally distributed random variable and
let 
	$ \XRV_{n, j} \colon \Omega \to [a,b]^d $ 
be a continuously uniformly distributed random variable,  
let
$\defaultStochLoss\colon\R^\defaultParamDim\times [0,T] \times \R^d \to\R$
satisfy for all
	$\theta\in\R^\d$,
	$w \in \R^d$,
	$x \in [a,b]^d$
that
\begin{equation}
\label{dkm_derivation:eq4}
\begin{split}
	\defaultStochLoss(\theta, w, x)
&=
	\abs[\big]{ \varphi( \varrho w  + x ) - \localAnn{\theta}(x) }^2,
\end{split}
\end{equation}
and
let
$\Theta = (\Theta_n)_{n \in \N_0} \colon\N_0\times\Omega\to\R^\d$ 
satisfy for all 
$n\in\N$ that
\begin{equation}
\label{dkm_derivation:eq5}
	\Theta_0= \xi
\qquad\text{and}\qquad
	\Theta_n
=
	\Theta_{n-1}-\gamma_n\br*{\frac1{J}\sum_{j=1}^{J}(\nabla_\theta\defaultStochLoss)(\Theta_{n-1}, \WRV_{n, j}, \XRV_{n, j} )}.
\end{equation}
Finally, the idea of \DKMs\ is to consider
for large enough $n \in \N$ the realization function
$\localAnn{\Theta_n}$ as an approximation
\begin{equation}
\label{dkm_derivation:eq6}
\begin{split} 
	\pr[\big]{\localAnn{\Theta_n}}|_{[a,b]^d}\approx u(T, \cdot)|_{[a,b]^d} 
\end{split}
\end{equation}
of the unknown solution $u$ of the \PDE\ in \cref{dkm_derivation:eq1} at time $T$ restricted to $[a,b]^d$.

An implementation in the case of a two-dimensional heat equation of the \DKMs\ derived above that employs the more sophisticated \Adam\ \SGD\ optimization method instead of the \SGD\ optimization method can be found in the next section.

\endgroup

\section{Implementation of DKMs}
\label{sec:impl_dKM}

In \cref{lst.kolmogorov} below we present a simple implementation of a \DKM,
as explained in \cref{sec:dkm_derivation} above, for finding an approximation
of a solution $u\in C^{1,2}([0,2]\times\R^2)$ of the two-dimensional
heat equation
\begin{align}
  \bpr{\tfrac{\partial u}{\partial t}}(t,x) = (\Delta_x u)(t,x)
\end{align}
with $u(0,x)=\cos(x_1)+\cos(x_2)$ for $t\in[0,2]$, $x=(x_1,x_2)\in\R^2$.
This implementation trains a fully connected feed-forward \ann\ with $2$ hidden layers
(with 50 neurons on each hidden layer)
and using the \ReLU\ activation function (cf.\ \cref{subsect:Relu}).
The training uses batches of size 256 with each batch 
consisting of 256 randomly chosen realizations of the random variable $(\mathcal T,\mathcal X)$,
where $\mathcal T$ is continuously uniformly distributed random variable on $[0,2]$
and where
$\mathcal X$ is a continuously uniformly distributed random variable on $[-5,5]^2$.
The training is performed using the \Adam\ \SGD\ optimization method
(cf.\ \cref{sect:adam}).
A plot of the resulting approximation of the solution $u$ after
3000 training steps is shown in \cref{fig:pinn}.

\begingroup
  \filelisting{lst.kolmogorov}{code/kolmogorov.py}{A simple implementation
  in \textsc{PyTorch} of the deep Kolmogorov method based on \cref{cor:kolmogorovtime},
   computing
  an approximation of the function $u\in C^{1,2}(\br{0,2}\times\R^{2},\R)$
  which satisfies for all $t\in\br{0,2}$, $x=(x_1,x_2)\in\R^{2}$ that
  $\bpr{\tfrac{\partial u}{\partial t}}(t,x) = (\Delta_x u)(t,x)$
  and
  $u(0,x)=\cos(x_1) + \cos(x_2)$.
  }
\endgroup

\begin{figure}[!ht]
	\centering
	\includegraphics[width=\linewidth]{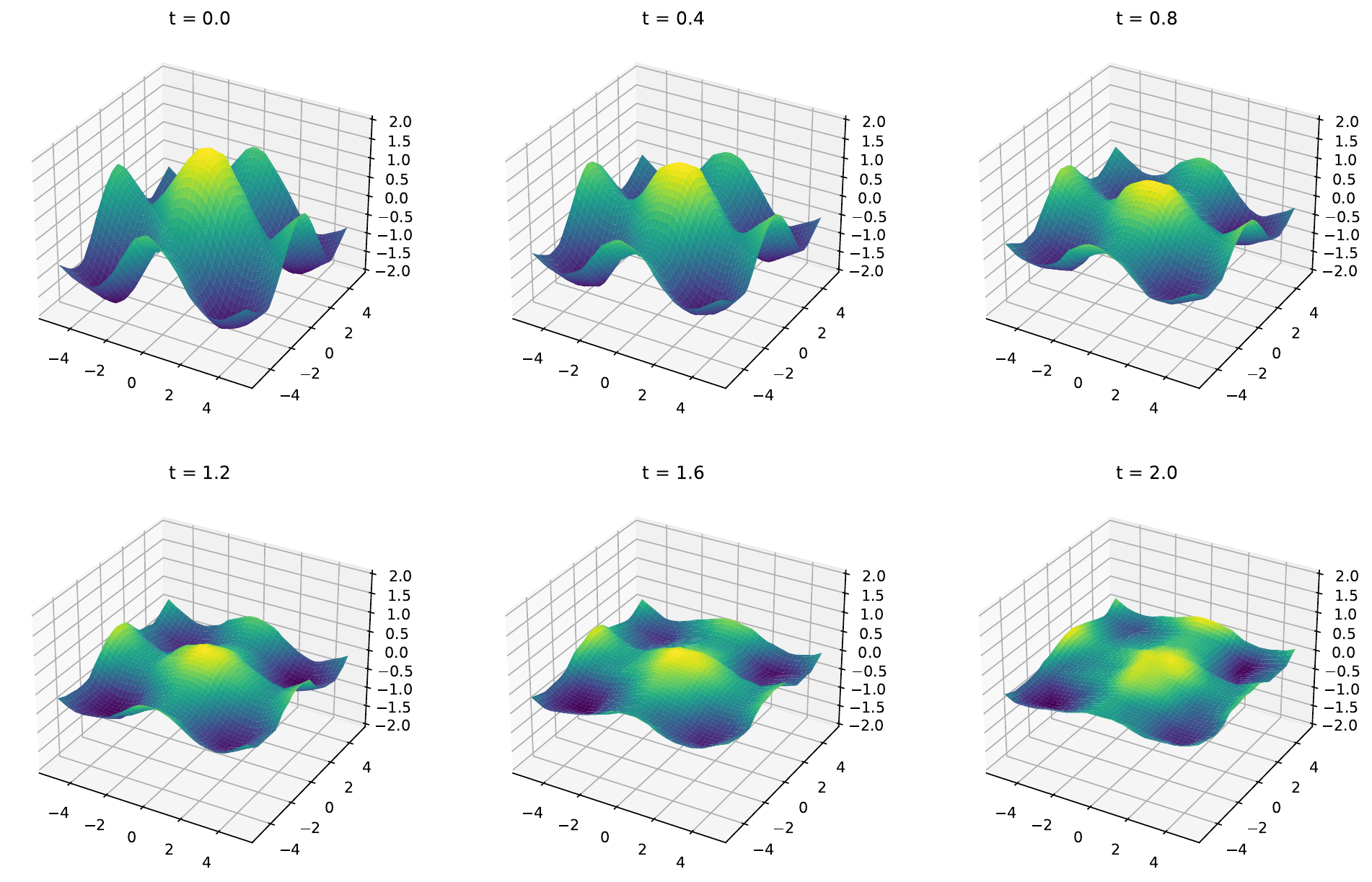}
	\caption{\label{fig:kolmogorov}Plots for the functions
  $[-5,5]^2\ni x\mapsto U(t,x)\in\R$, where
  $t\in\{0,0.4,0.8,1.2,1.6,2\}$
  and where
  $U\in C([0,2]\times\R^{2},\R)$
  is an approximation for the function 
  $u\in C^{1,2}([0,2]\times\R^{2},\R)$
  satisfies for all $t\in[0,2]$, $x=(x_1,x_2)\in\R^{2}$ that
  $\bpr{\tfrac{\partial u}{\partial t}}(t,x) = (\Delta_x u)(t,x)$
  and
  $u(0,x)=\cos(x_1) + \cos(x_2)$
  computed by means of \cref{lst.kolmogorov}.
  }
\end{figure}

%% file: parts/Further_DL_methods.tex
\cchapter{Further deep learning methods for PDEs}{sect:further_DL_for_PDEs}

\todosecond{Cite \cite{Paszke2017} somewhere}

Besides \PINNs, \DGMs, and \DKMs\ reviewed in \cref{sect:deepKolmogorov,subsec:dgm} above
there are also a large number of other works which propose and study deep learning based approximation methods for various classes of \PDEs.
In the following we mention a selection of such methods from the literature roughly grouped into three classes.
Specifically,  
	we consider deep learning methods for \PDEs\ which employ \emph{strong formulations} of \PDEs\ to set up learning problems in \cref{sec:DL_strong},
	we consider deep learning methods for \PDEs\ which employ \emph{weak or variational formulations} of \PDEs\ to set up learning problems in \cref{sec:DL_weak},
	and
	we consider deep learning methods for \PDEs\ which employ intrinsic \emph{stochastic representations} of \PDEs\ to set up learning problems in \cref{sec:DL_stochastic}.
Finally, in \cref{sec:ANN_approx_PDEs} we also point to several theoretical results and error analyses for deep learning methods for \PDEs\ in the literature.

Our selection of references for methods as well as theoretical results is by no means complete.
For more complete reviews of the literature on deep learning methods for \PDEs\ and corresponding theoretical results
we refer, \eg, to the overview articles
	\cite{beck2020overview,Germain2021,Ruf2019,Cuomo2022,karniadakis2021physics,Brunton2023,Weinan2021}.

\section{Deep learning methods based on strong formulations of PDEs}
\label{sec:DL_strong}

There are a number of deep learning based methods for \PDEs\ in the literature that employ residuals of strong formulations of \PDEs\ to set up learning problems
(cf., \eg, \cref{thm:dgm} and \cref{PINNS_derivation:eq6} for the residual of the strong formulation in the case of semilinear heat \PDEs).
Basic methods in this category include
	the \PINNs\ (see Raissi et al.~\cite{raissi2019physics}) and \DGMs\ (see Sirignano \& Spiliopoulos~\cite{Sirignano2018dgm}) reviewed in \cref{subsec:dgm} above,
	the approach proposed in Berg \& Nyström~\cite{Berg2018},
	the \TGNNs\ proposed in Wang et al.~\cite{Wang2020},
	and 
	the two early methods proposed in \cite{Dissanayake1994,Lagaris1998}.
There are also many refinements and adaptions of these basic methods in the literature including
\begin{itemize}
\item 
the  \cPINNs\ methodology for conservation laws in
	Jagtap et al.~\cite{Jagtap2020} 
which relies on multiple \anns\ representing a \PDE\ solution on respective sub-domains,
\item 
the \XPINNs\ methodology in
	Jagtap \& Karniadakis~\cite{Jagtap2020a}
which generalizes the domain decomposition idea of Jagtap et al.~\cite{Jagtap2020} to other types of \PDEs,
\item 
the \NSFnets\ methodology in
	Jin et al.~\cite{Jin2021}
which explores the use of \PINNs\ for the incompressible Navier-Stokes \PDEs,
\item 
the  \emph{Bayesian \PINNs} methodology in
	Yang et al.~\cite{Yang2021}
which combines \PINNs\ with \BNNs\ from Bayesian learning (cf., \eg, \cite{Neal1996,Luo2020}),
\item 
the \PPINNs\ methodology for time-dependent \PDEs\ with long time horizons in
	Meng et al.~\cite{Meng2020}
which combines the \PINNs\ methodology with ideas from parareal algorithms (cf., \eg, \cite{Maday2002,Blumers2019}) in order to split up long-time problems into many independent short-time problems,
\item 
the \emph{SelectNets} methodology in
	Gu et al.~\cite{Gu2021}
which extends the \PINNs\ methodology by employing a second \ann\ to adaptively \emph{select} during the training process the points at which the residual of the \PDE\ is considered,
and
\item 
the \fPINNs\ methodology in
	Pang et al.~\cite{Pang2019}
which extends the \PINNs\ methodology to \PDEs\ with fractional derivatives such as space-time fractional advection-diffusion equations.
\end{itemize}
We also refer to the article Lu et al.~\cite{Lu2021a} which introduces an elegant {\sc Python} library for \PINNs\ called \emph{DeepXDE} and also provides a good introduction to \PINNs.

\section{Deep learning methods based on weak formulations of PDEs}
\label{sec:DL_weak}

Another group of deep learning methods  for \PDEs\ relies on weak or variational formulations of \PDEs\ to set up learning problems.
Such methods include
\begin{itemize}
\item
the \VPINNs\ methodology in 
	Kharazmi et al.~\cite{Kharazmi2019,Kharazmi2021}
which use the residuals of weak formulations of \PDEs\ for a fixed set of test functions to set up a learning problem,
\item
the \emph{VarNets} methodology in 
	Khodayi-Mehr \& Zavlanos~\cite{KhodayiMehr2020}
which employs a similar methodology than \VPINNs\ but also consider parametric \PDEs,
\item
the \emph{weak form \TGNN} methodology in
	Xu et al.~\cite{Xu2021}
which further extend the \VPINNs\ methodology by (amongst other adaptions) considering test functions in the weak formulation of \PDEs\ tailored to the considered problem, 
\item
the \emph{deep fourier residual method} in
	Taylor et al.~\cite{Taylor2023}
which is based on minimizing the dual norm of the weak-form residual operator of \PDEs\ by employing Fourier-type representations of this dual norm which can efficiently be approximated using the \DST\ and \DCT,
\item
the \WANs\ methodology in 
	Zang et al.~\cite{Zang2020} (cf.\ also Bao et al.~\cite{Bao2020})
which is based on approximating both the solution of the \PDE\ and the test function in the weak formulation of the \PDE\ by \anns\ and on using an adversarial approach (cf., \eg, Goodfellow et al.~\cite{Goodfellow2014}) to train both networks to minimize and maximize, respectively, the weak-form residual of the \PDE,
\item
the \emph{Friedrichs learning} methodology in
	Chen et al.~\cite{Chen2023}
which is similar to the \WAN\ methodology but uses a different minimax formulation for the weak solution related to Friedrichs' theory on symmetric system of \PDEs\ (see Friedrichs~\cite{Friedrichs1958}),
\item
the \emph{deep Ritz} method for elliptic \PDEs\ in
	E \& Yu~\cite{E2018}
which employs variational minimization problems associated to \PDEs\ to set up a learning problem,
\item
the \emph{deep Nitsche} method in
	Liao \& Ming~\cite{Liao2021}
which refines the deep Ritz method using Nitsche's method (see Nitsche~\cite{Nitsche1971}) to enforce boundary conditions,
and 
\item
the \DDDM\ in
	Li et al.~\cite{Li2020} 
which refines the deep Ritz method using domain decompositions.
\end{itemize}
We also refer to the \MscaleDNNs\ in
	Cai et al.~\cite{Liu2020,Cai2019}
for a refined \ann\ architecture which can be employed in both the strong-form-based \PINNs\ methodology and the variational-form-based deep Ritz methodology.

\section{Deep learning methods based on stochastic representations of PDEs}
\label{sec:DL_stochastic}

\todosecond{Cite \cite{Markidis2021},\cite{Meer2022}, cite Deep Quadratic Hedging}

A further class of deep learning based methods for \PDEs\ are based on intrinsic links between \PDEs\ and probability theory
such as Feynman--Kac-type formulas; cf., \eg, \cite[Section 8.2]{Oeksendal2003}, \cite[Section 4.4]{Karatzas1991} for linear Feynman--Kac formulas based on (forward) \SDEs\ and cf., \eg, \cite{Pardoux1992,Pardoux1999,Pardoux1990,Cheridito2007} for nonlinear Feynman--Kac-type formulas based on \BSDEs.
The \DKMs\ for linear \PDEs\ (see Beck et al.~\cite{BeckJafaari21}) reviewed in \cref{sect:deepKolmogorov} are one type of such methods based on linear Feynman--Kac formulas.
Other methods based on stochastic representations of \PDEs\ include
\begin{itemize}
\item 
the \emph{deep BSDE} methodology in
	E et al.~\cite{EHanJentzen2017Science,EHanJentzen17}
which 
suggests to approximate solutions of \emph{semilinear parabolic \PDEs} 
by approximately solving the \BSDE\ associated to the considered \PDE\ through the nonlinear Feyman-Kac formula (see Pardoux \& Peng~\cite{Pardoux1992,Pardoux1990})
using a new deep learning methodology based on
\begin{itemize}
\item 
reinterpreting the \BSDE\ as a stochastic control problem in which the objective is to minimize the distance between the terminal value of the controlled process and the terminal value of the \BSDE,
\item
discretizing the control problem in time,
and
\item
approximately solving the discrete time control problem by approximating the policy functions at each time steps by means of \anns\ as proposed in E \& Han~\cite{Han2016},
\end{itemize}
\item 
the generalization of the deep BSDE methodology in
	Han \& Long~\cite{Han2020}
for semilinear and quasilinear parabolic \PDEs\ based on \FBSDEs\,
\item 
the refinements of the deep BSDE methodology in
	\cite{Fujii2019,Raissi2018a,ChanWaiNam2019,HenryLabordere2017,Nuesken2021}
which explore different nontrivial variations and extensions of the original deep BSDE methodology including different \ann\ architectures, initializations, and loss functions,
\item 
the extension of the deep BSDE methodology to fully nonlinear parabolic \PDEs\ in
	Beck et al.~\cite{BeckWeinanJentzen2019}
which is based on a nonlinear Feyman-Kac formula involving second order \BSDEs\ (see Cheridito et al.~\cite{Cheridito2007}),
\item 
the \emph{deep backward schemes} for semilinear parabolic \PDEs\ in
	Hur\'{e} et al.~\cite{Hure2020}
which also rely on \BSDEs\ but set up many separate learning problems which are solved inductively backwards in time
instead of one single optimization problem,
\item 
the \emph{deep backward schemes} in
	Pham et al.~\cite{Pham2021}
which extend the methodology in Hur\'{e} et al.~\cite{Hure2020} to fully nonlinear parabolic \PDEs,
\item 
the \emph{deep splitting} method for semilinear parabolic \PDEs\ in
	 Beck et al.~\cite{Beck2021}
which iteratively solve for small time increments linear approximations of the semilinear parabolic \PDEs\ using \DKMs,
\item 
the extensions of the deep backwards schemes to \PIDEs\ in
	\cite{Castro2022,Gnoatto2022},
\item 
the extensions of the deep splitting method to \PIDEs\ in
	\cite{Frey2022a,Boussange2022},
\item 
the methods in 
	Nguwi et al.~\cite{Nguwi2023,Nguwi2022,Nguwi2022a}
which are based on representations of \PDE\ solutions involving \emph{branching-type processes} (cf., \eg, also \cite{HenryLabordere2012,Nguwi2023a,HenryLabordere2021} and the references therein for nonlinear Feynman--Kac-type formulas based on such branching-type processes), and

\item 
the methodology for elliptic \PDEs\ in
	Kremsner et al.~\cite{Kremsner2020}
which relies on suitable representations of elliptic \PDEs\ involving \BSDEs\ with random terminal times.
\end{itemize}

\section{Error analyses for deep learning methods for PDEs}
\label{sec:ANN_approx_PDEs}

Until today there is not yet any complete error analysis for a
\GD/\SGD\ based \ann\ training approximation scheme for \PDEs\
in the literature (cf.\ also \cref{suboptimal_points} above).
However, there are now several partial error analysis results for deep learning methods for \PDEs\ in the literature (cf., \eg, \cite{Germain2022,Han2020,Belak2021,Mishra2020,Mishra2020a,Gonon2023,Frey2022} and the references therein).

In particular, there are nowadays a number of results which rigorously establish that \anns\ have the fundamental capacity to approximate solutions of certain classes of \PDEs\ without the \COD\ (cf., \eg, \cite{Bellman2010} and \cite[Chapter 1]{Novak08}) in the sense that the
number of parameters of the approximating \ann\ grows at most polynomially in both the reciprocal $\nicefrac{1}{\varepsilon}$ of the prescribed approximation accuracy $\varepsilon \in (0,\infty)$ and the \PDE\ dimension $d \in \N$.
We refer, \eg, to 
\cite{Beznea2022,Baggenstos2023,Berner2020,Elbraechter2022,Gonon2022,Gonon2021,Grohs2021,Grohs2023Aproof,Grohs2022,HornungJentzenSalimova2020arXiv,JentzenSalimovaWelti2021,Kutyniok2022,Reisinger2020}
for such and related \ann\ approximation results for solutions of linear \PDEs\
and we refer, \eg, to 
\cite{CioicaLicht2022,Grohs2021a,HutzenthalerJentzenKruseNguyen2019,Ackermann2023,Neufeld2023a}
for such and related \ann\ approximation results for solutions of nonlinear \PDEs.

The proofs in the above named \ann\ approximation results are usually based,
first,
	on considering a suitable algorithm which approximates the considered \PDEs\ without the \COD\ and,
thereafter, 
	on constructing \anns\ which approximate the considered approximation algorithm.
In the context of linear \PDEs\ the employed approximation algorithms are typically standard Monte Carlo methods (cf., \eg, \cite{Graham2013,Korn2010,Gobet2016} and the references therein)
and in the context of nonlinear \PDEs\ the employed approximation algorithms are typically nonlinear Monte Carlo methods of the mulitlevel-Picard-type (cf., \eg, \cite{Neufeld2022,Hutzenthaler2020,Hutzenthaler2020a,Beck2020,Giles2019,Hutzenthaler2022,Beck2020a,Hutzenthaler2020b,Hutzenthaler2021a,Neufeld2023} and the references therein).

In the literature the above named polynomial growth property
in both the reciprocal $\nicefrac{1}{\varepsilon}$ of the prescribed approximation accuracy $\varepsilon \in (0,\infty)$ and the \PDE\ dimension $d \in \N$ 
is also referred to as \emph{polynomial tractability} 
(cf., \eg, \cite[Definition 4.44]{Novak08}, \cite{Novak2010}, and \cite{Novak2012}).